\documentclass[11pt]{gsasthesis} %

\usepackage[margin={1.2in}]{geometry}
\usepackage[titletoc]{appendix}

\usepackage{float}
\usepackage{microtype}
\usepackage{graphicx}
\usepackage{booktabs} %
\usepackage{enumitem,multirow,adjustbox,array}
\usepackage[utf8]{inputenc} %
\usepackage[T1]{fontenc}    %
\usepackage{amsfonts}
\usepackage{tikz}
\usetikzlibrary{bayesnet} %
\usetikzlibrary{arrows}
\usepackage{arydshln} %
\usepackage{wrapfig, ragged2e}
\usepackage{siunitx, hhline, colortbl}
\usepackage{subfigure,wrapfigure}
\usepackage{framed}
\definecolor{shadecolor}{rgb}{0.94, 0.97, 1.0}
\definecolor{purple}{HTML}{7D2882}
\usepackage{CJK}

\usepackage{amsmath,amsfonts,bm}

\newcommand{\norm}[1]{\left\lVert#1\right\rVert}
\newcommand{\ind}{\perp\!\!\!\!\perp}

\def\eqref#1{equation~\ref{#1}}

\def\floor#1{\lfloor #1 \rfloor}
\def\1{\bm{1}}
\newcommand{\train}{\mathcal{D_{\mathrm{tr}}}}

\def\rv{{\textnormal{v}}}

\def\rvc{{\mathbf{c}}}

\def\rvg{{\mathbf{g}}}

\def\rvv{{\mathbf{v}}}
\def\rvw{{\mathbf{w}}}
\def\rvx{{\mathbf{x}}}

\def\rmH{{\mathbf{H}}}

\def\rmW{{\mathbf{W}}}

\def\vg{{\bm{g}}}

\def\vp{{\bm{p}}}

\def\mL{{\bm{L}}}

\def\mT{{\bm{T}}}

\def\mW{{\bm{W}}}
\def\mX{{\bm{X}}}
\def\mY{{\bm{Y}}}

\DeclareMathAlphabet{\mathsfit}{\encodingdefault}{\sfdefault}{m}{sl}
\SetMathAlphabet{\mathsfit}{bold}{\encodingdefault}{\sfdefault}{bx}{n}

\def\gA{{\mathcal{A}}}

\def\gC{{\mathcal{C}}}
\def\gD{{\mathcal{D}}}
\def\gE{{\mathcal{E}}}
\def\gF{{\mathcal{F}}}
\def\gG{{\mathcal{G}}}
\def\gH{{\mathcal{H}}}
\def\gI{{\mathcal{I}}}

\def\gL{{\mathcal{L}}}

\def\gN{{\mathcal{N}}}
\def\gO{{\mathcal{O}}}
\def\gP{{\mathcal{P}}}

\def\gS{{\mathcal{S}}}

\def\gW{{\mathcal{W}}}
\def\gX{{\mathcal{X}}}
\def\gY{{\mathcal{Y}}}
\def\gZ{{\mathcal{Z}}}

\def\sP{{\mathbb{P}}}

\def\sZ{{\mathbb{Z}}}

\newcommand{\E}{\mathbb{E}}

\newcommand{\R}{\mathbb{R}}

\newcommand{\sigmoid}{\sigma}

\newcommand{\KL}{D_{\mathrm{KL}}}

\DeclareMathOperator*{\argmax}{arg\,max}
\DeclareMathOperator*{\argmin}{arg\,min}

\DeclareMathOperator{\sign}{sign}

\usepackage{algorithm,algorithmic}

\usepackage{amsmath}
\usepackage{amssymb}
\usepackage{mathtools}
\usepackage{amsthm}

\usepackage[capitalize,noabbrev]{cleveref}

\theoremstyle{plain}
\newtheorem{theorem}{Theorem}[section]
\newtheorem{nono-theorem}{Theorem}[]
\newtheorem{proposition}[theorem]{Proposition}
\newtheorem{lemma}[theorem]{Lemma}
\newtheorem{corollary}[theorem]{Corollary}
\newtheorem{definition}[theorem]{Definition}
\newtheorem{assumption}[theorem]{Assumption}

\usepackage{xspace}
\newcommand{\irml}[0]{\text{IRMv1}\xspace}
\newcommand{\erm}[0]{\text{ERM}\xspace}
\newcommand{\Rad}{\textrm{Rad}}
\newcommand{\inv}{\textrm{inv}}
\newcommand{\spu}{\textrm{spu}}

\newcommand{\dobed}[0]{\textsc{DomainBed}\xspace}
\newcommand{\wilds}[0]{\textsc{Wilds}\xspace}
\newcommand{\pacs}[0]{\textsc{PACS}\xspace}
\newcommand{\terra}[0]{\textsc{TerraIncognita}\xspace}

\newcommand{\ours}[0]{\text{FAT}\xspace}

\newcommand{\ourst}[0]{\text{FAT}\xspace}	%
\newcommand{\oursfull}[0]{\textbf{F}eature \textbf{A}ugmented \textbf{T}raining \xspace}

\newcommand{\rfc}[0]{\text{Bonsai}\xspace}
\newcommand{\irmx}[0]{\text{IRMX}\xspace}
\newcommand{\vrex}[0]{\text{VREx}\xspace}
\newcommand{\cmnist}[0]{\textsc{ColoredMNIST}\xspace}
\newcommand{\dataset}{{\cal D}}
\newcommand{\ood}{\text{ood}}
\newcommand{\env}{{\gE}}
\newcommand{\envtrain}{{\gE_{\text{tr}}}}
\newcommand{\envtest}{{\gE_{\text{te}}}}
\newcommand{\envall}{{\gE_{\text{all}}}}
\newcommand{\envmix}{{\gE^{\text{mix}}_{\text{tr}}}}
\newcommand{\gen}{{{\text{gen}}}}
\newcommand{\invrat}{{{InvRAT}}}
\newcommand{\ego}{\text{ego}}
\newcommand{\envalpha}{{\gE_{\alpha}}}
\newcommand{\rad}{{\text{Rad}}}
\newcommand{\var}{{\text{var}}}
\newcommand{\des}{\text{des}}
\newcommand{\std}[1]{\textit{(\scriptsize{$\pm$#1}})}
\newcommand{\abs}[1]{\left\lvert#1\right\rvert}
\newcommand{\vL}{\bm{L}}
\newcommand{\vG}{\bm{G}}

\newcommand{\PA}{\text{PA}}

\newcommand{\doop}{{\text{do}}}

\newcommand{\dir}[0]{{DIR}\xspace}
\newcommand{\grea}[0]{{GREA}\xspace}

\newcommand{\gsat}[0]{{GSAT}\xspace}
\newcommand{\disc}[0]{{DisC}\xspace}
\newcommand{\mole}[0]{{MoleOOD}\xspace}
\newcommand{\gil}[0]{{GIL}\xspace}
\newcommand{\trainvirtual}{\mathcal{D}^v_{\mathrm{tr}}}
\newcommand{\ciga}[0]{{CIGA}\xspace}
\newcommand{\cigafull}[0]{Causality Inspired Invariant Graph LeArning\xspace}
\newcommand{\gala}[0]{{GALA}\xspace}
\newcommand{\galafull}[0]{Graph invAriant Learning Assistant\xspace}
\newcommand{\pair}[0]{{PAIR}\xspace}
\newcommand{\pairo}[0]{{PAIR-o}\xspace}
\newcommand{\pairs}[0]{{PAIR-s}\xspace}
\newcommand{\pairfull}[0]{PAreto Invariant Risk Minimization\xspace}
\newcommand{\feat}[0]{{FeAT}\xspace}
\newcommand{\feati}[0]{{iFeAT}\xspace}
\newcommand{\featfull}[0]{Feature Augmented Training\xspace}
\newcommand{\hao}[0]{{HAO}\xspace}
\newcommand{\haofull}[0]{Harmonious Adversarial Objective\xspace}
\newcommand{\gmt}[0]{{GMT}\xspace}
\newcommand{\gmtl}[0]{{GMT-lin}\xspace}

\newcommand{\gmts}[0]{{GMT-sam}\xspace}

\newcommand{\gmtt}[0]{\text{GMT}\xspace}	%
\newcommand{\gmtfull}[0]{Graph Multilinear neT\xspace}
\newcommand{\smtfull}[0]{subgraph multilinear extension\xspace}
\newcommand{\smt}[0]{SubMT\xspace}
\newcommand{\acts}{ActsTrack\xspace}
\newcommand{\taumu}{Tau3Mu\xspace}
\newcommand{\pdbbind}{PLBind\xspace}
\newcommand{\synbind}{SynMol\xspace}
\newcommand{\pge}{BernMask-P\xspace}
\newcommand{\gexp}{BernMask\xspace}
\newcommand{\gradpos}{GradGeo\xspace}
\newcommand{\gradcam}{GradGAM\xspace}
\newcommand{\pointmask}{PointMask\xspace}
\newcommand{\xgnns}[0]{\text{XGNNs}\xspace}
\newcommand{\xgnn}[0]{\text{XGNN}\xspace}

\newcommand{\lri}[0]{LRI\xspace}

\newcommand{\topspace}{\vspace{-0.25in}}
\newcommand{\atk}{\text{atk}}
\newcommand{\gma}{\text{GMA}}
\newcommand{\gia}{\text{GIA}}
\newcommand{\concat}{\mathbin{\|}}

\usepackage{pifont}
\newcommand{\cmark}{\ding{51}}%
\newcommand{\xmark}{\ding{55}}%
\newenvironment{myquotation}{\setlength{\leftmargini}{0em}\quotation}{\endquotation}
\usepackage{mathabx}
\newcommand{\pred}[1]{\widehat{#1}\xspace}
\newcommand{\pa}[0]{\triangle\xspace}
\newcommand{\pc}[0]{\widebar{\triangle}\xspace}

\newcommand{\irms}[0]{$\text{IRM}_\gS$\xspace}
\newcommand{\irm}[0]{\text{IRM}\xspace}
\newcommand{\lsq}{\ell_{\textup{sq}}}
\newcommand{\llog}{\ell_{\textup{log}}}
\newcommand{\vrexz}{VREx\textsubscript{0}}
\newcommand{\Ix}{\mathcal{I}_X}
\newcommand{\Is}{\mathcal{I}_\mathcal{S}}
\newcommand{\Ede}{\mathbb{E}_{\mathcal{D}_e}}
\newcommand{\Prde}{\Pr_{\mathcal{D}_e}}
\newcommand{\hL}{\widehat{\mathcal{L}}}

\newcommand{\Peo}{\bm{P}^\epsilon_\ourst}

\newcommand{\Peeo}{\widehat{\bm{P}}^\epsilon_\ourst}

\newcommand{\absbig}[1]{\big\lvert#1\big\rvert}

\usepackage{rotating}

\usepackage{longtable}

\usepackage{comment}
\usepackage{makecell}

\usepackage{soul}
\usepackage[normalem]{ulem}

\newcolumntype{L}[1]{>{\raggedright\let\newline\\\arraybackslash\hspace{0pt}}m{#1}}
\newcolumntype{C}[1]{>{\raggedright\let\newline\\\arraybackslash\hspace{0pt}}m{#1}}

\usepackage{pdflscape}
\usepackage[english]{babel}
\usepackage[sc]{mathpazo}
\usepackage{courier}

\RequirePackage[font=small,format=plain,labelfont=bf]{caption}

\title{Learning Causality for Modern Machine Learning} 
\author{CHEN, Yongqiang} 
\degreename{Doctor of Philosophy}
\degreefield{Computer Science and Engineering} %
\department{Computer Science and Engineering} %
\degreemonth{July} 
\degreeyear{2024}
\principaladvisor{Professor CHENG James}
\committee{
	\centering
	Professor LI Yu (Chair)  \\
	Professor CHENG James (Thesis Supervisor)  \\
	Professor DOU Qi (Committee Member)  \\
	Professor RIBEIRO Bruno  (External Examiner)\\
}

\begin{document}

\pagenumbering{roman}
\thesistitlepage
	\committeepage
	\newpage
	\begin{abstract}
        \noindent
        Abstract of thesis entitled:\\
    	Learning Causality for Modern Machine Learning\\
    	\noindent
    	Submitted by CHEN, Yongqiang\\
    	\noindent
    	for the degree of Doctor of Philosophy in Computer Science and Engineering\\
    	at The Chinese University of Hong Kong in July 2024
    	\vspace{1cm}
    	
		\noindent In the past decades, machine learning with Empirical Risk Minimization (ERM) has demonstrated great capability in learning and exploiting the statistical patterns from data, or even surpassing humans.
		Despite the success, ERM avoids the modeling of \textit{causality}\textemdash the way of understanding and handling \textit{changes}, which is fundamental to human intelligence. 
		When deploying models beyond the training environment, distribution shifts are everywhere.
		For example, an autopilot system often needs to deal with new weather conditions that have not been seen during training; An AI-aided drug discovery system needs to predict the biochemical properties of molecules with respect to new viruses such as COVID-19.
		It renders the problem of Out-of-Distribution (OOD) generalization challenging to conventional machine learning.
		
		In this thesis, we investigate how to incorporate and realize the causality for broader tasks in modern machine learning. 
		In particular, we exploit the \textit{invariance} implied by the principle of independent causal mechanisms (ICM), that is, the causal mechanisms generating the effects from causes do not inform or influence each other. Therefore, the conditional distribution between the target variable given its causes is \textit{invariant} under distribution shifts.
		With the causal invariance principle, we first instantiate it to graphs\textemdash a general data structure ubiquitous in many real-world industry and scientific applications, such as financial networks and molecules.
		Then, we shall see how learning the causality benefits many of the desirable properties of modern machine learning, in terms of (i) OOD generalization capability; (ii) interpretability; and (iii) robustness to adversarial attacks.
		
		Realizing the causality in machine learning, on the other hand, raises a dilemma for optimization in conventional machine learning, as it often contradicts the objective of ERM. Therefore, we characterize how the contradicts affect the feature learning and optimization, and propose new representation learning, and optimization paradigms, that properly handle the optimization dilemma. 
		
		With proper objectives and effective realization schemes of causal invariance learning, this thesis marks the first steps toward building foundations for modern paradigms of machine learning.
	\end{abstract}
	
	\begin{cabstract}
		\begin{CJK*}{UTF8}{bkai}
        \noindent
        摘要：

        \vspace{1cm}
		在過去幾十年裡，機器學習配合經驗風險最小化已經展現了在學習和利用數據統計模式方面的巨大能力，甚至超越了人類。儘管取得了成功，但經驗風險最小化避免了對因果性的建模——理解和處理變化的方式，這對人類智能來說是基本的。當模型部署到訓練環境之外時，處處都是分佈變化。例如，自動駕駛系統經常需要處理在訓練期間未曾遇到的新天氣條件；一個人工智能輔助的藥物發現系統需要預測分子對於新型病毒如2019冠狀病毒的生化特性。這使得分佈外泛化問題對於傳統機器學習來說充滿挑戰。
		
		在這篇論文中，我們探討如何在現代機器學習中的廣泛任務中納入和實現因果性。特別是，我們利用獨立因果機制原則所暗示的不變性，即，產生效果的因果機制不會相互通報或影響。因此，目標變量給定其原因的條件分佈在分佈變化下是不變的。有了因果不變原則，我們首先將其實例化到圖上——一種在許多真實世界工業和科學應用中無處不在的通用數據結構，如金融網絡和分子。然後，我們進一步展示學習因果性如何有利於現代機器學習的許多理想特性，如(i)分佈外泛化能力；(ii)可解釋性；以及(iii)對抗性攻擊的魯棒性。
		
		另一方面，實現機器學習中的因果性，對於傳統機器學習中的優化提出了一個難題，因為它經常與經驗風險最小化的目標相矛盾。因此，我們描述了這些矛盾如何影響特徵學習和優化，並提出了新的表示學習和優化範式，適當處理優化難題。通過適當的目標和有效的因果不變學習實現方案，這篇論文搭建了邁向現代機器學習新基礎的第一步。
		\end{CJK*}
	\end{cabstract}
	\newpage
	\noindent
	\vspace*{\fill}
	\begin{center}
		This work is dedicated to the people \\in pursuit of "why"\\for intelligence and causality.
	\end{center}
	\vspace*{\fill}
	\begin{acknowledgments}
	First, I would like to express my sincere gratitude to my supervisor, Prof. James Cheng, for providing unreserved support, intellectual freedom, and wise advice during my PhD study.
    I would also like to especially thank Prof. Bo Han and Kaiwen Zhou, as well as Prof. Tongliang Liu, Han Yang, Yonggang Zhang, Jin-Ge Yao, and B\"{o}jre Karlsoon who hosted my undergraduate internship at Microsoft Research Asia, for walking me into the palace of research.
    Meanwhile, I am very grateful to Yatao Bian and Peilin Zhao for providing an open and supportive environment for my research.
    I would not have survived the PhD grind without their patient guidance and kind support.
    Also, I would like to thank my committee members Prof. Bruno Ribeiro, Prof. Yu Li, and Prof. Qi Dou for their valuable time, interest, and comments on my research and thesis.

    I feel very fortunate to have worked together with and been advised by Prof. Kun Zhang, Prof. Tong Zhang, Prof. Masashi Sugiyama, Prof. Ludwig Schmidt, Dr. Gang Niu, and the pioneers in machine learning and causality area, for opening up my mind. I shall be always inspired by their passion, wisdom, and vision in pursuit of top research in my entire career.

    I would also like to thank my amazing collaborators,
    especially (randomly sorted) Kaiwen Zhou, Wei Huang, Yong Lin, and Zeyu Qin for their enthusiasm, insights, and help that really shaped my research and thesis.
    This thesis would not be possible without the important contributions of (in alphabetical order):
    Prof. Yuan Cao, Xinyan Dai, Prof. Mingming Gong, Andi Han, Kaili Ma, Barakeel Fanseu Kamhoua, Chenxi Liu, Weiwen Liu, Yu Rong, Prof. Taiji Suzuki, Jiaqi Wang, Qizhou Wang, Zihao Wang, Bingzhe Wu, Xiao Yan, Binghui Xie, Prof. Zhiqiang Xu, Han Yang, Haochen Yang, Prof. Zhiqiang Xu, Tianjun Yao, Lin Zhang and Yonggang Zhang.
    I am also very lucky to know and have had many insightful discussions with great researchers and friends like Jianyu Zhang and Prof. Han Zhao, where some are developed as exciting ongoing projects (Let us keep the full list secret for now).
    Besides, it is my great pleasure to meet and learn from many fascinating guys in the Husky data lab, TMLR group, MLC center, Causality/CleaR Group, and RIKEN-AIP.

    Finally, I wish to thank my family for their continuous support and love.
\end{acknowledgments}
\renewcommand{\contentsname}{\protect\centering\protect\Large Contents}
\renewcommand{\listtablename}{\protect\centering\protect\Large List of Tables}
\renewcommand{\listfigurename}{\protect\centering\protect\Large List of Figures}

\tableofcontents %
\listoffigures
\listoftables

\newpage
\pagenumbering{arabic} %
\setcounter{page}{1}
\pagestyle{headings}
\chapter{Introduction} \label{CH:intro}

In the past decades, machine learning with Empirical Risk Minimization (ERM) has demonstrated great capability in learning and exploiting the statistical patterns from data~\citep{erm}, or even surpassing humans in a variety of tasks such as object recognition, natural language translation, games of GO and StarCraft.
As a recent pinnacle of the ERM-based machine learning paradigm, when trained on a massive amount of data available on the Internet, ERM enables unprecedented large-scale neural networks to demonstrate human-like zero-shot or few-shot generalization capabilities on a wide range of cognitive tasks. The emergence of large pre-trained models is even considered to be an early spark of artificial general intelligence~\citep{spark_AGI}.

Despite the success, ERM avoids the modeling of \textit{causality}\textemdash a fundamental capability of human intelligence and an essential component of science~\citep{Hanson1958PatternsOD}. Causality uncovers the underlying cause-effect relationships of the observable. Causal knowledge provides the way of understanding and handling \textit{changes}, which can not be implied by the statistical correlations~\citep{common_cause}. For example, it is observed that the increase in chocolate consumption highly correlates to the increase of Nobel awardees in a country. However, to further increase the number of Nobel laureates, causality suggests policymakers invest more in developing the economy instead of feeding the people with more chocolates as implied by the correlations.
When deploying models beyond the training environment, \textit{the changes, appeared as distribution shifts, are everywhere}.
For example, an autopilot system often needs to deal with new weather conditions that have not been seen during training; An AI-aided drug discovery system needs to predict the biochemical properties of molecules with respect to new viruses such as COVID-19.
It raises the challenge of Out-of-Distribution (OOD) generalization, which requires machine learning models to perform well on data from a different distribution during training. 
Conventional ERM-based machine learning paradigms are shown to often exploit the statistical shortcuts in the training data, and fail catastrophically when there are distribution shifts during testing.

The theme of this thesis is to investigate how to incorporate and realize the causality for broader tasks in modern machine learning. 
In particular, we exploit the \textit{invariance} implied by the principle of independent causal mechanisms (ICM), that is, the causal mechanisms generating the effects $Y$ from causes $\PA(Y)$ do not inform or influence each other~\citep{elements_ci}. Therefore, the conditional distribution $P(Y|\PA(Y))$ between the target variable given its causes is \textit{invariant} under distribution shifts.
Leveraging merely the underlying causes to predict the target label is immune to the shifts of the observables $P(X)$.
Recently, a promising framework called invariant risk minimization (IRM) has been proposed to implement the causal invariance and has been demonstrated useful in linear data.
we first consider extending IRM to a general data structure, i.e., graphs.
In addition to images and natural languages, graphs are also ubiquitous in many real-world industry and scientific applications, such as e-commerce networks, molecules, physical systems, etc.
The complex nature of graphs poses unique challenges to learning the causal invariance. In particular, distribution shifts on graphs can appear in a variety of forms such as attributes and structures, making it difficult to identify the invariance. Moreover, environment partitions,	which are often required by IRM-based methods, could be highly expensive to obtain for graphs.
To tackle these challenges, we propose a series of new frameworks and architectures to learn the causal invariance on graph data.
We show that explicitly learning the causality can significantly improve the existing graph machine learning paradigms with better \textit{(i) OOD generalization capability; (ii) interpretability; and (iii) robustness to adversarial attacks.} 

Despite the promising objective of learning causal invariance from the data, making it work in deep learning is still challenging.
The additional regularization required for learning causal invariance is intrinsically contradicted with the existing paradigm of empirical risk minimization, which leads to a dilemma for optimization in conventional machine learning. 
If the regularization is too strong, it destroys the normal optimization routine.
If it is too weak, the invariance cannot be guaranteed.
Therefore, we conduct an in-depth analysis of how the contradicts affect feature learning and optimization.
Our theoretical results further motivate us to propose new representation learning, and optimization paradigms, that properly handle the optimization dilemma. 

In the following sections, we will first introduce the backgrounds and related work of OOD generalization. Then, since we study the problem from a more general perspective, i.e., graphs, we will also give a brief introduction to the neural networks operating on graphs, i.e., Graph Neural Networks (GNNs). It establishes the necessary preliminaries to appreciate the results of this thesis.

\section{Out-of-Distribution Generalization} 
The problem of OOD generalization typically considers
a supervised learning setting based on the data $\dataset=\{\dataset^e\}_{e\in\envall}$
collected from multiple causally related environments $\envall$,
where a subset of samples $\dataset^e=\{x^e_i,y^e_i\}$ from a single environment $e\in\envall$
are drawn independently from an identical distribution $\sP^e(X,Y)$~\citep{inv_principle}.
Given the data from training environments $\{\dataset^e\}_{e\in\envtrain}$,
the goal of OOD generalization is to find a predictor $f:\gX\rightarrow\gY$
that generalizes well to all (unseen) environments, i.e., to minimize
\begin{equation}
    \max_{e\in\envall}\gL_e(f), 
\end{equation}
where $\gL_e$ is the empirical risk under environment $e$.
The predictor $f=w\circ\varphi$ is usually composed of a featurizer $\varphi:\gX\rightarrow\gZ$ that learns to extract useful features, and a classifier $w:\gZ\rightarrow\gY$ that makes predictions from the extracted features.

There exists a rich literature aiming to overcome the OOD generalization challenge, which usually appears as \emph{additional regularizations} of ERM~\citep{erm}.
The first line is the Domain Generalization works~\citep{DANN,CORAL,deep_DG,DouCKG19} that tries to regularize the learned features to be \textbf{domain-invariant}. However, \citet{DG_issue} show that the domain invariant features solely are not sufficient for guaranteed good OOD generalization. We refer readers to~\citet{domainbed} for more details of the literature about Domain Generalization.
Moreover, \citet{dro,DRSL,groupdro} aim to regularize the models to be \textbf{robust to mild distributional perturbations} of the training distributions such that the models are expected to perform well in unseen test environments. Following the line of distributional robustness, \citet{jtt,cnc,lisa} further propose advanced strategies to improve the robustness by assuming that models trained with ERM have strong reliance to spurious features.

Recently there is increasing interest in adopt theory of causality~\citep{causality,elements_ci,towards_causality} and introduce the \textbf{causal invariance} to the learned representations~\citep{inv_principle,causal_transfer,irmv1}.
The causal invariance is inspired by the assumption of Independent Causal Mechanism (ICM) in causality~\citep{elements_ci}. ICM assumes that conditional distribution of each variable given its causes (i.e., its mechanism) does not inform or influence the other conditional distributions~\citep{causality,elements_ci}. \citet{inv_principle} introduce the concept of environments which are generated by different interventions on certain variables involved in the underlying data generation process of $(X,Y)$. Despite of the changes to the intervened variables, the conditional distribution of intervened variables (they usually are the direct parents of $Y$ in the underlying causal graph) and $Y$ is invariant. Therefore, the invariant relationship can be leveraged to predict $Y$ and generalize to different environments. We refer interested readers to~\citet{inv_principle,towards_causality,ib-irm} for more details.
Inspired by the causal invariance principle, \citet{irmv1} propose the framework of Invariant Risk Minimization (IRM) that allows the adoption of the causal invariance in neural networks.
It further inspires plentiful invariant learning works~\citep{andmask,causal_matching,env_inference,clove,ib-irm,zin}.
At the heart of these works is the intuition that: When a predictor $w$ acting on $\varphi$ minimizes the risks in all of the environments simultaneously,
$\varphi$ is expected to discard the spurious signals while keeping the causally invariant signals.
Additionally, there can be more definitions and implementations of the invariance~\citep{iga,vrex,fish,fishr} which further encourage \textbf{agreements} at various levels across different environments. We refer interested readers to~\citet{fishr} for a detailed comparison and discussion.
As shown that most of the existing approaches encounter the optimization dilemma when learning the causal invariance, this work mainly focuses on resolving the optimization issue in learning the causal invariance defined by the framework of Invariant Risk Minimization~\citep{irmv1}, which is different from the literature of IRM variants or other OOD objectives that focus on proposing better objectives to learn the causal invariance.

\section{Graph Neural Networks}
Graph Neural Networks (GNNs), as a generalization of deep learning models for graph-structured data, have gained great success in tasks involving relational information~\citep{jure_grlsurvey, peter_survey, zhou_survey, wu_survey}.
Consider a graph $G=(A,X)$  with node set $V=\{v_1,v_2,...,v_n\}$ and edge set $E=\{e_1,e_2,...,e_m\}$,
where  $A \in \{0,1\}^{n\times n}$  is the adjacency matrix and $X\in \R^{n \times d}$ is the node feature matrix.
GNNs are widely applied in node-level, link-level, and graph-level tasks. In this thesis, we will be focusing on node-level and graph-level tasks.

In node-level tasks, we are mainly interested in semi-supervised node classification. Given the set of labels $Y\in \{0,1,..,c-1\}^n$ from $c$ classes,
we can train a graph neural network $f_\theta$ parameterized by $\theta$ on the training (sub)graph $G_{\text{train}}$
by minimizing a classification loss $\gL_{\text{train}}$ (e.g., cross-entropy).
Then the trained $f_\theta$ can predict the labels of nodes in test graph $G_{\text{test}}$.
A GNN typically follows a neighbor aggregation scheme to recursively update the node representations as:
\begin{equation}
	\label{eq:gnn}
	h{(k)}_u = \sigma(W_k\cdot a(\{h{(k-1)}_v\}| v\in\mathcal{N}(u)\cup\{u\})),
\end{equation}
where $\mathcal{N}(u)$ is the set of neighbors of node $u$, $h{(0)}_u=X_u, \forall u \in V$, $h{(k)}_u$ is the hidden representation of node $u$ after the $k$-th aggregation,
$\sigma(\cdot)$ is an activation function, e.g., $\text{ReLU}$, and $a(\cdot)$ is an aggregation function over neighbors, e.g., $\text{MEAN}$ or $\text{SUM}$~\citep{gcn,sage,sgc,gin}.

Node-level tasks with GNNs are often performed as the semi-supervised node classification: we split the set of nodes $V$ into labeled nodes $V_L$ (or training set $V_{\text{train}}$) and unlabeled nodes $V_U$ (or test set $V_{\text{test}}$), and use $V_L$ to train $f_\theta$ which would map each node to one class from the $c$ classes by minimizing a loss function $\gL_{\text{train}}$ (e.g., cross-entropy) over the training sets. The learning can be performed in a transductive manner:
\[
\theta^* = \argmin_{\theta}\gL_{\text{train}}(f_\theta(G))=\frac{1}{|V_\text{train}|}\sum_{u\in V_\text{train}}L_{\text{train}}(f_\theta(G)_u,y_u),                       
\]
where the whole graph including the unlabeled nodes $L_U$ can be used. The learning can also be performed in a inductive manner:
\[
\theta^* = \argmin_{\theta}\gL_{\text{train}}(f_\theta(G_\text{train}))=\frac{1}{|V_\text{train}|}\sum_{u\in V_\text{train}}L_{\text{train}}(f_\theta(G_\text{train})_u,y_u),                     
\]
where only labeled nodes with edges among them can be seen during training. 

In graph-level tasks, we focus on graph classification, where we are given a set of $N$ graphs $\{G_i\}_{i=1}^N\subseteq \gG$
and their labels $\{Y_i\}_{i=1}^N\subseteq\gY=\R^c$ from $c$ classes.
Then, we train a GNN $f_\theta=\rho \circ h$ with an encoder $h:\gG\rightarrow\R^h$ that learns a meaningful representation $h_G$ for each graph $G$ to help predict their labels $y_G=\rho(h_G)$ with a downstream classifier $\rho:\R^h\rightarrow\gY$.
The representation $h_G$ is typically obtained by performing pooling with a $\text{READOUT}$ function on the learned node representations:
\begin{equation}
	\label{eq:gnn_pooling}
	h_G = \text{READOUT}(\{h^{(K)}_u|u\in V\}),
\end{equation}
where the $\text{READOUT}$ is a permutation invariant function (e.g., $\text{SUM}$, $\text{MEAN}$)~\citep{gin,diff_pooling,relation_pooling,can_gnn_count,wl_goml}.

\section{Thesis Organization}
\bgroup
\def\arraystretch{2}
\begin{table}[H]
	\caption{Organizations and categorization of each chapter in the thesis.}
	\label{summary_table}
	\footnotesize
	\centering
	\begin{tabular}{|c|c|c|}
		\hline
		\textbf{Category} &\textbf{Method} & \textbf{Chapter} \\\hhline{|===|}
          \multirow{2}{*}{Foundation} & \ciga:\cigafull & Chapter~\ref{CH:CIGA} \\
         &\gala:\galafull &Chapter~\ref{CH:GALA} \\
		\hline
  \multirow{2}{*}{Implication} & \gmt:\gmtfull & Chapter~\ref{CH:GMT} \\
            & \hao:\haofull &Chapter~\ref{CH:HAO} \\
		\hline
  \multirow{2}{*}{Optimization} & \pair:\pairfull & Chapter~\ref{CH:PAIR} \\
            & \feat:\featfull &Chapter~\ref{CH:FeAT} \\
		\hline
	\end{tabular}
\end{table}
\egroup

This thesis is organized as follows (relations are given in Table \ref{summary_table}):
\begin{itemize}
    \item Part~\ref{P1} Chapter~\ref{CH:CIGA} presents the basic framework \ciga, including architectures and learning objectives for learning causal invariance on graphs. The derived method \ciga demonstrates strong OOD generalizability in more than $30$ synthetic and real-world benchmarks. 
    \item Part~\ref{P1} Chapter~\ref{CH:GALA} investigates the feasibility and minimal assumptions for learning the causal invariance on graphs. The proposed hardness results motivate a set of minimal assumptions as well as a new method called \gala that achieves a better OOD generalization performance in more general settings.
    \item Part~\ref{P2} Chapter~\ref{CH:GMT} presents the implication of causal learning to the interpretability. 
    \item Part~\ref{P2} Chapter~\ref{CH:HAO} presents the implication of causal learning to the adversarial robustness.
    \item Part~\ref{P3} Chapter~\ref{CH:PAIR} investigates the optimization-generalization dilemma in OOD generalization and presents new optimization schemes to mitigate the dilemma.
    \item Part~\ref{P3} Chapter~\ref{CH:FeAT} investigates the optimization-generalization dilemma from the feature learning perspective, and presents a new representation learning to learn rich feature representations ready for OOD generalization.
\end{itemize}

\section{Publications Related to This Thesis}
The results in this thesis are based on the following papers (* denotes equal contributions): 

\begin{itemize}
	\item Part~\ref{P1} Chapter~\ref{CH:CIGA} is based on the publication \cite{ciga}:
 
	\textbf{Chen, Y.}, Zhang, Y., Bian, Y., Yang, H., Ma, K., Xie, B., Liu, T., Han, B., and Cheng, J.\newblock Learning Causally Invariant Representations for Out-of-Distribution Generalization on Graphs, \textbf{Spotlight} in \textit{Neural Information Processing Systems} (NeurIPS), pages 22131--22148, 2022.

    \item Part~\ref{P1} Chapter~\ref{CH:GALA} is based on the publication \cite{gala}:
 
	\textbf{Chen, Y.}, Bian, Y., Zhou, K., Xie, B., Han, B., and Cheng, J.\newblock Does Invariant Graph Learning via Environment Augmentation Learn Invariance? in \textit{Neural Information Processing Systems} (NeurIPS), pages 71486--71519, 2023.

    \item Part~\ref{P2} Chapter~\ref{CH:GMT} is based on the publication \cite{gmt}:
 
	\textbf{Y. Chen}, Y. Bian, B. Han, and J. {Cheng}.\newblock How Interpretable Are Interpretable Graph Neural Networks? in \textit{International Conference on Machine Learning} (ICML), 2024.

    \item Part~\ref{P2} Chapter~\ref{CH:HAO} is based on the publication \cite{hao}:
 
	\textbf{Y. Chen}, H. Yang, Y. Zhang, K. Ma, T. Liu, B. Han, and J. {Cheng}.\newblock Understanding and Improving Graph Injection Attack by Promoting Unnoticeability, in \textit{International Conference on Learning Representations} (ICLR), 2022.

    \item Part~\ref{P3} Chapter~\ref{CH:PAIR} is based on the publication \cite{pair}:
 
	\textbf{Y. Chen}, Zhou, K., Bian, Y., Xie, B., Wu, B., Zhang, Y., Ma, K., Yang, H., Zhao, P., Han, B., and J. {Cheng}.\newblock Pareto Invariant Risk Minimization: Towards Mitigating the Optimization Dilemma in OOD Generalization, in \textit{International Conference on Learning Representations} (ICLR), 2023.

    \item Part~\ref{P3} Chapter~\ref{CH:FeAT} is based on the publication \cite{feat}:
 
	\textbf{Chen, Y.}*, W. Huang*, K. Zhou*, Y. Bian, B. Han, and J. {Cheng}.\newblock Understanding and Improving Feature Learning for Out-of-Distribution Generalization, in \textit{Neural Information Processing Systems} (NeurIPS), pages 68221--68275, 2023.
\end{itemize}

Besides, the following lists the other publications related to this thesis, but left out for a clearer story (in chronological order):

\begin{itemize}
    \item A new principle called spurious infomax to improve graph invariance learning.
    
    T.~Yao*, \textbf{Y.~Chen}*, Z.~Chen, K.~Hu, Z.~Shen and K.~Zhang.
	Empowering Graph Invariance Learning with Deep Spurious Infomax, in \textit{International Conference on Machine Learning} (ICML), 2024.
 
    \item Improving the OOD generalization of decision transformer in solving the network collapse problem.
    
    K.~Ma, H.~Yang, S.~Yang, K.~Zhao, L.~Li, \textbf{Y.~Chen}, J.~Huang, J.~Cheng and Y.~Rong.
	Solving the Non-Submodular Network Collapse Problems via Decision Transformer, \textit{Neural Networks}, pages 106328, 2024.
 
    \item Incorporating the symmetry to design more generalizable neural nets for subset selection:
    
    B. Xie, Y. Bian, K. Zhou, \textbf{Y. Chen}, P. Zhao, B. Han, W. Meng, and J. {Cheng}. Enhancing Neural Subset Selection: Integrating Background Information into Set Representations, in \textit{International Conference on Learning Representations} (ICLR), 2024.
    
	\item Evolving domain generalization that leverages both domain invariant and domain related features:
 
	B. Xie, \textbf{Y. Chen}, J. Wang, K. Zhou, B. Han, W. Meng, and J. {Cheng}. Enhancing Evolving Domain Generalization through Dynamic Latent Representations, \textbf{Oral presentation} in \textit{Thirty-Eighth AAAI Conference on Artificial Intelligence} (AAAI), 2024.

    \item Improving the generalization of self-supervised graph contrastive learning via calibration:
    
    Kaili. Ma, H. Yang, H. Yang, \textbf{Y. Chen}, and J. {Cheng}.
	Calibrating and Improving Graph Contrastive Learning, in \textit{Transactions on Machine Learning Research} (TMLR), 2023.

    \item Benchmarking OOD generalization for predicting properties of chemical kinetics:
    
    \textbf{Y. Chen}*, Z. Wang*, Y. Duan, W. Li, B. Han, J. {Cheng}, and H. Tong.
	Towards Out-of-Distribution Generalizable Predictions of Chemical Kinetics Properties, \textbf{Oral presentation} in \textit{NeurIPS Workshop on AI for Science}, 2023.

    \item A benchmark and systems for OOD generalization of extraction and normalization of temporal and numerical expressions across different languages:
    
    S. Chen, \textbf{Y. Chen}, and B. Karlsson.
	\newblock Dataset and Baseline System for Multi-lingual Extraction and Normalization of Temporal and Numerical Expressions, in \textit{Microsoft Research Technical Report MSR-TR-2023-9}, 2023.

    \item Improving OOD generalization of GNNs in shape correspondence:
    
    B. Kamhoua, L. Zhang, \textbf{Y. Chen}, H. Yang, K. Ma, B. Han, B. Li, and J. {Cheng}.
	Exact Shape Correspondence via 2D graph convolution, \textbf{Spotlight presentation} in \textit{Advances in Neural Information Processing Systems} (NeurIPS), pages 18072--18087, 2022.

    \item Improving the generalization of GNNs in a self-teaching manner:
    
    H. Yang, X. Yan, X. Dai, \textbf{Y. Chen}, and J. {Cheng}.
	Self-enhanced gnn: Improving graph neural networks using model outputs, in \textit{International Joint Conference on Neural Networks} (IJCNN), 2021.
\end{itemize}

\part{Foundations}\label{P1}
\chapter{Frameworks for Causal Invariance Learning on Graphs} \label{CH:CIGA}
Graph is a general data structure and is ubiquitous in many real-world applications.
This part of the thesis aims to establish a general foundations, including the architectures, objectives and theories for learning causal invariance on graphs. The following two chapters start by introducing the challenges, and developing the principles, basic theoretical and practical framework for learn causal invariance to tackle the Out-of-Distribution Generalization problem on graphs. Then, Chapter~\ref{CH:GALA} extends the discussion of necessary theoretical assumptions for the feasibility of learning invariant graph representations.

\section{Motivations}
Graph representation learning with graph neural networks (GNNs) has gained great success in tasks involving relational information~\citep{gcn,sage,gat,jknet,gin}.
However, it assumes that the training and test graphs are drawn from the same distribution, which
is often violated in reality~\citep{ogb,wilds,TDS,drugood}.
The mismatch between training and test distributions, i.e., \textit{distribution shifts},
introduced by some underlying environmental factors related to data collection or processing,
could seriously degrade the performance of deployed models~\citep{camel_example,covid19_application}.
Such \textit{out-of-distribution} (OOD) generalization failures become the major roadblock for practical applications of graph representation learning~\citep{drugood}.

Meanwhile, enabling OOD generalization on regular Euclidean data has received surging attention and several solutions were proposed~\citep{irmv1,groupdro,meta-transfer,vrex,env_inference,ood_max_inv,ib-irm}.
In particular, the invariance principle from causality is at the heart of those works~\citep{inv_principle,causality,causal_transfer}.
The principle leverages the Independent Causal Mechanism (ICM) assumption~\citep{causality,elements_ci} and implies that,
model predictions that only focus on the causes of the label can stay invariant to a large class of distribution shifts~\citep{inv_principle,irmv1}.

Despite the success of the invariance principle on Euclidean data,
the complex nature of graphs raises several new challenges that
prohibit direct adoptions of the principle.
First, distribution shifts on graphs are more complicated.
They can happen at both attribute-level and  structure-level,
and be observed in multiple forms such as graph sizes, subgraph densities and homophily~\citep{size_gen1,size_gen2,reliable_gnn_survey}.
On the other hand, each of the shifts can spuriously correlate with labels in different modes~\citep{irmv1,failure_modes,ib-irm}.
Consequently, the entangled complex distribution shifts make it more difficult to identify and capture the invariance on graphs.
Second, OOD algorithms developed and analyzed on Euclidean data
often require additional environment (or domain) labels for distinguishing the sources of distribution shifts~\citep{irmv1}.
However, the environment labels could be highly expensive to obtain and thus often unavailable for graphs,
as collecting the labels usually requires expert knowledge due to the abstraction of graphs~\citep{ogb}.
These challenges render the problem studied in this chapter even more challenging:
\begin{myquotation}
	\emph{How could one generalize the invariance principle to enable OOD generalization on graphs?}
\end{myquotation}

\begin{figure}[t]
	\centering
	\includegraphics[width=0.8\textwidth]{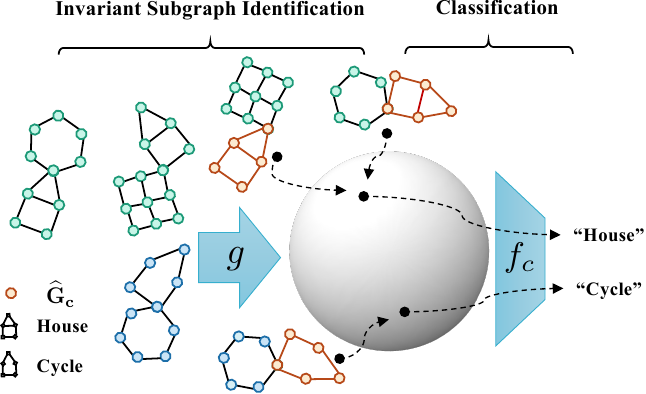}
	\caption[Illustration of \cigafull (\ciga).]{Illustration of \cigafull (\ciga):
		GNNs need to classify graphs based on the specific motif (``House'' or ``Cycle'').
		The featurizer $g$ will extract an (orange-colored) subgraph $\widehat{G}_c$ from each input
		for the classifier $f_c$ to predict the label.
		The training objective of $g$ is implemented in a contrastive strategy where
		the distribution of $\widehat{G}_c$ at the latent sphere
		will be optimized to maximize the intra-class mutual information, hence predictions will be invariant to distribution shifts.}
	\label{CH:CIGA:fig:motivation}
\end{figure}

To solve the above problem, we propose \cigafull (\ciga), a new framework for capturing the invariance of graphs to enable guaranteed OOD generalization under different distribution shifts.
Specifically, we build three Structural Causal Models (SCMs)~\citep{causality}
to characterize the distribution shifts that could happen on graphs:
one is to model the graph generation process, and the other two are to model two possible
interactions between invariant and spurious features during the graph generation,
i.e., Fully Informative Invariant Feature (FIIF)
and Partially Informative Invariant Feature (PIIF) (Sec.~\ref{CH:CIGA:sec:data_gen}).
Then, we generalize the invariance principle to graphs for OOD generalization:
GNN models are invariant to distribution shifts
if they focus only on an invariant and critical subgraph $G_c$
that contains most of the information in $G$ about the underlying causes of the label.
Thus, the problem of achieving OOD generalization on graphs can be rephrased into two processes:
invariant subgraph identification and label prediction.
Accordingly, shown as Fig.~\ref{CH:CIGA:fig:motivation}, we introduce a prototypical invariant graph learning algorithm that decomposes a GNN into:
a) a featurizer $g$ for identifying the underlying invariant subgraph $G_c$ from $G$;
b) a classifier $f_c$ for making predictions based on $G_c$.
To extract the desired subgraph $G_c$, we derive an information-theoretic objective for the featurizer to identify subgraphs that maximally preserve the invariant intra-class information
across a set of different (unknown) environments.
We theoretically show that this approach can provably identify the underlying $G_c$ under mild assumptions (Sec.~\ref{CH:CIGA:sec:good_framework}).

Experiments on $16$ synthetic and real-world datasets with various distribution shifts,
including a challenging setting from AI-aided drug discovery~\citep{drugood},
show that \ciga can significantly outperform all of the existing methods
up to $10\%$, demonstrating its promising OOD generalization ability (Sec.~\ref{CH:CIGA:sec:exp}).

\section{OOD Generalization on Graphs through the Lens of Causality}
\label{CH:CIGA:sec:graph_ood_causal_lens}
It is known that OOD generalization is impossible without assumptions on the environments $\envall$~\citep{causality,ib-irm}.
Thus, we will first formulate the data generation process with structural causal models
and latent-variable model~\citep{causality,elements_ci,ssl_isolate},
to characterize the distribution shifts that could happen on graphs.
Then, we investigate whether the existing methods are generalizable under these distribution shifts.
We also provide a more detailed introduction of the necessary background in Appendix~\ref{CH:CIGA:sec:background_appdx}.

\begin{figure}[t]
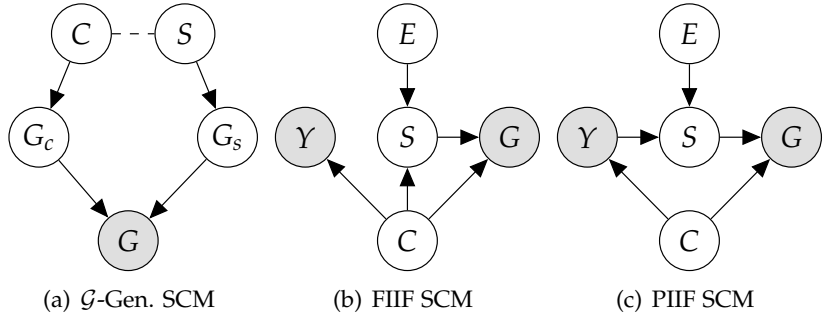

	\centering
	\subfigure[$\gG$-Gen. SCM]{\label{CH:CIGA:fig:graph_gen}
		\resizebox{!}{0.23\textwidth}{\tikz{
				\node[latent] (S) {$S$};%
				\node[latent,left=of S,xshift=0.5cm] (C) {$C$};%
				\node[latent,below=of C,xshift=-0.5cm,yshift=0.5cm] (GC) {$G_c$}; %
				\node[latent,below=of S,xshift=0.5cm,yshift=0.5cm] (GS) {$G_s$}; %
				\node[obs,below=of GC,xshift=1.05cm,yshift=0.5cm] (G) {$G$}; %
				\edge[dashed,-] {C} {S}
				\edge {C} {GC}
				\edge {S} {GS}
				\edge {GC,GS} {G}
			}}}
	\subfigure[FIIF SCM]{\label{CH:CIGA:fig:scm_fiif}
		\resizebox{!}{0.23\textwidth}{\tikz{
				\node[latent] (E) {$E$};%
				\node[latent,below=of E,yshift=0.5cm] (S) {$S$}; %
				\node[obs,below=of E,xshift=-1.2cm,yshift=0.5cm] (Y) {$Y$}; %
				\node[obs,below=of E,xshift=1.2cm,yshift=0.5cm] (G) {$G$}; %
				\node[latent,below=of Y,xshift=1.2cm,yshift=0.5cm] (C) {$C$}; %
				\edge {E} {S}
				\edge {C} {Y,G}
				\edge {S} {G}
				\edge {C} {S}
			}}}
	\subfigure[PIIF SCM]{\label{CH:CIGA:fig:scm_piif}
		\resizebox{!}{0.23\textwidth}{\tikz{
				\node[latent] (E) {$E$};%
				\node[latent,below=of E,yshift=0.5cm] (S) {$S$}; %
				\node[obs,below=of E,xshift=-1.2cm,yshift=0.5cm] (Y) {$Y$}; %
				\node[obs,below=of E,xshift=1.2cm,yshift=0.5cm] (G) {$G$}; %
				\node[latent,below=of Y,xshift=1.2cm,yshift=0.5cm] (C) {$C$}; %
				\edge {E} {S}
				\edge {C} {Y,G}
				\edge {S} {G}
				\edge {Y} {S}
			}}}
	\caption{
		SCMs on graph distribution shifts.}
	\label{CH:CIGA:fig:scm}
\end{figure}

\subsection{Graph Generation Process}
\label{CH:CIGA:sec:data_gen}
We take a latent-variable model perspective on the graph generation process and assume
that the graph is generated through a mapping $f_\gen:\gZ\rightarrow \gG$,
where $\gZ\subseteq\R^n$ is the latent space and $\gG=\cup_{N=1}^\infty\{0,1\}^N\times \R^{N\times d}$ is the graph space.
Let $E$ denote environments.
Following previous works~\citep{ssl_isolate,ib-irm},
we partition the latent variable from $\gZ$ into an invariant part $C\in\gC=\R^{n_c}$ %
and a varying part $S\in\gS=\R^{n_s}$, s.t., $n=n_c+n_s$,
according to whether they are affected by $E$ or not.
Similarly in images, $C$ and $S$ can represent content and style
while $E$ can refer to the locations where the images are taken~\citep{camel_example,adv_causal_lens,ssl_isolate}.
Furthermore, $C$ and $S$ control the generation of the observed graphs (Assumption~\ref{CH:CIGA:assump:graph_gen}) and
can have multiple types of interactions at the latent space (Assumptions~\ref{CH:CIGA:assump:scm_fiif},~\ref{CH:CIGA:assump:scm_piif}).

\paragraph{Graph generation model.} We elaborate the SCM for the graph generation process in Assumption~\ref{CH:CIGA:assump:graph_gen}
and Fig.~\ref{CH:CIGA:fig:graph_gen}, where noises in the structural equations are omitted for simplicity~\citep{elements_ci}.
\begin{assumption}[Graph Generation Structural Causal Model]
	\label{CH:CIGA:assump:graph_gen}
	\[
		G_c:=f_\gen^{G_c}(C),\qquad G_s:=f_\gen^{G_s}(S),\qquad G:=f_\gen^G(G_c,G_s).
	\]
\end{assumption}
In Assumption~\ref{CH:CIGA:assump:graph_gen},
$f_\gen$ is decomposed into $f_\gen^{G_c}$, $f_\gen^{G_s}$ and $f_\gen^G$ to
control the generation of $G_c$, $G_s$, and $G$, respectively.
Among them, $G_c$ inherits the invariant information of $C$ that would not be affected by the interventions (or changes) of $E$~\citep{causality,elements_ci}.
For example, certain properties of a molecule can usually be described by a sub-molecule, or a functional group,
which is invariant across different species or assays~\citep{art_drug,zinc15,drugood}.
On the contrary, the generation of $G_s$ and $G$ will be affected by
the environment $E$ through $S$.
Thus, graphs collected from different environments (or domains) can have different distributions of
structure-level properties (e.g., graph sizes~\citep{size_gen2,reliable_gnn_survey})
as well as feature-level properties (e.g., homophily~\citep{homophily1_birds,hao}).
Therefore, the subgraph $G_s$ inherits the spurious feature about $Y$~\citep{adv_causal_lens}.
In fact, Assumption~\ref{CH:CIGA:assump:graph_gen} is compatible with many graph generation models
by specifying the function classes of $f_\gen^{G_c}$, $f_\gen^{G_s}$ and $f_\gen^G$~\citep{sbm,graphon,graphrnn,graphdf}.
Since our goal is to characterize the potential distribution shifts
in Assumption~\ref{CH:CIGA:assump:graph_gen},
we focus on building a general SCM that is compatible to many graph families
and leave graph family specifications and their implications to OOD generalization in future works. More discussions are provided in Appendix~\ref{CH:CIGA:sec:full_scm_appdx}.

\paragraph{Interactions at latent space.}
Following previous works~\citep{irmv1,ib-irm},
we categorize the latent interactions between $C$ and $S$
into Fully Informative Invariant Features (FIIF, Fig.~\ref{CH:CIGA:fig:scm_fiif})
and Partially Informative Invariant Features (PIIF, Fig.~\ref{CH:CIGA:fig:scm_piif})\footnote{Note that FIIF and PIIF can be mixed as Mixed Informative Invariant Features (Appendix~\ref{CH:CIGA:fig:scm_miif_appdx}) in several ways, while our analysis will focus on the axiom ones for the purpose of generality.},
depending on whether the latent invariant part $C$
is fully informative about label $Y$, i.e., $(S,E)\ind Y|C$.
Formal definitions of the corresponding SCMs are given as follows,
where noises are omitted for simplicity~\citep{causality,elements_ci}.
\begin{assumption}[FIIF Structural Causal Model]
	\label{CH:CIGA:assump:scm_fiif}\[Y:= f_\inv(C),\ S:=f_\spu(C,E),\ G:= f_\gen(C,S).\]
\end{assumption}

\begin{assumption}[PIIF Structural Causal Model]
	\label{CH:CIGA:assump:scm_piif}\[Y:= f_\inv(C),\ S:=f_\spu(Y,E),\ G:= f_\gen(C,S).\]
\end{assumption}
In the two SCMs above,
$f_\gen$ corresponds to the graph generation process in Assumption~\ref{CH:CIGA:assump:graph_gen}, and
$f_\spu$ is the mechanism describing how $S$ is affected by $C$ and $E$ at the latent space.
By definition,
$S$ is directly controlled by $C$ in FIIF and
indirectly controlled by $C$ through $Y$  in PIIF,
which can exhibit different behaviors in the observed distribution shifts.
In practice, performances of OOD algorithms can degrade dramatically if one of FIIF or PIIF is excluded~\citep{aubin2021linear,failure_modes}.
This issue can be more serious in graphs, since different distribution shifts can have different interaction modes at the latent space.
Moreover, $f_\inv:\gC\rightarrow\gY$ indicates the labelling process,
which assigns labels $Y$ for the corresponding $G$ merely based on $C$.
Consequently, $\gC$ is better clustered than $\gS$ when given $Y$~\citep{cluster_assump,cluster_assump2,causality4ml,towards_causality},
which also serves as the necessary separation assumption for a classification task~\citep{svm1,svm2,lda}.
\begin{assumption}[Better Clustered Invariant Features]
	\label{CH:CIGA:assump:latent_sep}$H(C|Y)\leq H(S|Y)$.
\end{assumption}

\subsection{Challenges of OOD Generalization on Graphs}
\label{CH:CIGA:sec:limitation_prev}

Built upon the graph generation process,
we can formally derive the desired GNN that is able to generalize to OOD graphs under different distribution shifts, which implies the invariant GNN below\footnote{A discussion on Def. 2.5 and its relation to the SCMs is provided in Appendix~\ref{CH:CIGA:sec:inv_gnn_discuss_appdx}.}.
\begin{definition}[Invariant GNN]
	\label{CH:CIGA:def:inv_gnn}
	Given a set of graph datasets $\{\dataset^e\}_e$ %
	and environments $\envall$ that follow the same graph generation process in Sec.~\ref{CH:CIGA:sec:data_gen},
	considering a GNN $\rho \circ h$ that has a permutation invariant graph
	encoder $h:\gG\rightarrow\R^h$ and a downstream classifier $\rho:\R^h\rightarrow\gY$,
	$\rho \circ h$ is an invariant GNN if it minimizes the worst case  risk
	among all environments, i.e., $\min \max_{e\in\envall}R^e$.
\end{definition}

Can existing methods produce a desired invariant GNN model?
We find the answers to be negative, unfortunately.
\begin{figure}[t]
	\subfigure[Failure cases for existing methods.]{
		\includegraphics[width=0.36\textwidth]{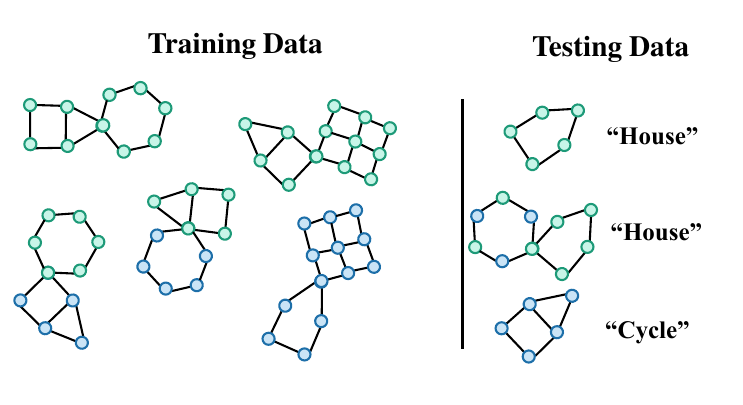}
		\label{CH:CIGA:fig:good_fail_cases}
	}
	\subfigure[Structure and attribute shifts.]{
		\includegraphics[width=0.28\textwidth]{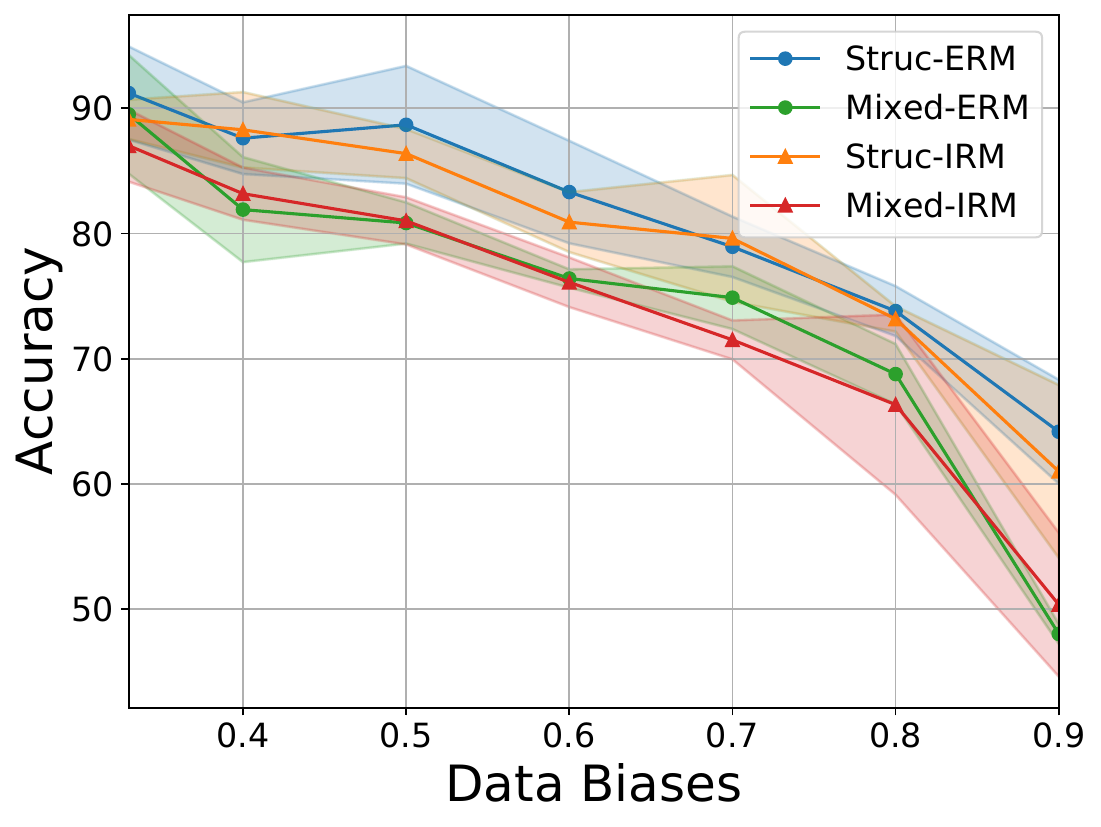}
		\label{CH:CIGA:fig:ood_failure_wosize}
	}
	\subfigure[Mixed with graph size shifts.]{
		\includegraphics[width=0.28\textwidth]{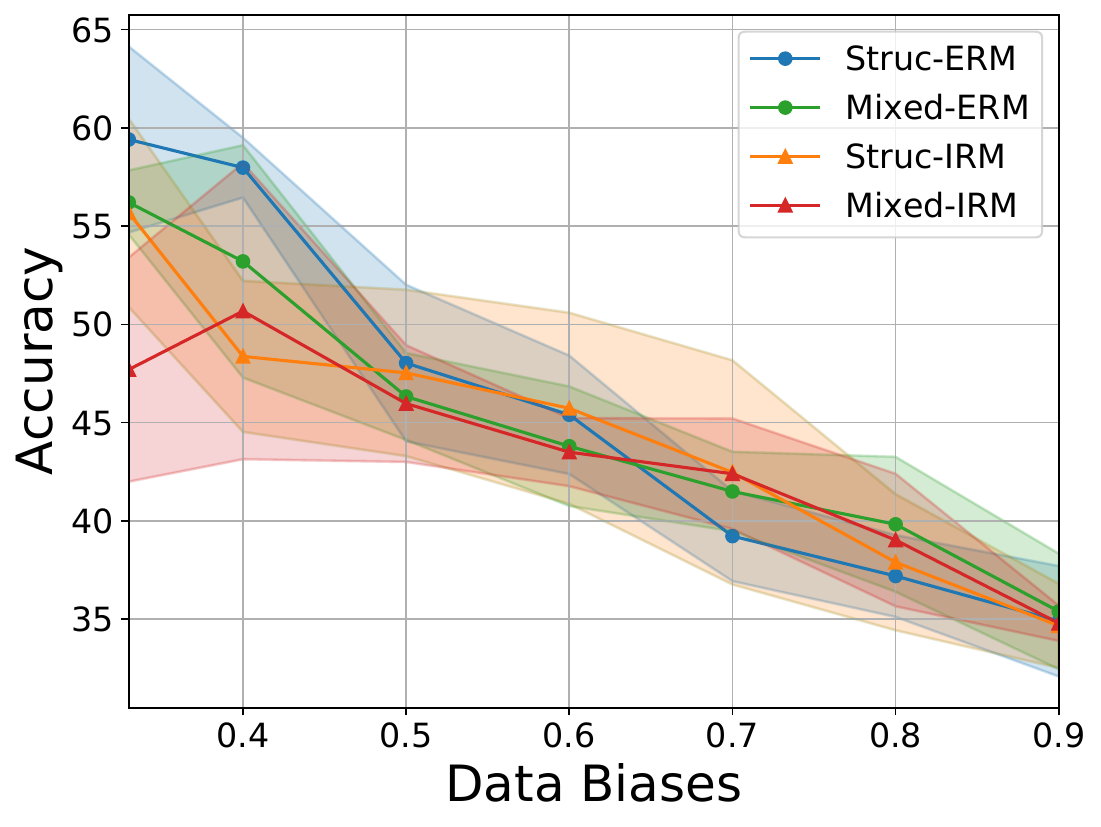}
		\label{CH:CIGA:fig:ood_failure_size}
	}
	\caption[Failures of OOD generalization on graphs.]{Failures of OOD generalization on graphs:
		(a) GNNs are required to classify whether the graph contains a ``house'' or ``cycle'' motif,
		where the colors represent node features.
		However, distribution shifts in the training data exist at both structure-level (from left to right: ``house'' mostly co-occur with a hexagon),
		attribute-level (from upper to lower: nodes are mostly  colored green if the graph contains a ``house'', or  colored blue if the graph contains a ``cycle''),
		and graph sizes, making GNNs hard to capture the invariance. Consequently,
		\textit{ERM can fail} for leveraging the shortcuts and predicting graphs that have a hexagon or have nodes mostly colored green as ``house''.
		\textit{IRM can fail} as the test data are not sufficiently supported by the training data.
		(b) GCNs optimized with neither ERM nor IRM can generalize to OOD graphs under
		structure-level shifts (Struc-) or mixed with feature shifts (Mixed-).
		(c) When more complex shifts are presented, GNNs can fail more seriously.}
	\label{CH:CIGA:fig:good_fail}
\end{figure}
Based on the synthetic BAMotif graph classification task~\citep{pge,dir} shown in Fig.~\ref{CH:CIGA:fig:good_fail},
we theoretically and empirically analyze whether existing methods
could produce an invariant GNN, through the investigation of the following aspects.
More details and results are given in Appendix~\ref{CH:CIGA:sec:good_fail_setting_appdx}.

\paragraph{Can GNNs trained with ERM generalize to OOD graphs?}
As shown in Fig.~\ref{CH:CIGA:fig:good_fail}, we find that GNNs trained with the
standard empirical risk minimization (ERM) algorithm~\citep{erm}
are not able to generalize to OOD graphs.
As the data biases grows stronger,
the performances of GNNs drop dramatically.
Furthermore, when graph size shifts are mixed in the data,
GNNs can have larger variance at low data biases, indicating
the instability of learning the desired relationships for the task.
The reason is that ERM tends to overfit to the shortcuts
or spurious correlations presented in specific substructures or attributes
in the graphs~\citep{shortcut_dl}.
This phenomenon has also been shown to exist
in GNNs equipped with more sophisticated architectures such as attention mechanisms~\citep{gat},
under graph size shifts~\citep{understand_att}.

\paragraph{Can OOD objectives improve OOD generalization of GNNs?}
Meanwhile, as shown in Fig.~\ref{CH:CIGA:fig:good_fail}, OOD objectives primarily developed on Euclidean data such as
invariant risk minimization (IRM)~\citep{irmv1}
also cannot alleviate the problem. On the contrary, IRM can
fail catastrophically at non-linear regime if without sufficient support overlap
for the test environments,
i.e., $\cup_{e\in\envtest}\text{supp}(\sP^e)\not\subseteq\cup_{e\in\envtrain}\text{supp}(\sP^e)$~\citep{risk_irm}.
In addition to IRM, the failure would also happen for alternative objectives~\citep{vrex,gen_inv_conf,ib-irm} as proved by~\citet{risk_irm}.
Besides, different distribution shifts on graphs can be nested with each other where
each one can have distinct spurious correlation type, e.g., FIIF or PIIF.
OOD objectives will also fail seriously if either of the correlation types is not supported~\citep{aubin2021linear,failure_modes}.
Moreover, non-trivial environment partitions or labels
are required for performance guarantee of these OOD objectives~\citep{irmv1,vrex,groupdro,ib-irm}.
However, collecting meaningful environment partitions of graphs
requires expert knowledge about graph data.
Thus, the environment labels can be expensive to obtain
and are usually not available~\citep{tudataset,benchmark_gnn,ogb}.
Alternative options such as random partitions
tend not to alleviate the issue~\citep{env_inference,zin}, as it can be trivially deemed as mini-batching.

\paragraph{Challenges of OOD generalization on graphs.}
The aforementioned failure analysis reveals that existing methods or objectives fail to elicit an invariant GNN primarily due to the following two challenges:
a) Distribution shifts on graphs are more complicated where different types of spurious correlations can be entangled via different graph properties;
b) Environment labels are usually not available due to the abstraction of graphs.
Despite these challenges, we are still highly motivated to address the following research  question:
\emph{Would it be possible to learn an invariant GNN that is generalizable under various distribution shifts by lifting the invariance principle to the graph data?}

\section{Invariance Principle for OOD Generalization on Graphs}
\label{CH:CIGA:sec:good_framework}
We provide affirmative answers to the previous question
by proposing a new framework, \ciga: \cigafull. Specifically,
built upon the SCMs in Sec.~\ref{CH:CIGA:sec:data_gen},
we generalize the invariance principle to graphs
and instantiate the principle with theoretical guarantees.

\subsection{Invariance for OOD Generalization on Graphs}
Towards extending the invariance principle to graphs under SCMs in Sec.~\ref{CH:CIGA:sec:data_gen},
we need to identify a set of variables that have stable causal relationship with $Y$
under both FIIF and PIIF (Assumption~\ref{CH:CIGA:assump:scm_fiif},~\ref{CH:CIGA:assump:scm_piif}).
According to the ICM assumption~\citep{elements_ci}, the labeling process $C\rightarrow Y$ is not informed nor influenced by other processes, implying that the conditional distribution $P(Y|C)$ remains invariant to the interventions on the environment latent variable $E$~\citep{causality}.
Consequently, for a GNN with a permutation invariant encoder $h:\gG\rightarrow \R^h$
and a downstream classifier $\rho:\R^h\rightarrow\gY$,
if $h$ can recover the information of $C$ from $G$ in the learned graph representations,
then the learning of $\rho$ resembles traditional ERM~\citep{erm}
and can achieve the desired min-max optimality required by an invariant GNN (Def.~\ref{CH:CIGA:def:inv_gnn}).
However, recovering $C$ from $G$ is particularly difficult,
since the generation of $G$ from $C$ involves two causal mechanisms $f_\gen^{G_c}$ and $f_\gen^G$
in Assumption~\ref{CH:CIGA:assump:graph_gen}.
The unavailability of $E$ further adds up the difficulty of enforcing the
independence between the learned representations and $E$.

\subsection{Invariant Graph Learning Framework}
\label{CH:CIGA:sec:good_framework_prac}

\paragraph{Causal algorithmic alignment.}
To enable a GNN to learn to extract the information about $C$ from $G$, we propose the \ciga framework that \mbox{\textit{explicitly aligns with}} the
two causal mechanisms $f_\gen^{G_c}$ and $f_\gen^G$ in Assumption~\ref{CH:CIGA:assump:graph_gen}.
The idea of alignment in \ciga is motivated by the algorithmic reasoning results
that a neural network can learn a reasoning process better if its computation structure aligns with the process better~\citep{what_nn_reason,nn_extrapo}.
Specifically, we realize the alignment by decomposing a GNN into two sub-components:\footnote{The encoder of the GNN in \ciga can be regarded as the composition of $g$ and the graph encoder in $f_c$.}
a) a featurizer GNN $g:\gG\rightarrow\gG_c$ aiming to identify the desired $G_c$;
b) a classifier GNN $f_c:\gG_c\rightarrow\gY$ that predicts the label $Y$ based on the estimated $G_c$,
where $\gG_c$ refers to the space of subgraphs of $G$.
Formally, the learning objectives of $f_c$ and $g$ can be formulated as:
\begin{equation}
	\label{CH:CIGA:good_opt}
	\text{$\max$}_{f_c, \; g} \ I(\widehat{G}_{c};Y), \ \text{s.t.}\ \widehat{G}_{c}\ind E,\ \widehat{G}_{c}=g(G),
\end{equation}
where maximizing $I(\widehat{G}_c;Y)$ is equivalent
to minimizing a variational upper bound of $R(f_c(\widehat{G}_{c}))$~\citep{vib,gib} that takes $\widehat{G}_{c}$ as inputs to predict label $Y$ for $G$ through $f_c$ and $g$,
and $\widehat{G}_{c}$ is the estimated subgraph containing the information about $C$ and hence needs to be independent of $E$.
Moreover, the extracted $G_c$ can either shares the same graph space with input $G$ or
has its own space with latent node and edge features, depending on the specific graph generation process.
In practice, architectures from the literature of interpretable GNNs are compatible with \ciga~\citep{xgnn_tax}, hence can serve as practical choices for the implementation of \ciga.
More details are given in Appendix~\ref{CH:CIGA:sec:good_impl_appdx}.

Although we can technically align with the two causal mechanisms with $g$ and $f_c$,
trivially optimizing this architecture cannot satisfy $\widehat{G}_{c}\ind E$.
Formally, merely maximizing $I(\widehat{G}_{c};Y)$
may include a subgraph from $G_s$ in $\widehat{G}_{c}$ since $G_s$ also shares certain mutual information with $Y$.
Moreover, the unavailability of $E$ prevents the direct usage of $E$ in enforcing the independence
that is often adopted by previous methods~\citep{irmv1,vrex,groupdro,DANN,CORAL}, making the identification of $G_c$ more challenging.

\paragraph{Optimization objective.}
To mitigate this issue,
we need to find and translate other properties of $G_c$
into some differentiable and equivalent objectives to satisfy the independence constraint $\widehat{G}_{c}\ind E$.

We begin by considering a simplistic setting where all the invariant subgraphs $G_c$ have the
same size $s_c$, i.e., $|G_c|=s_c$\footnote{Throughout the paper, we use generalized set operators for the ease of understanding. They can have multiple implementations in terms of nodes, edges or attributes.}.
When maximizing $I(\widehat{G}_{c};Y)$ in Eq.~\ref{CH:CIGA:good_opt},
both FIIF and PIIF can introduce part of $G_s$ into $\widehat{G}_{c}$.
In FIIF (Fig.~\ref{CH:CIGA:fig:scm_fiif}), as $G_c$ already contains the maximal possible information in $G$ about $Y$,
$G_c$ is a solution to $\max I(\widehat{G}_{c};Y)$. However, some subgraph of $G_c$ can be replaced by some subgraph of $G_s$ that is equally informative about $Y$.
In PIIF (Fig.~\ref{CH:CIGA:fig:scm_piif}), there also exists some subgraph of $G_s$ that
contains additional information about $Y$ than $G_c$, hence $\widehat{G}_{c}$ is more likely
to involve some subgraph of $G_s$.
Thus, the new objective needs to eliminate the auxiliary subgraphs of $\widehat{G}_{c}$ from $G_s$ such that the estimated $\widehat{G}_{c}$ can only contain $G_c$.

\paragraph{An important property of $G_c$.}
Under both FIIF and PIIF SCMs (Fig.~\ref{CH:CIGA:fig:scm}),
for $G_c^{e_1}$, $G_c^{e_2}$ that relate to the same causal factor $c$
under two environments $e_1$ and $e_2$,
the desired $\widehat{G}_c^{e_1}, \widehat{G}_c^{e_2}$
in $e_1$ and $e_2$ tend to have high mutual information, i.e.,
$(G_c^{e_1},G_c^{e_2})\in \argmax I(\widehat{G}_c^{e_1}; \widehat{G}_c^{e_2})$.
While for $G_c^{e_1}$ and another $G_{c'}^{e_1}$ corresponding to a different $c'\neq c$,
under the same environment $e_1$,
including any subgraph from $G_s^{e_1}$ in $\widehat{G}_c^{e_1},\widehat{G}_{c'}^{e_1}$
will enlarge their mutual information, or in other words,
$(G_c^{e_1},G_{c'}^{e_1})\in \argmin I(\widehat{G}_c^{e_1}; \widehat{G}_{c'}^{e_1})$.
Thus, we can derive an important property of $G_c$, that is, $\forall e_1,e_2\in\envall$,
\begin{equation}
	\label{CH:CIGA:good_cond_1}
	G_c^{e_1}\in \text{$\argmax$}_{\widehat{G}_c^{e_1}}\  I(\widehat{G}_c^{e_1};\widehat{G}_c^{e_2}|C=c)-I(\widehat{G}_c^{e_1};\widehat{G}_{c'}^{e_2}|C=c',c'\neq c),
\end{equation}
where $\widehat{G}_c^{e_1}$ and $\widehat{G}_c^{e_2}$ are the estimated
invariant subgraphs corresponding to the same causal factor $c$ under
environment $e_1$ and $e_2$, respectively,
while $\widehat{G}_{c'}^{e_2}$ corresponds to a different causal factor $c'$.

\paragraph{Deriving CIGAv1 based on the identified property of $G_c$.}
In practice, $C$ is not given.
Nevertheless, since $C$ and $Y$ shares a stable causal relationship in both FIIF and PIIF SCMs,
$Y$ can serve as a proxy of $C$ in Eq.~\ref{CH:CIGA:good_cond_1}.
Moreover, as Eq.~\ref{CH:CIGA:good_cond_1} holds for any $\forall e_1,e_2\in\envall$,
the environment superscripts can be eliminated without affecting Eq.~\ref{CH:CIGA:good_cond_1}.
Furthermore,
when both $I(\widehat{G}^{e_1}_{c};\widehat{G}^{e_2}_c|C=c)$ and $I(\widehat{G}_{c};Y)$ are maximized,
$I(\widehat{G}^{e_1}_{c};\widehat{G}^{e_1}_{c'}|C=c',c'\neq c)$ is automatically minimized,
otherwise, all classes will collapse to trivial solutions which is contradictory given
$I(\widehat{G}_{c};Y)$ being maximized.
Therefore, we can derive an alternative objective to Eq.~\ref{CH:CIGA:good_opt}
by leveraging Eq.~\ref{CH:CIGA:good_cond_1}
to replace the independence condition:
\begin{snugshade}
	\vspace{-0.1in}
	\begin{equation}
		\label{CH:CIGA:good_opt_contrast_v1}
		(\text{CIGAv1})\qquad\qquad\qquad \ %
		\max_{f_c, g} \ I(\widehat{G}_{c};Y), \ \text{s.t.}\
		\widehat{G}_{c}\in\argmax_{\widehat{G}_{c}=g(G), |\widehat{G}_{c}|\leq s_c} I(\widehat{G}_{c};\widetilde{G}_c|Y),
		\qquad%
	\end{equation}
\end{snugshade}
\noindent where $\widetilde{G}_c=g(\widetilde{G})$ and $\widetilde{G}\sim \sP(G|Y)$,
i.e., $\widetilde{G}$ is sampled from training graphs that share the same label $Y$ as $G$.
In Theorem~\ref{CH:CIGA:thm:good_inv_gnn_new}, we show how Eq.~\ref{CH:CIGA:good_opt_contrast_v1}
is equivalent to Eq.~\ref{CH:CIGA:good_opt}.
Nevertheless, Eq.~\ref{CH:CIGA:good_opt_contrast_v1} requires a strong
assumption on the size of $G_c$.
However, the size of $G_c$ is usually unknown or changes for different $C$s.
In this circumstance, maximizing Eq.~\ref{CH:CIGA:good_cond_1} without additional constraints will lead to the presence of part of $G_s$ in $\widehat{G}_{c}$.
For instance, $\widehat{G}_{c}=G$ is a trivial solution to Eq.~\ref{CH:CIGA:good_opt_contrast_v1}
when $s_c = \infty$.

\paragraph{Deriving CIGAv2 by resolving size constraint on $G_c$ in CIGAv1.}
To this end, we further resort to the properties of $G_s$.
In both FIIF and PIIF SCMs (Fig.~\ref{CH:CIGA:fig:scm}), $G_s$ and $G_c$ can share certain
overlapped information about $Y$. When maximizing $I(\widehat{G}_{c};\widetilde{G}_c|Y)$ and $I(\widehat{G}_{c};Y)$,
the appearance of partial $G_s$ in $\widehat{G}_{c}$ will not affect the optimality.
However, it can reduce the mutual information between the
left part $\widehat{G}_s=G-\widehat{G}_{c}$ and $Y$, i.e., $I(\widehat{G}_s;Y)$.
Therefore, by maximizing $I(\widehat{G}_s;Y)$, we can reduce including part of $G_s$ into $\widehat{G}_{c}$.
Meanwhile, to avoid trivial solution that $G_c\subseteq\widehat{G}_s$ during maximizing $I(\widehat{G}_s;Y)$,
we can leverage the better clustering property of $G_c$ implied by Assumption~\ref{CH:CIGA:assump:latent_sep}
to derive the constraint $I(\widehat{G}_s;Y)\leq I(\widehat{G}_{c};Y)$.
Thus, we can obtain a new objective {\ciga}v2
as follows:
\begin{snugshade}\vspace{-0.1in}
	\begin{equation}
		\label{CH:CIGA:good_opt_contrast_v3}
		\begin{aligned}
			\text{$\max$}_{f_c, g} \  I(\widehat{G}_{c};Y)+I(\widehat{G}_s;Y), \ \text{s.t.}\
			 & \widehat{G}_{c}\in
			\text{$\argmax$}_{\widehat{G}_{c}=g(G)} I(\widehat{G}_{c};\widetilde{G}_c|Y),\
			\\
			(\text{CIGAv2})\qquad\qquad\qquad\qquad\qquad\qquad\ %
			 & I(\widehat{G}_s;Y)\leq I(\widehat{G}_{c};Y),\ \widehat{G}_s=G-g(G),
			\qquad\ \ %
		\end{aligned}
	\end{equation}
\end{snugshade}
\noindent where $\widehat{G}_{c}=g(G),\widetilde{G}_c=g(\widetilde{G})$ and $\widetilde{G}\sim \sP(G|Y)$,
i.e., $\widetilde{G}$ is sampled from training graphs that share the same label $Y$ as $G$.
We also prove the equivalence between Eq.~\ref{CH:CIGA:good_opt_contrast_v3} and Eq.~\ref{CH:CIGA:good_opt}
in Theorem~\ref{CH:CIGA:thm:good_inv_gnn_new}.

\subsection{Theoretical Analysis and Practical Discussions}
\label{CH:CIGA:sec:good_theory}
\begin{theorem}[\ciga Induces Invariant GNNs]
	\label{CH:CIGA:thm:good_inv_gnn_new}
	Given a set of graph datasets $\{\dataset^e\}_e$ %
	and environments $\envall$ that follow the same graph generation process in Sec.~\ref{CH:CIGA:sec:data_gen},
	assuming that \textup{(a)} $f_\gen^G$ and $f_\gen^{G_c}$ in Assumption~\ref{CH:CIGA:assump:graph_gen} are invertible,
	\textup{(b)} samples from each training environment are equally distributed,
	i.e.,$|\dataset_{\hat{e}}|=|\dataset_{\tilde{e}}|,\ \forall \hat{e},\tilde{e}\in\envtrain$, then:
	\begin{enumerate}[label=(\roman*).,wide]
		\item If $\forall G_c, |G_c|=s_c$,
		      then each solution to Eq.~\ref{CH:CIGA:good_opt_contrast_v1},
		      elicits an invariant GNN (Def.~\ref{CH:CIGA:def:inv_gnn}).
		\item Each solution to Eq.~\ref{CH:CIGA:good_opt_contrast_v3},
		      elicits an invariant GNN (Def.~\ref{CH:CIGA:def:inv_gnn}).
	\end{enumerate}
\end{theorem}
We prove Theorem~\ref{CH:CIGA:thm:good_inv_gnn_new} (i) and (ii)
in Appendix~\ref{proof:good_inv_gnn_new_appdx},~\ref{proof:good_inv_gnn_v3_appdx}, respectively.

\paragraph{Practical implementations of \ciga objectives.}
After showing the power of \ciga,
we introduce the practical implementations of {\ciga}v1 and {\ciga}v2 objectives.
Specifically,  an exact estimate of the second term $I(\widehat{G}_{c};\widetilde{G}_c|Y)$ could be highly expensive~\citep{infoNCE,mine}.
However, contrastive learning with supervised sampling provides a practical solution for the approximation~\citep{sup_contrastive,contrast_loss1,contrast_loss2,infoNCE,mine}:
\begin{equation} \label{CH:CIGA:good_opt_contrast}
	I(\widehat{G}_{c};\widetilde{G}_c|Y) \approx
	\mathbb{E}_{
	\substack{
	\{\widehat{G}_{c},\widetilde{G}_c\} \sim \sP_g(G|\gY=Y)\\\
	\{G^i_c\}_{i=1}^{M} \sim \sP_g(G|\gY \neq Y)
	}
	}
	\log\frac{e^{\phi(h_{\widehat{G}_{c}},h_{\widetilde{G}_c})}}
	{e^{\phi(h_{\widehat{G}_{c}},h_{\widetilde{G}_c})} +
		\sum_{i=1}^M e^{\phi(h_{\widehat{G}_{c}},h_{G^i_c})}},
\end{equation}
where positive samples $(\widehat{G}_{c},\widetilde{G}_c)$ are the extracted subgraphs of graphs that share the same label as $G$,
negative samples are those having different labels, $\sP_g(G|\gY=Y)$ is the push-forward distribution of $\sP(G|\gY=Y)$ by featurizer $g$,
$\sP(G|\gY=Y)$ refers to the distribution of $G$ given the label $Y$,
$\sP(G|\gY\neq Y)$ refers to the distribution of $G$ given the label that is different from $Y$,
$h_{\widehat{G}_{c}},h_{\widetilde{G}_c},h_{G^i_c}$ are the graph presentations of the estimated subgraphs,
and $\phi$ is the similarity metric for graph representations.
As $M\rightarrow \infty$, Eq.~\ref{CH:CIGA:good_opt_contrast} approximates $I(\widehat{G}_{c};\widetilde{G}_c|Y)$,
which can be regarded as a
non-parameteric resubstitution entropy estimator via the von Mises-Fisher
kernel density~\citep{feat_dist_entropy,vMF_entropy,align_uniform}.
Thus, plugging it into Eq.~\ref{CH:CIGA:good_opt_contrast_v1} and Eq.~\ref{CH:CIGA:good_opt_contrast_v3}
can relieve the issue of approximating $I(\widehat{G}_{c};\widetilde{G}_c|Y)$ in practice.

For the implementation of $I(\widehat{G}_s;Y)$ and the constraint $I(\widehat{G}_s;Y)\leq I(\widehat{G}_{c};Y)$ in {\ciga}v2,
a practical choice is to follow the idea of hinge loss, $I(\widehat{G}_s;Y)\approx \frac{1}{N} R_{\widehat{G}_s}\cdot \mathbb{I}(R_{\widehat{G}_c}\leq R_{\widehat{G}_{s}})$,
where $N$ is the number of samples, $\mathbb{I}$ is an indicator function that outputs $1$ when the
inner condition is satisfied otherwise $0$, and
$R_{\widehat{G}_s}$ and $R_{\widehat{G}_{c}}$ are the empirical risk vector of the predictions
for each sample based on the corresponding $\widehat{G}_s$ and $\widehat{G}_{c}$.
More implementation details can be found in Appendix~\ref{CH:CIGA:sec:good_impl_appdx}.

\paragraph{Discussions and implications of \ciga.}
Although using contrastive learning to improve OOD generalization is not new
in the literature~\citep{DouCKG19,causal_matching,cnc},
previous methods cannot yield OOD guarantees in graph circumstances due to  the highly non-linearity and the unavailability of domain labels $E$.
In particular, \ciga can \textit{be reduced to directly applying contrastive learning}
when without the decomposition for causal algorithmic alignment.
However, in the experiments we found that merely using the contrastive objective, i.e., CNC~\citep{cnc},
yields unsatisfactory OOD performance,
which further implies the necessity of the decomposition in \ciga.

Moreover, the architecture of \ciga can have multiple other implementations
for both the featurizer and classifier, such as identifying $G_c$ at the latent space~\citep{causality4ml,towards_causality}.
Since we cannot enumerate every possible implementation, in this work
we choose interpretable GNN architectures as a prototype validation for \ciga and leave more sophisticated architectures as future works.
In particular,
when optimized with ERM objective,
\ciga can \textit{be reduced to interpretable GNNs}.
However, merely using interpretable GNNs such as ASAP~\citep{asap}, GIB~\citep{gib} or DIR~\citep{dir} cannot yield satisfactory OOD performance.
As discussed in Appendix.~\ref{CH:CIGA:sec:discussion_ood_obj_appdx},
GIB can only work for FIIF,
while DIR \textit{cannot} yield OOD guarantees for neither FIIF nor PIIF SCMs.
These results are also empirically validated in the experiments.
We provide more detailed discussions in Appendix~\ref{CH:CIGA:sec:discuss_future_appdx}.

\section{Empirical Studies}
\label{CH:CIGA:sec:exp}

We conduct extensive experiments with $16$ datasets
to verify the effectiveness of \ciga.

\paragraph{Datasets.}
We use the SPMotif datasets from DIR~\citep{dir} where
artificial structural shifts and graph size shifts are nested (SPMotif-Struc).
Besides, we  construct a harder version mixed with attribute shifts (SPMotif-Mixed).
To examine \ciga in real-world scenarios with more complicated relationships and distribution shifts,
we also use DrugOOD~\citep{drugood} from AI-aided Drug Discovery with Assay, Scaffold, and Size splits,
convert the ColoredMNIST from IRM~\citep{irmv1} using the algorithm from~\citet{understand_att} to inject attribute shifts,
and split Graph-SST~\citep{xgnn_tax} to inject degree biases.
To compare with previous specialized OOD methods for graph size shifts~\citep{size_gen1,size_gen2},
we use the datasets in~\citet{size_gen2} that are converted from TU benchmarks~\citep{tudataset}.
More details can be found in Appendix~\ref{CH:CIGA:sec:exp_data_appdx}.

\paragraph{Baselines and our methods.} Besides the ERM, we also compare with SOTA interpretable GNNs,
GIB~\citep{gib}, ASAP Pooling~\citep{asap}, and DIR~\citep{dir}, to validate the effectiveness of the optimization objective in \ciga.
We use the same selection ratio (i.e., $s_c$) for all models.
Moreover, to validate the effectiveness of the decomposition in \ciga,
we compare \ciga with SOTA OOD objectives including IRM~\citep{irmv1},
vrex~\citep{vrex} and IB-IRM~\citep{ib-irm}, for which we apply random environment partitions following~\citep{env_inference}.
We also compare \ciga with
EIIL~\citep{env_inference} and CNC~\citep{cnc} that do not require environment labels,
where CNC~\citep{cnc} has a more sophisticated contrastive sampling strategy for combating subpopulation shifts.
More implementation and comparison details are deferred to Appendix~\ref{CH:CIGA:sec:exp_impl_appdx}.

\paragraph{Evaluation.} We report the classification accuracy for all datasets,
except for DrugOOD datasets where we use ROC-AUC following~\citep{drugood}, and
for TU datasets where we use Matthews correlation coefficient following~\citep{size_gen2}.
We repeat the evaluation multiple times, select models based on the validation performances,
and report the mean and standard deviation of the corresponding metric. For each dataset, we also report the ``Oracle'' performances that run ERM on the randomly shuffled data.

\begin{table}[ht]
	\scriptsize
	\caption{OOD generalization performances on structural and mixed shifts for synthetic graphs.}
		\vspace{-0.1in}
	\label{CH:CIGA:tab:sythetic}
	\begin{sc}
		\begin{center}
			\resizebox{\textwidth}{!}{\begin{tabular}{l|rrr|rrr|r}
					\toprule
					                                            &
					\multicolumn{3}{c}{SPMotif-Struc$^\dagger$}
					                                            &
					\multicolumn{3}{c}{SPMotif-Mixed$^\dagger$} &                                                                           \\

					                                            &
					\multicolumn{1}{c}{bias=$0.33$}
					                                            &
					\multicolumn{1}{c}{bias=$0.60$}
					                                            &
					\multicolumn{1}{c}{bias=$0.90$}
					                                            &
					\multicolumn{1}{c}{bias=$0.33$}
					                                            &
					\multicolumn{1}{c}{bias=$0.60$}
					                                            &
					\multicolumn{1}{c}{bias=$0.90$}             & \multicolumn{1}{c}{Avg}
					\\

					\cmidrule(lr){2-4}\cmidrule(lr){5-7}\cmidrule(lr){8-8}
					ERM                                         & 59.49 (3.50)            & 55.48 (4.84)
					                                            & 49.64 (4.63)
					                                            & 58.18 (4.30)            & 49.29 (8.17)
					                                            & 41.36 (3.29)            & 52.24                                           \\
					ASAP                                        & 64.87 (13.8)            & 64.85 (10.6)
					                                            & \textbf{57.29 (14.5)}
					                                            & 66.88 (15.0)            & 59.78 (6.78)
					                                            & \textbf{50.45 (4.90)}   & 60.69                                           \\

					DIR                                         & 58.73 (11.9)            & 48.72 (14.8)
					                                            & 41.90 (9.39)
					                                            & 67.28 (4.06)            & 51.66 (14.1)
					                                            & 38.58 (5.88)            & 51.14                                           \\

					\hline
					\rule{0pt}{8pt}IRM                          & 57.15 (3.98)            & 61.74 (1.32)
					                                            & 45.68 (4.88)
					                                            & 58.20 (1.97)            & 49.29 (3.67)
					                                            & 40.73 (1.93)            & 52.13                                           \\
					vrex                                        & 54.64 (3.05)            & 53.60 (3.74)
					                                            & 48.86 (9.69)
					                                            & 57.82 (5.93)            & 48.25 (2.79)
					                                            & 43.27 (1.32)            & 51.07                                           \\
					EIIL                                        & 56.48 (2.56)            & 60.07 (4.47)
					                                            & 55.79 (6.54)
					                                            & 53.91 (3.15)            & 48.41 (5.53)
					                                            & 41.75 (4.97)            & 52.73                                           \\
					IB-IRM                                      & 58.30 (6.37)            & 54.37 (7.35)
					                                            & 45.14 (4.07)
					                                            & 57.70 (2.11)            & 50.83 (1.51)
					                                            & 40.27 (3.68)            & 51.10                                           \\

					CNC                                         & 70.44 (2.55)            & \textbf{66.79 (9.42)}
					                                            & 50.25 (10.7)
					                                            & 65.75 (4.35)            & 59.27 (5.29)
					                                            & 41.58 (1.90)            & 59.01                                           \\

					\hline
					\rule{0pt}{8pt}\textbf{{\ciga}v1}           & \textbf{71.07 (3.60)}   & 63.23 (9.61)
					                                            & 51.78 (7.29)
					                                            & \textbf{74.35 (1.85)}   & \textbf{64.54 (8.19)}
					                                            & 49.01 (9.92)            & \textbf{62.33}                                  \\
					\textbf{{\ciga}v2}                          & \textbf{77.33 (9.13)}   & \textbf{69.29 (3.06)}
					                                            & \textbf{63.41 (7.38)}
					                                            & \textbf{72.42 (4.80)}   & \textbf{70.83 (7.54)}
					                                            & \textbf{54.25 (5.38)}   & \textbf{67.92}                                  \\
					\hline
					\rule{0pt}{8pt}Oracle (IID)                 &                         & 88.70 (0.17)          &  &  & 88.73 (0.25) &  & \\
					\bottomrule
					\multicolumn{8}{l}{\rule{0pt}{8pt}$^\dagger$\text{\normalfont \small Higher accuracy and lower variance indicate better OOD generalization ability.}  }
				\end{tabular}}
		\end{center}
	\end{sc}
\end{table}
\begin{table}[t]
	\small\centering
	\caption{OOD generalization performances on complex distribution shifts for real-world graphs.}
	\label{table:other_graph}
	\resizebox{\columnwidth}{!}{
		\begin{sc}
			\begin{tabular}{lrrrrrrc}
				\toprule
				Datasets                          & \multicolumn{1}{c}{Drug-Assay} & \multicolumn{1}{c}{Drug-Sca} & \multicolumn{1}{c}{Drug-Size} & \multicolumn{1}{c}{CMNIST-sp} & \multicolumn{1}{c}{Graph-SST5} & \multicolumn{1}{c}{Twitter} & \multicolumn{1}{c}{Avg (Rank)$^\dagger$} \\
				\midrule
				ERM                               & 71.79 (0.27)                   & 68.85 (0.62)                 & 66.70 (1.08)                  & 13.96 (5.48)                  & 43.89 (1.73)                   & 60.81 (2.05)                & 54.33 (6.00)                             \\
				ASAP                              & 70.51 (1.93)                   & 66.19 (0.94)                 & 64.12 (0.67)                  & 10.23 (0.51)                  & 44.16 (1.36)                   & 60.68 (2.10)                & 52.65 (8.33)                             \\
				GIB                               & 63.01 (1.16)                   & 62.01 (1.41)                 & 55.50 (1.42)                  & 15.40 (3.91)                  & 38.64 (4.52)                   & 48.08 (2.27)                & 47.11 (10.0)                             \\
				DIR                               & 68.25 (1.40)                   & 63.91 (1.36)                 & 60.40 (1.42)                  & 15.50 (8.65)                  & 41.12 (1.96)                   & 59.85 (2.98)                & 51.51 (9.33)                             \\\hline
				\rule{0pt}{8pt}IRM                & 72.12 (0.49)                   & 68.69 (0.65)                 & 66.54 (0.42)                  & 31.58 (9.52)                  & 43.69 (1.26)                   & 63.50 (1.23)                & 57.69 (4.50)                             \\
				vrex                              & 72.05 (1.25)                   & 68.92 (0.98)                 & 66.33 (0.74)                  & 10.29 (0.46)                  & 43.28 (0.52)                   & 63.21 (1.57)                & 54.01 (6.17)                             \\
				EIIL                              & 72.60 (0.47)                   & 68.45 (0.53)                 & 66.38 (0.66)                  & 30.04 (10.9)                  & 42.98 (1.03)                   & 62.76 (1.72)                & 57.20 (5.33)                             \\
				IB-IRM                            & 72.50 (0.49)                   & 68.50 (0.40)                 & 66.64 (0.28)                  & \textbf{39.86 (10.5)}         & 40.85 (2.08)                   & 61.26 (1.20)                & 58.27 (5.33)                             \\
				CNC                               & 72.40 (0.46)                   & 67.24 (0.90)                 & 65.79 (0.80)                  & 12.21 (3.85)                  & 42.78 (1.53)                   & 61.03 (2.49)                & 53.56 (7.50)                             \\
				\hline
				\rule{0pt}{8pt}\textbf{{\ciga}v1} & \textbf{72.71 (0.52)}          & \textbf{69.04 (0.86)}        & \textbf{67.24 (0.88)}         & 19.77 (17.1)                  & \textbf{44.71 (1.14)}          & \textbf{63.66 (0.84)}       & \textbf{56.19 (2.50)}                    \\
				\textbf{{\ciga}v2}                & \textbf{73.17 (0.39)}          & \textbf{69.70 (0.27)}        & \textbf{67.78 (0.76)}         & \textbf{44.91 (4.31)}         & \textbf{45.25 (1.27)}          & \textbf{64.45 (1.99)}       & \textbf{60.88 (1.00)}                    \\
				\hline
				\rule{0pt}{8pt}Oracle             & 85.56 (1.44)                   & 84.71 (1.60)
				                                  & 85.83 (1.31)
				                                  & 62.13 (0.43)                   & 48.18 (1.00)
				                                  & 64.21 (1.77)                   &                                                                                                                                                                                                        \\
				\bottomrule
				\multicolumn{8}{l}{\rule{0pt}{8pt}$^\dagger$\text{\normalfont \small Averaged rank is also reported in the blankets because of dataset heterogeneity. A lower rank is better.}  }
			\end{tabular}
		\end{sc}
	}
\end{table}

\paragraph{OOD generalization performance on the structure and mixed shifts.}
In Table~\ref{CH:CIGA:tab:sythetic}, we report the test accuracy of each method, where we omit GIB due to its poor convergence.
Different biases indicate different strengths of the distribution shifts.
Although the training accuracy of most methods converges to more than $99\%$,
the test accuracy decreases dramatically as the bias increases and as more distribution shifts are mixed,
which concurs with our discussions in Sec.~\ref{CH:CIGA:sec:limitation_prev} and Appendix~\ref{CH:CIGA:sec:good_fail_setting_appdx}.

\begin{table}[t]
	\scriptsize
	\caption{OOD generalization performance on graph size shifts for real-world graphs in terms of Matthews correlation coefficient.}
	\vspace{-0.1in}\centering
	\label{table:graph_size}
	\resizebox{0.7\columnwidth}{!}{
		\begin{sc}
			\begin{tabular}{lrrrrr}
				\toprule
				Datasets                          & \multicolumn{1}{c}{NCI1} & \multicolumn{1}{c}{NCI109} & \multicolumn{1}{c}{PROTEINS} & \multicolumn{1}{c}{DD} & \multicolumn{1}{c}{Avg} \\
				\midrule
				ERM                               & 0.15 (0.05)              & 0.16 (0.02)                & 0.22 (0.09)                  & 0.27 (0.09)            & 0.20                    \\
				ASAP                              & 0.16 (0.10)              & 0.15 (0.07)                & 0.22 (0.16)                  & 0.21 (0.08)            & 0.19                    \\
				GIB                               & 0.13 (0.10)              & 0.16 (0.02)                & 0.19 (0.08)                  & 0.01 (0.18)            & 0.12                    \\
				DIR                               & 0.21 (0.06)              & 0.13 (0.05)                & 0.25 (0.14)                  & 0.20 (0.10)            & 0.20                    \\\hline
				\rule{0pt}{8pt}IRM                & 0.17 (0.02)              & 0.14 (0.01)                & 0.21 (0.09)                  & 0.22 (0.08)            & 0.19                    \\
				vrex                              & 0.15 (0.04)              & 0.15 (0.04)                & 0.22 (0.06)                  & 0.21 (0.07)            & 0.18                    \\
				EIIL                              & 0.14 (0.03)              & 0.16 (0.02)                & 0.20 (0.05)                  & 0.23 (0.10)            & 0.19                    \\
				IB-IRM                            & 0.12 (0.04)              & 0.15 (0.06)                & 0.21 (0.06)                  & 0.15 (0.13)            & 0.16                    \\
				CNC                               & 0.16 (0.04)              & 0.16 (0.04)                & 0.19 (0.08)                  & 0.27 (0.13)            & 0.20                    \\
				\hline
				\rule{0pt}{8pt}WL kernel          & \textbf{0.39 (0.00)}     & 0.21 (0.00)                & 0.00 (0.00)                  & 0.00 (0.00)            & 0.15                    \\
				GC kernel                         & 0.02 (0.00)              & 0.00 (0.00)                & 0.29 (0.00)                  & 0.00 (0.00)            & 0.08                    \\
				$\Gamma_\text{1-hot}$             & 0.17 (0.08)              & \textbf{0.25 (0.06)}       & 0.12 (0.09)                  & 0.23 (0.08)            & 0.19                    \\
				$\Gamma_\text{GIN}$               & 0.24 (0.04)              & 0.18 (0.04)                & 0.29 (0.11)                  & \textbf{0.28 (0.06)}   & 0.25                    \\
				$\Gamma_\text{RPGIN}$             & 0.26 (0.05)              & 0.20 (0.04)                & 0.25 (0.12)                  & 0.20 (0.05)            & 0.23                    \\
				\hline
				\rule{0pt}{8pt}\textbf{{\ciga}v1} & 0.22 (0.07)              & \textbf{0.23 (0.09)}       & \textbf{0.40 (0.06)}         & \textbf{0.29 (0.08)}   & \textbf{0.29}           \\
				\textbf{{\ciga}v2}                & \textbf{0.27 (0.07)}     & 0.22 (0.05)                & \textbf{0.31 (0.12)}         & 0.26 (0.08)            & \textbf{0.27}           \\
				\hline
				\rule{0pt}{8pt}Oracle (IID)       & 0.32 (0.05)              & 0.37 (0.06)                & 0.39 (0.09)                  & 0.33 (0.05)            &                         \\
				\bottomrule
			\end{tabular}
		\end{sc}
	}
\end{table}
Due to the simplicity of the task as well as the relatively high support overlap between training and test distributions,
interpretable GNNs and OOD objectives can improve certain OOD performance, while they can have \emph{high variance} since they do not have OOD generalization guarantees.
In contrast, {\ciga}v1 and {\ciga}v2 outperform all of the baselines by a significant margin up to $10\%$ with \emph{lower variance},
which demonstrates the effectiveness and excellent OOD generalization ability of \ciga.

\paragraph{OOD generalization performance on realistic shifts.}
In Table~\ref{table:other_graph} and Table~\ref{table:graph_size},
we examine the effectiveness of \ciga in real-world data and more complicated distribution shifts. Both averaged accuracy and ranks are reported because of the dataset heterogeneity.
Since the tasks are harder than synthetic ones, interpretable GNNs and OOD objectives perform similar to or even under-perform the ERM baselines, which is also consistent to the observations in non-linear benchmarks~\citep{domainbed,drugood}.
However, both {\ciga}v1 and {\ciga}v2 consistently and significantly outperform previous methods,
including previous specialized methods $\Gamma$ GNNs~\citep{size_gen2} for combating graph size shifts,
demonstrating the generality and superiority of \ciga.

\paragraph{Comparisons with advanced ablation variants.}
As discussed in Sec.~\ref{CH:CIGA:sec:good_theory}, \ciga can be reduced to interpretable GNNs and contrastive learning approaches.
However, across all experiments, we can observe that neither the advanced interpretable GNNs (DIR) nor sophisticated contrastive objectives with specialized sampling strategy (CNC) can yield satisfactory OOD performance, which serves as \emph{strong evidence} for the necessities of the decomposition as well as the objective in \ciga.
Furthermore, although {\ciga}v1 can outperform {\ciga}v2 when we may have a relatively accurate $s_c$,
the improvements in {\ciga}v1 are not as stable as {\ciga}v2 or even unsatisfactory when the assumption is violated.
This phenomenon also reveals the superiority of {\ciga}v2 in practice.

\begin{figure}[t]
	\centering
	\subfigure[SPMotif-Mixed (bias=$0.9$)]{
		\includegraphics[width=0.31\textwidth]{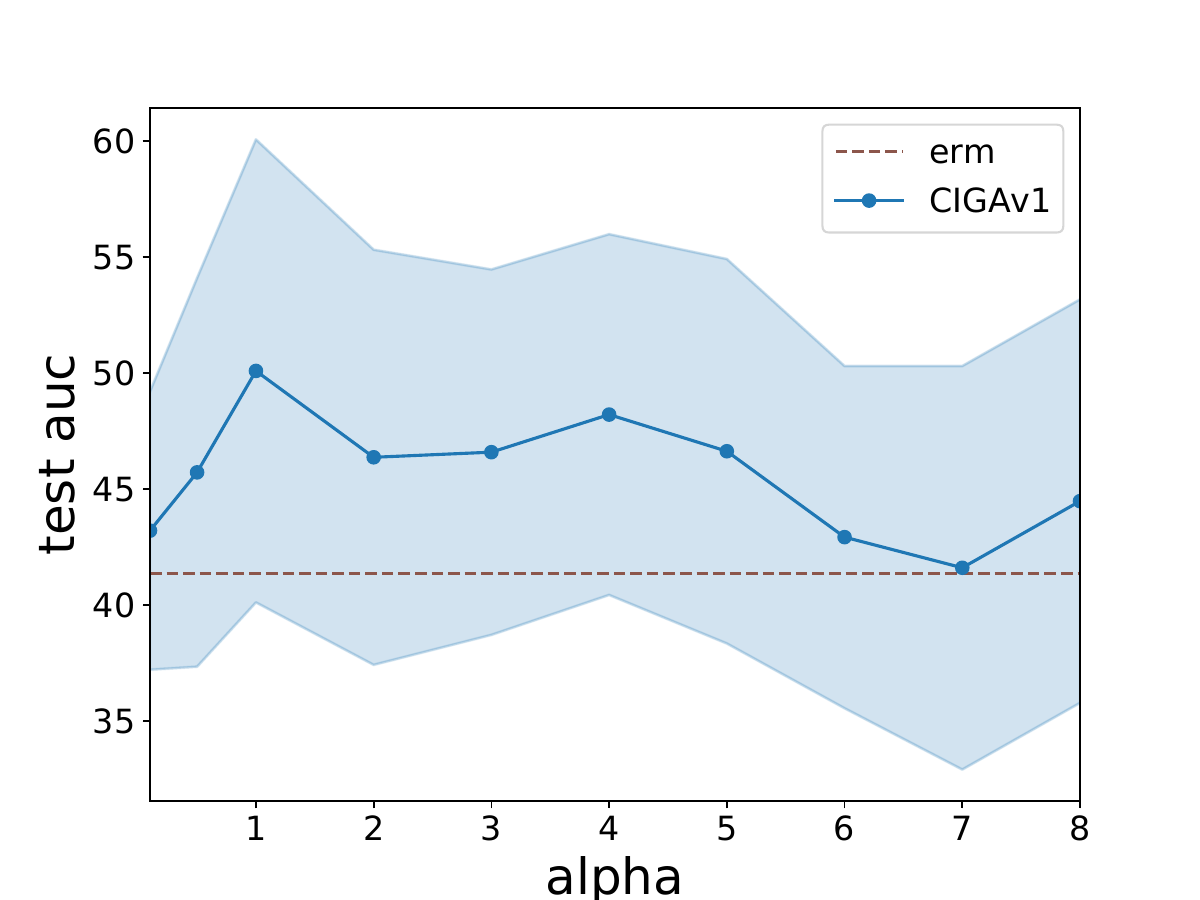}
	}
	\subfigure[DrugOOD-Scaffold]{
		\includegraphics[width=0.31\textwidth]{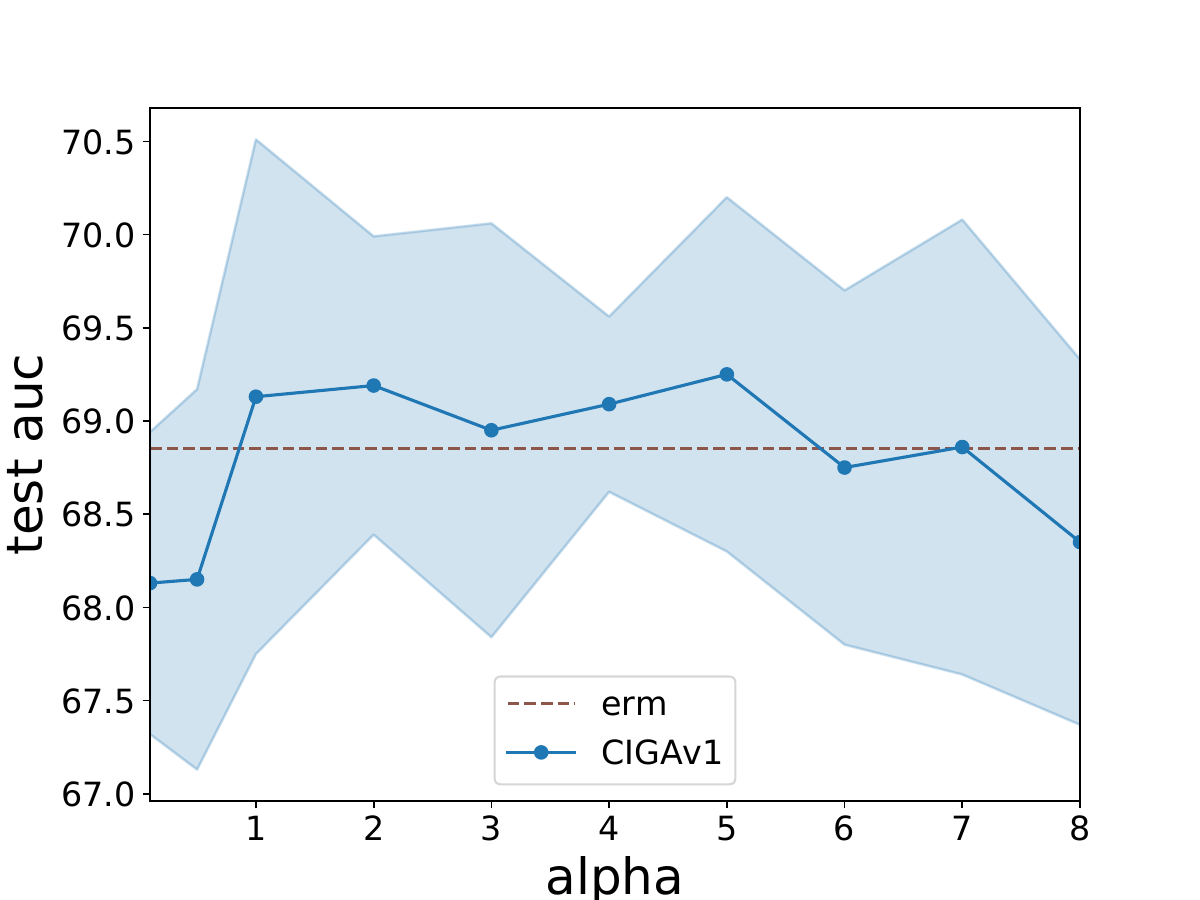}
	}
	\subfigure[NCI109]{
		\includegraphics[width=0.31\textwidth]{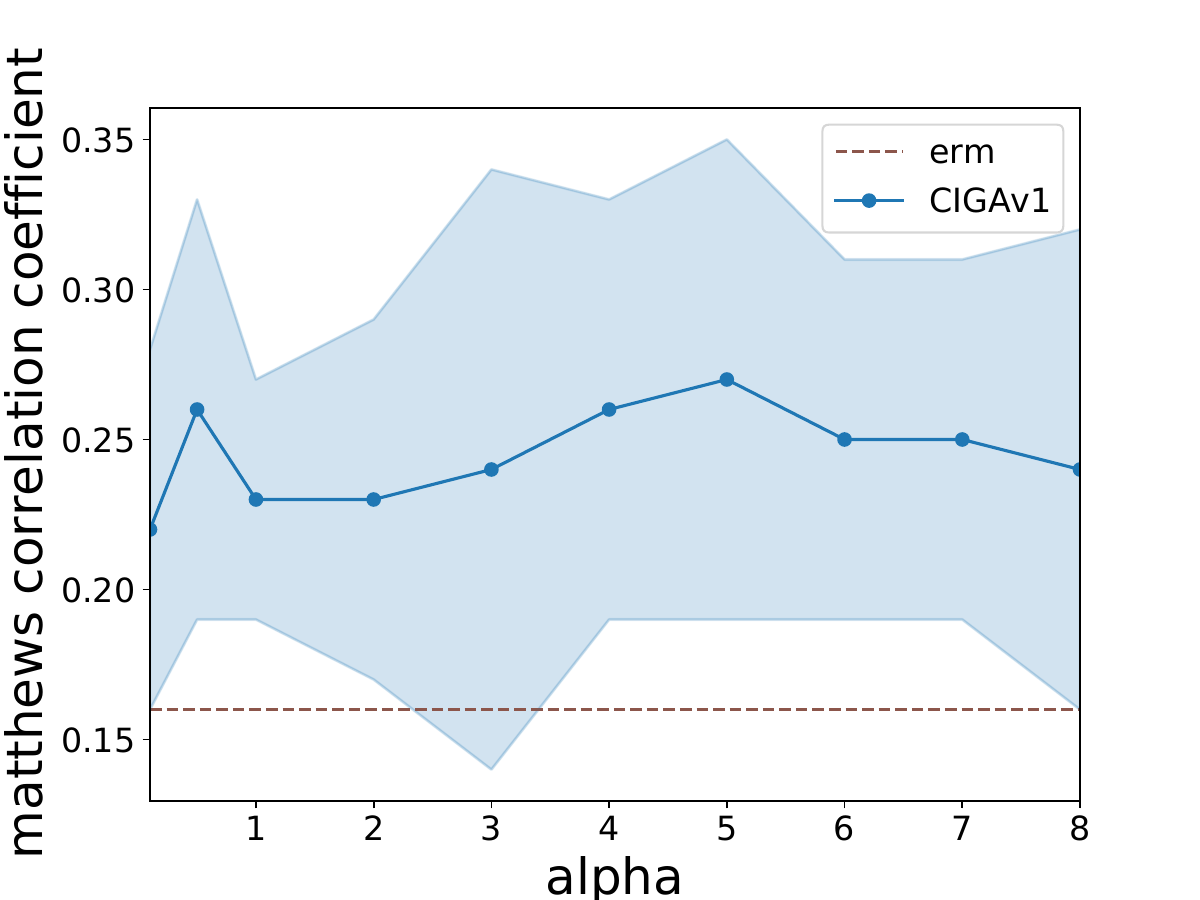}
	}
	\caption{
		Hyperparameter sensitivity analysis on the coefficient of contrastive loss ($\alpha$).}
	\label{CH:CIGA:fig:hp_sen_alpha}
\end{figure}

\begin{figure}[ht]
	\centering
	\subfigure[SPMotif-Mixed (bias=$0.9$, $\alpha$=$4$)]{
		\includegraphics[width=0.31\textwidth]{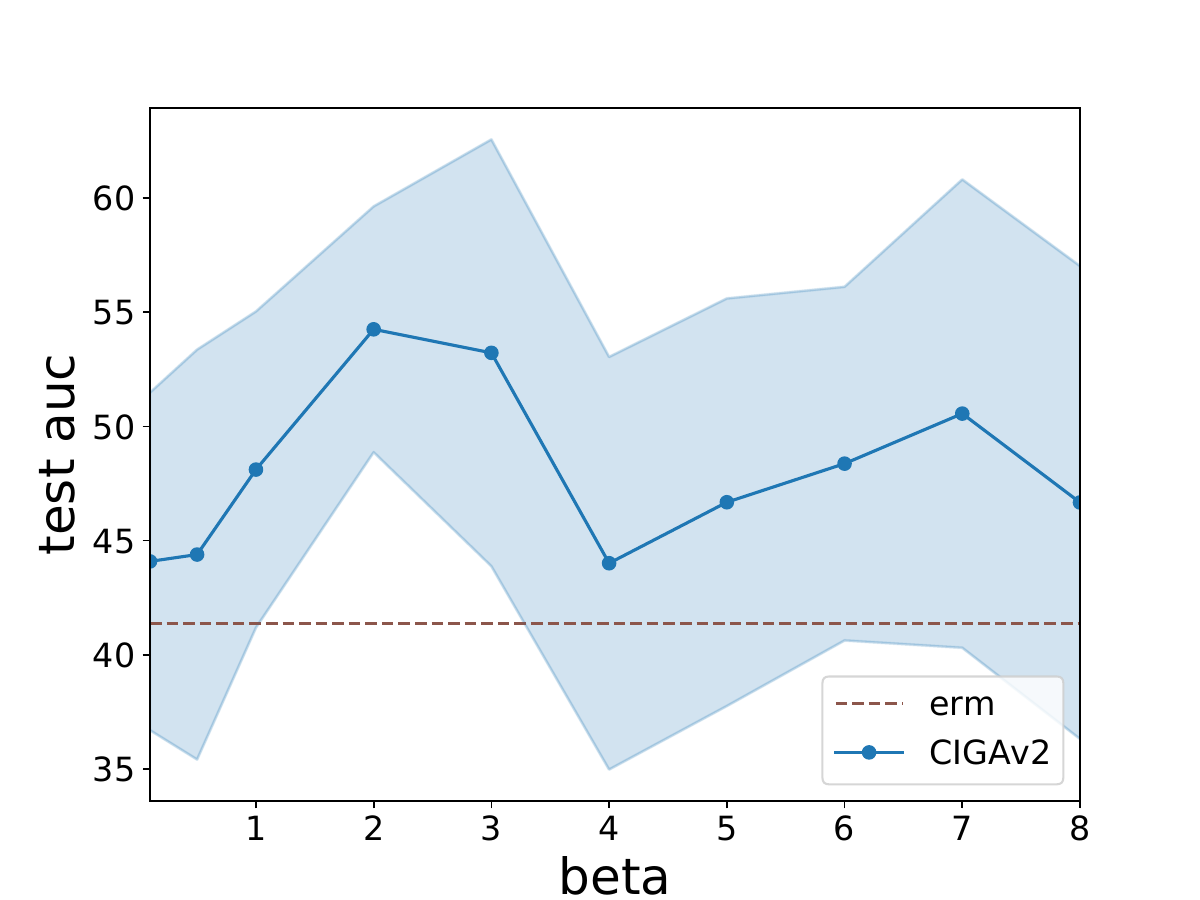}
	}
	\subfigure[DrugOOD-Scaffold ($\alpha$=$1$)]{
		\includegraphics[width=0.31\textwidth]{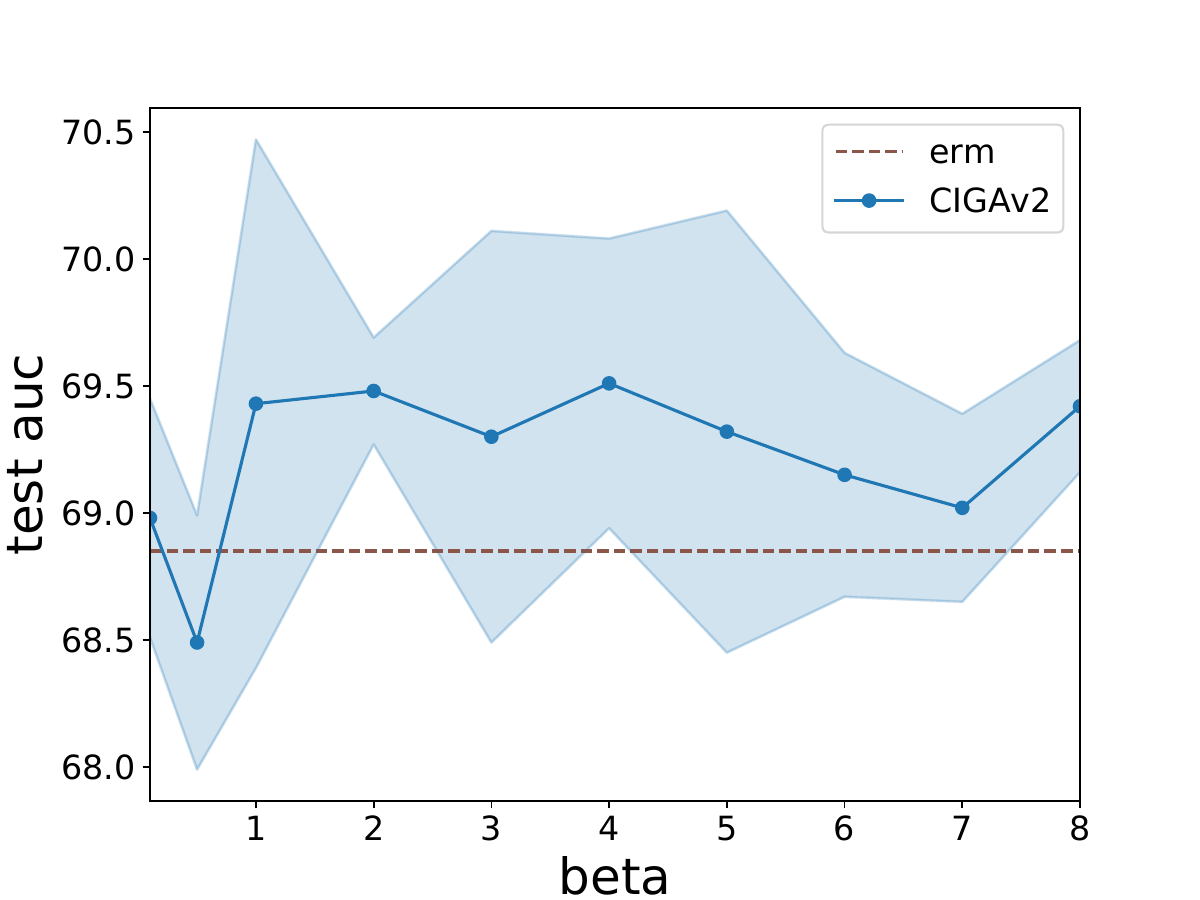}
	}
	\subfigure[NCI109 ($\alpha$=$1$)]{
		\includegraphics[width=0.31\textwidth]{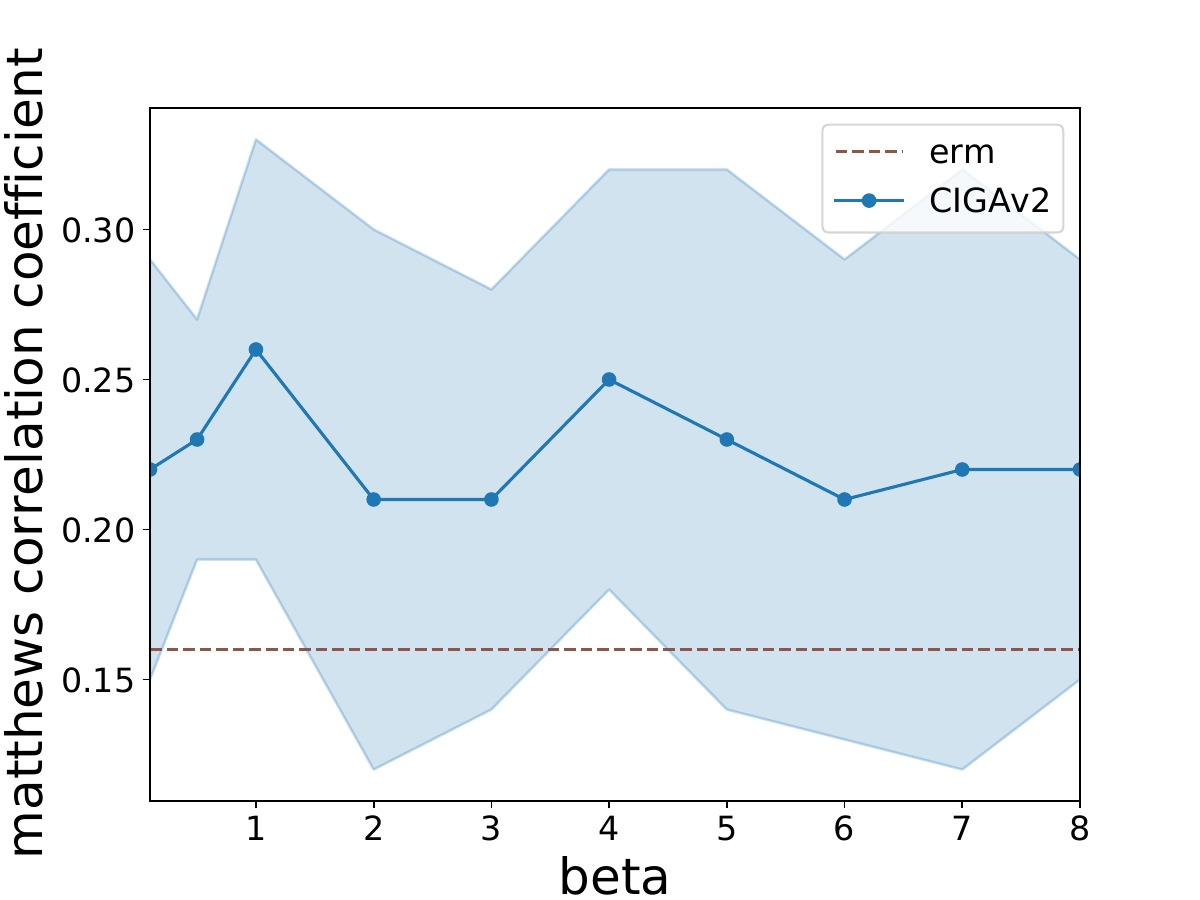}
	}
	\caption{
		Hyperparameter sensitivity analysis on the coefficient of hinge loss ($\beta$).}
	\label{CH:CIGA:fig:hp_sen_beta}
\end{figure}

\paragraph{Hyperparameter sensitivity analysis.}
To examine how sensitive \ciga is to the hyperparamters $\alpha$ and $\beta$ for contrastive loss and hinge loss, respectively.
We conduct experiments based on the hardest datasets from each table (i.e., SPMotif-Mixed with the bias of $0.9$, DrugOOD-Scaffold and the NCI109 datasets from Table~\ref{CH:CIGA:tab:sythetic}, Table~\ref{table:other_graph}, and Table~\ref{table:graph_size}, respectively.) with different $\alpha$ and $\beta$.
When changing the value of $\beta$, we fix the $\alpha$ to a specific value under which the model has a relatively good performance (but not the best, to fully examine the robustness of \ciga in practice).

The results are shown in Fig.~\ref{CH:CIGA:fig:hp_sen_alpha} and Fig.~\ref{CH:CIGA:fig:hp_sen_beta}.
It can be found that both {\ciga}v1 and {\ciga}v2 are robust to different values of $\alpha$ and $\beta$, respectively, across different datasets and distribution shifts.
Besides, the results also reflect the effects of the additional penalty terms in \ciga.
For example, in Fig.~\ref{CH:CIGA:fig:hp_sen_alpha_appdx}, when  $\alpha$ is too small, the invariance of the identified invariant subgraphs $\widehat{G}_c$ may not be guaranteed, resulting worse performances.
Similarly, as shown in Fig.~\ref{CH:CIGA:fig:hp_sen_beta_appdx}, when $\beta$ becomes too small, some part of the spurious subgraph may still appear in the estimated invariant subgraphs, which yields worse performances.
Besides, when $\alpha$ and $\beta$ become too large, the optimization of \ciga can be affected due to their intrinsic conflicts with ERM, hence a better optimization scheme for \ciga can be a promising future direction.
We provide more details and additional analysis on the efficiency of \ciga and single environment OOD generalization performance of \ciga in Appendix~\ref{CH:CIGA:sec:additional_exp_appdx}, as well as the visualization examples of the identified invariant subgraph in Appendix~\ref{CH:CIGA:sec:interpret_visualize_appdx}.

\chapter{Assumptions for Causal Invariance Learning on Graphs} \label{CH:GALA}

\section{Motivations}
As discussed in Chapter~\ref{CH:CIGA}, graph representation learning with graph neural networks (GNNs) has proven to be highly successful in tasks involving relational information~\citep{gcn,sage,gat,jknet,gin}.
However, it assumes that the training and test graphs are independently drawn from the identical distribution (iid.), which can hardly hold for many graph applications such as in Social Network, and Drug Discovery~\citep{ogb,wilds,TDS,ai4sci,reactionOOD,gdl_ds}.
The performance of GNNs could be seriously degenerated by \emph{graph distribution shifts}, i.e., mismatches between the training and test graph distributions caused by some underlying environmental factors during the graph data collection process~\citep{adv_causal_lens,closer_look_ood,drugood,good_bench,reactionOOD,gdl_ds}.
To overcome the Out-of-Distribution (OOD) generalization failure, recently there has been a growing surge of interest in incorporating the invariance principle from causality~\citep{inv_principle} into GNNs~\citep{handle_node,dir,ciga,gsat,dps,grea,gil,disc,moleood}.
The rationale of the invariant graph learning approaches is to identify the underlying \emph{invariant subgraph} of the input graph, which shares an invariant correlation with the target labels across multiple graph distributions from different environments~\citep{handle_node,ciga}.
Thus, the predictions made merely based on the invariant subgraphs can be generalized to OOD graphs that come from a new environment~\citep{inv_principle}.

As the environment labels or partitions on graphs are often expensive to obtain~\citep{ciga}, augmenting the environment information, such as generating new environments~\citep{handle_node,dir,grea} and inferring the environment labels~\citep{gil,moleood}, has become the \emph{de facto} approach for invariant graph learning.
However, little attention has been paid to verifying the \emph{fidelity} (or \emph{faithfulness}\footnote{The \emph{fidelity} or \emph{faithfulness} refers to whether the augmented environment information can actually improve the OOD generalization on graphs.}) of the augmented environment information. For example, if the generated environments or inferred environment labels induce a higher bias or noise, it would make the learning of graph invariance even harder.
Although it looks appealing to \emph{learn both}
the environment information and the graph invariance,
the existing approaches could easily run into the ``no free lunch'' dilemma~\citep{no_free_lunch}.
In fact, \citet{zin} found that there exist negative cases in the Euclidean regime where it is impossible to identify the invariant features without environment partitions.
When it comes to the graph regime where the OOD generalization
is fundamentally more difficult~\citep{ciga} than the Euclidean regime, it raises a challenging research question:
\begin{myquotation}\centering
    \emph{When and how could one learn graph invariance without the environment labels?}
\end{myquotation}
In this work, we present a theoretical investigation of the problem and seek a set of \emph{minimal assumptions} on the underlying environments for feasible invariant graph learning.
Based on a family of simple graph examples (Def.~\ref{def:twobit_graph}),
we show that existing environment generation approaches can fail to generate faithful environments,
when the underlying environments are not sufficient to uncover all the variations of the spurious subgraphs (Prop.~\ref{CH:GALA:thm:env_gen_fail}).
On the contrary, incorporating the generated environments can even lead to a worse OOD performance. The failure of faithful environment generation implies the necessity of \emph{variation sufficiency} (Assumption~\ref{CH:GALA:assump:var_sufficiency}).
Moreover, even with sufficient environments, inferring faithful environment labels remains impossible.
Since invariant and spurious subgraphs can have an arbitrary degree of correlation with labels, there exist multiple sets of training environments that have the same joint distribution of $P(G, Y)$ but different invariant subgraphs. Any invariant graph learning algorithms will inevitably fail to identify the invariant subgraph in at least one set of training environments (Prop.~\ref{CH:GALA:thm:env_infer_fail}).
Therefore, we need to additionally ensure the \emph{variation consistency} (Assumption.~\ref{CH:GALA:assump:var_consistency}), that is, the invariant and spurious subgraphs should have a consistent relationship in the correlation strengths with the labels.

\begin{figure}[t]
    \centering
    \includegraphics[width=0.9\columnwidth]{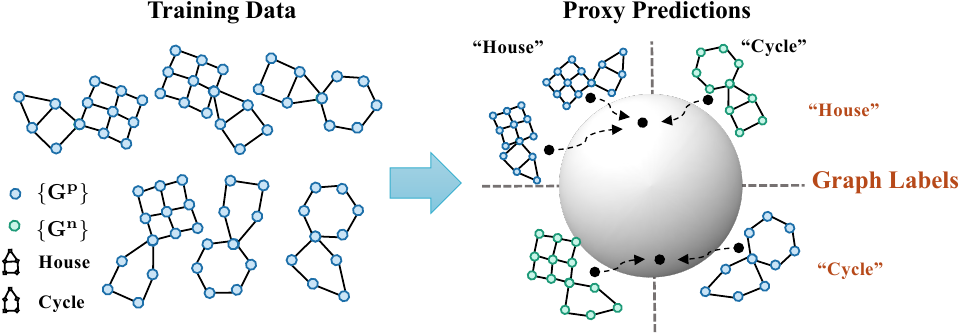}
    \caption[Illustration of \galafull (\gala).]{An illustration of \gala with the task of classifying graphs according to whether there exists a ``House'' or ``Cycle'' motif.
        Given the training data where the ``House'' subgraph often co-occurs with a ``Grid'' and the ``Cycle'' subgraph often co-occurs with a ``Hexagon''.
        An ERM trained environment assistant model will fit the spurious subgraph and therefore yield proxy predictions ``House'' or ``Cycle'' for any graphs containing a ``Grid'' (left half) or ``Hexagon'' (right half), respectively.
        \gala first separates the samples according to the correctness of the proxy predictions into the sets of positive graphs $\{G^p\}$ (correct, in blue) and negative graphs $\{G^n\}$ (incorrect, in green).
        Then, \gala extracts the maximally invariant subgraph among $\{G^p\}$ and $\{G^n\}$, i.e., pulling graphs with the same graph label but from $\{G^p\}$ and $\{G^n\}$ closer in the latent space, hence identifies the invariant subgraph.}
    \label{CH:GALA:fig:illustration}
\end{figure}

To resolve the OOD generalization challenge under the established assumptions, we propose a new framework
\galafull (\gala). \gala incorporates an additional assistant model
that needs to be prone to distribution shifts, to generate proxy predictions of the training samples.
Different from previous environment inferring approaches~\citep{moleood,gil}, \gala does not require explicit environment labels but merely proxy predictions to differentiate the variations in the spurious subgraphs.
As shown in Fig.~\ref{CH:GALA:fig:illustration}, we first fit an environment assistant model to the training distribution and then divide the training graphs into a positive set $\{G^p\}$ and a negative $\{G^n\}$, according to whether the proxy predictions are correct or not, respectively.
As spurious correlations tend to vary more easily than invariant correlations, the variations in spurious subgraphs are further differentiated and increased between $\{G^p\}$ and $\{G^n\}$.
Then, only the invariant subgraph holds an invariant correlation with the label among $\{G^p\}$ and $\{G^n\}$,
and hence can be identified by extracting the subgraphs that maximize the intra-class subgraph mutual information among $\{G^p\}$ and $\{G^n\}$ (Theorem~\ref{CH:GALA:thm:gala_success}).

We conduct extensive experiments to validate the effectiveness of \gala using $12$ datasets with various graph distribution shifts.
Notably, \gala brings improvements up to $30\%$ in multiple graph datasets.

The contributions of this chapter can be summarized as follows:
\begin{itemize}[leftmargin=*]
    \item We identify failure cases of existing invariant graph learning approaches and establish the minimal assumptions for feasible invariant graph learning;
    \item We develop a novel framework \gala with provable identifiability of the invariant subgraph for OOD generalization on graphs under the assumptions;
    \item We conduct extensive experiments to verify both our theoretical results and the superiority of \gala;
\end{itemize}
Notably, both our theory and solution differ from~\citet{zin} fundamentally, as we do not rely on the auxiliary information and are compatible with the existing interpretable and generalizable GNN architecture for OOD generalization on graphs. Meanwhile, we provide a new theoretical framework that resolves the counterexample in~\citet{zin} while enjoying provable identifiability.

\section{Background and Preliminaries}
\label{CH:GALA:sec:prelim}
As a supplementary to Sec.~\ref{CH:CIGA:sec:data_gen}, we begin by introducing the additional key concepts and backgrounds
of invariant graph learning, and leave more details in Appendix~\ref{CH:GALA:sec:prelim_appdx}. 
The notations used in the paper are given in Appendix~\ref{CH:GALA:sec:notations_appdx}.

\begin{wrapfigure}{r}{0.55\textwidth}
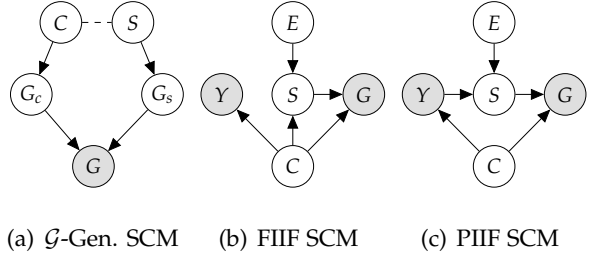

    \vspace{-0.15in}
    \centering
    \subfigure[$\gG$-Gen. SCM]{\label{CH:GALA:fig:graph_gen}
        \resizebox{!}{0.16\textwidth}{\tikz{
                \node[latent] (S) {$S$};%
                \node[latent,left=of S,xshift=0.5cm] (C) {$C$};%
                \node[latent,below=of C,xshift=-0.5cm,yshift=0.5cm] (GC) {$G_c$}; %
                \node[latent,below=of S,xshift=0.5cm,yshift=0.5cm] (GS) {$G_s$}; %
                \node[obs,below=of GC,xshift=1.05cm,yshift=0.5cm] (G) {$G$}; %
                \edge[dashed,-] {C} {S}
                \edge {C} {GC}
                \edge {S} {GS}
                \edge {GC,GS} {G}
            }}}
    \subfigure[FIIF SCM]{\label{CH:GALA:fig:scm_fiif}
        \resizebox{!}{0.16\textwidth}{\tikz{
                \node[latent] (E) {$E$};%
                \node[latent,below=of E,yshift=0.5cm] (S) {$S$}; %
                \node[obs,below=of E,xshift=-1.2cm,yshift=0.5cm] (Y) {$Y$}; %
                \node[obs,below=of E,xshift=1.2cm,yshift=0.5cm] (G) {$G$}; %
                \node[latent,below=of Y,xshift=1.2cm,yshift=0.5cm] (C) {$C$}; %
                \edge {E} {S}
                \edge {C} {Y,G}
                \edge {S} {G}
                \edge {C} {S}
            }}}
    \subfigure[PIIF SCM]{\label{CH:GALA:fig:scm_piif}
        \resizebox{!}{0.16\textwidth}{\tikz{
                \node[latent] (E) {$E$};%
                \node[latent,below=of E,yshift=0.5cm] (S) {$S$}; %
                \node[obs,below=of E,xshift=-1.2cm,yshift=0.5cm] (Y) {$Y$}; %
                \node[obs,below=of E,xshift=1.2cm,yshift=0.5cm] (G) {$G$}; %
                \node[latent,below=of Y,xshift=1.2cm,yshift=0.5cm] (C) {$C$}; %
                \edge {E} {S}
                \edge {C} {Y,G}
                \edge {S} {G}
                \edge {Y} {S}
            }}}
    \vspace{-0.1in}
    \caption{SCMs on graph distribution shifts.}
    \label{CH:GALA:fig:scm}
    \vspace{-0.15in}
\end{wrapfigure}

\paragraph{OOD generalization on graphs.}
Basically, we follow the same data generation assumptions as Sec.~\ref{CH:CIGA:sec:data_gen}, which is inspired by real-world drug discovery task~\citep{fragment} and covers a broad case of graph distribution shifts.
As shown in Fig.~\ref{CH:GALA:fig:scm}, the generation of the observed graphs $G$ and labels $Y$ are controlled by a latent causal variable $C$ and a spurious variable $S$.
$C$ and $S$ control $Y$ and $G$ by controlling the generation of the underlying invariant subgraph $G_c$ and spurious subgraph $G_s$, respectively.
Since $S$ can be affected by the environment $E$, the correlation between $Y$ and $G_s$ can change arbitrarily when the environment changes.
Besides, the interaction among $C$, $S$ and $Y$ at the latent space can be further categorized into \emph{Full Informative Invariant Features} (\emph{FIIF})
when $Y\ind S|C$, and \emph{Partially Informative Invariant Features} (\emph{PIIF}) when $Y \not\ind S|C$.

To tackle the OOD generalization challenge on graphs from Fig.~\ref{CH:GALA:fig:scm}, the existing invariant graph learning approaches (including those proposed concurrently and after \ciga) are generically designed to identify the underlying invariant subgraph $G_c$ to predict the label $Y$~\citep{handle_node,ciga}.
Specifically, the goal of OOD generalization on graphs
is to learn an \emph{invariant GNN} $f\coloneqq f_c\circ g$,
which is composed of:
a) a featurizer $g:\gG\rightarrow\gG_c$ that estimates the invariant subgraph $\widehat{G}_c$;
b) a classifier $f_c:\gG_c\rightarrow\gY$ that predicts the label $Y$ based on the extracted $\widehat{G}_c$,
where $\gG_c$ refers to the space of subgraphs of $\gG$.
The learning objectives of $f_c$ and $g$ are formulated as
\begin{equation}
    \label{CH:GALA:eq:inv_cond}
    \text{$\max$}_{f_c, \; g} \ I(\widehat{G}_{c};Y), \ \text{s.t.}\ \widehat{G}_{c}\ind E,\ \widehat{G}_{c}=g(G).
\end{equation}
Since $E$ is not observed, many strategies are proposed to
impose the independence of $\widehat{G}_c$ and $E$.
A prevalent approach is to augment the environment information.
Based on the estimated invariant subgraphs $\widehat{G}_c$ and spurious subgraphs $\widehat{G}_s$,
\citet{dir,grea,handle_node} propose to generate new environments, while \citet{moleood,gil} propose to infer the underlying environment labels.
However, we show that they all fail to augment faithful environment information in Sec.~\ref{CH:GALA:sec:env_aug_failure}.

Besides, \citet{gib,vgib,gsat,dps,lri} adopt graph information bottleneck to tackle FIIF graph shifts, but they cannot generalize to PIIF shifts, while Our work focuses on PIIF shifts as it is more challenging when without environment labels~\citep{zin}.
\citet{disc} generalize~\citep{ldd} to tackle severe graph biases, i.e., when $H(S|Y)< H(C|Y)$. \citet{ciga} propose a contrastive framework to tackle both FIIF and PIFF graph shifts, but is limited to $H(S|Y)> H(C|Y)$.
In practice, as it is usually unknown which correlation is stronger, we need a unified solution to tackle both cases.

\paragraph{Invariant learning without environment labels.}
In the Euclidean regime, there are plentiful studies in invariant learning without environment labels.
\citet{eiil} propose a minmax formulation to infer the environment labels.
\citet{hrm} propose a self-boosting framework based on the estimated invariant and variant features.
\citet{jtt,cnc,pde,xrm} propose to infer labels based on the failures of an ERM model.
However, \citet{zin} find failure cases of the aforementioned approaches that it is impossible to identify the invariant features without given environment labels in Euclidean data, and propose a solution that leverages auxiliary environment information for invariant learning.
As the OOD generalization on graphs poses more challenges~\citep{ciga}, whether it is feasible to learn invariant graph representations without any auxiliary environment information remains elusive.

\section{Pitfalls of Environment Augmentation}\label{CH:GALA:sec:env_aug_failure}
Given only the mixed training data without environment partitions,
is it possible to learn to generate faithful environments or
infer the underlying environment labels that facilitate OOD generalization on graphs?
In the discussion below, we adopt the two-piece graphs to instantiate the problem, which is the simplistic version of the PIIF distribution shifts in Fig.~\ref{CH:GALA:fig:scm_piif}, motivated by~\citet{irm_aistats}.
\begin{definition}[Two-piece graphs]
    \label{def:twobit_graph}
    Each environment $e$ is defined with two parameters, $\alpha_e,\beta_e\in[0,1]$,  and the dataset $(G^e,Y^e)\in\dataset_e$ is generated as follows:
    \begin{enumerate}[label=(\alph*),leftmargin=*]
        \item Sample $Y^e\in\{-1,1\}$ uniformly;
        \item Generate $G_c$ and $G_s$ via :
              $G_c\coloneqq f_\gen^{G_c}(Y^e\cdot\rad(\alpha_e)),\ G_s\coloneqq f_\gen^{G_s}(Y^e\cdot\rad(\beta_e)),$
              where $f_\gen^{G_c},f_\gen^{G_s}$ map the input $\{-1,1\}$ to a corresponding graph selected from a given set,
              and $\rad(\alpha)$ is a random variable taking value $-1$ with probability $\alpha$ and $+1$
              with $1-\alpha$;
        \item Synthesize $G^e$ by randomly assembling $G_c$ and $G_s$:
              $G^e\coloneqq f_\gen^{G}(G_c,G_s).$
    \end{enumerate}
\end{definition}

We denote an environment $e$ with $(\alpha,\beta_e)$ for simplicity.
Different environments will have a different $\beta_e$,
thus $P(Y|G_s)$ will change across different environments, while $P(Y|G_c)$ remains invariant.

\subsection{Pitfalls of environment generation}
\label{CH:GALA:sec:var_sufficiency}
We begin by discussing the cases where there are few environments,
and generating new environments is necessary~\citep{handle_node,dir,grea}.
Environment generation aims to provide some additional ``virtual'' environments $\env_v$ such that the invariant subgraph can be identified via applying an OOD risk to the joint dataset with the augmented data $\trainvirtual=\{\dataset_e|e\in\envtrain\cup\env_v\}$.

The generation of ``virtual'' environments is primarily based on the intermediate estimation of the invariant and spurious subgraphs, denoted as $\widehat{G}_c$ and $\widehat{G}_s$, respectively.
\citet{dir,grea} propose \dir and \grea to construct new graphs
by assembling $\widehat{G}_c$ and $\widehat{G}_s$ from different graphs.
Specifically, given $n$ samples $\{G^i,Y^i\}_{i=1}^n$,\footnote{We slightly abuse the superscript and subscript when denoting the $i$th sample to avoid confusion of double superscripts or subscripts.}
the new graph samples in $\env_v$ is generated as follows:
\[
    G^{i,j}=f_\gen^G(\widehat{G}_c^i,\widehat{G}_s^j),\ \forall i,j \in\{1...n\},\ Y^{i,j}=Y^i,
\]
which generates a new environment $\env_v$ with $n^2$ samples.
Although both \dir and \grea gain some empirical success,
the faithfulness of $\env_v$ remains questionable,
as the generation is merely based on \emph{inaccurate} estimations of the invariant and spurious subgraphs.
Specifically, when $\widehat{G}_c$ contains parts of $G_s$, assigning the same labels to the generated graph is more likely to \textit{strengthen} the spurious correlation between $G_s$ and $Y$.
For example, when the model yields a reversed estimation, i.e., $\widehat{G}_c=G_s$ and $\widehat{G}_s=G_c$,
the generated environment will destroy the invariant correlations.
\begin{proposition}\label{CH:GALA:thm:env_gen_fail}
    Consider the two-piece graph dataset $\envtrain=\{(\alpha,\beta_1),(\alpha,\beta_2)\}$ with $\alpha\geq\beta_1,\beta_2$
    (e.g., $\envtrain=\{(0.25,0.1),(0.25,0.2)\}$),
    and its corresponding mixed environment $\envmix=\{(\alpha,(\beta_1+\beta_2)/2)\}$ (e.g., $\envmix=\{(0.25,0.15)\}$).
    When $\widehat{G}_c=G_s$ and $\widehat{G}_s=G_c$, it holds that the augmented environment $\env_v$ is also a two-piece graph dataset with
    \[
        \mathcal{E}_v = \{(0.5,(\beta_1 + \beta_2)/2)\}\text{ (e.g., $\mathcal{E}_v = \{(0.5,0.15)\}$)}.
    \]
\end{proposition}
The proof is given in Appendix~\ref{CH:GALA:proof:env_gen_fail_appdx}.
This also extends to the adversarial augmentation~\citep{handle_node,dps}, which will destroy the actual $\widehat{G}_c$.
As both \dir and \grea adopt the same environment generation procedure,
we verify the failures of environment generation with \grea in Table~\ref{CH:GALA:tab:syn_graph} of Sec.~\ref{CH:GALA:sec:exp}, where \grea can perform comparably with ERM.
In fact, when the underlying environments are insufficient to differentiate the variations of the spurious features,
it is fundamentally impossible to identify the underlying invariant graph from the spurious subgraph.
More formally, if $\exists G_s$, such that $P^{e_1}(Y|G_s)=P^{e_2}(Y|G_s)$ for any $e_1,e_2\in \envtrain$, where $P^e(Y|G_s)$ is the conditional distribution $P(Y|G_s)$ under environment $e\in \envall$, it is impossible for any graph learning algorithm to identify $G_c$.
We provide a formal discussion in Appendix~\ref{CH:GALA:proof:var_sufficiency_appdx}. The failure implies a fundamental requirement that $\envtrain$ should uncover all the potential variations in the spurious subgraph.
\begin{assumption}(Variation sufficiency)\label{CH:GALA:assump:var_sufficiency}
    For graphs generated following Fig.~\ref{CH:GALA:fig:scm},
    for any $G_s$, $\exists e_1,e_2\in \envtrain$,
    such that $P^{e_1}(Y|G_s)\neq P^{e_2}(Y|G_s)$, and $P^{e_1}(Y|G_c) = P^{e_2}(Y|G_c)$.
\end{assumption}
Assumption~\ref{CH:GALA:assump:var_sufficiency} aligns with the definition of invariance~\citep{irm_aistats,ciga} that the invariant subgraph $G_c$ is expected to satisfy $P^{e_1}(Y|G_c) = P^{e_2}(Y|G_c)$ for $e_1,e_2\in \envall$. If there exists $G_s$ satisfying the invariance condition as well, then it is impossible to tell $G_c$ from $G_s$ even with environment labels.

\subsection{Pitfalls of environment inferring}
\label{CH:GALA:sec:var_consistency}
Although environment sufficiency (Assumption~\ref{CH:GALA:assump:var_sufficiency}) relieves the need for generating new environments,
is it possible to infer the underlying environment labels via approaches such as \mole~\citep{moleood} and \gil~\citep{gil}, to facilitate invariant graph learning?
Unfortunately, we find a negative answer.

Considering the two-piece graph examples $\envtrain=\{(0.2,0.1),(0.2,0.3)\}$, when given the underlying environment labels, it is easy to identify the invariant subgraphs from spurious subgraphs.
However, when the environment labels are not available, we have the mixed data as $\envtrain=\{(0.2,0.2)\}$, where $P(Y|G_c)=P(Y|G_s)$.
The identifiability of $G_s$ is \emph{ill-posed}, as it does not affect the $\envtrain$ even if we swap $G_c$ and $G_s$.
More formally, considering the environment mixed from two two-piece graph environments $\{(\alpha,\beta_1)\}$ and $\{(\alpha,\beta_2)\}$, then we have $\envtrain=\{(\alpha,(\beta_1+\beta_2)/2\}$.
For each $\envtrain$, we can also find a corresponding $\envtrain'=\{((\beta'_1+\beta'_1)/2,\alpha')\}$ with $\{(\beta'_1,\alpha')\}$ and $\{(\beta'_2,\alpha')\}$. Then, let
\begin{equation}\label{CH:GALA:eq:env_infer_fail}
    \alpha=(\beta'_1+\beta'_1)/2=\alpha'=(\beta_1+\beta_2)/2.
\end{equation}
We now obtain $\envtrain$ and $\envtrain'$ which share the same joint distribution $P(Y, G)$ while the underlying $G_c$ is completely different.
More generally, we have the following proposition.
\begin{proposition}\label{CH:GALA:thm:env_infer_fail}
    There exist $2$ two-piece graph training environments $\envtrain$ and $\envtrain'$ that share the same joint distribution $P(Y, G)$. Any learning algorithm will fail in either $\envtrain$ or $\envtrain'$.
\end{proposition}
The proof is given in Appendix~\ref{CH:GALA:proof:env_infer_fail_appdx}.
The experiments in Sec.~\ref{CH:GALA:sec:exp} validate that both \mole and \gil fail to infer faithful environment labels and even underperform ERM.
It implies that whenever it allows the existence of an identical training distribution by mixing the environments, invariant graph learning is impossible.
Therefore, we need an additional assumption that excludes the unidentifiable case.
We propose to constrain the relationship between $\alpha$ (i.e., $H(Y|G_c)$ ) and $\beta_e$ (i.e., $H(Y|G_s)$).

\begin{assumption}(Variation consistency)\label{CH:GALA:assump:var_consistency}
    For all environments in $\envtrain$, $H(C|Y)\neq H(S|Y)$.
\end{assumption}
Intuitively, Assumption~\ref{CH:GALA:assump:var_consistency} imposes the consistency requirement on the correlation strengths between invariant and spurious subgraphs with labels.
For two-piece graphs with consistent variations, mixing up the environments will yield a new environment with the same variation strength relationships. Thus, Assumption~\ref{CH:GALA:assump:var_consistency} gets rid of the previous unidentifiable cases.
Moreover, Assumption~\ref{CH:GALA:assump:var_consistency} also aligns with many realistic cases. For example, the relation of a specific functional group (e.g., -OH) with a molecule can hardly be reversed to that held upon the scaffold of the molecule, due to the data collection process.
Therefore, Assumption~\ref{CH:GALA:assump:var_consistency} also resolves the counterexample proposed by~\citet{zin}.
Different from our work, \citet{zin} propose to incorporate additional auxiliary information that satisfies certain requirements to mitigate the unidentifiable case. However, such auxiliary information is often unavailable and expensive to obtain on graphs.
More importantly, the requirements are also unverifiable without more assumptions, which motivates us to consider the relaxed case implied by Assumption~\ref{CH:GALA:assump:var_consistency}.

\subsection{Challenges of environment augmentation}
\label{CH:GALA:sec:env_aug_challenge}

To summarize, the two assumptions constitute the minimal assumptions for feasible invariant graph learning. Failing to satisfy either one of them while lacking additional inductive biases will result in the ``no free lunch'' dilemma~\citep{no_free_lunch} and suffer from the unidentifiability issue.

\begin{corollary}(No Free Graph OOD Lunch)\label{CH:GALA:thm:var_consistency}
    Without Assumption~\ref{CH:GALA:assump:var_sufficiency} or Assumption~\ref{CH:GALA:assump:var_consistency},
    there does not exist a learning algorithm that captures the invariance of the two-piece graph environments.
\end{corollary}
\begin{wraptable}{r}{0.5\textwidth}
    \begin{center}
        \vskip -0.3in
        \caption{Challenges of invariant graph learning: no existing works can handle both cases.}
        \scalebox{0.75}{
            \begin{tabular}{c|c|c}
                \hline
                             & $H(S|Y)<H(C|Y)$ &
                $H(S|Y)>H(C|Y)$
                \\\hline
                \disc        & \cmark          & \xmark \\\hline
                \ciga        & \xmark          & \cmark \\\hline
                \gala (Ours) & \cmark          & \cmark \\
                \hline
            \end{tabular}}
        \label{tab:env_challenge}
    \end{center}
	\vskip -0.3in
\end{wraptable}
Corollary~\ref{CH:GALA:thm:var_consistency} is a natural conclusion from the previous discussion. The proof is straightforward and  given in Appendix~\ref{CH:GALA:proof:var_consistency_appdx}.
Assumption~\ref{CH:GALA:assump:var_sufficiency} and Assumption~\ref{CH:GALA:assump:var_consistency}
establish the minimal premises for identifying the underlying invariant subgraphs.
However, it also raises new challenges, as shown in Table.~\ref{tab:env_challenge}.
\citet{ciga} propose \ciga to maximize the intra-class mutual information
of the estimated invariant subgraphs to tackle the case when $H(C|Y)< H(S|Y)$.
While for the case when $H(S|Y)< H(C|Y)$,
\citet{disc} propose \disc that adopts GCE loss~\citep{ldd} to
extract the spurious subgraph with a larger learning step size
such that the left subgraph is invariant.
However, both of them can fail when there is no prior knowledge about the relations between $H(C|Y)$ and $H(S|Y)$.
We verify the failures of \disc and \ciga in Table.~\ref{CH:GALA:tab:syn_graph}.
The failure thus raises a challenging question:
\begin{myquotation}
    \emph{Given the established minimal assumptions,
        is there a unified framework that tackles both cases when $H(C|Y)< H(S|Y)$ and $H(C|Y)> H(S|Y)$?}
\end{myquotation}
\section{Environment Assistant for Invariant Graph Representations}
\label{CH:GALA:sec:gala_sol}
We give an affirmative answer by proposing a new framework,  \gala: \galafull,
which adopts an assistant model to provide proxy information
about the environments.

\subsection{Learning with An Environment Assistant}\label{CH:GALA:sec:gala_der}
Intuitively, a straightforward approach to tackle the aforementioned challenge is to extend the framework of either \disc~\citep{disc} or \ciga~\citep{ciga}
to resolve the other case.
As \disc always destroys the first learned features and tends to be more difficult to extend (which is empirically verified in Sec.~\ref{CH:GALA:sec:exp}), we are motivated to extend the framework of \ciga
to resolve the case when $H(S|Y)<H(C|Y)$.

\paragraph{Understanding the success and failure of \ciga.}
The principle of \ciga lies in maximizing the intra-class mutual information
of the estimated invariant subgraphs, i.e.,
\begin{equation}
    \label{CH:GALA:eq:cigav1_sol}
    \max_{f_c, g} \ I(\pred{G}_{c};Y), \ \text{s.t.}\
    \pred{G}_{c}\in\argmax_{\pred{G}_{c}=g(G), |\pred{G}_{c}|\leq s_c} I(\pred{G}_{c};\pred{G}_c^s|Y),
\end{equation}
where $\pred{G}_c^s=g(G^s)$ and $G^s\sim \sP(G|Y)$,
i.e., $\pred{G}$ is sampled from training graphs that share the same label $Y$ as $\pred G$.
The key reason for the success of Eq.~\ref{CH:GALA:eq:cigav1_sol} is that, given the data generation process as in Fig.~\ref{CH:GALA:fig:scm}
and the same $C$, the underlying invariant subgraph $G_c$ maximizes the
mutual information of subgraphs from any two environments, i.e., $\forall e_1,e_2\in\envall$,
\begin{equation}
    \label{CH:GALA:eq:ciga_cond}
    G_c^{e_1}\in \text{$\argmax$}_{\pred{G}_c^{e_1}}\  I(\pred{G}_c^{e_1};\pred{G}_c^{e_2}|C),
\end{equation}
where $\pred{G}_c^{e_1}$ and $\pred{G}_c^{e_2}$ are the estimated
invariant subgraphs corresponding to the same latent causal variable $C=c$ under the environments $e_1, e_2$, respectively.
Since $C$ is not observable, \ciga adopts $Y$ as a proxy for $C$, as when $H(S|Y)>H(C|Y)$, $G_c$ maximizes $I(\pred{G}_c^{e_1};\pred{G}_c^{e_2}|Y)$
and thus $I(\pred{G}_c;\pred{G}_c^s|Y)$.
However, when $H(S|Y)<H(C|Y)$, the proxy no longer holds.
Given the absence of $E$, simply maximizing intra-class mutual information
favors the spurious subgraph $G_s$ instead, i.e.,
\begin{equation}\label{CH:GALA:eq:cigav1_fail}
    G_s\in\text{$\argmax$}_{\pred{G}_c}I(\pred{G}_c;\pred{G}_c^s|Y).
\end{equation}

\paragraph{Invalidating spuriousness dominance.}
To mitigate the issue, we are motivated to find a new proxy
that samples $\pred{G}_c$ for Eq.~\ref{CH:GALA:eq:cigav1_fail},
while preserving only the $G_c$ as the solution under both cases.

To begin with, we consider the case of $H(S|Y)<H(C|Y)$. Although the correlation between $G_s$ and $Y$ dominates the intra-class mutual information, Assumption~\ref{CH:GALA:assump:var_sufficiency} implies that
there exists a subset of training data where $P(Y|G_s)$ varies,
while $P(Y|G_c)$ remains invariant.
Therefore, the dominance of spurious correlations no longer holds for samples from the subset.
Incorporating samples from the subset into Eq.~\ref{CH:GALA:eq:cigav1_sol} as $\pred{G}_c^s$ invalidates the dominance of $G_s$.
Denote the subset as $\{\pred{G}_{c}^n\}$, then
\begin{equation}\label{CH:GALA:eq:gala_sol_spu}
    G_c\in\text{$\argmax$}_{\pred{G}_c^p}I(\pred{G}_c^p;\pred{G}_c^n|Y),
\end{equation}
where $\pred{G}_c^p\in \{\pred{G}_{c}^p\}$
is sampled from the subset $\{\pred{G}_{c}^p\}$ dominated by spurious correlations,
while $\pred{G}_c^n\in \{\pred{G}_{c}^n\}$
is sampled from the subset $\{\pred{G}_{c}^n\}$ where spurious correlation no long dominates, or is dominated by invariant correlations.
We prove the effectiveness of Eq.~\ref{CH:GALA:eq:gala_sol_spu} in Theorem~\ref{CH:GALA:thm:gala_success}.

\paragraph{Environment assistant model $A$.}
To find the desired subsets $\{\pred{G}_{c}^p\}$ and $\{\pred{G}_{c}^n\}$,
inspired by the success in tackling spuriousness-dominated OOD generalization via learning from a biased predictors~\citep{lff,ldd,jtt,cnc},
we propose to incorporate an assistant model $A$ that is prone to spurious correlations.
Simply training $A$ with ERM using the spuriousness-dominated data enables $A$ to learn spurious correlations, and hence identifies the subsets where the spurious correlations hold or shift,
according to whether the predictions of $A$ are correct or not, respectively. Let $A=\argmax_{\pred{A}} I(\pred{A}(G);Y)$, we have
\begin{equation}
    \begin{aligned}
         & \{\pred{G}_{c}^p\}=\{g(G^p_i)|A(G^p_i)=Y_i\},       \
        \{\pred{G}_{c}^n\}=\{g(G^n_i)|A(G^n_i)\neq Y_i\}.
    \end{aligned}
\end{equation}

\paragraph{Reducing to invariance dominance case.}
After showing that Eq.~\ref{CH:GALA:eq:gala_sol_spu} resolves the spuriousness dominance case, we still need to show that Eq.~\ref{CH:GALA:eq:gala_sol_spu} preserves $G_c$ as the only solution when $H(S|Y)>H(C|Y)$.
Considering training $A$ with ERM using the invariance-dominated data,
$A$ will learn both invariant correlations and spurious correlations~\citep{disc,feat}.
Therefore, $\{\pred{G}_{c}^n\}$ switches to the subset that
is dominated by spurious correlations,
while $\{\pred{G}_{c}^p\}$ switches to the subset dominated by invariant correlations.
Then, Eq.~\ref{CH:GALA:eq:gala_sol_spu} establishes a lower bound for the intra-class mutual information, i.e.,
\begin{equation}\label{CH:GALA:eq:gala_sol_inv}
    I(\pred{G}_c^p;\pred{G}_c^n|Y)\leq I(\pred{G}_c;\pred{G}_c^s|Y),
\end{equation}

\begin{algorithm}[H]
    \caption{\textbf{\gala}: \galafull }
    \label{alg:gala}
    \begin{algorithmic}[1]
        \STATE \textbf{Input:} Training data $\train$;
        environment assistant $A$;
        featurizer GNN $g$; classifier GNN $f_c$;
        length of maximum training epochs $e$; batch size $b$;
        \STATE Initialize environment assistant $A$;
        \FOR{$p \in [1,\ldots, e]$}
        \STATE Sample a batch of data $\{G_i,Y_i\}_{i=1}^b$ from $\train$;
        \STATE Obtain Environment Assistant predictions $\{\hat{y}^e_i\}_{i=1}^b$;
        \FOR{each sample $G_i,y_i \in \{G_i,Y_i\}_{i=1}^b$}
        \STATE Find \emph{positive} graphs with same $y_i$ and different $\hat{y}^e_i$;
        \STATE Find \emph{negative} graphs with different $y_i$ but same assistant prediction $\hat{y}^e_i$;
        \STATE Calculate \gala risk via Eq.~\ref{CH:GALA:eq:gala_sol};
        \STATE Update $f_c, g$ via gradients from \gala risk;
        \ENDFOR\ENDFOR
        \STATE \textbf{return} final model $f_c\circ g$;
    \end{algorithmic}
\end{algorithm}
where $\pred{G}_c^p\in\{\pred{G}_{c}^p\}, \pred{G}_c^n\in \{\pred{G}_{c}^n\}$,
and $\pred{G}_c, \pred{G}_c^s$ are the same as in Eq.~\ref{CH:GALA:eq:cigav1_sol}.
The inequality in Eq.~\ref{CH:GALA:eq:gala_sol_inv} holds as any subgraph maximizes the left hand side can also be incorporated in right hand side, while the sampling space of $\widehat{G}_c$ and $\widehat{G}^s_c$ in the right hand side (i.e., both $\pred{G}_c$ and $\pred{G}_c^s$ are sampled from the whole train set) is larger than that of the left hand side.
The equality is achieved by taking the ground truth $G_c$ as the solution for the featurizer $g$. We verify the correctness of Eq.~\ref{CH:GALA:eq:gala_sol_spu} and Eq.~\ref{CH:GALA:eq:gala_sol_inv} in Fig.~\ref{CH:GALA:fig:corr_change}.

\subsection{Practical implementations.}
The detailed algorithm description of \gala is shown as in Algorithm~\ref{alg:gala}.
In practice, the environment assistant can have multiple implementation choices so long as it is prone to distribution shifts.
As discussed in Sec.~\ref{CH:GALA:sec:gala_der}, ERM trained model can
serve as a reliable environment assistant, since ERM tends to learn the dominant features no matter whether the features are invariant or spurious.
For example,
when $H(S|Y)<H(C|Y)$, ERM will first learn to use spurious subgraphs $G_s$ to make predictions.
Therefore, we can obtain $\{G^p\}$ by finding samples where ERM correctly predicts the labels, and $\{G^n\}$ for samples where ERM predicts incorrect labels.
In addition to label predictions,
the clustering predictions of the hidden representations yielded by environment assistant models can also be used for sampling $\{G^p\}$ and $\{G^n\}$~\citep{cnc}.
Besides, we can also incorporate models that are easier to overfit to the first dominant features to better differentiate $\{G^p\}$ from  $\{G^n\}$.
When the number of positive or negative samples is imbalanced, we can upsample the minor group to avoid trivial solutions.
In addition, the final \gala objective is given in Eq.~\ref{CH:GALA:eq:gala_sol} and implemented as in Eq.~\ref{CH:GALA:eq:gala_impl}.
We provide more discussions about the implementation options in Appendix~\ref{CH:GALA:sec:gala_impl_appdx}.

\subsection{Theoretical analysis}\label{CH:GALA:sec:gala_theory}
In the following theorem, we show that the \gala objective derived in Sec.~\ref{CH:GALA:sec:gala_der} can identify the underlying invariant subgraph and yields an invariant GNN defined in Sec.~\ref{CH:GALA:sec:prelim}.
\begin{theorem}\label{CH:GALA:thm:gala_success}
    Given i) the same data generation process as in Fig.~\ref{CH:GALA:fig:scm};
    ii) $\train$ that satisfies variation sufficiency (Assumption~\ref{CH:GALA:assump:var_sufficiency})
    and variation consistency (Assumption~\ref{CH:GALA:assump:var_consistency});
    iii) $\{G^p\}$ and $\{G^n\}$ are distinct subsets of $\train$ such that
    $I(G_s^p;G_s^n|Y)=0$,
    $\forall G_s^p =\argmax_{\pred{G}_s^p}I(\pred{G}_s^p;Y)$ under $\{G^p\}$, and
    $\forall G_s^n =\argmax_{\pred{G}_s^n}I(\pred{G}_s^n;Y)$ under $\{G^n\}$;
    suppose $|G_c|=s_c,\ \forall G_c$,
    resolving the following \gala objective elicits an invariant GNN defined via Eq.~\ref{CH:GALA:eq:inv_cond},
    \begin{equation}
        \label{CH:GALA:eq:gala_sol}
        \max_{f_c, g} \ I(\pred{G}_{c};Y), \ \text{s.t.}\
        g\in\argmax_{\hat{g},|\pred{G}_c^p|\leq s_c}I(\pred{G}_c^p;\pred{G}_c^n|Y),
    \end{equation}
    where $\pred{G}_c^p\in \{\pred{G}_{c}^p=g({G}^p)\}$
    and $\pred{G}_c^n\in \{\pred{G}_{c}^n=g({G}^n)\}$
    are the estimated invariant subgraphs via $g$ from $\{G^p\}$ and $\{G^n\}$, respectively.
\end{theorem}
The proof is given in Appendix~\ref{CH:GALA:proof:gala_success_appdx}.
Essentially, assumption iii) in Theorem~\ref{CH:GALA:thm:gala_success}
is an implication of the variation sufficiency (Assumption~\ref{CH:GALA:assump:var_sufficiency}).
When given the distinct subsets $\{G^p\}$ and $\{G^n\}$
with different relations of $H(C|Y)$ and $H(S|Y)$,
since $H(C|Y)$ remains invariant across different subsets,
the variation happens mostly to the spurious correlations between $S$ and $Y$.
By differentiating spurious correlations into distinct subsets,
maximizing the intra-class mutual information helps identify the true invariance.
The fundamental rationale for why \gala resolves two seemingly conversed cases essentially relies on the commutative law of mutual information.

\section{Empirical Studies}
\label{CH:GALA:sec:exp}
We evaluated \gala with both synthetic and realistic graph distribution shifts.
Specifically, we are interested in the following two questions:
(a) Can \gala improve over the state-of-the-art invariant graph learning methods
when the spurious subgraph has a stronger correlation with the labels?
(b) Will \gala affect the performance when the invariant correlations are stronger?

\subsection{Datasets and experiment setup}
We prepare both synthetic and realistic graph datasets containing various distribution shifts to evaluate \gala. We will briefly introduce each dataset and leave more details in Appendix~\ref{CH:GALA:sec:dataset_appdx}.

\paragraph{Two-piece graph datasets.} We adopt BA-2motifs~\citep{pge} to implement $4$ variants of $3$-class two-piece graph (Def.~\ref{def:twobit_graph}) datasets. The datasets contain different relationships of $H(C|Y)$ and $H(S|Y)$ by controlling the $\alpha$ and $\beta$ in the mixed environment, respectively. We consider $4$ cases of $\alpha-\beta$, ranging from $\{+0.2,+0.1,-0.1,-0.2\}$, to verify our discussion in Sec.~\ref{CH:GALA:sec:gala_theory}.

\paragraph{Realistic datasets.} We also adopt datasets containing various realistic graph distribution shifts to comprehensively evaluate the OOD performance of \gala.
We adopt $6$ datasets from DrugOOD benchmark~\citep{drugood}, which focuses on the challenging real-world task of AI-aided drug affinity prediction.
The DrugOOD datasets include splits using Assay, Scaffold, and Size from the EC50 category (denoted as \textbf{EC50-*}) and the Ki category (denoted as \textbf{Ki-*}).
We also adopt graphs converted from the ColoredMNIST dataset~\citep{irmv1} using the algorithm from~\citet{understand_att}, which contains distribution shifts in node attributes (denoted as \textbf{CMNIST-sp}).
In addition, we adopt \textbf{Graph-SST2}~\citep{xgnn_tax}, where we split graphs with a larger average degree in the training set while smaller in the test set.

\paragraph{Experiment setup.}
We adopt the state-of-the-art OOD methods from the Euclidean regime, including IRMv1~\citep{irmv1}, VREx~\citep{vrex}, EIIL~\citep{env_inference} and IB-IRM~\citep{ib-irm},
and from the graph regime, including GREA~\citep{grea}, GSAT~\citep{gsat}, CAL~\citep{cal}, MoleOOD~\citep{moleood}, GIL~\citep{gil}, DisC~\citep{disc} and CIGA~\citep{ciga}.
We exclude DIR~\citep{dir} and GIB~\citep{gib} as GREA and GSAT are their sophisticated variants.
In addition to the ERM baseline that trained a vanilla GNN with ERM objective, in two-piece motif datasets, we also include XGNN to demonstrate the failures of previous approaches, which is an interpretable GNN trained with ERM.
We also exclude CIGAv2~\citep{ciga} as \gala focuses on improving the contrastive sampling via environment assistant for the objective in CIGAv1.
All methods use the same GIN backbone~\citep{gin}, the same interpretable GNN architecture as in~\citep{gsat}, and optimization protocol for fair comparisons. We tune the hyperparameters following the common practice. Details are given in Appendix~\ref{CH:GALA:sec:eval_appdx}.

\begin{table}
    \vspace{-0.2in}
    \center
    \caption[OOD generalization performance of \gala under various invariant and spurious correlation degrees in the two-piece graph datasets.]{OOD generalization performance under various invariant and spurious correlation degrees in the two-piece graph datasets.
        Each dataset is generated from a variation of a two-piece graph model, denoted as $\{a,b\}$, where $a$ refers to the invariant correlation strength and $b$ refers to the spurious correlation strength.}
    \vspace{-0.1in}
    \label{CH:GALA:tab:syn_graph}
    \resizebox{0.65\textwidth}{!}{
        \begin{tabular}{lccccc}
            \toprule
            \textbf{Datasets}      & $\{0.8,0.6\}$            & $\{0.8,0.7\}$            & $\{0.8,0.9\}$                                    & $\{0.7,0.9\}$                                    & Avg.           \\\midrule
            ERM                    & 77.33\std{0.47}          & 75.65\std{1.62}          & 51.37\std{1.20}                                  & 42.73\std{3.82}                                  & 61.77          \\
            IRM                    & 78.32\std{0.70}          & 75.13\std{0.77}          & 50.76\std{2.56}                                  & 41.32\std{2.50}                                  & 61.38          \\
            V-Rex                  & 77.69\std{0.38}          & 74.96\std{1.40}          & 49.47\std{3.36}                                  & 41.65\std{2.78}                                  & 60.94          \\
            IB-IRM                 & 78.00\std{0.68}          & 73.93\std{0.79}          & 50.93\std{1.87}                                  & 42.05\std{0.79}                                  & 61.23          \\
            EIIL                   & 76.98\std{1.24}          & 74.25\std{1.74}          & 51.45\std{4.92}                                  & 39.71\std{2.64}                                  & 60.60          \\
            \hline
            \rule{0pt}{12pt}XGNN   & 83.84\std{0.59}          & 83.05\std{0.20}          & 53.37\std{1.32}                                  & 38.28\std{1.71}                                  & 64.63          \\
            GREA                   & 82.86\std{0.50}          & 82.72\std{0.50}          & 50.34\std{1.74}                                  & 39.01\std{1.21}                                  & 63.72          \\
            GSAT                   & 80.54\std{0.88}          & 78.11\std{1.23}          & 48.63\std{2.18}                                  & 36.62\std{0.87}                                  & 63.32          \\
            CAL                    & 76.98\std{6.03}          & 62.95\std{8.58}          & 51.57\std{6.33}                                  & 46.23\std{3.93}                                  & 59.43          \\
            MoleOOD                & 49.93\std{2.25}          & 49.85\std{7.31}          & 38.49\std{4.25}                                  & 34.81\std{1.65}                                  & 43.27          \\
            GIL                    & 83.51\std{0.41}          & 82.67\std{1.18}          & 51.76\std{4.32}                                  & 40.07\std{2.61}                                  & 64.50          \\
            DisC                   & 60.47\std{17.9}          & 54.29\std{15.0}          & 45.06\std{7.82}                                  & 39.42\std{8.59}                                  & 50.81          \\
            CIGA                   & 84.03\std{0.53}          & 83.21\std{0.30}          & 57.87\std{3.38}                                  & 43.62\std{3.20}                                  & 67.18          \\
            \textbf{\ourst}        & \textbf{84.27\std{0.34}} & \textbf{83.65\std{0.44}} & \textbf{76.42\std{3.53}} & \textbf{72.50\std{1.06}} & \textbf{79.21} \\\hline
            \rule{0pt}{12pt}Oracle & 84.73\std{0.36}          & 85.42\std{0.25}          & 84.28\std{0.15}                                  & 78.38\std{0.19}                                                   \\
            \bottomrule
        \end{tabular}
    }
\end{table}

\subsection{Experimental results and analysis}

\paragraph{Proof-of-concept study.}
The results in two-piece graph datasets are reported in Table~\ref{CH:GALA:tab:syn_graph}.
It can be found that the previous environment augmentation approaches fail either in datasets where the invariant correlations dominate or where the spurious correlations dominate, aligned with our discussions in Sec.~\ref{CH:GALA:sec:env_aug_failure}.
In particular, \grea, \ciga and \gil achieve high performance when the invariant correlation dominates, but suffer great performance decrease when the spurious correlations are stronger.
Although \disc is expected to succeed when spurious correlations dominate, \disc fails to outperform others because of its excessive destruction of the learned information.
\mole also yields degraded performance, which could be caused by the failures to infer reliable environment labels.
In contrast, \gala achieves consistently high performance under \emph{both} cases and improves CIGA up to $30\%$ under $\{0.7,0.9\}$ and $13\%$ in average, which validates our theoretical results in Sec.~\ref{CH:GALA:sec:gala_theory}.

\begin{table}[H]
    \center
    \caption[OOD generalization performance of \gala under realistic graph distribution shifts.]{OOD generalization performance under realistic graph distribution shifts.}
    \vspace{-0.15in}
    \label{CH:GALA:tab:realistic_graph}
    \resizebox{\textwidth}{!}{
        \begin{tabular}{lccccccccc}
            \toprule
            \textbf{Datasets}      & EC50-Assay               & EC50-Sca                                         & EC50-Size                & Ki-Assay                                         & Ki-Sca                                           & Ki-Size                  & CMNIST-sp                                        & Graph-SST2               & Avg.(Rank)$^\dagger$  \\\midrule
            ERM                    & 76.42\std{1.59}          & 64.56\std{1.25}                                  & 61.61\std{1.52}          & 74.61\std{2.28}                                  & 69.38\std{1.65}                                  & 76.63\std{1.34}          & 21.56\std{5.38}                                  & 81.54\std{1.13}          & 65.79 (6.50)          \\
            IRM                    & 77.14\std{2.55}          & 64.32\std{0.42}                                  & 62.33\std{0.86}          & 75.10\std{3.38}                                  & 69.32\std{1.84}                                  & 76.25\std{0.73}          & 20.25\std{3.12}                                  & 82.52\std{0.79}          & 65.91 (6.13)          \\
            V-Rex                  & 75.57\std{2.17}          & 64.73\std{0.53}                                  & 62.80\std{0.89}          & 74.16\std{1.46}                                  & 71.40\std{2.77}                                  & 76.68\std{1.35}          & 30.71\std{11.8}                                  & 81.11\std{1.37}          & 67.15 (5.25)          \\
            IB-IRM                 & 64.70\std{2.50}          & 62.62\std{2.05}                                  & 58.28\std{0.99}          & 71.98\std{3.26}                                  & 69.55\std{1.66}                                  & 70.71\std{1.95}          & 23.58\std{7.96}                                  & 81.56\std{0.82}          & 62.87 (10.6)          \\
            EIIL                   & 64.20\std{5.40}          & 62.88\std{2.75}                                  & 59.58\std{0.96}          & 74.24\std{2.48}                                  & 69.63\std{1.46}                                  & 76.56\std{1.37}          & 23.55\std{7.68}                                  & 82.46\std{1.48}          & 64.14 (8.00)          \\\hline
            \rule{0pt}{12pt}XGNN   & 72.99\std{2.56}          & 63.62\std{1.35}                                  & 62.55\std{0.81}          & 72.40\std{3.05}                                  & 72.01\std{1.34}                                  & 73.15\std{2.83}          & 20.96\std{8.00}                                  & 82.55\std{0.65}          & 65.03 (7.13)          \\
            GREA                   & 66.87\std{7.53}          & 63.14\std{2.19}                                  & 59.20\std{1.42}          & 73.17\std{1.80}                                  & 67.82\std{4.67}                                  & 73.52\std{2.75}          & 12.77\std{1.71}                                  & 82.40\std{1.98}          & 62.36 (10.1)          \\
            GSAT                   & 76.07\std{1.95}          & 63.58\std{1.36}                                  & 61.12\std{0.66}          & 72.26\std{1.76}                                  & 70.16\std{0.80}                                  & 75.78\std{2.60}          & 15.24\std{3.72}                                  & 80.57\std{0.88}          & 64.35 (8.63)          \\
            CAL                    & 75.10\std{2.71}          & 64.79\std{1.58}                                  & 63.38\std{0.88}          & 75.22\std{1.73}                                  & 71.08\std{4.83}                                  & 72.93\std{1.71}          & 23.68\std{4.68}                                  & 82.38\std{1.01}          & 66.07 (5.38)          \\
            DisC                   & 61.94\std{7.76}          & 54.10\std{5.69}                                  & 57.64\std{1.57}          & 54.12\std{8.53}                                  & 55.35\std{10.5}                                  & 50.83\std{9.30}          & 50.26\std{0.40}                                  & 76.51\std{2.17}          & 56.59 (12.4)          \\
            MoleOOD                & 61.49\std{2.19}          & 62.12\std{1.91}                                  & 58.74\std{1.73}          & 75.10\std{0.73}                                  & 60.35\std{11.3}                                  & 73.69\std{2.29}          & 21.04\std{3.36}                                  & 81.56\std{0.35}          & 61.76 (10.0)          \\
            GIL                    & 70.56\std{4.46}          & 61.59\std{3.16}                                  & 60.46\std{1.91}          & 75.25\std{1.14}                                  & 70.07\std{4.31}                                  & 75.76\std{2.23}          & 12.55\std{1.26}                                  & 83.31\std{0.50}          & 63.69 (8.00)          \\
            CIGA                   & 75.03\std{2.47}          & 65.41\std{1.16}                                  & 64.10\std{1.08}          & 73.95\std{2.50}                                  & 71.87\std{3.32}                                  & 74.46\std{2.32}          & 15.83\std{2.56}                                  & 82.93\std{0.63}          & 65.45 (5.88)          \\
            \textbf{\ourst}        & \textbf{77.56\std{2.88}} & \textbf{66.28\std{0.45}} & \textbf{64.25\std{1.21}} & \textbf{77.92\std{2.48}} & \textbf{73.17\std{0.88}} & \textbf{77.40\std{2.04}} & \textbf{68.94\std{0.56}} & \textbf{83.60\std{0.66}} & \textbf{73.64 (1.00)} \\\hline
            \rule{0pt}{12pt}Oracle & 84.77\std{0.58}          & 82.66\std{1.19}                                  & 84.53\std{0.60}          & 91.08\std{1.43}                                  & 88.58\std{0.64}                                  & 92.50\std{0.53}          & 67.76\std{0.60}                                  & 91.40\std{0.26}          &                       \\
            \bottomrule
            \multicolumn{8}{l}{\rule{0pt}{12pt}$^\dagger$\text{\normalfont Averaged rank is also reported in the parentheses because of dataset heterogeneity. A lower rank is better.}  }
        \end{tabular}
    }
\end{table}

\paragraph{OOD generalization in realistic graphs.}
The results in realistic datasets are reported in Table~\ref{CH:GALA:tab:realistic_graph}.
Aligned with our previous discussion, existing environment augmentation approaches sometimes yield better performance than ERM, such as CAL in EC50-Size, \mole in Ki-Assay, \gil in Graph-SST2, or \ciga in EC50-Size, however, inevitably fail to bring consistent improvements than ERM, due to the existence of failure cases.
\disc is suspected to work only for graph distribution shifts on node features and bring impressive improvements in CMNIST-sp, but can destroy the learned information under more challenging settings.
In contrast, \gala consistently outperforms ERM by a non-trivial margin in all datasets. Notably, \gala achieves near oracle performance in CMNIST-sp and improves \ciga by $53\%$.
The consistent improvements of \gala confirm the effectiveness of \gala.

\begin{figure}[H]
    \centering
    \subfigure[Correlation strengths]{
        \includegraphics[width=0.32\textwidth]{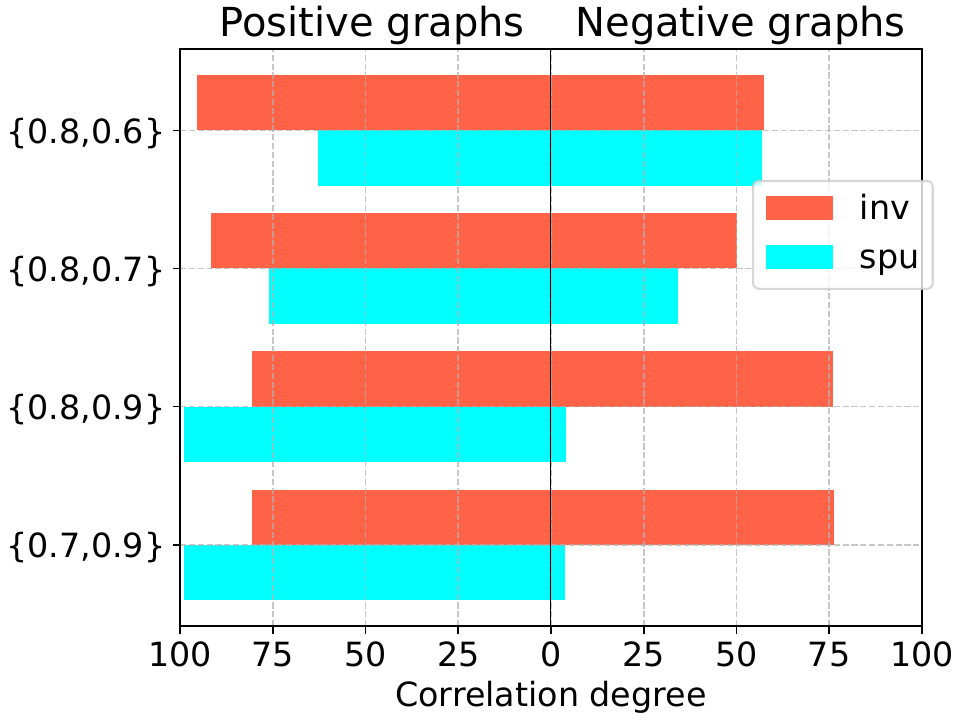}
        \label{CH:GALA:fig:corr_change}
    }
    \subfigure[CIGAv2 compatibility]{
        \includegraphics[width=0.32\textwidth]{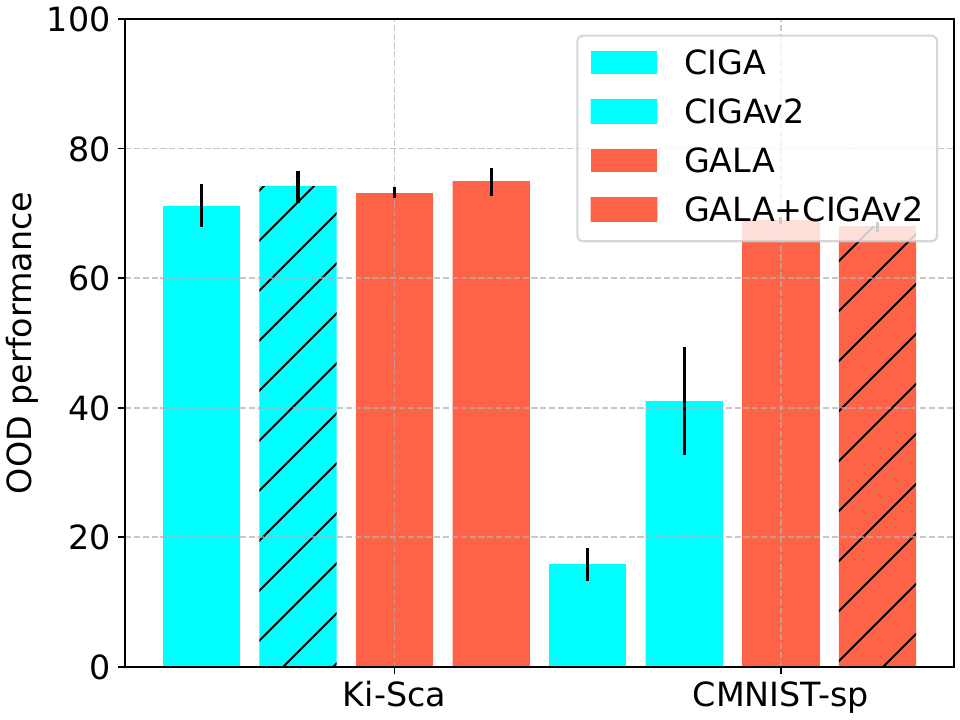}
        \label{CH:GALA:fig:cigav2_comp}
    }
    \subfigure[Hyperparameter sensitivity]{
        \includegraphics[width=0.28\textwidth]{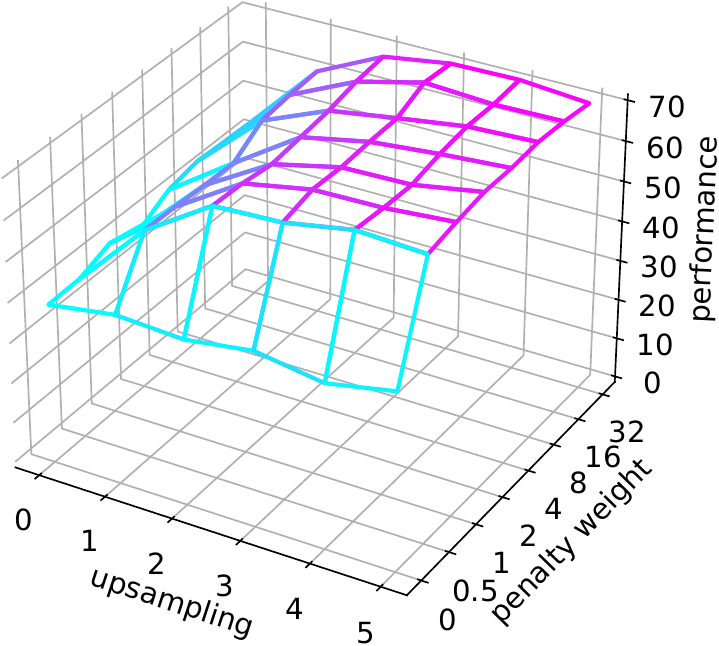}
        \label{CH:GALA:fig:hyper_sens}
    }
    \caption[Ablation studies of \gala.]{Ablation studies.}
    \label{CH:GALA:fig:ablation}
\end{figure}

\paragraph{Correlation strengths of $\{G^p\}$ and $\{G^n\}$.}
We conduct experiments with the two-piece graph datasets evaluated in Table~\ref{CH:GALA:tab:syn_graph} to verify the correctness of Eq.~\ref{CH:GALA:eq:gala_sol_spu} and Eq.~\ref{CH:GALA:eq:gala_sol_inv}.
Eq.~\ref{CH:GALA:eq:gala_sol_spu} and Eq.~\ref{CH:GALA:eq:gala_sol_inv} imply that the underlying invariant subgraph will be the subgraph that maximizes the mutual information among subgraphs from $\{G^p\}$ and $\{G^n\}$, no matter whether the dominant correlation is spurious or not.
We measure the invariant and spurious correlation strengths in terms of co-occur probability of the invariant and spurious subgraphs with the labels.
The results are shown in Fig.~\ref{CH:GALA:fig:corr_change}. It can be found that, under both cases, the underlying invariant subgraph maintains the predictivity with the label in an invariant manner.
Hence, maximizing the intra-class subgraph mutual information between $\{G^p\}$ and $\{G^n\}$ in \gala succeeds in identifying the underlying invariant subgraph.

\paragraph{CIGAv2 compatibility.}
Although \gala focuses on the contrastive term in CIGA, both \gala and CIGA are compatible with the additional CIGAv2 term that facilitates constraining the graph sizes.
To verify, we compare the OOD performances of CIGA, CIGAv2, \gala, and \gala+CIGAv2 using two challenging datasets, Ki-Scaffold and CMNIST-sp.
The results are given in Fig.~\ref{CH:GALA:fig:cigav2_comp}.
It can be found that, despite incorporating the additional CIGAv2 constraint, CIGA can not outperform \gala, while \gala can bring more improvements with the additional CIGAv2 constraint.
In CMNIST-sp, since \gala already achieve the upper bound, incorporating CIGAv2 can only achieve a similar result.

\paragraph{Hyperparameter sensitivity.}
We also test the hyperparameter sensitivity of \gala to the contrastive penalty weights as well as the upsampling times that are introduced to mitigate the imbalance of positive and negative graphs. We conduct the experiments with two-piece graph dataset $\{0.7,0.9\}$.
As shown in Fig.~\ref{CH:GALA:fig:hyper_sens}, it can be found that \gala is generically robust to different hyperparameter choices.
In addition, when the penalty weight or the upsampling times turn to $0$, the performance will decrease a lot, which serves as strong evidence for the effectiveness of \gala.

\paragraph{Computational analysis.} We also conduct computational analysis of \gala and other methods, and defer the results to Table.~\ref{CH:GALA:tab:time_analysis} in Appendix~\ref{CH:GALA:sec:time_appdx}, due to space constraints. The results show that \gala costs only a competitive training time as environment generation based methods, while achieving much better OOD generalization performance.

\part{Implications}\label{P2}
\chapter{Causality in Interpretability} \label{CH:GMT}

\section{Motivations}
Graph Neural Networks (GNNs) have been widely used in scientific applications~\citep{ai4sci0,ai4sci} such as Physics~\citep{ai4sci_phy}, Chemistry~\citep{ai4sci_qchem,ai4sci_bchem}, Quantum mechanics~\citep{ai4sci_qua}, Materials~\citep{ai4sci_mat} and Cosmology~\citep{ai4sci_cos}.
In pursuit of scientific discoveries, it often requires GNNs to be able to generalize to \textit{unseen or Out-of-Distribution} (OOD) graphs~\citep{good_bench,drugood,ai4sci}, and also provide \textit{interpretations} of the predictions that are crucial for scientists to collect insights~\citep{ai4sci_xgnn0,ai4sci_xgnn1,ai4sci_xgnn2} and promote better scientific practice~\citep{xgnn_ai4sci0,xgnn_ai4sci1}.
Recently there has been a surge of interest in developing intrinsically interpretable and generalizable GNNs (\xgnns)~\citep{gib,gsat,dir,ciga,lri}.
In contrast to \textit{post-hoc} explanations~\citep{gnn_explainer,xgnn,pgm_explainer,pge,subgraphxgn,gen_xgnn,orphicx} which are shown to be suboptimal in interpretation and sensitive to pre-trained GNNs performance~\citep{gsat,lri}, \xgnns can provide both reliable explanations and (OOD) generalizable predictions under the proper guidance such as information bottleneck~\citep{gib} and causality~\citep{ciga}.

Indeed, the faithful interpretation and the reliable generalization are the \textit{two sides of the same coin} for \xgnns.
Grounded in the causal assumptions of data generation processes, \xgnns assume that there exists a causal subgraph which holds a causal relation with the target label. Predictions made solely based on the causal subgraph are generalizable under various graph distribution shifts~\citep{handle_node,gsat,ciga}.
Therefore, \xgnns typically adopt a two-step paradigm that first extracts a subgraph of the input graph and then predicts the label.
To circumvent the inherent discreteness of subgraphs, \xgnns often learn the sampling probability for each edge or node with the attention mechanism and extract the subgraph with high attention scores~\citep{gsat}.
Predictions are then made via a weighted message passing scheme with the attention scores.
Despite the success of the paradigm in enhancing both interpretability and out-of-distribution (OOD) generalization
~\citep{gsat,lri,ciga},
there is limited theoretical understanding of the representational properties and limitations of \xgnns, and whether they can provide faithful interpretations.

\begin{figure*}[t]
  \centering
  \includegraphics[width=0.9\textwidth]{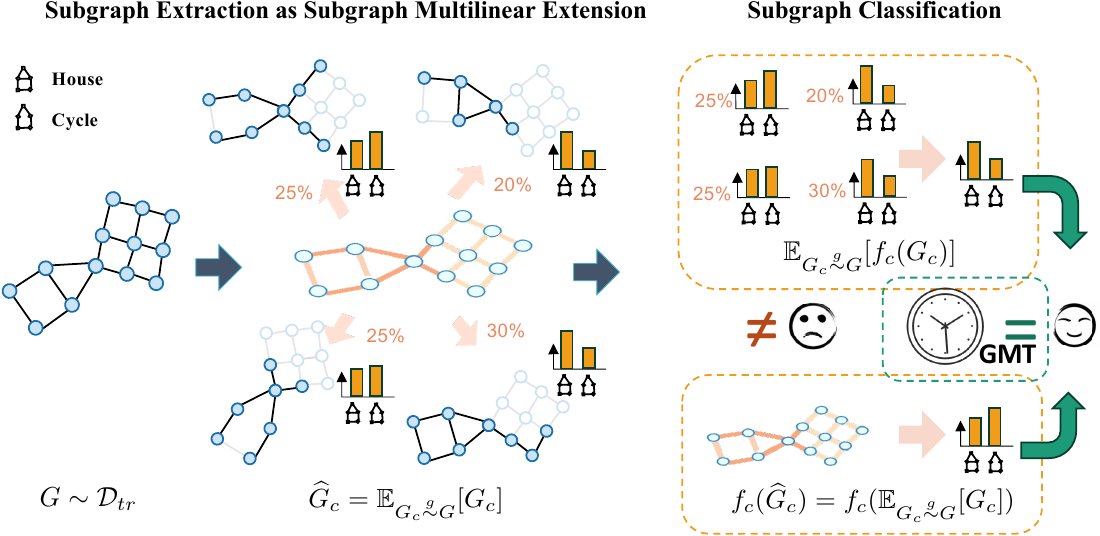}
  \caption[Illustration of Subgraph Multilinear Extension (\smt).]{Illustration of Subgraph Multilinear Extension (\smt).
    The task is to classify whether a graph contains a specific ``house'' or ``cycle'' motif.
    An \xgnn $f=f_c\circ g$ predicts the label with the classifier $f_c$ based on the extracted soft subgraph $\widehat{G}_c=g(G)$, denoted as the central graph. Different intensities of edge colors refer to the sampling probability of the edge appearing in the interpretation.
    $\widehat{G}_c$ corresponds to a subgraph distribution
    with respect to the sampling probability of each subgraph $G_c$ (i.e., subgraphs with solid lines in the figure).
    \smt extends GNNs to accept soft subgraph inputs by estimating the subgraph conditional prediction
    as the expectation of each possible subgraph $\E [f_c(G_c)]$.
    Interpretable subgraph learning requires an accurate estimation of the subgraph conditional prediction distribution.
    Yet existing \xgnns that directly input the soft subgraph $\widehat{G}_c$ to the classifier GNN will lead to a biased estimation of \smt.
    \gmt is designed to bridge the gap by learning a neural \smt to approximate \smt.
  }
  \label{CH:GMT:fig:gmt_illustration}
\end{figure*}

Inspired by the close connection between interpretable subgraph learning and multilinear extension~\citep{mt}, we present a framework to analyze the expressiveness and evaluate the faithfulness of \xgnns.
In fact, the subgraph learning in \xgnns naturally resembles the multilinear extension of the subgraph predictivity, which we term as \textit{\smtfull} (\smt).
The extracted interpretable subgraph is faithful if the associated prediction is highly correlated with the sampling probability of the subgraph.
However, we show that the prevalent attention-based paradigm can fail to reliably approximate \smt (Sec.~\ref{CH:GMT:sec:expressivity_issue}).
Consequently, the \smt approximation failure will decrease the interpretability of the subgraph for predicting the target label.
More specifically, we instantiate the issue via a causal framework and propose a new interpretability measure called \textit{counterfactual fidelity}, i.e., the sensitivity of the prediction to small perturbations to the extracted subgraphs (Sec.~\ref{CH:GMT:sec:counterfactual_faith}).
Although faithful interpretation should have a high counterfactual fidelity with the prediction, we find that \xgnns implemented with the prevalent paradigm only have a low counterfactual fidelity.

To bridge the gap, we propose a simple yet effective \xgnn architecture called \gmtfull (\gmt). Motivated by the \smt formulation, \gmt first performs random subgraph sampling onto the subgraph distribution to approximate \smt, which is provably more powerful in approximating \smt (Sec.~\ref{CH:GMT:sec:gmt_sol}). Then, we will train a new classifier onto the trained subgraph extractor without random subgraph sampling, to obtain the final approximator of neural \smt.
Our contributions can be summarized as follows:
\begin{itemize}[leftmargin=*]
  \item We propose the first theoretical framework through the notion of \smt for the expressivity of XGNNs (Sec.~\ref{CH:GMT:sec:expressivity});
  \item We propose a new XGNN architecture \gmt that is provably more powerful than previous XGNNs. The key differentiator of \gmt is a new paradigm to effectively approximate \smt with random subgraph sampling (Sec.~\ref{CH:GMT:sec:gmt_sol}).
  \item We validate both our theory and the solution through extensive experiments with $12$ regular and geometric graph benchmarks. The results show that \gmt significantly improves the state-of-the-art up to $10\%$ in both interpretability and generalizability (Sec.~\ref{CH:GMT:sec:exp}).
\end{itemize}

\section{Preliminaries and Related Work}
We begin by introducing preliminary concepts of \xgnns and leave more details to Appendix~\ref{CH:GMT:sec:related_appdx}, and also provide a table of notations for key concepts in Appendix~\ref{CH:GMT:sec:notations_appdx}.

\textbf{Interpretable GNNs.}
Let $G=(A,X)$ be a graph with node set $V=\{v_1,v_2,...,v_n\}$ and edge set $E=\{e_1,e_2,...,e_m\}$,
where  $A \in \{0,1\}^{n\times n}$  is the adjacency matrix and $X\in \R^{n \times d}$ is the node feature matrix.
In this work, we focus on interpretable GNNs (or \xgnns) for the graph classification task, while the results can be generalized to node-level tasks as well~\citep{gib_node}.
Given each sample from training data $\train=(G^i,Y^i)$,
an interpretable GNN $f:=f_c\circ g$ aims to identify a (causal) subgraph $G_c\subseteq G$ via a subgraph extractor GNN $g:\gG\rightarrow\gG_c$, and then predicts the label via a subgraph classifier GNN $f_c:\gG_c\rightarrow\gY$, where $\gG,\gG_c,\gY$ are the spaces of graphs, subgraphs, and the labels, respectively~\citep{gib}.
Although \textit{post-hoc} explanation approaches also aim to find an interpretable subgraph as the explanation for the model prediction~\citep{gnn_explainer,xgnn,pgm_explainer,pge,subgraphxgn,gen_xgnn,orphicx}, they are shown to be suboptimal in interpretation performance and sensitive to the performance of the pre-trained GNNs~\citep{gsat}.
Therefore, this work focuses on \textit{intrinsic interpretable} GNNs (XGNNs).

A predominant approach to implement \xgnns is to incorporate the idea of information bottleneck~\citep{ib}, such that $G_c$ keeps the minimal sufficient information of $G$ about $Y$~\citep{gib,vgib,gsat,lri,gib_hiera},
which can be formulated as
\begin{equation}
  \text{$\max$}_{G_c}I(G_c;Y)-\lambda I(G_c;G),\ G_c\sim g(G),
\end{equation}
where the maximizing $I(G_c;Y)$ endows the interpretability of $G_c$ while minimizing $I(G_c;G)$ ensures $G_c$ captures only the most necessary information, $\lambda$ is a hyperparamter trade off between the two objectives.
In addition to minimizing $I(G_c;G)$, there are also alternative approaches that impose different constraints such as causal invariance~\citep{ciga,gil} or disentanglement~\citep{dir,cal,grea,disc} to identify the desired subgraphs.
When extracting the subgraph, \xgnns adopts the attention mechanism to learn the sampling probability of each edge or node, which avoids the complicated Monte Carlo tree search used in other alternative implementations~\citep{protGNN}.
Specifically, given node representation learned by message passing $H_i\in\R^h$ for each node $i$, \xgnns either learns a \textbf{node attention} $\alpha_i\in\R_+=\sigma(a(H_i))$ via the attention function $a:\R^h\rightarrow\R_+$, or the \textbf{edge attention} $\alpha_e\in\R_+=\sigma(a([H_u,H_v]))$ for each edge $e=(u,v)$ via the attention function $a:\R^{2h}\rightarrow\R_+$, where $\sigma(\cdot)$ is a sigmoid function. $\boldsymbol{\alpha}=[\alpha_1,...,\alpha_m]^T$ essentially elicits a subgraph distribution of the interpretable subgraph. In this work, we focus on edge-centric subgraph sampling as it is most widely used in \xgnns while our method can be easily generalized to node-centric approaches.

\paragraph{Faithful interpretation and (OOD) generalization.}
The faithfulness of interpretation is critical to all interpretable and explainable methods~\citep{fidelity,mythos_inter,robust_xnn,att_not_exp}. There are several metrics developed to measure the faithfulness of graph explanations, such as fidelity~\citep{xgnn_tax,GraphFramEx}, counterfactual robustness~\citep{RCExplainer,counterfactual_xgnn_sur,clear}, and equivalence~\citep{xgnn_equi}, which are however limited to post-hoc graph explanation methods. In contrast, we develop the first faithfulness measure for \xgnns in terms of counterfactual invariance.

In fact, the generalization ability and the faithfulness of the interpretation are naturally intertwined in \xgnns. \xgnns need to extract the underlying ground-truth subgraph in order to make correct predictions on unseen graphs~\citep{gsat}. When distribution shifts are present during testing, the underlying subgraph that has a causal relationship with the target label (or causal subgraphs) naturally becomes the ground-truth subgraph that needs to be learned by \xgnns~\citep{ciga}.

\paragraph{Multilinear extension}
serves as a powerful tool
for maximizing combinatorial functions, especially for submodular set function maximization \citep{mt,Vondrak08,optimal_drsub,sets2multisets,neural_set}.
It is the expected value of a set function under the fully factorized Bernoulli distribution.
Our work is the first to identify subgraph multilinear extension as the factorized subgraph distribution for interpretable subgraph learning.

\section{On the Expressivity of Interpretable GNNs}
\label{CH:GMT:sec:expressivity}
In this section, we present our theoretical framework for characterizing the expressivity of \xgnns. Since all existing methods need to maximize $I(G_c;Y)$ regardless of the regularization on $G_c$, we focus on modeling the subgraph distribution that maximizes $I(G_c;Y)$.

\subsection{Subgraph multilinear extension}
The need for maximizing $I(G_c;Y)$ originates from extracting information in $G$ to predict $Y$ with $f_c$. %
The estimating and maximizing $I(G_c;Y)$ in \xgnns can be formulated as:
\begin{equation}\label{CH:GMT:eq:GI}
  \begin{aligned}
    \text{$\argmax$}_{f_c} I(G;Y) & = \text{$\argmax$}_{f_c} [H(Y)-H(Y|G)] \\&=\text{$\argmin$}_{f_c} H(Y|G),
  \end{aligned}
\end{equation}
where the last equality is due to the irrelevance of $H(Y)$ and $f_c$.
For each sample $(G,Y)$, \xgnn then adopts the subgraph extractor $g$ to extract a subgraph $G_c\sim g(G)$, and take $G_c$ as the input of $f_c$ to predict $Y$.
Then, Eq.~\ref{CH:GMT:eq:GI} is realized as follows\footnote{With a bit of abuse of notations, we will omit the unnecessary superscript of samples for the sake of clarity.}: let $L(\cdot)$ be the cross-entropy loss, then
\begin{equation}\label{CH:GMT:eq:GCE}
  \begin{aligned}
    \text{$\argmin$}_{g,f_c} & \  \E_{(G,Y)\sim\train}[-\log P(Y|\E_{G_c\stackrel{g}{\sim}G}G_c)] \\
                             & =\E_{(G,Y)\sim\train} [L(f_c(\boldsymbol{\alpha};G),Y)],
  \end{aligned}
\end{equation}
where $\boldsymbol{\alpha}\in\R^m_+$ is the attention score elicited from the subgraph extractor $g$.
We leave more details about the deduction of Eq.~\ref{CH:GMT:eq:GCE} in Appendix~\ref{CH:GMT:sec:GCE_deduce_appdx}.
Note that $f_c$ is a GNN defined only for \textit{discrete} graph-structured inputs (i.e., $\boldsymbol{\alpha}\in\{0,1\}^m$), while Eq.~\ref{CH:GMT:eq:GCE} imposes continuous inputs to $f_c$.
Considering $f_c(G_c)$ is a \textit{set function} with respect to node/edge index subsets of $G$ (i.e., subgraphs $G_c$),
and the parameterization of $P(G)$ in \xgnns~\citep{gsat},
we resort to the \textit{multilinear extension} of $f_c(G_c)$.
Multilinear extension for set functions has been extensively studied in the domain of solving classical combinatorial optimization problems ~\citep{mt,neural_set}.

\begin{definition}[Subgraph multilinear extension (\smt)]\label{def:sub_mt}
  Given the attention score $\boldsymbol{\alpha}\in [0, 1]^m$ as sampling probability of $G_c$, \xgnns factorize $P(G)$ as independent Bernoulli distributions on edges:
  \[P(G_c|G)=\prod_{e\in G_c}\alpha_e\prod_{e\in G/ G_c}(1-\alpha_e),\]
  which elicits the \textit{multilinear extension} of $f_c(G_c)$ in Eq.~\ref{CH:GMT:eq:GCE}:
  \begin{equation}\label{CH:GMT:eq:smt}
    \begin{aligned}
      F_c(\boldsymbol{\alpha}; G) & :=\sum_{G_c\in G}f_c(G_c)\prod_{e\in G_c}\alpha_e\prod_{e\in G/G_c}(1-\alpha_e) \\&=\E_{G_c\stackrel{g}{\sim}G}f_c(G_c).
    \end{aligned}
  \end{equation}
\end{definition}

The parameterization of $P(G)$ is widely employed in \xgnns~\citep{gsat,ciga}, which implicitly assumes the random graph data model~\citep{er_graph}. Def.~\ref{def:sub_mt} can also be generalized to other graph models with the corresponding parameterization of $P(G)$~\citep{sbm,graphon}.
When a \xgnn approximates \smt well, we have:
\begin{definition}[$\epsilon$-\smt approximation]\label{def:submt_approx}
  Let $d(\cdot,\cdot)$ be a distribution distance metric, a \xgnn $f=f_c\circ g$ $\epsilon$-approximates \smt (Def.~\ref{def:sub_mt}), if there exists $\epsilon\in\R_+$ such that $d(P_f(Y|G),P(Y|G))\leq\epsilon$
  where $P(Y|G)\in\R^{|\gY|}$ is the ground truth conditional label distribution, and $P_f(Y|G)\in\R^{|\gY|}$ is the predicted label distribution for $G$ via a \xgnn $f$, i.e., $P_f(Y|G)=f_c(\E_{G_c\stackrel{g}{\sim}G}G_c)$.
\end{definition}
Def.~\ref{def:submt_approx} is a natural requirement for \xgnn that approximates \smt properly.
With the definition of \smt, we can write the objective in Eq.~\ref{CH:GMT:eq:GCE} as the following:
\begin{equation}\label{CH:GMT:eq:GCE_mt}
  \begin{aligned}
     & \ \E_{(G,Y)\sim\train} [L(\E_{G_c\stackrel{g}{\sim}G}f_c(G_c),Y)] \\
     & =\E_{(G,Y)\sim\train}L(F_c(\boldsymbol{\alpha}; G), Y),
  \end{aligned}
\end{equation}
from which it suffices to know that optimizing for $g,f_c$ in Eq.~\ref{CH:GMT:eq:GCE} requires an accurate estimation of \smt.

\subsection{Issues of existing approaches}
\label{CH:GMT:sec:expressivity_issue}
In general, evaluating \smt requires $\gO(2^m)$ calls of $f_c(G_c)$.
Nonetheless, existing \xgnns introduce a soft subgraph $\widehat{G}_c$ with the adjacency matrix as the attention matrix $\widehat{A}$ where $\widehat{A}_{u,v}\!=\!\alpha_e, \forall e\!=\!(u,\!v)\!\in\! E$, to solve Eq.~\ref{CH:GMT:eq:GCE} via weighted message passing~\citep{gsat}:
\begin{equation}\label{CH:GMT:eq:GCE_att}
  \begin{aligned}
     & \E_{(G,Y)\sim\train} [L(\E_{G_c\stackrel{g}{\sim}G}f_c(G_c),Y)] \\
     & = \E_{(G,Y)\sim\train}[L(f_c(\widehat{G}_c),Y)].
  \end{aligned}
\end{equation}
From the edge-centric perspective, introducing $\widehat{G}_c$ seems to be natural at first glance, as:
\begin{equation}\label{CH:GMT:eq:GCE_att_exp}
  \widehat{G}_c=\E_{G_c\stackrel{g}{\sim}G}G_c.%
\end{equation}
However, Eq.~\ref{CH:GMT:eq:GCE_att} holds only when $f_c$ is \textit{linear}. In other words, if Eq.~\ref{CH:GMT:eq:GCE_att} holds, we need the following to hold:
\begin{equation}\label{CH:GMT:eq:exp_issue}
  f_c(\widehat{G}_c)=f_c(\E[G_c])=\E[f_c(G_c)],
\end{equation}
where the last equality adheres to the equality of Eq.~\ref{CH:GMT:eq:GCE_att}.
Obviously $f_c(\cdot)$ is a non-linear function even with a linearized GNN~\citep{sgnn} with more than $1$ layers:
\begin{equation}\label{CH:GMT:eq:linear_gnn}
  f_c(\widehat{G}_c)=\rho(\widehat{A}^kX \mW),
\end{equation}
where $\rho$ is the pooling, $k$ is the number of layers and $\mW\in\R^{h\times h}$ are the learnable weights.
We prove the \smt approximation failure in Appendix~\ref{proof:submt_gap}.
\begin{proposition}\label{CH:GMT:thm:submt_gap}
  An XGNN based on linear GNN with $k>1$ cannot satisfy
  Eq.~\ref{CH:GMT:eq:exp_issue}, thus cannot  approximate \smt.
\end{proposition}
When given more complicated GNNs, the approximation error to \smt can be even higher, as verified in Appendix~\ref{CH:GMT:sec:smt_gap_viz_appdx}.
For example, when $k=2$ and $|\gY|=1$, Eq.~\ref{CH:GMT:eq:linear_gnn} is convex, and we have $f_c(\E[A])\leq\E[f_c(A)]$ due to Jensen's inequality, which introduces the Jensen gap as $\E[f_c(A)]-f_c(\widehat{A})$ when fitting \smt.

\begin{figure*}[t]
  \centering
  \subfigure[SCM of \xgnns.]{\label{CH:GMT:fig:scm_ber}
    \raisebox{0.4\height}{\includegraphics[width=0.35\textwidth]{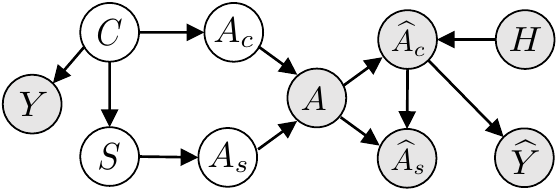}}
  }
  \subfigure[\smt on BA-2Motifs.]{\label{CH:GMT:fig:counterfactual_fidelty_ba}
    \includegraphics[width=0.29\textwidth]{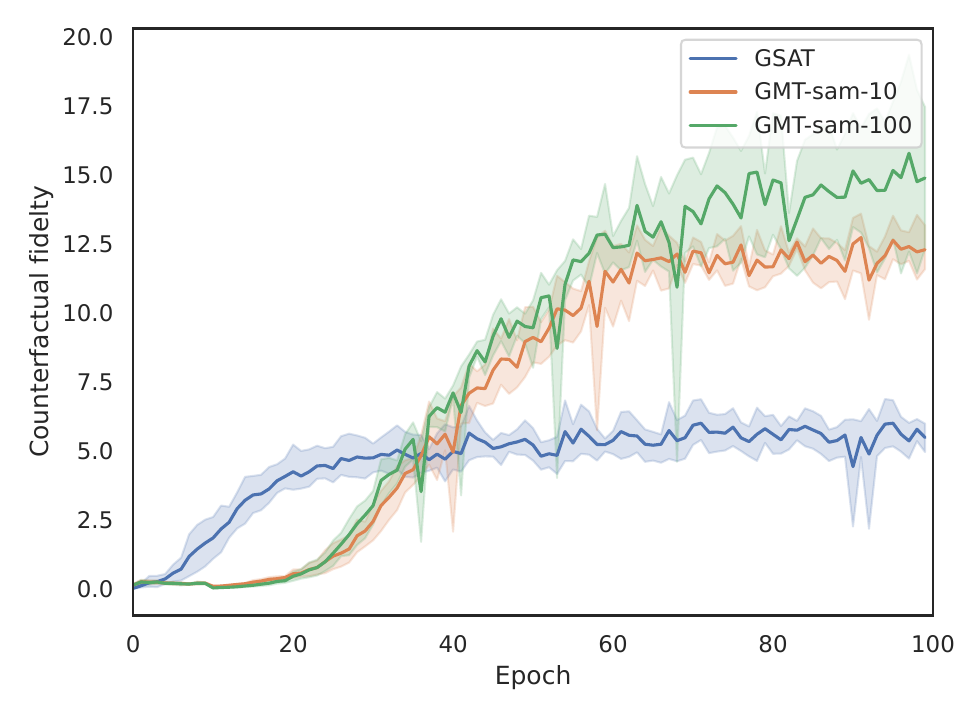}
  }
  \subfigure[\smt on Mutag.]{\label{CH:GMT:fig:counterfactual_fidelty_mutag}
    \includegraphics[width=0.29\textwidth]{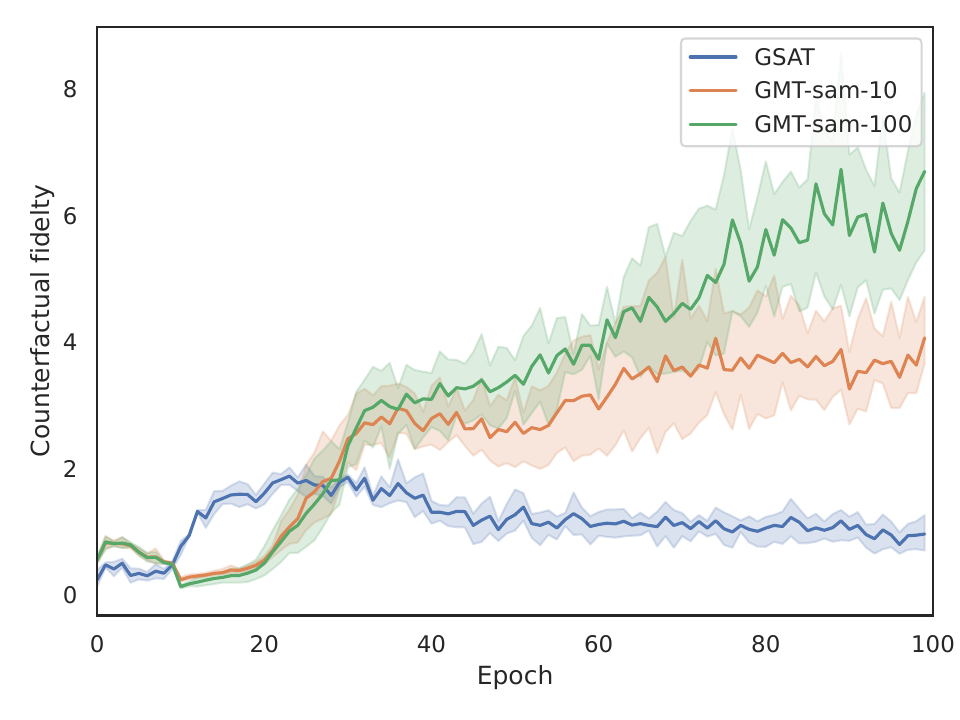}
  }
  \caption{Illustration of counterfactual faithfulness.}
\end{figure*}
\section{On the Generalization and Interpretability: A Causal View}
\label{CH:GMT:sec:causal_view}
To understand the consequences of the \smt approximation issue,
we conduct a causal analysis of the interpretation faithfulness in \xgnns.
Without loss of generality, we will focus on the edge-centric data generation and interpretation.
\subsection{Causal model of interpretable GNNs}
\label{CH:GMT:sec:scm_xgnn}

\textbf{Data generation.} We consider the same data model as previous works~\citep{size_gen2,gsat,ciga},
where the underlying causal subgraph $G_c$ and the spurious subgraph $G_s$ will be assembled via some underlying assembling process. As we focus on the edge-centric view, our following discussion will focus on the graph structures $A_c$ and $A_s$ of the subgraphs. Full details of the structural causal model are deferred to  Appendix~\ref{CH:GMT:sec:scm_xgnn_appdx}.

As shown in Fig.~\ref{CH:GMT:fig:scm_ber}, there are latent causal and spurious variables $C$ and $S$ that have invariant and spurious correlations with the label $Y$ across training and test distributions, respectively.
$C$ and $S$ correspondingly control the generation of causal subgraph $G_c$, and the spurious subgraph $G_s$. For example, when generating $A_c$ and $A_s$, $C$ and $S$ will specify the number of nodes in  $A_c$ and $A_s$ and also the edge sampling probability for edges in $A_c$ and $A_s$, respectively.

\textbf{Interpretation.} Correspondingly, \xgnns uses a subgraph extractor to predict the causal and spurious subgraphs $\widehat{G}_c$ and $\widehat{G}_s$, respectively.
The extraction aims to reverse the generation and recover the structure of the underlying causal subgraph $A_c$.  We denote the \xgnn architecture and the hyperparameter settings as $H$.
$H$ takes $A$ as inputs to learn the edge sampling probability via the attention mechanism and then obtain $\widehat{A}_c$.
Once $\widehat{A}_c$ is determined, $\widehat{A}_s\!=\!A\!-\!\widehat{A}_c$ is also obtained by taking the complementary part. Then, the extracted causal and spurious subgraphs are obtained with $\widehat{G}_c\!=\!(X,\widehat{A}_c)$ and $\widehat{G}_s\!=\!(X,\widehat{A}_s)$, respectively. The classifier then uses $\widehat{G}_c$ to make the prediction $\widehat{Y}$.

\subsection{Causal faithfulness of \xgnns}
\label{CH:GMT:sec:counterfactual_faith}

With the aforementioned causal model, we are able to specify the causal desiderata for faithful \xgnns.
When a \xgnn fails to accurately approximate \smt, the estimated label conditional probability will have a huge gap from the ground truth.
The failure will bias the optimization of the subgraph extractor $g$ and lead to the degenerated interpretability of $\widehat{A}$.
More concretely, the recovery of $\widehat{A}$ to the underlying $A$ will be worse, which further affects the extraction of $G_c$ and brings both worse interpretation and (OOD) generalization performance.
As a single measure such as the interpretation or generalization may not fully reflect the consequence or even exhibit conflicted information\footnote{For example, in the experiments of~\citet{gsat}, higher interpretation performance does not necessarily correlate with higher generalization performance.}, we consider a direct notion that jointly consider the interpretability and generalizabiliy to measure the causal faithfulness of \xgnns, inspired by~\citet{att_not_exp}.
\begin{definition}[$(\delta,\epsilon)$-counterfactual fidelity]\label{def:counterfactual_x}
  Given a meaningful minimal distance $\delta>0$,
  let $d(\cdot,\cdot)$ be a distribution distance metric ,
  if a \xgnn $f=f_c\circ g$ commits to the $\epsilon-$counterfactual fidelity, then there exist $\epsilon>0$ such that, $\forall G, \widetilde{G}$ that $d(P(Y|G),P(Y|\widetilde{G}))\geq \delta$, the following holds:
  \[d(P_f(Y|\widetilde{G}),P_f(Y|G))\geq \epsilon\delta.\]
\end{definition}
Intuitively, if the extracted interpretable subgraph $\widehat{G}_c$ is faithful to the target label, then the predictions made based on $\widehat{G}_c$ are sensitive to any perturbations on $\widehat{G}_c$.
Different from counterfactual interpretability~\citep{counterfactual_xgnn_sur,counterfactual_gl} that seeks minimum modifications to change the predictions,
$(\delta,\epsilon)$-counterfactual fidelity measures how sensitive are the predictions to the changes of the interpretable subgraphs. A higher fidelity implies better interpretability and is also a natural behavior of a \xgnn that approximates \smt well.
\begin{proposition}\label{CH:GMT:thm:smt_fidelty}
  If a \xgnn $f$ $\epsilon$-approximates \smt,  $f$ satisfies $(\delta,1\!-\!\frac{2\epsilon}{\delta})$-counterfactual fidelity.
\end{proposition}
The proof is given in Appendix~\ref{proof:smt_fidelty}. Intuitively, Proposition~\ref{CH:GMT:thm:smt_fidelty} implies that the counterfactual fidelity is an effective measure for the approximation ability of \smt.

\textbf{Practical estimation of counterfactual fidelity.}
Since it is hard to enumerate every possible $\widetilde{G}$,
to verify Def.~\ref{def:counterfactual_x}, we consider a random attention matrix $\widetilde{A}\sim\sigmoid(\gN(\mu_{\widehat{H}_A},\sigma_{\widehat{H}_A}))$,
where $\mu_{\widehat{H}_A}$ and $\sigma_{\widehat{H}_A}$ are the mean and standard deviation
of the pre-attention matrix $\widehat{H}_A$ (The adjacency matrix with the unnormalized attention).
Each non-symmetric entry in $\widetilde{A}$ is sampled independently following the factorization of $P(G)$.
We randomly sample $\widetilde{A}$ by $k$ times and obtain
\begin{equation}
  c_{\widehat{G}_c} = \frac{1}{k}\sum_{i=1}^k d(f_c(Y|\widetilde{G}^i_c),f_c(Y|\widehat{G}_c)),
\end{equation}
where $\widetilde{G}^i_c=(X,\widetilde{A}^i_c)$ and $d$ is total variation distance.
We compute $c_{\widehat{G}_c}$ for the state-of-the-art \xgnn \gsat~\citep{gsat}.
Shown as in Fig.~\ref{CH:GMT:fig:counterfactual_fidelty_ba},~\ref{CH:GMT:fig:counterfactual_fidelty_mutag},
we plot the counterfactual fidelity of \gsat on BA-2Motifs and Mutag datasets against
is $2$ to $3$ times lower than the simulated \smt with $10$ and $100$ sampling rounds.
We provide a more detailed discussion in Appendix~\ref{CH:GMT:sec:pratical_cf_appdx} and Appendix~\ref{CH:GMT:sec:cf_viz_appdx}.
\begin{table*}[t]
    \caption[Interpretation Performance (AUC) of \gmt on regular graphs.]{Interpretation Performance (AUC) on regular graphs. %
    }
    \small\sc\centering
    \resizebox{\textwidth}{!}{
        \begin{tabular}{llcccccc}
            \toprule
            \multirow{2}{*}{GNN} & \multirow{2}{*}{Method} & \multirow{2}{*}{Ba-2motifs}                        & \multirow{2}{*}{Mutag}                             & \multirow{2}{*}{MNIST-75sp}                        & \multicolumn{3}{c}{Spurious-motif}                                                                                                                           \\
                                 &                         &                                                    &                                                    &                                                    & $b=0.5$                                            & $b=0.7$                                            & $b=0.9$                                            \\
            \midrule\multirow{5}{*}{GIN}
                                 & GNNExplainer            & $67.35$\std{3.29}                                  & $61.98$\std{5.45}                                  & $59.01$\std{2.04}                                  & $62.62$\std{1.35}                                  & $62.25$\std{3.61}                                  & $58.86$\std{1.93}                                  \\
                                 & PGExplainer             & $84.59$\std{9.09}                                  & $60.91$\std{17.10}                                 & $69.34$\std{4.32}                                  & $69.54$\std{5.64}                                  & $72.33$\std{9.18}                                  & $72.34$\std{2.91}                                  \\
                                 & GraphMask               & $92.54$\std{8.07}                                  & $62.23$\std{9.01}                                  & $73.10$\std{6.41}                                  & $72.06$\std{5.58}                                  & $73.06$\std{4.91}                                  & $66.68$\std{6.96}                                  \\
                                 & IB-Subgraph             & $86.06$\std{28.37}                                 & $91.04$\std{6.59}                                  & $51.20$\std{5.12}                                  & $57.29$\std{14.35}                                 & $62.89$\std{15.59}                                 & $47.29$\std{13.39}                                 \\
                                 & DIR                     & $82.78$\std{10.97}                                 & $64.44$\std{28.81}                                 & $32.35$\std{9.39}                                  & $78.15$\std{1.32}                                  & $77.68$\std{1.22}                                  & $49.08$\std{3.66}                                  \\
            \midrule\multirow{3}{*}{GIN}
                                 & GSAT                    & $98.85$\std{0.47}                                  & $99.35$\std{0.95}                                  & $80.47$\std{1.86}                                  & $74.49$\std{4.46}                                  & $72.95$\std{6.40}                                  & $65.25$\std{4.42}                                  \\
                                 & \gmtl                   & $98.36$\std{0.56}                                  & $99.86$\std{0.09}                                  & $82.98$\std{1.49}                                  & $76.06$\std{6.39}                                  & $76.50$\std{5.63}                                  & \cellcolor{lightskyblue}$\mathbf{80.57}$\std{2.59} \\
                                 & \gmts                   & \cellcolor{lightskyblue}$\mathbf{99.62}$\std{0.11} & \cellcolor{lightskyblue}$\mathbf{99.87}$\std{0.11} & \cellcolor{lightskyblue}$\mathbf{86.50}$\std{1.80} & \cellcolor{lightskyblue}$\mathbf{85.50}$\std{2.40} & \cellcolor{lightskyblue}$\mathbf{84.67}$\std{2.38} & $73.49$\std{5.33}                                  \\
            \midrule\multirow{3}{*}{PNA}
                                 & GSAT                    & $89.35$\std{5.41}                                  & $99.00$\std{0.37}                                  & $85.72$\std{1.10}                                  & $79.84$\std{3.21}                                  & $79.76$\std{3.66}                                  & $80.70$\std{5.45}                                  \\
                                 & \gmtl                   & $95.79$\std{7.30}                                  & $99.58$\std{0.17}                                  & $85.02$\std{1.03}                                  & $80.19$\std{2.22}                                  & $84.74$\std{1.82}                                  & $85.08$\std{3.85}                                  \\
                                 & \gmts                   & \cellcolor{lightskyblue}$\mathbf{99.60}$\std{0.48} & \cellcolor{lightskyblue}$\mathbf{99.89}$\std{0.05} & $\mathbf{87.34}$\std{1.79}                         & \cellcolor{lightskyblue}$\mathbf{88.27}$\std{1.71} & \cellcolor{lightskyblue}$\mathbf{86.58}$\std{1.89} & \cellcolor{lightskyblue}$\mathbf{85.26}$\std{1.92} \\
            \bottomrule
            \label{table:Interpretation}
        \end{tabular}}
\end{table*}

\begin{table*}[t]
    \caption{Prediction Performance (Acc.) of \gmt on regular graphs.}
    \small\sc\centering
    \resizebox{\textwidth}{!}{
        \begin{tabular}{llcccccc}
            \toprule
            \multirow{2}{*}{GNN} & \multirow{2}{*}{Method} & \multirow{2}{*}{MolHiv (AUC)}                      & \multirow{2}{*}{Graph-SST2}                        & \multirow{2}{*}{MNIST-75sp}                        & \multicolumn{3}{c}{Spurious-motif}                                                                                                                           \\
                                 &                         &                                                    &                                                    &                                                    & $b=0.5$                                            & $b=0.7$                                            & $b=0.9$                                            \\
            \midrule
            \multirow{3}{*}{GIN}
                                 & GIN                     & $76.69$\std{1.25}                                  & $82.73$\std{0.77}                                  & $95.74$\std{0.36}                                  & $39.87$\std{1.30}                                  & $39.04$\std{1.62}                                  & $38.57$\std{2.31}                                  \\
                                 & IB-subgraph             & $76.43$\std{2.65}                                  & $82.99$\std{0.67}                                  & $93.10$\std{1.32}                                  & $54.36$\std{7.09}                                  & $48.51$\std{5.76}                                  & $46.19$\std{5.63}                                  \\
                                 & DIR                     & $76.34$\std{1.01}                                  & $82.32$\std{0.85}                                  & $88.51$\std{2.57}                                  & $45.49$\std{3.81}                                  & $41.13$\std{2.62}                                  & $37.61$\std{2.02}                                  \\
            \midrule\multirow{3}{*}{GIN}
                                 & GSAT                    & $76.12$\std{0.91}                                  & $83.14$\std{0.96}                                  & $96.20$\std{1.48}                                  & $47.45$\std{5.87}                                  & $43.57$\std{2.43}                                  & $45.39$\std{5.02}                                  \\
                                 & \gmtl                   & $76.87$\std{1.12}                                  & $83.19$\std{1.28}                                  & $96.01$\std{0.25}                                  & $47.69$\std{4.93}                                  & $53.11$\std{4.12}                                  & $46.22$\std{4.18}                                  \\
                                 & \gmts                   & \cellcolor{lightskyblue}$\mathbf{77.22}$\std{0.93} & $\mathbf{83.62}$\std{0.50}                         & \cellcolor{lightskyblue}$\mathbf{96.50}$\std{0.19} & \cellcolor{lightskyblue}$\mathbf{60.09}$\std{2.40} & \cellcolor{lightskyblue}$\mathbf{54.34}$\std{4.04} & \cellcolor{lightskyblue}$\mathbf{55.83}$\std{5.68} \\
            \midrule\multirow{4}{*}{PNA}
                                 & PNA                     & $78.91$\std{1.04}                                  & $79.87$\std{1.02}                                  & $87.20$\std{5.61}                                  & $68.15$\std{2.39}                                  & $66.35$\std{3.34}                                  & $61.40$\std{3.56}                                  \\
                                 & GSAT                    & $79.82$\std{0.67}                                  & $80.90$\std{0.37}                                  & $93.69$\std{0.73}                                  & $68.41$\std{1.76}                                  & $67.78$\std{3.22}                                  & $51.51$\std{2.98}                                  \\
                                 & \gmtl                   & $80.05$\std{0.71}                                  & $81.18$\std{0.47}                                  & $94.44$\std{0.49}                                  & $69.33$\std{1.42}                                  & $64.49$\std{3.51}                                  & $58.30$\std{6.61}                                  \\
                                 & \gmts                   & $\mathbf{80.58}$\std{0.83}                         & \cellcolor{lightskyblue}$\mathbf{82.36}$\std{0.96} & \cellcolor{lightskyblue}$\mathbf{95.75}$\std{0.42} & \cellcolor{lightskyblue}$\mathbf{71.98}$\std{3.44} & $\mathbf{69.68}$\std{3.99}                         & \cellcolor{lightskyblue}$\mathbf{67.90}$\std{3.60} \\
            \bottomrule
            \label{table:Generalization}
        \end{tabular}}
\end{table*}

\section{Building Reliable \xgnns}
\label{CH:GMT:sec:gmt_sol}
The aforementioned gap motivates us to propose a new \xgnn architecture, called \gmtfull (\gmt), to provide both faithful interpretability and reliable (OOD) generalizability.
\gmt have two variants, i.e., \gmtl and \gmts, motivated by resolving the failures in Sec.~\ref{CH:GMT:sec:expressivity_issue}.

\subsection{Linearized \gmtt}
\label{CH:GMT:sec:gmt_lin}
Recall that the main reason for the failure of Eq.~\ref{CH:GMT:eq:exp_issue} is because of the non-linearity of the expectation to the $k$ weighted message passing with $k>1$. If $k$ can be reduced to $1$, then the linearity can be preserved to ensure a better approximation of \smt, which naturally motivates the following variant:
\begin{equation}\label{CH:GMT:eq:gmt_linear}
  (\text{\gmtl})\qquad\qquad\qquad \qquad %
  f^l(\widehat{G}_c)=\rho(\widehat{A}\odot A^{k-1}X \mW),
  \qquad \qquad \qquad \qquad \qquad
\end{equation}
Compared to the previous weighted message passing scheme with linearized GNN (Eq.~\ref{CH:GMT:eq:linear_gnn}), \gmtl improves the linearity by reducing the number of weighted message passing rounds to $1$.
If $\exists \mT\in\R^{|\gY|\times |\gY|}$ such that $\mT\cdot f_c(G_c)=P(Y|G_c)$ ($f_c$ is linear), then,we can incorporate \gmtl into Eq.~\ref{CH:GMT:eq:exp_issue} and have
\[f^l(\widehat{G}_c)=\mT\cdot f(\widehat{G}_c)=\E [f_c(G_c)],\]
due to the linearity of $f^l(G_c)$ with respect to $G_c$.
During training, $\mT$ can be further absorbed into $\mW$, which implies that \gmtl is able to fit to \smt.
Empirically, we find that the simple strategy of \gmtl already yields better interpretability than the state-of-the-art methods even with non-linear GNNs in experiments.

\subsection{\gmtt with random subgraph sampling}
\label{CH:GMT:sec:gmt_sam}

To generalize \gmt to more general cases, inspired by the \smt formulation, we propose a random subgraph sampling approach, that performs Markov Chain Monte Carlo (MCMC) sampling to approximate \smt. More concretely, given the attention matrix $\widehat{A}$, we perform $t$ rounds of random subgraph sampling from the subgraph distribution elicited by $\widehat{A}$ (or equivalently $\widehat{G}_c=(X,\widehat{A})$ as in \smt(Def.~\ref{def:sub_mt})),
and obtain $t$ i.i.d. random subgraph samples $\{G_c^i\}_{i=1}^t$ for estimating \smt as the following:
\begin{equation}\label{CH:GMT:eq:gmt_sampling}
  (\text{\gmts})\qquad\qquad\qquad \qquad %
  f_c^s(\widehat{G}_c)=\frac{1}{t}\sum_{i=1}^tf_c(Y|G_c^i),
  \qquad \qquad \qquad \qquad \qquad
\end{equation}
where $f_c$ is the classifier taking discrete subgraphs as inputs.

\begin{table*}[t]
    \caption[Interpretation performance of \gmt on geometric graphs.]{Interpretation performance on geometric graphs.
        }
    \small
    \centering\sc
    \label{CH:GMT:tab:lri_inter}
    \resizebox{\textwidth}{!}{%
        \begin{tabular}{lcccccccc}
            \toprule
            \multicolumn{1}{c}{\multirow{2}{*}{}}
                                 & \multicolumn{2}{c}{\textsc{\acts}}                 & \multicolumn{2}{c}{\textsc{\taumu}}                & \multicolumn{2}{c}{\textsc{\synbind}}              & \multicolumn{2}{c}{\textsc{\pdbbind}}                        \\
            \cmidrule(l{5pt}r{5pt}){2-3} \cmidrule(l{5pt}r{5pt}){4-5} \cmidrule(l{5pt}r{5pt}){6-7} \cmidrule(l{5pt}r{5pt}){8-9}
            \multicolumn{1}{c}{} & ROC AUC                                            & Prec@12                                            & ROC AUC
                                 & Prec@12                                            & ROC AUC                                            & Prec@12                                            & ROC AUC                                            & Prec@12 \\
            \midrule
            Random               & 50                                                 & 21                                                 & 50                                                 & 35
                                 & 50                                                 & 31                                                 & 50                                                 & 45                                                           \\
            \gradpos             & $69.31 $\std{0.89}                                 & $33.54$\std{1.23}                                  & $78.04$\std{0.57}                                  & $64.18$\std{1.25 }
                                 & $76.38$\std{4.96}                                  & $64.72$\std{3.75 }                                 & $58.11$\std{2.91 }                                 & $64.78$\std{4.73 }
            \\
            \gexp                & $54.23 $\std{4.31}                                 & $20.46$\std{5.46}                                  & $71.58$\std{0.69}                                  & $60.51$\std{0.76 }
                                 & $76.38$\std{4.96}                                  & $64.72$\std{3.75 }                                 & $52.23$\std{4.45 }                                 & $41.50$\std{9.77 }
            \\
            \pge                 & $22.87 $\std{3.33}                                 & $11.29 $\std{5.46}                                 & $70.72$\std{5.10 }                                 & $55.50$\std{6.26}
                                 & $87.06$\std{7.12 }                                 & $77.11$\std{7.58 }                                 & $51.98 $\std{4.66}                                 & $59.20 $\std{5.48}
            \\
            \pointmask           & $49.20$\std{1.51}                                  & $20.54 $\std{1.71}                                 & $55.93$\std{4.85 }                                 & $39.65$\std{7.14 }
                                 & $66.46$\std{6.86 }                                 & $53.93$\std{1.94 }                                 & $50.00 $\std{0.00}                                 & $45.10 $\std{0.00}
            \\
            \gradcam             & $75.19$\std{1.91}                                  & $75.94$\std{2.16}                                  & $76.18$\std{2.62 }                                 & $62.05$\std{2.16 }
                                 & $60.31$\std{4.95 }                                 & $52.35$\std{11.02 }                                & $48.61$\std{2.34}                                  & $55.10$\std{10.57 }
            \\
            \midrule
            LRI-Bernoulli        & $74.38$\std{4.33}                                  & $81.42$\std{1.52}                                  & $78.23$\std{1.11}                                  & $65.64$\std{2.44}
                                 & $89.22$\std{3.58}                                  & $68.76$\std{7.35}                                  & $54.87$\std{1.89}                                  & $72.12$\std{2.60}
            \\
            \gmtl                & \cellcolor{lightskyblue}$\mathbf{77.45}$\std{1.69} & $\mathbf{81.81}$\std{1.57}                         & \cellcolor{lightskyblue}$\mathbf{79.17}$\std{0.82} & \cellcolor{lightskyblue}$\mathbf{68.94}$\std{1.08}
                                 & \cellcolor{lightskyblue}$\mathbf{96.17}$\std{1.44} & \cellcolor{lightskyblue}$\mathbf{86.33}$\std{6.16} & $59.70$\std{1.10}                                  & $70.62$\std{3.59}
            \\
            \gmts                & $75.61$\std{1.86}                                  & $81.96$\std{1.35}                                  & $78.28$\std{1.34}                                  & $65.69$\std{2.61}
                                 & $93.93$\std{3.59}                                  & $83.20$\std{4.74}                                  & \cellcolor{lightskyblue}$\mathbf{60.03}$\std{1.02} & $\mathbf{72.56}$\std{2.27}
            \\
            \bottomrule
        \end{tabular}%
    }
\end{table*}

\begin{table}[t]
    \caption{Prediction performance (AUC) of \gmt on geometric graphs.}
    \centering
    \label{CH:GMT:tab:lri_gen}
    \small\sc
    \begin{tabular}{lcccc}
        \toprule
        \multicolumn{1}{c}{\multirow{1}{*}{}} & \multicolumn{1}{c}{\text{\acts}}                   & \multicolumn{1}{c}{\text{\taumu}}                  & \multicolumn{1}{c}{\text{\synbind}}                 & \multicolumn{1}{c}{\text{\pdbbind}}
        \\
        \midrule
        ERM                                   & $97.40$\std{0.32}                                  & $82.75$\std{0.16}                                  & $99.30$\std{0.20}                                   & $85.31$\std{2.21 }                  \\
        \gmtl                                 & $93.92 $\std{0.98}                                 & $82.60 $\std{0.17}                                 & $99.26 $\std{0.27}                                  & $86.29 $\std{0.80}
        \\
        LRI-Bernoulli                         & $94.00 $\std{0.78}                                 & $86.36$\std{0.06}                                  & $99.30 $\std{0.15}                                  & $85.80 $\std{0.70}                  \\
        \gmts                                 & \cellcolor{lightskyblue}$\mathbf{98.55}$\std{0.11} & \cellcolor{lightskyblue}$\mathbf{86.42}$\std{0.08} & \cellcolor{lightskyblue}$\mathbf{99.89} $\std{0.03} & $\mathbf{87.19} $\std{1.86}
        \\
        \bottomrule
    \end{tabular}%
\end{table}

\begin{theorem}\label{CH:GMT:thm:gmt_success}
  Given the attention matrix $\widehat{A}$,
  and the distribution distance metric $d$ as the total variation distance,
  let $C=|\gY|$,
  for a \gmts with $t$ i.i.d. samples of
  $G_c^i\sim P(G_c|G)$, then,
  there exists $\epsilon\in\R_+$ such that,
  with a probability at least $1-e^{-t\epsilon^2/4}$, \gmts $\frac{\epsilon C}{2}$-approximates \smt and satisfies $(\delta,1-\frac{\epsilon C}{\delta})$ counterfactual fidelity.
\end{theorem}
The proof for Theorem~\ref{CH:GMT:thm:gmt_success} is given in Appendix~\ref{proof:gmt_success}.
Intuitively, with more random subgraph samples drawn from $P(G_c|G)$, \gmts obtains a more accurate estimation of \smt.
However, it will incur more practical challenges such as the a) gradient of discrete sampling and b) computational overhead. To overcome the challenges a) and b), we incorporate the following two techniques.

\textbf{Backpropagation of discrete sampling.}
To enable gradient backpropagation with the sampled subgraphs, we also incorporate gradient estimation techniques such as Gumbel softmax and straight-through estimator~\citep{gumbel,gumbel2}.
Compared to the state-of-the-art \xgnn \gsat~\citep{gsat},
this scheme brings two additional benefits: (i) reduces the gradient biases in discrete sampling with Gumbel softmax; (ii) avoids weighted message passing and alleviates the input distribution gap to the graph encoder when shared in both $f_c$ and $g$ as in \gsat.

\begin{figure*}[t]
  \centering
  \subfigure[Counterfactual fidelity.]{
    \includegraphics[width=0.27\textwidth]{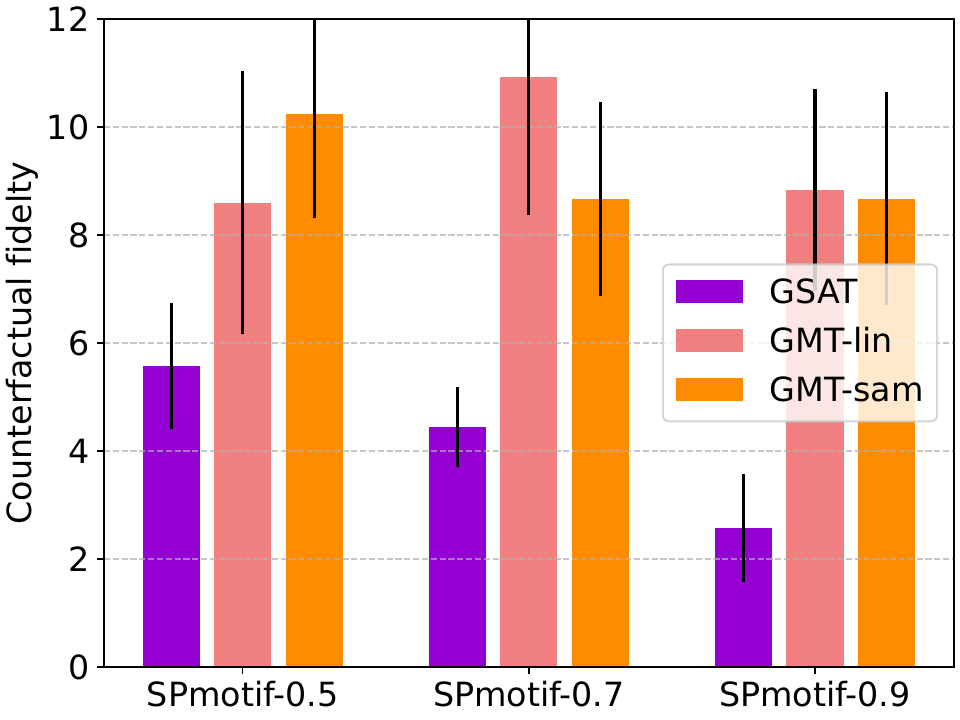}
    \label{CH:GMT:fig:cf_results}
  }
  \subfigure[Interpretation sensitivity.]{
    \includegraphics[width=0.3\textwidth]{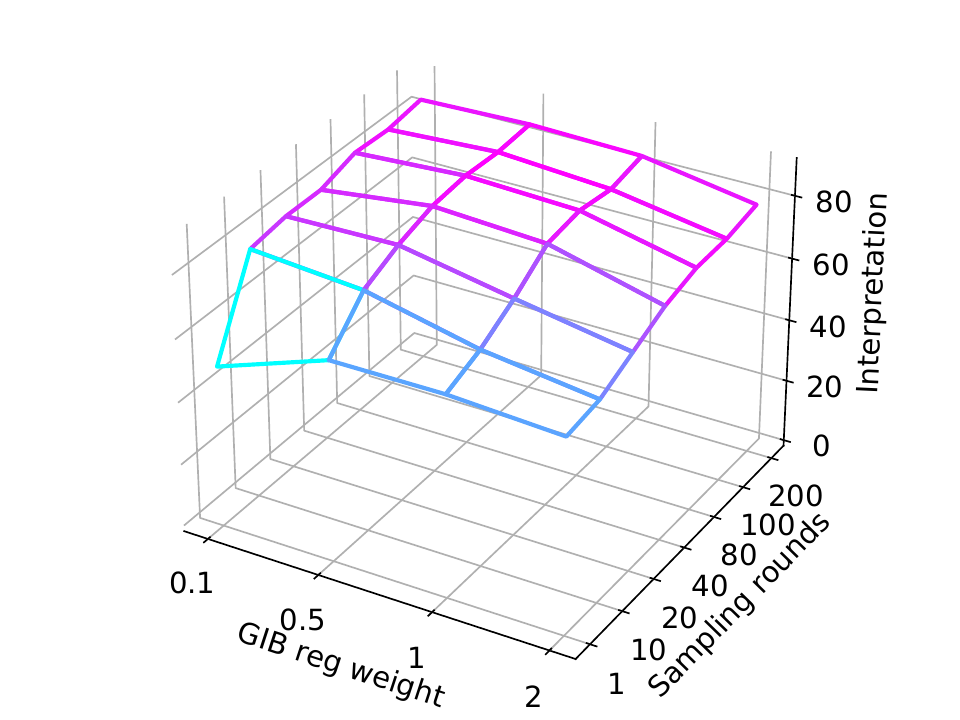}
    \label{CH:GMT:fig:inter_sens}
  }
  \subfigure[Generalization sensitivity.]{
    \includegraphics[width=0.3\textwidth]{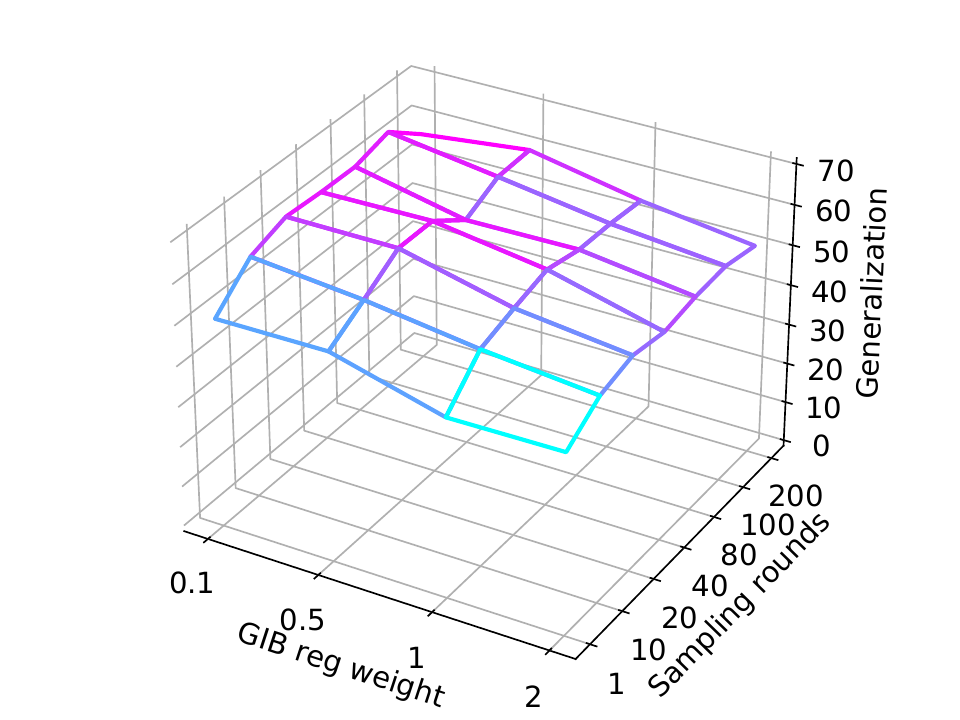}
    \label{CH:GMT:fig:gen_sens}
  }
  \caption[Ablation studies of \gmt.]{Ablation studies.}
  \label{CH:GMT:fig:ablation}
\end{figure*}

\textbf{The number of sampling rounds.} Although the estimation of \smt will be more accurate with the increased sampling rounds, it unnecessarily brings improvements. First, as shown in Fig.~\ref{CH:GMT:fig:ablation}, the performance may be saturated with moderately sufficient samplings. Besides, the performance may degenerate as more sampling rounds can affect the optimization, as discussed in Appendix~\ref{CH:GMT:sec:gmt_impl_appdx}.

\subsection{Learning neural subgraph multilinear extension}
Although \gmt trained with \gmts improve interpretability, \gmts still requires multiple random subgraph sampling to approximate \smt and costs much additional overhead.
To this end, we propose to learn a neural \smt that only requires single sampling, based on the trained subgraph extractor $g$ with \gmts.

Learning the neural \smt is essentially to approximate the MCMC with a neural network, though it is inherently challenging to approximate MCMC~\citep{no_free_lunch_MCMC,papamarkou2022a}.
Nevertheless, the feasibility of neural \smt learning is backed by the inherent causal subgraph assumption of~\citep{ciga}, once the causal subgraph is correctly identified, simply learning the statistical correlation between the subgraph and the label is sufficient to recover the causal relation.

Therefore, we propose to simply re-train a new classifier GNN with the frozen subgraph extractor, to distill the knowledge contained in $\widehat{G}_c$ about $Y$.
This scheme also brings additional benefits over the originally trained classifier, which avoid to learn all the available statistical correlations between $G_c$ and $Y$ that can be spurious.
More details and discussions on the implementations are given in Appendix~\ref{CH:GMT:sec:gmt_impl_dis_appdx}.

\section{Experimental Evaluations}\label{CH:GMT:sec:exp}
We conduct extensive experiments to evaluate \gmt with different backbones and on multiple  benchmarks, and compare both the interpretability and (OOD) generalizability against the baselines.
We will briefly introduce the datasets, baselines, and setups, and leave more details in Appendix~\ref{CH:GMT:sec:exp_appdx}.

\subsection{Experimental settings}
\label{CH:GMT:sec:exp_setting}

\textbf{Datasets.}
We consider both the regular and geometric graph classification benchmarks following the \xgnn  literature~\citep{gsat,lri}.
For regular graphs, we include \textsc{BA-2Motifs}~\citep{pge}, \textsc{Mutag}~\citep{mutag}, \textsc{MNIST-75sp}~\citep{understand_att}, which are widely evaluated by post-hoc explanation approaches~\citep{xgnn_tax}, as well as \textsc{Spurious-Motif}~\citep{dir}, \textsc{Graph-SST2}~\citep{sst25,xgnn_tax} and \textsc{OGBG-Molhiv}~\citep{ogb} where there exist various graph distribution shifts.
For geometric graphs, we consider \textsc{ActsTrack}, \textsc{Tau3Mu}, \textsc{SynMol} and \textsc{PLBind} curated by~\citet{lri}.

\textbf{Baselines.}
For post-hoc methods, we mainly adopt the results from the previous works~\citep{gsat,lri}, including \text{ GNNExplainer}~\citep{gnn_explainer}, \text{PGExplainer}~\citep{pge}, \text{GraphMask}~\citep{graphmask} for regular graph benchmarks, and \text{BernMask}, \text{BernMask-P}, that are modified from \text{GNNExplainer} and \text{PGExplainer}, GradGeo~\citep{gradgeo}, and GradCam~\citep{gradcam} that are extended for geometric data, as well as PointMask~\citep{pointmask} developed specifically for geometric data.
For \xgnns, since we focus on the interpretation performance, we mainly compared with \xgnns that have the state-of-the-art interpretation abilities, i.e., \gsat~\citep{gsat} and \lri~\citep{lri}, which also have excellent OOD generalizability than other \xgnns~\citep{good_bench}.
We also include two representative \xgnns baselines, \text{DIR}~\citep{dir} and \text{IB-subgraph}~\citep{gib} for regular graphs.

\textbf{Training and evaluation.}
We consider three backbones GIN~\citep{gin} and PNA~\citep{pna} for regular graph data, EGNN~\citep{egnn} for geometric data.
All methods adopted the identical graph encoder, and optimization protocol for fair comparisons.
We tune the hyperparameters as recommended by previous works.
More details are given in Appendix~\ref{CH:GMT:sec:eval_appdx}.

\subsection{Experimental results and analysis}
\label{CH:GMT:sec:results}

\textbf{Interpretation performance.} As shown in Table.~\ref{table:Interpretation}, compared to post-hoc methods (in the first row) and \gsat, both \gmtl and \gmts lead to non-trivial improvements for interpretation performance.
Especially, in challenging Spurious-Motif datasets with distribution shifts, \gmts brings improvements than \gsat up to $15\%$ with GIN, and up to $8\%$ with PNA. In challenging realistic dataset MNIST-75sp, \gmts also improves \gsat up to $6\%$.

\textbf{Generalization performance.} Table \ref{table:Generalization} illustrates the prediction accuracy on regular graph datasets.  We again observe consistent improvements by \gmt spanning from molecule graphs to image-converted datasets. Despite distribution shifts, \gmts still brings improvements up to $13\%$ with GIN, and up to $16\%$ against \gsat in Spurious-Motif.

\textbf{Results on geometric graphs.}
Tables \ref{CH:GMT:tab:lri_inter} and \ref{CH:GMT:tab:lri_gen} show the interpretation and generalization performances of various methods. Again, we observe consistent non-trivial improvements of \gmtl and \gmts in most cases than \gsat and post-hoc methods.
Interestingly, \gmtl brings more improvements than \gmts in terms of interpretation performance despite its simplicity. In terms of generalization performance, \gmts remains the best method.
The results on geometric datasets further demonstrate the strong generality of \gmt across different tasks and backbones.

\textbf{Ablation studies.}
In complementary to the interpretability and generalizability study, we conduct further ablation studies to better understand the results. Fig.~\ref{CH:GMT:fig:cf_results} shows the counterfactual fidelity of \gsat, \gmtl and \gmts in Spurious-Motif (SPmotif) test sets.
As shown in Fig.~\ref{CH:GMT:fig:cf_results} that \gsat achieves a lower counterfactual fidelity.
In contrast, \gmtl and \gmts improve a higher counterfactual fidelity, which explains the reason for the improved interpretability of \gmt.
We also examine the hyperparameter sensitivity of \gmts in SPMotif-0.5 dataset.
As shown in Fig.~\ref{CH:GMT:fig:inter_sens},~\ref{CH:GMT:fig:gen_sens}, \gmts maintains strong robustness against the hyperparameter choices.
The interpretation performance gets improved along with the sampling rounds,
while a too larger GIB information regularizer weights will affect the optimization of \gmt and the generalizability.

\textbf{More baseline results in PNA backbones} are given in Appendix~\ref{CH:GMT:sec:more_x_appdx}, including two representative post-hoc methods GNNExplainer and PGExplainer, and one representative XGNN baseline DIR.
The results show that most of the baselines still significantly underperform \gsat and \gmt.

\textbf{Computational analysis} is given in Appendix~\ref{CH:GMT:sec:comp_appdx}. Although \gmts takes a longer time for training, but the absolute values are not high even for the largest dataset MNIST-75sp. When compared to other intrinsic interpretable methods,  \gmts consumes a similar training time around 6 hours on MNIST-75sp as DIR.
As for inference,  \gmts enjoys a similar latency as others.

\chapter{Causality in Adversarial Robustness} \label{CH:HAO}

\section{Motivations}
Graph Neural Networks (GNNs), as a generalization of deep learning models for graph-structured data, have gained great success in tasks involving relational information~\citep{jure_grlsurvey, peter_survey, zhou_survey, wu_survey,gcn,sage,gat,jknet,gin}.
Nevertheless, GNNs are shown to be inherently vulnerable to adversarial attacks~\citep{sun_gadv_survey,deeprobust},
or small intentional perturbations on the input~\citep{intriguing}.
Previous studies show that moderate changes to the existing topology or node features of the input graph, i.e., Graph Modification Attacks (GMA),
can dramatically degenerate the performance of GNNs~\citep{rls2v,nettack,metattack,topo_atk,restrict_blk_adv}.
Since in many real-world scenarios, it is prohibitively expensive to modify the original graph,
recently there has been an increasing \mbox{attention} paid to Graph Injection Attack (GIA),
where the adversary can merely inject few \mbox{malicious} nodes to perform the attack~\citep{fakenodes,nipa,afgsm,tdgia}.

\begin{figure}[t]
	\topspace
	\centering
	\subfigure[Attack without defense]{
		\includegraphics[width=0.31\textwidth]{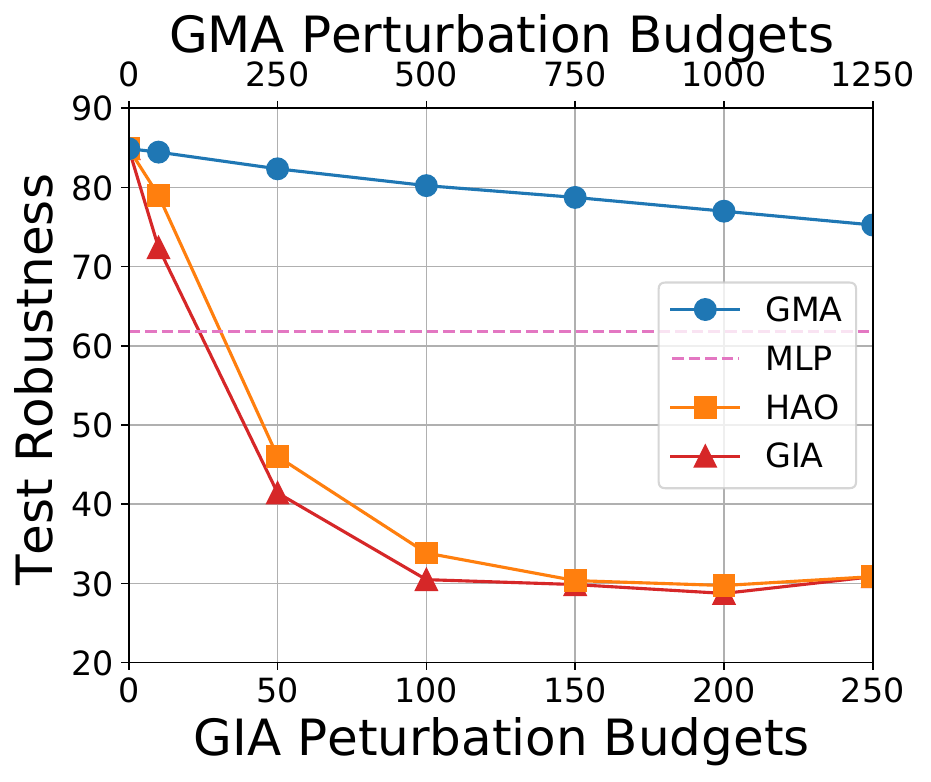}
		\label{CH:HAO:fig:motivate_wo_defense}
	}
	\subfigure[Attack with defense]{
		\includegraphics[width=0.31\textwidth]{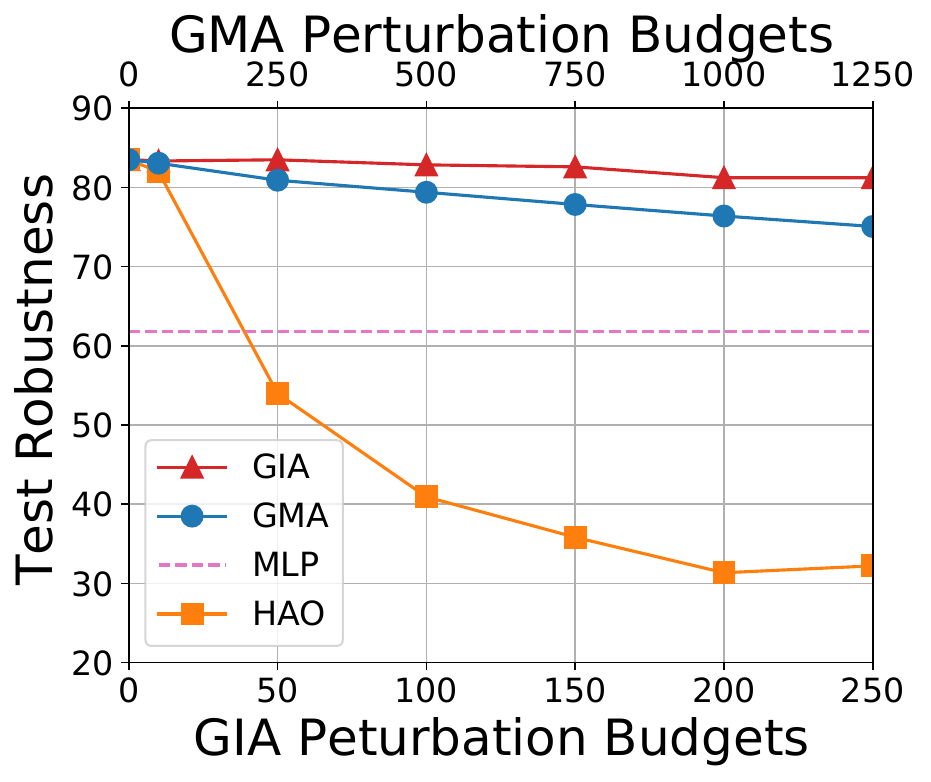}
		\label{CH:HAO:fig:motivate_w_defense}
	}
	\subfigure[Illustration of GIA at node $u$]{
		\includegraphics[width=0.30\textwidth]{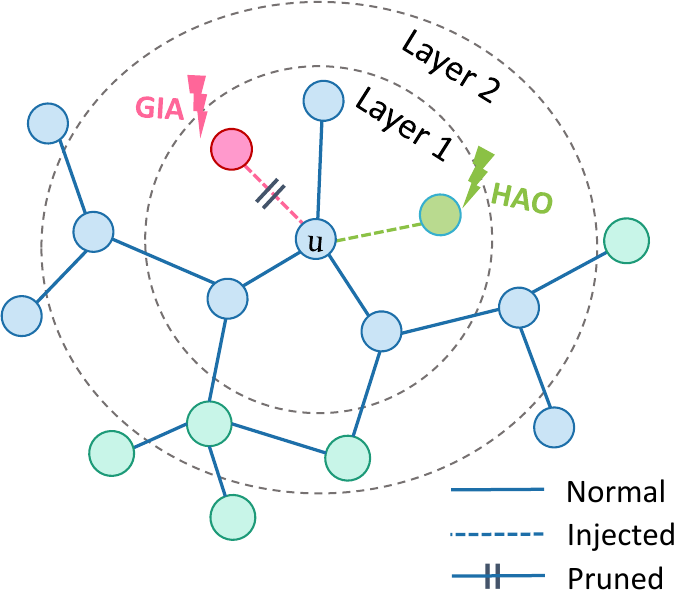}
		\label{CH:HAO:fig:hao_illustration}
	}
	\caption[Illustration of \hao.]{
		The lower test robustness indicates better attack performance. (a) Without defenses: GIA  performs consistently better than GMA; (b) With defenses: GIA without HAO performs consistently worse than GMA, while GIA with HAO performs the best; (c) Homophily indicates the tendency of similar nodes connecting with each other (blue \& green nodes).
		The malicious (red) nodes and edges injected by GIA without HAO will greatly break the homophily and hence can be easily identified and pruned by homophily defenders.
		GIA with HAO is aware of preserving homophily that attacks the targets by injecting unnoticeable (more similar) but still adversarial (dark green) nodes and edges, which will not be easily pruned hence effectively causing the damage.
	}
	\label{CH:HAO:fig:motivation_fig}
\end{figure}

Despite the promising empirical results, why GIA is booming and whether there is any pitfall behind the success remain elusive.
To bridge this gap, we investigate both the advantages and limitations of GIA by comparing it with GMA in a unified setting (Sec.~\ref{CH:HAO:sec:adv_setting}).
Our theoretical results show that, in this setting when there is no defense, GIA can be provably more harmful than GMA due to its relatively high flexibility.
Such flexibility enables GIA to map GMA perturbations into specific GIA perturbations and to further optimize the mapped perturbations to amplify the damage (Fig.~\ref{CH:HAO:fig:motivate_wo_defense}).
However, according to the principle of no free lunch, we further find that the power of GIA is built upon the severe damage to the homophily of the original graph.
Homophily indicates the tendency of nodes to connect to others with similar features or labels,
which is important for the success of most existing GNNs~\citep{homophily1_birds,homophily2_collective, klicpera2018combining, peter_survey, Hou2020Measuring, beyond_homophily, self-enhance}.
The severe damage to homophily will disable the effectiveness of GIA in evaluating robustness
because non-robust models can easily mitigate or even prevent GIA merely by exploiting the property of homophily damage.

Specifically, having observed the destruction of homophily, it is straightforward to devise a defense mechanism aiming to recover the homophily,
which we term \emph{homophily defenders}.
Homophily defenders are shown to have strong robustness against GIA attacks.
Theoretically, they can effectively reduce the harm caused by GIA to be lower than GMA.
Empirically, simple implementations of homophily defenders with edge pruning~\citep{gnnguard} can deteriorate even the state-of-the-art GIA attacks~\citep{tdgia} (Fig.~\ref{CH:HAO:fig:motivate_w_defense}).
Therefore, overlooking the damage to homophily will make GIA powerless and further limit its applications for evaluating the robustness of GNNs.

To enable the effectiveness of GIA in evaluating various robust GNNs, it is necessary to be aware of preserving the homophily when developing GIA.
To this end, we introduce a novel constraint -- \mbox{\emph{homophily unnoticeability}} that enforces GIA to retain the homophily of the original graph,
which can serve as a supplementary for the unnoticeability constraints in graph adversarial learning.
To instantiate the homophily unnoticeability, we propose the Harmonious Adversarial Objective (HAO) for GIA (Fig.~\ref{CH:HAO:fig:hao_illustration}).
Specifically, HAO introduces a novel differentiable realization of homophily constraint by regularizing the homophily distribution shift during the attack.
In this way, adversaries will not be easily identified by homophily defenders while still performing effective attacks (Fig.~\ref{CH:HAO:fig:motivate_w_defense}).
Extensive experiments with $38$ defense models on $6$ benchmarks demonstrate that GIA with HAO
can break homophily defenders and significantly outperform all previous works across all settings,
including both non-target attack and targeted attack\footnote{Code is available in \url{https://github.com/LFhase/GIA-HAO}.}.
Our contributions are summarized as follows:
\begin{itemize}
	\item We provide a formal comparison between GIA and GMA in a unified setting and find that GIA can be provably more harmful than GMA due to its high flexibility (Theorem~\ref{CH:HAO:thm:gia_threat}).

	\item However, the flexibility of GIA will also cause severe damage to the homophily distribution which makes GIA easily defendable by homophily defenders (Theorem~\ref{CH:HAO:thm:gia_easy_defend}).
	\item To mitigate the issue, we introduce the concept of homophily unnoticeability and a novel objective HAO to conduct homophily unnoticeable attacks (Theorem~\ref{CH:HAO:thm:hao_break_limit}).
\end{itemize}

\section{Preliminaries}
\subsection{Graph Neural Networks}
Consider a graph $\gG=(A,X)$  with node set $V=\{v_1,v_2,...,v_n\}$ and edge set $E=\{e_1,e_2,...,e_m\}$,
where  $A \in \{0,1\}^{n\times n}$  is the adjacency matrix and $X\in \R^{n \times d}$ is the node feature matrix.
We are interested in the semi-supervised node classification task~\citep{deeprobust}. That is, given the set of labels $Y\in \{0,1,..,C-1\}^n$ from $C$ classes,
we can train a graph neural network $f_\theta$ parameterized by $\theta$ on the training (sub)graph $\gG_{\text{train}}$
by minimizing a classification loss $\mathcal{L}_{\text{train}}$ (e.g., cross-entropy).
Then the trained $f_\theta$ can predict the labels of nodes in test graph $\gG_{\text{test}}$.
A GNN typically follows a neighbor aggregation scheme to recursively update the node representations as:
\begin{equation}
	\label{CH:HAO:eq:gnn}
	H^{(k)}_u = \sigma(W_k\cdot\rho(\{H^{(k-1)}_v\}| v\in\mathcal{N}(u)\cup\{u\})),
\end{equation}

where $\mathcal{N}(u)$ is the set of neighbors of node $u$, $H^{(0)}_u=X_u, \forall u \in V$, $H^{(k)}_u$ is the hidden representation of node $u$ after the $k$-th aggregation,
$\sigma(\cdot)$ is an activation function, e.g., $\text{ReLU}$, and $\rho(\cdot)$ is an aggregation function over neighbors, e.g., $\text{MEAN}$ or $\text{SUM}$.%

\subsection{Graph Adversarial Attack}
\label{CH:HAO:sec:adv_setting}
The goal of a graph adversarial attack is to fool a GNN model, $f_{\theta^*}$, trained on a graph $\gG=(A,X)$ by constructing a graph $\gG'=(A',X')$ with limited budgets $\lVert \gG'-\gG\rVert \leq \triangle$.\footnote{
	We leave more details and reasons about the setting used in this work in Appendix~\ref{CH:HAO:sec:adv_setting_appdx}.}
Given a set of victim nodes $V_{\text{c}}\subseteq V$, the graph adversarial attack can be generically formulated as:
\begin{equation}
	\label{CH:HAO:eq:gia_obj}
	\min \ \mathcal{L}_{\text{atk}}(f_{\theta^*}(\gG')),
	\ \text{s.t.}\ {\lVert \gG'-\gG\rVert \leq \triangle},
\end{equation}
where $\theta^* =\argmin_{\theta}\mathcal{L}_{\text{train}}(f_{\theta}(\gG_{\text{train}}))$ and $\mathcal{L}_{\text{atk}}$ is usually taken as $-\mathcal{L}_{\text{train}}$.
Following previous works~\citep{nettack,tdgia}, Graph adversarial attacks can be characterized into graph modification attacks and graph injection attacks by their perturbation constraints.

\textbf{Graph Modification Attack (GMA).} GMA generates $\gG'$ by modifying the graph structure $A$ and the node features $X$ of the original graph $\gG$.
Typically the constraints in GMA are to limit the number of perturbations on $A$ and $X$, denoted by $\triangle_A$ and $\triangle_X$, respectively, as:
\begin{equation}
	\label{CH:HAO:eq:gma_cons}
	\triangle_A+\triangle_X\leq \triangle\in\sZ, \norm{A'-A}_0 \leq \triangle_A\in \sZ, \norm{X'-X}_\infty \leq \epsilon \in \R,
\end{equation}
where the perturbation on $X$ is bounded by $\epsilon$ via L-p norm, since we are using continuous features.

\textbf{Graph Injection Attack (GIA).} Differently, GIA generates $\gG'$ by injecting a set of malicious nodes $V_{\text{atk}}$ as
$X'=\begin{bmatrix}
		\ X\hfill        \\
		\ X_{\text{atk}} \\
	\end{bmatrix},
	A'=\begin{bmatrix}
		\ A\hfill          & A_{\text{atk}} \\
		\ A_{\text{atk}}^T & O_{\text{atk}} \\
	\end{bmatrix},
$
where $X_{\text{atk}}$ is the features of the injected nodes, $O_{\text{atk}}$ is the adjacency matrix among injected nodes, and $A_{\text{atk}}$ is the adjacency matrix between the injected nodes and the original nodes. Let $d_u$ denote the degree of node $u$, the constraints in GIA are:
\begin{equation}
	\label{CH:HAO:eq:gia_cons}
	|V_{\text{atk}}| \leq \triangle\in\sZ, \ 1\leq d_u\leq b\in\sZ,
	X_{u} \in \mathcal{D}_X \subseteq \R^d,
	\forall u\in V_{\text{atk}},
\end{equation}
where the number and degrees of the injected nodes are limited, $\mathcal{D}_X=\{C\in\R^d,\min(X)\cdot\mathbf{1}\leq C\leq \max(X)\cdot\mathbf{1} \}$ where $\min(X)$ and $\max(X)$ are the minimum and maximum entries in $X$ respectively.

\textbf{Threat Model.}
We adopt a unified setting, i.e., evasion, inductive and black-box, which is also used by Graph Robustness Benchmark~\citep{grb}.
Evasion: The attack only happens at test time, i.e., $\gG_{\text{test}}$, rather than attacking $\gG_{\text{train}}$.
Inductive:  Test nodes are invisible during training.
Black-box: The adversary can not access the architecture or the parameters of the target model.

\section{Power and Pitfalls of Graph Injection Attack}
\label{CH:HAO:sec:gia_comparison}
Based on the setting above, we investigate both the advantages and limitations of GIA by comparing it with GMA.
While we find GIA is more harmful than GMA when there is no defense (Theorem~\ref{CH:HAO:thm:gia_threat}),
we also find pitfalls in GIA that can make it easily defendable (Theorem~\ref{CH:HAO:thm:gia_easy_defend}).

\subsection{Power of Graph Injection Attack}
Following previous works~\citep{nettack}, we use a linearized GNN, i.e., $H^{(k)}=\hat{A}^kX\Theta$, to track the changes brought by attacks.
Firstly we will elaborate the threats of an adversary as follows.%

\begin{definition}[Threats]
	\label{CH:HAO:def:threats}
	Consider an adversary $\mathcal{A}$, given a perturbation budget $\triangle$,
	the threat of $\mathcal{A}$ to a GNN $f_\theta$ is defined as $\min_{\norm{\gG'-\gG}\leq \triangle}\mathcal{L}_\atk(f_\theta(\gG'))$, i.e., the optimal objective value of Eq.~\ref{CH:HAO:eq:gia_obj}.
\end{definition}

With Definition~\ref{CH:HAO:def:threats}, we can quantitatively compare the threats of different adversaries.
\begin{theorem}
	\label{CH:HAO:thm:gia_threat}
	Given moderate perturbation budgets $\triangle_\gia$ for GIA and $\triangle_\gma$ for GMA, that is,
	let $\triangle_\gia \leq \triangle_\gma \ll |V|\leq |E|$,
	for a fixed linearized GNN $f_\theta$ trained on $\gG$,
	assume that $\gG$ has no isolated nodes,
	and both GIA and GMA follow the optimal strategy,
	then, \mbox{$\forall \triangle_\gma\geq0, \exists \triangle_\gia\leq \triangle_\gma$}, %
	\[
		\mathcal{L}_{\atk}(f_\theta(\gG'_\gia))-\mathcal{L}_{\atk}(f_\theta(\gG'_\gma))\leq 0,
	\]
	where $\gG'_\gia$ and $\gG'_\gma$ are the perturbed graphs generated by GIA and GMA, respectively.
\end{theorem}
We prove Theorem~\ref{CH:HAO:thm:gia_threat} in Appendix~\ref{CH:HAO:proof:gia_threat}.
Theorem~\ref{CH:HAO:thm:gia_threat} implies that GIA can cause more damage than GMA with equal or fewer budgets, which is also verified empirically as shown in Fig.~\ref{CH:HAO:fig:motivate_wo_defense}.

Intuitively, the power of GIA mainly comes from its relatively high flexibility in perturbation generation.
Such flexibility enables us to find a mapping that can map any GMA perturbations to GIA perturbations, leading the same influences to the predictions of $f_\theta$.
We will give an example below.
\begin{figure}[t]
	\topspace
	\subfigure[Illustration of $\mathcal{M}_2$ mapping]{
		\includegraphics[width=0.30\textwidth]{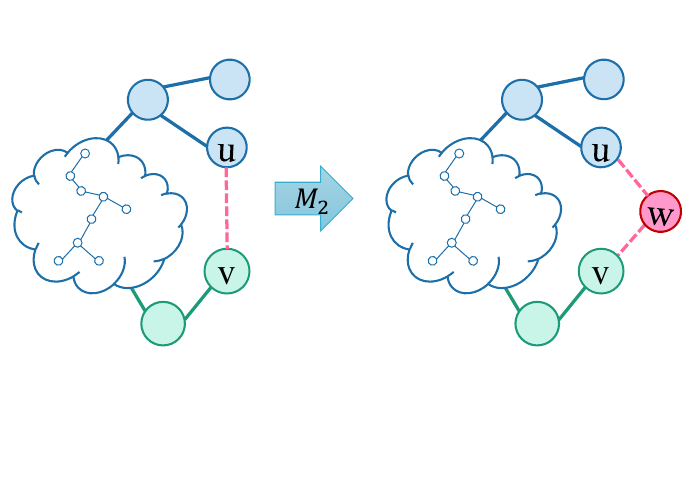}
		\label{CH:HAO:fig:m2_illustration}
	}
	\subfigure[GMA v.s. GIA with $\mathcal{M}_2$]{
		\includegraphics[width=0.31\textwidth]{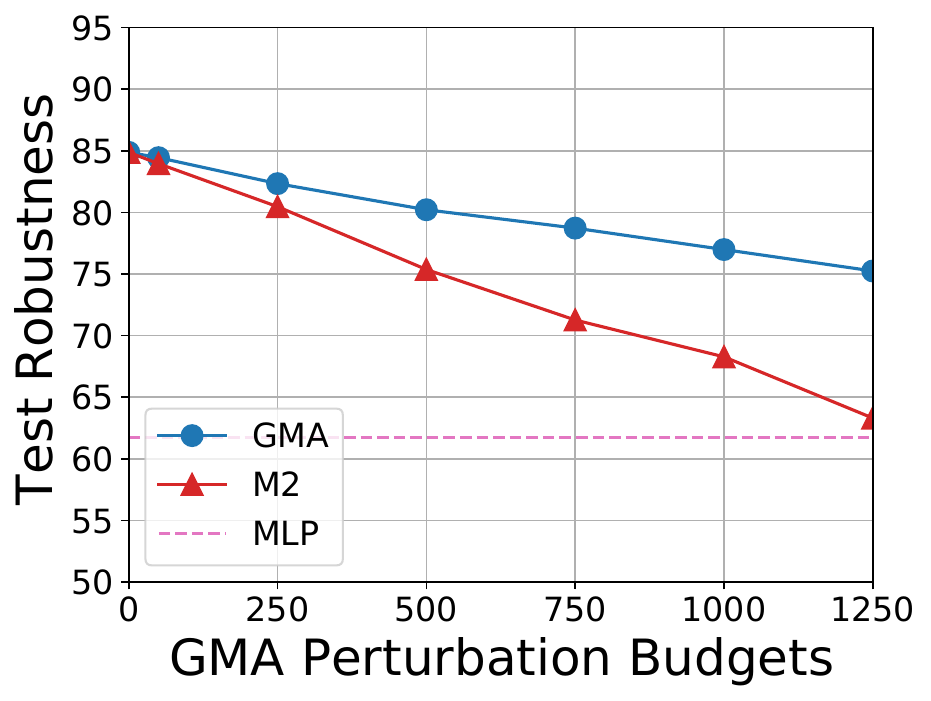}
		\label{CH:HAO:fig:gia_vs_gma_mapping}
	}
	\subfigure[Homophily changes]{
		\includegraphics[width=0.31\textwidth]{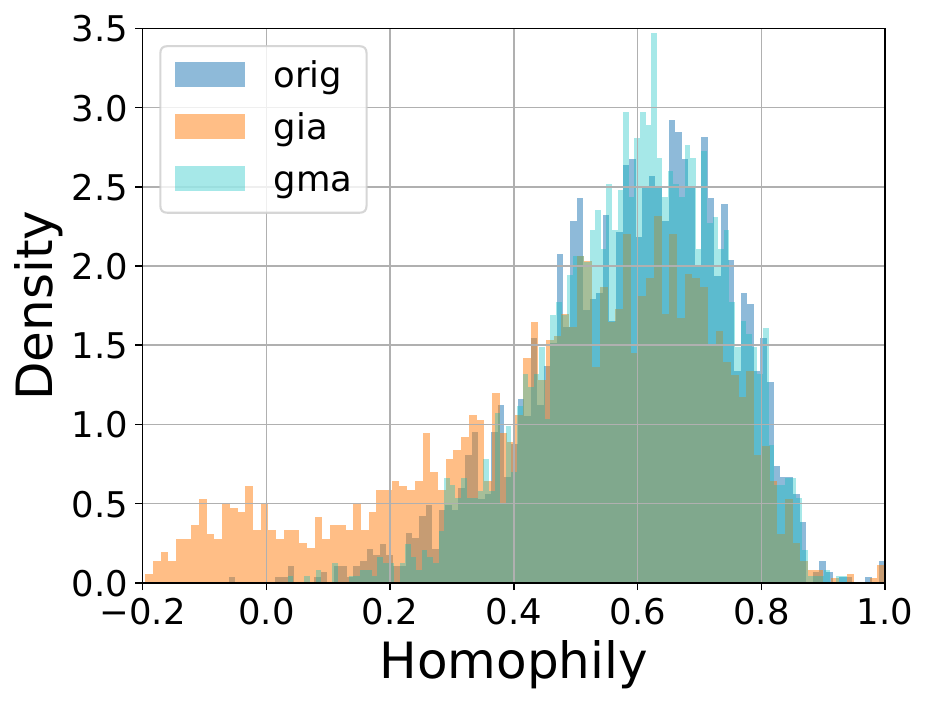}
		\label{CH:HAO:fig:gia_homophily_dist_node_atk}
	}
	\caption{Power and pitfalls of Graph Injection Attack}
	\label{CH:HAO:fig:gia_vs_gma}
\end{figure}

\begin{definition}[Plural Mapping $\mathcal{M}_2$]
	\label{CH:HAO:def:m2_mapping}
	$\mathcal{M}_2$ maps a perturbed graph $\gG'_\gma$ generated by GMA  with only edge addition perturbations,\footnote{We focus on edge addition in later discussions since \cite{advsample_deepinsights} observed that it produces the most harm in GMA. Discussions about the other GMA operations can be found in Appendix~\ref{CH:HAO:sec:m2_gma_appdx}.}
	to a GIA perturbed graph $\gG'_\gia=\mathcal{M}_2(\gG'_\gma)$,
	such that:
	\[
		f_\theta(\gG'_\gia)_u=f_\theta(\gG'_\gma)_u,\forall u \in V.
	\]
\end{definition}
As illustrated in Fig.~\ref{CH:HAO:fig:m2_illustration}, the procedure of $\mathcal{M}_2$ is,
for each edge $(u,v)$ added by GMA to attack node $u$,
$\mathcal{M}_2$ can inject a new node $w$ to connect $u$ and $v$, and change $X_w$ to make the same effects to the prediction on $u$.
Then GIA can be further optimized to bring more damage to node $u$.
We also empirically verify the above procedure in Fig.~\ref{CH:HAO:fig:gia_vs_gma_mapping}. Details about the comparison are in Appendix~\ref{CH:HAO:sec:gma_gia_comparison_appdx}.

\subsection{Pitfalls in Graph Injection Attack}
Through $\mathcal{M}_2$, we show that the flexibility in GIA can make it more harmful than GMA when there is no defense,
however, we also find a side-effect raised in the optimization trajectory of $X_w$ from the above example.
Assume GIA uses PGD~\citep{pgd} to optimize $X_w$ iteratively, we find:
\begin{equation}
	\label{CH:HAO:eq:gia_lower_sim}
	\text{sim}(X_u,X_w)^{(t+1)} \leq \text{sim}(X_u,X_w)^{(t)},
\end{equation}
where $t$ is the number of optimization steps and $\text{sim}(X_u,X_v)=\frac{X_u\cdot X_v}{\norm{X_u}_2\norm{X_v}_2}$.
We prove the statement in Appendix~\ref{CH:HAO:proof:gia_lower_sim}.
It implies that, under the mapping, $\mathcal{M}_2$, the similarity between injected nodes and targets continues to decrease as the optimization processes,
and finally becomes lower than that in GMA. We find this is closely related to the loss of \emph{homophily} of the target nodes.

Before that, we will elaborate on the definition of homophily in graph adversarial setting.
Different from typical definitions that rely on the label information~\citep{homophily1_birds,homophily2_collective,geom_gcn,beyond_homophily},
as the adversary does not have the access to all labels,
we provide another instantiation of homophily based on node feature similarity as follows:
\begin{definition}[Node-Centric Homophily]
	\label{CH:HAO:def:homophily_node}
	The homophily of a node $u$ can be defined with the \mbox{similarity} between the features of node $u$ and the aggregated features of its neighbors:
	\begin{equation}
		\label{CH:HAO:eq:homophily_node}
		h_u=\text{sim}(r_u,X_u),\ r_u=\sum_{j \in \mathcal{N}(u)}\frac{1}{\sqrt{d_j}\sqrt{d_u}}X_j,
	\end{equation}
	where $d_u$ is the degree of node $u$ and $\text{sim}(\cdot)$ is a similarity metric, e.g., cosine similarity.
\end{definition}
We also define edge-centric homophily while we will focus primarily on node-centric homophily. Details and reasons are in Appendix~\ref{CH:HAO:sec:homophily_edge_appdx}.
With Definition~\ref{CH:HAO:def:homophily_node}, combining Eq.~\ref{CH:HAO:eq:gia_lower_sim}, we have:
\[
	h_u^\gia \leq h_u^\gma,
\]
where $h_u^\gia$ and $h_u^\gma$ denote the homophily of node $u$ after GIA and GMA attack, respectively.
It implies that GIA will cause more damage to the homophily of the original graph than GMA.
To verify the discovery for more complex cases,
we plot the homophily distributions in Fig.~\ref{CH:HAO:fig:gia_homophily_dist_node_atk}.
The blue part denotes the original homophily distribution. Notably, there is an outstanding out-of-distribution (orange) part caused by GIA, compared to the relatively minor (canny) changes caused by GMA.
The same phenomenon also appears in other datasets that can be found in Appendix~\ref{CH:HAO:sec:homophily_appdx}.

Having observed the huge homophily damage led by GIA, it is straightforward to devise a defense mechanism aiming to recover the original homophily, which we call \textit{homophily defenders}.
We theoretically elaborate such defenses in the form of edge pruning\footnote{Actually, homophily defenders can have many implementations other than pruning edges as given in Appendix~\ref{CH:HAO:sec:more_homo_defender_appdx}, while we will focus on the design above in our discussion.}, adapted from Eq.~\ref{CH:HAO:eq:gnn}:
\begin{equation}
	\label{CH:HAO:eq:homo_defender}
	H^{(k)}_u = \sigma(W_k\cdot\rho(\{\mathbb{1}_{\text{con}}(u,v)\cdot H^{(k-1)}_v\}| \ v \in\mathcal{N}(u)\cup\{u\}).
\end{equation}
We find that simply pruning the malicious edges identified by a proper condition can empower homophily defenders with strong theoretical robustness against GIA attacks.
\begin{theorem}
	\label{CH:HAO:thm:gia_easy_defend}
	Given conditions in Theorem~\ref{CH:HAO:thm:gia_threat},
	consider a GIA attack, which \textup{(i)} is mapped by $\mathcal{M}_2$ (Def.~\ref{CH:HAO:def:m2_mapping}) from a GMA attack that only performs edge addition perturbations,
	and \textup{(ii)} uses a linearized GNN trained with at least one node from each class in $\gG$ as the surrogate model, and \textup{(iii)} optimizes the malicious node features with PGD.
	Assume that $\gG$ has no isolated node, and has node features as $X_u=\frac{C}{C-1}e_{Y_u}-\frac{1}{C-1}\mathbf{1}\in \R^d$,
	where $Y_u$ is the label of node $u$ and $e_{Y_u}\in\R^d$ is a one-hot vector with the $Y_u$-th entry being $1$ and others being $0$.
	Let the minimum similarity for any pair of nodes connected in $\gG$ be $s_{\gG}=\min_{(u,v)\in E}\text{sim}(X_u,X_v)$ with $\text{sim}(X_u,X_v)=\frac{X_u\cdot X_v}{\norm{X_u}_2\norm{X_v}_2}$.
	For a homophily defender $g_\theta$ that prunes edges $(u,v)$ if $\text{sim}(X_u,X_v)\leq s_{\gG}$,
	we have:
	\[
		\mathcal{L}_{\text{atk}}(g_\theta(\mathcal{M}_2(\gG'_{\text{GMA}})))-\mathcal{L}_{\text{atk}}(g_\theta(\gG'_{\text{GMA}}))\geq 0.
	\]
\end{theorem}
We prove Theorem~\ref{CH:HAO:thm:gia_easy_defend} in Appendix~\ref{CH:HAO:proof:gia_easy_defend}.
It implies that, by specifying a mild pruning condition, the homophily defender can effectively reduce the harm caused by GIA to be lower than that of GMA.

Considering a more concrete example with $\mathcal{M}_2$,
$X_w$ is generated to make $\mathcal{L}_{\text{atk}}(f_\theta(\mathcal{M}_2(\gG'_{\text{GMA}})))=\mathcal{L}_{\text{atk}}(f_\theta(\gG'_{\text{GMA}}))$ on node $u$ at first.
Then, due to the flexibility in GIA, $X_w$ can be optimized to some $X'_w$ that greatly destroys the homophily of node $u$, i.e., having a negative cosine similarity score with $u$.
Thus, for a graph with relatively high homophily, i.e., $s_{\gG}\geq 0$,
a mild pruning condition such as $\mathbb{1}_{\text{sim}(u,v)\leq 0}(u,v)=0$ could prune all the malicious edges generated by GIA while possibly keeping some of those generated by GMA,
which makes GIA less threatful than GMA.

In the literature, we find that GNNGuard~\citep{gnnguard} serves well for an implementation of homophily defenders as Eq.~\ref{CH:HAO:eq:homo_defender}.
With GNNGuard, we verify the strong empirical robustness of homophily defenders against GIA.
As Fig.~\ref{CH:HAO:fig:motivate_w_defense} depicts,
when with homophily defenders, GIA can only cause little-to-no damage, while GMA can still effectively perturb the predictions of the target model on some nodes.
To fully demonstrate the power of homophily defenders, we also prove its certified robustness for a concrete GIA case in Appendix~\ref{CH:HAO:sec:gia_cer_robo}.

\section{Homophily Unnoticeable Graph Injection Attack}
\label{CH:HAO:sec:gia_hao}

\subsection{Harmonious Adversarial Objective}
\label{CH:HAO:sec:gia_unnoticeability}
As shown in Sec.~\ref{CH:HAO:sec:gia_comparison}, the flexibility of GIA makes it powerful
while dramatically hinders its performance when combating against homophily defenders,
because of the great damage to the homophily distribution brought by GIA.
This observation motivates us to introduce the concept of \emph{homophily unnoticeability} that enforces GIA to preserve the original homophily distribution during the attack.
\begin{definition}[Homophily Unnoticeability]
	\label{CH:HAO:def:homo_unnoticeable}
	Let the node-centric homophily distribution for a graph $\gG$ be $\mathcal{H}_\gG$.
	Given the upper bound for the allowed homophily distribution shift $\triangle_{\mathcal{H}}\geq 0$,
	an attack $\mathcal{A}$ is homophily unnoticeable if:
	\[
		m(\mathcal{H}_\gG,\mathcal{H}_{\gG'})\leq \triangle_{\mathcal{H}},
	\]
	where $\gG'$ is the perturbed graph generated by $\mathcal{A}$,
	and $m(\cdot)$ is a distribution distance measure.
\end{definition}
Intuitively, homophily unnoticeability can be a supplementary for the unnoticeability in graph adversarial attack that requires a GIA adversary to consider
how likely the new connections between the malicious nodes and target nodes will appear \textit{naturally}.
Otherwise, i.e., unnoticeability is broken, the malicious nodes and edges can be easily detected and removed by database administrators or homophily defenders.
However, homophily unnoticeability can not be trivially implemented as a rigid constraint and be inspected incrementally like that for degree distribution~\citep{nettack}.
For example, a trivial implementation such as clipping all connections that do not satisfy the constraint (Def.~\ref{CH:HAO:def:homo_unnoticeable}) will trivially clip all the injected edges due to the unconstrained optimization in GIA.

Considering the strong robustness of homophily defenders, we argue that they can directly serve as external examiners for homophily unnoticeability check.
Satisfying the homophily constraint can be approximately seen as bypassing the homophily defenders.
Obviously, GIA with constraints as Eq.~\ref{CH:HAO:eq:gia_cons} can not guarantee homophily unnoticeability,
since it will only optimize towards maximizing the damage by minimizing the homophily of the target nodes.
Hence, we propose a novel realization of the homophily constraint for GIA that enforces it to meet the homophily unnoticeability \textit{softly}.

\begin{definition}[Harmonious Adversarial Objective (HAO)]
	Observing the homophily definition in Eq.~\ref{CH:HAO:eq:homophily_node} is differentiable with respect to $X$, we can integrate it into the objective of Eq.~\ref{CH:HAO:eq:gia_obj} as:\footnote{Note that we only use HAO to solve for $\gG'$ while still using the original objective to evaluate the threats.}
	\begin{equation}
		\label{CH:HAO:eq:gia_obj_harmo}
		\begin{aligned}
			\min _{\lVert \gG'-\gG\rVert \leq \triangle} \ \mathcal{L}^{\text{h}}_{\text{atk}}(f_{\theta^*}(\gG')) & =
			\mathcal{L}_{\text{atk}}(f_{\theta^*}(\gG'))
			-\lambda C(\gG,\gG'),                                                                                      \\
		\end{aligned}
	\end{equation}
	where $C(\gG,\gG')$ is a regularization term based on homophily and $\lambda\geq 0$ is the corresponding weight.
\end{definition}
One possible implementation is to maximize the homophily for each injected node as:
\begin{equation}
	\label{CH:HAO:eq:harmo_reg_1}
	C(\gG,\gG') = \frac{1}{|V_{\text{atk}}|}\sum_{u\in V_{\text{atk}}}{h_u}.
\end{equation}

HAO seizes the possibility of
retaining homophily unnoticeability, while still performing effective attacks.
Hence, given the homophily distribution distance measure $m(\cdot)$ in Def.~\ref{CH:HAO:def:homo_unnoticeable}, we can infer:
\begin{theorem}
	\label{CH:HAO:thm:hao_break_limit}
	Given conditions in Theorem~\ref{CH:HAO:thm:gia_easy_defend},
	we have
	$m(\mathcal{H}_\gG,\mathcal{H}_{\gG'_{\text{HAO}}})\!\leq\! m(\mathcal{H}_\gG,\mathcal{H}_{\gG'_{\text{GIA}}})$, hence:
	\[
		\mathcal{L}_{\text{atk}}(g_\theta(\gG'_{\text{HAO}}))-\mathcal{L}_{\text{atk}}(g_\theta(\gG'_{\text{GIA}}))\leq 0,
	\]
	where $\gG'_{\text{HAO}}$ is generated by GIA with HAO, and $\gG'_{\text{GIA}}$ is generated by GIA without HAO.
\end{theorem}
We prove Theorem~\ref{CH:HAO:thm:hao_break_limit} in Appendix~\ref{CH:HAO:proof:hao_break_limit}.
Intuitively, since GIA with HAO can reduce the damage to homophily,
it is more likely to bypass the homophily defenders, thus being more threatful than GIA without HAO.
We also empirically verify Theorem~\ref{CH:HAO:thm:hao_break_limit} for more complex cases in the experiments.

\subsection{Adaptive Injection Strategies}
\label{CH:HAO:sec:gia_algorithm}
GIA is generically composed of two procedures, i.e., node injection and feature update, to solve for $\gG'=(A',X')$,
where node injection leverages either the gradient information or heuristics to solve for $A'$, and feature update usually uses PGD~\citep{pgd} to solve for $X'$.
Most previous works separately optimize $A'$ and $X'$ in a greedy manner,
which implicitly assumes that the other will be optimized to maximize the harm.
However, HAO does not follow the assumption but stops the optimization when the homophily is overly broken.
Thus, a more suitable injection strategy for HAO shall \textit{be aware of} retaining the original homophily.
To this end, we propose to optimize $A'$ and $X'$ alternatively and introduce three adaptive injection strategies to coordinate with HAO.

\textbf{Gradient-Driven Injection.}
We propose a novel bi-level formulation of HAO to perform the alternative optimization using gradients,
where we separate the optimization of $\gG'=\!(A',X')$ as:
\begin{equation}
	\label{CH:HAO:eq:gia_bilevel_obj}
	\begin{aligned}
		X'^*       & =\argmin_{X'\in \Phi(X')}\mathcal{L}_{\text{atk}}(f_{\theta^*}(A'^*,X'))-\lambda_A C(\gG',\gG), \\
		s.t.\ A'^* & = \argmin_{A'\in \Phi(A')}\mathcal{L}_{\text{atk}}(f_{\theta^*}(A',X'))-\lambda_X C(\gG',\gG) , \\
	\end{aligned}
\end{equation}
where $\Phi(A')$ and $\Phi(X')$ are the corresponding feasible regions for $A'$ and $X'$ induced by the original constraints.
Here we use different homophily constraint weights $\lambda_A$ and $\lambda_X$ for the optimizations of $A'$ and $X'$, since $A'$ is discrete while $X'$ is continuous.
We can either adopt Meta-gradients like Metattack~\citep{metattack} (\textbf{MetaGIA}) or directly optimize edge weights to solve for $A'$ (\textbf{AGIA}).
The detailed induction of meta-gradients and algorithms are given in Appendix~\ref{CH:HAO:sec:gia_gradient_appdx}.

\textbf{Heuristic-Driven Injection.}
As the state-of-the-art GIA methods are leveraging heuristics to find $A'$, based on TDGIA~\citep{tdgia}, we also propose a variant (\textbf{ATDGIA}) using heuristics as:
\begin{equation}
	\label{CH:HAO:eq:tdgia+_score}
	s_u = ((1-p_u)\mathbb{1}(\argmax{(p)}=y'_u))(\frac{0.9}{\sqrt{bd_u}}+\frac{0.1}{d_u}),
\end{equation}
where $s_u$ indicates the vulnerability of node $u$ and $\mathbb{1}(\cdot)$ is to early stop destroying homophily.

\textbf{Sequential Injection} for large graphs. Since gradient methods require huge computation overhead,
we propose a novel divide-and-conquer strategy (\textbf{SeqGIA}) to iteratively select some of the most vulnerable targets with Eq.~\ref{CH:HAO:eq:tdgia+_score} to attack.
Detailed algorithm is given in Appendix~\ref{CH:HAO:sec:gia_apgd_seq_appdx}.
\section{Experiments}
\label{CH:HAO:sec:experiments}

\subsection{Setup \& Baselines}
\label{CH:HAO:sec:exp_setups}
\textbf{Datasets.}
We comprehensively evaluate our methods with $38$ defense models on $6$ datasets.
We select two classic citation networks Cora and Citeseer~\citep{cora,citeseer} refined by GRB~\citep{grb}.
We also use Aminer and Reddit~\citep{aminer,sage,saint} from GRB, Arxiv from OGB~\citep{ogb}, and a co-purchasing network Computers~\citep{computers_photo} to cover more domains and scales.
Details are in Appendix~\ref{CH:HAO:sec:datasets_appdx}.

\textbf{Comparing with previous attack methods.}
We incorporate HAO into several existing GIA methods as well as our proposed injection strategies to verify its effectiveness and versatility.
First of all, we select \textbf{PGD}~\citep{pgd} as it is one of the most widely used adversarial attacks.
We also select \textbf{TDGIA}~\citep{tdgia} which is the state-of-the-art GIA method.
We adopt the implementations in GRB~\citep{grb} for the above two methods.
We exclude FGSM~\citep{fgsm} and AFGSM \citep{afgsm}, since PGD is better at dealing with non-linear models than FGSM~\citep{pgd}, and AFGSM performs comparably with FGSM but is worse than TDGIA as demonstrated by~\cite{tdgia}.
For GMA methods, we adopt \textbf{Metattack}~\citep{metattack} as one of the bi-level implementations. We exclude Nettack~\citep{nettack} as it is hard to perform incremental updates with GCN (the surrogate model used in our experiments) and leave reinforcement learning methods such as RL-S2V \citep{rls2v} and NIPA \citep{nipa} for future work. More details are given in Appendix~\ref{CH:HAO:sec:exp_rl_appdx}.

\textbf{Categories and complexity analysis of attack methods.}
We provide categories and complexity analysis of all attack methods used in our experiments in Table~\ref{CH:HAO:tab:complexity}, Appendix~\ref{CH:HAO:sec:gia_alg_complexity}.

\textbf{Competing with different defenses.} We select both popular GNNs and robust GNNs as the defense models.
For popular GNNs, we select the three most frequently used baselines, i.e., \textbf{GCN}~\citep{gcn}, \textbf{GraphSage}~\citep{sage}, and \textbf{GAT}~\citep{gat}.
For robust GNNs, we select \textbf{GCNGuard}~\citep{gnnguard} for graph purification approach, and \textbf{RobustGCN}~\citep{robustgcn} for stabilizing hidden representation approach,
as representative ones following the surveys~\citep{sun_gadv_survey,deeprobust}.
Notably, the author-released GCNGuard implementation requires $O(n^2)$ complexity, which is hard to scale up.
To make the comparison fair, following the principle of homophily defenders, we implement two efficient robust alternatives, i.e., Efficient GCNGuard (\textbf{EGuard}) and Robust Graph Attention Network (\textbf{RGAT}).
More details are given in Appendix~\ref{CH:HAO:sec:rgat_appdx}.
Besides, we exclude the robust GNNs learning in a transductive manner like ProGNN~\citep{prognn} that can not be adapted in our setting.

\textbf{Competing with the extremely robust defenses.} %
To make the evaluation for attacks more reliable,
we also adopt two widely used robust tricks Layer Normalization (\textbf{LN})~\citep{layer_norm} and an efficient adversarial training~\citep{fgsm,pgd} method \textbf{FLAG}~\citep{flag}.
Here, as FLAG can effectively enhance robustness, we exclude other adversarial training methods for efficiency consideration.
More details are given in Appendix~\ref{CH:HAO:sec:baseline_detail_appdx}.

\begin{table}[t]
	\caption{Performance of non-targeted attacks with \hao against different models.}
	\label{CH:HAO:tab:eval_non_targeted}
	\resizebox{\textwidth}{!}{
		\scriptsize
		\begin{tabular}{@{}{l}*{13}{c}@{}}
			\toprule
			                                         &              & \multicolumn{3}{c}{{\small Cora ($\downarrow$)}} & \multicolumn{3}{c}{{\small Citeseer($\downarrow$)}} & \multicolumn{3}{c}{{\small Computers($\downarrow$)}} & \multicolumn{3}{c}{{\small Arxiv($\downarrow$)}}                                                                                                                                                                                                                                                                                                          \\\cmidrule(lr){3-5}\cmidrule(lr){6-8}\cmidrule(lr){9-11}\cmidrule(lr){12-14}
			                                         & HAO          & \multicolumn{1}{c}{\textbf{Homo}}                & \multicolumn{1}{c}{\textbf{Robust}}                 & \multicolumn{1}{c}{\textbf{Combo}}                   & \multicolumn{1}{c}{\textbf{Homo}}                & \multicolumn{1}{c}{\textbf{Robust}} & \multicolumn{1}{c}{\textbf{Combo}} & \multicolumn{1}{c}{\textbf{Homo}} & \multicolumn{1}{c}{\textbf{Robust}} & \multicolumn{1}{c}{\textbf{Combo}} & \multicolumn{1}{c}{\textbf{Homo}} & \multicolumn{1}{c}{\textbf{Robust}} & \multicolumn{1}{c}{\textbf{Combo}} \\ \midrule
			Clean                                    &              & $85.74$                                          & $86.00$                                             & $87.29$                                              & $74.85$                                          & $75.46$                             & $75.87$                            & $93.17$                           & $93.17$                             & $93.32$                            & $70.77$                           & $71.27$                             & $71.40$                            \\  \hdashline[0.5pt/1pt]
			\rule{0pt}{10pt}PGD                      &              & $83.08$                                          & $83.08$                                             & $85.74$                                              & $74.70$                                          & $74.70$                             & $75.19$                            & $84.91$                           & $84.91$                             & $91.41$                            & $68.18$                           & $68.18$                             & $71.11$                            \\
			PGD                                      & $\checkmark$ & $52.60$                                          & $62.60$                                             & $77.99$                                              & $69.05$                                          & $69.05$                             & $73.04$                            & $79.33$                           & $79.33$                             & $87.83$                            & $55.38$                           & $\underline{62.89}$                 & $68.68$                            \\\hdashline[0.5pt/1pt]
			\rule{0pt}{10pt}MetaGIA$^\dagger$        &              & $83.61$                                          & $83.61$                                             & $85.86$                                              & $74.70$                                          & $74.70$                             & $75.15$                            & $84.91$                           & $84.91$                             & $91.41$                            & $68.47$                           & $68.47$                             & $71.09$                            \\
			MetaGIA$^\dagger$                        & $\checkmark$ & $49.25$                                          & $\underline{69.83}$                                 & $76.80$                                              & $68.04$                                          & $68.04$                             & $\underline{71.25}$                & $78.96$                           & $78.96$                             & $90.25$                            & $57.05$                           & $63.30$                             & $69.97$                            \\
			AGIA$^\dagger$                           &              & $83.44$                                          & $83.44$                                             & $85.78$                                              & $74.72$                                          & $74.72$                             & $75.29$                            & $85.21$                           & $85.21$                             & $91.40$                            & $68.07$                           & $68.07$                             & $71.01$                            \\
			AGIA$^\dagger$                           & $\checkmark$ & $\underline{47.24}$                              & $\mathbf{61.59}$                                    & $\mathbf{75.25}$                                     & $70.24$                                          & $70.24$                             & $71.80$                            & $\underline{75.14}$               & $\underline{75.14}$                 & $\underline{86.02}$                & $59.32$                           & $65.62$                             & $69.92$                            \\\hdashline[0.5pt/1pt]
			\rule{0pt}{10pt}TDGIA                    &              & $83.44$                                          & $83.44$                                             & $85.72$                                              & $74.76$                                          & $74.76$                             & $75.26$                            & $88.32$                           & $88.32$                             & $91.40$                            & $64.49$                           & $64.49$                             & $70.97$                            \\
			TDGIA                                    & $\checkmark$ & $56.95$                                          & $73.38$                                             & $79.45$                                              & $\mathbf{60.91}$                                 & $\mathbf{60.91}$                    & $72.51$                            & $\mathbf{74.77}$                  & $\mathbf{74.77}$                    & $90.42$                            & $\underline{49.36}$               & $\mathbf{60.72}$                    & $\mathbf{63.57}$                   \\
			ATDGIA                                   &              & $83.07$                                          & $83.07$                                             & $85.39$                                              & $74.72$                                          & $74.72$                             & $75.12$                            & $86.03$                           & $86.03$                             & $91.41$                            & $66.95$                           & $66.95$                             & $71.02$                            \\
			ATDGIA                                   & $\checkmark$ & $\mathbf{42.18}$                                 & $70.30$                                             & $\underline{76.87}$                                  & $\underline{61.08}$                              & $\underline{61.08}$                 & $\mathbf{71.22}$                   & $80.86$                           & $80.86$                             & $\mathbf{84.60}$                   & $\mathbf{45.59}$                  & $63.30$                             & $\underline{64.31}$                \\\hdashline[0.5pt/1pt]
			\hdashline[0.5pt/3pt]\rule{0pt}{10pt}MLP &              & \multicolumn{3}{c}{{$61.75$}}                    & \multicolumn{3}{c}{{$65.55$}}                       & \multicolumn{3}{c}{{$84.14$}}                        & \multicolumn{3}{c}{{$52.49$}}                                                                                                                                                                                                                                                                                                                             \\ \bottomrule
			\multicolumn{14}{l}{$^\downarrow$The lower number indicates better attack performance.  $^\dagger$Runs with SeqGIA framework on Computers and Arxiv.  }                                                                                                                                                                                                                                                                                                                                                                                                                             \\
		\end{tabular}}
\end{table}

\begin{table}[t]
	\caption{Performance of targeted attacks with \hao against different models.}
	\label{CH:HAO:tab:eval_targeted}
	\resizebox{\textwidth}{!}{
		\scriptsize
		\begin{tabular}{@{}{l}*{13}{c}@{}}
			\toprule
			                                         &              & \multicolumn{3}{c}{{\small Computers($\downarrow$)}} & \multicolumn{3}{c}{{\small Arxiv($\downarrow$)}} & \multicolumn{3}{c}{{\small Aminer($\downarrow$)}} & \multicolumn{3}{c}{{\small Reddit($\downarrow$)}}                                                                                                                                                                                                                                                                                                          \\\cmidrule(lr){3-5}\cmidrule(lr){6-8}\cmidrule(lr){9-11}\cmidrule(lr){12-14}
			                                         & HAO          & \multicolumn{1}{c}{\textbf{Homo}}                    & \multicolumn{1}{c}{\textbf{Robust}}              & \multicolumn{1}{c}{\textbf{Combo}}                & \multicolumn{1}{c}{\textbf{Homo}}                 & \multicolumn{1}{c}{\textbf{Robust}} & \multicolumn{1}{c}{\textbf{Combo}} & \multicolumn{1}{c}{\textbf{Homo}} & \multicolumn{1}{c}{\textbf{Robust}} & \multicolumn{1}{c}{\textbf{Combo}} & \multicolumn{1}{c}{\textbf{Homo}} & \multicolumn{1}{c}{\textbf{Robust}} & \multicolumn{1}{c}{\textbf{Combo}} \\ \midrule
			Clean                                    &              & $92.68$                                              & $92.68$                                          & $92.83$                                           & $69.41$                                           & $71.59$                             & $72.09$                            & $62.78$                           & $66.71$                             & $66.97$                            & $94.05$                           & $97.15$                             & $97.13$                            \\  \hdashline[0.5pt/1pt]
			\rule{0pt}{10pt}PGD                      &              & $88.13$                                              & $88.13$                                          & $91.56$                                           & $69.19$                                           & $69.19$                             & $71.31$                            & $53.16$                           & $53.16$                             & $56.31$                            & $92.44$                           & $92.44$                             & $93.03$                            \\
			PGD                                      & $\checkmark$ & $71.78$                                              & $71.78$                                          & $85.81$                                           & $\mathbf{36.06}$                                  & $\mathbf{37.22}$                    & $69.38$                            & $34.62$                           & $34.62$                             & $39.47$                            & $\underline{56.44}$               & $\underline{86.12}$                 & $\mathbf{84.94}$                   \\\hdashline[0.5pt/1pt]
			\rule{0pt}{10pt}MetaGIA$^\dagger$        &              & $87.67$                                              & $87.67$                                          & $91.56$                                           & $69.28$                                           & $69.28$                             & $71.22$                            & $48.97$                           & $48.97$                             & $52.35$                            & $92.40$                           & $92.40$                             & $93.97$                            \\
			MetaGIA$^\dagger$                        & $\checkmark$ & $\underline{70.21}$                                  & $\underline{71.61}$                              & $85.83$                                           & $38.44$                                           & $\underline{38.44}$                 & $48.06$                            & $41.12$                           & $41.12$                             & $45.16$                            & $\mathbf{46.75}$                  & $90.06$                             & $90.78$                            \\
			AGIA$^\dagger$                           &              & $87.57$                                              & $87.57$                                          & $91.58$                                           & $66.19$                                           & $66.19$                             & $70.06$                            & $50.50$                           & $50.50$                             & $53.69$                            & $91.62$                           & $91.62$                             & $93.66$                            \\
			AGIA$^\dagger$                           & $\checkmark$ & $\mathbf{69.96}$                                     & $\mathbf{71.58}$                                 & $85.72$                                           & $38.84$                                           & $38.84$                             & $68.97$                            & $35.94$                           & $35.94$                             & $42.66$                            & $80.69$                           & $88.84$                             & $90.44$                            \\\hdashline[0.5pt/1pt]
			\rule{0pt}{10pt}TDGIA                    &              & $87.21$                                              & $87.21$                                          & $91.56$                                           & $63.66$                                           & $63.66$                             & $71.06$                            & $51.34$                           & $51.34$                             & $54.82$                            & $92.19$                           & $92.19$                             & $93.62$                            \\
			TDGIA                                    & $\checkmark$ & $71.39$                                              & $71.62$                                          & $\mathbf{77.15}$                                  & $42.56$                                           & $42.56$                             & $\underline{42.53}$                & $\underline{25.78}$               & $\underline{25.78}$                 & $\underline{29.94}$                & $78.16$                           & $\mathbf{85.06}$                    & $\underline{88.66}$                \\
			ATDGIA                                   &              & $87.85$                                              & $87.85$                                          & $91.56$                                           & $66.12$                                           & $66.12$                             & $71.16$                            & $50.87$                           & $50.87$                             & $53.68$                            & $91.25$                           & $91.25$                             & $93.03$                            \\
			ATDGIA                                   & $\checkmark$ & $72.00$                                              & $72.53$                                          & $\underline{78.35}$                               & $\underline{38.28}$                               & $40.81$                             & $\mathbf{39.47}$                   & $\mathbf{22.50}$                  & $\mathbf{22.50}$                    & $\mathbf{28.91}$                   & $64.09$                           & $89.06$                             & $88.91$                            \\\hdashline[0.5pt/1pt]
			\hdashline[0.5pt/3pt]\rule{0pt}{10pt}MLP &              & \multicolumn{3}{c}{{$84.11$}}                        & \multicolumn{3}{c}{{$52.49$}}                    & \multicolumn{3}{c}{{$32.80$}}                     & \multicolumn{3}{c}{{$70.69$}}                                                                                                                                                                                                                                                                                                                              \\ \bottomrule
			\multicolumn{14}{l}{$^\downarrow$The lower number indicates better attack performance. $^\dagger$Runs with SeqGIA framework.  }                                                                                                                                                                                                                                                                                                                                                                                                                                                    \\
		\end{tabular}}
\end{table}
\textbf{Evaluation protocol.} %
We use a $3$-layer GCN as the surrogate model to generate perturbed graphs with various GIA attacks, and report the mean accuracy of defenses from multiple runs.
Details are in Appendix~\ref{CH:HAO:sec:model_setting_appdx}.
For in-detail analysis of attack performance, we categorize all defenses into three folds by their robustness: Vanilla, \textbf{Robust}, and Extreme Robust (\textbf{Combo}) (Table~\ref{CH:HAO:tab:defense_category}).
To examine how much an attack satisfies the homophily unnoticeability and its upper limits, we report \textit{maximum test accuracy} of both homophily defenders (\textbf{Homo}) and defenses from the last two categories.

\begin{table}[t]
\caption{Averaged attack performance of various attacks with or without HAO against both homophily defenders and other defense models. }
	\label{CH:HAO:tab:averaged_performance}
	\centering
	\centering\resizebox{0.8\textwidth}{!}{
		\begin{tabular}{lcccccc}
			\toprule
			\textbf{Model} & \textbf{Cora}$^\dagger$ & \textbf{Computers}$^\dagger$ & \textbf{Arxiv}$^\dagger$ & \textbf{Computers}$^\ddagger$ & \textbf{Aminer}$^\ddagger$ & \textbf{Reddit}$^\ddagger$ \\\midrule
			Clean          & $84.74$                 & $92.25$                      & $70.44$                  & $91.68$                       & $62.39$                    & $95.51$                    \\
			PGD            & $61.09$                 & $61.75$                      & $54.23$                  & $62.41$                       & $26.13$                    & $62.72$                    \\
			\hfill +HAO    & $\underline{56.63}$     & $\underline{59.16}$          & $\underline{45.05}$      & $59.09$                       & $\underline{22.15}$        & $56.99$                    \\
			MetaGIA        & $60.56$                 & $61.75$                      & $53.69$                  & $62.08$                       & $32.78$                    & $60.14$                    \\
			\hfill +HAO    & $58.51$                 & $60.29$                      & $48.48$                  & $\underline{58.63}$           & $29.91$                    & $\mathbf{54.14}$           \\
			AGIA           & $60.10$                 & $60.66$                      & $48.86$                  & $61.98$                       & $31.06$                    & $59.96$                    \\
			\hfill +HAO    & $\mathbf{53.79}$        & $\mathbf{58.71}$             & $48.86$                  & $\mathbf{58.37}$              & $26.51$                    & $56.36$                    \\
			TDGIA          & $66.86$                 & $66.79$                      & $49.73$                  & $62.47$                       & $32.37$                    & $57.97$                    \\
			\hfill +HAO    & $65.22$                 & $65.46$                      & $49.54$                  & $59.67$                       & $22.32$                    & $\underline{54.32}$        \\
			ATDGIA         & $61.14$                 & $65.07$                      & $46.53$                  & $64.66$                       & $24.72$                    & $61.25$                    \\
			\hfill +HAO    & $58.13$                 & $63.31$                      & $\mathbf{44.40}$         & $59.27$                       & $\mathbf{17.62}$           & $56.90$                    \\\bottomrule
			\multicolumn{7}{l}{The lower is better. $^\dagger$Non-targeted attack. $^\ddagger$Targeted attack.  }
		\end{tabular}}
\end{table}

\subsection{Empirical Performance}
In Table~\ref{CH:HAO:tab:eval_non_targeted} and Table~\ref{CH:HAO:tab:eval_targeted}, we report the non-targeted and targeted attack performance of various GIA methods, respectively.
We bold out the best attack and underline the second-best attack when combating defenses from each category.
Full results are in Appendix~\ref{CH:HAO:sec:eval_nontarget_detailed} and Appendix~\ref{CH:HAO:sec:eval_target_detailed}.

\begin{wrapfigure}{r}{0.5\textwidth}
    \centering
	\includegraphics[width=0.5\textwidth]{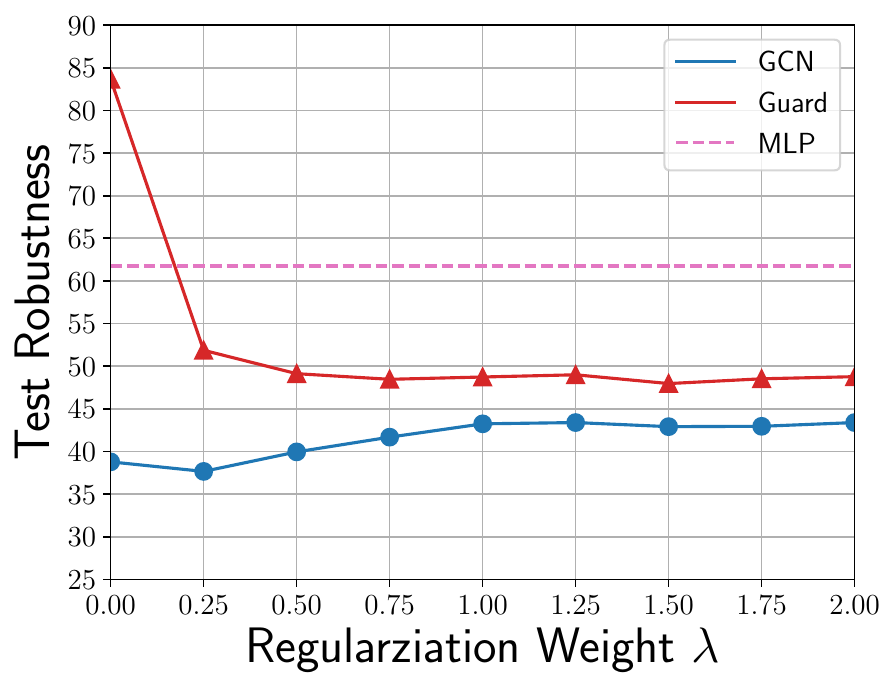}
	\caption{Effects of HAO with different weights.}
		\label{CH:HAO:fig:ablation_hao_coe}
    \vspace{-0.3in}
\end{wrapfigure}
\paragraph{Performance of non-targeted attacks.}
In Table~\ref{CH:HAO:tab:eval_non_targeted}, we can see that HAO significantly improves the performance of \textit{all} attacks on \textit{all} datasets up to $30\%$, which implies the effectiveness and versatility of HAO.
Especially, even coupled with a random injection strategy (PGD), HAO can attack robust models to be comparable with or inferior to simple MLP which does not consider relational information.
Meanwhile, adaptive injection strategies outperform previous methods PGD and TDGIA by a non-trivial margin for most cases, which further indicates that they are more suitable for HAO.

\paragraph{Performance of targeted attack on large-scale graphs.}
In Table~\ref{CH:HAO:tab:eval_targeted}, HAO also improves the targeted attack performance of \textit{all} attack methods on \textit{all} datasets by a significant margin of up to $15\%$,
which implies that the benefits of incorporating HAO are universal. Besides, adaptive injections can further improve the performance of attacks and establish the new state-of-the-art coupled with HAO.

\subsection{Analysis and Discussions}

\begin{figure}[t]
	\centering
	\subfigure[Homophily change]{
		\includegraphics[width=0.30\textwidth]{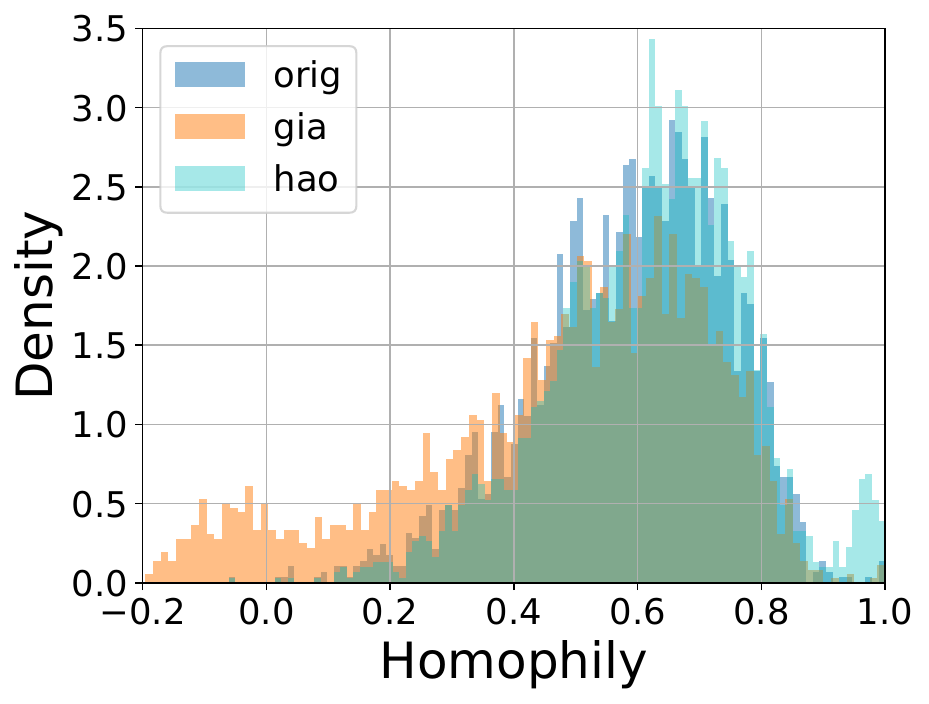}
		\label{CH:HAO:fig:gia_homophily_dist_node_atk_hao}
	}
	\subfigure[Varying node budgets]{
		\includegraphics[width=0.31\textwidth]{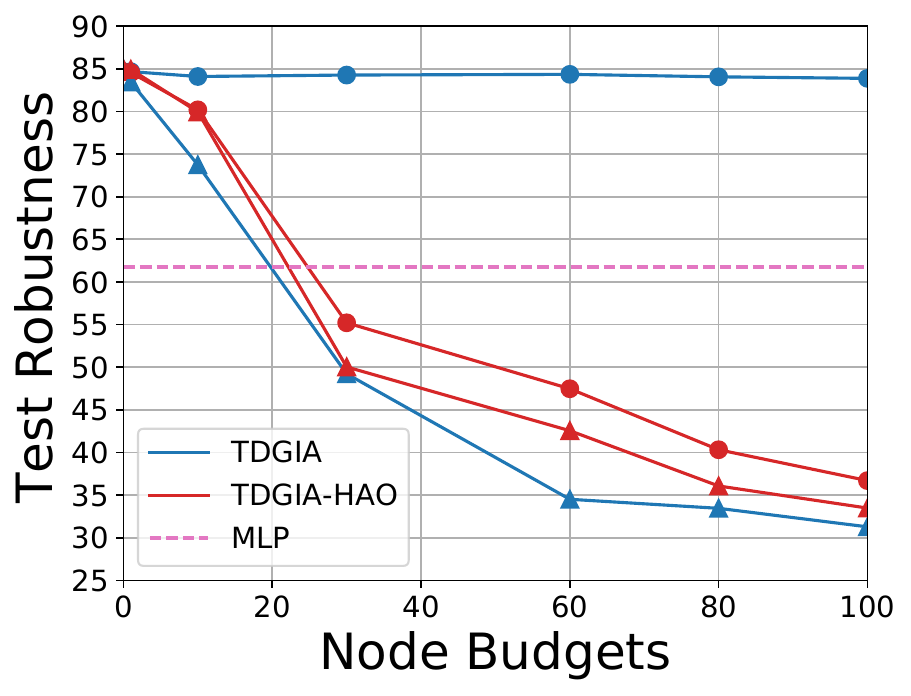}
		\label{CH:HAO:fig:ablation_node_budgets}
	}
	\subfigure[Varying edge budgets]{
		\includegraphics[width=0.31\textwidth]{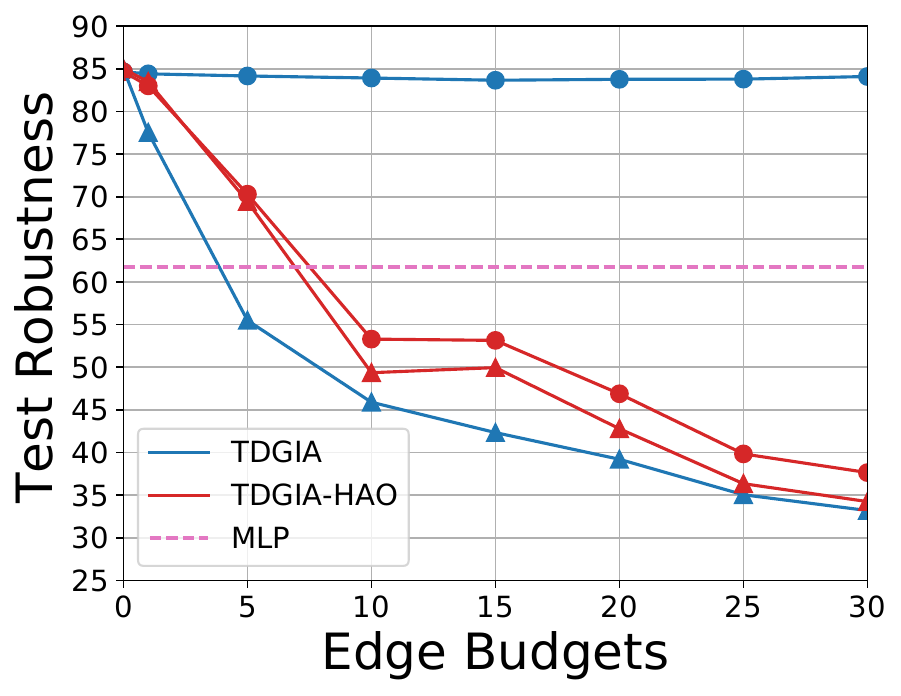}
		\label{CH:HAO:fig:ablation_edge_budgets}
	}
	\caption[Ablation studies of \hao.]{(a) Homophily changes after attacked by GIA without HAO (orange) and GIA with HAO (canny); (b), (c) Attack performance against GCN and EGuard with different node and edge budgets. $\bullet$ indicates attack with defenses and $\blacktriangle$ indicates attack without defenses;}
	\label{CH:HAO:fig:abalation_attacked_graph}
\end{figure}

\paragraph{Effects of HAO.} Though HAO can substantially improve GIA methods under defenses, we find it essentially trades with the performance under no defenses.
In Fig.~\ref{CH:HAO:fig:ablation_hao_coe}, as the weight for regularization term $\lambda$ increases, HAO trades slightly more of the performance against GCN for the performance against homophily defenders.
Finally, GIA reduces the performance of both GNNs with defenses and without defenses to be inferior to MLP.
Additionally, as also shown in Table~\ref{CH:HAO:tab:averaged_performance}, the trade-off will not hurt the overall performance while consistently brings benefits up to $5\%$.

\paragraph{Analysis of the perturbed graphs.}
In Fig.~\ref{CH:HAO:fig:gia_homophily_dist_node_atk_hao}, we also analyze the homophily distribution changes after the attack. It turns out that GIA with HAO can effectively preserve the homophily while still conducting effective attacks.
Similar analysis on other datasets can be found in Appendix~\ref{CH:HAO:sec:homophily_appdx}.

\paragraph{Attacks with limited budgets.}
We also examine the performance of GIA methods with or without HAO varying different node and edge budgets.
Fig.~\ref{CH:HAO:fig:ablation_node_budgets} and Fig.~\ref{CH:HAO:fig:ablation_edge_budgets} show that HAO can consistently improve the overall performance by slightly trading with the performance under no defenses.

\section{Causal Models of Unnoticeability in Graph Adversarial Attacks}
\label{CH:HAO:sec:causal_understand}

\subsection{Causal Models of the Graph Adversarial Attacks}
\label{CH:HAO:sec:scm_hao}
\begin{figure}[t]
	\centering
	\subfigure[Graph generation.]{
		\includegraphics[width=0.35\textwidth]{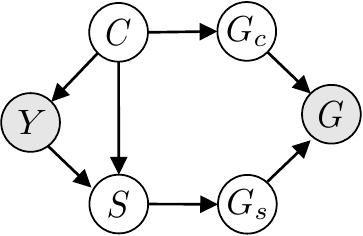}
		\label{CH:HAO:fig:hao_scm_p1}
	}
 \subfigure[Adversarial graph generation.]{
		\includegraphics[width=0.3\textwidth]{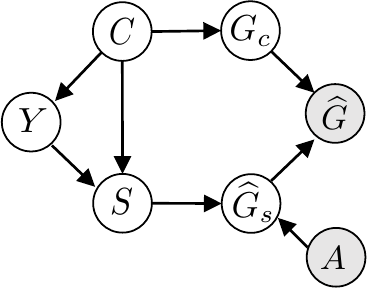}
		\label{CH:HAO:fig:hao_scm_p2}
	}
 	\caption{SCMs of the graph adversarial attacks.}
	\label{CH:HAO:fig:hao_scm}
\end{figure}
To gain a better understanding of the unnoticeability issue, we construct causal models of graph adversarial attacks. As shown in Fig.~\ref{CH:HAO:fig:hao_scm}, the left figure presents the graph generative models under regular conditions. 
In our causal models, we convert the task of node classification into graph classification. Given a GNN with $k$ rounds of message passing, for a target node $u$, it will take the information of the $k$-hop neighbors of the target node $u$ as inputs to make classifications. Hence classifying $u$ can be considered as classifying the ego-graph of node $u$, i.e., $k$-hop neighbors of the node $u$.
Inspired by~\cite{handle_node,ciga}, the generative processes of the observed graph $G$ and the target label $Y$ are controlled by latent variables $C$ and $S$. 

\paragraph{Graph Models.} For the generation of the $k$-hop ego-graph of node $u$, $G$, more specifically, $C$ retains the causality of the class information $Y$ and controls the generation of an invariant subgraph $G_c$ of the observed graph $G$. On the other hand, $S$ retains the other correlative information about class $Y$ and further controls the generation of a spurious subgraph $G_s$ of the observed graph $G$. For example, in a social network, for an influencer $u$, $G_c$ can be considered as the subgraph consisting of genuine friends that reflect the interests of the user $u$. $G_s$ can be considered as a subgraph consisting of families or relatives of node $u$ whose interests are different from $u$.  
In addition, at the latent space, $C$ and $S$ may have some interactions such as $C\rightarrow S$ or $Y\rightarrow S$. The interactions create correlations between $S$ and $Y$.
To determine the interests of node $u$, we need to leverage only the information of $C$ and avoid relying on $S$ to make decisions.

\paragraph{Unnoticeability in Adversarial Graph Attacks.} While for the adversary $A$, the unnoticeability constraints expect $A$ only to perturb the information about $S$ by peturbing the underlying spurious subgraph $G_s$. Otherwise, if the adversary perturbs the causal subgraph $G_c$, it will destroy the causal relationship between $C$ and $G$, thus the underlying label which is not visible to the adversary $A$, no longer aligns with the graph. However, the defender can access both the label and the graph information. The mismatch can easily be detected as shown by the robustness of the homophily defenders.

\subsection{Causality-Inspired Graph Adversarial Training}
\label{CH:HAO:sec:causal_defense}
Motivated by the aforementioned discussion, it is natural to leverage \hao to further improve the robustness of GNNs. Intuitively, when leveraging the vanilla graph adversarial attacks to generate the adversarial examples for training, it will allocate severe label noises as the vanilla graph adversarial attacks destroy the causal information. While incorporating \hao into the generation of the adversarial examples, the noises can be mitigated. Therefore, we conduct further experiments with GCN~\cite{gcn} and GNNGuard\cite{gnnguard} on Cora~\cite{cora}. We set the adversarial training epochs as $200$ in order to avoid overoptimization of vanilla graph adversarial attacks.

\begin{table}[t]
    \caption{Performance of adversarial training methods under various graph adversarial attacks.}
    \label{CH:HAO:tab:adv_train}
    \centering
    \scriptsize
    \begin{tabular}{@{}{l}*{10}{c}@{}}
        \toprule
        &              & Clean     & \multicolumn{2}{c}{PGD}  & \multicolumn{2}{c}{TDGIA} & \multicolumn{2}{c}{MetaGIA}                                                                                                                                                         \\
                                                  & HAO          &                               &                          & $\checkmark$              &                             & $\checkmark$             &                          & $\checkmark$             & \multicolumn{1}{l}{mean}     & \multicolumn{1}{l}{worst}             \\\midrule
        GCN                                       &              & $ 84.95                     $ & $ 38.55                $ & $ 38.55                $  & $ 40.67                $    & $ 43.78                $ & $ 38.43                $ & $ 38.80                 $ & $ 46.25              $       & $ 38.43           $                   \\
        GCN+FLAG                                  &              & $ 81.84                     $ & $ 59.95                $ & $ 57.71                $  & $ 59.82                $    & $ 54.60                 $ & $ 59.82                $ & $ 54.72                $ & $ 61.21              $       & $ 54.60            $                   \\
        GCN+PGD                                   &              & $ 86.19                     $ & $ \mathbf{72.76}       $ & $ \mathbf{72.13}       $  & $ \mathbf{80.34}       $    & $ \mathbf{75.49}       $ & $ \mathbf{70.77}       $ & $ 64.92                $ & $ 74.66     $                & $ 64.92                    $          \\
        GCN+PGD                                   & $\checkmark$ & $ \mathbf{86.94}            $ & $ \mathbf{72.88}       $ & $ \mathbf{72.63}       $  & $ \mathbf{81.21}       $    & $ \mathbf{79.22}       $ & $ \mathbf{72.01}       $ & $ \mathbf{68.78}       $ & $ 76.24     $                & $ \mathbf{68.78}                    $ \\
        GCN+TDGIA                                 &              & $ 85.69                     $ & $ 66.29                $ & $ 65.29                $  & $ 75.74                $    & $ 71.76                $ & $ 64.92                $ & $ 58.83                $ & $ 69.79              $       & $ 58.83           $                   \\
        GCN+TDGIA                                 & $\checkmark$ & $ \mathbf{86.56}            $ & $ 70.14                $ & $ 69.40                 $  & $ 79.35                $    & $ 75.87                $ & $ 69.02                $ & $ \mathbf{65.42}       $ & $ 73.68                    $ & $ 65.42           $                   \\\hdashline[0.5pt/1pt]
        \hdashline[0.5pt/3pt]\rule{0pt}{10pt}GNNGuard &              & $ 85.07                     $ & $ 84.20                 $ & $ 84.70                 $  & $ 84.45                $    & $ 53.73                $ & $ 84.82                $ & $ 43.15                $ & $ 74.30              $       & $ 43.15           $                   \\
        GNNGuard+FLAG                                 &              & $ 84.57                     $ & $ 84.32                $ & $ 84.32                $  & $ 84.32                $    & $ 69.77                $ & $ 84.45                $ & $ 64.92                $ & $ 79.52              $       & $ 64.92           $                   \\
        GNNGuard+PGD                                  &              & $ \mathbf{86.44}            $ & $ \mathbf{86.69}       $ & $ \mathbf{85.69}       $  & $ \mathbf{86.56}       $    & $ 71.51                $ & $ \mathbf{86.19}       $ & $ 57.08                $ & $ 80.02              $       & $ 57.08           $                   \\
        GNNGuard+PGD                                  & $\checkmark$ & $ \mathbf{86.44}            $ & $ \mathbf{86.31}       $ & $ \mathbf{86.06}       $  & $ \mathbf{86.19}       $    & $ \mathbf{77.86}       $ & $ \mathbf{86.31}       $ & $ \mathbf{69.77}       $ & $ \mathbf{82.71}     $       & $ \mathbf{69.77}                    $ \\
        GNNGuard+TDGIA                                &              & $ 85.94                     $ & $ 85.94                $ & $ 85.57                $  & $ 85.82                $    & $ 71.14                $ & $ 85.69                $ & $ 56.46                $ & $ 79.51              $       & $ 56.46           $                   \\
        GNNGuard+TDGIA                                & $\checkmark$ & $ 85.57                     $ & $ 85.69                $ & $ 85.57                $  & $ 85.32                $    & $ \mathbf{76.61}       $ & $ 85.57                $ & $ \mathbf{65.17}       $ & $ \mathbf{81.36}     $       & $ 65.17                    $          \\\bottomrule
    \end{tabular}%
\end{table}

As given in Table~\ref{CH:HAO:tab:adv_train}, when incorporated \hao into the adversarial training, the robustness of both GCN and GNNGuard increase significantly by $10\%$. On the other hand, due to the overoptimization and the noises brought by adversarial training with vanilla adversarial attacks, the trained models will underperform a simple baseline FLAG~\cite{flag}, which injects mild adversarial noises into both node features and graph structures for adversarial training. The empirical results serve as strong evidence for the benefits of incorporating the causality to understand and improve the robustness of GNNs.

\part{Optimizations}\label{P3}
\chapter{Optimization Dilemma in Causal Invariance Learning} \label{CH:PAIR}
Although learning the causality demonstrates great potential as shown in previous chapters, it remains unexplored to what extent one could realize the desired causal invariance learning objectives. Therefore, Chapter~\ref{CH:PAIR} characterizes the optimization dilemma in realizing the causal invariance learning with the traditional empirical risk minimization, and proposes a new optimization strategy to mitigate the dilemma. Furthermore, Chapter~\ref{CH:FeAT} delves deeper into the feature learning dynamics under the optimization dilemma and proposes a new representation learning framework to resolve the dilemma.

\section{Motivations}
The interplay between optimization and generalization is crucial to the success of deep learning~\citep{memorization,arora_dlt,beyond_2l,ntk,featlearn}.
Guided by empirical risk minimization (ERM)~\citep{erm}, simple optimization algorithms can find uneventful descent paths in the non-convex loss landscape of deep neural networks~\citep{hess_analysis}.
However, when distribution shifts are present, the optimization is usually biased by spurious signals such that the learned models can fail dramatically in
\emph{Out-of-Distribution} (OOD) data~\citep{camel_example,covid19_application,shortcut_dl}.
Therefore, overcoming the OOD generalization challenge has drawn much attention recently. Most efforts are devoted to proposing better \emph{optimization objectives}~\citep{causal_transfer,iga,andmask,vrex,env_inference,jtt,sd,ib-irm,clove,fish,fishr,ciga} that regularize the gradient signals produced by ERM,
while it has been long neglected that the interplay between optimization and generalization under distribution shifts has already changed its nature.

\begin{figure}[t]
	\subfigure[Theoretical failure case.]{
		\includegraphics[width=0.25\textwidth]{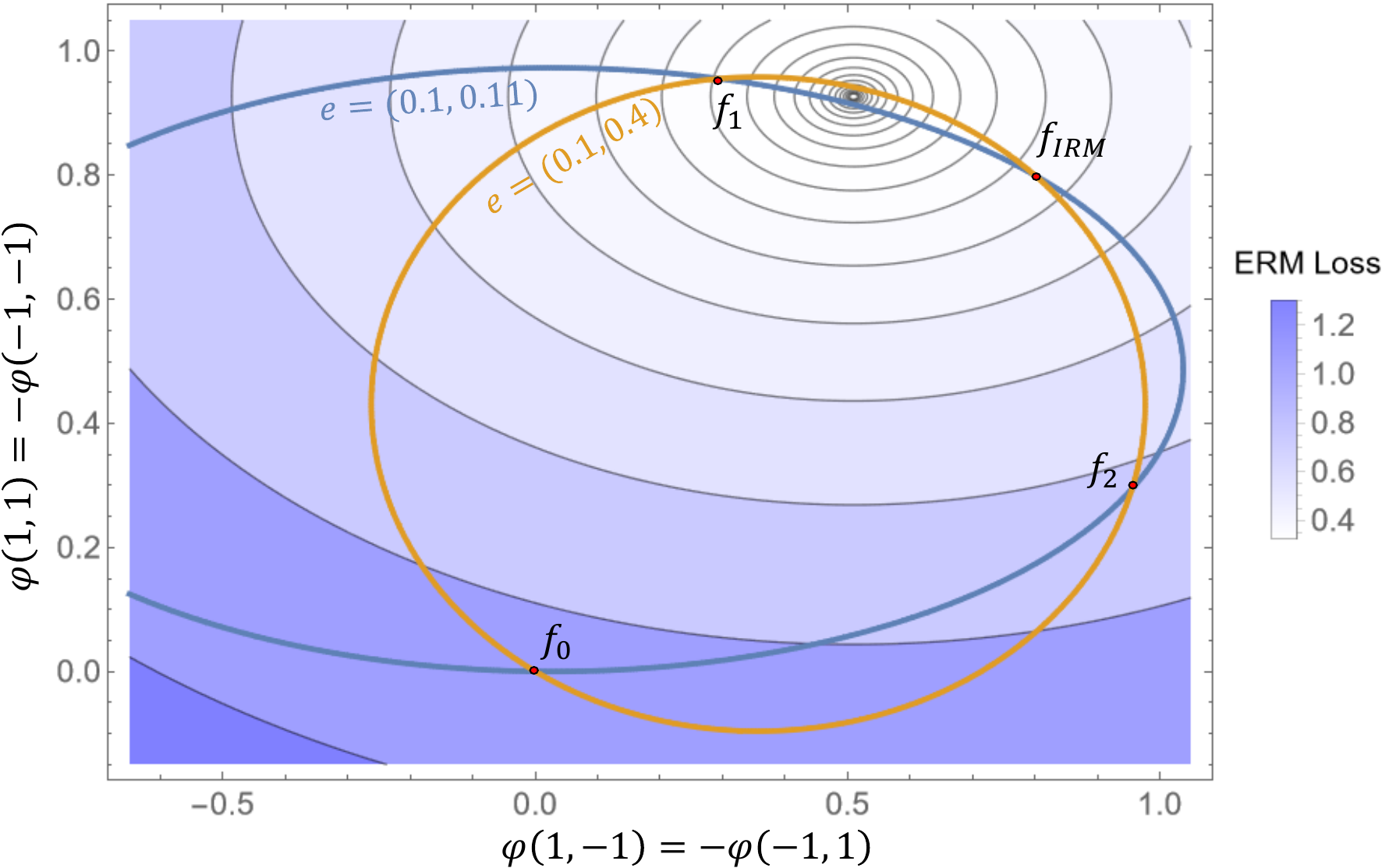}
		\label{CH:PAIR:fig:aistats_fail}
	}
	\subfigure[Gradient conflicts.]{
		\includegraphics[width=0.221\textwidth]{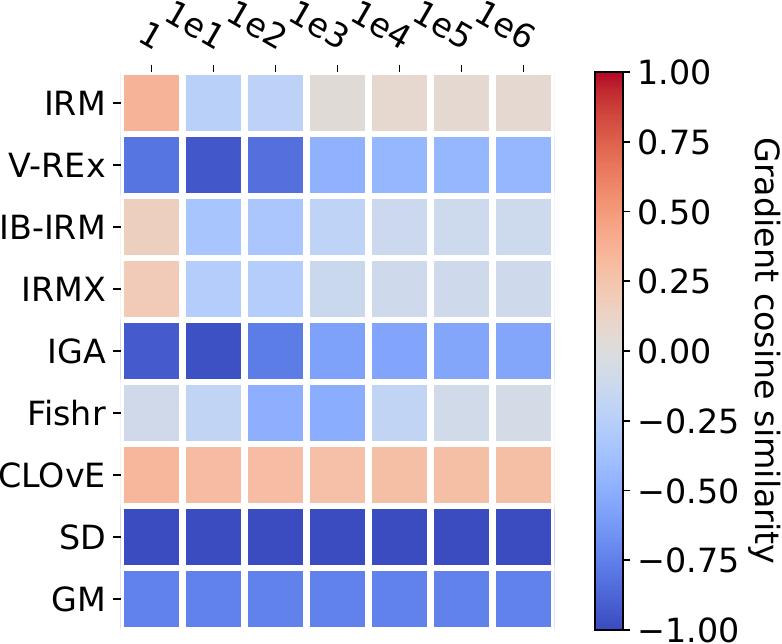}
		\label{CH:PAIR:fig:grad_conflict}
	}
	\subfigure[Unreliable opt. scheme.]{
		\includegraphics[width=0.22\textwidth]{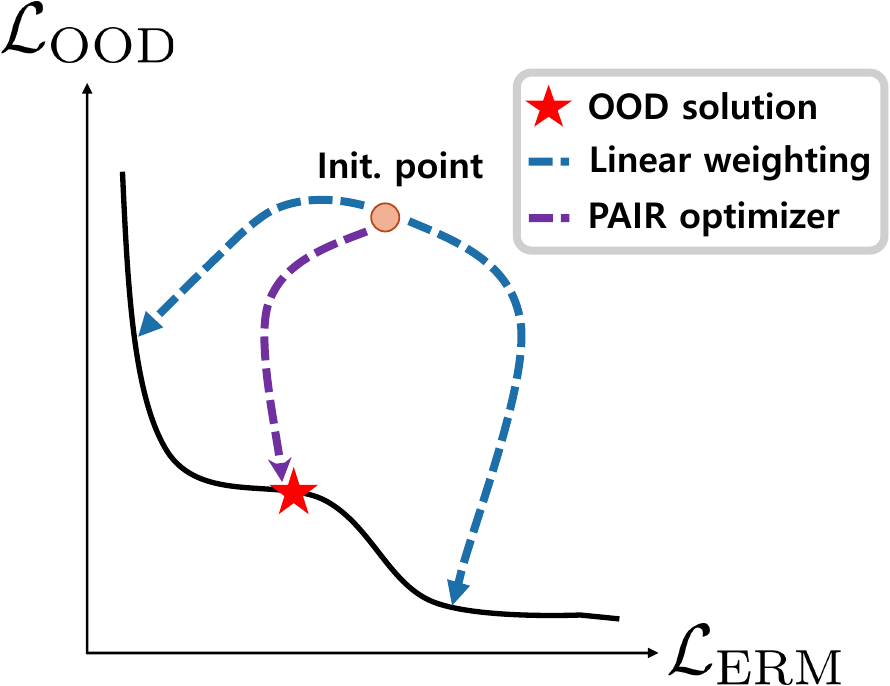}
		\label{CH:PAIR:fig:bad_scalar}
	}
	\subfigure[Exhaustive tuning.]{
		\includegraphics[width=0.203\textwidth]{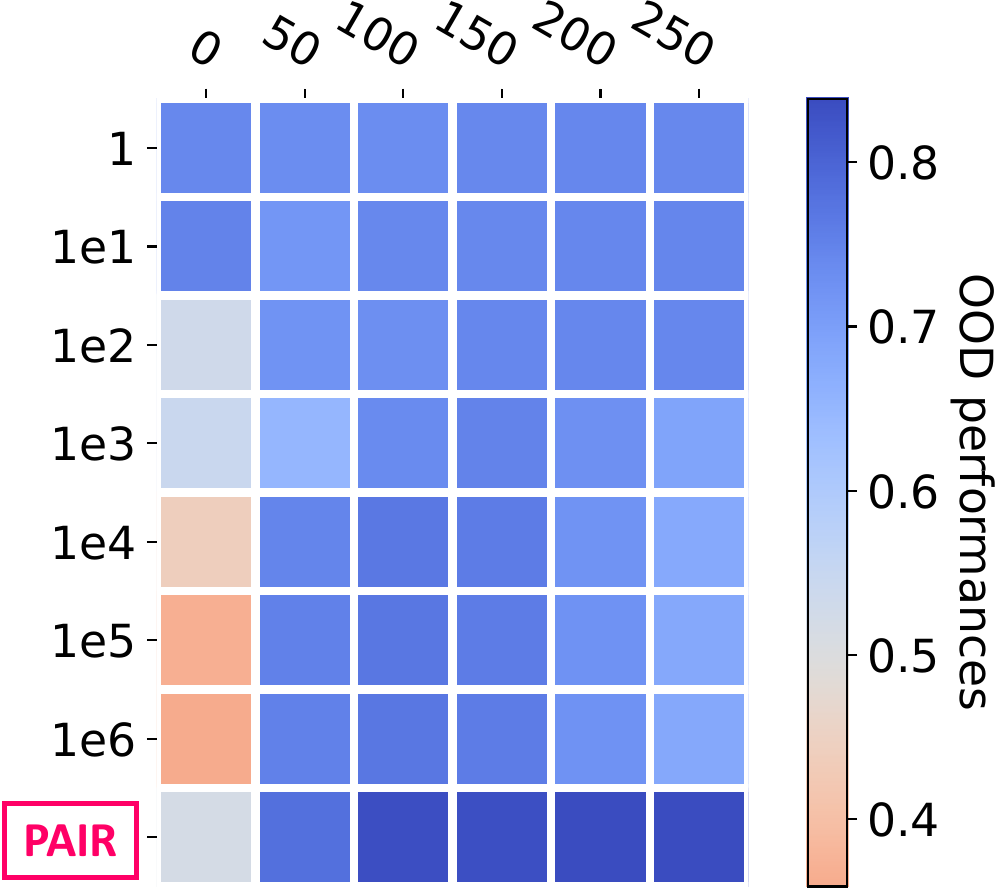}
		\label{CH:PAIR:fig:sweep_acc}
	}
	\caption[Optimization issues in OOD algorithms.]{Optimization issues in OOD algorithms. (a) OOD objectives such as \irm usually require several relaxations for the ease of optimization, which however introduces huge gaps. The ellipsoids denote solutions that satisfy the invariance constraints of practical \irm variant \irml. When optimized with ERM, \irml prefers $f_1$ instead of $f_\irm$ (The predictor produced by \irm).\! (b) The gradient conflicts between ERM and OOD objectives generally exist for different objectives at different penalty weights ($x$-axis).
		(c) The typically used linear weighting scheme to combine ERM and OOD objectives requires careful tuning of the weights to approach the solution. However, the scheme cannot reach any solutions in the non-convex part of the Pareto front.
		In contrast, \pair finds an adaptive descent direction under gradient conflicts that leads to the desired solution. (d) Due to the optimization dilemma, the best OOD performance (e.g., \irml w.r.t. a modified \cmnist from Sec.~\ref{CH:PAIR:sec:experiments}) usually requires exhaustive tuning of hyperparameters ($y$-axis: penalty weights; $x$-axis: pretraining epochs), while \pair robustly yields top performances by resolving the compromises.}
	\label{CH:PAIR:fig:ood_opt_issue}
\end{figure}

In fact, the \emph{optimization process} of the OOD objectives turns out to be substantially more challenging than ERM.
There are often compromises when applying the OOD objectives in practice.
Due to the optimization difficulty,
many OOD objectives have to be relaxed as penalty terms of ERM in practice~\citep{irmv1,iga,vrex,sd,ib-irm,fishr},
but the relaxed formulations can behave very differently from the original objective~\citep{irm_aistats} (Fig.~\ref{CH:PAIR:fig:aistats_fail}).
Moreover, due to the generally existing gradient conflicts between ERM and OOD objectives (Fig.~\ref{CH:PAIR:fig:grad_conflict}), trade-offs among ERM and OOD performance during the optimization are often needed.
\citet{groupdro,gen_reweighted} suggest that ERM performance usually needs to be sacrificed for better OOD generalization.
On the other hand,
it usually requires careful tuning of the OOD penalty hyperparameters \citep{rfc} (Fig.~\ref{CH:PAIR:fig:sweep_acc}),
which however either weakens the power of OOD objectives or makes them too strong that preventing models from capturing all desirable patterns.
Consequently, using the traditional optimization wisdom to train and select models can \emph{easily lead to suboptimal performance} of either ERM or OOD objectives.
Most OOD objectives remain struggling with distribution shifts or even underperform ERM~\citep{domainbed,wilds}.
This phenomenon calls for a better understanding of the optimization in OOD generalization, and raises a challenging question:
\begin{myquotation}
	\emph{How can one obtain a desired OOD solution under the conflicts of ERM and OOD objectives?}
\end{myquotation}

To answer this question, we take a multi-objective optimization (MOO) perspective of OOD optimization. Specifically, using the representative OOD objective IRM~\citep{irmv1} as an example, we find that the failures in OOD optimization can be attributed to two issues.
The first one is the compromised robustness of OOD objectives due to the relaxation in the practical variants.
In fact, it can even eliminate
the desired invariant solution from the Pareto front w.r.t. the ERM and the OOD penalty (Fig.~\ref{CH:PAIR:fig:aistats_fail}). Therefore, merely optimizing the ERM and the relaxed OOD penalty can hardly approach the desired solution.
On the other hand, when the Pareto front contains the desired solution, as shown in Fig.~\ref{CH:PAIR:fig:bad_scalar}, using the traditional linear weighting scheme that linearly reweights the ERM and OOD objectives, cannot reach the solution if it lies in the non-convex part of the front~\citep{bv_convex}. Even when the OOD solution is reachable (i.e., lies in the convex part), it still requires careful tuning of the OOD penalty weights to approach the solution, as shown in Fig.~\ref{CH:PAIR:fig:sweep_acc}.

To address these issues, we propose a new optimization scheme for OOD generalization, called \pairfull (\pair), which includes a new optimizer (\pairo) and a new model selection criteria (\pairs).
Owing to the MOO formulation, \pairo allows for cooperative optimization with other OOD objectives to improve the robustness of practical OOD objectives.
Despite the huge gaps between \irml and \irm, we show that incorporating VREx~\citep{vrex} into \irml provably recovers the causal invariance~\citep{irmv1} for some group of problem instances (Sec.~\ref{CH:PAIR:sec:irm_pair_sol}).
When given robust OOD objectives, \pairo finds a descent path with adaptive penalty weights, which leads to a Pareto optimal solution that trades off ERM and OOD performance properly (Sec.~\ref{CH:PAIR:sec:pair_solution}).
In addition, the MOO analysis also motivates \pairs, which facilitates the OOD model selection by considering the trade-offs between ERM and OOD objectives.

We conducted extensive experiments on challenging OOD benchmarks. Empirical results show that \pairo successfully alleviates the objective conflicts and empowers \irml to achieve high performance in $6$ datasets from \wilds~\citep{wilds}.\!
\pairs effectively improves the performance of selected OOD models up to $10\%$ across $3$ datasets from \dobed~\citep{domainbed}, demonstrating the significance of considering the ERM and OOD trade-offs in optimization.

\section{Background and related work}
\label{CH:PAIR:sec:related_work}
We first briefly introduce the background of our work (more details are given in Appendix~\ref{CH:PAIR:sec:related_work_appdx}.

\paragraph{Problem setup.}
The problem of OOD generalization typically considers
a supervised learning setting based on the data $\dataset=\{\dataset^e\}_{e\in\envall}$
collected from multiple causally related environments $\envall$,
where a subset of samples $\dataset^e=\{X^e_i,Y^e_i\}$ from a single environment $e\in\envall$
are drawn independently from an identical distribution $\sP^e$~\citep{inv_principle}.
Given the data from training environments $\{\dataset^e\}_{e\in\envtrain}$,
the goal of OOD generalization is to find a predictor $f:\gX\rightarrow\gY$
that generalizes well to all (unseen) environments, i.e., to minimize
$\max_{e\in\envall}\gL_e(f)$, where $\gL_e$ is the empirical risk under environment $e$.
The predictor $f=w\circ\varphi$ is usually composed of a featurizer $\varphi:\gX\rightarrow\gZ$ that learns to extract useful features, and a classifier $w:\gZ\rightarrow\gY$ that makes predictions from the extracted features.

\paragraph{Existing solutions to OOD generalization.}
There exists a rich literature aiming to overcome the OOD generalization challenge, which usually appear as \emph{additional regularizations} of ERM~\citep{erm}.
\citet{DANN,CORAL,deep_DG,DouCKG19} regularize the learned features to be \textbf{domain-invariant}.
\citet{dro,DRSL,groupdro} regularize the models to be \textbf{robust to mild distributional perturbations} of the training distributions, and~\citet{adv_causal_lens,jtt,cnc,lisa} improve the robustness with additional assumptions.
Recently there is increasing interest in adopting the causality theory~\citep{causality,towards_causality} and introducing the \textbf{causal invariance} to representation learning~\citep{inv_principle,irmv1,env_inference,andmask,clove,ib-irm}. They require $\varphi$ to learn causally invariant representations such that a predictor $w$ acting on $\varphi$ minimizes the risks of all the environments simultaneously. This work focuses on resolving the optimization issue in learning the causal invariance.
In addition, \citet{iga,vrex,fish,fishr} implement the invariance by encouraging \textbf{agreements} at various levels across environments.
However, they mostly focus on developing better objectives while neglecting the optimization process of the objectives.

\paragraph{Optimization dilemma in OOD generalization.}
Along with the development of OOD methods, the OOD optimization dilemma is gradually perceived in the literature.
\citet{domainbed} find it hard to select a proper model in OOD generalization given ERM performance at different environments.
\citet{groupdro,gen_reweighted} find the ERM performance needs to be sacrificed for  satisfactory OOD performance.
Some initial trials are proposed.
\citet{pareto_da} use the guidance of the data from similar distributions with the test environment in MOO to resolve gradient conflicts and achieve better performance in domain adaption.
\citet{rfc} propose to construct diverse initializations for stabilizing OOD performance under the dilemma.
However, why there exists such a dilemma in OOD generalization and whether we can resolve it remain elusive.

\textbf{Multi-Objective Optimization (MOO).}
MOO considers solving $m$ objectives w.r.t. $\{\gL_i\}_{i=1}^m$ losses,
i.e., $\min_\theta\mL(\theta)\!=\!(\gL_1(\theta),...,\gL_m(\theta))^T$~\citep{moo_book}.
A solution $\theta$ dominates another $\bar{\theta}$, i.e., $\mL(\theta)\preceq\mL(\bar{\theta})$, if $\gL_i(\theta)\leq\gL_i(\bar{\theta})$ for all $i$ and $\mL(\theta)\neq\mL(\bar{\theta})$.
A solution $\theta^*$ is called \textbf{Pareto optimal} if no other $\theta$ dominates $\theta^*$. The set of Pareto optimal solutions is called Pareto set ($\gP$) and its image is called \textbf{Pareto front}.
In practice, it is usual that one cannot find a global optimal solution for all objectives, hence Pareto optimal solutions are of particular value.
Although MOO has been widely applied to improving multi-task learning~\citep{mtl_moo},
it remains underexplored on how to model and mitigate objective conflicts in OOD generalization from the MOO perspective.

\section{Optimization Challenges in IRM and its Effective Fix}
\label{CH:PAIR:sec:irm_usecase}
This work focus on one of the most representative OOD objectives in learning the causal invariance \!--\! IRM, %
\textbf{}to show how we can understand and mitigate the optimization dilemma through the MOO lens.
\subsection{Drawbacks of IRM in Practice}
\label{CH:PAIR:sec:irm_drawback}
We first introduce the basics of IRM and the drawbacks of its practical variants, and leave theoretical details in Appendix~\ref{CH:PAIR:sec:irm_failure_appdx}.
Specifically, the IRM framework approaches OOD generalization by finding an invariant representation $\varphi$,
such that there exists a classifier acting on $\varphi$ that is
simultaneously optimal in $\envtrain$.
Hence, IRM leads to a challenging bi-level optimization problem as
\begin{equation}
	\label{CH:PAIR:eq:irm}
	\min_{w,\varphi}  \ \sum_{e\in\envtrain}\gL_e(w\circ\varphi),
	\text{s.t.}
	\ w\in\argmin_{\bar{w}:\gZ\rightarrow\gY} \gL_e(\bar{w}\circ\varphi),\ \forall e\in\envtrain.
\end{equation}
Given the training environments $\envtrain$, and functional spaces $\gW$ for $w$ and $\varPhi$ for $\varphi$,
predictors $f=w\circ\varphi$ satisfying the constraint in Eq.~\ref{CH:PAIR:eq:irm} are called invariant predictors,
denoted as $\gI(\envtrain)$.
When solving for invariant predictors,
characterizing $\gI(\envtrain)$ is particularly difficult in practice,
hence it is natural to restrict $\gW$ to be the space of linear functions on $\gZ=\R^d$~\citep{ntk}.
Furthermore, \citet{irmv1} argue that linear classifiers actually do not provide additional representation power than \emph{scalar} classifiers, i.e., $d=1,\gW=\gS=\R^1$. The scalar restriction elicits a practical variant \irms as
\begin{equation}
	\label{CH:PAIR:eq:irms}
	\min_{\varphi} \ \sum_{e\in\envtrain}\gL_e(\varphi),
	\text{s.t.}
	\ \nabla_{w|w=1}\gL_e(w\cdot\varphi)=0,\ \forall e\in\envtrain.
\end{equation}

Since Eq.~\ref{CH:PAIR:eq:irms} remains a constrained programming. \citet{irmv1} further introduce a soften-constrained variant, called \irml, as the following
\begin{equation}
	\label{CH:PAIR:eq:irml}
	\min_{\varphi}  \sum_{e\in\envtrain}\gL_e(\varphi)+\lambda|\nabla_{w|w=1}\gL_e(w\cdot\varphi)|^2.
\end{equation}

\textbf{Theoretical failure of practical IRM variants.}
Although the practical variants seem promising,
the relaxations introduce huge gaps between IRM and the practical variants, so that both \irms and \irml can fail to capture the invariance~\citep{irm_aistats}.
The failure case is illustrated by the two-bit environment with $\alpha_e,\beta_e\in[0,1]$.
Each environment $\dataset_e=\{X^e,Y^e\}$ is generated following
\begin{equation}
	\label{CH:PAIR:eq:twobit_env}
	Y^e:=\rad(0.5),\ X^e:=(X^e_1,X^e_2),\ X_1^e:=Y^e{\cdot}\rad(\alpha_e),\ X^e_2:=Y^e{\cdot}\rad(\beta_e),
\end{equation}
where $\rad(\sigma)$ is a random variable taking value $-1$ with probability $\sigma$ and $+1$ with probability $1-\sigma$.
Each environment is denoted as $\envalpha=\{(\alpha,\beta_e):0<\beta_e<1\}$ where $X^e_1$ is the invariant feature as $\alpha$ is fixed for different environment $e$, and $X^e_2$ is the spurious feature as $\beta_e$ varies across different $e$.

Let $\gI_\gS(\envtrain)$ denote the set of invariant predictors elicited by the relaxed constraint in \irms. It follows that $\gI (\envtrain)\subseteq \gI_\gS(\envtrain)$.
Consequently, there exist some undesired predictors but considered ``invariant'' by \irms and \irml.
For example, in $\envtrain\!=\!\{(0.1,0.11),(0.1,0.4)\}$, the solutions satisfying the constraint in \irms are those intersected points in Fig.~\ref{CH:PAIR:fig:aistats_fail} (The ellipsoids are the constraints).
Although $f_1,f_\irm\in\gI_\gS(\envtrain)$, both \irms and \irml prefer $f_1$ instead of $f_\irm$ (the predictor produced by \irm), as $f_1$ has the smallest ERM loss.
In fact, \citet{irm_aistats} show that the failure can happen in a wide range of environments even given \emph{infinite} amount of environments and samples,
demonstrating the huge gap between the practical and the original IRM variants.

\textbf{Empirical drawback of practical IRM variants.}
In addition,
the optimization of \irml introduces more challenges due to the conflicts between the \irml penalty and ERM objective.
As shown in Fig.~\ref{CH:PAIR:fig:sweep_acc},
it often requires significant efforts to tune the hyperparameters such as pretraining epochs and penalty weights $\lambda$ in Eq.~\ref{CH:PAIR:eq:irml}.
Otherwise, the \irml penalty could be either too weak to enforce the invariance as required by \irm, or too strong that prevents ERM from learning all desirable patterns.

\subsection{Pareto Optimization for IRM}
\label{CH:PAIR:sec:irm_pair_sol}
As shown that both \irms and \irml fail to properly trade off between ERM and \irm objectives, we switch to a new perspective, i.e., the lens of MOO, to understand the failures of \irm in practice.

\begin{wrapfigure}{r}{0.3\textwidth}
	\includegraphics[width=0.3\textwidth]{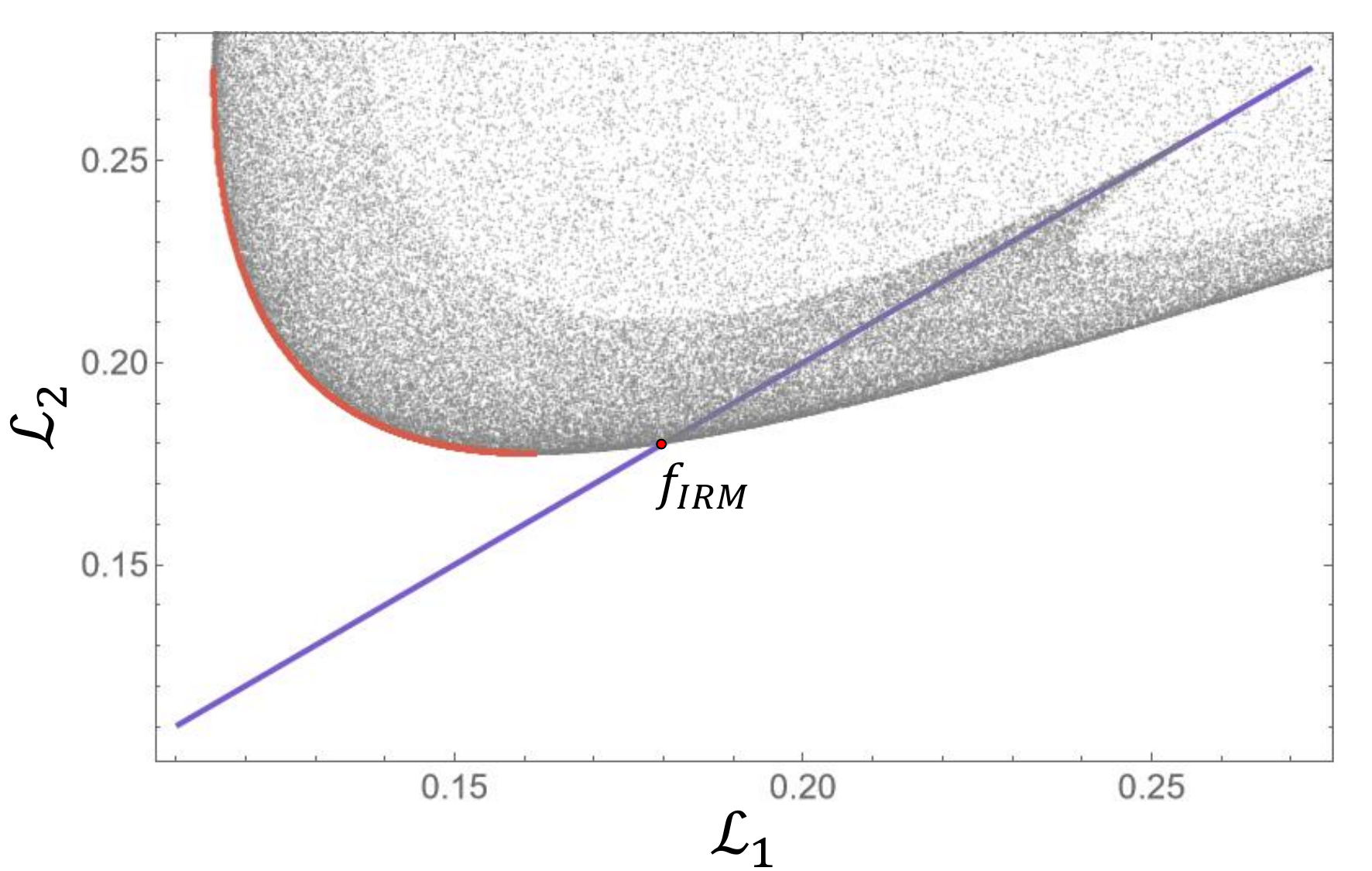}
	\caption{Pareto front of ERM losses w.r.t. environments.}
	\label{CH:PAIR:fig:pareto_front_mse}
\end{wrapfigure}

\textbf{Understanding the \irm failures through the MOO perspective.}
To begin with, it is natural to reformulate the practical \irm problem (Eq.~\ref{CH:PAIR:eq:irml}) as a MOO problem:
\begin{equation}
	\label{CH:PAIR:eq:irm_moo}
	\min_{\varphi} (\gL_\erm,\gL_\irm)^T,
\end{equation}
where $\gL_\erm =\frac{1}{|\envtrain|}\sum_{e\in\envtrain} \gL_e$ denotes the ERM loss, and
$\gL_\irm=\sum_e |\nabla_{w|w=1}\gL_e(w\cdot\varphi)|^2$ denotes the practical \irml loss.
To understand the behaviors of solutions to Eq.~\ref{CH:PAIR:eq:irm_moo},
We visualize the Pareto front w.r.t. $\{\gL_e\}_{e\in\envtrain}$ using the
previous failure case in Fig.~\ref{CH:PAIR:fig:aistats_fail}.

Let $\gP(\gL_1(\theta),...,\gL_m(\theta))$ denote the set of Pareto optimal solutions w.r.t. $(\gL_1(\theta),...,\gL_m(\theta))$.
As shown in Fig.~\ref{CH:PAIR:fig:pareto_front_mse}, at first, we can find that $f_\irm\notin\gP(\gL_1, \gL_2)$.
In other words, solving any environment-reweighted ERM losses cannot obtain $f_\irm$.
Moreover, together with Fig.~\ref{CH:PAIR:fig:aistats_fail}, the failure remains even combined with the \irms or \irml, i.e., $f_\irm\notin\gP(\gL_1,\gL_2,\gL_\irm)$, hence $f_\irm\notin\gP(\gL_\erm,\gL_\irm)$, as $f_\irm$ is dominated by $f_1$.
Therefore, no matter how we carefully control the optimization process, we cannot obtain $f_\irm$ by merely minimizing the objectives in Eq.~\ref{CH:PAIR:eq:irm_moo}.
This is essentially because of the weakened OOD robustness of \irms and \irml caused by the relaxations.
Thus, choosing robust objectives for optimization is of great importance to OOD generalization. The ideal objectives should at least constitute a Pareto front that contains the desired OOD solution.

\textbf{Improving OOD robustness of practical IRM variants.}
In pursuit of proper optimization objectives,
we resort to the OOD extrapolation explanation of IRM~\citep{bottou_talk}.
\begin{wrapfigure}{r}{0.33\textwidth}
	\includegraphics[width=0.33\textwidth]{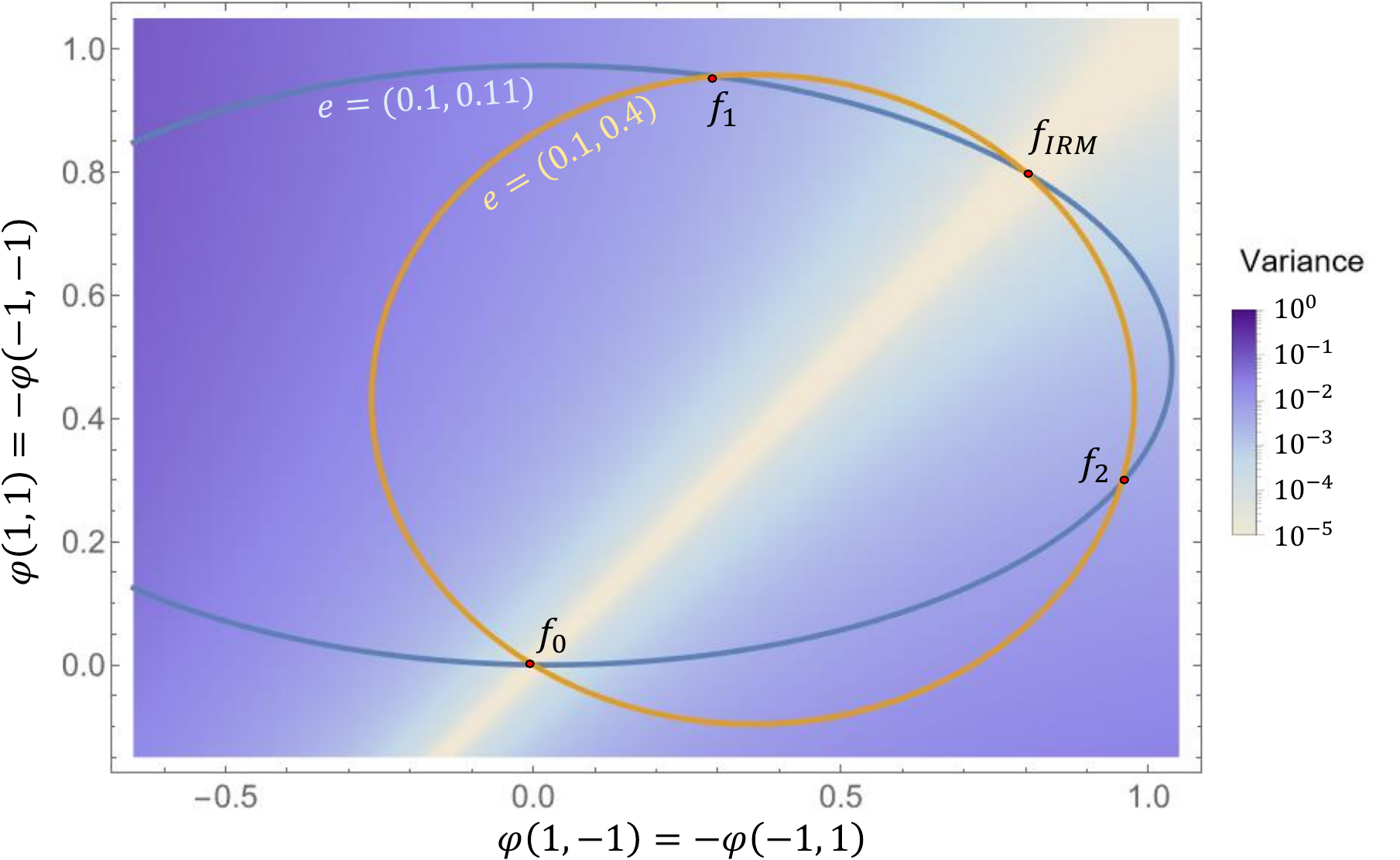}
	\caption{Variance distribution.}
	\label{CH:PAIR:fig:aistats_var_mse}
\end{wrapfigure}
A solution that is simultaneously optimal to all training environments (i.e., satisfying the original \irm constraints) is also a stationary point of ERM loss w.r.t. some OOD distribution:
\begin{equation}
	\label{CH:PAIR:eq:irm_dro}
	\partial \gL_t /\partial  f_\irm = \mathbf{0},\ \gL_t \in \{\text{$\sum$}_{e\in\envtrain}\lambda_e\gL_e|\text{$\sum$}_{e\in\envtrain}\lambda_e=1\},
\end{equation}
where $\gL_t$ is the ERM loss under the OOD distribution.
Different from Distributionally Robust Optimization approaches~\citep{dro}, Eq.~\ref{CH:PAIR:eq:irm_dro} allows for some negative $\lambda_e$ and hence its solutions are expected to extrapolate better~\citep{bottou_talk}.

The previous failure case implies that both \irms and \irml fail in the extrapolation due to the relaxations,
nevertheless, we can introduce additional objectives to directly improve the OOD extrapolation power of the practical \irm variants.
To this end, we introduce the REx objective to \irml,
which is derived by directly minimizing the worst-case ERM loss under all OOD distributions up to a certain distance from the training distributions~\citep{vrex}.
More formally, REx minimizes the worst case $\gL_t$ under an additional constraint of $\{\lambda_e\}_{e\in\envtrain}\geq -\beta$ in Eq.~\ref{CH:PAIR:eq:irm_dro}.
For the ease of optimization, they also propose an alternative objective as $\gL_\vrex := \var(\{\gL_e\}_{e\in\envtrain})$.
In Fig.~\ref{CH:PAIR:fig:aistats_var_mse}, we plot the distribution of $\gL_\vrex$ in the
the failure case of Fig.~\ref{CH:PAIR:fig:aistats_fail}.
It can be found that, $f_\irm$ lies in the low variance region.
Similarly, in Fig.~\ref{CH:PAIR:fig:pareto_front_mse}, the zero variance solutions (shown as the purple line in the middle) point out the underlying $f_\irm$ beyond the Pareto front.
Therefore, incorporating $\gL_\vrex$ in Eq.~\ref{CH:PAIR:eq:irm_moo} can relocate $f_\irm$ into the Pareto front, which implies the desirable objectives as the following
\begin{equation}
	\label{CH:PAIR:eq:irmx_moo}
	(\text{\irmx})\qquad\qquad\qquad\qquad\min_{\varphi} (\gL_\erm,\gL_\irm,\gL_\vrex)^T.\qquad\qquad\qquad\qquad
\end{equation}
By resolving a large class of failure cases of \irms and \irml~\citep{irm_aistats}, solutions to Eq.~\ref{CH:PAIR:eq:irmx_moo} are more powerful than those to \irms and \irml
in OOD extrapolation.
In fact, we have
\begin{proposition}(Informal)\label{CH:PAIR:thm:recover_IRM_paper} Under Setting A (\citet{irm_aistats}), for all $\alpha\in (0,1)$,
	let $\mathcal{E} \coloneqq \{(\alpha, \beta_e): \beta_e \in (0,1)\}$ be any instance of the two-bit environment (Eq.~\ref{CH:PAIR:eq:twobit_env}),
	$\Ix$ denote the invariant predictors produced by Eq.~\ref{CH:PAIR:eq:irmx_moo},
	it holds that $\Ix(\mathcal{E}) = \mathcal{I}(\mathcal{E})$.\footnote{Readers might be interested in the necessities of keeping \irml in the objectives. Proposition~\ref{CH:PAIR:thm:recover_IRM_paper}  considers only the ideal case, we additionally provide more empirical reasons in  Appendix~\ref{CH:PAIR:sec:discuss_pair_objs_appdx}; Our results can also be extended to multi-class following typical machine learning theory practice.}
\end{proposition}
The formal description and proof of Proposition~\ref{CH:PAIR:thm:recover_IRM_paper} are given in Appendix~\ref{proof:recover_IRM}. Proposition~\ref{CH:PAIR:thm:recover_IRM_paper} implies that Eq.~\ref{CH:PAIR:eq:irmx_moo} are the ideal objectives for optimization.
However, Eq.~\ref{CH:PAIR:eq:irmx_moo} can even add up the difficulty of OOD penalty tunning.
It introduces one more penalty to the overall objective that makes the Pareto front more complicated for the linear weighting scheme to find the desired solution.

\textbf{Pareto optimization for \irmx.}
Ideally, the set of Pareto optimal solutions is small such that each $f\in\gP(\gL_\erm,\gL_\irm,\gL_\vrex)$ satisfies the invariance constraints of \irml and \vrex, i.e., $\gL_\irm=0$ and $\gL_\vrex=0$, and with a minimal $\gL_\erm$,
thereby eliciting the desired OOD solutions.
However, the ideal constraints might be too strong to be achieved when there are noises among invariant features and labels~\citep{dro_relax,cs_irm}, which will future enlarge the set of Pareto optimal solutions.
Therefore, it is natural to relax the constraints as $\gL_\irm\leq \epsilon_\irm$ and $\gL_\vrex\leq \epsilon_\vrex$.
When $\epsilon_\irm\rightarrow 0,\epsilon_\vrex\rightarrow 0$, it recovers the ideal invariance.
To obtain a desired solution under these circumstances, the optimization process is expected to meet the following two necessities:
\begin{enumerate}[label=(\roman*).,wide]
	\item The additional objective in \irmx can make the Pareto front more complicated such that the desired solutions are more likely to appear in the non-convex part, which are however not reachable by the linear weighting scheme~\citep{bv_convex}.
	      Therefore, the optimizer needs to be able to reach any Pareto optimal solutions in the front, e.g., MGDA algorithms~\citep{mgda}.\footnote{We leave more sophisticated Pareto front exploration methods~\citep{nonlinear_scalar,pareto_exp_mtl} to future investigation.}

	\item When both $\epsilon_\irm, \epsilon_\vrex> 0$, there can be multiple Pareto optimal solutions while there are few desired OOD solutions. Hence a preference of ERM and OOD objectives is usually needed.
	      As the optimality of each OOD objective usually appears as a necessary condition for satisfactory OOD performance, the preferences for OOD objectives are expected to be higher.

\end{enumerate}

Given the two requirements, we leverage a preference-aware MOO solver to solve \irmx for the desired Pareto optimal solution~\citep{epo}.
We summarize the overall solution as \pairfull (\pair).
When assigning a high preference to $\gL_\irm$ and $\gL_\vrex$ in \irmx (Eq.~\ref{CH:PAIR:eq:irmx_moo}),
\pair approaches a Pareto optimal solution that minimizes the OOD losses while not sacrificing the ERM performance too much, and has good OOD performance, shown as in Table.~\ref{CH:PAIR:tab:cmnist}.

\begin{figure}[t]
	\subfigure[Ground truth.]{
		\includegraphics[width=0.22\textwidth]{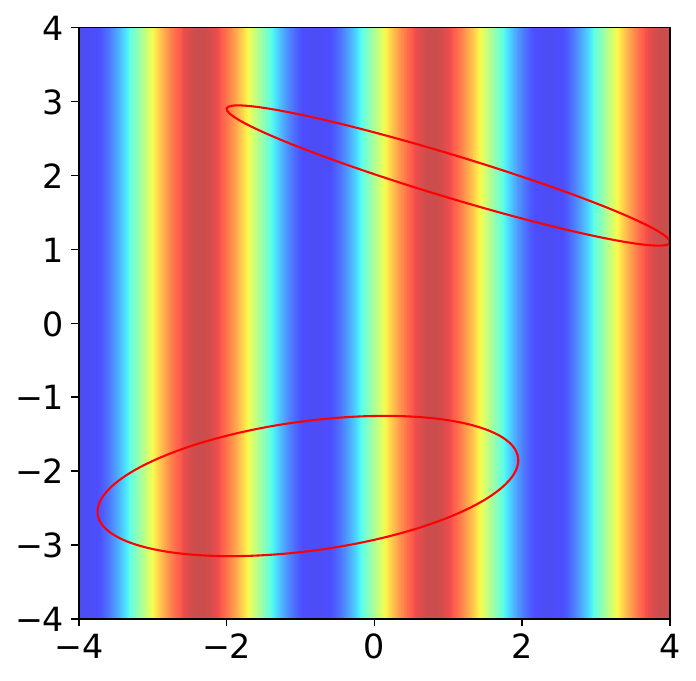}
		\label{CH:PAIR:fig:linextra_ground}
	}
	\subfigure[\irml.]{
		\includegraphics[width=0.22\textwidth]{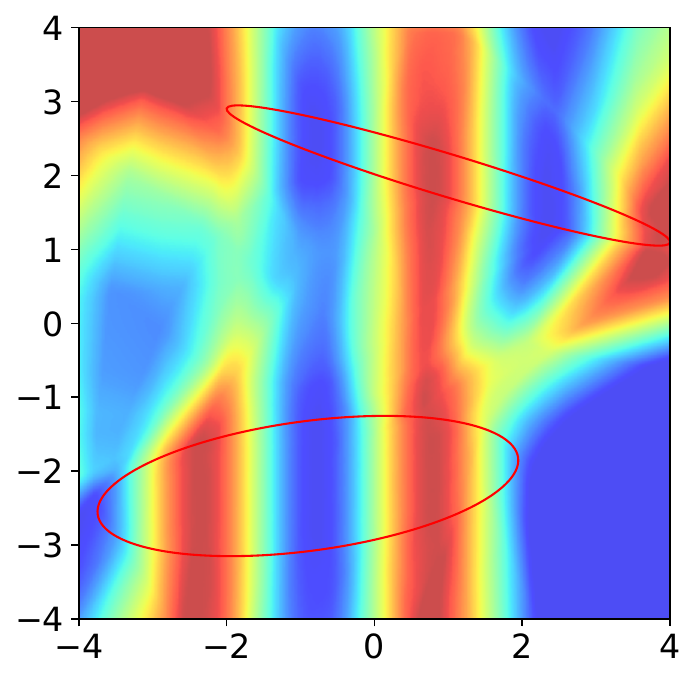}
		\label{CH:PAIR:fig:linextra_irm}
	}
	\subfigure[VREx.]{
		\includegraphics[width=0.22\textwidth]{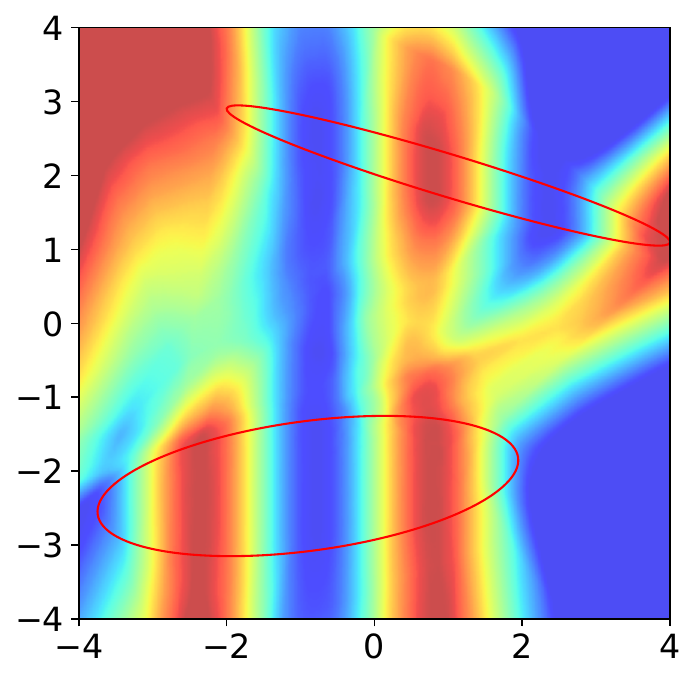}
		\label{CH:PAIR:fig:linextra_vrex}
	}
	\subfigure[\pair.]{
		\includegraphics[width=0.22\textwidth]{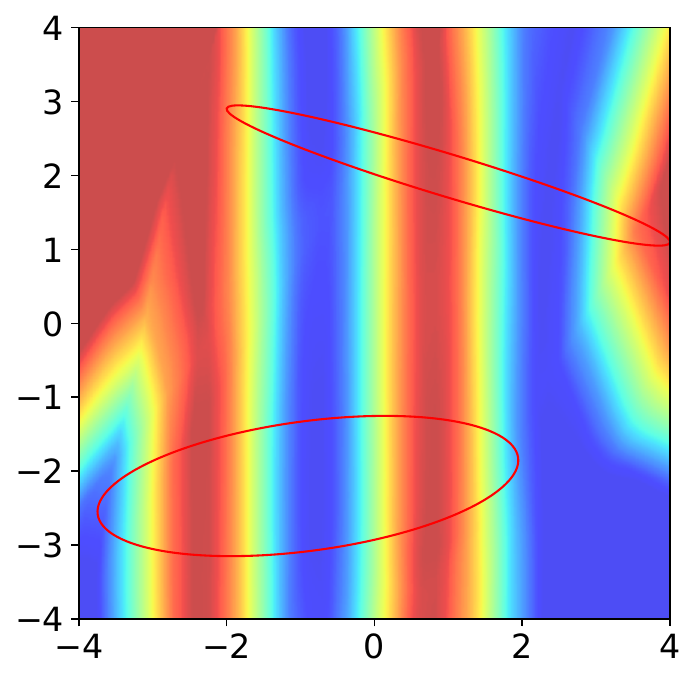}
		\label{CH:PAIR:fig:linextra_pair}
	}
	\caption[Recovery of causal invariance.]{Recovery of causal invariance. The causal invariance (Definition.~\ref{def:causal_inv}) requires the model predictions to be independent of the spurious features within the overlapped invariant features. In this example, intuitively it requires the colored belts to be perpendicular to $x$-axis within $[-2,2]$. It can be found that \pair succeeds out of \irml and \vrex in recovering the causal invariance.}
	\label{CH:PAIR:fig:extrapolation}
\end{figure}

\subsection{Recovery of Causal Invariance}
\label{CH:PAIR:sec:causal_extrapolate}
To better understand how \pair bridges the gaps between the practical and original \irm objectives, we examine to what extent \pair can recover the causal invariance specified by~\citet{irmv1} in a more difficult case. More formally, the causal invariance is defined as follows.
\begin{definition}(Causal Invariance)\label{def:causal_inv}
	Given a predictor $f:=w\circ\varphi$, the representation produced by the featurizer $\varphi$ is invariant over $\envall$ if and only if for all $e_1,e_2\in\envall$, it holds that
	\[
		\mathbb{E}_{\gD_{e_1}}[Y|\varphi(X)=z]=\mathbb{E}_{\gD_{e_2}}[Y|\varphi(X)=z],
	\]
	for all $z\in\gZ_\varphi^{e_1}\cap\gZ_\varphi^{e_2}$,
	where
	$\gZ_\varphi^e:=\{\varphi(X)|(X,Y)\in\text{supp}(\gD_e)\}$.
\end{definition}

Following Definition~\ref{def:causal_inv}, we construct a regression problem. As shown in Fig.~\ref{CH:PAIR:fig:extrapolation}, $Y =\sin(X_1)+1$ is solely determined by $X_1$, i.e., the values of the $x$-axis, while $X_2$ is the values of $y$-axis and does not influence the values of $Y$.
Different colors indicate different values of $Y$.
In this problem, the invariant representation $\varphi$ should only take $X_1$ and discard $X_2$.
We sampled two training environments as denoted by the ellipsoids colored in red, among which the overlapped region of the invariant features $X_1$ is $[-2,2]$. Hence the prediction produced by the invariant predictor following Definition~\ref{def:causal_inv} is expected to be independent of $X_2$. In other words, the plotted belts need to be perpendicular to the $x$-axis within the overlapped invariant features $[-2,2]$. More details can be found in Appendix~\ref{CH:PAIR:sec:causal_extrapolate_appdx}.

We plot predictions with the best MSE losses of \irml and \vrex in Fig.~\ref{CH:PAIR:fig:linextra_irm} and Fig.~\ref{CH:PAIR:fig:linextra_vrex}, respectively.
Although both \irml and \vrex fail to achieve the causal invariance as expected, perhaps surprisingly, \pair almost recovers the causal invariance, as shown in Fig.~\ref{CH:PAIR:fig:linextra_pair}.

\section{Pareto Invariant Risk Minimization}
\label{CH:PAIR:sec:pair_solution}
The success of \pair in empowering unrobust \irml to achieve the causal invariance of \irm demonstrates the significance of considering the trade-offs between ERM and OOD objectives in the optimization.
In the next, we will summarize our findings and elaborate \pair in more details.
\subsection{Methodology Outcomes}
\label{CH:PAIR:sec:pair_method}
\textbf{Key takeaways from the \irm example.}
To summarize, the failures of OOD optimization can be attributed to:
i) Using unrobust objectives for optimization; ii) Using unreliable schemes to approach the desired solution.
Nevertheless, we can improve the robustness of the OOD objectives by introducing additional guidance such that the desired solution is relocated in the Pareto front w.r.t. the new objectives.
After obtaining robust objectives to optimize, we then leverage a preference-aware MOO solver to find the Pareto optimal solutions that maximally satisfy the invariance constraints by assigning the OOD objective a higher preference while being aware of retaining ERM performance.

More formally, let $f_\ood$ be the desired OOD solution and $\gF$ be the functional class of $f_\ood$, a group of OOD objectives $\vL_\ood=\{\gL_\ood^i\}_{i=1}^m$ are robust if their composite objective $\vL_\ood$ satisfies that
\begin{equation}\label{CH:PAIR:eq:robust_obj}
	\vL_\ood(f_\ood) \preceq \vL_\ood (f), \forall f\neq f_\ood\in\gF,
\end{equation}
When given a robust OOD objective $\vL_\ood$, our target is to solve the following MOO problem
\begin{equation}
	\label{CH:PAIR:eq:pair_moo}
	\text{$\min$}_f (\gL_\erm,\vL_\ood)^T,
\end{equation}
where $\vL_\ood$ corresponds to an $\bm{\epsilon}_\ood$-relaxed invariance constraint as $\vL_\ood(f_\ood)=\bm{\epsilon}_\ood\preceq\vL_\ood(f),\forall f\neq f_\ood\in\gF$.
Denote the $\epsilon_\inv$ as empirical loss of using the underlying invariant features to predict labels, then the optimal values of the desired OOD solution w.r.t. Eq.~\ref{CH:PAIR:eq:pair_moo} are $(\epsilon_\inv,\bm{\epsilon}_\ood)^T=(\gL_\erm(f_\ood),\vL_\ood(f_\ood))^T$, which corresponds to an ideal preference (or OOD preference) for the objectives, that is $\vp_\ood=(\epsilon^{-1}_\inv,\bm{\epsilon}^{-1}_\ood)^T$.
The optimal solutions of Eq.~\ref{CH:PAIR:eq:pair_moo} that satisfy the exact Pareto optimality, i.e.,$\text{$\vp_\ood$}_i\gL_i=\text{$\vp_\ood$}_j\gL_j,\forall \gL_i,\gL_j \in\vL$, are expected to recover $f_\ood$ in Eq.~\ref{CH:PAIR:eq:robust_obj}.

\textbf{\pairo as an optimizer for OOD generalization.}
To find a desired Pareto optimal solution specified by $\vp_\ood$,
we adopt a 2-stage optimization scheme, which consists of two phases, i.e., the ``descent'' and the ``balance'' phase, following the common practice~\citep{domainbed}.

In the ``descent'' phase, we train the model with the ERM loss such that it approaches the Pareto front by merely minimizing $\gL_\erm$ first.
Then, in the ``balance'' phase, we adjust the solution to maximally satisfy the exact Pareto optimality specified by $\vp_\ood$.
We adopt the off-the-shelf preference-aware MOO solver EPO~\citep{epo} to find the desired Pareto optimal solutions with the given $\vp_\ood$.
Specifically, at each step, $\vp_\ood$ implies a descent direction $\vg_b$ that maximally increases the satisfaction to the exact Pareto optimality.
Then, we will find an objective weight vector to reweight both the ERM and OOD objectives (thus their gradients), such that the reweighted descent direction $\vg_\text{dsc}$ has a maximum angle with $\vg_b$.
Meanwhile, to avoid divergence, $\vg_\text{dsc}$ also needs to guarantee that it has a positive angle with the objective that diverges from the preferred direction most.
We provide detailed descriptions and theoretical discussions of the algorithm in Appendix~\ref{CH:PAIR:sec:pair_optimizer_appdx}.

\textbf{\pairs for OOD model selection.}
Model selection in OOD generalization is known to be challenging, as the validation data used to evaluate the model performance is no longer necessarily identically distributed to the test data~\citep{domainbed}.
The IRM example also implies that the traditional model selection methods that merely depend on the validation performance, i.e., the ERM performance, can easily compromise OOD performance due to the conflicts with the ERM objective, especially when the validation set has a large gap between the test set (cf. CMNIST in Table~\ref{CH:PAIR:tab:dobed_select}).

When given no additional assumption, we posit that the OOD loss values can serve as a proxy for OOD performance, which essentially corresponds to the \emph{underlying prior} assumed in the OOD methods.
It naturally resembles \pair optimization and therefore motivates \pairs.
\pairs jointly considers and trades off the ERM and OOD performance in model selection, and select models that maximally satisfy the exact Pareto optimality. We leave more details and discussions in Appendix~\ref{CH:PAIR:sec:pair_selection_appdx}.

\subsection{Theoretical Discussions and Practical Considerations}
\label{CH:PAIR:sec:pair_discussion}
Essentially both \pairo and \pairs aim to solve Eq.~\ref{CH:PAIR:eq:pair_moo} up to the exact Pareto optimality. However, in practice, the ideal preference is usually unknown and the exact Pareto optimality could be too strict to achieve.
Therefore,
we develop an $\epsilon$-approximated formulation of Eq.~\ref{CH:PAIR:eq:pair_moo}, i.e.,$|\text{$\vp_\ood$}_i\gL_i-\text{$\vp_\ood$}_j\gL_j|\leq \epsilon,\forall \gL_i,\gL_j \in\vL$, which might be of independent interest.
Built upon the relaxed variant, we analyze the OOD performance of \pair in terms of sample complexity, given the empirical risk and imprecise OOD preference, and prove the following Theorem in Appendix~\ref{proof:pair_theory_appdx}.

\begin{theorem}(Informal)\label{CH:PAIR:thm:pair_theory}
	For $\gamma\in(0,1)$ and any $\epsilon,\delta>0$, if $\gF$ is a finite hypothesis class, both ERM and OOD losses are bounded above, let $I_\pair$ be the index of all  losses, $p_{\max} \coloneqq \max_{i\in I_\pair} {p_i}$ and $L_\textup{max} \coloneqq \max_{i\in I_\pair}{L_i}$, if the number of training samples 
 \[\abs{D} \geq (32L_\textup{max}^2p_{\max}^2/\delta^2)\log[{2(m+1)\abs{\mathcal{F}}/\gamma},]\],\ then with probability at least $1 - \gamma$,
	\pairo and \pairs yield an $\epsilon$-approximated solution of $f_\ood$.
\end{theorem}

\textbf{Practical Considerations.} Theorem~\ref{CH:PAIR:thm:pair_theory} establishes the theoretical guarantee of \pairo and \pairs given only an imprecise OOD preference.
Empirically, we find that assigning a large enough preference to the OOD objectives is generally sufficient for \pairo to find a desired OOD solution.
For example, in most experiments \pairo yields a satisfactory OOD solution
with a relative preference of $(1,1e10,1e12)$ for \erm, \irml, and \vrex.
For \pairs, we can estimate the empirical upper bounds of ($\epsilon_\inv$, $\bm{\epsilon}_\ood$) from the running history and adjust OOD preference to be slightly larger.
We provide a detailed discussion on the preference choice in practice in Appendix~\ref{CH:PAIR:sec:pair_discussion_pc_appdx}.

Besides, the requirement of whole network gradients in \pairo can be a bottleneck when deployed to models that have a prohibitively large number of parameters~\citep{mtl_moo}.
To this end, we can use only the gradients of classifier $w$ to solve for the objective weights, or freeze the featurizer after the ``descent'' phase to further reduce the resource requirement~\citep{rfc}.
We discuss more practical options and how \pair can be applied to other OOD methods in Appendix~\ref{CH:PAIR:sec:pair_discussion_appdx}.

\section{Experiments}
\label{CH:PAIR:sec:experiments}
We conduct extensive experiments on \cmnist, \wilds, and \dobed to verify the effectiveness of \pairo and \pairs in finding a better OOD solution under objective conflicts.

\begin{table}[t]%
	\centering
	\small
	\caption{\small OOD Performance of \pair on \cmnist.}
	\label{CH:PAIR:tab:cmnist}
	\resizebox{0.45\columnwidth}{!}{
		\begin{tabular}{ l r  r r}
			\toprule
			Method                           & CMNIST                  & CMNIST-m               & Avg.            \\
			\midrule
			ERM                              & $17.1 \pm 0.9$          & $73.3 \pm 0.9$         & $45.2$          \\
			IRMv1                            & $67.3 \pm 1.9$          & $76.8 \pm 3.2$         & $72.1$          \\
			V-REx                            & $68.6 \pm 0.7$          & $82.9 \pm 1.3$         & $75.8$          \\
			\irmx                            & $65.8 \pm 2.9$          & $81.6 \pm 2.0$         & $73.7$          \\\midrule
			\textbf{$\text{\pairo}_f$}       & $68.6 \pm 0.9$          & $\mathbf{83.7}\pm 1.2$ & $76.2$          \\
			\textbf{$\text{\pairo}_\varphi$} & $68.6 \pm 0.8$          & $\mathbf{83.7}\pm 1.2$ & $76.2$          \\
			\textbf{$\text{\pairo}_w$}       & $\mathbf{69.2 \pm 0.7}$ & $\mathbf{83.7}\pm 1.2$ & $\mathbf{76.5}$ \\
			\midrule
			Oracle \!                        & $72.2 \pm 0.2$          & $86.5 \pm 0.3$         & $79.4$          \\
			Optimum                          & $75$                    & $90$                   & $82.5$          \\
			Chance                           & $50$                    & $50$                   & $50$            \\
			\bottomrule
		\end{tabular}}
\end{table}

\textbf{Proof of concept on \cmnist.}
In Table~\ref{CH:PAIR:tab:cmnist}, we compare \pairo implemented with \irmx to other strong baselines on \cmnist (CMNIST) and the failure case variant~\citep{irm_aistats} (CMNIST-m). We follow the evaluation setup as in IRM~\citep{irmv1} and report the results from $10$ runs. We assign a relative preference $(1,1e10,1e12)$ to \erm, \irml and \vrex objectives, respectively. It can be found that \pairo significantly improves over \irml across all environment settings, while \irmx using the linear weighting scheme performs worse than \pairo, confirming the effectiveness of \pairo.
Interestingly, using only the gradients of the classifier $w$ in \pairo can yield competitive performance as that uses $f$ or $\varphi$, while the former has better scalability. Therefore, we will use $\text{\pairo}_w$ in the following experiments.
More details are given in Appendix~\ref{CH:PAIR:sec:cmnist_appdx}.

\bgroup
\def\arraystretch{1.1}
\begin{table}[ht]
    \centering
    \caption{OOD generalization performances with \pair on \wilds benchmark.}
    \label{CH:PAIR:tab:wilds_results}
    \resizebox{\textwidth}{!}{
        \small
        \begin{tabular}{@{}{l}*{7}{c}@{}}    \toprule
                                    & {
            \textsc{Camelyon17}}    & {
            \textsc{CivilComments}} & {
            \textsc{FMoW}}          & {
            \textsc{iWildCam}}      & {
            \textsc{PovertyMap}}    & {
            \textsc{RxRx1}}         & {\multirow{2.5}{*}{\scriptsize{\textsc{Avg. Rank}($\downarrow$)$^\dagger$}}
            }                                                                                                                                                                                                                                                                        \\
            \cmidrule(lr){2-2} \cmidrule(lr){3-3} \cmidrule(lr){4-4} \cmidrule(lr){5-5} \cmidrule(lr){6-6} \cmidrule(lr){7-7}
                                    & Avg. acc. (\%)                                                              & Worst acc. (\%)             & Worst acc. (\%)            & Macro F1                  & Worst Pearson r            & Avg. acc. (\%)                               \\
            \midrule
            ERM                     & $70.3$ \std{6.4}                                                            & $56.0$ \std{3.6}            & $32.3$ \std{1.25}          & $30.8$ \std{1.3}          & $0.45$ \std{0.06}          & $29.9$  \std{0.4}          & $4.50$          \\\hdashline[0.5pt/1pt]
            \rule{0pt}{10pt}CORAL   & $59.5$ \std{7.7}                                                            & $65.6$ \std{1.3}            & $31.7$ \std{1.24}          & $\mathbf{32.7}$ \std{0.2} & $0.44$ \std{0.07}          & $28.4$  \std{0.3}          & $5.50$          \\
            GroupDRO                & $68.4$ \std{7.3}                                                            & $70.0$ \std{2.0}            & $30.8$ \std{0.81}          & $23.8$ \std{2.0}          & $0.39$ \std{0.06}          & $23.0$  \std{0.3}          & $6.83$          \\
            IRMv1                   & $64.2$ \std{8.1}                                                            & $66.3$ \std{2.1}            & $30.0$ \std{1.37}          & $15.1$ \std{4.9}          & $0.43$ \std{0.07}          & $8.2 $ \std{0.8}           & $7.67$          \\
            V-REx                   & $71.5$ \std{8.3}                                                            & $64.9$ \std{1.2}            & $27.2$ \std{0.78}          & $27.6$ \std{0.7}          & $0.40$ \std{0.06}          & $7.5 $ \std{0.8}           & $7.00$          \\
            Fish                    & $74.3$ \std{7.7}                                                            & $73.9$ \std{0.2}            & $34.6$ \std{0.51}          & $24.8$ \std{0.7}          & $0.43$ \std{0.05}          & $10.1$  \std{1.5}          & $4.33$          \\
            LISA                    & $\mathbf{74.7}$ \std{6.1}                                                   & $70.8$ \std{1.0}            & $33.5$ \std{0.70}          & $24.0$ \std{0.5}          & $\mathbf{0.48}$ \std{0.07} & $\mathbf{31.9}$  \std{0.8} & $2.67$          \\\hdashline[0.5pt/1pt]
            \rule{0pt}{10pt}\irmx   & $67.0$ \std{6.6}                                                            & $74.3$ \std{0.8}            & $33.7$ \std{0.78}          & $26.6$ \std{0.9}          & $0.45$ \std{0.04}          & $28.7$  \std{0.2}          & $4.00$          \\
            \textbf{\pairo}         & $74.0$ \std{7.0}                                                            & {$\mathbf{75.2}$ \std{0.7}} & $\mathbf{35.5}$ \std{1.13} & $27.9$ \std{0.7}          & $0.47$ \std{0.06}          & $28.8$  \std{0.1}          & $\mathbf{2.17}$ \\
            \bottomrule
            \multicolumn{8}{l}{\rule{0pt}{8pt}$^\dagger$\text{\normalfont \small Averaged rank is reported because of the dataset heterogeneity. A lower rank is better.}  }
        \end{tabular}
    }
\end{table}
\egroup

\textbf{Can \pairo effectively find better OOD solutions under realistic distribution shifts?}
We evaluate \pairo implemented with \irmx on $6$ challenging datasets from \wilds benchmark~\citep{wilds}, and compare \pairo with other state-of-the-art OOD methods from different lines (Sec.~\ref{CH:PAIR:sec:related_work}), including CORAL~\citep{CORAL}, GroupDRO~\citep{groupdro}, IRM~\citep{irmv1}, V-REx~\citep{vrex}, Fish~\citep{fish} and an advanced importance-aware data augmentation method LISA~\citep{lisa}.
By default, we assign a relative preference  $(1,1e10,1e12)$ to \erm, \irml and \vrex objectives, respectively, and restrict the search space of the preference.
Our implementation and evaluation protocol follow the exact configuration as previous works~\citep{wilds,fish,lisa}. Details can be found in Appendix~\ref{CH:PAIR:sec:wilds_appdx}.

Table~\ref{CH:PAIR:tab:wilds_results} shows that \pairo substantially improves over \irml as well as \irmx and yields top-ranking OOD performance among all state-of-the-art methods across different realistic distribution shifts, demonstrating the effectiveness and significance of resolving the optimization dilemma in OOD generalization.
Besides, the advances of \pair over \irmx also confirm the effectiveness of \pairo in finding a better trade-off between ERM and OOD objectives.

\begin{figure}[t]
	\subfigure[\pair v.s. \irmx.]{
		\includegraphics[width=0.21\textwidth]{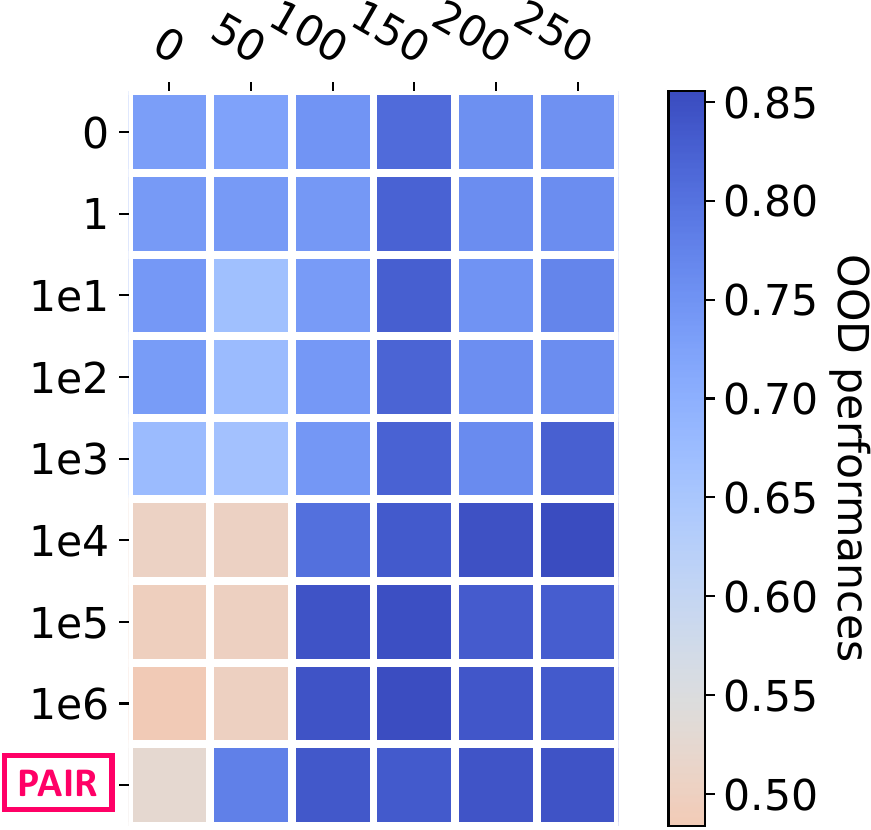}
		\label{CH:PAIR:fig:scalar}
	}
	\subfigure[Penalty trajectory.]{
		\includegraphics[width=0.24\textwidth]{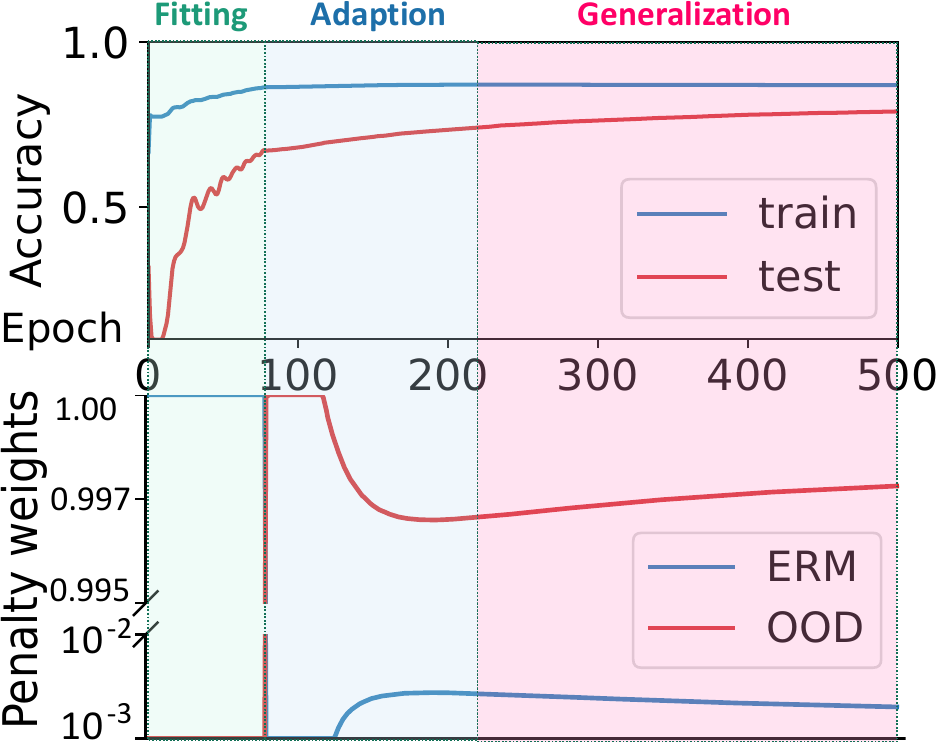}
		\label{CH:PAIR:fig:trajectory}
	}
	\subfigure[Normalized losses.]{
		\includegraphics[width=0.22\textwidth]{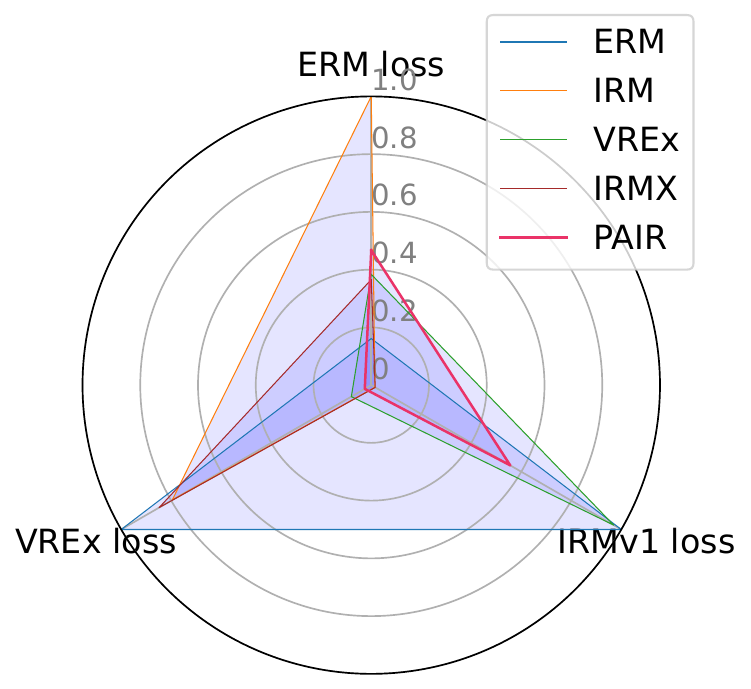}
		\label{CH:PAIR:fig:loss_radar}
	}
	\subfigure[Preference sensitivity.]{
		\includegraphics[width=0.21\textwidth]{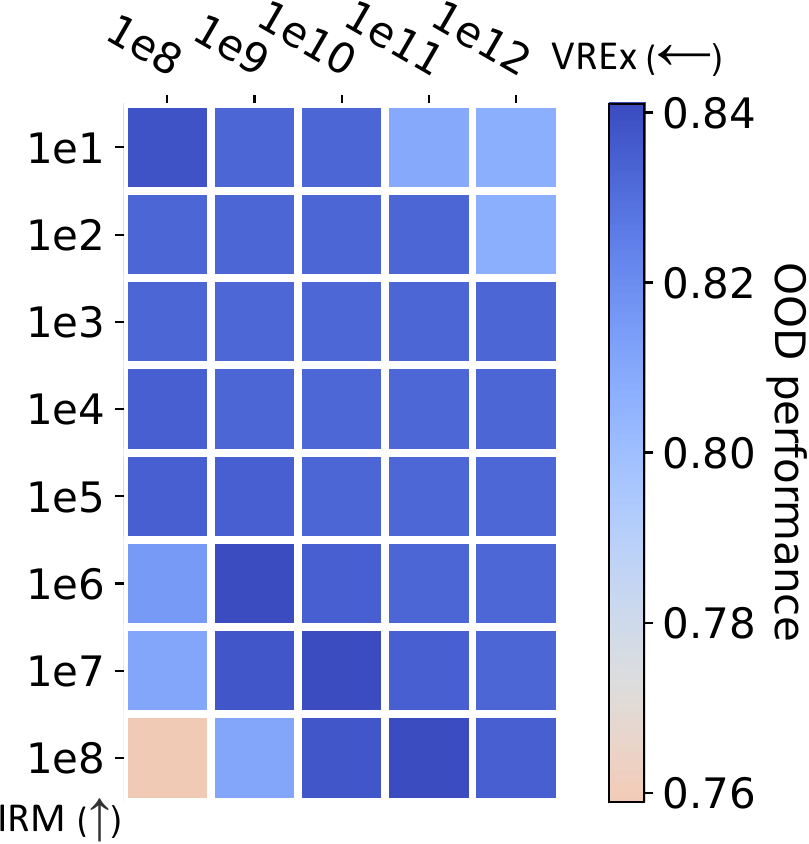}
		\label{CH:PAIR:fig:pc_sensitivity}
	}
	\caption[Ablation study of \pair.]{(a) Each point is the \emph{best} performed \irmx among corresponding pretraining epoch ($x$-axis), the \irml penalty weights ($y$-axis) and \emph{all} possible \vrex penalty weights. Despite the substantial tunning efforts, \irmx performs no better than \pair. That is because (b) \pair can adaptively adjust the penalty weights during the optimization process, and leads to a (c) Pareto optimal solution. (d) The robustness of \pairo to different preference choices enables it adaptable to various scenarios.}
	\label{CH:PAIR:fig:ablation_exp}
\end{figure}

\textbf{How can \pairo mitigate the objective conflicts?}
We conduct ablation studies with the modified \cmnist (More details and results are given in Appendix~\ref{CH:PAIR:sec:ablation_appdx}).
First, as shown in Fig.~\ref{CH:PAIR:fig:scalar}, \pairo effectively finds a better solution than exhaustive tuning of penalty weights in \irmx.
That is because \pair can adaptively adjust the penalty weights (Fig.~\ref{CH:PAIR:fig:trajectory}), which leads to a Pareto optimal solution that has lower OOD losses while not compromising the ERM loss too much (Fig.~\ref{CH:PAIR:fig:loss_radar}).
The other reason is that, \pairo is generally robust to different choices of preference choices (Fig.~\ref{CH:PAIR:fig:pc_sensitivity}), which makes it adaptable to various scenarios, confirming our discussions in Sec.~\ref{CH:PAIR:sec:pair_discussion}.

\textbf{Can \pairs effectively select better OOD solutions under realistic distribution shifts?}
To verify the effectiveness of \pairs,
we apply \pairs to multiple representative OOD methods as discussed in Sec.~\ref{CH:PAIR:sec:related_work}, and examine whether \pairs can improve the model selections under rigorous hyperparameters tunning~\citep{domainbed} on \cmnist~\citep{irm_aistats}, \pacs~\citep{pacs} and \terra~\citep{camel_example}. Intuitively, models selected merely based on ERM performance tend to have a high preference or better performance on environments that have a similar distribution of the corresponding validation set, which will lead to higher variance of performances at different environments or a lower worst environment performance.
Hence we use training-domain validation accuracy for \cmnist and \terra, and test-domain validation accuracy for \pacs to validate the existence of this issue under different scenarios~\citep{trainval_issue}. More details and results are provided in Appendix~\ref{CH:PAIR:sec:dobed_full_appdx}.

\begin{table}[h]
	\caption{OOD generalization performances with \pair using \dobed evaluation protocol.}
	\label{CH:PAIR:tab:dobed_select}
	\resizebox{\textwidth}{!}{
		\small
		\begin{tabular}{@{}{l}*{15}{c}@{}}
			\toprule
			                         &                 & \multicolumn{4}{c}{{\small\cmnist$^\dagger$}} & \multicolumn{5}{c}{{\small \pacs$^\ddagger$}} & \multicolumn{5}{c}{{\small \terra$^\dagger$}}                                                                                                                                                                                                                         \\\cmidrule(lr){3-6}\cmidrule(lr){7-11}\cmidrule(lr){12-16}
			                         & \textbf{\pairs} & \textbf{+90\%}                                & \textbf{+80\%}                                & \textbf{10\%}                                 & \textbf{$\Delta$ wr.} & \textbf{A}      & \textbf{C}      & \textbf{P}      & \textbf{S}      & \textbf{$\Delta$ wr.} & \textbf{L100}   & \textbf{L38}    & \textbf{L43}    & \textbf{L46}    & \textbf{$\Delta$ wr.} \\ \midrule
			ERM                      &                 & $71.0$                                        & $\mathbf{73.4}$                               & $10.0$                                        &                       & $87.2$          & $79.5$          & $95.5$          & $76.9$          &                       & $46.7$          & $\mathbf{41.8}$ & $57.4$          & $39.7$          &                       \\  \hdashline[0.5pt/1pt]
			\rule{0pt}{10pt}DANN     &                 & $71.0$                                        & $\mathbf{73.4}$                               & $10.0$                                        &                       & $86.5$          & $79.9$          & $97.1$          & $75.3$          &                       & $46.1$          & $41.2$          & $56.7$          & $35.6$          &                       \\
			DANN                     & $\checkmark$    & $71.6$                                        & $73.3$                                        & $10.9$                                        & $+0.9$                & $87.0$          & $81.4$          & $96.8$          & $77.5$          & $+2.2$                & $43.1$          & $41.1$          & $55.2$          & $38.7$          & $+3.1$                \\\hdashline[0.5pt/1pt]
			\rule{0pt}{10pt}GroupDRO &                 & $72.6$                                        & $73.1$                                        & $9.9$                                         &                       & $87.7$          & $82.1$          & $98.0$          & $79.6$          &                       & $48.4$          & $40.3$          & $57.9$          & $40.0$          &                       \\
			GroupDRO                 & $\checkmark$    & $\mathbf{72.7}$                               & $73.2$                                        & $13.0$                                        & $+3.1$                & $86.7$          & $\mathbf{83.2}$ & $\mathbf{97.8}$ & $81.4$          & $+1.8$                & $48.4$          & $40.3$          & $57.9$          & $40.0$          & $+0.0$                \\\hdashline[0.5pt/1pt]
			\rule{0pt}{10pt}IRMv1    &                 & $72.3$                                        & $72.6$                                        & $9.9$                                         &                       & $82.3$          & $80.8$          & $95.8$          & $78.9$          &                       & $48.4$          & $35.6$          & $55.4$          & $40.1$          &                       \\
			IRMv1                    & $\checkmark$    & $67.4$                                        & $64.8$                                        & $\mathbf{24.2}$                               & $+14.3$               & $85.3$          & $81.7$          & $97.4$          & $79.7$          & $+0.8$                & $40.4$          & $38.3$          & $48.8$          & $37.0$          & $+1.4$                \\\hdashline[0.5pt/1pt]
			\rule{0pt}{10pt}Fishr    &                 & $72.2$                                        & $73.1$                                        & $9.9$                                         &                       & $\mathbf{88.4}$ & $82.2$          & $97.7$          & $81.6$          &                       & $49.2$          & $40.6$          & $57.9$          & $40.4$          &                       \\
			Fishr                    & $\checkmark$    & $69.1$                                        & $70.9$                                        & $22.6$                                        & $+12.7$               & $87.4$          & $82.6$          & $97.5$          & $\mathbf{82.2}$ & $+0.6$                & $\mathbf{51.0}$ & $40.7$          & $\mathbf{58.2}$ & $\mathbf{40.8}$ & $+0.3$                \\
			\bottomrule
			\multicolumn{15}{l}{\rule{0pt}{8pt}$^\dagger$\text{\normalfont \small Using the training domain validation accuracy.} $^\ddagger$\text{\normalfont \small Using the test domain validation accuracy.} }
		\end{tabular}}
\end{table}

Table~\ref{CH:PAIR:tab:dobed_select} shows that there is a high variance in the performances at different environments of the models selected only based on the validation accuracy. In contrast, by jointly considering and trading off the ERM and OOD performances in model selection, \pairs substantially mitigate the variance by improving the worst environment performance of all methods under all setups up to $10\%$. It could serve as strong evidence for the importance of considering ERM and OOD trade-offs.

\chapter{Feature Learning in Causal Invariance Learning} \label{CH:FeAT}

\section{Motivations}
Understanding feature learning in neural networks is crucial to understanding how they generalize to different data distributions~\citep{mlp,ib,sgd_fl1,simple_bias,understand_ensemble,understand_benign}.
Deep networks trained with empirical risk minimization (ERM) learn highly predictive features that generalize surprisingly well to in-distribution (ID) data~\citep{erm,dl_book}. However, ERM also tends to learn \emph{spurious} features or shortcuts such as image backgrounds~\citep{camel_example,shortcut_dl,covid19_application,adv_causal_lens} whose correlations with labels do not hold in the out-of-distribution (OOD) data, and suffers from serious performance degeneration~\citep{wilds}.
Therefore, it is widely believed that the reason for the OOD failures of deep networks is that ERM fails to learn the desired features that have \emph{invariant} correlations with labels across different distributions~\citep{camel_example}.

\begin{figure}[t]
	\centering
	\subfigure{
		\includegraphics[width=0.55\textwidth]{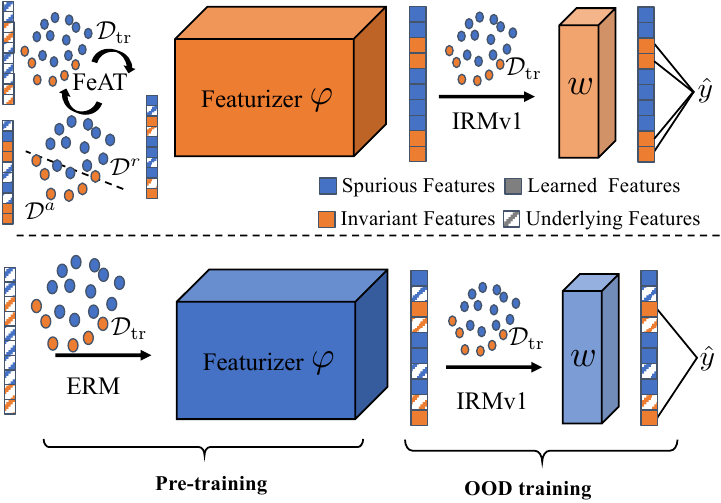}
		\label{CH:FeAT:fig:fat_illustration}
	}
	\subfigure{
		\includegraphics[width=0.4\textwidth]{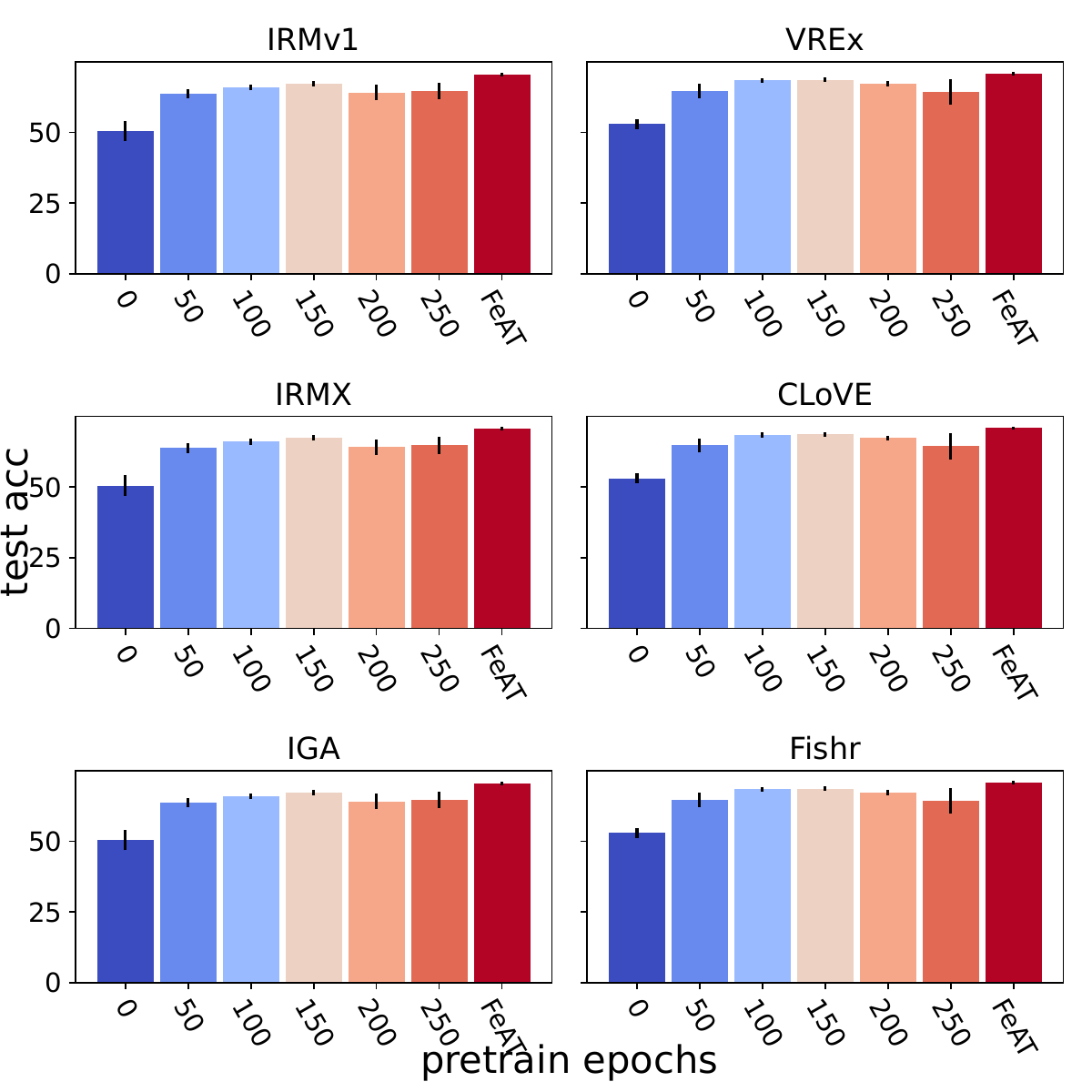}
		\label{CH:FeAT:fig:ood_sweep}
	}
	\caption[Illustration of \featfull (\feat).]{\textit{(a) An illustration of \feat (top row) compared to ERM (bottom row).}
		Different colors in samples denote the respective dominant features.
		As the original data is dominated by spurious features (blue), ERM tends to learn more spurious features but limited invariant features (orange).
		Thus the OOD training with \irml can only leverage limited invariant features and achieve limited performance.
		In contrast, iteratively, \feat divides $\train$ into augmentation $D^a$ and retention sets $D^r$ that contain features not learned and already learned by the current model at the round, respectively. In each round, \feat augments the model with new features contained in the growing augmentation sets while retaining the already learned features contained in the retention sets, which will lead the model to learn richer features for OOD training and obtain a better OOD performance.
		Then \feat augments the model with new features while retaining the already learned features, which leads to richer features for OOD training and better OOD performance.
		\textit{(b) OOD Performance vs. the number of ERM pre-training epochs in \cmnist-025.} The performance of various OOD objectives largely relies on the quality of ERM-learned features. When there exist underlying useful features poorly learned by ERM, the OOD performance will be limited. In contrast, \feat learns richer features with $2$ rounds (or $300$ epochs) and improves the OOD performance.}
	\label{CH:FeAT:fig:flood_phenon}
\end{figure}

However,
several recent works find that ERM-trained models have \emph{already learned sufficiently good features} that are able to generalize to OOD data~\citep{dare,dfr,dfrlearn}.
{In addition, when optimizing various penalty terms~\citep{causal_transfer,iga,andmask,vrex,sd,ib-irm,clove,fish,fishr,maple,ciga} that aim to regularize ERM to capture the invariant features (termed as OOD objectives)},
there also exists a curious phenomenon that the performance of OOD objectives largely relies on the pre-training with ERM before applying the OOD objectives~\citep{rfc,pair}.
As shown in Fig.~\ref{CH:FeAT:fig:ood_sweep},
the number of ERM pre-training epochs \emph{has a large influence} on the final OOD performance.
These seemingly contradicting phenomena raise a challenging research question:
\begin{myquotation}
	\emph{What features are learned by ERM and OOD objectives, respectively, and how do the learned features generalize to in-distribution and out-of-distribution data?}
\end{myquotation}

To answer the question,
we conduct a theoretical investigation of feature learning in a two-layer CNN network, when trained with ERM and a widely used OOD objective, \irml~\citep{irmv1}, respectively.
We use a variation of the data models proposed in~\cite{understand_ensemble,understand_benign},
and include features with different correlation degrees to the labels to simulate invariant and spurious features~\citep{risk_irm}.

First, we find that ERM essentially learns \emph{both} spurious features and invariant features (Theorem~\ref{CH:FeAT:thm:erm_learn_feat}).
The degrees of spurious and invariant feature learning are mostly controlled by their correlation strengths with labels.
Moreover, merely training with \irml \emph{cannot learn new} features (Theorem~\ref{CH:FeAT:thm:irmv1_not_learn}).
Therefore, the \emph{quality} of ERM feature learning affects the final OOD performance significantly.
Hence, as the number of ERM pre-training epochs increases, the model learns invariant features better and thus the final OOD performance will increase (Fig.~\ref{CH:FeAT:fig:flood_phenon}).
However, when ERM does not capture \emph{all} useful features for OOD generalization, i.e., there exist some useful features that are poorly learned by ERM, the model can hardly learn these features during OOD training and the OOD performance will be limited.
Given a limited number of pre-training steps, it could often happen due to low invariant correlation strength, the feature learning biases of ERM~\citep{simple_bias}, or the model architectures~\citep{what_shape}.
Consequently, ERM feature learning can be a \emph{bottleneck} to OOD generalization~\citep{imagenetv2}.

To remedy the issue, we propose  \featfull (\feat), an iterative strategy to enforce the model to learn richer features.
As shown in Fig.~\ref{CH:FeAT:fig:fat_illustration}, in each round, \feat separates the train set into two subsets according to whether the underlying features in each set are already learned (Retention set $\dataset^r$) or not (Augmentation set $\dataset^a$), by examining whether the model yields correct ($\dataset^r$) or incorrect ($\dataset^a$) predictions for samples from the subsets, respectively.
Intuitively, $\dataset^a$ and $\dataset^r$ will contain distinct features that are separated in different rounds.
Then, \feat performs distributionally robust optimization (DRO)~\citep{dro,rfc} on all subsets, which \emph{augments} the model to learn new features by minimizing the maximal ERM losses on all $\dataset^a$ and \emph{retains} the already learned features by minimizing ERM losses on all $\dataset^r$.
Along with the growth of the augmentation and retention sets, \feat is able to learn richer features for OOD training and obtain a better OOD performance.
\feat terminates when the model cannot learn any new predictive features (Algorithm~\ref{alg:fat}).

We conduct extensive experiments on both \cmnist~\citep{irmv1,pair} and $6$ datasets from the challenging benchmark, \wilds~\citep{wilds},
and show that \feat effectively learns richer features and thus consistently improves the OOD performance when applied to various OOD objectives (Sec.~\ref{CH:FeAT:sec:exp}).
\section{Related Work}%
We discuss the most related work to ours and leave more details in Appendix~\ref{CH:FeAT:sec:related_work_appdx}.

\textbf{On Feature Learning and Generalization.}
Understanding feature learning in deep networks is crucial to understanding their generalization~\citep{mlp,ib,sgd_fl1,sgd_fl2,understand_ensemble,understand_benign,huang2023graph}.
Beyond the empirical probing~\citep{saliency,new_saliency,what_shape,toy_model}, \citet{understand_ensemble} proposed a new theoretical framework for analyzing the feature learning process of deep networks, which has been widely adopted to study various deep learning phenomena~\citep{understand_ssl,understand_adam,understand_benign,huang2023graph}.
However, how the learned features from ID data can generalize to OOD data remains elusive. The only exceptions are \citep{understand_da} and \citep{feat_distort}. \citet{feat_distort} find fine-tuning can distort the pre-trained features while fine-tuning can be considered as a special case in our framework. \citet{understand_da} focus on how data augmentation helps promote good but hard-to-learn features and improve OOD generalization.
\citet{pde} finds neural networks tend to learn spurious features under imbalanced groups.
In contrast, we study the direct effects of ERM and OOD objectives to feature learning and provide a theoretical explanation for the curious phenomenon~\citep{dare,dfrlearn}.
To the best of our knowledge, we are the \textit{first} to analyze the feature learning of ERM and OOD objectives and their interactions in the general OOD generalization setting.

\textbf{Rich Feature Learning.}
Recently many OOD objectives have been proposed to regularize  ERM such that the model can focus on learning invariant features~\citep{irmv1,vrex,sd,clove,fishr}.
However, the final OOD performance has a large dependence on the number of ERM pre-training epochs~\citep{rfc,pair}.
To remedy the issue, \citet{rfc} proposed \rfc to construct rich feature representations as network initialization for OOD training. Although both \rfc and \feat perform DRO on grouped subsets, \rfc rely on multiple initializations of the whole network to capture diverse features from the subsets, and complicated ensembling of the features, which requires more training epochs for convergence. In contrast, \feat relieves the requirements via direct augmentation-retention on the grouped subsets, and thus obtains better performance.
More crucially, although rich feature learning algorithms such as \rfc and weight averaging~\citep{diwa,eoa} have gained some successes, explanations about the reliance of OOD performance on ERM pre-training and why rich feature learning mitigates the issue remain elusive. In addition to a new rich feature learning algorithm, our work provides theoretical explanations for the success of rich feature learning in OOD generalization.

\section{Preliminaries and Problem Definition}
\label{CH:FeAT:sec:prelim}
\textbf{Notations.} We use old-faced letters for vectors and matrices otherwise for scalar;
$\| \cdot \|_2$ to denote the Euclidean norm of a vector or the spectral norm of a matrix,
while $\| \cdot \|_F$ for the Frobenius norm of a matrix.
${\bf I}_d$ refers to the identity matrix in $\mathbb{R}^{d \times d}$.
Full details are deferred to Appendix~\ref{CH:FeAT:sec:notations_appdx}.

Our data model $\dataset=\{ \rvx_i, y_i \}_{i=1}^n$ is adapted from~\citep{understand_ensemble,understand_benign} and further characterizes each data point $\rvx_i$ as invariant and spurious feature patches from the two-bit model~\citep{risk_irm,pair}.
\begin{definition}\label{CH:FeAT:def:risk_irm}
	$\dataset=\{\dataset_e\}_{e\in\envall}$ is composed of multiple subsets $\dataset_e$ from different environments $e\in\envall$, where each $\dataset_e=\{(\rvx_i^e,y_i^e)\}_{i=1}^{n_e}$ is composed of i.i.d. samples $(\rvx_i^e,y_i^e)\sim \mathbb{P}^e$.
	Each data $(\rvx^e,y^e)\in\dataset_e$ with $\rvx^e \in \mathbb{R}^{2d}$ and $y^e \in \{-1 , 1 \}$ is generated as follows:
	\begin{enumerate}[label=(\alph*),leftmargin=*]
		\item Sample $y^e\in\{-1,1\}$ uniformly;\vspace{-0.02in}
		\item Given $y^e$, each input $\rvx^e=[ \rvx^e_1 , \rvx^e_2  ]$ contains a feature patch $\rvx_1$ and a noise patch $\rvx_2$, that are sampled as:\vspace{-0.02in}
		      \begin{equation*}
			      \begin{aligned}
				      \rvx_1 & = y \cdot \Rad(\alpha) \cdot  \rvv_1 + y \cdot \Rad(\beta) \cdot  \rvv_2  \quad
				      \rvx_2 = \boldsymbol{\xi}
			      \end{aligned}
		      \end{equation*}
		      where $\Rad(\delta)$ is a random variable taking value $-1$ with probability $\delta$ and $+1$  with probability $1-\delta$, $\rvv_1 = [1, 0,\ldots 0]^\top$ and $\rvv_2 = [0, 1, 0, \ldots 0]^\top$.\vspace{-0.02in}
		\item  A noise vector $\boldsymbol{\xi}$ is generated from the Gaussian distribution $\mathcal{N}(\mathbf{0}, \sigma_p^2 \cdot (\mathbf{I}_d - \rvv_1 \rvv_1^\top - \rvv_2 \rvv_2^\top  ))$
	\end{enumerate}
\end{definition}
Definition \ref{CH:FeAT:def:risk_irm} is inspired by the structure of image data in image classification with CNN~\citep{understand_ensemble}, where the inputs consist of different patches, some of the patches consist of features that are related to the class label of the image, and the others are noises that are irrelevant to the label. In particular, $\rvv_1$ and $\rvv_2$ are feature vectors that simulate the invariant and spurious features, respectively.
Although our data model focuses on two feature vectors,
the discussion and results can be further generalized to multiple invariant and spurious features with fine-grained characteristics~\citep{understand_da}.
Following previous works~\citep{understand_benign}, we assume that the noise patch is generated from the Gaussian distribution
such that the noise vector is orthogonal to the signal vector $\rvv$.
Each environment is denoted as $\envalpha\!=\!\{(\alpha,\beta_e):0<\beta_e<1\}$, where $\rvv_1$ is the invariant feature as $\alpha$ is fixed while $\rvv_2$ is the spurious feature as $\beta_e$ varies across $e$.

\paragraph{CNN model.} We consider training a two-layer convolutional neural network with a hidden layer width of $m$. The filters are applied to $\rvx_1$, $\rvx_2$, respectively,\footnote{When the environment $e$ is not explicitly considered, we will omit it for clarity.} and the second layer parameters of the network are fixed as $\frac{1}{m}$ and $-\frac{1}{m}$,
respectively. Then the network can be written as $f (\rmW, \rvx)\!=\!F_{+1}(\rmW_{+1}, \rvx) - F_{-1}(\rmW_{-1}, \rvx)$, where $F_{+1}(\rmW_{+1}, \rvx)$ and $F_{-1}(\rmW_{-1}, \rvx)$ are defined as follows:
\begin{equation}\label{CH:FeAT:eq:cnn}
	\begin{aligned}
		F_j(\rmW_j, \rvx ) = \frac{1}{m}\sum_{r=1}^m \left[  {\psi}(\rvw_{j,r}^\top \rvx_1) +  {\psi}(\rvw_{j,r}^\top \rvx_2) \right],
	\end{aligned}
\end{equation}
where $ {\psi(x)}$ is the activation function. We assume that all network weights are initialized as $\mathcal{N}(0,\sigma_0^2)$.

\paragraph{ERM objective.}
We train the CNN model by minimizing the empirical cross-entropy loss function:
\begin{equation}\label{CH:FeAT:eq:logistic_loss}
	\begin{aligned}
		\vspace{-0.05in}
		{L}(\rmW ) & = \sum_{e \in \mathcal{E}_{\mathrm{tr}}} \frac{1}{n_e}  \sum_{i=1}^{n_e} \ell( y^e_i \cdot f(\rmW, \rvx^e_i)),
	\end{aligned}
\end{equation}
where $\ell(z)\!=\!\log(1\!+\!\exp(-z))$ and $\{\dataset_e\}_{e\in\envtrain}\!=\!\{\{ \rvx_i^e, y_i^e \}_{i=1}^{n_e}\}_{e\in\envtrain}$ is the trainset with $\sum_{e\in\envtrain}n_e\!=\!n$.

\textbf{OOD objective.}
The goal of OOD generalization is,
given the data from training environments $\{\dataset_e\}_{e\in\envtrain}$,
to find a predictor $f:\gX\rightarrow\gY$
that generalizes well to all (unseen) environments, or minimizes
$\max_{e\in\envall}L_e(f)$, where $L_e$ is the empirical risk under environment $e$.
The predictor $f=w\circ\varphi$ is usually composed of a featurizer $\varphi:\gX\rightarrow\gZ$ that learns to extract useful features, and a classifier $w:\gZ\rightarrow\gY$ that makes predictions from the extracted features.

Since we are interested in cases where the OOD objective succeeds in learning the invariant features. In the discussion below, without loss of generality, we study one of the most widely discussed OOD objective, \irml objective, from IRM framework \cite{irmv1}, and the data model where \irml succeeds.
Specifically, the IRM framework approaches OOD generalization by finding an invariant representation $\varphi$,
such that there exists a classifier acting on $\varphi$ that is
simultaneously optimal in $\envtrain$.
Hence, IRM leads to a challenging bi-level optimization problem as
\begin{equation}%
	\label{CH:FeAT:eq:irm}
	\min_{w,\varphi}  \ \sum_{e\in\envtrain}L_e(w\circ\varphi),
	\text{s.t.}
	\ w\in\argmin_{\bar{w}:\gZ\rightarrow\gY} L_e(\bar{w}\circ\varphi),\ \forall e\in\envtrain.
\end{equation}

Due to the optimization difficulty of Eq.~(\ref{CH:FeAT:eq:irm}), \citet{irmv1} relax Eq.~(\ref{CH:FeAT:eq:irm}) into \irml as follows:
\begin{equation}%
	\label{CH:FeAT:eq:irml}
	\min_{\varphi}  \sum_{e\in\envtrain}L_e(\varphi)+\lambda|\nabla_{w|w=1}L_e(w\cdot\varphi)|^2.
\end{equation}
Given the convolutional neural network (Eq.~\ref{CH:FeAT:eq:cnn}) and logistic loss (Eq.~\ref{CH:FeAT:eq:logistic_loss}), \irml can be written as
\begin{equation}\label{CH:FeAT:eq:irml_cnn}
	\begin{aligned}
		L_{\irml}(\rmW) & =   \sum_{e \in \mathcal{E}_{\mathrm{tr}}} \frac{1}{n_e} \sum_{i=1}^{n_e} \ell \left( y_i^e \cdot f(\rmW, \rvx_i^e) \right)                           +   \sum_{e \in \mathcal{E}_{\mathrm{tr}}}  \frac{\lambda }{n_e^2 }  \left( \sum_{i=1}^{n_e}  {\ell_i'^e}  \cdot y_i^e \cdot f(\rmW, \rvx_i^e)  \right)^2,
	\end{aligned}
\end{equation}
where $ {\ell_i'^e} = \ell'( y_i^e \cdot f(\rmW, \rvx_i^e)) =-  \frac{ \exp(-y_i^e \cdot f(\rmW, \rvx_i^e))}{1 + \exp(-y_i^e \cdot f(\rmW, \rvx_i^e))} $.
Due to the complexity of \irml, in the analysis below, we introduce $C_\irml^e$ for the ease of expressions. Specifically, we define $C_\irml^e$ as
\[
	C_\irml^e \triangleq \frac{1}{n_e}\sum_{i=1}^{n_e} {\ell'\big(y^e_i \hat{y}^e_i\big) \cdot y^e_i \hat{y}^e_i},
\]
where $\hat{y}^e_i \triangleq f(\rmW, \rvx^e_i)$ is the logit of sample $\rvx_i$ from environment $e$. The convergence of $C_\irml$ indicates the convergence of \irml penalty. The following lemma will be useful in our analysis.

\begin{lemma}(\citet{understand_benign})
	Let $\rvw_{j,r}(t)$\footnote{We use $\rvw_{j,r}(t)$, $\rvw_{j,r}^{(t)}$ and $\rvw_{j,r}^t$ interchangeably.} for $j \in\{+1, -1 \}$ and $r \in \{1,2,\ldots,m\}$ be the convolution filters of the CNN at $t$-th iteration of gradient descent. Then there exists unique coefficients {$\gamma^{inv}_{j,r}(t), \gamma^{spu}_{j,r}(t) \ge 0 $} and $\rho_{j,r,i}(t)$ such that,\vspace{-0.1in}
	\begin{align}
		\rvw_{j,r}(t) & = \rvw_{j,r}(0) + j \cdot \gamma^{inv}_{j,r}(t) \cdot \rvv_1 + j \cdot \gamma^{spu}_{j,r}(t) \cdot  \rvv_2    + \sum_{i=1}^n \rho_{j,r,i}(t) \cdot \| \boldsymbol{\xi}_i \|^{-2}_2 \cdot \boldsymbol{\xi}_i. \label{CH:FeAT:eq:decompostion}
	\end{align}
\end{lemma}
We refer Eq.~(\ref{CH:FeAT:eq:decompostion}) as the \textit{signal-noise decomposition} of $\rvw_{j,r}{(t)}$~\citep{understand_benign}. We add normalization factor $\|\boldsymbol{\xi}_{i}\|_{2}^{-2}$ in the definition so that  $\rho_{j,r}^{(t)} \approx \langle \rvw_{j,r}^{(t)}, \boldsymbol{\xi}_{i} \rangle$. Note that $\| \rvv_1\|_2 = \| \rvv_2 \|_2 = 1 $, the corresponding normalization factors are thus neglected.
Furthermore,{$\gamma^{inv}_{j,r}\approx\langle\rvw_{j,r},\rvv_1\rangle$ and $\gamma^{spu}_{j,r}\approx\langle\rvw_{j,r},\rvv_2\rangle$} respectively denote the degrees of invariant and spurious feature learning. %

\section{Understanding Feature Learning in OOD Generalization}
\label{CH:FeAT:sec:understand}
\begin{figure}[t]
	\centering
	\subfigure[$C_\irml$, w/ PT]{\includegraphics[width=0.22\textwidth]{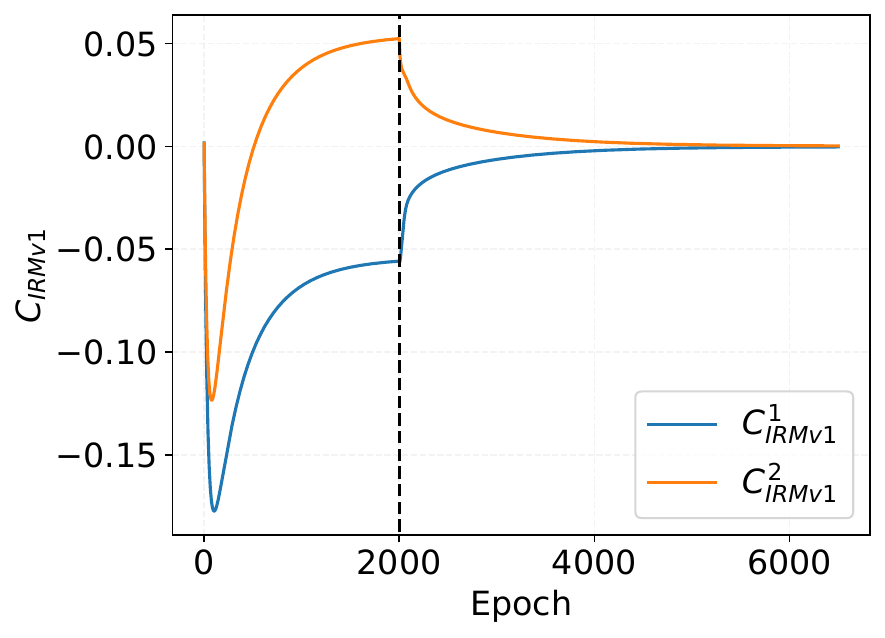}
		\label{f5a}
	}
	\subfigure[FL, w/ PT]{\includegraphics[width=0.248\textwidth]{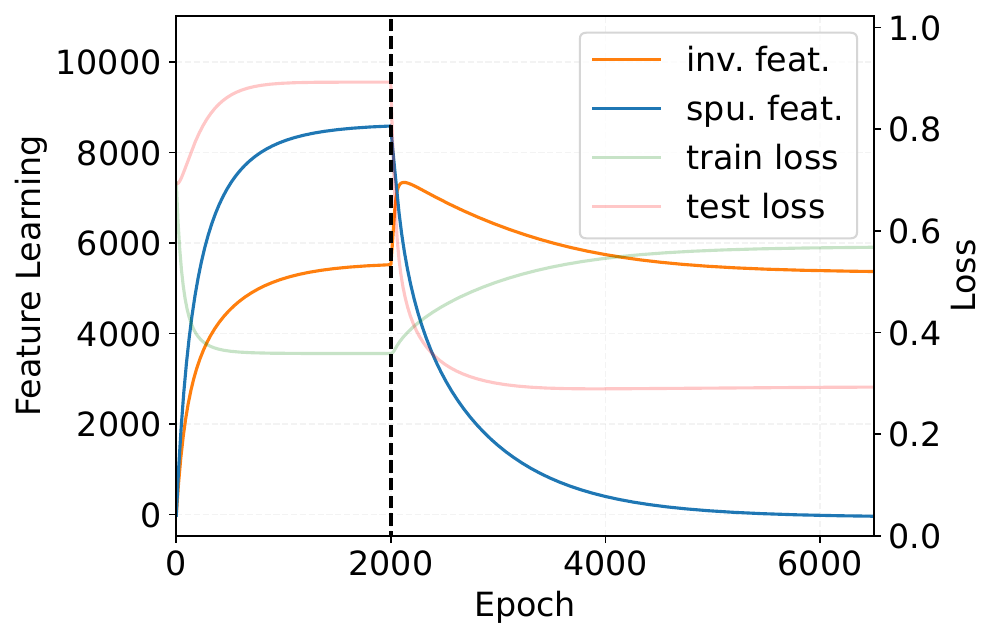}
		\label{f5b}}
	\subfigure[$C_\irml$, w/o PT]{\includegraphics[width=0.225\textwidth]{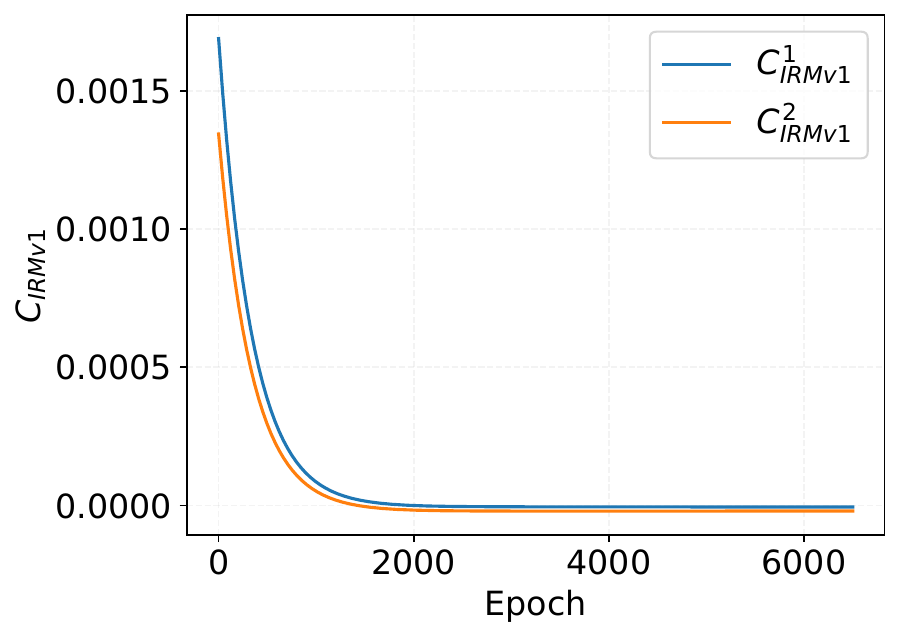} \label{f5c}}
	\subfigure[FL, w/o PT]{\includegraphics[width=0.25\textwidth]{
			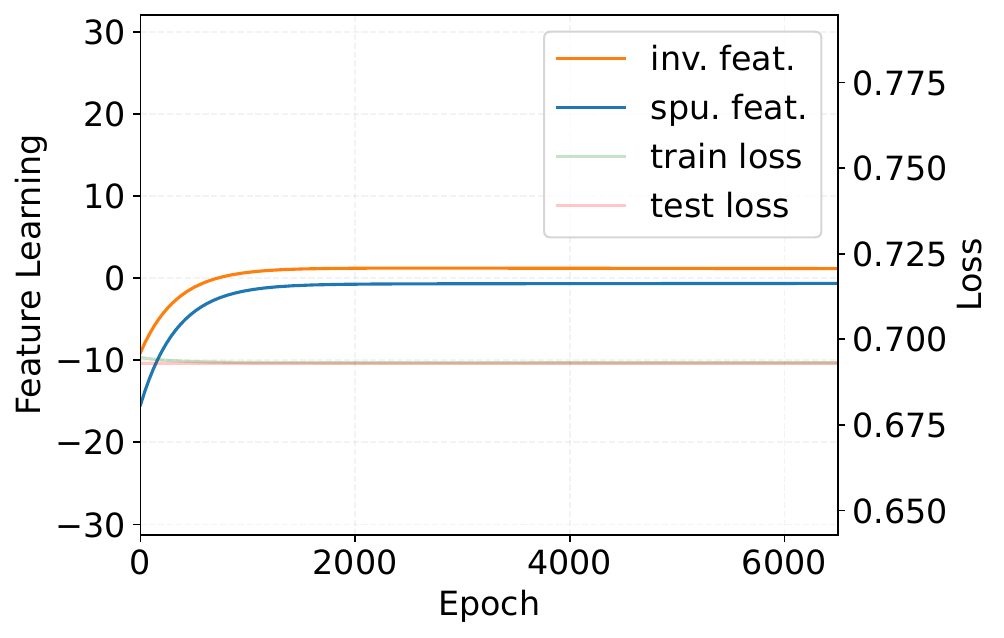} \label{f5d}}
	\caption[Analysis of feature learning with \erm and \irml.]{The convergences of $C_\irml$ and feature learning coefficients (FL) with or without ERM pre-training (PT).
		The invariant and spurious feature learning terms are the mean of $\langle \mathbf{w}_{j,r}, j\mathbf{v}_1 \rangle$
		and $\langle \mathbf{w}_{j,r}, j\mathbf{v}_2 \rangle$ for $j\in \lbrace\pm 1\rbrace, r\in[m]$, respectively.
		The training environments are $\mathcal{E}_{tr}= \{(0.25, 0.1), (0.25, 0.2) \}$. The black dashed line indicates the end of pre-training. More details are given in Appendix \ref{exp:CNN_synthetic}.}
	\label{f5}
\end{figure}
\subsection{ERM Feature Learning}
\label{CH:FeAT:sec:erm_learn_feat}

With the setup in Sec.~\ref{CH:FeAT:sec:prelim}, we first study the feature learning of the ERM objective. We consider a two training environments setup $\mathcal{E}_{tr} = \{(\alpha, \beta_1), (\alpha, \beta_2)\}$ where the signal of invariant feature is weaker than the average of  spurious signals (i.e., $\alpha > \frac{\beta_1 + \beta_2}{2}$), which corresponds to Figure \ref{f5}. For a precise characterization of the training dynamic, we adopted a minimal setup where $\psi(x) = x$ in Figure \ref{f5} and the following theorem, which already captures the key phenomenon in ERM feature learning. We study ERM feature learning with \textit{non-linear} activations in Appendix \ref{CH:FeAT:sec:erm_non_linear}.

\begin{theorem}\label{CH:FeAT:thm:erm_learn_feat} (Informal) For $\rho>0$, let $\underline{n} \triangleq \min_{e\in\mathcal{E}_{tr}}{n_e}$. Suppose that we run $T$ iterations of GD for the ERM objective. With sufficiently large $\underline{n}$ and $\psi(x)=x$,  assuming that (i) $\alpha, \beta_1, \beta_2<\frac{1}{2}$, and (ii) $\alpha > \frac{\beta_1 + \beta_2}{2}$,
	with properly chosen $\sigma_0^2$ and $\sigma_p^2$,
	there exists a constant $\eta$, such that with probability at least $1-2\rho$, both invariant and spurious features are converging and the increment of the spurious feature is larger than that of the invariant feature at any iteration $t\in\{0,\ldots,T-1\}$ (the detailed quantitative result of this gap can be found at (\ref{erm_eq7}) in Appendix \ref{CH:FeAT:sec:proof_erm_feat}).
\end{theorem}

As the formal statement of Theorem~\ref{CH:FeAT:thm:erm_learn_feat} is too complicated and lengthy, we leave it and its proof  in Appendix~\ref{CH:FeAT:sec:proof_erm_feat}, while giving an informal but more intuitive version here.
Theorem~\ref{CH:FeAT:thm:erm_learn_feat} states that ERM training learns both invariant feature and spurious feature at the same time, and if the average of  spurious signals is stronger, the coefficient of spurious feature learning will dominate that of invariant feature learning in the whole training process, corresponding to Figure \ref{f5b}. We establish the proof based on inspecting a novel recursive equation, which might be of independent interest. Note that Theorem \ref{CH:FeAT:thm:erm_learn_feat} can be directly generalized to handle any number of environments.

Speaking of implications, Theorem~\ref{CH:FeAT:thm:erm_learn_feat} provides answers to the seemingly contradicting phenomena that ERM fails in OOD generalization~\citep{camel_example,covid19_application}
but still learns the invariant features~\citep{dare,dfr,dfrlearn}.
In fact, ERM fails since it learns the spurious features more quickly, when spurious correlations are stronger than invariant correlations.
Nevertheless, invariant feature learning also happens, even when the spurious correlations are strong, so long as the invariant feature has a non-trivial correlation strength with the labels.
Therefore, simply re-training a classifier based on a subset of unbiased data on top of the ERM-trained featurizer achieves impressive OOD generalization performance~\citep{dare,dfr,dfrlearn}.
Theorem~\ref{CH:FeAT:thm:erm_learn_feat} also provides an explanation for the ID-OOD performance correlations when fine-tuning or training neural networks (especially large pre-trained models like CLIP~\citep{clip}, GPT~\citep{gpt3})~\citep{assaying_transfer,effective_robustness,robust_finetune,forgetting}. We provide a detailed discussion in Appendix~\ref{CH:FeAT:sec:related_work_appdx}.

\subsection{IRM Feature Learning}
\label{CH:FeAT:sec:ood_not_learn}
Although Theorem~\ref{CH:FeAT:thm:erm_learn_feat} states that ERM learns both invariant and spurious features,
the following questions remain unanswered: (1) whether \irml learns new features or simply amplifies the already learned ERM features, and (2) how the quality of the ERM-learned features affects the feature learning when \irml is incorporated.
We first study \irml training from scratch (w/o pre-training). %
\begin{theorem}\label{CH:FeAT:thm:irmv1_not_learn}
	Consider training a CNN model (\ref{CH:FeAT:eq:cnn}) with data model (\ref{CH:FeAT:def:risk_irm}), define
	\[\rvc(t) \triangleq \left[ C^1_\irml(\rmW,t),  C^2_\irml(\rmW,t), \cdots,  C^{|\envtrain|}_\irml(\rmW,t) \right],\]
	and $\lambda_0 =  \lambda_{\min}(\mathbf{H}^\infty)$, where \[ \mathbf{H}^\infty_{e,e'} \triangleq \frac{  1}{2m n_e n_{e'}} \sum_{i=1}^{n_e} \psi'(\langle \rvw_{j,r}(0) ,  \mathbf{x}^{e}_{1,i} \rangle ) \mathbf{x}^{e\top}_{1,i} \sum_{i'=1}^{n_{e'}} \psi'(\langle \rvw_{j,r}(0) ,  \rvx_{1,i'}^{e'} \rangle )    \rvx_{1,i'}^{e'}\]. Suppose that dimension $d
		=  \Omega(\log(m/\delta))$, network width $m = \Omega(1/\delta)$, regularization factor $\lambda \ge 1/(\sigma_0  {|\mathcal{E}_{tr}|}^{3/2} )$, noise variance $\sigma_p = O(d^{-2})$, weight initial scale $\sigma_0 = O( \frac{ |\mathcal{E}_{tr} |^{7/2} \beta^3 L }{ d^{1/2}m^2\lambda^2_0 \log(1/\epsilon)} )$, then with probability at least $1-\delta$, after training time $T = \Omega \left( \frac{ \log(1/\epsilon)}{ \eta  \lambda \lambda_0 } \right) $, we have
	$\| \rvc(T )\|_2 \le  \epsilon, \
		{\gamma_{j,r}^{inv}}(T)\!=\!{o(1)}, \  {\gamma_{j,r}^{spu}}(T)\!=\!{o(1)}.$
\end{theorem}
The proof is given in Appendix~\ref{CH:FeAT:sec:proof_irmv1_not_learn}. We highlight that Theorem \ref{CH:FeAT:thm:irmv1_not_learn} allows any number of training environments, which indicates a fundamental limitation of pure \irml training.
Intuitively, Theorem~\ref{CH:FeAT:thm:irmv1_not_learn} implies that,
when a heavy regularization of \irml is applied, the model will not learn any features, corresponding to Figure \ref{f5d}.
Instead, \irml suppresses any feature learning,
even at the beginning of the training.
Then, what would happen when given a properly pre-trained network?

After ERM pre-training, according to Theorem \ref{CH:FeAT:thm:erm_learn_feat}, we have $\abs{\langle  \rvw_{j,r}, \rvv_1 \rangle} = \Omega(1)$, $\abs{\langle \rvw_{j,r}, \rvv_2 \rangle} = \Omega(1)$, and $\abs{\langle \rvw_{j,r}, \boldsymbol{\xi} \rangle} = O(\sigma_0 \sigma_p \sqrt{d})$. Then, we have the following hold.

\begin{proposition}
	\label{pro:irmv1_with_erm_feat}
	Given the same setting as Theorem~\ref{CH:FeAT:thm:irmv1_not_learn}, suppose that $\psi(x) = x$, $\gamma_{j,r}^{inv}(t_1) =  \gamma_{j,r}^{inv}(t_1-1)$, and $ \gamma_{j,r}^{spu}(t_1 ) =  \gamma_{j,r}^{spu}(t_1-1)$ at the end of ERM pre-train $t_1$, $\delta>0$, and $ n > C \log(1/\delta)$, with $C$ being a positive constant, then with a high probability at least $1-\delta$, we have $ \sum_e C^e_\irml(t_1) =0,$ $\gamma_{j,r}^{inv}(t_1 + 1) >  \gamma_{j,r}^{inv}(t_1),$ and $\gamma_{j,r}^{spu}(t_1 + 1) <  \gamma_{j,r}^{spu}(t_1).$
\end{proposition}
The proof is given in Appendix~\ref{CH:FeAT:sec:proof_irmv1_with_erm_feat}, which takes converged feature learning terms from Theorem \ref{CH:FeAT:thm:erm_learn_feat} as the inputs.
Proposition \ref{pro:irmv1_with_erm_feat} demonstrates that with sufficient ERM pre-training, \irml can enhance the learning of invariant features while suppressing the learning of spurious features, which is verified in Figure \ref{f5b} and \ref{f5a}.
Thus, when given the initialization with better learned invariant features, i.e., longer ERM pre-training epochs,
\irml improves the invariant feature better.
Proposition~\ref{pro:irmv1_with_erm_feat} explains why the final OOD performance highly depends on the ERM pre-training~\citep{rfc,pair}.

\subsection{Limitations of ERM Feature Learning}\label{CH:FeAT:sec:erm_limits}
Combining results from both Sec.~\ref{CH:FeAT:sec:erm_learn_feat} and Sec.~\ref{CH:FeAT:sec:ood_not_learn},
we know that the invariant features will be learned during ERM pre-training and discovered during OOD training.
However, given poorly learned invariant features, can \irml still improve it?
In practice, there often exist some invariant features that are not properly learned by ERM.
For example, in our data model Def.~\ref{CH:FeAT:def:risk_irm} when the invariant correlation is much weaker than the spurious correlation, given a limited number of training steps, the spurious feature learning can dominate the invariant feature learning.
Besides, when considering other factors such as the simplicity bias of ERM~\citep{simple_bias} or the inductive biases of the network architecture~\citep{what_shape}, it is more likely that there exist invariant features that are not properly learned~\citep{imagenetv2}. Then we have:

\begin{corollary}
	\label{cor:irmv1_with_bad_feat}
	Consider training the CNN with the data generated from Def.~\ref{CH:FeAT:def:risk_irm},
	suppose that $\psi(x) = x$, $\gamma_{j,r}^{inv}(t_1 ) =  o(1) $, and $\gamma_{j,r}^{spu}(t_1 ) =  \Theta(1)$ at the end of ERM pre-training $t_1$. Suppose that $\delta>0$, and $ n > C \log(1/\delta)$, with $C$ being a positive constant, then with a high probability at least $1-\delta$, we have $\gamma_{j,r}^{inv}(t_1 + 1) <  \gamma_{j,r}^{inv}(t_1).$
\end{corollary}

Corollary~\ref{cor:irmv1_with_bad_feat} shows that \irml
requires sufficiently well-learned features for OOD generalization.
It is also consistent with the experimental results in Fig.~\ref{f5b},~\ref{f5c}, and Fig.~\ref{CH:FeAT:fig:flood_phenon}, where all the OOD objectives only achieve a performance comparable to random guesses.
\section{Feature Augmentated Training}
\label{CH:FeAT:sec:fat_sol}

\subsection{Rich Features for OOD Generalization}
\label{CH:FeAT:sec:fat_theory}
The results in Sec.~\ref{CH:FeAT:sec:understand} imply the necessity of learning
all potentially useful features during the pre-training stage for OOD generalization.
Otherwise, the OOD training is less likely to enhance the poorly learned features.
It also explains the success of learning diverse and rich features by weight averaging~\citep{diwa,eoa} and rich feature construction (or \rfc)~\citep{rfc}, and other approaches~\citep{fft,recycling}.

Despite the empirical success, however, the learning of rich features in both \rfc and weight averaging is unstable and expensive. On the one hand, they may discard previously learned useful features or fail to explore all the desired features as it is hard to evaluate the quality of the intermediate learned features.
\begin{wrapfigure}{r}{0.6\textwidth}
	\vskip-0.3in
	\begin{minipage}{0.6\textwidth}
		\begin{algorithm}[H]
			\caption{\feat: \featfull }
			\label{alg:fat}
			\begin{algorithmic}[1]
				\STATE \textbf{Input:} Training data $\train$; the maximum augmentation rounds $K$; predictor $f:=w\circ\varphi$; length of inner training epochs $t$; termination threshold $p$;
				\STATE Initialize groups $G^a\leftarrow {\train}, G^r\leftarrow \{\}$;
				\FOR{$k \in [1,\ldots, K]$}
				\STATE Randomly initialize $w_k$;
				\FOR{$j \in [1,\ldots, t]$}
				\STATE Obtain $\ell_\feat$ with $G$ via Eq.~\ref{CH:FeAT:eq:fat_upd};
				\STATE Update $w_k, \varphi$ with $\ell_\feat$;
				\ENDFOR
				\STATE \texttt{// Early Stop if $f_k=w_k\circ\varphi$ fails to find new features.}
				\IF{Training accuracy of $f_k$ is smaller than $p$}
				\STATE Set $K=k-1$ and terminate the loop;
				\ENDIF
				\STATE Split $\train$ into groups $\dataset_k^a,\dataset_k^r$ according to whether $f_k$ classifies the examples in $\train$ correctly or not;
				\STATE Update groups $G^a \leftarrow G^a \cup \{\dataset_k^a\}, G^r\leftarrow G^r\cup\{\dataset_k^r\}$;
				\ENDFOR
				\STATE Synthesize the final classifier $w\leftarrow\frac{1}{K}\sum_{i=1}^{K}w_i$;
				\STATE \textbf{return} $f=w\circ\varphi$;
			\end{algorithmic}
		\end{algorithm}
	\end{minipage}
	\vskip-0.2in
\end{wrapfigure}
On the other hand, they also need multiple initializations and training of the whole networks with different random seeds to encourage the diversity of feature learning, which brings more instability and computational overhead, especially when applied to large and deep networks.

\subsection{The FeAT Algorithm}
\label{CH:FeAT:sec:fat_alg}
To overcome the limitations of previous rich feature learning algorithms, we propose  \featfull (\feat), that directly augment the feature learning in an iterative manner.

Intuitively, the potentially useful features presented in the training data are features that have non-trivial correlations with labels, or using the respective feature to predict the labels is able to achieve a \emph{non-trivial training performance}.
Moreover, the invariance principle assumes that the training data comes from different environments~\citep{irmv1}, which implies that each set of features can only dominate the correlations with labels in a \emph{subset} of data.
Therefore, it is possible to differentiate the distinct sets of useful features entangled in the training data into different subsets, where ERM can effectively learn the dominant features presented in the corresponding subset as shown in Theorem~\ref{CH:FeAT:thm:erm_learn_feat}.

The intuition naturally motivates an iterative rich feature learning algorithm, i.e., \feat, that identifies the subsets containing distinct features and explores to learn new features in multiple rounds. The details of \feat are given in Algorithm~\ref{alg:fat}, where we are given a randomly initialized or pre-trained model $f=w\circ\varphi$ that consists of a featurizer $\varphi$ and a classifier $w$.
In round $k$, \feat first identifies the subset that contains the already learned features by collecting the samples where $f$ yields the correct prediction, denoted as $G^r_k$, and the subset of samples that contains the features that have not been learned, denoted as $G^a_k$.

At the $k$-th round, given the grouped subsets $G=\{G^r,G^a\}$ with $2k-1$ groups, where $G^a=\{\dataset_i^a\}_{i=0}^{k-1}$ is the grouped sets for new feature augmentation, and $G^r=\{\dataset_i^r\}_{i=1}^{k-1}$ is the grouped sets for already learned feature retention (notice that $\dataset_0^r$ is the empty set), where $\dataset_i^a$ and $\dataset_i^r$ are the corresponding augmentation and retention set elicited at $i$-th round.
\feat performs distributionally robust optimization (DRO)~\citep{dro,rfc} on $G^a$ to  explore new features that have not been learned in previous rounds.
Meanwhile, \feat also needs to \emph{retain} the already learned features by minimizing the empirical risk at $G^r$, for which we store and use the historical classifiers $w_i$ with the current featurizer to evaluate the feature retention degree.
Then, the \feat objective at round $k$  is
\begin{equation}\label{CH:FeAT:eq:fat_upd}
	\ell_\feat =
	\max_{\dataset_i^a\in G^a}\ell_{\dataset_i^a}(w_k\circ\varphi)+\lambda\cdot \sum_{\dataset_i^r\in G^r}\ell_{\dataset_i^r}(w_{i}\circ\varphi),
\end{equation}
where $\ell_{\dataset_i}(w\circ\varphi)$ refers to the empirical risk of $w\circ\varphi$ evaluated at the subset $\dataset_i$, and $\{w_i|1\leq i\leq k-1\}$ are the historical classifiers trained in round $i$.
The final classifier is the average of all historical classifiers as they already capitalize all the learned features in each round.

\textbf{Relations with previous rich feature learning algorithms.}
Compared with previous rich feature learning algorithms,
\feat \emph{directly} trades off the exploration of new features and the retention of the already learned features.
Although \rfc also adopts DRO to explore new features,
the isolation of new feature exploration and already learned feature synthesis makes the feature learning in \rfc more unstable.
In other words, \rfc can not evaluate the intermediate feature learning results due to the \emph{indirect} feature exploration and synthesis.
Consequently, \rfc can not control when to stop the new feature exploration, and thus may fail to explore all of the desired features or discard important features.
Besides, the multiple re-initializations and re-training of the whole network in \rfc could also lead to suboptimal performance and meanwhile require more computational overhead.

\textbf{Practical implementations.}
Algorithm~\ref{alg:fat} requires to store $2K-1$ subsets and a larger memory cost in training the network, which may cause additional storage burden when $\varphi$ contains a massive amount of parameters~\citep{wilds}. Hence, we propose a lightweight variant of \feat (denoted as \feati) which only retains the latest subsets and historical classifiers ($\dataset_{k-1}^a, \dataset_{k-1}^r, w_{k-1}$ at the $k$-th round). Throughout the whole experiment, we will use \feati and find that \feati already achieves state-of-the-art. More details are given in Appendix~\ref{CH:FeAT:sec:ifat_appdx}.

As \feati stores only the latest augmentation and retention subsets, inspecting the training performance for termination check (line 10 of Algorithm~\ref{alg:fat}) may not be suitable. However, one can still
inspect the performance in either an OOD validation set to check the quality of the intermediate feature representations, or the retention set to check whether learning new features leads to a severe contradiction of the already learned features (\feat should terminate if so).

Compared to ERM, the additional computational and memory overhead introduced in \feat mainly lie in the \feat training and partitioning. At each training step, \feat needs $(k-1)$ additional forward and backward propagation, the same as Bonsai, while \feat only needs $\min(1,k-1)$ additional propagation. Besides, Bonsai additionally require another round of training with $(K-1)$ additional propagation, given $K$ total rounds. More details are given in Appendix~\ref{CH:FeAT:sec:comput_analysis_appdx}.
\section{Empirical Study}
\label{CH:FeAT:sec:exp}
\begin{table}[t]
    \caption{OOD performance with \feat on \cmnist datasets.}
    \small\center\sc
    \label{tab:sythetic}
    \begin{center}
        \resizebox{\textwidth}{!}{
            \begin{tabular}{l|rrrr|rrrr}
                \toprule
                                           &
                \multicolumn{4}{c}{\cmnist-025}
                                           &
                \multicolumn{4}{c}{\cmnist-01}                                                                                                                                           \\
                \rule{0pt}{8pt}            &
                \multicolumn{1}{c}{ERM-nf} &
                \multicolumn{1}{c}{ERM}
                                           &
                \multicolumn{1}{c}{\rfc}
                                           &
                \multicolumn{1}{c}{\ours}
                                           &
                \multicolumn{1}{c}{ERM-nf}
                                           &
                \multicolumn{1}{c}{ERM}
                                           &
                \multicolumn{1}{c}{\rfc}
                                           &
                \multicolumn{1}{c}{\ours}
                \\
                \cmidrule(lr){2-5}\cmidrule(lr){6-9}
                ERM                        & $17.14$ \std{0.73}                     & $12.40$ \std{0.32}                     & $11.21$ \std{0.49}          & $\mathbf{17.27}$ \std{2.55}
                                           & $73.06$ \std{0.71}                     & $73.75$ \std{0.49}                     & $70.95$ \std{0.93}          & $\mathbf{76.05}$ \std{1.45} \\
                IRMv1                      & $67.29$ \std{0.99}                     & $59.81$ \std{4.46}                     & $70.28$ \std{0.72}          & $\mathbf{70.57}$ \std{0.68}
                                           & $76.89$ \std{3.25}                     & $73.84$ \std{0.56}                     & $76.71$ \std{4.10}          & $\mathbf{82.33}$ \std{1.77} \\
                V-Rex                      & $68.62$ \std{0.73}                     & $65.96$ \std{1.29}                     & $70.31$ \std{0.66}          & $\mathbf{70.82}$ \std{0.59}
                                           & $83.52$ \std{2.52}                     & $81.20$ \std{3.27}                     & $82.61$ \std{1.76}          & $\mathbf{84.70}$ \std{0.69} \\
                \irmx                      & $67.00$ \std{1.95}                     & $64.05$ \std{0.88}                     & $70.46$ \std{0.42}          & $\mathbf{70.78}$ \std{0.61}
                                           & $81.61$ \std{1.98}                     & $75.97$ \std{0.88}                     & $80.28$ \std{1.62}          & $\mathbf{84.34}$ \std{0.97} \\
                IB-IRM                     & $56.09$ \std{2.04}                     & $59.81$ \std{4.46}                     & $70.28$ \std{0.72}          & $\mathbf{70.57}$ \std{0.68}
                                           & $75.81$ \std{0.63}                     & $73.84$ \std{0.56}                     & $76.71$ \std{4.10}          & $\mathbf{82.33}$ \std{1.77} \\
                CLOvE                      & $58.67$ \std{7.69}                     & $65.78$ \std{0.00}                     & $65.57$ \std{3.02}          & $\mathbf{65.78}$ \std{2.68}
                                           & $75.66$ \std{10.6}                     & $74.73$ \std{0.36}                     & $72.73$ \std{1.18}          & $\mathbf{75.12}$ \std{1.08} \\
                IGA                        & $51.22$ \std{3.67}                     & $62.43$ \std{3.06}                     & $\mathbf{70.17}$ \std{0.89} & $67.11$ \std{3.40}
                                           & $74.20$ \std{2.45}                     & $73.74$ \std{0.48}                     & $74.72$ \std{3.60}          & $\mathbf{83.46}$ \std{2.17} \\
                Fishr                      & $69.38$ \std{0.39}                     & $67.74$ \std{0.90}                     & $68.75$ \std{1.10}          & $\mathbf{70.56}$ \std{0.97}
                                           & $77.29$ \std{1.61}                     & $82.23$ \std{1.35}                     & $84.19$ \std{0.66}          & $\mathbf{84.26}$ \std{0.93} \\\hline
                \rule{0pt}{8pt}Oracle      & \multicolumn{4}{c}{$71.97$ \std{0.34}} & \multicolumn{4}{c}{$86.55$ \std{0.27}}                                                             \\
                \bottomrule
            \end{tabular}
        }
    \end{center}
\end{table}

We conduct extensive experiments on \cmnist~\citep{pair} and \wilds~\citep{wilds} to verify the effectiveness of \feat in learning richer features than ERM and the state-of-the-art algorithm \rfc~\citep{rfc}.

\textbf{Proof-of-concept study on \cmnist.}
We first conduct a proof-of-concept study using \cmnist~\citep{pair} and examine the feature learning performance of \feat under various conditions.
We consider both the original \cmnist with $\envtrain=\{(0.25,0.1),(0.25,0.2)\}$ (denoted as \cmnist-025), where spurious features are better correlated with labels, and the modified \cmnist (denoted as \cmnist-01) with $\envtrain=\{(0.1,0.2),(0.1,0.25)\}$, where invariant features are better correlated with labels.
We compare the OOD performance of the features learned by \feat, with that of \erm and the state-of-the-art rich feature learning algorithm \rfc~\citep{rfc}.
Based on the features learned by ERM, \rfc, and \feat,
we adopt various state-of-the-art OOD objectives including \irml~\citep{irmv1}, \vrex~\citep{vrex}, \irmx~\citep{pair}, IB-IRM~\citep{ib-irm}, CLOvE~\citep{clove}, IGA~\citep{iga} and Fishr~\citep{fishr} for OOD training, in order to evaluate the practical quality of the learned features.
The feature representations are frozen once initialized for the OOD training as fine-tuning the featurizer can distort the pre-trained features~\citep{finetuning_distorts}.
We also compare \feat with the common training approach that uses unfrozen \erm features, denoted as \erm-\textsc{NF}.
For \rfc, we trained $2$ rounds following~\citet{rfc}, while for \feat the automatic termination stopped at round $2$ in \cmnist-025 and round $3$ in \cmnist-01.
For \erm, we pre-trained the model with the same number of overall epochs as \feat in \cmnist-01, while early stopping at the number of epochs of $1$ round in \cmnist-025 to prevent over-fitting.
All methods adopted the same backbone and the same training protocol following previous works~\citep{rfc,pair}. More details are given in Appendix~\ref{CH:FeAT:sec:cmnist_appdx}.

The results are reported in Table~\ref{tab:sythetic}. It can be found that
ERM will learn insufficiently good features under both stronger spurious correlations and invariant correlations, confirming our discussion in Sec.~\ref{CH:FeAT:sec:erm_limits}.
Besides, \rfc learns richer features in \cmnist-025 and boosts OOD performance, but \rfc sometimes leads to suboptimal performances in \cmnist-01, which could be caused by the unstable feature learning in \rfc.
In contrast, \feat consistently improves the OOD performance of all OOD objectives for all the \cmnist datasets, demonstrating the advances of direct feature learning control in \feat than \rfc and ERM.

\bgroup
\begin{table}[t]
    \centering
    \caption{OOD generalization performances with \feat on \wilds benchmark.}
    \label{tab:wilds_results}
    \resizebox{\textwidth}{!}{
        \small
        \begin{tabular}{@{}{c}|{c}|*{6}{c}@{}}    \toprule
            \multirow{2.5}{*}{\textsc{Init.}} & \multirow{2.5}{*}{\textsc{Method}}
                                              & {
            \textsc{Camelyon17}}              & {
            \textsc{CivilComments}}           & {
            \textsc{FMoW}}                    & {
            \textsc{iWildCam}}                & {
            \textsc{Amazon}}                  & {
            \textsc{RxRx1}}                                                                                                                                                                                                                                             \\
            \cmidrule(lr){3-3} \cmidrule(lr){4-4} \cmidrule(lr){5-5} \cmidrule(lr){6-6} \cmidrule(lr){7-7}\cmidrule(lr){8-8}
                                              &                                    & Avg. acc. (\%)              & Worst acc. (\%)             & Worst acc. (\%)             & Macro F1                    & 10-th per. acc. (\%)        & Avg. acc. (\%)               \\
            \midrule
            ERM                               & DFR$^\dagger$                      & $95.14$ \std{1.96}          & $\mathbf{77.34}$ \std{0.50} & $41.96$ \std{1.90}          & $23.15$ \std{0.24}          & $48.00$ \std{0.00}          & -                            \\
            ERM                               & DFR-s$^\dagger$                    & -                           & $82.24$ \std{0.13}          & $56.17$ \std{0.62}          & $52.44$ \std{0.34}          & -                           & -                            \\\hdashline[0.5pt/1pt]
            \rule{0pt}{8pt}\rfc               & DFR$^\dagger$                      & $95.17$ \std{0.18}          & $77.07$ \std{0.85}          & $43.26$ \std{0.82}          & $21.36$ \std{0.41}          & $46.67$ \std{0.00}          & -                            \\
            \rfc                              & DFR-s$^\dagger$                    & -                           & $81.26$ \std{1.86}          & $58.58$ \std{1.17}          & $50.85$ \std{0.18}          & -                           & -                            \\\hdashline[0.5pt/1pt]
            \rule{0pt}{8pt}\ours              & DFR$^\dagger$                      & $\mathbf{95.28}$ \std{0.19} & $\mathbf{77.34}$ \std{0.59} & $\mathbf{43.54}$ \std{1.26} & $\mathbf{23.54}$ \std{0.52} & $\mathbf{49.33}$ \std{0.00} & -                            \\
            \ours                             & DFR-s$^\dagger$                    & -                           & $79.56$ \std{0.38}          & $57.69$ \std{0.78}          & $52.31$ \std{0.38}          & -                           & -                            \\
            \midrule\midrule
            ERM                               & ERM                                & $74.30$ \std{5.96}          & $55.53$ \std{1.78}          & $33.58$ \std{1.02}          & $28.22$ \std{0.78}          & $51.11$ \std{0.63}          & $30.21$  \std{0.09}          \\
            ERM                               & GroupDRO                           & $76.09$ \std{6.46}          & $69.50$ \std{0.15}          & $33.03$ \std{0.52}          & $28.51$ \std{0.58}          & $52.00$ \std{0.00}          & $29.99$  \std{0.13}          \\
            ERM                               & IRMv1                              & $75.68$ \std{7.41}          & $68.84$ \std{0.95}          & $33.45$ \std{1.07}          & $28.76$ \std{0.45}          & $52.00$ \std{0.00}          & $30.10$ \std{0.05}           \\
            ERM                               & V-REx                              & $71.60$ \std{7.88}          & $69.03$ \std{1.08}          & $33.06$ \std{0.46}          & $28.82$ \std{0.47}          & $52.44$ \std{0.63}          & $29.88$ \std{0.35}           \\
            ERM                               & \irmx                              & $73.49$ \std{9.33}          & $68.91$ \std{1.19}          & $33.13$ \std{0.86}          & $28.82$ \std{0.47}          & $52.00$ \std{0.00}          & $30.10$  \std{0.05}          \\\hdashline[0.5pt/1pt]
            \rule{0pt}{8pt}\rfc               & ERM                                & $73.98$ \std{5.30}          & $63.34$ \std{3.49}          & $31.91$ \std{0.51}          & $28.27$ \std{1.05}          & $48.58$ \std{0.56}          & $24.22$  \std{0.44}          \\
            \rfc                              & GroupDRO                           & $72.82$ \std{5.37}          & $70.23$ \std{1.33}          & $33.12$ \std{1.20}          & $27.16$ \std{1.18}          & $42.67$ \std{1.09}          & $22.95$  \std{0.46}          \\
            \rfc                              & IRMv1                              & $73.59$ \std{6.16}          & $68.39$ \std{2.01}          & $32.51$ \std{1.23}          & $27.60$ \std{1.57}          & $47.11$ \std{0.63}          & $23.35$ \std{0.43}           \\
            \rfc                              & V-REx                              & $76.39$ \std{5.32}          & $68.67$ \std{1.29}          & $33.17$ \std{1.26}          & $25.81$ \std{0.42}          & $48.00$ \std{0.00}          & $23.34$ \std{0.42}           \\
            \rfc                              & \irmx                              & $64.77$ \std{10.1}          & $69.56$ \std{0.95}          & $32.63$ \std{0.75}          & $27.62$ \std{0.66}          & $46.67$ \std{0.00}          & $23.34$ \std{0.40}           \\\hdashline[0.5pt/1pt]
            \rule{0pt}{8pt}\ours              & ERM                                & $77.80$ \std{2.48}          & $68.11$ \std{2.27}          & $33.13$ \std{0.78}          & $28.47$ \std{0.67}          & $\mathbf{52.89}$ \std{0.63} & $\mathbf{30.66}$  \std{0.42} \\
            \ours                             & GroupDRO                           & $\mathbf{80.41}$ \std{3.30} & $\mathbf{71.29}$ \std{0.46} & $33.55$ \std{1.67}          & $28.38$ \std{1.32}          & $52.58$ \std{0.56}          & $29.99$  \std{0.11}          \\
            \ours                             & IRMv1                              & $77.97$ \std{3.09}          & $70.33$ \std{1.14}          & $\mathbf{34.04}$ \std{0.70} & $\mathbf{29.66}$ \std{1.52} & $\mathbf{52.89}$ \std{0.63} & $29.99$ \std{0.19}           \\
            \ours                             & V-REx                              & $75.12$ \std{6.55}          & $70.97$ \std{1.06}          & $34.00$ \std{0.71}          & $29.48$ \std{1.94}          & $\mathbf{52.89}$ \std{0.63} & $30.57$ \std{0.53}           \\
            \ours                             & \irmx                              & $76.91$ \std{6.76}          & $71.18$ \std{1.10}          & $33.99$ \std{0.73}          & $29.04$ \std{2.96}          & $\mathbf{52.89}$ \std{0.63} & $29.92$  \std{0.16}          \\
            \bottomrule
            \multicolumn{8}{l}{\rule{0pt}{8pt}
                $^\dagger$\text{\normalfont \small DFR/DFR-s use an additional OOD dataset to evaluate invariant and spurious feature learning, respectively.} }
        \end{tabular}
    }
\end{table}
\egroup

\textbf{Experiments on real-world benchmarks.}
We also compare \feat with \erm and \rfc in $6$ real-world OOD generalization datasets curated by~\citet{wilds} that contain complicated features and distribution shifts.
The learned features are evaluated with several representative state-of-the-art OOD objectives in \wilds, including GroupDro~\citep{groupdro}, \irml~\citep{irmv1}, \vrex~\citep{vrex} as well as \irmx~\citep{pair}.
By default, we train \erm, \rfc and \feat the same number of steps, and kept the rounds of \rfc and \feat the same (though \rfc still requires one more round for feature synthesis).
The only exception is in \textsc{RxRx1} where both \rfc and \feat required more steps than ERM to converge.
We use the same evaluation protocol following the practice in the literature~\citep{wilds,fish,rfc,pair} to ensure a fair comparison.
More details are given in Appendix~\ref{CH:FeAT:sec:wilds_appdx}.

In addition to OOD objectives, we evaluate the learned features with Deep Feature Reweighting (DFR)~\citep{dfr}.
DFR uses an additional OOD validation set where the \textit{spurious correlation does not hold}, to perform logistic regression based on the learned features. Intuitively, DFR can serve as a proper measure for the quality of learned invariant features~\citep{dfrlearn}. When the original dataset does not provide a proper OOD validation set, e.g., \textsc{Camelyon17}, we use an alternative implementation based on a random split of the training and test data to perform the invariant feature quality measure~\citep{dare}.
Similarly, we also report DFR-s by regression with the environment labels (when available) to evaluate the spurious feature learning of different methods.
More details are given in Appendix~\ref{CH:FeAT:sec:eval_detail_appdx}.

The results are presented in Table~\ref{tab:wilds_results}.
Similarly, when the tasks grow more challenging and neural architectures become more complicated, the ERM learned features can have a lower quality as discussed Sec.~\ref{CH:FeAT:sec:erm_limits}.
For example, ERM can not sufficiently learn all useful features in FMoW, while ERM can learn more spurious correlations in CivilComments.
Moreover, it can also be observed the instability of \rfc in learning richer features that \rfc even underperforms ERM in rich feature learning and OOD generalization in multiple datasets.
In contrast, \feat consistently achieves the best invariant feature learning performance across various challenging realistic datasets. Meanwhile, compared to \erm and \rfc, \feat also reduces over-fitting to the spurious feature learning led by spurious correlations.
As a result, \feat achieves consistent improvements when the learned features are applied to various OOD objectives.
\begin{table}[t]%
	\caption{Performances at different \feat rounds.}
	\label{tab:termin_check}
	\begin{center}
		\begin{scriptsize}
			\begin{sc}
				\begin{tabular}{lcccc}
					\toprule
					\cmnist-025    & Round-1         & Round-2         & Round-3         \\
					\midrule
					Training Acc.  & 85.08$\pm$ 0.14 & 71.87$\pm$ 0.96 & 84.93$\pm$ 1.26 \\
					Retention Acc. & -               & 88.11$\pm$ 4.28 & 43.82$\pm$ 0.59 \\
					OOD Acc.       & 11.08$\pm$ 0.30 & 70.64$\pm$ 0.62 & 10.07$\pm$ 0.26 \\
					\bottomrule
				\end{tabular}
			\end{sc}
		\end{scriptsize}
	\end{center}
\end{table}

\textbf{The termination check in \feat.}
As elaborated in Sec.~\ref{CH:FeAT:sec:fat_alg}, a key difference between \feat and previous rich feature learning algorithms such as \rfc is that \feat is able to access the intermediate feature representations and thus can perform the automatic termination check and learn the desired features stably.
To verify, we list the \feat performances in various subsets of \cmnist-025 at different rounds in Table~\ref{tab:termin_check}. By inspecting the retention accuracy, after \feat learns sufficiently good features at Round $2$, it is not necessary to proceed with Round $3$ as it will destroy the already learned features and lead to degenerated retention and OOD performance. More details and results are given in Appendix~\ref{CH:FeAT:sec:cmnist_appdx}.

\textbf{Computational analysis.} We also analyze the computational and memory overhead of different methods, for which the details are given in Appendix~\ref{CH:FeAT:sec:comput_analysis_appdx}.
Compared to ERM and Bonsai, \feati achieves the best performance without introducing too much additional overhead.

\begin{figure}[t!]
	\centering
	\subfigure[ERM 300 epochs]{
		\includegraphics[width=0.45\textwidth]{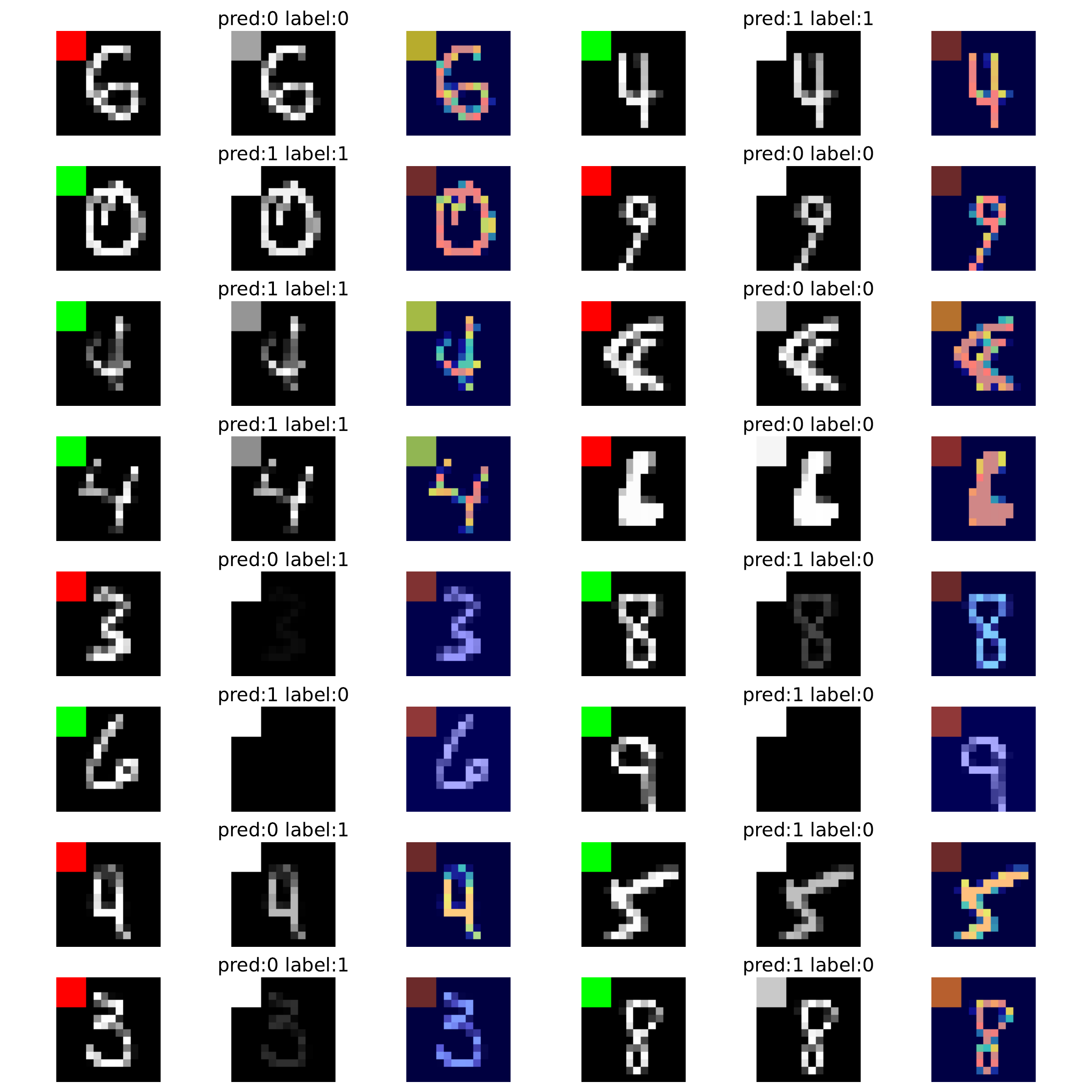}
		\label{CH:FeAT:fig:s4_erm_ep300_appdx}
	}%
	\subfigure[FeAT 2 rounds]{
		\includegraphics[width=0.45\textwidth]{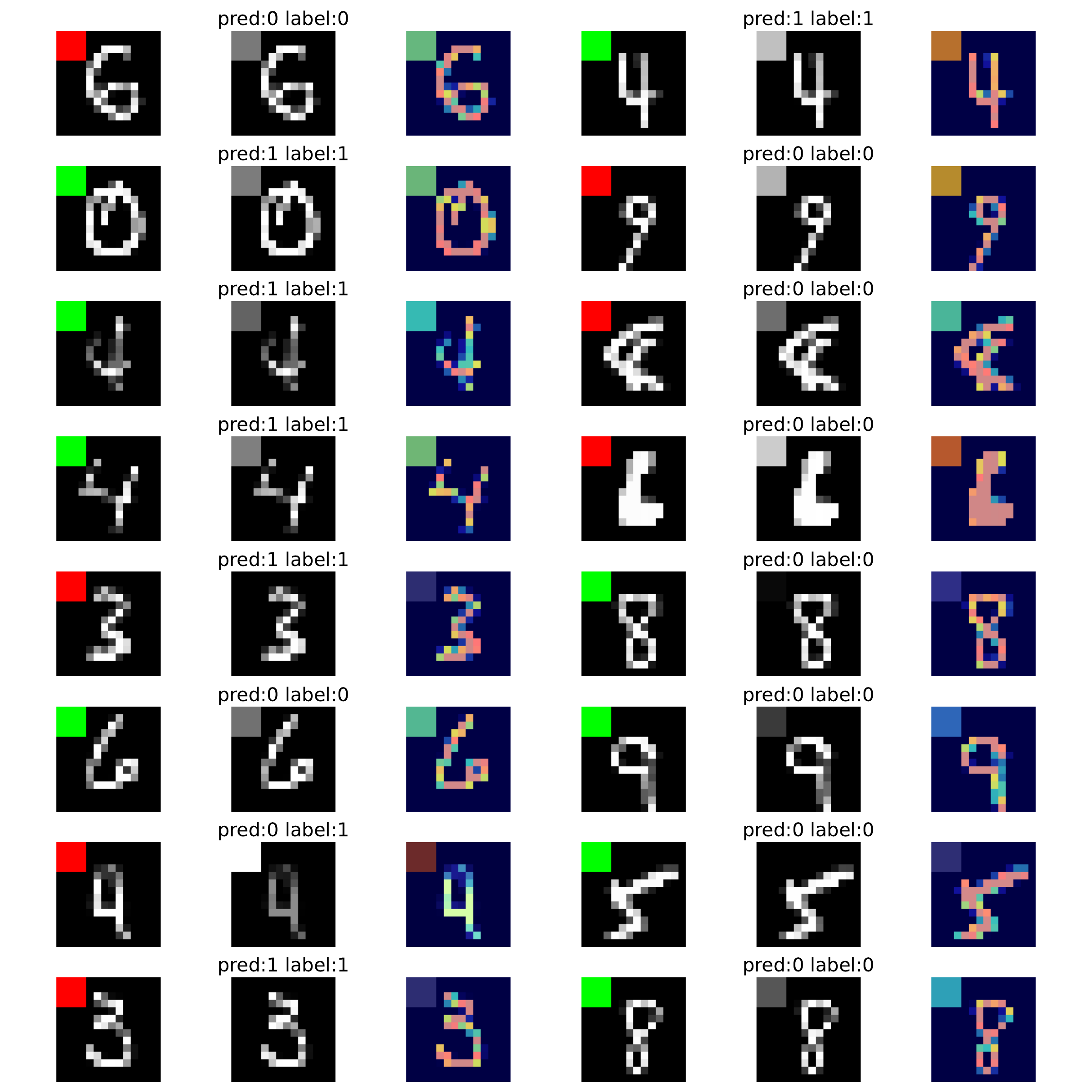}
		\label{CH:FeAT:fig:s4_ifat_r2_appdx}
	}
	\caption[GradCAM visualization on \cmnist-025.]{GradCAM visualization on \cmnist-025, where the shortcuts are now concentrated to a colored path at the up left. Three visualizations are drawn for each sample: the original figure, the gray-colored gradcam, and the gradcam. It can be found that ERM can not properly capture the desired features while \feat can stably capture the desired features.}
	\label{CH:FeAT:fig:gradcam_cmnist}
\end{figure}

\textbf{Feature learning analysis.} We visualize the feature learning of ERM and \feat on ColoredMNIST-025.
As shown in Fig.~\ref{CH:FeAT:fig:gradcam_cmnist}, ERM can learn both invariant and spurious features to predict the label, aligned with our theory.
However, ERM focuses more on spurious features and even forgets certain features with longer training epochs, which could be due to multiple reasons such as the simplicity biases of ERM.
Hence predictions based on ERM learned features fail to generalize to OOD examples.
In contrast, \feat effectively captures the meaningful features for all samples and generalizes to OOD examples well.
More analysis including results on \wilds benchmark can be found in Appendix~\ref{CH:FeAT:sec:feat_analysis_appdx}.

\chapter{Conclusion} \label{ch:sec:conclusion}
This thesis is about establishing proper objectives and feasible optimization frameworks to learn causality for a variety of modern machine learning problems.
In Part~\ref{P1}, we proposed the basic frameworks and necessary assumptions for learning causal invariance on a general data structure, i.e., graphs. Then, we discussed the implications of causal invariance to interpretability and robustness. 
In Part~\ref{P2}, we investigated the optimization of causal invariance learning and proposed new optimization and representation learning schemes. 

Going beyond the thesis: currently, transformers have gained huge success in learning and modeling complex relations in images and sequential data. In fact, transformers can be considered as a family of GNNs, yet taking a different paradigm in optimization and generalization. 
How can we understand and improve the inner mechanisms of transformers by relating the insights from message-passing-based GNNs? Is it possible to combine the strengths of both families of GNNs to derive a better architecture for learning and utilizing causality with artificial intelligence?

\appendix
\chapter{Appendices of \ciga}\label{APP:CIGA}

\section{More Discussions on Related Work}
\label{CH:CIGA:sec:discuss_future_appdx}

\subsection{More backgrounds}
\label{CH:CIGA:sec:background_appdx}
We give more background introduction about GNNs and Invariant Learning in this section.

\textbf{Graph Neural Networks.} Let $G=(A,X)$ denote a graph with $n$ nodes and $m$ edges,
where $A \in \{0,1\}^{n\times n}$ is the adjacency matrix, and $X\in \R^{n \times d}$ is the node feature matrix
with a node feature dimension of $d$.
In graph classification, we are given a set of $N$ graphs $\{G_i\}_{i=1}^N\subseteq \gG$
and their labels $\{Y_i\}_{i=1}^N\subseteq\gY=\R^c$ from $c$ classes.
Then, we train a GNN $\rho \circ h$ with an encoder $h:\gG\rightarrow\R^h$ that learns a meaningful representation $h_G$ for each graph $G$ to help predict their labels $y_G=\rho(h_G)$ with a downstream classifier $\rho:\R^h\rightarrow\gY$.
The representation $h_G$ is typically obtained by performing pooling with a $\text{READOUT}$ function on the learned node representations:
\begin{equation}
	\label{CH:CIGA:gnn_pooling}
	h_G = \text{READOUT}(\{h^{(K)}_u|u\in V\}),
\end{equation}
where the $\text{READOUT}$ is a permutation invariant function (e.g., $\text{SUM}$, $\text{MEAN}$)~\citep{gin,diff_pooling,relation_pooling,gin,can_gnn_count,wl_goml},
and $h^{(K)}_u$ stands for the node representation of $u\in V$ at $K$-th layer that is obtained by neighbor aggregation:
\begin{equation}
	\label{CH:CIGA:gnn}
	h^{(K)}_u = \sigma(W_K\cdot a(\{h^{(K-1)}_v\}| v\in\mathcal{N}(u)\cup\{u\})),
\end{equation}
where $\mathcal{N}(u)$ is the set of neighbors of node $u$,
$\sigma(\cdot)$ is an activation function, e.g., $\text{ReLU}$, and $a(\cdot)$ is an aggregation function over neighbors, e.g., $\text{MEAN}$.

\textbf{Invariant Learning.}
Invariant learning typically considers
a supervised learning setting based on the data $\dataset=\{\dataset^e\}_e$ %
collected from multiple environments $\envall$,
where $\dataset^e=\{G^e_i,y^e_i\}$ is the dataset from environment $e\in\envall$.
$(G^e_i,y^e_i)$ from a single
environment $e$ are considered as drawn independently from an identical distribution $\sP^e$.
The goal of OOD generalization is to train a GNN $\rho \circ h:\gG\rightarrow\gY$
with data from training environments $\train=\{\dataset^e\}_{e\in\envtrain\subseteq\envall}$, and
generalize well to all (unseen) environments, i.e., to minimize:
\begin{equation}
	\label{CH:CIGA:ood}
	\min_{\rho,h}\max_{e\in\envall}R^e(\rho \circ h),
\end{equation}
where $R^e$ is the empirical risk under environment $e$~\citep{erm,inv_principle,irmv1}.
More details can be referred in~\citep{ib-irm}.

\subsection{Detailed related work}
\label{CH:CIGA:sec:related_work_appdx}

\textbf{GNN Explainability.}
Works in GNN explainability  aim to find a subgraph of the input graph as the explanation
for the prediction of a GNN model~\citep{gnn_explainer,xgnn_tax}.
Although some may leverage causality in explanation generation~\citep{gen_xgnn},
they mostly focus on understanding the predictions of GNNs in a post-hoc manner instead of OOD generalization.
Recently there are two works aiming to provide robust explanations under distribution shifts, i.e., GIB~\citep{gib} and DIR~\citep{dir},
and both of them focus on tackling FIIF spurious correlations (Assumption~\ref{CH:CIGA:assump:scm_fiif_appdx}).
The theoretical guarantees of GIB follows the theory of information bottleneck~\citep{ib},
while GIB can not solve PIIF spurious correlations (Assumption~\ref{CH:CIGA:assump:scm_piif_appdx}).
As both FIIF and PIIF widely exist in realistic scenarios, failing to solve either of them could result in severe performance degradation in practice~\citep{irmv1,ib-irm,aubin2021linear,failure_modes}.
While for DIR, though as a generalization of~\citet{inv_rat} to graphs, can not provide any theoretical guarantees under FIIF spurious correlations as shown in Appendix~\ref{CH:CIGA:sec:discussion_ood_obj_appdx},
nor under PIIF spurious correlations.

\textbf{GNN Extrapolation.}
Recently there is a surge of attention in improving the extrapolation ability of GNNs and apply them to various applications,
such as mathematical reasoning~\citep{math_reasoning1,math_reasoning2},
physics~\citep{physics_reasoning1,physics_reasoning2},
and graph algorithms~\citep{size_inv_extra,neural_exe,what_nn_reason,transfer_alg}.
\citet{nn_extrapo} study the neural network extrapolation ability from a geometrical perspective.
\citet{reliable_drug} improve OOD drug discovery by mitigating
the overconfident misprediction issue.
\citet{understand_att,size_gen1} focus on the extrapolation of GNNs in terms of
graph sizes, while making additional assumptions on the knowledge about
ground truth attentions and access to test inputs.
\citet{size_gen2} study the graph size extrapolation problem of GNNs through a causal lens,
while the induced  invariance principle is built upon assumptions on the specific family of graphs.
Different from these works, we consider the GNN extrapolation as a causal
problem, establish generic SCMs that are compatible with several graph generation models,
as well as, more importantly, different types of distribution shifts.
Hence, the induced the invariance principle and provable algorithms built upon the SCMs in our work can generalize to multiple graph families and distribution shifts.

Additionally, \citet{handle_node} propose causal models as well as specialized objectives to extrapolate
nodes with different neighbors. However, their formulation is limited to node classification task and specific spurious correlation type.
In contrast, the induced invariance principle in~\citet{handle_node},
can be seen as a extension of \ciga for node classification,
where we cab identify an invariant subgraph
from the $K$-hop neighbor graph of each node, and
making predictions based on it, i.e., $Y\ind E|G_c^\ego\subseteq G_u^\ego$ for node $u$.
We leave specific formulation and implementation to future works.

\textbf{Causality and OOD Generalization.}
Causality comes to the stage for demystifying and improving the huge success
of machine learning algorithms to further advances~\citep{seven_tools_causality,causality4ml,towards_causality}.
One of the most widely applied concept from causality is the Independent Causal Mechanism (ICM)
that assumes conditional distribution of each variable given its causes (i.e., its mechanism)
does not inform or influence the other conditional distributions~\citep{causality,elements_ci}.
The invariance principle is also induced from the ICM assumption.
Once proper assumptions about the underlying data generation process
via Structural Causal Models (SCM) are established,
it is promising to apply the invariance principle to machine learning models for finding
an invariant representation about the causal relationship between the underlying causes and the label~\citep{inv_principle,irmv1}.
Consequently, models built upon the invariant representation can generalize to unseen environments or
domains with guaranteed performance~\citep{inv_principle,causal_transfer,irmv1,groupdro,meta-transfer,ood_max_inv,domainbed,vrex,env_inference,ib-irm}.
The arguably first formulation of invariance principle was introduced by~\citet{inv_principle}.
\citet{irmv1} propose a novel formulation of learning causal invariance in representation learning, i.e., IRM,
show how it connects with existing areas such as  distributional robust optimization~\citep{dro} and generalization~\citep{memorization},
and prove its effectiveness in addressing PIIF spurious correlations (Assumption~\ref{CH:CIGA:assump:scm_piif_appdx}).
However, in practice, both PIIF and FIIF (Assumption~\ref{CH:CIGA:assump:scm_fiif_appdx}) can appear in data, while IRM can fail in these cases~\citep{aubin2021linear,failure_modes}.
\citet{ib-irm} then propose to add information bottleneck criteria into the IRM formulation to address the issue.
However, their results are restricted to linear regime and also require environment partitions to distinguish the sources of distribution shifts.
Recently, \citet{env_inference} and \citet{zin} propose new OOD objectives to relieve the needs for environment partitions, but limited to PIIF spurious types and linear regime.
Besides, \citet{BayesianIRM} identify the overfitting problem as a key challenge when applying IRM on large neural networks. \citet{SparseIRM} propose to alleviate this problem by imposing sparsity constrain.

In parallel invariant learning approaches,
\citet{groupdro} propose to regularize the worst group in group distributionally robust optimization (GroupDro).
\citet{cnc} propose a contrastive approach to tackle GroupDro when the group partitions are not available.
However, minimizing the gap between worst group risk and averaged risk can not yield a OOD generalizable predictors in our circumstances.
Besides, traditional approaches to tackle OOD generalization also include Domain Adaption, Transfer Learning and Domain Generalization\citep{causal_transfer,domain_gen_inv,DANN,CORAL,deep_DG,DouCKG19,causal_matching,DG_survey},
which aim to learn the class conditional invariant representation shared across source domain and target domain.
However, they all require a stronger assumption on the availability of target domain data or the ground truth predictors~\citep{domainbed,ib-irm},
hence are not able to yield predictors with OOD generalization guarantees.
We refer interested readers to~\citet{seven_tools_causality,causality4ml,towards_causality} for an in-depth understanding,
and \citet{domainbed,ib-irm} for a thorough overview.

\subsection{More discussions on connections of \ciga with existing work}
\label{CH:CIGA:sec:good_connection_appdx}

Although primarily serving for graph OOD generalization
problem, our theory complements the identifiability study on graphs
through contrastive learning, and aligns with the discoveries
in the image domain that contrastive learning learns to isolate
the content ($C$) and style ($S$)~\citep{contrast_inverts,ssl_isolate}.
Moreover, our results also partially explain the success of
graph contrastive learning~\citep{graphcl,contrast_reg,graphcl_auto},
where GNNs may implicitly learn to identify the underlying invariant subgraphs for prediction.

\textbf{On expressivity of graph encoder in \ciga.}
The expressivity of \ciga is essentially constrained by the
encoders embedded for learning graph representations.
During isolating $G_c$ from $G$,
if the encoder can not differentiate two isomorphic graphs
$G_c$ and $G_c\cup G_s^p$ where $G_s^p\subseteq G_s$,
then the featurizer will fail to identify the underlying invariant subgraph.
Moreover, the classifier will also fail if the encoder
can not differentiate two non-isomorphic $G_c$s from different classes.
Thus, adopting more powerful graph representation encoders into \ciga
can improve the OOD generalization.

\textbf{On \ciga and graph information bottleneck.}
Under the FIIF assumption on latent interaction,
the independence condition derived from causal model can
also be rewritten as $Y\ind S|C$ (similar to that in DIR~\citep{dir} as they also focus on FIIF),
which further implies $Y\ind S|\widehat{G}_c$.
Hence it is natural to use Information Bottleneck (IB) objective~\citep{ib}
to solve for $G_c$:
\begin{equation}
	\label{CH:CIGA:good_ib}
	\begin{aligned}
		\min_{f_c,g} & \ R_{G_c}(f_c(\widehat{G}_c)),                                                        \\
		\text{s.t.}  & \ G_c=\argmax_{\widehat{G}_c=g(G)\subseteq G}I(\widehat{G}_c,Y)-I(\widehat{G}_c,\gG), \\
	\end{aligned}
\end{equation}
which explains the success of many existing
works in finding predictive subgraphs through IB~\citep{gib}.
However, the estimation of $I(\widehat{G}_c,G)$
is notoriously difficult due to the complexity of the graph,
which can lead to unstable convergence as observed in our experiments.
In contrast, optimization with contrastive objective in \ciga as Eq.~\ref{CH:CIGA:good_opt_contrast}
induces more stable convergence.

\textbf{On \ciga for node classifications.}
The task of node classification can be viewed as
graph classification based on the ego-graphs of a node,
our analysis and discoveries can be generalized to
node classification.
More specifically, the invariance principle for node classification
can be implemented by identifying an invariant subgraph
from the $K$-hop neighbor graph of each node, and
making predictions based on it, i.e., $Y\ind E|G_c^\ego\subseteq G_u^\ego$ for node $u$~\citep{handle_node}.

\section{Full Structural Causal Models on Graph Generation}
\label{CH:CIGA:sec:full_scm_appdx}
\begin{figure*}[t]
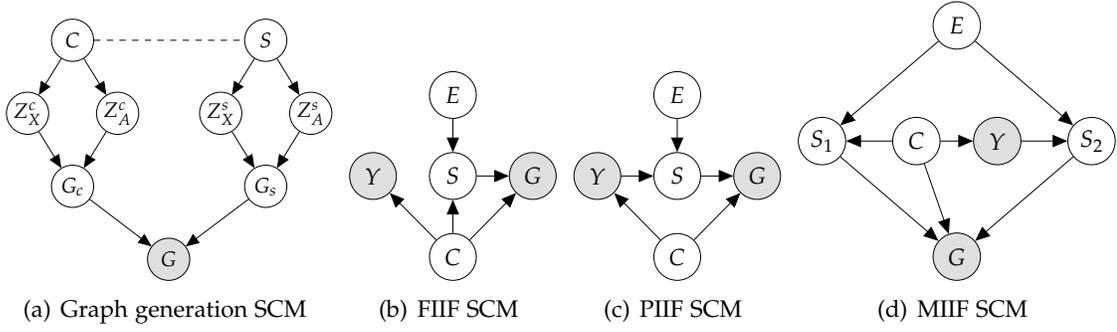

	\centering\hfill
	\subfigure[Graph generation SCM]{\label{CH:CIGA:fig:graph_gen_appdx}
		\resizebox{!}{0.225\textwidth}{\tikz{
				\node[latent] (S) {$S$};%
				\node[latent,left=of S,xshift=-1.5cm] (C) {$C$};%
				\node[latent,below=of C,xshift=-0.75cm,yshift=0.5cm] (ZCA) {$Z_X^c$}; %
				\node[latent,below=of C,xshift=0.75cm,yshift=0.5cm] (ZCX) {$Z_A^c$}; %
				\node[latent,below=of S,xshift=-0.75cm,yshift=0.5cm] (ZSA) {$Z_X^s$}; %
				\node[latent,below=of S,xshift=0.75cm,yshift=0.5cm] (ZSX) {$Z_A^s$}; %
				\node[latent,below=of ZCX,xshift=-0.75cm,yshift=0.5cm] (GC) {$G_c$}; %
				\node[latent,below=of ZSX,xshift=-0.75cm,yshift=0.5cm] (GS) {$G_s$}; %
				\node[obs,below=of GC,xshift=1.6cm,yshift=0.5cm] (G) {$G$}; %
				\edge[dashed,-] {C} {S}
				\edge {C} {ZCX,ZCA}
				\edge {S} {ZSX,ZSA}
				\edge {ZCX,ZCA} {GC}
				\edge {ZSX,ZSA} {GS}
				\edge {GC,GS} {G}
			}}}
	\subfigure[FIIF SCM]{\label{CH:CIGA:fig:scm_fiif_appdx}
		\resizebox{!}{0.18\textwidth}{\tikz{
				\node[latent] (E) {$E$};%
				\node[latent,below=of E,yshift=0.5cm] (S) {$S$}; %
				\node[obs,below=of E,xshift=-1.2cm,yshift=0.5cm] (Y) {$Y$}; %
				\node[obs,below=of E,xshift=1.2cm,yshift=0.5cm] (G) {$G$}; %
				\node[latent,below=of Y,xshift=1.2cm,yshift=0.5cm] (C) {$C$}; %
				\edge {E} {S}
				\edge {C} {Y,G}
				\edge {S} {G}
				\edge {C} {S}
			}}}
	\subfigure[PIIF SCM]{\label{CH:CIGA:fig:scm_piif_appdx}
		\resizebox{!}{0.18\textwidth}{\tikz{
				\node[latent] (E) {$E$};%
				\node[latent,below=of E,yshift=0.5cm] (S) {$S$}; %
				\node[obs,below=of E,xshift=-1.2cm,yshift=0.5cm] (Y) {$Y$}; %
				\node[obs,below=of E,xshift=1.2cm,yshift=0.5cm] (G) {$G$}; %
				\node[latent,below=of Y,xshift=1.2cm,yshift=0.5cm] (C) {$C$}; %
				\edge {E} {S}
				\edge {C} {Y,G}
				\edge {S} {G}
				\edge {Y} {S}
			}}}
	\subfigure[MIIF SCM]{\label{CH:CIGA:fig:scm_miif_appdx}
		\resizebox{!}{0.24\textwidth}{\tikz{
				\node[latent] (E) {$E$};%
				\node[latent,below=of E,xshift=-2cm] (S1) {$S_1$}; %
				\node[latent,below=of E,xshift=-0.6cm] (C) {$C$}; %
				\node[latent,below=of E,xshift=2cm] (S2) {$S_2$}; %
				\node[obs,below=of E,xshift=0.6cm] (Y) {$Y$}; %
				\node[obs,below=of C,xshift=0.6cm] (G) {$G$}; %
				\edge {E} {S1,S2}
				\edge {C} {S1,Y,G}
				\edge {Y} {S2}
				\edge {S1,S2} {G}
			}}}
	\caption{Full SCMs on Graph Distribution Shifts.}
	\label{CH:CIGA:fig:scm_appdx}
\end{figure*}

Due to the space constraints in the main paper, we make some simplifications when giving the SCMs on the graph generation process.
Hence in this section,
supplementary to the graph generation process in Sec.~\ref{CH:CIGA:sec:data_gen}, we provide full
SCMs on the graph generation process in this section as shown in Fig.~\ref{CH:CIGA:fig:scm_appdx}.
Formal descriptions are given as Assumptions~\ref{CH:CIGA:assump:graph_gen_appdx},~\ref{CH:CIGA:assump:scm_fiif_appdx},~\ref{CH:CIGA:assump:scm_piif_appdx},~\ref{CH:CIGA:assump:scm_miif_appdx}.

To begin with, we take a latent-variable model perspective on the graph generation process and assume
that the graph is generated through a mapping $f_\gen:\gZ\rightarrow \gG$,
where $\gZ\subseteq\R^n$ is the latent space and $\gG=\cup_{N=1}^\infty\{0,1\}^N\times \R^{N\times d}$ is the graph space.
Let $E$ denote environments.
Following previous works~\citep{ssl_isolate,ib-irm},
we partition the latent variable from $\gZ$ into an invariant part $C\in\gC=\R^{n_c}$
and a varying part $S\in\gS=\R^{n_s}$, s.t., $n=n_c+n_s$,
according to whether they are affected by $E$.
Similarly in images, $C$ and $S$ can represent content and style
while $E$ can refer to the locations where the images are taken~\citep{camel_example,adv_causal_lens,ssl_isolate}.
While in graphs, $C$ can be the latent variable that controls the generation of functional groups in
a molecule, which can not be affected by the changes of environments, such as species (or scaffolds), experimental environment for
examining the chemical property (or assays)~\citep{drugood}. On the contrary, the other latent variable $S$
inherits environment-specific information thus can further affect the finally generated graphs.
Besides, $C$ and $S$ can have multiple types of interactions at the latent space with environments $E$ and labels $Y$,
which will generate different types of spurious correlations~\citep{ib-irm}.

\begin{assumption}[Graph generation SCM]
	\label{CH:CIGA:assump:graph_gen_appdx}
	\[
		\begin{aligned}
			 & (Z^c_A,Z^c_X):=f_\gen^{(A,X)^c}(C),\ G_c:=f_\gen^{G_c}(Z^c_A,Z^c_X), \\
			 & (Z^s_A,Z^s_X):=f_\gen^{(A,X)^s}(S),\ G_s:=f_\gen^{G_s}(Z^s_A,Z^s_X), \\
			 & G:=f_\gen^G(G_c,G_s).
		\end{aligned}
	\]
\end{assumption}

Specifically, the graph generation process is shown as Fig.~\ref{CH:CIGA:fig:graph_gen_appdx}.
The generation mapping $f_\gen$ is decomposed into $f_\gen^{(A,X)^c}$,$f_\gen^{G_c}$, $f_\gen^{(A,X)^s}$,$f_\gen^{G_s}$ and $f_\gen^G$ to
control the generation of $(Z^c_A,Z^c_X)$, $G_c$, $(Z^s_A,Z^s_X)$, $G_s$, and $G$, respectively.
Given the variable partitions $C$ and $S$ at the latent space $\gZ$,
they control the generation of the adjacency matrix and features for the invariant subgraph $G_c$ and spurious subgraph $G_s$
through two pairs of latent variables $(Z^c_A,Z^c_X)$ and $(Z^s_A,Z^s_X)$, respectively.
$Z^c_A$ and $Z^s_A$ will control the structure-level properties in the generated graphs, such as degrees, sizes, and subgraph densities.
While $Z^c_X$ and $Z^s_X$ mainly control the attribute-level properties in the generated graphs, such as homophily.
Then, $G_c$ and $G_s$ are entangled into the observed graph $G$ through $f_\gen^{G}$.
It can be a simply $\text{JOIN}$ of
a $G_c$ with one or multiple $G_s$,
or more complex generation processes controlled by the latent variables~\citep{sbm,graphon,graphrnn,graphdf,size_gen2}.
Note that since our focus is to describe the potential distribution shifts with SCMs,
in Assumption~\ref{CH:CIGA:assump:graph_gen}, we aim to build a SCM that is compatible to many graph generation processes~\citep{sbm,graphon,graphrnn,graphdf}.
In fact, in Appendix~\ref{CH:CIGA:sec:case_scm_appdx}, we showcase how our SCMs can generalize to specific graph families studied in the literature~\citep{size_gen2,dir,handle_node}, when given more additional knowledge about the graph generation process.
Nevertheless, we believe integrating specific graph generation processes and their implications to improving OOD generalization on graphs would be a promising future direction.

Due to the correlation between $E$ and $G$, graphs collected from different environments can have different structure-level properties such as degrees, graph sizes, and subgraph densities,
as well as feature-level properties such as homophily~\citep{understand_att,size_gen1,size_gen2,hao}.
Meanwhile, all of them can spuriously correlated with the labels depending on how the underlying latent
variables are interacted with each others. The interaction types can be further
divided into two axiom types FIIF and PIIF, as well as the mixed one MIIF.
Previous OOD methods such as GIB~\citep{gib} and DIR~\citep{dir} mainly focus on FIIF case,
while others such as IRM~\citep{irmv1} mainly focuses on the PIIF case.
Evidences show that failing to model either of them when developing the OOD objectives
can have serious performance degenerations in practice~\citep{aubin2021linear,failure_modes}.
That is why we aim to model both of them in our solution.

\begin{assumption}[FIIF SCM]
	\label{CH:CIGA:assump:scm_fiif_appdx}
	\[\begin{aligned}
			Y:= f_\inv(C),\ S:=f_\spu(C,E),\ G:= f_\gen(C,S).
		\end{aligned}\]
\end{assumption}

\begin{assumption}[PIIF SCM]
	\label{CH:CIGA:assump:scm_piif_appdx}
	\[\begin{aligned}
			Y:= f_\inv(C),\ S:=f_\spu(Y,E),\ G:= f_\gen(C,S).
		\end{aligned}\]
\end{assumption}

\begin{assumption}[MIIF SCM]
	\label{CH:CIGA:assump:scm_miif_appdx}
	\[\begin{aligned}
			Y:= f_\inv(C),\ S_1:=f_\spu(C,E),\ S_2:=f_\spu(Y,E),\ G:= f_\gen(C,S_1,S_2).
		\end{aligned}\]
\end{assumption}

As for the interactions between $C$ and $S$ at the latent space,
we categorize the interaction modes into Fully Informative Invariant Features (FIIF, Fig.~\ref{CH:CIGA:fig:scm_fiif_appdx}), and Partially Informative Invariant Features (PIIF, Fig.~\ref{CH:CIGA:fig:scm_piif_appdx}),
depending on whether the latent invariant part $C$
is fully informative about label $Y$, i.e., $(S,E)\ind Y|C$.
It is also possible that FIIF and PIIF are entangled into a Mixed Informative Invariant Features (MIIF,Fig.~\ref{CH:CIGA:fig:scm_miif_appdx}).
We follow ~\citet{irmv1,ib-irm} to formulate the SCMs for FIIF and PIIF, where we omit noises for simplicity~\citep{causality,elements_ci}.
Since MIIF is built upon FIIF and PIIF, we will focus on the axiom interaction modes (FIIF and PIIF) in this paper,
while most of our discussions can be extended to MIIF or more complex interactions built upon FIIF and PIIF.

Among all of the interaction modes,
$f_\gen$ corresponds to the graph generation process in Assumption~\ref{CH:CIGA:assump:graph_gen_appdx}.
$f_\spu$ is the mechanism describing how $S$ is affected by $C$ and $E$ at the latent space.
In FIIF, $S$ is directly controlled by $C$
while in PIIF,
indirectly controlled by $C$ through $Y$,
which can exhibit different behaviors in practice~\citep{ib-irm,failure_modes}.
Additionally, in MIIF, $S$ is further partitioned into $S_1$ and $S_2$ depending on whether it is directly or indirectly controlled by $C$, respectively.
Moreover, $f_\inv:\gC\rightarrow\gY$ indicates the labeling process,
which assigns labels $Y$ for the corresponding $G$ merely based on $C$.
Consequently, $\gC$ is better clustered than $\gS$ when given $Y$~\citep{cluster_assump,cluster_assump2,causality4ml,towards_causality},
which also serves as the necessary separation assumption for a classification task~\citep{svm1,svm2,lda}.
\begin{assumption}[Latent Separability]
	\label{CH:CIGA:assump:latent_sep_appdx}$H(C|Y)\leq H(S|Y)$.
\end{assumption}

\subsection{Discussions on specific cases of the SCMs}
\label{CH:CIGA:sec:case_scm_appdx}
Although our primary focus in this work is to characterize general graph distribution shifts that could happen in practice without any additional knowledge about the underlying graph family, and derive the corresponding solutions, our SCMs (Fig.~\ref{CH:CIGA:fig:scm_appdx}) can generalize to specific cases studied in previous works, when incorporating more inductive biases about the underlying graph family~\citep{size_gen2,dir,handle_node}.

\begin{figure}[ht]
	\centering
	\subfigure[$\gG$-Gen. SCM]{\label{CH:CIGA:fig:graph_gen_case_appdx}
		\resizebox{!}{0.2\textwidth}{\tikz{
				\node[latent] (S) {$S$};%
				\node[latent,left=of S,xshift=0.5cm] (C) {$C_W$};%
				\node[latent,below=of C,xshift=-0.5cm,yshift=0.5cm] (GC) {$G_c$}; %
				\node[latent,below=of S,xshift=0.5cm,yshift=0.5cm] (GS) {$G_s$}; %
				\node[obs,below=of GC,xshift=1.05cm,yshift=0.5cm] (G) {$G$}; %
				\edge[dashed,-] {C} {S}
				\edge {C} {GC}
				\edge {S} {GS}
				\edge {GC,GS} {G}
			}}}
	\hfill
	\subfigure[FIIF SCM]{\label{CH:CIGA:fig:scm_fiif_case_appdx}
		\resizebox{!}{0.2\textwidth}{\tikz{
				\node[latent] (E) {$E$};%
				\node[latent,below=of E,yshift=0.5cm] (S) {$S$}; %
				\node[obs,below=of E,xshift=-1.2cm,yshift=0.5cm] (Y) {$Y$}; %
				\node[obs,below=of E,xshift=1.2cm,yshift=0.5cm] (G) {$G$}; %
				\node[latent,below=of Y,xshift=1.2cm,yshift=0.5cm] (C) {$C_W$}; %
				\edge {E} {S}
				\edge {C} {Y,G}
				\edge {S} {G}
				\edge {C} {S}
			}}}
	\hfill
	\subfigure[Graphon SCM]{\label{CH:CIGA:fig:scm_graphon_case_appdx}
		\includegraphics[width=0.5\textwidth]{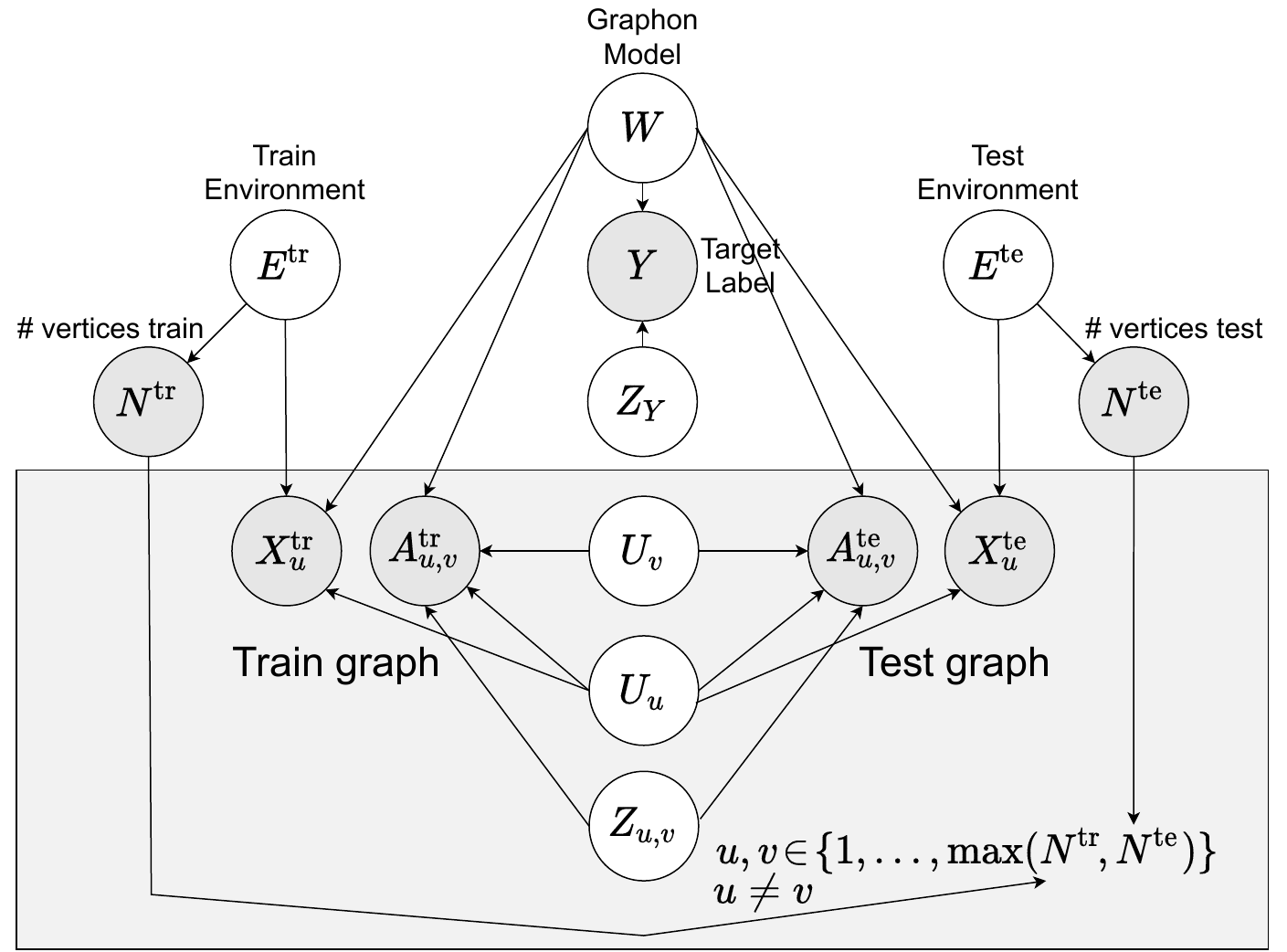}
	}
	\caption{
		Specialized graph generation SCMs when incorporating additional knowledge from graphon~\citep{size_gen2}.}
	\label{CH:CIGA:fig:case_scm_appdx}
\end{figure}

Specifically, we illustrate the specialized SCMs in Fig.~\ref{CH:CIGA:fig:case_scm_appdx} for the SCM studied in~\citep{size_gen2} which assumes the graphs are generated following a graphon model~\citep{graphon}.

When with the additional knowledge about the underlying graph generative model, the graph generation SCM (Fig.~\ref{CH:CIGA:fig:graph_gen_appdx}) and the FIIF SCM (Fig.~\ref{CH:CIGA:fig:scm_fiif_appdx}) together generalizes to the graphon SCM studied in~\citep{size_gen2}. We now give a brief description in the below.

Specifically, shown as in Fig.~\ref{CH:CIGA:fig:graph_gen_case_appdx}, $C$ now is instantiated as a graphon model $C_W\sim \mathbb{P}(C_W)$, where $C_W:[0,1]^2\rightarrow[0,1]$ is a random symmetric measurable function sampled from the set of all symmetric measurable functions~\citep{graphon}. Besides, the label $Y$ is determined according to $C_W$. Then, $C_W$ will further control the generation of the adjcency matrix $G_c=A^c$ through graphon generative process:
\[A^c_{u,v}:=\mathbb{I}(Z_{u,v}>C_W(U_u,U_v)),\ \forall u,v\in V,\]
where $Z_{u,v}$ is an independent uniform noises on $[0,1]$ for each possible edge $(u,v)$ in the graph. Bascially, $Z$ and $U$ are inherited from the graphon SCM as Fig.~\ref{CH:CIGA:fig:scm_graphon_case_appdx}.

On the other hand, as $S$ does not imply any information about $Y$ in this case, it resembles the FIIF SCM (Fig.~\ref{CH:CIGA:fig:scm_fiif_appdx}). In other words, $(S,E)\ind Y|C$ still holds. Moreover, the node attributes $G_s=X^s$ are generated jointly influenced by the environment $E$ and the graphon $C_W$ through $S$:
\[X_v:=f_\gen^s(S),\ S:=f_\spu(E,C_W),\]
which resembles the attribute generation in Fig.~\ref{CH:CIGA:fig:scm_graphon_case_appdx}.

Then, both $G_c$ and $G_s$ are concatenated together. In a simplistic case intuitively, we can regard $G_c$ only contains the edges in $G$ and $G_s$ only contains the node attributes. Since the graphon model mainly controls the edge connection, the edge connection patterns, e.g., motif appearance frequency or subgraph densities, acts as a informative indicator for the label $Y$. In contrast, the node attributes and its numbers would be affected by the environments. A GNN model is prone to the changes of the environments if it overfits to some spurious patterns about the graph sizes or the attributes. While if the GNN model can leverage the connection patterns to make predictions, it remain invariant to the changes of environments, or the spurious patterns such as graph sizes and node attributes, which resembles the solutions derived in~\citep{size_gen1,size_gen2}. Besides, it also partially explains why \ciga can generalize to OOD graphs studied in these works~\citep{size_gen1,size_gen2}.

In addition to the graphon SCM, essentially, the SCM studied in~\citep{dir} resembles the FIIF SCM, and that of~\citep{handle_node} resembles PIIF SCM, which also serves as partial evidence for the superiority OOD generalization performances of \ciga.

\section{More Details about Failure Case Studies in Sec.~\ref{CH:CIGA:sec:limitation_prev}}
\label{CH:CIGA:sec:good_fail_setting_appdx}
In this section, we provide details on failure case studies in Sec.~\ref{CH:CIGA:sec:limitation_prev}.
We first elaborate on the empirical evaluation setting where we construct synthetic graph datasets to probe the
behaviors of existing methods in OOD generalization on graphs.

\subsection{More empirical details about failure case study in Sec.~\ref{CH:CIGA:sec:limitation_prev}}
\label{CH:CIGA:sec:more_emp_fail_case_appdx}

\begin{figure}[H]
	\centering
	\includegraphics[width=0.8\textwidth]{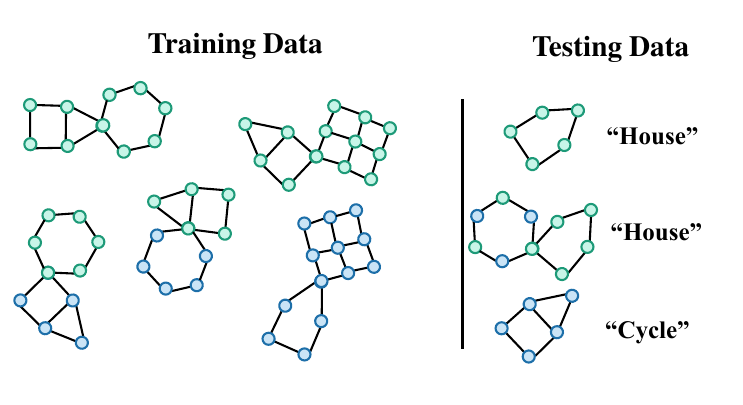}
	\caption[Failure cases of existing methods in OOD generalization on graphs.]{Failure cases of existing methods. GNNs are required to classify whether the graph contains a ``house'' or ``cycle'',
		where the colors represent node features.
		However, distribution shifts in the training exist at both structure level (From left to right: ``house'' mostly co-occurs with a hexagon),
		attribute level (From upper to lower: graphs nodes are mostly green colored if they contain ``house'', or blued colored if they contain ``cycle''),
		and graph sizes, making it hard to capture the invariance.
		\textit{ERM can fail} for leveraging the shortcuts and predicting graphs that have a hexagon or have mostly green nodes as ``house''.
		\textit{IRM can fail} when test data is not sufficiently supported by the training data.}
	\label{CH:CIGA:fig:good_fail_cases_appdx}
\end{figure}

To begin with,
we construct 3-class synthetic datasets based on BAMotif~\citep{pge} and follow~\citet{dir} to inject spurious correlations
between motif graph and base graph during the generation.
In this graph classification task, the model needs to tell which motif the graph contains, e.g., ``House'' or ``Cycle'' motif, as shown in Fig.~\ref{CH:CIGA:fig:good_fail_cases_appdx}.
We inject the distribution shifts in the training data while keeping the test data and validation data without the biases.
For structure-level shifts, we introduce the artificial bias based on FIIF, where the motif and the base graph are spuriously correlated with a probability of various biases.
For mixed shifts, we additionally introduced attribute-level shifts based on FIIF, where all of the node features are spuriously correlated with a probability of various biases.
The number of training graphs is $600$ for each class and the number of graphs in validation and test set is $200$ for each class.
More construction details are given in Appendix~\ref{CH:CIGA:sec:exp_appdx}.

For the GNN encoders, by default,
we use $3$-layer GCN~\citep{gcn} with mean readout, a hidden dimension of $64$, and JK jump connections~\citep{jknet} at the last layer.
During training, we use a batch size of $32$, learning rate of $1e-3$ with Adam optimizer~\citep{adam}, and
batch normalization between hidden layers~\citep{batch_norm}.
Meanwhile, to stabilize the training,
we also use dropout~\citep{dropout} of $0.1$ and early stop the training when the validation accuracy does not increase till $5$ epoch after the first $20$ epochs.
All of the experiments are repeated $5$ times, and the mean accuracy as well as variance are reported and plotted.
When using IRM objective~\citep{irmv1}, as the environment partitions are not available, we generate $2$ environments with random partitions.

\subsection{More discussions about failure cases in Sec.~\ref{CH:CIGA:sec:limitation_prev}}
In Fig.~\ref{CH:CIGA:fig:good_fail_wosize_appdx},~\ref{CH:CIGA:fig:good_fail_size_appdx},~\ref{CH:CIGA:fig:good_fail_piif_wosize_appdx},~\ref{CH:CIGA:fig:good_fail_piif_appdx},
we investigate whether existing training objectives (ERM and IRM), adding more message passing, as well as using expressive GNNs, can improve the OOD generalization ability
on graphs. Here we also provide an additional discussion
in complementary to the discussions on OOD generalization performance of ERM and IRM objectives in Sec.~\ref{CH:CIGA:sec:limitation_prev}.
\begin{myquotation}
	\emph{Can better architectures improve OOD generalization of GNNs?}
\end{myquotation}
\textbf{Adding more message passing turns.} It is a common practice in GNNs to denoise the signals by aggregating more neighbors with higher layers,
or enhance the expressive power with more powerful readout functions~\citep{jknet,gin,p-reg}.
Aggregating neighbor information with more layers to denoise the input signal,
or enhancing the expressivity with more powerful readout functions,
are two common choices in GNNs to improve the generalization ability~\citep{jknet,oversmoothing,gin,p-reg}.
However, in the experiments next, we empirically found that GCNs with more layers and more powerful readout operations are still sensitive to distribution shifts.
In particular, stacking more layers helps denoising certain shifts,
while the OOD performance would drop more sharply when the bias increases.
Intuitively, if the spurious features from nodes cannot be eliminated by the denoising property of a deeper GNN,
they would spread among the whole graph more widely, which in turn leads to stronger spurious correlations.
Besides, the spurious correlations would be more difficult to be disentangled
if there are distribution shifts at both structure-level and attribute-level.
Since the node representations from hidden layers can also encode graph topology features~\citep{gin},
distribution shifts introduced through $Z_A^s$ and $Z_X^s$ will doubly mix at the learned features.
In the worst case, the information about $Z_A^c$ and $Z_X^c$ could be partially covered by or even replaced by $Z_A^s$ and $Z_X^s$.
This will make OOD generalization of message passing GNNs trained through ERM much more difficult or even impossible.
Besides, as the node representations of $1\leq i\leq k$-th layer can also encode graph topology features~\citep{gin}, which, if spuriously correlated with labels through $Z^s_A$ and entangled with part of invariant node features, i.e., $Z^c_X$, in the worst case, can greatly improve the difficulty or even make the OOD generalization impossible for neighbor aggregation GNNs trained with ERM.

\textbf{Using more expressive GNNs.} Previous results on the expressivity of GNNs show that GNNs are limited to
distinguishing isomorphic graphs at most as 1-WL/2-WL test can distinguish~\cite{gin}. After that, many follow-up variants are proposed
to improve the expressivity of GNNs~\citep{wl_goml}. However, if the labels are spuriously correlated with certain subgraphs,
even the GNN has high expressivity can still be prone to distribution shifts.
In a idealistic case, when classifying a graph with a highly expressive GNN, it reduces to the linear or discrete feature case on the Euclidean regime.
In this case, there exists many evidences showing that neural networks can fail to generalize to OOD data without a proper objective~\citep{camel_example,covid19_application,irmv1,groupdro,meta-transfer,vrex,env_inference,ood_max_inv,ib-irm}.
Empirically, we use $k$-GNNs~\citep{kgnn} to verify the intuition and observe similar failures for this provably more expressive GNN as basic GNN variants.

\subsection{More empirical results about failure case study in Sec.~\ref{CH:CIGA:sec:limitation_prev}}

\begin{figure}[ht]
	\centering
	\subfigure[Failures of training objectives.]{
		\includegraphics[width=0.3\textwidth]{Figures/CIGA/ood_failure_wosize.pdf}
	}
	\subfigure[Failures of deeper GNNs.]{
		\includegraphics[width=0.3\textwidth]{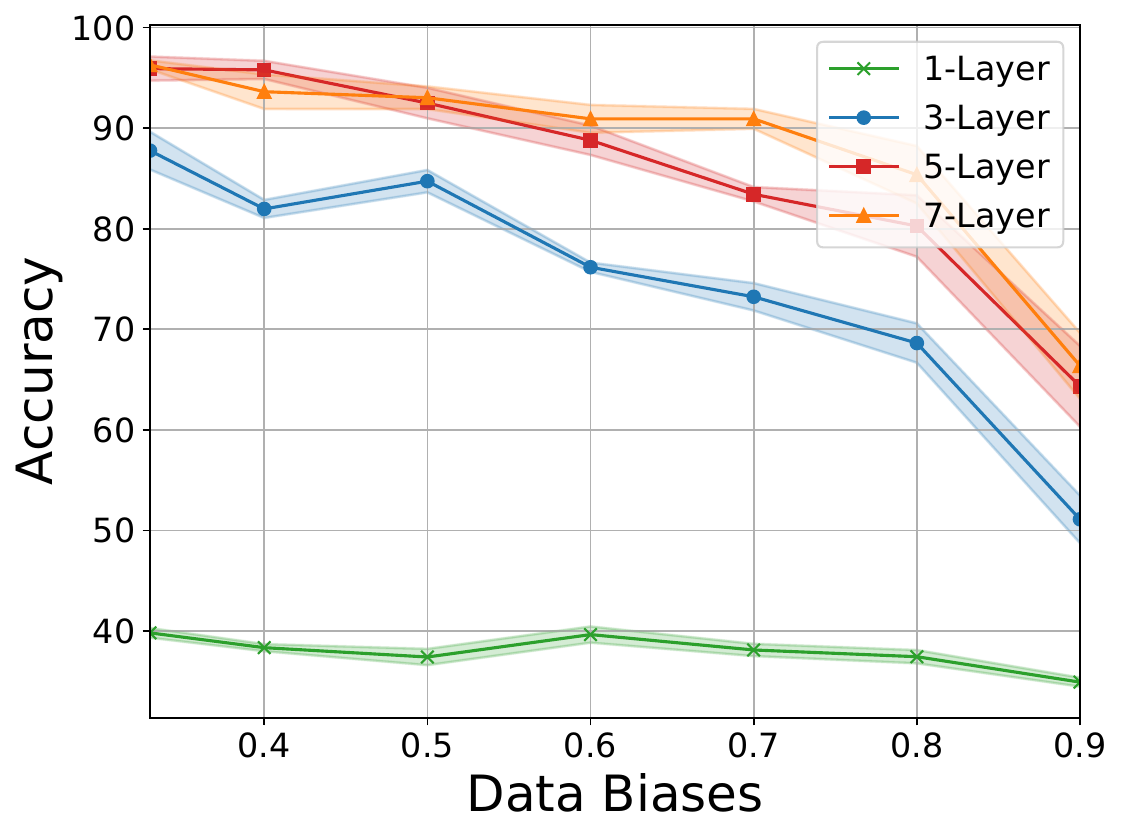}
	}
	\subfigure[Failures of expressive GNNs.]{
		\includegraphics[width=0.3\textwidth]{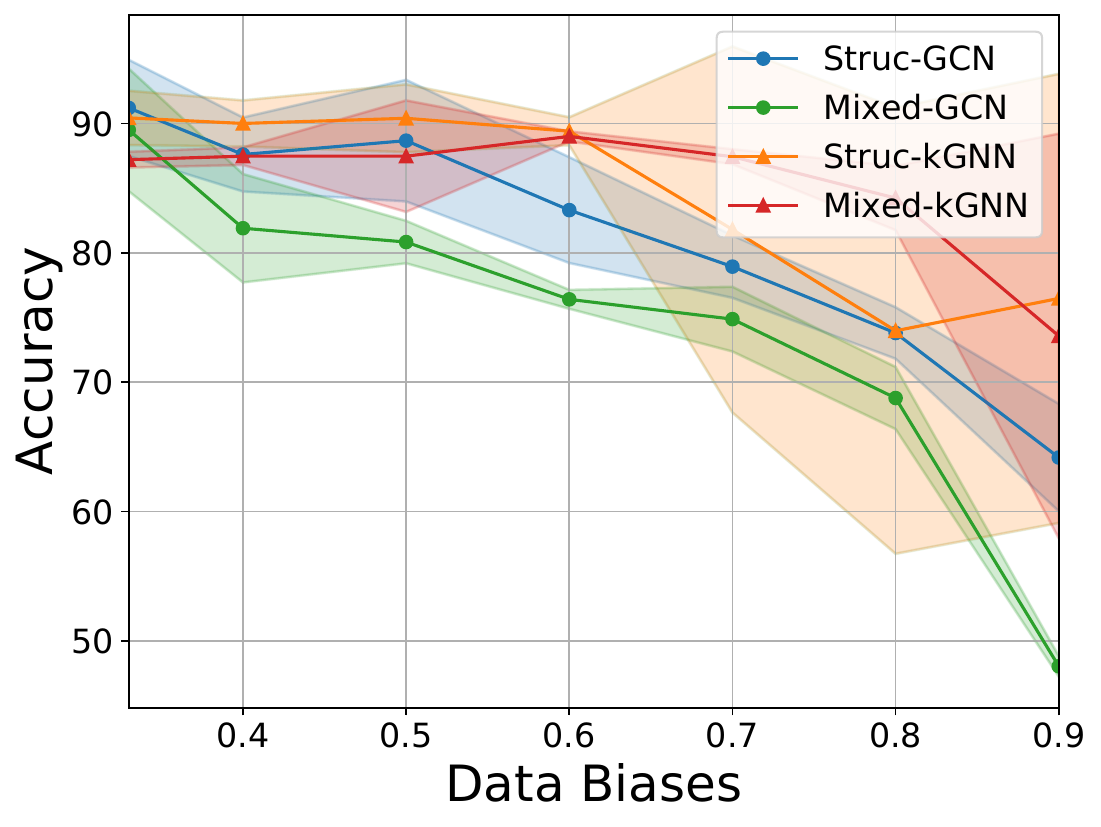}
	}
	\caption{
		Failure of existing methods on SPMotif with FIIF attribute shifts.}
	\label{CH:CIGA:fig:good_fail_wosize_appdx}
\end{figure}

\begin{figure}[ht]
	\centering
	\subfigure[Failures of training objectives.]{
		\includegraphics[width=0.3\textwidth]{Figures/CIGA/ood_failure_size.pdf}
	}
	\subfigure[Failures of deeper GNNs.]{
		\includegraphics[width=0.3\textwidth]{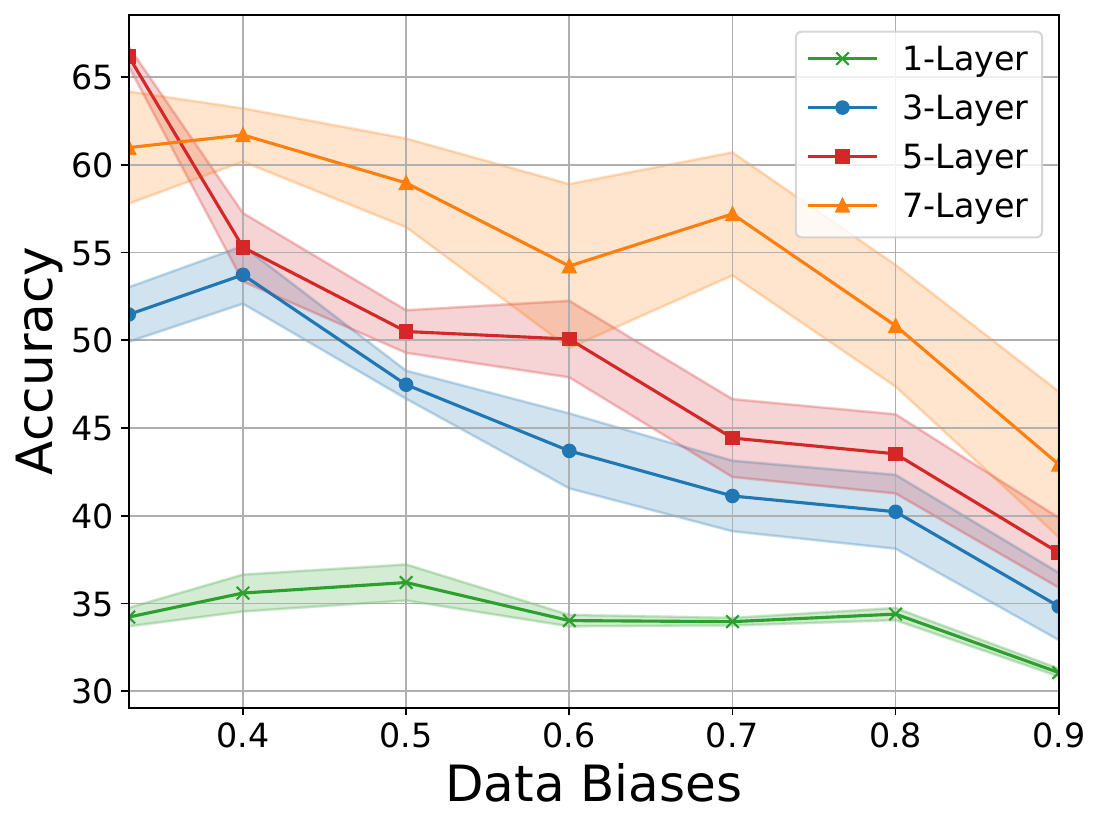}
	}
	\subfigure[Failures of expressive GNNs.]{
		\includegraphics[width=0.3\textwidth]{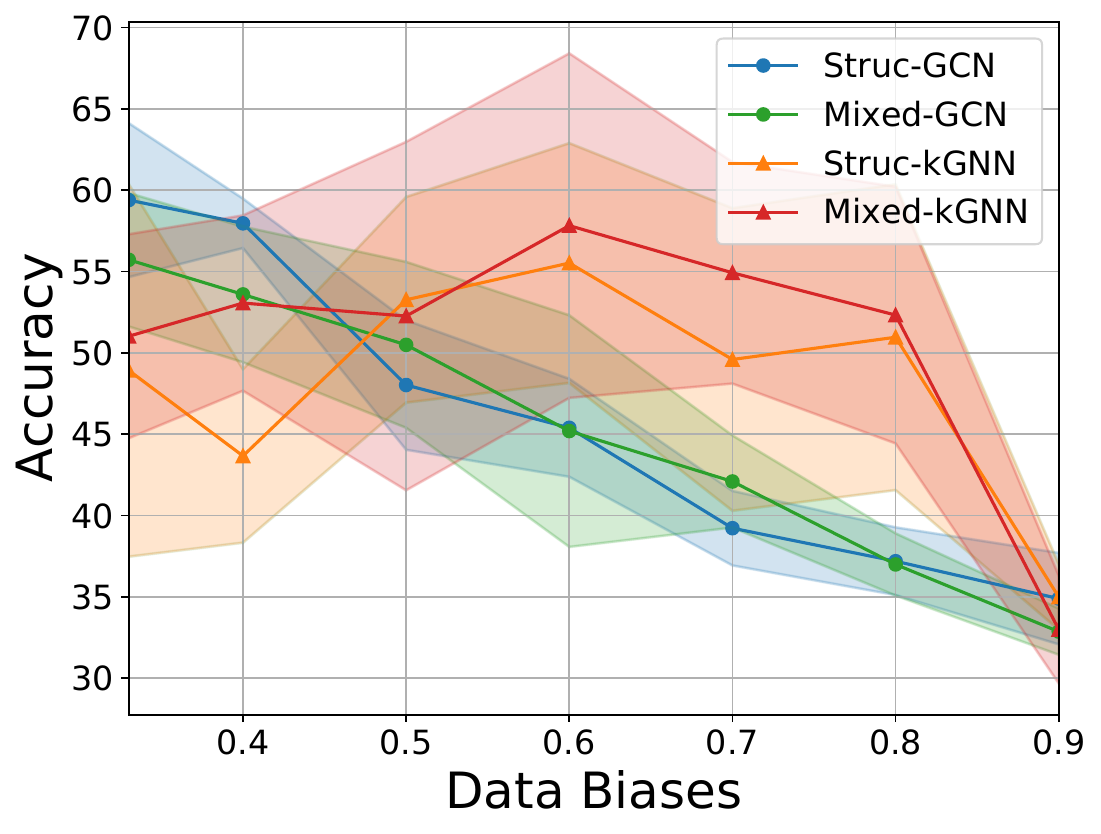}
	}
	\caption{
		Failure of existing methods on SPMotif with FIIF attribute shifts and graph size shifts.}
	\label{CH:CIGA:fig:good_fail_size_appdx}
\end{figure}

\begin{figure}[H]
	\centering
	\subfigure[Failures of training objectives.]{
		\includegraphics[width=0.3\textwidth]{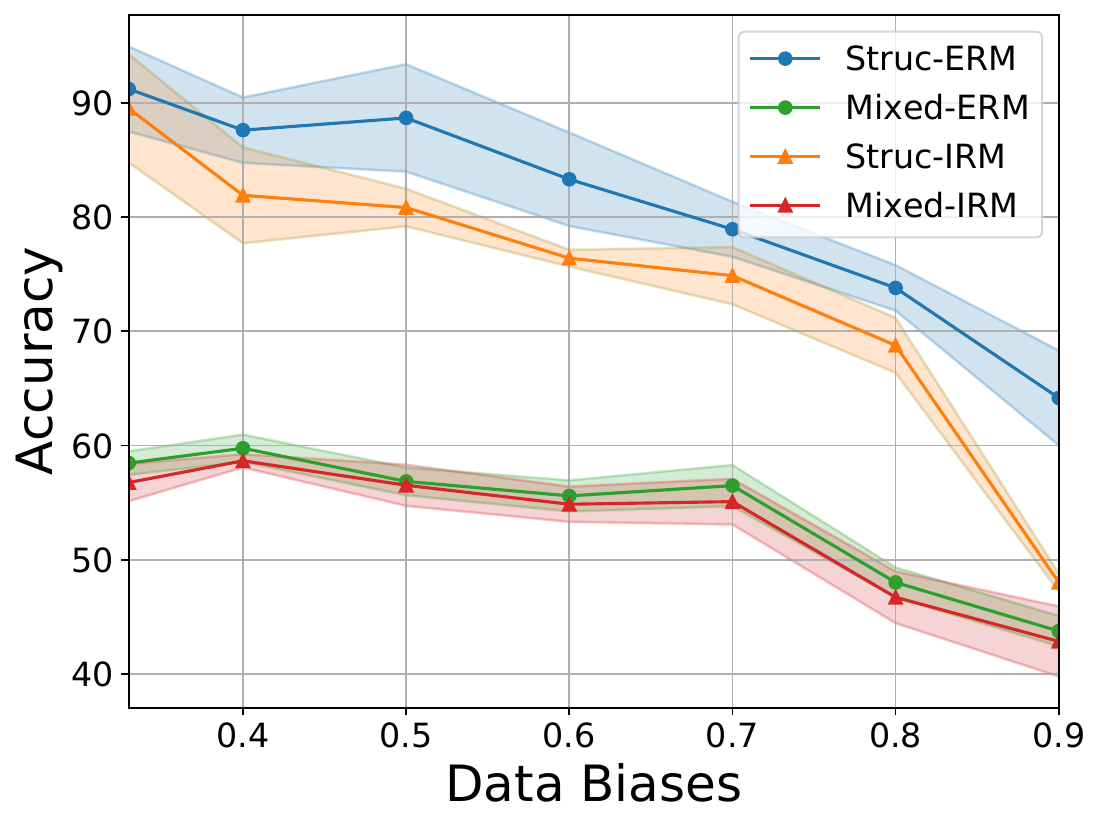}
	}
	\subfigure[Failures of deeper GNNs.]{
		\includegraphics[width=0.3\textwidth]{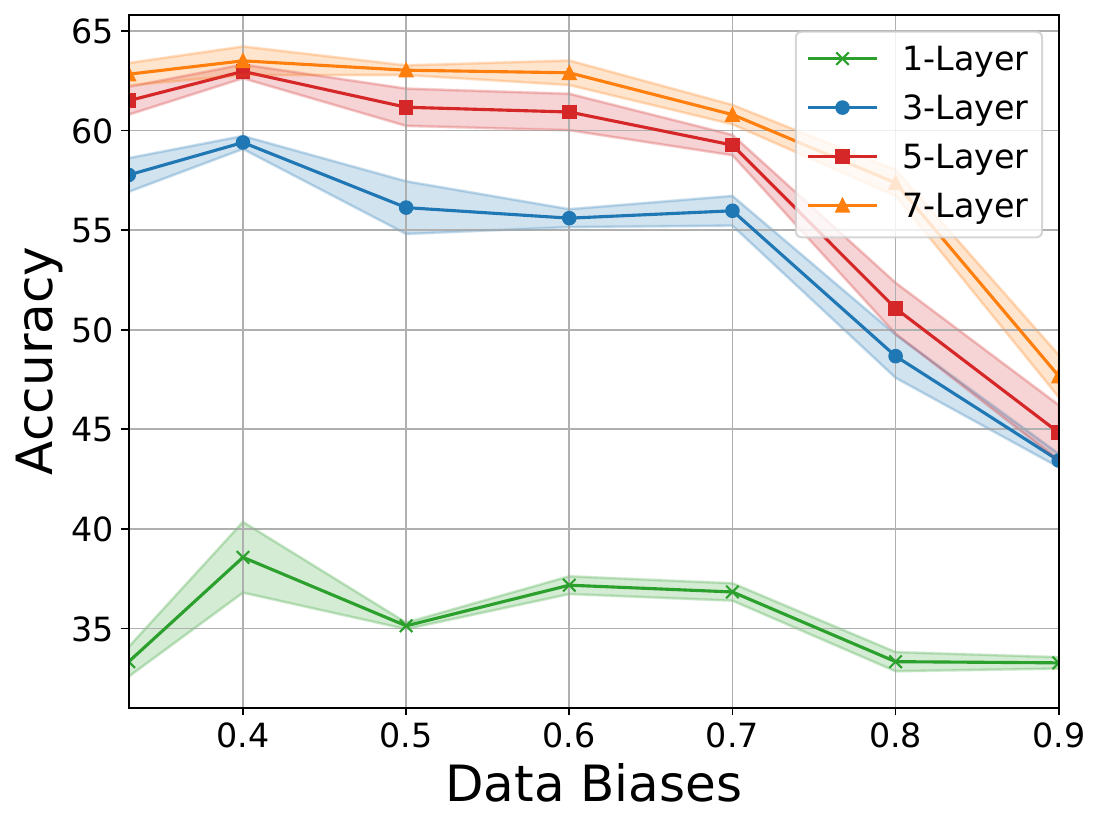}
	}
	\subfigure[Failures of expressive GNNs.]{
		\includegraphics[width=0.3\textwidth]{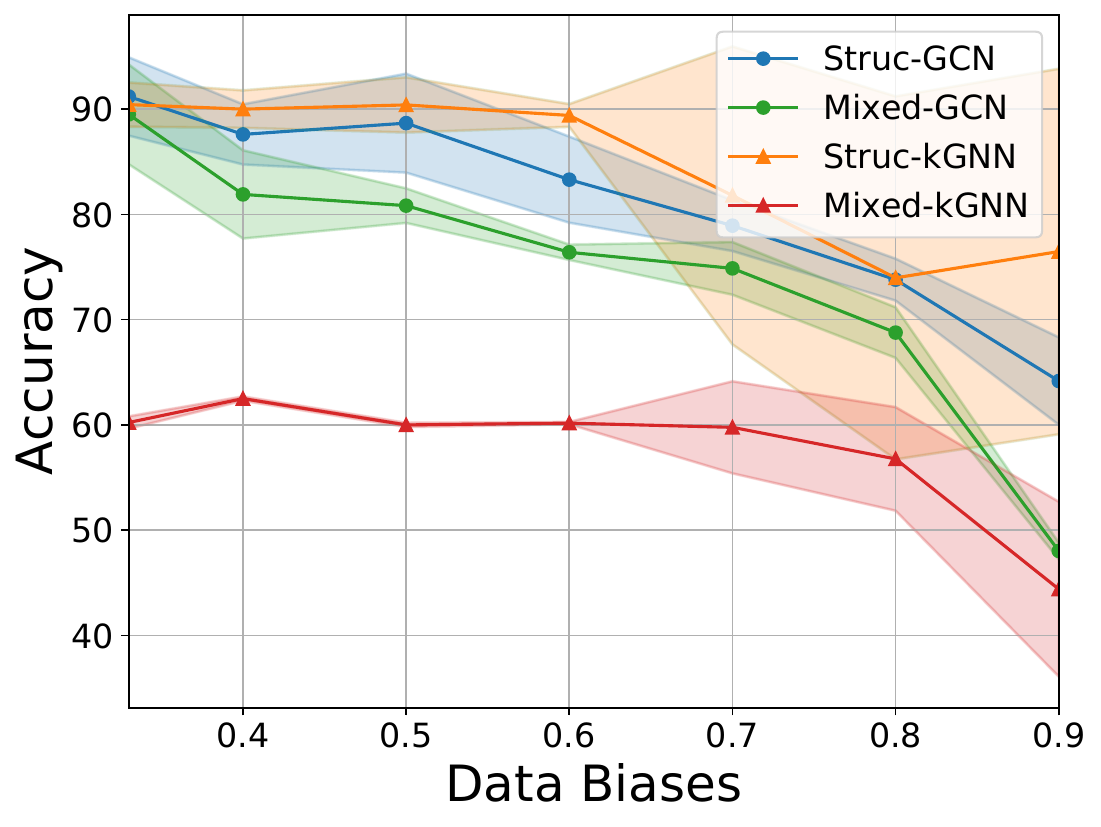}
	}
	\caption{
		Failure of existing methods on SPMotif with PIIF attribute shifts.}
	\label{CH:CIGA:fig:good_fail_piif_wosize_appdx}
\end{figure}

\begin{figure}[H]
	\centering
	\subfigure[Failures of training objectives.]{
		\includegraphics[width=0.3\textwidth]{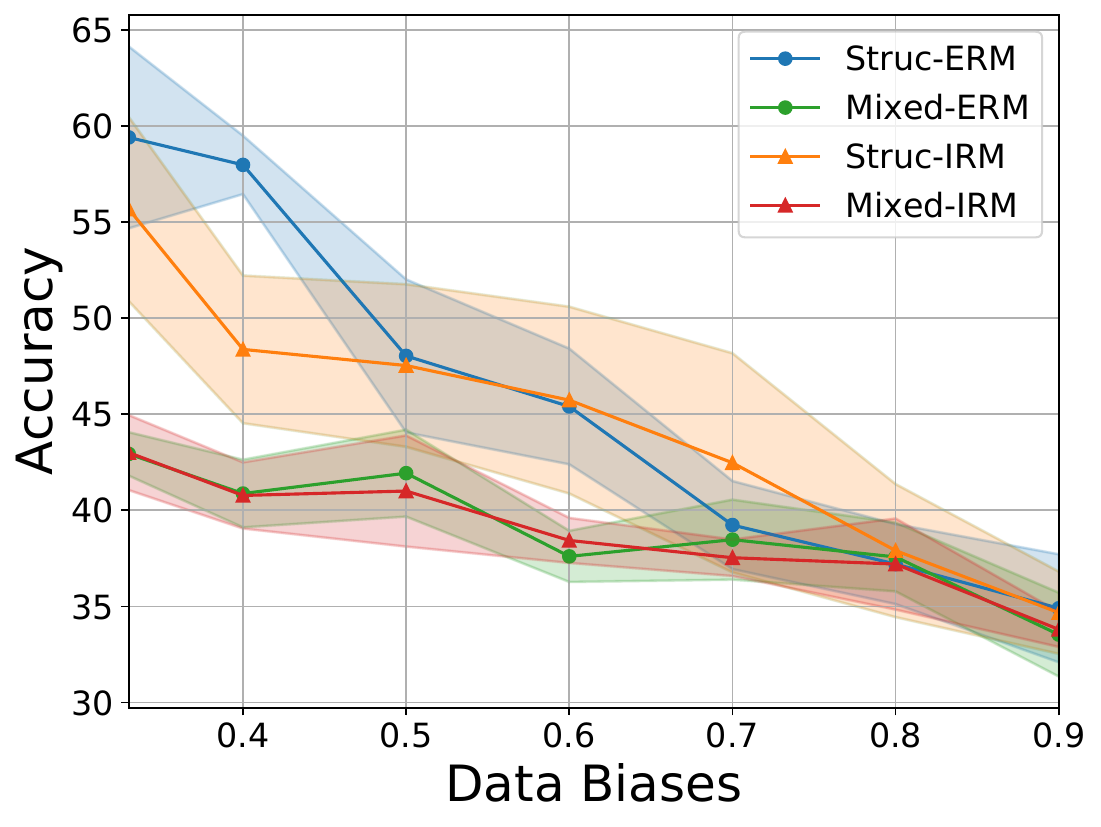}
	}
	\subfigure[Failures of deeper GNNs.]{
		\includegraphics[width=0.3\textwidth]{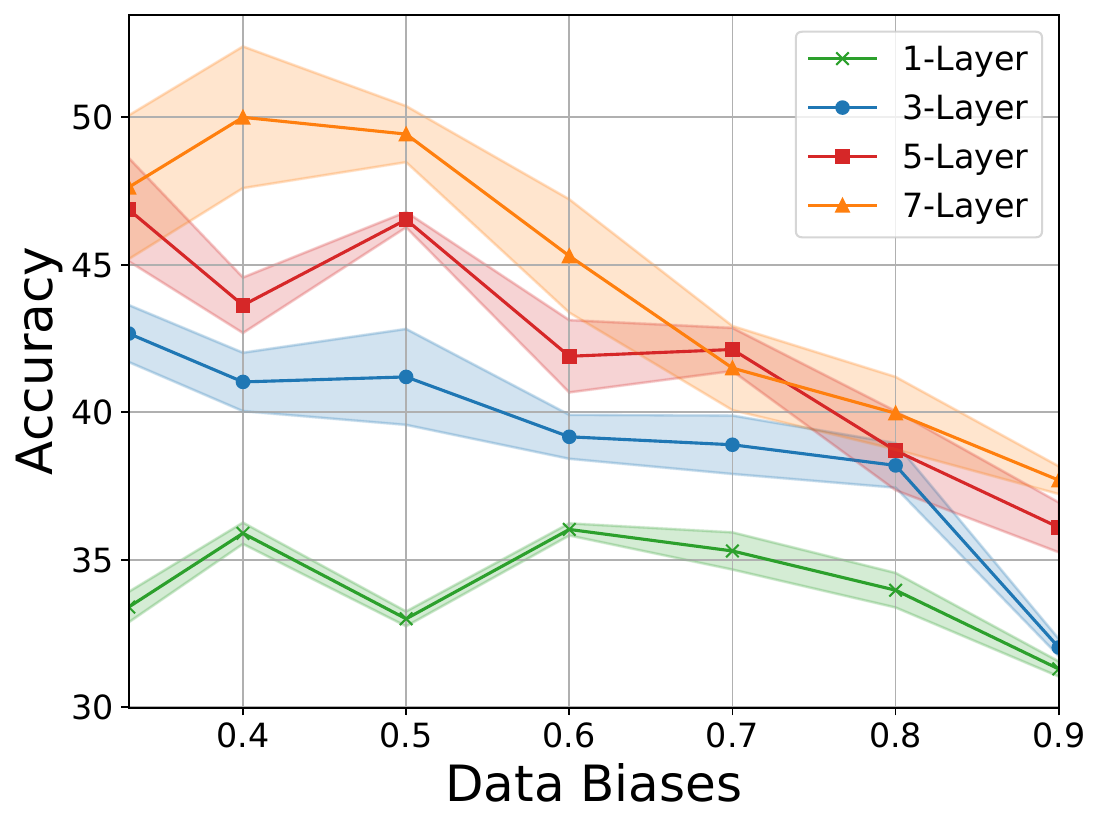}
	}
	\subfigure[Failures of expressive GNNs.]{
		\includegraphics[width=0.3\textwidth]{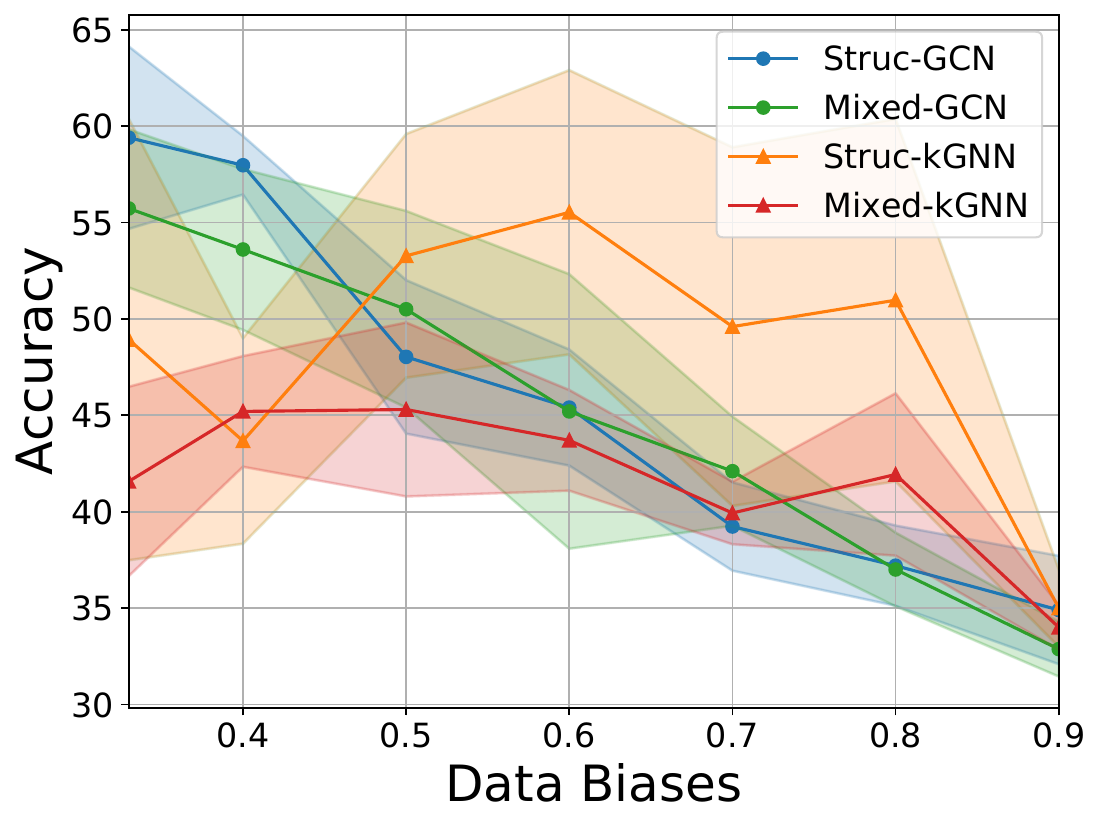}
	}
	\caption{
		Failure of existing methods on SPMotif PIIF attribute shifts with graph size shifts.}
	\label{CH:CIGA:fig:good_fail_piif_appdx}
\end{figure}

To explore the behaviors of aforementioned methods against complicated distribution shifts on graphs,
we first modify construction method in~\citet{dir} to construct dataset for Fig.~\ref{CH:CIGA:fig:good_fail_wosize_appdx}, where only
FIIF structure-level spurious correlations are injected.
Then we also inject FIIF attribute-level shifts, by setting the node attributes to constant vectors which is spuriously correlated with the labels.
Furthermore, in Fig.~\ref{CH:CIGA:fig:good_fail_size_appdx}, graph size shifts are added, which is exactly the SPMotif datasets used in DIR~\citep{dir}.
Besides, in Fig.~\ref{CH:CIGA:fig:good_fail_piif_wosize_appdx}, we can also change the FIIF attribute-level shifts to PIIF attribute-level shifts, where we flip the labels by a probability of $5\%$
and let the flipped label to be spuriously correlated with the node features, following the PIIF SCM in Fig.~\ref{CH:CIGA:fig:scm_appdx}.
Graph size shifts can also be injected in this case, shown as Fig.~\ref{CH:CIGA:fig:good_fail_piif_appdx}. Next, we summarize our findings from the experiments.

\textbf{Observation I: All existing methods are sensitive to distribution shifts.}
From the Fig.~\ref{CH:CIGA:fig:good_fail_wosize_appdx},~\ref{CH:CIGA:fig:good_fail_size_appdx},~\ref{CH:CIGA:fig:good_fail_piif_wosize_appdx},~\ref{CH:CIGA:fig:good_fail_piif_appdx},
we can observe that \emph{all} GNNs are sensitive to distribution shifts.
As the intensity of spurious correlation grows, GNNs are more likely to overfit to shortcuts
presented either in the structure-level or attribute-level, which is similar to general deep learning models~\citep{shortcut_dl}.

\textbf{Observation II: Higher variance also indicates unstable OOD performance.}
Although GNNs show certain robustness against single distribution shifts, e.g., performances do not decrease sharply at the beginning in Fig.~\ref{CH:CIGA:fig:good_fail_wosize_appdx},
when the spurious correlation grows stronger, the OOD performance become more \emph{unstable}, e.g., higher variance.
The reason is that, GNNs sometimes can directly learn about the desired information at some random initializations, since the task is relatively simple
compared to reality. Hence the performance will be highly sensitive to the quality of initialized points at the beginning.
Consequently, the performances from multiple runs would exhibit high variance.
However, when the task becomes more difficult, GNNs will consistently be prone to distribution shifts, and the variance will be smaller,
as shown in experiments (Sec.~\ref{CH:CIGA:sec:exp}).

\textbf{Observation III: Entangling more distribution shifts can degenerate more GNN performance.}
As implied by the graph generation SCMs in Fig.~\ref{CH:CIGA:fig:scm_appdx}, distribution shifts can happen at both structure-level and attribute-level,
and each of them can have different type of spurious correlation with the label.
In Fig.~\ref{CH:CIGA:fig:good_fail_wosize_appdx}, we can find that, when the attribute-level distribution shifts are mixed,
the performance will be worse and more unstable.
When the graph size shifts are mixed, this phenomenon will be more obvious, as shown in Fig.~\ref{CH:CIGA:fig:good_fail_size_appdx}.
This phenomenon also verifies the observations in~\citet{understand_att} that attention mechanism in GNN
is also sensitive to graph size shifts and can hardly learn the desired attention distributions without further guidance.
Moreover, when the structure-level and attribute-level shifts have different spurious correlation types,
i.e., when FIIF structure-level shifts and PIIF attribute-level shifts are both presented,
the performance drop will be more serious, by comparing Fig.~\ref{CH:CIGA:fig:good_fail_wosize_appdx} to Fig.~\ref{CH:CIGA:fig:good_fail_piif_wosize_appdx},
as well as Fig.~\ref{CH:CIGA:fig:good_fail_size_appdx} to Fig.~\ref{CH:CIGA:fig:good_fail_piif_appdx}.

\textbf{Observation IV: Using more powerful architectures can not improve the OOD performance.}
From the sub-figures (b) and (c) in
Fig.~\ref{CH:CIGA:fig:good_fail_wosize_appdx},~\ref{CH:CIGA:fig:good_fail_size_appdx},~\ref{CH:CIGA:fig:good_fail_piif_wosize_appdx},~\ref{CH:CIGA:fig:good_fail_piif_appdx},
we can also observe that neither adding more message passing turns nor using more expressive GNN architectures can
be immune to distribution shifts. On the contrary, they also exhibit similar behaviors like basic GNN architectures.
Specifically, adding more message passing runs show certain robustness against distribution shifts since they are more likely to
learn the desired information during the optimization~\citep{gnn_opt}.
However, when the intensity of spurious correlation grows stronger,
deeper GNNs are more likely to overfit to shortcuts hence their performances will drop more sharply.
On the other hand, using provably more expressive GNN architectures can not improve the OOD performance, either.
In Fig.~\ref{CH:CIGA:fig:good_fail_wosize_appdx},~\ref{CH:CIGA:fig:good_fail_size_appdx},~\ref{CH:CIGA:fig:good_fail_piif_wosize_appdx},~\ref{CH:CIGA:fig:good_fail_piif_appdx}
we use $1$-$2$-$3$-GNN following the algorithm of $k$-GNNs which is provably more expressive than $2$-WL test~\citep{kgnn}.
When there are no graph size shifts, $k$-GNNs will have higher performance at the beginning.
When there are graph size shifts, $k$-GNNs will have a lower initial performance at the beginning.
Then, as the spurious strength grows, $k$-GNNs can suddenly become seriously unstable, though
$k$-GNNs can have higher averaged performance, which reflects unsatisfactory OOD performance as Observation II implies.
When the intensity of spurious correlations grows even stronger, similar to deeper GNNs,
OOD performances of $k$-GNNs will be more unstable and go down to similar level as that of normal GNN architectures.
Hence, it calls for better optimization objectives as well as a suitable architectures to help improve the OOD generalization performance.

Beyond the empirical studies in previous section, we aim to accompany more formal discussions for explaining the failures of existing optimization
objectives and architectures in the next sections.

\subsection{Theoretical discussions for failure case study in Sec.~\ref{CH:CIGA:sec:limitation_prev}}
\label{CH:CIGA:sec:discussion_ood_obj_appdx}

\textbf{A motivating example.}
To begin with, we follow~\citet{ib-irm} to introduce a formal example on the failures of GNNs optimized with ERM or IRM~\citep{erm,irmv1} via a linear binary classification problem:
\begin{definition}[Linear classification structural equation model (FIIF)]
	\label{CH:CIGA:def:linear_fiif_appdx}
	\[
		\begin{aligned}
			 & Y:= (w_\inv^* \cdot C)\oplus N,\ N\sim \text{Ber}(q),\ N\ind (C,S), \\
			 & X\leftarrow S(C,S),
		\end{aligned}
	\]
	where $w_\inv^*\in\R^{n_c}$ with $\norm{w_\inv^*}=1$ is the labeling hyperplane,
	$C\in \R^{n_c},\ S\in\R^{n_s}$ are the corresponding invariant and varying latent variables, $N$ is Bernoulli binary noise with a parameter of $q$ and identical across all environments, $\oplus$ is the $\text{XOR}$ operator, $S$ is invertible.
\end{definition}

Given data generation process as Assumption~\ref{CH:CIGA:assump:graph_gen_appdx}, and latent space interaction as Assumption~\ref{CH:CIGA:assump:scm_fiif_appdx} or ~\ref{CH:CIGA:assump:scm_piif_appdx}, and strictly separable invariant features~\ref{CH:CIGA:assump:latent_sep},
consider a $k$-layer linearized GNN $\rho \circ h$ using $\text{mean}$ as $\text{READOUT}$ for binary graph classification,
if $\cup_{e\in\envtest}\text{supp}(\sP^e)\not\subseteq\cup_{e\in\envtrain}\text{supp}(\sP^e)$:
\begin{enumerate}[label=(\roman*),nosep]
	\item  For graphs features  generated as Definition~\ref{CH:CIGA:def:linear_fiif_appdx},
	      $\rho \circ h$ optimized with ERM or IRM will fail to generalize OOD (Eq.~\ref{CH:CIGA:ood}) almost surely;
	\item For graphs with more than two nodes, globally same node features generated as Definition~\ref{CH:CIGA:def:linear_fiif_appdx}, and graph labels that are the same as global node labels, $\rho \circ h$ optimized with ERM or IRM will fail to generalize OOD (Eq.~\ref{CH:CIGA:ood}) almost surely;
\end{enumerate}

For graph classification, if the number of nodes is fixed to one, it covers the linear classification as above. When  $\cup_{e\in\envtest}\text{supp}(\sP^e)\not\subseteq\cup_{e\in\envtrain}\text{supp}(\sP^e)$,
it implies the $S$ from training environments $\envtrain$ does not cover $S$ from testing environments, while $C$ can be covered. Moreover, the condition of strictly separable training data now can be formulated as $\min_{C\in\cup_{e\in\envtrain}(C\subseteq G^e)}\text{sgn}(w^*_\inv\cdot C)(w^*_\inv \cdot C)>0$. Recall that ERM trains the model by minimizing the empirical risk (e.g., 0-1 loss) over all training data, and IRM formulates OOD generalization as:
\begin{equation}
	\label{CH:CIGA:irmv1_formula_appdx}
	\begin{aligned}
		\min_{\theta, f_c} & \frac{1}{|\envtrain|}\sum_{e\in\envtrain}R^e(\rho \circ h)                   \\
		\text{s.t.}\       & \rho\in\argmin_{\hat{\rho}}R^e(\hat{\rho}\circ h),\ \forall e \in \envtrain.
	\end{aligned}
\end{equation}
However, both ERM and IRM can not enable OOD generalization, i.e., finding the ground truth $w^*_\inv$, following the Theorem 3 from~\citet{ib-irm}:
\begin{theorem}[Insufficiency of ERM and IRM]
	Suppose each $e\in\envall$ follows Definition.~\ref{CH:CIGA:def:linear_fiif_appdx}, $C$ are strictly separable, bounded and satisfy the support overlap between $\envtrain$ and $\envtest$, and $S$ are bounded, if $S$ does not support the overlap, then both ERM and IRM fail at solving the OOD generalization problem.
\end{theorem}

The reason is that, when $C$ from all environments are strictly separable, there can be infinite many Bayes optimal solutions given training data $\{G^e,y^e\}_{e\in\envtrain}$, while there is only one optimal solution that does not rely on $S$. Hence, the probability of generalization to OOD (finding the optimal solution) tends to be $0$ in probability.

As for case (ii), when the GNN uses mean readout to classify more than one node graphs, assuming the graph label is determined by the node label and all of the nodes have the same label that are determined as Definition~\ref{CH:CIGA:def:linear_fiif_appdx}, then GNN optimized with ERM and IRM will also fail because of the same reasons as case (i).

\textbf{Discussions on the failures of previous OOD related solutions.}
First of all, for IRM or similar objectives~\citep{groupdro,vrex,ib-irm,gen_inv_conf} that require environment information or non-trivial data partitions,
they can hardly be applied to graphs due to the lack of such information.
The reason is that obtaining such information can be expensive due to the abstraction of graphs.
Moreover, as proved in Theorem 5.1 of~\citet{risk_irm}, when there is not sufficient support overlap between training environments and testing environments, the IRM or similar objectives can fail catastrophically when being applied to non-linear regime.
The only OOD objective EIIL~\citep{env_inference} that does not require environment labels, also rely on similar assumptions on the support overlap.
We also empirically verify their failing behaviors in our experiments.

Moreover, since part of explainability works also try to find a subset of the inputs for interpretable prediction robustly against distribution shifts.
Here we also provide a discussion for these works.
The first work following this line is $\invrat$~\citep{inv_rat},
which develops an information-theoretic objective (we re-formulate it to suit with OOD generalization problem on graphs):
\begin{equation}
	\label{CH:CIGA:inv_rat_appdx}
	\min_{g,f_c} \max_{f_s}R(f_c \circ g,Y)+\lambda h(R(f_c \circ g,Y)-R_e(f_s\circ g,Y,E)).
\end{equation}
However, it also requires extra environment labels for optimization that are often unavailable in graphs.
Besides, the corresponding assumption on the data generation for guaranteed performance is essentially PIIF if applied to our case,
while it can not provide any theoretical guarantee on FIIF.

We also notice a recent work, $\dir$~\citep{dir}, as a generalization of  $\invrat$ to graphs while studying FIIF spurious correlations,
that proposes an alternative objective that does not require environment label:
\begin{equation}
	\label{CH:CIGA:dir_appdx}
	\min \mathbb{E}_s[R(h,Y|\doop(S=s))]+\\\lambda \var_s(\{R(h,Y|\doop(S=s))\}).
\end{equation}
However, the theoretical justification established for $\dir$ (Theorem 1 to Corollary 1 in~\citet{dir}) essentially depends on the
quality of the generator $g$ which can be prone to spurious correlations.
Thus, $\dir$ can hardly provide any theoretical guarantees when applied to our case, neither for FIIF nor PIIF.
In experiments, we empirically find the unstable and relatively high sensitivity of DIR to spurious correlations,
which verifies our finding.
More details about the empirical behaviors of DIR can be found in Appendix~\ref{CH:CIGA:sec:exp_appdx}.

In contrast to $\dir$, GIB~\citep{gib} which focuses on discovering an informative subgraph for explanation,
essentially can provide theoretical guarantees for FIIF spurious correlations. Theoretically,
(we copy the discussion in Appendix~\ref{CH:CIGA:sec:good_impl_appdx} here to provide an overview of relationships
between GIB and DIR.)
Under the FIIF assumption on latent interaction,
the independence condition derived from causal model can
also be rewritten as $Y\ind S|C$ (similar to that in DIR~\citep{dir} as they also focus on FIIF),
which further implies $Y\ind S|\widehat{G}_c$.
Hence it is natural to use Information Bottleneck (IB) objective~\citep{ib}
to solve for $G_c$ (rewritten for Eq.~\ref{CH:CIGA:good_ib}):
\begin{equation}
	\begin{aligned}
		\min_{f_c,g} & \ R_{G_c}(f_c(\widehat{G}_c)),                                                        \\
		\text{s.t.}  & \ G_c=\argmax_{\widehat{G}_c=g(G)\subseteq G}I(\widehat{G}_c,Y)-I(\widehat{G}_c,\gG), \\
	\end{aligned}
\end{equation}
which explains the success of many existing
works in finding predictive subgraph through IB~\citep{gib}.
However, the estimation of $I(\widehat{G}_c,G)$
is notoriously difficult due to the complexity of graph,
which can lead to unstable convergence as observed in our experiments.
In contrast, optimization with contrastive objective in \ciga as Eq.~\ref{CH:CIGA:good_opt_contrast}
induces more stable convergence.
\subsection{Challenges of OOD generalization on graphs.}
From the aforementioned analysis, we can summarize some key challenges revealed by the failures of both existing optimization
objectives and GNN architectures.
In particular, we are facing two main challenges
a) Distribution shifts on graphs are more complicated where different types of spurious correlations can be entangled via different graph properties;
b) Environment labels are usually not available due to the abstract graph data structure.

\section{Theory and Discussions}
\label{CH:CIGA:sec:theory_appdx}
In this section, we provide proofs for propositions and theorems mentioned in the main paper.

\subsection{More discussions on Definition~\ref{CH:CIGA:def:inv_gnn} for Invariant GNNs}
\label{CH:CIGA:sec:inv_gnn_discuss_appdx}
Definition~\ref{CH:CIGA:def:inv_gnn} is motivated by applying the invariance principle to the established SCMs in Sec.~\ref{CH:CIGA:sec:data_gen}, following the literature of invariant learning~\citep{inv_principle}. In this section, we will present Proposition~\ref{CH:CIGA:thm:causal_minmax_appdx} and Proposition~\ref{CH:CIGA:thm:minmax_causal_appdx}
to illustrate how satisfying the minmax objective in Definition~\ref{CH:CIGA:def:inv_gnn_appdx} is equivalent to identifying the underlying invariant subgraph $G_c$ that contains all of the information about causal factor $C$ in $G$, under both FIIF and PIIF SCMs (Fig.~\ref{CH:CIGA:fig:scm_fiif} and Fig.~\ref{CH:CIGA:fig:scm_piif}).

\begin{definition}[Invariant GNN]
	\label{CH:CIGA:def:inv_gnn_appdx}
	Given a set of graph datasets $\{\dataset^e\}_e$ %
	and environments $\envall$ that follow the same graph generation process in Sec.~\ref{CH:CIGA:sec:data_gen},
	considering a GNN $\rho \circ h$ that has a permutation invariant graph
	encoder $h:\gG\rightarrow\R^h$ and a downstream classifier $\rho:\R^h\rightarrow\gY$,
	$\rho \circ h$ is an invariant GNN if it minimizes the worst case  risk
	among all environments, i.e., $\min \max_{e\in\envall}R^e$.
\end{definition}

First, we show that using the invariant subgraphs $G_c$ to predict $Y$ can satisfy the minmax objective $\min \max_{e\in\envall}R^e$ in Proposition~\ref{CH:CIGA:thm:causal_minmax_appdx}.

\begin{proposition}
	\label{CH:CIGA:thm:causal_minmax_appdx}
	Let $\gG_c$ denote the subgraph space for $G_c$,
	given a set of graphs with their labels $\dataset=\{G^{(i)},y^{(i)}\}_{i=1}^N$ and $\envall$ that
	follow the graph generation process in Sec.~\ref{CH:CIGA:sec:data_gen} (or Sec.~\ref{CH:CIGA:sec:full_scm_appdx}),
	a GNN $\rho\circ h:\gG_c\rightarrow\gY$ that takes $G_c$ of $G$ as the input to predict $Y$, and solves the following objective can generalize to OOD graphs, i.e., solving the minmax objective in Def.~\ref{CH:CIGA:def:inv_gnn_appdx}:
	\[\min_{\theta}R_{\gG_c}(\rho\circ h),\]
	where $R_{\gG_c}$ is the empirical risk over $\{G^{(i)}_c,y^{(i)}\}_{i=1}^N$
	and $G^{(i)}_c$ is the underlying invariant subgraph $G_c$ for $G^{(i)}$.
\end{proposition}

\begin{proof}
	\label{proof:causal_minmax_appdx}
	We establish the proof with independent causal mechanism (ICM) assumption in SCM~\citep{causality,elements_ci}.
	In particular, given the data generation assumption, i.e.,
	for both FIIF (Assumption~\ref{CH:CIGA:assump:scm_fiif}) and PIIF (Assumption~\ref{CH:CIGA:assump:scm_piif}),
	we have: $\forall e,$
	\begin{equation}
		\label{CH:CIGA:icm_appdx}
		\begin{aligned}
			P(Y|C)                     & =P(Y|C,E=e)                     \\
			P(Y|G_c)\sum_{G_c}P(G_c|C) & =P(Y|G_c)\sum_{G_c}P(G_c|C,E=e) \\
			P(Y|G_c)\sum_{G_c}P(G_c|C) & =P(Y|G_c,E=e)\sum_{G_c}P(G_c|C) \\
			P(Y|G_c)                   & =P(Y|G_c,E=e),                  \\
		\end{aligned}
	\end{equation}
	where we use ICM for the first three equalities.
	From Eq.~\ref{CH:CIGA:icm_appdx}, it suffices to know $P(Y|G_c)$ is invariant across different environments.
	Hence, a GNN predictor $\rho\circ h:\gG_c\rightarrow\gY$ optimized with
	empirical risk given $G_c$, essentially minimizes the empirical risk
	across all environments, i.e., $\min R_{\gG_c} = \min \max R^e$.
	Thus, if $\rho\circ h$ solves $\min R_{\gG_c}$, it also solves $\min \max R^e$,
	hence it elicits a invariant GNN predictor according to Definition.~\ref{CH:CIGA:def:inv_gnn_appdx}.
\end{proof}

Besides, we show in Proposition~\ref{CH:CIGA:thm:minmax_causal_appdx} that only using the underlying invariant subgraphs $G_c$ to make predictions can satisfy the minmax objectives. Or equivalently, a GNN predictor solving the minmax objective can only rely on the underlying invariant subgraph $G_c$ to predict $Y$.

\begin{proposition}
	\label{CH:CIGA:thm:minmax_causal_appdx}
	Given a set of graph datasets $\{\dataset^e\}_e$ %
	and environments $\envall$ that follow the same graph generation process in Sec.~\ref{CH:CIGA:sec:data_gen},
	considering a GNN $\rho \circ h$ that has a permutation invariant graph
	encoder $h:\gG\rightarrow\R^h$ and a downstream classifier $\rho:\R^h\rightarrow\gY$,
	$\rho \circ h$ that minimizes the worst case  risk
	among all environments, i.e., $\min \max_{e\in\envall}R^e$,
	can not rely on any part of $G_s$, i.e., $\rho \circ h (G) \ind G_s$.
\end{proposition}

\begin{proof}
	\label{proof:minmax_causal_appdx}
	The proof for Proposition~\ref{CH:CIGA:thm:minmax_causal_appdx} is straightforward. Assuming that $\rho \circ h (G) \not\ind G_s$, as $E$ is influenced by the changes of $E$ through $S$ in both FIIF and PIIF SCMs (Fig.~\ref{CH:CIGA:fig:scm_fiif} and Fig.~\ref{CH:CIGA:fig:scm_piif}), then $\rho \circ h (G) \not\ind E$ as well.
	Consequently, there exists some graph $G$ corresponding to $G_c,G_s^e$ and $\rho \circ h (G)=Y$ under an environment $e$, such that we can always find a proper $e'$ to make $\rho \circ h (G) \neq Y$.
	In contrast, the prediction of a GNN that satisfies $\rho \circ h (G) \ind G_s$ remains invariant against arbitrary changes of environments.
	Thus, it leads to a contradiction to the condition that $\min \max_{e'\in\envall}R^{e'}$. Therefore, a GNN that solves $\min \max_{e\in\envall}R^e$ must satisfy $\rho \circ h (G) \ind G_s$.
\end{proof}

Combining Proposition~\ref{CH:CIGA:thm:causal_minmax_appdx} and Proposition~\ref{CH:CIGA:thm:minmax_causal_appdx}, we are highly motivated to find the underlying invariant subgraphs to make predictions about the original graphs, which converges to Eq.~\ref{CH:CIGA:good_opt}. Tackling Eq.~\ref{CH:CIGA:good_opt} under the unavailability of $E$ brings us two variants of CIGA solutions, as illustrated in Section~\ref{CH:CIGA:sec:good_framework}.

\subsection{Proof for theorem~\ref{CH:CIGA:thm:good_inv_gnn_new} (i)}

\begin{theorem}[{\ciga}v1 Induces Invariant GNNs]
	\label{CH:CIGA:thm:good_inv_gnn_new_appdx}
	Given a set of graph datasets $\{\dataset^e\}_e$ %
	and environments $\envall$ that follow the same graph generation process in Sec.~\ref{CH:CIGA:sec:data_gen},
	assuming that \textup{(a)} $f_\gen^G$ and $f_\gen^{G_c}$ in Assumption~\ref{CH:CIGA:assump:graph_gen} are invertible,
	\textup{(b)} samples from each training environment are equally distributed,
	i.e.,$|\dataset_{\hat{e}}|=|\dataset_{\tilde{e}}|,\ \forall \hat{e},\tilde{e}\in\envtrain$,
	if $\forall G_c, |G_c|=s_c$,
	then a GNN $f_c\circ g$ solves Eq.~\ref{CH:CIGA:good_opt_contrast_v3},
	is an invariant GNN (Def.~\ref{CH:CIGA:def:inv_gnn}).
\end{theorem}

\textit{Proof.}
We re-write the objective as follows:
\label{proof:good_inv_gnn_new_appdx}
\begin{equation}
	\label{CH:CIGA:good_opt_contrast_new_appdx}
	\max_{f_c,g} \ I(\widehat{G}_c;Y), \ \text{s.t.}\
	\widehat{G}_c\in\argmax_{\widehat{G}_c=g(G),|\widehat{G}_c|\leq s_c} I(\widehat{G}_c;\widetilde{G}_c|Y),
\end{equation}
where $\widehat{G}_c=g(G),\widetilde{G}_c=g(\widetilde{G})$ and $\widetilde{G}\sim \sP(G|Y)$,
i.e., $\widetilde{G}$ and $G$ have the same label.

The proof of Theorem~\ref{CH:CIGA:thm:good_inv_gnn_new_appdx} is essentially
to show the estimated $\widehat{G}_c$ through Eq.~\ref{CH:CIGA:good_opt_contrast_new_appdx}
is the underlying $G_c$, then
the maximizer of $I(\widehat{G}_c;Y)$ in Eq.~\ref{CH:CIGA:good_opt_contrast_new_appdx}
can produce most informative and stable predictions about $Y$ based on $G$,
hence is an invariant GNN (Definition.~\ref{CH:CIGA:def:inv_gnn_appdx}).

In the next, we are going to take an information-theoretic view of
the first term $I(\widehat{G}_c;Y)$ and
the second term $I(\widehat{G}_c;\widetilde{G}_c|Y)$ to conclude the proof.
We begin by introducing the following lemma:
\begin{lemma}
	\label{CH:CIGA:thm:env_eq_appdx}
	Given the same conditions as Thm.~\ref{CH:CIGA:thm:good_inv_gnn_new_appdx},
	$I(\widehat{G}_c;Y)$ is maximized if and only if
	$I(\widehat{G}_c;Y|E=e)$ is maximized, $\forall e\in\envtrain$.
\end{lemma}
The proof for Lemma~\ref{CH:CIGA:thm:env_eq_appdx} is straightforward, given the
condition that samples from each training environment are equally distributed,
i.e.,$|\dataset_{\hat{e}}|=|\dataset_{\tilde{e}}|,\ \forall \hat{e},\tilde{e}\in\envtrain$.
Obviously, $\widehat{G}_c=G_c$ is a maximizer of $I(\widehat{G}_c;Y)=I(C;Y)=H(Y)$, since
$f_\gen^c:\gC\rightarrow\gG_c$ is invertible and $C$ causes $Y$.
However, there might be some subset $G_s^p\subseteq G_s$ from the underlying $G_s$
that entail the same information about label, i.e., $I(G_c^p\cup G_s^p;Y)=I(G_c;Y)$
where $\widehat{G}_c=G_c^p\cup G_s^p$ and $G_c^p= G_c\cap\widehat{G}_c$.
For FIIF (Assumption~\ref{CH:CIGA:fig:scm_fiif_appdx}), it can not happen, otherwise,
let $G_c^l= G_c-G_c^p$, then we have:
\begin{equation}
	\begin{aligned}
		I(\widehat{G}_c;Y)=I(G_c^p\cup G_s^p;Y) & =I(G_c^p\cup G_c^l;Y)=I(G_c;Y) \\
		I(G_c^p;Y)+I(G_s^p;Y|G_c^p)             & =I(G_c^p;Y)+I(G_c^l;Y|G_c^p)   \\
		I(G_s^p;Y|G_c^p)                        & =I(G_c^l;Y|G_c^p)              \\
		H(Y|G_c^p)-H(Y|G_c^p,G_s^p)             & =H(Y|G_c^p)-H(Y|G_c^p,G_c^l)   \\
		H(Y|G_c^p)-H(Y|G_c^p,G_s^p)             & =H(Y|G_c^p),                   \\
		H(Y|G_c^l,G_s^p)                        & =0,                            \\
	\end{aligned}
\end{equation}
where the second last equality is due to $C\rightarrow Y$ and the invertibility of $f_\gen^c:\gC\rightarrow\gG_c$ in FIIF,
i.e., $H(Y|C)=H(Y|G_c)=H(Y|G_c^p,G_c^l)=0$.
However, in PIIF,
it can hold since conditioning on $G_c^p,G_s^p$ can not determine $Y$, as $S\not\ind Y|C$. In other words, $G_s\not\ind Y|G_c$, which
means $G_s$ can imply some information about $Y$
that is equivalent to $I(G_c^l;Y|G_c^p)$.

To avoid the presence of spuriously correlated $G_s$ in $\widehat{G}_c$, we will
use the second term to eliminate it:
\begin{equation}
	\label{CH:CIGA:good_opt_contrast_cond_appdx_yg2}
	\begin{aligned}
		\max_{f_c,g} & \ I(\widehat{G}_c;\widetilde{G}_c|Y),                    \\
		             & = H(\widehat{G}_c|Y)-H(\widehat{G}_c|\widetilde{G}_c,Y), \\
	\end{aligned}
\end{equation}
where $\widehat{G}_c=g(G)$, $\widetilde{G}_c=g(\widetilde{G})$ are two positive samples drawn from the same class (i.e., condition on the same $Y$).
Since the all of the training environments are equally distributed,
maximizing $I(\widehat{G}_c;\widetilde{G}_c|Y)$ is
essentially maximizing $I(\widehat{G}_c,E=\hat{e};\widetilde{G}_c,E=\tilde{e}|Y)$, $\forall \hat{e},\tilde{e}\in\envtrain$.
Hence, we have:
\begin{equation}
	\label{CH:CIGA:good_opt_contrast_env_appdx_i}
	\begin{aligned}
		\max_{f_c,g} & \ I(\widehat{G}_c;\widetilde{G}_c|Y),                                                    \\
		             & =I(\widehat{G}_c,E=\hat{e};\widetilde{G}_c,E=\tilde{e}|Y)                                \\
		             & = H(\widehat{G}_c,E=\hat{e}|Y)-H(\widehat{G}_c,E=\hat{e}|\widetilde{G}_c,E=\tilde{e},Y). \\
	\end{aligned}
\end{equation}
We claim Eq.~\ref{CH:CIGA:good_opt_contrast_env_appdx_i}
can eliminate any potential
subsets from $G_s$ in the estimated $\widehat{G}_c$.
\begin{figure}[H]\centering
	\includegraphics[width=0.5\textwidth]{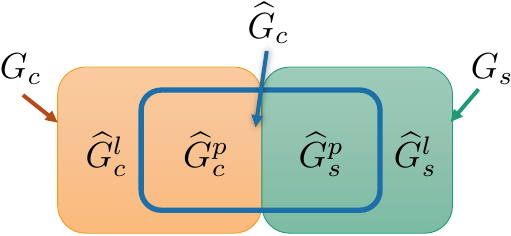}
	\label{CH:CIGA:fig:good_set_venn_1_appdx}
	\caption[Illustration of the notation in the proofs for {\ciga}v1.]{Illustration of the notation. $G_c$ and $G_s$ are two
		disjoint sets. $\widehat{G}_c$ may contain certain subsets from $G_c$ and $G_s$.
		The subsets from $G_c$ and $G_s$ contained in $\widehat{G}_c$ are denoted as $\widehat{G}_c^p$ and $\widehat{G}_s^p$, respectively.
		While the left subsets in $G_c$ and $G_s$ are denoted as $\widehat{G}_c^l$ and $\widehat{G}_s^l$, respectively.}
\end{figure}

Otherwise, suppose there are some subsets $\widehat{G}_s^p\subseteq \widehat{G}_s$
and $\widetilde{G}_s^p\subseteq \widetilde{G}_s$
contained in the estimated $\widehat{G}_c$, $\widetilde{G}_c$, where
$\widehat{G}_s,\widetilde{G}_s$ be the corresponding underlying $G_s$s for $\widehat{G}_c,\widetilde{G}_c$.
Let $\widehat{G}_c^*$ and $\widetilde{G}_c^*$ be the ground truth invariant subgraph $G_c$s
of $\widehat{G}$ and $\widetilde{G}$,
$\widehat{G}_c^l=\widehat{G}_c^*-\widehat{G}_c$
and $\widetilde{G}_c^l=\widetilde{G}_c^*-\widetilde{G}_c$ be the \textbf{l}eft (un-estimated) subsets
from corresponding ground truth $G_c$s,
and $\widehat{G}_c^p=\widehat{G}_c^*-\widehat{G}_c^l$ and
$\widetilde{G}_c^p=\widetilde{G}_c^*-\widetilde{G}_c^l$ be the complement,
or equivalently,
the \textbf{p}artial $\widehat{G}_c^*,\widetilde{G}_c^*$ that are estimated in $\widehat{G}_c,\widetilde{G}_c$, respectively.
We can also define similar counterparts for $G_s$: $\widehat{G}_s^p,\widetilde{G}_s^p$
are the partial $\widehat{G}_s,\widetilde{G}_s$s contained in the estimated $\widehat{G}_c,\widetilde{G}_c$
while $\widehat{G}_s^l,\widetilde{G}_s^l$ are the left subsets $\widehat{G}_s,\widetilde{G}_s$, respectively.

Recall the constraint that $|G_c|=s_c$, hence if $\widehat{G}_c^p\subseteq \widehat{G}_c$,
then a corresponding $\widehat{G}_c^l=\widehat{G}_c^*-\widehat{G}_c^p$ will be
replaced by $\widehat{G}_s^p$ in $\widehat{G}_c$.
In this case, we have:
\begin{equation}
	\label{CH:CIGA:good_opt_contrast_mi_t1_appdx_i}
	\begin{aligned}
		H(\widehat{G}_c,E=\hat{e}|Y) & =H(E=\hat{e}|\widehat{G}_c,Y)+H(\widehat{G}_c|E=\hat{e},Y)                      \\
		                             & =H(\widehat{G}_c^p\cup \widehat{G}_s^p|E=\hat{e},Y)                             \\
		                             & =H(\widehat{G}_c^p|E=\hat{e},Y) +H(\widehat{G}_s^p|\widehat{G}_c^p,E=\hat{e},Y) \\
	\end{aligned}
\end{equation}
where the second equality is due to $E=\hat{e}$ is determined so that $H(E=\hat{e}|\widehat{G}_c,Y)=0$.
Compared Eq.~\ref{CH:CIGA:good_opt_contrast_mi_t1_appdx_i} to that
when $\widehat{G}_c=\widehat{G}_c^*$, we have the entropy change as:
\begin{equation}
	\label{CH:CIGA:good_opt_contrast_env_t1_delta_appdx}
	\begin{aligned}
		\Delta H(\widehat{G}_c,E=\hat{e}|Y) & =H(\widehat{G}_c,E=\hat{e}|Y)-H(\widehat{G}_c^*,E=\hat{e}|Y),                                   \\
		                                    & =H(\widehat{G}_s^p|\widehat{G}_c^p,E=\hat{e},Y)-H(\widehat{G}_c^l|\widehat{G}_c^p,E=\hat{e},Y).
	\end{aligned}
\end{equation}
Let $\epsilon=H(\widehat{G}_s^p|\widehat{G}_c^p,E=\hat{e},Y)$.
In a idealistic setting, when the noise of the generation process $S:=f_\spu(Y,E)$ in PIIF
tends to be $0$, i.e., $\epsilon\rightarrow 0$,
$S$ is determined conditioned on $E,Y$,
hence $G_s$ and any subsets of $G_s$ are all determined.
Then, it suffices to know that in Eq.~\ref{CH:CIGA:good_opt_contrast_env_t1_delta_appdx},
$H(\widehat{G}_s^p|\widehat{G}_c^p,E=\hat{e},Y)=0$ while $H(\widehat{G}_c^l|\widehat{G}_c^p,E=\hat{e},Y)>0$
since $\widehat{G}_c^l$ can not be determined when given $\widehat{G}_c^p,E=\hat{e},Y$.
Thus, when some subset from $G_s$ is included in $\widehat{G}_c$, it will minimize $H(\widehat{G}_c,E=\hat{e}|Y)$.

However in practice, it is usual that $\epsilon>0$.
Therefore, in the next, we will show how $\epsilon=H(\widehat{G}_s^p|\widehat{G}_c^p,E=\hat{e},Y)$
can be cancelled thus leading to a smaller $H(\widehat{G}_c,E=\hat{e}|Y)$, by considering
the second term $H(\widehat{G}_c,E=\hat{e}|\widetilde{G}_c,E=\tilde{e},Y)$.

As for $H(\widehat{G}_c,E=\hat{e}|\widetilde{G}_c,E=\tilde{e},Y)$, without loss of generality,
we can divide all of the possible cases into two:
\begin{enumerate}[label=(\roman*)]
	\item One of $\widehat{G}_c$ and $\widetilde{G}_c$ contains some subset of $G_s$, i.e.,
	      $\widehat{G}_c$ contains some $\widehat{G}_s^p\subseteq \widehat{G}_s$;
	\item Both $\widehat{G}_c$ and $\widetilde{G}_c$ contain some $\widehat{G}_s^p\subseteq \widehat{G}_s$ and $\widetilde{G}_s^p\subseteq \widetilde{G}_s$, respectively.
\end{enumerate}
For (i), we have:
\begin{equation}
	\begin{aligned}
		H(\widehat{G}_c,E=\hat{e}|\widetilde{G}_c,E=\tilde{e},Y) & =H(\widehat{G}_c^p, \widehat{G}_s^p,E=\hat{e}|\widetilde{G}_c,E=\tilde{e},Y)                                                              \\
		                                                         & =H( \widehat{G}_s^p|\widetilde{G}_c,E=\tilde{e},Y,\widehat{G}_c^p,E=\hat{e})+H( \widehat{G}_c^p,E=\hat{e}|\widetilde{G}_c,E=\tilde{e},Y), \\
	\end{aligned}
\end{equation}
Thus, we can write the change of $H(\widehat{G}_c,E=\hat{e}|\widetilde{G}_c,E=\tilde{e},Y)$ between $\widehat{G}_c=\widehat{G}_c^p\cup\widehat{G}_s^p$ and $\widehat{G}_c=\widehat{G}_c^*$ as:
\begin{equation}
	\begin{aligned}
		\Delta H(\widehat{G}_c,E=\hat{e}|\widetilde{G}_c,E=\tilde{e},Y) & =H(\widehat{G}_c,E=\hat{e}|\widetilde{G}_c,E=\tilde{e},Y)-H(\widehat{G}_c^*,E=\hat{e}|\widetilde{G}_c,E=\tilde{e},Y), \\
		                                                                & =H(\widehat{G}_s^p|\widetilde{G}_c,E=\tilde{e},Y,\widehat{G}_c^p,E=\hat{e})                                           \\
		                                                                & \qquad -H( \widehat{G}_c^l|\widetilde{G}_c,E=\tilde{e},Y,\widehat{G}_c^p,E=\hat{e}).
	\end{aligned}
\end{equation}
Combing $\Delta H(\widehat{G}_c,E=\hat{e}|Y)$, we have:
\begin{equation}
	\begin{aligned}
		\Delta I(\widehat{G}_c,E=\hat{e};\widetilde{G}_c,E=\tilde{e}|Y) & =\Delta H(\widehat{G}_c,E=\hat{e}|Y)-\Delta H(\widehat{G}_c,E=\hat{e}|\widetilde{G}_c,E=\tilde{e},Y)                                                \\
		                                                                & =\left\{H(\widehat{G}_s^p|\widehat{G}_c^p,E=\hat{e},Y)-H(\widehat{G}_s^p|\widetilde{G}_c,E=\tilde{e},Y,\widehat{G}_c^p,E=\hat{e})\right\}           \\
		                                                                & \qquad +\left\{-H(\widehat{G}_c^l|\widehat{G}_c^p,E=\hat{e},Y)+H( \widehat{G}_c^l|\widetilde{G}_c,E=\tilde{e},Y,\widehat{G}_c^p,E=\hat{e})\right\}, \\
		                                                                & =-H(\widehat{G}_c^l|\widehat{G}_c^p,E=\hat{e},Y)+H(\widehat{G}_c^l|\widetilde{G}_c,E=\tilde{e},Y,\widehat{G}_c^p,E=\hat{e}),
	\end{aligned}
\end{equation}
where the last equality is because of the independence of $\widehat{G}_s^p$ between
$\widetilde{G}_c,E=\tilde{e}$ conditioned on $Y,E=\hat{e}$.
Since conditioning will lower the entropy for both discrete and continuous variables~\citep{elements_info,network_coding},
we have:
\begin{equation}
	\Delta I(\widehat{G}_c,E=\hat{e};\widetilde{G}_c,E=\tilde{e}|Y) <0,
\end{equation}
which implies the existence of $\widehat{G}_s^p$ in $\widehat{G}_c$ will lower down the second term in Eq.~\ref{CH:CIGA:good_opt_contrast_new_appdx}
for the case (i).

For (ii), we have:
\begin{equation}
	\begin{aligned}
		H(\widehat{G}_c,E=\hat{e}|\widetilde{G}_c,E=\tilde{e},Y) & =H(\widehat{G}_c^p, \widehat{G}_s^p,E=\hat{e}|\widetilde{G}_c^p,\widetilde{G}_s^p,E=\tilde{e},Y) \\
		                                                         & =H( \widehat{G}_s^p|\widetilde{G}_c^p,\widetilde{G}_s^p,E=\tilde{e},Y,\widehat{G}_c^p,E=\hat{e}) \\
		                                                         & \qquad +H(\widehat{G}_c^p,E=\hat{e}|\widetilde{G}_c^p,\widetilde{G}_s^p,E=\tilde{e},Y),          \\
	\end{aligned}
\end{equation}
Similar to (i),
$H( \widehat{G}_s^p|\widetilde{G}_c^p,\widetilde{G}_s^p,E=\tilde{e},Y,\widehat{G}_c^p,E=\hat{e})$
can be cancelled out with $H(\widehat{G}_s^p|\widehat{G}_c^p,E=\hat{e},Y)$.
Then, we have:
\begin{equation}
	\begin{aligned}
		\Delta I(\widehat{G}_c,E=\hat{e};\widetilde{G}_c,E=\tilde{e}|Y) & =\Delta H(\widehat{G}_c,E=\hat{e}|Y)-\Delta H(\widehat{G}_c,E=\hat{e}|\widetilde{G}_c,E=\tilde{e},Y)                                             \\
		                                                                & =-H(\widehat{G}_c^l|\widehat{G}_c^p,E=\hat{e},Y)+H(\widehat{G}_c^l|\widetilde{G}_c^p,\widetilde{G}_s^p,E=\tilde{e},\widehat{G}_c^p,Y,E=\hat{e}).
	\end{aligned}
\end{equation}
Since additionally conditioning on $\widehat{G}_s^p$ in $H(\widehat{G}_c^l,E=\hat{e}|\widetilde{G}_c^p,\widetilde{G}_s^p,E=\tilde{e},Y)$
can not lead to new information about $\widehat{G}_c^l$, we have:
\begin{equation}
	\begin{aligned}
		H(\widehat{G}_c^l|\widetilde{G}_c^p,\widetilde{G}_s^p,E=\tilde{e},\widehat{G}_c^p,Y,E=\hat{e}) & =H(\widehat{G}_c^l|\widetilde{G}_c^p,E=\tilde{e},\widehat{G}_c^p,Y,E=\hat{e}) \\
		                                                                                               & < H(\widehat{G}_c^l|\widehat{G}_c^p,Y,E=\hat{e}),                             \\
	\end{aligned}
\end{equation}
which follows that $\Delta I(\widehat{G}_c,E=\hat{e};\widetilde{G}_c,E=\tilde{e}|Y)<0$.

To summarize, the ground truth $G_c$ is the only maximizer of
the objective (Eq.~\ref{CH:CIGA:good_opt_contrast_new_appdx}),
hence solving for the objective (Eq.~\ref{CH:CIGA:good_opt_contrast_new_appdx}) can elicit an invariant GNN.

\subsection{Proof for theorem~\ref{CH:CIGA:thm:good_inv_gnn_new} (ii)}
\begin{theorem}[{\ciga}v2 Induces Invariant GNNs]
	\label{CH:CIGA:thm:good_inv_gnn_v3_appdx}
	Given a set of graph datasets $\{\dataset^e\}_e$ %
	and environments $\envall$ that follow the same graph generation process in Sec.~\ref{CH:CIGA:sec:data_gen},
	assuming that \textup{(a)} $f_\gen^G$ and $f_\gen^{G_c}$ in Assumption~\ref{CH:CIGA:assump:graph_gen} are invertible,
	\textup{(b)} samples from each training environment are equally distributed,
	i.e.,$|\dataset_{\hat{e}}|=|\dataset_{\tilde{e}}|,\ \forall \hat{e},\tilde{e}\in\envtrain$,
	a GNN $f_c\circ g$ solves Eq.~\ref{CH:CIGA:good_opt_contrast_v3},
	is an invariant GNN (Def.~\ref{CH:CIGA:def:inv_gnn}).
\end{theorem}

\textit{Proof.}
We re-write the objective as follows:
\label{proof:good_inv_gnn_v3_appdx}
\begin{equation}
	\label{CH:CIGA:good_opt_contrast_v3_appdx}
	\begin{aligned}
		\begin{aligned}
			\max_{f_c,g} \  I(\widehat{G}_c;Y)+I(\widehat{G}_s;Y), \ \text{s.t.}\
			 & \widehat{G}_c\in
			\argmax_{\widehat{G}_c=g(G), \widetilde{G}_c=g(\widetilde{G})} I(\widehat{G}_c;\widetilde{G}_c|Y),\
			\\
			 & I(\widehat{G}_s;Y)\leq I(\widehat{G}_c;Y),\ \widehat{G}_s=G-g(G).
		\end{aligned}
	\end{aligned}
\end{equation}
where $\widehat{G}_c=g(G),\widetilde{G}_c=g(\widetilde{G})$ and $\widetilde{G}\sim \sP(G|Y)$,
i.e., $\widetilde{G}$ and $G$ have the same label.

Similar to the proof for Theorem~\ref{CH:CIGA:thm:good_inv_gnn_new_appdx},
to prove Theorem~\ref{CH:CIGA:thm:good_inv_gnn_v3_appdx} is essentially
to show the estimated $\widehat{G}_c$ through Eq.~\ref{CH:CIGA:good_opt_contrast_v3_appdx}
is the underlying $G_c$, hence the minimizer of Eq.~\ref{CH:CIGA:good_opt_contrast_v3_appdx}
elicits an invariant GNN predictor (Definition.~\ref{CH:CIGA:def:inv_gnn_appdx}).

In the next, we also begin with a lemma:
\begin{lemma}
	\label{CH:CIGA:thm:mi_gc_larger_than_gs_appdx}
	Given data generation process as Theorem~\ref{CH:CIGA:thm:good_inv_gnn_v3_appdx}, for both FIIF and PIIF, we have:
	\[I(C;Y)\geq I(S;Y),\]
	hence $I(G_c;Y)\geq I(G_s;Y)$.
\end{lemma}
\begin{proof}[Proof for Lemma~\ref{CH:CIGA:thm:mi_gc_larger_than_gs_appdx}]
	\label{proof:mi_gc_larger_than_gs_appdx}
	For both FIIF and PIIF, Assumption~\ref{CH:CIGA:assump:latent_sep}
	implies that $H(C|Y)\leq H(S|Y)$.
	It follows that $I(C;Y)=H(Y)-H(C|Y)\geq H(Y)-H(S|Y)=I(S;Y)$.
	Then, since $f^{G_c}_\gen:\gC\rightarrow\gG_c$ is invertible,
	we have $I(G_c;Y)=I(C;Y)\geq I(S;Y)\geq I(G_s;Y)$.
\end{proof}
Given Lemma~\ref{CH:CIGA:thm:mi_gc_larger_than_gs_appdx}, we know $\widehat{G}_c$ at least contains some subset of the underlying $G_c$,
otherwise the constraint $I(\widehat{G}_s;Y)\leq I(\widehat{G}_c;Y)$ will be violated since $G_c\subseteq \widehat{G}_s$ in this case.

Assuming there are some subset of $G_s$ contained in $\widehat{G}_c$,
without loss of generality,
we can divide all of the possible cases about $\widehat{G}_c$ into two:
\begin{enumerate}[label=(\roman*)]
	\item $\widehat{G}_c$ only contains a subset of the underlying $G_c$;
	\item $\widehat{G}_c$ contains a subset of the underlying $G_c$ as well as part of the underlying $G_s$;
\end{enumerate}

\begin{wrapfigure}{r}{0.5\textwidth}
	\includegraphics[width=0.5\textwidth]{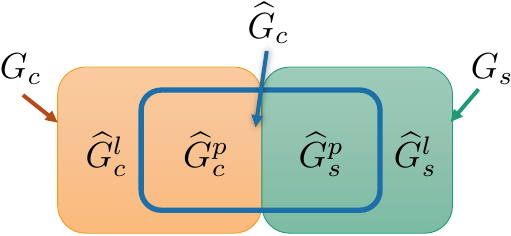}
	\label{CH:CIGA:fig:good_set_venn_2_appdx}
	\caption[Illustration of the notation in the proofs for {\ciga}v2.]{Illustration of the notation for estimated $\widehat{G}_c$ from $G$.
		$G_c$ and $G_s$ are two
		disjoint sets. $\widehat{G}_c$ may contain certain subsets from $G_c$ and $G_s$.
		The subsets from $G_c$ and $G_s$ contained in $\widehat{G}_c$ are denoted as $\widehat{G}_c^p$ and $\widehat{G}_s^p$, respectively.
		While the left subsets in $G_c$ and $G_s$ are denoted as $\widehat{G}_c^l$ and $\widehat{G}_s^l$, respectively.
		Similar notations are also applicable for the estimated $\widetilde{G}_c$ from $\widetilde{G}$.
	}
	\vspace{-0.3in}
\end{wrapfigure}

Before the discussion, let us inherit the notations of subsets of $G_c,G_s$
from the proof for Theorem~\ref{CH:CIGA:thm:good_inv_gnn_new_appdx}:
Let $\widehat{G}_c^*$ and $\widetilde{G}_c^*$ be the ground truth invariant subgraph $G_c$s
of $\widehat{G}$ and $\widetilde{G}$,
$\widehat{G}_c^l=\widehat{G}_c^*-\widehat{G}_c$
and $\widetilde{G}_c^l=\widetilde{G}_c^*-\widetilde{G}_c$ be the \textbf{l}eft (un-estimated) subsets
from corresponding ground truth $G_c$s,
and $\widehat{G}_c^p=\widehat{G}_c^*-\widehat{G}_c^l$ and
$\widetilde{G}_c^p=\widetilde{G}_c^*-\widetilde{G}_c^l$ be the complement,
or equivalently,
the \textbf{p}artial $\widehat{G}_c^*,\widetilde{G}_c^*$ that are estimated in $\widehat{G}_c,\widetilde{G}_c$, respectively.
Similarly,  $\widehat{G}_s^p,\widetilde{G}_s^p$
are the partial $\widehat{G}_s,\widetilde{G}_s$s contained in the estimated $\widehat{G}_c,\widetilde{G}_c$
while $\widehat{G}_s^l,\widetilde{G}_s^l$ are the left subsets $\widehat{G}_s,\widetilde{G}_s$, respectively.

First of all, case (i) cannot hold because,
when maximizing $I(\widehat{G}_c;\widetilde{G}_c|Y)$,
if $\exists \widehat{G}_c^l=\widehat{G}_c^*-\widehat{G}_c$,
as shown in the proof for Theorem~\ref{CH:CIGA:thm:good_inv_gnn_new_appdx},
including $\widehat{G}_c^l$ into $\widehat{G}_c$ can always enlarge $I(\widehat{G}_c;\widetilde{G}_c|Y)$,
while not affecting the optimality of $I(\widehat{G}_s;Y)+I(\widehat{G}_c;Y)$
by re-distributing $\widehat{G}_c^l$ from $\widehat{G}_s$ to $\widehat{G}_c$.
Consequently, $\widehat{G}_c^*$ must be included in $\widehat{G}_c$, i.e., $\widehat{G}_c^*\subseteq \widehat{G}_c$.

As for case (ii),
recall that, by the condition of equally distributed training samples from each training environment,
maximizing $I(\widehat{G}_c;\widetilde{G}_c|Y)$ is
essentially maximizing $I(\widehat{G}_c,E=\hat{e};\widetilde{G}_c,E=\tilde{e}|Y)$, $\forall \hat{e},\tilde{e}\in\envtrain$,
hence, we have:
\begin{equation}
	\label{CH:CIGA:good_opt_contrast_env_appdx_ii}
	\begin{aligned}
		\max_{g,f_c} & \ I(\widehat{G}_c;\widetilde{G}_c|Y),                                                    \\
		             & =I(\widehat{G}_c,E=\hat{e};\widetilde{G}_c,E=\tilde{e}|Y)                                \\
		             & = H(\widehat{G}_c,E=\hat{e}|Y)-H(\widehat{G}_c,E=\hat{e}|\widetilde{G}_c,E=\tilde{e},Y). \\
	\end{aligned}
\end{equation}
We claim Eq.~\ref{CH:CIGA:good_opt_contrast_env_appdx_ii}
can eliminate any potential
subsets in the estimated $\widehat{G}_c$.
Similarly, we have:
\begin{equation}
	\label{CH:CIGA:good_opt_contrast_mi_t1_appdx_ii}
	\begin{aligned}
		H(\widehat{G}_c,E=\hat{e}|Y) & =H(E=\hat{e}|\widehat{G}_c,Y)+H(\widehat{G}_c|E=\hat{e},Y)                      \\
		                             & =H(\widehat{G}_c^*\cup \widehat{G}_s^p|E=\hat{e},Y)                             \\
		                             & =H(\widehat{G}_c^*|E=\hat{e},Y) +H(\widehat{G}_s^p|\widehat{G}_c^*,E=\hat{e},Y) \\
		                             & =H(\widehat{G}_c^*|Y) +H(\widehat{G}_s^p|\widehat{G}_c^*,E=\hat{e},Y)           \\
	\end{aligned}
\end{equation}
where the second equality is due to $E=\hat{e}$ is determined.
Compared to the case that $\widehat{G}_c=\widehat{G}_c^*$, we have:
\begin{equation}
	\begin{aligned}
		\Delta H(\widehat{G}_c,E=\hat{e}|Y) & =H(\widehat{G}_c,E=\hat{e}|Y)-H(\widehat{G}_c^*,E=\hat{e}|Y), \\
		                                    & =H(\widehat{G}_s^p|\widehat{G}_c^*,E=\hat{e},Y).
	\end{aligned}
\end{equation}

Then, as for $H(\widehat{G}_c,E=\hat{e}|\widetilde{G}_c,E=\tilde{e},Y)$,
without loss of generality, we can divide all of the possible cases into two:
\begin{enumerate}[label=(\alph*)]
	\item $\widehat{G}_c$ contains some $\widehat{G}_s^p\subseteq \widehat{G}_s$;
	\item Both $\widehat{G}_c$ and $\widetilde{G}_c$ contain some $\widehat{G}_s^p\subseteq \widehat{G}_s$ and $\widetilde{G}_s^p\subseteq \widetilde{G}_s$, respectively.
\end{enumerate}
For (a), we have:
\begin{equation}
	\begin{aligned}
		H(\widehat{G}_c,E=\hat{e}|\widetilde{G}_c,E=\tilde{e},Y) & =H(\widehat{G}_c^*, \widehat{G}_s^p,E=\hat{e}|\widetilde{G}_c,E=\tilde{e},Y)                                                              \\
		                                                         & =H( \widehat{G}_s^p|\widetilde{G}_c,E=\tilde{e},Y,\widehat{G}_c^*,E=\hat{e})+H( \widehat{G}_c^*,E=\hat{e}|\widetilde{G}_c,E=\tilde{e},Y), \\
	\end{aligned}
\end{equation}
Similarly to the proof for Theorem~\ref{CH:CIGA:thm:good_inv_gnn_new_appdx},
when considering $\Delta I(\widehat{G}_c;\widetilde{G}_c|Y)$, the effects of
$H( \widehat{G}_s^p|\widetilde{G}_c,E=\tilde{e},Y,\widehat{G}_c^*,E=\hat{e})$ is cancelled out by
$H(\widehat{G}_s^p|\widehat{G}_c^*,E=\hat{e},Y)$.
Hence, we have:
\[\Delta I(\widehat{G}_c;\widetilde{G}_c|Y)=0.\]

For (b), we have:
\begin{equation}
	\begin{aligned}
		H(\widehat{G}_c,E=\hat{e}|\widetilde{G}_c,E=\tilde{e},Y) & =H(\widetilde{G}_c^*, \widetilde{G}_s^p,E=\hat{e}|\widetilde{G}_c^*,\widetilde{G}_s^p,E=\tilde{e},Y) \\
		                                                         & =H( \widehat{G}_s^p|\widetilde{G}_c^*,\widetilde{G}_s^p,E=\tilde{e},Y,\widehat{G}_c^*,E=\hat{e})     \\
		                                                         & \qquad+H(\widehat{G}_c^*|\widetilde{G}_c^*,\widetilde{G}_s^p,E=\tilde{e},Y,E=\hat{e}),               \\
	\end{aligned}
\end{equation}
Similarly, $H( \widehat{G}_s^p|\widetilde{G}_c^*,\widetilde{G}_s^p,E=\tilde{e},Y,\widehat{G}_c^*,E=\hat{e})=0$
can also be cancelled out by $H(\widehat{G}_s^p|\widehat{G}_c^*,E=\hat{e},Y)$.
Moreover, for $H(\widehat{G}_c^*|\widetilde{G}_c^*,\widetilde{G}_s^p,E=\tilde{e},Y,E=\hat{e})$,
$\widetilde{G}_s^p$ can not bring no additional information about $\widehat{G}_c^*$,
when conditioning on $\widetilde{G}_c^*,Y,E=\tilde{e}$. Hence, we also have:
\[\Delta I(\widehat{G}_c;\widetilde{G}_c|Y)=0.\]

To summarize, when maximizing $I(\widehat{G}_c;\widetilde{G}_c|Y)$,
including any $\widehat{G}_s^p\subseteq \widehat{G}_s^*$ can not
bring additional benefit while affecting the optimality of $I(\widehat{G}_s;Y)+I(\widehat{G}_c;Y)$.
More specifically, when considering the changes to $I(\widehat{G}_s;Y)+I(\widehat{G}_c;Y)$,
$\forall G_s^p\subseteq G_s$, we have
\[I(G-\widehat{G}_c^*-G_s^p;Y)\leq I(G-\widehat{G}_c^*;Y),\ \forall G_s^p\subseteq G_s,\]
while $I(Y;\widehat{G}_c^*,G_s^p)=I(Y;\widehat{G}_c^*)+I(Y;\widehat{G}_s^p|\widehat{G}_c^*),\ \forall e\in\envtrain$.
Consequently,
\begin{equation}
	\begin{aligned}
		\Delta I(\widehat{G}_s;Y)+I(\widehat{G}_c;Y) & = -I(\widehat{G}_s^p;Y|\widehat{G}_s^l)+I(\widehat{G}_s^p;Y|\widehat{G}_c^*) \\
		                                             & =-I(\widehat{G}_s^p;Y)+I(\widehat{G}_s^p;Y|\widehat{G}_c^*)\leq 0.           \\
	\end{aligned}
\end{equation}

Hence, only the underlying $G_c$ is the solution to Eq.~\ref{CH:CIGA:good_opt_contrast_v3_appdx},
which implies that solving for the objective (Eq.~\ref{CH:CIGA:good_opt_contrast_v3_appdx})
can elicit an invariant GNN.

\section{Details of Prototypical \ciga Implementation}
\label{CH:CIGA:sec:good_impl_appdx}
In fact, the \ciga framework introduced in Sec.~\ref{CH:CIGA:sec:good_framework} can have multiple implementations.
We choose interpretable architectures in our experiments for the purpose of concept verification.
More sophisticated architectures can be incorporated.
Experimental results in Sec.~\ref{CH:CIGA:sec:exp} also demonstrates that, even equipped with basic GNN architectures,
\ciga already has the excellent OOD generalization ability, hence
it is promising to incorporate more advanced architectures from the prosperous GNN literature.

We now introduce the details of the architectures used in our experiments. Recall that \ciga decomposes a GNN model for graph classification into two modules,
i.e., a featurizer: $g:\gG\rightarrow\gG_c$ and a classifier $f_c:\gG_c\rightarrow\gY$.
Specifically, for the implementation of Featurizer, we choose one of the common practices GAE~\citep{gae} for calculating the sampled weights for each edge. More formally, the soft mask is predicted through the following equation:
\begin{equation}\label{CH:CIGA:gae_appdx}\nonumber
	Z=\text{GNN}(G)\in\R^{n\times h},\ M=\sigma(ZZ^T)\in\R^{n\times n}.
\end{equation}

\begin{wrapfigure}{r}{0.5\textwidth}
	\includegraphics[width=0.5\textwidth]{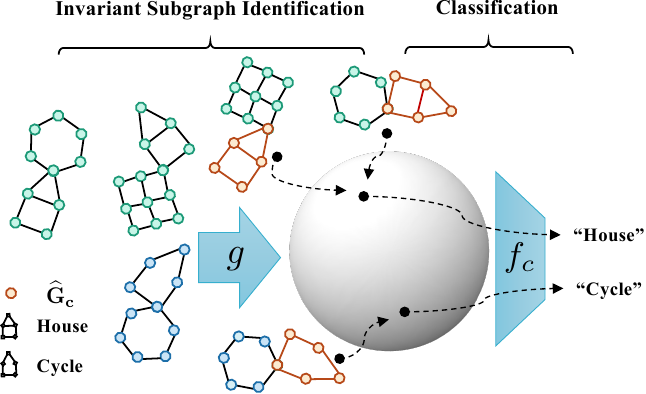}
	\caption[Illustration of \cigafull (\ciga).]{Illustration of \cigafull (\ciga):
		GNNs need to classify graphs based on the specific motif (``House'' or ``Cycle'').
		The featurizer $g$ will extract an (orange colored) subgraph $\widehat{G}_c$ from each input
		for the classifier $f_c$ to predict the label.
		The training objective of $g$ is implemented in a contrastive strategy where
		the distribution of $\widehat{G}_c$ at the latent sphere
		will be optimized to maximize the intra-class mutual information.
		With the identified invariant subgraph $G_c$,
		the predictions made by classifier $f_c$ based on $G_c$ are invariant to distribution shifts;
	}
	\label{CH:CIGA:fig:motivation_appdx}
\end{wrapfigure}

If a sampling ratio $s_c$ is predetermined, we sample $s_c$ of total edges with the largest predicted weights as a soft estimation of $\widehat{G}_c$.
Then, the estimated $\widehat{G}_c$ will be forwarded to the classifier $f_c$ for predicting the labels of the original graph.
Although Theorem~\ref{CH:CIGA:thm:good_inv_gnn_new_appdx} assumes $s_c$ is known, in real applications we do not know the specific $s_c$.
Hence, in experiments, we select $s_c$ according to the validation performance.
To thoroughly study the effects of $I(\widehat{G}_s;Y)$ comparing to {\ciga}v1, we stick to using the same $s_c$ and sampling process for {\ciga}v2,
while  {\ciga}v2 essentially requires less specific knowledge about ground truth $r_c$ hence achieving better empirical performance.
Moreover, once the sampled edges are determined, the classifier GNN can take either the original feature of the input graph
or the learned feature from the featurizer as the new node attributes for $\widehat{G}_c$.
We select the architecture according to the validation performance from some random runs.

For the implementation of the information theoretic objectives,
we will use {\ciga}v2 for elaboration while the implementation of {\ciga}v1 can be obtained via removing the third term from {\ciga}v2.
Recall that {\ciga}v2 has the following formulation:
\begin{equation}
	\label{CH:CIGA:good_opt_contrast_v3_appdx2}
	\begin{aligned}
		\max_{f_c,g} \  I(\widehat{G}_c;Y)+I(\widehat{G}_s;Y), \ \text{s.t.}\
		 & \widehat{G}_c\in
		\argmax_{\widehat{G}_c=g(G), \widetilde{G}_c=g(\widetilde{G})} I(\widehat{G}_c;\widetilde{G}_c|Y),\
		\\
		 & I(\widehat{G}_s;Y)\leq I(\widehat{G}_c;Y),\ \widehat{G}_s=G-g(G).
	\end{aligned}
\end{equation}
where $\widehat{G}_c=g(G),\widetilde{G}_c=g(\widetilde{G})$ and $\widetilde{G}\sim P(G|Y)$, i.e., $\widetilde{G}$ and $G$ have the same label.
In Sec.~\ref{CH:CIGA:sec:good_theory}, we introduce a contrastive approximation for $I(\widehat{G}_c;\widetilde{G}_c|Y)$:
\begin{equation} \label{CH:CIGA:good_opt_contrast_appdx2}
	I(\widehat{G}_c;\widetilde{G}_c|Y) \approx
	\mathbb{E}_{
	\substack{
	\{\widehat{G}_c,\widetilde{G}_c\} \sim \sP_g(G|\gY=Y)\\\
	\{G^i_c\}_{i=1}^{M} \sim \sP_g(G|\gY \neq Y)
	}
	}
	\log\frac{e^{\phi(h_{\widehat{G}_c},h_{\widetilde{G}_c})}}
	{e^{\phi(h_{\widehat{G}_c},h_{\widetilde{G}_c})} +
		\sum_{i}^M e^{\phi(h_{\widehat{G}_c}h_{G^i_c})}},
\end{equation}
where positive samples $(\widehat{G}_c,\widetilde{G}_c)$ are the extracted subgraphs of graphs that have the same label of $G$,
negative samples are those with different labels, $\sP_g(G|\gY=Y)$ is the pushforward distribution of $\sP(G|\gY=Y)$ by featurizer $g$,
$\sP(G|\gY=Y)$ refers to the distribution of $G$ given the label $Y$,
$h_{\widehat{G}_c},h_{\widetilde{G}_c},h_{G^i_c}$ are the graph presentations of the estimated subgraphs,
and $\phi$ is the similarity metric for the graph presentations.
As $M\rightarrow \infty$, Eq.~\ref{CH:CIGA:good_opt_contrast_appdx2} approximates $I(\widehat{G}_c;\widetilde{G}_c|Y)$ which can be regarded as a
non-parameteric resubstitution entropy estimator via the von Mises-Fisher
kernel density~\citep{feat_dist_entropy,vMF_entropy,align_uniform}.

While for the third term $I(\widehat{G}_s;Y)$ and the constraint $I(\widehat{G}_s;Y)\leq I(\widehat{G}_c;Y)$,
a straightforward implementation is to imitate the hinge loss:
\begin{equation} \label{CH:CIGA:good_opt_hinge_appdx2}
	I(\widehat{G}_s;Y)\approx \frac{1}{N} R_{\widehat{G}_s}\cdot \mathbb{I}(R_{\widehat{G}_s}\leq R_{\widehat{G}_c}),
\end{equation}
where $N$ is the number of samples, $\mathbb{I}$ is a indicator function that outputs $1$ when the
interior condition is satisfied otherwise $0$, and
$R_{\widehat{G}_s}$ and $R_{\widehat{G}_c}$ are the empirical risk vector of the predictions
for each sample based on $\widehat{G}_s$ and $\widehat{G}_c$ respectively.
One can also formulate Eq.~\ref{CH:CIGA:good_opt_contrast_v3_appdx2} from game-theoretic perspective~\citep{inv_rat}.

Finally, we can derive the specific loss for the optimization of {\ciga}v2 combining Eq.~\ref{CH:CIGA:good_opt_contrast_appdx2} and Eq.~\ref{CH:CIGA:good_opt_hinge_appdx2}:
\begin{equation} \label{CH:CIGA:good_opt_loss_impl_appdx}
	\begin{aligned}
		 & R_{\widehat{G}_c} +\alpha
		\mathbb{E}_{
		\substack{
		\{\widehat{G}_c,\widetilde{G}_c\} \sim \sP_g(G|\gY=Y)                                             \\\
		\{G^i_c\}_{i=1}^{M} \sim \sP_g(G|\gY \neq Y)
		}
		}
		\log\frac{e^{\phi(h_{\widehat{G}_c},h_{\widetilde{G}_c})}}
		{e^{\phi(h_{\widehat{G}_c},h_{\widetilde{G}_c})} +
		\sum_{i}^M e^{\phi(h_{\widehat{G}_c}h_{G^i_c})}}                                                  \\
		 & +\beta \frac{1}{N} R_{\widehat{G}_s}\cdot \mathbb{I}(R_{\widehat{G}_c}\leq R_{\widehat{G}_s}), \\
	\end{aligned}
\end{equation}
where $R_{\widehat{G}_c},R_{\widehat{G}_s}$ are the empirical risk when using $\widehat{G}_c,\widehat{G}_s$ to predict $Y$ through the classifier.
Typically, we use a additional MLP downstream classifier $\rho_s$ for $\widehat{G}_s$ in the classifier GNN.
$h_{\widehat{G}_c}$ is the graph representation of $\widehat{G}_c$ which can be induced from the GNN encoder either in the featurizer or in the classifier.
$\alpha,\beta$ are the weights for $I(\widehat{G}_c;\widetilde{G}_c|Y)$ and $I(\widehat{G}_s;Y)$, and $\phi$ is implemented as cosine similarity.
The optimization loss for {\ciga}v1 merely contains the first two terms in Eq.~\ref{CH:CIGA:good_opt_loss_impl_appdx}.

The detailed algorithm for \ciga is given in the Algorithm~\ref{CH:CIGA:alg:good_appdx}, assuming the $h_{\widehat{G}_c}$ is obtained via the graph encoder in $f_c$. Fig.~\ref{CH:CIGA:fig:motivation_appdx} also shows a illustration of the working procedure of \ciga.

\begin{algorithm}[tb]
	\caption{Pseudo code for \ciga framework.}
	\label{CH:CIGA:alg:good_appdx}
	\begin{algorithmic}
		\STATE {\bfseries Input:} Training graphs and labels $\train=\{G_i,Y_i\}_{i=1}^N$; learning rate $l$; loss weights $\alpha,\beta$ required by Eq.~\ref{CH:CIGA:good_opt_loss_impl_appdx}; number of training epochs $e$; batch size $b$;
		\STATE Randomly initialize parameters of $g,f_c$;
		\FOR{$i=1$ {\bfseries to} $e$}
		\STATE Sample a batch of graphs $\{G^j,Y^j\}_{j=1}^b$;
		\STATE Estimate the invariant subgraph for the batch: $\{\widehat{G}_c^j\}_{j=1}^b=g(\{G^j,Y^j\}_{j=1}^b)$;
		\STATE Make predictions based the estimated invariant subgraph: $\{\widehat{Y}^j\}_{j=1}^b=f_c(\{\widehat{G}_c^j\}_{j=1}^b)$;
		\STATE Calculate the empirical loss $R_{\widehat{G}_c}$ with $\{\widehat{Y}^j\}_{j=1}^b$;
		\STATE Fetch the graph representations of invariant subgraphs from $f_c$ as $\{h_{\widehat{G}_c^j}\}_{j=1}^b$;
		\STATE Calculate the contrastive loss $R_c$ with Eq.~\ref{CH:CIGA:good_opt_contrast_appdx2}, where positive samples and negative samples are constructed from the batch;
		\STATE Obtain $\widehat{G}_s$
		for the batch: $\{\widehat{G}_c^j\}_{j=1}^b=\{G^j-\widehat{G}_c^j\}_{j=1}^b$;
		\STATE Make predictions based on the $\widehat{G}_s$: $\{\widehat{Y}_s^j\}_{j=1}^b=f_c(\{\widehat{G}_c^j\}_{j=1}^b)$;
		\STATE Calculate the empirical loss $R_{\widehat{G}_s}$ with $\{\widehat{Y}_s^j\}_{j=1}^b$, and weighted as Eq.~\ref{CH:CIGA:good_opt_hinge_appdx2};
		\STATE Update parameters of $g,f_c$ with respect to $R_{\widehat{G}_c}+\alpha R_c+\beta R_{\widehat{G}_s}$ as Eq.~\ref{CH:CIGA:good_opt_loss_impl_appdx};
		\ENDFOR
	\end{algorithmic}
\end{algorithm}

\section{Detailed Experimental Settings}
\label{CH:CIGA:sec:exp_appdx}
In this section, we provide more details about our experimental settings in Sec.~\ref{CH:CIGA:sec:exp}, including the dataset preparation, dataset statistics, implementations of baselines, selection of models and hyperparameters as well as evaluation protocols.

\bgroup
\def\arraystretch{1.2}
\begin{table}[ht]
	\centering
	\caption[Information about the datasets used in experiments of \ciga.]{Information about the datasets used in experiments. The number of nodes and edges are taking average among all graphs. MCC indicates the Matthews correlation coefficient.}\small\sc
	\label{CH:CIGA:tab:datasets_stats_appdx}
	\resizebox{\textwidth}{!}{
		\begin{small}
			\begin{tabular}{lccccccc}
				\toprule
				\textbf{Datasets} & \textbf{\# Training} & \textbf{\# Validation} & \textbf{\# Testing} & \textbf{\# Classes} & \textbf{ \# Nodes} & \textbf{ \# Edges}
				                  & \textbf{  Metrics}                                                                                                                            \\\midrule
				SPMotif           & $9,000$              & $3,000$                & $3,000$             & $3$                 & $44.96$            & $65.67$            & ACC     \\
				PROTEINS          & $511$                & $56$                   & $112$               & $2$                 & $39.06$            & $145.63$           & MCC     \\
				DD                & $533$                & $59$                   & $118$               & 2                   & $284.32$           & $1,431.32$         & MCC     \\
				NCI1              & $1,942$              & $215$                  & $412$               & $2$                 & $29.87$            & $64.6$             & MCC     \\
				NCI109            & $1,872$              & $207$                  & $421$               & $2$                 & $29.68$            & $64.26$            & MCC     \\
				SST5              & $6,090$              & $1,186$                & $2,240$             & $5$                 & $19.85$            & $37.70$            & ACC     \\
				Twitter           & $3,238$              & $694$                  & $1,509$             & $3$                 & $21.10$            & $40.20$            & ACC     \\
				CMNIST-sp         & $40,000$             & $5,000$                & $15,000$            & $2$                 & $56.90$            & $373.85$           & ACC     \\
				DrugOOD-Assay     & $34,179$             & $19,028$               & $19,032$            & $2$                 & $32.27$            & $70.25$            & ROC-AUC \\
				DrugOOD-Scaffold  & $21,519$             & $19,041$               & $19,048$            & $2$                 & $29.95$            & $64.86$            & ROC-AUC \\
				DrugOOD-Size      & $36,597$             & $17,660$               & $16,415$            & $2$                 & $30.73$            & $66.90$            & ROC-AUC \\
				\bottomrule
			\end{tabular}	\end{small}}
\end{table}
\egroup

\begin{table*}[ht]
	\caption[Detailed statistics of selected TU datasets.]{Detailed statistics of selected TU datasets. Table from~\citet{size_gen1,size_gen2}.}
	\label{CH:CIGA:tab:datasets_stats_tu_appdx}
	\begin{small}
		\begin{sc}
			\begin{center}
				\resizebox{\textwidth}{!}{
					\centering
					\begin{tabular}{|l|r|r|r|r|r|r|}
						\cline{2-7}
						\multicolumn{1}{c|}{}   & \multicolumn{3}{c|}{\textbf{NCI1}} & \multicolumn{3}{c|}{\textbf{NCI109}}                                                                                                                          \\
						\cline{2-7}
						\multicolumn{1}{c|}{}   & \textbf{all}                       & \textbf{Smallest} $\mathbf{50\%}$    & \textbf{Largest $\mathbf{10\%}$} & \textbf{all} & \textbf{Smallest} $\mathbf{50\%}$ & \textbf{Largest $\mathbf{10\%}$} \\
						\hline
						\textbf{Class A}        & $49.95\%$                          & $62.30\%$                            & $19.17\%$                        & $49.62\%$    & $62.04\%$                         & $21.37\%$                        \\
						\hline
						\textbf{Class B}        & $50.04\%$                          & $37.69\%$                            & $80.82\%$                        & $50.37\%$    & $37.95\%$                         & $78.62\%$                        \\
						\hline
						\textbf{Num of graphs}  & 4110                               & 2157                                 & 412                              & 4127         & 2079                              & 421                              \\
						\hline
						\textbf{Avg graph size} & 29                                 & 20                                   & 61                               & 29           & 20                                & 61                               \\
						\hline
					\end{tabular}
				}

				\bigskip

				\resizebox{\textwidth}{!}{
					\centering
					\begin{tabular}{|l|r|r|r|r|r|r|}
						\cline{2-7}
						\multicolumn{1}{c|}{}   & \multicolumn{3}{c|}{\textbf{PROTEINS}} & \multicolumn{3}{c|}{\textbf{DD}}                                                                                                                           \\
						\cline{2-7}
						\multicolumn{1}{c|}{}   & \textbf{all}                           & \textbf{Smallest} $\mathbf{50\%}$ & \textbf{Largest $\mathbf{10\%}$} & \textbf{all} & \textbf{Smallest} $\mathbf{50\%}$ & \textbf{Largest $\mathbf{10\%}$} \\
						\hline
						\textbf{Class A}        & $59.56\%$                              & $41.97\%$                         & $90.17\%$                        & $58.65\%$    & $35.47\%$                         & $79.66\%$                        \\
						\hline
						\textbf{Class B}        & $40.43\%$                              & $58.02\%$                         & $9.82\%$                         & $41.34\%$    & $64.52\%$                         & $20.33\%$                        \\
						\hline
						\textbf{Num of graphs}  & 1113                                   & 567                               & 112                              & 1178         & 592                               & 118                              \\
						\hline
						\textbf{Avg graph size} & 39                                     & 15                                & 138                              & 284          & 144                               & 746                              \\
						\hline
					\end{tabular}
				}
			\end{center}
		\end{sc}
	\end{small}
\end{table*}

\subsection{Details about the datasets}
\label{CH:CIGA:sec:exp_data_appdx}
We provide more details about the motivation and construction method of the datasets that are used in our experiments. Statistics of the datasets are presented in Table~\ref{CH:CIGA:tab:datasets_stats_appdx}.

\textbf{SPMotif datasets.} We construct 3-class synthetic datasets based on BAMotif~\citep{gnn_explainer,pge} following~\cite{dir},
where the model needs to tell which one of three motifs (House, Cycle, Crane) that the graph contains.
For each dataset, we generate $3000$ graphs for each class at the training set, $1000$ graphs for each class at the validation set and testing set, respectively.
During the construction, we merely inject the distribution shifts in the training data while keep the testing data and validation data without the biases.
For structure-level shifts (\textbf{SPMotif-Struc}), we introduce the bias based on FIIF, where the motif and one of the three base graphs (Tree, Ladder, Wheel) are artificially (spuriously) correlated with a probability of various biases, and equally correlated with the other two. Specifically, given a predefined bias $b$, the probability of a specific motif (e.g., House) and a specific base graph (Tree) will co-occur is $b$ while for the others is $(1-b)/2$ (e.g., House-Ladder, House-Wheel). We use random node features for SPMotif-Struc, in order to study the influences of structure level shifts.
Moreover, to simulate more realistic scenarios where both structure level and topology level have distribution shifts, we also construct \textbf{SPMotif-Mixed} for mixed distribution shifts.
We additionally introduced FIIF attribute-level shifts based on SPMotif-Struc, where all of the node features are spuriously correlated with a probability of various biases by setting to the same number of corresponding labels. Specifically, given a predefined bias $b$, the probability that all of the node features of a graph has label $y$ (e.g., $y=0$) being set to $y$ (e.g., $\mX=\mathbf{0}$) is $b$ while for the others is $(1-b)/2$ (e.g., $P(\mX=\mathbf{1})=P(\mX=\mathbf{2})=(1-b)/2$). More complex distribution shift mixes can be studied following our construction approach, which we will leave for future works.

\textbf{TU datasets.} To study the effects of graph sizes shifts, we follow~\citet{size_gen1,size_gen2} to study the OOD generalization abilities of various methods on four of TU datasets~\citep{tudataset}, i.e., \textbf{PROTEINS, DD, NCI1, NCI109}. Specifically, we use the data splits generated by~\citet{size_gen1} and use the Matthews correlation coefficient  as evaluation metric following~\cite{size_gen2} due to the class imbalance in the splits. The splits are generated as follows: Graphs with sizes smaller than the $50$-th percentile are assigned to training, while graphs with sizes larger than the $90$-th percentile are assigned to test. A validation set for hyperparameters tuning consists of $10\%$ held out examples from training. We also provide a detailed statistics about these datasets in table~\ref{CH:CIGA:tab:datasets_stats_tu_appdx}.

\textbf{Graph-SST datasets.} Inspired by the data splits generation for studying distribution shifts on graph sizes, we split the data curated from sentiment graph data~\citep{xgnn_tax}, that converts sentiment sentence classification datasets \textbf{SST5} and \textbf{SST-Twitter}~\citep{sst25,sst_twitter} into graphs, where node features are generated using BERT~\citep{bert} and the edges are parsed by a Biaffine parser~\citep{biaffine}. Our splits are created according to the averaged degrees of each graph. Specifically, we assign the graphs as follows: Those that have smaller or equal than $50$-th percentile averaged degree are assigned into training, those that have averaged degree large than $50$-th percentile while smaller than $80$-th percentile are assigned to validation set, and the left are assigned to test set. For SST5 we follow the above process while for Twitter we conduct the above split in an inversed order to study the OOD generalization ability of GNNs trained on large degree graphs to small degree graphs.

\textbf{CMNIST-sp.} To study the effects of PIIF shifts, we select the ColoredMnist dataset created in IRM~\citep{irmv1}. We convert the ColoredMnist into graphs using super pixel algorithm introduced by~\citet{understand_att}. Specifically, the original Mnist dataset are assigned to binary labels where images with digits $0-4$ are assigned to $y=0$ and those with digits $5-9$ are assigned to $y=1$. Then, $y$ will be flipped with a probability of $0.25$. Thirdly, green and red colors will be respectively assigned to images with labels $0$ and $1$ an averaged probability of $0.15$ (since we do not have environment splits) for the training data. While for the validation and testing data the probability is flipped to $0.9$.

\textbf{DrugOOD datasets.} To evaluate the OOD performance in realistic scenarios with realistic distribution shifts, we also include three datasets from DrugOOD benchmark.
DrugOOD is a systematic OOD benchmark for AI-aided drug discovery, focusing on the task of drug target binding affinity prediction for both macromolecule (protein target) and small-molecule (drug compound).
The molecule data and the notations are curated from realistic ChEMBL database~\citep{chembl}.
Complicated distribution shifts can happen on different assays, scaffolds and molecule sizes.
In particular, we select \texttt{DrugOOD-lbap-core-ic50-assay}, \texttt{DrugOOD-lbap-core-ic50-scaffold}, and \texttt{DrugOOD-lbap-core-ic50-size}, from the task of Ligand Based Affinity Prediction which uses \texttt{ic50} measurement type and contains \texttt{core} level annotation noises.
For more details, we refer interested readers to~\citet{drugood}.

\subsection{Training and Optimization in Experiments}
\label{CH:CIGA:sec:exp_impl_appdx}
During the experiments, we do not tune the hyperparameters exhaustively while following the common recipes for optimizing GNNs.
Details are as follows.

\textbf{GNN encoder.} For fair comparison, we use the same GNN architecture as graph encoders for all methods.
By default, we use $3$-layer GNN with Batch Normalization~\citep{batch_norm} between layers and JK residual connections at last layer~\citep{jknet}.
For the architectures we use the GCN with mean readout~\citep{gcn} for all datasets except Proteins where we empirically observe better validation performance
with a GIN and max readout~\citep{gin}, and for DrugOOD  datasets where we follow the backbone used in the paper~\citep{drugood}, i.e., $4$-layer GIN with sum readout.
The hidden dimensions are fixed as $32$ for SPMotif, TU datasets, CMNIST-sp, and $128$ for SST5, Twitter and DrugOOD datasets.

\textbf{Optimization and model selection.}
By default, we use Adam optimizer~\citep{adam} with a learning rate of $1e-3$ and a batch size of $32$ for all models at all datasets.
Except for DrugOOD datasets, we use a batch size of $128$ following the original paper~\citep{drugood}.
To avoid underfitting, we pretrain models for $20$ epochs for all datasets, except
for CMNIST and Twitter where we pretrain $5$ epochs and for SST5 we pretrain $10$ epochs, because of the dataset size and the difficulty of the task.
To avoid overfitting, we also employ an early stopping of $5$ epochs according to the validation performance.
Meanwhile, dropout~\citep{dropout} is also adopted for some datasets.
Specifically, we use a dropout rate of $0.5$ for CMNIST, SST5, Twitter, DrugOOD-Assay and DurgOOD-Scaffold,
$0.1$ for DrugOOD-Size according to the validation performance,
and $0.3$ for TU datasets following the practice of~\citet{size_gen2}.

\textbf{Implementations of baselines.}
For implementations of the interpretable GNNs, we use the author released codes~\citep{gib,asap}, where we use the codes provided by the authors\footnote{\url{https://anonymous.4open.science/r/DIR/}} for DIR~c\citep{dir}
which is the same as the author-released codes.
During the implementation, we use the same $s_c$ for all interpretable GNN baselines, chosen from $\{0.1,0.2,0.25,0.3,0.4,0.5,0.6,0.7,0.8,0.9\}$ according to the validation performances,
and set to $0.25$ for SPMotif following~\citet{dir}, $0.3$ for Proteins and DD,
$0.6$ for NCI1, $0.7$ for NCI109, $0.8$ for CMNIST-sp, $0.5$ for SST5 and Twitter, and $0.8$ for DrugOOD datasets, respectively.
Empirically, we observe that the optimization process in GIB can be unstable during its nested optimization for approximating the mutual information of the predicted subgraph and the input graph.
We use a larger batch size of $128$ or reduce the nested optimization steps to be lower than $20$ for stabilizing the performance.
If the optimization fails due to instability during training, we will select the results with the best validation accuracy as the final outcomes.
Although SPMotif-Struc is also evaluated in DIR, we find the results are inconsistent with the results reported by the author, because
DIR adopts \texttt{Last Epoch Model Selection} which is \emph{different} from the claim that they select models according to \texttt{the validation performance},
i.e., \texttt{line $264$} to \texttt{line $278$} in \texttt{train/spmotif\_dir.py} from the commit \texttt{4b975f9b3962e7820d8449eb4abbb4cc30c1025d} of \url{https://github.com/Wuyxin/DIR-GNN}.
We select the hyperparameters for the proposed DIR regularization from $\{0.01,0.1,1,10\}$ according to the validation performances at the datasets, while we stick to the authors' claimed hyperparameters for the datasets they also experimented with.

For invariant learning,
we refer to the implementations in DomainBed~\citep{domainbed} for IRM~\citep{irmv1}, vrex~\citep{vrex} and IB-IRM~\citep{ib-irm}.
Since the environment information is not available, we perform random partitions on the training data to obtain two equally large environments for these objectives.
Moreover, we select the weights for the corresponding regularization from $\{0.01,0.1,1,10,100\}$ for these objectives according to the validation performances of IRM and stick to it for others,
since we empirically observe that they perform similarly with respect to the regularization weight choice.
For EIIL~\citep{env_inference}, we use the author-released implementations about assigning different samples the weights for being put in each environment and calculating the IRM loss.

Besides, for CNC~\citep{cnc}, we follow the algorithm description to modify the sampling strategy in supervised contrastive loss~\citep{sup_contrastive} based on a pre-trained GNN optimized with ERM and choose the weight for contrastive loss using the same grid search as for \ciga.

\textbf{Implementations of \ciga.}
For a fair comparison, \ciga uses the same GNN architecture for GNN encoders as the baseline methods.
We did not do exhaustive hyperparameters tuning for the loss Eq.~\ref{CH:CIGA:good_opt_loss_impl_appdx}.
By default, we fix the temperature to be $1$ in the contrastive loss,
and merely search $\alpha$ from $\{0.5,1,2,4,8,16,32\}$ and $\beta$ from $\{0.5,1,2,4\}$ according to the validation performances.
For CMNIST-sp, we find larger $\beta$ are required to get rid of intense spurious node features hence we expand the search range for $\beta$ to $\{0.5,1,2,4,16,32\}$,
For Graph-SST datasets, we search $\alpha$ from $\{0.5,1,2,4\}$ as we empirically find that increasing $\alpha$ does not help increase the performance with few random runs.
Besides, we also have various implementation options for obtaining the features in $\widehat{G}_c$, for obtaining $h_{\widehat{G}_c}$, as well as for obtaining predictions based on $\widehat{G}_s$.
By default, we feed the graph representations of featurizer GNN to the classifier GNN, as well as to the contrastive loss.
For classifying $G$ based on $\widehat{G}_s$, we use a separate MLP downstream classifier in the classifier GNN $f_c$.
The only exception is for the CMNIST-sp dataset where the spurious correlation is stronger than the invariant signal.
Directly feeding the graph representations from the featurizer GNN can easily overfit to the shortcuts hence we instead feed the original features to the downstream classifier GNN.
There can be more other options, such as using separate graph convolutions on $\widehat{G}_s$ or $\widehat{G}_c$, which we leave for future work.

\textbf{Evaluation protocol.} We run each experiment $10$ on TU datasets and $5$ times for others where the random seeds start from $1$ to the number of total repeated times.
During each run, we select the model according to the validation performance and report the mean and standard deviation of the corresponding metrics.

\subsection{Software and Hardware}
\label{CH:CIGA:sec:exp_software_appdx}
We implement our methods with PyTorch~\citep{pytorch} and PyTorch Geometric~\citep{pytorch_geometric}.
We ran our experiments on Linux Servers with 40 cores Intel(R) Xeon(R) Silver 4114 CPU @ 2.20GHz, 256 GB Memory, and Ubuntu 18.04 LTS installed.
GPU environments are varied from 4 NVIDIA RTX 2080Ti graphics cards with CUDA 10.2, 2 NVIDIA RTX 2080Ti and 2 NVIDIA RTX 3090Ti graphics cards with CUDA 11.3, and NVIDIA TITAN series with CUDA 11.3.

\subsection{Additional Analysis}
\label{CH:CIGA:sec:additional_exp_appdx}

\begin{figure}[H]
	\centering
	\subfigure[SPMotif-Mixed under bias$=0.9$]{
		\includegraphics[width=0.31\textwidth]{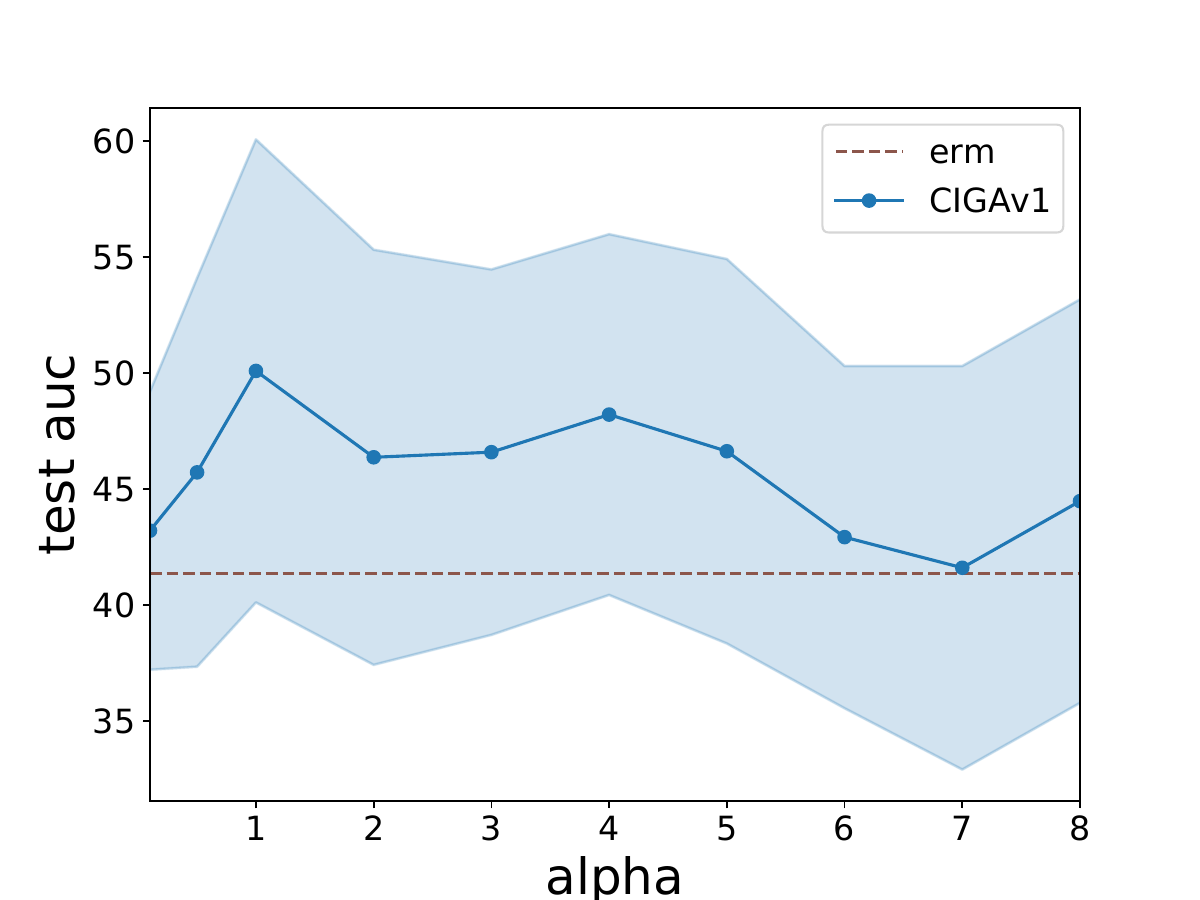}
	}
	\subfigure[DrugOOD-Scaffold]{
		\includegraphics[width=0.31\textwidth]{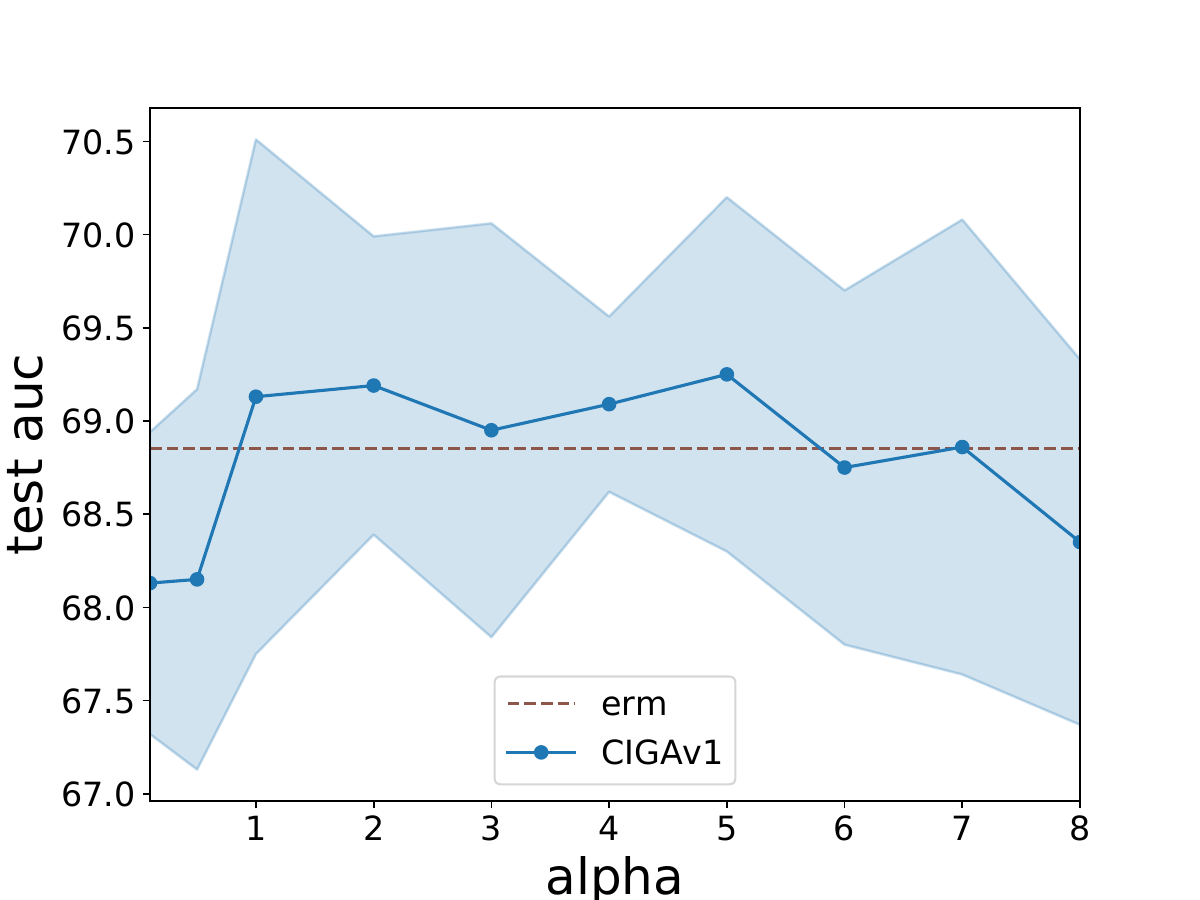}
	}
	\subfigure[NCI109]{
		\includegraphics[width=0.31\textwidth]{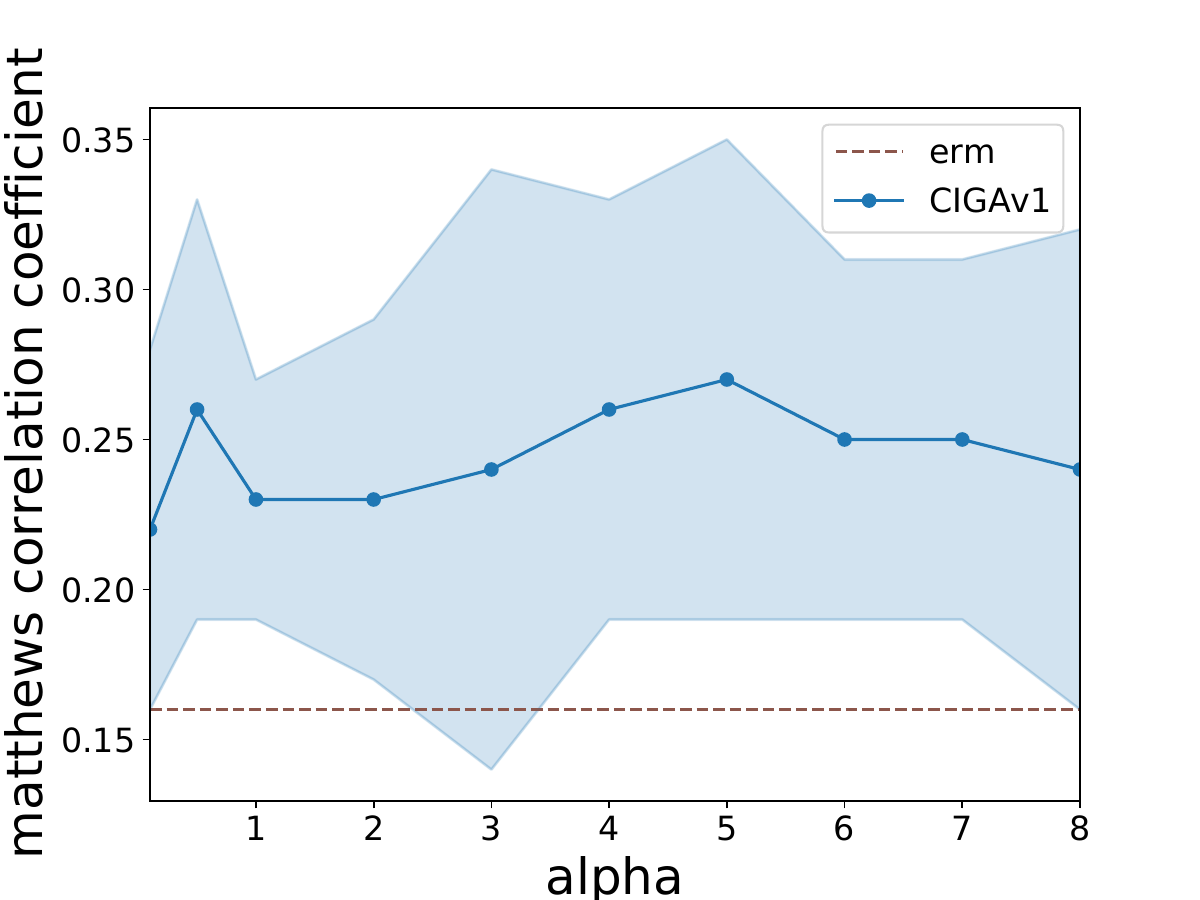}
	}
	\caption{
		Hyperparameter sensitivity analysis on the coefficient of contrastive loss ($\alpha$).}
	\label{CH:CIGA:fig:hp_sen_alpha_appdx}
\end{figure}

\begin{figure}[H]
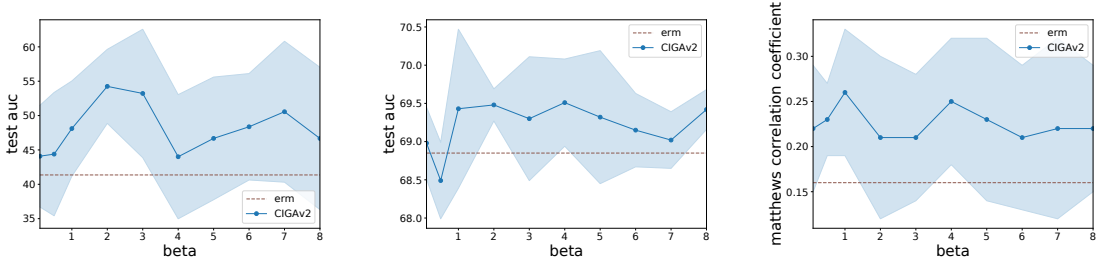

	\centering
	\subfigure[SPMotif-Mixed under bias$=0.9$ with $\alpha=4$]{
		\includegraphics[width=0.31\textwidth]{Figures/CIGA/spmotif_ab_conf.pdf}
	}
	\subfigure[DrugOOD-Scaffold with $\alpha=1$]{
		\includegraphics[width=0.31\textwidth]{Figures/CIGA/drugood_scaffold_ab_conf.pdf}
	}
	\subfigure[NCI109 with $\alpha=1$]{
		\includegraphics[width=0.31\textwidth]{Figures/CIGA/nci_ab_conf.pdf}
	}
	\caption{
		Hyperparameter sensitivity analysis on the coefficient of hinge loss ($\beta$).}
	\label{CH:CIGA:fig:hp_sen_beta_appdx}
\end{figure}

\textbf{Hyperparameter sensitivity analysis.} To examine how sensitive \ciga is to the hyperparameters $\alpha$ and $\beta$ for contrastive loss and hinge loss, respectively, under different distribution shifts. We conduct experiments based on the hardest datasets from each table (i.e., SPMotif-Mixed with the bias of $0.9$, DrugOOD-Scaffold, and the NCI109 datasets from Table~\ref{CH:CIGA:tab:sythetic}, Table~\ref{table:other_graph}, and Table~\ref{table:graph_size}, respectively.) To increase the difficulty, we search for more fine-grained spaces for both parameters, i.e., $\{0.1,0.5,1,2,3,4,5,6,7,8\}$. During changing the value of $\beta$, we will fix the $\alpha$ to a specific value under which the model has a relatively good performance (but not the best, to fully examine the robustness of \ciga in practice). During the sensitivity tests, we follow the evaluation protocol as that used for the main experiments. The results are shown in Fig.~\ref{CH:CIGA:fig:hp_sen_alpha_appdx} and Fig.~\ref{CH:CIGA:fig:hp_sen_beta_appdx}.

From the results above, we can see that both {\ciga}v1 and {\ciga}v2 are robust to different values of $\alpha$ and $\beta$, respectively, across different datasets and distribution shifts. Notably, in Fig.~\ref{CH:CIGA:fig:hp_sen_alpha_appdx}, when the coefficient $\alpha$ for the contrastive loss becomes too small, the invariance of the identified invariant subgraphs $\widehat{G}_c$ may not be guaranteed, resulting worse performances. Moreover, when $\alpha$ becomes too large, it may affect the optimization and yield worse performances. In SPMotif datasets, the worse performances can be observed via the large variances as well. Similarly for $\beta$, as shown in Fig.~\ref{CH:CIGA:fig:hp_sen_beta_appdx}, when $\beta$ becomes too small, some part from the spurious subgraph may still be contained in the estimated invariant subgraphs. While if $\beta$ becomes too large, there might be part of $\widehat{G}_c$ being eliminated. Although both {\ciga}v1 and {\ciga}v2 are robust to the changes of $\alpha$ and $\beta$, the intrinsic difficult optimization in OOD generalization algorithms including the proposed \ciga in our work, still require a more proper and smooth optimization process~\citep{pair}.

\begin{table}[ht]
	\centering
	\caption{Averaged training time (sec.) per epoch of various methods on DrugOOD-Scaffold.}\small\sc
	\label{CH:CIGA:tab:running_time_appdx}
	\resizebox{\textwidth}{!}{
		\begin{small}
			\begin{tabular}{lccccccccc}
				\toprule
				Methods         & ERM   & ASAP   & GIB     & DIR     & IRM   & EIIL   & CNC   & {\ciga}v1 & {\ciga}v2 \\\midrule
				Running time    & 8.055 & 15.578 & 300.304 & 106.919 & 8.73  & 69.664 & 9.795 & 40.065    & 46.181    \\
				OOD Performance & 68.85 & 66.19  & 62.01   & 63.91   & 68.69 & 68.45  & 67.24 & 69.04     & 69.7      \\
				Avg. Rank       & 2     & 5.5    & 9       & 8       & 3     & 6      & 4.5   & 3.5       & 3.5       \\
				\bottomrule
			\end{tabular}	\end{small}}
\end{table}

\textbf{Running time analysis.} To examine how much computational overhead is induced by the architecture and the additional objectives in \ciga, we analyze and compare the averaged training time of different methods on DrugOOD-Scaffold. Factors that could affect the running time such as GNN backbone, batch size, and the running devices (NVIDIA RTX 2080Ti, Linux Servers with 40 cores Intel(R) Xeon(R) Silver 4114 CPU @ 2.20GHz, 256 GB Memory, and Ubuntu 18.04 LTS), are fixed the same during the testing. The results are shown as in Table.~\ref{CH:CIGA:tab:running_time_appdx}. It can be found that \ciga is the only OOD method that outperforms ERM by a non-trivial margin with a relatively low additional computational overhead.

\begin{table}[ht]
	\centering
	\caption{Performances of different methods on Drug-Assay under single environment OOD generalization (i).}\small\sc
	\label{CH:CIGA:tab:single_dg_appdx}
	\resizebox{\textwidth}{!}{
		\begin{small}
			\begin{tabular}{lccccccc}
				\toprule
				Methods         & ERM         & ASAP        & GIB         & DIR         & {\ciga}v1             & {\ciga}v2             & Oracle (IID) \\\midrule
				OOD Performance & 63.29(2.67) & 63.41(0.70) & 62.72(0.59) & 62.56(0.79) & \textbf{63.86 (0.57)} & \textbf{64.31 (0.92)} & 84.71 (1.60) \\
				Rank            & 5           & 4           & 8           & 9           & 2                     & 1                     &              \\
				\bottomrule
			\end{tabular}	\end{small}}
\end{table}

\begin{table}[ht]
	\centering
	\caption{Performances of different methods on Drug-Assay under single environment OOD generalization (ii).}\small\sc
	\label{CH:CIGA:tab:single_dg_2_appdx}
	\resizebox{\textwidth}{!}{
		\begin{small}
			\begin{tabular}{lccccccccc}
				\toprule
				Methods         & ERM         & IRM         & vrex        & EIIL        & IB-IRM      & CNC         & {\ciga}v1             & {\ciga}v2             & Oracle (IID) \\\midrule
				OOD Performance & 63.29(2.67) & 63.25(1.45) & 62.18(1.71) & 62.95(1.37) & 61.95(1.72) & 63.61(0.96) & \textbf{63.86 (0.57)} & \textbf{64.31 (0.92)} & 84.71 (1.60) \\
				Rank            & 5           & 6           & 10          & 7           & 11          & 3           & 2                     & 1                     &              \\
				\bottomrule
			\end{tabular}	\end{small}}
\end{table}

\textbf{Single environment OOD generalization.} The theory of invariant learning fundamentally assume the presence of multiple environments~\citep{inv_principle,irmv1}. However in practice, it does not always hold, which would inevitably fail all of the invariant learning solutions~\citep{irmv1,vrex,env_inference,ib-irm}, including \ciga.

Nevertheless, to examine how \ciga performs under various realistic scenarios, we conduct an additional experiment based on DrugOOD-Assay. We select samples that are from the largest assay group (i.e., the biochemical functionalities of these molecules are tested and reported under the same experimental setup in the lab)~\citep{drugood}. The results are separated and shown in Table~\ref{CH:CIGA:tab:single_dg_appdx} and Table~\ref{CH:CIGA:tab:single_dg_2_appdx}. Besides the baselines, we also show the ``Oracle'' performances from the main table, to demonstrate the performance gaps.

From Table~\ref{CH:CIGA:tab:single_dg_appdx} and Table~\ref{CH:CIGA:tab:single_dg_2_appdx}, we can see that, both {\ciga}v1 and {\ciga}v2 maintain their state-of-the-art performances even in the single training environment setting. We hypothesize that enforcing the mutual information between the estimated $\widehat{G}_c$ also helps to retain the invariance even under the single training environment setting. That may partially explain why CNC can bring some improvements. We believe it is an interesting and promising future direction to develop an in-depth understanding and better solutions under this circumstance.

\subsection{Interpretation Visualization}
\label{CH:CIGA:sec:interpret_visualize_appdx}
Since we use the interpretable GNN architecture to implement \ciga\footnote{We use the code provided by~\citep{gsat}.}, it brings an additional benefit that provides certain interpretation for the predictions automatically, which may facilitate human understanding in practice.

\begin{figure}[H]
	\centering
	\subfigure[]{
		\includegraphics[width=0.31\textwidth]{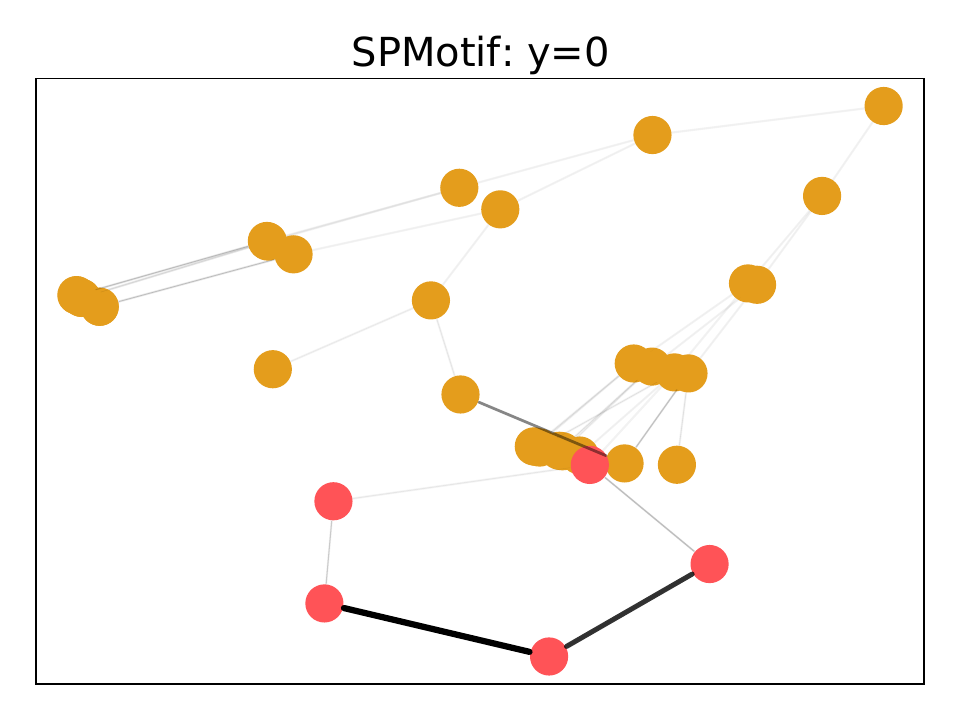}
	}
	\subfigure[]{
		\includegraphics[width=0.31\textwidth]{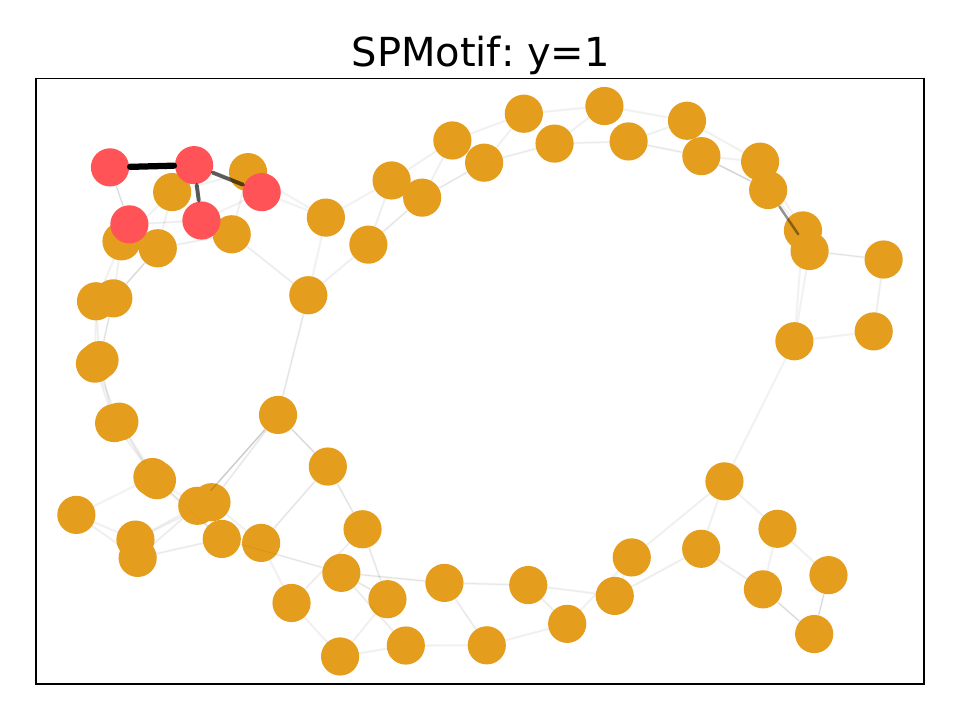}
	}
	\subfigure[]{
		\includegraphics[width=0.31\textwidth]{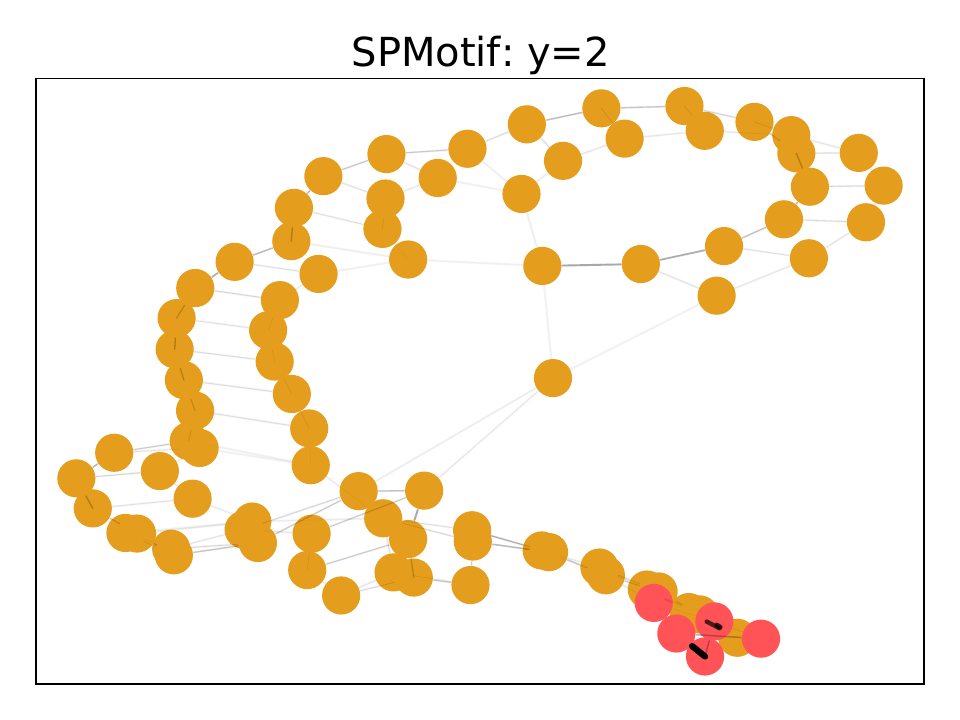}
	}
	\caption{
		Interpretation visualization of examples from SPMotif-Struc under bias$=0.6$.}
	\label{CH:CIGA:fig:spmotif_b6_appdx}
\end{figure}

\begin{figure}[H]
	\centering
	\subfigure[]{
		\includegraphics[width=0.31\textwidth]{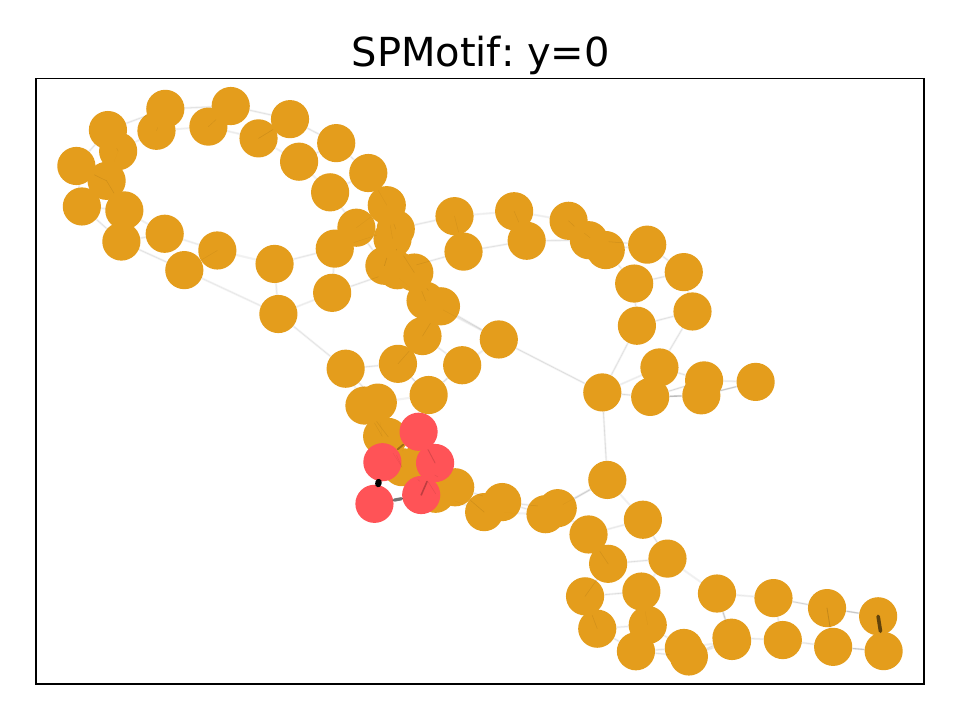}
	}
	\subfigure[]{
		\includegraphics[width=0.31\textwidth]{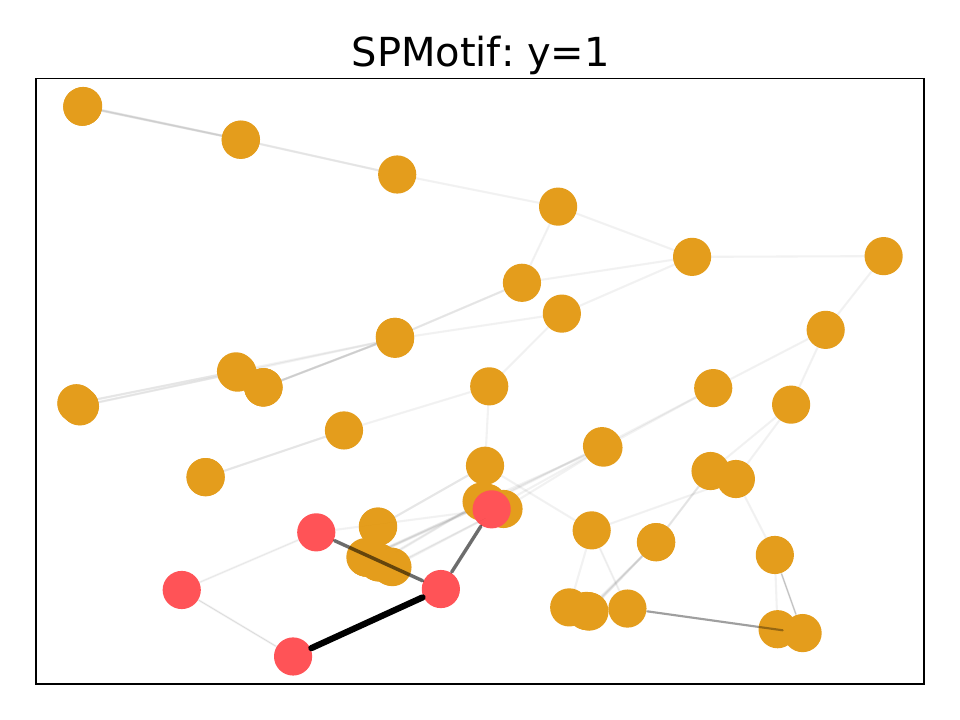}
	}
	\subfigure[]{
		\includegraphics[width=0.31\textwidth]{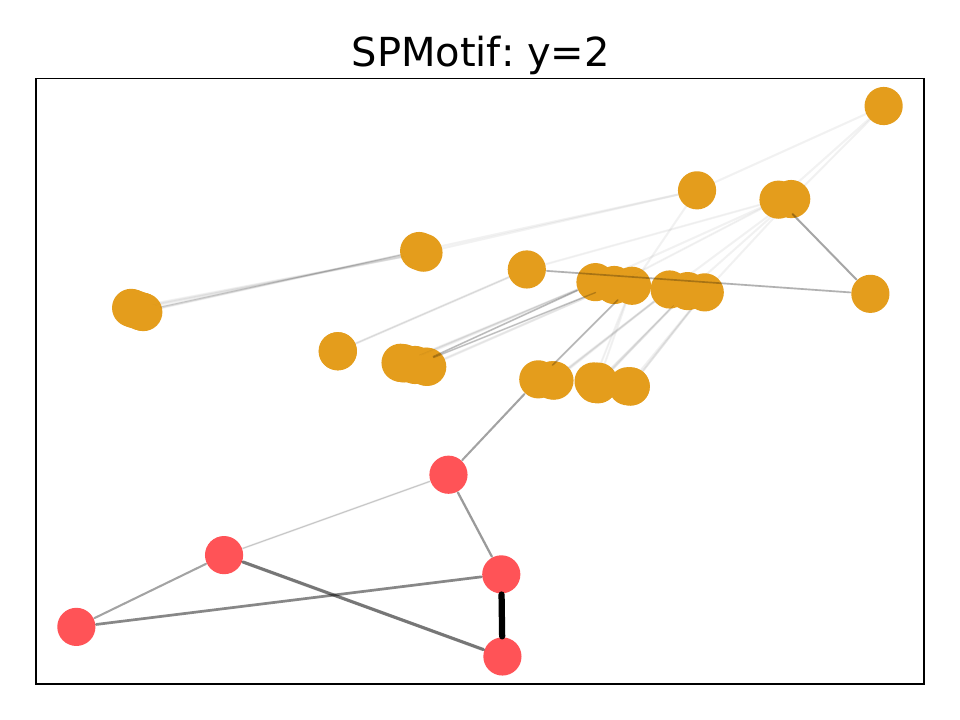}
	}
	\caption{
		Interpretation visualization of examples from SPMotif-Struc under bias$=0.9$.}
	\label{CH:CIGA:fig:spmotif_b9_appdx}
\end{figure}

\begin{figure}[H]
	\centering
	\subfigure[]{
		\includegraphics[width=0.31\textwidth]{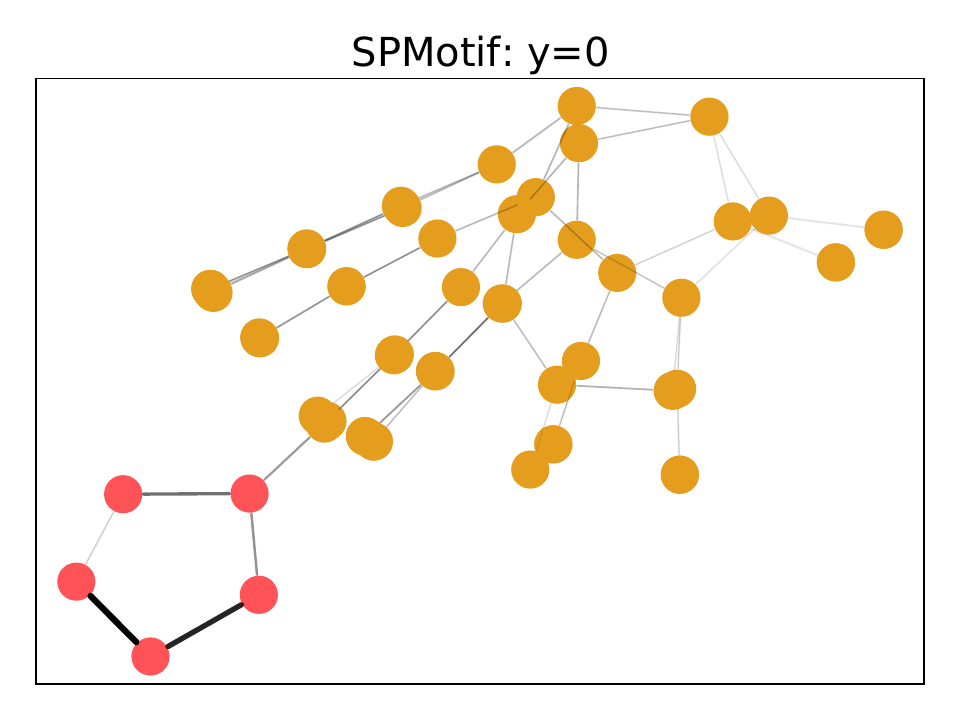}
	}
	\subfigure[]{
		\includegraphics[width=0.31\textwidth]{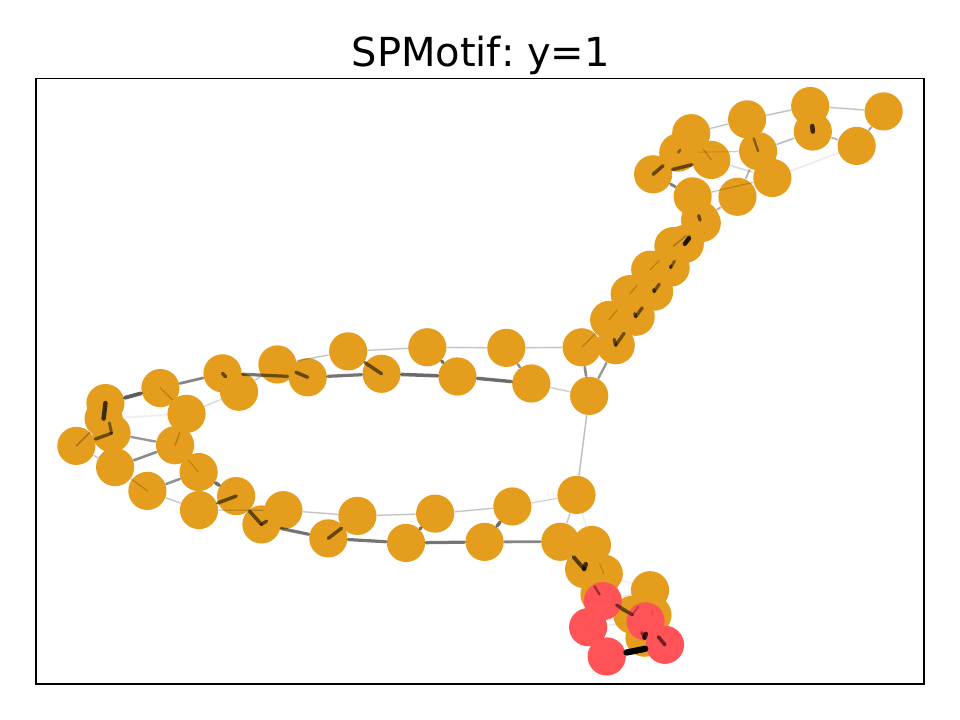}
	}
	\subfigure[]{
		\includegraphics[width=0.31\textwidth]{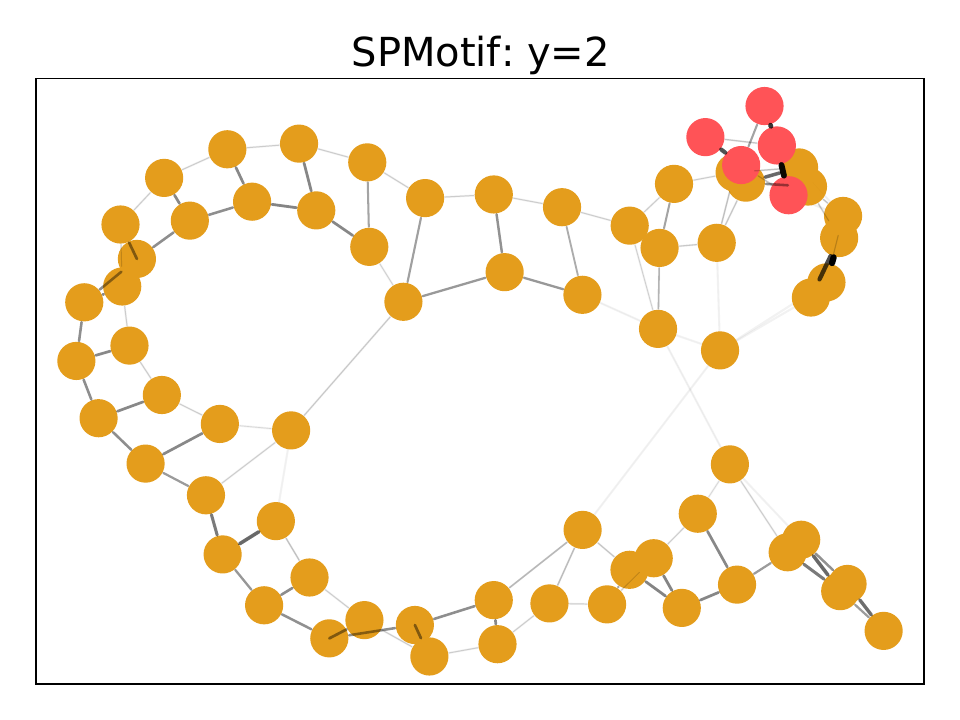}
	}
	\caption{
		Interpretation visualization of examples from SPMotif-Mixed under bias$=0.6$.}
	\label{CH:CIGA:fig:spmotifm_b6_appdx}
\end{figure}

\begin{figure}[H]
	\centering
	\subfigure[]{
		\includegraphics[width=0.31\textwidth]{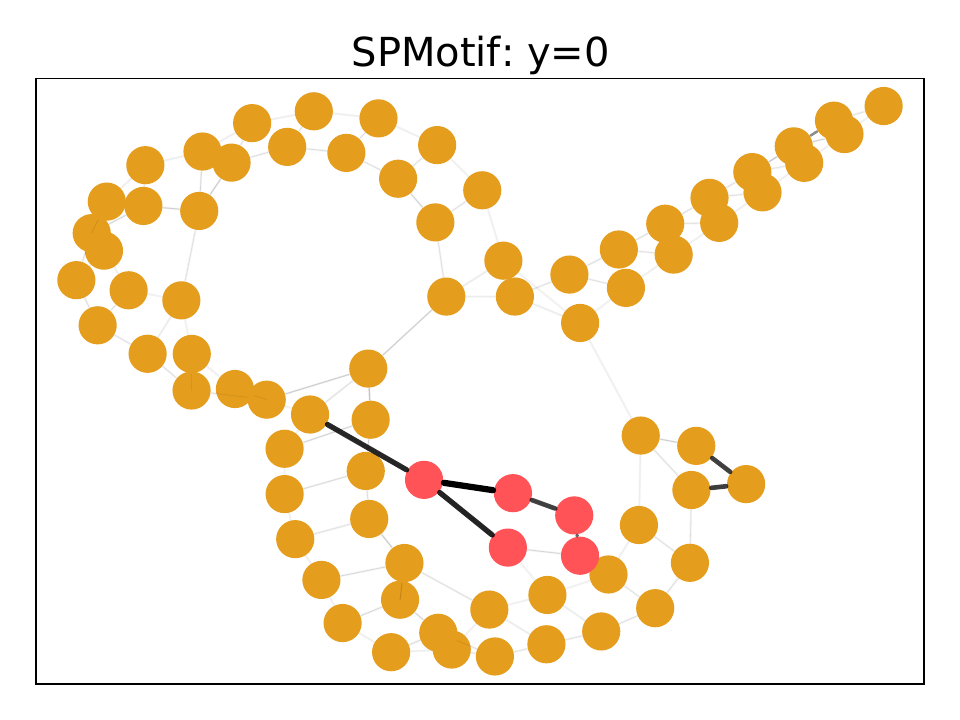}
	}
	\subfigure[]{
		\includegraphics[width=0.31\textwidth]{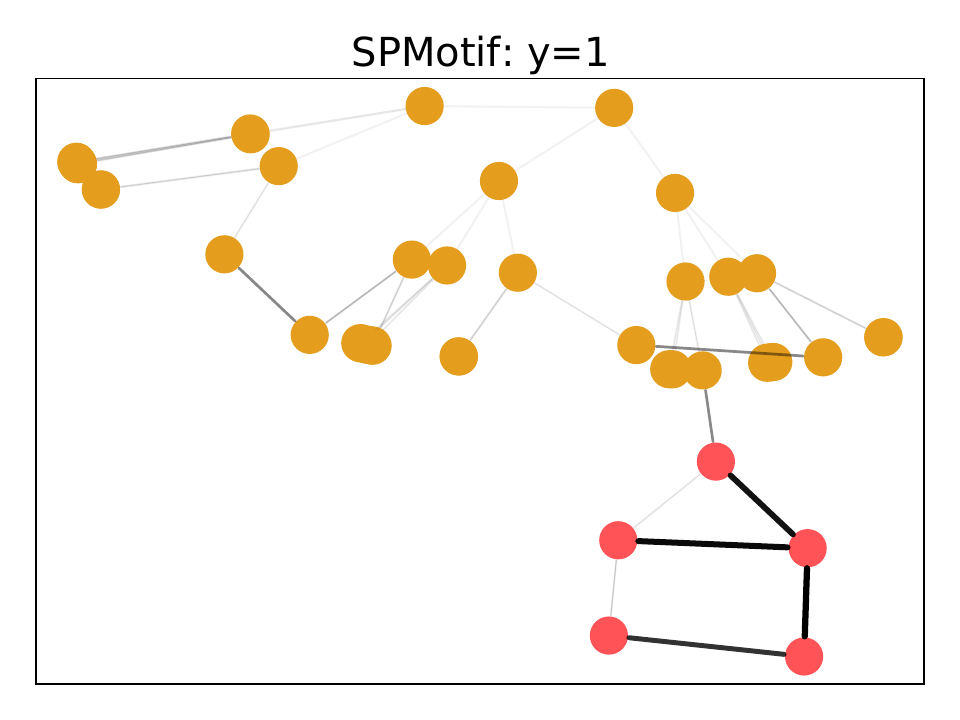}
	}
	\subfigure[]{
		\includegraphics[width=0.31\textwidth]{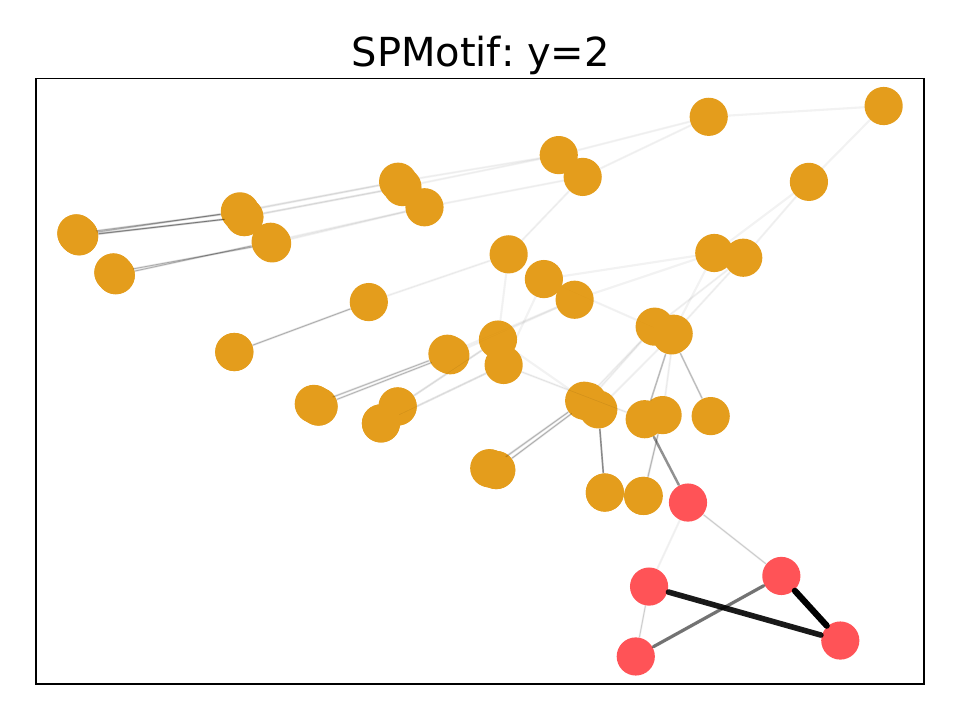}
	}
	\caption{
		Interpretation visualization of examples from SPMotif-Mixed under bias$=0.9$.}
	\label{CH:CIGA:fig:spmotifm_b9_appdx}
\end{figure}

First, we provide some interpretation visualizations in SPMotif-Struc and SPMotif-Mixed datasets, under the biases of $0.6$ and $0.9$. Shown in Fig.~\ref{CH:CIGA:fig:spmotif_b6_appdx} to Fig.~\ref{CH:CIGA:fig:spmotifm_b9_appdx}, we use pink to color the ground truth nodes in $G_c$, and denote the relative attention strength with edge color intensities.

\begin{figure}[H]
	\centering
	\subfigure[]{
		\includegraphics[width=0.31\textwidth]{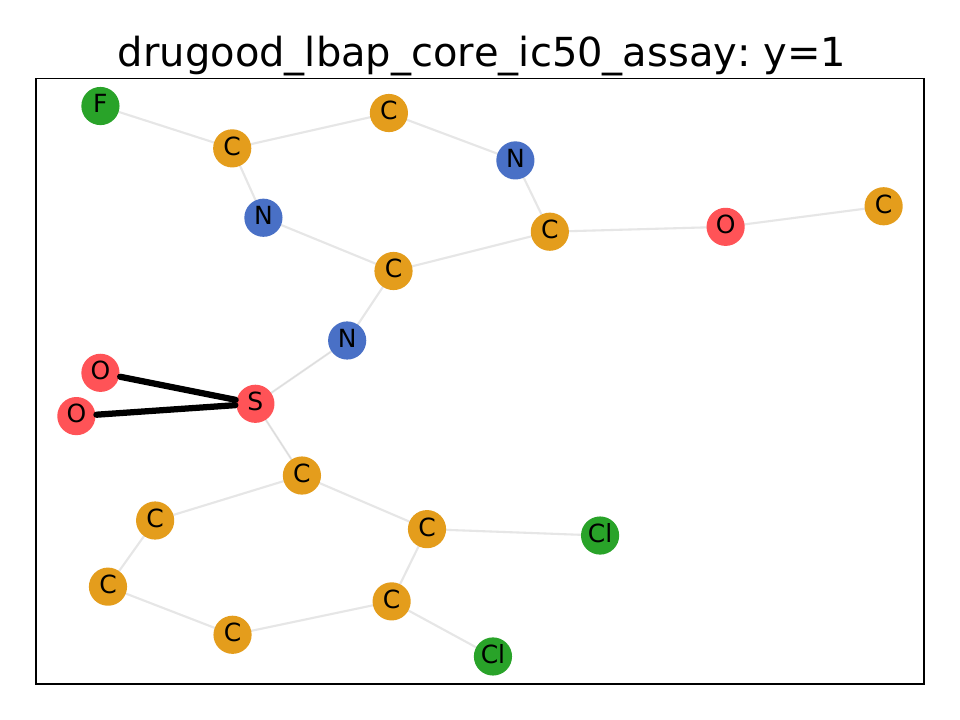}
	}
	\subfigure[]{
		\includegraphics[width=0.31\textwidth]{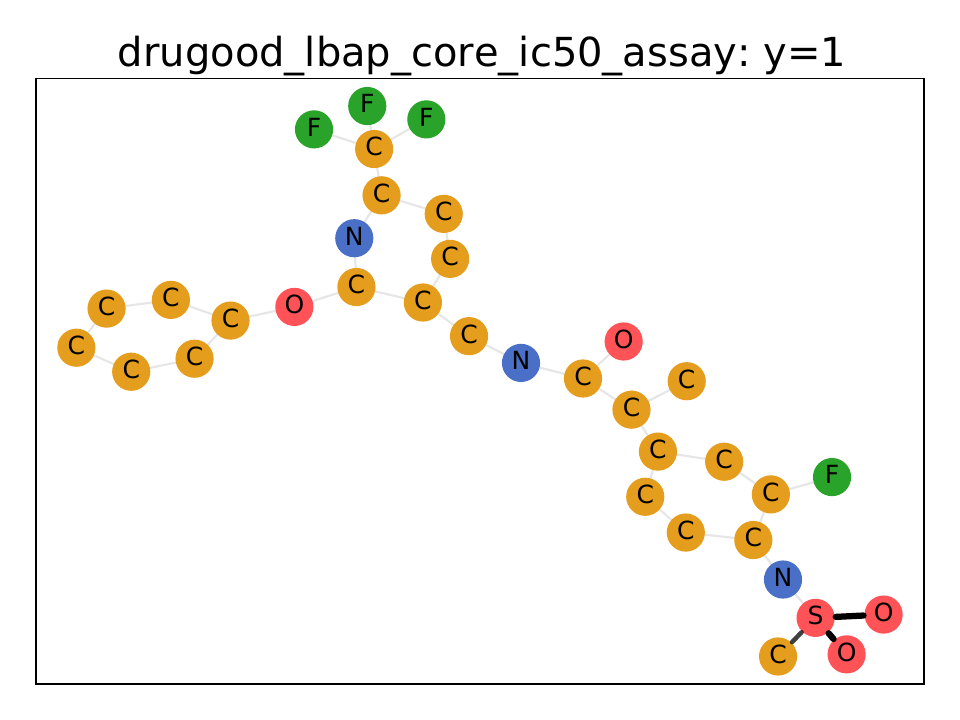}
	}
	\subfigure[]{
		\includegraphics[width=0.31\textwidth]{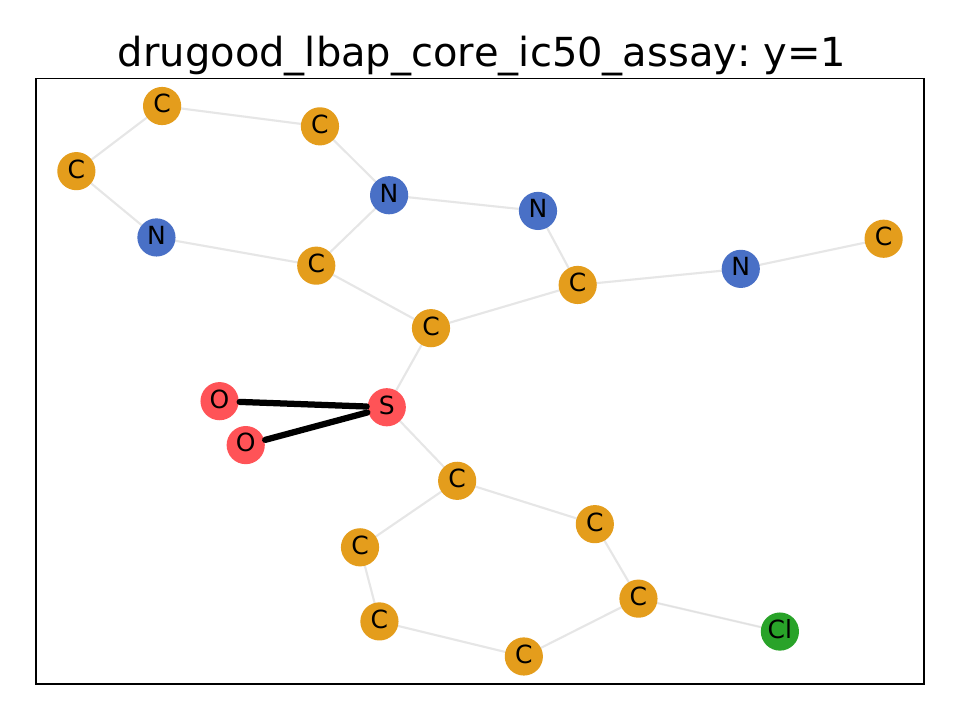}
	}
	\caption{
		Interpretation visualization of activate examples ($y=1$) from DrugOOD-Assay.}
	\label{CH:CIGA:fig:assay_viz_act_appdx}
\end{figure}

\begin{figure}[H]
	\centering
	\subfigure[]{
		\includegraphics[width=0.31\textwidth]{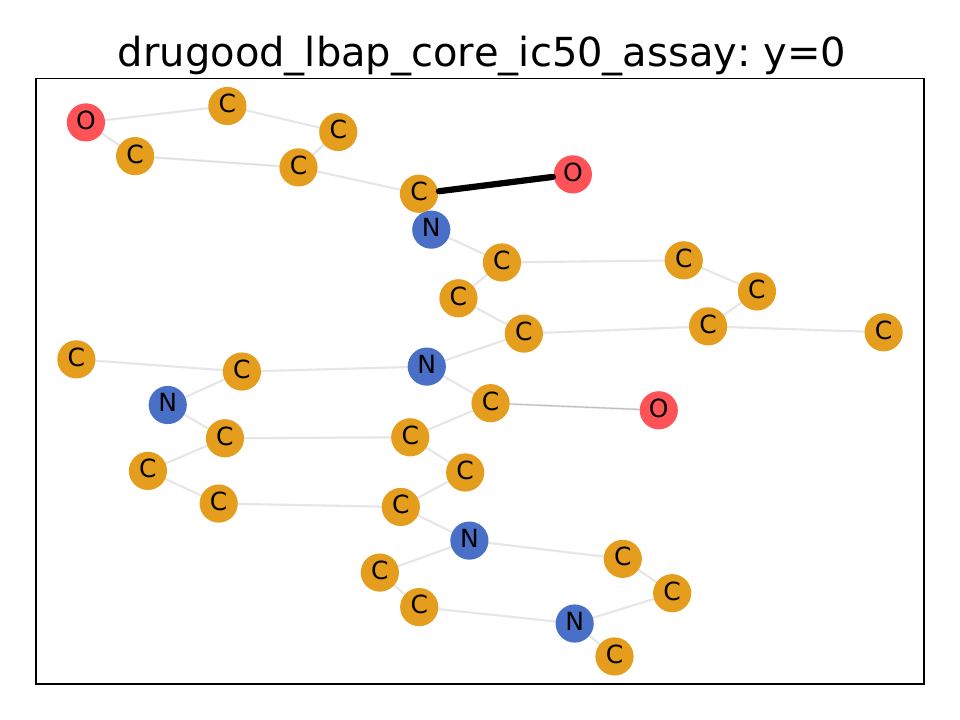}
	}
	\subfigure[]{
		\includegraphics[width=0.31\textwidth]{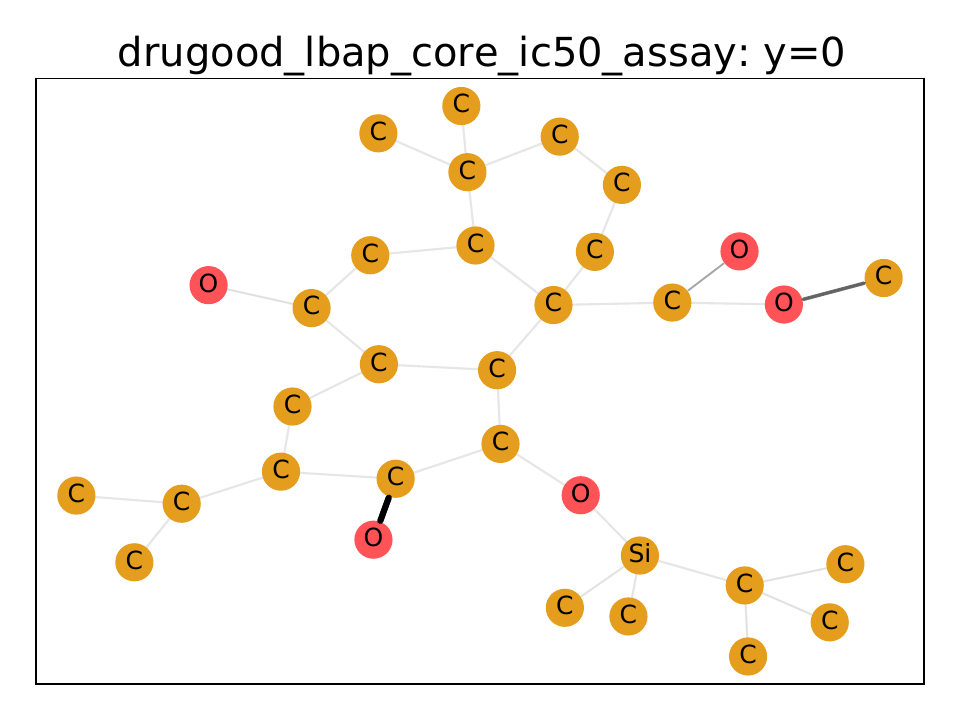}
	}
	\subfigure[]{
		\includegraphics[width=0.31\textwidth]{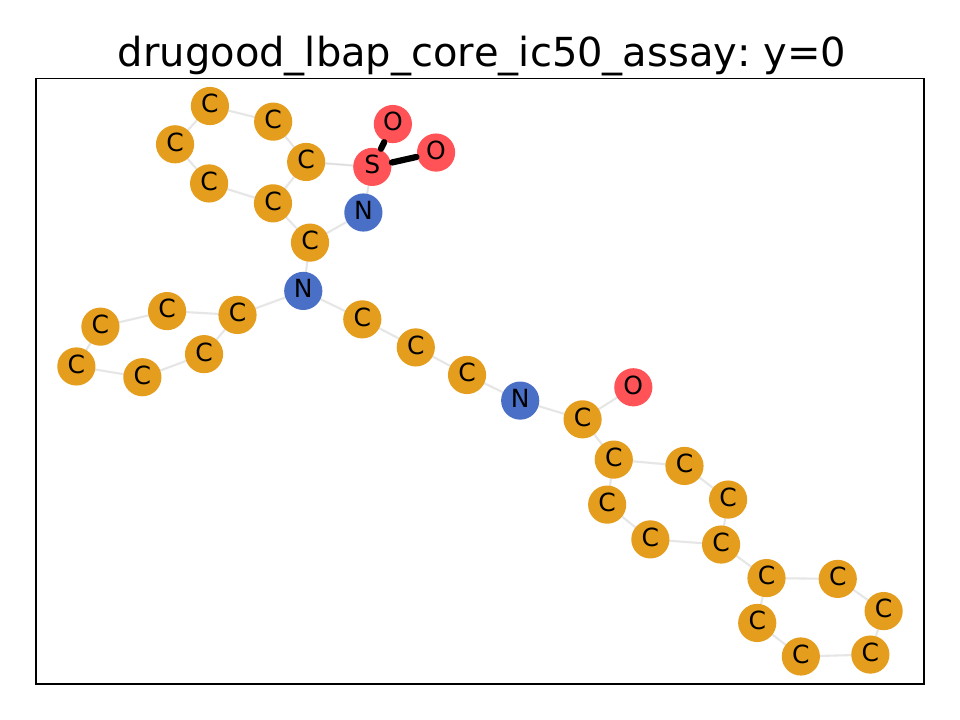}
	}
	\caption{
		Interpretation visualization of inactivate examples ($y=0$) from DrugOOD-Assay.}
	\label{CH:CIGA:fig:assay_viz_inact_appdx}
\end{figure}

\begin{figure}[H]
	\centering
	\subfigure[]{
		\includegraphics[width=0.31\textwidth]{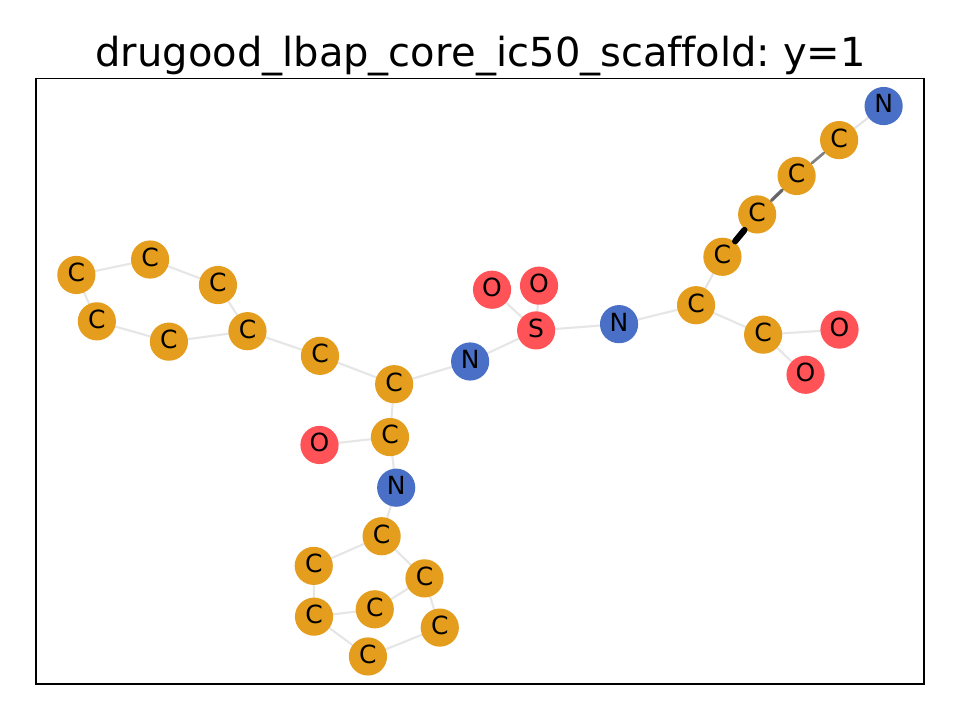}
	}
	\subfigure[]{
		\includegraphics[width=0.31\textwidth]{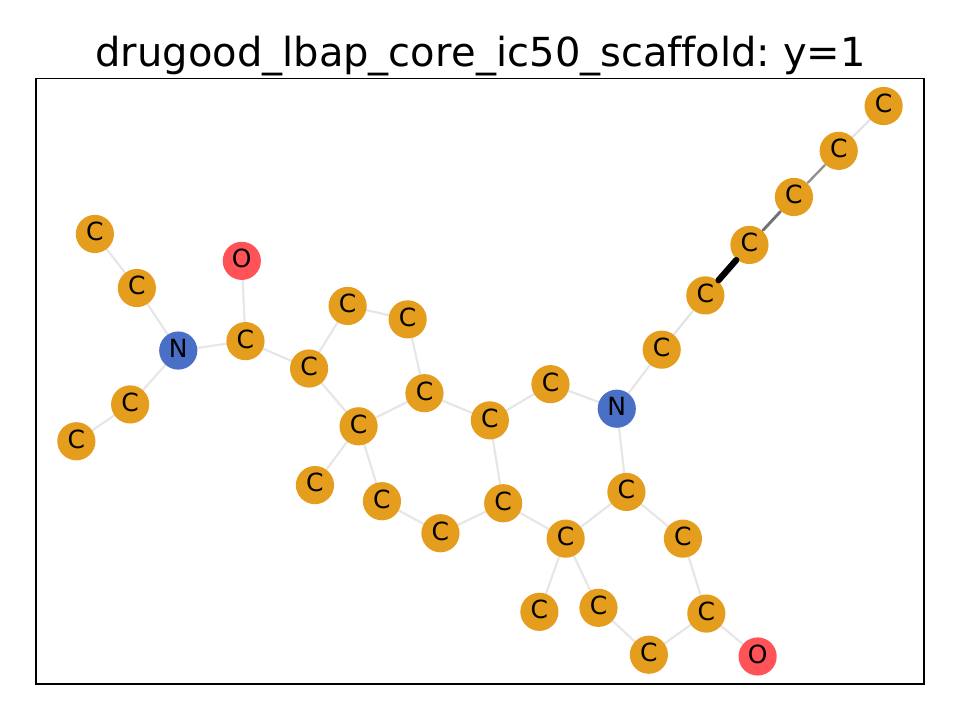}
	}
	\subfigure[]{
		\includegraphics[width=0.31\textwidth]{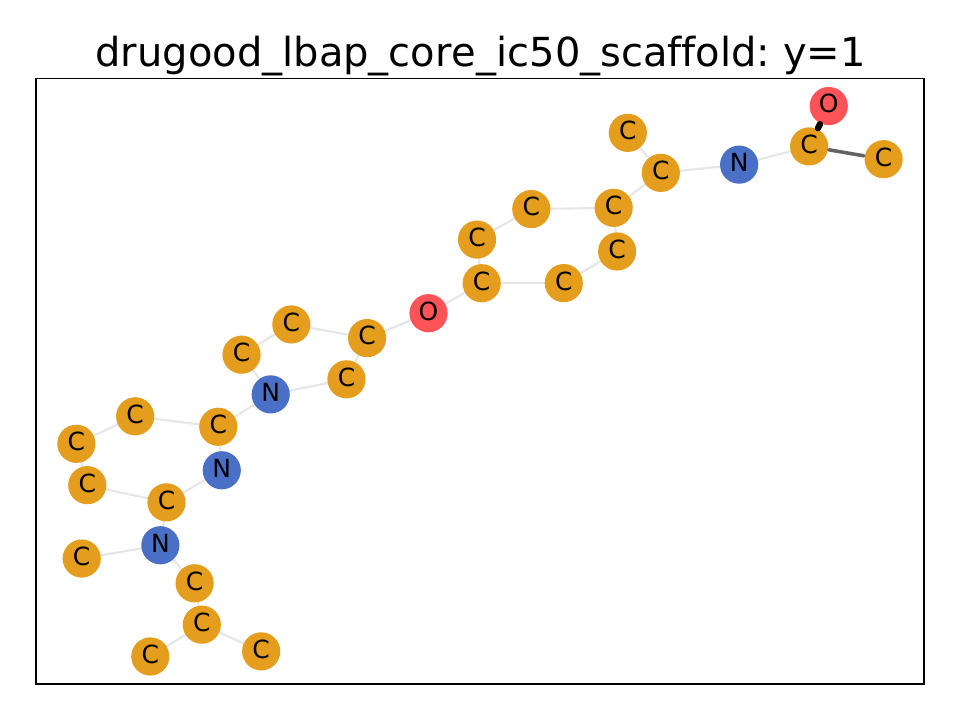}
	}
	\caption{
		Interpretation visualization of activate examples ($y=1$) from DrugOOD-Scaffold.}
	\label{CH:CIGA:fig:scaffold_viz_act_appdx}
\end{figure}

\begin{figure}[H]
	\centering
	\subfigure[]{
		\includegraphics[width=0.31\textwidth]{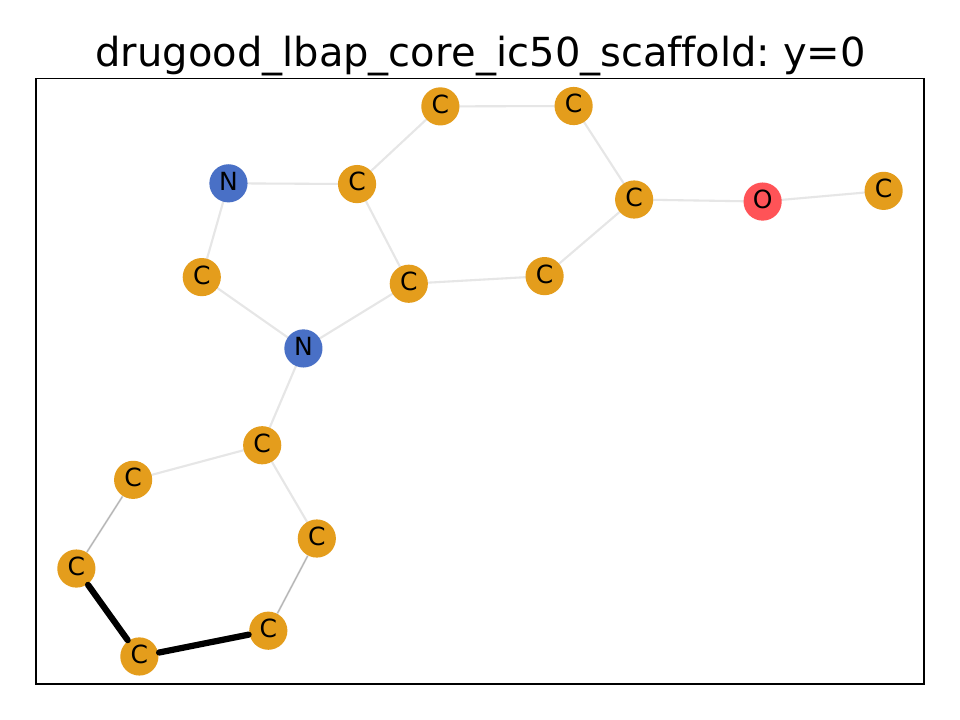}
	}
	\subfigure[]{
		\includegraphics[width=0.31\textwidth]{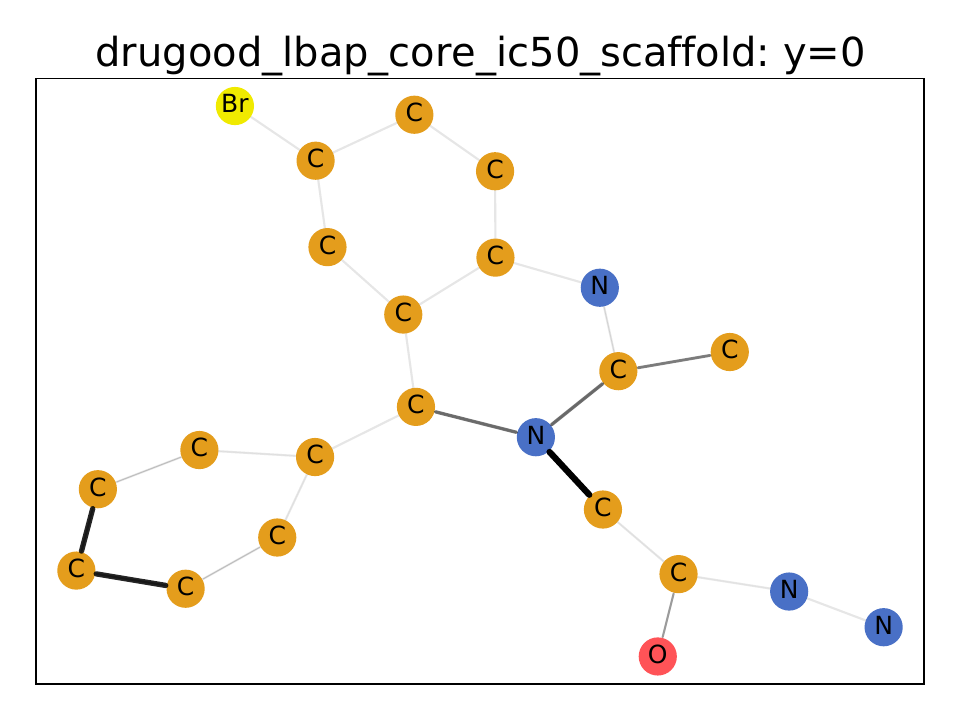}
	}
	\subfigure[]{
		\includegraphics[width=0.31\textwidth]{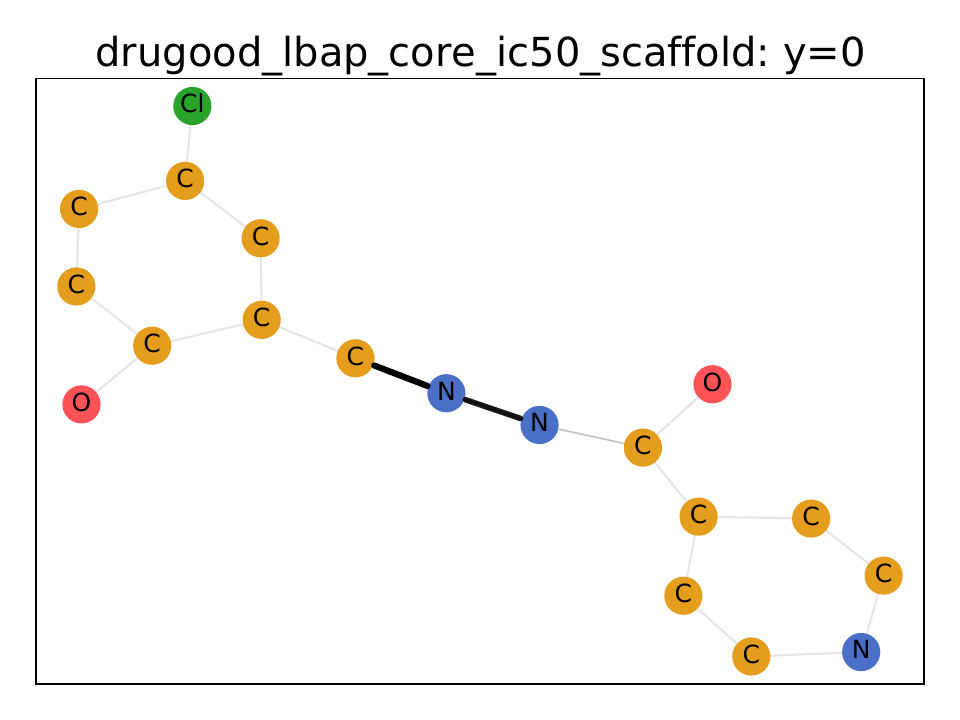}
	}
	\caption{
		Interpretation visualization of inactivate examples ($y=0$) from DrugOOD-Scaffold.}
	\label{CH:CIGA:fig:scaffold_viz_inact_appdx}
\end{figure}

\begin{figure}[H]
	\centering
	\subfigure[]{
		\includegraphics[width=0.31\textwidth]{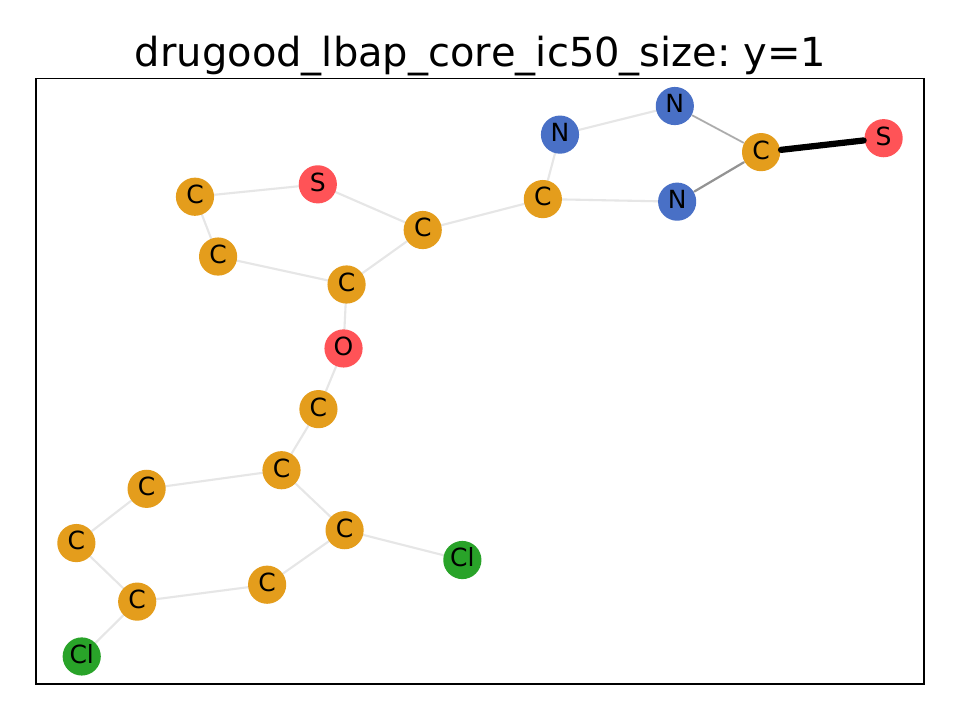}
	}
	\subfigure[]{
		\includegraphics[width=0.31\textwidth]{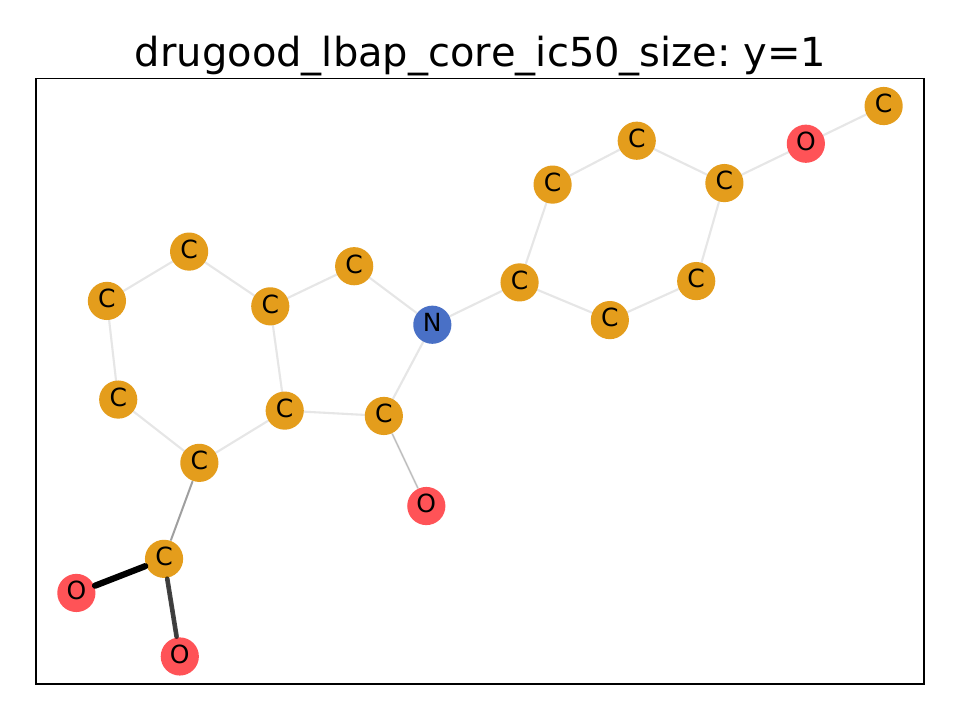}
	}
	\subfigure[]{
		\includegraphics[width=0.31\textwidth]{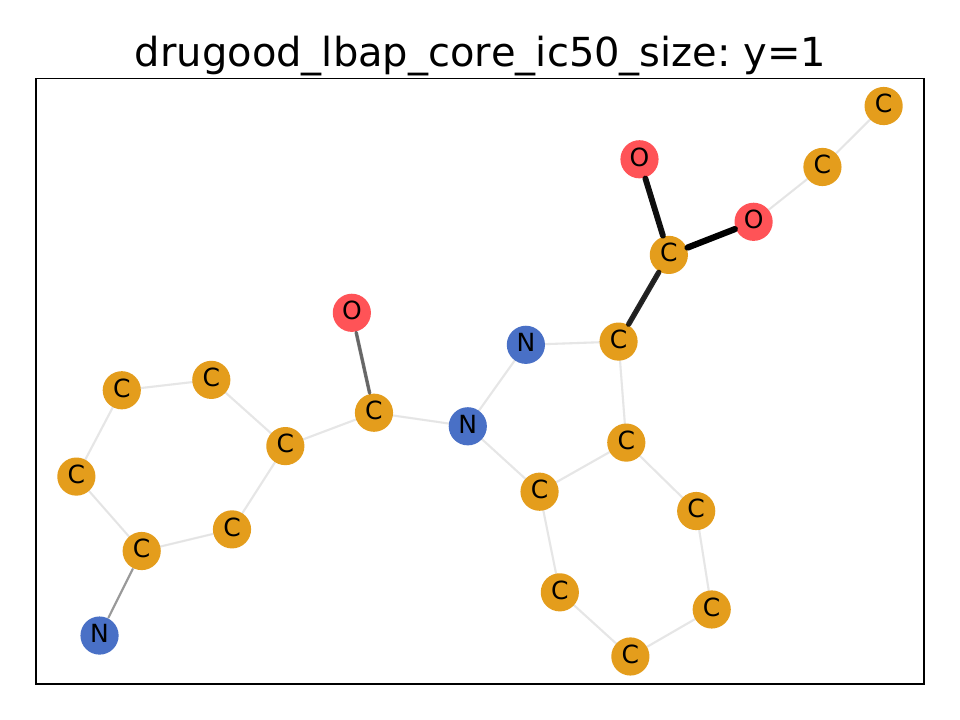}
	}
	\caption{
		Interpretation visualization of activate examples ($y=1$) from DrugOOD-Size.}
	\label{CH:CIGA:fig:size_viz_act_appdx}
\end{figure}

\begin{figure}[H]
	\centering
	\subfigure[]{
		\includegraphics[width=0.31\textwidth]{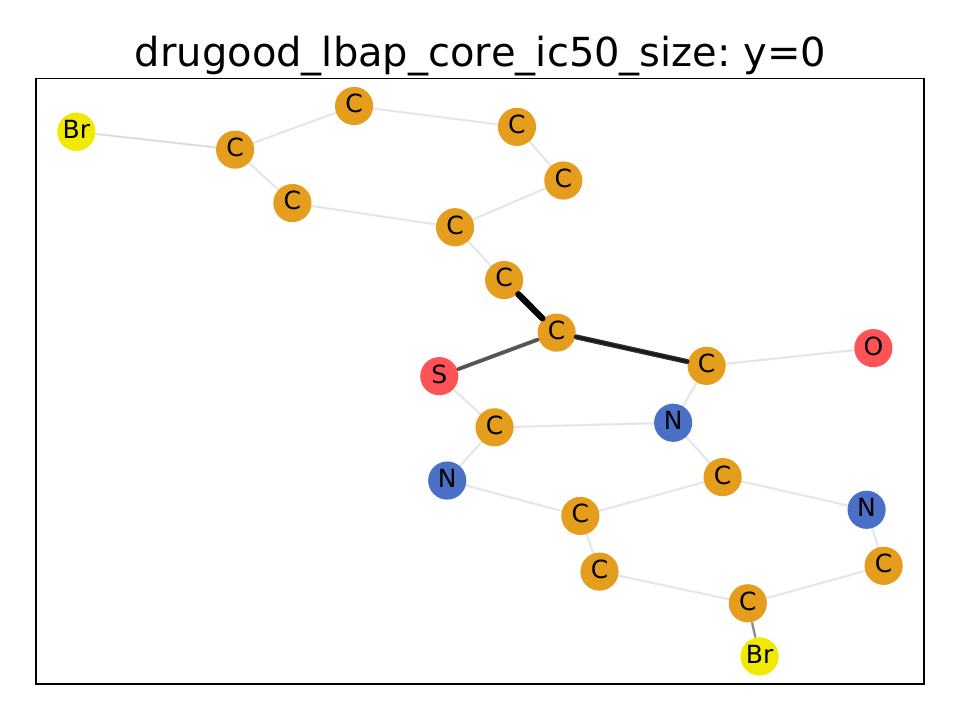}
	}
	\subfigure[]{
		\includegraphics[width=0.31\textwidth]{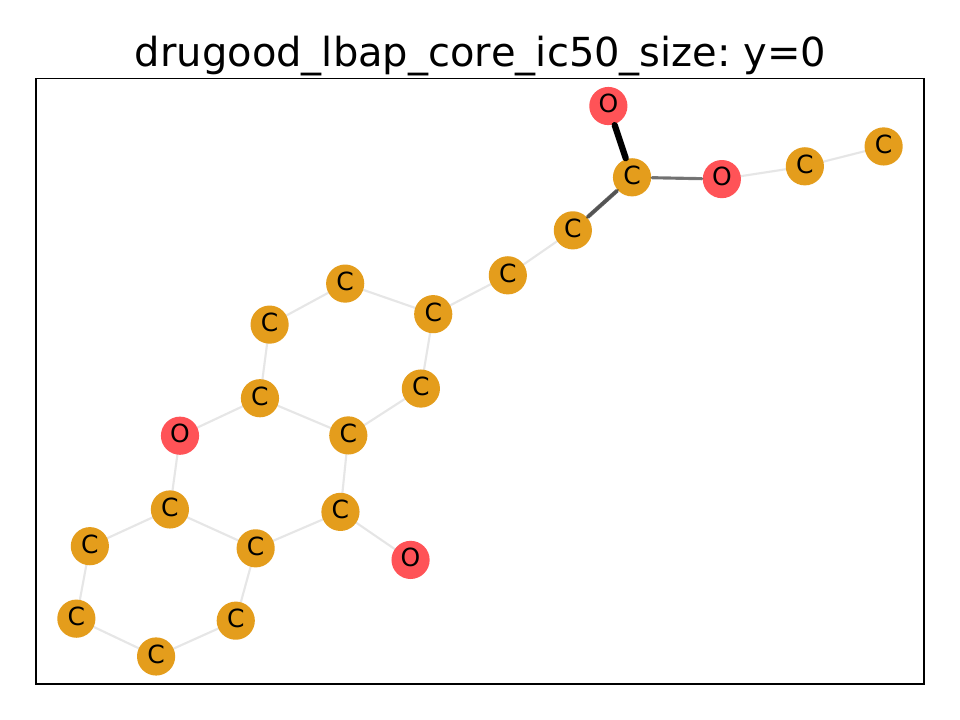}
	}
	\subfigure[]{
		\includegraphics[width=0.31\textwidth]{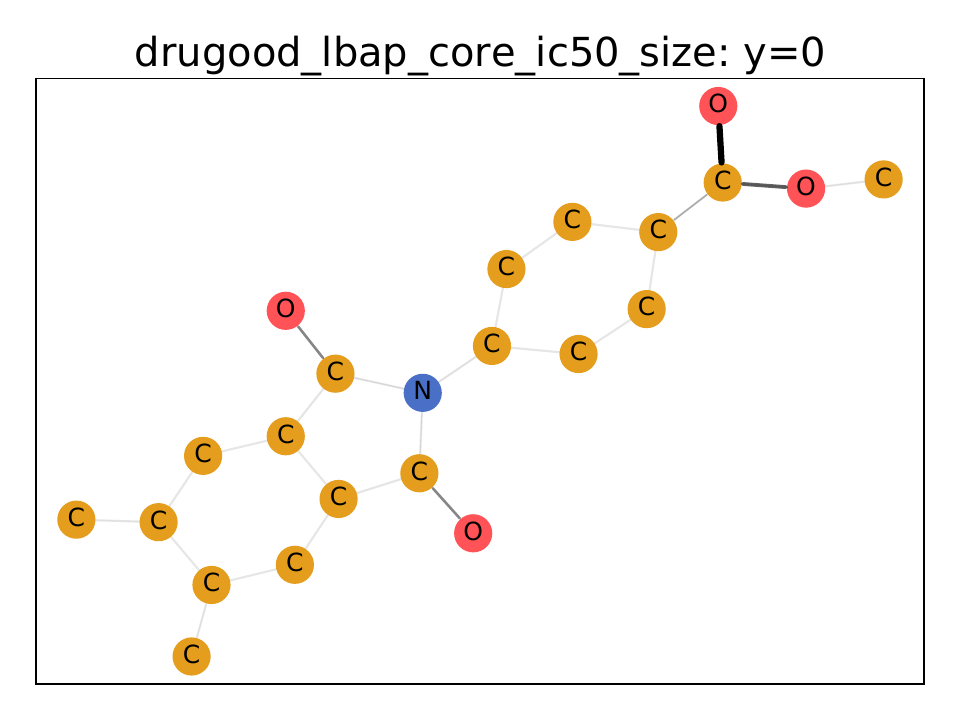}
	}
	\caption{
		Interpretation visualization of inactivate examples ($y=0$) from DrugOOD-Size.}
	\label{CH:CIGA:fig:size_viz_inact_appdx}
\end{figure}

Besides, we also provide some interpretation visualization examples in DrugOOD datasets.
Shown in Fig.~\ref{CH:CIGA:fig:assay_viz_act_appdx} to Fig.~\ref{CH:CIGA:fig:size_viz_inact_appdx}, we use the edge color intensities to denote the attentions of models that pay to the corresponding edge.
Some interesting patterns can be found in the molecules shared with the same label, which could provide insights to the domain experts when developing new drugs.
We believe that, because of its superior OOD generalization performance on graphs, \ciga can have high potential to push forward the developments of AI-Assisted Drug Discovery and enrich the AI tools for facilitating the fundamental practice of science in the future.

\chapter{Appendices of \gala}\label{APP:GALA}

\section{Notations}
\label{CH:GALA:sec:notations_appdx}
Typically, for graphs that appeared in the discussion, we will use the superscript to denote the sampling process (e.g., $G^p$ is the positive graph), and the subscript to denote the specific invariant (i.e., $G_c$) or spurious subgraph (i.e., $G_s$). Graph symbols with $\pred{G}$ are the predicted graphs of a model (i.e., the estimated invariant subgraph $\pred{G}_c$. Below, we list some examples of graphs involved in this paper.
\begin{table}[ht]%
    \caption{Notations for graphs involved in \gala.}
    \centering
    \resizebox{\textwidth}{!}{
        \begin{tabular}{ll}
            \toprule
            \textbf{Symbols}                     & \textbf{Definitions}                                                                                     \\\midrule
            \(\gG\)                              & the graph space                                                                                          \\
            \(\gG_c\)                            & the space of subgraphs with respect to the graphs from $\gG$                                             \\
            \(\gY\)                              & the label space                                                                                          \\
            \(G\in\gG\)                          & a graph                                                                                                  \\
            \(G=(A,X)\)                          & a graph with the adjacency matrix $A\in\{0,1\}^{n\times n}$ and node feature matrix $X\in\R^{n\times d}$ \\
            \(\{G\}\)                            & a set of graphs                                                                                          \\\midrule
            \(G^p\)                              & a graph sampled as positive samples                                                                      \\
            \(G^n\)                              & a graph sampled as negative samples                                                                      \\
            \(G^s\)                              & a graph sampled according to CIGA~\cite{ciga}                                                            \\
            \(G_c\)                              & the invariant subgraph with respect to $G$                                                               \\
            \(G_s\)                              & the spurious subgraph with respect to $G$                                                                \\
            \(G_c^p\)                            & the invariant subgraph of a positive graph $G^p$                                                         \\
            \(G_s^p\)                            & the spurious subgraph of a positive graph $G^p$                                                          \\\midrule
            \(\pred{G}_c\)                       & the estimated invariant subgraph                                                                         \\
            \(\pred{G}_s\)                       & the estimated spurious subgraph                                                                          \\
            \(\pred{G}_c^p\)                     & the estimated invariant subgraph of a positive graph $G^p$                                               \\
            \(\pred{G}_s^p\)                     & the estimated spurious subgraph of a positive graph $G^p$                                                \\
            \(\pa\pred{G}_c\subseteq G_c\)       & the part of the underlying invariant subgraph $G_c$ appeared in $\pred{G}_c$                             \\
            \(\pc\pred{G}_c= G_c-\pa\pred{G}_c\) & the complementary part of $\pa\pred{G}_c$ with respect to the invariant subgraph $G_c$                   \\\midrule
            \bottomrule
        \end{tabular}}
\end{table}

\section{Limitations and Future Directions}
Although our work establishes a set of minimal assumptions for feasible invariant graph learning when the environment partitions and auxiliary information about the environment are both not available, our work is built upon the minimal availability of the environment knowledge.
Nevertheless, there could exist some additional information that may be helpful for environment augmentation.
Therefore, it remains interesting to explore more theoretically grounded strategies to discover and leverage more environment information for identifying graph invariance. When the direct environment augmentation is not feasible, \gala provides a suitable framework that one could easily manipulate the environment assistant model or the partitioning of the positive and negative graphs, to select the spurious features via the additional information and better identify the graph invariance.

In addition to the correlation strengths discussed in this work, there exist other factors, such as the size of spurious and invariant subgraphs, that affect the fitting of spurious and invariant patterns,  another promising future direction is to discuss the influence of these factors to the design of environment assistant model and OOD generalization on graphs.

Besides, a better data partitioning strategy can be developed with uncertainty measures~\citep{contrast_reg}.

\section{Full Details of the Background}
\label{CH:GALA:sec:prelim_appdx}
We give a more detailed background introduction about GNNs and Invariant Learning in this section.

\textbf{Graph Neural Networks.} Let $G=(A,X)$ denote a graph with $n$ nodes and $m$ edges,
where $A \in \{0,1\}^{n\times n}$ is the adjacency matrix, and $X\in \R^{n \times d}$ is the node feature matrix
with a node feature dimension of $d$.
In graph classification, we are given a set of $N$ graphs $\{G_i\}_{i=1}^N\subseteq \gG$
and their labels $\{Y_i\}_{i=1}^N\subseteq\gY=\R^c$ from $c$ classes.
Then, we train a GNN $\rho \circ h$ with an encoder $h:\gG\rightarrow\R^h$ that learns a meaningful representation $h_G$ for each graph $G$ to help predict their labels $y_G=\rho(h_G)$ with a downstream classifier $\rho:\R^h\rightarrow\gY$.
The representation $h_G$ is typically obtained by performing pooling with a $\text{READOUT}$ function on the learned node representations:
\begin{equation}
    \label{CH:GALA:eq:gnn_pooling}
    h_G = \text{READOUT}(\{h^{(K)}_u|u\in V\}),
\end{equation}
where the $\text{READOUT}$ is a permutation invariant function (e.g., $\text{SUM}$, $\text{MEAN}$)~\citep{gin},
and $h^{(K)}_u$ stands for the node representation of $u\in V$ at $K$-th layer that is obtained by neighbor aggregation:
\begin{equation}
    \label{CH:GALA:eq:gnn}
    h^{(K)}_u = \sigma(W_K\cdot a(\{h^{(K-1)}_v\}| v\in\mathcal{N}(u)\cup\{u\})),
\end{equation}
where $\mathcal{N}(u)$ is the set of neighbors of node $u$,
$\sigma(\cdot)$ is an activation function, e.g., $\text{ReLU}$, and $a(\cdot)$ is an aggregation function over neighbors, e.g., $\text{MEAN}$.

\begin{figure*}[ht]
	\centering\hfill
	\subfigure[Graph generation SCM]{\label{CH:GALA:fig:graph_gen_appdx}
		\resizebox{!}{0.225\textwidth}{\tikz{
				\node[latent] (S) {$S$};%
				\node[latent,left=of S,xshift=-1.5cm] (C) {$C$};%
				\node[latent,below=of C,xshift=-0.75cm,yshift=0.5cm] (ZCA) {$Z_X^c$}; %
				\node[latent,below=of C,xshift=0.75cm,yshift=0.5cm] (ZCX) {$Z_A^c$}; %
				\node[latent,below=of S,xshift=-0.75cm,yshift=0.5cm] (ZSA) {$Z_X^s$}; %
				\node[latent,below=of S,xshift=0.75cm,yshift=0.5cm] (ZSX) {$Z_A^s$}; %
				\node[latent,below=of ZCX,xshift=-0.75cm,yshift=0.5cm] (GC) {$G_c$}; %
				\node[latent,below=of ZSX,xshift=-0.75cm,yshift=0.5cm] (GS) {$G_s$}; %
				\node[obs,below=of GC,xshift=1.6cm,yshift=0.5cm] (G) {$G$}; %
				\edge[dashed,-] {C} {S}
				\edge {C} {ZCX,ZCA}
				\edge {S} {ZSX,ZSA}
				\edge {ZCX,ZCA} {GC}
				\edge {ZSX,ZSA} {GS}
				\edge {GC,GS} {G}
			}}}
	\subfigure[FIIF SCM]{\label{CH:GALA:fig:scm_fiif_appdx}
		\resizebox{!}{0.18\textwidth}{\tikz{
				\node[latent] (E) {$E$};%
				\node[latent,below=of E,yshift=0.5cm] (S) {$S$}; %
				\node[obs,below=of E,xshift=-1.2cm,yshift=0.5cm] (Y) {$Y$}; %
				\node[obs,below=of E,xshift=1.2cm,yshift=0.5cm] (G) {$G$}; %
				\node[latent,below=of Y,xshift=1.2cm,yshift=0.5cm] (C) {$C$}; %
				\edge {E} {S}
				\edge {C} {Y,G}
				\edge {S} {G}
				\edge {C} {S}
			}}}
	\subfigure[PIIF SCM]{\label{CH:GALA:fig:scm_piif_appdx}
		\resizebox{!}{0.18\textwidth}{\tikz{
				\node[latent] (E) {$E$};%
				\node[latent,below=of E,yshift=0.5cm] (S) {$S$}; %
				\node[obs,below=of E,xshift=-1.2cm,yshift=0.5cm] (Y) {$Y$}; %
				\node[obs,below=of E,xshift=1.2cm,yshift=0.5cm] (G) {$G$}; %
				\node[latent,below=of Y,xshift=1.2cm,yshift=0.5cm] (C) {$C$}; %
				\edge {E} {S}
				\edge {C} {Y,G}
				\edge {S} {G}
				\edge {Y} {S}
			}}}
	\subfigure[MIIF SCM]{\label{CH:GALA:fig:scm_miif_appdx}
		\resizebox{!}{0.24\textwidth}{\tikz{
				\node[latent] (E) {$E$};%
				\node[latent,below=of E,xshift=-2cm] (S1) {$S_1$}; %
				\node[latent,below=of E,xshift=-0.6cm] (C) {$C$}; %
				\node[latent,below=of E,xshift=2cm] (S2) {$S_2$}; %
				\node[obs,below=of E,xshift=0.6cm] (Y) {$Y$}; %
				\node[obs,below=of C,xshift=0.6cm] (G) {$G$}; %
				\edge {E} {S1,S2}
				\edge {C} {S1,Y,G}
				\edge {Y} {S2}
				\edge {S1,S2} {G}
			}}}
	\caption{Full SCMs on Graph Distribution Shifts~\citep{ciga}.}
	\label{CH:GALA:fig:scm_appdx}
\end{figure*}

\paragraph{Graph generation process.}
This work focuses on graph classification, while the results generalize
to node classification as well using the same setting as in~\citet{handle_node}.
Specifically,
we are given a set of graph datasets $\dataset=\{\dataset_e\}_e$ collected from multiple environments $\envall$.
Samples $(G^e_i, Y^e_i)\in \dataset^e$ from the same
environment are considered as drawn independently from an identical distribution $\sP^e$.
We consider the graph generation process proposed
by~\citet{ciga} that covers a broad case of graph distribution shifts.
Fig.~\ref{CH:GALA:fig:scm_appdx} shows the full graph generation process considered in~\citet{ciga}.
The generation of the observed graph $G$ and labels $Y$
are controlled by a set of latent causal variable $C$ and spurious variable $S$, i.e.,
\[G\coloneqq f_\gen(C,S).\]
$C$ and $S$ control the generation of $G$ by controlling the underlying invariant subgraph $G_c$
and spurious subgraph $G_s$, respectively.
Since $S$ can be affected by the environment $E$,
the correlation between $Y$, $S$ and $G_s$ can change arbitrarily
when the environment changes.
$C$ and $S$ control the generation of the underlying invariant subgraph $G_c$
and spurious subgraph $G_s$, respectively.
Since $S$ can be affected by the environment $E$,
the correlation between $Y$, $S$ and $G_s$ can change arbitrarily
when the environment changes.
Besides, the latent interaction among $C$, $S$ and $Y$
can be further categorized into \emph{Full Informative Invariant Features} (\emph{FIIF})
when $Y\ind S|C$ and \emph{Partially Informative Invariant Features} (\emph{PIIF}) when $Y \not\ind S|C$. Furthermore, PIIF and FIIF shifts can be mixed together and yield \emph{Mixed Informative Invariant Features} (\emph{MIIF}), as shown in Fig.~\ref{CH:GALA:fig:scm_appdx}.
We refer interested readers to~\citet{ciga} for a detailed introduction of the graph generation process.

\paragraph{Invariant graph representation learning.}
To tackle the OOD generalization challenge
on graphs from Fig.~\ref{CH:GALA:fig:scm_appdx},
the existing invariant graph learning approaches generically
aim to identify the underlying invariant subgraph $G_c$ to predict the label $Y$~\citep{handle_node,ciga}.
Specifically, the goal of OOD generalization on graphs
is to learn an \emph{invariant GNN} $f\coloneqq f_c\circ g$,
which is composed of two modules:
a) a featurizer $g:\gG\rightarrow\gG_c$ that extracts the invariant subgraph $G_c$;
b) a classifier $f_c:\gG_c\rightarrow\gY$ that predicts the label $Y$ based on the extracted $G_c$,
where $\gG_c$ refers to the space of subgraphs of $\gG$.
The learning objectives of $f_c$ and $g$ are formulated as
\begin{equation}
    \label{CH:GALA:eq:inv_cond_appdx}
    \text{$\max$}_{f_c, \; g} \ I(\pred{G}_{c};Y), \ \text{s.t.}\ \pred{G}_{c}\ind E,\ \pred{G}_{c}=g(G).
\end{equation}
Since $E$ is not observed, many strategies are proposed to
impose the independence of $\pred{G}_c$ and $E$.
A common approach is to augment the environment information.
For example, based on the estimated invariant subgraphs $\pred{G}_c$ and spurious subgraphs $\pred{G}_s$,
\citet{dir,grea,handle_node} proposed to generate new environments, while \citet{moleood,gil} proposed to infer the underlying environment labels.
However, we show that it is fundamentally impossible to augment faithful environment information in Sec.~\ref{CH:GALA:sec:env_aug_failure}.
\citet{gib,vgib,gsat,dps,lri} adopt graph information bottleneck to tackle FIIF graph shifts, and they cannot generalize to PIIF shifts.
Our work focuses on PIIF shifts, as it is more challenging when without environment labels~\citep{zin}.
\citet{disc} generalized~\citep{ldd} to tackle severe graph biases, i.e., when $H(S|Y)< H(C|Y)$.
\citet{ciga} proposed a contrastive framework to tackle both
FIIF and PIFF graph shifts, but are limited to $H(S|Y)> H(C|Y)$.
However, in practice, it is usually unknown whether $H(S|Y)< H(C|Y)$ or $H(S|Y)> H(C|Y)$ without environment information.

\paragraph{More OOD generalization on graphs.}
In addition to the aforementioned invariant learning approaches, \citet{size_gen1,size_gen2,size_gen3,graph_extrapolation} study the OOD generalization as an extrapolation from small graphs to larger graphs in the task of graph classification and link prediction. In contrast, we study OOD generalization against various graph distribution shifts formulated in Fig.~\ref{CH:GALA:fig:scm_appdx}. In addition to the standard OOD generalization tasks studied in this paper, \citet{nn_extrapo,OOD_CLRS} study the OOD generalization in tasks of algorithmic reasoning on graphs. \citet{graph_ttt} study the test-time adaption in the graph regime. \citet{shape_matching} study the 3D shape matching under the presence of noises. \citet{LECI} propose an independence constraint onto the target label and environment label to improve the OOD generalization when environment labels are available.
\citet{flood}  adopt a flexible framework to tackle shifting graph distributions.
\citet{hao,zhou2023combating,zhou2023mcgra,Tao2023IDEAIC} study the OOD generalization on graphs from the adversarial robustness perspective.

In addition to graph classification, \citet{handle_node,causality_bianode} study node classification. \citet{struc_reweight} propose a structural reweighting strategy to improve the OOD generalization of node classification. \citet{multimodule_gnn} propose to incorporate multiple modules to handle different degree modes in OOD node classification. \citet{gda,conda} study unsupervised graph domain adaption.\citet{zhou2022ood,Gao2023DoubleEF,Zhou2023AnOM} study the OOD link prediction.

Besides, \citet{cfxgnn} aims to find counterfactual subgraphs for explaining GNNs, which focuses on post-hoc explainability while this work focuses on intrinsic interpretability.

\paragraph{Invariant learning without environment labels.}
There are also plentiful studies in invariant learning without environment labels.
\citet{eiil} proposed a min-max formulation to infer the environment labels.
\citet{hrm} proposed a self-boosting framework based on the estimated invariant and variant features.
\citet{jtt,cnc} proposed to infer labels based on the predictions of an ERM trained model.
\citet{xrm,pde} improve the inference of group labels based on feature learning and prediction correctness.
However, \citet{zin} found failure cases in Euclidean data
where it is impossible to identify the invariant features without given environment labels.
Moreover, as the OOD generalization on graphs is fundamentally more difficult than Euclidean data~\citep{ciga}, the question about the feasibility of learning invariant subgraphs without environment labels remains unanswered.

\section{More Details about the Failure Cases}
\label{CH:GALA:sec:fail_appdx}

\begin{figure}[ht]
    \centering
    \subfigure[Failures of env. generation]{
        \includegraphics[width=0.31\textwidth]{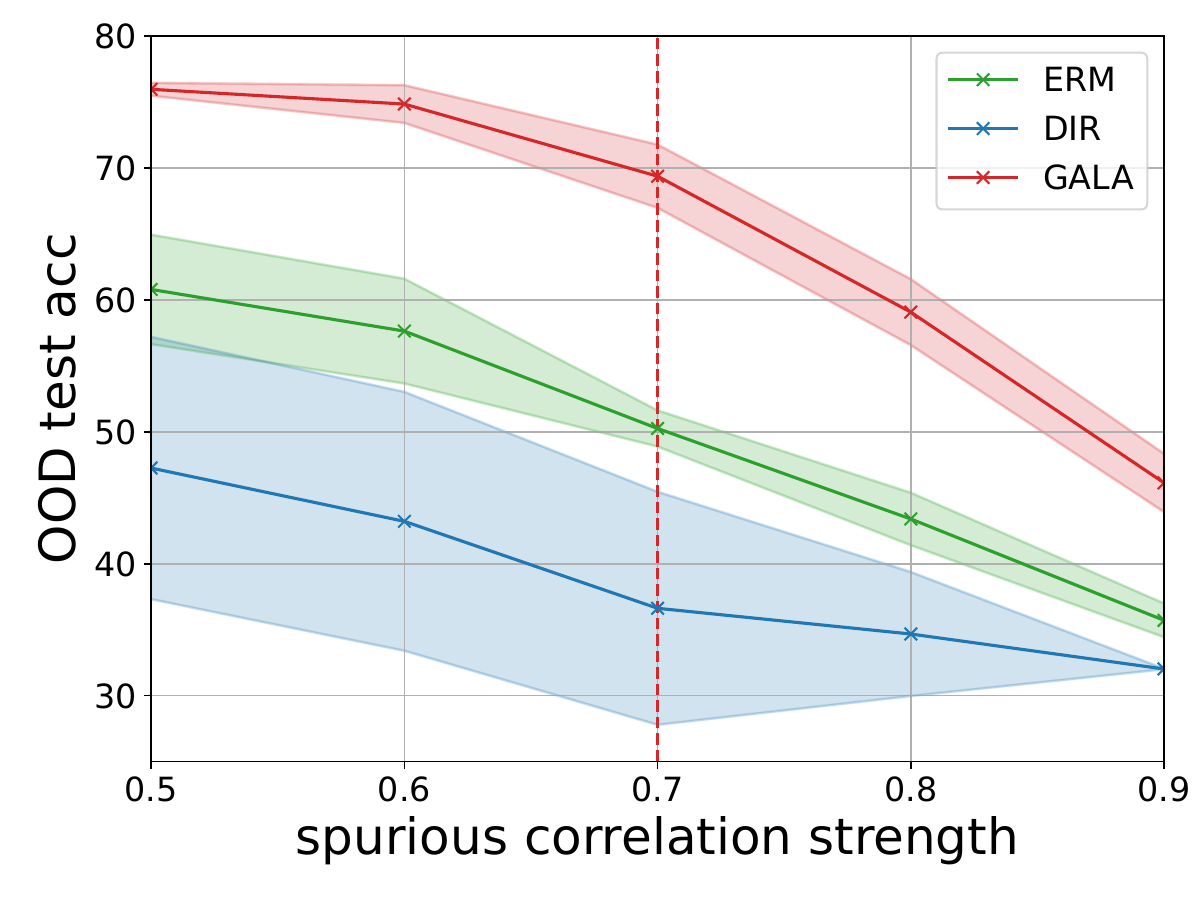}
        \label{CH:GALA:fig:grea_fail_p1_appdx}
    }
    \subfigure[Failures of env. inferring]{
        \includegraphics[width=0.31\textwidth]{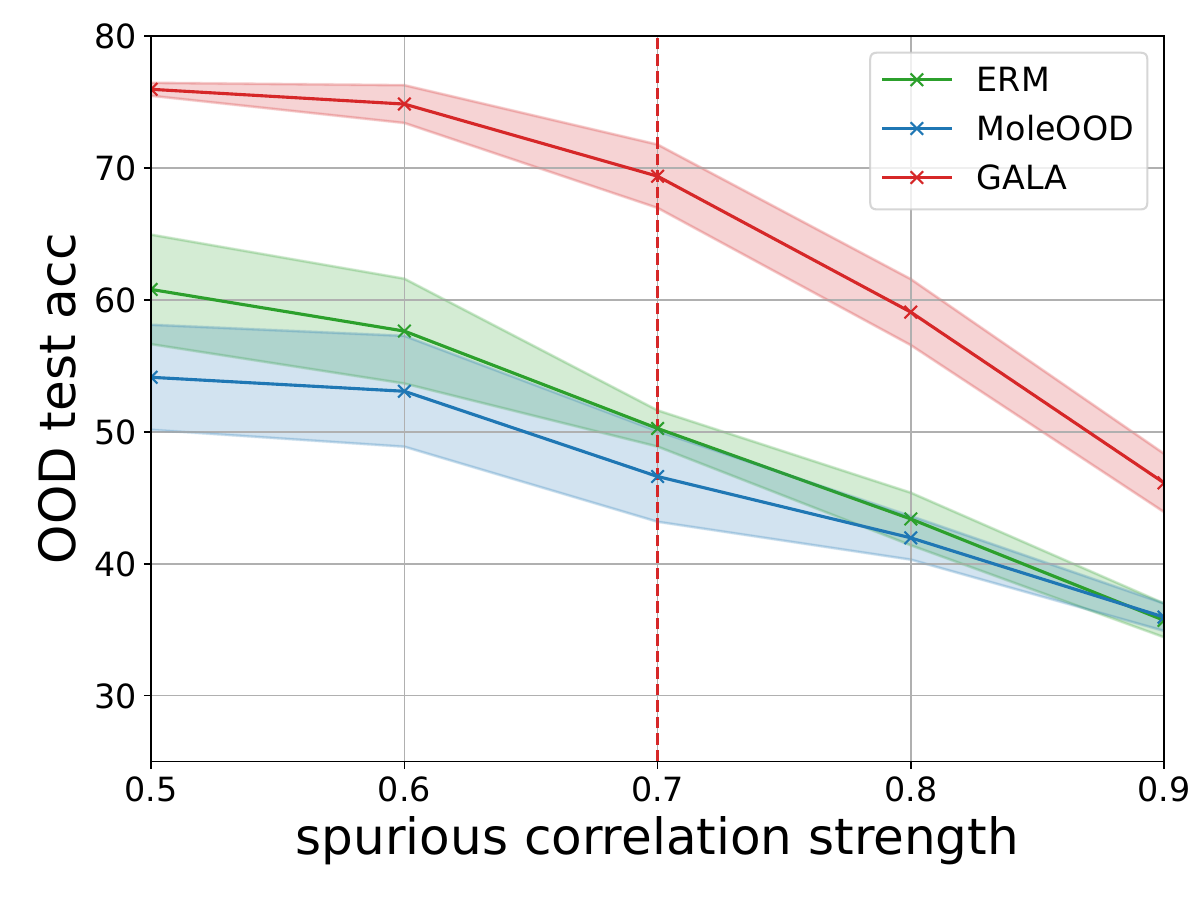}
        \label{CH:GALA:fig:gil_fail_p1_appdx}
    }
    \subfigure[Failures of resolving env. consistency]{
        \includegraphics[width=0.31\textwidth]{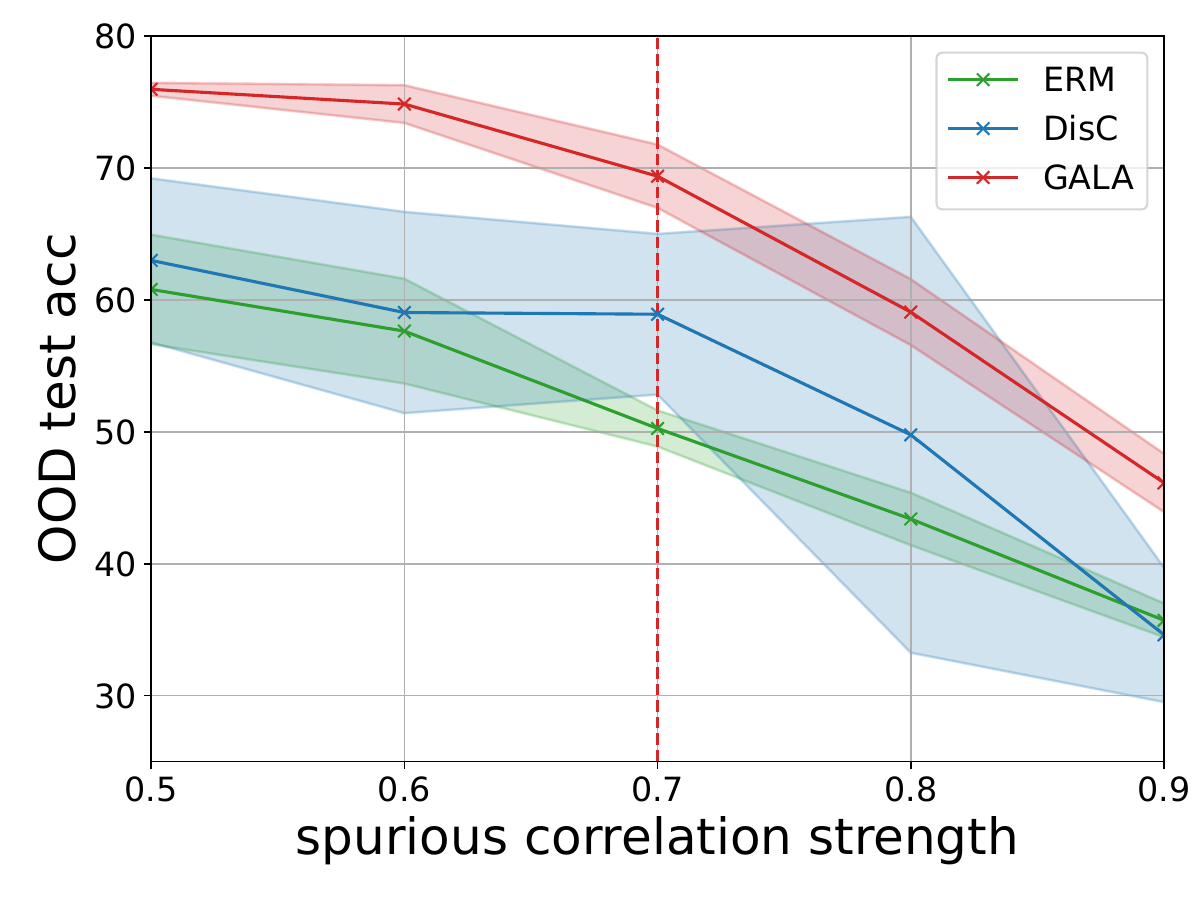}
        \label{CH:GALA:fig:disc_fail_p1_appdx}
    }
    \subfigure[Failures of env. generation]{
        \includegraphics[width=0.31\textwidth]{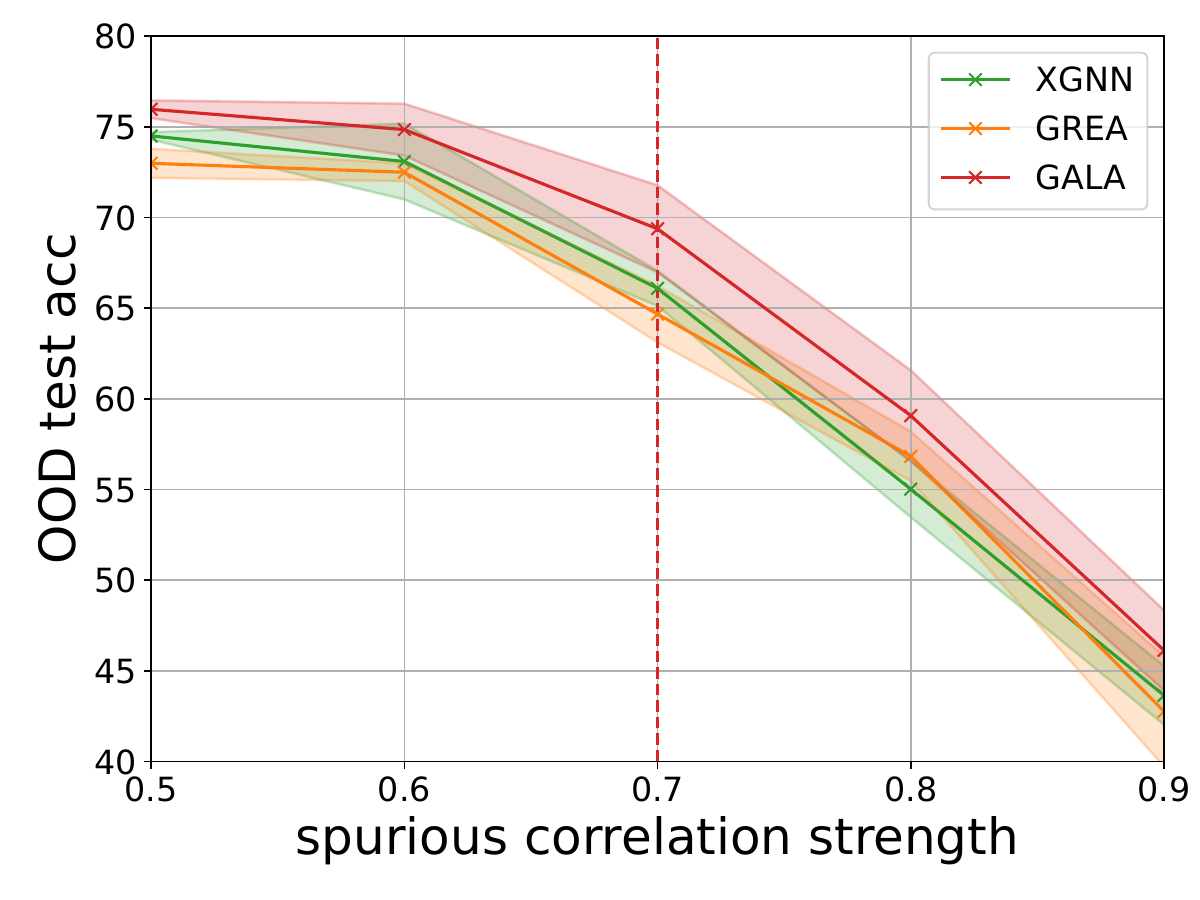}
        \label{CH:GALA:fig:grea_fail_p2_appdx}
    }
    \subfigure[Failures of env. inferring]{
        \includegraphics[width=0.31\textwidth]{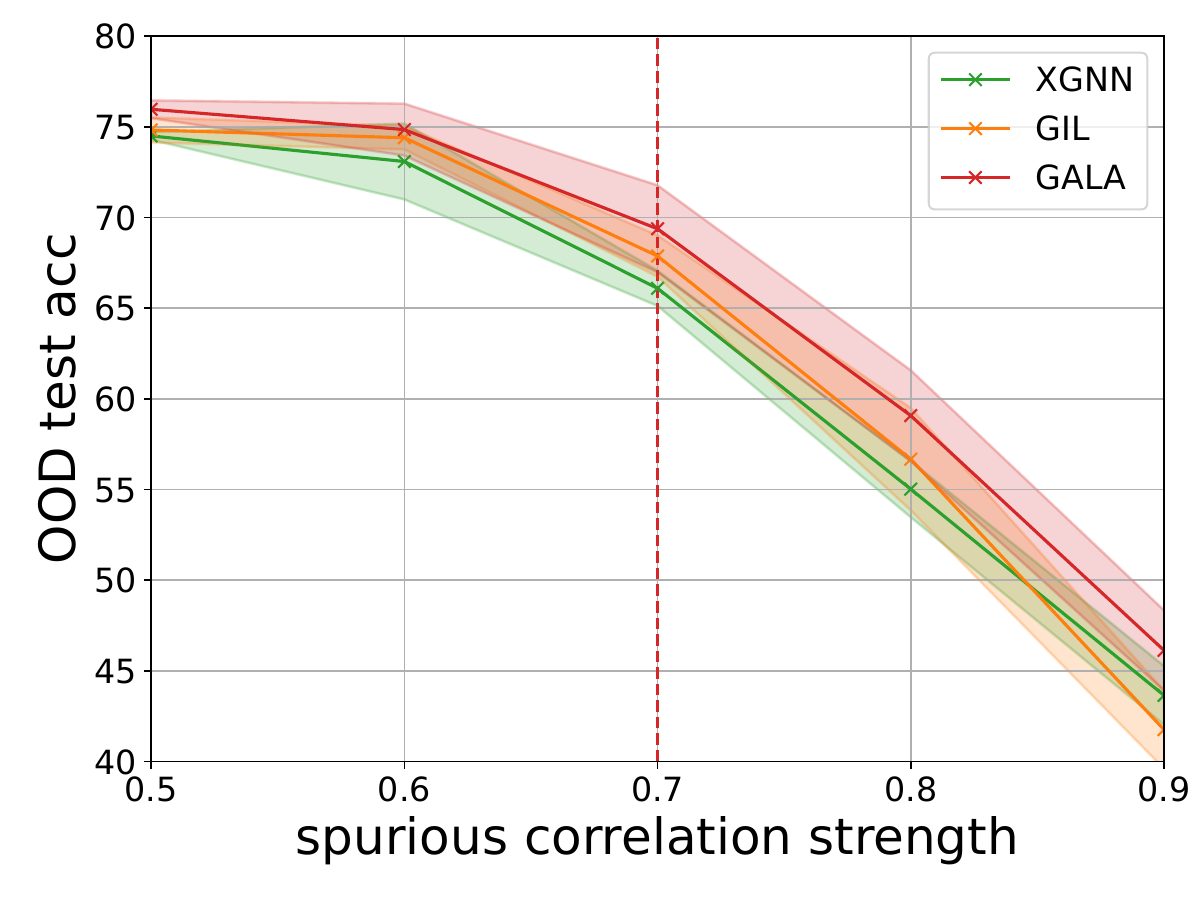}
        \label{CH:GALA:fig:gil_fail_p2_appdx}
    }
    \subfigure[Failures of resolving env. consistency]{
        \includegraphics[width=0.31\textwidth]{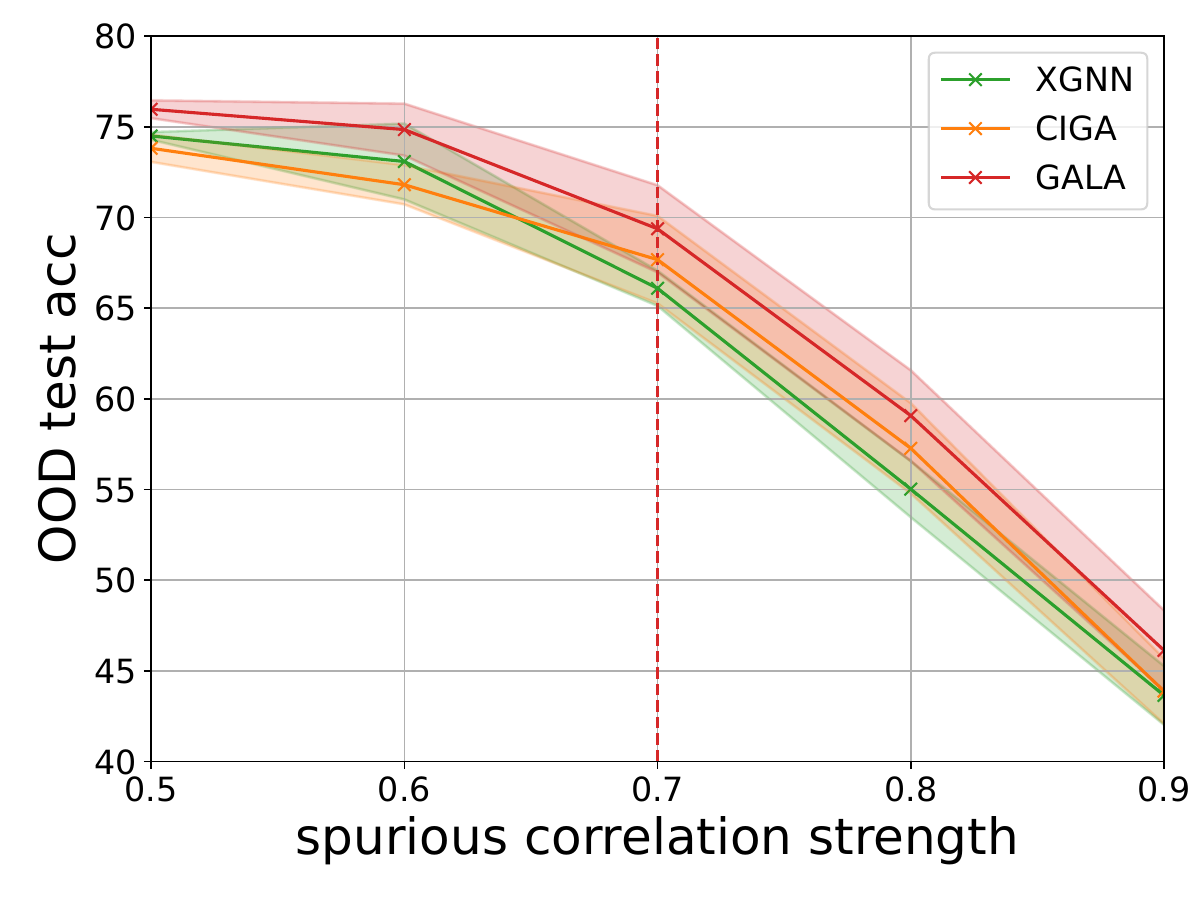}
        \label{CH:GALA:fig:disc_fail_p2_appdx}
    }
    \caption[Failures of finding faithful environment information for OOD generalization on graphs.]{
        Failures of finding faithful environment information.
        Results shown in the figure are based on the $3$ class two-piece graphs (Def.~\ref{def:twobit_graph_appdx}),
        where the invariant correlation strength is fixed as $0.7$
        while the spurious correlation strength is varied from $0.5$ to $0.7$. We can find that both environment augmentation and inferring approaches suffer from severe performance decreases or even underperform ERM and XGNN when the dominated correlation is not suitable for the method. In contrast, \gala maintains strong OOD performance for both cases.}
    \label{CH:GALA:fig:fail_appdx}
\end{figure}

We provide more empirical results and details about the failure case verification experiments in complementary to Sec.~\ref{CH:GALA:sec:env_aug_failure}.
The results are shown in Fig.~\ref{CH:GALA:fig:fail_appdx}. We compared different environment augmentation approaches the vanilla GNN model trained with ERM (termed ERM), and an interpretable GNN model trained with ERM (termed XGNN).

The failure cases are constructed according to the two-piece graph generation models. The specific description is given as the following.

\begin{definition}[$3$-class two-piece graphs]
    \label{def:twobit_graph_appdx}
    Each environment is defined with two parameters, $\alpha_e,\beta_e\in[0,1]$,  and the dataset $\dataset_e$ is generated as follows:
    \begin{enumerate}[label=(\alph*)]
        \item Sample $y^e\in\{0,1,2\}$ uniformly;
        \item Generate $G_c$ and $G_s$ via :
              \[G_c\coloneqq f_\gen^{G_c}(Y\cdot\rad(\alpha_e)),\ G_s\coloneqq f_\gen^{G_s}(Y\cdot\rad(\beta_e)),\]
              where $f_\gen^{G_c},f_\gen^{G_s}$
              respectively map input $\{0,1,2\}$ to a specific graph selected from a given set,
              and $\rad(\alpha)$ is a random variable with probability $\alpha$ taking a uniformly random value from $\{0,1,2\}$, and
              a probability of $1-\alpha$ taking the value of $+1$;
        \item Sythesize $G$ by randomly concatenating $G_c$ and $G_s$:
              \[G\coloneqq f_\gen^{G}(G_c,G_s).\]
    \end{enumerate}
\end{definition}
In experiments, we implement the $3$-class two-piece graphs with the BA-motifs~\citep{pge} model.

In experiments, we adopt a $3$-layer GIN~\citep{gin} with a hidden dimension of $32$ and a dropout rate of $0.0$ as the GNN encoder. The XGNN architecture is implemented via two GNNs following the original implementation as CIGA.
The optimization is proceeded with Adam~\citep{adam} using a learning rate of $1e-3$. All experiments are repeated with $5$ different random seeds of $\{1,2,3,4,5\}$. The mean and standard deviation are reported from the $5$ runs.

We implement DIR~\citep{dir}, GREA~\citep{grea}, MoleOOD~\citep{moleood}, GIL~\citep{gil}, DisC~\citep{disc}, and CIGA~\citep{ciga}, according to the author provided codes (if available).
As for the hyperparameters in each method, we use a penalty weight of $1e-2$ for DIR following its original experiment in spurious motif datasets generated similarly using BA-motifs~\citep{dir}. We use a penalty weight of $1$ for GREA as we empirically it does not affect the performance by changing to different weights.
For MoleOOD and GIL, we set the number of environments as $3$. We tune the penalty weights of MoleOOD with values from $\{1e-2,1e-1,1,10\}$ but did not observe much performance differences. We tune the penalty weights of GIL with values from $\{1e-5,1e-3,1e-1\}$ recommended by the authors. For DisC, we tune only the $q$ weight from $\{0.9,0.7,0.5\}$ in the GCE loss as we did not observe performance differences by changing the weight of the other term. We tune the penalty weight of CIGA with values from $\{0.5,1,2,4,8,16,32\}$ as recommended by the authors.

\section{Proofs for Theorems and Propositions}
\label{CH:GALA:sec:theory_appdx}

\subsection{Proof of Proposition~\ref{CH:GALA:thm:env_gen_fail}}\label{CH:GALA:proof:env_gen_fail_appdx}
\begin{proposition}(Restatement of Proposition~\ref{CH:GALA:thm:env_gen_fail})\label{CH:GALA:thm:env_gen_fail_appdx}
    Consider the two-piece graph dataset $\envtrain=\{(\alpha,\beta_1),(\alpha,\beta_2)\}$ with $\alpha\geq\beta_1,\beta_2$
    (e.g., $\envtrain=\{(0.25,0.1),(0.25,0.2)\}$),
    and its corresponding mixed environment $\envmix=\{(\alpha,(\beta_1+\beta_2)/2\}$ (e.g., $\envmix=\{(0.25,0.15)\}$).
    When $\widehat{G}_c=G_s$ and $\widehat{G}_s=G_c$, it holds that the augmented environment $\env_v$ is also a two-piece graph dataset with
    \[
        \mathcal{E}_v = \{(0.5,(\beta_1 + \beta_2)/2)\}\text{ (e.g., $\mathcal{E}_v = \{(0.5,0.15)\}$)}.
    \]
\end{proposition}
\begin{proof}
    From Definition \ref{def:twobit_graph}, we known that for each graph $G_i \sim \envmix = \{(\alpha, (\beta_1 + \beta_2)/2)\}$, $G_i$ is the concatenation of the $G^i_c$ and $G^i_s$ defined as
    \[
        G^i_c \coloneqq f^{G_c}_\gen (Y_i \cdot \rad(\alpha)_i), \quad G^i_s \coloneqq f^{G_s}_\gen (Y_i \cdot \rad((\beta_1 + \beta_2)/2)_i),
    \]
    where $\rad(\cdot)_i$ denotes the $i$th sample of the random variable $\rad(\cdot)$.

    Denote
    \[G_A=f^{G_c}_\gen (+1),\ G_B=f^{G_c}_\gen(-1),\]
    and
    \[G_C=f^{G_s}_\gen (+1),\ G_D=f^{G_s}_\gen(-1),\]
    Considering applying the augmentation to $2n$ samples randomly sampled from $\envmix$,
    since the featurizer $g$ separates each $G\in\envmix$ into $\pred{G}_c=G_s$ and $\pred{G}_s=G_c$,
    and the augmented graph $G^i$ is obtained by
    \[G^{i,j}=f_\gen^G(\pred{G}^i_c,\ \pred{G}^j_s),\forall i,j\in \{1...n\}.\]
    Then, the new $\alpha_v, \beta_v$ in $\env_v$ can be obtained by summing up the overall numbers
    of $G_A,G_B,G_C,G_D$ concatenated into $2n^2$ samples in $\env_v$.

    Specifically, we can inspect the changes in the distributions of motifs and labels.
    Let $\bar{\beta}=(\beta_1 + \beta_2)/2$, without loss of generality, we focus on inspecting the changes given $Y=+1$,
    since the changes given $Y=-1$ is symmetric as $Y=+1$. The original distribution is shown as follows:
    \begin{center}
        \begin{tabular}{ |c|c|c| }
            \hline
            $Y=+1$ & $G_A$                        & $G_B$                    \\\hline
            $G_C$  & $(1-\alpha)(1-\bar{\beta})n$ & $\alpha(1-\bar{\beta})n$ \\\hline
            $G_D$  & $(1-\alpha)\bar{\beta}n$     & $\alpha\bar{\beta}n$     \\
            \hline
        \end{tabular}
    \end{center}
    Then, new distributions of the motifs and labels are determined by
    the number of original motifs identified as $\pred{G}_c$ and $\pred{G}_s$, respectively.
    When $\pred{G}_c=G_s$ and $\pred{G}_s=G_c$,
    in the new environment $\env_v$, given $Y=+1$, $G_C$ contributes $(1-\bar{\beta})n*2n$ samples as the ``invariant'' subgraph.
    More specifically, $G_C$ will be concatenated with $G_A$ and $G_B$ by $n$ times, respectively.
    Then we have the new distribution tables shown as follows:
    \begin{center}
        \begin{tabular}{ |c|c|c| }
            \hline
            $Y=+1$ & $G_A$                & $G_B$                \\\hline
            $G_C$  & $(1-\bar{\beta})n^2$ & $(1-\bar{\beta})n^2$ \\\hline
            $G_D$  & $\bar{\beta}n^2$     & $\bar{\beta}n^2$     \\
            \hline
        \end{tabular}
    \end{center}
    Since given the same $Y$, the spurious subgraph $G_C$ and $G_D$ will still have the same chance of being flipped, we have $\beta_v=\bar{\beta}$.
    While $G_A$ and $G_B$ appear the same times given the same $Y$, it suffices to know that $\alpha_v=0.5$.
\end{proof}

\subsection{Complementary discussion for Sec.~\ref{CH:GALA:sec:var_sufficiency}}\label{CH:GALA:proof:var_sufficiency_appdx}
\begin{proposition}\label{CH:GALA:thm:var_sufficiency_appdx}
    Given the same graph generation process as in Fig.~\ref{CH:GALA:fig:scm},
    when there exists spurious subgraph $G_s$
    such that $P^{e_1}(Y|G_s)=P^{e_2}(Y|G_s)$ for any two environments $e_1,e_2\in \envtrain$,
    where $P^e(Y|G_s)$ is the conditional distribution $P(Y|G_s)$ under environment $e\in \envall$,
    it is impossible for any learning algorithm applied to $f_c\circ g$ to differentiate $G_c$ from $G_s$.
\end{proposition}
\begin{proof}
    Let $G_s^*$ be the spurious subgraph such that $P^{e_1}(Y|G_s)=P^{e_2}(Y|G_s)$ for any two environments $e_1,e_2\in \envtrain$,
    and $G_c$ be the invariant subgraph which $P^{e_1}(Y|G_c)=P^{e_2}(Y|G_c),\ \forall e_1,e_2\in \envtrain$ by definition.
    Consider a learning algorithm applied to $f_c\circ g$ that accepts the input of $\envmix$,
    and extracts a subgraph $\pred{G}_c=g(Y)$ as an estimation of the invariant subgraph for any $G$ to
    predict $Y$ via $f_c(\pred{G}_c)$ in a deterministic manner.
    If the algorithm succeed to extract $G_c$ from $\envmix$, then there always exists a $\envmix'$ with the desired spurious subgraph $G_s'$
    and a underlying invariant subgraph $G_c'$, such that $G_s'=G_c$ and $G_c'=G_s^*$.
    Due to the deterministic nature, the algorithm fails to identify $G_c'$ in $\envmix'$.
\end{proof}

\subsection{Proof of Proposition \ref{CH:GALA:thm:env_infer_fail}}\label{CH:GALA:proof:env_infer_fail_appdx}
\begin{proposition}(Restatement of Proposition~\ref{CH:GALA:thm:env_infer_fail})\label{CH:GALA:thm:env_infer_fail_appdx}
    There exist $2$ two-piece graph training environments $\envtrain$ and $\envtrain'$ that share the same joint distribution $P(Y, G)$. Any learning algorithm will fail in either $\envtrain$ or $\envtrain'$.
\end{proposition}
\begin{proof}
    Let the mixed training environment of $\envtrain$ and $\envtrain'$ be $\envmix = \{(\alpha, \beta)\}$. Based on the definition of two-piece graphs (Definition \ref{def:twobit_graph}), the joint distribution of the mixed training dataset $(G = \textup{Concat}[G_c, G_s], Y)$ can be computed as
    \[
        \begin{cases}
            Y=+1,                                                     & \text{with probability } 0.5,                     \\
            Y=-1,                                                     & \text{with probability } 0.5,                     \\
            \textup{Bit}^{G_c}(G_c) = \textup{Bit}^{G_s}(G_s) = Y,    & \text{with probability } (1 - \alpha)(1 - \beta), \\
            \textup{Bit}^{G_c}(G_c)\neq \textup{Bit}^{G_s}(G_s) = Y,  & \text{with probability } \alpha(1 - \beta),       \\
            \textup{Bit}^{G_s}(G_s)\neq \textup{Bit}^{G_c}(G_c) = Y,  & \text{with probability } (1 - \alpha)\beta,       \\
            \textup{Bit}^{G_c}(G_c) = \textup{Bit}^{G_s}(G_s) \neq Y, & \text{with probability } \alpha\beta.
        \end{cases}
    \]
    Here we use $\textup{Bit}^{G_c}(G_c)$ to obtain the input bit of a subgraph $G_c$ (or $(f^{G_c}_\gen)^{-1}$),
    and $\textup{Bit}^{G_s}(G_s)$ for $G_s$, respectively.

    Any learning algorithm that tries to identify the invariant subgraph from this training dataset will compute a model that uses subgraph $G_c$, or subgraph $G_s$, or both $G_c$ and $G_s$ to
    predict $Y$ deterministically. Thus, as long as the joint distribution does not change, the resulting model will always identify the same invariant subgraph. Without loss of generality, let us assume that the model correctly identifies $G_c$ as the invariant subgraph for $\envtrain = \{(\alpha, \beta_1), (\alpha, \beta_2)\}$ with $\beta = (\beta_1 + \beta_2)/2$.

    Now let the other training environment be $\envtrain' = \{(\alpha_1, \beta),(\alpha_2, \beta)\}$ with $\alpha = (\alpha_1 + \alpha_2)/2$. It is clear that since the mixed training environment of $\envtrain'$ is still $\{(\alpha, \beta)\}$,
    the model keeps regarding $G_c$ as the invariant subgraph. However, for $\envtrain'$, the model fails to identify the invariance since now the invariant subgraph is $G_s$.

\end{proof}

\subsection{Proof of Corollary~\ref{CH:GALA:thm:var_consistency}}\label{CH:GALA:proof:var_consistency_appdx}
\begin{corollary}(Restatement of Corollary~\ref{CH:GALA:thm:var_consistency})\label{CH:GALA:thm:var_consistency_appdx}
    Without Assumption~\ref{CH:GALA:assump:var_sufficiency} or Assumption~\ref{CH:GALA:assump:var_consistency},
    there does not exist a learning algorithm that captures the invariance of the two-piece graph environments.
\end{corollary}
\begin{proof}
    The proof for lacking Assumption~\ref{CH:GALA:assump:var_sufficiency} is identical to the proof for Proposition~\ref{CH:GALA:thm:var_sufficiency_appdx}.
    Consider a learning algorithm applied to $f_c\circ g$ that accepts the input of $\envmix$,
    and extracts a subgraph $\pred{G}_c=g(Y)$ as an estimation of the invariant subgraph for any $G$ to
    predict $Y$ via $f_c(\pred{G}_c)$ in a deterministic manner.
    Without the holding of Assumption~\ref{CH:GALA:assump:var_consistency}, due to Proposition~\ref{CH:GALA:thm:env_infer_fail}, there exists $\envmix'$ for each $\envmix$ that have the identical joint distribution but different underlying invariant subgraph. Thus, any learning algorithm that succeeds in either $\envmix$ or $\envmix'$ will fail in the other.
\end{proof}

\subsection{Proof of Theorem~\ref{CH:GALA:thm:gala_success}}\label{CH:GALA:proof:gala_success_appdx}
\begin{theorem}(Restatement of Theorem~\ref{CH:GALA:thm:gala_success})\label{CH:GALA:thm:gala_success_appdx}
    Given, i) the same data generation process as in Fig.~\ref{CH:GALA:fig:scm};
    ii) $\train$ that satisfies variation sufficiency (Assumption~\ref{CH:GALA:assump:var_sufficiency})
    and variation consistency (Assumption~\ref{CH:GALA:assump:var_consistency});
    iii) $\{G^p\}$ and $\{G^n\}$ are distinct subsets of $\train$ such that
    $I(G_s^p;G_s^n|Y)=0$,
    $\forall G_s^p =\argmax_{\pred{G}_s^p}I(\pred{G}_s^p;Y)$ under $\{G^p\}$, and
    $\forall G_s^n =\argmax_{\pred{G}_s^n}I(\pred{G}_s^n;Y)$ under $\{G^n\}$;
    suppose $|G_c|=s_c,\ \forall G_c$,
    resolving the following \gala objective elicits an invariant GNN defined via Eq.~\ref{CH:GALA:eq:inv_cond_appdx},
    \begin{equation}
        \label{CH:GALA:eq:gala_sol_appdx}
        \max_{f_c, g} \ I(\pred{G}_{c};Y), \ \text{s.t.}\
        g\in\argmax_{\hat{g},|\pred{G}_c^p|\leq s_c}I(\pred{G}_c^p;\pred{G}_c^n|Y),
    \end{equation}
    where $\pred{G}_c^p\in \{\pred{G}_{c}^p=g({G}^p)\}$
    and $\pred{G}_c^n\in \{\pred{G}_{c}^n=g({G}^n)\}$
    are the estimated invariant subgraphs via $g$ from $\{G^p\}$ and $\{G^n\}$, respectively.
\end{theorem}

\begin{proof}

    Without loss of generality, we assume that $\{G^p\}$ has the same spurious dominance situation as $\envtrain$.
    In other words, when $H(S|Y)<H(C|Y)$, the data distribution in $\{G^p\}$ also follows $H(S|Y)<H(C|Y)$, while $H(S|Y)>H(C|Y)$ in $\{G^n\}$.
    To proceed, we will use the language of~\citet{ciga}.

    We begin by discussing the case of $H(S|Y)<H(C|Y)$. Given $H(S|Y)<H(C|Y)$, we have $H(S|Y)<H(C|Y)$ in $\{G^p\}$ and $H(S|Y)>H(C|Y)$ in $\{G^n\}$.
    Then, we claim that
    \begin{equation}\label{CH:GALA:eq:gala_sol_spu_appdx}
        G_c\in\argmax_{\pred{G}_c^p,|\pred{G}_c^p|\leq s_c}I(\pred{G}_c^p;\pred{G}_c^n|Y).
    \end{equation}
    Otherwise, consider there exists a subgraph of the spurious subgraph $\pa\pred{G}_s^p\subseteq G_s^p$ in $\pred{G}_c^p$,
    which takes up the space of $\pa\pred{G}_c^p\subseteq G_c^p$ from $\pred{G}_c^p$.
    Then, let $\pc\pred{G}_c^p=G_c^p-\pa\pred{G}_c^p$
    we can inspect the changes to $I(\pred{G}_c^p;\pred{G}_c^n|Y)$ led by $\pa\pred{G}_s^p$:
    \begin{equation}
        \label{CH:GALA:eq:pa_delta_appdx}
        \begin{aligned}
             & \triangle I(\pred{G}_c^p;\pred{G}_c^n|Y)                                                                       \\& = \triangle H(\pred{G}_c^p|Y)-\triangle H(\pred{G}_c^p|\pred{G}_c^n,Y) \\
             & =\left[H(\pc\pred{G}_c^p,\pa\pred{G}_s^p|Y)-H(\pc\pred{G}_c^p,\pa\pred{G}_c^p|Y)\right]-
            \left[H(\pc\pred{G}_c^p,\pa\pred{G}_s^p|\pred{G}_c^n,Y)-H(\pc\pred{G}_c^p,\pa\pred{G}_c^p|\pred{G}_c^n,Y)\right]  \\
             & =\left[H(\pa\pred{G}_s^p|\pc\pred{G}_c^p,Y)-H(\pa\pred{G}_c^p|\pc\pred{G}_c^p,Y)\right]-
            \left[H(\pa\pred{G}_s^p|\pc\pred{G}_c^p,\pred{G}_c^n,Y)-H(\pa\pred{G}_c^p|\pc\pred{G}_c^p,\pred{G}_c^n,Y)\right], \\
        \end{aligned}
    \end{equation}
    where the last equality is obtained via expanding the conditional entropy.
    Then, considering the contents in $\pred{G}_c^n$, without loss of generality,
    we can divide all of the possible cases into two:
    \begin{enumerate}[label=(\roman*)]
        \item $\pred{G}_c^n$ contains only the corresponding invariant subgraph $G_c^n$;
        \item $\pred{G}_c^n$ contains subgraph from the corresponding spurious subgraph $G_s^n$, denoted as $\pa\pred{G}_s^n\subseteq G_s^n$;
    \end{enumerate}
    For case (i), it is easy to write Eq.~\ref{CH:GALA:eq:pa_delta_appdx} as:
    \begin{equation}
        \label{CH:GALA:eq:pa_delta_case_i_appdx}
        \begin{aligned}
             & \triangle I(\pred{G}_c^p;\pred{G}_c^n|Y)                                                                       \\
             & = \left[H(\pa\pred{G}_s^p|\pc\pred{G}_c^p,Y)-H(\pa\pred{G}_c^p|\pc\pred{G}_c^p,Y)\right]-
            \left[H(\pa\pred{G}_s^p|\pc\pred{G}_c^p,\pred{G}_c^n,Y)-H(\pa\pred{G}_c^p|\pc\pred{G}_c^p,\pred{G}_c^n,Y)\right], \\
             & =-H(\pa\pred{G}_c^p|\pc\pred{G}_c^p,Y)+H(\pred{G}_c^p|\pc\pred{G}_c^p,\pred{G}_c^n,Y),                         \\
        \end{aligned}
    \end{equation}
    since $H(\pa\pred{G}_s^p|\pc\pred{G}_c^p,Y)=H(\pa\pred{G}_s^p|\pred{G}_c^n,\pc\pred{G}_c^p,Y)=H(\pa\pred{G}_s^p|Y)$ given $C\ind S|Y$ for PIIF shifts.
    Then, it suffices to know that $\triangle I(\pred{G}_c^p;\pred{G}_c^n|Y)\leq 0$
    as conditioning on new variables will not increase the entropy~\citep{network_coding}.

    For case (ii), we have :
    \begin{equation}
        \label{CH:GALA:eq:pa_delta_case_ii_appdx}
        \begin{aligned}
             & \triangle I(\pred{G}_c^p;\pred{G}_c^n|Y)                                                                       \\
             & =\left[H(\pa\pred{G}_s^p|\pc\pred{G}_c^p,Y)-H(\pa\pred{G}_c^p|\pc\pred{G}_c^p,Y)\right]-
            \left[H(\pa\pred{G}_s^p|\pc\pred{G}_c^p,\pred{G}_c^n,Y)-H(\pa\pred{G}_c^p|\pc\pred{G}_c^p,\pred{G}_c^n,Y)\right], \\
             & =\left[-H(\pa\pred{G}_c^p|\pc\pred{G}_c^p,Y)+H(\pa\pred{G}_c^p|\pc\pred{G}_c^p,\pred{G}_c^n,Y)\right]+
            \left[H(\pa\pred{G}_s^p|\pc\pred{G}_c^p,Y)-H(\pa\pred{G}_s^p|\pc\pred{G}_c^p,\pred{G}_c^n,Y)\right],              \\
        \end{aligned}
    \end{equation}
    where we claim that $H(\pa\pred{G}_s^p|\pc\pred{G}_c^p,Y)-H(\pa\pred{G}_s^p|\pc\pred{G}_c^p,\pred{G}_c^n,Y)=0$,
    and similarly conclude that $\triangle I(\pred{G}_c^p;\pred{G}_c^n|Y)\leq 0$.
    More specifically,
    we can rewrite the first term in Eq.~\ref{CH:GALA:eq:pa_delta_case_ii_appdx} as
    \begin{align*}
        H(\pa\pred{G}_s^p|\pc\pred{G}_c^p,Y)-H(\pa\pred{G}_s^p|\pc\pred{G}_c^p,\pred{G}_c^n,Y) & =
        H(\pa\pred{G}_s^p|Y)-H(\pa\pred{G}_s^p|\pa\pred{G}_s^n,Y)                                                                         \\
                                                                                               & =I(\pa\pred{G}_s^p;\pa\pred{G}_s^n|Y)=0,
    \end{align*}
    using the variation condition (i.e., assumption iii)) for $\pa\pred{G}_s^p$ under $\{G^p\}$, and $\pa\pred{G}_s^n$ under $\{G^n\}$.

    After showing the success of \gala in tackling $H(S|Y)<H(C|Y)$,
    it also suffices to know that the aforementioned discussion also generalizes to the other case, i.e., when
    $H(S|Y)>H(C|Y)$ in $\{G^p\}$ and $H(S|Y)<H(C|Y)$ in $\{G^n\}$.
\end{proof}
\clearpage
\section{More Discussions on Practical Implementations of \gala}
\label{CH:GALA:sec:gala_impl_appdx}
In this section, we provide more implementation discussions about \gala in complementary to Sec.~\ref{CH:GALA:sec:gala_sol}.

\paragraph{Objective implementation.}
As the estimation of mutual information could be highly expensive~\citep{infoNCE,mine}, inspired by~\citet{ciga},
we adopt the contrastive learning to approximate the mutual information between subgraphs in Eq.~\ref{CH:GALA:eq:gala_sol}~\citep{sup_contrastive,contrast_loss1,contrast_loss2,infoNCE,mine}:
\begin{equation} \label{CH:GALA:eq:gala_impl}
    \begin{aligned}
        I(\pred{G}_{c}^p;\pred{G}_c^n|Y) \approx
         & \mathbb{E}_{
        \substack{
        \{\pred{G}_{c}^p,\pred{G}_c^n\} \sim \sP_g(G|\gY=Y)         \\\
        \{G^i_c\}_{i=1}^{M} \sim \sP_g(G|\gY \neq Y)
        }
        }                                                           \\
         & \log\frac{e^{\phi(h_{\pred{G}_{c}^p},h_{\pred{G}_c^n})}}
        {e^{\phi(h_{\pred{G}_{c}^p},h_{\pred{G}_c^n})} +
            \sum_{i=1}^M e^{\phi(h_{\pred{G}_{c}},h_{G^i_c})}},
    \end{aligned}
\end{equation}
where $(\pred{G}_{c}^p,\pred{G}_c^n)$ are subgraphs extracted by $g$ from $\{G^p\},\{G^n\}$ that share the same label, respectively.
$\{G^i_c\}_{i=1}^{M}$ are subgraphs extracted by $g$ from $G$ that has a different label.
$\sP_g(G|\gY=Y)$ is the push-forward distribution of $\sP(G|\gY=Y)$ by featurizer $g$,
$\sP(G|\gY=Y)$ refers to the distribution of $G$ given the label $Y$,
$\sP(G|\gY\neq Y)$ refers to the distribution of $G$ given the label that is different from $Y$,
$\pred{G}_{c}=g(\pred{G}), \pred{G}_c=g(\pred{G}), G^{i}_c=g(G^{i})$ are the estimated subgraphs,
$h_{\pred{G}_{c}^p},h_{\pred{G}_c^n},h_{G^i_c}$ are the graph presentations of the extracted subgraphs.
$\phi$ is a similarity measure.
As $M\rightarrow \infty$, Eq.~\ref{CH:GALA:eq:gala_impl} approximates $I(\pred{G}_{c}^p;\pred{G}_c^n|Y)$~\citep{feat_dist_entropy,vMF_entropy,align_uniform}.

\paragraph{Environment assistant implementation.}
Theorem~\ref{CH:GALA:thm:gala_success} shows the effectiveness of \gala when given proper
subsets of $\{G^p\}$ and $\{G^n\}$.
In practice, we can implement the environment assistant into multiple forms.
As discussed in Sec.~\ref{CH:GALA:sec:gala_der}, ERM trained model can
serve as a reliable proxy. Since ERM tends to learn the first dominant features,
when $H(S|Y)<H(C|Y)$, ERM will firstly learn to extract spurious subgraphs $G_s$ to make predictions.
Therefore, we can obtain $\{G^p\}$ by finding samples where ERM correctly predicts the labels,
while $\{G^n\}$ for samples that ERM predicts an incorrect label.
In addition to direct label predictions,
we can also adopt clustering~\citep{cnc} to yield environment assistant predictions
for better contrastive sampling. We provide the detailed description of the clustering based variant of \gala in Algorithm~\ref{alg:gala_cl}.

\begin{algorithm}[ht]
    \caption{\textbf{\gala}: Clustering based \galafull }
    \label{alg:gala_cl}
    \begin{algorithmic}[1]
        \STATE \textbf{Input:} Training data $\train$;
        environment assistant $A$;
        featurizer $g$; classifier $f_c$;
        length of maximum training epochs $e$; batch size $b$;
        \STATE Initialize environment assistant $A$;
        \FOR{$p \in [1,\ldots, e]$}
        \STATE Sample a batch of data $\{G_i,Y_i\}_{i=1}^b$ from $\train$;
        \STATE Obtain Environment Assistant predictions $\{\hat{c}^e_i\}_{i=1}^b$
        using $k$-means clustering on the graph representations yielded by $A$;
        \FOR{each sample $G_i,y_i \in \{G_i,Y_i\}_{i=1}^b$}
        \STATE Find \emph{postive} graphs with same $y_i$ and different $\hat{c}^e_i$;
        \STATE Find \emph{negative} graphs with different $y_i$ but same environment assistant prediction $\hat{c}^e_i$;
        \STATE Calculate \gala risk via Eq.~\ref{CH:GALA:eq:gala_impl};
        \STATE Update $f_c, g$ via gradients from \gala risk;
        \ENDFOR
        \ENDFOR
        \STATE \textbf{return} final model $f_c\circ g$;
    \end{algorithmic}
\end{algorithm}

\begin{figure}[H]
    \centering
    \subfigure[Colored by environment labels.]{
        \includegraphics[width=0.31\textwidth]{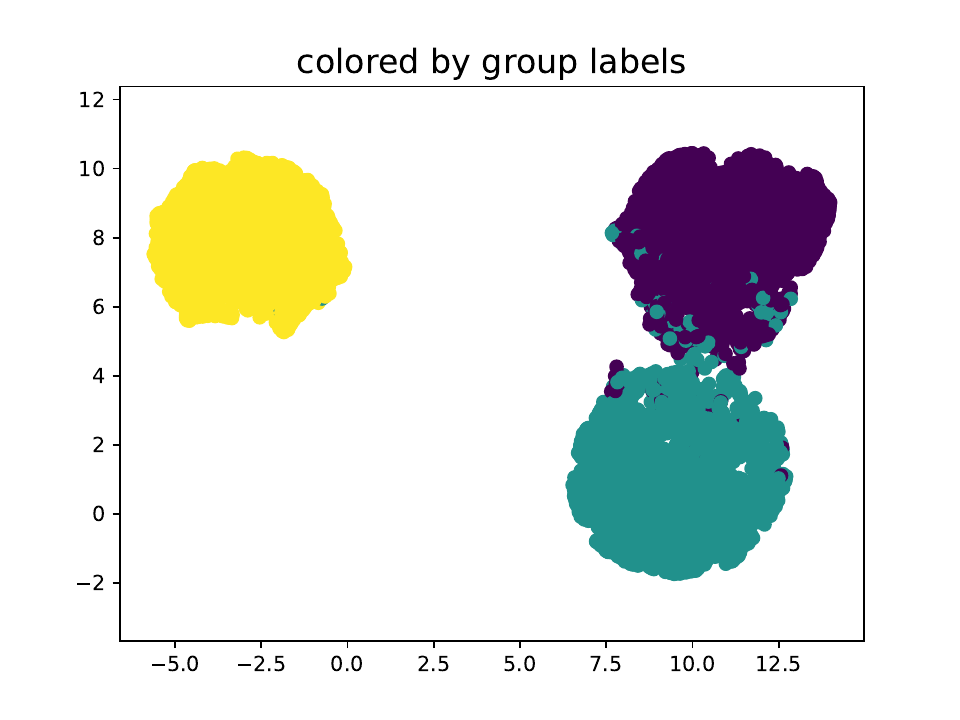}
    }
    \subfigure[Colored by label predictions.]{
        \includegraphics[width=0.31\textwidth]{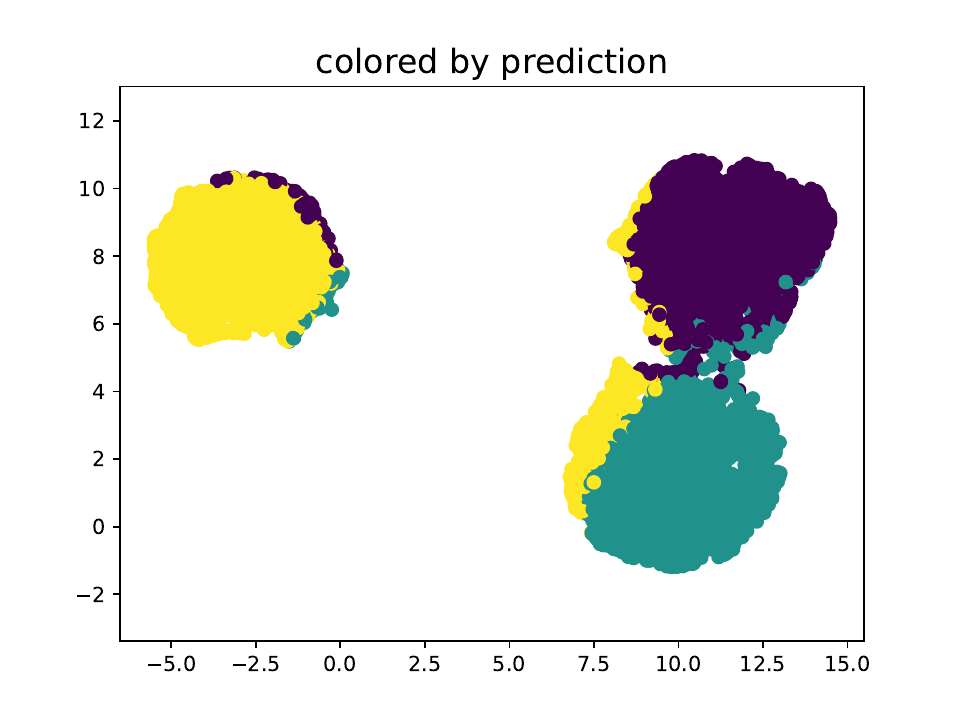}
    }
    \subfigure[Colored by cluster predictions.]{
        \includegraphics[width=0.31\textwidth]{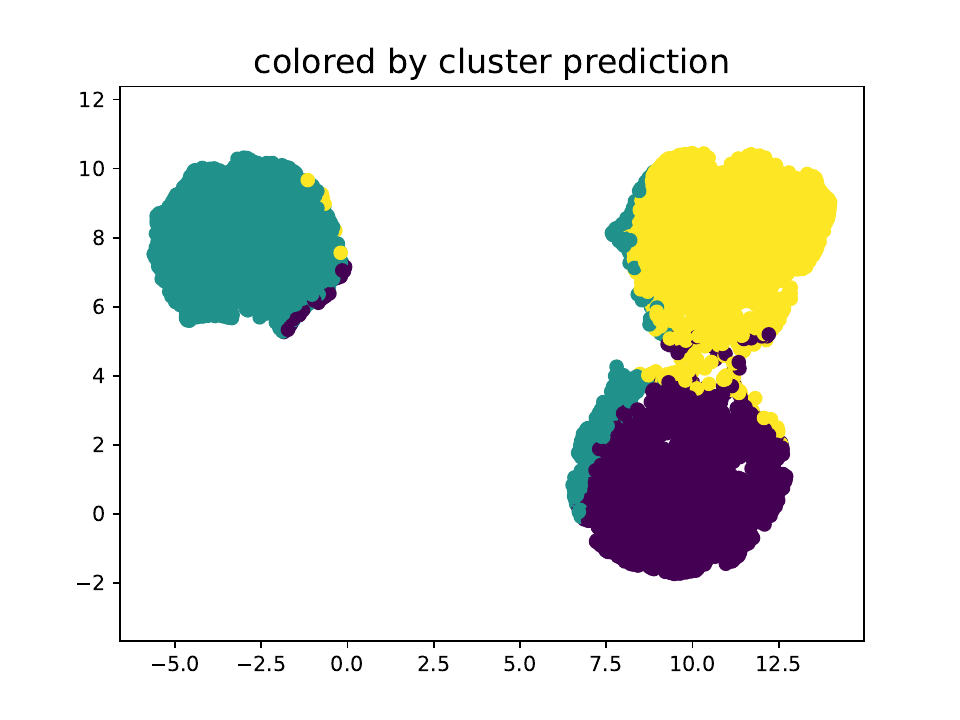}
    }
    \caption{
        Umap visualizations of learned graph representations in ERM trained environment assistant model
        based on the $3$-class two-piece graph $\{0.7,0.9\}$.}
    \label{CH:GALA:fig:ea_cluster}
\end{figure}

Empirically, we find clustering based variants can provide better performance
when the spurious correlations are well learned by the environment assistant model.
More concretely, we plot the umap visualizations~\citep{umap} of ERM trained environment assistant model
as in Fig.~\ref{CH:GALA:fig:ea_cluster},
where we can find that clustering predictions provide better approximations
to the underlying group labels.

Besides, we can also incorporate models
that are  easier to overfit to the first dominant features to better differentiate $\{G^p\}$ from  $\{G^n\}$.
To demonstrate the influence of different environment assistant implementations,
we conduct more studies with interpretable GNNs with an interpretable ratio of $30\%$ trained with ERM and also with a CIGAv1 penalty of $4$.

\begin{figure}[H]
    \centering
    \subfigure[Colored by environment labels.]{
        \includegraphics[width=0.31\textwidth]{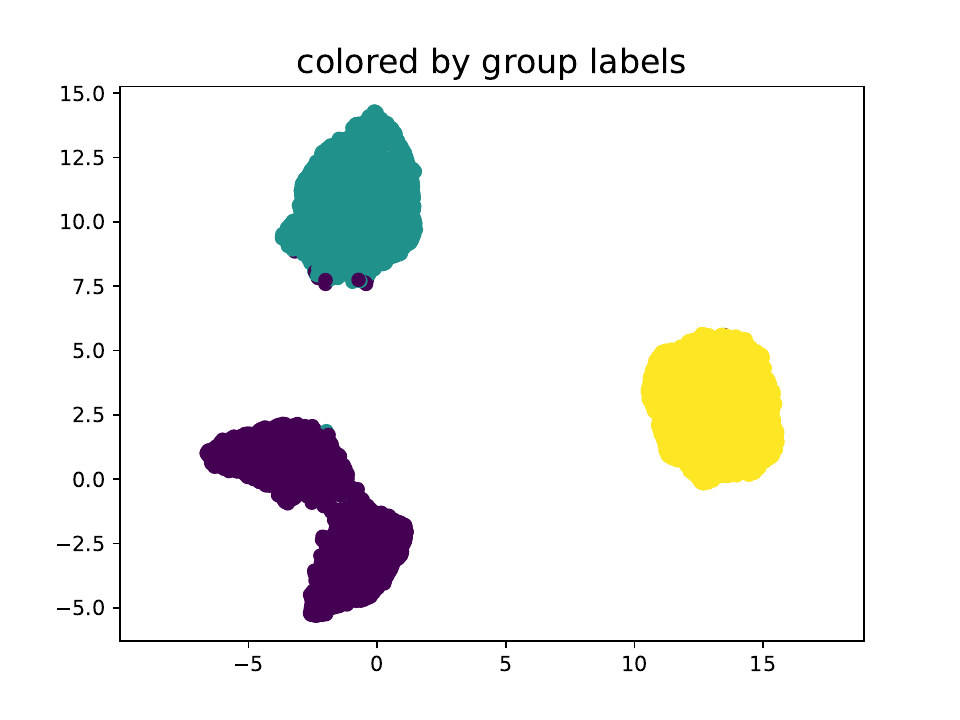}
    }
    \subfigure[Colored by label predictions.]{
        \includegraphics[width=0.31\textwidth]{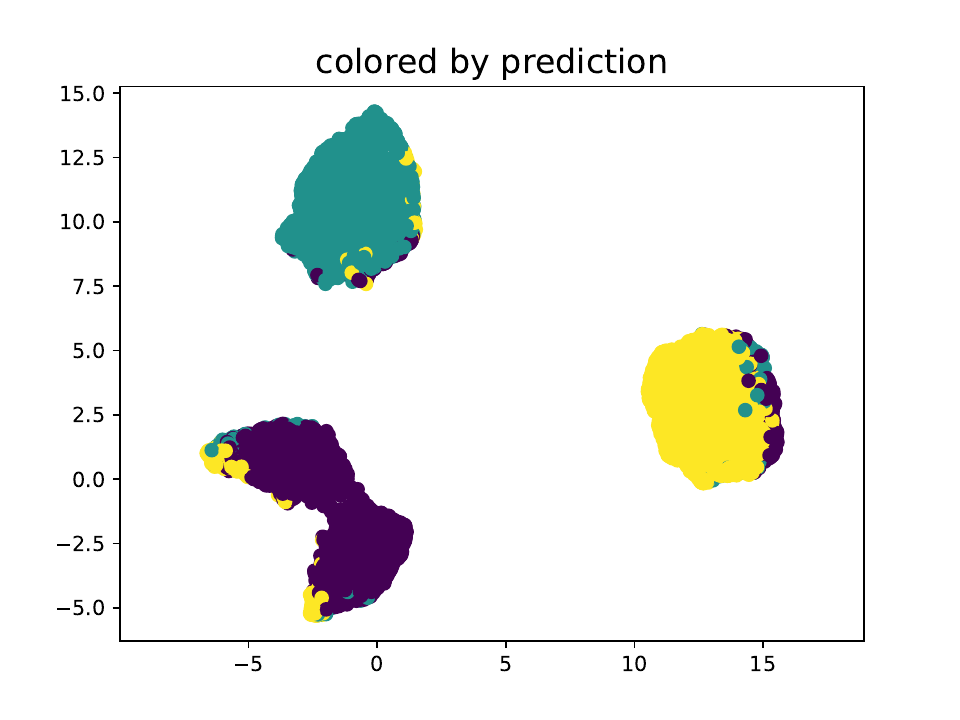}
    }
    \subfigure[Colored by cluster predictions.]{
        \includegraphics[width=0.31\textwidth]{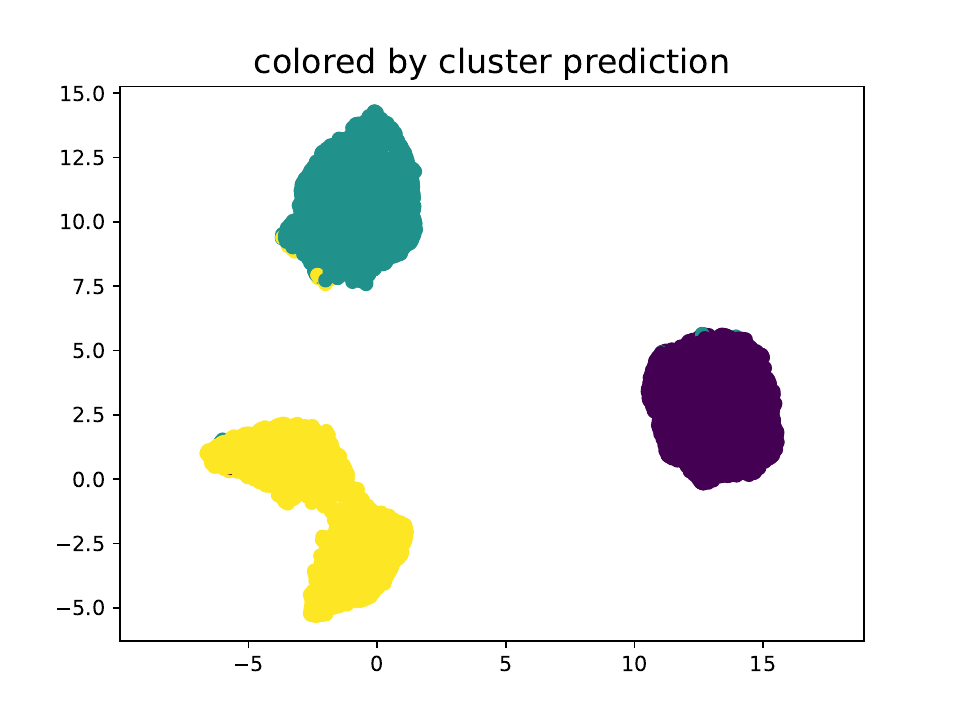}
    }
    \caption{
        Umap visualizations of learned graph representations
        in an interpretable GNN model (ratio=$30\%$) trained with ERM
        based on the $3$-class two-piece graph $\{0.7,0.9\}$.}
    \label{CH:GALA:fig:ea_cluster_xgnn}
    \vskip -0.15in
\end{figure}

\begin{figure}[H]
    \centering
    \subfigure[Colored by environment labels.]{
        \includegraphics[width=0.31\textwidth]{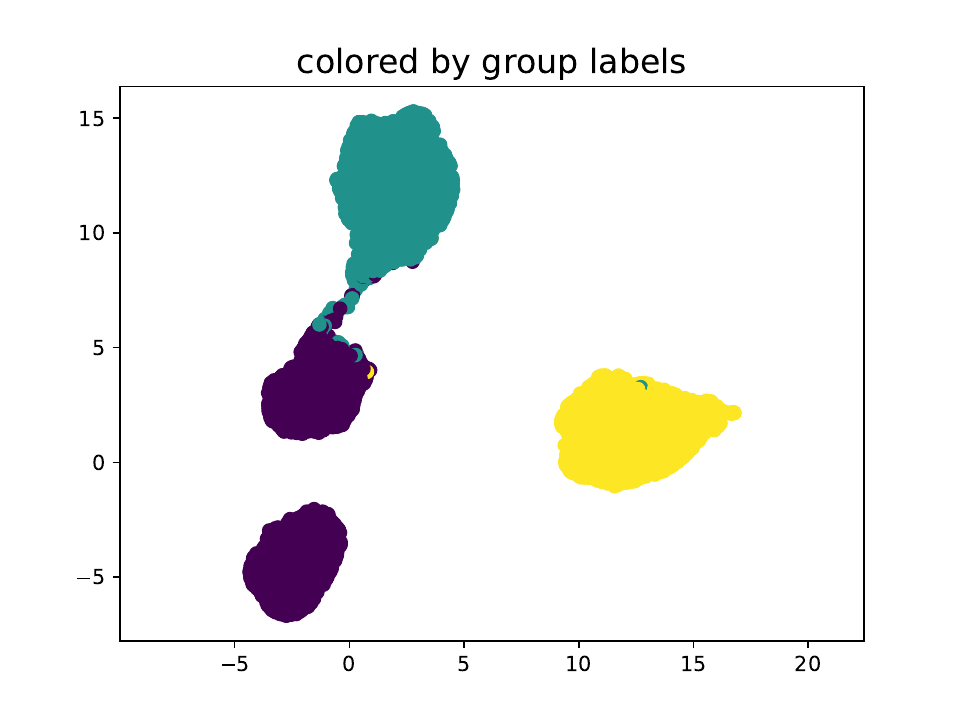}
    }
    \subfigure[Colored by label predictions.]{
        \includegraphics[width=0.31\textwidth]{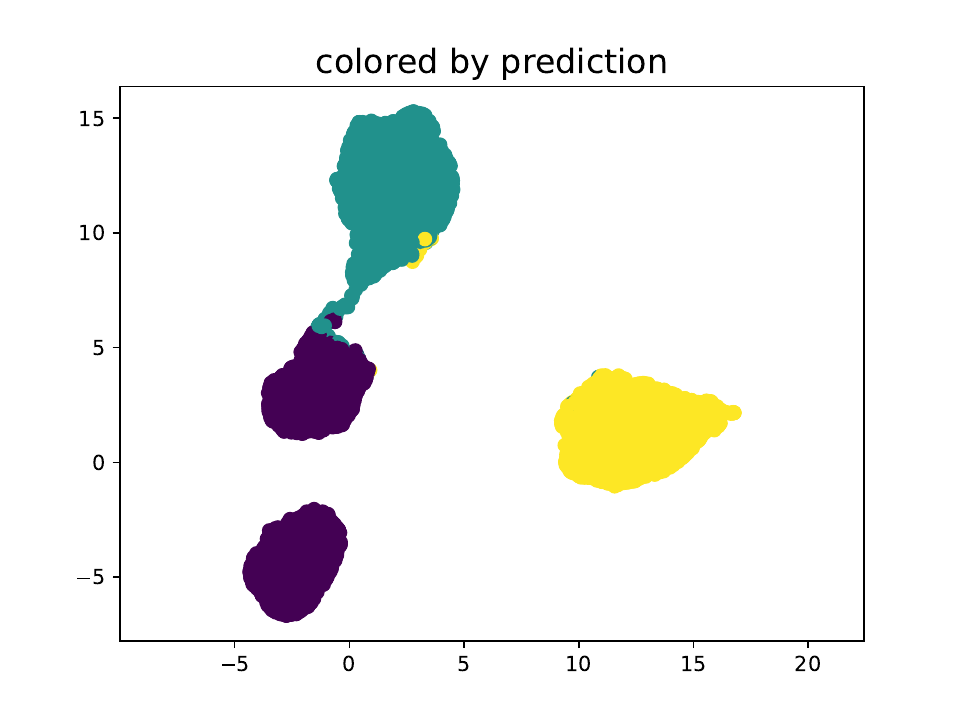}
    }
    \subfigure[Colored by cluster predictions.]{
        \includegraphics[width=0.31\textwidth]{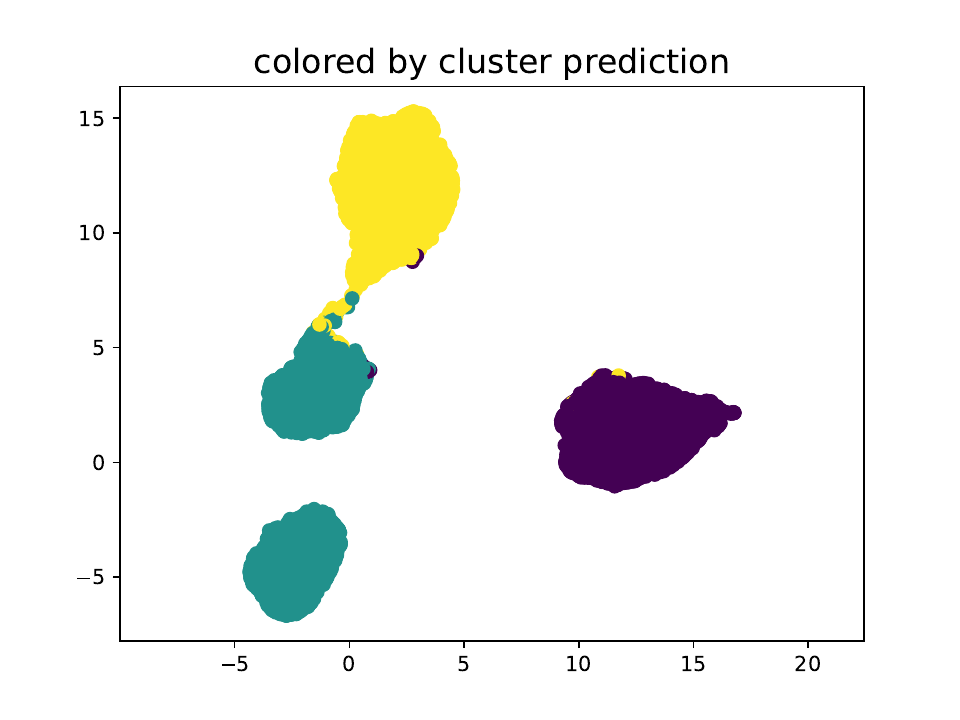}
    }
    \caption{
        Umap visualizations of learned graph representations
        in an interpretable GNN model (ratio=$30\%$) trained with ERM
        based on the $3$-class two-piece graph $\{0.7,0.9\}$.}
    \label{CH:GALA:fig:ea_cluster_xgnn_c4}
    \vskip -0.15in
\end{figure}

\begin{figure}[H]
    \centering
    \subfigure[Colored by environment labels.]{
        \includegraphics[width=0.31\textwidth]{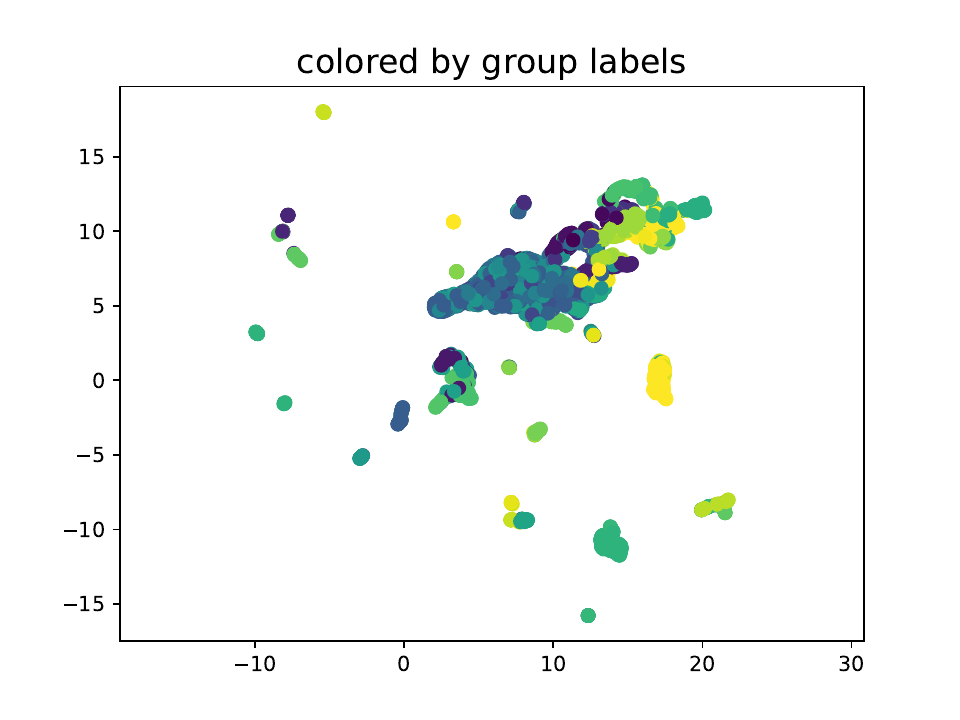}
    }
    \subfigure[Colored by label predictions.]{
        \includegraphics[width=0.31\textwidth]{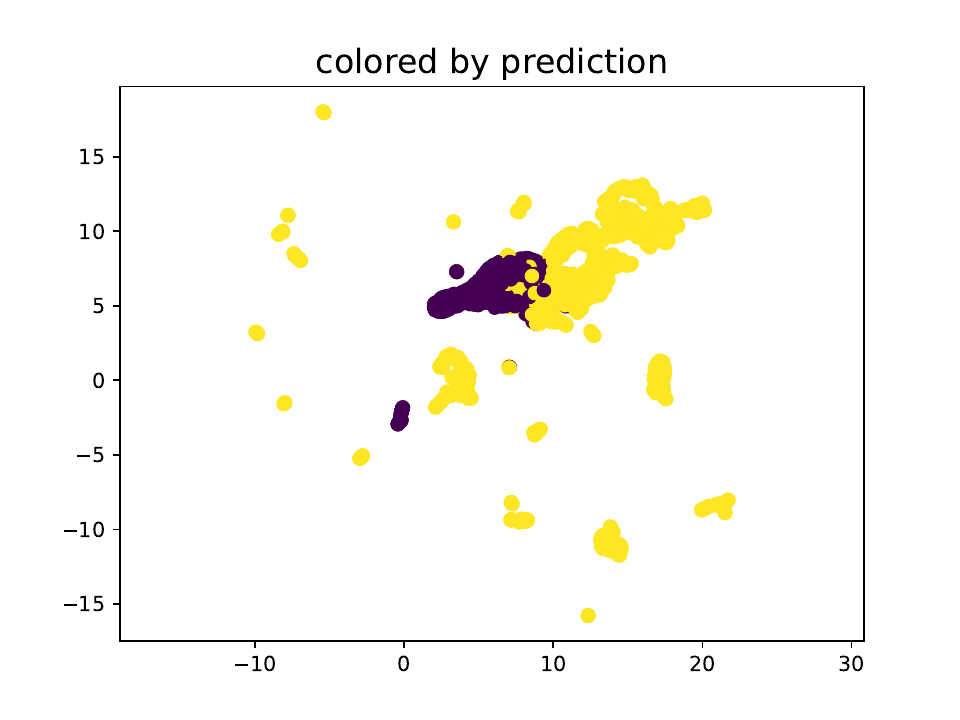}
    }
    \subfigure[Colored by cluster predictions.]{
        \includegraphics[width=0.31\textwidth]{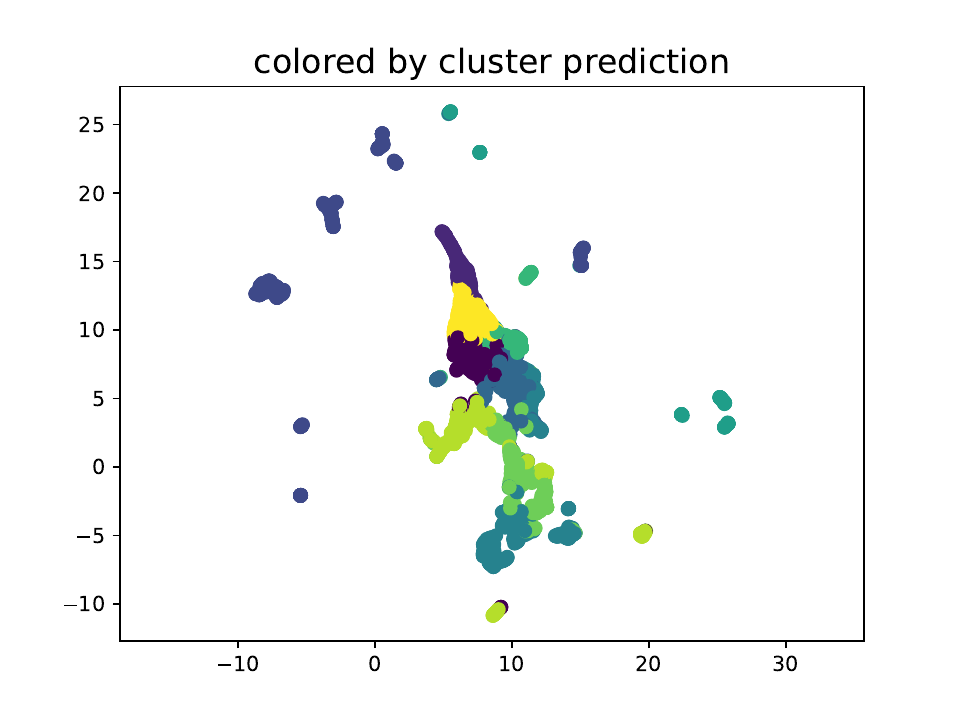}
    }
    \caption{
        Umap visualizations of learned graph representations of
        a interpretable GNN trained by ERM on EC50-Assay.}
    \label{CH:GALA:fig:ea_cluster_assay}
    \vskip -0.15in
\end{figure}

\begin{figure}[H]
    \centering
    \subfigure[Colored by environment labels.]{
        \includegraphics[width=0.31\textwidth]{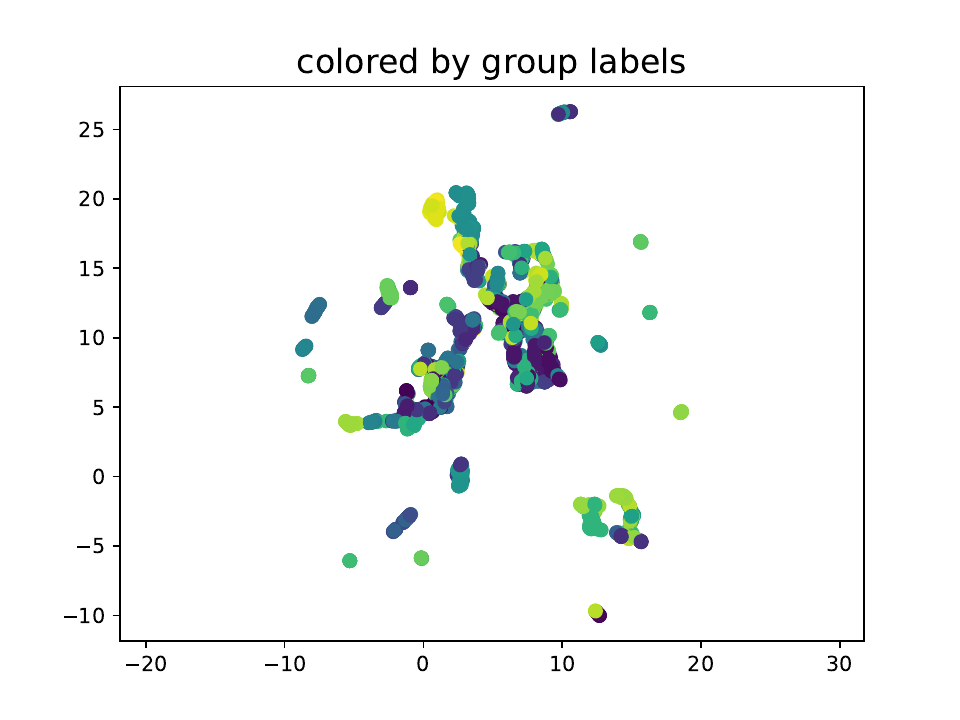}
    }
    \subfigure[Colored by label predictions.]{
        \includegraphics[width=0.31\textwidth]{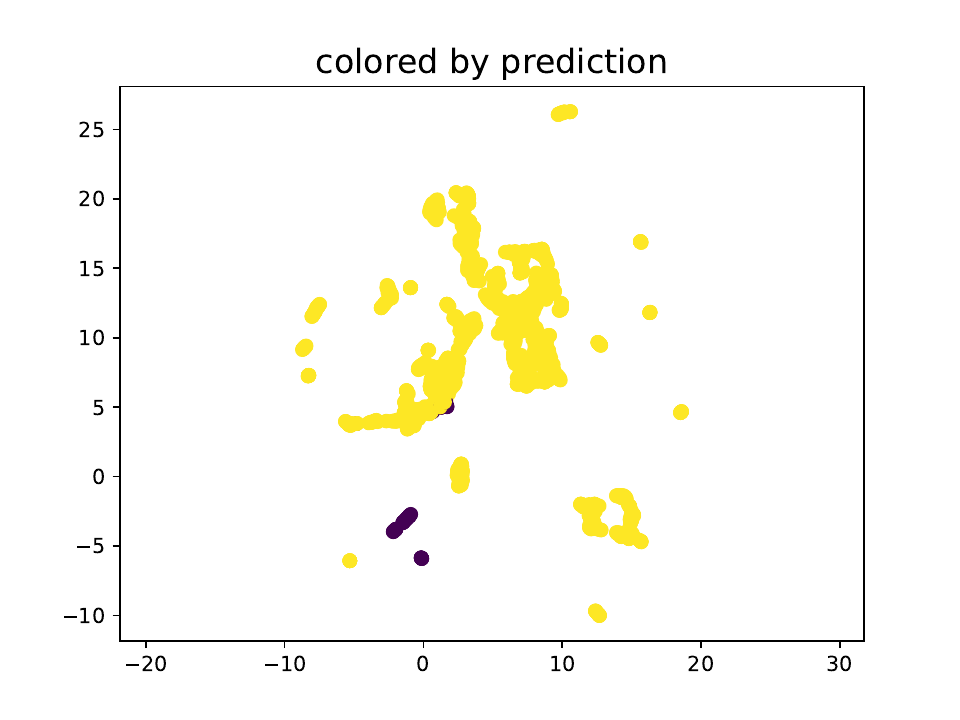}
    }
    \subfigure[Colored by cluster predictions.]{
        \includegraphics[width=0.31\textwidth]{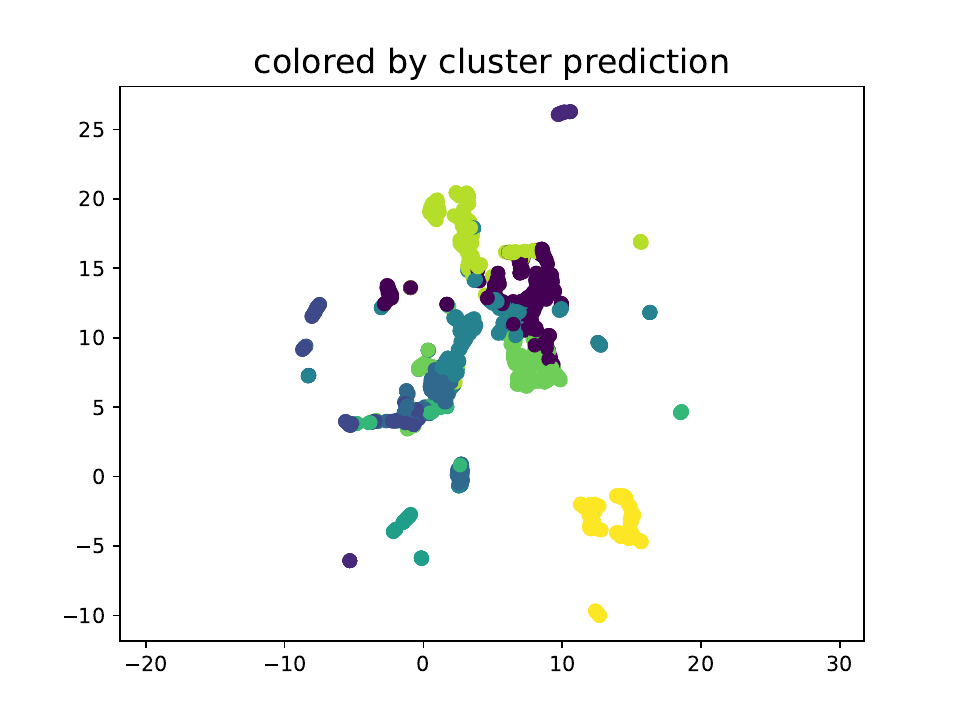}
    }
    \caption{
        Umap visualizations of learned graph representations of
        an interpretable GNN trained by ERM on EC50-Scaffold.}
    \label{CH:GALA:fig:ea_cluster_sca}
    \vskip -0.15in
\end{figure}

\begin{figure}[H]
    \centering
    \subfigure[Colored by environment labels.]{
        \includegraphics[width=0.31\textwidth]{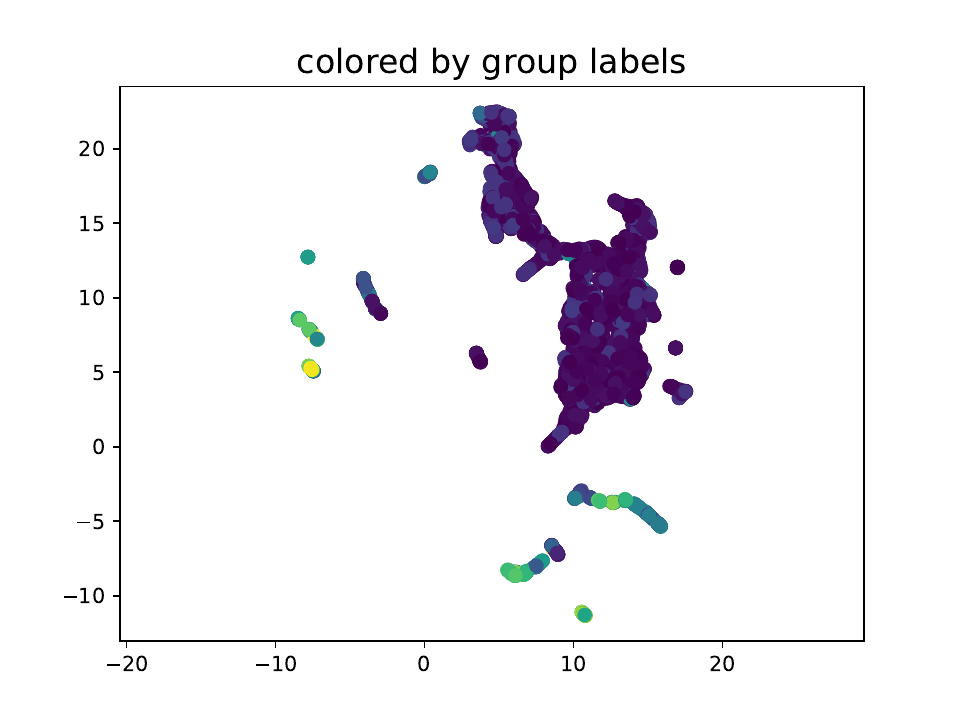}
    }
    \subfigure[Colored by label predictions.]{
        \includegraphics[width=0.31\textwidth]{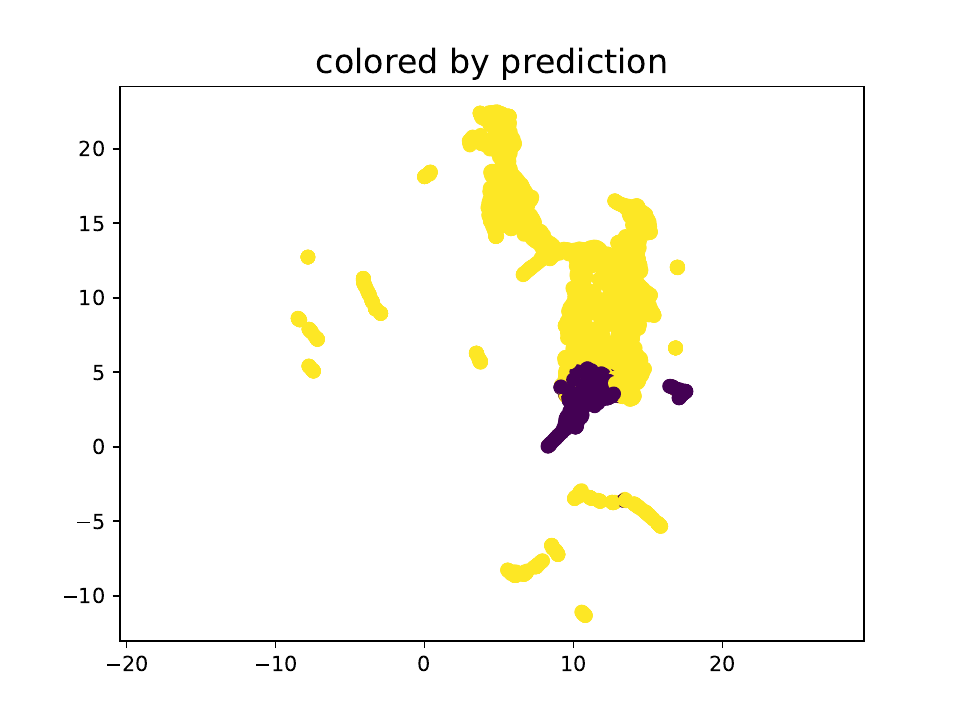}
    }
    \subfigure[Colored by cluster predictions.]{
        \includegraphics[width=0.31\textwidth]{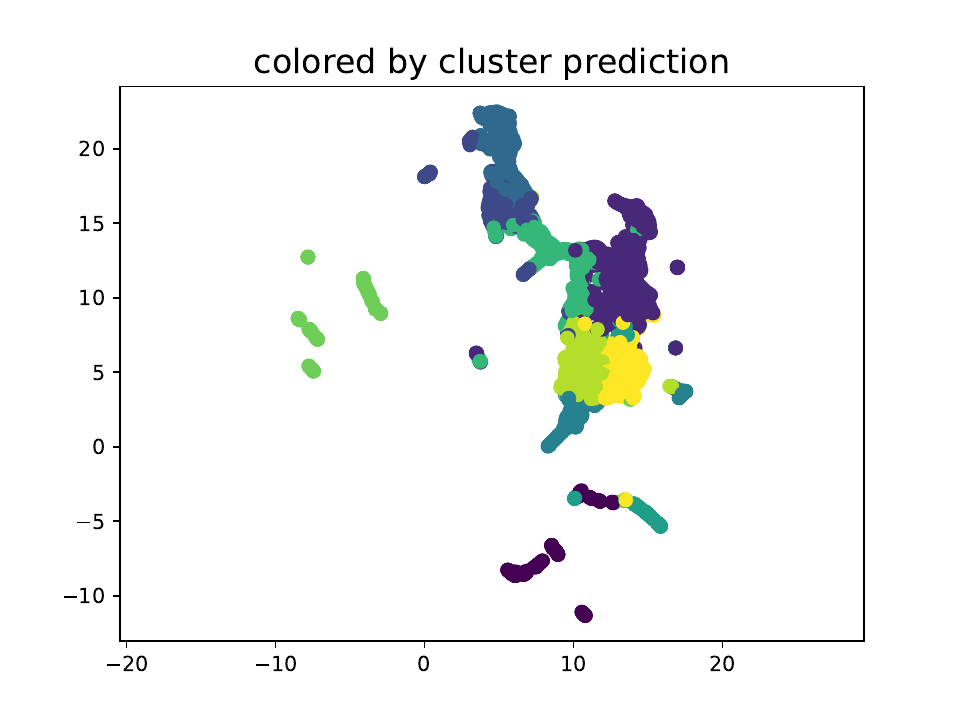}
    }
    \caption{
        Umap visualizations of learned graph representations of
        a interpretable GNN trained by ERM on EC50-Size.}
    \label{CH:GALA:fig:ea_cluster_size}
\end{figure}

In Fig.~\ref{CH:GALA:fig:ea_cluster_xgnn} and Fig.~\ref{CH:GALA:fig:ea_cluster_xgnn_c4},
it can be found that the interpretable GNN learns hidden representations
that are better clustered with group labels.
The clustering based predictions yields a better approximation of the
underlying environment labels.
Furthermore, when implementing the environment assistant model
using a interpretable GNN as well as a CIGAv1 penalty,
which facilitates the overfitting to the spurious correlations,
then the vanilla label predictions can also yield a good approximation of the
underlying environment labels.

Although using the clustering predictions seem to be promising,
we also find negative cases. For example, in DrugOOD datasets,
the number of curated environment labels are much larger that
learning a well clusterd hidden representations for the environment labels
appears to be difficult. Shown as in Fig.~\ref{CH:GALA:fig:ea_cluster_assay} to Fig.~\ref{CH:GALA:fig:ea_cluster_size},
the learned representations have poor quality for approximating the underlying
environment labels.
Empirically, we also find that direct using label predictions in DrugOOD datasets
generically yield better performance.

\paragraph{One-side contrastive sampling.}
The original supervised contrastive implementation~\citep{sup_contrastive}
takes positive and negative samples within the batch
using two-side contrastive sampling. That is,
all the samples will be considered as anchor points.
However, when it is used to contrast samples from $\widehat{G}_{c}^p$ and $\widetilde{G}_c^n$,
there could be undesired behaviors.
First, it can often happen that there are few to no negative cases when the spurious correlations are too strong.
The samples from $\{G^p\}$ in a batch may pull the representations of
samples from $\{G^n\}$ to even closer, which makes the model further overfitted to the spurious correlations.
Second, the sampling over $\widehat{G}_{c}^p$ and $\widetilde{G}_c^n$, can be
seen as hard positive and negative samples, that may impose a too strong regularizations that preventing
the learning of any correlations.
Therefore, we propose to use one-side sampling. That is, only using the
incorrectly predicted samples as anchor points.
We empirically observe one-side sampling could yield better performance in two-piece graphs.

\paragraph{Upsampling of minority group samples.}
It is possible that the number of positive and negative graphs is imbalanced, especially when adopting the label predictions to sample positive and negative graphs. For example, when the ERM trained assistant model overfits the training distribution under the spuriousness-dominated case, the number of negative graphs will be extremely small. Given an extremely small number of negative samples for contrastive learning, the resulting mutual information estimation will be collapsed to trivial solutions. Therefore, we propose a simple strategy to mitigate the issue. We directly upsample the minority group samples. The minority group of samples will be repeated $k$ times within the training set.

\section{More Details about the Experiments}
\label{CH:GALA:sec:exp_appdx}
In this section, we provide more details about the experiments, including the dataset preparation, baseline implementations, models and hyperparameters selection as well as the evaluation protocols.

\subsection{Datasets}
\label{CH:GALA:sec:dataset_appdx}

We provide more details about the motivation and construction method of the datasets that are used in our experiments. Statistics of the datasets are presented in Table~\ref{tab:datasets_stats_appdx}.

\bgroup
\def\arraystretch{1.2}
\begin{table}[H]
    \centering
    \caption[Information about the datasets used in experiments of \gala.]{Information about the datasets used in experiments of \gala. The number of nodes and edges are respectively taking average among all graphs.}%
    \label{tab:datasets_stats_appdx}
    \resizebox{\textwidth}{!}{
        \begin{small}
            \begin{tabular}{l|ccccccc}
                \toprule
                \textbf{Datasets}              & \textbf{\# Training} & \textbf{\# Validation} & \textbf{\# Testing} & \textbf{\# Classes} & \textbf{ \# Nodes} & \textbf{ \# Edges}
                                               & \textbf{  Metrics}                                                                                                                            \\\midrule
                Two-piece graphs $\{0.8,0.6\}$ & $9,000$              & $3,000$                & $3,000$             & $3$                 & $26.14$            & $36.21$            & ACC     \\
                Two-piece graphs $\{0.8,0.7\}$ & $9,000$              & $3,000$                & $3,000$             & $3$                 & $26.18$            & $36.27$            & ACC     \\
                Two-piece graphs $\{0.8,0.9\}$ & $9,000$              & $3,000$                & $3,000$             & $3$                 & $26.13$            & $36.22$            & ACC     \\
                Two-piece graphs $\{0.7,0.9\}$ & $9,000$              & $3,000$                & $3,000$             & $3$                 & $26.13$            & $36.22$            & ACC     \\\midrule
                CMNIST-sp                      & $40,000$             & $5,000$                & $15,000$            & $2$                 & $56.90$            & $373.85$           & ACC     \\
                Graph-SST2                     & $24,881 $            & $7,004 $               & $12,893$            & $2$                 & $10.20$            & $18.40$            & ACC     \\\midrule
                EC50-Assay                     & $4,978$              & $ 2,761 $              & $2,725$             & $2$                 & $40.89$            & $87.18$            & ROC-AUC \\
                EC50-Scaffold                  & $2,743$              & $ 2,723 $              & $2,762$             & $2$                 & $35.54$            & $75.56$            & ROC-AUC \\
                EC50-Size                      & $5,189$              & $2,495$                & $2,505$             & $2$                 & $35.12$            & $75.30$            & ROC-AUC \\
                Ki-Assay                       & $8,490$              & $ 4,741 $              & $4,720$             & $2$                 & $32.66$            & $71.38$            & ROC-AUC \\
                Ki-Scaffold                    & $5,389$              & $ 4,805 $              & $4,463$             & $2$                 & $29.96$            & $65.11$            & ROC-AUC \\
                Ki-Size                        & $8,605$              & $4,486$                & $4,558$             & $2$                 & $30.35$            & $66.49$            & ROC-AUC \\
                \bottomrule
            \end{tabular}	\end{small}}
\end{table}
\egroup
\textbf{Two-piece graph datasets.} We construct 3-class synthetic datasets based on BAMotif~\citep{pge} following Def.~\ref{def:twobit_graph_appdx},
where the model needs to tell which one of three motifs (House, Cycle, Crane) the graph contains.
For each dataset, we generate $3000$ graphs for each class at the training set, $1000$ graphs for each class at the validation set and testing set, respectively.
Each dataset is defined with two variables $\{a,b\}$ referring to the strength of invariant and spurious correlations.
Given $\{a,b\}$, we generate the training data following the percise generation process as Def.~\ref{def:twobit_graph_appdx}. While for the generation of validation sets, we use a $b_v=\max(1/3,b-0.2)$ that facilitates the model selection for OOD generalization~\citep{domainbed,pair}. While for the generation of test datasets, we merely use a $b=0.33$ that contains no distribution shifts, to fully examine to what extent the model learns the invariant correlations.
During the construction, we merely inject the distribution shifts in the training data while keeping the testing data and validation data without the biases.

\textbf{CMNIST-sp.} To study the effects of PIIF shifts, we select the ColoredMNIST dataset created in IRM~\citep{irmv1}. We convert the ColoredMnist into graphs using the superpixel algorithm introduced by~\citet{understand_att}. Specifically, the original Mnist dataset is assigned to binary labels where images with digits $0-4$ are assigned to $y=0$ and those with digits $5-9$ are assigned to $y=1$. Then, $y$ will be flipped with a probability of $0.25$. Thirdly, green and red colors will be respectively assigned to images with labels $0$ and $1$ an averaged probability of $0.15$ (since we do not have environment splits) for the training data. While for the validation and testing data, the probability is flipped to $0.9$.

\textbf{Graph-SST2.} Inspired by the data splits generation for studying distribution shifts on graph sizes, we split the data curated from sentiment graph data~\citep{xgnn_tax}, that converts sentiment sentence classification datasets \textbf{Graph-SST2}~\citep{sst25} into graphs, where node features are generated using BERT~\citep{bert} and the edges are parsed by a Biaffine parser~\citep{biaffine}. Our splits are created according to the averaged degrees of each graph. Specifically, we assign the graphs as follows: Those that have smaller or equal to $50$-th percentile averaged degree are assigned to training, those that have averaged degree large than $50$-th percentile while smaller than $80$-th percentile are assigned to the validation set, and the left are assigned to test set.

\textbf{DrugOOD datasets.} To evaluate the OOD performance in realistic scenarios with realistic distribution shifts, we also include three datasets from DrugOOD benchmark~\citep{drugood}.
DrugOOD is a systematic OOD benchmark for AI-aided drug discovery, focusing on the task of drug target binding affinity prediction for both macromolecule (protein target) and small-molecule (drug compound).
The molecule data and the notations are curated from realistic ChEMBL database~\citep{chembl}.
Complicated distribution shifts can happen on different assays, scaffolds, and molecule sizes.
In particular, we select \texttt{lbap-core-ec50-assay}, \texttt{lbap-core-ec50-scaffold}, \texttt{lbap-core-ec50-size}, \texttt{lbap-core-ki-assay}, \texttt{lbap-core-ki-scaffold}, and \texttt{lbap-core-ki-size},
from the task of Ligand Based Affinity Prediction which uses \texttt{ic50} measurement type and contains \texttt{core} level annotation noises.
We directly use the data files provided by the authors.\footnote{https://drugood.github.io/}
For more details, we refer interested readers to~\citet{drugood}.

\subsection{Baselines and Evaluation Setup}
\label{CH:GALA:sec:eval_appdx}

During the experiments, we do not tune the hyperparameters exhaustively while following the common recipes for optimizing GNNs.
Details are as follows.

\textbf{GNN encoder.} For a fair comparison, we use the same GNN architecture as graph encoders for all methods.
By default, we use $3$-layer GIN~\citep{gin} with Batch Normalization~\citep{batch_norm} between layers and JK residual connections at the last layer~\citep{jknet}.
The hidden dimension is set to $32$ for Two-piece graphs, CMNIST-sp, and $128$ for SST2, and DrugOOD datasets.
The pooling is by default a mean function over all nodes. The only exception is DrugOOD datasets, where we follow the backbone used in the paper~\citep{drugood}, i.e., $4$-layer GIN with sum readout.

\textbf{Interpretable GNN backbone.} As mentioned in Sec.~\ref{CH:GALA:sec:prelim} that most of the existing invariant graph learning approaches adopt the interpretable GNN as the basic backbone model for the whole predictor $f=f_c\circ g$, where
$g:\gG\rightarrow\gG_c$ is a featurizer GNN and  $f_c:\gG_c\rightarrow\gY$ is a classifier GNN.
$g$ first calculates the sampling weights as in $\widehat{G}_c$ for each edge. More formally, given a graph $G$ containing $n$ nodes, a soft mask is predicted through the following equation:
\begin{equation}\label{CH:GALA:eq:gae_appdx}\nonumber
    Z=\text{GNN}(G)\in\R^{n\times h},\ M=\text{a}(Z,A)\in\R^{n\times n},
\end{equation}
where $a$ calculates the sampling weights for each edge using a MLP: $M_{ij}=\text{MLP}([Z_i,Z_j])$.
Based on the continuous sampling score $M$, $g$ could sample discrete edges according to the predicted scores~\citep{gsat}.
For two-piece graph datasets and DrugOOD datasets, we will directly use the score to reweight the messaging passing process along the edge, as we empirically find it yields more stable performance. While for CMNIST-sp and Graph-SST2,  we will sample a ratio $r\%$ of all edges for each graph. The ratios adopted are $80\%$ and $60\%$, respectively, following previous works~\citep{ciga,drugood}.
Meanwhile, to improve the stability of the subgraph extractor, we adopt a layernorm~\citep{instancenorm} following the practice of~\citep{gsat}.

Besides, we also have various implementation options for obtaining the features in $\widehat{G}_c$, for further obtaining $h_{\widehat{G}_c}$, as well as for obtaining predictions based on $\widehat{G}_s$.
Following previous works~\citep{gsat}, we will adopt the same GNN encoder for the two GNNs in the interpretable GNN backbone, and feed the raw graph inputs to the classifier GNN. The contrastive loss is obtained via the graph representations of the sampled subgraph by the classifier GNN.
For classifying $G$ based on $\widehat{G}_s$, we use a separate MLP downstream classifier in the classifier GNN $f_c$.

\textbf{Optimization and model selection.}
By default, we use Adam optimizer~\citep{adam} with a learning rate of $1e-3$ and a batch size of $128$ for all models at all datasets.
Except for CMNIST-sp, we use a batch size of $256$ to facilitate the evaluation following previous works~\citep{gsat}.
To avoid underfitting, we pre-train models for $20$ epochs for all datasets by default. While in two-piece graphs, we find pre-training by $100$ epochs yields more stable performance.
To avoid overfitting, we also employ an early stopping of $5$ epochs according to the validation performance.
Meanwhile, dropout is also adopted for some datasets.
Specifically, we use a dropout rate of $0.5$ for all of the realistic graph datasets, following previous works~\citep{ciga,drugood}.

The final model is selected according to the performance at the validation set. All experiments are repeated with $5$ different random seeds of $\{1,2,3,4,5\}$. The mean and standard deviation are reported from the $5$ runs.

\textbf{Implementations of Euclidean OOD methods.}
When implementing IRM~\citep{irmv1}, vrex~\citep{vrex} and IB-IRM~\citep{ib-irm}, we refer the implementations from DomainBed~\citep{domainbed}.
Since the environment information is not available, we perform random partitions on the training data to obtain two equally large environments for these objectives following previous works~\citep{eiil,ciga}.
Moreover, we select the weights for the corresponding regularization from $\{0.01,0.1,1,10,100\}$ for these objectives according to the validation performances of IRM and stick to it for others,
since we empirically observe that they perform similarly with respect to the regularization weight choice.
For EIIL~\citep{env_inference}, we use the author-released implementations about assigning different samples the weights for being put in each environment and calculating the IRM loss.

\textbf{Implementations of invariant graph learning methods.}
We implement GSAT~\citep{gsat}, GREA~\citep{grea}, CAL~\citep{cal}, MoleOOD~\citep{moleood}, GIL~\citep{gil}, DisC~\citep{disc}, and CIGA~\citep{ciga}, according to the author provided codes (if available).
\begin{itemize}[leftmargin=*]
    \item GREA~\citep{grea}: We use a penalty weight of $1$ for GREA as we empirically it does not affect the performance by changing to different weights.
          \begin{itemize}[leftmargin=*]
              \item Interpretable ratio: same as others;
              \item Penalty weight: $1$;
              \item Number of environments: N/A;
          \end{itemize}
    \item GSAT~\citep{gsat}: We follow the recommendations of the released implementations by the authors.
          \begin{itemize}[leftmargin=*]
              \item Interpretable ratio: $70\%$;
              \item Penalty weight: $1$;
              \item Decay ratio: $10\%$;
              \item Decay interval: \texttt{pretrain epoch}$//2$;
              \item Number of environments: N/A;
          \end{itemize}
    \item CAL~\citep{cal}: We follow the recommendations of the released implementations by the authors.
          \begin{itemize}[leftmargin=*]
              \item Interpretable ratio: same as others;
              \item Penalty weight: $\{0.1,0.5,1.0\}$;;
              \item Number of environments: N/A;
          \end{itemize}
    \item MoleOOD~\citep{moleood}: We tune the penalty weights of MoleOOD with values from $\{1e-2,1e-1,1,10\}$ but did not observe much performance differences. Hence we stick the penalty weight as $1$ for all datasets.
          \begin{itemize}[leftmargin=*]
              \item Interpretable ratio: N/A;
              \item Penalty weight: $1$;
              \item Number of environments: same as others;
          \end{itemize}
    \item GIL~\citep{gil}: We follow the recommendations of the paper.
          \begin{itemize}[leftmargin=*]
              \item Interpretable ratio: same as others;
              \item Penalty weight: $\{1e-5,1e-3,1e-1\}$;
              \item Number of environments: same as others;
          \end{itemize}
    \item DisC~\citep{disc}: We tune only the $q$ weight from $\{0.9,0.7,0.5\}$ in the GCE loss as we did not observe performance differences by changing the weight of the other terms.
          \begin{itemize}[leftmargin=*]
              \item Interpretable ratio: same as others;
              \item $q$ weight: $\{0.9,0.7,0.5\}$;
              \item Number of environments: same as others;
          \end{itemize}
    \item CIGA~\citep{ciga}: We follow the recommendations of the released implementations by the authors..
          \begin{itemize}[leftmargin=*]
              \item Interpretable ratio: same as others;
              \item Penalty weight: $\{0.5,1,2,4,8,16,32\}$;
              \item Number of environments: N/A;
          \end{itemize}
\end{itemize}
\begin{itemize}[leftmargin=*]
    \item \gala:
          \begin{itemize}[leftmargin=*]
              \item Interpretable ratio: same as others;
              \item Penalty weight: $\{0.5,1,2,4,8,16,32\}$;
              \item Environment assistant: $\{\texttt{vanilla GNN}, \texttt{XGNN}\}$;
              \item Sampling proxy: $\{\texttt{label predictions}, \texttt{cluster predictions}\}$;
              \item Number of environments: same as others;
          \end{itemize}
\end{itemize}

All of the graph learning methods adopt an interpretable GNN as the backbone by default. The only exception is MoleOOD, we follow the original implementation while using a shared GNN encoder for the variational losses to ensure the fairness of comparison. Besides, for DisC, we find the soft masking implementation in two-piece graphs will incur a severe performance degeneration hence we use a ratio of $25\%$ for the interpretable GNN backbone.

For environment inferring methods, we search the number of environments
\begin{itemize}[leftmargin=*]
    \item Two-piece graphs: fixed as $3$ (since there are $3$ spurious graphs);
    \item CMNIST-sp: $2$ (since there are $2$ environments);
    \item Graph-SST2: $\{2,3,4\}$ following previous practice~\citep{gil};
    \item DrugOOD datasets: $\{2,3,5,10,20\}$ following previous practice~\citep{moleood}.;
\end{itemize}

\textbf{Implementations of \gala.}
For a fair comparison, \gala uses the same GNN architecture for GNN encoders as the baseline methods.
By default, we fix the temperature to be $1$ in the contrastive loss,
and merely search the penalty weight of the contrastive loss from $\{0.5,1,2,4,8,16,32\}$ according to the validation performances, following the \ciga implementations~\citep{ciga}.
By default, we implement the environment assistant as a ERM model, and adopt directly the environment assistant predictions to sample possible and negative graph pairs.
Nevertheless, as discussed in Sec.~\ref{CH:GALA:sec:gala_sol} that there could be multiple implementation choices for the environment assistant and the use of its predictions. We hence also try with XGNN based environment assistant model and clustering based proxy predictions.
By default, the selection of the environment assistant model is performed via best training performance, as which encourages a better fit to the dominant subgraph patterns, while we also try the model selection with best validation performance in DrugOOD datasets and find it empirically sometimes leads to better performance.
All the options for the selection of the environment assistant models depend on the validation performance.
For Two piece graphs, EC50-Scaffold, EC50-Size, Ki-Assay, Ki-Scaffold, CMNIST-sp and Graph-SST2, we find implementing the environment assistant as a ERM model already yield impressive improvements.
While for the other DrugOOD datasets, we implement the environment assistant as an interpretable GNN trained with ERM and cluster the learned graph representations of the model to sample positive and negative pairs.

Since \gala imposes a strong regularization to the data that may hinder the learning of graph representations, we pre-train the model by $10$ epochs using ERM and then impose the \gala penalty implemented as one-side contrastive loss as discussed in Sec.~\ref{CH:GALA:sec:gala_impl_appdx}.
When the numbers of positive and negative pairs are extremely imbalanced, we will upsample the minor groups by a factor of $\{2,3,4\}$, depending on the validation performance.

\subsection{Software and Hardware}
\label{CH:GALA:sec:exp_software_appdx}
We implement our methods with PyTorch~\citep{pytorch} and PyTorch Geometric~\citep{pytorch_geometric}.
We ran our experiments on Linux Servers installed with V100 graphics cards and CUDA 10.2.

\subsection{Computational analysis}
\label{CH:GALA:sec:time_appdx}
\begin{table}[H]
    \center\small
    \caption{Averaged total training time of different methods.}
    \label{CH:GALA:tab:time_analysis}
    \begin{tabular}{lcccc}
        \toprule
        \textbf{Datasets} & Two-piece graphs   & EC50-Assay        & CMNIST-sp          & Graph-SST2          \\\midrule
        ERM               & 435.85\std{2.14}   & 80.45\std{10.27}  & 315.84\std{5.55}   & 374.31\std{1.28}    \\
        XGNN              & 673.82\std{0.81}   & 126.65\std{17.57} & 591.09\std{11.48}  & 722.44\std{48.51}   \\
        GREA              & 1128.28\std{34.57} & 210.30\std{21.23} & 902.06\std{8.49}   & 979.15\std{18.3114} \\
        GSAT              & 1205.67\std{62.54} & 228.88\std{25.04} & 791.55\std{15.67}  & 949.57\std{97.68}   \\
        DisC              & 1244.68\std{4.76}  & 207.50\std{17.72} & 932.40\std{76.99}  & 1280.77\std{551.97} \\
        MoleOOD           & 714.06\std{6.53}   & 136.39\std{17.87} & 439.49\std{9.10}   & 712.31\std{81.62}   \\
        GIL               & 533.46\std{11.42}  & 279.30\std{25.39} & 919.53\std{14.15 } & 733.36\std{147.08}  \\
        CIGA              & 873.49\std{16.21}  & 167.63\std{1.10}  & 650.94\std{5.01}   & 792.10\std{59.12}   \\
        GALA-cluster      & 811.41\std{3.20}   & 147.97\std{2.05}  & 756.41\std{21.63}  & 765.32\std{20.86}   \\
        GALA-pred         & 793.27\std{8.58}   & 149.89\std{2.71}  & 644.78\std{53.58}  & 764.69\std{30.98}   \\
        \bottomrule
    \end{tabular}
\end{table}

We calculate the average total training time of different methods at various datasets in seconds. As shown in Table.~\ref{CH:GALA:tab:time_analysis}, the training of GALA (no matter with clustering based sampling or prediction based sampling) does not bring much additional overhead than its counterpart CIGA.
When considering the additional training time of the assistant model with ERM, GALA costs only a competitive training time as environment generation based methods such as GREA and DisC. Notably, some methods such as DisC and GIL sometimes may be slow to converge even with the same early stop setting, which will cost even more time than the time cost by GALA plus the ERM training.
Besides, the ERM training time (for a assistant model) is not much long and usually around 5mins (or 300seconds in the table).

\chapter{Appendices of \gmt}\label{APP:GMT}

\section{Notations}
\label{CH:GMT:sec:notations_appdx}
In the following, we list notations for key concepts that have appeared in this paper.
\begin{table}[t]%
	\caption{Notations for key concepts involved in \gmt.}
	\centering
	\resizebox{\textwidth}{!}{
		\begin{tabular}{ll}
			\toprule
			\(\gG\)                     & the graph space                                                                                          \\\midrule
			\(\gG_c\)                   & the space of subgraphs with respect to the graphs from $\gG$                                             \\\midrule
			\(\gY\)                     & the label space                                                                                          \\\midrule
			\(\rho\)                    & the pooling function of the GNN                                                                          \\\midrule
			\(d(\cdot,\cdot)\)          & a distribution distance metric                                                                           \\\midrule
			\(L(\cdot,\cdot)\)          & the loss       function                                                                                  \\\midrule
			\(G\in\gG\)                 & a graph                                                                                                  \\\midrule
			\(G=(A,X)\)                 & a graph with the adjacency matrix $A\in\{0,1\}^{n\times n}$ and node feature matrix $X\in\R^{n\times d}$ \\
			                            & for brevity, we also use $G$ and $Y$ to denote the random variables as the graphs and labels             \\\midrule
			\(f=f_c\circ g\)            & a \xgnn with a subgraph extractor $g$ and a classifier $f_c$                                             \\\midrule
			\(g\)                       & a subgraph extractor $g:\gG\rightarrow\gG_c$                                                             \\\midrule
			\(f_c\)                     & a classifier GNN $f_c:\gG_c\rightarrow\gY$                                                               \\\midrule
			\(G_c\)                     & the invariant subgraph with respect to $G$                                                               \\\midrule
			\(G_s\)                     & the spurious subgraph with respect to $G$                                                                \\\midrule
			\(\pred{A}_c,\pred{A}\)     & the weighted adjacency matrix for causal subgraph with entries $A_{u,v}=\alpha_e$                        \\
			                            & as the sampling probability predicted by $g$                                                             \\\midrule
			\(\pred{A}_s\)              & the weighted adjacency matrix for spurious subgraph with entries $A_{u,v}=1-\alpha_e$                    \\
			                            & as the sampling probability predicted by $g$                                                             \\\midrule
			\(\pred{G}_c\)              & the estimated invariant subgraph produced by $g$                                                         \\
			                            & if the subgraph partitioning is conducted in an edge-centric view, then $\pred{G}_c=(X,\pred{A}_c)$      \\\midrule
			\(\pred{G}_s\)              & the estimated spurious subgraph produced by tacking the complementary of $\pred{G}_c$                    \\
			                            & if the subgraph partitioning is conducted in an edge-centric view, then $\pred{G}_s=(X,\pred{A}_s)$      \\\midrule
			\(I(G_c;Y)\)                & mutual information between the extracted subgraph $G_c$ and $Y$, specialized for maximizing $I(G;Y)$     \\\midrule
			\(P(G_c|G)\in\R_+\)         & the probability for sampling $G_c$ from $G$ with the subgraph extractor $g$                              \\\midrule
			\(P(Y|G)\in\R^{|\gY|}_+\)   & the label distribution of $Y$ conditioned on $G$                                                         \\\midrule
			\(P_f(Y|G)\in\R^{|\gY|}_+\) & the predicted label distribution of $Y$ conditioned on $G$                                               \\\midrule
			\(f_c(G_c)\in\R^{|\gY|}_+\) & the predicted label distribution of $Y$ with $f_c$ by taking the input $G_c$                             \\\midrule
			\bottomrule
		\end{tabular}}
\end{table}
\clearpage
\section{More Details about the Background}
\label{CH:GMT:sec:prelim_appdx}
We begin by introducing related works in Appendix~\ref{CH:GMT:sec:related_appdx} and then more backgrounds about graph information bottleneck in Appendix~\ref{CH:GMT:sec:GCE_deduce_appdx}, especially for how to obtain the formulas in the main text.
\subsection{More related works}
\label{CH:GMT:sec:related_appdx}
We give a more detailed background introduction of interpretable and generalizable GNNs (\xgnns) in this section.

\paragraph{Graph Neural Networks.}
We use $G=(A,X)$ to denote a graph with $n$ nodes and $m$ edges.
Within $G$, $A \in \{0,1\}^{n\times n}$ is the adjacency matrix, and $X\in \R^{n \times d}$ is the node feature matrix with a node feature dimension of $d$.
This work focuses on the task of graph classification.
Specifically, we are given a set of $N$ graphs $\{G_i\}_{i=1}^N\subseteq \gG$
and their labels $\{Y_i\}_{i=1}^N\subseteq\gY=\R^c$ from $c$ classes.
Then, we need to train a GNN $\rho \circ h$ with an encoder $h:\gG\rightarrow\R^h$ that learns a meaningful representation $h_G$ for each graph $G$ to help predict their labels $y_G=\rho(h_G)$ with a downstream classifier $\rho:\R^h\rightarrow\gY$.
The representation $h_G$ is typically obtained by performing pooling with a $\text{READOUT}$ function on the learned node representations:
\begin{equation}
	\label{CH:GMT:eq:gnn_pooling}
	h_G = \text{READOUT}(\{h^{(K)}_u|u\in V\}),
\end{equation}
where the $\text{READOUT}$ is a permutation invariant function (e.g., $\text{SUM}$, $\text{MEAN}$)~\citep{gin},
and $h^{(K)}_u$ stands for the node representation of $u\in V$ at $K$-th layer that is obtained by neighbor aggregation:
\begin{equation}
	\label{CH:GMT:eq:gnn}
	h^{(K)}_u = \sigma(W_K\cdot a(\{h^{(K-1)}_v\}| v\in\mathcal{N}(u)\cup\{u\})),
\end{equation}
where $\mathcal{N}(u)$ is the set of neighbors of node $u$,
$\sigma(\cdot)$ is an activation function, e.g., $\text{ReLU}$, and $a(\cdot)$ is an aggregation function over neighbors, e.g., $\text{MEAN}$.

\paragraph{Interpretable GNNs.}
Let $G=(A,X)$ be a graph with node set $V=\{v_1,v_2,...,v_n\}$ and edge set $E=\{e_1,e_2,...,e_m\}$,
where  $A \in \{0,1\}^{n\times n}$  is the adjacency matrix and $X\in \R^{n \times d}$ is the node feature matrix.
In this work, we focus on interpretable GNNs (or \xgnns) for the graph classification task, while the results can be generalized to node-level tasks as well~\citep{gib_node}.
Given each sample from training data $\train=(G^i,Y^i)$,
an interpretable GNN $f:=h\circ g$ aims to identify a (causal) subgraph $G_c\subseteq G$ via a subgraph extractor GNN $g:\gG\rightarrow\gG_c$, and then predicts the label via a subgraph classifier GNN $f_c:\gG_c\rightarrow\gY$, where $\gG,\gG_c,\gY$ are the spaces of graphs, subgraphs, and the labels, respectively~\citep{gib}.
Although \textit{post-hoc} explanation approaches also aim to find an interpretable subgraph as the explanation for the model prediction~\citep{gnn_explainer,xgnn,pgm_explainer,pge,subgraphxgn,gen_xgnn,orphicx}, they are shown to be suboptimal in interpretation performance and sensitive to the performance of the pre-trained GNNs~\citep{gsat}.
Therefore, this work focuses on \textit{intrinsic interpretable} GNNs (XGNNs).

A predominant approach to implement \xgnns is to incorporate the idea of information bottleneck~\citep{ib}, such that $G_c$ keeps the minimal sufficient information of $G$ about $Y$~\citep{gib,vgib,gsat,lri,gib_hiera},
which can be formulated as
\begin{equation}
	\max_{G_c}I(G_c;Y)-\lambda I(G_c;G),\ G_c\sim g(G),
\end{equation}
where  maximizing the mutual information between $G_c$ and $Y$ endows the interpretability of $G_c$ while minimizing $I(G_c;G)$ ensures $G_c$ captures only the most necessary information, $\lambda$ is a hyperparamter trade off between the two objectives.
In addition to minimizing $I(G_c;G)$, there are also alternative approaches that impose different constraints such as causal invariance~\citep{ciga,gil} or disentanglement~\citep{dir,cal,grea,disc} to identify the desired subgraphs.
When extracting the subgraph, \xgnns adopts the attention mechanism to learn the sampling probability of each edge or node, which avoids the complicated Monte Carlo tree search used in other alternative implementations~\citep{protGNN}.
Specifically, given node representation learned by message passing $H_i\in\R^h$ for each node $i$, \xgnns either learns a \textbf{node attention} $\alpha_i\in\R_+=\sigma(a(H_i))$ via the attention function $a:\R^h\rightarrow\R_+$, or the \textbf{edge attention} $\alpha_e\in\R_+=\sigma(a([H_u,H_v]))$ for each edge $e=(u,v)$ via the attention function $a:\R^{2h}\rightarrow\R_+$, where $\sigma(\cdot)$ is a sigmoid function. $\boldsymbol{\alpha}=[\alpha_1,...,\alpha_m]^T$ essentially elicits a subgraph distribution of the interpretable subgraph. In this work, we focus on edge attention-based subgraph distribution as it is most widely used in \xgnns while our method can be easily generalized to node attention-based subgraph approaches as demonstrated in the experiments with geometric learning datasets.

Besides, \citet{gatv2,aero_gnn} find the failures of graph attention networks in properly propagating messages with the attention mechanism. They differ from our work as they focus on node classification tasks.

\textbf{Faithful interpretation and (OOD) generalization.}
The faithfulness of interpretation is critical to all interpretable and explainable methods~\citep{fidelity,mythos_inter,robust_xnn,Rudin2018StopEB,att_not_exp,relation_exp_pred}.
Yet, there are many failure cases found especially when with attention mechanisms. For example, \citet{att_not_exp} reveals that in NLP, randomly shuffling or imposing adversarial noises will not affect the predictions too much, highlighting a weak correlation between attention and prediction.
\citet{relation_exp_pred} present a causal analysis showing the hyperparameters and the architecture setup could be a cofounder that affects the causal analysis. \citet{inv_rat} show interpretations will fail when  distribution shifts are presented.
Although the faithfulness of explanation/interpretations has been widely a concern for Euclidean data, whether and how GNNs and \xgnns suffer from the same issue is under-explored.

Talking about the progress in graph data, there are several metrics developed to measure the faithfulness of graph explanations, such as fidelity~\citep{xgnn_tax,GraphFramEx}, counterfactual robustness~\citep{RCExplainer,counterfactual_xgnn_sur,clear}, and equivalence~\citep{xgnn_equi}, which are however limited to post-hoc graph explanation methods.
In fact, post-hoc explanation methods are mostly developed to adhere the faithfulness measures such as fidelity. However, as shown by~\citet{gsat}, the post-hoc methods are suboptimal in finding the interpretable subgraph and sensitive to the pre-trained model, which highlights a drawback of the existing faithfulness measure.
In contrast, we develop the first faithfulness measure for \xgnns in terms of counterfactual invariance.

Although \citet{RCExplainer,counterfactual_xgnn_sur,clear} also adopt the concept of counterfactual to develop post-hoc explanation methods, they focus on finding the minimal perturbations that will change the predictions. Counterfactual is also widely used to improve graph representation learning~\citep{counterfactual_gl}.
In contrast, we adopt the concept of counterfactual to measure the sensitivity of the \xgnns predictions to the predicted attention.

\begin{figure*}[ht]
	\centering\hfill
	\subfigure[Graph generation SCM]{\label{CH:GMT:fig:graph_gen_appdx}
		\resizebox{!}{0.225\textwidth}{\tikz{
				\node[latent] (S) {$S$};%
				\node[latent,left=of S,xshift=-1.5cm] (C) {$C$};%
				\node[latent,below=of C,xshift=-0.75cm,yshift=0.5cm] (ZCA) {$Z_X^c$}; %
				\node[latent,below=of C,xshift=0.75cm,yshift=0.5cm] (ZCX) {$Z_A^c$}; %
				\node[latent,below=of S,xshift=-0.75cm,yshift=0.5cm] (ZSA) {$Z_X^s$}; %
				\node[latent,below=of S,xshift=0.75cm,yshift=0.5cm] (ZSX) {$Z_A^s$}; %
				\node[latent,below=of ZCX,xshift=-0.75cm,yshift=0.5cm] (GC) {$G_c$}; %
				\node[latent,below=of ZSX,xshift=-0.75cm,yshift=0.5cm] (GS) {$G_s$}; %
				\node[obs,below=of GC,xshift=1.6cm,yshift=0.5cm] (G) {$G$}; %
				\edge[dashed,-] {C} {S}
				\edge {C} {ZCX,ZCA}
				\edge {S} {ZSX,ZSA}
				\edge {ZCX,ZCA} {GC}
				\edge {ZSX,ZSA} {GS}
				\edge {GC,GS} {G}
			}}}
	\subfigure[FIIF SCM]{\label{CH:GMT:fig:scm_fiif_appdx}
		\resizebox{!}{0.18\textwidth}{\tikz{
				\node[latent] (E) {$E$};%
				\node[latent,below=of E,yshift=0.5cm] (S) {$S$}; %
				\node[obs,below=of E,xshift=-1.2cm,yshift=0.5cm] (Y) {$Y$}; %
				\node[obs,below=of E,xshift=1.2cm,yshift=0.5cm] (G) {$G$}; %
				\node[latent,below=of Y,xshift=1.2cm,yshift=0.5cm] (C) {$C$}; %
				\edge {E} {S}
				\edge {C} {Y,G}
				\edge {S} {G}
				\edge {C} {S}
			}}}
	\subfigure[PIIF SCM]{\label{CH:GMT:fig:scm_piif_appdx}
		\resizebox{!}{0.18\textwidth}{\tikz{
				\node[latent] (E) {$E$};%
				\node[latent,below=of E,yshift=0.5cm] (S) {$S$}; %
				\node[obs,below=of E,xshift=-1.2cm,yshift=0.5cm] (Y) {$Y$}; %
				\node[obs,below=of E,xshift=1.2cm,yshift=0.5cm] (G) {$G$}; %
				\node[latent,below=of Y,xshift=1.2cm,yshift=0.5cm] (C) {$C$}; %
				\edge {E} {S}
				\edge {C} {Y,G}
				\edge {S} {G}
				\edge {Y} {S}
			}}}
	\subfigure[MIIF SCM]{\label{CH:GMT:fig:scm_miif_appdx}
		\resizebox{!}{0.24\textwidth}{\tikz{
				\node[latent] (E) {$E$};%
				\node[latent,below=of E,xshift=-2cm] (S1) {$S_1$}; %
				\node[latent,below=of E,xshift=-0.6cm] (C) {$C$}; %
				\node[latent,below=of E,xshift=2cm] (S2) {$S_2$}; %
				\node[obs,below=of E,xshift=0.6cm] (Y) {$Y$}; %
				\node[obs,below=of C,xshift=0.6cm] (G) {$G$}; %
				\edge {E} {S1,S2}
				\edge {C} {S1,Y,G}
				\edge {Y} {S2}
				\edge {S1,S2} {G}
			}}}
	\caption{Full SCMs on Graph Distribution Shifts~\citep{ciga}.}
	\label{CH:GMT:fig:scm_appdx}
\end{figure*}

\paragraph{On the natural connection of \xgnns and OOD generalization on graphs.}
In the context of graph classification, the generalization ability and the faithfulness of the interpretation are naturally intertwined in \xgnns.
In many realistic graph classification practices such as drug property prediction~\citep{drugood,ai4sci}, the property of a drug molecule can naturally be represented by a subgraph, termed as causal subgraph. The causal subgraph, in return, holds a causal relationship with the drug property. Therefore, it is natural to identify the underlying causal subgraph to provide OOD generalizable predictions and interpretations.

Typically, \xgnns need to extract the underlying ground truth subgraph in order to make correct predictions on unseen test graphs~\citep{gsat}. When distribution shifts are presented in the test data, it is critical to find the underlying subgraph that has a causal relationship with the target label (or causal subgraphs) ~\citep{inv_rat,ciga}.

We now briefly introduce the background of causal subgraph and OOD generalization.
Specifically,
we are given a set of graph datasets $\dataset=\{\dataset_e\}_e$ collected from multiple environments $\envall$.
Samples $(G^e_i, Y^e_i)\in \dataset^e$ from the same
environment are considered as drawn independently from an identical distribution $\sP^e$.
We consider the graph generation process proposed
by~\citet{ciga} that covers a broad case of graph distribution shifts.
Fig.~\ref{CH:GMT:fig:scm_appdx} shows the full graph generation process considered in~\citet{ciga}.
The generation of the observed graph $G$ and labels $Y$
are controlled by a set of latent causal variable $C$ and spurious variable $S$, i.e.,
\[G\coloneqq f_\gen(C,S).\]
$C$ and $S$ control the generation of $G$ by controlling the underlying invariant subgraph $G_c$
and spurious subgraph $G_s$, respectively.
Since $S$ can be affected by the environment $E$,
the correlation between $Y$, $S$ and $G_s$ can change arbitrarily
when the environment changes.
$C$ and $S$ control the generation of the underlying invariant subgraph $G_c$
and spurious subgraph $G_s$, respectively.
Since $S$ can be affected by the environment $E$,
the correlation between $Y$, $S$ and $G_s$ can change arbitrarily
when the environment changes.
Besides, the latent interaction among $C$, $S$ and $Y$
can be further categorized into \emph{Full Informative Invariant Features} (\emph{FIIF})
when $Y\ind S|C$ and \emph{Partially Informative Invariant Features} (\emph{PIIF}) when $Y \not\ind S|C$. Furthermore, PIIF and FIIF shifts can be mixed together and yield \emph{Mixed Informative Invariant Features} (\emph{MIIF}), as shown in Fig.~\ref{CH:GMT:fig:scm_appdx}.
We refer interested readers to~\citet{ciga} for a detailed introduction to the graph generation process.

To tackle the OOD generalization challenge on graphs generated following in Fig.~\ref{CH:GMT:fig:scm_appdx},
the existing invariant graph learning approaches generically
aim to identify the underlying invariant subgraph $G_c$ to predict the label $Y$~\citep{handle_node,ciga}.
Specifically, the goal of OOD generalization on graphs
is to learn an invariant \xgnn $f\coloneqq f_c\circ g$, with the following objective:
\begin{equation}
	\label{CH:GMT:eq:inv_cond_appdx}
	\text{$\max$}_{f_c, \; g} \ I(\pred{G}_{c};Y), \ \text{s.t.}\ \pred{G}_{c}\ind E,\ \pred{G}_{c}=g(G).
\end{equation}
Since $E$ is not observed, many strategies are proposed to
impose the independence of $\pred{G}_c$ and $E$.
A common approach is to augment the environment information.
For example, based on the estimated invariant subgraphs $\pred{G}_c$ and spurious subgraphs $\pred{G}_s$,
\citet{dir,grea,handle_node,dps} propose to generate new environments, while \citet{gil} propose to infer the underlying environment labels via clustering.
\citet{moleood} propose a variational framework to infer the environment labels.
\citet{joint_causal_indep} propose to learn causal independence between labels and environments.
\citet{gib,vgib,gsat,lri,gib_hiera} adopt graph information bottleneck to tackle FIIF graph shifts, and they cannot generalize to PIIF shifts.
Nevertheless, since most of the existing works adopt the backbone of \xgnns, and \xgnns with information bottleneck is the state-of-the-art method with both high interpretation performance and OOD generalization performance, the focus in this work will be around tackling FIIF shifts with the principle of graph information bottleneck. More details are given in the next section.

In addition to the aforementioned approaches, \citet{size_gen1,size_gen2,size_gen3} study the OOD generalization as an extrapolation from small graphs to larger graphs in the task of graph classification and link prediction. In contrast, we study OOD generalization against various graph distribution shifts formulated in Fig.~\ref{CH:GMT:fig:scm_appdx}.
\citet{graph_joint_extra} propose an extrapolation strategy to improve OOD generalization on graphs.
In addition to the standard OOD generalization tasks studied in this paper, \citet{nn_extrapo,OOD_CLRS} study the OOD generalization in tasks of algorithmic reasoning on graphs. \citet{graph_ttt} study the test-time adaption in the graph regime. \citet{shape_matching} study the 3D shape matching under the presence of noises.

\textbf{Multilinear extension.}
Multilinear extension serves as a powerful tool
for maximizing combinatorial functions, especially for submodular set function maximization \citep{mt_game,mt,Vondrak08,Calinescu11,Chekuri14,Chekuri15,optimal_drsub,sets2multisets,bian2022energybased,neural_set}.
For example, \citet{Vondrak08,Calinescu11} study the multilinear extension in the context of social welfare. \citet{bian2022energybased} study the multilinear extension for cooperative games.
It is the expected value of a set function under the fully factorized i.i.d. Bernoulli distribution.
The closest work to ours is~\citet{neural_set} that builds neural set function extensions for multiple discrete functions.
Nevertheless, to the best of our knowledge, the notion of multilinear extensions for \xgnns is yet underexplored.
In contrast, in this work, we are the first to identify subgraph multilinear extension as the factorized subgraph distribution for interpretable subgraph learning.

\subsection{Variational bounds and realization of the IB principle}
\label{CH:GMT:sec:GCE_deduce_appdx}
We first introduce how to derive Eq.~\ref{CH:GMT:eq:GCE} in the main text, and then discuss how to implement the graph information bottleneck regularization $\min I(G_c;G)$ following the state-of-the-art architecture \gsat~\citep{gsat,lri}.

\paragraph{Variational bounds for $I(G;Y)$.}
For the term $I(G;Y)$, notice that

\begin{equation}\label{}
	\begin{aligned}
		I(G;Y) = \E_{G, Y} \left[ \log \frac{P(Y|G)}{P(Y)}  \right]
	\end{aligned}
\end{equation}
Since the true $P(Y|G)$ is intractable, through XGNN modelling we introduce a variational approximation $P_{f_c, g}(Y|G)$. Then,
\begin{align}
	I(G;Y) & = \E_{G, Y} \left[ \log \frac{P_{f_c, g}(Y|G)}{P(Y)} \right]  +  \E_{G, Y} \left[ \log \frac{P(Y|G)}{P_{f_c, g}(Y|G)} \right] \\
	       & = \E_{G, Y} \left[ \log \frac{P_{f_c, g}(Y|G)}{P(Y)} \right]  +  \KL (P(Y|G) || P_{f_c, g}(Y|G))                              \\
	       & \geq  \E_{G, Y} \left[ \log P_{f_c, g}(Y|G)  \right]  + H(Y)
\end{align}
Since the optimization does not involve $H(Y) $, we continue with
$\E_{G, Y} \left[ \log P_{f_c, g}(Y|G)  \right] $,
\begin{align}
	\E_{G, Y} \left[ \log P_{f_c, g}(Y|G)  \right] & =  \E_{G, Y} \left[ \log \sum_{G_c} P_{f_c, g}(Y, G_c|G)  \right]                   \\
	                                               & =  \E_{G, Y} \left[ \log \sum_{G_c} P_{f_c, g}(Y|G, G_c)  P_{f_c, g}(G_c|G) \right] \\  \label{overall_xgnn_imple}
	                                               & = \E_{G, Y} \left[ \log \sum_{G_c} P_{f_c}(Y|G_c)  P_{g}(G_c|G) \right]
\end{align}
where Eq. \ref{overall_xgnn_imple} is due to the implementation of XGNNs.
Eq. \ref{overall_xgnn_imple} can also be written with expectations:
\[
	\E_{G, Y} \left[ \log \sum_{G_c} P_{f_c}(Y|G_c)  P_{g}(G_c|G) \right]=
	\E_{G, Y} \left[ \log \E_{G_c\sim \mathbb{P}(G_c|G)} P_{f_c}(Y|G_c) \right].
\]
Maximizing $I(G;Y)$ is then equivalent to minimizing $-I(G;Y)$, and further minimizing $\E_{G,Y}[-\log P_{f_c,g}(Y|G)]$.
This achieves to Eq.~\ref{CH:GMT:eq:GCE} in the main text, i.e.,
\[
	\begin{aligned}
		\E_{(G,Y)\sim\train}[-\log P(Y|\E_{G_c\stackrel{g}{\sim}G}G_c)]
		=\E_{(G,Y)\sim\train} [L(f_c(\boldsymbol{\alpha};G),Y)],
	\end{aligned}
\]
with $L$ as the cross entropy loss, and $\boldsymbol{\alpha}$ as the predicted sampling probability for edges. $\boldsymbol{\alpha}$ factorizes the sampling probability of the subgraphs as independent Bernoulli distributions on edges $e\sim \text{Bern}(\alpha_e), \forall e\in E$:
\[
	P(G_c|G)=\prod_{e\in G_c}\alpha_e\prod_{e\in G/G_c}(1-\alpha_e).
\]

\paragraph{Variational bounds for $I(G_c;G)$.}
For the term $I(G_c;G)$, since we factorize graph distribution as multiple independent Bernoulli distributions on edges, we are able to calculate the KL divergence to upper bound $I(G_c;G)$:
\begin{align}
	I(G_c;G)\leq D_\text{KL}(P(G_c|G)||Q(G_c)),
\end{align}
where $Q(G_c)$ is a variational approximation to $P(G_c)$. $D_\text{KL}$ can be obtained via
\begin{align}
	D_\text{KL}(P(G_c|G)||Q(G_c))=\sum_{e\in G_c}D_\text{KL}(\text{Bern}(\alpha_e)||\text{Bern}(r))+c(n,r),
\end{align}
where $c(n,r)$ is a small constant, $r$ is a hyperparameter to specify the prior for subgraph distributions. To minimize $I(G_c;G)$ is essentially to minimize $D_\text{KL}(\text{Bern}(\alpha_e)||\text{Bern}(r))$. The KL divergence can be directly calculated as
\begin{equation}\label{CH:GMT:eq:gib_reg_appdx}
	D_\text{KL}(\text{Bern}(\alpha_e)||\text{Bern}(r))=\sum_{e}\alpha_e\log\frac{\alpha_e}{r}+(1-\alpha_e)\log\frac{(1-\alpha_e)}{(1-r)}.
\end{equation}
\citet{gsat} find the mutual information based regularization can effectively regularize the information contained in $G_c$ than previous implementations such as vanilla size constraints with the norm of attention scores or connectivity constraints~\citep{gib}.

Besides, we would like to note that  \gsat implementation provided by the author  does not exactly equal to the mathematical formulation, i.e., they directly take the unormalized attention to Eq.~\ref{CH:GMT:eq:gib_reg_appdx}, as acknowledged by the authors~\footnote{\url{https://github.com/Graph-COM/GSAT/issues/10}}.
The reason for using another form of information regularization is because the latter empirically performs better.
Nevertheless, \lri adopts the mathematically correct form and obtains better empirical performance. In our experiments, we adopt the mathematically correct form for both regular and geometric learning tasks, in order to align with the theory. Empirically, we find the two forms perform competitively well with the suggested hyperparemters and hence stick to the mathematically correct form.

\section{On the Generalization and Interpretability: A Causal View}
\subsection{Structural Causal Model for \xgnns}
\label{CH:GMT:sec:scm_xgnn_appdx}
We provide a detailed description and the full structural causal model of \xgnns in complementary to the causal analysis in Sec.~\ref{CH:GMT:sec:causal_view}.
\begin{figure}[H]
	\centering
	\includegraphics[scale=0.8]{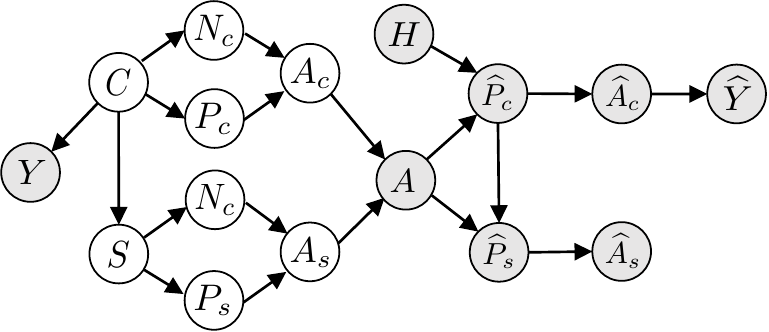}
	\caption{Bernoulli Parameterized SCM for interpretable GNN}
	\label{CH:GMT:fig:scm_ber_full_appdx}
\end{figure}
\paragraph{Data generation.} We consider the same data model as previous works~\citep{size_gen2,gsat,ciga},
where the underlying causal subgraph $G_c$ and the spurious subgraph $G_s$ will be assembled via some underlying assembling process $G = f_g(G_c,G_s)$,
as illustrated in Appendix~\ref{CH:GMT:sec:prelim_appdx} Fig.~\ref{CH:GMT:fig:scm_appdx}.

We focus on the FIIF distribution shifts (Fig.~\ref{CH:GMT:fig:scm_fiif_appdx}) that can be resolved by graph information bottleneck~\citep{gsat,ciga}.
As shown in the figure, there are latent causal and spurious variables $C$ and $S$ that have an invariant and spurious correlation with the label $Y$, respectively. $C$ and $S$ further control the generation of the graph structure of the causal subgraph $G_c$, and the spurious subgraph $G_s$.
Specifically, $C$ and $S$ will specify the number of nodes in $G_c$ and $G_s$ as $N_c$ and $N_s$. Then, $C$ and $S$ further control the underlying Bernoulli distributions on edges, by specifying the sampling probability as $P_c$ and $P_s$. With $N_c$ and $P_c$ (or $N_s$ and $P_s$), $A_c$ (or $A_s$) can be sampled and then assembled into the observed graph structure $A$.
As we focus on the edge-centric view, our discussion focuses on the graph structures $A_c$ and $A_s$ of the subgraphs.
Nevertheless, a similar generation model can also be developed for the node-centric view.

\paragraph{Interpretation.} Correspondingly, \xgnns first uses a subgraph extractor to predict the causal and spurious subgraphs $\widehat{G}_c$ and $\widehat{G}_s$, respectively.
The extraction aims to reverse the generation and recover the underlying $P_c$, by learning the $\widehat{P}_c$ via the attention $\boldsymbol{\alpha}$. We denote the architecture and the hyperparameter settings  as $H$.
Once $\widehat{P}_c$ is determined, $\widehat{P}_s=1-\widehat{P}_c$ is also obtained by finding the complementary part. Then, the estimated causal and spurious subgraphs are sampled from $\widehat{P}_c$ and $\widehat{P}_s$, respectively.
With the estimated causal subgraph $\widehat{G}_c=(X,\widehat{A}_c)$, the classifier GNN $c(\cdot)$ will use it to make a prediction $\widehat{Y}$.

\subsection{Practical Estimation of Counterfactual Fidelity}
\label{CH:GMT:sec:pratical_cf_appdx}
Since it is prohibitively expensive to enumerate all possible $\widetilde{G}$ and the distance $\delta$ to examine the counterfactual fidelity. We instead consider an alternative notion that adopts random perturbation onto the learned attention score.
Specifically, we consider a random attention matrix $\widetilde{A}\sim\sigmoid(\gN(\mu_{\widehat{H}_A},\sigma_{\widehat{H}_A}))$,
where $\mu_{\widehat{H}_A}$ and $\sigma_{\widehat{H}_A}$ are the mean and standard deviation of the pre-attention matrix $\widehat{H}_A$ (The adjacency matrix with the unnormalized attention).
Since each non-symmetric entry in the attention is generated independently,
each non-symmetric entry in $\widetilde{A}$ is sampled independently following the factorization of $P(G)$.
We randomly sample $\widetilde{A}$ by $k$ times and calculate the following:
\begin{equation}\label{CH:GMT:eq:cf_appdx}
	c_{\widehat{G}_c} = \frac{1}{k}\sum_{i=1}^k d(f_c(Y|\widetilde{G}^i_c),f_c(Y|\widehat{G}_c)),
\end{equation}
where $\widetilde{G}^i_c=(X,\widetilde{A}^i_c)$ and $d$ is total variation distance.
The detailed computation of the practical counterfactual fidelity is provided in Algorithm~\ref{alg:counterfactual_fidelity}.

\begin{algorithm}[ht]
	\caption{Practical estimation of counterfactual fidelity. }
	\label{alg:counterfactual_fidelity}
	\begin{algorithmic}[1]
		\STATE \textbf{Input:} Training data $\train$;
		a trained \xgnn $f$ with subgraph extractor $g$, and classifier $f_c$;
		sampling times $e_s$;
		batch size $b$;
		total variation distance $d(\cdot)$;
		\STATE \texttt{// Minibatch sampling.}
		\FOR{$j=1$ to $|\train|/b$}
		\STATE Sample a batch of data $\{G^i,Y^i\}_{i=1}^b$ from $\train$;
		\STATE Obtain the pre-attention matrix $\widehat{H}_A$;
		\STATE Obtain the attention matrix $\widehat{A}=\sigmoid(\widehat{H}_A)$;
		\STATE Obtain the original prediction with $f_c$ based on the attention matrix $\widehat{A}$ as $\{\hat{y}^i\}_{i=1}^b$;
		\STATE \texttt{// Random noises injection.}
		\FOR{$k=1$ to $e_s$}
		\STATE Sample a random attention matrix $\widetilde{A}\sim\sigmoid(\gN(\mu_{\widehat{H}_A},\sigma_{\widehat{H}_A}))$;
		\STATE Obtain sampling attention $\{\boldsymbol{\alpha}^i\}_{i=1}^b$;
		\STATE Obtain the perturbed prediction with $f_c$ based on the attention matrix $\widetilde{A}$ as $\{\hat{y}^i_k\}_{i=1}^b$;
		\ENDFOR
		\STATE Calculate $\{c_{\widehat{G}_c}^i\}_{i=1}^b$ with $k$ groups of $\{\hat{y}^i_k\}_{i=1}^b$ and $\{\hat{y}^i\}_{i=1}^b$;
		\STATE Obtain the averaged $c_{\widehat{G}_c}^j$ within the batch;
		\ENDFOR
		\STATE Obtain the averaged $c_{\widehat{G}_c}$ within the training data;
		\STATE \textbf{Return} estimated $c_{\widehat{G}_c}$;
	\end{algorithmic}
\end{algorithm}

\begin{figure}[ht]
    \centering
    \subfigure[\smt on BA-2Motifs trainset.]{
        \includegraphics[width=0.3\textwidth]{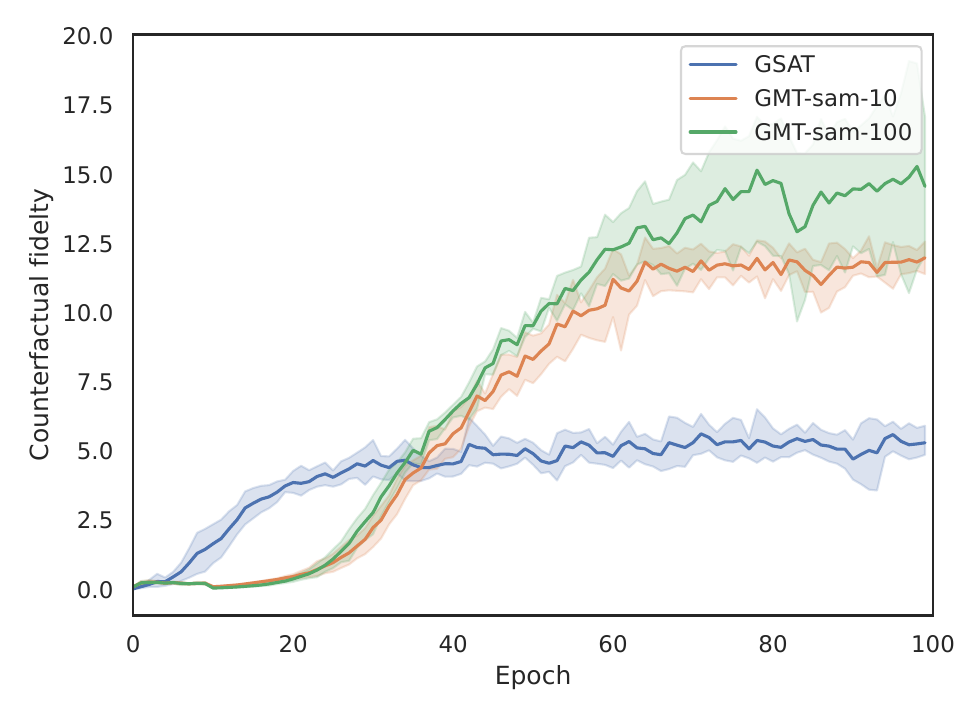}
        \label{CH:GMT:fig:cf_ba_train_appdx}
    }
    \subfigure[\smt on BA-2Motifs valset.]{
        \includegraphics[width=0.3\textwidth]{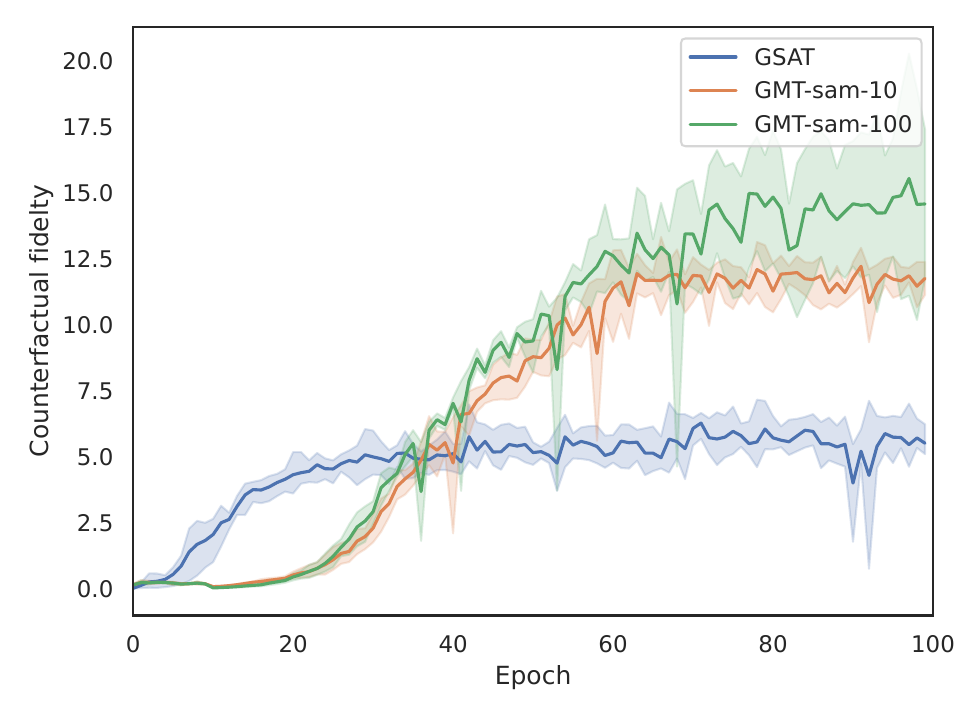}
        \label{CH:GMT:fig:cf_ba_val_appdx}
    }
    \subfigure[\smt on BA-2Motifs test set.]{
        \includegraphics[width=0.3\textwidth]{Figures/GMT/cf_ba_test_logits_tvd.pdf}
        \label{CH:GMT:fig:cf_ba_test_appdx}
    }
    \caption{Counterfactual fidelity on BA-2Motifs.}
    \label{CH:GMT:fig:cf_ba_appdx}
\end{figure}
\begin{figure}[ht]
    \centering
    \subfigure[\smt on Mutag trainset.]{
        \includegraphics[width=0.3\textwidth]{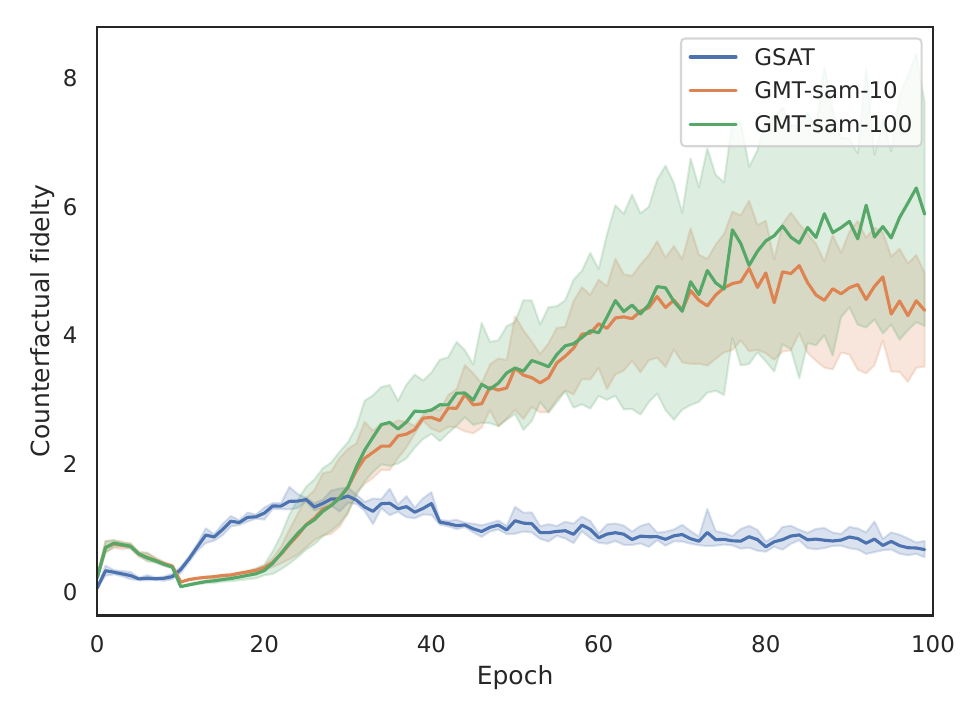}
        \label{CH:GMT:fig:cf_mu_train_appdx}
    }
    \subfigure[\smt on Mutag validation set.]{
        \includegraphics[width=0.3\textwidth]{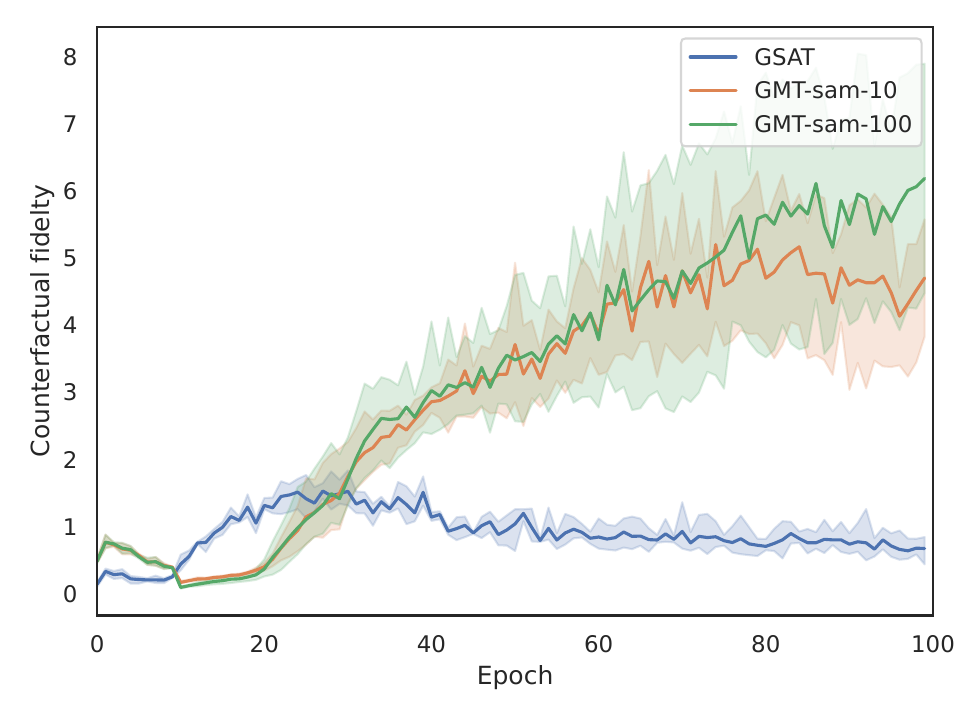}
        \label{CH:GMT:fig:cf_mu_val_appdx}
    }
    \subfigure[\smt on Mutag test set.]{
        \includegraphics[width=0.3\textwidth]{Figures/GMT/cf_mu_test_logits_tvd.pdf}
        \label{CH:GMT:fig:cf_mu_test_appdx}
    }
    \caption{Counterfactual fidelity on Mutag.}
    \label{CH:GMT:fig:cf_mu_appdx}
\end{figure}

Shown as in Fig.~\ref{CH:GMT:fig:cf_ba_appdx},~\ref{CH:GMT:fig:cf_mu_appdx},
we plot the counterfactual fidelity of \gsat and the simulated \smt with $10$ and $100$ sampling rounds on BA-2Motifs and Mutag datasets.
The \smt is approximated via \gmts with different sampling rounds.
It can be found that \gsat achieves a counterfactual fidelity that is $2$ to $3$ times lower than the simulated \smt via \gmts with $10$ and $100$ sampling rounds.
Moreover, in simple tasks such as BA-2Motifs and Mutag, using larger sampling rounds like $100$ does not necessarily bring more counterfactual fidelity. One reason can be using small sampling rounds to touch the upper bounds of counterfactual fidelity measured in our work.
We also provide a discussion on why the counterfactual fidelity grows slowly at the initial epochs in BA-2Motif datasets in Appendix~\ref{CH:GMT:sec:gmt_impl_appdx}.
More counterfactual fidelty studies can be found in Appendix~\ref{CH:GMT:sec:cf_viz_appdx}.

\section{Theories and Proofs}
\label{CH:GMT:sec:proof_appdx}

\subsection{Useful definitions}
We give the relevant definitions here for ease of reference when reading our proofs.
\begin{definition}[Subgraph multilinear extension (\smt)]\label{def:sub_mt_appdx}
	Given the attention $\boldsymbol{\alpha}\in\R^m_+$ as edge sampling probability of $G_c$, \xgnns factorize $P(G)$ as independent Bernoulli distributions on edges:
	\[P(G_c|G)=\prod_{e\in G_c}\alpha_e\prod_{e\in G/ G_c}(1-\alpha_e),\]
	which elicits the \textit{multilinear extension} of $f_c(G_c)$ in Eq.~\ref{CH:GMT:eq:GCE} as:
	\begin{equation}\label{CH:GMT:eq:smt_appdx}
		\begin{aligned}
			F_c(\boldsymbol{\alpha}; G) :=\sum_{G_c\in G}f_c(G_c)\prod_{e\in G_c}\alpha_e\prod_{e\in G/G_c}(1-\alpha_e) =\E_{G_c\stackrel{g}{\sim}G}f_c(G_c).
		\end{aligned}
	\end{equation}
\end{definition}

\begin{definition}[$\epsilon$-\smt approximation]\label{def:submt_approx_appdx}
	Let $d(\cdot,\cdot)$ be a distribution distance metric, a \xgnn $f=f_c\circ g$ $\epsilon$-approximates \smt (Def.~\ref{def:sub_mt}), if there exists $\epsilon\in\R_+$ such that $d(P_f(Y|G),P(Y|G))\leq\epsilon$
	where $P(Y|G)\in\R^{|\gY|}$ is the ground truth conditional label distribution, and $P_f(Y|G)\in\R^{|\gY|}$ is the predicted label distribution for $G$ via a \xgnn $f$, i.e., $P_f(Y|G)=f_c(\E_{G_c\stackrel{g}{\sim}G}G_c)$.
\end{definition}

\begin{definition}[$(\delta,\epsilon)$-counterfactual fidelity]\label{def:counterfactual_x_appdx}
	Given a meaningful minimal distance $\delta>0$,
	let $d(\cdot,\cdot)$ be a distribution distance metric ,
	if a \xgnn $f=f_c\circ g$ commits to the $\epsilon-$counterfactual fidelity, then there exist $\epsilon>0$ such that, $\forall G, \widetilde{G}$ that $d(P(Y|G),P(Y|\widetilde{G}))\geq \delta$, the following holds:
	\[d(P_f(Y|\widetilde{G}),P_f(Y|G))\geq \epsilon\delta.\]
\end{definition}

\subsection{Proof for Proposition~\ref{CH:GMT:thm:submt_gap}}
\label{proof:submt_gap}
\begin{proposition}\label{CH:GMT:thm:submt_gap_appdx}
	Consider a linearized GNN~\citep{sgnn} with number of message passing layers $k>1$, linear activations and pooling,
	\begin{equation}\label{CH:GMT:eq:linear_gnn_appdx}
		f_c(G_c)=\rho(\widehat{A}^kX \mW),
	\end{equation}
	if there exists $1\leq i,j\leq n$ that $0<\widehat{A}_{i,j}<1$,
	Eq.~\ref{CH:GMT:eq:exp_issue} can not hold, thus Eq.~\ref{CH:GMT:eq:linear_gnn_appdx}
	can not  approximate \smt (Def.~\ref{def:sub_mt}).
\end{proposition}
\begin{proof}
	To begin with, given a linear pooling function $\rho$, one could write the outcomes of $f_c(A)=\rho(A^kXW)$ as a summation in $A^k_{i,j}v_{i,j}$, with $v_{i,j}$ is the weight that accounting for the pooling as well as $XW$:
	\begin{align}
		f_c(A)=\sum_{i}\sum{j}A_{i,j}v_{i,j}.
	\end{align}
	Given the linearity of expectations, the comparison between $E[f_c(A)]$ and $f_c(E[A])$ now turns into the comparison between $E[A^k_{i,j}v_j]$ and $(E[A_{i,j}])^kv_j$.
	Since $A_ij$ is drawn from the Bernoulli distribution, with the expectation as $\widehat{A}_{i,j}$, it suffices to know that \begin{align}
		E[A^k_{i,j}v_j]=1^k\widehat{A}_{i,j}+0^k(1-\widehat{A}_{i,j})=\widehat{A}_{i,j},
	\end{align}
	while $(E[A_{i,j}])^k=\widehat{A}_{i,j}^k$. Then, we know that $E[f_c(A)] \neq f_c(E[A])$.
\end{proof}
We also conduct empirical verifications with \gsat implemented in GIN and SGC with various layers in Appendix~\ref{CH:GMT:sec:smt_gap_viz_appdx}.

\subsection{Proof for Proposition~\ref{CH:GMT:thm:smt_fidelty}}
\label{proof:smt_fidelty}
\begin{proposition}\label{CH:GMT:thm:smt_fidelty_appdx}
	If a \xgnn $f$ $\epsilon$-approximates \smt (Def.~\ref{def:submt_approx_appdx}),  then $f$ also satisfies $(\delta,1\!-\!\frac{2\epsilon}{\delta})$-counterfactual fidelity (Def.~\ref{def:counterfactual_x_appdx}).
\end{proposition}
\begin{proof}
	Considering any two graphs $G$ and $\widetilde{G}$ that $d(P(Y|G),P(Y|\widetilde{G})\geq \delta$,
	since $d$ is a distance metric, we have the following inequality holds:
	\begin{align}
		d(P(Y|G),P_f(Y|\widetilde{G})) \leq d(P_f(Y|G),P(Y|G))+d(P_f(Y|G),P_f(Y|\widetilde{G})),
	\end{align}
	by the triangle inequality.
	Furthermore, we have
	\begin{equation}
		\begin{aligned}
			d(P(Y|G),P_f(Y|\widetilde{G}))-d(P_f(Y|G),P(Y|G)) & \leq d(P_f(Y|G),P_f(Y|\widetilde{G}))
		\end{aligned}
	\end{equation}
	As \xgnn $f$ that $\epsilon$-approximates \smt,
	we have the following by definition:
	\[ d(P_f(Y|\widetilde{G}),P(Y|\widetilde{G}))\leq \epsilon, d(P_f(Y|G),P(Y|G))\leq \epsilon.\]
	Then, call the triangle inequality again, we have
	\begin{equation}
		\begin{aligned}
			d(P(Y|G),P(Y|\widetilde{G}))                                            & \leq d(P_f(Y|\widetilde{G}),P(Y|G))+d(P_f(Y|\widetilde{G}),P(Y|\widetilde{G})) \\
			d(P(Y|G),P(Y|\widetilde{G}))-d(P_f(Y|\widetilde{G}),P(Y|\widetilde{G})) & \leq d(P_f(Y|\widetilde{G}),P(Y|G))                                            \\
			\delta-d(P_f(Y|\widetilde{G}),P(Y|\widetilde{G}))                       & \leq d(P_f(Y|\widetilde{G}),P(Y|G))                                            \\
			\delta-\epsilon                                                         & \leq d(P_f(Y|\widetilde{G}),P(Y|G)).                                           \\
		\end{aligned}
	\end{equation}
	Combining the aforementioned three inequalities, we have
	\[d(P_f(Y|\widetilde{G}),P(Y|G))-d(P_f(Y|G),P(Y|G))\geq \delta-2\epsilon,\]
	Then, it suffices to know that
	\begin{equation}
		\begin{aligned}
			\delta-2\epsilon & \leq d(P_f(Y|G),P_f(Y|\widetilde{G})).
		\end{aligned}
	\end{equation}
\end{proof}

\subsection{Proof for Theorem~\ref{CH:GMT:thm:gmt_success}}
\label{proof:gmt_success}
\begin{theorem}\label{CH:GMT:thm:gmt_success_appdx}
	Given the attention matrix $\widehat{A}$,
	and the distribution distance metric $d$ as the total variation distance,
	let $C=|\gY|$,
	for a \gmts with $t$ i.i.d. samples of
	$G_c^i\sim P(G_c|G)$, then,
	there exists $\epsilon\in\R_+$ such that,
	with a probability at least $1-e^{-t\epsilon^2/4}$, \gmts $\frac{\epsilon C}{2}$-approximates \smt (Def.~\ref{def:submt_approx_appdx}) and satisfies $(\delta,1-\frac{\epsilon C}{\delta})$ counterfactual fidelity (Def.~\ref{def:counterfactual_x_appdx}).
\end{theorem}
\begin{proof}
	Recall the \smt objective:
	\[F_c(\boldsymbol{\alpha}; G) :=\sum_{G_c\in G}f_c(G_c)\prod_{e\in G_c}\alpha_e\prod_{e\in G/G_c}(1-\alpha_e),\]
	which is the expanded form of  $\E[f_c(G_c)]$, $G_c\sim P(G_c|\widehat{A})$.
	Then, denote $M=\max|f_c(G_c)|$, $f_c(G_c)$ can be considered as a random variable within the range of $[-M,M]$.
	Considering $t$ random i.i.d. examples of $\{G_c^i\}_{i=1}^t$ drawn from $P(G_c|\widehat{A})$,
	and the predicted probability for each class, denoted as $Y_i=\frac{1}{M}f_c(G_c^i)$,
	we then have $Y_i\in[-1,1]$ and $\sum_{i=1}^t\E[Y_i]=\frac{t}{M}F(\boldsymbol{\alpha};G)$.
	It allows us to adopt the Markov's inequality and obtain the following Chernoff bound:
	\[
		\text{Pr}(|\sum_{i=1}^tY_i-\frac{t}{M}F(\boldsymbol{\alpha};G)>t\epsilon|)<e^{-t^2\epsilon^2/4t}=e^{-t\epsilon^2/4}.
	\]
	Since by definition of \gmts, i.e.,
	\[f_c^s(\widehat{G}_c)=\frac{1}{t}\sum_{i=1}^tf_c(Y|G_c^i),\]
	we have \[\sum_{i=1}^tY_i=\frac{t}{M}\sum_{i=1}^tf_c(G_c^i)=\frac{t}{M}f_c^s(\widehat{G}_c),\]
	the bound can be written as:
	\begin{equation}
		\begin{aligned}
			\text{Pr}(|\frac{t}{M}f_c^s(\widehat{G}_c)-\frac{t}{M}F(\boldsymbol{\alpha};G)>t\epsilon|) & <e^{-t^2\epsilon^2/4t}=e^{-t\epsilon^2/4} \\
			\text{Pr}(|f_c^s(\widehat{G}_c)-F(\boldsymbol{\alpha};G)>\epsilon M|)                      & <e^{-t\epsilon^2/4}                       \\
			\text{Pr}(|f_c^s(\widehat{G}_c)-F(\boldsymbol{\alpha};G)\leq\epsilon M|)                   & \geq 1-e^{-t\epsilon^2/4}.
		\end{aligned}
	\end{equation}
	In other words, with a probability at least $1-e^{-t\epsilon^2/4}$, we have the following holds:
	\begin{equation}
		|f_c^s(\widehat{G}_c)-F_c(\boldsymbol{\alpha}; G)]|\leq \epsilon M.
	\end{equation}
	Since $M$ is defined as the maximal probability for each class,
	\[M= \max\E[f_cP(Y|G_c)],\]
	it suffices to know that $M\leq 1$.
	Therefore, it follows that
	\[|f_c^s(\widehat{G}_c)-F_c(\boldsymbol{\alpha}; G)]|\leq \epsilon,\]
	for each class, which further implies that
	\[|f_c^s(\widehat{G}_c)-F_c(\boldsymbol{\alpha}; G)]|\leq \epsilon |\gY|=\epsilon C,\]
	which commits to the $\frac{\epsilon C}{2}$ \smt approximation under the total variation distance.
	Then, using the results of Proposition~\ref{CH:GMT:thm:smt_fidelty}, we know \gmts also commit to the $1-\frac{\epsilon C}{\delta}$ counterfactual fidelity.
\end{proof}
\section{More Discussions on Practical Implementations of \gmt}
\label{CH:GMT:sec:gmt_impl_dis_appdx}
We provide more discussion complementary to the description of Sec.~\ref{CH:GMT:sec:gmt_sol} in the main text.

\subsection{Algorithms of \gmt}
\label{CH:GMT:sec:gmt_alg_appdx}
\paragraph{Training subgraph extractor with random subgraph sampling.}
We focus on discussing the implementation details of \gmts since \gmtl differs from \gsat only in the number of weighted message passing times.
\gmts contains two stages: i) subgraph extractor training, and ii) neural subgraph extension learning. The first stage aims to train the subgraph extractor to extract the desired subgraphs, while the second stage aims to reduce the additional computation overhead of the random subgraph sampling, and further better learn the correlations between the soft subgraphs and the labels. The algorithm for stage i) is given in Algorithm~\ref{alg:gmt_pretraining} and for stage ii) is given in Algorithm~\ref{alg:gmt_finetune}, respectively.

\begin{algorithm}[H]
    \caption{Subgraph extractor training algorithm of \oursfull (\textbf{\ourst}). }
    \label{alg:gmt_pretraining}
    \begin{algorithmic}[1]
        \STATE \textbf{Input:} Training data $\train$;
        a \xgnn $f$ with subgraph extractor $g$, and classifier $f_c$;
        subgraph sampling epochs $e_s$;
        length of maximum subgraph learning epochs $e_l$;
        batch size $b$;
        loss function $l(\cdot)$;
        subgraph regularization $o(\cdot)$;
        subgraph regularization weight $\gamma$;
        \STATE Randomly initialize $f$;
        \STATE \texttt{// Stage I:  subgraph learning.}
        \FOR{$j=1$ to $e_l$}
        \STATE Sample a batch of data $\{G^i,Y^i\}_{i=1}^b$ from $\train$;
        \STATE Obtain sampling attention $\{\boldsymbol{\alpha}^i\}_{i=1}^b$ via Eq.~\ref{CH:GMT:eq:subgraph_att};
        \STATE \texttt{// MCMC subgraph sampling.}
        \FOR{$k=1$ to $e_s$}
        \STATE Obtain the sampling probability $\{\boldsymbol{\beta}^i\}_{i=1}^b$ via Eq.~\ref{CH:GMT:eq:gumbel_appdx} using Gumbel-softmax;
        \STATE Randomly sample subgraphs $\{G_c^i\sim\text{Ber}(\boldsymbol{\beta}^i)\}_{i=1}^b$ via Eq.~\ref{CH:GMT:eq:subgraph_sampling_appdx};
        \STATE Obtain predictions as logits $\{\hat{y}^i_k\}_{i=1}^b$;
        \ENDFOR
        \STATE Obtain simulated prediction $\{\hat{y}^i=\frac{1}{e_s}\sum_{k=1}^{e_s}\hat{y}_k^i\}_{i=1}^b$;
        \STATE Obtain prediction loss $l_p$ with $l(\cdot)$ and $\{\hat{y}^i\}_{i=1}^b$;
        \STATE Obtain subgraph regularization loss $l_o$ with $o(\cdot)$ and  $\{\boldsymbol{\alpha}^i\}_{i=1}^b$;
        \STATE Obtain the final loss $l_f = l_p+\eta \cdot l_o$;
        \STATE Updated model via backpropagation with $l_f$;
        \ENDFOR
        \STATE \textbf{Return} trained subgraph extraction model $f_c\circ g$;
    \end{algorithmic}
\end{algorithm}

For each input graph along with the label $(G,Y)$, the subgraph extractor $g$ first propagates among $G$ and obtains the node representations $H_i\in\R^h$ for each node.
Then, the (edge-centric) sampling attention is obtained as the following
\begin{equation}\label{CH:GMT:eq:subgraph_att}
	\alpha_e=a([H_u,H_v]),
\end{equation}
for each edge $e=(u,v)\in E$, where $a(\cdot)$ is the attention function and can be simply implemented as an MLP. Note that $\alpha_e$ is slightly different from that in the main text, since we will discuss in detail the discrete sampling process in the implementation.

To enable the gradient backpropagation along with the discrete sampling of subgraphs, we will adopt the Gumbel-softmax trick and straight-through estimator~\citep{gumbel,gumbel2}. With the attention from Eq.~\ref{CH:GMT:eq:subgraph_att}, the sampling probability $\boldsymbol{\beta}$ is then obtained as follows
\begin{equation}\label{CH:GMT:eq:gumbel_appdx}
	\beta_e=\sigma((\alpha_e+D)/\tau),
\end{equation}
where $\tau$ is the temperature, $\sigma$ is the sigmoid function, and
\[
	D=\log U-\log(1-U),
\]
with $U\sim\text{Uniform}(0,1)$. To sample the discrete subgraph, we sample from the Bernoulli distributions on edges independently
\[A_e\sim \text{Bern}(\beta_e)\]
and obtain the discrete subgraph with each entry as
\begin{equation}\label{CH:GMT:eq:subgraph_sampling_appdx}
	A_e = \text{StopGrad}(A_e-\alpha_e)+\alpha_e,
\end{equation}
which allows computing the gradients along with the subgraph sampling probability. Although the trick works empirically well, the estimated gradients are approximated ones that have biases from the ground truth.
It might be of independent interest to analyze whether the random subgraph sampling in \gmts can also reduce the gradient estimator biases during discrete sampling.

\begin{algorithm}[H]
    \caption{Subgraph classifier training algorithm of \oursfull (\textbf{\ourst}). }
    \label{alg:gmt_finetune}
    \begin{algorithmic}[1]
        \STATE \textbf{Input:} Training data $\train$;
        trained \xgnn $f$ with subgraph extractor $g$, and classifier $f_c$ by Alg.~\ref{alg:gmt_pretraining};
        length of maximum subgraph classifier training epochs $e_l$;
        batch size $b$;
        loss function $l(\cdot)$;
        subgraph regularization $o(\cdot)$;
        subgraph regularization weight $\gamma$;
        \STATE Initialize $f_c$;  Keep $g$ frozen;
        \STATE \texttt{// Stage II:  subgraph classifier learning.}
        \FOR{$j=1$ to $e_l$}
        \STATE Sample a batch of data $\{G^i,Y^i\}_{i=1}^b$ from $\train$;
        \STATE Obtain sampling attention $\{\boldsymbol{\alpha}^i\}_{i=1}^b$ via Eq.~\ref{CH:GMT:eq:subgraph_att};
        \STATE \texttt{// Soft subgraph propagation.}
        \STATE Obtain edge sampling probability $\{\boldsymbol{\beta}^i=\text{StopGrad}(\boldsymbol{\alpha}^i)\}_{i=1}^b$; \texttt{//  subgraph extractor frozen}
        \STATE Obtain prediction with subgraph $\{\hat{y}^i\}_{i=1}^b$ via weighted message passing with $\{\boldsymbol{\beta}^i\}_{i=1}^b$;
        \STATE Obtain prediction loss $l_p$ with $l(\cdot)$ and $\{\hat{y}^i\}_{i=1}^b$;
        \STATE Obtain final loss $l_f = l_p$;
        \STATE Updated model via backpropagation with $l_f$;
        \ENDFOR
        \STATE \textbf{Return} final model $f_c\circ g$;
    \end{algorithmic}
\end{algorithm}

\paragraph{Learning neural subgraph multilinear extension.}
When the subgraph extractor is trained, we then enter into stage two, which focuses on extracting the learned subgraph information for better predicting the label with a single pass forward.
More concretely, although \ours trained with \gmts improves interpretability, \gmts still requires multiple random subgraph sampling to approximate \smt and costs much additional overhead.
To this end, we propose to learn a neural \smt that only requires a single sampling, based on the trained subgraph extractor $g$ with \gmts.

Learning the neural \smt is essentially to approximate the MCMC with a neural network, though it is inherently challenging to approximate MCMC~\citep{no_free_lunch_MCMC,papamarkou2022a}.
Nevertheless, the feasibility of neural \smt learning is backed by the inherent causal subgraph assumption of~\citep{ciga}, once the causal subgraph is correctly identified, simply learning the statistical correlation between the subgraph and the label is sufficient to recover the causal relation.

Therefore, we propose to simply re-train a new classifier GNN with the frozen subgraph extractor, to distill the knowledge contained in $\widehat{G}_c$ about $Y$.
The implementation is simply to stop the gradients of the subgraph extractor, while only optimizing the classifier GNN with the predicted sampling probability. Note that it breaks the shared encoder structure of the \gsat, which could avoid potential representation conflicts for a graph encoder shared by both the subgraph extractor and the classifier.
Under this consideration, we also enable the BatchNorm~\citep{batch_norm} in the subgraph extractor to keep count of the running stats when training the new classifier.

Empirically, the weighted message passing can effectively capture the desired information from $g$ and lead to a performance increase.
This scheme also brings additional benefits over the originally trained classifier, which focuses on providing the gradient guidance for finding proper $G_c$ instead of learning all the available statistical correlations between $G_c$ and $Y$.

\subsection{Discussions on  \gmt Implementations}
\label{CH:GMT:sec:gmt_impl_appdx}
With the overall algorithm training the subgraph extractor and the classifier, we then discuss in more detail the specific implementation choices of \gmts.

\paragraph{Transforming node-centric random subgraph sampling}
In the task of geometric learning, the input graphs are initially represented as point clouds. The graph structures are built upon the node features and geometric knowledge. Therefore, \lri adopts the node-centric sampling and learns sampling probabilities for nodes when implementing the graph information bottleneck.
However, when sampling concrete subgraphs from a node-centric view, it will often lead to a too-aggressive sampling. Otherwise, one has to increase the sampling probability $r$ of the variational distribution $Q(G_c)$ in Eq.~\ref{CH:GMT:eq:gib_reg_appdx}.
To this end, we can transform the node-centric sampling to edge-centric sampling. Let $\alpha_i$ denote the sampling probability for node $i$, then the edge sampling probabilities can be obtained via:
\begin{align}
	\beta_e = \alpha_u \cdot \alpha_v,
\end{align}
for each edge $e=(u,v)\in E$. It thus enables the subgraph sampling from the node-centric view. Empirically, in geometric datasets, we observe a lower variance when transforming the node-centric sampling to edge-centric sampling.

\paragraph{Warmup of \gmts.}
Although more sampling rounds can improve the approximation precision of \gmts to \smt, it would also affect the optimization of the interpretable subgraph learning, in addition to the additional unnecessary computational overhead. For example, at the beginning of the interpretable subgraph learning, the subgraph extractor will yield random probabilities like $0.5$.
\begin{itemize}
	\item First, a more accurate estimation based on random \smt is unnecessary.
	\item Second, at such random probabilities, every subgraph gets a nearly equal chance of being sampled, and gets gradients backpropagated. Since neural networks are universal approximators, the whole network can easily be misled by the noises, which will slow down the learning speed of the meaningful subgraphs.
	\item Third, when spurious correlations exist between subgraphs and the labels, the learning process will be more easily misled by the potential spurious correlations at the beginning of the subgraph learning.
\end{itemize}
More importantly, sampling multiple times can lead to trivial solutions with degenerated performance in the \gsat objective. Specifically, the formulation of the mutual information regularizer in \gsat has a trivial solution where all $\alpha_e$ directly collapses to the given $r$. More formally, let $\alpha_e=r$ in the following objective obviously lead to zero loss that appears to be a Pareto optimal solution~\citep{pair} that can be selected as the output:
\[
	D_\text{KL}(\text{Bern}(\alpha_e)||\text{Bern}(r))=\sum_{e}\alpha_e\log\frac{r}{r}+(1-r)\log\frac{(1-\alpha_e)}{(1-r)}=0.
\]
The trivial solutions can occur to \ours more easily with more rounds of subgraph sampling, especially in too simple or too complicated tasks.

To tackle the above problem, we propose two warmup strategies:
\begin{itemize}
	\item Larger initial prior $r$ of $Q(G_c)$ in Eq.~\ref{CH:GMT:eq:gib_reg_appdx}: \gsat achieves the objective of graph information bottleneck with a schedule of $r$ in $Q(G_c)$ as $0.9$, which could promote the random sampling probabilities to meaningful subgraph signals. As the random subgraph sampling will slow the optimization, we can warm up the initial subgraph learning with a larger initial $r$. In experiments, we try with $r=1.0$ and $r=0.9$, and find $r=0.1$ can effectively warm up and speed up the subgraph learning, which is especially meaningful for too simple tasks where \xgnns can easily overfit to, or too hard tasks where \xgnns learns the meaningful subgraph signals in a quite slow speed. We can also use a larger regularization penalty at the initial stage to speed up meaningful subgraph learning.
	\item Single subgraph sampling: As sampling too many subgraphs can bring many drawbacks such as overfitting and slow learning, we propose warm up the initial subgraph learning with a single sampling during the first stage of $r$ (i.e., when $r$ still equals to the initial $r$ in the schedule of \gsat). The single subgraph sampling also implicitly promotes meaningful subgraph learning, as it encourages a higher chance even for a small difference in the sampling probability.
\end{itemize}

In addition to helping with the warmup of the interpretable subgraph, single subgraph sampling also has some additional benefits and effectively tackles the trivial solution of \gsat objective. It also brings more variance between meaningful subgraph learning and noisy subgraph learning, and we find using a single random subgraph learning is extremely helpful for simple tasks such as BA\_2motifs in our experiments. The implicit variance of single random subgraph sampling also brings additional benefits to maintaining high variance between the signal subgraph and noisy subgraph, which might be of independent interest.
It turns out the variance in single subgraph learning can have an implicit regularization preventing the trivial solution.

In experiments, we will use all of the warmup strategies together (i.e., a larger initial $r$, a larger penalty score, and single subgraph sampling) when we observe a performance degeneration in the validation set. Otherwise, we will stick to the original receipt. More details are given in Sec.~\ref{CH:GMT:sec:eval_appdx}.

\paragraph{Single weighted message passing in \gmtl.}
Although it has been shown that propagation with the attention only once can effectively reduce the \smt approximation error, it remains unknown which layer the attention should be applied.
Empirically, we examine the following three strategies:
\begin{itemize}
	\item Weighted message passing on the first layer;
	\item Weighted message passing on the last layer;
	\item Single weighted message passing of all layers, and then average the logits;
\end{itemize}
We find applying weighted message passing to the first layer outperforms the other two strategies in experiments, and thus we stick to the first layer weighted message passing scheme. Exploring the reasons behind the intriguing phenomenon will be an interesting future extension.

\paragraph{Subgraph sampling for neural \smt.}
Although the weighted message passing with $\boldsymbol{\alpha}$ produced by the trained subgraph extractor already achieves better performance, it may not maximally extract the full underlying information of the learned subgraph and the labels, since the original function is a MCMC that is not easy to be fitted~\citep{no_free_lunch_MCMC}. Besides, the weighted message passing itself may not be expressive enough due to the expressivity constraints of GNNs~\citep{gin}, and also the limitations of the attention-based GNNs~\citep{gatv2,aero_gnn}.

Therefore, we propose more subgraph sampling strategies along with alternative architecture of the new classifier, in order to best fit the underlying MCMC function.
Specifically, we consider the following aspects:
\begin{itemize}
	\item Initialization: the graph encoder of the new classifier can be initialized from scratch and avoids overfitting, or initialized from the random subgraph sampling trained models;
	\item Architecture: weighted message passing, or single weighted message passing as that of \gmtl;
	\item Attention sampling: set the minimum $p\%$ attention scores directly to $0$; set the maximum $p\%$ attention scores directly to $1$; set the maximum $p\%$ attention scores directly to $1$ while set the minimum $(1-p)\%$ attention scores directly to $0$;
\end{itemize}
We examine the aforementioned strategies and choose the one according to the validation performance in experiments. We exhibit the detailed hyperparameter setup in Appendix~\ref{CH:GMT:sec:eval_appdx}.

\section{More Details about the Experiments}
\label{CH:GMT:sec:exp_appdx}
In this section, we provide more details about the experiments, including the dataset preparation, baseline implementations, models and hyperparameters selection as well as the evaluation protocols.

\bgroup
\def\arraystretch{1}
\begin{table}[ht]
	\centering
	\caption{Information about the datasets used in experiments. The number of nodes and edges are respectively taking average among all graphs.}\small\sc
	\label{CH:GMT:tab:datasets_stats_appdx}
	\resizebox{\textwidth}{!}{
		\begin{small}
			\begin{tabular}{l|ccccccc}
				\toprule
				\textbf{Datasets}      & \textbf{\# Training} & \textbf{\# Validation} & \textbf{\# Testing} & \textbf{\# Classes} & \textbf{ \# Nodes} & \textbf{ \# Edges}
				                       & \textbf{  Metrics}                                                                                                                        \\\midrule
				BA-2Motifs             & $800$                & $100$                  & $100$               & $2$                 & $25$               & $50.96$            & ACC \\
				Mutag                  & $2,360$              & $591$                  & $1,015$             & $2$                 & $30.13$            & $60.91$            & ACC \\
				Suprious-Motif $b=0.5$ & $9,000$              & $3,000$                & $6,000$             & $3$                 & $45.05$            & $65.72$            & ACC \\
				Suprious-Motif $b=0.7$ & $9,000$              & $3,000$                & $6,000$             & $3$                 & $46.36$            & $67.10$            & ACC \\
				Suprious-Motif $b=0.9$ & $9,000$              & $3,000$                & $6,000$             & $3$                 & $46.58$            & $67.59$            & ACC \\
				MNIST-75sp             & $20,000$             & $5,000$                & $10,000$            & $10$                & $70.57$            & $590.52$           & ACC \\
				Graph-SST2             & $28,327$             & $3,147 $               & $12,305$            & $2$                 & $10.20$            & $18.40$            & ACC \\
				OGBG-MolHiv            & $32,901 $            & $4,113 $               & $4,113$             & $2$                 & $25.51$            & $54.94$            & AUC \\
				\bottomrule
			\end{tabular}	\end{small}}
\end{table}
\egroup

\begin{table}[t]
	\caption{Statistics of the four geometric datasets from~\citet{lri}.}
	\label{CH:GMT:tab:lri_stat}
	\centering
	\resizebox{\textwidth}{!}{%
		\begin{tabular}{lccccccccc}
			\toprule
			                  & \# Classes & \# Features in $\mathbf{X}$ & \# Dimensions in $\mathbf{r}$ & \# Samples & Avg. \# Points/Sample & Avg. \# Important Points/Sample & Class Ratio & Split Scheme & Split Ratio \\
			\midrule
			\texttt{\acts}    & 2          & 0                           & 3                             & 3241       & 109.1                 & 22.8                            & 39/61       & Random       & 70/15/15    \\
			\texttt{\taumu}   & 2          & 1                           & 2                             & 129687     & 16.9                  & 5.5                             & 24/76       & Random       & 70/15/15    \\
			\texttt{\synbind} & 2          & 1                           & 3                             & 8663       & 21.9                  & 6.6                             & 18/82       & Patterns     & 78/11/11    \\
			\texttt{\pdbbind} & 2          & 3                           & 3                             & 10891      & 339.8                 & 132.2                           & 29/71       & Time         & 92/6/2      \\
			\bottomrule
		\end{tabular}%
	}
\end{table}

\subsection{Datasets}
\label{CH:GMT:sec:dataset_appdx}

We provide more details about the motivation and construction method of the datasets that are used in our experiments. Statistics of the regular graph datasets are presented in Table~\ref{CH:GMT:tab:datasets_stats_appdx}, and statistics of the geometric graph datasets are presented in Table~\ref{CH:GMT:tab:lri_stat}.

\paragraph{BA-2Motifs}~\citep{pge} is a synthetic dataset that adopts the Barab\'asi-Albert (BA) graph data model to generate subgraphs in specific shapes. Each graph contains a motif subgraph that is either a five-node cycle or a house. The class labels are determined by the motif, and the motif itself serves as the interpretation of ground truth. The motif is then attached to a large base graph.

\paragraph{Mutag}~\citep{mutag} is a typical molecular property prediction dataset. The nodes represent atoms and the edges represent chemical bonds. The label of each graph is binary and is determined based on its mutagenic effect. Following~\citet{pge,gsat}, -NO2 and -NH2 in mutagen graphs are labeled as ground-truth explanations.

\paragraph{MNIST-sp}~\citep{understand_att} is a graph dataset converted from MNIST dataset via superpixel transformation. The nodes of MNIST-75sp graphs are the superpixels and the edges are generated according to the spatial distance of nodes in the original image. The ground truth explanations of MNIST-75sp are simply the non-zero pixels. As the original digits are hand-written, the interpretation subgraphs could be in varying sizes.

\paragraph{Suprious-Motif datasets}~\citep{dir} is a 3-class synthetic datasets based on BA-2Motifs~\citep{gnn_explainer,pge} with structural distribution shifts.
The model needs to tell which one of three motifs (House, Cycle, Crane) the graph contains.
For each dataset, $3000$ graphs are generated for each class at the training set, $1000$ graphs for each class at the validation set and testing set, respectively.
During the construction of the training data, the motif and one of the three base graphs (Tree, Ladder, Wheel) are artificially (spuriously) correlated with a probability of various biases, and equally correlated with the other two. Specifically, given a predefined bias $b$, the probability of a specific motif (e.g., House) and a specific base graph (Tree) will co-occur is $b$ while for the others is $(1-b)/2$ (e.g., House-Ladder, House-Wheel).
The test data does not have spurious correlations with the base graphs, however, test data will use larger base graphs that contain graph size distribution shifts.
Following~\citet{gsat}, we select datasets with a bias of $b=0.5$, $b=0.7$, and $b=0.9$.  The interpretation ground truth is therefore the motif itself.

\paragraph{Graph-SST2}~\citep{sst25,xgnn_tax} is converted from a sentiment analysis dataset in texts. Each sentence in SST2 will be converted to a graph. In the converted graph, the nodes are the words and the edges are the relations between different words. Bode features are generated using BERT~\citep{bert} and the edges are parsed by a Biaffine parser~\citep{biaffine}.
Following previous works~\citep{dir,gsat,ciga}, our splits are created according to the averaged degrees of each graph.
Specifically, we assign the graphs as follows: Those that have smaller or equal to $50$-th percentile averaged degree are assigned to training, those that have averaged degree larger than $50$-th percentile while smaller than $80$-th percentile are assigned to the validation set, and the left are assigned to test set.
Since the original dataset does not have the ground truth interpretations, we report only the classification results.

\paragraph{OGBG-Molhiv}~\citep{ogb} is also a molecular property prediction dataset. The nodes represent atoms and the edges represent chemical bonds. The label of each graph is binary and is determined based on whether a molecule inhibits HIV virus replication or not. The training, validation and test splits are constructed according to the scaffolds~\citep{ogb} hence there also exist distribution shifts across different splits. Since the original dataset does not have the ground truth interpretations, we report only the classification results.

In what follows we continue to introduce the four geometric learning datasets. We refer interested readers to~\citet{lri} for more details.

\paragraph{ActsTrack} dataset~\citep{lri}:
\begin{itemize}[leftmargin=*]
	\item Background: \textbf{ActsTrack} involves a fundamental resource in High Energy Physics (HEP), employed for the purpose of reconstructing various properties, including the kinematics, of charged particles based on a series of positional measurements obtained from a tracking detector. Within the realm of HEP experimental data analysis, particle tracking is an essential procedure, and it also finds application in medical contexts, such as proton therapy~\citep{act}. \text{ActsTrack} is formulated differently by~\citet{lri} from traditional track reconstruction tasks: It requires predicting the existence of a $z\rightarrow \mu\mu$ decay and using the set of points from the $\mu$'s to verify model interpretation, which can be used to reconstruct $\mu$ tracks.
	\item Construction: In the \textbf{ActsTrack} dataset, each data point corresponds to a detector hit left by a particle, and it is associated with a 3D coordinate. Notably, the data points in ActsTrack lack any features in the X dimension, necessitating the use of a placeholder feature with all values set to one during model training. Additionally, the dataset provides information about the momenta of particles as measured by the detectors, which has the potential to be employed for assessing fine-grained geometric patterns in the data; however, it is not utilized as part of the model training process. Given that, on average, each particle generates approximately 12 hits, and a model can perform well by capturing the trajectory of any one of the $\mu$ (muon) particles resulting from the decay, we report performance metrics in precision@12 following~\citet{lri}. The dataset was randomly split into training, validation, and test sets, maintaining a distribution ratio of 70\% for training, 15\% for validation, and 15\% for testing.
\end{itemize}

\paragraph{Tau3Mu} dataset~\citep{lri}:
\begin{itemize}[leftmargin=*]
	\item Background: \textbf{Tau3Mu} involves another application in High Energy Physics (HEP) dedicated to identifying a particularly challenging signature - charged lepton flavor-violating decays, specifically $\tau\rightarrow\mu\mu\mu$ decay.
	      This task involves the analysis of simulated muon detector hits resulting from proton-proton collisions. It's worth noting that such decays are heavily suppressed within the framework of the Standard Model (SM) of particle physics~\citep{tau3mu}, making their detection a strong indicator of physics phenomena beyond the Standard Model~\citep{tau3mu_discovery}.
	      Unfortunately, $\tau\rightarrow\mu\mu\mu$ decay involves particles with extremely low momentum, rendering them technically impossible to trigger using conventional human-engineered algorithms. Consequently, the online detection of these decays necessitates the utilization of advanced models that explore the correlations between signal hits and background hits, particularly in the context of the Large Hadron Collider. Our specific objective is twofold: to predict the occurrence of $\tau\rightarrow\mu\mu\mu$ decay and to employ the detector hits generated by the $\mu$ (muon) particles to validate the model's interpretations.

	\item Construction: \textbf{Tau3Mu} uses the data simulated via the PYTHIA generator~\citep{Bierlich2022ACG}.

	      The interpretation labels are using the signal sample with the background samples on a per-event basis (per point cloud) while preserving the ground-truth labels. The hits originating from $\mu$ (muon) particles resulting from the $\tau\rightarrow\mu\mu\mu$ decay are designated as ground-truth interpretation. The training data only include hits from the initial layer of detectors, ensuring that each sample in the dataset contains a minimum of three detector hits. Each data point in the samples comprises measurements of a local bending angle and a 2D coordinate within the pseudorapidity-azimuth ($\eta-\phi$) space.

	      Given that, in the most favorable scenario, the model is required to capture hits from each $\mu$ particle, we report precision@3 following~\citet{lri}.  Lastly, the dataset is randomly split into training, validation, and test sets, maintaining a distribution ratio of 70\% for training, 15\% for validation, and 15\% for testing.

\end{itemize}

\paragraph{SynMol} dataset~\citep{lri}:
\begin{itemize}[leftmargin=*]
	\item Background: \textbf{SynMol} is a molecular property prediction task. While prior research efforts have explored model interpretability within this domain~\citep{McCloskey2018UsingAT}, their emphasis has been primarily on examining chemical bond graph representations of molecules, often overlooking the consideration of geometric attributes. In our present study, we shift our attention towards 3D representations of molecules. Our specific objective is to predict a property associated with two functional groups, namely carbonyl and unbranched alkane (as defined by~\citet{McCloskey2018UsingAT}), and subsequently employ the atoms within these functional groups to validate our model's interpretations.
	\item Construction: \textbf{SynMol} is constructed based on ZINC~\citep{zin} following~\citet{McCloskey2018UsingAT} that creates synthetic properties based on the existence of certain functional groups. The labeling criteria involve classifying a molecule as a positive sample if it contains both an unbranched alkane and a carbonyl group. Conversely, molecules lacking this combination are categorized as negative samples. Consequently, the atoms within branched alkanes and carbonyl groups serve as the designated ground-truth interpretation.

	      In addition to specifying a 3D coordinate, each data point within a sample is also associated with a categorical feature signifying the type of atom it represents. While the combined total of atoms in the two functional groups may be limited to just five, it is important to note that certain molecules may contain multiple instances of such functional groups. Consequently, we report precision metric at precision@5 following~\citet{lri}.

	      Finally, to mitigate dataset bias, the dataset is split into training, validation, and test sets using a distribution strategy following~\citet{McCloskey2018UsingAT,lri}. This approach ensures a uniform distribution of molecules containing or lacking either of these functional groups.
\end{itemize}

\paragraph{PLBind} dataset~\citep{lri}:
\begin{itemize}[leftmargin=*]
	\item Background: \textbf{PLBind} is to predict protein-ligand binding affinities leveraging the 3D structural information of both proteins and ligands. This task holds paramount significance in the field of drug discovery, as a high affinity between a protein and a ligand is a critical criterion in the drug selection process~\citep{Wang2017ImprovingSP,Karimi2018DeepaffinityID}. The accurate prediction of these affinities using interpretable models serves as a valuable resource for rational drug design and contributes to a deeper comprehension of the underlying biophysical mechanisms governing protein-ligand binding~\citep{Du2016InsightsIP}. Our specific mission is to forecast whether the affinity surpasses a predefined threshold, and we achieve this by examining the amino acids situated within the binding site of the test protein to corroborate our model's interpretations.
	\item Construction: \textbf{PLBind} is constructed protein-ligand complexes from PDBind~\citep{PDBind}. PDBind annotates binding affinities for a subset of complexes in the Protein Data Bank (PDB)~\citep{PDB}, therefore, a threshold on the binding affinity between a pair of protein and ligand can be used to construct a binary classification task. The ground-truth interpretation is then the part of the protein that are within 15A of the ligand to be the binding site~\citep{Liu2022Generating3M}. Besides, PLBind also includes all atomic contacts (hydrogen bond and hydrophobic contact) for every protein-ligand pair from PDBsum~\citep{Laskowski2001PDBsumSS}, where the ground-truth interpretations are the corresponding amino acids in the protein.

	      Every amino acid in a protein is linked to a 3D coordinate, an amino acid type designation, the solvent-accessible surface area (SASA), and the B-factor. Likewise, each atom within a ligand is associated with a 3D coordinate, an atom type classification, and Gasteiger charges. The whole dataset is then partitioned into training, validation, and test sets, adopting a division based on the year of discovery for the complexes, following~\citet{Strk2022EquiBindGD}.
\end{itemize}

\subsection{Baselines and Evaluation Setup}
\label{CH:GMT:sec:eval_appdx}

During the experiments, we do not tune the hyperparameters exhaustively while following the common recipes for optimizing GNNs, and also the recommendation setups by previous works.
Details are as follows.

\textbf{GNN encoder.} For fair comparison, we use the same GNN architecture as graph encoders for all methods, following~\citet{gsat,lri}.
For the backbone of GIN, we use $2$-layer GIN~\citep{gin} with Batch Normalization~\citep{batch_norm} between layers, a hidden dimension of $64$ and a dropout ratio of $0.3$.
For the backbone of PNA, we use $4$-layer PNA~\citep{pna} with Batch Normalization~\citep{batch_norm} between layers, a hidden dimension of $80$ and a dropout ratio of $0.3$. The PNA network does not use scalars, while using (mean, min, max, std aggregators.
For the backbone of EGNN~\citep{egnn}, we use $4$-layer EGNN with Batch Normalization~\citep{batch_norm} between layers, a hidden dimension of $64$ and a dropout ratio of $0.2$. The pooling functions are all sum pooling.

\textbf{Dataset Splits.} We follow previous works~\citep{pge,gsat} to split BA-2Motifs randomly into three sets as (80\%/10\%/10\%), Mutag randomly into 80\%/20\% as train and validation sets where the test data are the mutagen molecules with -NO2 or -NH2. We use the default split for MNIST-75sp given by~\citep{understand_att} with a smaller sampling size following~\citep{gsat}. We use the default splits for Graph-SST2~\citep{xgnn_tax}, Spurious-Motifs~\citep{dir} and OGBG-Molhiv~\citep{ogb} datasets.
For geometric datasets, we use the author provided default splits.

\textbf{Baseline implementations.} We use the author provided codes to implement the baselines \gsat~\citep{gsat}\footnote{\url{https://github.com/Graph-COM/GSAT}} and \lri~\citep{lri}\footnote{\url{https://github.com/Graph-COM/LRI}}.
We re-run \gsat and \lri under the same environment using the author-recommended hyperparameters for a fair comparison.
Specifically, BA-2Motif, Mutag and PLBind use $r = 0.5$, and all other datasets use $r = 0.7$. The $\lambda$ of information regularizer is set to be $1$ for regular graphs, $0.01$ for \taumu, and $0.1$ for \acts, \synbind and \pdbbind as recommended by the authors. $r$ will initially be set to $0.9$ and gradually decay to the tuned value. We adopt a step decay, where $r$ will decay $0.1$ for every $10$ epochs.
As for the implementation of explanation methods,
for regular graphs, we directly adopt the results reported.
For geometric graphs, we re-run the baselines to obtain the results, as previous results are obtained according to the best validation interpretation performance that may mismatch the practical scenario where the interpretation labels are usually not available.

\textbf{Optimization and model selection.}
Following previous works, by default, we use Adam optimizer~\citep{adam} with a learning rate of $1e-3$ and a batch size of $128$ for all models at all datasets, except for Spurious-Motif with GIN and PNA, Graph-SST2 with PNA that we will use a learning rate of $3e-3$.
When GIN is used as the backbone model, MNIST-75sp is trained for 200 epochs, and all other datasets are trained for 100 epochs, as we observe that 100 epochs are sufficient for convergence at OGBG-Molhiv.
When PNA is used, Mutag and Ba-2Motifs are trained for 50 epochs and all other datasets are trained for 200 epochs. We report the performance of the epoch that achieves the best validation prediction performance and use the models that achieve such best validation performance as the pre-trained models.
All datasets use a batch size of 128; except for MNIST-75sp with GIN, we use a batch size of 256 to speed up training due to its large size in the graph setting.

The final model is selected according to the best validation classification performance. We report the mean and standard deviation of $10$ runs with random seeds from $0$ to $9$.

\textbf{Implementations of \gmt.}
For a fair comparison, \ours uses the same GNN architecture for GNN encoders as the baseline methods.
We search for the hyperparameters of $r$ from $[r_0-0.1,r_0,r_0+0.1]$ according to the default $r_0$ given by~\citet{gsat,lri}. We search the weights of graph information regularizers from $[0.1,0.5,1,2]$ for regular graphs and from $[0.01,0.1,1]$ for geometric datasets.
To avoid trivial solutions of the subgraph extractor at the early stage, we search for warm-up strategies mentioned in Appendix~\ref{CH:GMT:sec:gmt_impl_appdx}. Besides, we also search for the decay epochs of the $r$ scheduler to avoid trivial solutions.
We search for the sampling rounds from $[1,20,40,80,100,200]$ when the memory allows.
In experiments, we find \ours already achieves the state-of-the-art results in most of the set-ups without the warm-up. Only in BA-2Motifs and MNIST-75sp with GIN, and in Tau3Mu with EGNN, \ours needs the warmups.

\begin{table}[H]
	\caption{Sensitivity to different subgraph decoding strategies.}
	\label{CH:GMT:tab:decoding_strategy}
	\resizebox{\textwidth}{!}{
		\begin{tabular}{@{}lll|llllll@{}}
			\toprule
			               &              &           & Generalization             &                            &                            & Interpretation             &                            &                            \\
			Initialization & Architecture & Attention & spmotif-0.5                & spmotif-0.7                & spmotif-0.9                & spmotif-0.5                & spmotif-0.7                & spmotif-0.9                \\\midrule
			               &              & \gsat     & $47.45$\std{5.87}          & $43.57$\std{3.05}          & $45.39$\std{5.02}          & $74.49$\std{4.46}          & $72.95$\std{6.40}          & $65.25$\std{4.42}          \\
			new            & mul          & min0      & $\mathbf{60.09}$\std{5.57} & $54.34$\std{4.04}          & $\mathbf{55.83}$\std{5.68} & $85.50$\std{2.40}          & $\mathbf{84.67}$\std{2.38} & $73.49$\std{5.33}          \\
			old            & mul          & min0      & $58.83$\std{7.22}          & $\mathbf{55.04}$\std{4.73} & $55.77$\std{5.97}          & $\mathbf{85.52}$\std{2.41} & $84.65$\std{2.42}          & $73.49$\std{5.33}          \\
			new            & mul          & max1      & $44.49$\std{2.65}          & $49.77$\std{2.31}          & $50.22$\std{2.79}          & $85.50$\std{2.39}          & $84.66$\std{2.37}          & $73.50$\std{5.31}          \\
			old            & mul          & max1      & $45.91$\std{2.86}          & $49.11$\std{3.04}          & $50.30$\std{2.07}          & $85.49$\std{2.39}          & $84.64$\std{2.39}          & $73.50$\std{5.35}          \\
			old            & mul          & min0max1  & $51.21$\std{6.46}          & $50.91$\std{6.50}          & $53.13$\std{4.46}          & $\mathbf{85.52}$\std{2.41} & $84.66$\std{2.43}          & $73.49$\std{5.34}          \\
			new            & mul          & normal    & $47.69$\std{5.72}          & $44.12$\std{5.44}          & $40.69$\std{4.84}          & $84.69$\std{2.40}          & $80.08$\std{5.37}          & $73.48$\std{5.34}          \\
			old            & mul          & normal    & $45.36$\std{2.65}          & $44.25$\std{5.41}          & $43.43$\std{5.44}          & $83.52$\std{3.41}          & $80.07$\std{5.35}          & $73.49$\std{5.36}          \\
			new            & lin          & normal    & $43.54$\std{5.02}          & $47.59$\std{4.78}          & $46.53$\std{3.27}          & $85.47$\std{2.39}          & $80.07$\std{5.37}          & $\mathbf{73.52}$\std{5.34} \\
			old            & lin          & normal    & $46.18$\std{3.03}          & $46.42$\std{5.63}          & $49.00$\std{3.34}          & $83.51$\std{3.39}          & $80.09$\std{5.34}          & $73.46$\std{5.35}          \\\bottomrule
		\end{tabular}}
\end{table}
To better extract the subgraph information, we also search for subgraph sampling strategies mentioned in Appendix~\ref{CH:GMT:sec:gmt_impl_appdx}.
Note that the hyperparameter search and training of the classifier is independent of the hyperparameter search of the subgraph extractor. Once could select the best subgraph extractor and train the new classifier onto it. When training the classifier, we search for the following $9$ subgraph decoding strategies as shown in Table~\ref{CH:GMT:tab:decoding_strategy}.
Specifically,
\begin{itemize}[leftmargin=*]
	\item Initialization: "new" refers to that  the classifier is initialized from scratch; "old" refers to that the classifier is initialized from the subgraph extractor;
	\item Architecture: "mul" refers to the default message passing architecture; "lin" refers to the \gmtl architecture;
	\item Attention: "normal" refers to the default weighted message passing scheme; "min0" refers to setting the minimum $p\%$ attention scores directly to $0$; "max0" refers to setting the maximum $p\%$ attention scores directly to $1$; "min0max1" refers to setting the maximum $p\%$ attention scores directly to $1$ while set the minimum $(1-p)\%$ attention scores directly to $0$;
\end{itemize}
Table~\ref{CH:GMT:tab:decoding_strategy} demonstrates the generalization and interpretation performance of \gmts in spurious motif datasets~\citep{dir}, denoted as "spmotif" with different levels of spurious correlations. It can be found that \gmts is generically robust to the different choices of the decoding scheme and leads to improvements in terms of OOD generalizability and interpretability.

\subsection{More interpretation results}
\label{CH:GMT:sec:more_x_appdx}
We additionally conduct experiments with post-hoc explanation methods based on the PNA backbone.
Specifically, we selected two representative post-hoc methods GNNExplainer and PGExplainer, and one representative intrinsic interpretable baseline DIR.
The results are given in the table below. It can be found that most of the baselines still significantly underperform GSAT and GMT.
One exception is that DIR obtains highly competitive (though unstable) interpretation results in spurious motif datasets,
nevertheless, the generalization performance of DIR remains highly degenerated ($53.03$\std{8.05} on spmotif\_0.9).

\begin{table}[ht]\caption{More interpretation results of baselines using PNA}\centering\small
	\begin{tabular}{@{}lllllll@{}}
		\toprule
		       & BA\_2Motifs      & Mutag            & MNIST-75sp       & spmotif\_0.5    & spmotif\_0.7    & spmotif\_0.9    \\\midrule
		GNNExp & 54.14\std{3.30}  & 73.10\std{7.44}  & 53.91\std{2.67}  & 59.40\std{3.88} & 56.20\std{6.30} & 57.39\std{5.95} \\
		PGE    & 48.80\std{14.58} & 76.02\std{7.37}  & 56.61\std{3.38}  & 59.46\std{1.57} & 59.65\std{1.19} & 60.57\std{0.85} \\
		DIR    & 72.33\std{23.87} & 87.57\std{27.87} & 43.12\std{10.07} & 85.90\std{2.24} & 83.13\std{4.26} & 85.10\std{4.15} \\
		GSAT   & 89.35\std{5.41}  & 99.00\std{0.37}  & 85.72\std{1.10}  & 79.84\std{3.21} & 79.76\std{3.66} & 80.70\std{5.45} \\
		\gmtl  & 95.79\std{7.30}  & 99.58\std{0.17}  & 85.02\std{1.03}  & 80.19\std{2.22} & 84.74\std{1.82} & 85.08\std{3.85} \\
		\gmts  & 99.60\std{0.48}  & 99.89\std{0.05}  & 87.34\std{1.79}  & 88.27\std{1.71} & 86.58\std{1.89} & 85.26\std{1.92} \\\bottomrule
	\end{tabular}
\end{table}

\subsection{Computational analysis}
\label{CH:GMT:sec:comp_appdx}
We provide more discussion and analysis about the computational overhead required by \ours, when compared to \gsat.
As \gmtl differs only in the number of weighted message passing rounds from \gsat, and has the same number of total message passing rounds, hence \gmtl and \gsat have the same time complexity as $O(E)$ for each epoch, or for inference.
When comparing \gmts to \gmtl and \gsat,
During training, \gmts needs to process $k$ rounds of random subgraph sampling, resulting in $O(k|E|)$ time complexity;
During inference, \gmts with normal subgraph decoding methods requires the same complexity as \gmtl and \gsat, as $O(|E|)$. When with special decoding strategy such as setting part of the attention entries to $1$ or $0$,
\gmts additionally needs to sort the attention weights, and requires $O(|E|+|E|\log |E|)$ time complexity.

\begin{table}[ht]
	\centering\small
	\begin{tabular}{@{}llllll@{}}
		\toprule
		               & BA\_2Motifs    &                 & MNIST-75sp       &                  & ActsTrack      \\
		Training       & GIN            & PNA             & GIN              & PNA              & EGNN           \\  \midrule
		\gsat          & 0.70\std{0.12} & 1.00\std{0.13}  & 41.28\std{0.61}  & 80.98\std{10.5}5 & 3.57\std{1.41} \\
		\gmtl          & 0.68\std{0.12} & 1.02\std{0.15}  & 41.12\std{0.69}  & 81.11\std{10.4}4 & 3.69\std{0.93} \\
		\gmts          & 6.25\std{0.48} & 17.03\std{0.91} & 136.60\std{1.21} & 280.77\std{4.00} & 5.38\std{0.59} \\ \midrule
		Inference      &                &                 &                  &                  &                \\ \midrule
		\gsat          & 0.07\std{0.05} & 0.11\std{0.12}  & 18.69\std{0.35}  & 24.40\std{2.06}  & 0.84\std{0.38} \\
		\gmtl          & 0.08\std{0.07} & 0.07\std{0.01}  & 18.72\std{0.41}  & 23.81\std{1.89}  & 0.80\std{0.21} \\
		\gmts (normal) & 0.05\std{0.01} & 0.12\std{0.01}  & 18.72\std{0.35}  & 18.01\std{1.47}  & 0.50\std{0.13} \\
		\gmts (sort)   & 0.07\std{0.01} & 0.21\std{0.06}  & 19.07\std{0.55}  & 18.69\std{3.35}  & 0.54\std{0.10} \\ \bottomrule
	\end{tabular}
\end{table}

In the table above, we benchmarked the real training/inference time of \gsat, \gmtl and \gmts in different datasets,
where each entry demonstrates the time in seconds for one epoch.
We benchmark the latency of \gsat, \gmtl and \gmts based on GIN, PNA and EGNN on different scales of datasets.
The sampling rounds of \gmts are set to $20$ for PNA on MNIST-sp, $10$ for EGNN, and $100$ to other setups.
From the table, it can be found that, although \gmts takes longer time for training, but the absolute values are not high even for the largest dataset MNIST-sp.
As for inference, \gmts enjoys a similar latency as others, aligned with our discussion.

\subsection{More counterfactual fidelity studies}
\label{CH:GMT:sec:cf_viz_appdx}

To better understand the results, we provide more counterfactual fidelity results in supplementary to Sec.~\ref{CH:GMT:sec:expressivity_issue} and Fig.~\ref{CH:GMT:fig:cf_ba_appdx} and~\ref{CH:GMT:fig:cf_mu_appdx}.

\begin{figure}[H]
	\centering
	\subfigure[\smt on BA-2Motifs trainset.]{
		\includegraphics[width=0.31\textwidth]{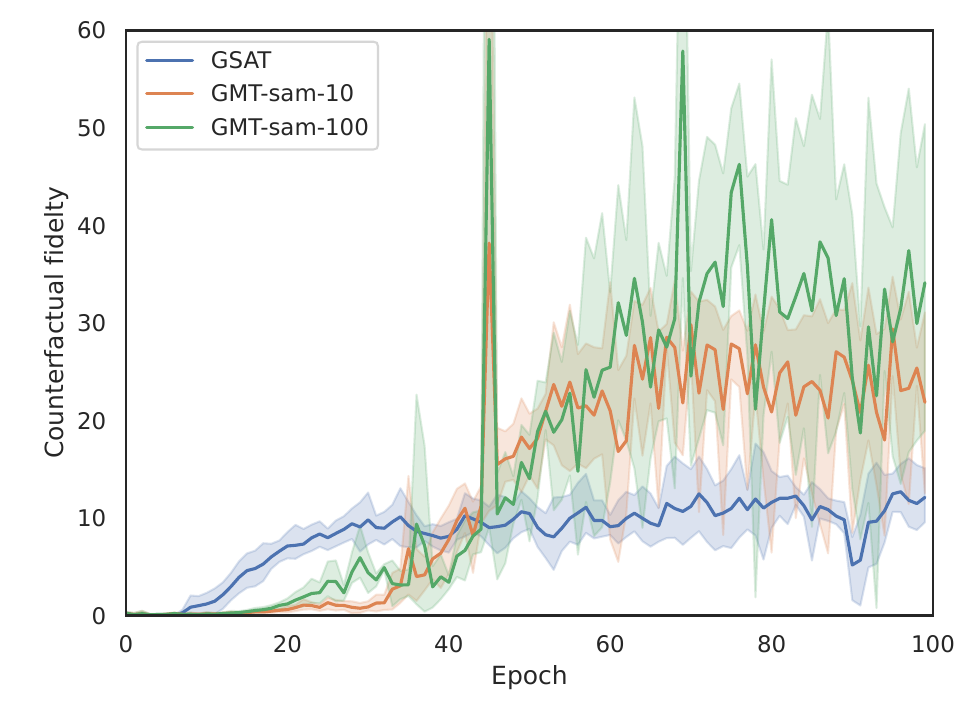}
	}
	\subfigure[\smt on BA-2Motifs valset.]{
		\includegraphics[width=0.31\textwidth]{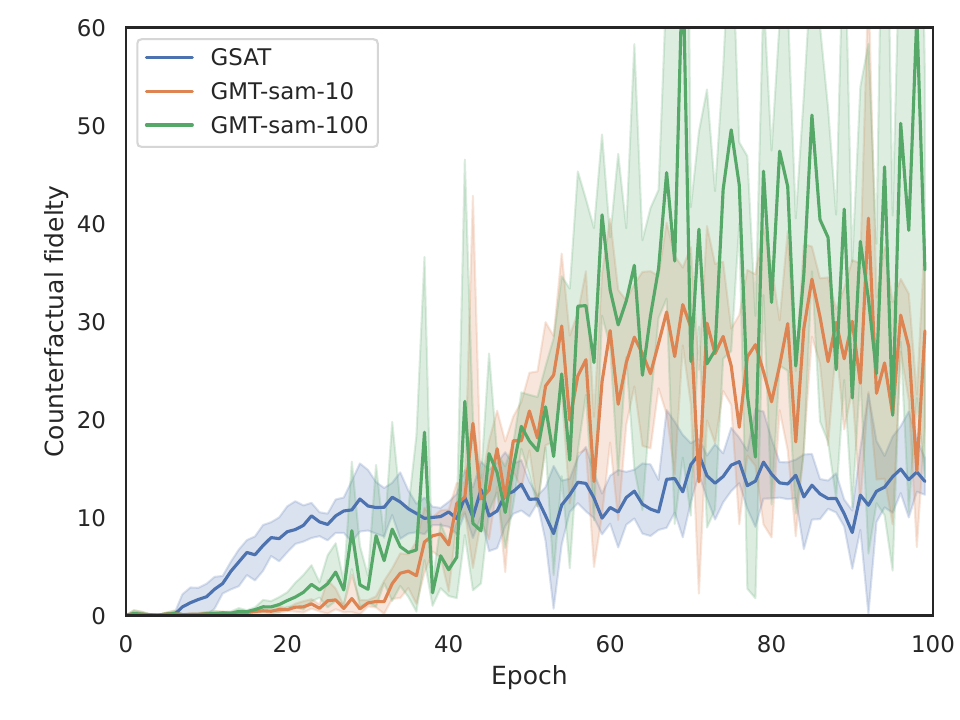}
	}
	\subfigure[\smt on BA-2Motifs test set.]{
		\includegraphics[width=0.31\textwidth]{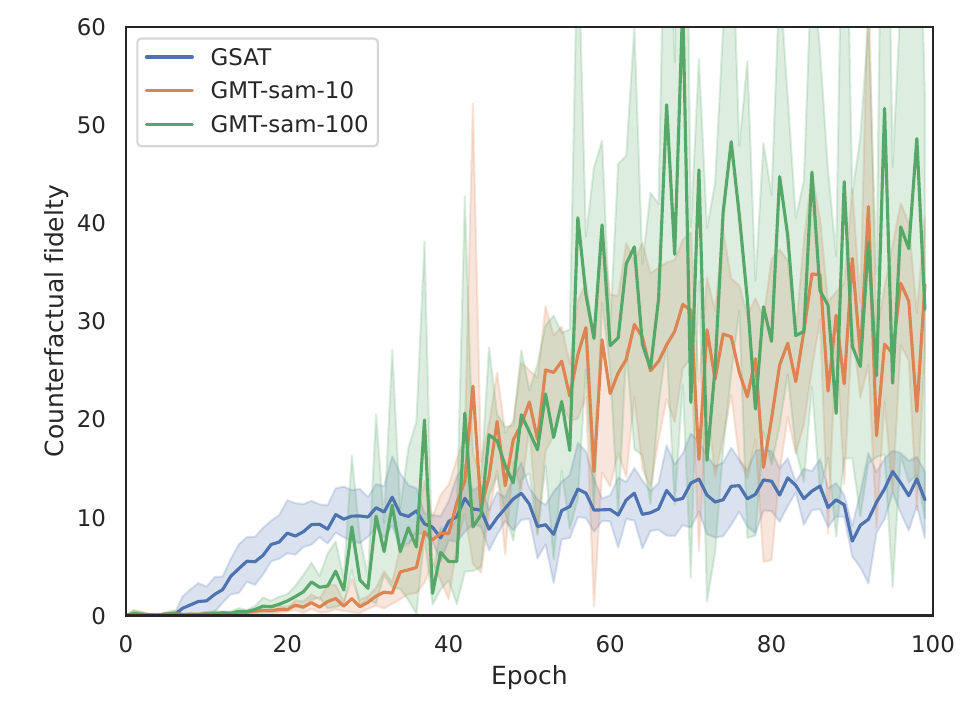}
	}
	\caption{Counterfactual fidelity on BA-2Motifs with the distance measure as KL divergence.}
	\label{CH:GMT:fig:cf_ba_kl_appdx}
\end{figure}

\begin{figure}[ht]
	\centering
	\subfigure[\smt on Mutag trainset.]{
		\includegraphics[width=0.31\textwidth]{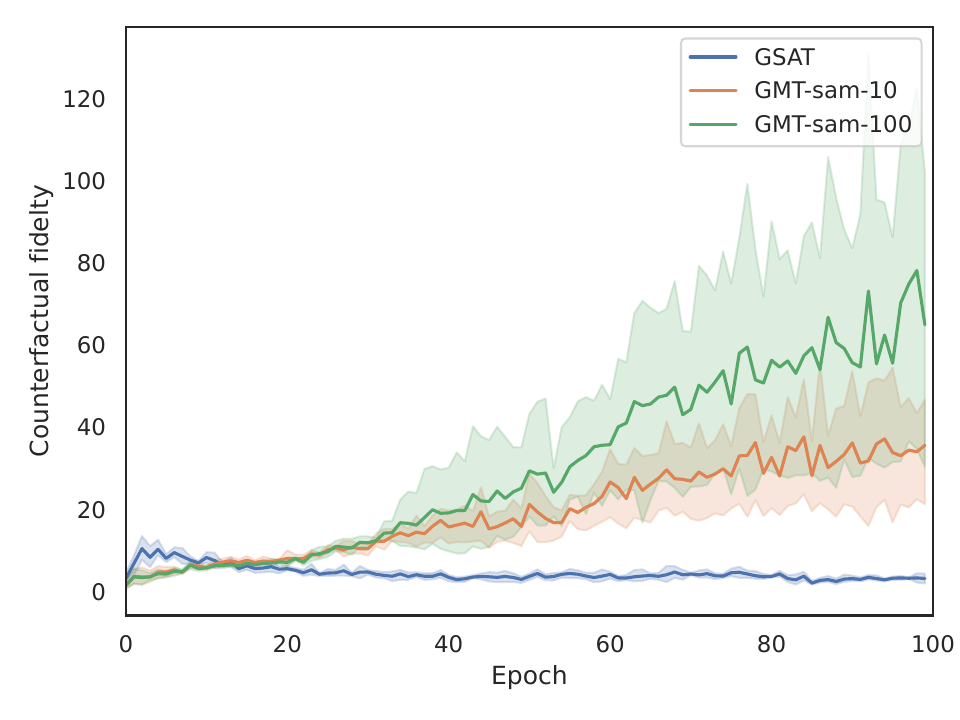}
	}
	\subfigure[\smt on Mutag validation set.]{
		\includegraphics[width=0.31\textwidth]{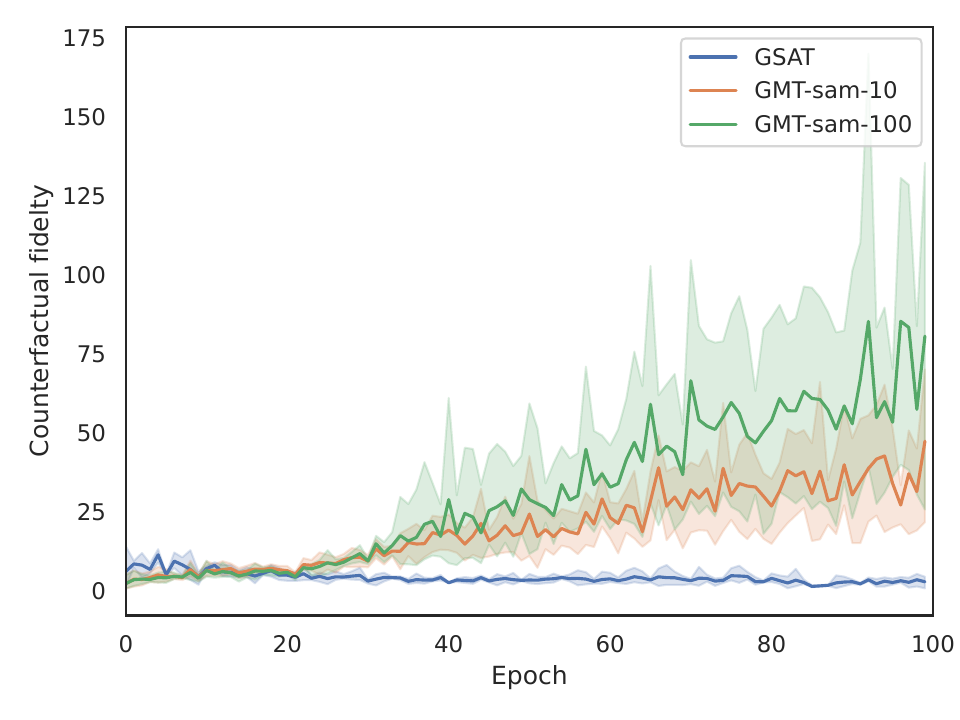}
	}
	\subfigure[\smt on Mutag test set.]{
		\includegraphics[width=0.31\textwidth]{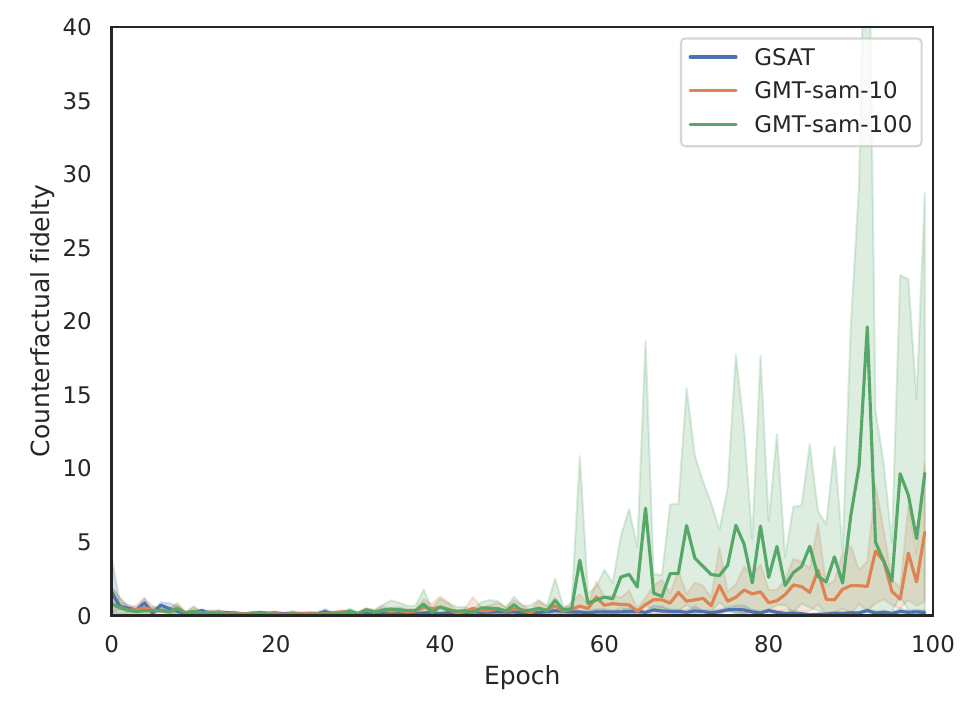}
	}
	\caption{Counterfactual fidelity on Mutag with the distance measure as KL divergence.}
	\label{CH:GMT:fig:cf_mu_kl_appdx}
\end{figure}

\begin{figure}[H]
	\centering
	\subfigure[\smt on BA-2Motifs trainset.]{
		\includegraphics[width=0.31\textwidth]{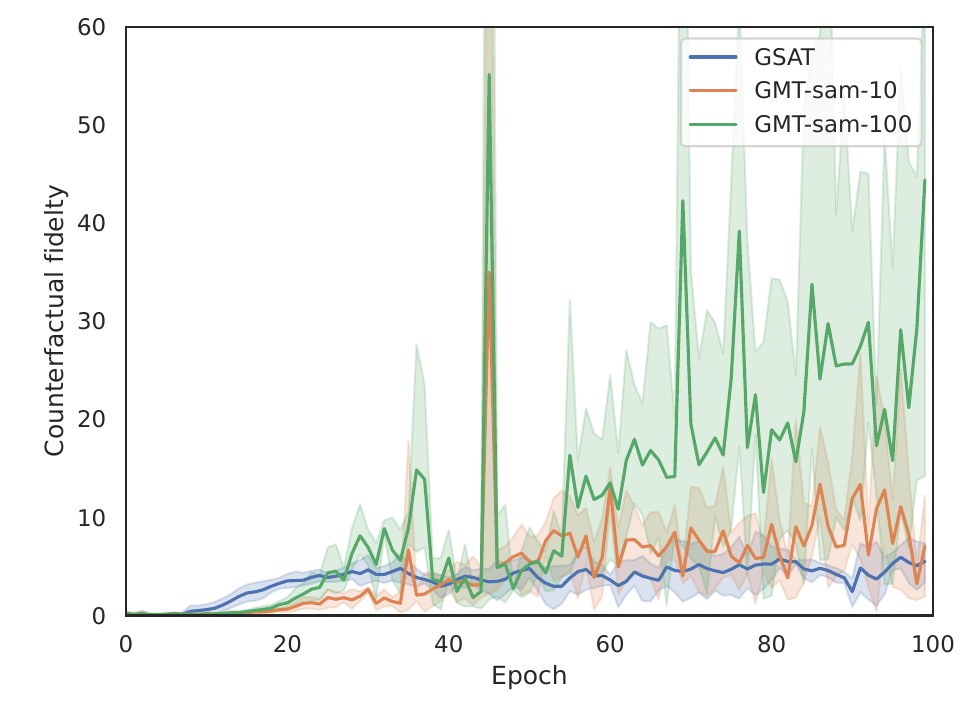}
	}
	\subfigure[\smt on BA-2Motifs valset.]{
		\includegraphics[width=0.31\textwidth]{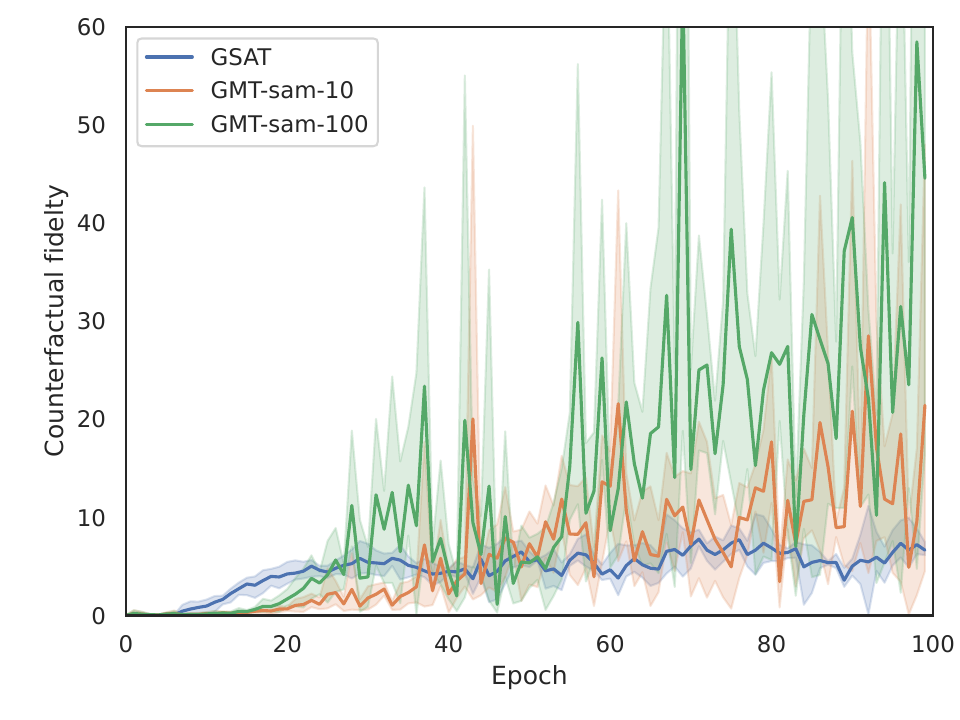}
	}
	\subfigure[\smt on BA-2Motifs test set.]{
		\includegraphics[width=0.31\textwidth]{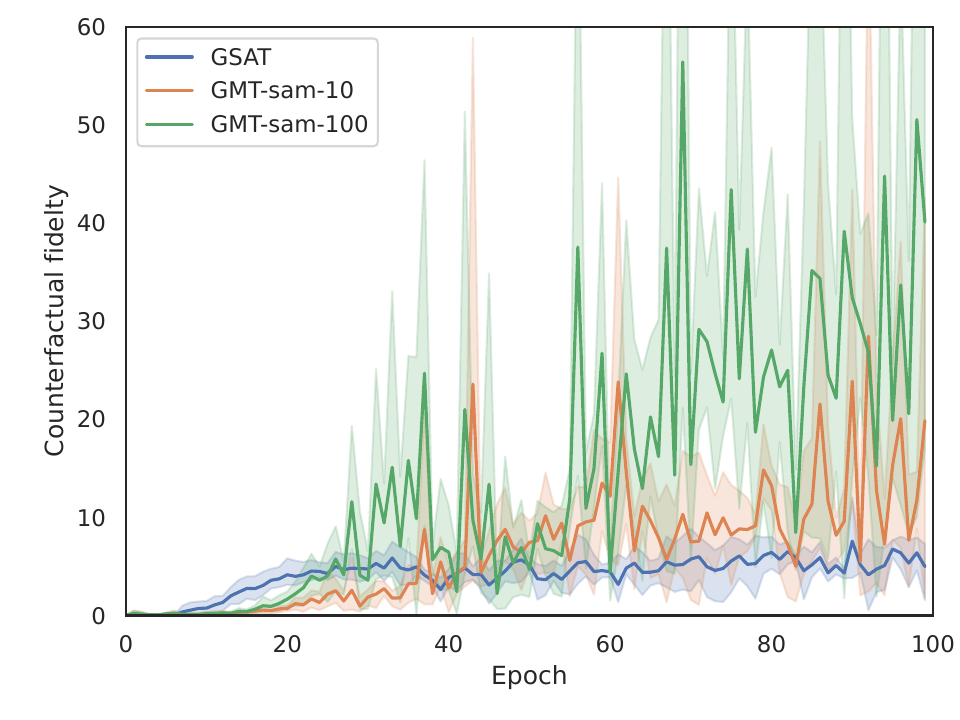}
	}
	\caption{Counterfactual fidelity on BA-2Motifs with the distance measure as JSD divergence.}
	\label{CH:GMT:fig:cf_ba_jsd_appdx}
\end{figure}

\begin{figure}[ht]
	\centering
	\subfigure[\smt on Mutag trainset.]{
		\includegraphics[width=0.31\textwidth]{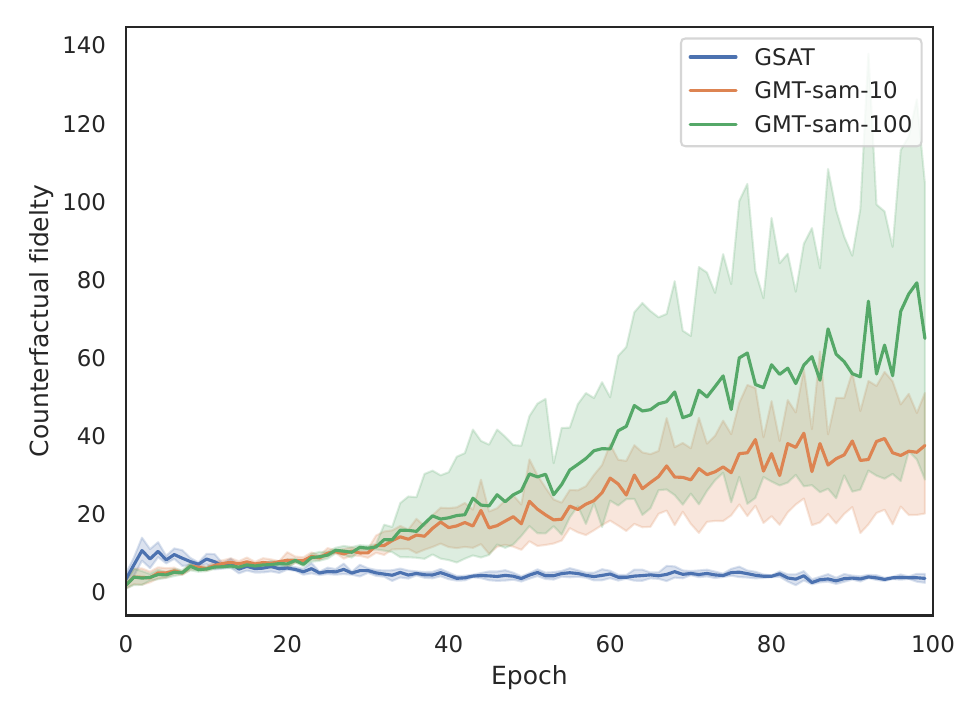}
	}
	\subfigure[\smt on Mutag validation set.]{
		\includegraphics[width=0.31\textwidth]{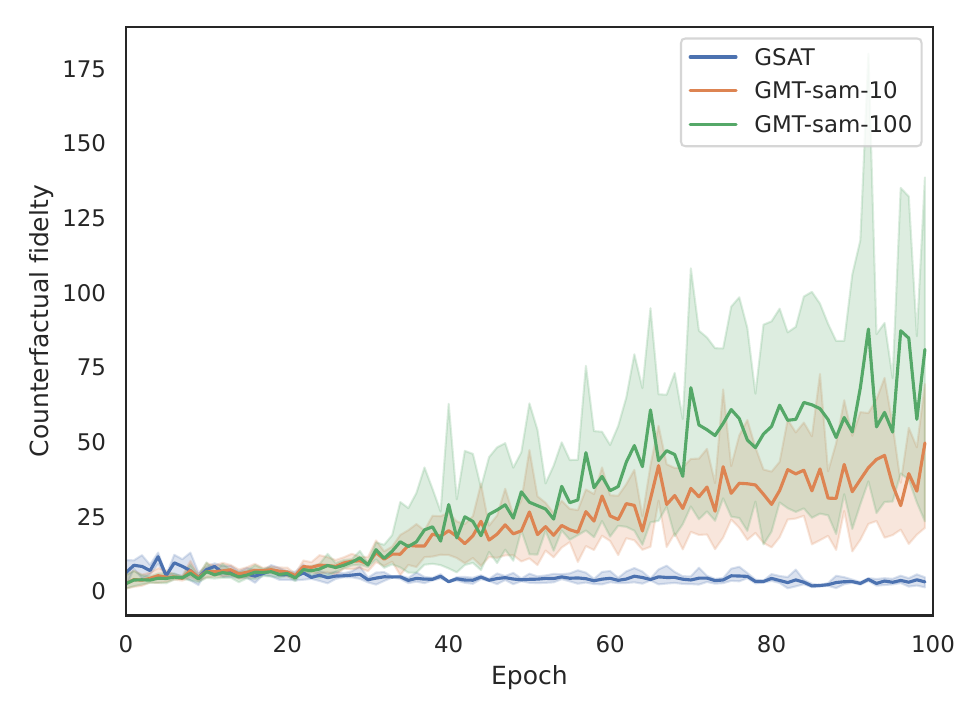}
	}
	\subfigure[\smt on Mutag test set.]{
		\includegraphics[width=0.31\textwidth]{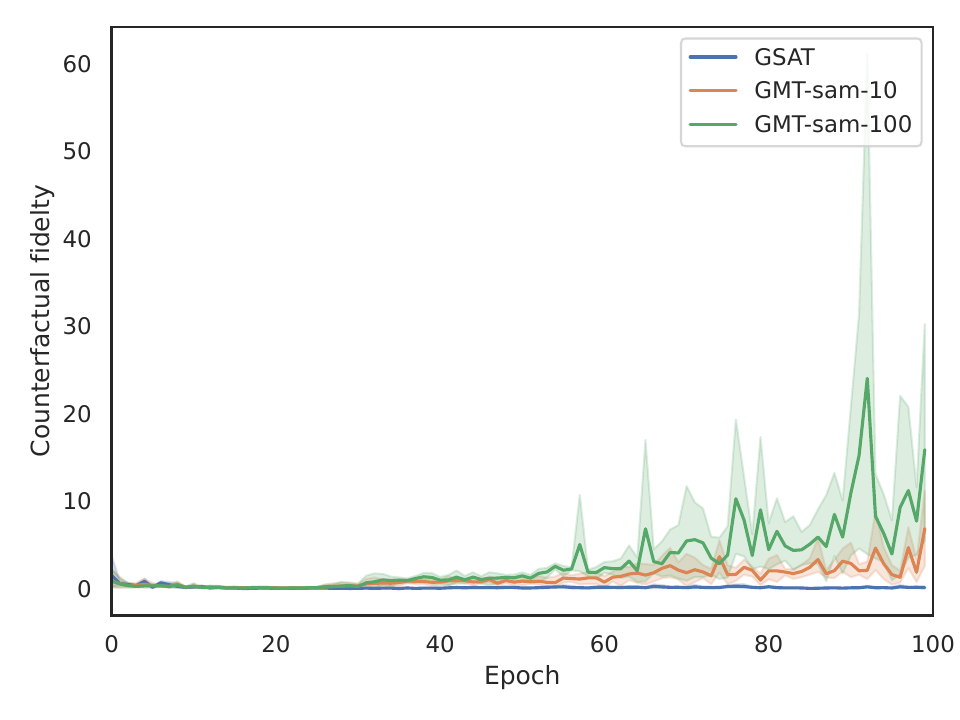}
	}
	\caption{Counterfactual fidelity on Mutag with the distance measure as JSD divergence.}
	\label{CH:GMT:fig:cf_mu_jsd_appdx}
\end{figure}

Shown as in Fig.~\ref{CH:GMT:fig:cf_ba_kl_appdx},~\ref{CH:GMT:fig:cf_mu_kl_appdx},
we plot the counterfactual fidelity results of \gsat and the simulated \smt via \gmts with $10$ and $100$  on BA-2Motifs and Mutag datasets with the distance measure as KL divergence.
Fig.~\ref{CH:GMT:fig:cf_ba_jsd_appdx},~\ref{CH:GMT:fig:cf_mu_jsd_appdx} show the counterfactual fidelity results of \gsat and the simulated \smt via \gmts with $10$ and $100$
on BA-2Motifs and Mutag datasets with the distance measure as JSD divergence.
It can be found that, the gap in counterfactual fidelity measured in KL divergence or JSD divergence can be even larger between \gsat and \smt,
growing up to $10$ times.
These results can serve as strong evidence for the degenerated interpretability caused by the failure of \smt approximation.


\begin{figure}[ht]
	\centering
	\subfigure[BA-2Motifs trainset.]{
		\includegraphics[width=0.31\textwidth]{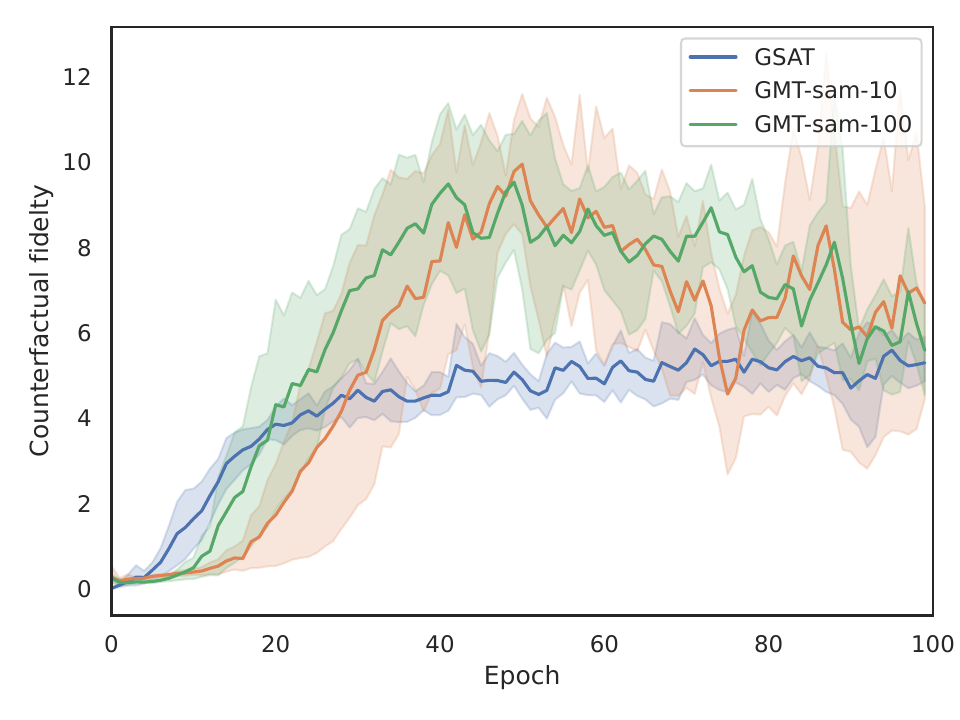}
	}
	\subfigure[BA-2Motifs valset.]{
		\includegraphics[width=0.31\textwidth]{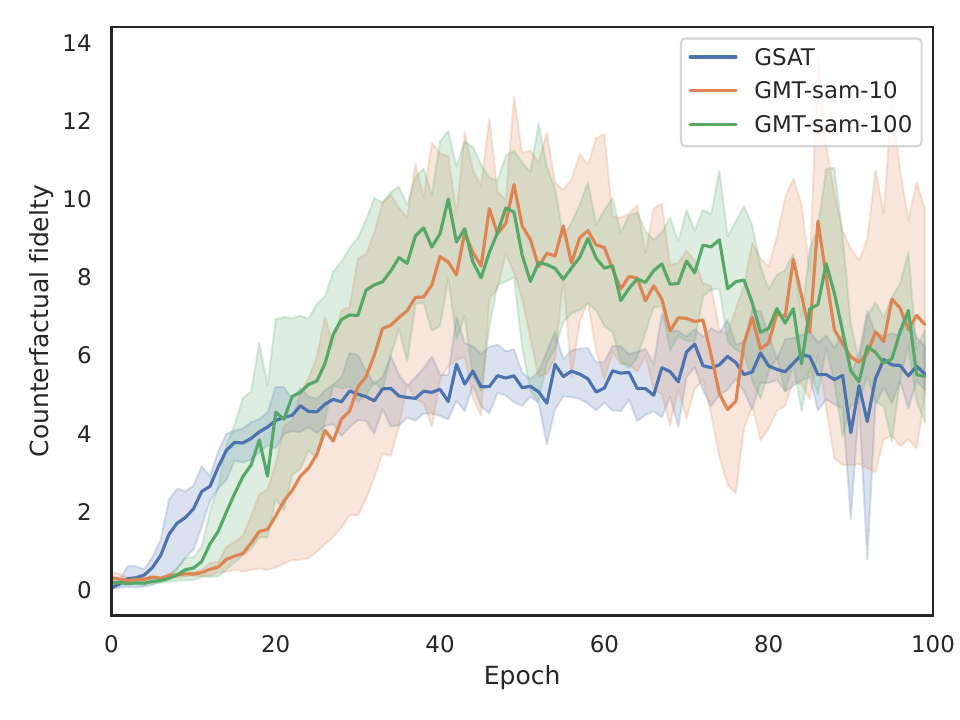}
	}
	\subfigure[BA-2Motifs test set.]{
		\includegraphics[width=0.31\textwidth]{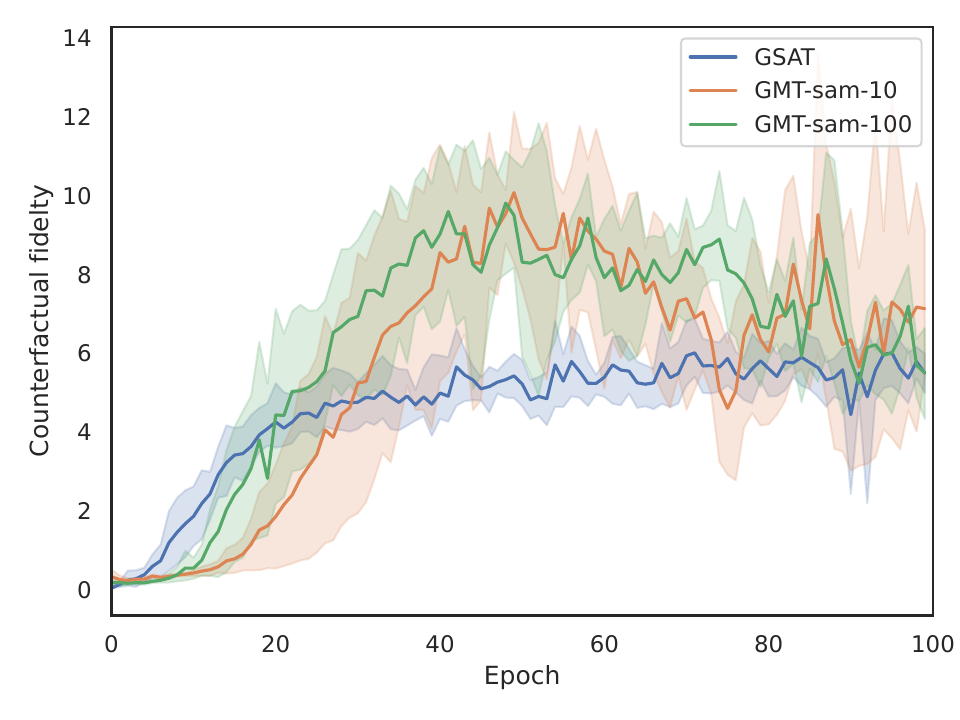}
	}
	\caption{The \ours optimization issue in terms of counterfactual fidelity on BA-2Motifs.}
	\label{CH:GMT:fig:cf_ba_opt_appdx}
\end{figure}

\begin{figure}[H]
	\centering
	\subfigure[Mutag trainset.]{
		\includegraphics[width=0.31\textwidth]{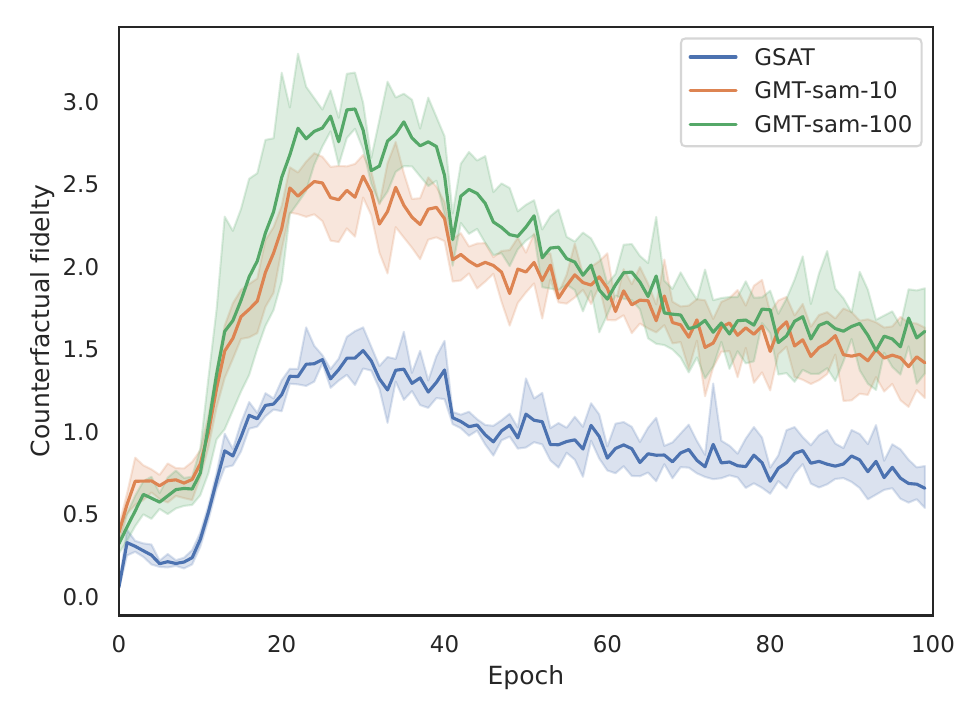}
	}
	\subfigure[Mutag validation set.]{
		\includegraphics[width=0.31\textwidth]{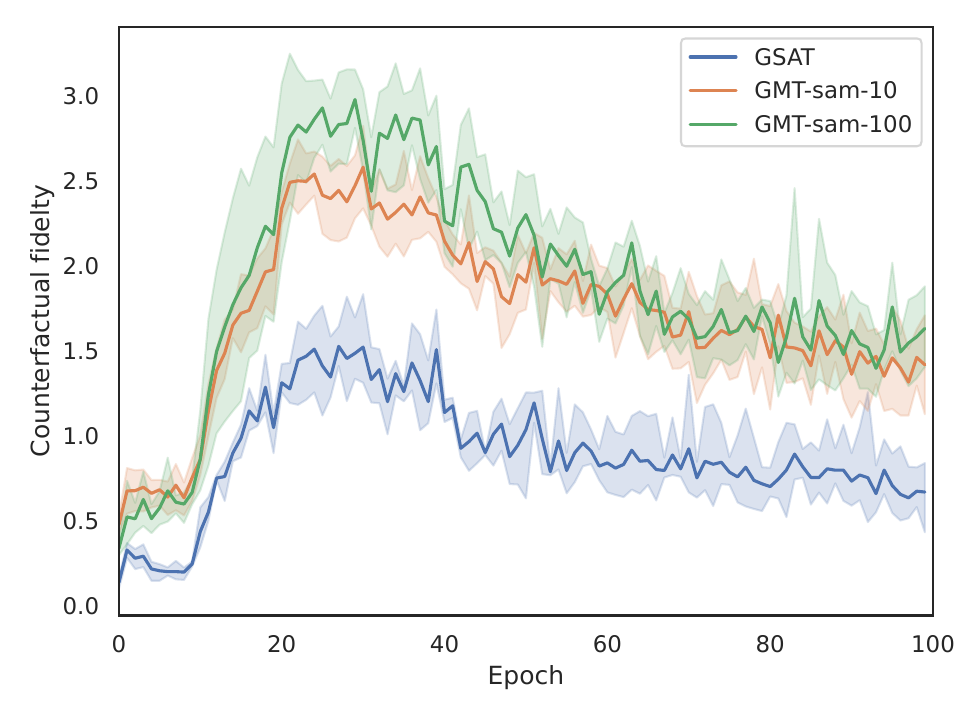}
	}
	\subfigure[Mutag test set.]{
		\includegraphics[width=0.31\textwidth]{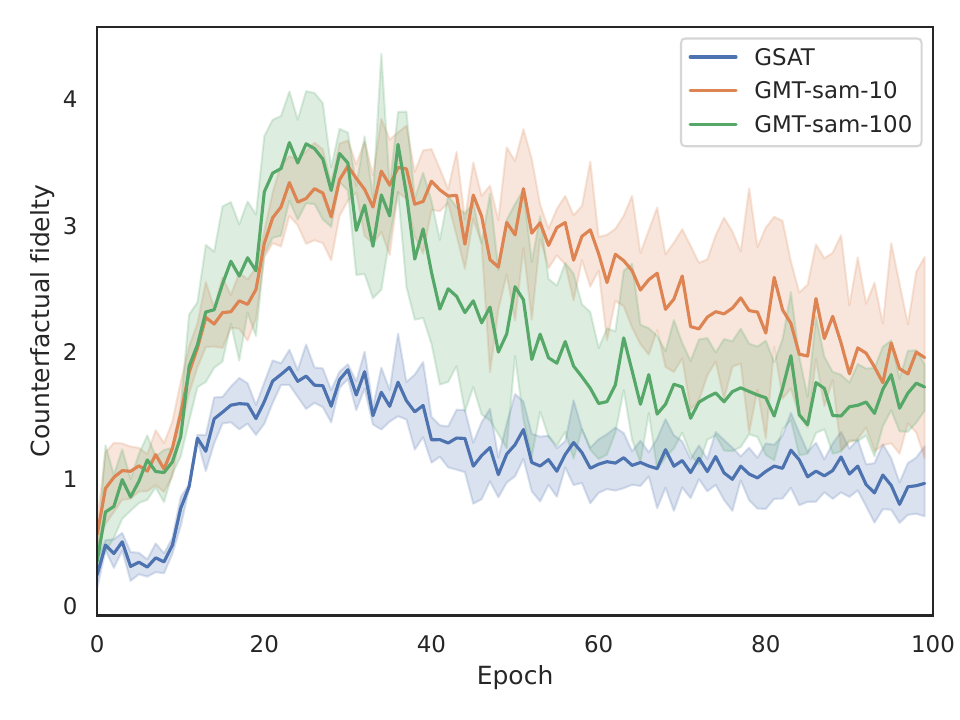}
	}
	\caption{The \ours optimization issue in terms of counterfactual fidelity on Mutag.}
	\label{CH:GMT:fig:cf_mu_opt_appdx}
\end{figure}

Shown as in Fig.~\ref{CH:GMT:fig:cf_ba_opt_appdx},~\ref{CH:GMT:fig:cf_mu_opt_appdx},
we plot the counterfactual fidelity results of \gsat and the simulated \smt via \gmts with $10$ and $100$  on BA-2Motifs and Mutag datasets.
Compared to previous results, the \gmts in Fig.~\ref{CH:GMT:fig:cf_ba_opt_appdx},~\ref{CH:GMT:fig:cf_mu_opt_appdx} does not use any warmup strategies
that may suffer from the optimization issue as discussed in Sec.~\ref{CH:GMT:sec:gmt_impl_dis_appdx}.
It can be found that, at the begining of the optimization, \gmts demonstrates increasing counterfactual fidelity.
However, as the optimization keeps proceeding, the counterfactual fidelity of \gmts will degenerate, because of fitting to the trivial solution of the \gsat objective.
Consequently, the interpretation results will degenerate too at the end of the optimization.

\subsection{\smt approximation gap analysis}
\label{CH:GMT:sec:smt_gap_viz_appdx}

\begin{figure}[H]
	\centering
	\subfigure[BA-2Motifs trainset.]{
		\includegraphics[width=0.31\textwidth]{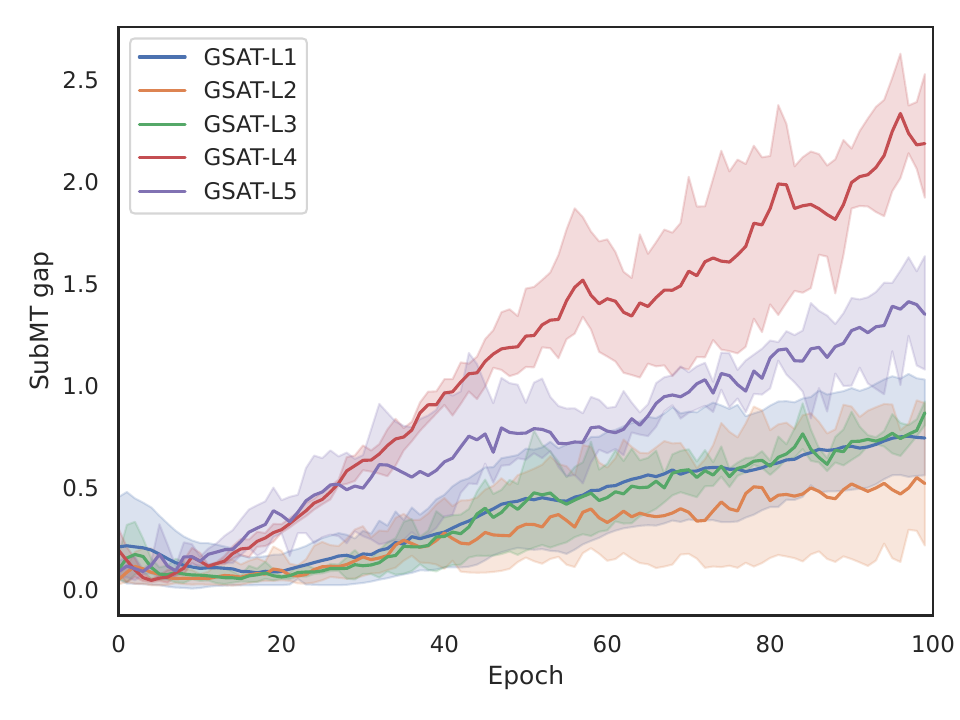}
	}
	\subfigure[BA-2Motifs valset.]{
		\includegraphics[width=0.31\textwidth]{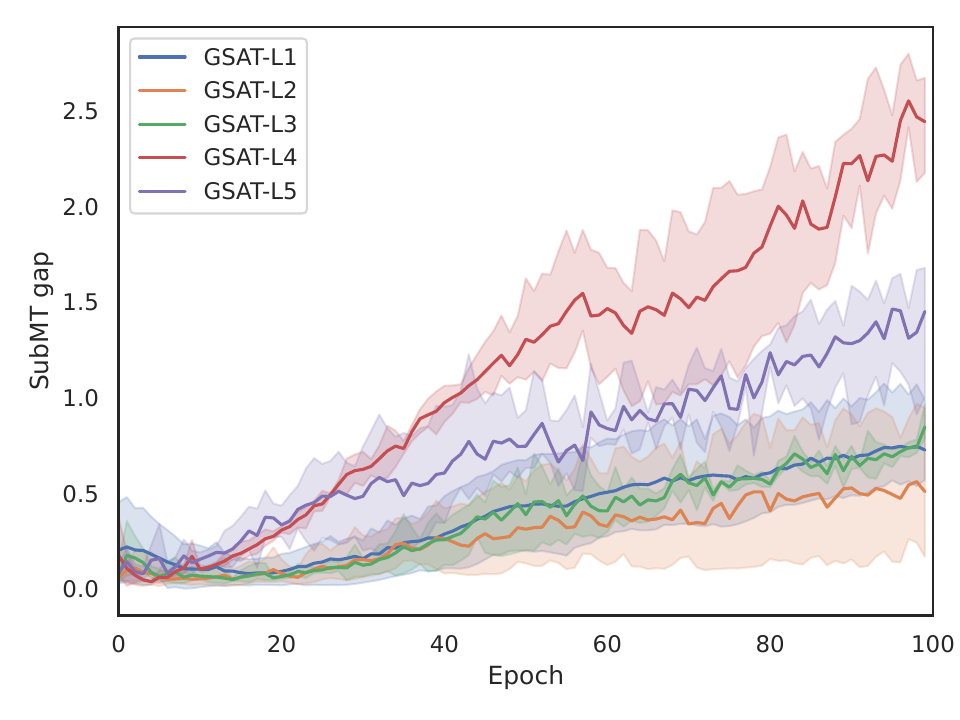}
	}
	\subfigure[BA-2Motifs test set.]{
		\includegraphics[width=0.31\textwidth]{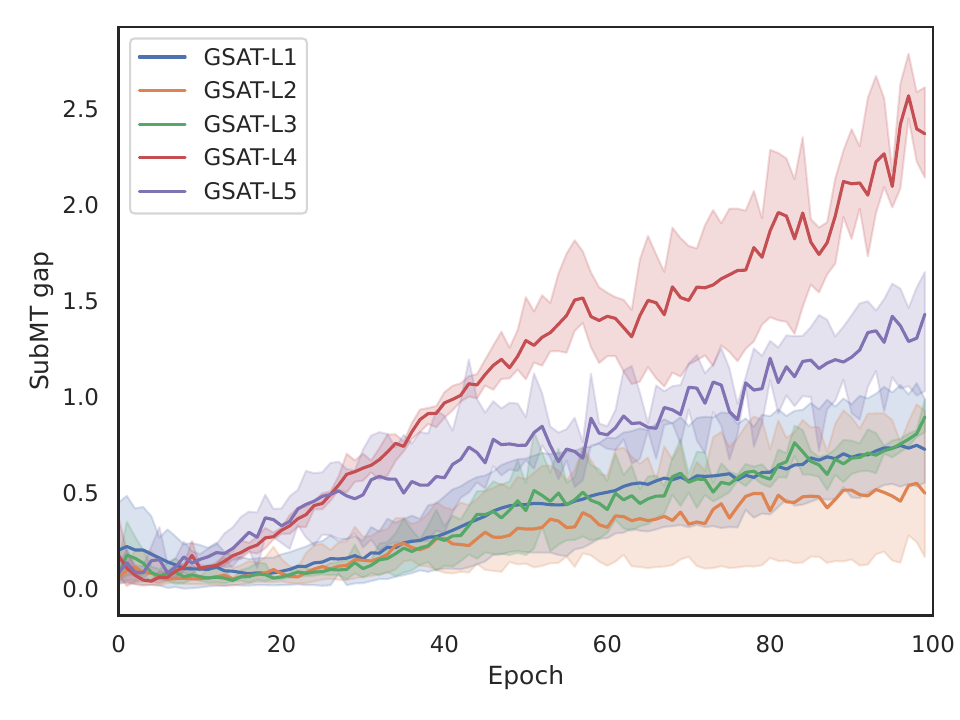}
	}
	\caption{The \smt approximation gap of \gsat with SGC on BA-2Motifs.}
	\label{CH:GMT:fig:gap_ba2_sgc_appdx}
\end{figure}
\begin{figure}[H]
	\centering
	\subfigure[BA-2Motifs trainset.]{
		\includegraphics[width=0.31\textwidth]{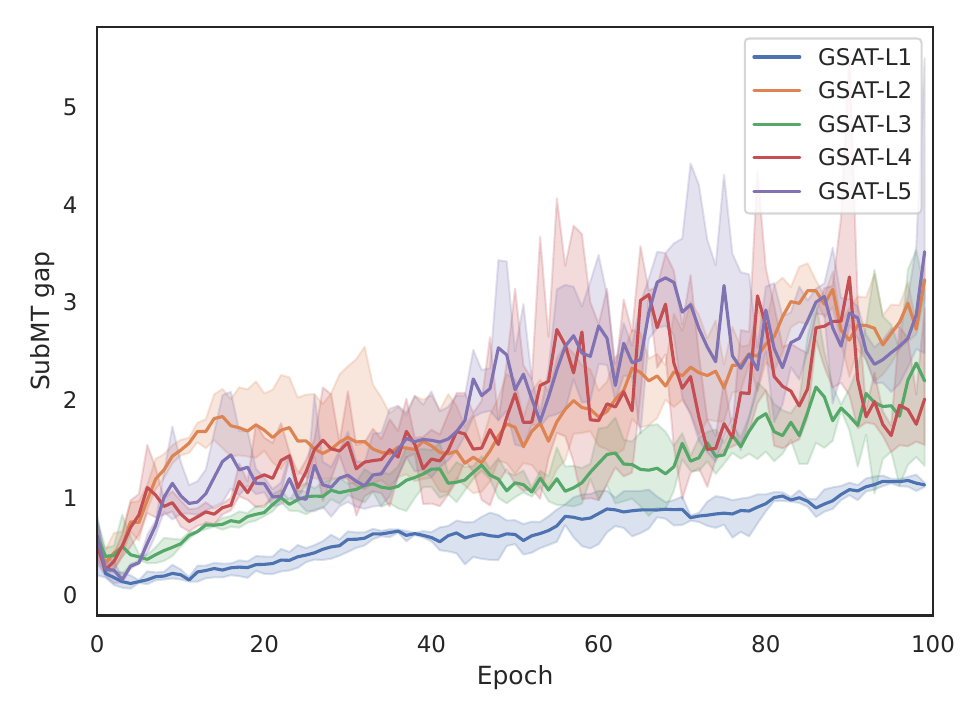}
	}
	\subfigure[BA-2Motifs valset.]{
		\includegraphics[width=0.31\textwidth]{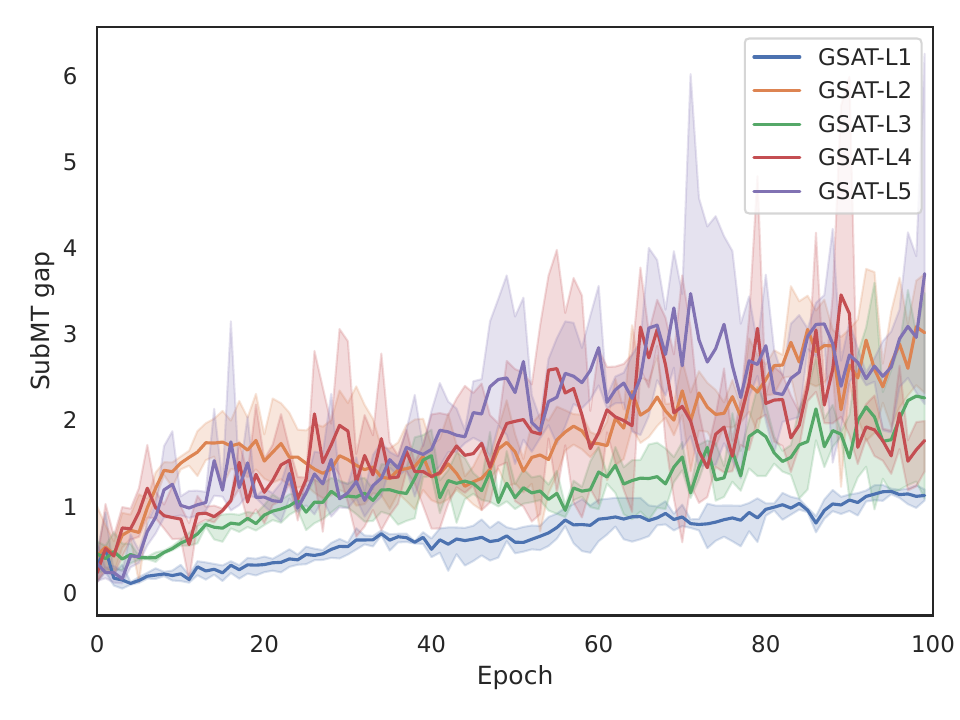}
	}
	\subfigure[BA-2Motifs test set.]{
		\includegraphics[width=0.31\textwidth]{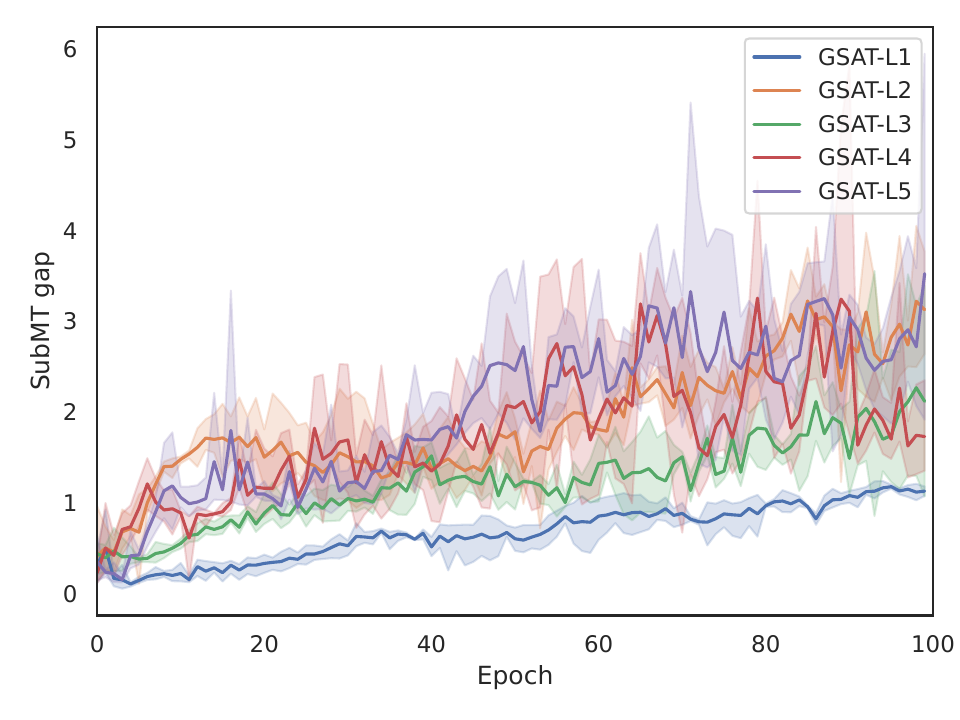}
	}
	\caption{The \smt approximation gap of \gsat with GIN on BA-2Motifs.}
	\label{CH:GMT:fig:gap_ba2_gin_appdx}
\end{figure}
\begin{figure}[H]
	\centering
	\subfigure[Mutag trainset.]{
		\includegraphics[width=0.31\textwidth]{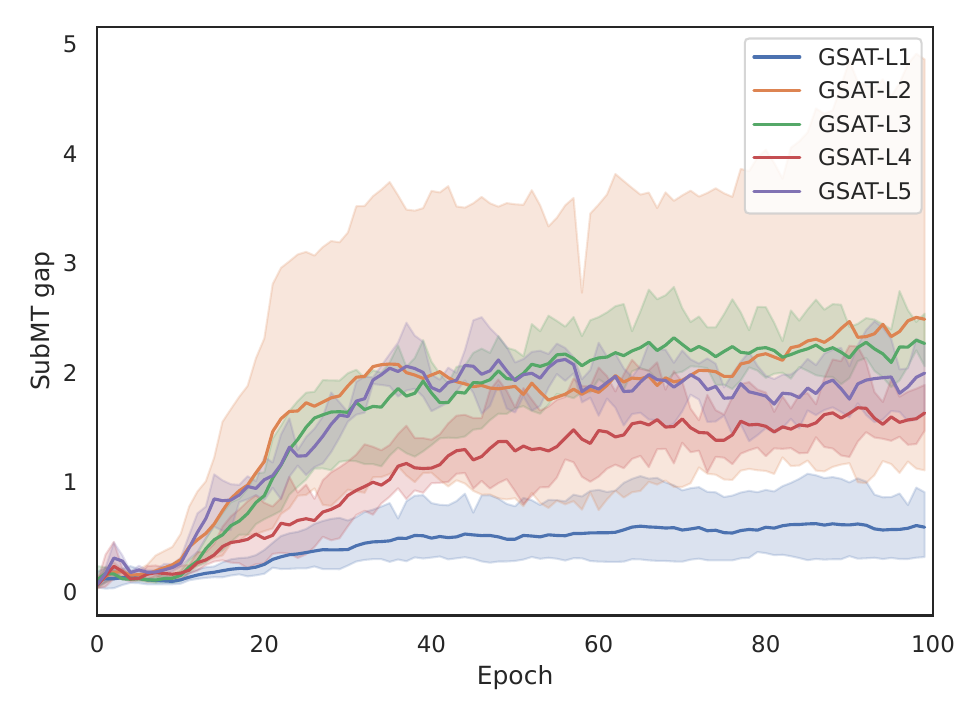}
	}
	\subfigure[Mutag validation set.]{
		\includegraphics[width=0.31\textwidth]{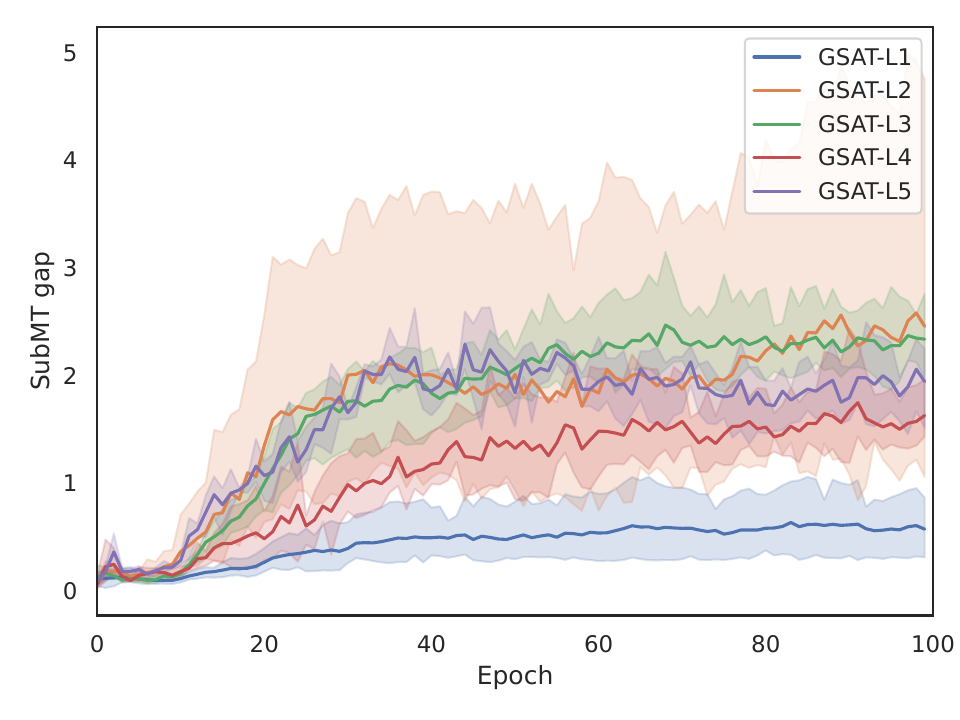}
	}
	\subfigure[Mutag test set.]{
		\includegraphics[width=0.31\textwidth]{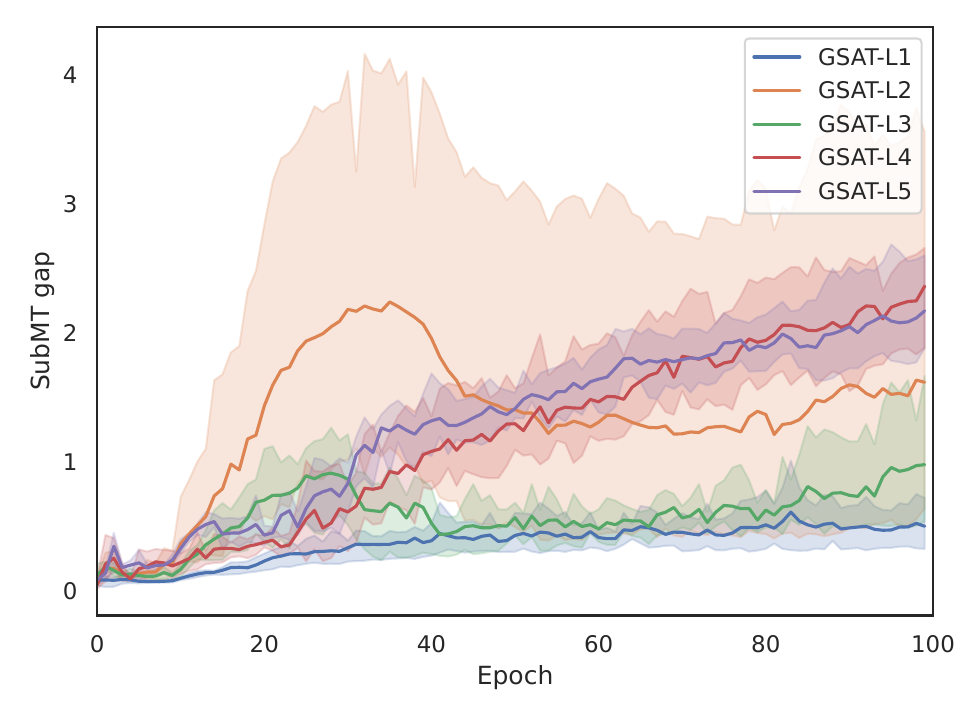}
	}
	\caption{The \smt approximation gap of \gsat with SGC on Mutag.}
	\label{CH:GMT:fig:gap_mu_sgc_appdx}
\end{figure}
\begin{figure}[H]
	\centering
	\subfigure[Mutag trainset.]{
		\includegraphics[width=0.31\textwidth]{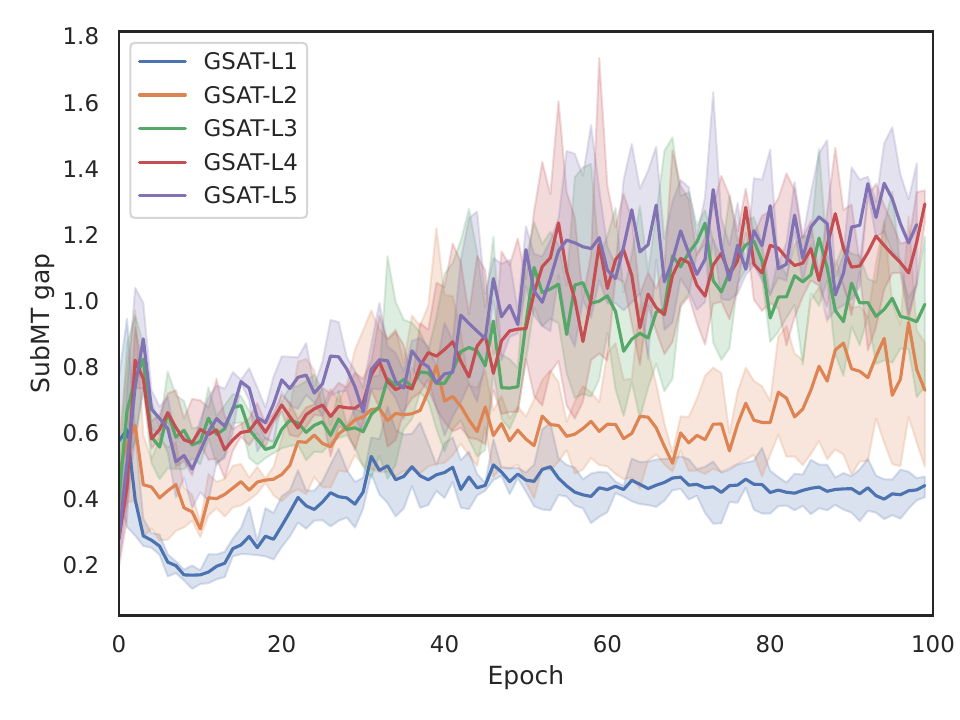}
	}
	\subfigure[Mutag validation set.]{
		\includegraphics[width=0.31\textwidth]{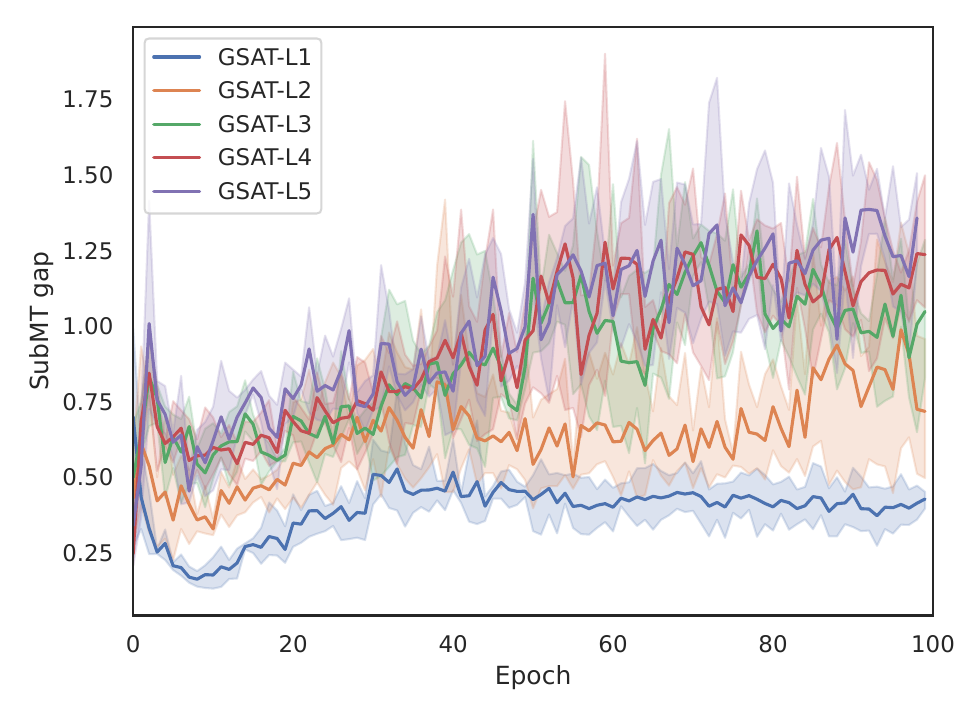}
	}
	\subfigure[Mutag test set.]{
		\includegraphics[width=0.31\textwidth]{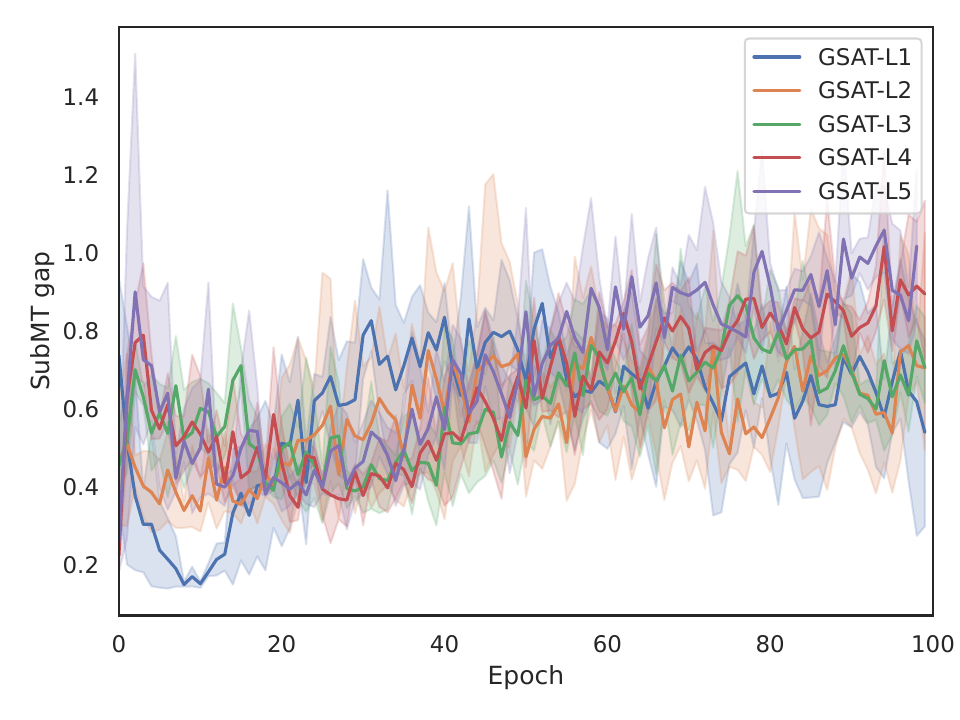}
	}
	\caption{The \smt approximation gap of \gsat with GIN on Mutag.}
	\label{CH:GMT:fig:gap_mu_gin_appdx}
\end{figure}

Fig.~\ref{CH:GMT:fig:gap_ba2_sgc_appdx} and ~\ref{CH:GMT:fig:gap_ba2_gin_appdx}, Fig.~\ref{CH:GMT:fig:gap_mu_sgc_appdx} and ~\ref{CH:GMT:fig:gap_mu_gin_appdx} demonstrate the \smt approximation gaps of \gsat implemented
in GIN and SGC on BA\_2Motifs and Mutag respectively.
To fully verify Proposition~\ref{CH:GMT:thm:submt_gap_appdx}, we range the number of layers of GIN and SGC from $1$ to $5$.
It can be found that the results are well aligned with Proposition~\ref{CH:GMT:thm:submt_gap_appdx}.
When the number of layers is $1$, the \smt approximation gap is smallest, because of more ``linearity'' in the network.
While along with the growing number of GNN layers, the network becomes more ``unlinear'' such that the \smt approximation gap will be larger.

\subsection{Software and Hardware}
\label{CH:GMT:sec:exp_software_appdx}
We implement our methods with PyTorch~\citep{pytorch} and PyTorch Geometric~\citep{pytorch_geometric} 2.0.4.
We ran our experiments on Linux Servers installed with V100 graphics cards and CUDA 11.3.

\subsection{Interpretation Visualization}
\label{CH:GMT:sec:inter_viz_appdx}
To better understand the results, we provide visualizations of the learned interpretable subgraphs by \gsat and \gmts in the Spurious-Motif datasets, as well as the learned interpretable subgraphs by \gmts in OGBG-Molhiv dataset.

\begin{figure}[H]
	\centering
	\subfigure[Spurious-Motif class $0$ under bias$=0.5$ by \gsat.]{
		\includegraphics[width=\textwidth]{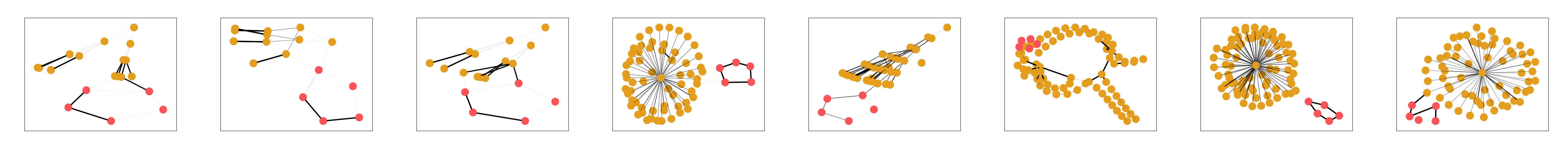}
	}
	\subfigure[Spurious-Motif class $0$ under bias$=0.5$ by \gmts.]{
		\includegraphics[width=\textwidth]{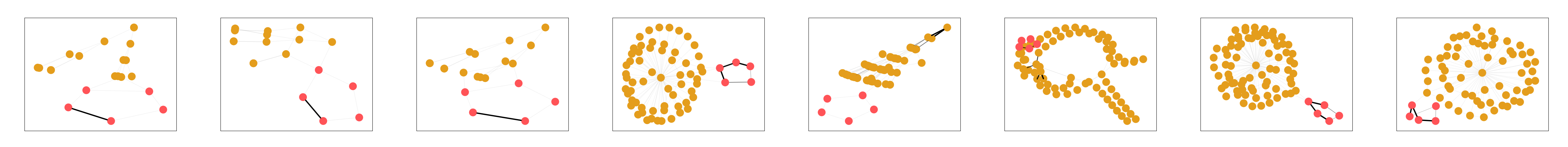}
	}
	\subfigure[Spurious-Motif class $1$ under bias$=0.5$ by \gsat.]{
		\includegraphics[width=\textwidth]{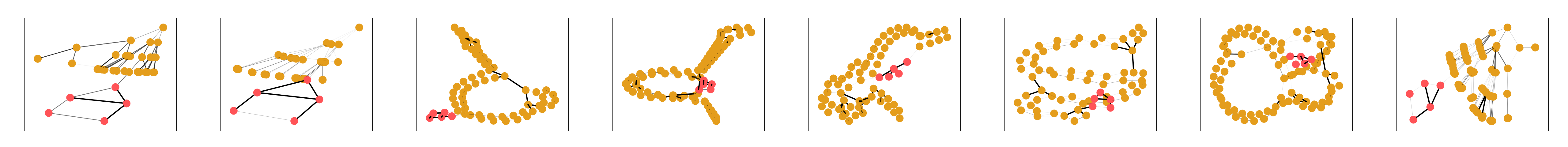}
	}
	\subfigure[Spurious-Motif class $1$ under bias$=0.5$ by \gmts.]{
		\includegraphics[width=\textwidth]{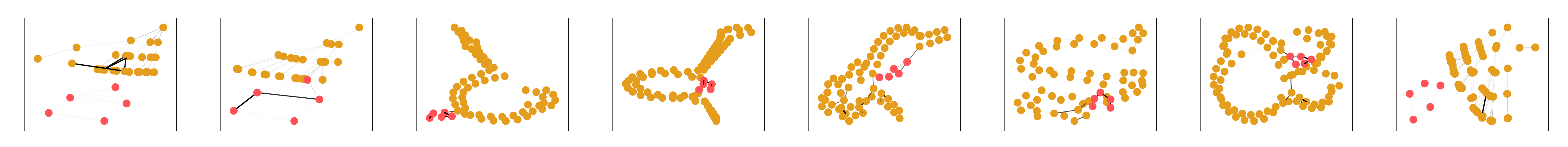}
	}
	\subfigure[Spurious-Motif class $2$ under bias$=0.5$ by \gsat.]{
		\includegraphics[width=\textwidth]{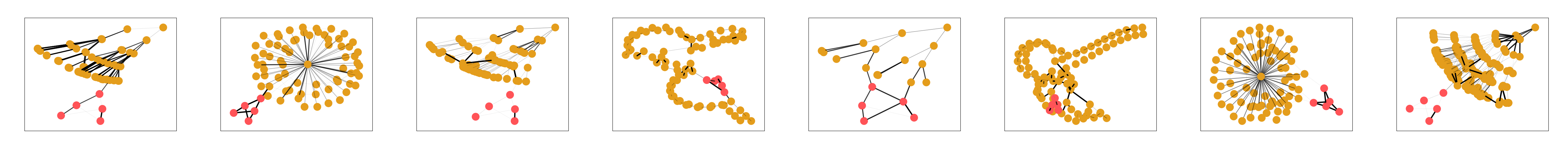}
	}
	\subfigure[Spurious-Motif class $2$ under bias$=0.5$ by \gmts.]{
		\includegraphics[width=\textwidth]{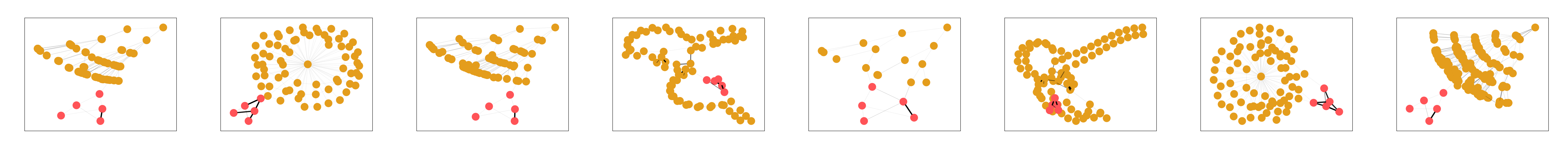}
	}
	\caption{
		Learned interpretable subgraphs by \gsat and \gmts on Spurious-Motif $b=0.5$.}
	\label{CH:GMT:fig:sp5_viz_appdx}
\end{figure}

\begin{figure}[H]
	\centering
	\subfigure[Spurious-Motif class $0$ under bias$=0.5$ by \gsat.]{
		\includegraphics[width=\textwidth]{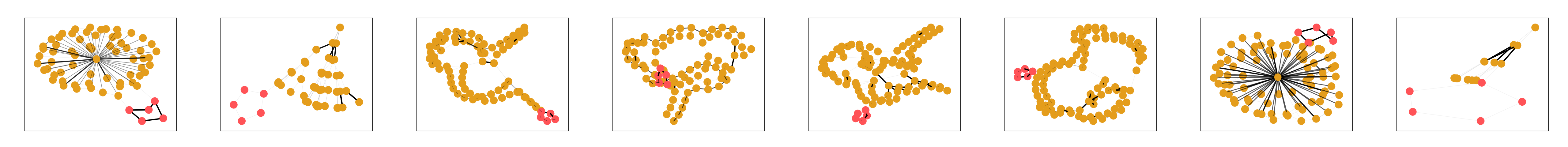}
	}
	\subfigure[Spurious-Motif class $0$ under bias$=0.5$ by \gmts.]{
		\includegraphics[width=\textwidth]{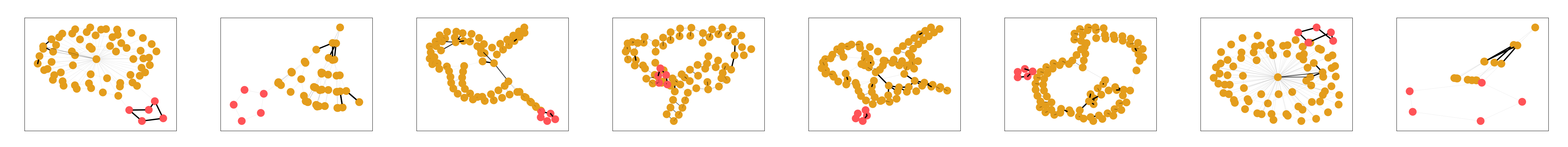}
	}
	\subfigure[Spurious-Motif class $1$ under bias$=0.5$ by \gsat.]{
		\includegraphics[width=\textwidth]{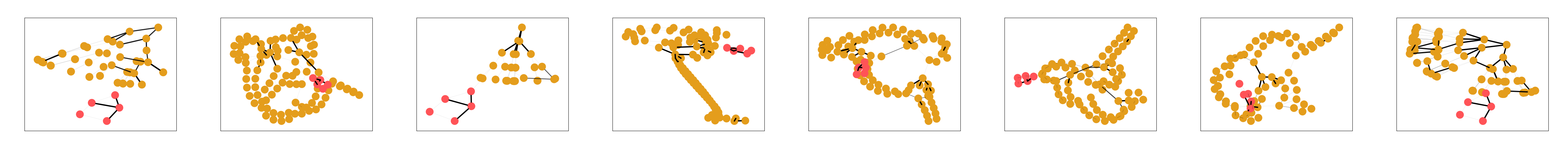}
	}
	\subfigure[Spurious-Motif class $1$ under bias$=0.5$ by \gmts.]{
		\includegraphics[width=\textwidth]{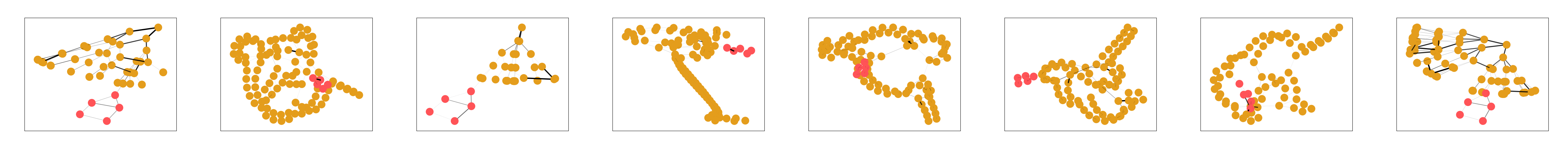}
	}
	\subfigure[Spurious-Motif class $2$ under bias$=0.5$ by \gsat.]{
		\includegraphics[width=\textwidth]{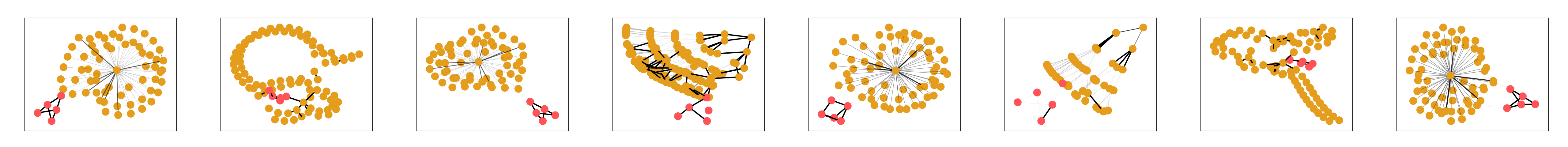}
	}
	\subfigure[Spurious-Motif class $2$ under bias$=0.5$ by \gmts.]{
		\includegraphics[width=\textwidth]{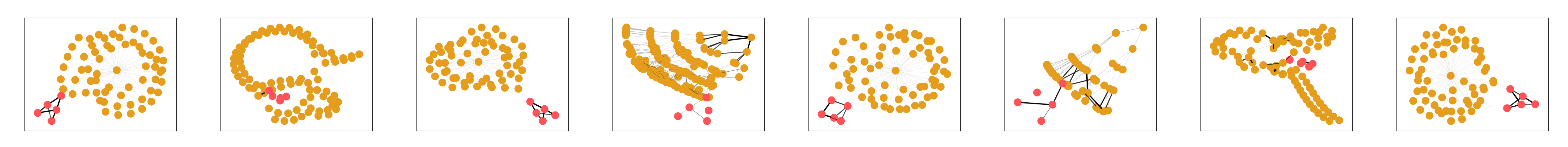}
	}
	\caption{
		Learned interpretable subgraphs by \gsat and \gmts on Spurious-Motif $b=0.7$.}
	\label{CH:GMT:fig:sp7_viz_appdx}
\end{figure}

\begin{figure}[H]
	\centering
	\subfigure[Spurious-Motif class $0$ under bias$=0.5$ by \gsat.]{
		\includegraphics[width=\textwidth]{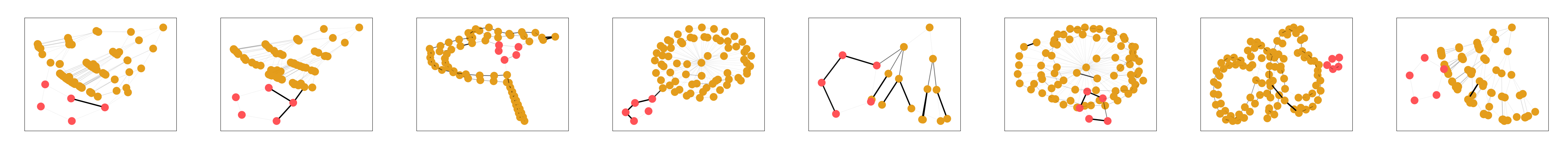}
	}
	\subfigure[Spurious-Motif class $0$ under bias$=0.5$ by \gmts.]{
		\includegraphics[width=\textwidth]{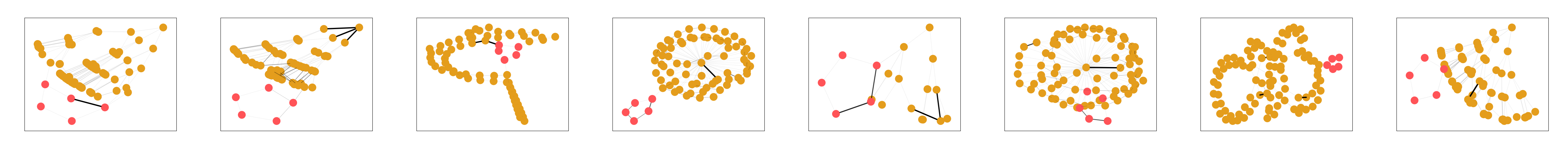}
	}
	\subfigure[Spurious-Motif class $1$ under bias$=0.5$ by \gsat.]{
		\includegraphics[width=\textwidth]{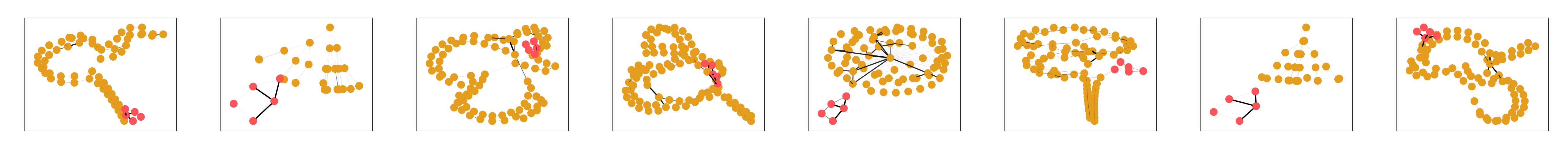}
	}
	\subfigure[Spurious-Motif class $1$ under bias$=0.5$ by \gmts.]{
		\includegraphics[width=\textwidth]{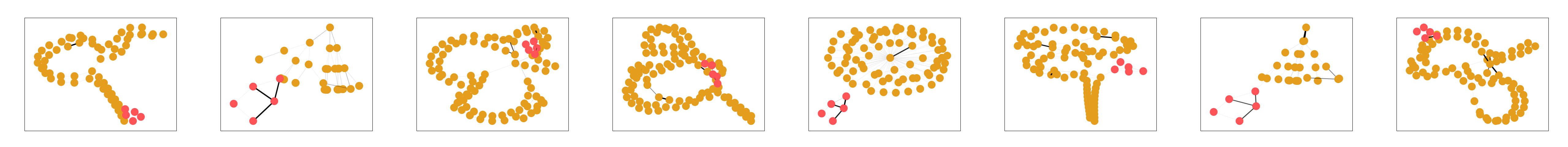}
	}
	\subfigure[Spurious-Motif class $2$ under bias$=0.5$ by \gsat.]{
		\includegraphics[width=\textwidth]{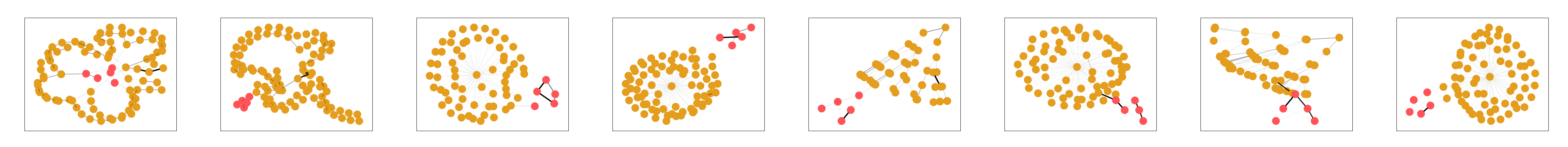}
	}
	\subfigure[Spurious-Motif class $2$ under bias$=0.5$ by \gmts.]{
		\includegraphics[width=\textwidth]{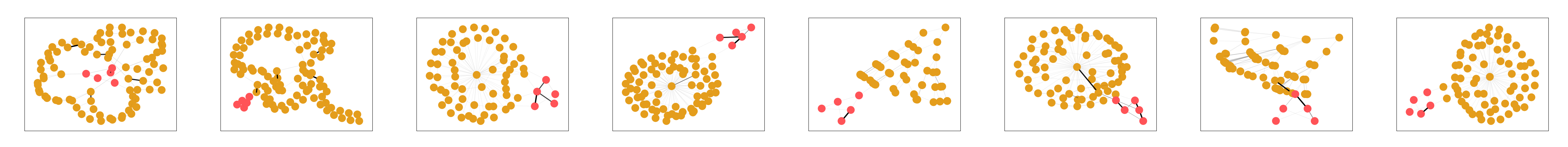}
	}
	\caption{
		Learned interpretable subgraphs by \gsat and \gmts on Spurious-Motif $b=0.9$.}
	\label{CH:GMT:fig:sp9_viz_appdx}
\end{figure}

The results on Spurious-Motif datasets are given in Fig.~\ref{CH:GMT:fig:sp5_viz_appdx},~\ref{CH:GMT:fig:sp7_viz_appdx},\ref{CH:GMT:fig:sp9_viz_appdx} for $b=0.5$, $b=0.7$ and $b=0.9$, respectively. The red nodes are the ground-truth interpretable subgraphs. It can be found that \gmts indeed learns the interpretable subgraph better than \gsat, which also explains the excellent OOD generalization ability of \gmts on Spurious Motif datasets.

\begin{figure}[H]
	\centering
	\subfigure[OGBG-Molhiv class $0$ by \gmts.]{
		\includegraphics[width=\textwidth]{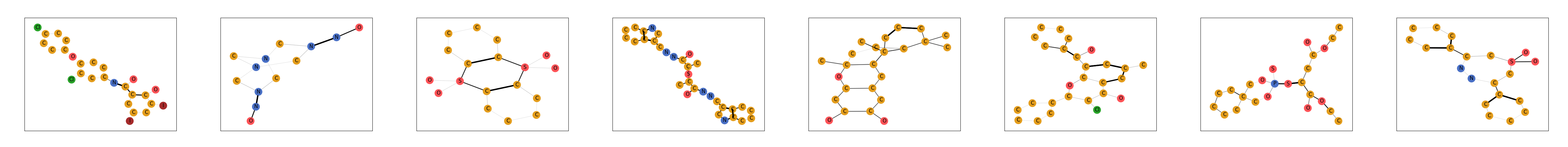}
	}
	\subfigure[OGBG-Molhiv class $1$ by \gmts.]{
		\includegraphics[width=\textwidth]{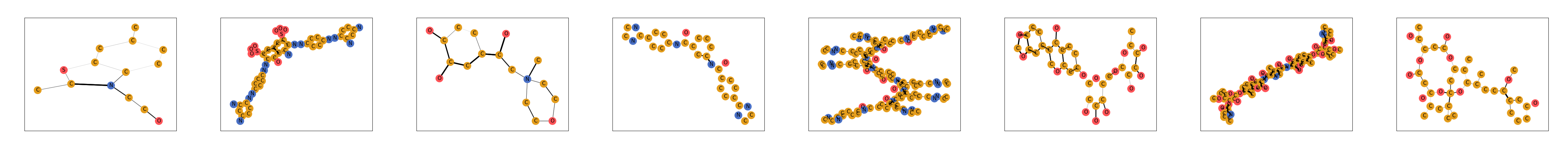}
	}
	\caption{
		Learned interpretable subgraphs by \gmts on OGBG-Molhiv.}
	\label{CH:GMT:fig:oh_viz_appdx}
\end{figure}

In addition, we also provide the visualization of interpretable subgraphs learned by \gmts on OGBG-Molhiv, given in Fig.~\ref{CH:GMT:fig:oh_viz_appdx}.

\chapter{Appendices of \hao}\label{APP:HAO}

\section{Additional Discussions and Future Directions}
\label{CH:HAO:sec:discussion_future_appdx}
\subsection{Discussions on HAO and its Limitations}
\label{CH:HAO:sec:discussion_limitation_appdx}
\textbf{Discussions on HAO  and future implications.} It is widely received that it is difficult to give a proper definition of unnoticeability for graphs (More details are also given in Appendix~\ref{CH:HAO:sec:threat_model_reason_appdx}).
Based on earliest unnoticeability constraints on degree distribution changes~\citep{nettack,metattack},
empirical observations that graph adversarial attacks may change some feature statistics and connect dissimilar neighbors are identified,
and leveraged as heuristics to develop robust GNNs~\citep{advsample_deepinsights,gnn_svd,gnnguard,prognn}.
Though empirically effective, however, few of them provide theoretical explanations or relate this phenomenon to unnoticeability.
In this work, starting from the comparison of GMA and GIA, we identified GIA would lead to severe damage to the original homophily.
Furthermore, the relatively high flexibility of GIA amplifies the destruction and finally results in the break of homophily unnoticeability.
The main reason for this phenomenon is mainly because of the poorly defined unnoticeability in graph adversarial attack.
Without a proper definition, the developed attacks tend to the shortcut to incur damage instead of capturing the true underlying vulnerability of GNNs.
Consequently, using these attacks to evaluate robustness of GNNs will bring unreliable results thus hindering the development of trustworthy GNNs.

To be more specific, due to the poor unnoticeability constraint for graph adversarial learning, the developed attacks tend to leverage the shortcuts to greatly destroy the original homophily, which leads to the break of unnoticeability. Thus, using homophily defenders can easily defend these seemingly powerful attacks, even with a simple design, which however brings us unreliable conclusions about the robustness of homophily defenders.
Essentially, HAO penalizes GIA attacks that take the shortcuts, and retain their unnoticeability in terms of homophily.
Thus, HAO mitigates the shortcut issue of GIA attacks, urges the attacks to capture the underlying vulnerability of GNNs
and brings us a more reliable evaluation result, from which we know simple homophily defenders are essentially not robust GNNs.

In addition, the proposed realization of unnoticeability check for adversarial attacks provides another approach to instantiate the unnoticeability. Especially for the domains that we can hardly leverage inductive bias from human, we can still try to identify their homophily, or the underlying rationales/causality of the data generation process, e.g., grammar correctness, fluency and semantics for natural languages, to instantiate the unnoticeability constraint with the help of external examiners. Since people are likely to be more sensitive to quantitative numbers like accuracy, those external examiners can be conveniently leveraged to the corresponding benchmark or leaderboards to further benefit the community.

\textbf{Limitations of HAO and future implications.} Since HAO are mostly developed to preserve the homophily unnoticeability, it counters the greedy strategy of attacks without HAO that destroys the homophily to incur more damage.
Therefore, it will inevitably reduce the damage of the attacks without HAO against vanilla GNNs.
As observed from the experiments, we find HAO essentially trades the attack performance when against vanilla GNNs for the performance when against homophily defenders.
As Fig.~\ref{CH:HAO:fig:ablation_hao_coe} shows, the trade-off effects can be further amplified with a large coefficient lambda in HAO.
As also shown by Fig.~\ref{CH:HAO:fig:ablation_node_budgets} and Fig.~\ref{CH:HAO:fig:ablation_edge_budgets}, when against vanilla GNNs, compared with GIA without HAO, GIA with HAO show fewer threats.
In certain cases, the trade-off might generate the performance of attacks.
Thus, it calls for more tailored optimization methods to solve for better injection matrix and node features in the future. Moreover, the trade-off effects also reflect the importance of homophily to the performance of node classifications and the utility of homophily unnoticeability, where we believe future theoretical works can further study this phenomenon and reveal the underlying causality for node classification or even more other downstream tasks. Thus, we can develop more robust and trustworthy neural graph models that do not depend on spurious correlations to perform the task.

In addition, as homophilous graph is the most common class of graph benchmarks for node classification~\citep{cora,citeseer,ogb,grb}, our discussions are mostly focused on this specific class of graphs.
However, when applying HAO to other classes of graphs such as non-attributed graphs, a direct adaption of HAO may not work.
Nevertheless, if the underlying information for making correct predictions still resemble the homophily property,
for example, in a non-attributed graph, nodes with similar structures tend to have similar labels,
it is still promising to introduce the node features with node embeddings, derive a new definition of homophily and apply HAO.
Moreover, recently disassortative graphs appear to be interesting to the community~\citep{geom_gcn,beyond_homophily},
which exhibit heterophily property that neighbors tend to have dissimilar labels, in contrast to homophily.
We conduct an investigation on this specific class of graphs and detailed results are given in Table~\ref{CH:HAO:tab:eval_non_targeted_nonhomo_appdx},
from which we surprisingly find HAO still maintains the advance when incorporating various GIA attacks.
The reason might be that GNNs and GIA with HAO can still implicitly learn the homophily such as the similarity between class label distributions~\citep{is_homophily_necessity}, even without explicit definitions.
To summarize, we believe future extension of HAO to other classes of graphs is another interesting direction.

Besides, the discussions in this paper are only considered the relationship between adversarial robustness and homophily.
However, label noises are another widely existing threats that are deserved to be paid attention to~\citep{label_noise_reweight,coteaching,sigua,label_noise_survey}.
Essentially, our discussions in Appendix~\ref{CH:HAO:sec:Memorization Effects of GNNs}
are also closely related to the vulnerability of GNNs to label noises, where GNNs can still achieve
near-perfect fitting to the datasets with full label noises. Thus, it is desirable to broaden the attention
and discussion to include the label noises when developing trustworthy GNNs.

\subsection{More Future Directions}
\label{CH:HAO:sec:future_direction_appdx}

Besides the future implications inspired by the limitations of HAO, we believe there are also many promising future works that could be built upon HAO.

\textbf{Rethinking the definition of unnoticeability in adversarial robustness.}
Though the study of adversarial robustness was initially developed around the deep learning models on image classification~\citep{intriguing,fgsm,pgd},
images and classification are far from the only data and the task we want to build neural networks for.
Deep learning models are widely applied to other types of data, such as natural languages and graphs,
where human inductive bias can hardly be leveraged to elaborate a proper definition of unnoticeability.
Moreover, for more complicated tasks involving implicit reasoning, even in the domain of images,
the original definition of unnoticeability, i.e., L-p norm, may not be sufficient to
secure all shortcuts that can be leveraged by adversaries.
How to establish and justify a proper definition of unnoticeability in these domains and tasks,
is critical for developing trustworthy deep learning models.

\textbf{Applications to other downstream tasks.}
Given the wide applications of GNNs, we believe the studies on the robustness of GNNs should be extended to other downstream tasks, such as link predictions and graph clustering.
Specifically, when with a different task objective, it is interesting to find whether the underlying task still depends on the homophily property and
how the different optimization objectives affect the attack optimization trajectory.

\textbf{Attack with small budgets.}
In real-life scenarios, the budgets of the adversary may be limited to a small number.
It is interesting to study how to maximize the damage given limited budgets and its interplay between homophily.
For example, how to locate the most vulnerable targets. We show an initial example through ATDGIA.

\textbf{Mix-up attack of GMA and GIA.}
In real-life scenarios, both GMA and GIA could happen with different budget limits.
It is interesting to see whether and how they could be combined to conduct more powerful attacks.

\textbf{Injection for defense.}
Actually, not only attackers can inject extra nodes, but defenders can also inject some nodes to
promote the robustness of the networks. For example, according to the Proposition.~\ref{CH:HAO:thm:gia_cer_robo},
nodes with higher degrees, higher MLP decision margin, and higher homophily tend more unlikely to be attacked.
Hence, defenders may directly inject some nodes the promote the above properties of vulnerable nodes.

\textbf{Attacks on more complicated and deep GNNs.}
Most existing graph adversarial works focus on analyzing linearized GNNs and apply the discoveries to more complex cases.
However, with the development of deep learning and GNNs, some models with complicated structures fail to fit those theories. For example, methods developed by studying linearized GNNs can hardly adapt to GNNs with normalizations as also revealed from our experiments. Then they can even more hardly be adapted to more complex models such as Transformers. On the other hand, most graph adversarial studies only focus on relatively shallow GNNs. Different from other deep learning models, as GNNs go deep, besides more parameters, they also require an exponentially growing number of neighbors as inputs. How the number of layers would affect their robustness and the threats of attacks remain unexplored.
From both theoretical and empirical perspectives, we believe it is very interesting to study the interplay between the number of GNN layers and homophily, in terms of adversarial robustness and threats, and how to leverage the discoveries to probe the weakness of complicated models.

\textbf{Reinforcement Learning based GIA.}
Reinforcement learning based approaches are shown to exhibit promising performances in previously mixed settings~\citep{rls2v,nipa}.
Though we exclude them for the efforts needed to adapt them to our setting, we believe it is promising and interesting to
incorporate reinforcement learning to develop more tailored injection strategies and vulnerable node selection.
Meanwhile, it is also interesting to explore how to leverage the idea of SeqGIA proposed in Sec.~\ref{CH:HAO:sec:gia_hao} to reduce the computation overhead of
reinforcement learning approaches and enhance their scalability.

\section{More Details and Reasons about the Graph Adversarial Attack Setting}
\label{CH:HAO:sec:adv_setting_appdx}

We provide more details about the perturbation constraints
and the threat model used in Sec.~\ref{CH:HAO:sec:adv_setting}.

\subsection{Perturbation Constraints}
Following previous works~\citep{nettack,tdgia}, Graph adversarial attacks can be characterized into graph modification attacks and graph injection attacks by their perturbation constraints.
Moreover, we adopt standardization methods (i.e., arctan transformation) following Graph Robustness Benchmark~\citep{grb} on input features $X$.

\textbf{Graph Modification Attack (GMA).} GMA generates $\gG'$ by modifying the graph structure $A$ and the node features $X$ of the original graph $\gG$. The most widely adopted constraint in GMA is to limit the number of perturbations on $A$ and $X$, denoted by $\triangle_A$ and $\triangle_X$, respectively, as:
\begin{equation}
	\label{CH:HAO:eq:gma_cons_appdx}
	\triangle_A+\triangle_X\leq \triangle\in\sZ, \norm{A'-A}_0 \leq \triangle_A\in \sZ, \norm{X'-X}_\infty \leq \epsilon \in \R,
\end{equation}
where the perturbation on $X$ is bounded by $\epsilon$ via L-p norm, since we are using continuous features.

\textbf{Graph Injection Attack (GIA).} Differently, GIA generates $\gG'$ by injecting a set of malicious nodes $V_{\text{atk}}$ as:
\begin{equation}
	X'=\begin{bmatrix}
		\ X\hfill        \\
		\ X_{\text{atk}} \\
	\end{bmatrix},
	A'=\begin{bmatrix}
		\ A\hfill          & A_{\text{atk}} \\
		\ A_{\text{atk}}^T & O_{\text{atk}} \\
	\end{bmatrix},
\end{equation}
where $X_{\text{atk}}$ is the features of the injected nodes, $O_{\text{atk}}$ is the adjacency matrix among injected nodes, and $A_{\text{atk}}$ is the adjacency matrix between the injected nodes and the original nodes. Let $d_u$ denote the degree of node $u$, the constraints in GIA are:
\begin{equation}
	\label{CH:HAO:eq:gia_cons_appdx}
	|V_{\text{atk}}| \leq \triangle\in\sZ, \ 1\leq d_u\leq b\in\sZ,
	X_{u} \in \mathcal{D}_X \subseteq \R^d,
	\forall u\in V_{\text{atk}},
\end{equation}
where the number and degrees of the injected nodes are limited, $\mathcal{D}_X=\{C\in\R^d,\min(X)\cdot\mathbf{1}\leq C\leq \max(X)\cdot\mathbf{1} \}$ where $\min(X)$ and $\max(X)$ are the minimum and maximum entries in $X$ respectively.
In other words, each entry of the injected node features are bounded within the minimal entry and maximal entry of the original node feature matrix, following the previous setting~\citep{tdgia}.

\subsection{Threat Model}
\label{CH:HAO:sec:threat_model_appdx}
We adopt a unified setting which is also used by Graph Robustness Benchmark~\citep{grb},
that is evasion, inductive, and black-box. Next, we will elaborate on more details and reasons for adopting the setting.

\subsubsection{Details of the Threat Model}
\label{CH:HAO:sec:threat_model_detail_appdx}
\textbf{Evasion.} The attack only happens at test time, which means that defenders are able to obtain the original clean graph $\gG_{\text{train}}$ for training, while testing on a perturbed graph $\gG'$.
The reasons for adopting the evasion setting is as shown in Appendix~\ref{CH:HAO:sec:threat_model_reason_appdx}.

\textbf{Inductive.} The training and testing of GNNs is performed in an inductive manner. Specifically, $f_\theta$ is trained on the (sub) training graph $\gG_{\text{train}}$, which consists of
the training nodes with their labels and the edges among training nodes. While during testing, the model will access the whole graph $\gG_{\text{test}}=\gG$ for inferring the labels of test nodes.
In particular, $\gG$ consists of all of the nodes and the edges, including $\gG_{\text{train}}$, the test nodes, the edges among test nodes, and the edges between training nodes and the test nodes.
In contrast, if the training and testing are performed in a transductive manner, the model can access the whole graph during both training and testing, i.e., $\gG_{\text{train}}=\gG_{\text{test}}=\gG$.
Since we adopt the evasion setting where the adversary may modify the $\gG_{\text{test}}$ during testing,
the GNN has to be learned in an inductive manner. More reasons are as elaborated in Appendix~\ref{CH:HAO:sec:threat_model_reason_appdx}.

\textbf{Black-box.} The adversary has no information about the target model, but the adversary may obtain the graph and training labels to train a surrogate model for generating perturbed graph $\gG'$.

Combining all of the above, conducting effective attacks raises special challenges to adversaries, since defenders can adapt the information extracted from training graph $\gG_{\text{train}}$ to learn more robust hidden representations~\citep{robustgcn}, or learn to drop noisy edges~\citep{advsample_deepinsights,gnnguard,prognn}, or even perform adversarial training~\citep{laten_adv,dyadv} which is known as one of the strongest defense mechanisms in the domain of images~\citep{fgsm,pgd}.

\subsubsection{Discussions about the Threat Model}
\label{CH:HAO:sec:threat_model_reason_appdx}
Different from images where we can adopt the inductive bias from the human vision system to use numerical constraints, i.e., L-p norm, to bound the perturbation range~\citep{fgsm,pgd}, we cannot use similar numerical constraints to define the unnoticeability for graphs, as they are \textit{weakly correlated} to the information required for node classification. For example, previous work~\citep{nettack} tries to use node degree distribution changes as the unnoticeability constraints. However, given the same degree distribution, we can shuffle the node features to generate multiple graphs with completely different semantic meanings, which disables the functionality of unnoticeability.

Because of the difficulty of properly defining the unnoticeability of graphs, adopting a poisoning setting in graph adversarial attack will enlarge the gap between research and practice. Specifically, poisoning attacks require an appropriate definition of unnoticeability so that the defenders are able to distinguish highly poisoned data from unnoticeable poisoned data and the original data. Otherwise, attackers can always leverage some underlying shortcuts implied by the poorly defined unnoticeability, i.e., homophily in our case, to perform the attacks, since the defenders are blind to these shortcuts. On the other hand, leveraging shortcuts may generate data that is unlikely to appear in real-world applications. For example, in a citation network, medical papers are unlikely to cite or be cited by linguistic papers while the attacks may modify the graphs or inject malicious nodes to make medical papers cite or be cited by lots of linguistic papers, which is apparently impractical. Using these attacks to evaluate the robustness of GNNs may bring unreliable conclusions, i.e., homophily defenders in our case, which will greatly hinder the development of trustworthy GNNs.

Moreover, under a poor unnoticeability definition, without the presence of the original data, defenders have no idea to what extent the data is poisoned and whether the original labels remain the correspondence. Furthermore, it is well-known that neural networks have universal approximation power~\citep{mlp_approx}, thus can easily overfit the training set~\citep{dl_book}, or even \textit{memorize} the labels appeared during training~\citep{memorization}. As a generalization from deep learning models to graphs, GNNs tend to exhibit similar behaviors, which is shown empirically in our experiments (See Appendix \ref{CH:HAO:sec:Memorization Effects of GNNs} for details). Thus, even trained on a highly poisoned graph, GNNs may still converge to 100\% training accuracy, even though the correspondence between the data and the underlying labels might be totally corrupted. In this case, defenders can hardly distinguish whether the training graph is perturbed hence unlikely to make any effective defenses. Besides, studying the robustness of GNNs trained from such highly poisoned graphs seems to be impractical, since real-world trainers are unlikely to use such highly poisoned data to train GNNs.

While in an evasion setting, the defenders are able to use the training graph to tell whether the incoming data is heavily perturbed and make some effective defenses, even simply leveraging some feature statistics~\citep{advsample_deepinsights,prognn}. Notably, A recent benchmark~\citep{grb} also has similar positions. Thus, we will focus on the evasion setting in this paper.

Given the evasion setting, GNNs can only perform inductive learning where the test nodes and edges are not visible during training. The reason is that, transductive learning (i.e., the whole graph except test labels is available), requires the training graph and test graph to be the same. However, it can not be satisfied as the adversary will modify the test graph, i.e., changing some nodes or edges during GMA attack, or injecting new malicious nodes during GIA attack. Additionally, inductive learning has many practical scenarios. For example, in an academic network, the graph grows larger and larger day by day as new papers are published and added to the original network. GNN models must be inductive to be applied to such evolving graphs.

\subsection{Memorization Effects of Graph Neural Networks}
\label{CH:HAO:sec:Memorization Effects of GNNs}
\begin{figure}[h]
	\centering
	\subfigure[Original labels]{
		\includegraphics[width=0.3\textwidth]{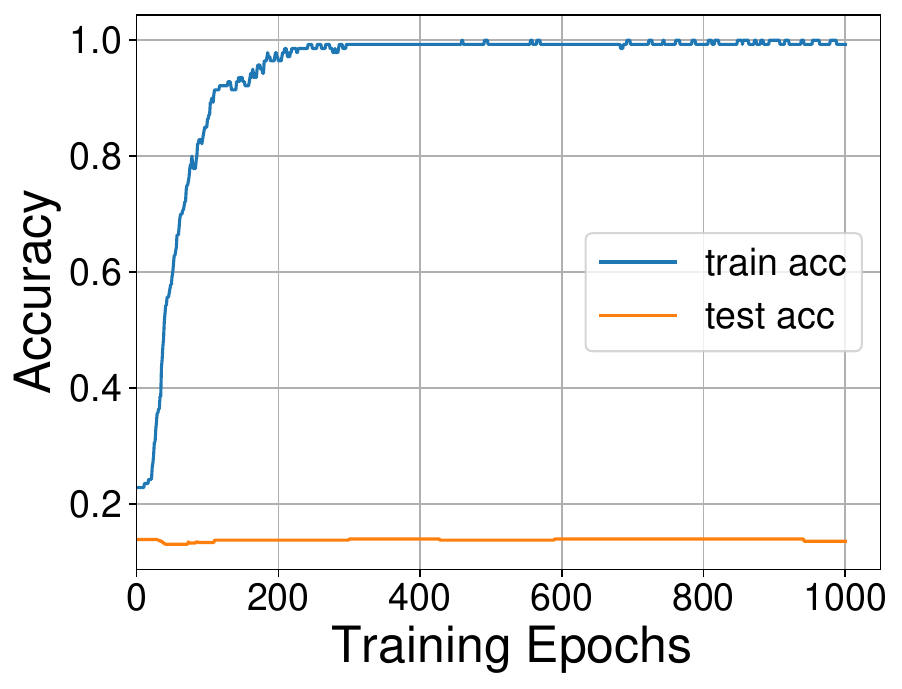}
	}
	\hfill
	\subfigure[Random labels]{
		\includegraphics[width=0.3\textwidth]{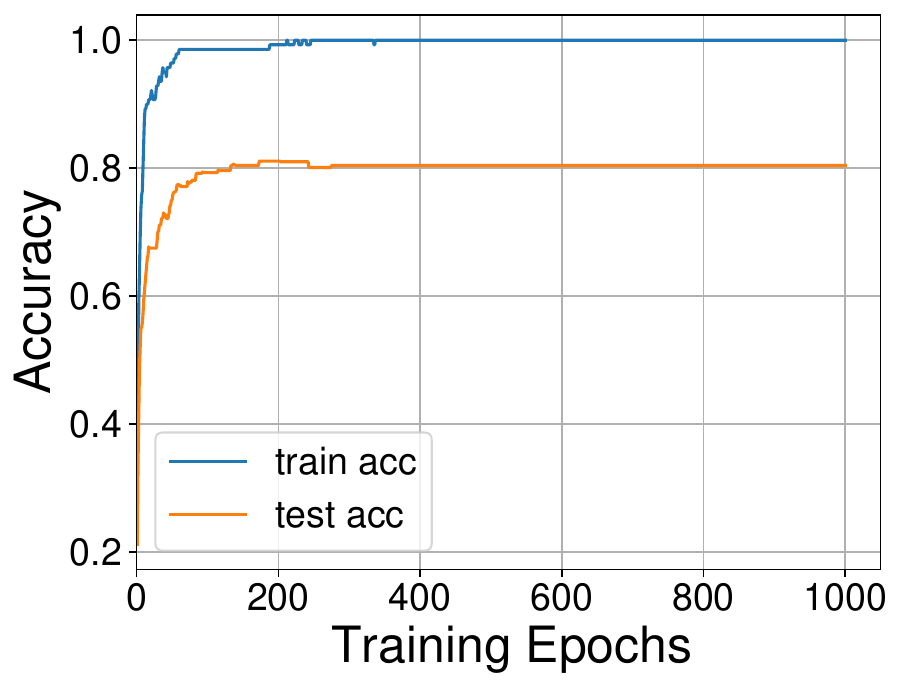}
	}
	\hfill
	\subfigure[Partial random labels]{
		\includegraphics[width=0.3\textwidth]{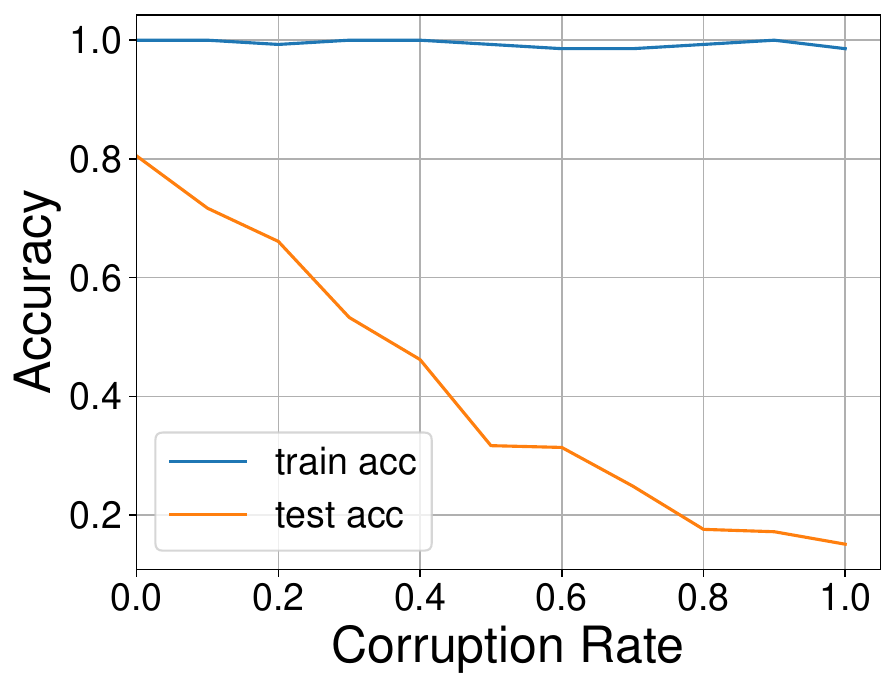}
	}
	\caption{Training curve of GCN on Cora with random labels}
	\label{CH:HAO:fig:Memorization Effects of GNNs}
\end{figure}
We conduct experiments with GCN \citep{gcn} on Cora \citep{cora}. The architecture we select is a 2-layer GCN with 16 hidden units, optimized using Adam \citep{adam} with a learning rate of $0.01$ and a $L_2$ weight decay of $5\times10^{-4}$ for the first layer. We train $1000$ epochs and report the training accuracy and test accuracy according to the best validation accuracy. We randomly sample a certain percentage of nodes from the whole graph and reset their labels. It can be seen from Fig.~\ref{CH:HAO:fig:Memorization Effects of GNNs} (b) and (c) that even with all random labels, the training accuracy can reach to nearly $100\%$, which serves as strong evidence for the existence of memorization effects in GNNs. In other words, even if a GNN is trained on a heavily poisoned graph (changes dramatically in the sense of semantics), it can still achieve good training accuracy while the defender has no way to explicitly find it or do anything about it. That is against the original setting and purpose of adversarial attacks \citep{intriguing,fgsm,pgd}. Thus, it urges the community for a proper solution to the ill-defined unnoticeability in current graph adversarial learning. Till the appearance of a silver bullet for unnoticeability on graphs, an evasion attack can serve as a better solution than a poisoning attack.

\section{More Details about GIA and GMA Comparison}
\label{CH:HAO:sec:gma_gia_comparison_appdx}
\subsection{Implementation of Graph Modification Attack}
Following Metattack~\citep{metattack}, we implement Graph Modification Attack by taking $A$ as a hyper-parameter.
Nevertheless, since we are conducting the evasion attack, we do not have meta-gradients but the gradient of $A$ with respect to $\mathcal{L}_\atk$, or $\nabla_{A}\mathcal{L}_\atk$.
In each step, we take the maximum entry in $\nabla_{A}\mathcal{L}_\atk$, denoted with $\max(\nabla_{A}\mathcal{L}_\atk)$, and change the corresponding edge, if it is not contained in the training graph.
Then we perform the perturbation as follows:
\begin{enumerate}[label=(\alph*)]
	\item If $\max(\nabla_{A}\mathcal{L}_\atk)\leq 0$ and the corresponding entry in $A$ is $0$, i.e., the edge does not exist before, we will add the edge.
	\item If $\max(\nabla_{A}\mathcal{L}_\atk)\geq 0$ and the corresponding entry in $A$ is $1$, i.e., the edge exists before, we will remove the edge.
\end{enumerate}
If the selected entry can satisfy neither of the above conditions, we will take the next maximum entry to perform the above procedure until we find one that satisfies the conditions.
Here we exclude perturbations on node features given limited budgets, since \cite{advsample_deepinsights} observed the edge perturbations produce more harm than node perturbations.
Besides, as shown in the proof, the damage brought by perturbations on node features is at most the damage brought by a corresponding injection to the targets in GIA,
hence when given the same budgets to compare GMA and GIA, we can exclude the perturbations on nodes without loss of generality.
Note that given the definitions of direct attack and influencer attack in Nettack~\citep{nettack}, our theoretical discussions are applicable to both direct GMA attack and indirect/influencer GMA attack, since the results are derived by establishing mappings between each kind of perturbations in GMA attack that are agnostic to these two types of GMA attacks. Moreover, the GMA attack evaluated in our experiments is exactly the direct attack.
As in our case, all of the test nodes become victim nodes and the adversary is allowed to modify the connections and features of these nodes to perform the attack.

\subsection{Implementation of Graph Injection Attack with Plural Mapping}
GIA with $\mathcal{M}_2$ is implemented based on the GMA above.
For each edge that appears in the perturbed graph produced by GMA but does not exist in the original graph, in GIA,
we will inject a node to connect with the corresponding nodes of the edge. After injecting all of the nodes, then we use PGD~\citep{pgd} to optimize the features of the injected nodes.

\section{More Homophily Distributions}
\subsection{Edge-Centric Homophily}
\label{CH:HAO:sec:homophily_edge_appdx}
\begin{figure}[ht]
	\centering
	\subfigure[Cora]{
		\includegraphics[width=0.315\textwidth]{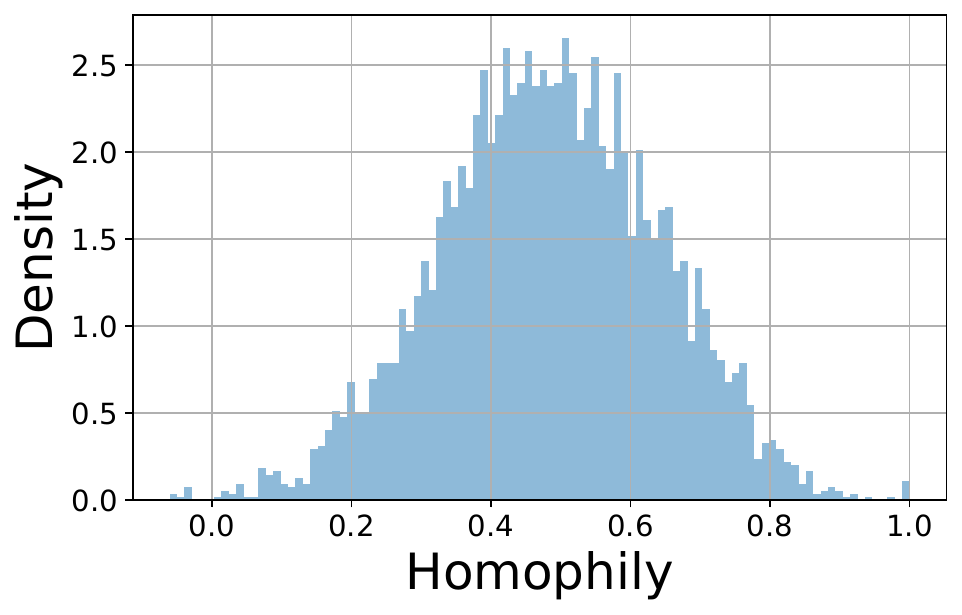}
	}
	\subfigure[Computers]{
		\includegraphics[width=0.32\textwidth]{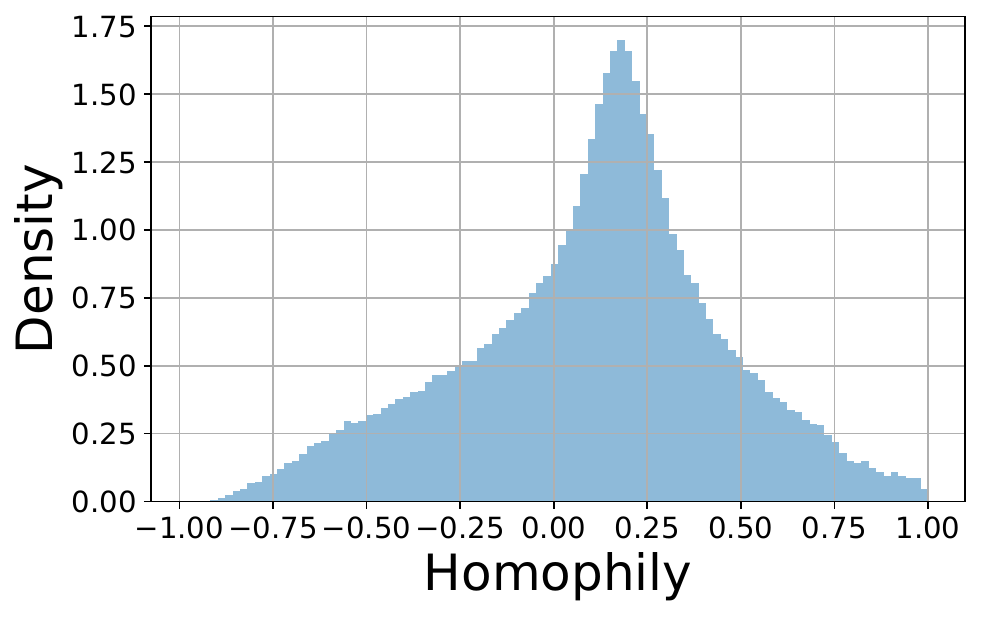}
	}
	\subfigure[Arxiv]{
		\includegraphics[width=0.305\textwidth]{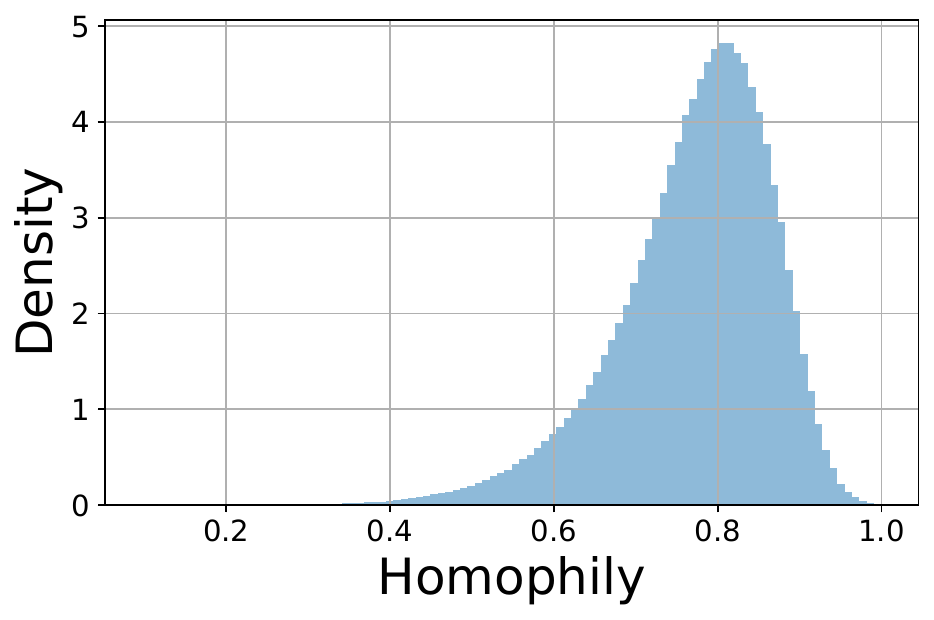}
	}
	\caption{Edge-Centric homophily distributions}
	\label{CH:HAO:fig:gia_homophily_dist_edge}
\end{figure}

In addition to node-centric homophily (Def.~\ref{CH:HAO:eq:homophily_node}), we can also define edge-centric homophily as:
\begin{definition}[Edge-Centric Homophily]
	The homophily for an edge $(u,v)$ can be defined as.
	\begin{equation}
		\label{CH:HAO:eq:homophily_edge}
		h_e=\text{sim}(X_u,X_v),
	\end{equation}
	where $\text{sim}(\cdot)$ is also a distance metric, e.g., cosine similarity.
\end{definition}
With the definition above, we can probe the natural edge-centric homophily distribution of real-world benchmarks, as shown in Fig.~\ref{CH:HAO:fig:gia_homophily_dist_edge}.
It turns out that the edge-centric homophily distributes follows a Gaussian prior.
However, it seems to be improper to utilize edge-centric homophily to instantiate the homophily unnoticeability for several reasons.
On the one hand, edge similarity does not consider the degrees of the neighbors which is misaligned with the popular aggregation scheme of GNNs.
On the other hand, edge-centric and node-centric homophily basically perform similar functionality to retain the homophily,
but if considering the future extension to high-order neighbor relationships, edge similarity might be harder to extend than node-centric homophily.
Thus, we utilize the node-centric homophily for most of our discussions.
\begin{figure}[ht]
	\centering
	\subfigure[Cora]{
		\includegraphics[width=0.31\textwidth]{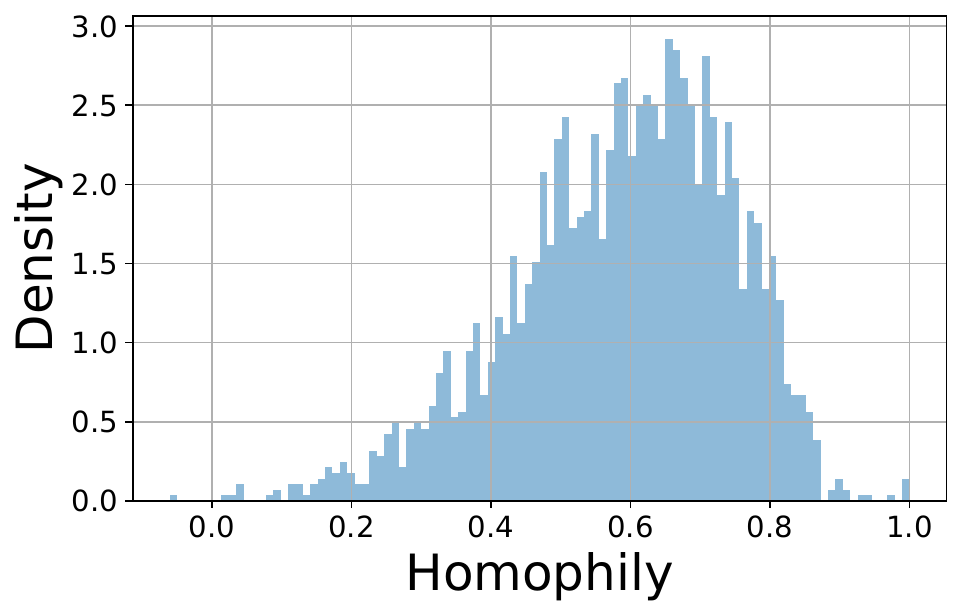}
	}
	\subfigure[Computers]{
		\includegraphics[width=0.31\textwidth]{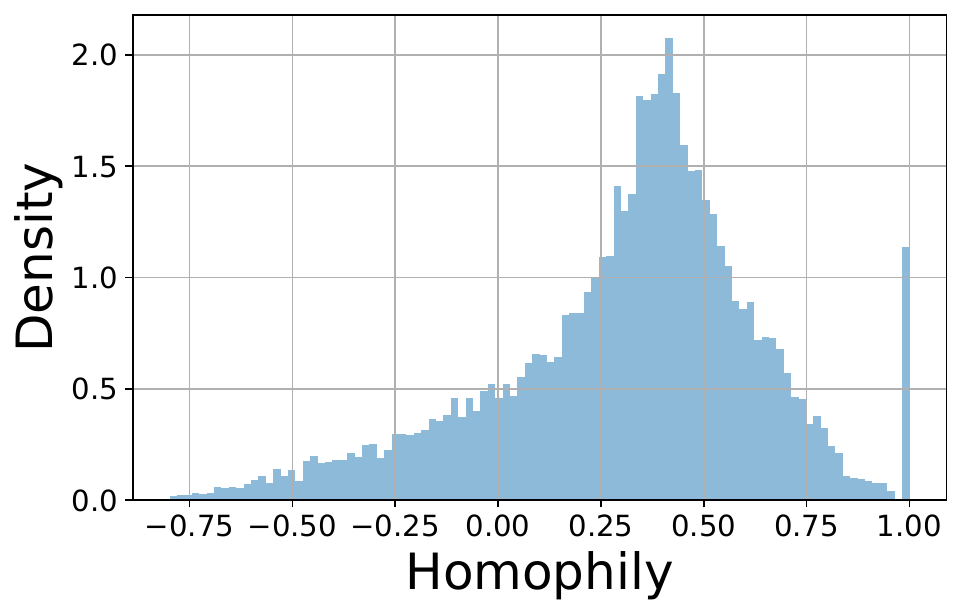}
	}
	\subfigure[Arxiv]{
		\includegraphics[width=0.31\textwidth]{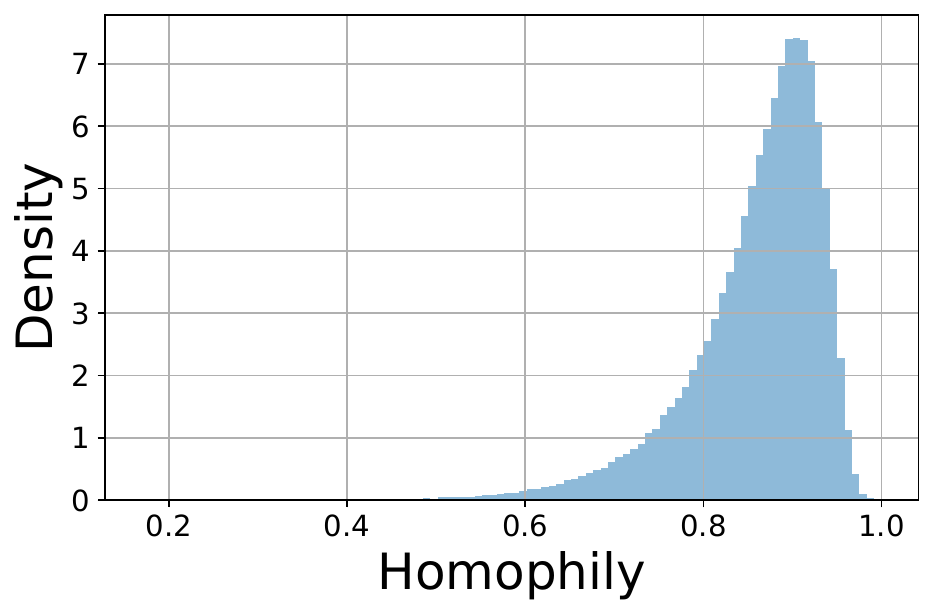}
	}
	\caption{Homophily distributions before attack}
	\label{CH:HAO:fig:gia_homophily_dist_node_atk_appdx_1}
\end{figure}

\begin{figure}[ht]
	\centering
	\subfigure[Cora]{
		\includegraphics[width=0.32\textwidth]{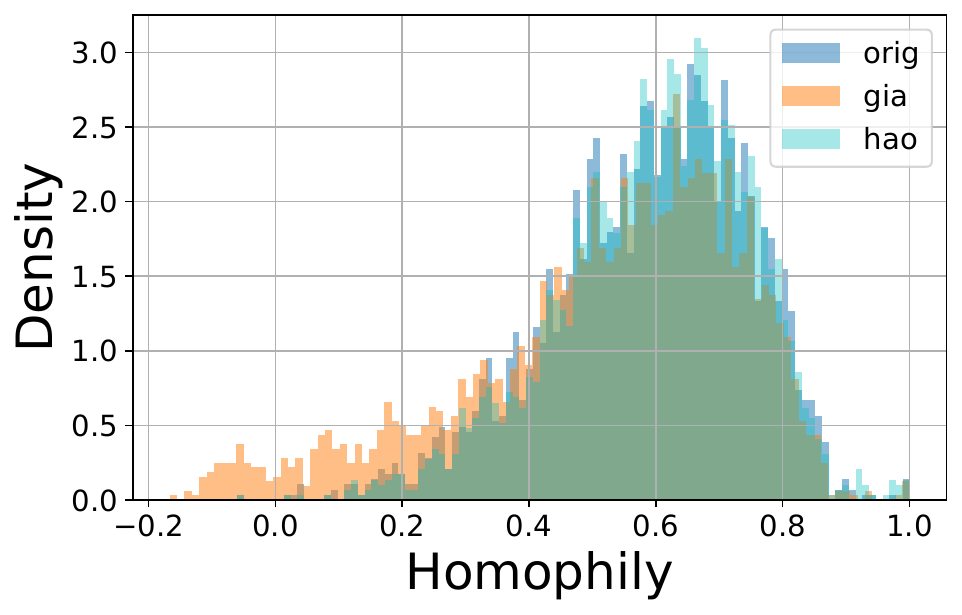}
	}
	\subfigure[Computers]{
		\includegraphics[width=0.32\textwidth]{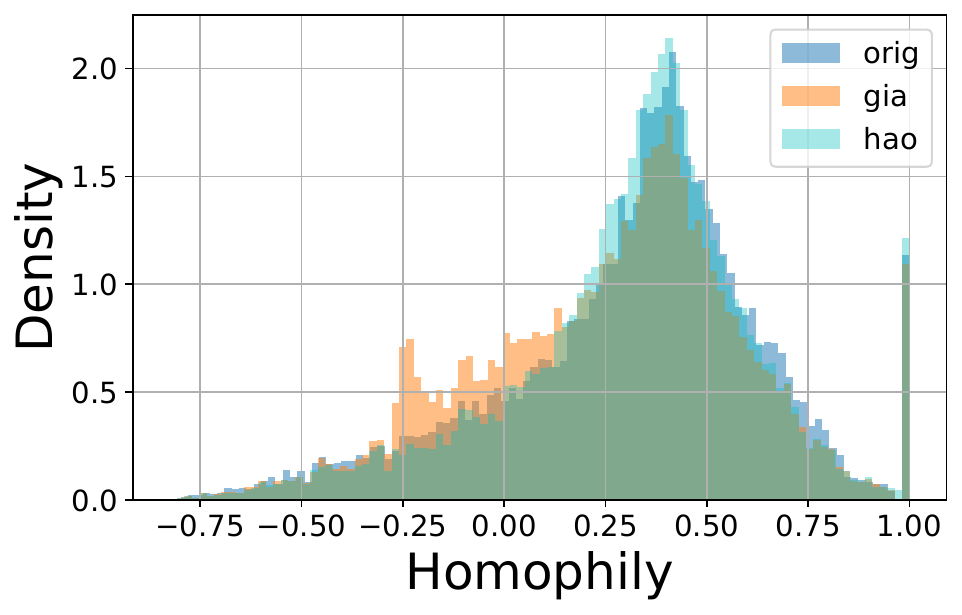}
	}
	\subfigure[Arxiv]{
		\includegraphics[width=0.30\textwidth]{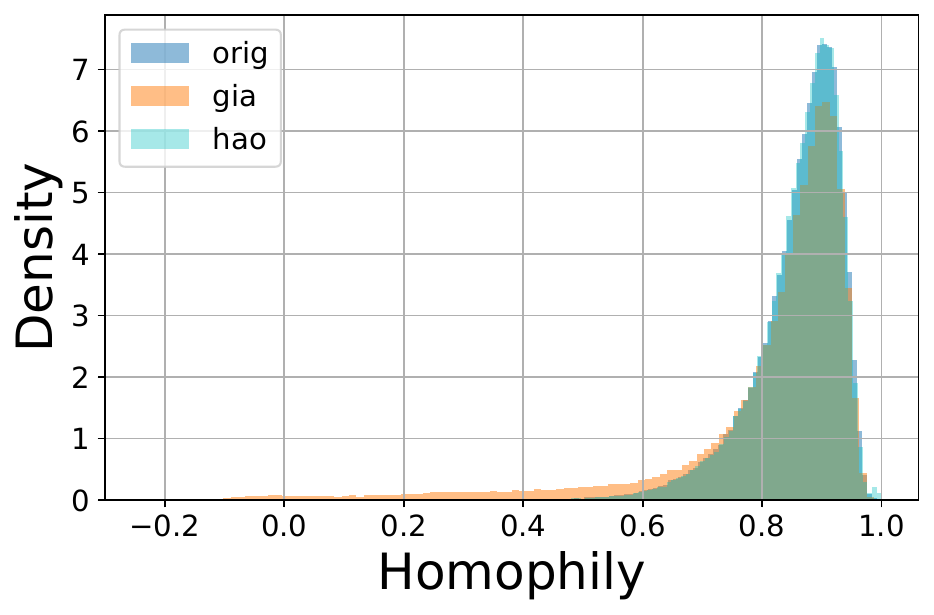}
	}
	\caption{Homophily distributions after attack}
	\label{CH:HAO:fig:gia_homophily_dist_node_atk_appdx_2}
\end{figure}
\subsection{More Homophily Distributions Changes}
\label{CH:HAO:sec:homophily_appdx}
We provide more homophily distribution results of the benchmarks we used in the experiments for Cora, Computers and Arxiv, shown as in Fig.~\ref{CH:HAO:fig:gia_homophily_dist_node_atk_appdx_1} and Fig.~\ref{CH:HAO:fig:gia_homophily_dist_node_atk_appdx_2}, respectively.
GIA is implemented with TDGIA~\citep{tdgia}.
Note that the budgets for TDGIA here is different from that in the previous sections, which utilized the budgets resulting in the maximum harm when compared with GMA.
Similarly, GIA without HAO would severely break the original homophily distribution hence making GIA can be easily defended by homophily defenders.
While incorporated with HAO, GIA would retain the original homophily during attack.

\section{Proofs and Discussions of Theorems}
\label{CH:HAO:sec:proof_disscussion_theory}
\subsection{Proof for Theorem~\ref{CH:HAO:thm:gia_threat}}
\begin{nono-theorem}
Given moderate perturbation budgets $\triangle_\gia$ for GIA and $\triangle_\gma$ for GMA, that is,
let $\triangle_\gia \leq \triangle_\gma \ll |V|\leq |E|$,
for a fixed linearized GNN $f_\theta$ trained on $\gG$,
assume that $\gG$ has no isolated nodes,
and both GIA and GMA adversaries follow the optimal strategy,
then, \mbox{$\forall \triangle_\gma>0, \exists \triangle_\gia\leq \triangle_\gma$}, such that:
\[
	\mathcal{L}_{\atk}(f_\theta(\gG'_\gia))-\mathcal{L}_{\atk}(f_\theta(\gG'_\gma))\leq 0,
\]
where $\gG'_\gia$ and $\gG'_\gma$ are the perturbed graphs generated by GIA and GMA, respectively.
\end{nono-theorem}
\begin{proof}
	\label{CH:HAO:proof:gia_threat}

	The proof sketch is to show that,
	\begin{enumerate}[label=(\alph*)]
		\item Assume the given GNN model has $k$ layers, there exists a mapping, that when given the same budget, i.e., $\triangle_\gia= \triangle_\gma \ll |V|\leq |E|$, for each perturbation generated by GMA intended to attack node $u$ by perturbing edge $(u,v)$, or node attributes of node $u$ or some node $v$ that connects to $u$ within $k$ hops, we can always map it to a corresponding injection attack, that injects node $x_w$ to attack $u$, and lead to the same effects to the prediction.
		\item When the number of perturbation budget increases, the optimal objective values achieved of GIA is monotonically non-increasing with respect to $\triangle_\gia$, that is
		      \[
			      \mathcal{L}^{k+1}_{\text{atk}}(f_\theta(\gG'_{\text{GIA}}))\leq \mathcal{L}^{k}_{\text{atk}}(f_\theta(\gG'_{\text{GIA}})),
		      \]where $\mathcal{L}^{k}_{\text{atk}}(f_\theta(\gG'_{\text{GIA}}))$ is the optimal value achieved under the perturbation budget of $k$, which is obvious.
	\end{enumerate}
	Once we prove both (a) and (b), the $\mathcal{L}_{\text{atk}}(f_\theta(\gG'_{\text{GIA}}))$ will approach to $\mathcal{L}^{k}_{\text{atk}}(f_\theta(\gG'_{\text{GMA}}))$ from the above as $\triangle_\gia$ approaches to $\triangle_\gma$, hence proving Theorem~\ref{CH:HAO:thm:gia_threat}.
	Furthermore, for the flexibility of the constraints on $X_w$, we may adopt the gradient information of $X_w$ with respect to $\mathcal{L}_{\text{atk}}(f_\theta(\gG'_{\text{GIA}}))$ to further optimize $X_w$ and make more damages. Hence, we have $\mathcal{L}_{\text{atk}}(f_\theta(\gG'_{\text{GIA}}))\leq \mathcal{L}^{k}_{\text{atk}}(f_\theta(\gG'_{\text{GMA}}))$.

	To prove (a), the key technique is to show that, under a predefined mapping, there exist a corresponding injection matrix $A_\atk$ along with the features of the injected nodes $X_\atk$,  such that the GIA adversary can cause the same damage as GMA. The definition of the mapping mostly derives how the injection matrix is generated. While for the generation of $X_\atk$, note that all of the input features $X$ is normalized to a specific range within $[-f_l,f_r]$ where $f_l,f_r\geq 0$, following previous works~\citep{grb}. Thus, for any features $X_v \in \mathcal{D}_X$, $\alpha X_v \in \mathcal{D}_X$ when $0 \leq \alpha \leq 1$. We will use the statement multiple times during the derivation of $X_\atk$.

	Next, we will start to prove (a). Following~\citet{advsample_deepinsights}, in GMA, adding new connections between nodes from different classes produces the most benefit to the adversarial objective.
	Hence, given the limited perturbation budget, we give our primary focus to the action of connecting nodes from different classes and
	will prove (a) also holds for the remaining two actions, i.e., edge deletion and node attribute perturbation.

	We prove (a) by induction on the number of linearized layers.
	First of all, we will show prove (a) holds for 1-layer and 2-layer linearized GNN as a motivating example. The model is as $f_\theta=\hat{A}^2X\Theta$ with $H=\hat{A}X\Theta$ and $Z=f_\theta$.

	\textbf{Plural Mapping $\mathcal{M}_2$.} Here we define the mapping $\mathcal{M}_2$ for edge addition. For each edge perturbation pair $(u,v)$ generated by GMA, we can insert a new node $w$ to connect $u$ and $v$.
	The influence of adversaries can be identified as follows, as $\Theta$ is fixed, we may exclude it for simplicity:
	\\
	In layer (1):
	\begin{itemize}
		\item Clean graph:
		      \begin{equation}
			      H_i = \sum_{t \in \mathcal{N}(i)\cup \{i\}}\frac{1}{\sqrt{d_i d_t}}X_t
		      \end{equation}
		\item GMA:
		      \begin{equation}
			      H'_i = \left\{
			      \begin{aligned}
				       & \sum_{t\in \mathcal{N}(i)\cup \{i\}}\frac{1}{\sqrt{d_t (d_i+1)}}X_t + \frac{1}{\sqrt{d_v(d_i+1)}}X_v, & i \in \{u\}      \\
				       & \sum_{t\in \mathcal{N}(i)\cup \{i\}}\frac{1}{\sqrt{d_t (d_i+1)}}X_t + \frac{1}{\sqrt{d_u(d_i+1)}}X_u, & i \in \{v\}      \\
				       & H_i,                                                                                                  & i \notin \{u,v\} \\
			      \end{aligned}
			      \right.
		      \end{equation}
		\item GIA:
		      \begin{equation}H''_i = \left\{
			      \begin{aligned}
				       & \sum_{t \in \mathcal{N}(i)\cup \{i\}}\frac{1}{\sqrt{d_t (d_i+1)}}X_t+\frac{1}{\sqrt{3(d_i+1)}}X_w, & i \in \{u,v\}      \\
				       & H_i,                                                                                               & u \notin \{u,v,w\} \\
				       & \frac{1}{\sqrt{3}}(\frac{1}{\sqrt{d_u+1}}X_u+\frac{1}{\sqrt{d_v+1}}X_v+\frac{1}{\sqrt{3}}X_w),     & i \in \{w\}        \\
			      \end{aligned}
			      \right.
		      \end{equation}
	\end{itemize}
	where $d_i$ refers to the degree of node $i$ with self-loops added for simplicity.
	Thus, in layer (1), to make the influence from GMA and GIA on node $u$ equal, the following constraint has to be satisfied:
	\begin{equation}
		\frac{1}{\sqrt{3(d_u+1)}}X_w = \frac{1}{\sqrt{(d_v+1)(d_u+1)}}X_v, \\
	\end{equation}
	which is trivially held by setting
	\begin{equation}X_w =
		\frac{\sqrt{3}}{\sqrt{d_v+1}}X_v.
	\end{equation}
	Normally, GMA does not consider isolated nodes~\citep{nettack,metattack} hence we have $d_v\geq 2$ and $X_w\in\mathcal{D}_X$.
	Note that we can even change $X_w$ to make more affects to node $u$ with gradient information, then we may generate a more powerful perturbation in this way. \\
	Then, we go deeper to layer 2.
	In layer (2):
	\begin{itemize}
		\item Clean graph:
		      \begin{equation}
			      Z_i = \sum_{t \in \mathcal{N}(i)\cup \{i\}}\frac{1}{\sqrt{d_i d_t}}H_t
		      \end{equation}
		\item GMA:
		      \begin{equation}
			      Z'_i = \left\{
			      \begin{aligned}
				       & \sum_{t\in \mathcal{N}(i)}\frac{H_t}{\sqrt{d_t (d_i+1)}} +\frac{H'_i}{d_i+1}+ \frac{H'_v}{\sqrt{(d_v+1)(d_i+1)}}, & u \in \{u\}                              \\
				       & \sum_{t\in \mathcal{N}(i)}\frac{H_t}{\sqrt{d_t (d_i+1)}} +\frac{H'_i}{d_i+1}+ \frac{H'_u}{\sqrt{(d_u+1)(d_i+1)}}, & u \in \{v\}                              \\
				       & \sum_{t\in \mathcal{N}(i)}\frac{H'_t}{\sqrt{d_t (d_i+1)}},                                                        & u \in \mathcal{N}(u) \cup \mathcal{N}(v) \\
				       & Z_u,                                                                                                              & \text{otherwise}                         \\
			      \end{aligned}
			      \right.
		      \end{equation}
		\item GIA:
		      \begin{equation}Z''_i = \left\{
			      \begin{aligned}
				       & \sum_{t \in \mathcal{N}(i)}\frac{1}{\sqrt{d_t (d_i+1)}}H_t+\frac{1}{d_i+1}H''_i+\frac{1}{\sqrt{3(d_i+1)}}H''_w, & i \in \{u,v\}      \\
				       & H_u,                                                                                                            & i \notin \{u,v,w\} \\
				       & \frac{1}{\sqrt{3}}(\frac{1}{\sqrt{d_u+1}}H''_u+\frac{1}{\sqrt{d_v+1}}H''_v+\frac{1}{\sqrt{3}}H''_w),            & i \in \{w\}        \\
			      \end{aligned}
			      \right.
		      \end{equation}
	\end{itemize}
	Similarly, to make $Z'_u=Z''_u$, we have to satisfy the following constraint:
	\begin{equation}
		\begin{aligned}
			\frac{1}{d_u+1}H''_u+\frac{1}{\sqrt{3(d_u+1)}}H''_w                                                & = \frac{1}{d_u+1}H'_i+ \frac{1}{\sqrt{(d_v+1)(d_u+1)}}H'_v, \\
			\frac{\sqrt{3}}{d_u+1}\sum_{t \in \mathcal{N}(u)\cup \{u\}}\frac{1}{\sqrt{d_t }}X_t+\frac{4}{3}X_w & +
			\frac{1}{\sqrt{3}}(\frac{1}{\sqrt{d_u+1}}X_u+\frac{1}{\sqrt{d_v+1}}X_v)                                                                                          \\
			=                                                                                                                                                                \\
			\frac{\sqrt{3}}{d_u+1}\sum_{t\in \mathcal{N}(u)\cup \{u\}}\frac{X_t}{\sqrt{d_t}}                   & + \frac{\sqrt{3X_v}}{\sqrt{d_v+1}}  +                       \\
			\frac{\sqrt{3}}{\sqrt{d_v+1}}(\sum_{t\in \mathcal{N}(v)\cup \{v\}}\frac{X_t}{\sqrt{d_t (d_v+1)}}   & + \frac{X_u}{\sqrt{(d_u+1)(d_v+1)}}),                       \\
			\frac{4}{3}X_w                                                                                                        +
			\frac{1}{\sqrt{3}}(\frac{1}{\sqrt{d_u+1}}X_u                                                       & +\frac{1}{\sqrt{d_v+1}}X_v)                                 \\
			=                                                                                                                                                                \\
			\frac{\sqrt{3X_v}}{\sqrt{d_v+1}}                                                                                      +
			\frac{\sqrt{3}}{\sqrt{d_v+1}}(\sum_{t\in \mathcal{N}(v)\cup \{v\}}\frac{X_t}{\sqrt{d_t (d_v+1)}}   & + \frac{X_u}{\sqrt{(d_u+1)(d_v+1)}}),                       \\
		\end{aligned}
	\end{equation}
	then we let $X_w=\frac{3}{4}(\text{RHS}-\frac{1}{\sqrt{3}}(\frac{1}{\sqrt{d_u+1}}X_u+\frac{1}{\sqrt{d_v+1}}X_v))$ to get the solution of $X_w$ that makes the same perturbation.
	Similarly, we can infer $X_w \in \mathcal{D}_X$. The following proof also applies to layer 2.

	Next, we will prove that, for a linearized GNN with $k$ layers ($k\geq 1$), i.e., $H^{(k)}=\hat{A}^kX\Theta$, once $\exists X_w$, such that the predictions for node $u$ is the same to that perturbed by GMA, i.e., $H^{(k-1)}_u=E^{(k-1)}_u$, then $\exists X'_w$, such that $H^{(k)}_u=E^{(k)}_u$. Here we use $H$ to denote the prediction of GNN attacked by GMA and $E$ for that of GIA. Note that, once the theorem holds, as we have already proven the existence of such $X_w$, it naturally generalizes to an arbitrary number of layers.

	To be more specific, when $H^{(k-1)}_u=E^{(k-1)}_u$, we need to show that, $\exists X_w$, s.t.,
	\begin{equation}
		\label{CH:HAO:eq:klayer_con}
		\begin{aligned}
			H^{(k)}_u & = \sum_{j \in \mathcal{N}(u)}\frac{1}{\sqrt{d_u+1}\sqrt{d_j}}H_{j}^{(k-1)}+\frac{1}{d_u+1}H^{(k-1)}_u+\frac{1}{\sqrt{d_u+1}\sqrt{d_v+1}}H_v^{(k-1)}, \\
			E^{(k)}_u & = \sum_{j \in \mathcal{N}(u)}\frac{1}{\sqrt{d_u+1}\sqrt{d_j}}E_{j}^{(k-1)}+\frac{1}{d_u+1}E^{(k-1)}_u+\frac{1}{\sqrt{d_u+1}\sqrt{3}}E_w^{(k-1)},     \\
			H^{(k)}_u & = E^{(k)}_u.                                                                                                                                         \\
		\end{aligned}
	\end{equation}
	Here we make a simplification to re-write Eq.~\ref{CH:HAO:eq:klayer_con} by defining the influence score.
	\begin{definition}[Influence Score]
		The influence score from node $v$ to $u$ after $k$ neighbor aggregations with a fixed GNN following Eq.~\ref{CH:HAO:eq:gnn},
		is the weight for $X_v$ contributing to $H^{(k)}_u$:
		\begin{equation}
			\label{CH:HAO:eq:influence_score_appd}
			H^{(k)}_{u} = \sum_{j\in \mathcal{N}(u)\cup\{u\}}I^{k}_{uj}\cdot X_j,
		\end{equation}
		which can be calculated recursively through:
		\begin{equation}
			I^{k}_{uw} = \sum_{j\in \mathcal{N}(u)\cup\{u\}}(I_{uj}\cdot I^{(k-1)}_{jw})+I^{(k-1)}_{uw}.
		\end{equation}
	\end{definition}
	As $\Theta$ is fixed here, we can simply regard $I^k_{uv}=\hat{A}^k_{uv}$.
	Compared to the predictions after $k$-th propagation onto the clean graph, in GMA, $H^{(k)}_u$ is additionally influenced by node $v$, while in GIA, $H_u^{(k)}$ is additionally influenced by node $v$ and node $w$. Without loss of generality, we may absorb the influence from neighbors of node $v$ into that of node $v$.
	Hence we can rewrite Eq.~\ref{CH:HAO:eq:klayer_con} as the following:
	\begin{equation}
		\begin{aligned}
			\Delta H^{(k)}_u & = I^{k}_{\text{GMA}_{uv}}X_{v},                              \\
			\Delta E^{(k)}_u & = I^{k}_{\text{GIA}_{uv}}X_{v}+I^{k}_{\text{GIA}_{uw}}X_{w}, \\
			\Delta H^{(k)}_u & = \Delta E^{(k)}_u,                                          \\
		\end{aligned}
	\end{equation}
	where
	\[
		I^{k}_{\text{GIA}_{uv}} =
		\sum_{j \in \mathcal{N}(u)\cup\{u\}}I_{\text{GIA}_{uj}}\cdot I^{(k-1)}_{\text{GIA}_{jv}} +
		I_{\text{GIA}_{uw}}\cdot I^{(k-1)}_{\text{GIA}_{wv}}.
	\]
	Then we can further simplify it as,
	\begin{equation}
		\label{CH:HAO:eq:xw_sol}
		(I^{k}_{\text{GMA}_{uv}}-I^{k}_{\text{GIA}_{uv}})X_{v}=I^{k}_{\text{GIA}_{uw}}X_{w}.
	\end{equation}
	To show the existence of $X_w$ that solves the above equation, it suffices to show $I^{k}_{\text{GIA}_{uw}}\neq 0$ and $X_w\in \mathcal{D}_X$.
	Note that $\exists X_w$ s.t.,
	\begin{equation}
		(I^{(k-1)}_{\text{GMA}_{uv}}-I^{(k-1)}_{\text{GIA}_{uv}})X_{v}=I^{(k-1)}_{\text{GIA}_{uw}}X_{w}.
	\end{equation}
	Since $\hat{A}^k \geq \mathbf{0}, \forall k \geq 0$, so we have $I^{(k-1)}_{\text{GIA}_{uw}}> 0$. Moreover,
	\[
		I^{k}_{uw} = \sum_{j\in \mathcal{N}(u)\cup\{u\}}(\hat{A}_{uj}\cdot\hat{A}^{(k-1)}_{jw})+I^{(k-1)}_{uw},
	\]
	then it is obvious that the $I^{k}_{uw} > 0$.
	Moreover, with the definition of $I^k_{uv}=\hat{A}^k_{uv}$, it is obvious that
	$I^{(k-1)}_{\text{GIA}_{uw}}\geq I^{(k-1)}_{\text{GMA}_{uv}}$ for $v$ with a degree not less than $1$ (i.e., $v$ is not an isolated node).
	Hence, we have $(I^{(k-1)}_{\text{GMA}_{uv}}-I^{(k-1)}_{\text{GIA}_{uv}})/I^{(k-1)}_{\text{GIA}_{uw}} \leq 1$ and $X_w\in \mathcal{D}_X$.

	Now we have proved (a) holds for edge addition. For the remaining actions of GMA, we can use a new mapping $\mathcal{M}_1$ that injects one node $w$ to node $u$ to prove (a).

	For an edge deletion of $(u,v)$, given $\mathcal{M}_1$, one may rewrite Eq.~\ref{CH:HAO:eq:klayer_con} for the left nodes other than $v$, as well as the equation involving $I^{k}_{uw}$, and derive the same conclusions similarly. Intuitively, for edge deletion, considering the classification probability, removing an edge is equivalent to enlarging the predicted classification probability for other classes, hence it fictionalizes likewise the edge addition and we can use a similar proof for this action.

	Besides, $\mathcal{M}_1$ can also apply to the perturbation of features to node $u$ or the other neighbor nodes of $u$ within $k$ hops, where we inject one node $w$ to make the same effect. In this case, we can rewrite Eq.~\ref{CH:HAO:eq:klayer_con} and simplify it as following:
	\begin{equation}
		\begin{aligned}
			\Delta H^{(k)}_u & = I^{k}_{\text{GMA}_{uv}}\Delta X_{v}, \\
			\Delta E^{(k)}_u & = I^{k}_{\text{GIA}_{uw}}X_{w},        \\
			\Delta H^{(k)}_u & = \Delta E^{(k)}_u,                    \\
		\end{aligned}
	\end{equation}
	where $v\in \{\mathcal{N}^k(u)\cup u\}$, i.e., node $u$ or its $k$-hop neighbor, and $\Delta X_{v}$ is the perturbation to the attributes of node $v$.
	Similarly, by the definition of $I^k_{uv}$, for node $v$ with a degree not less than $1$ (i.e., $v$ is not an isolated node), we have $I^{k}_{\text{GIA}_{uw}}\geq I^{k}_{\text{GMA}_{uv}}$, hence we have $I^{k}_{\text{GMA}_{uv}}/I^{k}_{\text{GIA}_{uw}}\leq 1$ and $X_{w}\in \mathcal{D}_X$.

	Thus, we complete the whole proof.
\end{proof}

\textbf{Theorem~\ref{CH:HAO:thm:gia_threat} for other GNNs}. We can extend Theorem~\ref{CH:HAO:thm:gia_threat} to other GNNs such as GCN, GraphSage, etc.
Recall the theorem 1 in~\cite{jknet}:
\begin{lemma}
	\label{CH:HAO:thm:random_walk}
	Given a k-layer GNN following the neighbor aggregation scheme via Eq.~\ref{CH:HAO:eq:gnn}, assume that all paths in the computation graph of the model are activated with the same probability of success $p$. Then the influence distribution $I_x$ for any node $x\in V$ is equivalent, in expectation, to the $k$-step random walk distribution on $\tilde{\gG}$ starting at node $x$.
\end{lemma}
To apply Lemma~\ref{CH:HAO:thm:random_walk}, we observe that the definition of $I^{k}_{uw}$ is analogous to random walk starting from node $u$.
Thus, one may replace the definition of $I^{k}_{uw}$ here to the influence score defined by~\cite{jknet}, conduct a similar proof above with a random walk score, and obtain the same conclusions,
given the mapping $\mathcal{M}_2$, for each edge addition $(u,v)$, $\exists X_w$, such that
\begin{equation}
	\mathbb{E}(\mathcal{L}^{k}_{\text{atk}}(f_\theta(\gG'_{\text{GIA}}))) = \mathbb{E}(\mathcal{L}^{k}_{\text{atk}}(f_\theta(\gG'_{\text{GIA}}))).
\end{equation}
Though the original theorem only proves Lemma~\ref{CH:HAO:thm:random_walk} for GCN and GraphSage, it is obvious one can easily extend the proof in \cite{jknet} for aggregation scheme as Eq.~\ref{CH:HAO:eq:gnn}.

\textbf{Cases for Less GIA Budget}. We can reduce GIA budgets in two ways.
\begin{enumerate}[label=(\alph*)]
	\item For GMA that performs both node feature perturbation and edge addition, considering an edge perturbation $(u,v)$, $\mathcal{M}_2$  essentially also applies for node feature perturbations on $u$ or $v$ without additional budgets.
	\item It is very likely that with the mapping above, GIA will produce many similar nodes. Hence, with one post-processing step to merge similar nodes together and re-optimize them again,
	      GIA tends to require less budget to make the same or more harm than GMA. That is also reflected in our experiments as shown in Fig.~\ref{CH:HAO:fig:motivate_w_defense}.
\end{enumerate}

\subsection{GIA with Plural Mapping for More GMA Operations}
\label{CH:HAO:sec:m2_gma_appdx}
Here we explain how our theoretical results also apply to the remaining actions, i.e., edge deletion and node feature perturbation, of GMA with $\mathcal{M}_2$ (Def.~\ref{CH:HAO:def:m2_mapping}). In the proof for Theorem~\ref{CH:HAO:thm:gia_threat}, we have proved the existence of mappings for edge removal and node feature perturbation.
Once the injected node features are set to have the same influence on the predictions on the targets, they can be further optimized for amplifying the damage, thus all of our theoretical results can be derived similarly to that for the edge addition operation.

\subsection{Proof for Theorem~\ref{CH:HAO:thm:gia_easy_defend}}
\begin{nono-theorem}
Given conditions in Theorem~\ref{CH:HAO:thm:gia_threat},
consider a GIA attack, which \textup{(i)} is mapped by $\mathcal{M}_2$ (Def.~\ref{CH:HAO:def:m2_mapping}) from a GMA attack that only performs edge addition perturbations,
and \textup{(ii)} uses a linearized GNN trained with at least one node from each class in $\gG$ as the surrogate model, and \textup{(iii)} optimizes the malicious node features with PGD.
Assume that $\gG$ has no isolated node, and has node features as $X_u=\frac{C}{C-1}e_{Y_u}-\frac{1}{C-1}\mathbf{1}\in \R^d$,
where $Y_u$ is the label of node $u$ and $e_{Y_u}\in\R^d$ is a one-hot vector with the $Y_u$-th entry being $1$ and others being $0$.
Let the minimum similarity for any pair of nodes connected in $\gG$ be $s_{\gG}=\min_{(u,v)\in E}\text{sim}(X_u,X_v)$ with $\text{sim}(X_u,X_v)=\frac{X_u\cdot X_v}{\norm{X_u}_2\norm{X_v}_2}$.
For a homophily defender $g_\theta$ that prunes edges $(u,v)$ if $\text{sim}(X_u,X_v)\leq s_{\gG}$,
we have:
\[
	\mathcal{L}_{\text{atk}}(g_\theta(\mathcal{M}_2(\gG'_{\text{GMA}})))-\mathcal{L}_{\text{atk}}(g_\theta(\gG'_{\text{GMA}}))\geq 0.
\]
\end{nono-theorem}
\begin{proof}
	\label{CH:HAO:proof:gia_easy_defend}
	We prove Theorem~\ref{CH:HAO:thm:gia_easy_defend} by firstly show the following lemma.
	\begin{lemma}
		\label{CH:HAO:thm:gia_lower_sim}
		Given conditions in Theorem~\ref{CH:HAO:thm:gia_easy_defend}, as the optimization on $X_w$ with respect to $\mathcal{L}_{\text{atk}}$ by PGD approaches,
		we have:
		\[
			\text{sim}(X_u,X_w)^{(t+1)} \leq \text{sim}(X_u,X_w)^{(t)},
		\]
		where $t$ is the number of optimization steps.
	\end{lemma}
	We prove Lemma~\ref{CH:HAO:thm:gia_lower_sim} in the follow-up section, i.e., Appendix~\ref{CH:HAO:proof:gia_lower_sim}.
	With Lemma~\ref{CH:HAO:thm:gia_lower_sim}, known that GIA is mapped from GMA with $\mathcal{M}_2$, $X_w$ will be optimized to have the same effects as GMA at first and continue being optimized to a more harmful state,
	hence for the unit perturbation case as Fig.~\ref{CH:HAO:fig:m2_illustration}, we know:
	\begin{equation}
		\label{CH:HAO:eq:gia_edge_less_sim}
		\text{sim}(X_u,X_w)\leq \text{sim}(X_u,X_v),
	\end{equation}
	as the optimization on $X_w$ approaches. Furthermore, it follows:
	\begin{equation}
		h_u^\gia \leq h_u^\gma,
	\end{equation}
	where $h_u^\gia$ and $h_u^\gma$ denote the homophily of node $u$ after GIA and GMA attack, respectively.
	Now if we go back to the homophily defender $g_\theta$, for any threshold specified to prune the edge $(u,v)$, as Lemma~\ref{CH:HAO:thm:gia_lower_sim} and Eq.~\ref{CH:HAO:eq:gia_edge_less_sim} indicates,
	direct malicious edges in GIA are more likely to be pruned by $g_\theta$. Let $\tau_\gia$ and $\tau_\gma$ denote the corresponding similarity between $(u,w)$ in GIA and $(u,v)$ in GMA,
	we have several possibilities compared with $s_{\gG}=\min_{(u,v)\in E}\text{sim}(X_u,X_v)$:
	\begin{enumerate}[label=(\alph*)]
		\item $\tau_\gia\leq \tau_\gma \leq s_{\gG}$: all the malicious edges will be pruned, Theorem~\ref{CH:HAO:thm:gia_easy_defend} holds;
		\item $\tau_\gia\leq  s_{\gG}\leq \tau_\gma$: all the GIA edges will be pruned, Theorem~\ref{CH:HAO:thm:gia_easy_defend} holds;
		\item $ s_{\gG}\leq \tau_\gia\leq \tau_\gma$: this is unlikely to happen, otherwise $\tau_\gia$ can be optimized to even worse case, Theorem~\ref{CH:HAO:thm:gia_easy_defend} holds;
	\end{enumerate}
	Thus, we complete our proof.
\end{proof}
Interestingly, we can also set a specific threshold $\tau_h$ for homophily defender s.t., $\tau_h-s_{\gG}\leq \epsilon\geq 0$, where some of the original edges will be pruned, too.
However, some previous works indicate promoting the smoothness or slightly dropping some edges will bring better performance~\citep{dropedge,p-reg,aug_gnn,self-enhance}.
A similar discussion can also be applied to this case and obtain the same conclusions.

\subsection{Proof for Lemma~\ref{CH:HAO:thm:gia_lower_sim}}
\label{CH:HAO:proof:gia_lower_sim}
\begin{proof}
	To begin with, without loss of generality, we may assume the number of classes is $2$ and $Y_u=0$, which can be similarly extended to the case of multi-class.
	With the feature assignment in the premise, let the label of node $u$ be $Y_u$, we have:
	\begin{equation}
		\label{CH:HAO:eq:xw_indicator}
		X_u = \left\{
		\begin{aligned}
			 & [1, -1]^T, & Y_u =0, \\
			 & [-1, 1]^T, & Y_u =1. \\
		\end{aligned}
		\right.
	\end{equation}
	After setting it to having the same influence as that in GMA following Eq.~\ref{CH:HAO:eq:xw_sol}, we have:
	\begin{equation}
		X_{w}=\frac{(I^{k}_{\text{GMA}_{uv}}-I^{k}_{\text{GIA}_{uv}})}{I^{k}_{\text{GIA}_{uw}}}X_{v}.
	\end{equation}
	Then, let $\mathcal{L}_u$ denote the training loss $\mathcal{L}_{\text{train}}$ on node $u$, we can calculate the gradient of $X_w$:
	\begin{equation}
		\frac{\partial \mathcal{L}_u}{\partial X_u}=\frac{\partial \mathcal{L}_u}{\partial H^{(k)}_u}\cdot \frac{\partial H^{(k)}_u}{\partial X_w}=\frac{\partial \mathcal{L}_u}{\partial H^{(k)}_u}\cdot I^{k}_{\text{GIA}_{uw}}\cdot \Theta.
	\end{equation}
	With Cross-Entropy loss, we further have:
	\begin{equation}
		\frac{\partial \mathcal{L}_u}{\partial H^{(k)}_u}=[-1, 1]^T.
	\end{equation}
	Then, we can induce the update step of optimizing $X_w$ with respect to $\mathcal{L_\atk}=-\mathcal{L}_{\text{train}}$ by PGD:
	\begin{equation}
		X^{(t+1)}_w = X^{(t)}_w+\epsilon \sign(I^{k}_{\text{GIA}_{uw}}\cdot [-1, 1]^T\cdot \Theta),
	\end{equation}
	where $t$ is the number of update steps.
	As the model is trained on at least nodes with indicator features following Eq.~\ref{CH:HAO:eq:xw_indicator} from each class,
	without loss of generality, here we may assume $\Theta \geq \mathbf{0}$,
	the optimal $\Theta$ would converge to $\Theta \geq \mathbf{0}$. Thus,
	\[
		\sign(I^{k}_{\text{GIA}_{uw}}\cdot [-1, 1]^T\cdot \Theta) = \sign(I^{k}_{\text{GIA}_{uw}}\cdot [-1, 1]^T).
	\]
	Let us look into the change of cosine similarity between node $u$ and node $v$ as:
	\begin{equation}
		\Delta \text{sim}(X_u,X_w) = \alpha (X_u\cdot X^{(t+1)}_w-X_u\cdot X^{(t)}_w),
	\end{equation}
	where $\alpha\geq 0$ is the normalized factor. To determine the sign of $\Delta \text{sim}(X_u,X_w)$, we may compare $X_u\cdot X^{(t+1)}$ with $X_u\cdot X^{(t)}_w$.
	Here we expand $X_u\cdot X^{(t+1)}_w$. Let $X_{u0}, X_{u1}$ to denote the first and second element in $X_u$ respectively, we have:
	\begin{equation}
		\label{CH:HAO:eq:x_t_cos}
		\begin{aligned}
			X_u\cdot X^{(t+1)}_w & =\frac{X_u\cdot X_w+\epsilon \sign(I^{k}_{\text{GIA}_{uw}}\cdot [-1, 1]^T)X_u}{\norm{X_u}_2\cdot\norm{X^{(t+1)}_w}_2},   \\
			                     & =\frac{X_u\cdot X_w+\epsilon (X_{u1}-X_{u0})}{\norm{X_u}_2\sqrt{X^2_{w0}+X^2_{w1}+\epsilon^2+2\epsilon(X_{w1}-X_{w0})}}, \\
		\end{aligned}
	\end{equation}
	where we omit the sign of $I^{k}_{\text{GIA}_{uw}}$ for $I^{k}_{\text{GIA}_{uw}}\geq 0$ according to the definition. Recall that we let $Y_u=0$, hence we have $(X_{u1}-X_{u0})\leq 0$.
	Besides, following Eq.~\ref{CH:HAO:eq:xw_sol}, we have $\sign(X_{w1}-X_{w0})=\sign(X_{v1}-X_{v0})$. As GMA tends to connect nodes from different classes, we further have $\sign(X_{w1}-X_{w0})\geq 0$.
	Comparing to $X_u\cdot X^{(t)}_w$, we know in Eq.~\ref{CH:HAO:eq:x_t_cos}, the numerator decreases and the denominator increases, as $\epsilon\geq0$, so the overall scale decreases. In other words, we have:
	\begin{equation}
		\Delta \text{sim}(X_u,X_w) = \alpha (X_u\cdot X^{(t+1)}_w-X_u\cdot X^{(t)}_w)\leq 0,
	\end{equation}
	which means that the cosine similarity between node $u$ and node $v$ decreases as the optimization of $X_w$ with respect to $\mathcal{L}_\atk$ processes. Thus, we complete our proof for Lemma~\ref{CH:HAO:thm:gia_lower_sim}.
\end{proof}

\subsection{Proof for Theorem~\ref{CH:HAO:thm:hao_break_limit}}
\begin{nono-theorem}
Given conditions as Theorem~\ref{CH:HAO:thm:gia_easy_defend},
when $\lambda>0$, we have
$m(\mathcal{H}_\gG,\mathcal{H}_{\gG'_{\text{HAO}}})\leq m(\mathcal{H}_\gG,\mathcal{H}_{\gG'_{\text{GIA}}})$,
hence: %
\[
	\mathcal{L}_{\text{atk}}(g_\theta(\gG'_{\text{HAO}}))-\mathcal{L}_{\text{atk}}(g_\theta(\gG'_{\text{GIA}}))\leq 0,
\]
where $\gG'_{\text{HAO}}$ is generated by GIA with HAO, and $\gG'_{\text{GIA}}$ is generated by GIA without HAO.
\end{nono-theorem}
\begin{proof}
	\label{CH:HAO:proof:hao_break_limit}
	Similar to the proof for Theorem~\ref{CH:HAO:thm:gia_easy_defend}, we begin with binary classification, without loss of generality.
	With the feature assignment in the premise, let the label of node $u$ be $Y_u$, we have:
	\begin{equation}
		X_u = \left\{
		\begin{aligned}
			 & [1, -1]^T, & Y_u =0, \\
			 & [-1, 1]^T, & Y_u =1. \\
		\end{aligned}
		\right.
	\end{equation}
	Let $\mathcal{L}_u$ denote the training loss $\mathcal{L}_{\text{train}}$ on node $u$, we look into the gradient of $X_w$ with respect to $\mathcal{L}_u$:
	\begin{equation}
		\frac{\partial \mathcal{L}_u}{\partial X_u}=\frac{\partial \mathcal{L}_u}{\partial H^{(k)}_u}\cdot \frac{\partial H^{(k)}_u}{\partial X_w}=\frac{\partial \mathcal{L}_u}{\partial H^{(k)}_u}\cdot I^{k}_{\text{GIA}_{uw}}\cdot \Theta.
	\end{equation}
	With Cross-Entropy loss, we further have:
	\begin{equation}
		\frac{\partial \mathcal{L}_u}{\partial H^{(k)}_u}=[-1, 1]^T.
	\end{equation}
	Together with HAO, we can infer the update step of optimizing $X_w$ with respect to \mbox{$\mathcal{L_\atk}=-\mathcal{L}_{\text{train}}+\lambda C(\gG,\gG')$} by PGD:
	\begin{equation}
		X^{(t+1)}_w = X^{(t)}_w+\epsilon \sign((I^{k}_{\text{GIA}_{uw}}\cdot [-1, 1]^T+\lambda[1,-1]^T)\cdot \Theta),
	\end{equation}
	where $t$ is the number of update steps. Similarly, without loss of generality, we may assume $\Theta\geq \mathbf{0}$.
	As the optimization approaches, given $\lambda>0$, GIA with HAO will early stop to some stage that $(I^{k}_{\text{GIA}_{uw}}\cdot [-1, 1]^T+\lambda[1,-1]^T)=\mathbf{0}$,
	hence similar to the proof of Theorem~\ref{CH:HAO:thm:gia_easy_defend}, it follows:
	\begin{equation}
		h_u^\gia \leq h_u^\text{HAO},
	\end{equation}
	where $h_u^\gia$ and $h_u^\text{HAO}$ denote the homophily of node $u$ after GIA and GIA with HAO attack, respectively.
	Likewise, we can infer that:
	\[
		\mathcal{L}_{\text{atk}}(g_\theta(\gG'_{\text{HAO}}))-\mathcal{L}_{\text{atk}}(g_\theta(\gG'_{\text{GIA}}))\leq 0.
	\]
	Thus, we complete our proof.
\end{proof}

\subsection{Certified Robustness of Homophily Defender}
Here we prove the certified robustness of homophily for a concrete GIA case. We prove via the decision margin as follows:
\label{CH:HAO:sec:gia_cer_robo}
\begin{definition}[Decision Margin]
	\label{CH:HAO:def:decision_margin}
	Given a $k$-layer GNN,
	let $H^{(k)}_{[u,c]}$ denote the corresponding entry in $H^{(k)}_{u}$ for the class $c$,
	the decision margin on node u with class label $Y_u$ can be denoted by:
	\[
		m_u = 	H^{(k)}_{[u,y_u]}-\max_{c\in \{0,..,C-1\}}H^{(k)}_{[u,c]}.
	\]
\end{definition}
A Multi-Layer Perceptron (MLP) can be taken as a $0$-layer GNN which the definition also applies.
Then, we specify the certified robustness as follows:
\begin{proposition}[Certified Robustness of Homophily Defender]
	\label{CH:HAO:thm:gia_cer_robo}
	Consider a direct GIA attack uses a linearized GNN trained with at least one node from each class in $\gG$,
	that targets at node $u$ by injecting a node $w$ connecting to $u$,
	let node features $x_u=\frac{C}{C-1}\text{onehot}(Y_u)-\frac{1}{C-1}\mathbf{1}$,
	the homophily of $u$ be $\tau$, the decision margin of a MLP on $u$ be $\gamma$,
	the minimum similarity for any pair of nodes connected in the original graph be $s_{\gG}=\min_{(u,v)\in E}\text{sim}(X_u,X_v)$,
	homophily defender $g_\theta$ can defend such attacks, if $-\alpha\frac{1}{\sqrt{1+1/d_u}}(\tau+\beta \gamma)\leq s_{\gG}$,
	and $g_\theta$ prunes edges $(u,v)$ s.t.,
	\[
		\text{sim}(X_u,X_w)\leq -\alpha\sqrt{\frac{1}{1+1/d_u}}(\tau+\beta\gamma),
	\]
	where $\alpha,\beta\geq 0$ are corresponding normalization factors.
\end{proposition}
Intuitively, effective attacks on a node with higher degrees, homophily, or decision margin require a lower similarity between node $w$ and $u$ hence more destruction to the homophily of node $u$.
GIA without any constraints tends to optimize $\text{sim}(X_u,X_w)$ to an even lower value.
Thus, it becomes easier to find a suitable condition for $g_\theta$, with which it can painlessly prune all vicious edges while keeping all original edges.
\begin{proof}
	\label{CH:HAO:proof:gia_cer_robo}
	Analogous to the proof for Lemma~\ref{CH:HAO:thm:gia_lower_sim}, without loss of generality, we begin with
	binary classification, normalized indicator features, and $Y_u=0$ as follows:
	\begin{equation}
		X_u = \left\{
		\begin{aligned}
			 & [1, -1]^T, & Y_u =0, \\
			 & [-1, 1]^T, & Y_u =1. \\
		\end{aligned}
		\right.
	\end{equation}
	The decision margin based on $k$-th layer representation can be denoted by
	\begin{equation}
		m=H^{(k)}_{[u,y_u]}-\max_{c\in \{0,..,C-1\}}H^{(k)}_{[u,c]},
	\end{equation}
	follows the Definition~\ref{CH:HAO:def:decision_margin}. In our binary classification case, we have
	\begin{equation}
		\gamma=H^{(0)}_{[u,0]}-H^{(0)}_{[u,1]},
	\end{equation}
	where $H^{(0)}$ is the output of a $0$-layer GNN, or MLP (Multi-Layer Perceptron). A $k$-layer GNN can be regarded as
	generating new hidden representation for node $u$ by aggregating its neighbors, hence, we may induce the decision margin
	for a $k$-layer GNN at node $u$ as
	\begin{equation}
		m=H^{(k)}_{[u,0]}-H^{(k)}_{[u,1]}=([\sum_{j\in \mathcal{N}(u)}I_{uj}X_j]_{[0]}-[\sum_{j\in \mathcal{N}(u)}I_{uj}X_j]_{[1]})+I_{uu}^{(k)}\gamma,
	\end{equation}
	where we can replace the influence from neighbors with homophily of node $u$. Observe that $h_u$ essentially indicates how much neighbors of node $u$ contribute to $H_{[u,0]}^{(k)}$, for example, in binary case, let $\zeta> 0$ be the corresponding normalization factor,
	\[
		h_u = \frac{1}{\zeta}([\sum_{j\in \mathcal{N}(u)}I_{uj}X_j]_{[0]}[X_u]_{[0]}+[\sum_{j\in \mathcal{N}(u)}I_{uj}X_j]_{[1]}[X_u]_{[1]}),
	\]
	which means,
	\[
		[\sum_{j\in \mathcal{N}(u)}I_{uj}X_j]_{[1]} = \frac{1}{[X_u]_{[1]}} (\zeta h_u-[\sum_{j\in \mathcal{N}(u)}I_{uj}X_j]_{[0]}[X_u]_{[0]}),
	\]
	replaced with $X_u=[1,-1]^T$,
	\begin{equation}
		\begin{aligned}
			m & =H^{(k)}_{[u,0]}-H^{(k)}_{[u,1]}                                                                                                                           \\
			  & =([\sum_{j\in \mathcal{N}(u)}I_{uj}X_j]_{[0]}-[\sum_{j\in \mathcal{N}(u)}I_{uj}X_j]_{[1]})+                                                                \\
			  & =([\sum_{j\in \mathcal{N}(u)}I_{uj}X_j]_{[0]}-\frac{1}{[X_u]_{[1]}} (\zeta h_u-[\sum_{j\in \mathcal{N}(u)}I_{uj}X_j]_{[0]}[X_u]_{[0]}))+I_{uu}^{(k)}\gamma \\
			  & =\zeta h_u+I_{uu}^{(k)}\gamma.                                                                                                                             \\
		\end{aligned}
	\end{equation}
	Hence, we have:
	\[
		m=H^{(k)}_{[u,0]}-H^{(k)}_{[u,1]}=\zeta h_u+I_{uu}^{(k)}\gamma,
	\]
	where $\zeta\geq 0$ is the factor of $h_u$.
	With node $w$ injected, the margin can be rewritten as:
	\begin{equation}
		m'=\sqrt{\frac{d_u}{d_u+1}}m+I^{(k)}_{uw}(X_{[w,0]}-X_{[w,1]}).
	\end{equation}
	To perturb the prediction of node $u$, we make $m\leq 0$, hence, we have
	\begin{equation}
		\label{CH:HAO:eq:homo_req}
		\begin{aligned}
			m'                                & =\sqrt{\frac{d_u}{d_u+1}}m+I^{(k)}_{uw}(X_{[w,0]}-X_{[w,1]})\leq 0,                \\
			I^{(k)}_{uw}(X_{[w,1]}-X_{[w,0]}) & \geq \sqrt{\frac{d_u}{d_u+1}}m,                                                    \\
			(X_{[w,1]}-X_{[w,0]})             & \geq \frac{1}{I^{(k)}_{uw}}\sqrt{\frac{d_u}{d_u+1}}(\zeta h_u+I_{uu}^{(k)}\gamma). \\
		\end{aligned}
	\end{equation}
	Observe that $\text{sim}(X_u,X_w)=(X_{[w,0]}-X_{[w,1]})$ and $h_u=\tau$, hence, we can write Eq.~\ref{CH:HAO:eq:homo_req} in a clean form as
	\begin{equation}
		\text{sim}(X_u,X_w)\leq -\alpha\sqrt{\frac{d_u}{d_u+1}}(\tau+\beta\gamma),
	\end{equation}
	where $\alpha,\beta$ are corresponding normalization factors whose signs are determined by signs of $I^k_{uw}$ and $I^k_{uu}$ respectively.
	In other words, GIA has to optimize $X_w$ satisfying the above requirement to make the attack \textit{effective},
	however, given the premise that all $s_\gG=\min_{(u,v)\in E}\text{sim}(X_u,X_v)\geq -\alpha\sqrt{\frac{d_u}{d_u+1}}(\tau+\beta\gamma)$,
	a defense model $g_\theta$ will directly prune all of the vicious edges satisfying the above requirement and make the attack \textit{ineffective},
	which is exactly what we want to prove.
\end{proof}

\section{More Implementations of Homophily Defender}
\label{CH:HAO:sec:more_homo_defender_appdx}
There are many ways to design homophily defenders, inheriting the spirit of recovering the original homophily.
In addition to edge pruning, one could leverage variational inference to learn the homophily distribution or the similarity distribution among neighbors.
Then we use adversarial training to train the model to denoise.
Similarly, learning to promote the smoothness of the graph can also be leveraged to build homophily defenders~\citep{aug_gnn,p-reg,self-enhance}.
Besides, outlier detection can also be adopted to remove or reduce the aggregation weights of malicious edges or nodes.
In the following two subsections, we will present two variants that perform better than GNNGuard~\citep{gnnguard}.

\subsection{Details of Efficient GNNGuard}
\label{CH:HAO:sec:egnnguard_appdx}
The originally released GNNGuard requires $O(n^2)$ computation for node-node similarity, making it prohibitive to run on large graphs.
To this end, we implement an efficient alternative of GNNGuard adopting a similar message passing scheme, let $\tau$ be the threshold to prune an edge:
\begin{equation}
	\label{CH:HAO:eq:egnngaurd}
	H^{(k)}_u = \sigma(W_k \cdot \sum_{j \in \mathcal{N}(u)\cup\{u\}} \alpha_{uj} H^{(k-1)}_j),
\end{equation}
where
\[
	\alpha_{uj}=\text{softmax}(\frac{z_{uj}}{\sum_{v\in \mathcal{N}(u)\cup \{u\}}z_{uv}}),
\]
and
\[
	z_{uj}  = \left\{
	\begin{aligned}
		 & \frac{\mathbf{1}\{\text{sim}(H^{(k-1)}_j, H^{(k-1)}_u)> \tau\}\cdot\text{sim}(H^{(k-1)}_j\cdot H^{(k-1)}_u)}{\sum_{v \in \mathcal{N}(u)}\mathbf{1}\{\text{sim}(H^{(k-1)}_v, H^{(k-1)}_u)> \tau\}\cdot\text{sim}(H^{(k-1)}_v, H^{(k-1)}_u)}, & u\neq j, \\
		 & \frac{1}{\sum_{v \in \mathcal{N}(u)}\mathbf{1}\{\text{sim}(H^{(k-1)}_v\cdot H^{(k-1)}_u)> \tau\}+1}                                                                                                                                         & u=j.     \\
	\end{aligned}
	\right.
\]
Essentially, it only requires $O(E)$ complexity. We will present the performance of Efficient GNNGuard (EGNNGuard) in table~\ref{CH:HAO:tab:homo_defender_performance}.

\subsection{Details of Robust Graph Attention Network (RGAT)}
\label{CH:HAO:sec:rgat_appdx}
We introduce another implementation of the Robust Graph Attention Network (RGAT).
We adopt the same spirit of GCNGuard \citep{gnnguard}, which eliminates, unlike neighbors during message passing based on neighbor similarity. Specifically, we change the standard GAT \citep{gat} attention mechanism as
\[
	\alpha_{i,j} = \frac{\mathbb{1}\{\text{sim}(x_i, x_j)\geq\tau\}}{\sum_{k\in \mathcal{N}(i)\cup \{i\}}\mathbb{1}\{\text{sim}(x_i, x_k)\geq\tau\}},
\]
Additionally, we also adopt the idea of RobustGCN \citep{robustgcn} that stabilizes the hidden representations between layers, so we add Layer Normalization \citep{layer_norm} among layers of RGAT. Empirical experiments show that RGAT is a more robust model with or without GIA attacks. For more details, we refer readers to Table~\ref{CH:HAO:tab:homo_defender_performance}.

\subsection{Performance of Homophily Defenders}
\bgroup
\def\arraystretch{1.1}
\begin{table}[!h]
	\centering\small
	\caption{Performance of homophily defenders used in experiments.}
	\label{CH:HAO:tab:homo_defender_performance}
	\begin{tabular}{lccc}
		\toprule
		\multicolumn{1}{l}{\textbf{Model}} & \textbf{Natural Accuracy} & \textbf{Test Robustness} & \textbf{Running Time}         \\\midrule
		GNNGuard                           & $83.58$                   & $64.96$                  & $1.76\times 10^{-3}$
		\\\hline
		EGNNGuard                          & $84.45$                   & $64.27$                  & $\mathbf{5.39\times 10^{-5}}$
		\\\hline
		RGAT                               & $\mathbf{85.75}$          & $\mathbf{66.57}$         & $6.03\times 10^{-5}$          \\\hline
		GCN                                & $84.99$                   & $36.62$                  & $5.87\times 10^{-5}$
		\\\bottomrule
	\end{tabular}
\end{table}
\egroup

We test the performance of different homophily defenders on Cora. Natural Accuracy refers to the test accuracy on clean graph. Test Robustness refers to their averaged performance against all the attacks.
Running time refers to their averaged running time for one training epoch. We repeat the evaluation $10$ times to obtain the average accuracy.
We can see that EGNNGuard has competitive performance with GNNGuard while $20\times$ faster. RGAT performs slightly better and $10\times$ faster.
Hence, for large graphs and adversarial training of GNNGuard, we will use EGNNGuard instead.

\section{More Details about Algorithms used}
\label{CH:HAO:sec:gia_alg_desc}
Here we provide detailed descriptions of algorithms mentioned in Section.~\ref{CH:HAO:sec:gia_algorithm}.

\subsection{Details of MetaGIA and AGIA}
\label{CH:HAO:sec:gia_gradient_appdx}

\subsubsection{Induction of Meta Gradients for MetaGIA}
\label{CH:HAO:sec:gia_meta_appdx}
With the bi-level optimization formulation of GIA, similar to meta-attack, we can infer the meta-gradients as follows:
\begin{equation}
	\nabla^{meta}_{A_{\atk}} = \nabla_{A_{\atk}}\mathcal{L}_{\atk}(f_{\theta^*}(A_{\atk},X^*_{\atk})),\quad
	X^*_{\atk} = \text{opt}_{X_{\atk}}\mathcal{L}_{\atk}(f_{\theta^*}(A_{\atk},X_{\atk})).
\end{equation}
Consider the $\text{opt}$ process, we have
\begin{equation}
	\label{x_opt}
	X_{\atk}^{(t+1)} = X_{\atk}^{(t)}-\alpha \nabla_{X_{\atk}^{(t)}}\mathcal{L}_{\atk}(f_{\theta^*}(A_{\atk},X^{(t)}_{\atk})).
\end{equation}
With that, we can derive the meta-gradient for $A_{\atk}$:
\begin{equation}
	\begin{aligned}
		\nabla^{\text{meta}}_{A_{\atk}} & = \nabla_{A_{\atk}}\mathcal{L}_{\atk}(f_{\theta^*}(A_{\atk},X^*_{\atk}))     \\
		                                & = \nabla_{X_{\atk}}\mathcal{L}_{\atk}(f_{\theta^*}(A_{\atk},X^{(t)}_{\atk}))
		\cdot[\nabla_{A_{\atk}}f_{\theta^*}(A_{\atk},X^{(t)}_{\atk})+
		\nabla_{X^{(t)}_{\atk}}f_{\theta^*}(A_{\atk},X^{(t)}_{\atk})\cdot
		\nabla_{A_{\atk}}X^{(t)}_{\atk}],                                                                              \\
	\end{aligned}
\end{equation}
where
\begin{equation}
	\nabla_{A_{\atk}}X^{(t+1)}_{\atk} = \nabla_{A_{\atk}}X^{(t)}_{\atk}
	-\alpha \nabla_{A_{\atk}} \nabla_{X^{(t)}_{\atk}}\mathcal{L}_{\atk}(f_{\theta^*}(A_{\atk},X^{(t)}_{\atk})).
\end{equation}

Note that $X^{(t)}_{\atk}$ depends on $A_{\atk}$ according to Eq.~\ref{x_opt}, so the derivative w.r.t. $A_{\atk}$ need to be traced back. Finally, the update schema for $A_{\atk}$ is as follows:
\begin{equation}
	A^{(t+1)}_{\atk} = A^{(t)}_{\atk} - \beta \nabla^{\text{meta}}_{A^{(t)}_{\atk}}.
\end{equation}
Directly computing the meta gradients is expensive, following Metattack, we adopt approximations like MAML~\citep{maml} for efficiency consideration.
We refer readers to the paper of Metattack for the detailed algorithms by replacing the corresponding variables with those above.

\subsection{Details of AGIA}
For optimizing weights of edge entries in $A_\atk$, we can use either Adam~\citep{adam}, PGD~\citep{pgd} or other optimization methods leveraging gradients.
For simplicity, we use PGD to illustrate the algorithm description of AGIA as follows:

\begin{algorithm}[!h]
	\caption{AGIA: Adaptive Graph Injection Attack with Gradient}
	\label{alg:gia_apgd}
	\begin{algorithmic}[1]
		\STATE \textbf{Input:}
		A graph $\gG=(A,X)$, a trained GNN model $f_{\theta^*}$, number of injected nodes $c$, degree budget $b$, outer attack epochs $e_{\text{outer}}$, inner attack epochs for node features and adjacency matrix $e^X_{\text{inner}},e^A_{\text{inner}}$, learning rate $\eta$, weight for sparsity penalty $\beta$, weight for homophily penalty $\lambda$ ;
		\STATE Perturbed graph $\gG'=(A', X')$;
		\STATE Random initialize injection parameters ($A_\atk, X_\atk$);
		\STATE $\mY_{\text{orig}} \leftarrow f_{\theta^*}(A,X)$ \texttt{// Obtain original predictions on clean graph}
		\FOR{epoch $\leftarrow$ $0$ to $e_{\text{outer}}$}
		\STATE Random initialize $X_\atk$;
		\FOR{epoch $\leftarrow$ $0$ to $e^X_{\text{inner}}$}
		\STATE $A'\leftarrow A\concat A_\atk, X'\leftarrow X\concat X_\atk$ ;
		\STATE $X_\atk \leftarrow \text{Clip}_{(x_{\min},x_{\max})}(X_\atk - \eta \cdot \nabla_{X_\atk}(\mathcal{L}^h_{\text{atk}}))$ ;
		\ENDFOR
		\FOR{epoch $\leftarrow$ $0$ to $e^A_{\text{inner}}$}
		\STATE $A'\leftarrow A \concat A_\atk, X'\leftarrow X\concat X_\atk$ ;
		\STATE $A_\atk \leftarrow \text{Clip}_{(0,1)}(A_\atk - \eta \cdot \nabla_{A_\atk}(\mathcal{L}^A_{\text{atk}}))$ ;
		\ENDFOR
		\STATE $A_\atk \leftarrow \left\lVert\right\rVert_{i=1}^{k}\argmax_{\text{top}\ b}(A_{\atk[i, :]})$ ;
		\ENDFOR
	\end{algorithmic}
\end{algorithm}
Here, $\mathcal{L}^h_{\text{atk}}$ refers to the objective of GIA with HAO for the optimization of $X_\atk$.
For the optimization of $A_\atk$, we empirically find the $\lambda_A$ would degenerate the performance, which we hypothesize that is because of the noises as $A_\atk$ is a discrete variable.
Hence, we set $\lambda_A=0$ in our experiments. Additionally, we introduce a sparsity regularization term for the optimization of $A_\atk$:
\begin{equation}
	\mathcal{L}^{A}_\atk = \mathcal{L}_\atk+\beta\frac{1}{|V_\atk|} \sum_{u\in V_\atk}|b-\lVert A_{\atk_{u,:}} \rVert_1|.
\end{equation}
Besides, we empirically observe that Adam performs better than PGD. Hence, we would use Adam for AGIA in our experiments, and leave other methods for future work.
Adopting Adam additionally brings the benefits to utilize momentum and history information to accelerate the optimization escape from the local optimum, which PGD fails to achieve.

\subsection{Details of SeqGIA}
\label{CH:HAO:sec:gia_apgd_seq_appdx}
Since gradient methods require huge computation overhead,
we propose a novel divide-and-conquer strategy to iteratively select some of the most vulnerable targets with Eq.~\ref{CH:HAO:eq:tdgia+_score} to attack.
Note that it is different from traditional sequential injection methods which still connect the targets in full batch.
For simplicity, we also illustrate the algorithm with PGD, and one may switch to other optimizer such as Adam to optimize $A_\atk$.
The detailed algorithm is as follows:

\begin{algorithm}[H]
	\caption{SeqGIA: Sequential Adaptive Graph Injection Attack}
	\label{alg:gia_apgd_seq}
	\begin{algorithmic}[1]
		\STATE \textbf{Input:} A graph $\gG=(A,X)$, a trained GNN model $f_{\theta^*}$, number of injected nodes $k$, degree budget $b$, outer attack epochs $e_{\text{outer}}$, inner attack epochs for node features and adjacency matrix $e^X_{\text{inner}},e^A_{\text{inner}}$, learning rate $\eta$, weight for sparsity penalty $\beta$, weight for homophily penalty $\lambda$,
		sequential step for vicious nodes $\gamma_\atk$, sequential step for target nodes $\gamma_c$ ;
		\STATE Initialize injection parameters ($A_\atk, X_\atk$);
		$\mY_{\text{orig}} \leftarrow f_{\theta^*}(A,X)$ \texttt{// Obtain original predictions on clean graph};
		\WHILE{Not Injecting All Nodes}
		\STATE $n_\atk \leftarrow  \gamma_\atk*|V_\atk|$; $n_c \leftarrow \gamma_c*|V_c|$ ;
		\STATE Ranking and selecting $n_c$ targets with Eq.~\ref{CH:HAO:eq:tdgia+_score};
		\STATE Random initialize $A_\atk^{(\text{cur})}\in\R^{n_c\times n_\atk},X_\atk^{(\text{cur})} \in \R^{n_\atk\times d}$ ;
		\FOR{epoch $\leftarrow$ $0$ to $e_{\text{outer}}$}
		\FOR{epoch $\leftarrow$ $0$ to $e^X_{\text{inner}}$}
		\STATE $A'\leftarrow A\concat A_\atk \concat A_\atk^{(\text{cur})}, X'\leftarrow X\concat X_\atk \concat X_\atk^{(\text{cur})} $;
		\STATE $X^{(\text{cur})}_\atk \leftarrow \text{Clip}_{(x_{\min},x_{\max})}(X_\atk^{(\text{cur})} - \eta \cdot \nabla_{X^{(\text{cur})}_\atk}(\mathcal{L}^h_{\text{atk}}))$ ;
		\ENDFOR
		\FOR{epoch $\leftarrow$ $0$ to $e^A_{\text{inner}}$}
		\STATE $A'\leftarrow A\concat A_\atk \concat A_\atk^{(\text{cur})}, X'\leftarrow X\concat X_\atk \concat X_\atk^{(\text{cur})} $;
		\STATE $A_\atk^{(\text{cur})} \leftarrow \text{Clip}_{(0,1)}(A_\atk^{(\text{cur})} - \eta \cdot \nabla_{A_\atk^{(\text{cur})}}(\mathcal{L}^A_{\text{atk}}))$ ;
		\ENDFOR
		\STATE $A_\atk^{(\text{cur})} \leftarrow \left\lVert\right\rVert_{i=1}^{n_\atk}\argmax_{\text{top}\ b}(A_{\atk[i, :]}^{(\text{cur})})$ ;
		\ENDFOR
		\STATE $A_\atk$=$A_\atk$ $\Vert$ $A_\atk^{(\text{cur})}$; $X_\atk$=$X_\atk$ $\Vert$ $X_\atk^{(\text{cur})}$;
		\ENDWHILE
		\RETURN Perturbed graph $\gG'=(A', X')$;
	\end{algorithmic}
\end{algorithm}

Actually, one may also inject few nodes via heuristic based algorithms first, then inject the left nodes with gradients sequentially.
Assume that $\alpha$ nodes are injected by heuristic, we may further optimize the complexity
from
\[
	O(\frac{1}{\gamma_\atk}(|V_c|\log|V_c|+ e_{\text{outer}}(e^A_{\text{inner}}|V_c|\gamma_c|V_\atk|+e^X_{\text{inner}}|V_\atk| d))N_{V_c})
\]
to
\[
	\begin{aligned}
		O(\alpha\frac{1}{\gamma_\atk}(|V_c|\log|V_c|+   & |V_\atk|b+e^X_{\text{inner}}|V_\atk|d)N_{V_c}+                                                  \\
		(1-\alpha)\frac{1}{\gamma_\atk}(|V_c|\log|V_c|+ & e_{\text{outer}}(e^A_{\text{inner}}|V_c|\gamma_c|V_\atk|+e^X_{\text{inner}}|V_\atk| d))N_{V_c})
	\end{aligned}
\]
in Table~\ref{CH:HAO:tab:complexity}.

\section{More Details about the Experiments}
\label{CH:HAO:sec:experiment_appdx}

\subsection{Statistics and Budgets of Datasets}
\label{CH:HAO:sec:datasets_appdx}
Here we provide statistics of datasets used in the experiments as Sec.~\ref{CH:HAO:sec:exp_setups}.
The label homophily utilizes the previous homophily definition~\citep{beyond_homophily},
while the avg. homophily utilizes the node-centric homophily based on node similarity.

\begin{table}[!ht]
	\centering\small
	\caption{Statistics of datasets used in \hao.}
	\label{CH:HAO:tab:datasets}%
	\begin{tabular}{lcccccc}
		\toprule
		\textbf{Datasets} & \textbf{Nodes} & \textbf{Edges} & \textbf{Classes} & \textbf{Avg. Degree} & \textbf{Label Homophily} & \textbf{Avg. Homophily} \\\midrule
		Cora              & $2680$         & $5148$         & $7$              & $3.84$               & $0.81$                   & $0.59$                  \\
		Citeseer          & $3191$         & $4172$         & $6$              & $2.61$               & $0.74$                   & $0.90$                  \\
		Computers         & $13,752$       & $245,861$      & $10$             & $35.76$              & $0.77$                   & $0.31$                  \\
		Arxiv             & $169,343$      & $1,166,243$    & $40$             & $13.77$              & $0.65$                   & $0.86$                  \\
		Aminer            & $659,574$      & $2,878,577$    & $18$             & $8.73$               & $0.65$                   & $0.38$                  \\
		Reddit            & $232,965$      & $11,606,919$   & $41$             & $99.65$              & $0.78$                   & $0.31$                  \\\bottomrule
	\end{tabular}%
\end{table}
Following previous works~\citep{tdgia,grb}, we heuristically specify the budgets for each dataset according to the the number of target nodes and average degrees.

\begin{table}[!ht]
	\centering\small
	\caption{Budgets for non-targeted attacks on different datasets.}
	\label{CH:HAO:tab:datasets_budget_nontarget}
	\begin{tabular}{lcccccc}
		\toprule
		\textbf{Datasets} & \textbf{Nodes} & \textbf{Degree} & \textbf{Node Per.($\%$)} & \textbf{Edge Per.($\%$)} \\\midrule
		Cora              & $60$           & $20$            & $2.24\%$                 & $23.31\%$                \\
		Citeseer          & $90$           & $10$            & $2.82\%$                 & $21.57\%$                \\
		Computers         & $300$          & $150$           & $2.18\%$                 & $18.30\%$                \\
		Arxiv             & $1500$         & $100$           & $0.71\%$                 & $10.29\%$                \\\bottomrule
	\end{tabular}
\end{table}

For targeted attack, we follow previous works~\citep{nettack} to select $800$ nodes as targets according to the classification margins of the surrogate model.
Specifically, we select $200$ nodes with the highest classification margin, $200$ nodes with lowest classification margin and $400$ randomly.
For the budgets, we scale down the number of injected nodes and the maximum allowable degrees accordingly.
\begin{table}[!h]
	\centering\small
	\caption{Budgets of targeted attacks on different datasets}
	\label{CH:HAO:tab:datasets_budget_target}
	\begin{tabular}{lcccccc}
		\toprule
		\textbf{Datasets} & \textbf{Nodes} & \textbf{Degree} & \textbf{Node Per.($\%$)} & \textbf{Edge Per.($\%$)} \\\midrule
		Computers         & $100$          & $150$           & $0.73\%$                 & $6.1\%$                  \\
		Arxiv             & $120$          & $100$           & $0.07\%$                 & $1.03\%$                 \\
		Aminer            & $150$          & $50$            & $0.02\%$                 & $0.26\%$                 \\
		Reddit            & $300$          & $100$           & $0.13\%$                 & $0.26\%$                 \\\bottomrule
	\end{tabular}
\end{table}

\subsection{Additional Discussions about Attack Baselines}
\label{CH:HAO:sec:exp_rl_appdx}
For the selection of attack baselines, from the literature reviews~\citep{sun_gadv_survey,deeprobust},
existing reinforcement learning (RL) based approaches adopt different settings from ours, which either focus on the
poisoning attack, transductive learning, edge perturbation or other application tasks.
Even for NIPA~\citep{nipa} which has the closest setting to ours,
since it focuses on poisoning and transductive attack, and the features of the injected nodes are generated heuristically according to the labels assigned by the RL agent,
without author released code, the adaption requires lots of efforts including redesigning the markov decision process in NIPA,
hence we would like to leave them for future work. More discussions on RL based future works are given in Appendix~\ref{CH:HAO:sec:future_direction_appdx}.

\subsection{Complexity of Algorithms}
\label{CH:HAO:sec:gia_alg_complexity}
Here we provide complexity analyses of the GIA algorithms used in the experiments as discussed and selected in Sec.~\ref{CH:HAO:sec:exp_setups}.
As also defined in algorithm description section from Appendix~\ref{CH:HAO:sec:gia_alg_desc}, $e^X_{\text{inner}}$ is the number of epochs optimized for node features,
$b$ is the number of maximum degree of vicious nodes, $d$ is the number of feature dimension,
$N_{V_c}$ is the number of $k$-hop neighbors of the victim nodes for perform one forwarding of a $k$-layer GNN,
$e_{\text{outer}}$ is the number of epochs for optimizing $A_\atk$,
$\gamma_c$ is the ratio of target nodes to attack in one batch,
$\gamma_\atk$ is the ratio of vicious nodes to inject in one batch.
\bgroup
\def\arraystretch{2}
\begin{table}[!h]
	\centering
	\caption{Complexity of various attacks.}
	\label{CH:HAO:tab:complexity}
	\resizebox{\textwidth}{!}{
		\begin{tabular}{clcc}
			\toprule
			\textbf{Type} & \multicolumn{1}{c}{\textbf{Algorithm}}                                      & \textbf{Time Complexity}                                                                                                                   & \textbf{Space Complexity} \\ \hhline{----}
			\multirow{3}{*}{\small{Gradient}}
			              & MetaGIA                                                                     & $O(|V_\atk|b(|V_c||V_\atk|\log(|V_c||V_\atk|)+e^X_{\text{inner}}d(|V_\atk|+N_{V_c})))$
			              & $O(|V_c||V_\atk|+e^X_{\text{inner}}d(|V_\atk|+N_{V_c}))$
			\\\hhline{~|-|-|-}
			              & AGIA                                                                        & $O(e_{\text{outer}}(e^A_{\text{inner}}|V_c||V_\atk|+(e^A_{\text{inner}}+e^X_{\text{inner}})d(N_{V_c}+|V_\atk|)))$
			              & $O(|V_c||V_\atk|+e^X_{\text{inner}}d(|V_\atk|+N_{V_c}))$
			\\\hhline{~|-|-|-}
			              & AGIA-SeqGIA                                                                 & $O(e_{\text{outer}}(|V_c|\log(|V_c|)+e^A_{\text{inner}}\gamma_c|V_c||V_\atk|+(e^A_{\text{inner}}+e^X_{\text{inner}})d(N_{V_c}+|V_\atk|)))$
			              & $O(\gamma_c|V_c|\gamma_\atk|V_\atk|+e^X_{\text{inner}}d(|V_\atk|+N_{V_c}))$
			\\\hhline{====}
			\multirow{4}{*}{\small{Heuristic}}
			              & PGD
			              & $O(|V_\atk|b+e^X_{\text{inner}}d(|V_\atk|+N_{V_c}))$
			              & $O(|V_\atk|b+e^X_{\text{inner}}d(|V_\atk|+N_{V_c}))$
			\\\hhline{~|-|-|-}
			              & TDGIA
			              & $O((|V_c|\log|V_c|+|V_\atk|b+e^X_{\text{inner}}d(|V_\atk|+N_{V_c}))$
			              & $O(|V_\atk|b+e^X_{\text{inner}}d(|V_\atk|+N_{V_c}))$
			\\\hhline{~|-|-|-}
			              & ATDGIA                                                                      & $O(|V_c|\log|V_c|+|V_\atk|b+e^X_{\text{inner}}d(|V_\atk|+N_{V_c}))$
			              & $O(|V_\atk|b+e^X_{\text{inner}}d(|V_\atk|+N_{V_c}))$
			\\\bottomrule
		\end{tabular}}
\end{table}
\egroup

\subsection{Details of Defense Baselines}
\label{CH:HAO:sec:baseline_detail_appdx}
Here we provide the categories of defense models used in the experiments as Sec.~\ref{CH:HAO:sec:exp_setups}.
We categorize all models into Vanilla, Robust and Extreme Robust (Combo).
Basically, popular GNNs are belong to vanilla category, robust GNNs are belong to robust categorty,
and a robust trick will enhance the robust level by one to the next Category.
Consistenly to the observation in GRB~\citep{grb}, we find adding Layer Normalization~\citep{layer_norm} before or between convulotion layers can enhance the model robustness.
We use LN to denote adding layer norm before the first convulotion layer and LNi to denote adding layer norm between convulotion layers.
\bgroup
\def\arraystretch{1.5}
\begin{table}[H]
	\caption{Defense model categories.}
	\label{CH:HAO:tab:defense_category}
	\resizebox{\textwidth}{!}{
		\begin{tabular}{lc|lc|lc|lc}
			\textbf{Model} & \textbf{Category} & \textbf{Model} & \textbf{Category} & \textbf{Model}  & \textbf{Category} & \textbf{Model}     & \textbf{Category} \\ \midrule
			GCN            & Vanilla           & GCN+LN         & Robust            & GCN+LNi         & Robust            & GCN+FLAG           & Robust            \\
			GCN+LN+LNi     & Combo             & GCN+FLAG+LN    & Combo             & GCN+FLAG+LNi    & Combo             & GCN+FLAG+LN+LNi    & Combo             \\\hline
			Sage           & Vanilla           & Sage+LN        & Robust            & Sage+LNi        & Robust            & Sage+FLAG          & Robust            \\
			Sage+LN+LNi    & Combo             & Sage+FLAG+LN   & Combo             & Sage+FLAG+LNi   & Combo             & Sage+FLAG+LN+LNi   & Combo             \\\hline
			GAT            & Vanilla           & GAT+LN         & Robust            & GAT+LNi         & Robust            & GAT+FLAG           & Robust            \\
			GAT+LN+LNi     & Combo             & GAT+FLAG+LN    & Combo             & GAT+FLAG+LNi    & Combo             & GAT+FLAG+LN+LNi    & Combo             \\\hline
			Guard          & Robust            & Guard+LN       & Combo             & Guard+LNi       & Combo             & EGuard+FLAG        & Combo             \\
			Guard+LN+LNi   & Combo             & EGuard+FLAG+LN & Combo             & EGuard+FLAG+LNi & Combo             & EGuard+FLAG+LN+LNi & Combo             \\\hline
			RGAT           & Robust            & RGAT+LN        & Combo             & RGAT+FLAG       & Combo             & RGAT+FLAG+LN       & Combo             \\\hline
			RobustGCN      & Robust            & RobustGCN+FLAG & Combo             &                 &                   &                    &                   \\
		\end{tabular}
	}
\end{table}
\egroup

\subsection{Details of Evaluation and Model Settings}
\label{CH:HAO:sec:model_setting_appdx}
\subsubsection{Model Setting}
By default, all GNNs used in our experiments have $3$ layers, a hidden dimension of $64$ for Cora, Citeseer, and Computers, a hidden dimension of $128$ for the rest medium to large scale graphs. We also adopt dropout~\citep{dropout} with dropout rate of $0.5$ between each layer. The optimizer we used is Adam~\citep{adam} with a learning rate of $0.01$. By default, we set total training epochs as $400$ and employ the early stop of $100$ epochs according to the validation accuracy. For the set of threshold in homophily defenders, we use PGD~\citep{pgd} to find the threshold which performs well on both the clean data and perturbed data. By default, we set the threshold as $0.1$, while for Computers and Reddit, we use $0.15$ for Guard and EGuard, and for Citeseer and Arxiv we use $0.2$ for RGAT.

For adversarial training with FLAG~\citep{flag}, we set the step size be $1\times 10^{-3}$, and train $100$ steps for Cora, $50$ steps for Citeseer, $10$ steps for the rest datasets. We empirically observe that FLAG can enhance both the natural accuracy and robustness of GNNs. We refer readers to the results for more details in Sec.~\ref{CH:HAO:sec:eval_nontarget_detailed} and Sec.~\ref{CH:HAO:sec:eval_target_detailed}.

\subsubsection{Evaluation Setting}
For final model selection, we select the final model with best validation accuracy.
For data splits, we follow the split methods in GRB~\citep{grb} which splits the datasets according to the node degrees, except for non-targeted attack on Arxiv where we use the official split to probe the performances of various methods in a natural setting.
For non-targeted attack, following previous works~\citep{tdgia,grb}, we select all test nodes as targets.
While for targeted attacks, we follow previous works~\citep{nettack} to select $200$ nodes with highest classification margin and lowest classification margin of the surrogate model. Then we randomly select $400$ nodes as targets. In other words, there are $800$ target nodes in total for targeted attack. Note for targeted attack, the natural accuracy on the target nodes might be different from normal test accuracy.
We also follow previous works to specify the attack budgets as Table.~\ref{CH:HAO:tab:datasets_budget_nontarget} for non-targeted attack and Table.~\ref{CH:HAO:tab:datasets_budget_target} for targeted attack.

During evaluation, we follow the black-box setting. Specifically, we firstly use the surrogate model to generate the perturbed graph, then we let the target models which has trained on the clean graph to test on the perturbed graph. We repeat the evaluation for $10$ times on Cora, Citeseer, Computers, and Arxiv, and $5$ times for Aminer and Reddit since model performs more stably on large graphs. Then we report mean test accuracy of the target models on the target nodes and omit the variance due to the space limit.

\subsubsection{Attacks Setting}
By default, we use PGD~\citep{pgd} to generate malicious node features. The learning step is $0.01$ and the default training epoch is $500$. We also employ the early stop of $100$ epochs according to the accuracy of the surrogate model on the target nodes. While for heuristic approaches such as TDGIA~\citep{tdgia} and ATDGIA, we follow the setting of TDGIA to update the features. Empirically, we find the original TDGIA feature update suits better for heuristic approaches while they show no advance over PGD for other approaches. Besides, as Table~\ref{CH:HAO:tab:complexity} shows, MetaGIA requires huge amount of time originally. Thus, to scale up, we use a batch update which updates the injected edges by a step size of $b$, i.e., the maximum degree of injected nodes, and limit the overall update epochs by $|V_\atk|/6$, where we empirically observe this setting performs best in Cora hence we stick it for the other datasets.

For the setting of $\lambda$ for HAO, we search the parameters within $0.5$ to $8$ by a step size of $0.5$ such that the setting of $\lambda$ will not degenerate the performance of the attacks on surrogate model. Besides heuristic approaches, we additionally use a hinge loss to stabilize the gradient information from $\mathcal{L}_\atk$ and $C(\gG,\gG')$, where the former can be too large that blurs the optimization direction of the latter. Take Cross Entropy with $\log\_\text{softmax}$ as an example, we adopt the following to constrict the magnitude of $\mathcal{L}_\atk$:
\begin{equation}
	\label{CH:HAO:eq:cross_entropy_hinge}
	\begin{aligned}
		{\mathcal{L}_\atk}_{[u]} & = (-H^{(k)}_{[u,Y_u]})\cdot \mathbf{1}\{\frac{\exp(H^{(k)}_{[u,Y_u]})}{\sum_{i}\exp(H^{(k)}_{[u,i]})}\geq \tau\}              \\
		                         & +\log(\sum_{i}\exp(H^{(k)}_{[u,i]}\cdot \mathbf{1}\{\frac{\exp(H^{(k)}_{[u,i]})}{\sum_{j}\exp(H^{(k)}_{[u,j]})}\geq \tau\})),
	\end{aligned}
\end{equation}
where $\mathbf{1}\{\frac{\exp(H^{(k)}_{[u,Y_u]})}{\sum_{i}\exp(H^{(k)}_{[u,i]})}\geq \tau\}$ can be taken as the predicted probability for $Y_u=u$ and $\tau$ is the corresponding threshold for hinge loss that we set as $1\times 10^{-8}$.

For the hyper-parameter setting of our proposed strategies in Sec.~\ref{CH:HAO:sec:gia_algorithm}, we find directly adopting $\lambda$ in PGD for $\lambda_X$ and setting $\lambda_A=0$ performs empirically better. Hence we stick to the setting for $\lambda_A$ and $\lambda_X$. For the weight of sparsity regularization term in AGIA, we directly adopt $1/b$. For the hyper-parameters in heuristic methods, we directly follow TDGIA~\citep{tdgia}. For SeqGIA, we set $\gamma_\atk$ be $\min(0.2,\floor{|V_c|/2b})$ and $\gamma_c=\min(|V_c|,\gamma_\atk|V_\atk|b)$ by default.

\subsection{Software and Hardware}
We implement our methods with PyTorch~\citep{pytorch} and PyTorch Geometric~\citep{pytorch_geometric}.
We ran our experiments on Linux Servers with 40 cores Intel(R) Xeon(R) Silver 4114 CPU @ 2.20GHz, 256 GB Memory, and Ubuntu 18.04 LTS installed. One has 4 NVIDIA RTX 2080Ti graphics cards with CUDA 10.2 and the other has 2 NVIDIA RTX 2080Ti and 2 NVIDIA RTX 3090Ti graphics cards with CUDA 11.3.

\section{More Experimental Results }
In this section, we provide more results from experiments about HAO to further validate its effectiveness.
Specifically, we provide full results of averaged attack performance across all defense models,
as well as initial experiments of HAO on two disassortative graphs.

\subsection{Full Results of Averaged Attack Performance}
In this section, we provide full results of averaged attack performance across all defense models, as a supplementary for Table~\ref{CH:HAO:tab:averaged_performance}.
\begin{table}[t]
	\caption{Full averaged performance across all defense models.}
	\label{CH:HAO:tab:averaged_performance_full_appdx}
	\scriptsize
	\centering\resizebox{\textwidth}{!}{
		\begin{tabular}{lcccccccc}
			\toprule
			\textbf{Model} & \textbf{Cora}$^\dagger$ & \textbf{Citeseer}$^\dagger$ & \textbf{Computers}$^\dagger$ & \textbf{Arxiv}$^\dagger$ & \textbf{Arxiv}$^\ddagger$ & \textbf{Computers}$^\ddagger$ & \textbf{Aminer}$^\ddagger$ & \textbf{Reddit}$^\ddagger$ \\\midrule
			Clean          & $84.74$                 & $74.10$                     & $92.25$                      & $70.44$                  & $70.44$                   & $91.68$                       & $62.39$                    & $95.51$                    \\
			PGD            & $61.09$                 & $54.08$                     & $61.75$                      & $54.23$                  & $36.70$                   & $62.41$                       & $26.13$                    & $62.72$                    \\
			\hfill +HAO    & $\underline{56.63}$     & $48.12$                     & $\underline{59.16}$          & $\underline{45.05}$      & $28.48$                   & $59.09$                       & $\underline{22.15}$        & $56.99$                    \\
			MetaGIA        & $60.56$                 & $53.72$                     & $61.75$                      & $53.69$                  & $28.78$                   & $62.08$                       & $32.78$                    & $60.14$                    \\
			\hfill +HAO    & $58.51$                 & $47.44$                     & $60.29$                      & $48.48$                  & $\underline{24.61}$       & $\underline{58.63}$           & $29.91$                    & $\mathbf{54.14}$           \\
			AGIA           & $60.10$                 & $54.55$                     & $60.66$                      & $48.86$                  & $32.68$                   & $61.98$                       & $31.06$                    & $59.96$                    \\
			\hfill +HAO    & $\mathbf{53.79}$        & $48.30$                     & $\mathbf{58.71}$             & $48.86$                  & $29.52$                   & $\mathbf{58.37}$              & $26.51$                    & $56.36$                    \\
			TDGIA          & $66.86$                 & $52.45$                     & $66.79$                      & $49.73$                  & $31.68$                   & $62.47$                       & $32.37$                    & $57.97$                    \\
			\hfill +HAO    & $65.22$                 & $\underline{46.61}$         & $65.46$                      & $49.54$                  & $\mathbf{22.04}$          & $59.67$                       & $22.32$                    & $\underline{54.32}$        \\
			ATDGIA         & $61.14$                 & $49.46$                     & $65.07$                      & $46.53$                  & $32.08$                   & $64.66$                       & $24.72$                    & $61.25$                    \\
			\hfill +HAO    & $58.13$                 & $\mathbf{43.41}$            & $63.31$                      & $\mathbf{44.40}$         & $29.24$                   & $59.27$                       & $\mathbf{17.62}$           & $56.90$                    \\\bottomrule
			\multicolumn{9}{l}{The lower is better. $^\dagger$Non-targeted attack. $^\ddagger$Targeted attack.  }
		\end{tabular}}
\end{table}

\subsection{More Results on Disassortative Graphs}
In this section, we provide initial investigation into the non-targeted attack performances of various GIA methods with or without HAO on disassortative graphs.
Specifically, we select Chameleon and Squirrel provided by~\cite{geom_gcn}. Statistics and budgets used for attack are given in Table~\ref{CH:HAO:tab:datasets_nonhomo} and Table~\ref{CH:HAO:tab:datasets_budget_nontarget_nonhomo}.

\begin{table}[!h]
	\centering\small
	\caption{Statistics of the disassortative datasets.}
	\label{CH:HAO:tab:datasets_nonhomo}%
	\begin{tabular}{lcccccc}
		\toprule
		\textbf{Datasets} & \textbf{Nodes} & \textbf{Edges} & \textbf{Classes} & \textbf{Avg. Degree} & \textbf{Label Homophily} & \textbf{Avg. Homophily} \\\midrule
		Chameleon         & $2277$         & $31,421$       & $5$              & $27.60$              & $0.26$                   & $0.62$                  \\
		Squirrel          & $5201$         & $198,493$      & $5$              & $76.33$              & $0.23$                   & $0.58$                  \\\bottomrule
	\end{tabular}%
\end{table}
We also heuristically specify the budgets for each dataset according the the number of target nodes and average degrees.

\begin{table}[!h]
	\centering\small
	\caption{Budgets for non-targeted attacks on disassortative datasets.}
	\label{CH:HAO:tab:datasets_budget_nontarget_nonhomo}
	\begin{tabular}{lcccccc}
		\toprule
		\textbf{Datasets} & \textbf{Nodes} & \textbf{Degree} & \textbf{Node Per.($\%$)} & \textbf{Edge Per.($\%$)} \\\midrule
		Chameleon         & $60$           & $100$           & $2.64\%$                 & $19.10\%$                \\
		Squirrel          & $90$           & $50$            & $1.73\%$                 & $2.27\%$                 \\\bottomrule
	\end{tabular}
\end{table}

For the settings of hyperparameters in attack methods and evaluation, we basically follow the same setup as given in Appendix~\ref{CH:HAO:sec:model_setting_appdx}.
In particular, we find using a threshold of $0.05$ for homophily defenders work best on Chameleon.
Besides, we also observe robust tricks can not always improve performances of GNNs on these graphs.
For example, we observe that using a large step-size of FLAG may degenerate the performances of GNNs on these datasets,
hence we use a smaller step-size of $5\times 10^{-4}$ as well as a small number of steps of $10$.
Moreover, using a LN before the first GNN layer may also hurt the performance.
For fair comparison, we remove these results from defenses.
Finally, in Table~\ref{CH:HAO:tab:eval_non_targeted_nonhomo_appdx},
we report both categorized defense results as Table~\ref{CH:HAO:tab:eval_non_targeted} as well as the averaged attack performance as Table~\ref{CH:HAO:tab:averaged_performance}.

\begin{table}[t]
	\caption{Results of non-targeted attacks on disassortative graphs.}
	\label{CH:HAO:tab:eval_non_targeted_nonhomo_appdx}
	\scriptsize\centering
	\begin{tabular}{@{}{l}*{10}{c}@{}}
		\toprule
		                                         &              & \multicolumn{4}{c}{{\small Chameleon ($\downarrow$)}} & \multicolumn{4}{c}{{\small Squirrel($\downarrow$)}}                                                                                                                                                                                                                               \\\cmidrule(lr){3-6}\cmidrule(lr){7-10}
		                                         & HAO          & \multicolumn{1}{c}{\textbf{Homo}}                     & \multicolumn{1}{c}{\textbf{Robust}}                 & \multicolumn{1}{c}{\textbf{Combo}} & \multicolumn{1}{c}{\textbf{AVG.}} & \multicolumn{1}{c}{\textbf{Homo}} & \multicolumn{1}{c}{\textbf{Robust}} & \multicolumn{1}{c}{\textbf{Combo}} & \multicolumn{1}{c}{\textbf{AVG.}} & \\ \midrule
		Clean                                    &              & $61.89$                                               & $65.18$                                             & $64.92$                            & $62.58$                           & $37.33$                           & $43.88$                             & $45.87$                            & $40.04$                             \\  \hdashline[0.5pt/1pt]
		\rule{0pt}{10pt}PGD                      &              & $61.89$                                               & $61.89$                                             & $63.61$                            & $\underline{33.24}$               & $35.66$                           & $36.28$                             & $40.54$                            & $\underline{26.03}$                 \\
		PGD                                      & $\checkmark$ & $52.78$                                               & $57.87$                                             & $59.31$                            & $38.00$                           & $33.32$                           & $39.36$                             & $35.83$                            & $26.37$                             \\\hdashline[0.5pt/1pt]
		\rule{0pt}{10pt}MetaGIA$^\dagger$        &              & $61.89$                                               & $61.89$                                             & $63.61$                            & $34.38$                           & $35.66$                           & $\mathbf{35.66}$                    & $39.40$                            & $26.09$                             \\
		MetaGIA$^\dagger$                        & $\checkmark$ & $49.25$                                               & $55.83$                                             & $55.73$                            & $33.63$                           & $34.07$                           & $38.26$                             & $\mathbf{35.24}$                   & $\mathbf{25.81}$                    \\
		AGIA$^\dagger$                           &              & $61.89$                                               & $61.89$                                             & $63.61$                            & $35.95$                           & $35.66$                           & $\underline{35.89}$                 & $39.93$                            & $26.93$                             \\
		AGIA$^\dagger$                           & $\checkmark$ & $\underline{43.98}$                                   & $\mathbf{48.88}$                                    & $\mathbf{53.33}$                   & $\mathbf{32.03}$                  & $35.69$                           & $36.31$                             & $36.40$                            & $26.77$                             \\\hdashline[0.5pt/1pt]
		\rule{0pt}{10pt}TDGIA                    &              & $61.95$                                               & $61.95$                                             & $63.76$                            & $41.17$                           & $35.66$                           & $\mathbf{35.66}$                    & $40.81$                            & $29.02$                             \\
		TDGIA                                    & $\checkmark$ & $46.36$                                               & $\underline{51.12}$                                 & $\underline{55.14}$                & $38.90$                           & $\mathbf{31.51}$                  & $38.21$                             & $\underline{35.63}$                & $28.65$                             \\
		ATDGIA                                   &              & $61.95$                                               & $61.95$                                             & $63.76$                            & $41.11$                           & $35.66$                           & $\mathbf{35.66}$                    & $41.62$                            & $29.62$                             \\
		ATDGIA                                   & $\checkmark$ & $\mathbf{36.93}$                                      & $57.75$                                             & $59.25$                            & $38.88$                           & $\underline{32.02}$               & $40.00$                             & $40.62$                            & $30.24$                             \\\hdashline[0.5pt/1pt]
		\hdashline[0.5pt/3pt]\rule{0pt}{10pt}MLP &              & \multicolumn{4}{c}{{$50.15$}}                         & \multicolumn{4}{c}{{$32.51$}}                                                                                                                                                                                                                                                     \\ \bottomrule
		\multicolumn{10}{l}{$^\downarrow$The lower number indicates better attack performance.  $^\dagger$Runs with SeqGIA framework on Computers and Arxiv.  }                                                                                                                                                                                                                                             \\                                                                                                                                                                                                                                                                                                                                              \\
	\end{tabular}%
\end{table}

From the results, we observe that, although our methods are not initially designed for disassortative graphs, HAO still brings empirical improvements.
Specifically, on Chameleon, HAO improves the attack performance up to $25\%$ against homophily defenders, up to $12\%$ against robust models, up to $10\%$ against extreme robust models,
and finally brings up to $3\%$ averaged test robustness of all models.
While on Squirrel, the improvements become relatively low while still non-trivial. For example, HAO improves the attack performance up to $4\%$ in terms of test robustness against homophily defenders.
We hypothesize the reason why HAO also works on disassortative graphs is because GNN can still learn the homophily information implicitly, e.g., similarity between class label distributions~\citep{is_homophily_necessity},
which we will leave the in-depth analyses to future work.

\section{Detailed Results of Attack Performance}

\subsection{Detailed Results of Non-Targeted Attacks}
\label{CH:HAO:sec:eval_nontarget_detailed}
In this section, we present the detailed non-targeted attack results of the methods and datasets used in our experiments for Table~\ref{CH:HAO:tab:eval_non_targeted}.
For simplicity, we only give the results of top $20$ robust models according to the averaged test accuracy against all attacks.

\bgroup
\def\arraystretch{1.6}
\begin{table}[!t]
	\caption{Detailed results of non-targeted attacks on Cora (1).}
	\label{CH:HAO:tab:eval_nontarget_detailed_cora1}
	\resizebox{\textwidth}{!}{
\begin{tabular}{l*{10}{>{$}c<{$}}}
			\toprule
			             & \text{\scriptsize EGuard+LNi+FLAG+LN} & \text{\scriptsize EGuard+FLAG+LN} & \text{\scriptsize EGuard+LNi+FLAG} & \text{\scriptsize Guard+LNi+LN} & \text{\scriptsize RGAT+FLAG+LN} & \text{\scriptsize GCN+LNi+FLAG+LN} & \text{\scriptsize RobustGCN+FLAG} & \text{\scriptsize RGAT+LN} & \text{\scriptsize Guard+LN} & \text{\scriptsize EGuard+FLAG} \\ \midrule
			Clean        & 83.48                                 & 84.17                             & 85.9                               & 79.56                           & 87.29                           & 86.37                              & 86.21                             & 85.29                      & 81.72                       & 85.56                          \\\hdashline[0.5pt/1pt]
			PGD          & 82.53                                 & 83.94                             & 85.74                              & 79.74                           & 76.78                           & 71.10                              & 69.57                             & 79.56                      & 81.44                       & 85.35                          \\
			\hfill +HAO  & 77.99                                 & 73.04                             & 66.25                              & 74.21                           & 68.09                           & 71.06                              & 70.3                              & 67.92                      & 68.12                       & 53.99                          \\
			MetaGIA      & 82.68                                 & 83.96                             & 85.86                              & 79.51                           & 75.18                           & 69.72                              & 69.4                              & 78.04                      & 81.59                       & 85.48                          \\
\bottomrule
		\end{tabular}}
\end{table}
\egroup

\bgroup
\def\arraystretch{1.5}
\begin{table}[t]
	\caption{Detailed results of non-targeted attacks on Cora (2).}
	\label{CH:HAO:tab:eval_nontarget_detailed_cora2}
	\resizebox{\textwidth}{!}{
		\begin{tabular}{l*{10}{>{$}c<{$}}}
			\toprule
			             & \text{\scriptsize RGAT+FLAG} & \text{\scriptsize Guard+LNi} & \text{\scriptsize RobustGCN} & \text{\scriptsize GCN+FLAG+LN} & \text{\scriptsize GCN+LNi+FLAG} & \text{\scriptsize RGAT} & \text{\scriptsize GAT+LNi+FLAG+LN} & \text{\scriptsize Sage+LNi+FLAG+LN} & \text{\scriptsize Guard} & \text{\scriptsize GCN+LNi+LN} \\ \midrule
			Clean        & 87.21                        & 83.18                        & 84.63                        & 85.86                          & 86.36                           & 85.74                   & 86.55                              & 84.95                               & 83.61                    & 84.47                         \\\hdashline[0.5pt/1pt]
			PGD          & 76.93                        & 83.11                        & 63.20                        & 62.55                          & 60.68                           & 79.28                   & 61.29                              & 61.84                               & 83.08                    & 58.46                         \\
			\hfill +HAO  & 62.35                        & 53.68                        & 62.60                        & 63.60                          & 61.69                           & 52.60                   & 62.81                              & 62.34                               & 44.02                    & 58.78                         \\
			MetaGIA      & 75.14                        & 83.08                        & 63.53                        & 59.18                          & 60.36                           & 77.97                   & 57.88                              & 61.01                               & 83.61                    & 58.10                         \\
\bottomrule
		\end{tabular}}
\end{table}
\egroup

\bgroup
\def\arraystretch{1.5}
\begin{table}[t]
	\caption{Detailed results of non-targeted attacks on Citeseer (1).}
	\label{CH:HAO:tab:eval_nontarget_detailed_citeseer1}
	\resizebox{\textwidth}{!}{
		\begin{tabular}{l*{10}{>{$}c<{$}}}
			\toprule
			             & \text{\scriptsize RGAT+LN} & \text{\scriptsize RGAT+FLAG+LN} & \text{\scriptsize EGuard+LNi+FLAG+LN} & \text{\scriptsize Guard+LNi+LN} & \text{\scriptsize RGAT} & \text{\scriptsize EGuard+FLAG+LN} & \text{\scriptsize RGAT+FLAG} & \text{\scriptsize EGuard+LNi+FLAG} & \text{\scriptsize EGuard+FLAG} & \text{\scriptsize GCN+LNi+FLAG+LN} \\ \midrule
			Clean        & 74.82                      & 75.72                           & 75.44                                 & 74.25                           & 74.85                   & 73.64                             & 75.56                        & 74.75                              & 73.57                          & 75.67                              \\\hdashline[0.5pt/1pt]
			PGD          & 71.00                      & 71.32                           & 75.19                                 & 74.21                           & 69.33                   & 73.55                             & 69.84                        & 74.83                              & 73.57                          & 57.97                              \\
			\hfill +HAO  & 71.00                      & 70.82                           & 66.07                                 & 73.04                           & 69.05                   & 61.55                             & 65.78                        & 50.01                              & 47.54                          & 58.77                              \\
			MetaGIA      & 70.32                      & 70.21                           & 75.15                                 & 74.21                           & 68.42                   & 73.55                             & 68.90                        & 74.83                              & 73.57                          & 56.36                              \\
\bottomrule
		\end{tabular}}
\end{table}
\egroup

\bgroup
\def\arraystretch{1.5}
\begin{table}[t]
	\caption{Detailed results of non-targeted attacks on Citeseer (2).}
	\label{CH:HAO:tab:eval_nontarget_detailed_citeseer2}
	\resizebox{\textwidth}{!}{
		\begin{tabular}{l*{10}{>{$}c<{$}}}
			\toprule
			             & \text{\scriptsize Guard+LN} & \text{\scriptsize Guard+LNi} & \text{\scriptsize RobustGCN+FLAG} & \text{\scriptsize Guard} & \text{\scriptsize GCN+LNi+FLAG} & \text{\scriptsize Sage+LNi+FLAG+LN} & \text{\scriptsize GAT+LNi+FLAG+LN} & \text{\scriptsize RobustGCN} & \text{\scriptsize GCN+LNi+LN} & \text{\scriptsize Sage+LNi+FLAG} \\ \midrule
			Clean        & 73.97                       & 74.41                        & 75.87                             & 74.78                    & 75.45                           & 73.89                               & 75.60                              & 75.46                        & 74.65                         & 73.70                            \\\hdashline[0.5pt/1pt]
			PGD          & 74.07                       & 74.28                        & 53.81                             & 74.70                    & 47.56                           & 46.82                               & 45.00                              & 39.77                        & 40.69                         & 40.11                            \\
			\hfill +HAO  & 48.48                       & 38.91                        & 51.10                             & 33.83                    & 49.19                           & 46.93                               & 44.06                              & 39.72                        & 40.79                         & 40.88                            \\
			MetaGIA      & 74.07                       & 74.28                        & 53.11                             & 74.70                    & 47.14                           & 46.13                               & 44.76                              & 39.84                        & 40.87                         & 40.13                            \\
\bottomrule
		\end{tabular}}
\end{table}
\egroup

\bgroup
\def\arraystretch{1.6}
\begin{table}[t]
	\caption{Detailed results of non-targeted attacks on Computers (1).}
	\label{CH:HAO:tab:eval_nontarget_detailed_computers1}
	\resizebox{\textwidth}{!}{
		\begin{tabular}{l*{10}{>{$}c<{$}}}
			\toprule
			             & \text{\scriptsize EGuard+LNi+FLAG+LN} & \text{\scriptsize Guard+LNi+LN} & \text{\scriptsize EGuard+FLAG+LN} & \text{\scriptsize Guard+LN} & \text{\scriptsize RGAT+FLAG+LN} & \text{\scriptsize RGAT+FLAG} & \text{\scriptsize EGuard+LNi+FLAG} & \text{\scriptsize Guard+LNi} & \text{\scriptsize RGAT+LN} & \text{\scriptsize RGAT} \\ \midrule
			Clean        & 91.04                                 & 90.88                           & 91.40                             & 91.23                       & 93.21                           & 93.32                        & 92.16                              & 91.95                        & 93.20                      & 93.17                   \\\hdashline[0.5pt/1pt]
			PGD          & 90.94                                 & 90.87                           & 91.41                             & 91.24                       & 81.59                           & 80.19                        & 88.24                              & 87.93                        & 79.68                      & 79.05                   \\
			\hfill +HAO  & 87.83                                 & 87.59                           & 80.41                             & 75.94                       & 81.80                           & 82.26                        & 64.18                              & 62.69                        & 79.29                      & 79.33                   \\
			MetaGIA      & 90.94                                 & 90.87                           & 91.41                             & 91.24                       & 81.58                           & 80.18                        & 88.23                              & 87.91                        & 79.68                      & 79.06                   \\
\bottomrule
		\end{tabular}}
\end{table}
\egroup

\bgroup
\def\arraystretch{1.6}
\begin{table}[t]
	\caption{Detailed results of non-targeted attacks on Computers (2).}
	\label{CH:HAO:tab:eval_nontarget_detailed_computers2}
	\resizebox{\textwidth}{!}{
		\begin{tabular}{l*{10}{>{$}c<{$}}}
			\toprule
			             & \text{\scriptsize GAT+FLAG+LN} & \text{\scriptsize EGuard+FLAG} & \text{\scriptsize Guard} & \text{\scriptsize RobustGCN+FLAG} & \text{\scriptsize GAT+LNi+FLAG+LN} & \text{\scriptsize RobustGCN} & \text{\scriptsize Sage+LNi+FLAG+LN} & \text{\scriptsize GAT+LNi+FLAG} & \text{\scriptsize GCN+LNi+FLAG+LN} & \text{\scriptsize GAT+LNi+LN} \\ \midrule
			Clean        & 92.17                          & 91.68                          & 91.55                    & 92.46                             & 92.40                              & 92.24                        & 91.71                               & 92.45                           & 93.22                              & 92.05                         \\\hdashline[0.5pt/1pt]
			PGD          & 82.31                          & 85.82                          & 84.91                    & 73.27                             & 77.91                              & 67.14                        & 63.83                               & 67.61                           & 54.96                              & 52.20                         \\
			\hfill +HAO  & 69.83                          & 55.62                          & 54.31                    & 72.73                             & 65.08                              & 68.80                        & 62.55                               & 54.93                           & 63.28                              & 69.19                         \\
			MetaGIA      & 82.31                          & 85.81                          & 84.91                    & 73.28                             & 77.91                              & 67.14                        & 63.83                               & 67.62                           & 54.96                              & 52.21                         \\
\bottomrule
		\end{tabular}}
\end{table}
\egroup

\bgroup
\def\arraystretch{1.6}
\begin{table}[t]
	\caption{Detailed results of non-targeted attacks on Arxiv (1).}
	\label{CH:HAO:tab:eval_nontarget_detailed_arxiv1}
	\resizebox{\textwidth}{!}{
		\begin{tabular}{l*{10}{>{$}c<{$}}}
			\toprule
			             & \text{\scriptsize Guard+LNi+LN} & \text{\scriptsize RGAT+LN} & \text{\scriptsize RGAT+FLAG+LN} & \text{\scriptsize EGuard+LNi+FLAG+LN} & \text{\scriptsize EGuard+FLAG+LN} & \text{\scriptsize Guard+LN} & \text{\scriptsize RobustGCN+FLAG} & \text{\scriptsize RobustGCN} & \text{\scriptsize GCN+LNi+FLAG+LN} & \text{\scriptsize Guard+LNi} \\ \hline
			             & 71.15                           & 70.95                      & 70.84                           & 69.50                                 & 69.46                             & 69.76                       & 67.85                             & 67.50                        & 71.40                              & 70.99                        \\\hdashline[0.5pt/1pt]
			PGD          & 71.11                           & 66.57                      & 66.61                           & 69.28                                 & 69.24                             & 69.62                       & 60.60                             & 60.81                        & 55.99                              & 70.26                        \\
			             & 68.68                           & 66.68                      & 66.60                           & 61.05                                 & 61.02                             & 58.92                       & 62.99                             & 62.89                        & 60.02                              & 47.84                        \\
			MetaGIA      & 71.09                           & 67.87                      & 67.67                           & 69.23                                 & 69.22                             & 69.59                       & 64.10                             & 64.10                        & 63.58                              & 70.40                        \\
\bottomrule
		\end{tabular}}
\end{table}
\egroup

\bgroup
\def\arraystretch{1.5}
\begin{table}[t]
	\caption{Detailed results of non-targeted attacks on Arxiv (2).}
	\label{CH:HAO:tab:eval_nontarget_detailed_arxiv2}
	\resizebox{\textwidth}{!}{
		\begin{tabular}{l*{10}{>{$}c<{$}}}
			\toprule
			             & \text{\scriptsize RGAT+FLAG} & \text{\scriptsize GCN+LNi+LN} & \text{\scriptsize RGAT} & \text{\scriptsize GCN+FLAG+LN} & \text{\scriptsize GAT+FLAG+LN} & \text{\scriptsize EGuard+LNi+FLAG} & \text{\scriptsize GCN+LN} & \text{\scriptsize EGuard+FLAG} & \text{\scriptsize GCN+LNi+FLAG} & \text{\scriptsize GAT+LNi+FLAG+LN} \\ \midrule
			             & 70.63                        & 71.38                         & 70.77                   & 70.00                          & 70.28                          & 69.37                              & 70.42                     & 69.34                          & 71.31                           & 71.00                              \\\hdashline[0.5pt/1pt]
			PGD          & 66.49                        & 54.46                         & 66.26                   & 54.21                          & 57.44                          & 68.04                              & 51.97                     & 68.03                          & 48.00                           & 57.65                              \\
			             & 57.18                        & 58.40                         & 55.38                   & 55.51                          & 59.16                          & 37.02                              & 52.45                     & 36.80                          & 52.75                           & 53.97                              \\
			MetaGIA      & 67.42                        & 62.88                         & 67.68                   & 58.54                          & 61.92                          & 68.48                              & 57.04                     & 68.40                          & 55.73                           & 61.56                              \\
\bottomrule
		\end{tabular}}
\end{table}
\egroup

\subsection{Detailed Results of Targeted Attacks}
\label{CH:HAO:sec:eval_target_detailed}
In this section, we present the detailed targeted attack results of the methods and datasets used in our experiments for Table~\ref{CH:HAO:tab:eval_targeted}.
For simiplicity, we only give the results of top $20$ robust models according to the averaged test accuracy against all attacks.

\bgroup
\def\arraystretch{1.5}
\begin{table}[t]
	\caption{Detailed results of targeted attacks on Computers (1).}
	\label{CH:HAO:tab:eval_target_detailed_computers1}
	\resizebox{\textwidth}{!}{
\begin{tabular}{l*{10}{>{$}c<{$}}}
			\toprule
			             & \text{\scriptsize EGuard+LNi+FLAG+LN} & \text{\scriptsize Guard+LNi+LN} & \text{\scriptsize EGuard+FLAG+LN} & \text{\scriptsize Guard+LN} & \text{\scriptsize Guard+LNi} & \text{\scriptsize EGuard+LNi+FLAG} & \text{\scriptsize RobustGCN+FLAG} & \text{\scriptsize RGAT+FLAG} & \text{\scriptsize RGAT+FLAG+LN} & \text{\scriptsize EGuard+FLAG} \\ \midrule
			Clean        & 90.96                                 & 90.76                           & 91.56                             & 91.11                       & 91.12                        & 91.29                              & 91.85                             & 92.83                        & 92.78                           & 90.75                          \\\hdashline[0.5pt/1pt]
			PGD          & 90.96                                 & 90.76                           & 91.56                             & 91.11                       & 89.38                        & 89.54                              & 72.36                             & 72.17                        & 74.28                           & 88.36                          \\
			\hfill +HAO  & 85.81                                 & 85.75                           & 79.51                             & 73.71                       & 65.01                        & 64.15                              & 72.58                             & 74.40                        & 74.08                           & 56.50                          \\
			MetaGIA      & 90.96                                 & 90.76                           & 91.56                             & 91.11                       & 88.93                        & 89.10                              & 73.81                             & 70.58                        & 72.24                           & 88.10                          \\
\bottomrule
		\end{tabular}}
\end{table}
\egroup

\bgroup
\def\arraystretch{1.5}
\begin{table}[t]
	\caption{Detailed results of targeted attacks on Computers (2).}
	\label{CH:HAO:tab:eval_target_detailed_computers2}
	\resizebox{\textwidth}{!}{
		\begin{tabular}{l*{10}{>{$}c<{$}}}
			\toprule
			             & \text{\scriptsize Guard} & \text{\scriptsize RobustGCN} & \text{\scriptsize RGAT} & \text{\scriptsize RGAT+LN} & \text{\scriptsize GAT+FLAG+LN} & \text{\scriptsize GAT+LNi+FLAG+LN} & \text{\scriptsize GAT+LNi+FLAG} & \text{\scriptsize GCN+LNi+FLAG+LN} & \text{\scriptsize GAT+LNi+LN} & \text{\scriptsize GCN+LNi+FLAG} \\ \midrule
			Clean        & 90.50                    & 92.07                        & 92.68                   & 92.76                      & 91.07                          & 91.90                              & 91.92                           & 92.25                              & 91.56                         & 92.35                           \\\hdashline[0.5pt/1pt]
			PGD          & 88.13                    & 70.40                        & 71.85                   & 72.65                      & 77.69                          & 75.25                              & 72.57                           & 63.08                              & 58.46                         & 60.79                           \\
			\hfill +HAO  & 54.96                    & 70.76                        & 71.78                   & 71.40                      & 71.03                          & 66.01                              & 62.46                           & 66.01                              & 70.49                         & 64.17                           \\
			MetaGIA      & 87.67                    & 71.78                        & 70.44                   & 71.33                      & 74.93                          & 73.12                              & 70.89                           & 62.54                              & 57.40                         & 60.71                           \\
\bottomrule
		\end{tabular}}
\end{table}
\egroup

\bgroup
\def\arraystretch{1.6}
\begin{table}[t]
	\caption{Detailed results of targeted attacks on Arxiv (1).}
	\label{CH:HAO:tab:eval_target_detailed_arxiv1}
	\resizebox{\textwidth}{!}{
		\begin{tabular}{l*{10}{>{$}c<{$}}}
			\toprule
			             & \multicolumn{1}{l}{Guard+LNi+LN} & \multicolumn{1}{l}{EGuard+LNi+FLAG+LN} & \multicolumn{1}{l}{Guard+LNi} & \multicolumn{1}{l}{EGuard+FLAG+LN} & \multicolumn{1}{l}{EGuard+LNi+FLAG} & \multicolumn{1}{l}{Guard+LN} & \multicolumn{1}{l}{EGuard+FLAG} & \multicolumn{1}{l}{Guard} & \multicolumn{1}{l}{RobustGCN+FLAG} & \multicolumn{1}{l}{RGAT} \\ \midrule
			             & 71.34                            & 71.22                                  & 71.22                         & 69.59                              & 70.59                               & 69.78                        & 68.88                           & 69.41                     & 67.28                              & 67.03                    \\\hdashline[0.5pt/1pt]
			PGD          & 71.31                            & 71.16                                  & 71.16                         & 69.47                              & 70.47                               & 69.69                        & 68.69                           & 69.19                     & 39.91                              & 39.13                    \\
			             & 69.38                            & 65.69                                  & 33.78                         & 47.41                              & 29.12                               & 38.00                        & 14.31                           & 13.94                     & 36.12                              & 36.06                    \\
			MetaGIA      & 71.03                            & 71.22                                  & 70.53                         & 69.59                              & 70.59                               & 69.78                        & 68.84                           & 69.28                     & 42.56                              & 41.81                    \\
\bottomrule
		\end{tabular}}
\end{table}
\egroup

\bgroup
\def\arraystretch{1.5}
\begin{table}[t]
	\caption{Detailed results of targeted attacks on Arxiv (2).}
	\label{CH:HAO:tab:eval_target_detailed_arxiv2}
	\resizebox{\textwidth}{!}{
		\begin{tabular}{l*{10}{>{$}c<{$}}}
			\toprule
			             & \multicolumn{1}{l}{RobustGCN} & \multicolumn{1}{l}{RGAT+LN} & \multicolumn{1}{l}{RGAT+FLAG+LN} & \multicolumn{1}{l}{GCN+LNi+FLAG} & \multicolumn{1}{l}{RGAT+FLAG} & \multicolumn{1}{l}{GCN+LNi+LN} & \multicolumn{1}{l}{GCN+LNi} & \multicolumn{1}{l}{GCN+LNi+FLAG+LN} & \multicolumn{1}{l}{GAT+LN} & \multicolumn{1}{l}{GAT+FLAG+LN} \\ \midrule
			             & 67.69                         & 72.06                       & 71.41                            & 71.34                            & 71.16                         & 71.97                          & 71.59                       & 71.75                               & 69.94                      & 69.94                           \\\hdashline[0.5pt/1pt]
			PGD          & 38.66                         & 40.31                       & 38.06                            & 32.19                            & 37.78                         & 29.09                          & 29.97                       & 29.72                               & 36.34                      & 38.84                           \\
			             & 37.22                         & 37.06                       & 34.28                            & 32.75                            & 23.69                         & 28.91                          & 29.56                       & 29.28                               & 28.88                      & 30.47                           \\
			MetaGIA      & 35.00                         & 42.56                       & 41.28                            & 30.28                            & 41.03                         & 28.91                          & 28.59                       & 28.50                               & 16.00                      & 14.84                           \\
\bottomrule
		\end{tabular}}
\end{table}
\egroup

\bgroup
\def\arraystretch{1.5}
\begin{table}[t]
	\caption{Detailed results of targeted attacks on Aminer (1).}
	\label{CH:HAO:tab:eval_target_detailed_aminer1}
	\resizebox{\textwidth}{!}{
		\begin{tabular}{l*{10}{>{$}c<{$}}}
			\toprule
			             & \text{\scriptsize EGuard+LNi+FLAG} & \text{\scriptsize EGuard+LNi+FLAG+LN} & \text{\scriptsize Guard+LNi} & \text{\scriptsize Guard+LNi+LN} & \text{\scriptsize EGuard+FLAG} & \text{\scriptsize Guard} & \text{\scriptsize RGAT+FLAG} & \text{\scriptsize Guard+LN} & \text{\scriptsize EGuard+FLAG+LN} & \text{\scriptsize RGAT} \\ \midrule
			             & 59.03                              & 58.06                                 & 60.72                        & 60.85                           & 57.06                          & 57.25                    & 61.75                        & 58.50                       & 58.81                             & 62.78                   \\\hdashline[0.5pt/1pt]
			PGD          & 55.25                              & 48.47                                 & 56.31                        & 49.40                           & 53.03                          & 53.16                    & 41.84                        & 49.72                       & 48.31                             & 40.72                   \\
			             & 39.06                              & 39.47                                 & 37.03                        & 39.40                           & 35.16                          & 34.62                    & 33.53                        & 29.69                       & 29.97                             & 31.75                   \\
			MetaGIA      & 52.09                              & 50.66                                 & 52.35                        & 49.81                           & 49.03                          & 48.97                    & 46.19                        & 48.34                       & 47.59                             & 45.81                   \\
\bottomrule
		\end{tabular}}
\end{table}
\egroup

\bgroup
\def\arraystretch{1.5}
\begin{table}[t]
	\caption{Detailed results of targeted attacks on Aminer (2).}
	\label{CH:HAO:tab:eval_target_detailed_aminer2}
	\resizebox{\textwidth}{!}{
		\begin{tabular}{l*{10}{>{$}c<{$}}}
			\toprule
			             & \text{\scriptsize RGAT+FLAG+LN} & \text{\scriptsize GCN+LNi+FLAG+LN} & \text{\scriptsize RGAT+LN} & \text{\scriptsize Sage+LNi+FLAG+LN} & \text{\scriptsize GCN+LNi+FLAG} & \text{\scriptsize GCN+LNi+LN} & \text{\scriptsize Sage+LNi+LN} & \text{\scriptsize GAT+LNi+LN} & \text{\scriptsize GAT+LNi+FLAG+LN} & \text{\scriptsize Sage+LNi+FLAG} \\ \midrule
			             & 62.66                           & 64.41                              & 63.78                      & 65.56                               & 63.91                           & 66.88                         & 65.44                          & 66.97                         & 65.78                              & 64.34                            \\\hdashline[0.5pt/1pt]
			PGD          & 31.97                           & 28.03                              & 29.75                      & 26.22                               & 26.81                           & 22.65                         & 23.78                          & 17.00                         & 16.66                              & 22.03                            \\
			             & 29.06                           & 28.16                              & 27.06                      & 26.44                               & 26.81                           & 23.17                         & 23.88                          & 17.58                         & 16.53                              & 22.06                            \\
			MetaGIA      & 41.38                           & 41.12                              & 40.78                      & 37.56                               & 36.72                           & 38.17                         & 36.56                          & 38.40                         & 37.31                              & 31.25                            \\
\bottomrule
		\end{tabular}}
\end{table}
\egroup

\bgroup
\def\arraystretch{1.5}
\begin{table}[t]
	\caption{Detailed results of targeted attacks on Reddit (1).}
	\label{CH:HAO:tab:eval_target_detailed_reddit1}
	\resizebox{\textwidth}{!}{
		\begin{tabular}{l*{10}{>{$}c<{$}}}
			\toprule
			             & \text{\scriptsize Guard+LNi+LN} & \text{\scriptsize RobustGCN} & \text{\scriptsize RobustGCN+FLAG} & \text{\scriptsize Guard+LNi} & \text{\scriptsize Guard+LN} & \text{\scriptsize EGuard+LNi+FLAG+LN} & \text{\scriptsize EGuard+FLAG+LN} & \text{\scriptsize Sage+LNi+FLAG+LN} & \text{\scriptsize Guard} & \text{\scriptsize EGuard+FLAG} \\ \midrule
			             & 94.47                           & 95.08                        & 95.30                             & 94.42                        & 94.61                       & 94.61                                 & 94.60                             & 97.10                               & 94.05                    & 94.08                          \\\hdashline[0.5pt/1pt]
			PGD          & 92.91                           & 84.81                        & 83.84                             & 93.03                        & 92.69                       & 92.69                                 & 92.53                             & 76.25                               & 92.44                    & 92.72                          \\
			             & 80.03                           & 86.12                        & 84.94                             & 75.53                        & 68.53                       & 69.31                                 & 69.34                             & 75.25                               & 56.44                    & 58.03                          \\
			MetaGIA      & 93.53                           & 88.25                        & 87.22                             & 93.28                        & 93.38                       & 93.66                                 & 93.59                             & 80.72                               & 92.40                    & 92.88                          \\
\bottomrule
		\end{tabular}}
\end{table}
\egroup

\bgroup
\def\arraystretch{1.5}
\begin{table}[t]
	\caption{Detailed results of targeted attacks on Reddit (2).}
	\label{CH:HAO:tab:eval_target_detailed_reddit2}
	\resizebox{\textwidth}{!}{
		\begin{tabular}{l*{10}{>{$}c<{$}}}
			\toprule
			             & \text{\scriptsize EGuard+LNi+FLAG} & \text{\scriptsize Sage+LNi+LN} & \text{\scriptsize GAT+LNi+FLAG+LN} & \text{\scriptsize Sage+FLAG+LN} & \text{\scriptsize Sage+LN} & \text{\scriptsize GAT+LNi+LN} & \text{\scriptsize GAT+FLAG+LN} & \text{\scriptsize GAT+LN} & \text{\scriptsize Sage+LNi+FLAG} & \text{\scriptsize GCN+LNi+FLAG+LN} \\ \midrule
			             & 94.07                              & 97.10                          & 95.19                              & 97.13                           & 97.11                      & 95.37                         & 94.49                          & 94.77                     & 97.09                            & 95.84                              \\\hdashline[0.5pt/1pt]
			PGD          & 92.69                              & 74.94                          & 75.91                              & 67.47                           & 63.75                      & 70.38                         & 73.53                          & 78.12                     & 57.16                            & 71.28                              \\
			             & 58.12                              & 73.91                          & 79.59                              & 67.72                           & 64.72                      & 72.91                         & 78.16                          & 76.47                     & 56.62                            & 70.91                              \\
			MetaGIA      & 92.84                              & 78.63                          & 68.16                              & 84.03                           & 82.53                      & 67.28                         & 59.91                          & 62.94                     & 65.69                            & 62.13                              \\
\bottomrule
		\end{tabular}}
\end{table}
\egroup

\chapter{Appendices of \pair}\label{APP:PAIR}

\section{Notations}
\label{app:notations}
We first list the notations for key concepts in \pair.
\begin{table}[ht]\small
	\caption{Notations for key concepts in \pair.}
	\centering
	\begin{tabular}{ll}
		\toprule
		\(\gX=\R^n\)              & the input space                                                                                                 \\\midrule
		\(\gY=\R\)                & the label space                                                                                                 \\\midrule
		\(\gZ=\R^d\)              & the latent space                                                                                                \\\midrule
		\(\varphi\)               & the featurizer $\varphi:\gX\rightarrow\gZ$ learns a latent representation for each input example                \\\midrule
		\(w\)                     & the classifier $w:\gZ\rightarrow\gY$                                                                            \\\midrule
		\(f\in\gF\)               & the predictor $f=w\circ\varphi:\gX\rightarrow\gY$ is composed of a featurizer and classifier                    \\
		\(\)                      & when $w$ is linear, $f$ can be simply represented via dot product $w\cdot\varphi$                               \\\midrule
		\(\envall\)               & the set of indices for all environments                                                                         \\
\bottomrule
	\end{tabular}
\end{table}

\section{More Discussions on Background and Future Directions}

\subsection{Background and related work}
\label{CH:PAIR:sec:related_work_appdx}

In this section, we provide more details of the backgrounds and closely related works to ours, in complementary to Sec.~\ref{CH:PAIR:sec:related_work}.

\textbf{The problem of OOD generalization.}
The problem of OOD generalization typically considers
a supervised learning setting based on the data $\dataset=\{\dataset^e\}_{e\in\envall}$
collected from multiple causally related environments $\envall$,
where a subset of samples $\dataset^e=\{X^e_i,Y^e_i\}$ from a single environment $e\in\envall$
are drawn independently from an identical distribution $\sP^e$~\citep{inv_principle}.
Given the data from training environments $\{\dataset^e\}_{e\in\envtrain}$,
the goal of OOD generalization is to find a predictor $f:\gX\rightarrow\gY$
that generalizes well to all (unseen) environments, i.e., to minimize
$\max_{e\in\envall}\gL_e(f)$, where $\gL_e$ is the empirical risk~\citep{erm} under environment $e$, $\gX$ and $\gY$ are the input and labeling spaces, respectively.
The predictor $f=w\circ\varphi$ is usually composed of a featurizer $\varphi:\gX\rightarrow\gZ$ that learns to extract useful features, and a classifier $w:\gZ\rightarrow\gY$ that makes predictions from the extracted features.
In practice, $\varphi$ is commonly implemented as a deep feature extractor, while $w$ is generically implemented as a simple dense linear classifier~\citep{domainbed,wilds,fishr,dare}.

\textbf{Existing solutions to OOD generalization.}
There exists a rich literature aiming to overcome the OOD generalization challenge, which usually appear as \emph{additional regularizations} of ERM~\citep{erm}.
The first line is the Domain Generalization works~\citep{DANN,CORAL,deep_DG,DouCKG19} that tries to regularize the learned features to be \textbf{domain-invariant}. However, \citet{DG_issue} show that the domain invariant features solely are not sufficient for guaranteed good OOD generalization. We refer readers to~\citet{domainbed} for more details of the literature about Domain Generalization.
Moreover, \citet{dro,DRSL,groupdro} aim to regularize the models to be \textbf{robust to mild distributional perturbations} of the training distributions such that the models are expected to perform well in unseen test environments. Following the line of distributional robustness, \citet{jtt,cnc,lisa} further propose advanced strategies to improve the robustness by assuming that models trained with ERM have strong reliance to spurious features.

Recently there is increasing interest in adopt theory of causality~\citep{causality,elements_ci,towards_causality} and introduce the \textbf{causal invariance} to the learned representations~\citep{inv_principle,causal_transfer,irmv1}.
The causal invariance is inspired by the assumption of Independent Causal Mechanism (ICM) in causality~\citep{elements_ci}. ICM assumes that conditional distribution of each variable given its causes (i.e., its mechanism) does not inform or influence the other conditional distributions~\citep{causality,elements_ci}. \citet{inv_principle} introduce the concept of environments which are generated by different interventions on certain variables involved in the underlying data generation process of $(X,Y)$. Despite of the changes to the intervened variables, the conditional distribution of intervened variables (they usually are the direct parents of $Y$ in the underlying causal graph) and $Y$ is invariant. Therefore, the invariant relationship can be leveraged to predict $Y$ and generalize to different environments. We refer interested readers to~\citet{inv_principle,towards_causality,ib-irm} for more details.
Inspired by the causal invariance principle, \citet{irmv1} propose the framework of Invariant Risk Minimization (IRM) that allows the adoption of the causal invariance in neural networks.
It further inspires plentiful invariant learning works~\citep{andmask,causal_matching,env_inference,clove,ib-irm,ciga,zin}.
At the heart of these works is the intuition that: When a predictor $w$ acting on $\varphi$ minimizes the risks in all of the environments simultaneously,
$\varphi$ is expected to discard the spurious signals while keeping the causally invariant signals.
Additionally, there can be more definitions and implementations of the invariance~\citep{iga,vrex,fish,fishr} which further encourage \textbf{agreements} at various levels across different environments. We refer interested readers to~\citet{fishr} for a detailed comparison and discussion.
As shown that most of the existing approaches encounter the optimization dilemma when learning the causal invariance, this work mainly focuses on resolving the optimization issue in learning the causal invariance defined by the framework of Invariant Risk Minimization~\citep{irmv1}, which is different from the literature of IRM variants or other OOD objectives that focus on proposing better objectives to learn the causal invariance.

\begin{figure}[ht]
	\subfigure[Pareto Front under MSE loss.]{
		\centering
		\includegraphics[width=0.3\textwidth]{Figures/PAIR/New_Pareto_Front_Sqls.pdf}
		\label{CH:PAIR:fig:pareto_front_mse_appdx}
	}
	\subfigure[Failure case under MSE loss.]{
		\centering
		\includegraphics[width=0.3\textwidth]{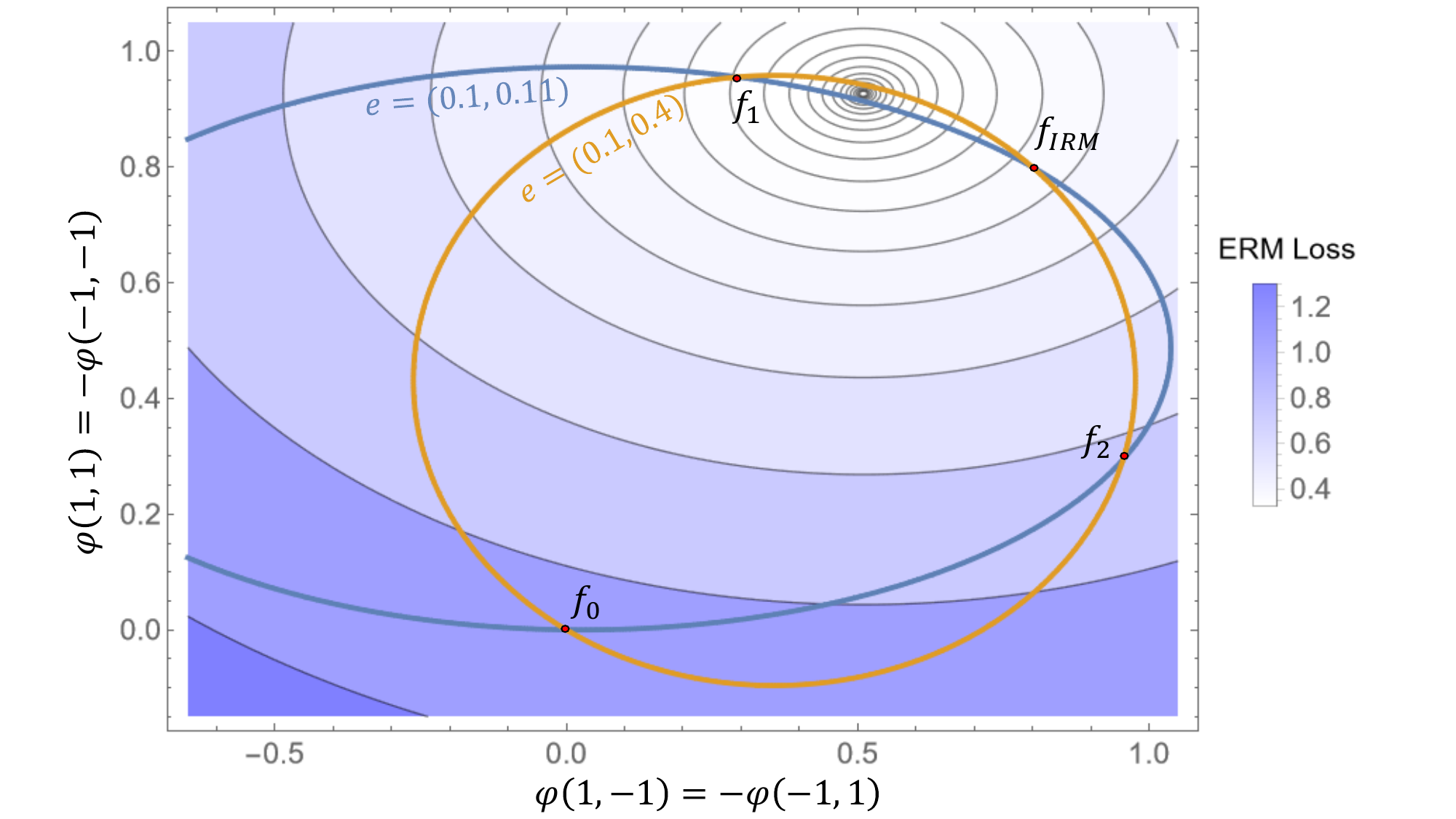}
		\label{CH:PAIR:fig:aistats_loss_mse_appdx}
	}
	\subfigure[Variance distribution under MSE loss.]{
		\centering
		\includegraphics[width=0.3\textwidth]{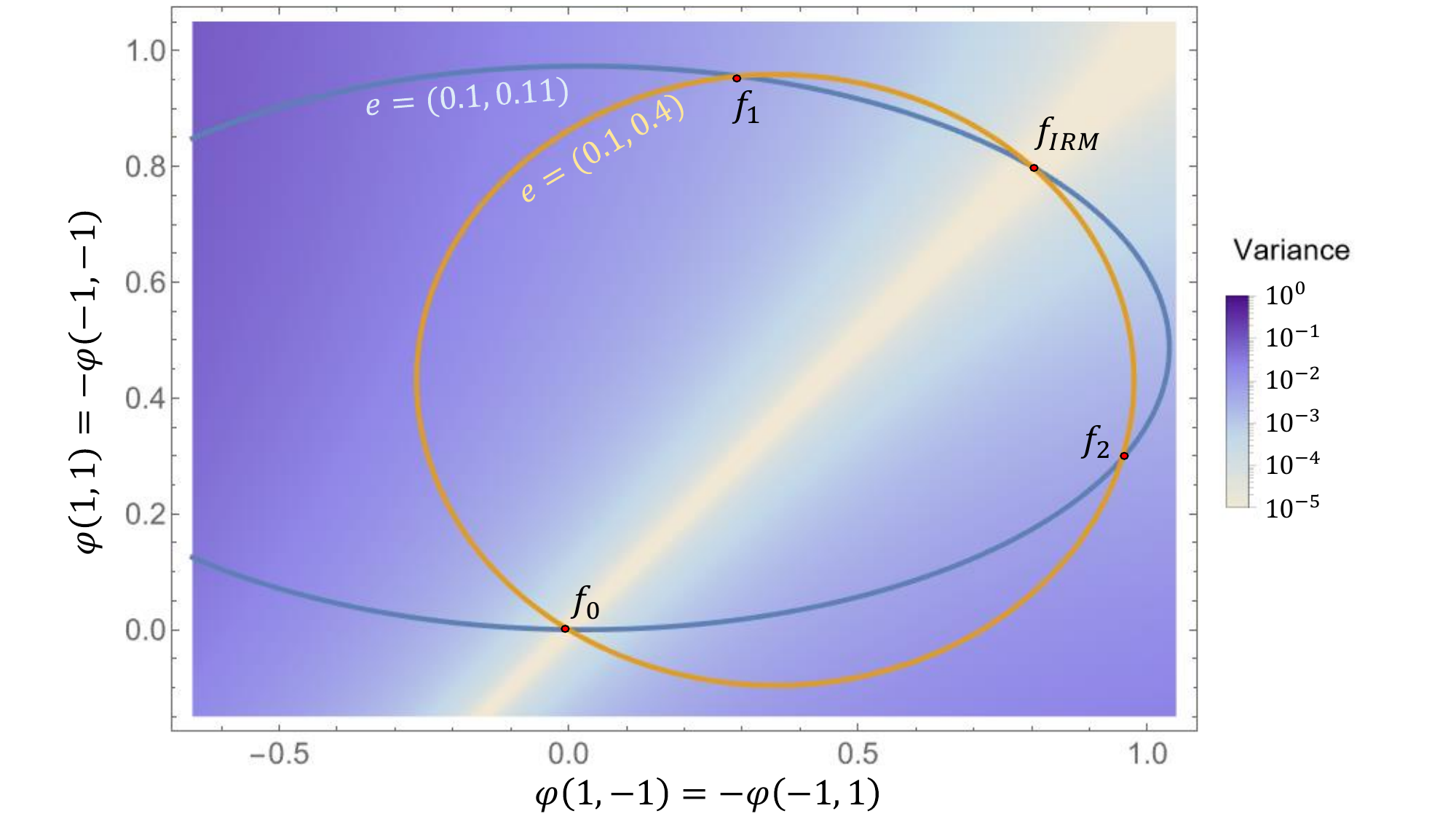}
		\label{CH:PAIR:fig:aistats_var_mse_appdx}
	}
	\vskip -0.15in
	\caption{Counterparts of Fig.~\ref{CH:PAIR:fig:aistats_fail}, Fig.~\ref{CH:PAIR:fig:aistats_var_mse} and Fig.~\ref{CH:PAIR:fig:pareto_front_mse} implemented in MSE loss.}
	\label{CH:PAIR:fig:aistats_fail_mse_appdx}
\end{figure}

\textbf{Optimization Dilemma in OOD Algorithms.}
Along with the developments of OOD methods, the optimization dilemma in OOD generalization is gradually perceived in the literature, and raises new puzzles to the community.
In fact, several recent works also notice the optimization dilemma in OOD algorithms, specifically, the trade-off between discovering the statistical correlations (i.e., ERM) and preventing the usage of spurious correlations (e.g., IRM).
Empirically, \citet{domainbed} observe that, with careful hyperparameter tuning and evaluation setting, many OOD algorithms cannot outperform ERM in domain generalization, demonstrating the difficulties of properly mitigating the trade-offs between OOD and ERM objectives in practice.
Moreover,
\citet{groupdro,gen_reweighted} find that, regularization on ERM, or sacrificing ERM performance, is usually needed for achieving satisfactory OOD performance.
A similar phenomenon has also been observed by~\citet{fund_tradeoff,innout,sadeghi2022on,sener2022domain,id_ood_inverse}, which aligns with our findings through Pareto front as shown in Fig.~\ref{CH:PAIR:fig:pareto_front_mse_appdx} and Fig.~\ref{CH:PAIR:fig:pareto_front_logl_appdx}.
Besides, \citet{BayesianIRM} find that IRM can easily overfit and learns unexpected features when applying IRM on large neural networks. \citet{SparseIRM} propose to alleviate this problem by imposing sparsity constraints. Orthogonal to \citet{BayesianIRM,SparseIRM} that focuses on the optimization consequences, we focus on the optimization process of OOD objectives.
In addition, \citet{rfc} find that, the performance of OOD algorithms largely relies on choosing proper pretraining epochs which aligns with our findings in Fig.~\ref{CH:PAIR:fig:sweep_acc}, hence propose to construct a ready-to-use features for stable OOD generalization performance.
Orthogonal to~\citet{rfc}, we focus on developing a better optimization scheme for OOD algorithms, including choosing the proper objectives and the achievability of the invariant predictors.
Besides, \citet{pareto_da} propose ParetoDA to leverage MOO to resolve the gradient conflicts amon the objectives in Domain Adaption. ParetoDA uses the guidance of validation loss based on the data that has the identical distribution to test distribution, to trade-off the conflicts in domain adaption objectives.
However, there can be multiple test domains, and the data that has identical distribution with the test domain is usually unavailable in OOD generalization. Therefore, ParetoDA is unsuitable for general OOD generalization methods.
Despite the increasing literature that perceives the OOD optimization dilemma, it remains an open problem on why there exists such a dilemma, and how to effectively mitigate the conflicts of ERM and OOD objectives and obtain a OOD generalizable solution.

\textbf{Further implications by the OOD optimization dilemma.} In addition to preventing finding a proper OOD solution, the OOD optimization dilemma also raises significant challenges for the model selection of OOD algorithms. \citet{domainbed} highlight this challenge with rigorous evaluation of OOD algorithms.
Similar to \pairo, \pairs resolves the dilemma by leveraging the OOD loss values and
explicitly considering the trade-offs of ERM and OOD performance.
We present more details in Sec.~\ref{CH:PAIR:sec:dobed_intro_appdx}.

\textbf{Multi-Objective Optimization (MOO) and its applications in Multi-Task Learning.}
MOO considers  solving $m$ objectives, w.r.t. $\{\gL_i\}_{i=1}^m$ losses,
i.e., \[\min_\theta\mL(\theta)=(\gL_1(\theta),...,\gL_m(\theta))^T\]~\citep{moo_book}.
A solution $\theta$ dominates another $\bar{\theta}$, i.e., $\mL(\theta)\preceq\mL(\bar{\theta})$, if $\gL_i(\theta)\leq\gL_i(\bar{\theta})$ for all $i$ and $\mL(\theta)\neq\mL(\bar{\theta})$.
A solution $\theta^*$ is called \textbf{Pareto optimal} if there exists no other solution that dominates $\theta^*$. The set of Pareto optimal solutions is called Pareto set, denoted as $\gP$, and its image is called \textbf{Pareto front}.
As it is usual that we cannot find a global optimal solution for all objectives in practice, hence Pareto optimal solutions are of particular value. The multiple-gradient descent algorithm (MGDA) is one of the commonly used approaches to efficiently find the Pareto optimal solutions~\citep{mgda} but is limited to low-dimensional data.
\citet{mtl_moo} then resolve the issue and apply MGDA to high-dimensional multi-task learning scenarios, where the objective conflicts may degenerate the performance when using linear scalarization.
As pure MGDA cannot find a Pareto optimal solution specified by certain objective preferences,
~\citet{pareto_mtl,nonlinear_scalar,pareto_exp_mtl} propose efficient methods to explore the Pareto set. \citet{epo} propose EPO to find the exact Pareto optimal solution with the specified objective preferences.
Although MOO has gained success in mitigating task conflicts in multi-task learning, it remains underexplored on whether and how we can leverage the MOO to model and resolve the ERM and OOD conflicts. Without a proper set of objectives and preference guidance, the existing MOO solvers are unable to obtain the desired solution for OOD generalization.

\subsection{Limitations and future directions}
\label{CH:PAIR:sec:future_appdx}
Although \pair effectively mitigates the objective conflicts and boosts the OOD performance via better optimization and model selection, the performance gain sometimes can decrease given the limitations of \pair. We believe future works can be built upon resolving the limitations of \pair.

From the optimizer perspective, the improvements of \pairo can decrease on some datasets. We hypothesize it is because of the inevitable stochastic gradient bias in all MGDA MOO solvers~\citep{moo_bias}, and potentially large variance in estimating the \irml penalties (e.g., \textsc{RxRx1} where both \irml and \vrex are shown to perform poor ), as we discussed in Appendix~\ref{CH:PAIR:sec:pair_opt_est_appdx}.

For \pairs, as discussed in Sec.~\ref{CH:PAIR:sec:pair_solution} that \pairs can mitigate the drawbacks of selecting models using an unreliable validation set (has a large gap from the test domain), the improvements will be a bit smaller when the gaps narrow down (e.g., \pacs using test domain validation accuracy). Besides, the estimation of satisfaction to Pareto optimality in \pairs can also be affected by the variances in estimating loss values in stochastic setting (e.g., \terra), as discussed in Appendix~\ref{CH:PAIR:sec:pair_selection_appdx}.

Additionally, \pair can also be applied to resolving OOD generalization issues in more complicated data domains~\citep{ciga,dps}, and other scenarios where gradient conflicts exist, such as the tradeoff between adversarial power and unnoticeability of the attacks~\citep{hao}, as well as improving the quality of representations in contrastive learning~\citep{contrast_reg}.

\section{More Details on IRM Failures and Fix}
\label{CH:PAIR:sec:irm_usecase_appdx}

In this section, we provide more details about the failure case of IRM and its effective fix from the perspective of MOO, in complementary to Sec.~\ref{CH:PAIR:sec:irm_usecase}.

\subsection{More detail about failure case of \irm}
\label{CH:PAIR:sec:irm_failure_appdx}
We follow~\citet{irm_aistats} to discuss the failure case of \irm. Specifically, given the problem setup as in Sec.~\ref{CH:PAIR:sec:related_work_appdx}, we are interested in the linear classification/regression following the setting. The loss values are measured as population loss in each environment.
\paragraph{Setting A (identical to (\citet{irm_aistats})):} $\hat{\mathcal{Y}} = \R, \mathcal{Y} \subseteq \R$, $\ell$ is either the square loss $\lsq(\hat{y}, y) \coloneqq \frac{1}{2}(\hat{y} - y)^2$, or the logistic loss $\llog(\hat{y}, y) \coloneqq \log{(1 + \exp{(-\hat{y}y)})}$ when $\mathcal{Y} =\{-1, 1\}$ (binary classification).

IRM approaches the problem by finding an invariant representation $\varphi:\gX\rightarrow\gZ$,
such that there exists a predictor $w: \gZ\rightarrow\gY$ acting on $\varphi$ that is
simultaneously optimal among $\envall$.
Hence, IRM leads to a challenging bi-level optimization problem~\citep{irmv1} as
\begin{equation}
	\label{CH:PAIR:eq:irm_appdx}
	\begin{aligned}
		\min_{w,\varphi} & \ \sum_{e\in\envtrain}\gL_e(w\circ\varphi),                                                   \\
		\text{s.t.}
		                 & \ w\in\argmin_{\bar{w}:\gZ\rightarrow\gY} \gL_e(\bar{w}\circ\varphi),\ \forall e\in\envtrain.
	\end{aligned}
\end{equation}
Given the training environments $\envtrain$, and functional spaces $\gW$ for $w$ and $\varPhi$ for $\varphi$,
predictors $w\circ\varphi$ satisfying the constraint are called invariant predictors,
denoted as $\gI(\envtrain)$.
When solving Eq.~\ref{CH:PAIR:eq:irm_appdx},
characterizing $\gI(\envtrain)$ is particularly difficult in practice,
given the access only to finite samples from a small subset of environments.
It is natural to introduce a restriction that $\gW$ is the space of linear functions on $\gZ=\R^d$~\citep{ntk}.
Furthermore, \citet{irmv1} argue that linear predictors actually do not provide additional representation power than \emph{scalar} predictors, i.e., $d=1,\gW=\gS=\R^1$. The scalar restriction on $\gW$ elicits a practical variant \irms as
\begin{equation}
	\label{CH:PAIR:eq:irms_appdx}
	\min_{\varphi} \ \sum_{e\in\envtrain}\gL_e(\varphi),
	\text{s.t.}
	\ \nabla_{w|w=1}\gL_e(w\cdot\varphi)=0,\ \forall e\in\envtrain.
\end{equation}
Let $\gI_\gS(\envtrain)$ denote the set of invariant predictors elicited by the relaxed constraint in \irms. It follows that $\gI (\envtrain)\subseteq \gI_\gS(\envtrain)$~\citep{irm_aistats}.
Yet, Eq.~\ref{CH:PAIR:eq:irms_appdx} remains a constrained programming. Hence, \citet{irmv1} introduce a soft-constrained variant \irml as
\begin{equation}
	\label{CH:PAIR:eq:irml_appdx}
	\min_{\varphi}  \sum_{e\in\envtrain}\gL_e(\varphi)+\lambda|\nabla_{w|w=1}\gL_e(w\cdot\varphi)|^2.
\end{equation}

\textbf{Theoretical Failure of Practical IRM Variants.}
Although the practical variants seem promising,
\citet{irm_aistats} show there exists huge gaps between the variants and the original IRM
such that both \irms and \irml can fail to capture the desired invariance, even being given the \emph{population loss} and \emph{infinite} amount of training environments.
The failure case, called two-bit environment~\citep{irm_aistats}, follows the setup of ColoredMNIST in IRM~\citep{irmv1}, and defines environments with two parameters $\alpha_e,\beta_e\in[0,1]$.
Each $\dataset_e$ is defined as
\begin{equation}
	\label{CH:PAIR:eq:twobit_env_appdx}
	Y\!:=\!\rad(0.5), X_1\!:=\!Y{\cdot}\rad(\alpha_e),X_2\!:=\!Y{\cdot}\rad(\beta_e),
\end{equation}
where $\rad(\sigma)$ is a random variable taking value $-1$ with probability $\sigma$ and $+1$
with probability $1-\sigma$. We denote an environment $e$ with $(\alpha_e,\beta_e)$ for simplicity.
The setup in IRM can be denoted as $\envalpha\!=\!\{(\alpha,\beta_e)\!:\!0\!<\!\beta_e\!<\!1\}$ where $X_1$ is the invariant feature as $\alpha$ is fixed for different $e$.

In the example given by~\citet{irmv1}, i.e.,
$\envtrain:=\{(0.25,0.1),(0.25,0.2)\}$,
\irms and \irml are shown to be able to learn the invariant predictor $f_\irm$ as the original IRM despite of the relaxation.
However, due to $\gI (\envtrain)\subseteq \gI_\gS(\envtrain)$, \citet{irm_aistats} show that the set of ``invariant predictors'' produced by \irms and \irml is broader than our intuitive sense.
For example, when given $\envtrain:=\{(0.1,0.11),(0.1,0.4)\}$, the solutions satisfying the constraint in \irms are those intersected points in Fig.~\ref{CH:PAIR:fig:aistats_fail} (The ellipsoids are the constraints).
Although $f_0,f_1,f_2,f_\irm\in\gI_\gS(\envtrain)$, both \irms and \irml prefer $f_1$ instead of $f_\irm$ (the predictor elicited by the original IRM), as $f_1$ has the smallest ERM loss.
In fact, \citet{irm_aistats} prove that, the failure can happen in a wide range of environments with $\alpha<0.1464$ and $\alpha>0.8356$,
even being given \emph{infinite} number of additional environments, under MSE loss.
It follows that $\gI(\envtrain)\subsetneq\gI_\gS(\envtrain)$. In other words,
the relaxation in \irms and \irml will introduce additional ``invariant predictors'' which however do not satisfy the original IRM constraint.
Both \irms and \irml will prefer those ``invariant predictors'' when they have lower ERM loss than $f_\irm$,
demonstrating the significant theoretical gap between the practical variants and the original IRM.

\begin{figure}[ht]
	\subfigure[Pareto Front under Logistic loss.]{
		\centering
		\includegraphics[width=0.28\textwidth]{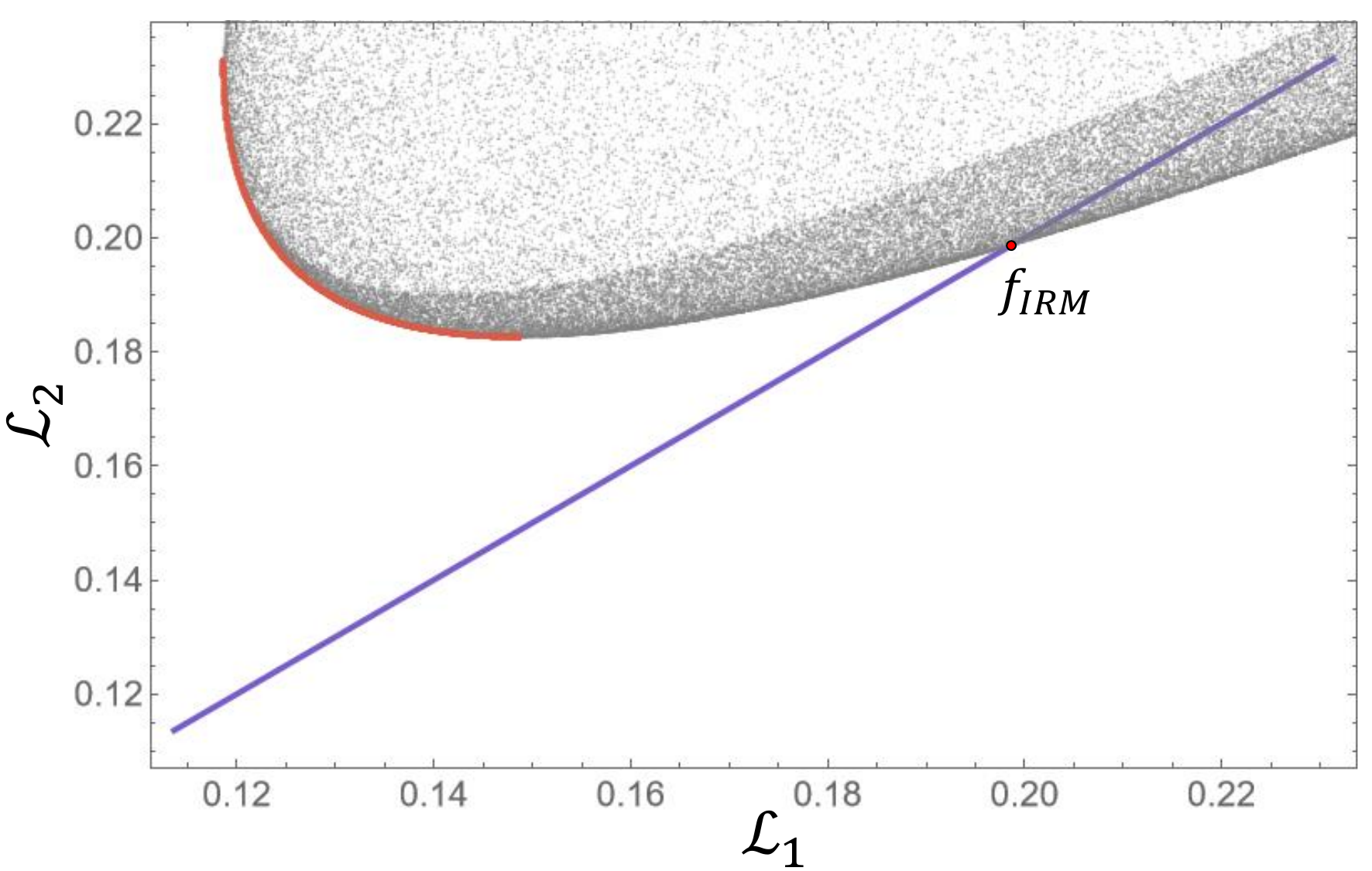}
		\label{CH:PAIR:fig:pareto_front_logl_appdx}
	}
	\subfigure[Failure case under Logistic loss.]{
		\centering
		\includegraphics[width=0.31\textwidth]{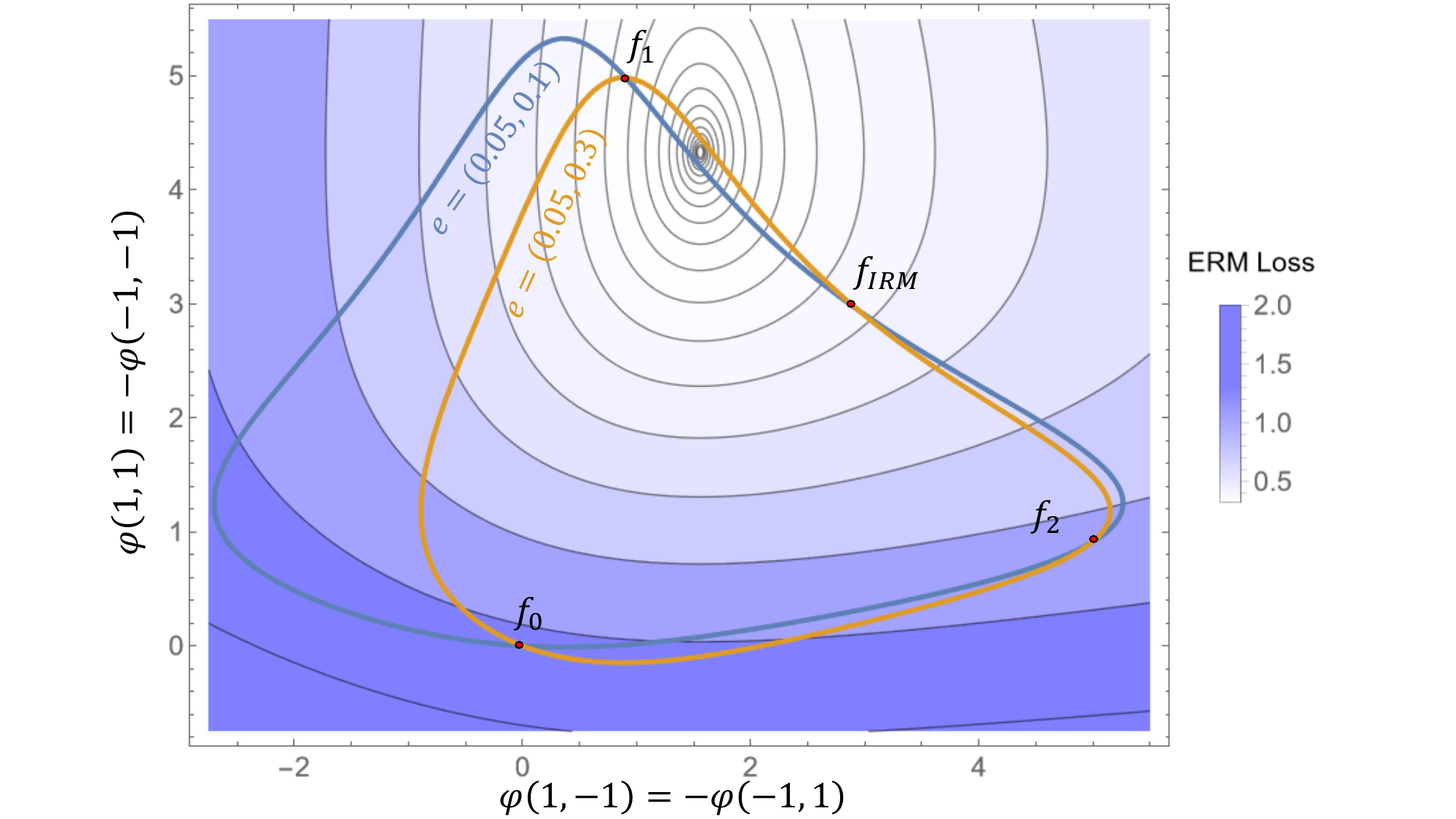}
		\label{CH:PAIR:fig:aistats_loss_logl_appdx}
	}
	\subfigure[Variance distribution under Logistic loss.]{
		\centering
		\includegraphics[width=0.32\textwidth]{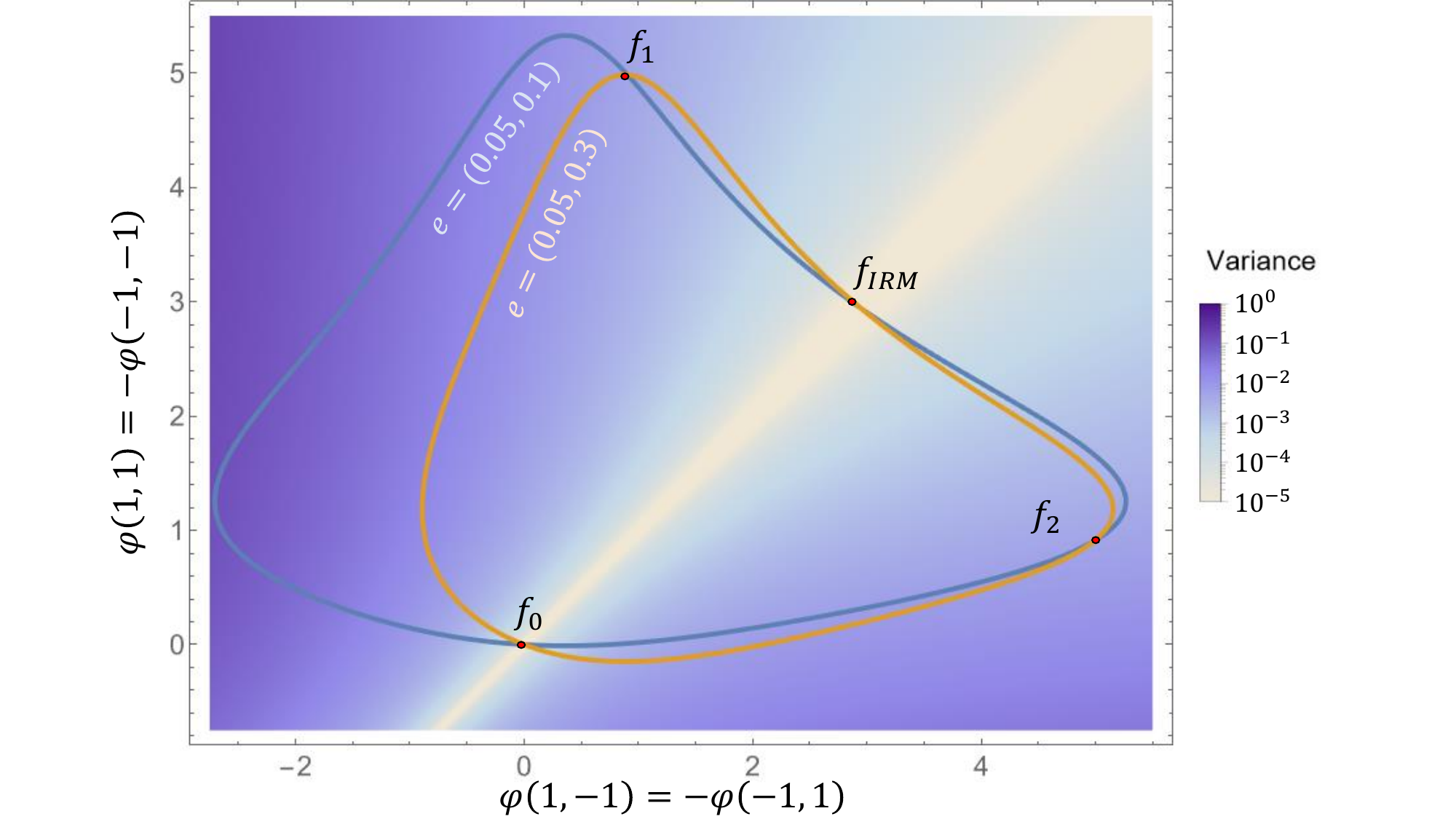}
		\label{CH:PAIR:fig:aistats_var_logl_appdx}
	}
	\vskip -0.15in
	\caption{Counterparts of Fig.~\ref{CH:PAIR:fig:aistats_fail}, Fig.~\ref{CH:PAIR:fig:aistats_var_mse} and Fig.~\ref{CH:PAIR:fig:pareto_front_mse} implemented in Logistic loss.}
	\label{CH:PAIR:fig:aistats_fail_logl_appdx}
\end{figure}

\textbf{More visualization results of the failure cases.}
In the main paper, we visualize the Pareto front, ERM loss distribution, and the variance distribution of the failure case given MSE losses, given the environment setup of $\envtrain:=\{(0.1,0.11),(0.1,0.4)\}$.
We plot Fig.~\ref{CH:PAIR:fig:aistats_fail} and Fig.~\ref{CH:PAIR:fig:aistats_var_mse} based on the Mathematica code provided by~\citet{irm_aistats}, where we focus on the odd predictors due to the symmetry in two-bit environments, i.e., predictors satisfying $\varphi(1,-1)=-\varphi(-1,1)$ and $\varphi(1,1)=-\varphi(-1,-1)$.
Since Fig.~\ref{CH:PAIR:fig:aistats_fail}, Fig.~\ref{CH:PAIR:fig:aistats_var_mse} and Fig.~\ref{CH:PAIR:fig:pareto_front_mse} are implemented in MSE loss, for completing the discussion under Setting A~\citep{irm_aistats}, we also give their logistic counterparts as in Fig.~\ref{CH:PAIR:fig:aistats_fail_logl_appdx}.

\begin{figure}[ht]
	\subfigure[\irml in the original CMNIST.]{
		\centering
		\includegraphics[width=0.31\textwidth]{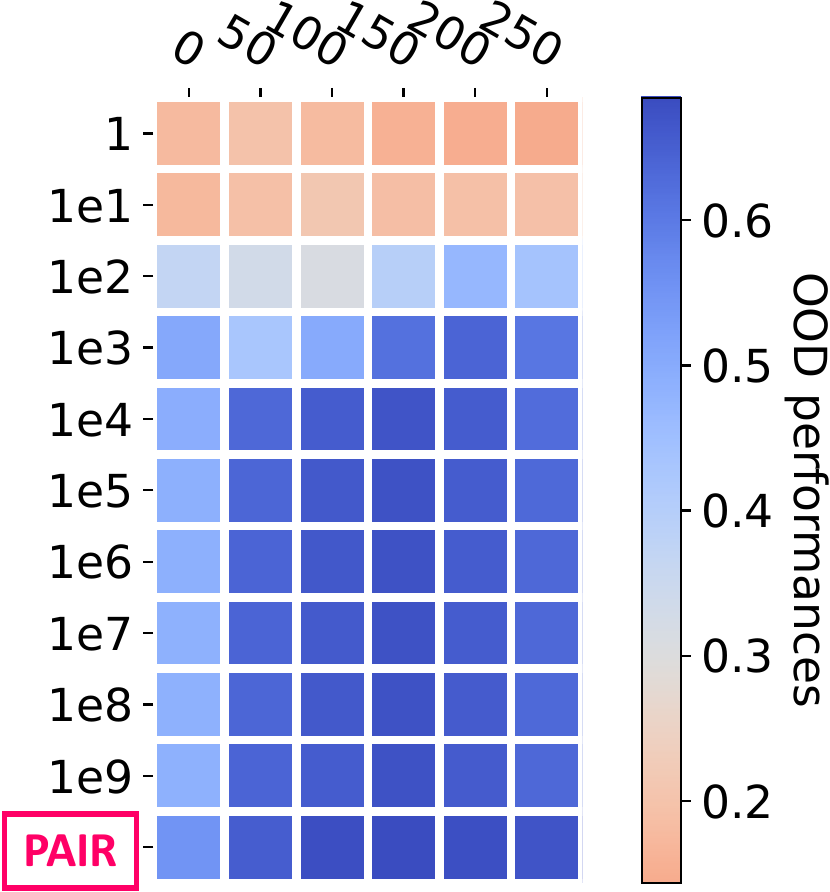}
	}
	\subfigure[\irml in CMNIST-m.]{
		\centering
		\includegraphics[width=0.31\textwidth]{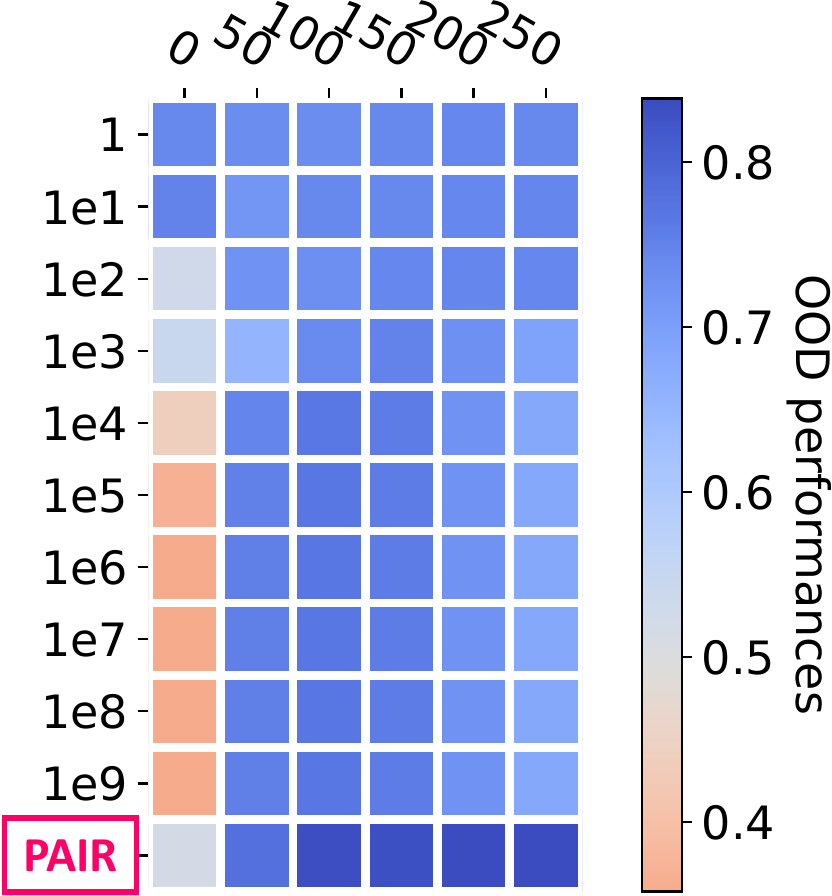}
	}
	\subfigure[Detailed performance of \irml.]{
		\centering
		\includegraphics[width=0.31\textwidth]{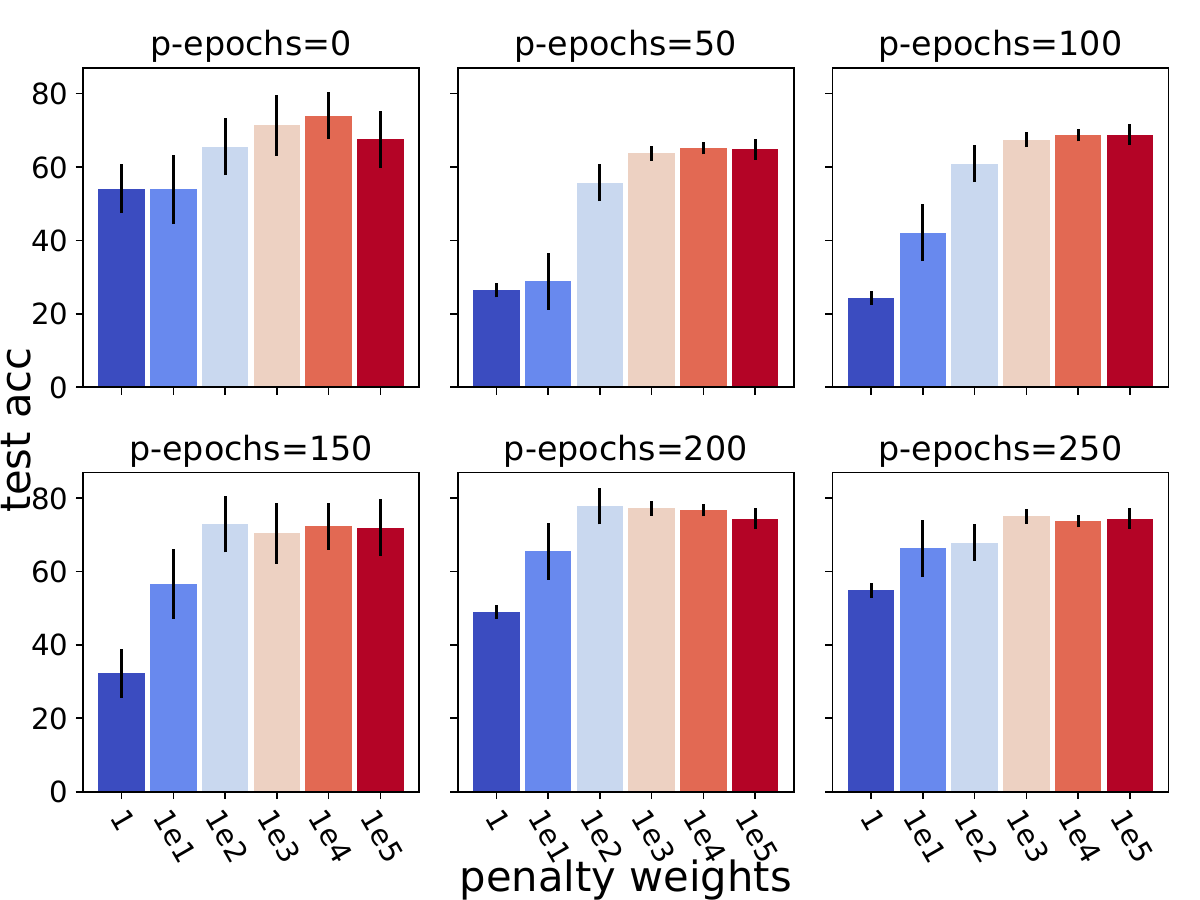}
		\label{CH:PAIR:fig:c12_hp_irm_appdx}
	}
	\caption{Performances of \irml in CMNIST and CMNIST-m under different hyperparameters.}
	\label{CH:PAIR:fig:sweep_irm_appdx}
\end{figure}

\textbf{Practical Drawback of Practical IRM Variants.}
In addition to the theoretical gap, the optimization of \irml is also difficult due to the conflicts between the IRM penalty and ERM penalty in Eq.~\ref{CH:PAIR:eq:irml_appdx}. It often requires significant efforts for choosing proper hyperparameters such as pretraining epochs and IRM penalty weights, i.e., $\lambda$. Otherwise, \irml may not enforce the constraint in \irms, hence will lead to unsatisfactory performance, as shown in Fig.~\ref{CH:PAIR:fig:sweep_acc}.
We argue that the gradient conflicts generally exist in OOD optimization for various objectives, in Fig.~\ref{CH:PAIR:fig:grad_conflict}, we visualize the cosine similarity between the gradients produced by ERM and OOD objectives, which is averaged from $50$ epochs after the pretraining. It can be found that, all of the OOD objectives~\citep{irmv1,vrex,ib-irm,iga,fishr,clove,sd} tend to yield gradients that have a lower cosine similarity with those of ERM. The generally existed conflicts can further lead to suboptimal performances of these OOD objective in practice even with exhaustive parameter tunning.

In complementary to Fig.~\ref{CH:PAIR:fig:sweep_acc}, we provide full results in Fig.~\ref{CH:PAIR:fig:sweep_irm_appdx}, where we show the results of \irml under different penalty weights ($y$-axis) and pretraining epochs ($x$-axis) on \cmnist~\citep{irmv1} (CMNIST) as well as the failure case~\citep{irm_aistats} (CMNIST-m), or $\envtrain:=\{(0.1,0.2),(0.1,0.25)\}$ described in two-bit environment.
It can be found that the performances of \irml are highly dependent on proper tuning of pretraining epochs and the penalty weights. The dependence grows stronger when \irml is shown to be unrobust on CMNIST-m.
We also provide a more detailed results of \irml on CMNIST-m in Fig.~\ref{CH:PAIR:fig:c12_hp_irm_appdx}, where the dependence can be clearly observed.
In contrast, \pair performs robustly well under different pretraining epochs, using a default preference $(1,1e10,1e12)$ to \erm, \irml and \vrex objectives, respectively. In Sec.~\ref{CH:PAIR:sec:experiments}, we provide more evidences to demonstrate the power of \pairo.

\subsection{Discussions of objectives in \pair}
\label{CH:PAIR:sec:discuss_pair_objs_appdx}

\begin{wrapfigure}{r}{0.33\textwidth}
	\vspace{-0.3in}
	\includegraphics[width=0.33\textwidth]{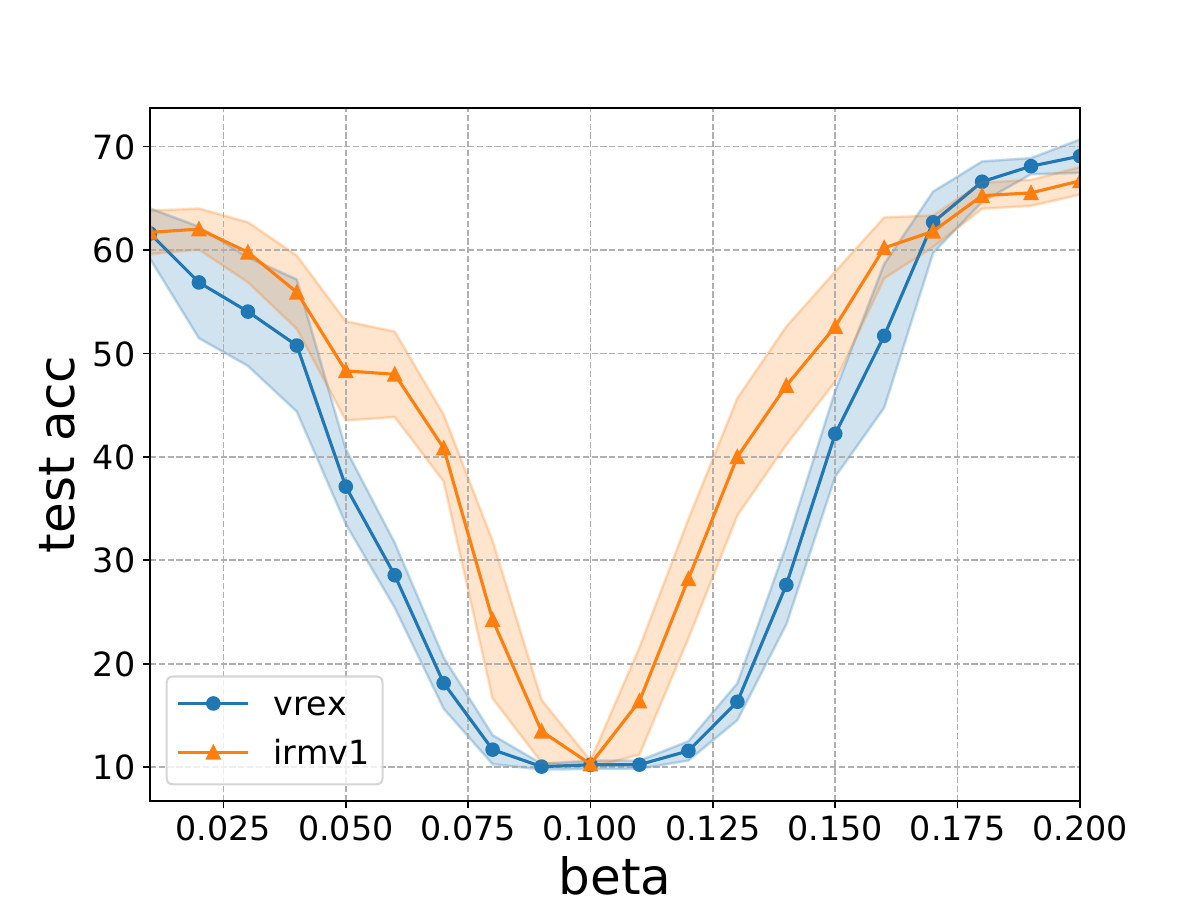}
	\vskip -0.1in
	\caption{Drawbacks of V-REx in practice.}
	\label{CH:PAIR:fig:vrex_failure}
	\vspace{-0.1in}
\end{wrapfigure}

In Sec.~\ref{CH:PAIR:sec:irm_pair_sol}, we derive a group of ideal objectives for improving the robustness of \irml, shown as the following:
\begin{equation}
	\label{CH:PAIR:eq:irmx_moo_appdx}
	(\text{\irmx})\qquad\qquad\qquad\qquad\min_{\varphi} (\gL_\erm,\gL_\irm,\gL_\vrex)^T.\qquad\qquad\qquad\qquad
\end{equation}
We prove in Proposition~\ref{CH:PAIR:thm:recovep_iRM_paper_appdx} that \irmx is able to solve a large number of failure cases of \irms and \irml, and recovers the set of invariant predictors produced by the original \irm. However, motivated readers might be interested in the reasons for keeping \irml in \irmx, since \vrex solely could resolve the two-bit environment failure case.

Theoretically, Proposition~\ref{CH:PAIR:thm:recovep_iRM_paper_appdx} requires also the invariant predictors produced by \irms, i.e., $\gI_\gS(\gE)$, to recover the invariant predictors yielded by \irm. Nevertheless, it considers only the ideal case.
In the next, we elaborate on a detailed discussion from the empirical side.

\textbf{Drawbacks of Robust Minimization in Practice.}
After showing REx~\citep{vrex} can help avoiding the failure cases of \irms, a natural question is that, does $\gL_\irm$ remain necessary? We find the answer is ``Yes''.
In Fig.~\ref{CH:PAIR:fig:vrex_failure}, we use a modified example of $\envtrain=\{(0.25,0.1),(0.25,\beta)\}$ with ColoredMNIST~\citep{irmv1}, where we change the variance between two environments through different $\beta$. It can be found that, as the variance between two environments getting closer, the performance of REx~\citep{vrex} (denoted as vrex) drops more sharply than \irml (denoted as irmv1). %
The main reason is that, as the variation of spurious signals in two environments tends to be smaller, the gradient signal of $\var(\{\gL_e\}_{e\in\envtrain})$ tends to vanish, while the signals from $\gL_\irm$ maintains. This issue can be more serious in stochastic gradient descent where the estimates of the variance of $\{\gL_e\}_{e\in\envtrain}$ in minibatches tend to be noisy, leading to weaker signals.

\subsection{More details on the extrapolation example}
\label{CH:PAIR:sec:causal_extrapolate_appdx}
In this section, we provide more details and results about the extrapolation example that examines the recovery of causal invariance, in complementary to Sec.~\ref{CH:PAIR:sec:causal_extrapolate}.

We first restate the definition of causal invariance specified by~\citet{inv_principle,irmv1,irm_aistats} as in Definition~\ref{def:causal_inv_appdx}.
\begin{definition}(Causal Invariance)\label{def:causal_inv_appdx}
	Given a predictor $f:=w\circ\varphi$, the representation produced by the featurizer $\varphi$ is invariant over $\envall$ if and only if for all $e_1,e_2\in\envall$, it holds that
	\[
		\mathbb{E}_{\gD_{e_1}}[Y|\varphi(X)=z]=\mathbb{E}_{\gD_{e_2}}[Y|\varphi(X)=z],
	\]
	for all $z\in\gZ_\varphi^{e_1}\cap\gZ_\varphi^{e_2}$,
	where
	$\gZ_\varphi^e:=\{\varphi(X)|(X,Y)\in\text{supp}(\gD_e)\}$.
	\vspace{-0.05in}
\end{definition}

Then, we construct a regression example from $\gX:\R^2\rightarrow\gY:\R$. The input $X$ is a two dimensional inputs, i.e., $X=(X_1,X_2)$.
$X_1$ is designed to be the invariant feature, i.e., $Y =\sin(X_1)+1$, while $X_2$ is designed to be the spurious feature that can be controlled to be spuriously correlated with label $Y$.
The environments are synthesized according to different sampling methods.

\begin{figure}[t]
	\subfigure[Uniform.]{
		\includegraphics[width=0.23\textwidth]{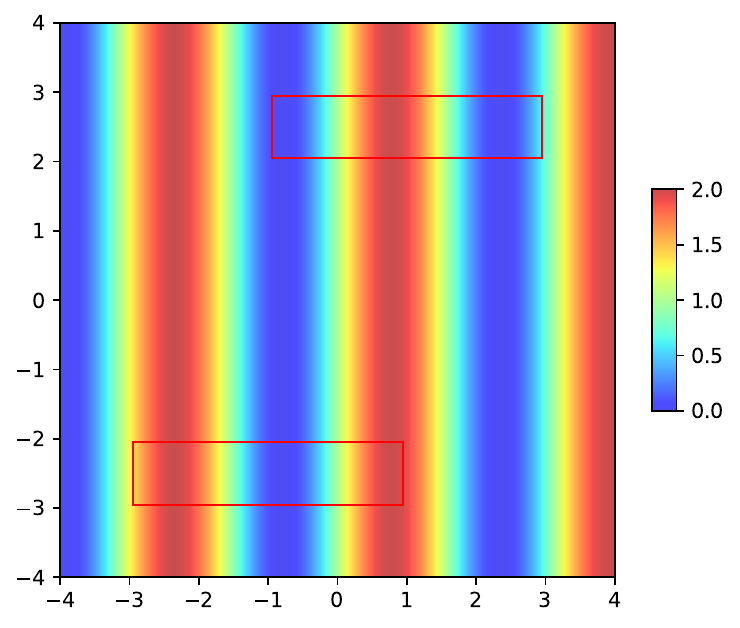}
		\label{CH:PAIR:fig:linextra_uniform_appdx}
	}
	\subfigure[ERM.]{
		\includegraphics[width=0.21\textwidth]{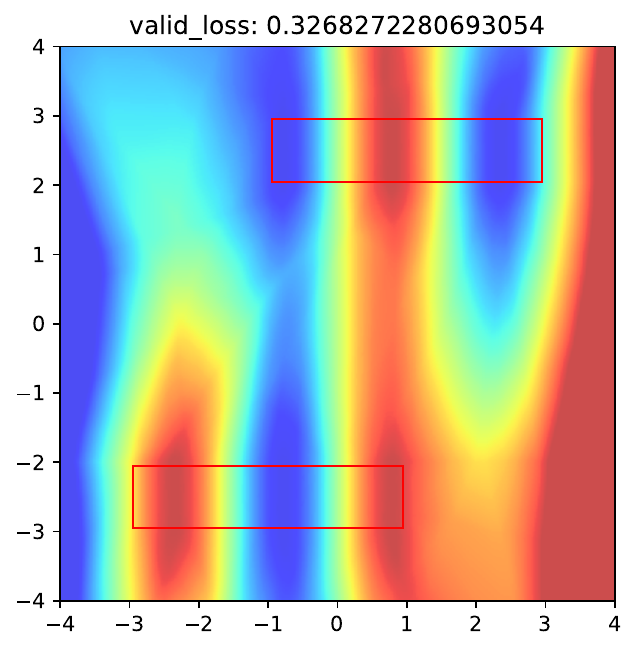}
		\label{CH:PAIR:fig:linextra_uniform_erm_appdx}
	}
	\subfigure[Gaussian.]{
		\includegraphics[width=0.23\textwidth]{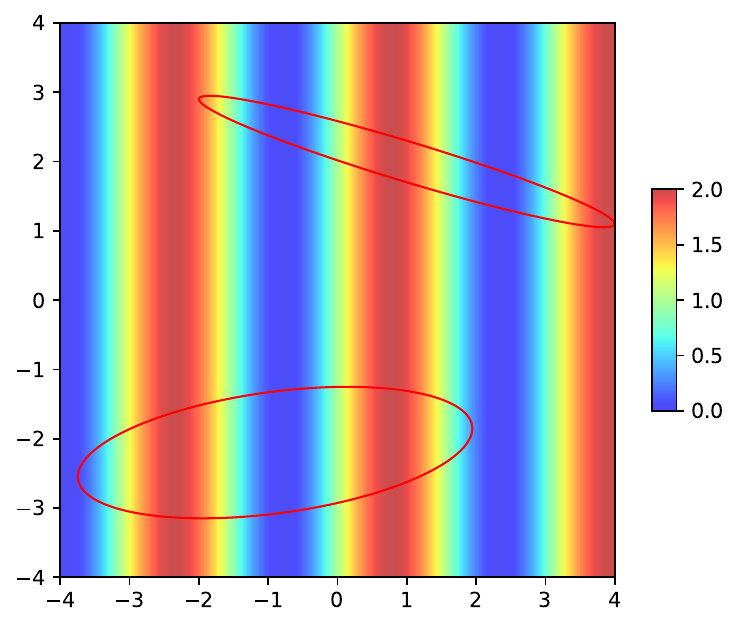}
		\label{CH:PAIR:fig:linextra_gau_appdx}
	}
	\subfigure[ERM.]{
		\includegraphics[width=0.21\textwidth]{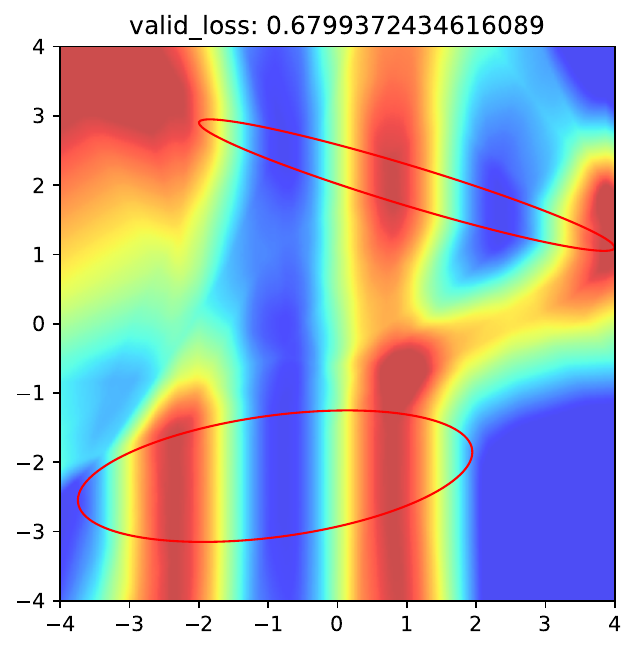}
		\label{CH:PAIR:fig:linextra_gau_erm_appdx}
	}
	\caption[Recovery of causal invariance via \pair.]{Recovery of causal invariance via \pair. (a), (c) We adopt two sampling methods where we sample the training data (mainly) from the regions marked in red, and evaluate the predictions across all regions from $(-4,-4)$ to $(4,4)$.
		The predictor following the invariance defined in \irm~\citep{irmv1} requires the predictions to be independent of spurious features within the overlapped invariant features. In this example, intuitively it requires the colored lines to be perpendicular to $x$-axis within $[-2,2]$. (b) and (d) show the performances of \erm under two sampling methods, it can be found that \erm fails to recover the causal invariance and incurs a high MSE loss.}
	\label{CH:PAIR:fig:extrapolation_sampling_appdx}
\end{figure}

Shown as in Fig.~\ref{CH:PAIR:fig:extrapolation_sampling_appdx}, we leverage two sampling methods: i) Uniform sampling and ii) Gaussian sampling, where the latter is more difficult than the former.
For Uniform sampling, we uniformly sample the rectangle regions $\{(-3,-3),(-2,1)\}$ as environment $1$ and $\{(-1,2),(3,3)\}$ as environment $2$, shown as the red regions marked in Fig.~\ref{CH:PAIR:fig:linextra_uniform_appdx}.
For Gaussian sampling, we sample from two Gaussian distributions: the first one has the center as $(-0.9,-2.2)$ with the covariance matrix as $\{(0.9,0.11),(0.11,0.1)\}$; the second one has the center as $(1,2)$ with the covariance matrix as $\{(1,-0.3),(-0.3,0.1)\}$, shown as the red regions marked in Fig.~\ref{CH:PAIR:fig:linextra_gau_appdx}.

Therefore, in these two examples, the invariant representation $\varphi$ should only take $X_1$ and discard the spurious features $X_2$ under the overlapped invariant features, i.e., $[-2,2]$.
As we use different colors to denote, the prediction produced by the invariant predictor following Definition~\ref{def:causal_inv_appdx} is expected be independent of $X_2$. In other words, the plotted lines need to be \emph{perpendicular} to the $x$-axis within the overlapped invariant features $[-2,2]$.

We implement the predictor with a $3$-layer linear perceptron that has a hidden dimension of $128$. We use the MSE loss and Adam~\citep{adam} to optimize the neural network. We sample $2500$ training data points from each environment and evaluate with $1000$ data points uniformly sampled across all regions. For a fair comparison, we train all algorithms $10000$ epochs until converge. Following the common practice~\citep{domainbed}, we use anneal iterations of the OOD penalties for all methods as $150$.
For \irml, \vrex, and \irmx, we search the penalty weights from $1e-4$ to $1e$ and find they generically perform well when with the penalty weights of $1e-2$ to $1e1$. While for $\pair$, we search the relative preferences across $6$ choices $(1,1e4,1e16),(1,1e4,1e12),(1,1e6,1e8),(1,1e8,1e4),(1,1e4,1e4),(1,1e8,1e8)$, and find \\$(1,1e4,1e12)$,$(1,1e8,1e4)$,$(1,1e4,1e4)$,$(1,1e8,1e8)$ have lower validation losses.

\begin{figure}[ht]
	\vskip -0.2in
	\subfigure[Uniform.]{
		\includegraphics[width=0.22\textwidth]{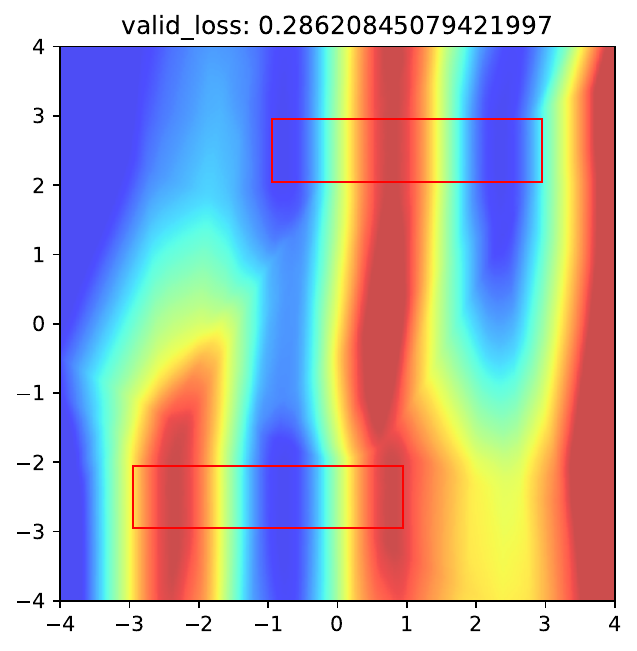}
		\label{CH:PAIR:fig:linextra_uniform_irm_appdx}
	}
	\subfigure[Uniform.]{
		\includegraphics[width=0.22\textwidth]{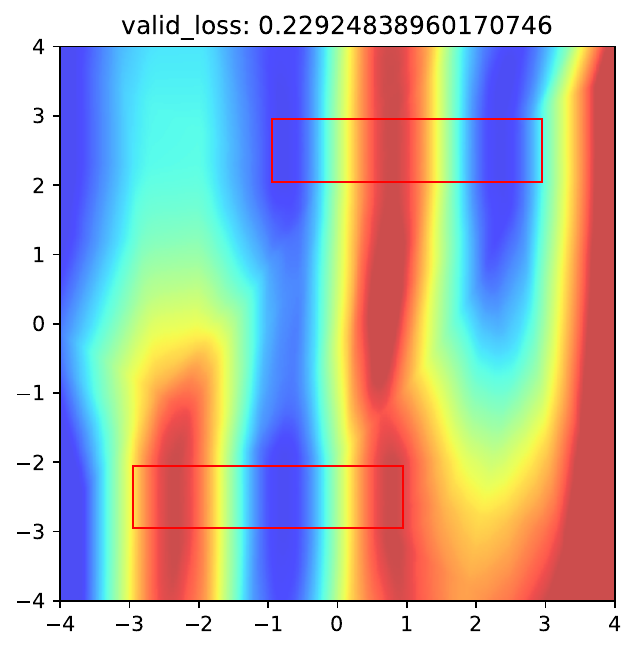}
		\label{CH:PAIR:fig:linextra_uniform_irm2_appdx}
	}
	\subfigure[Gaussian.]{
		\includegraphics[width=0.22\textwidth]{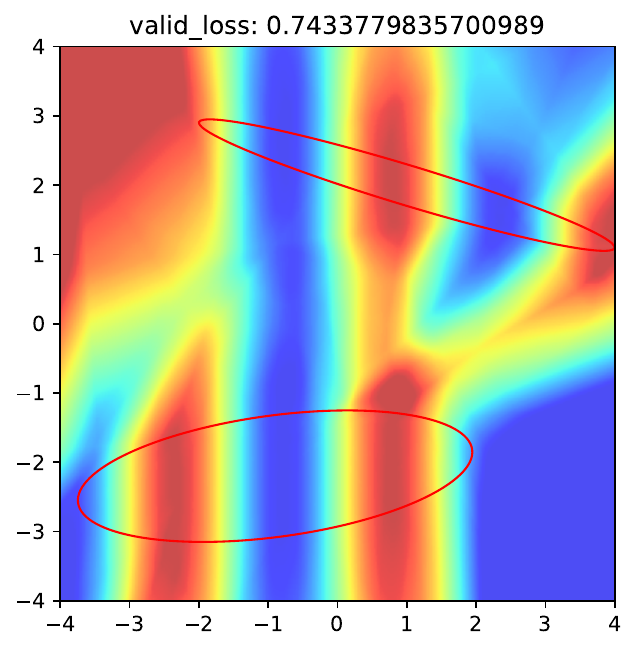}
		\label{CH:PAIR:fig:linextra_gau_irm_appdx}
	}
	\subfigure[Gaussian.]{
		\includegraphics[width=0.22\textwidth]{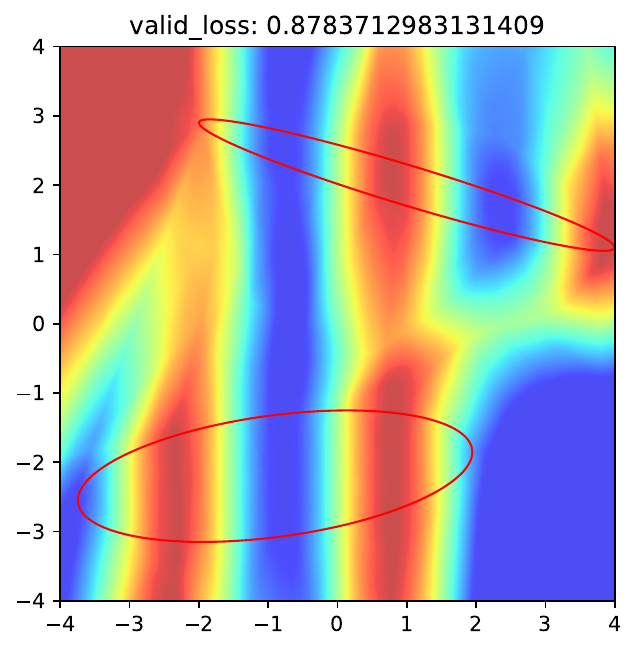}
		\label{CH:PAIR:fig:linextra_gau_irm2_appdx}
	}
	\caption{Recovery of causal invariance via \irml.}
	\label{CH:PAIR:fig:linextra_irm_appdx}
	\vskip -0.2in
\end{figure}

\begin{figure}[ht]
	\vskip -0.2in
	\subfigure[Uniform.]{
		\includegraphics[width=0.22\textwidth]{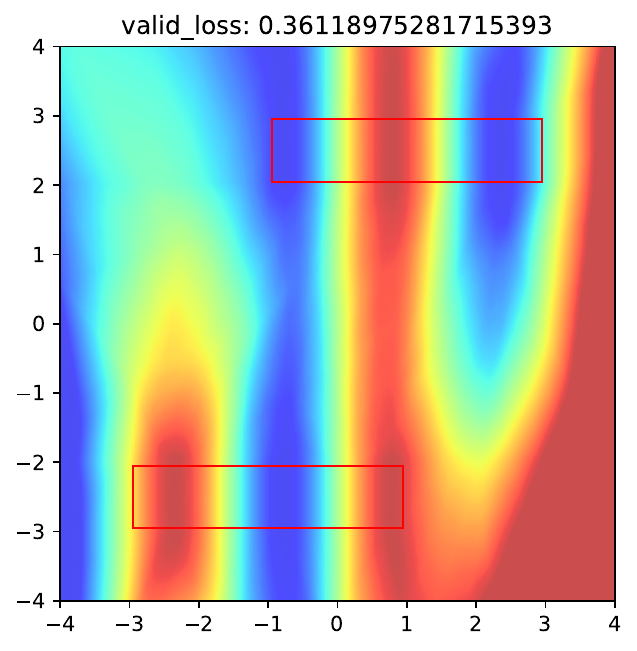}
		\label{CH:PAIR:fig:linextra_uniform_vrex_appdx}
	}
	\subfigure[Uniform.]{
		\includegraphics[width=0.22\textwidth]{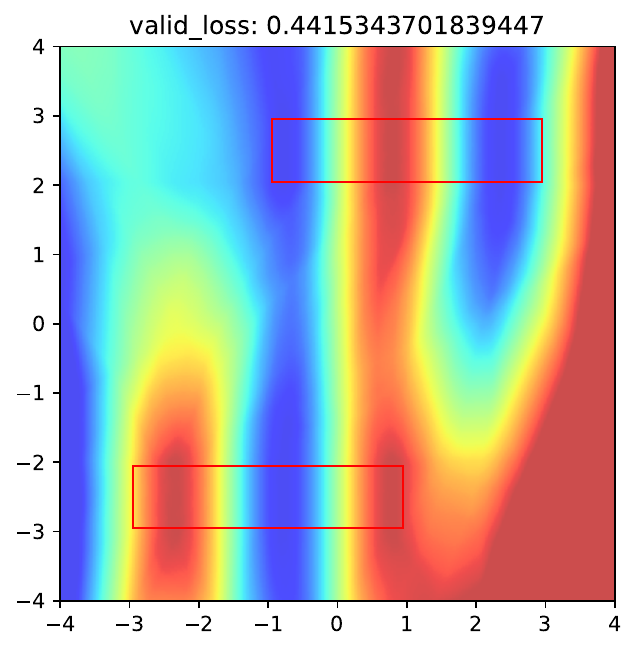}
		\label{CH:PAIR:fig:linextra_uniform_vrex2_appdx}
	}
	\subfigure[Gaussian.]{
		\includegraphics[width=0.22\textwidth]{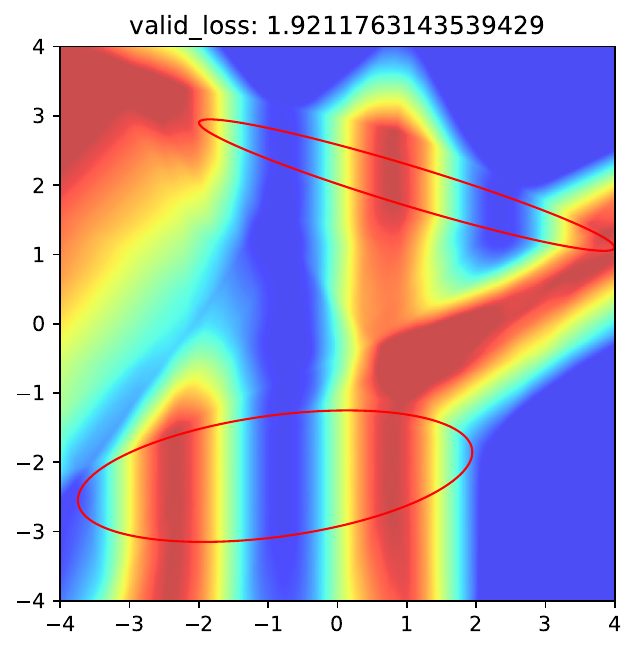}
		\label{CH:PAIR:fig:linextra_gau_vrex_appdx}
	}
	\subfigure[Gaussian.]{
		\includegraphics[width=0.22\textwidth]{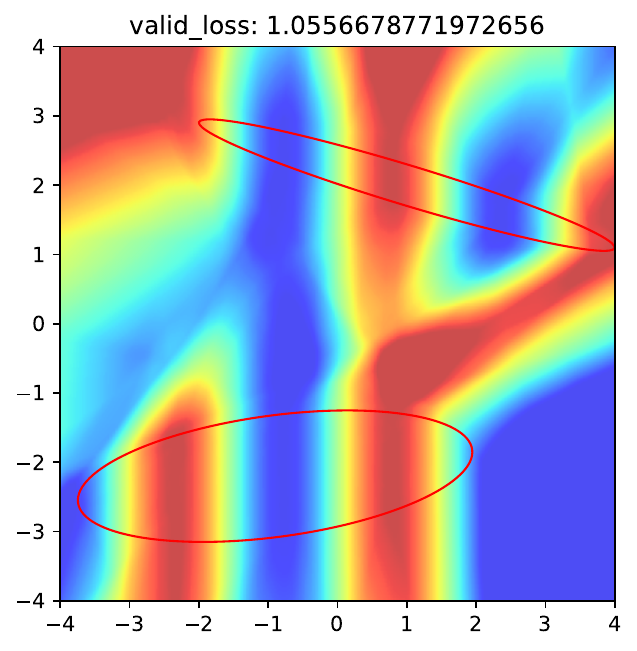}
		\label{CH:PAIR:fig:linextra_gau_vrex2_appdx}
	}
	\caption{Recovery of causal invariance via \vrex.}
	\label{CH:PAIR:fig:linextra_vrex_appdx}
\end{figure}

\begin{figure}[ht]
	\subfigure[Uniform.]{
		\includegraphics[width=0.22\textwidth]{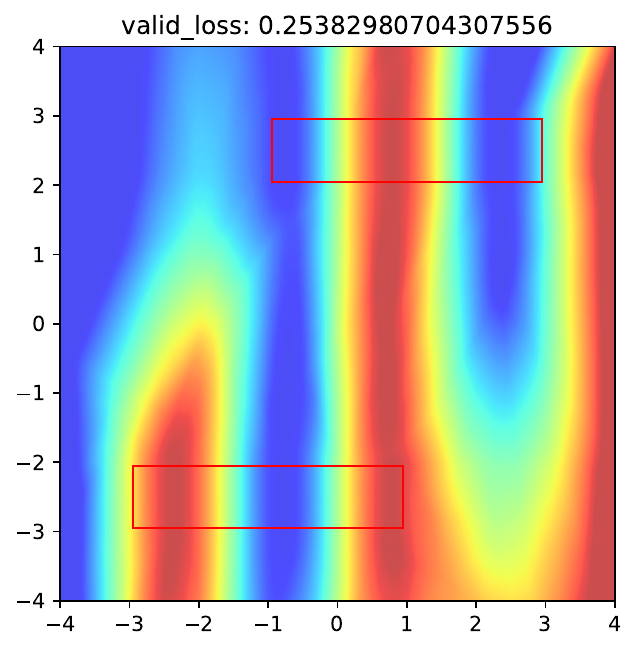}
		\label{CH:PAIR:fig:linextra_uniform_irmx_appdx}
	}
	\subfigure[Uniform.]{
		\includegraphics[width=0.22\textwidth]{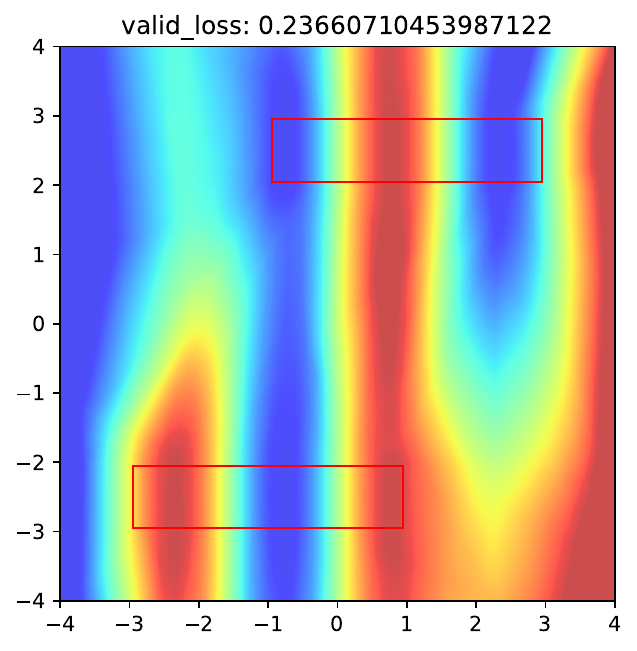}
		\label{CH:PAIR:fig:linextra_uniform_irmx2_appdx}
	}
	\subfigure[Gaussian.]{
		\includegraphics[width=0.22\textwidth]{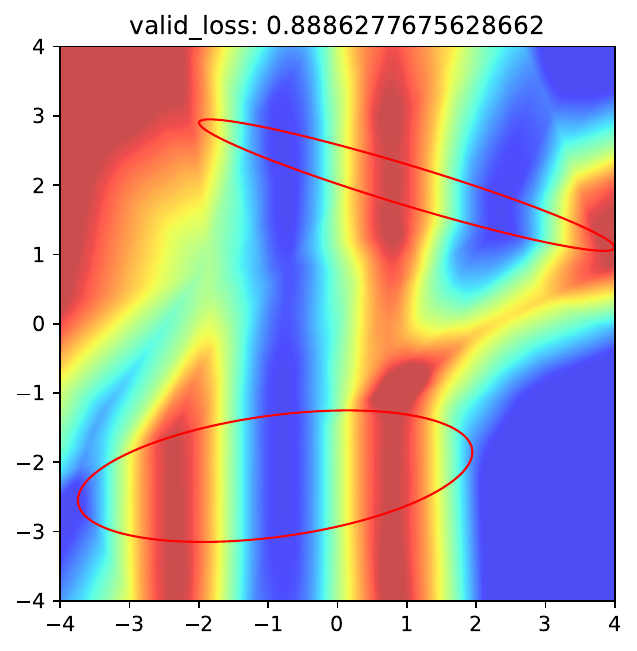}
		\label{CH:PAIR:fig:linextra_gau_irmx_appdx}
	}
	\subfigure[Gaussian.]{
		\includegraphics[width=0.22\textwidth]{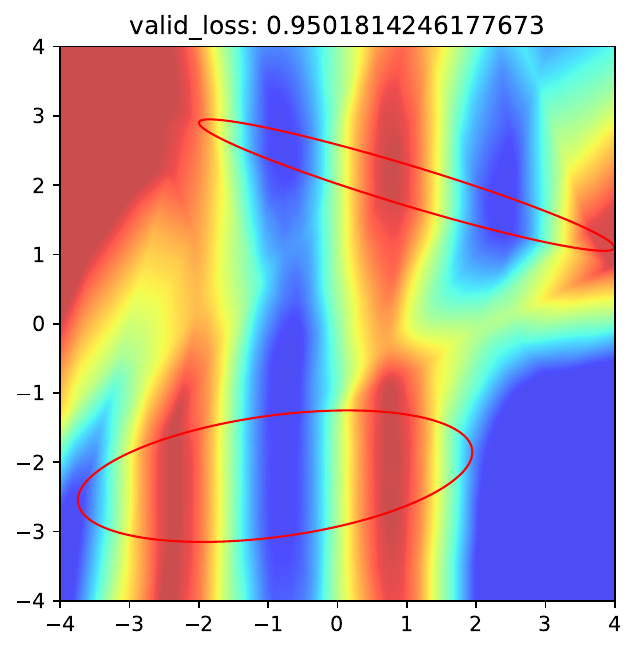}
		\label{CH:PAIR:fig:linextra_gau_irmx2_appdx}
	}
	\caption{Recovery of causal invariance via \irmx.}
	\label{CH:PAIR:fig:linextra_irmx_appdx}
\end{figure}

\begin{figure}[ht]
	\subfigure[Uniform.]{
		\includegraphics[width=0.22\textwidth]{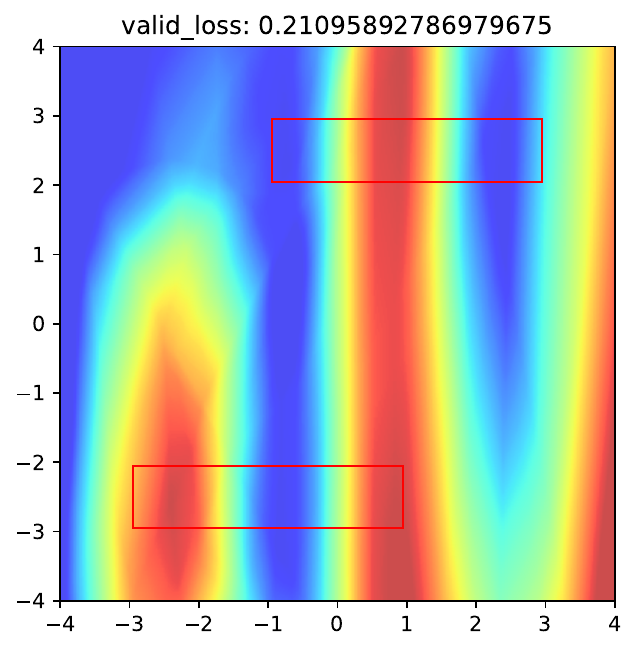}
		\label{CH:PAIR:fig:linextra_uniform_pair_appdx}
	}
	\subfigure[Uniform.]{
		\includegraphics[width=0.22\textwidth]{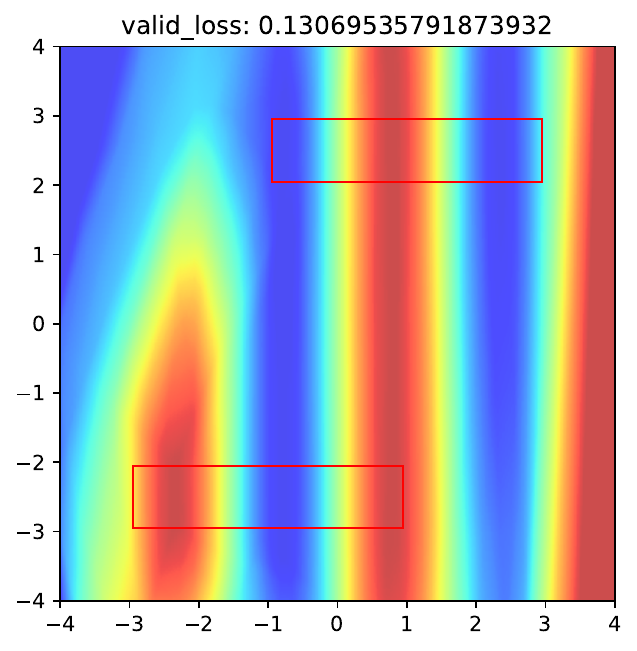}
		\label{CH:PAIR:fig:linextra_uniform_pair2_appdx}
	}
	\subfigure[Gaussian.]{
		\includegraphics[width=0.22\textwidth]{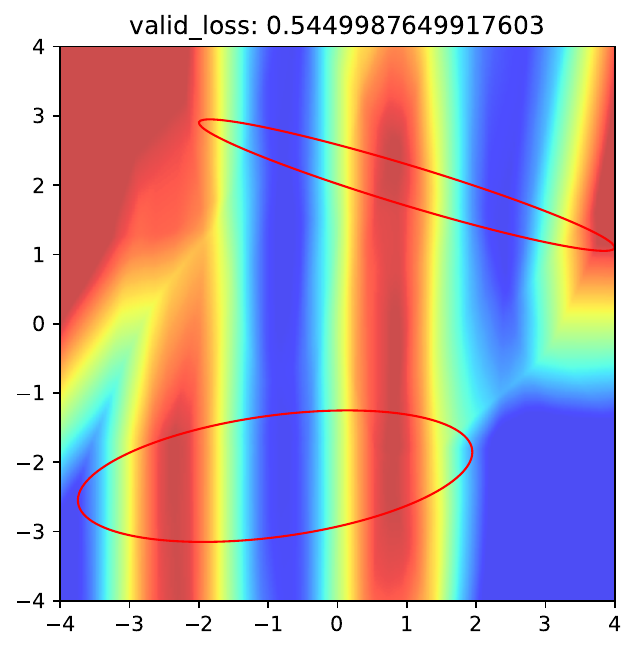}
		\label{CH:PAIR:fig:linextra_gau_pair_appdx}
	}
	\subfigure[Gaussian.]{
		\includegraphics[width=0.22\textwidth]{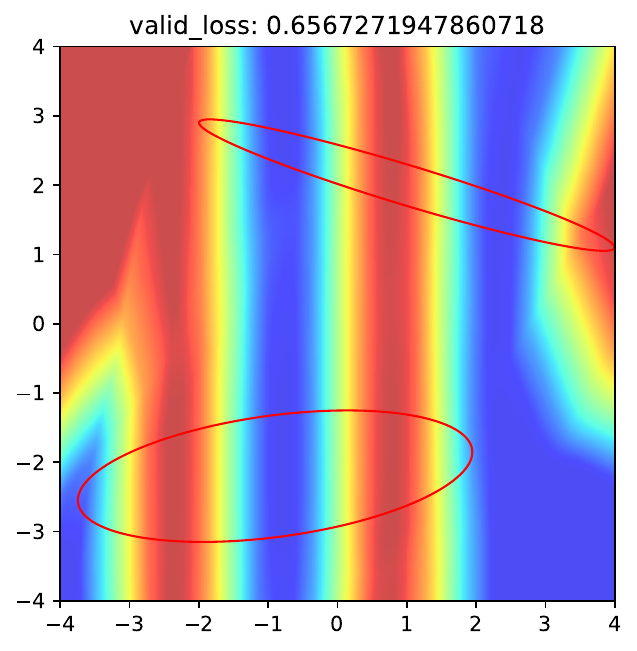}
		\label{CH:PAIR:fig:linextra_gau_pair2_appdx}
	}
	\caption{Recovery of causal invariance via \pair.}
	\label{CH:PAIR:fig:linextra_pair_appdx}
\end{figure}
We plot predictions with the best MSE losses of \irml, \vrex, \irmx and \pair in Fig.~\ref{CH:PAIR:fig:linextra_irm_appdx}, Fig.~\ref{CH:PAIR:fig:linextra_vrex_appdx}, Fig.~\ref{CH:PAIR:fig:linextra_irmx_appdx}, and Fig.~\ref{CH:PAIR:fig:linextra_pair_appdx} respectively.
We also plot the validation loss at the top of the image while \emph{it does not necessarily indicate a better recovery of causal invariance}.
It can be found that, when given the uniform sampled environments, the unrobust \irml, \vrex and \irmx can recover part of the causal invariance, while when switching to the Gaussian sampled environments, they can fail dramatically as expected.
In contrast, for both uniform sampling and Gaussian sampling, \pair manage to recover the causal invariance almost perfectly. Perhaps even more surprisingly, \pair achieve a lower extrapolation loss up to $0.06$ and $0.32$, which are essentially beyond the extrapolation requirement issued by the causal invariance. Hence we believe it is an interesting and promising future direction to probe the extrapolation ability within and beyond causal invariance.

\section{More Details on the Implementations of \pair}
\label{CH:PAIR:sec:pair_implementation_appdx}
In this section, we provide more details about the implementation of \pair as a optimizer and a model selection criteria, in complementary to Sec.~\ref{CH:PAIR:sec:pair_method}.

\textbf{Key takeaways from the \irm example.}
Recall that the key takeaways from the failures of OOD optimization can be attributed to:
i) using unrobust objectives for optimization; ii) using unreliable scheme to approach the desired solution.
Nevertheless, we can improve the robustness of the OOD objectives by introducing  additional guidance such that the desired solution can be relocated in the Pareto front w.r.t. to the new objectives.
After obtaining robust objectives to optimize, we then leverage a preference-aware MOO solver to find the Pareto optimal solutions that maximally satisfy the invariance constraints by assigning the OOD objective a higher preference while being aware of retaining ERM performance.

More formally, let $f_\ood$ be the desired OOD solution, a group of OOD objectives $\vL_\ood=\{\gL_\ood^i\}_{i=1}^m$ are robust if they satisfy that
\begin{equation}
	\vL_\ood(f_\ood) \preceq \vL_\ood (f), \forall f\neq f_\ood\in\gF,
\end{equation}
where $\gF$ denotes the functional class of possible predictors. When given a robust OOD objective $\vL_\ood$, our target is to solve the following MOO problem
\begin{equation}
	\label{CH:PAIR:eq:pair_moo_appdx}
	\text{$\min$}_f (\gL_\erm,\vL_\ood)^T,
\end{equation}
where $\vL_\ood$ corresponds to a $\bm{\epsilon}_\ood$-relaxed invariance constraint as $\vL_\ood(f_\ood)=\bm{\epsilon}_\ood\preceq\vL_\ood(f),\forall f\neq f_\ood\in\gF$.
Denote the $\epsilon_\inv$ as empirical loss of using the underlying invariant features to predict labels, then the optimal values of the desired OOD solution are $(\epsilon_\inv,\bm{\epsilon}_\ood)^T=(\gL_\erm(f_\ood),\vL_\ood(f_\ood))^T$, which corresponds to an ideal OOD preference for the objectives that is $\vp_\ood=(\frac{1}{\epsilon}_\inv,\bm{\frac{1}{\epsilon}}_\ood)^T$.
Then the solution of Eq.~\ref{CH:PAIR:eq:pair_moo} needs to maximally satisfy the OOD preference, i.e., maximize $\vL(f)^T\vp_\ood$.

\subsection{Detailed description of \pairo for OOD optimization}
\label{CH:PAIR:sec:pair_optimizer_appdx}

To find a Pareto optimal solution that satisfies the OOD preference $\vp_\ood$, we leverage the preference-aware MOO solver~\citep{epo}.
Different from~\citet{epo}, we adopt an explicit 2-stage ``descent'' and ``balance'' scheme, following the common practice in OOD generalization~\citep{domainbed}.

\begin{wrapfigure}{r}{0.33\textwidth}
	\vspace{-0.1in}
	\includegraphics[width=0.33\textwidth]{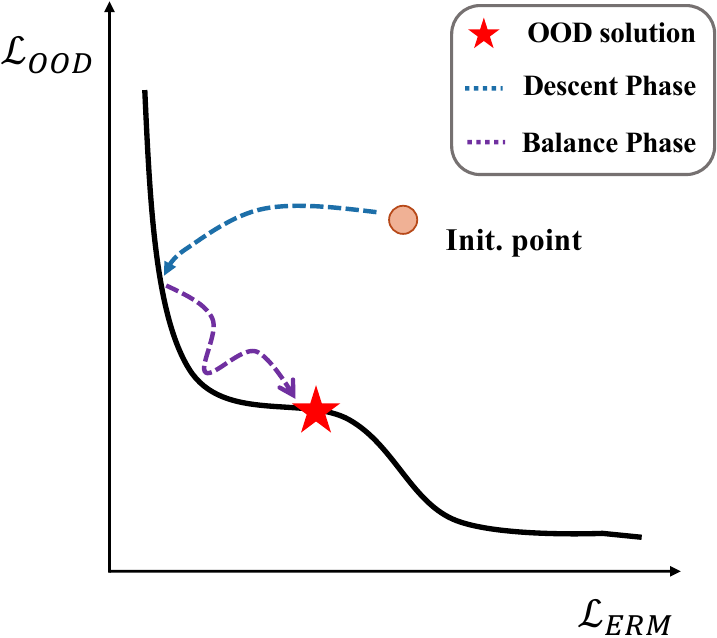}
	\vskip -0.1in
	\caption{Illustration of \pairo.}
	\label{CH:PAIR:fig:pair_opt_appdx}
	\vspace{-0.3in}
\end{wrapfigure}
Illustrated as in Fig.~\ref{CH:PAIR:fig:pair_opt_appdx}, in the ``descent'' phase, we train the model to minimize the ERM loss such that it approaches the Pareto front by merely minimizing $\gL_\erm$ first. Then, in the ``balance'' phase, we adjust the solution to maximally satisfy the OOD preference $\vp_\ood$.

Meanwhile, to avoid divergence from the Pareto front, at each  step, the descent direction $\vg_\des$ not only needs to maximize $\vL(f)^T\vp_\ood$, but also needs to avoid ascending all the loss values. More formally, let $\vG$ denote the gradient signals produced by $\vL$, at step $t$ of the  ``balance'' phase, it solves the following LP for the objective weights $\beta^*$,
\begin{equation}\label{CH:PAIR:eq:pair_lp_appdx}
	\begin{aligned}
		\beta^* = \text{$\arg\max$}_{\beta\in\gS^{m+1}} & \ (\vG\beta)^T\vg_b,                                             \\
		\ \text{s.t.}                                   & \ (\vG\beta)^T\vG_j \geq \vg_b^T\vG_j,\ \forall j\in\bar{J}-J^*, \\
		                                                & \ (\vG\beta)^T\vG_j \geq 0,\ \forall j\in J^*,
	\end{aligned}
\end{equation}
where $\gS^{m+1}=\{\beta\in\R^{m+1}_+|\sum_{i=1}^{m+1}\beta_i=1\}$, $\vg_b$ is the adjustment direction that leads to the preferred Pareto optimal solution by $\vp_\ood$, $J=\{j|G_j^T\vg_b>0\}$ are the indices of objectives which donot conflict with $\vg_b$ while $\bar{J}=\{j|G_j^T\vg_b\leq 0\}$ are those have conflicts with $\vg_b$, $J^*=\{j|L_j\text{$\vp_\ood$}_j=\max_{j'}(L_{j'}\text{$\vp_\ood$}_{j'})\}$ is the index of the objective which diverges from the preference most.

Specifically, \citet{epo} show that using the following $\vg_b$ could provably lead the solution converge to the desired preferred Pareto optimal solution, which is defined as follows
\begin{equation}\label{CH:PAIR:eq:pair_bal_appdx}
	\vg_b = \vp\odot(\log((m+1)\hat{\vL})-\mu(\vL)),
\end{equation}
where $\odot$ is the element-wise product operator, $\mu(\vL)$ is the quantitative divergence of the current solution from the preferred direction, calculated through the losses at the current step, as follows
\begin{equation}\label{CH:PAIR:eq:pair_mu_appdx}
	\mu(\vL)=\text{KL}(\hat{\vL}|\mathbf{1}/m)=\sum_{i}^{m+1}\hat{\vL}_i\log(m\hat{\vL}_i),
\end{equation}
where $\hat{\vL}$ is the normalized loss as
\[\hat{\vL}_i=\text{$\vp_\ood$}_i\vL_i/\sum_{j}^{m+1}\vp_j\vL_j.\]

Then, we elaborate the detailed algorithm of \pairo implemented via the EPO solver~\citep{epo} as in Algorithm~\ref{alg:pair_opt_appdx}.

\begin{algorithm}[tb]
	\caption{Pseudo code for \pairo.}
	\label{alg:pair_opt_appdx}
	\begin{algorithmic}[1]
		\STATE {\bfseries Input:} Training data $\train=\{X_i,Y_i\}_{i=1}^N$ with environment partitions $\train=\{\dataset^e\}_{e\in\envtrain}$; learning rate $\eta$; batch size $b$; number of sampled environments $d$; OOD preference $\vp_\ood$ for ERM loss $\gL_\erm$ and $m$ OOD losses $\vL_\ood=\{\gL_\ood^i\}_{i=1}^m$; pre-training epochs $e_p$; maximum training epochs for ``balance'' phase $e_b$; Trainable parameters at the ``balance'' phase $\theta$;
		\STATE Randomly initialize parameters in the model $f=w\circ\varphi$;
		\FOR{$i=1$ {\bfseries to} $e_p$}
		\STATE Sample batches of data $\{X_j,Y_j\}_{j=1}^b$;
		\STATE Make predictions with $f$: $\{\widehat{Y}_j\}_{j=1}^b=f(\{X_j\}_{j=1}^b)$;
		\STATE Calculate the empirical loss $L_\erm$ with $\{\widehat{Y}_j\}_{j=1}^b$;
		\STATE Update parameters of $f$ with the empirical loss $\gL_\erm$ using the learning rate $\eta$;
		\ENDFOR
		\FOR{$i=1$ {\bfseries to} $e_b$}
		\FOR{$D^e\in\texttt{permute}(\{\dataset^e\}_e\in\envtrain)$}
		\STATE Sample a batch of the data from $D^e$, $\{X_j^e,Y_j^e\}_{j=1}^b\sim D^e$;
		\STATE Make predictions with $f$: $\{\widehat{Y}_j^e\}_{j=1}^b=f(\{X_j^e\}_{j=1}^b)$;
		\ENDFOR
		\STATE Calculate empirical and OOD losses $\gL_\erm$ and $\gL_\ood$ and obtain the overall losses $\vL$;
		\STATE Obtain gradients $\vG=\partial\vL/\partial \theta$;
		\STATE Calculate the OOD divergence $\mu(\vL)$ using Eq.~\ref{CH:PAIR:eq:pair_mu_appdx};
		\STATE Obtain the adjustment direction $\vg_b$ using Eq.~\ref{CH:PAIR:eq:pair_bal_appdx};
		\STATE Obtain the index sets $J,J^*,\bar{J}$ required by Eq.~\ref{CH:PAIR:eq:pair_lp_appdx};
		\STATE Solve Eq.~\ref{CH:PAIR:eq:pair_mu_appdx} for the loss weights $\beta^*$;
		\STATE Update parameters $\theta^{i+1}= \theta^{i}-\eta\vG\beta^*$;
		\ENDFOR
	\end{algorithmic}
\end{algorithm}

We now state a informal version of the convergence guarantee.
\begin{theorem}(Informal)\label{CH:PAIR:thm:pair_converge_appdx}
	Given $L_\erm$ along with $m$ differentiable OOD losses $\vL_\ood$, at each step in the ``balance'' phase (line $9$ to line $21$ in Algorithm~\ref{alg:pair_opt_appdx}), there exists a step size $\eta_0$ such that,
	the set of new loss values $\vL^{(i+1)}=(L_\erm,L_i,...,L_m)^T$ with the updated parameters $\theta^{(t+1)}$ by any $\eta\in[0,\eta_0]$, denoted as $\gA^t$
	has the following properties:
	\begin{enumerate}[label=(\roman*).,wide]
		\item $\gA^t$ contains the exact Pareto optimal solution satisfying the OOD preference vector, i.e., $\vL^*\in\gA^t$;
		\item $\gA^t$ grows monotonically smaller and smaller.
	\end{enumerate}
\end{theorem}
From (i) and (ii) in Theorem~\ref{CH:PAIR:thm:pair_converge_appdx}, it suffices to know that as the optimization continues, $\gA^t$ converges to the losses of the exact Pareto optimal solution, hence for the parameters.
The proof for Theorem~\ref{CH:PAIR:thm:pair_converge_appdx} simply follows the Theorem 1 to Corollary 1 in~\citet{epo}.
Note that \pairo provides a general framework to find a better OOD solution that properly trades off ERM and OOD objectives.
In experiments, we find that using the simply modified variant of EPO solver~\citep{epo} in \pairo can effectively find a descent path under the gradient conflicts that leads to a better OOD solution.
Nevertheless, a more sophisticated preference-aware MOO solver can be developed and integrated into the framework of \pairo, which we believe is a promising future direction~\citep{importance_sampling,pmlr-v80-zhou18c,robust_momentum}.

\subsection{Detailed description of \pairs for OOD model selection}
\label{CH:PAIR:sec:pair_selection_appdx}

In this section, we provide a detailed description of \pairs for OOD model selection for Sec.~\ref{CH:PAIR:sec:pair_method}.
Before start, we also provide a detailed description of the critical reasons for designing \pairs in Appendix~\ref{CH:PAIR:sec:dobed_intro_appdx}.
From the \irm example, it is obvious that traditional model selection methods that merely use validation performance, i.e., ERM performance, are not suitable to select a desired solution for OOD generalization. Otherwise, the OOD performance would be easily compromised due to its conflicts with ERM objective. This issue is more serious when the validation set has a large gap between the test set (cf. Training-domain validation set selection for \cmnist in Table~\ref{CH:PAIR:tab:dobed_select}).
Intuitively, models selected merely based on ERM performance tend to have a high preference or better performance on environments that have a similar distribution of the corresponding validation set, which will lead to higher variance of performances at different environments or a lower worst environment performance.
Therefore, it is natural to jointly consider the ERM and OOD performances in model selection. Specifically, the selected model is expected to maximally satisfy the exact Pareto optimality.

Since our focus of \pairs is mainly to validate the existence of previous mode selection issues, we simply incorporate the \pair score as an additional model selection criteria. More specifically, given a OOD preference $\vp_\ood$, we can calculate the \pair selection score as
\begin{equation}\label{CH:PAIR:eq:pair_score_appdx}
	s_\pair = \vL^T\hat{\vp}_\ood,
\end{equation}
where $\hat{\vp}_\ood$ is the normalized OOD preference as $\vp_\ood/\sum_{i=1}^{m+1}\text{$\vp_\ood$}_i$.
With the \pair score, we then can apply it into the \dobed model selection algorithms~\citep{domainbed}. Specifically, the model selection in \dobed aims to select models from several rigorous hyperparameter trials according to the validation accuracy. For the model selection in each run, one can obtain all training domain validation accuracies but only one test domain validation accuracy for fairness.

The algorithm is detailed as in Algorithm~\ref{alg:pair_sel_appdx}. The \pair score is mainly used to select models among the logged steps within one run. To avoid trivial cases, we expect the models participated into the selection are converged. To this end, we heuristically use a threshold $c$ to filter out the first $c$ steps and find it empirically effective. To select models from different runs, we will first use the validation accuracy to filter out some unreliable cases, and then adopt the \pair to finalize the model selection. The only exception is the test domain validation accuracy, where the test domain validation accuracy is more likely to be a reliable indicator than the \pair score.

The main limitation of the \pair estimation is about the estimation of the loss values. In stochastic gradient descent, one could only obtain a stochastic estimate of loss values based on a minibatch sample of $\train$. When the stochastic estimates of the loss values are unbiased, the \pair is unbiased, too. However, there can exist certain variances in the stochastic estimates, which can severely affect the precision of the score thus the comparison of different models. Although Theorem~\ref{CH:PAIR:thm:pair_theory_appdx} establishes certain theoretical guarantees that allows for some degree of uncertainties, the variances are usually unavoidable. A instant fix for the issue is that one could afford some additional evaluation time to obtain a better estimate of the loss values. Besides, one could also jointly consider the uncertainty of the estimation and derive a more accurate model selection~\citep{clove}, which we leave for future work.

\begin{algorithm}[tb]
	\caption{Pseudo code for \pairs.}
	\label{alg:pair_sel_appdx}
	\begin{algorithmic}[1]
		\STATE {\bfseries Input:} Running history $\gH$ from $R$ runs, where each running history is consist of loss history $\gL=\{\gL_1^t,\gL_2^t,...,\gL_{(m+1)}^t\}_{t=1}^T$ of $(m+1)$ losses, i.e., $\gL_\erm$ and $\vL_\ood=\{\gL_\ood^i\}_{i=1}^m$, and training and validation accuracy history $\gA=\{A_{\text{tr}}^t,A_{\text{val}}^t\}_{t=1}^T$, from $T$ logging steps; OOD preference $\vp_\ood$; Convergence step $c$; Validation accuracy percentile $p$;
		\FOR{$r=1$ {\bfseries to} $R$}
		\STATE Calculate \pair score using $\vp_\ood$ for all $T$ steps as $\gS=\{s^t\}_{t=1}^T$ using Eq.~\ref{CH:PAIR:eq:pair_score_appdx};
		\STATE Filter out the first $c$ steps to avoid trivial cases and get $\widehat{\gS}=\{s^t\}_{t=c}^T$;
		\STATE Store the step with maximum \pair score as $s_*=\argmax_t \widehat{\gS}$;
		\ENDFOR
		\STATE Obtain the selected steps from $R$ runs as $\gS=\{s_*^r\}_{r=1}^R$;
		\STATE Obtain the validation accuracies for all selected steps $\gA_{\text{val}}=\{A_\text{val}^{s_*^r}\}_{r=1}^R$;
		\STATE Calculate the validation selection bar as $\bar{A}_\text{val}=(\max\gA_{\text{val}}-\min\gA_{\text{val}})*p+\min\gA_{\text{val}}$;
		\STATE Filter out all runs that have a validation accuracy lower than $\bar{A}_\text{val}$ and obtain $\bar{\gH}$;
		\STATE Find the run with highest \pair score as $r_*=\argmax_{r\in\bar{\gH}}s_*^r$;
		\STATE Return associated history of $r_*$;
	\end{algorithmic}
\end{algorithm}

\subsection{Discussion on the practical choices of OOD preference}
\label{CH:PAIR:sec:pair_discussion_pc_appdx}
Essentially, the performances of both \pairo and \pairs have certain dependence on the quality of the OOD preference $\vp_\ood$, however, it is often the case that the ideal OOD preference is usually unknown.
It is desirable to analyze the performances of \pairo and \pairs under a imprecise OOD preference. \citet{epo} discussed a bit that when the exact Pareto optimal solution under the preference does not exist, the EPO solver can still find a Pareto optimal solution that is closest to the preferred direction.
We discuss it in a more general way by developing a new MOO formulation of Eq.~\ref{CH:PAIR:eq:pair_moo_appdx} under a approximated preference up to some approximation error of $\epsilon$.
The theoretical discussion can be found in Sec.~\ref{proof:pair_theory_appdx}. In this section, we focus on the practical side of the choice of $\vp_\ood$.

We first discuss some heuristics that can be leveraged to obtain a proper OOD preference under two scenarios:
\begin{enumerate}[label=(\roman*).,wide]
	\item one has little-to-no knowledge about the OOD loss values;
	\item one has the access to some running histories that one has some empirical knowledge about the OOD loss values;
\end{enumerate}
In practice, i) mostly fits to \pairo while ii) mostly fits to \pairs.

When \textbf{i)} one has little-to-no knowledge about the OOD loss values, one can leverage certain theoretical inductive biases about the OOD losses. In fact, it is usual the case that the theoretical conditions for the optimality of OOD objectives do not hold in practice~\citep{DANN,groupdro,vrex,fish,fishr}. In this case, minimizing the OOD losses acts more like a necessary condition for a satisfactory OOD solution.
Therefore, one could assign a sufficiently larger preference to OOD objectives than ERM objective. For example, throughout all experiments in the paper, we mostly assign $(1,1e10,1e12)$ to \erm, \irml, and \vrex losses, which works under many scenarios.

Besides, among different OOD objectives, one could easily know which is more likely to be optimized than another. Therefore, to ensure all OOD losses are equally maximally optimized, we could assign the easily-optimizable OOD objectives higher preference. For example in \irmx, \vrex tends to be easier to optimize than \irml therefore we assign a higher preference to \vrex. Moreover, if one could know the performances of different OOD objectives, it is natural to assign a higher preference to those which solely perform better.

When \textbf{ii)} one has the access to some running histories that one has some empirical knowledge about the OOD loss values, one could obtain a empirical estimate of the OOD loss values w.r.t. ERM loss values at convergence. Since the estimate is obtained under gradient conflicts, one could expect the ratios of OOD loss w.r.t. ERM loss should be higher when one could resolve the gradient conflicts properly. Therefore, one could assign a slightly higher preference to OOD losses than the empirically estimated ratios. In the model selection experiments, we directly increase the ratio by $1e2$ and find it works well as expected.

In fact, both \textbf{i)} and \textbf{ii)} are discussed under minimal assumption about the external knowledge of the optimization process, the task and the data. We expect a better estimate of the OOD preference could be obtained when more external inductive biases are incorporated. For instance, \pairo generalize to ParetoDA~\citep{pareto_da} when one could obtain a validation set that has similar distribution to the test data. Even under the case that such data is not available, one could also adopt some techniques such as Mixup~\citep{mixup} to obtain an approximation. We believe that obtaining a better estimate of the ideal OOD preference would be a promising future development based on our work.

\subsection{Discussion on the use of \pair in practice}
\label{CH:PAIR:sec:pair_discussion_appdx}

\subsubsection{Scalability}
\label{CH:PAIR:sec:pair_opt_sca_appdx}
Similar to other MOO algorithms~\citep{mtl_moo,pareto_mtl,epo}, \pairo requires full gradients of the predictor to make an accurate derivation of the objective weights $\beta^*$, which could be a bottleneck when deployed to large-scale networks, as it usually involves a prohibitively massive number of parameters. \citet{mtl_moo} develops an approximation of the full gradients using the gradients w.r.t. the latent representation produced by the featurizer, i.e., $\partial \vL/\partial \varphi(X)$. However, it requires a strong assumption on the structure of the data and the model.
Moreover, when it involves complex network architectures such as DenseNet~\citep{densenet} or DistillBERT~\citep{distillbert} in \wilds, the approximation or even the full gradients can be even imprecise, as the gradients of the complex neural networks can not be directly concatenated as those of simple linear networks.

To this end, we develop another approximation that takes only the gradients of the classifier, which usually appears as a linear classification layer in the predictor. Interestingly, we empirically find $\partial \vL/\partial w$ can even produce more useful signals for OOD generalization than the gradients w.r.t. classifier, shown as in Table~\ref{CH:PAIR:tab:cmnist}.

When considering a more resource restricted scenarios, such as the iWildCam and RxRx1 in \wilds, we freeze the featurizer after the ``descent'' phase, which can further resolve the memory and computation overheads. It also aligns with some recent discoveries that the featurizer trained merely with ERM may already discovery all useful patterns~\citep{dare}. \citet{rfc} also find the technique useful in Camelyon17 dataset of \wilds.

\subsubsection{Loss value estimation}
\label{CH:PAIR:sec:pair_opt_est_appdx}
Similar to other MOO algorithms~\citep{mtl_moo,pareto_mtl,epo}, \pairo is described and analyzed in full batch setting, i.e., full gradient descent. However, in practice, stochastic setting tends to appear more often than vanilla gradient descent due to the scalability considerations. As also discussed in Sec.~\ref{CH:PAIR:sec:pair_method}, variances are unavoidable no matter the estimated values are biased or unbiased. Fortunately, the robustness of \pairo to the preference can partially mitigate the issue.

The another potential limitation in \pairo could be the possibly negative estimate of some OOD losses, such as the stochastic estimates of \irml, since general MOO algorithms together with \pairo only accept non-negative loss values as the inputs. To this end, we will use \irml as an example to explain how one could handle the potentially negative values in loss value estimation.

We will first introduce the unbiased empirical estimator of \irml, following~\citet{irmv1,cs_irm}. More specifically, considering the \irml objective,
\begin{equation}
	\label{CH:PAIR:eq:irml_bias_appdx}
	\min_{\varphi}  \sum_{e\in\envtrain}\gL_e(\varphi)+\lambda|\nabla_{w|w=1}\gL_e(w\cdot\varphi)|^2.
\end{equation}

Observe that $$\nabla_{w|w=1.0} \gL_e(w\cdot\varphi) = \frac{\partial \mathbb{E}^{e}\big[\ell(w\cdot\varphi(X^{e}), Y^{e})\big]}{\partial w}\Big|_{w=1.0}= \mathbb{E}^{e}\bigg[\frac{\partial \ell(w\cdot\varphi(X^{e}), Y^{e})}{\partial w}\Big|_{w=1.0}\bigg]$$ and
\begin{equation}
	\begin{aligned}\label{CH:PAIR:eq:irml_bias_prod_appdx}
		\|\nabla_{w|w=1.0} \gL_e(w\cdot\varphi)\|^2 & =\Big(\frac{\partial \mathbb{E}^{e}\big[\ell(w\cdot\varphi(X^{e}), Y^{e})\big]}{\partial w}\Big|_{w=1.0}\Big)^2     \\
		                                            & = \Big(\mathbb{E}^{e}\bigg[\frac{\partial \ell(w\cdot\varphi(X^{e}), Y^{e})}{\partial w}\Big|_{w=1.0}\bigg]\Big)^2, \\
	\end{aligned}
\end{equation}
for which the simplification is derived by taking the derivative inside the expectation, using the Leibniz integral rule. Obviously, the stochastic estimate of Eq.~\ref{CH:PAIR:eq:irml_bias_prod_appdx} is biased.

To obtain an unbiased estimate of \irml penalty, observe that
\[
	\mathbb{E}[X]^2 = \mathbb{E}[AB],
\]
if $A$, $B$ and $X$ are i.i.d. random variables w.r.t. the same distribution $\gX$.
Equipped with this observation, we can further write Eq.~\ref{CH:PAIR:eq:irml_bias_prod_appdx} as
\begin{equation}\label{CH:PAIR:eq:irml_unbias_prod_appdx}
	\begin{aligned}
		\|\nabla_{w|w=1.0} \gL_e(w\cdot\varphi)\|^2 & =  \mathbb{E}^{e}\bigg[\Big(\frac{\partial \ell(w\cdot\varphi(X^{e}), Y^{e})}{\partial w}\Big|_{w=1.0}\Big)\Big(\frac{\partial \ell(w\cdot\varphi(\tilde{X}^{e}), \tilde{Y}^{e})}{\partial w}\Big|_{w=1.0}\Big)\bigg],             \\
		                                            & =\bigg[\mathbb{E}^{e}\Big(\frac{\partial \ell(w\cdot\varphi(X^{e}), Y^{e})}{\partial w}\Big|_{w=1.0}\Big)\mathbb{E}^{e}\Big(\frac{\partial \ell(w\cdot\varphi(\tilde{X}^{e}), \tilde{Y}^{e})}{\partial w}\Big|_{w=1.0}\Big)\bigg],
	\end{aligned}
\end{equation}
where $(X^e, Y^e)\sim \mathbb{P}^{e}$ and $(\tilde{X}^{e}, \tilde{Y}^{e})\sim \mathbb{P}^{e}$ are i.i.d. samples from $\mathbb{P}^e$ of the environment $e$.
As $\mathbb{E}^{e}\Big(\frac{\partial \ell(w\cdot\varphi(X^{e}), Y^{e})}{\partial w}\Big|_{w=1.0}\Big)$ and $\mathbb{E}^{e}\Big(\frac{\partial \ell(w\cdot\varphi(\tilde{X}^{e}), \tilde{Y}^{e})}{\partial w}\Big|_{w=1.0}\Big)$ can separately be estimated in minibatches without bias, Eq.~\ref{CH:PAIR:eq:irml_unbias_prod_appdx} essentially provides a practical unbiased estimator of \irml.

However, different from \irml, Eq.~\ref{CH:PAIR:eq:irml_unbias_prod_appdx} does not have any guarantees for its non-negativity, though the expectation of Eq.~\ref{CH:PAIR:eq:irml_unbias_prod_appdx} is non-negative. To this end, we propose two heuristics to mitigate the issue.

The first heuristic is to add all minibatch estimates $\mathbb{E}^{e}\Big(\frac{\partial \ell(w\cdot\varphi(X^{e}), Y^{e})}{\partial w}\Big|_{w=1.0}\Big)$ by a sufficiently large constant $C$, such that the minimum value of $\mathbb{E}^{e}\Big(\frac{\partial \ell(w\cdot\varphi(X^{e}), Y^{e})}{\partial w}\Big|_{w=1.0}\Big)+C$ is non-negative. Moreover, as the constant does not affect the calculation of the gradients, when \irml is minimized to $0$, $\mathbb{E}^{e}\Big(\frac{\partial \ell(w\cdot\varphi(X^{e}), Y^{e})}{\partial w}\Big|_{w=1.0}\Big)$ is also optimized to $C$.

The other heuristic is to multiply the negative minibatch estimates $\mathbb{E}^{e}\Big(\frac{\partial \ell(w\cdot\varphi(X^{e}), Y^{e})}{\partial w}\Big|_{w=1.0}\Big)$ by a proper negative constant $-C$, which will make all estimations non-negative. On the other hand, however, it can dramatically affect the variances in the estimations. Essentially, this multiplication will enlarge the expectation of the estimated \irml, and may cause instability of the training, due to the unrobustness of \irml. Therefore, we can heuristically search the values $C$ from $1$ to $1e-4$ by observing the early training dynamics. If the training is unstable, then we heuristically tune $C$ to be smaller by $1e-2$.

Although both of the heuristics above can not rigorously recover a non-negative estimate of \irml penalty (which is essentially impossible for the formulations like \irml), we empirically find them effective, for which we hypothesize is because of the robustness of \pairo to the preference in OOD generalization.

\subsubsection{Generalizing to other OOD methods}
As shown in Fig.~\ref{CH:PAIR:fig:grad_conflict}, the gradient conflicts between ERM and OOD objectives generally exist~\citep{irmv1,vrex,clove,sd,fishr}. It implies that, on the one hand, the optimization dilemma generally exist for all OOD objectives. Meanwhile, both \pairo and \pairs are generically applicable to all OOD methods. In experiments (Sec.~\ref{CH:PAIR:sec:experiments}), we validate the generality of \pairs only for several OOD methods from the four main lines as discussed in related works (Sec.~\ref{CH:PAIR:sec:related_work_appdx}) though, \pairo essentially has similar generality as \pairs, for whose performances at real world datasets, we will leave for future verification due to the limited computational resources.
Nevertheless, we can theoretically discuss the implementation options about how \pairo can be applied to different OOD methods.

First, for Domain Generalization based methods~\citep{DANN,CORAL,deep_DG,DouCKG19}, such as DANN~\citep{DANN}, \pairo can directly take the domain classification loss and the label classification loss as the inputs.

Second, for Distributionally Robust Optimization methods~\citep{dro,DRSL,groupdro}, \pairo can take the worst group loss or some more sophisticated regularizations and the \erm loss as the inputs.

Third, for the causal invariance based methods~\citep{inv_principle,causal_transfer,irmv1,env_inference,andmask,clove,ib-irm,ciga} and agreement based methods~\citep{iga,vrex,fish,fishr}, they can be handled by \pairo similarly as \irmx.

\section{Theoretical Discussions}
\label{CH:PAIR:sec:theory_appdx}

\subsection{Proof for Proposition~\ref{CH:PAIR:thm:recover_IRM_paper}}
\label{proof:recover_IRM}
We first restate the proposition with formally defined Setting A by~\citet{irm_aistats}.

\paragraph{Setting A (identical to~\citet{irm_aistats}):} Considering the task of linear classification/regression $\gX\rightarrow\gY$ where the quality of predictors $f:\gX\rightarrow\widehat{\gY}$ is measured by population losses $l:\widehat{\gY}\times\gY\rightarrow\R_{\geq0}$, $\widehat{\mathcal{Y}} = \R, \mathcal{Y} \subseteq \R$, $\ell$ is either the square loss $\lsq(\hat{y}, y) \coloneqq \frac{1}{2}(\hat{y} - y)^2$, or the logistic loss $\llog(\hat{y}, y) \coloneqq \log{(1 + \exp{(-\hat{y}y)})}$ when $\mathcal{Y} =\{-1, 1\}$ (binary classification).

\begin{proposition}\label{CH:PAIR:thm:recovep_iRM_paper_appdx} Under Setting A (\citet{irm_aistats}), for all $\alpha\in (0,1)$,
	let $\mathcal{E} \coloneqq \{(\alpha, \beta_e): \beta_e \in (0,1)\}$ be any instance of the two-bit environment (Eq.~\ref{CH:PAIR:eq:twobit_env_appdx}),
	$\Ix$ denote the invariant predictors produced by Eq.~\ref{CH:PAIR:eq:irmx_moo},
	it holds that $\gI_{\gS\cap X}(\mathcal{E}) = \mathcal{I}(\mathcal{E})$.\footnote{Motivated readers might be interested in the necessities of keeping \irml in the objectives, for which we provide details in Appendix~\ref{CH:PAIR:sec:discuss_pair_objs_appdx}.}
\end{proposition}

Our proof is proceeded by discussing the set of invariant predictors elicited by an ideal V-REx~\citep{vrex} objective $\gI_X(\gE)$ (in a more general way), and then incorporating $\gI_X(\gE)$ into that elicited by \irms or \irml~\citep{irmv1} $\gI_\gS(\gE)$ for the two-bit failure case (Eq.~\ref{CH:PAIR:eq:twobit_env_appdx}).

We now first discuss the invariant predictors produced by the invariance constraints ideally elicited by V-REx. Recall that V-REx~\citep{vrex} aims to minimize the variances of ERM losses at different environments:
\[\gL_\vrex := \var(\{\gL_e\}_{e\in\envtrain}).\]
Therefore, when $\gL_\vrex$ is minimized, we have $\gL_{e_1}=\gL_{e_2},\ \forall e_1, e_2\in\envtrain$. Then, we can define the invariant predictors produced by V-REx, as the following.

\paragraph{\vrexz:} Define $\Ix(\mathcal{E}) \coloneqq \{f:\mathcal{X} \rightarrow \hat{\mathcal{Y}}\mid \mathcal{L}_{e_1}(f) = \mathcal{L}_{e_2}(f), \forall e_1, e_2 \in \mathcal{E}\}$. \vrexz{} is the objective: \[\min_{f\in \Ix(\mathcal{E}_{\textup{tr}})}{\sum_{e\in \mathcal{E}_{\textup{tr}}} {\mathcal{L}_e(f)}}.\]

Then, we characterize the set of $\gI_X$ through the following lemma.

\begin{lemma} \label{vrex0}
	Under Setting A, let $f = w\circ\varphi$ be the predictor elicited by $\mathcal{I}(\mathcal{E})$ and $(X_e, Y_e) \sim \mathcal{D}_e$.
	\[
		\text{If }\begin{cases}
			\ell = \lsq,\, \Ede[Y_e^2] \text{ is identical}, \text{the distribution of } \varphi(X_e) \text{ is identical (or }f\equiv0\text{)} \\
			\ell = \llog\text{ and }H(Y_e|\varphi(X_e)) \text{ is identical}
		\end{cases}\text{for all }e\in \mathcal{E},
	\]
	then $\mathcal{I}(\mathcal{E}) \subseteq \Ix(\mathcal{E})$.
\end{lemma}
\begin{proof}
	For any $f = w \circ\varphi\in \mathcal{I}(\mathcal{E})$, using Observation 2 in (Kamath et al, 2021), we have that
	\begin{equation}\label{Q1}
		{\mathbb{E}}_{\mathcal{D}_{e_1}}[Y\mid \varphi(X) = z] = {\mathbb{E}}_{\mathcal{D}_{e_2}}[Y\mid \varphi(X) = z],
	\end{equation}
	for all $e_1, e_2 \in \mathcal{E}$ and for all $z \in \mathcal{Z}$.\footnote{We assume that the support of $\varphi(X)$ (denoted as $\mathcal{Z}$) is identical in each environment for simplicity.}

	\noindent (i) For square loss $\lsq$,
	\[
		\begin{aligned}
			\mathcal{L}_{e}(f) & = \frac{1}{2}\Ede[(f(X) - Y)^2]          \\
			                   & = \frac{1}{2}\Ede[f(X)^2 - 2f(X)Y + Y^2] \\
			                   & =
			\frac{1}{2}\Ede\big[\Ede[w\circ\varphi(X)^2 - 2w\circ\varphi(X)Y\mid \varphi(X)]\big] + \frac{1}{2}\Ede[Y^2],
		\end{aligned}
	\]
	where $w$ is the simultaneously optimal classifier for all $e\in \mathcal{E}$.

	Then, note that for all $z \in \mathcal{Z}$, it holds that
	\[
		\Ede[w(z)^2 - 2w(z)Y\mid \varphi(X) = z]
		= w(z)^2 - 2 w(z) 	\Ede[Y\mid \varphi(X) = z].
	\]
	Using \eqref{Q1} and the assumptions that $\Ede[Y^2]$ is identical and the distribution of  $\varphi(X)$ is identical (or $f\equiv 0$) for all $e\in \mathcal{E}$, we can conclude that for all $e_1, e_2 \in \mathcal{E}$,
	$\mathcal{L}_{e_1}(f) = \mathcal{L}_{e_2}(f).$

	\noindent (ii) For logistic loss $\llog$, note that the simultaneously optimal $w$ has the form
	\[
		\begin{aligned}
			w(z) = \log{\left( \frac{\Prde[Y=1\mid \varphi(X) = z]}{\Prde[Y=-1\mid \varphi(X) = z]}\right)} = \log{\left( \frac{1 + \Ede[Y\mid \varphi(X) = z]}{1 - \Ede[Y\mid \varphi(X) = z]}\right)},
		\end{aligned}
	\]
	for all $e\in \mathcal{E}$ and all $z\in \mathcal{Z}$. We can thus conclude that in this case, $\mathcal{L}_e(f) = \Ede[H(Y|\varphi(X) = z)] =  H(Y|\varphi(X))$, which completes the proof.
\end{proof}

\paragraph{Remarks.} We formulate Lemma \ref{vrex0} in a general setting that covers Two-Bit-Env as a special case. It can be easily verified that the assumptions in this lemma are all satisfied in Two-Bit-Env (Eq.~\ref{CH:PAIR:eq:twobit_env_appdx}). Moreover, we can show that other environment settings (e.g., those in IB-IRM~\citep{ib-irm}) also satisfy the assumptions.

\begin{proposition}\label{recovep_iRM} Under Setting A, for all $\alpha\in (0,1)$, let $\mathcal{E} \coloneqq \{(\alpha, \beta_e): \beta_e \in (0,1)\}$ and $f$ be an odd (or linear) predictor. It holds that $\Ix(\mathcal{E}) \cap \Is(\mathcal{E}) = \mathcal{I}(\mathcal{E})$.
\end{proposition}
\begin{proof}
	From the proof of Proposition 5 in~\citet{irm_aistats}, we know that there are only two predictors in $\mathcal{I}(\mathcal{E})$: The zero predictor $f_0 \equiv 0$ (for both $\lsq$ and $\llog$) and $f_{\textup{IRM}} (x_1, x_2) = (1 - 2\alpha) \cdot x_1$ (for $\ell = \lsq$) or $f_{\textup{IRM}} (x_1, x_2) = \log{\frac{1 - \alpha}{\alpha}} \cdot x_1$ (for $\ell = \llog$).

	\noindent (i) For square loss $\lsq$, $\mathcal{L}_{e}(f) = \frac{1}{2}\Ede[f(X)^2 - 2f(X)Y + Y^2]$. Note that in Two-Bit-Env, $Y^2 \equiv 1$. Thus, in this case, $f\in \Ix(\mathcal{E})$ implies that $\Ede[f(X)^2 - 2f(X)Y]$ is identical for all $e\in \mathcal{E}$. Moreover,
	\[
		\begin{aligned}
			f\in \Is(\mathcal{E}) & \Rightarrow \nabla_{w\mid w=1} {\mathcal{L}_e(f) = 0}\text{ for all }e\in \mathcal{E} \\ &\Rightarrow \Ede{[f(X)^2]} = \Ede{[f(X)Y]}\text{ for all }e\in \mathcal{E}.
		\end{aligned}
	\]
	We can conclude that for any $f\in \Ix(\mathcal{E}) \cap \Is(\mathcal{E})$, it holds that
	\begin{align}
		 & \Ede{[f(X)^2]}\text{ and }\Ede{[f(X)Y]}\text{ are identical for all }e\in \mathcal{E}, \label{C1}
		\\
		 & \Ede{[f(X)^2]} = \Ede{[f(X)Y]}\text{ for all }e\in \mathcal{E}. \label{C2}
	\end{align}

	Denote $f_{(1,1)}\coloneqq f(X_1 = 1, X_2 = 1)$, and $f_{(1, -1)}, f_{(-1, 1)}, f_{(-1, -1)}$ are similarly defined. For condition~\eqref{C1},
	\begin{equation}\label{R1}
		\begin{aligned}
			\Ede{[f(X)^2]} ={} & \frac{1-\alpha}{2}\left(f_{(1,1)}^2 + f_{(-1,-1)}^2\right) + \frac{\alpha}{2} \left(f_{(1,-1)}^2 + f_{(-1,1)}^2\right) \\&+ \frac{\beta_e (1 - 2\alpha)}{2} \left(-f_{(1,1)}^2- f_{(-1,-1)}^2 + f_{(1,-1)}^2  + f_{(-1,1)}^2\right), \\
			\Ede{[f(X)Y]} ={}  & \frac{1 - \alpha}{2} \left(f_{(1,1)}  - f_{(-1,-1)}\right) + \frac{\alpha}{2} \left(f_{(-1,1)}  - f_{(1,-1)}\right)    \\
			                   & - \frac{\beta_e}{2} \left(f_{(1,1)} - f_{(-1,-1)} + f_{(-1,1)} - f_{(1,-1)}\right).
		\end{aligned}
	\end{equation}
	To enforce condition \eqref{C1} for any $\alpha,\beta_e \in (0,1)$, it is required that \[
		\begin{cases}
			f_{(1,1)} - f_{(-1,-1)} + f_{(-1,1)} - f_{(1,-1)} = 0, \\ -f_{(1,1)}^2- f_{(-1,-1)}^2 + f_{(1,-1)}^2  + f_{(-1,1)}^2 = 0.
		\end{cases} \Rightarrow
		\begin{cases}
			f_{(1,1)} - f_{(-1,-1)} = -\left(f_{(-1,1)} - f_{(1,-1)}\right), \\ f_{(1,1)}^2 + f_{(-1,-1)}^2 = f_{(1,-1)}^2  + f_{(-1,1)}^2.
		\end{cases}
	\]
	In this case, condition \eqref{C2} implies that $f_{(1,1)}^2 + f_{(-1,-1)}^2 = (1 - 2\alpha) \left(f_{(1,1)}  - f_{(-1,-1)}\right)$. Without restricting $f$ to be an odd predictor (or equivalently, linear predictor), this constraint is a circle passing through $f_0$ and $f_{\textup{IRM}}$. Requiring that $f$ is odd, i.e., $f_{(1,1)} = - f_{(-1,-1)}$ and $f_{(1,-1)} = - f_{(-1,1)}$, we can conclude that there are only two predictors left in $\Ix(\mathcal{E}) \cap \Is(\mathcal{E})$, which are $f_{(1,1)} = f_{(-1,-1)} = f_{(1,-1)} = f_{(-1,1)} = 0$ and
	\[
		\begin{cases}
			f_{(1,1)} = 1 - 2\alpha, \\ f_{(-1,-1)} = 2\alpha - 1,\\ f_{(1,-1)} = 1 - 2\alpha, \\ f_{(-1,1)} = 2\alpha - 1.
		\end{cases}
		\Rightarrow f (x_1, x_2) = (1 -2\alpha)\cdot x_1.
	\]

	\noindent (ii) For logistic loss $\llog$, $\mathcal{L}_{e}(f) = \Ede\big[\log{\big(1 + \exp{(-f(X)Y)}\big)}\big]$. Similarly, $f\in \Ix(\mathcal{E}) \cap \Is(\mathcal{E})$ implies that
	\begin{gather}
		\Ede\big[\log{\big(1 + \exp{(-f(X)Y)}\big)}\big] \text{ is identical for all }e\in \mathcal{E},\label{C3}\\
		\Ede\left[\frac{-f(X)Y}{1 + \exp{(f(X)Y)}}\right] = 0. \label{C4}
	\end{gather}

	From condition \eqref{C3} and that $f$ is an odd predictor ($f_{(1,1)} = - f_{(-1,-1)}$ and $f_{(1,-1)} = - f_{(-1,1)}$), we can conclude that
	\[
		\frac{(1+e^{f_{(1,1)}})^{2\alpha}}{(1 + e^{-f_{(1,1)}})^{2 - 2\alpha}} = \frac{(1+e^{f_{(1,-1)}})^{2\alpha}}{(1 + e^{-f_{(1,-1)}})^{2 - 2\alpha}} \Rightarrow f_{(1,1)} = f_{(1,-1)},
	\]
	which is due to that $\frac{(1+e^{x})^{2\alpha}}{(1 + e^{-x})^{2 - 2\alpha}}$ is a one-to-one function.

	In this case, condition \eqref{C4} can be simplified as
	\[
		e^{f_{(1,1)}} f_{(1,1)} \alpha - f_{(1,1)}(1 - \alpha) = 0 \Rightarrow f_{(1,1)} = 0 \text{ or } f_{(1,1)} = \log{\frac{1 - \alpha}{\alpha}}.
	\]

	Thus, the only predictors in $\Ix(\mathcal{E}) \cap \Is(\mathcal{E})$ are $f_0$ and $f_{\textup{IRM}}$.
\end{proof}

\begin{corollary}
	Under Setting A, for all $\alpha\in (0,1)$ and $\mathcal{E}_{\textup{tr}} = \{(\alpha, \beta_{e_1}),(\alpha, \beta_{e_2})\}$ for any two distinct $\beta_{e_1}, \beta_{e_2} \in (0,1)$, $\Ix(\mathcal{E}_{\textup{tr}}) \cap \Is(\mathcal{E}_{\textup{tr}}) = \Ix(\mathcal{E}) \cap \Is(\mathcal{E})$.
\end{corollary}
\begin{proof}
	This directly follows from the observation that in the proof of Proposition \ref{recovep_iRM}, enforcing condition \eqref{C1} and \eqref{C3} for two distinct $\beta_{e_1}, \beta_{e_2}$ impose the identical constraints on $f$.
\end{proof}

\subsection{Proof for Theorem~\ref{CH:PAIR:thm:pair_theory}}
\label{proof:pair_theory_appdx}
We first restate the informal version of the theorem as the following, while the formal description of Theorem~\ref{CH:PAIR:thm:pair_theory_appdx} will be given in Theorem~\ref{CH:PAIR:thm:sample_comp} with more formal definitions.
\begin{theorem}(Informal)\label{CH:PAIR:thm:pair_theory_appdx}
	For $\gamma\in(0,1)$ and any $\epsilon,\delta>0$, if $\gF$ is a finite hypothesis class, both ERM and OOD losses are bounded above, let $I_\pair$ be the index of all  losses, $p_{\max} \coloneqq \max_{i\in I_\pair} {p_i}$ and $L_\textup{max} \coloneqq \max_{i\in I_\pair}{L_i}$, if the number of training samples $\abs{D} \geq \frac{32L_\textup{max}^2p_{\max}^2}{\delta^2}\log{\frac{2(m+1)\abs{\mathcal{F}}}{\gamma}}$, then with probability at least $1 - \gamma$,
	\pairo and \pairs yield an $\epsilon$-approximated solution of $f_\ood$.
\end{theorem}
The proof for Theorem~\ref{CH:PAIR:thm:pair_theory} is also a theoretical discussion on the performances of \pairo and \pairs under an approximated OOD preference.
Essentially, the performances of both \pairo and \pairs have a certain dependence on the quality of the OOD preference $\vp_\ood$, however, it is often the case that the ideal OOD preference is usually unknown.
It is desirable to analyze the performances of \pairo and \pairs under an imprecise OOD preference. \citet{epo} discussed a bit that when the exact Pareto optimal solution under the preference does not exist, the EPO solver can still find a Pareto optimal solution that is closest to the preferred direction.
We discuss it in a more general way by developing a new MOO formulation of Eq.~\ref{CH:PAIR:eq:pair_moo_appdx} under an approximated preference up to some approximation error of $\epsilon$.

Without loss of generality, given a OOD preference $\vp_\ood=(p_\erm,p_1,...,p_m)^T=(\frac{1}{\epsilon_\inv},\bm{\frac{1}{\epsilon}_\ood})^T$, the ERM loss $\gL_\erm$ and $m$ OOD losses $\vL_\ood=(\gL_\ood^1,\gL_\ood^2,..,\gL_\ood^m)^T$, Eq.~\ref{CH:PAIR:eq:pair_moo_appdx} can be reformulated as
\begin{equation}\label{CH:PAIR:eq:constrained_pair_moo_appdx}
	\begin{aligned}
		\bm{f}_\pair \coloneqq{} & \argmin_{f \in \mathcal{F}} &  & {\mathcal{L}_\erm (f)} \\ &\ \ \ \ \text{s.t.} &&p_\erm\mathcal{L}_\erm (f) = p_1 \mathcal{L}_\ood^1 (f) = p_2 \mathcal{L}_\ood^2 (f) = \cdots = p_m \mathcal{L}_\ood^m(f).
	\end{aligned}
\end{equation}
We remark that under the ideal OOD preference, the optimal solution of Eq.~\ref{CH:PAIR:eq:constrained_pair_moo_appdx}, is also the optimal solution to Eq.~\ref{CH:PAIR:eq:pair_moo_appdx} (i.e., the unconstrained version). In other words, $ \bm{f}_\pair=f_\ood$. We will use $\bm{f}_\pair$ to differentiate from the solution to the unconstrained version.
We focus on Eq.~\ref{CH:PAIR:eq:constrained_pair_moo_appdx} for the reason that it is more convenient to establish the discussion on the approximated OOD preference, from the perspective of optimization constraints.

Exactly enforcing the above preference constraint is too restrictive {\it both practically and theoretically}, instead we incorporate the approximation by relaxing the constraint of the loss values w.r.t. the OOD preference. The $\epsilon$-approximated problem of Eq.~\ref{CH:PAIR:eq:constrained_pair_moo_appdx} is as the following
\begin{equation}\label{CH:PAIR:eq:ep_pair_moo_appdx}
	\begin{aligned}
		\bm{f}^\epsilon_\pair \coloneqq{} & \argmin_{f \in \mathcal{F}} &  & {\mathcal{L}_\erm (f)} \\ &\ \ \ \ \text{s.t.} && \forall i, j \in I_\pair, i\neq j, \,\abs{p_i\mathcal{L}_i (f) - p_j \mathcal{L}_j (f)} \leq \epsilon,
	\end{aligned}
\end{equation}
where $I_\pair \coloneqq \{\erm, \ood_1, \ood_2,\ldots, \ood_m\}$ is the index set of overall losses. We denote the relaxed constraint set in Eq.~\ref{CH:PAIR:eq:ep_pair_moo_appdx} as $\Peo \coloneqq \{f \mid \forall i, j \in I_\pair, i\neq j, \,\abs{p_i\mathcal{L}_i (f) - p_j \mathcal{L}_j (f)} \leq \epsilon\}$. Clearly, it holds that the solution sets satisfy $\bm{f}^0_\pair = \bm{f}_\pair$.

Then we define the empirical version of the $\epsilon$-approximated problem Eq.~\ref{CH:PAIR:eq:ep_pair_moo_appdx} with preference vector $\vp_\ood$ as follows.
\begin{equation}\label{CH:PAIR:eq:eep_pair_moo_appdx}
	\begin{aligned}
		\hat{\bm{f}}^\epsilon_\pair \coloneqq{} & \argmin_{f \in \mathcal{F}} &  & {\hL_\erm (f)} \\ &\ \ \ \ \text{s.t.} && \forall i, j \in I_\pair, i\neq j, \,\abs{p_i\hL_i (f) - p_j \hL_j (f)} \leq \epsilon.
	\end{aligned}
\end{equation}
Similarly, we denote the above constraint set as $\Peeo \coloneqq \{f \mid \forall i, j \in I_\pair, i\neq j, \,\abs{p_i\hL_i (f) - p_j \hL_j (f)} \leq \epsilon\}$.

Assume a finite hypothesis class $\mathcal{F}$ and define
\[
	\delta = \min_{f\in \mathcal{F},\forall i, j \in I_\pair, i\neq j} \absbig{\abs{p_i\mathcal{L}_i (f) - p_j \mathcal{L}_j (f)} - \epsilon}.
\]
First, we recall the definition of $\nu$-representative sample from~\citet{mlt}.

\begin{definition}
	\label{def:e-represent}
	(\citet{mlt}) A training set $S$ is called $\nu$-representative (w.r.t.\ domain $\mathcal{X}$, hypothesis $\mathcal{F}$, loss $\ell$ and distribution $\mathcal{D}$) if
	\[
		\forall f\in \mathcal{F}, \abs{\hL(f) - \mathcal{L}(f)} \leq \nu,
	\]
	where $\mathcal{L}(f) \coloneqq \mathbb{E}_{(X,Y)\sim\mathcal{D}}[\ell(f(X), Y)]$ and $\hL(f) \coloneqq \frac{1}{\abs{S}}\sum_{(X_i, Y_i) \in S}\ell(f(X_i), Y_i)$.
\end{definition}

Equipped with this definition, we can now characterize the condition under which the constraint sets in \eqref{CH:PAIR:eq:ep_pair_moo_appdx} and \eqref{CH:PAIR:eq:eep_pair_moo_appdx} contain exact the same predictors.

\begin{lemma}
	\label{lem:constraint_set}
	For any $\epsilon > 0$, assuming $\delta > 0$ and denoting $p_{\max} \coloneqq \max_{i\in I_\pair} {p_i}$, if the training set $\train$ is $\frac{\delta}{4p_{\max}}$-representative w.r.t.\ domain $\mathcal{X}$, hypothesis $\mathcal{F}$, distribution $\mathcal{D}$ and all the ERM and OOD losses $\{\mathcal{L}_\erm, \bm{L}_\ood\}$, then $\Peo = \Peeo$.
\end{lemma}
\begin{proof}
	We first show that $\Peo \subseteq \Peeo$. By the definition of $\delta$, for all $f\in \mathcal{F}$, and $\forall i, j \in I_\pair, i\neq j$ we have
	\begin{equation}\label{bullshit-condition}
		\abs{p_i\mathcal{L}_i (f) - p_j \mathcal{L}_j (f)} \leq \epsilon - \delta \,\text{ or }\,\abs{p_i\mathcal{L}_i (f) - p_j \mathcal{L}_j (f)} \geq \epsilon + \delta.
	\end{equation}

	Using this property, for any $f\in \Peo$, we can conclude that $\forall i, j \in I_\pair, i\neq j$,
	\[
		\abs{p_i\mathcal{L}_i (f) - p_j \mathcal{L}_j (f)} \leq \epsilon \Rightarrow \abs{p_i\mathcal{L}_i (f) - p_j \mathcal{L}_j (f)} \leq \epsilon - \delta.
	\]

	This inequality further implies that
	\[
		\begin{aligned}
			              & \abs{p_i\mathcal{L}_i (f) - p_i \hL_i(f) + p_j\hL_j(f) - p_j \mathcal{L}_j (f) + p_i\hL_i(f) - p_j\hL_j(f)} \leq \epsilon - \delta                 \\
			\Rightarrow{} & \absbig{ \abs{p_i\hL_i(f) - p_j\hL_j(f)} - \abs{p_i\mathcal{L}_i (f) - p_i \hL_i(f) + p_j\hL_j(f) - p_j \mathcal{L}_j (f)}} \leq \epsilon - \delta \\
			\Rightarrow{} & \abs{p_i\hL_i(f) - p_j\hL_j(f)} \leq \epsilon - \delta + \abs{p_i\mathcal{L}_i (f) - p_i \hL_i(f) + p_j\hL_j(f) - p_j \mathcal{L}_j (f)}           \\
			\Rightarrow{} & \abs{p_i\hL_i(f) - p_j\hL_j(f)} \leq \epsilon - \delta + p_i\abs{\mathcal{L}_i (f) -  \hL_i(f)} + p_j\abs{\hL_j(f) -  \mathcal{L}_j (f)},
		\end{aligned}
	\]
	which is based on the triangle inequality of the absolute value function.

	From the definition of $\frac{\delta}{4p_{\max}}$-representative, we have $\abs{\mathcal{L}_i (f) -  \hL_i(f)} \leq \frac{\delta}{4p_{\max}}, \forall i \in I_\pair$. Substituting this in the above inequality, we obtain
	\[
		\begin{aligned}
			\abs{p_i\hL_i(f) - p_j\hL_j(f)} & \leq \epsilon - \delta + \frac{p_i\delta}{4p_{\max}} + \frac{p_j\delta}{4p_{\max}} \\
			                                & \leq \epsilon - \frac{\delta}{2},
		\end{aligned}
	\]
	which implied that $f\in \Peeo$.

	Then, we prove that $\Peeo \subseteq \Peo$.

	For any $f\in \Peeo$, it holds that $\forall i, j \in I_\pair, i\neq j$,
	\[
		\begin{aligned}
			                             & \abs{p_i\hL_i (f) - p_j \hL_j (f)} \leq \epsilon                                                                                                            \\
			\Rightarrow{}                & \abs{p_i\hL_i (f) - p_i\mathcal{L}_i (f) + p_j\mathcal{L}_j (f) - p_j \hL_j (f) + p_i\mathcal{L}_i (f) - p_j\mathcal{L}_j (f)} \leq \epsilon                \\
			\Rightarrow{}                & \absbig{\abs{p_i\mathcal{L}_i (f) - p_j\mathcal{L}_j (f)} - \abs{p_i\hL_i (f) - p_i\mathcal{L}_i (f) + p_j\mathcal{L}_j (f) - p_j \hL_j (f)}} \leq \epsilon \\
			\Rightarrow{}                & \abs{p_i\mathcal{L}_i (f) - p_j\mathcal{L}_j (f)} \leq \epsilon + \abs{p_i\hL_i (f) - p_i\mathcal{L}_i (f) + p_j\mathcal{L}_j (f) - p_j \hL_j (f)}          \\
			\Rightarrow{}                & \abs{p_i\mathcal{L}_i (f) - p_j\mathcal{L}_j (f)} \leq \epsilon + p_i\abs{\hL_i (f) - \mathcal{L}_i (f)} + p_j\abs{\mathcal{L}_j (f) - \hL_j (f)}           \\
			\Rightarrow{}                & \abs{p_i\mathcal{L}_i (f) - p_j\mathcal{L}_j (f)} \leq \epsilon + \frac{p_i\delta}{4p_{\max}} + \frac{p_j\delta}{4p_{\max}}                                 \\
			{\color{white}\Rightarrow{}} & {\color{white}\abs{p_i\mathcal{L}_i (f) - p_j\mathcal{L}_j (f)}} \leq \epsilon + \frac{\delta}{2},
		\end{aligned}
	\]
	which is again based on the triangle inequality of the absolute value function and the definition of $\frac{\delta}{4p_{\max}}$-representative. Together with \eqref{bullshit-condition}, we conclude that $\abs{p_i\mathcal{L}_i (f) - p_j \mathcal{L}_j (f)} \leq \epsilon - \delta \Rightarrow f \in \Peo$, which implies $\Peeo \subseteq \Peo$.

	Based on the above discussion, we have proven that $\Peo = \Peeo$.
\end{proof}

\begin{assumption}
	\label{strong-assumption}
	For all $f\in \mathcal{F}, X\in \mathcal{X}, Y\in \mathcal{Y}$, the ERM loss is bounded, i.e., $\abs{\ell(f(X), Y)} \leq L_\erm < \infty$, and all the OOD objectives $\bm{L}_\ood$ can be written as the expectation of some bounded loss functions, i.e., $\forall i \in [m], \mathcal{L}_\ood^i(f) = \mathbb{E}_{(X,Y)\sim\mathcal{D}} [\ell_\ood^i(f(X), Y)]$ and $\abs{\ell_\ood^i(f(X), Y)} \leq L_\ood^i < \infty$.
\end{assumption}

We remark that the assumption is natural and generally holds for many OOD objectives including \irml~\citep{irmv1} and \vrex~\citep{vrex}.

\begin{theorem}
	\label{CH:PAIR:thm:sample_comp}
	For any $\epsilon > 0, \gamma \in (0, 1)$, if Assumption \ref{strong-assumption} holds and $\delta > 0$, denoting $p_{\max} \coloneqq \max_{i\in I_\pair} {p_i}$ and $L_\textup{max} \coloneqq \max_{i\in I_\pair}{L_i}$, if the number of training samples $\abs{\train} \geq \frac{32L_\textup{max}^2p_{\max}^2}{\delta^2}\log{\frac{2(m+1)\abs{\mathcal{F}}}{\gamma}}$, then with probability at least $1 - \gamma$, we have for any $f^\epsilon_\pair \in \bm{f}^\epsilon_\pair$ and $\hat{f}^\epsilon_\pair\in \hat{\bm{f}}^\epsilon_\pair$, $\mathcal{L}_\erm(f^\epsilon_\pair)\leq \mathcal{L}_\erm(\hat{f}^\epsilon_\pair) \leq \mathcal{L}_\erm(f^\epsilon_\pair) + \frac{\delta}{2p_{\max}}$.
\end{theorem}
\begin{proof}
	We proceed by first assuming that the training set $D$ is $\frac{\delta}{4p_{\max}}$-representative w.r.t.\ domain $\mathcal{X}$, hypothesis $\mathcal{F}$, distribution $\mathcal{D}$ and all the ERM and OOD losses $\{\mathcal{L}_\erm, \bm{L}_\ood\}$, and then we establish the sample complexity required for this condition. From Lemma \ref{lem:constraint_set}, we know that given this condition and the assumptions in the theorem, $\Peo = \Peeo$. Then, since the training set $\train$ is $\frac{\delta}{4p_{\max}}$-representative w.r.t.\ the ERM loss $\mathcal{L}_\erm$, we have for any $f^\epsilon_\pair \in \bm{f}^\epsilon_\pair$ and $\hat{f}^\epsilon_\pair\in \hat{\bm{f}}^\epsilon_\pair$,
	\[
		\begin{gathered}
			\absbig{\mathcal{L}_\erm(f^\epsilon_\pair) - \hL_\erm(f^\epsilon_\pair)} \leq \frac{\delta}{4p_{\max}},\\
			\absbig{\mathcal{L}_\erm(\hat{f}^\epsilon_\pair) - \hL_\erm(\hat{f}^\epsilon_\pair)} \leq \frac{\delta}{4p_{\max}}.
		\end{gathered}
	\] Moreover, based on the optimality of problem \eqref{CH:PAIR:eq:eep_pair_moo_appdx}, we can conclude that
	\[
		\begin{aligned}
			              & \mathcal{L}_\erm(\hat{f}^\epsilon_\pair) - \frac{\delta}{4p_{\max}}\leq \hL_\erm(\hat{f}^\epsilon_\pair) \leq \hL_\erm(f^\epsilon_\pair) \leq \mathcal{L}_\erm(f^\epsilon_\pair) + \frac{\delta}{4p_{\max}} \\
			\Rightarrow{} & \mathcal{L}_\erm(\hat{f}^\epsilon_\pair) \leq \mathcal{L}_\erm(f^\epsilon_\pair) + \frac{\delta}{2p_{\max}}.
		\end{aligned}
	\]
	Then, using the optimality of problem \eqref{CH:PAIR:eq:ep_pair_moo_appdx}, it holds that
	\[
		\mathcal{L}_\erm(f^\epsilon_\pair)\leq \mathcal{L}_\erm(\hat{f}^\epsilon_\pair) \leq \mathcal{L}_\erm(f^\epsilon_\pair) + \frac{\delta}{2p_{\max}}.
	\]

	It remains to analyze the sample complexity of ensuring that the training set $\train$ is $\frac{\delta}{4p_{\max}}$-representative w.r.t.\ $\mathcal{X}$, $\mathcal{F}$, $\mathcal{D}$ and all the ERM and OOD losses $\{\mathcal{L}_\erm, \bm{L}_\ood\}$.

	For any $i \in I_\pair$, based on Assumption \ref{strong-assumption}, we can write $\mathcal{L}_i(f) = \mathbb{E}_{(X,Y)\sim\mathcal{D}}[\ell_i(f(X), Y)]$ and $\hL_i(f) = \frac{1}{\abs{D}}\sum_{(X_j, Y_j)\in D}\ell_i(f(X_j), Y_j)$ with $\abs{\ell_i(f(X), Y)} \leq L_i \leq L_\textup{max}, \forall f, X, Y$. Using Hoeffding's inequality, we can conclude that for any $f\in \mathcal{F}$,
	\[
		\Pr\left[\absbig{\hL_i(f) - \mathcal{L}_i(f)} \geq \frac{\delta}{4p_{\max}} \right] \leq 2\exp{\left(\frac{-\abs{D}\delta^2}{32L_\textup{max}^2p_{\max}^2}\right)}.
	\]
	Thus, for any $\gamma\in (0,1)$, if we require
	\[
		\abs{D} \geq \frac{32L_\textup{max}^2p_{\max}^2}{\delta^2}\log{\frac{2(m+1)\abs{\mathcal{F}}}{\gamma}},
	\]
	it holds that
	\[
		\Pr\left[\exists f\in \mathcal{F}, \absbig{\hL_i(f) - \mathcal{L}_i(f)} \geq \frac{\delta}{4p_{\max}} \right] \leq \sum_{f\in \mathcal{F}}{\Pr\left[\absbig{\hL_i(f) - \mathcal{L}_i(f)} \geq \frac{\delta}{4p_{\max}} \right]} \leq \frac{\gamma}{m+1}.
	\]
	Thus,
	\[
		\begin{aligned}
			 & \Pr\left[\exists i\in I_\pair, \exists f\in \mathcal{F}, \absbig{\hL_i(f) - \mathcal{L}_i(f)} \geq \frac{\delta}{4p_{\max}} \right] \\ \leq{}& \sum_{i\in I_\pair}{\Pr\left[\exists f\in \mathcal{F},\absbig{\hL_i(f) - \mathcal{L}_i(f)} \geq \frac{\delta}{4p_{\max}} \right]} \leq \gamma.
		\end{aligned}
	\]
	Finally, we can conclude that with probability at least $1 - \gamma$, $\forall i\in I_\pair, \forall f\in \mathcal{F},$
	\[
		\absbig{\hL_i(f) - \mathcal{L}_i(f)} \leq \frac{\delta}{4p_{\max}},
	\]
	which completes the proof.
\end{proof}

\paragraph{Remarks.} The $\epsilon$-approximated formulation has a close relationship to another relaxation as the following.
\[
	\begin{aligned}
		\bm{f}_\pair \coloneqq{} & \argmin_{f \in \mathcal{F}} &  & {\mathcal{L}_\erm (f)}                                     \\
		                         & \ \ \ \ \text{s.t.}         &  & \mathcal{L}^i_\pair (f) \leq \epsilon_i, \forall i\in [m].
	\end{aligned}
\]
Essentially, both the $\epsilon$-approximated formulation and the above formulation are natural relaxation of the original problem (Eq.~\ref{CH:PAIR:eq:constrained_pair_moo_appdx} or Eq.~\ref{CH:PAIR:eq:pair_moo_appdx}). As the $\epsilon_i\rightarrow \text{$\epsilon_\ood$}_i$, the above formulation also yields the optimal solution $f_\ood$. In this work, since we focus on the approximations on the preference, $\epsilon$-approximated formulation essentially provides a convenient touch which could be of independent interests for future discussions.

\section{More Details on Experiments}
\label{CH:PAIR:sec:experiments_appdx}
In this section, we provide more details about the experiments (Sec.~\ref{CH:PAIR:sec:experiments}) in the main paper.

\subsection{More details on \cmnist experiments}
\label{CH:PAIR:sec:cmnist_appdx}
In the proof-of-concept experiments with \cmnist,
we follow the evaluation settings as IRM~\citep{irmv1} and the test-domain selection as DomainBed~\citep{domainbed}.
Specifically, we use a $4$-Layer MLP with a hidden dimension of $256$. By default, we use Adam~\cite{adam} optimizer with a learning rate of $1e-3$ and a weight decay of $1e-3$ to train the model with $500$ epochs and select the last epoch as the output model for each hyperparameter setting.
We choose the final model from different hyperparameter setups as the one that maximizes the accuracy on the validation that share the same distribution as test domain.
We then do grid search for the corresponding hyperparameters.
For pretraining epochs, we search from $\{0,50,100,150,200,250\}$.
For OOD penalty, we search from $\{1e1,1e2,1e3,1e4,1e5\}$. We evaluate each configuration of hyperparameters $10$ times and report the mean and standard deviation of the performances.
Besides, for \irml, we will refresh the history in Adam optimizer when the pretraining finishes, following the practice in~\citet{domainbed}. We also empirically find that refreshing the optimizer after pretraining can bring a better performance of \irml in \cmnist. While for \vrex, we find the refreshing is not needed.

For the implementation of \irmx, we change the penalty to be the sum of \irml and \vrex losses and conduct the same hyperparameter search as for \irml for fair comparison.
As for the implementation of \pair, we use SGD with a momentum of $0.9$~\citep{momentum} after pretraining, to avoid the interference of Adam to the gradient direction and convergence of EPO~\citep{epo} solver.
Moreover, we also empirically find that SGD requires larger learning rate (we search over two choices, i.e., $0.01$ and $0.1$) for approaching the direction.
This is because of the design in EPO solver that it first fits to the preference direction then does the ``pure'' gradient descent, while the intrinsically conflicting directions pointed by the objectives can make the loss surface more steep.
We will leave in-depth understanding of the above phenomenon and more sophisticated optimizer design in more complex tasks and network architectures to future works~\citep{importance_sampling,robust_momentum}.

\subsection{More details about ablation studies}
\label{CH:PAIR:sec:ablation_appdx}
\begin{figure}[t]
    \centering
	\subfigure[CMNIST.]{
		\includegraphics[width=0.4\textwidth]{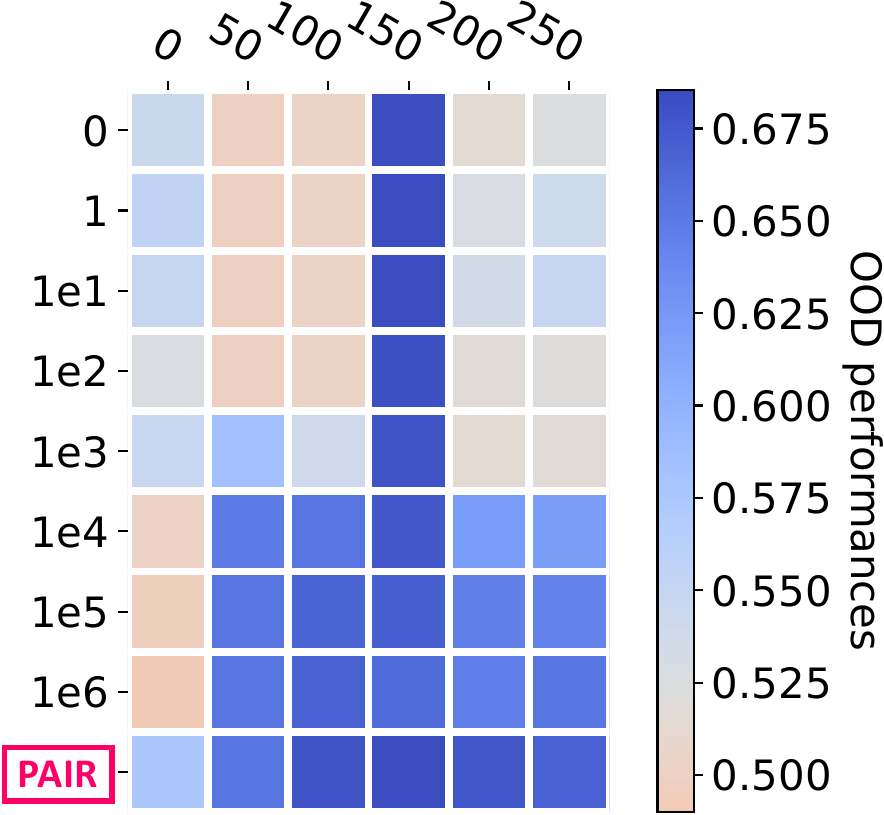}
		\label{CH:PAIR:fig:scalar_appdx}
	}
	\subfigure[CMNIST-m.]{
		\includegraphics[width=0.4\textwidth]{Figures/PAIR/sweep_acc_c01_short_color_crop.pdf}
		\label{CH:PAIR:fig:scalar_c01_appdx}
	}
	\subfigure[CMNIST losses.]{
		\includegraphics[width=0.4\textwidth]{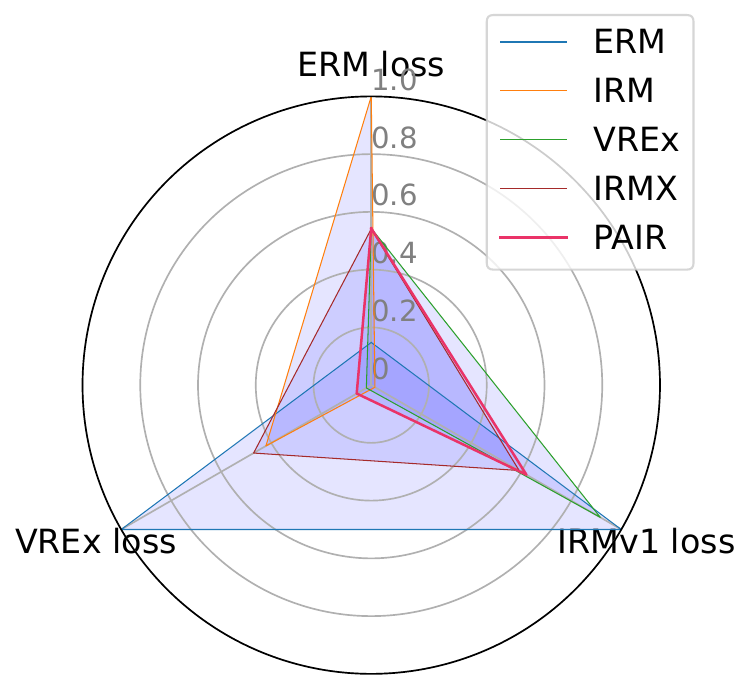}
		\label{CH:PAIR:fig:loss_radar_appdx}
	}
	\subfigure[CMNIST-m losses.]{
		\includegraphics[width=0.4\textwidth]{Figures/PAIR/loss_radar_c01.pdf}
		\label{CH:PAIR:fig:loss_radar_c01_appdx}
	}
	\caption[More abalation studies of \pair.]{(a),(b) \pair can effectively find a better solution than exhaustive tuning of penalty weights in \irmx. That is because \pair can adaptively adjust the penalty weights during the optimization process, and leads to a Pareto optimal solution, as shown in (c),(d).}
	\label{CH:PAIR:fig:ablation_exp_appdx}
\end{figure}

\textbf{Comparison between \pairo and the linear weighting scheme under exhaustive parameter search.}
In the main paper, to investigate how \pairo can find a better OOD solution under objective conflicts, we first conduct an ablation study to compare the OOD performances of \pairo and the exhaustive tuned \irmx. Specifically, we tune both \irml and \vrex penalty weights from a substantially larger scope, i.e., $\{1,1e1,1e2,1e3,1e4,1e5,1e6\}$. As for pretraining epochs, we search from $\{0,50,100,150,200,250\}$. The results of \irmx in \cmnist and the modified \cmnist are shown as in Fig.~\ref{CH:PAIR:fig:scalar_appdx} and Fig.~\ref{CH:PAIR:fig:scalar_c01_appdx}, respectively.
Each point represents the best performed \irmx with the configuration of the corresponding pretraining epoch, the \irml penalty weight and different \vrex penalty weights from $\{1,1e1,1e2,1e3,1e4,1e5,1e6\}$.

\begin{wrapfigure}{r}{0.64\textwidth}
	\subfigure[CMNIST.]{
		\includegraphics[width=0.301\textwidth]{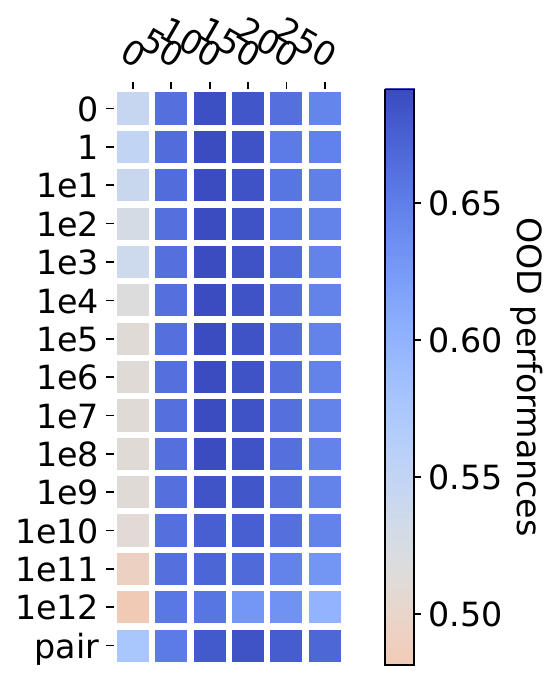}
	}
	\subfigure[CMNIST-m.]{
		\includegraphics[width=0.3\textwidth]{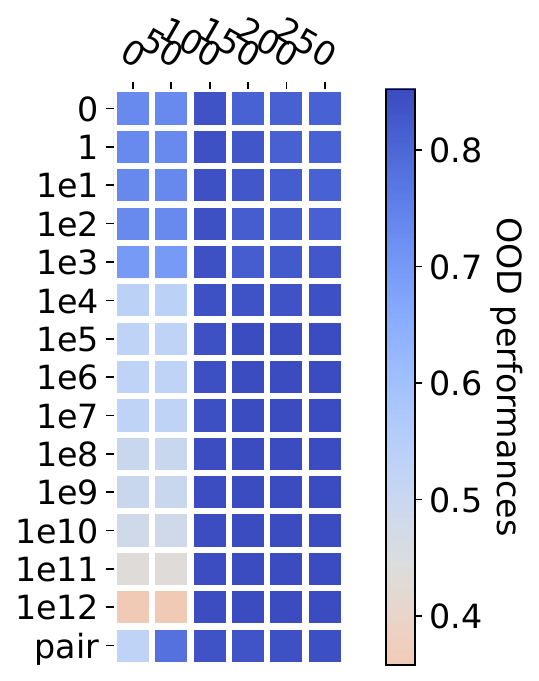}
	}
	\caption{Full exhaustive hyperparameter tunning study}
	\label{CH:PAIR:fig:scalar_full_appdx}
\end{wrapfigure}

We also present a full exhaustive hyperparameter tunning study based on a linear weighting scheme for \irmx, shown in Fig.~\ref{CH:PAIR:fig:scalar_full_appdx}, where we further enlarge the search space of penalty weights from $1e6$ to $1e12$ to better compare with \irmx optimized via \pairo. Similar to Fig.~\ref{CH:PAIR:fig:scalar_appdx} and Fig.~\ref{CH:PAIR:fig:scalar_c01_appdx}, each point in Fig.~\ref{CH:PAIR:fig:scalar_full_appdx} is selected from \textit{best performed} models trained with the corresponding \irml penalty weights, and pretraining epoch, and all possible \vrex penalty weights from $\{1,1e1,1e2,1e3,1e4,1e5,1e6, 1e7, 1e8, 1e9, 1e10, 1e11, 1e12\}$.

Compared to \irml shown as in Fig.~\ref{CH:PAIR:fig:sweep_irm_appdx}, \irmx can substantially improve the OOD performances in both \cmnist and the modified \cmnist, confirming our theoretical results. However, the OOD performances of \irmx turn out to be upper bounded by that optimized with \pairo at each pretraining epochs. In other words, \pairo requires substantially less parameter tuning efforts to achieve the top OOD performances, confirming the advances of \pairo.
In more complex tasks where the exhaustive parameter tunning is prohibitively expensive, such as in the experiments with \wilds~\citep{wilds}, \irmx performs worse than \pair, which further validates the effectiveness of \pairo.

To better demonstrate the advantages of \pairo over linear weighting scheme, we replicate the previous study in two datasets from \wilds, i.e., \textsc{CivilComments} and \textsc{FMoW}. Due to the computational resource limits, we limit the search scope of \irml and \vrex to $\{1e-2,1,1e2\}$, respectively. It can be found that, even with a broader hyperparameter search space, \irmx optimized via linear weighting scheme remain under-performed than \pairo.
\begin{table}[ht]
	\small\centering
	\caption{Comparison between linear weighting scheme and \pairo in \wilds.}
	\label{CH:PAIR:tab:irmx_pair_wilds_appdx}
	\resizebox{\textwidth}{!}{

		\begin{tabular}{llccc||llccc}
			\toprule
			\textbf{\textsc{CivilComments}} & IRMv1\textbackslash{}VREx & $1e-2$           & 1                        & $1e2$            & \textbf{\textsc{FMoW}} & IRMv1\textbackslash{}VREX & $1e-2$            & 1                         & $1e2$             \\\midrule
			                                & $1e-2$                    & $72.5$\std{2.00} & $73.8$\std{1.40}         & $73.1$\std{0.67} &                        & $1e-2$                    & $33.64$\std{0.59} & $34.20$\std{1.33}         & $34.43$\std{0.72} \\
			                                & 1                         & $73.5$\std{1.47} & $74.3$\std{0.83}         & $73.2$\std{0.67} &                        & 1                         & $30.25$\std{0.87} & $33.75$\std{0.78}         & $33.7$\std{0.78}  \\
			                                & $1e2$                     & $72.1$\std{0.59} & $70.1$\std{2.09}         & $74.3$\std{0.51} &                        & $1e2$                     & $21.33$\std{1.51} & $21.00$\std{2.41}         & $13.14$\std{1.63} \\\midrule
			\pairo                          &                           &                  & $\mathbf{75.2}$\std{0.7} &                  &                        &                           &                   & $\mathbf{35.5}$\std{1.13} &                   \\
			\bottomrule
		\end{tabular}}
\end{table}

\textbf{Loss values distribution at convergence.}
As for the loss distribution experiments (Fig.~\ref{CH:PAIR:fig:loss_radar_appdx},~\ref{CH:PAIR:fig:loss_radar_c01_appdx}), we plot the \erm,\irml and \vrex loss values at convergence of best performed algorithms. The plotted values are in log-scale and normalized to $[0,1]$.
It can be found that \pairo effectively find a better solution in terms of \irml and \vrex losses, while not generating the \erm performances too much, which confirms our motivations for the design of \pair.

\begin{wrapfigure}{r}{0.64\textwidth}
	\subfigure[CMNIST.]{
		\includegraphics[width=0.301\textwidth]{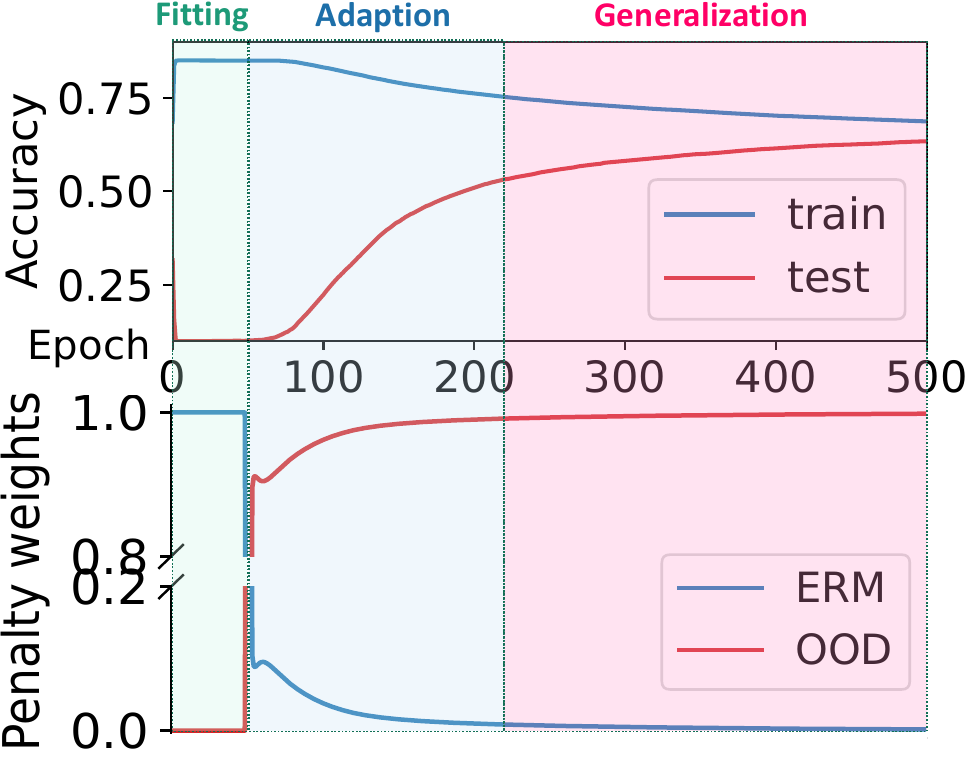}
	}
	\subfigure[CMNIST-m.]{
		\includegraphics[width=0.3\textwidth]{Figures/PAIR/trajectory_c01_crop.pdf}
	}
	\caption{Penalty weights trajectory}
	\label{CH:PAIR:fig:trajectory_appdx}
	\vspace{-0.2in}
\end{wrapfigure}

\textbf{Penalty weights trajectory.}
To examine whether \pairo can effectively adjust the penalty weights of ERM and OOD objectives, especially when the model has not arrived at the Pareto front (i.e., the gradient conflicts are expected to be more intense), we plot the trajectories of penalty weights generated by \pairo in both CMNIST and CMNIST-m, shown as in Fig.~\ref{CH:PAIR:fig:trajectory_appdx}.

It can be found that the whole training process can be divided into three phases: ``Fitting'' phase; ``Adaption'' phase; and ``Generalization'' phase.
In the ``Fitting'' phase, the model is trained with only the ERM objectives and is expected to approach the Pareto front first (cf. Fig.~\ref{CH:PAIR:fig:pair_opt_appdx}). It also corresponds to the ``descent'' phase in the \pairo algorithm, hence the penalty weight for ERM objective is $1$ while for OOD objective is $0$.
Then, when \pairo enters into the ``balance'' phase, \pairo begins to yield high weights to OOD objectives, while not diminishing the weights to ERM objectives.
That is the ``Adaption'' phase, where \pairo begins to adjust the solution towards the Pareto front as well as the preferred direction.
When the solution is close to the Pareto front, then \pairo enters into the ``Generalization'' phase. That is to incorporate the invariance into the features by assigning high weights to the OOD objectives.

\begin{wrapfigure}{r}{0.64\textwidth}
	\vspace{-0.3in}
	\subfigure[CMNIST.]{
		\includegraphics[width=0.3\textwidth]{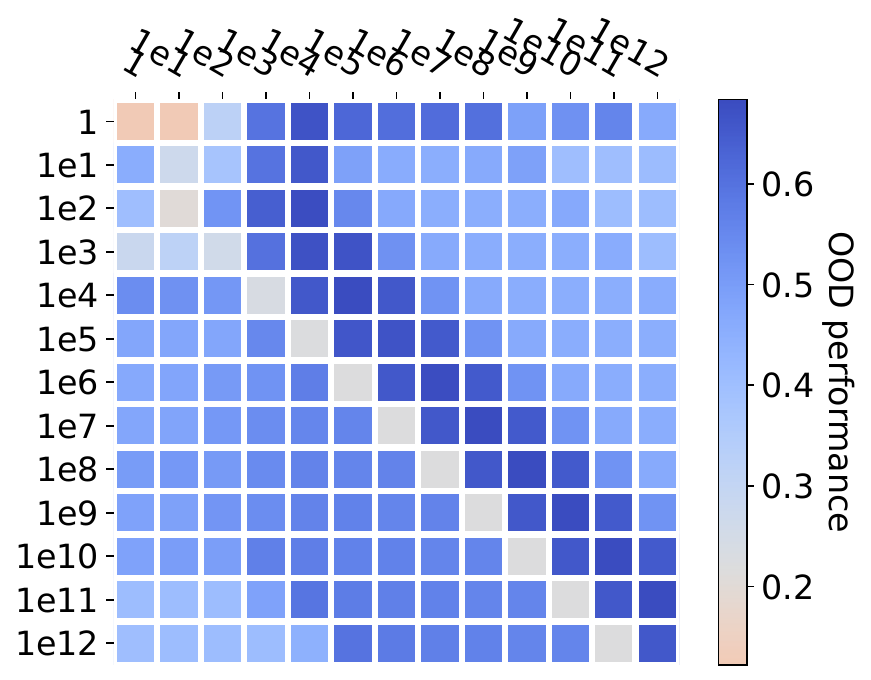}
	}
	\subfigure[CMNIST-m.]{
		\includegraphics[width=0.3\textwidth]{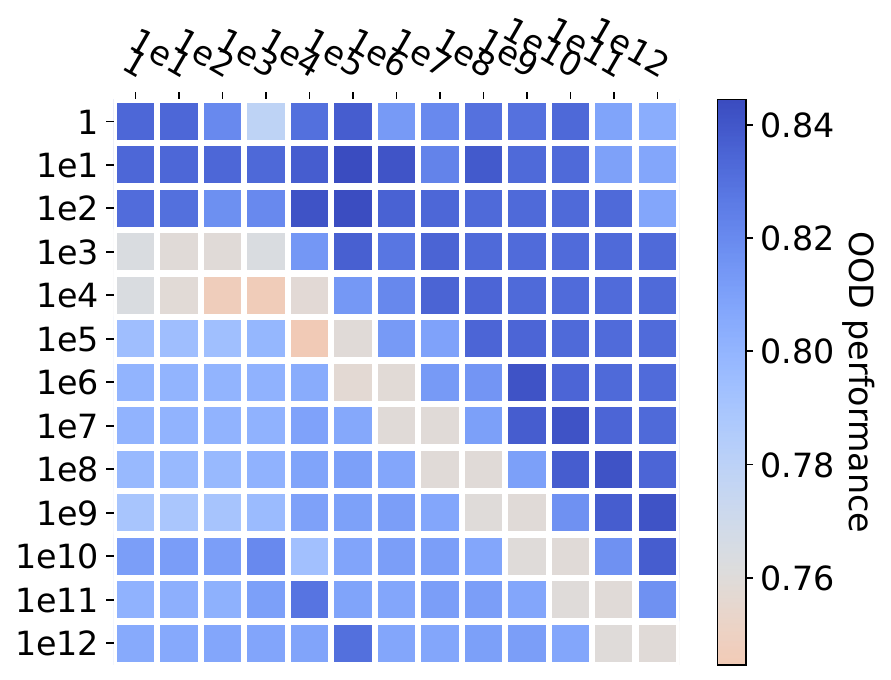}
	}
	\vspace{-0.1in}
	\caption[Preference sensitivity under \emph{strict} hyperparameter configuration.]{Preference sensitivity under \emph{strict} hyperparameter configuration. $x$-axis is the preference for \vrex while $y$-axis is the preference for \irml}
	\label{CH:PAIR:fig:pair_pc_sens_appdx}
	\vspace{-0.2in}
\end{wrapfigure}

\textbf{Preference sensitivity analysis under strict hyperparameter configuration.}
Another reason for the high performance of \pairo at both \cmnist and realistic datasets from \wilds is because of its robustness to different preference choices.
In complementary to the theoretical discussion in Theorem~\ref{CH:PAIR:thm:pair_theory_appdx}, we also conducted preference sensitivity analysis experiments under strict hyperparameter configurations.
In other words, the hyperparameter search space is restricted to \emph{single} point, i.e., a learning rate of $0.01$, and a pretraining epoch of $150$.
The results are shown in Fig.~\ref{CH:PAIR:fig:pair_pc_sens_appdx} for both the original and the modified \cmnist dataset. It can be found that, \pairo maintains high performance and robustness to different preference choices.

It also aligns with our discussion on preference choice in practice (Sec.~\ref{CH:PAIR:sec:pair_discussion_pc_appdx}), that we need to assign a higher preference to \emph{robust, and more easy-to-optimize} objectives, i.e., \vrex.
When the relative preferences are given within a reasonable scale, \pairo easily yields top OOD performances.

\begin{figure}[ht]
	\subfigure[``Perfect'' init. on CMNIST-10.]{
		\centering
		\includegraphics[width=0.3\textwidth]{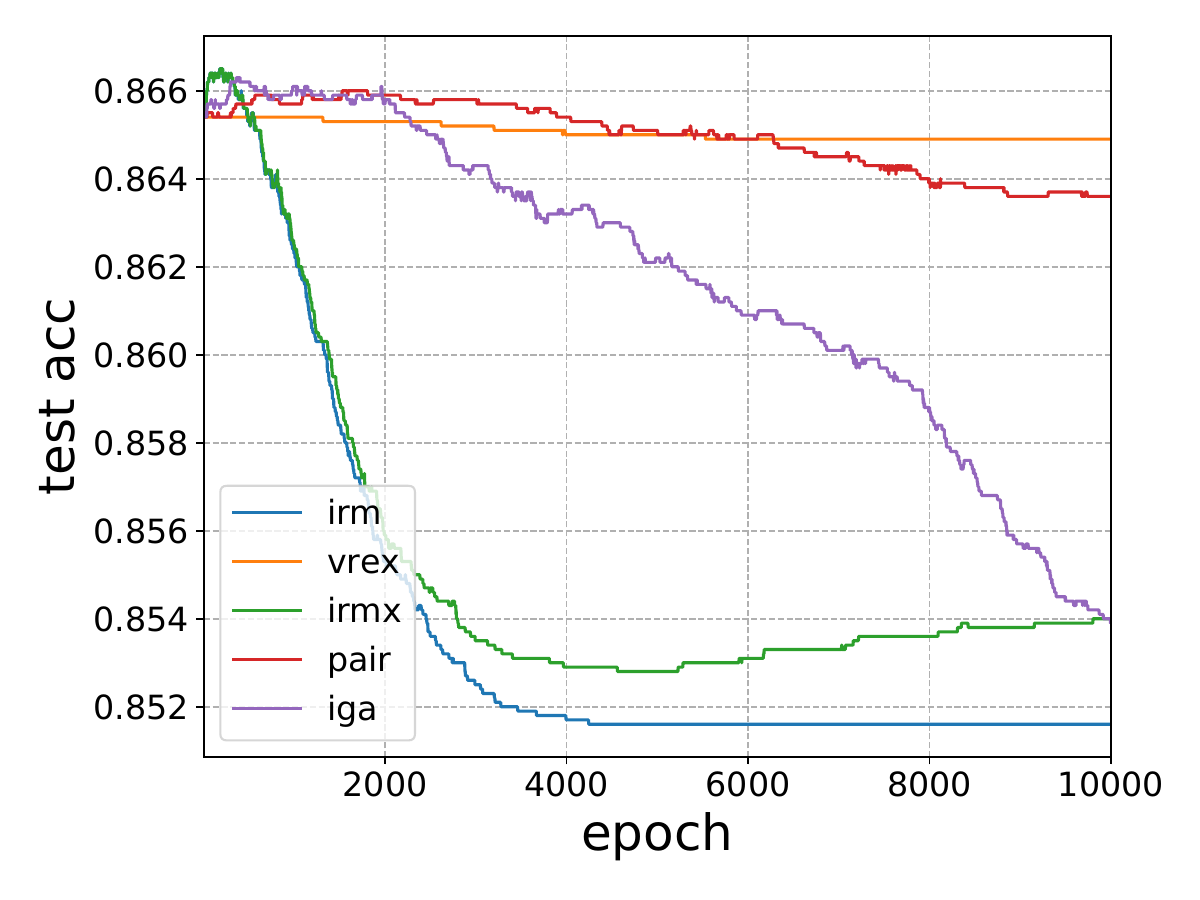}
		\label{CH:PAIR:fig:c10_perfect_init_appdx}
	}
	\hfill
	\subfigure[``Perfect'' init. on CMNIST-11.]{
		\centering
		\includegraphics[width=0.3\textwidth]{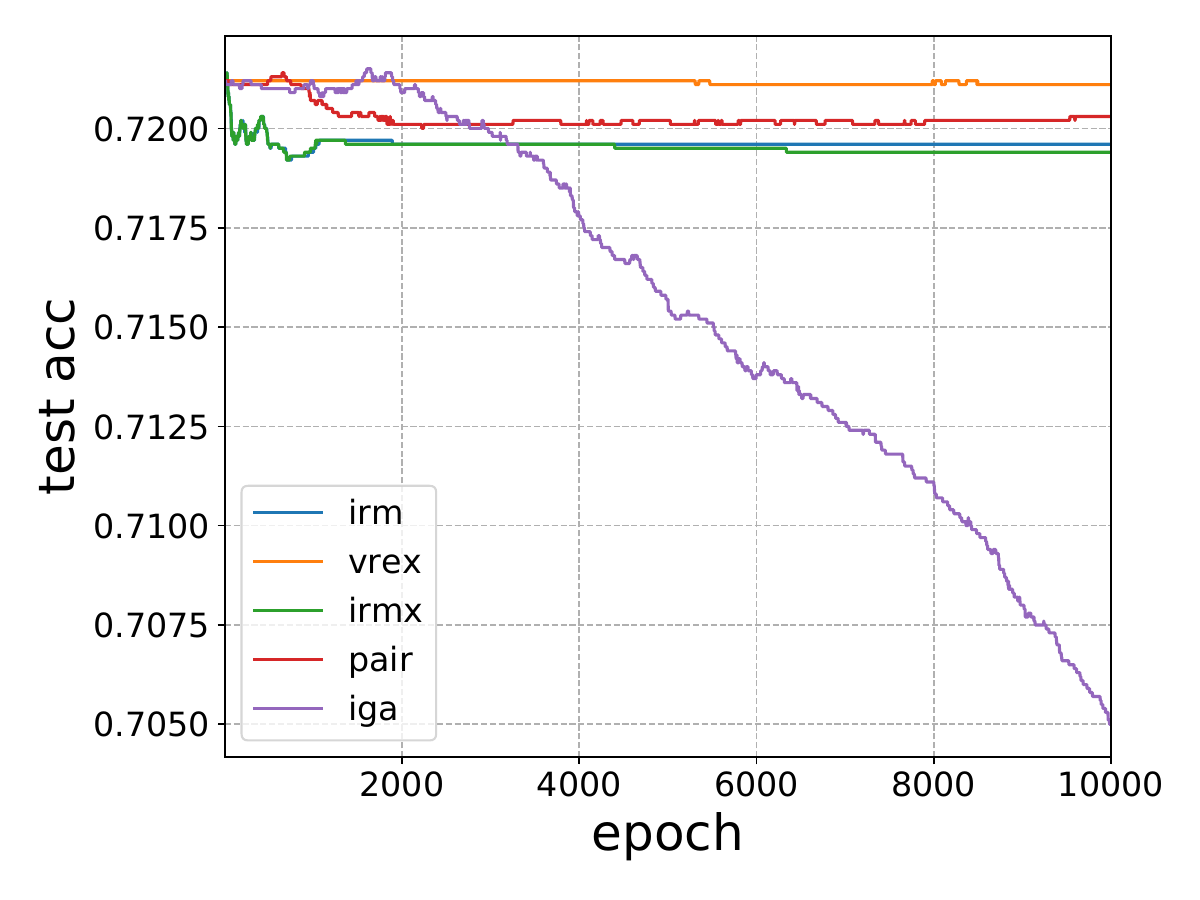}
		\label{CH:PAIR:fig:c11_perfect_init_appdx}
	}
	\hfil
	\subfigure[``Perfect'' init. on CMNIST-25.]{
		\centering
		\includegraphics[width=0.3\textwidth]{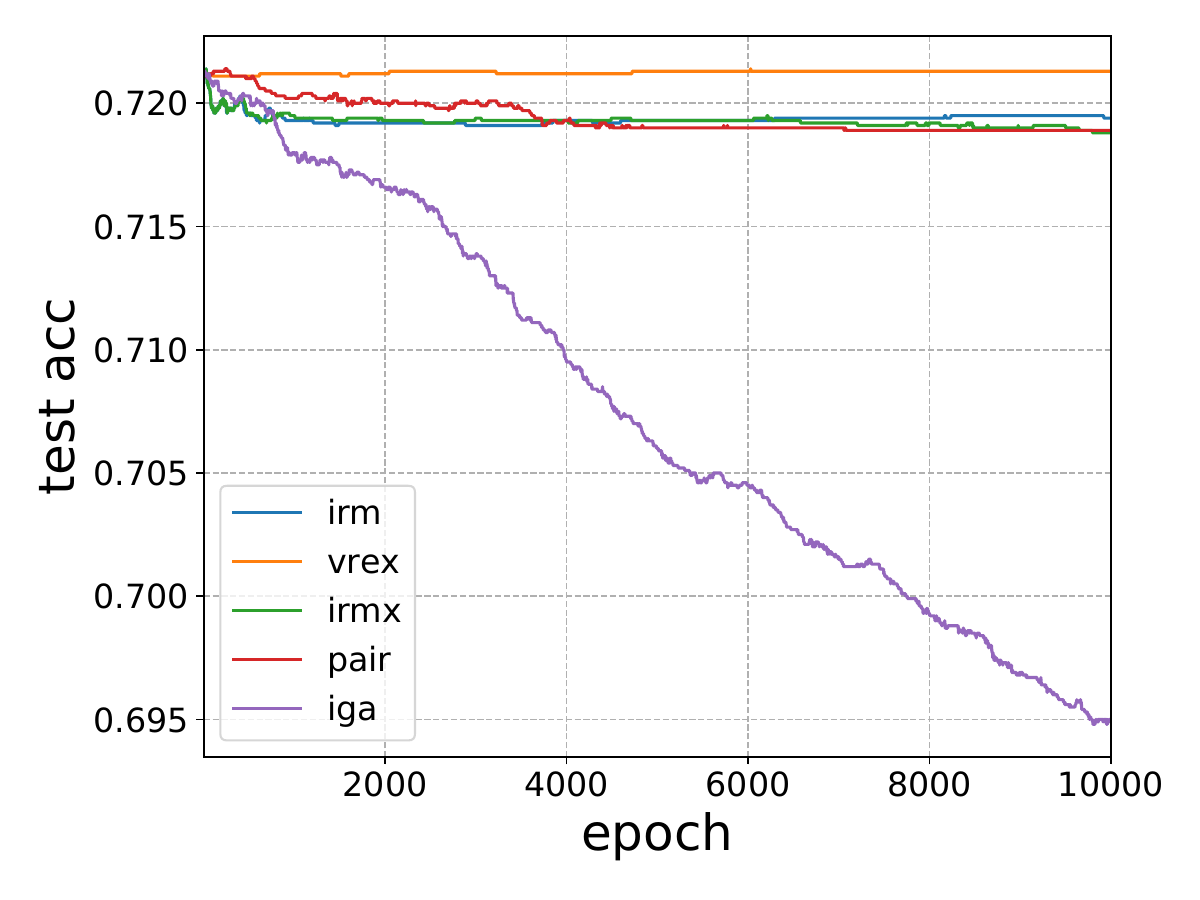}
		\label{CH:PAIR:fig:c25_perfect_init_appdx}
	}
	\caption{OOD performances with ``Perfect'' initializations.}
	\label{CH:PAIR:fig:perfect_init_appdx}
\end{figure}

\textbf{Additional ablation study on \cmnist with ``perfect'' initialization.}
We also conduct experiments with ``perfect'' initializations for different methods, to check whether the OOD constraints can enforce the invariance, following~\citet{rfc}. Besides the OOD methods used in the paper, we also include another OOD method IGA~\citep{iga} to give a more comprehensive overview of their performances with ``perfect'' initialization. We also introduce another variant of ColoredMNIST, i.e., \textbf{CMNIST-11}: $\{(0.25,0.10),(0.25,0.20)\}$ to complement more details.
All methods are initialized with a ERM model learned on gray-scale ColoredMNIST data which is expected to learn to use digit shapes in the image to make predictions.
The learning rate is $1e-3$ and the penalty weight is $1e5$.
Different from~\citet{rfc}, we use SGD to optimize the models, as Adam would generate larger step sizes when the gradients continue to be within a small range under the ``perfect'' initialization.
Results are shown as in Fig.~\ref{CH:PAIR:fig:perfect_init_appdx}.

It can be found that, in CMNIST-10, IRM, IRMx and IGA cannot enforce the invariance while V-REx and \pair maintain the invariance, which is consistent to our previous findings. Moreover, IGA fails to maintain the invariance in CMNIST-11 and CMNIST-25, demonstrating the relatively low robustness of IGA objective.
Besides, V-REx consistently maintain the invariance even in CMNIST-11, for the reason that the gradient signals of variance in ``perfect'' initialization tend to vanish. In contrast, \pair improve over both IRM and IRMx to maintain the invariance, confirming the effectiveness of \pair.

\textbf{Additional ablation study on the performance of \pairo and \pairs with more OOD objectives and their composite with \irml.}
Besides \vrex, we conduct additional ablation studies of \pair with IB~\citep{ib-irm}, Fishr~\citep{fishr}, CLOvE~\citep{clove}, IGA~\citep{iga} and SD~\citep{sd}, based on \cmnist and the modified \cmnist. We focus on the cases with no less than $2$ OOD objectives, as one could simply obtain a low OOD loss for single OOD objective, where linear weighting scheme is likely to approach the desired OOD solution as the Pareto front is simpler. However, it is often the case that a single OOD objective is not sufficiently robust to locate the desired OOD solution to the Pareto front.

In experiments, we follow the same evaluation protocol as previous experiments on \cmnist.
Due to the resource limits of NVIDIA RTX 3090Ti used for the original \cmnist experiments in previous sections, we switch the hardware and software platform to Linux servers with NVIDIA V100 graphics cards with CUDA 10.2, hence the results in Table~\ref{CH:PAIR:tab:pirm_appdx} and Table~\ref{CH:PAIR:tab:compirm_appdx} are not directly comparable with those in Table~\ref{CH:PAIR:tab:cmnist}.
Similar to previous experiments, for the stability of MOO solver under heterogeneous objectives, we search learning rate for \vrex and Fishr from $\{0.01,0.02,0.04,0.1,0.2\}$ at stage 2 while a larger scope $\{0.1,0.2,0.4,0.8,1\}$ for other objectives. Note that even considering the learning rate into the hyperparameter search space, \pair still uses a smaller  scope than that of linear weighting scheme.
Besides, we follow our previous discussion in Appendix~\ref{CH:PAIR:sec:pair_discussion_pc_appdx} to set up the preference of different OOD objectives. Specifically, for Fishr, we use a larger preference of $1e12$ than that for \irml ($1e8$), since the agreements based methods tend to have a smaller loss than \irml. While for the other objectives, we use a smaller preference of $1e8$ than that for \irml ($1e12$). Note that this is only a heuristic setup and the performance of \pair can be further improved if the preferences can be tuned.

\begin{table}[ht]
	\small\centering
	\caption{Generality study of \pair for \irml with other objectives in \cmnist.}
	\label{CH:PAIR:tab:pirm_appdx}
	\begin{tabular}{lccccccc}
		\toprule
		                       & \textbf{\scriptsize\irml} & \textbf{\scriptsize\pairo} & \textbf{\scriptsize\pairs} & CMNIST                        & CMNIST-M                      & Avg.                & $\Delta$Avg. \\\midrule
		ERM                    &                           &                            &                            & $17.14$\std{0.73}             & $73.30$\std{0.85}             & $45.22$             &              \\
		IRMv1                  &                           &                            &                            & $67.29$\std{0.99}             & $76.89$\std{3.23}             & $72.09$             & $+0.00$      \\\hdashline[0.5pt/1pt]
		\rule{0pt}{10pt}IB     &                           &                            &                            & $55.48$\std{3.67}             & $76.01$\std{0.58}             & $65.75$             &              \\
		                       & $\checkmark$              &                            &                            & $56.09$\std{2.04}             & $75.66$\std{10.6}             & $65.88$             & $-6.21$      \\
		                       & $\checkmark$              & $\checkmark$               &                            & $61.12$\std{2.33}             & $83.30$\std{3.00}             & $72.21$             & $+0.12$      \\
		                       & $\checkmark$              & $\checkmark$               & $\checkmark$               & $60.69$\std{2.26}             & $83.70$\std{1.79}             & $72.20$             & $+0.11$      \\\hdashline[0.5pt/1pt]
		\rule{0pt}{10pt}VREx   &                           &                            &                            & $68.62$\std{0.73}             & $83.52$\std{2.52}             & $76.07$             &              \\
		                       & $\checkmark$              &                            &                            & $66.19$\std{1.41}             & $81.75$\std{1.68}             & $73.97$             & $+1.88$      \\
		                       & $\checkmark$              & $\checkmark$               &                            & $68.89$\std{1.13}             & $\underline{83.80}$\std{1.60} & $\underline{76.35}$ & $+4.26$      \\
		                       & $\checkmark$              & $\checkmark$               & $\checkmark$               & $\underline{69.16}$\std{0.76} & $\mathbf{83.96}$\std{1.65}    & $\mathbf{76.56}$    & $+4.47$      \\\hdashline[0.5pt/1pt]
		\rule{0pt}{10pt}Fishr  &                           &                            &                            & $\mathbf{69.38}$\std{0.39}    & $77.29$\std{1.61}             & $73.34$             &              \\
		                       & $\checkmark$              &                            &                            & $66.20$\std{2.31}             & $81.07$\std{3.98}             & $73.63$             & $+1.54$      \\
		                       & $\checkmark$              & $\checkmark$               &                            & $68.90$\std{0.56}             & $82.70$\std{1.09}             & $75.80$             & $+2.49$      \\
		                       & $\checkmark$              & $\checkmark$               & $\checkmark$               & $68.78$\std{0.78}             & $84.02$\std{1.37}             & $76.40$             & $+3.31$      \\\hdashline[0.5pt/1pt]
		\rule{0pt}{10pt}CLOvE  &                           &                            &                            & $55.55$\std{9.97}             & $74.20$\std{2.45}             & $64.88$             &              \\
		                       & $\checkmark$              &                            &                            & $66.35$\std{1.51}             & $77.70$\std{1.00}             & $72.02$             & $-0.07$      \\
		                       & $\checkmark$              & $\checkmark$               &                            & $64.99$\std{2.29}             & $75.70$\std{1.05}             & $70.35$             & $-1.75$      \\
		                       & $\checkmark$              & $\checkmark$               & $\checkmark$               & $65.55$\std{2.17}             & $77.29$\std{1.55}             & $71.42$             & $-0.67$      \\\hdashline[0.5pt/1pt]
		\rule{0pt}{10pt}IGA    &                           &                            &                            & $58.67$\std{7.69}             & $76.27$\std{1.01}             & $68.97$             &              \\
		                       & $\checkmark$              &                            &                            & $51.22$\std{3.67}             & $74.20$\std{2.45}             & $62.71$             & $-9.38$      \\
		                       & $\checkmark$              & $\checkmark$               &                            & $66.17$\std{2.34}             & $81.84$\std{3.09}             & $74.01$             & $+1.91$      \\
		                       & $\checkmark$              & $\checkmark$               & $\checkmark$               & $66.51$\std{0.78}             & $82.12$\std{3.04}             & $74.32$             & $+2.23$      \\\hdashline[0.5pt/1pt]
		\rule{0pt}{10pt}SD     &                           &                            &                            & $62.31$\std{1.54}             & $76.73$\std{0.90}             & $69.52$             &              \\
		                       & $\checkmark$              &                            &                            & $62.48$\std{1.25}             & $81.24$\std{0.69}             & $71.86$             & $-0.23$      \\
		                       & $\checkmark$              & $\checkmark$               &                            & $59.52$\std{6.12}             & $82.82$\std{0.64}             & $71.17$             & $-0.92$      \\
		                       & $\checkmark$              & $\checkmark$               & $\checkmark$               & $65.54$\std{0.91}             & $83.57$\std{0.81}             & $74.56$             & $+2.47$      \\\hdashline[0.5pt/1pt]
		\rule{0pt}{10pt}Oracle &                           &                            &                            & $72.08$\std{0.24}             & $86.53$\std{0.14}             & $79.31$             & $79.31$      \\
		\bottomrule
	\end{tabular}%
\end{table}

The results are given in Table.~\ref{CH:PAIR:tab:pirm_appdx}. It can be found that, not all OOD objectives can improve \irml performance. For the OOD objectives that can enhance the OOD robustness when incorporated into \irml, \pair can further improve over the combined OOD objectives optimized via linear weighting scheme. While for unrobust combinations, intuitively it is hard to improve the OOD performance for the following reasons:
\begin{enumerate}[label=(\roman*).,wide]
	\item When the new objective combination is unrobust, the desired solution may not lie in the new Pareto optimal front;
	\item Eventhough the desired solution lies in the new Pareto optimal front, the weakened OOD robustness introduces more local minimals that have low OOD losses while worse OOD generalization performance;
	\item As an extra objective is involved, the OOD preference used in \pair tends to have a higher divergence from the ideal one;
\end{enumerate}
Therefore, given unrobust OOD objective combinations, the performance gain of \pair is not theoretically guaranteed. Nevertheless, \pairo can still improve some of the unrobust objective combinations, demonstrating its robustness. Notably, \pairs can further improve the performance of \pairo in most cases, demonstrating the generality of \pair.

To study what OOD objectives are suitable to be combined with \irml and whether using more OOD objectives can bring more performance improvements, additionally, we conduct experiments with all possible composites of \irml and IB~\citep{ib-irm}, Fishr~\citep{fishr} and \vrex~\citep{vrex}. In experiments, similar as in previous study, \pairo adopts a slightly broader learning rate search scope of $\{0.01,0.02,0.04,0.1,0.2\}$ at stage 2, in order to prevent divergence. Note that even considering the learning rate into the hyperparameter search space, \pair still uses a smaller search scope than that of linear weighting scheme. \pairs adopts the training domain validation accuracy to perform the model selection. Both \pairo and \pairs adopts a heuristic preference setup that uses a decreasing preference from $1e12$ to $1e8$ by a step size of $1e2$ for more objectives. For example, in the composite of IB, \irml and \vrex, we adopt the preference of $(1e8,1e10,1e12)$ for the OOD objectives. The choice of preference follows previous discussion in Appendix~\ref{CH:PAIR:sec:pair_discussion_pc_appdx}.

\begin{table}[ht]
	\small\centering
	\caption{Generality study of \pair for composite objectives in \cmnist.}
	\label{CH:PAIR:tab:compirm_appdx}

	\begin{tabular}{lccccccc}
		\toprule
		                                & IB           & VREx         & Fishr        & CMNIST                        & CMNIST-M                      & Avg.                & $\Delta$ Avg.       \\\midrule
		ERM                             &              &              &              & $17.14$\std{0.73}             & $73.30$\std{0.85}             & $45.22$             &                     \\
		IRMv1                           &              &              &              & $67.29$\std{0.99}             & $76.89$\std{3.23}             & $72.09$             & $+0.00$             \\\hdashline[0.5pt/1pt]
		\rule{0pt}{10pt}\texttt{Linear} & $\checkmark$ &              &              & $56.09$\std{2.04}             & $75.66$\std{10.6}             & $65.88$             & $-6.21$             \\
		+\pairo                         & $\checkmark$ &              &              & $61.12$\std{2.33}             & $83.30$\std{3.00}             & $72.21$             & $+0.12$             \\
		+\pairo+\pairs                  & $\checkmark$ &              &              & $60.69$\std{2.26}             & $83.70$\std{1.79}             & $72.20$             & $+0.11$             \\\hdashline[0.5pt/1pt]
		\rule{0pt}{10pt}\texttt{Linear} &              & $\checkmark$ &              & $66.19$\std{1.41}             & $81.75$\std{1.68}             & $73.97$             & $+1.88$             \\
		+\pairo                         &              & $\checkmark$ &              & $68.89$\std{1.13}             & $83.80$\std{1.60}             & $76.35$             & $+4.26$             \\
		+\pairo+\pairs                  &              & $\checkmark$ &              & $\mathbf{69.16}$\std{0.76}    & $\underline{83.96}$\std{1.65} & $\underline{76.56}$ & $\underline{+4.47}$ \\\hdashline[0.5pt/1pt]
		\rule{0pt}{10pt}\texttt{Linear} &              &              & $\checkmark$ & $66.20$\std{2.31}             & $81.07$\std{3.98}             & $73.63$             & $+1.54$             \\
		+\pairo                         &              &              & $\checkmark$ & $66.45$\std{0.90}             & $82.70$\std{1.09}             & $74.58$             & $+2.49$             \\
		+\pairo+\pairs                  &              &              & $\checkmark$ & $67.57$\std{0.81}             & $83.22$\std{2.10}             & $75.40$             & $+3.31$             \\\hdashline[0.5pt/1pt]
		\rule{0pt}{10pt}\texttt{Linear} & $\checkmark$ & $\checkmark$ &              & $52.61$\std{1.56}             & $63.84$\std{1.08}             & $58.23$             & $-13.9$             \\
		+\pairo                         & $\checkmark$ & $\checkmark$ &              & $68.35$\std{1.73}             & $81.25$\std{3.08}             & $74.80$             & $+2.71$             \\
		+\pairo+\pairs                  & $\checkmark$ & $\checkmark$ &              & $\underline{69.05}$\std{0.76} & $83.11$\std{1.46}             & $76.08$             & $+3.99$             \\\hdashline[0.5pt/1pt]
		\rule{0pt}{10pt}\texttt{Linear} & $\checkmark$ &              & $\checkmark$ & $51.91$\std{1.26}             & $68.88$\std{3.22}             & $60.39$             & $-11.7$             \\
		+\pairo                         & $\checkmark$ &              & $\checkmark$ & $59.70$\std{12.7}             & $74.59$\std{1.11}             & $67.15$             & $-4.94$             \\
		+\pairo+\pairs                  & $\checkmark$ &              & $\checkmark$ & $66.98$\std{2.66}             & $75.91$\std{3.50}             & $71.45$             & $-0.65$             \\\hdashline[0.5pt/1pt]
		\rule{0pt}{10pt}\texttt{Linear} &              & $\checkmark$ & $\checkmark$ & $64.83$\std{2.95}             & $79.34$\std{5.77}             & $72.09$             & $+0.00$             \\
		+\pairo                         &              & $\checkmark$ & $\checkmark$ & $67.96$\std{1.60}             & $81.44$\std{2.24}             & $74.70$             & $+2.61$             \\
		+\pairo+\pairs                  &              & $\checkmark$ & $\checkmark$ & $68.19$\std{1.58}             & $81.89$\std{3.01}             & $75.04$             & $+2.95$             \\\hdashline[0.5pt/1pt]
		\rule{0pt}{10pt}\texttt{Linear} & $\checkmark$ & $\checkmark$ & $\checkmark$ & $50.00$\std{0.32}             & $69.60$\std{2.33}             & $59.80$             & $-12.3$             \\
		+\pairo                         & $\checkmark$ & $\checkmark$ & $\checkmark$ & $66.89$\std{1.80}             & $83.46$\std{3.10}             & $75.18$             & $+3.08$             \\
		+\pairo+\pairs                  & $\checkmark$ & $\checkmark$ & $\checkmark$ & $68.59$\std{1.29}             & $\mathbf{85.30}$\std{0.64}    & $\mathbf{76.95}$    & $\mathbf{+4.85}$    \\\hdashline[0.5pt/1pt]
		\rule{0pt}{10pt}\texttt{Oracle} &              &              &              & $72.08$\std{0.24}             & $86.53$\std{0.14}             & $79.31$             &                     \\
		\bottomrule
	\end{tabular}%
\end{table}
The results are shown in Table~\ref{CH:PAIR:tab:compirm_appdx}. The best and second best results are in bold and underlined, respectively. It can be found that incorporating more OOD objectives does not necessarily bring more performance improvements into \irml. The linear weighting scheme can further exacerbate the performance drops of unrobust OOD objective combinations. For example, when incorporating IB objective into \irml, the OOD performance drops, since IB is proposed to mitigate a specific type of distribution shifts instead of directly improving learning the invariance in the original \irml setting. In contrast, it can be found that incorporating Fishr can bring performance increases in most cases. The reason is that minimizing Fishr loss can approximately minimize the \vrex loss, as shown by~\citet{fishr}. Therefore, we suspect that the reason for the performance drop could be that more objectives will make the Pareto front more complicated, and also lead to higher divergence of the OOD preference (since we are less likely to know the ideal preference given more objectives). Hence, the preferred composition of the objectives is preferred to those that have theoretical guarantees and are as concise as possible.

Interestingly, we also find that, although incorporating more objectives in \pairo does not necessarily bring performance increase, a combination of \pairo and \pairs can further improve the OOD performance, despite of the simple implementation of \pairo. It serves as strong evidence for the generality and significance of \pair.

\subsection{More details about experiments on \wilds}
\label{CH:PAIR:sec:wilds_appdx}

In this section, we provide more details about the \wilds datasets as well as the evaluation setups in the experiments.

\subsubsection{Dataset description.}
We select $6$ challenging datasets from \wilds~\citep{wilds} benchmark for evaluating \pairo performance in realistic distribution shifts. The datasets cover from domain distribution shifts, subpopulation shifts and the their mixed.
A summary of the basic information and statistics of the \textsc{Wilds} datasets can be found in Table.~\ref{CH:PAIR:tab:wilds_summary_appdx}, Table.~\ref{CH:PAIR:tab:wilds_stat_appdx}, respectively.
In the following, we will give a brief introduction to each of the datasets. More details can be found in the \wilds paper~\citep{wilds}.
\begin{table}[ht!]
	\centering
	\caption{A summary of datasets information from \wilds.}\label{CH:PAIR:tab:wilds_summary_appdx}
	\scalebox{0.65}{
		\begin{tabular}{lllllc}
			\toprule
			\textbf{Dataset}       & \textbf{Data ($x$)} & \textbf{Class information}         & \textbf{Domains}        & \textbf{Metric}       & \textbf{Architecture} \\
			\midrule
			\textsc{Camelyon17}    & Tissue slides       & Tumor (2 classes)                  & 5 hospitals             & Avg. acc.             & DenseNet-121          \\
			\textsc{CivilComments} & Online comments     & Toxicity (2 classes)               & 8 demographic groups    & Wr. group acc.        & DistillBERT           \\
			\textsc{FMoW}          & Satellite images    & Land use (62 classes)              & 16 years x 5 regions    & Wr. group acc.        & DenseNet-121          \\
			\textsc{iWildCam}      & Photos              & Animal species (186 classes)       & 324 locations           & Macro F1              & ResNet-50             \\
			\textsc{Poverty}       & Satellite images    & Asset (real valued)                & 23 countries            & Wr. group Pearson (r) & Resnet-18             \\
			\textsc{RxRx1}         & Cell images         & Genetic treatments (1,139 classes) & 51 experimental batches & Avg. acc              & ResNet-50             \\
			\bottomrule
		\end{tabular}
	}
\end{table}

\begin{table}[ht!]
	\centering
	\caption{A summary of datasets statistics from \wilds.}\label{CH:PAIR:tab:wilds_stat_appdx}
	\scalebox{0.8}{
		\small
		\begin{tabular}{lcccccccc}
			\toprule
			\multirow{2}{*}{Dataset} & \multicolumn{3}{c}{$\#$ Examples} &        & \multicolumn{3}{c}{$\#$ Domains}                         \\
			\cmidrule{2-4}\cmidrule{6-8}
			                         & train                             & val    & test                             &  & train & val & test \\
			\midrule
			\textsc{Camelyon17}      & 302,436                           & 34,904 & 85,054                           &  & 3     & 1   & 1    \\
			\textsc{CivilComments}   & 269,038                           & 45,180 & 133,782                          &  & -     & -   & -    \\
			\textsc{FMoW}            & 76,863                            & 19,915 & 22,108                           &  & 11    & 3   & 2    \\
			\textsc{iWildCam}        & 129,809                           & 14,961 & 42,791                           &  & 243   & 32  & 48   \\
			\textsc{Poverty}         & 10,000                            & 4,000  & 4,000                            &  & 13-14 & 4-5 & 4-5  \\
			\textsc{RxRx1}           & 40,612                            & 9,854  & 34,432                           &  & 33    & 4   & 14   \\
			\bottomrule
		\end{tabular}}
\end{table}

\textbf{Camelyon17.}
We follow the \wilds splits and data processing pipeline for the Camelyon17 dataset~\citep{camelyon}. It provides $450,000$ lymph-node scans from $5$ hospitals. The task in Camelyon17 is to take the input of $96\times96$ medical images to predict whether there exists a tumor tissue in the image. The domains $d$ refers to the index of the hospital where the image was taken. The training data are sampled from the first $3$ hospitals where the OOD validation and test data are sampled from the $4$-th and $5$-th hospital, respectively.
We will use the average accuracy as the evaluation metric and a DenseNet-121~\citep{densenet} as the backbone for the featurizer.

\textbf{CivilComments.}
We follow the \wilds splits and data processing pipeline for the CivilComments dataset~\citep{civil}. It provides $450,000$ comments collected from online articles. The task is to classify whether an online comment text is toxic or non-toxic. The domains $d$ are defined according to the demographic features, including male, female, LGBTQ, Christian, Muslim, other religions, Black, and White. CivilComments is used to study the subpopulation shifts, here we will use the worst group/domain accuracy as the evaluation metric. As for the backbone of the featurizer, we will use a DistillBert~\citep{distillbert} following \wilds~\citep{wilds}.

\textbf{FMoW.}
We follow the \wilds splits and data processing pipeline for the FMoW dataset~\citep{fmow}. It provides satellite images from $16$ years and $5$ regions. The task in FMoW is to classify the images into $62$ classes of building or land use categories. The domain is split according to the year that the satellite image was collected, as well as the regions in the image which could be Africa, America, Asia, Europe or Oceania. Distribution shifts could happen across different years and regions.
The training data contains data collected before $2013$, while the validation data contains images collected within $2013$ to $2015$, and the test data contains images collected after $2015$. The evaluation metric for FMoW is the worst region accuracy and the backbone model for the featurizer is a DenseNet-121~\citep{densenet}.

\textbf{iWildCam.}
We follow the \wilds splits and data processing pipeline for the iWildCam dataset~\citep{iwildcam}. It is consist of  $203,029$ heat or motion-activated photos of animal specifies from 323 different camera traps across different countries around the world. The task of iWildCam is to classify the corresponding animal specifies in the photos. The domains is split according to the locations of the camera traps which could introduce the distribution shifts. We will use the Macro F1 as the evaluation metric and a ResNet-50~\citep{resnet} as the backbone for the featurizer.

\textbf{PovertyMap.}
We follow the \wilds splits and data processing pipeline for the PovertyMap dataset~\citep{povertymap}. It consists of satellite imagery and survey data at $19,669$ villages from $23$ African countries between $2009$ and $2016$.
Different from other datasets, the task in PovertyMap is a regression task that asks the model to predict the real-valued asset wealth index computed from Demographic and Health Surveys (DHS) data. The domain is split according to the countries that the image was taken and whether the image is of an urban or rural area.
The relative small size of PoverMap allows for using cross-fold evaluation, where each fold defines a different set of OOD countries~\citep{wilds}.
We will use the Pearson correlation of the worst urban/rural subpopulation as the evaluation metric and a ResNet-18~\citep{resnet} as the featurizer.

\textbf{RxRx1.}
We follow the \wilds splits and data processing pipeline for the RxRx1 dataset~\citep{rxrx1}.
The input is an image of cells taken by fluorescent microscopy. The cells can be genetically perturbed by siRNA and the task of RxRx1 is to predict the class of the corresponding siRNA that have treated the cells.
There exists $1,139$ genetic treatments and the domain shifts are introduced by the experimental batches. We will use the average accuracy of the OOD experimental batches as the evaluation metric and a ResNet-50~\citep{resnet} as the backbone for the featurizer.

\subsubsection{Training and evaluation details.}
We follow previous works to implement and evaluate our models~\citep{wilds,fish,lisa}. The information of the referred paper and code is listed as in Table.~\ref{CH:PAIR:tab:referred_code_appdx}.

\begin{table}[ht]
	\centering
	\small
	\caption{The information of the referred paper and code in the experiments of \pair.}
	\label{CH:PAIR:tab:referred_code_appdx}
	\resizebox{\textwidth}{!}{
		\begin{tabular}{l|cc}
			\toprule
			Paper               & Commit                                            & Code                                    \\
			\midrule
			\wilds\citep{wilds} & v2.0.0                                            & \url{https://wilds.stanford.edu/}       \\
			Fish~\citep{fish}   & \texttt{333efa24572d99da0a4107ab9cc4af93a915d2a9} & \url{https://github.com/YugeTen/fish}   \\
			LISA~\citep{lisa}   & \texttt{bc424c47df6f072986b63cd906c44975bd34d9ff} & \url{https://github.com/huaxiuyao/LISA} \\
			\bottomrule
		\end{tabular}}
\end{table}

The general hyperparemter setting inherit from the referred codes and papers, and are shown as in Table~\ref{CH:PAIR:tab:hyper_wilds_appdx}.
We use the same backbone models to implement the featurizer~\citep{resnet,densenet,distillbert}.
By default, we repeat the experiments by $3$ runs with the random seeds of $0,1,2$. While for Camelyon17, we follow the official guide to repeat $10$ times with the random seeds from $0$ to $9$, and for PovertyMap, we repeat the experiments $5$ times with the random seeds from $0$ to $4$. Note that the PovertyMap use cross-fold validations hence each run will use different training and evaluation splits, following the \wilds official guideline.

For the evaluation of baselines, we refer the previous results from the literature~\citep{wilds,fish,lisa} by default, while we rerun Fish~\citep{fish} and LISA~\citep{lisa} to validate the reported results.
Since the original implementation of Fish does not support the evaluation of the updated PovertyMap dataset, we mildly adjust the hyperparameter settings to reproduce the corresponding results as shown in Table.~\ref{CH:PAIR:tab:hyper_wilds_appdx}.
We also reduce the batch size on FMoW due to the memory limits and we find it does not affect the reproducibility of Fish and LISA.
Besides, since the original implementation of LISA does not support PovertyMap, which differentiates as a regression task that could be not suitable with Mixup~\citep{mixup}, however we find the ``group by label'' strategy in LISA works particularly well and reaches to the state of the art.
For \irmx, we implement it as the simple addition of \irml and \vrex penalties based on the Fish implementation~\citep{fish}, and search the penalty weights using the same space as for other objectives~\citep{wilds} to ensure the fairness. Besides, since previously reported results did not cover the performance of \vrex in iWildCam and PovertyMap, we implement \vrex and report the results based on the Fish implementation~\citep{fish}.

\begin{table}[ht]
	\centering
	\small
	\caption{General hyperparameter settings for the experiments with \pair on \wilds.}
	\label{CH:PAIR:tab:hyper_wilds_appdx}
	\resizebox{\textwidth}{!}{
		\begin{tabular}{l|cccccc}
			\toprule
			Dataset        & \sc Camelyon17 & \sc  CivilComments & \sc FMoW    & \sc iWildCam & \sc PovertyMap & \sc RxRx1     \\
			\midrule
			Num. of seeds  & 10             & 3                  & 3           & 3            & 5              & 3             \\
			Learning rate  & 1e-4           & 2e-6               & 1e-4        & 1e-4         & 1e-4           & 1e-3          \\
			Weight decay   & 0              & 0.01               & 0           & 0            & 0              & 1e-5          \\
			Scheduler      & n/a            & n/a                & n/a         & n/a          & n/a            & Cosine Warmup \\
			Batch size     & 32             & 16                 & 32          & 16           & 64             & 72            \\
			Architecture   & DenseNet121    & DistilBert         & DenseNet121 & ResNet50     & ResNet18       & ResNet50      \\
			Optimizer      & SGD            & Adam               & Adam        & Adam         & Adam           & Adam          \\
			Pretraing Step & 10000          & 20000              & 24000       & 24000        & 5000           & 15000         \\
			Maximum Epoch  & 2              & 5                  & 12          & 9            & 200            & 90            \\
			\bottomrule
		\end{tabular}}
\end{table}

For \pairo, we implement it based on the Fish code~\citep{fish}. The detailed algorithm can be found in Algorithm.~\ref{alg:pair_opt_appdx}.
We leverage the same number of pretraining steps as in Fish to fulfill the first ``descent'' phase in \pairo algorithm.
Then, during the ``balance'' phase, at each training step, we sampled $k$ batches of data from different domains, calculate loss and conduct the back-propagation.
By default, we use only the gradients of the classifier to solve for the objective weights during the ``balance'' phase.
Except for iWildCam and RxRx1 datasets, due the memory limits, as discussed in Sec.~\ref{CH:PAIR:sec:pair_opt_sca_appdx}, we use the freeze technique to ensure the consistency of batch size and number of sampled domains as in Table.~\ref{CH:PAIR:tab:hyper_wilds_appdx}.
Moreover, as discussed in Sec.~\ref{CH:PAIR:sec:pair_opt_est_appdx}, the unbiased stochastic estimate of \irml penalties can not guarantee the non-negativity of the estimated loss values, which are however not compatible with MOO theory~\citep{moo_book} (thus the same for \pairo). Therefore, we will manually adjust the negative values to be positive, by multiplying it with a adjustment rate (short in Neg. \irml adj. rate in Table.~\ref{CH:PAIR:tab:hyper_pair_wilds_appdx}). The adjustment rate is tuned from $1$ to $1e-4$ with a step size of $1e-2$ to avoid the training divergence and instability.
Following the discussion as in Sec.~\ref{CH:PAIR:sec:pair_discussion_pc_appdx}, we tune the OOD relative preference by merely varying the preference for \irml objective from the default choice of $(1,1e10,1e12)$ by a step size of $1e2$. We find the performances of \irml and \vrex highly correlate to the corresponding relative preference weights.
We list hyperparameters of \pairo in Table~\ref{CH:PAIR:tab:hyper_pair_wilds_appdx}. Although we did not tune the hyperparameters heavily, we find that \pairo generically works well across different challenging datasets and realistic distribution shifts on \wilds.
As discussed in Sec.~\ref{CH:PAIR:sec:pair_discussion_pc_appdx}, there could be more sophisticated approaches to further improve the search and estimate of OOD preference, which we will leave for future developments based on \pair.

\begin{table}[ht]
	\centering
	\small
	\caption{Hyperparameter settings of \pairo for the experiments on \wilds.}
	\label{CH:PAIR:tab:hyper_pair_wilds_appdx}
	\resizebox{\textwidth}{!}{
		\begin{tabular}{l|cccccc}
			\toprule
			Dataset              & \sc Camelyon17 & \sc  CivilComments            & \sc FMoW               & \sc iWildCam   & \sc PovertyMap & \sc RxRx1            \\
			\midrule
			Gradients from       & Classifier     & Classifier                    & Classifier             & Classifier     & Classifier     & Classifier           \\
			Freeze featurizer    & No             & No                            & No                     & Yes            & No             & Yes                  \\
			Relative Preference  & (1,1e12,1e12)  & (1,1e8,1e12)                  & (1,1e12,1e12)          & (1,1e10,1e12)  & (1,1e8,1e12)   & (1,1e8,1e12)         \\
			Neg. \irml adj. rate & 1              & 1e-4                          & 1                      & 1e-2           & 1e-2           & 1                    \\
			Group by             & Hospitals      & Demographics$\times$ toxicity & Times $\times$ regions & Trap locations & Countries      & Experimental batches \\
			Sampled domains      & 3              & 5                             & 5                      & 10             & 5              & 10                   \\
			\bottomrule
		\end{tabular}}
\end{table}

\subsection{Software and hardware}
\label{CH:PAIR:sec:exp_software_appdx}
We implement our methods with PyTorch~\citep{pytorch}.
For the software and hardware configurations, we ensure the consistent environments for each datasets. Specifically, we run \cmnist experiments on Linux Servers with NVIDIA RTX 3090Ti graphics cards with CUDA 11.3, 40 cores Intel(R) Xeon(R) Silver 4114 CPU @ 2.20GHz, 256 GB Memory, and Ubuntu 18.04 LTS installed.
While for \wilds and \dobed experiments, we run on Linux servers with NVIDIA V100 graphics cards with CUDA 10.2.

\section{More Details of Model Selection Results on \dobed}
\label{CH:PAIR:sec:dobed_full_appdx}

\subsection{Introduction of difficult model selection in \dobed}
\label{CH:PAIR:sec:dobed_intro_appdx}

\dobed is proposed by~\citet{domainbed} to highlight the importance of model selection in OOD generalization. Specifically, they empirically show that, under rigorous hyperparameter tunning, ERM~\citep{erm} achieves the state-of-the-art performances. Although recently progress are made to outperform ERM under the rigorous \dobed evaluation protocol~\citep{fishr}, whether there exists a proper model selection for OOD algorithms remains elusive.

The difficulty of a proper model selection for OOD algorithms is mainly because of: We lack the access to a validation set that have a similar distribution with the test data. Therefore, \citet{domainbed} provide $3$ options to choose and construct a validation set from: training domain data; leave-one-out validation data; test domain data. However, all three validation set construction approaches have their own limitations, as they essentially posit different assumptions on the test distribution~\citep{domainbed,trainval_issue,fishr}.

\pairs tries to address the limitations caused by the difficulty of finding a proper validation set for model selection in domain generalization, by leveraging the \emph{prior} assumed within the OOD algorithm.
Essentially, different lines of OOD algorithms discussed in Sec.~\ref{CH:PAIR:sec:related_work_appdx} adopt different prior and assumptions on the causes of the distribution shifts. The main purpose of the OOD evaluation is to validate the correctness of the posed assumptions.
To this end, the selected models should properly reflect the preferences implied by the assumptions, i.e., the OOD loss values.
When considering the loss values during the model selection, it is natural to leverage the MOO perspective and explicitly consider the trade-offs between ERM and OOD performance.
The detailed description, implementation options, and potential leverages of \pairs are provided in Appendix~\ref{CH:PAIR:sec:pair_implementation_appdx}.

\subsection{Training and evaluation details}

Since our main purpose of the \dobed experiments is to validate the existence of the problem and the effectiveness of \pairs, we apply \pairs to the representative methods of the four discussed OOD solutions in Sec.~\ref{CH:PAIR:sec:related_work_appdx}.
Specifically, we choose the following four methods out of all implemented algorithms in \dobed (\url{https://github.com/facebookresearch/DomainBed}):

\begin{itemize}
	\item ERM: Empirical Risk Minimization \citep{erm}
	\item IRM: Invariant Risk Minimization \citep{irmv1}
	\item GroupDRO: Group Distributionally Robust Optimization \citep{groupdro}
	\item DANN: Domain Adversarial Neural Network \citep{DANN}
	\item Fishr: Invariant Gradient Variances for OOD Generalization \citep{fishr}
\end{itemize}

Due to the limits of computational resources, we select $3$ out of $7$ datasets from \dobed. We refer~\citet{fishr} to prescribe the detail, listed as follows:
\begin{enumerate}
	\item Colored MNIST \citep{irmv1} is a variant of the MNIST handwritten digit classification dataset \citep{mnist}. Domain $d\in\{90\%, 80\%, 10\%\}$ contains a disjoint set of digits colored: the correlation strengths between color and label vary across domains. The dataset contains 70,000 examples of dimension $(2,28,28)$ and 2 classes. Most importantly, the network, the hyperparameters, the image shapes, etc. are \textbf{not} the same as in the IRM setup for \cmnist experiments.
	\item PACS \citep{pacs} includes domains $d\in\{$art, cartoons, photos, sketches$\}$, with 9,991 examples of dimension $(3,224,224)$ and 7 classes.
	\item TerraIncognita \citep{camel_example} contains photographs of wild animals taken by camera traps at locations $d\in\{$L100, L38, L43, L46$\}$, with 24,788 examples of dimension $(3,224,224)$ and 10 classes.
\end{enumerate}
Note that CMNIST dataset in \dobed use a convolutional neural network as the featurizer, which is not the same MLP for \cmnist experiments.
By default, all real datasets leverage a ResNet-50~\citep{resnet} pretrained on ImageNet, with a dropout layer before the newly added dense layer and fine-tuned with frozen batch normalization layers.

During the training, we strictly follow the evaluation protocol in \dobed. Note that the hyperparameter configurations of Fishr have some differences from the default configurations hence we refer the configuration tables by~\citet{fishr} directly, shown as follows.

\begin{table}[ht]
	\caption[Hyperparameters used in the experiments of \pair.]{\textbf{Hyperparameters}, their default values and distributions for random search~\citep{domainbed,fishr}.}
	\centering
	\resizebox{\textwidth}{!}{%
		\begin{tabular}{llll}
			\toprule
			Condition              & Parameter                         & Default value & Random distribution                                                         \\
			\midrule
			\pacs /                & learning rate                     & $0.00005$     & $10^{\text{Uniform}(-5,-3.5)}$                                              \\
			\terra                 & batch size                        & 32            & $2^{\text{Uniform}(3,5.5)}$ if not DomainNet else $2^{\text{Uniform}(3,5)}$ \\
			                       & weight decay                      & 0             & $10^{\text{Uniform}(-6,-2)}$                                                \\
			                       & dropout                           & 0             & RandomChoice $([0,0.1,0.5])$                                                \\
			\midrule
			\cmnist                & learning rate                     & $0.001$       & $10^{\text{Uniform}(-4.5,-3.5)}$                                            \\
			                       & batch size                        & 64            & $2^{\text{Uniform}(3,9)}$                                                   \\
			                       & weight decay                      & 0             & 0                                                                           \\
			\midrule
			All                    & steps                             & 5000          & 5000                                                                        \\
			\midrule
			\midrule
			\multirow{3}{*}{Fishr} & regularization strength $\lambda$ & 1000          & $10^{\text{Uniform}(1,4)}$                                                  \\
			                       & ema $\gamma$                      & 0.95          & Uniform$(0.9,0.99)$                                                         \\
			                       & warmup iterations                 & 1500          & Uniform$(0,5000)$                                                           \\
			\bottomrule
		\end{tabular}}

	\label{CH:PAIR:tab:hyper_dobed_appdx}%
\end{table}

As for the construction of the validation set, we test with training domain validation set and test domain validation set, as leave-one-out domain selection requires more runs and more computational resources that are out of our limits.
Specifically, to construct the validation set, the data from each domain will be first splitted into $80\%$ (for training and evaluation) and $20\%$ (for validation and model selection).
For training domain validation set, the validation data is consist of the $20\%$ split from each training domain. While for the test domain validation set, the validation data is consist of the $20\%$ split from each test domain.

The whole evaluation will be repeated $3$ times where in each repeat, there will be $20$ samplings of hyperparameters from the distribution shown in Table~\ref{CH:PAIR:tab:hyper_dobed_appdx}. Therefore, there will be $20$ runs in each repeat and there will be $1$ model selected from the $20$ runs.

For the implementation of \pairs, we follow the algorithm as in Algorithm~\ref{alg:pair_sel_appdx}. Since training domain validation accuracy tends to be a more unreliable indicator than test domain validation accuracy, i.e., has a worse reflection of the OOD generalization performance due to the high similarity with the training data~\citep{trainval_issue}, during the selection within each run, we filter out the models before the last $5$ steps in \cmnist and the last $10$ steps in \pacs and \terra. During the selection within one repeat (across different runs), we use a percent of $50\%$ for step $9$ in Algorithm~\ref{alg:pair_sel_appdx} and finalize the selection according the \pair score. Except for GroupDRO and DANN of which the objective value tend to have higher variance and relatively low OOD robustness, we aggregate the models within each repeat by the validation accuracy.
In contrast, for the test domain validation accuracy, we filter out the models before the last $5$ steps for DANN while $10$ steps for others according to the robustness of the objectives during the selection within each run. During the selection within one repeat (across different runs), we directly adopt the validation accuracy to finalize the model selected. Note that \citet{domainbed} argue that test domain validation is more likely to be a invalid benchmarking methodology, since it requires access to the test domain which is usually inaccessible in realistic applications.

For the selection of loss values $\vL$, we use the values reported solely at each logging step, which is evaluated every $100$ steps with a minibtach of the training data, listed as follows:
\begin{itemize}
	\item ERM: N/A.
	\item IRM: \erm and \irml (\texttt{nll,penalty}).
	\item GroupDRO: Worst group \erm loss (\texttt{losses.min()}).
	\item DANN: Weighted \erm and domain discrimination loss (\texttt{gen\_loss}).
	\item Fishr: \erm and Fishr penalty (\texttt{nll,penalty}).
\end{itemize}

\subsection{Full \dobed Results}
\label{CH:PAIR:sec:dobed_full}
In this section, we provide full results of the \dobed experiments.
To begin with, we first present the overall results of the three datasets, including the averages and the improvements of the worst domain accuracies,
as in Table.~\ref{CH:PAIR:tab:dobed_select_train_appdx} and Table.~\ref{CH:PAIR:tab:dobed_select_test_appdx}.
From results we can seed that \pairs consistently improves the OOD performance across all datasets and validation set options.
Remarkably, in the most challenging setting that uses train domain validation set on \cmnist ,
\pairs improves the worst domain performances of \irml and Fishr by a large margin up tp $14.3\%$.
In the realistic dataset \pacs , \pairs improves the worst domain performances of \irml by a large margin up to $7.3\%$.
In \terra , \pairs improves the worst domain performances of DANN by a large margin up to $3.1\%$.
Besides the worst domain performance, \pairs improves the average domain performances up to $1.0\%$
and empower the OOD methods to reach new state-of-the-arts.

When using the test domain validation set, since the validation set itself could reflect the OOD generalization performance,
therefore the improvements could be lower. When comes to OOD objectives that have a relatively low robustness, the worst domain performance could be lower.

We also report the detailed results at each domain with the variance in the next section.

\subsubsection{Overall results}

\begin{table}[ht]
	\centering
	\caption{Overeall OOD generalization performances using training domain validation accuracy with \pairs.}
	\label{CH:PAIR:tab:dobed_select_train_appdx}
	\vskip 0.1in
	\resizebox{\textwidth}{!}{
		\small
		\begin{tabular}{@{}{l}*{8}{c}@{}}
			\toprule
			                         &                 & \multicolumn{2}{c}{{\small\cmnist}} & \multicolumn{2}{c}{{\small \pacs}} & \multicolumn{2}{c}{{\small \terra}} & \multicolumn{1}{c}{{Overall}}                                                                     \\\cmidrule(lr){3-4}\cmidrule(lr){5-6}\cmidrule(lr){7-8}
			                         & \textbf{\pairs} & \textbf{Avg. acc}                   & \textbf{$\Delta$ wr. acc}          & \textbf{Avg. acc}                   & \textbf{$\Delta$ wr. acc}     & \textbf{Avg. acc} & \textbf{$\Delta$ wr. acc} & \textbf{Avg. acc} \\ \midrule
			ERM                      &                 & 51.4 $\pm$ 1.0                      &                                    & 84.8 $\pm$ 0.3                      &                               & 44.6 $\pm$ 1.1    &                           & 60.2              \\  \hdashline[0.5pt/1pt]
			\rule{0pt}{10pt}DANN     &                 & 51.5 $\pm$ 0.1                      &                                    & 82.5 $\pm$ 0.8                      &                               & 44.9 $\pm$ 0.9    &                           & 59.6              \\
			DANN                     & $\checkmark$    & 51.9 $\pm$ 0.1                      & +0.9                               & 83.3 $\pm$ 0.5                      & +0.7                          & 44.5 $\pm$ 1.5    & +3.1                      & 59.9              \\\hdashline[0.5pt/1pt]
			\rule{0pt}{10pt}GroupDRO &                 & 51.8 $\pm$ 0.0                      &                                    & 84.1 $\pm$ 0.8                      &                               & 46.6 $\pm$ 1.1    &                           & 60.8              \\
			GroupDRO                 & $\checkmark$    & 53.0 $\pm$ 0.4                      & +3.1                               & 84.4 $\pm$ 0.7                      & +1.1                          & 46.6 $\pm$ 1.1    & +0.0                      & 61.3              \\\hdashline[0.5pt/1pt]
			\rule{0pt}{10pt}IRM      &                 & 51.6 $\pm$ 0.1                      &                                    & 83.5 $\pm$ 1.1                      &                               & 44.9 $\pm$ 0.3    &                           & 60.0              \\
			IRM                      & $\checkmark$    & 52.2 $\pm$ 0.5                      & +14.3                              & 85.1 $\pm$ 0.9                      & +7.3                          & 41.1 $\pm$ 3.8    & +1.4                      & 59.5              \\\hdashline[0.5pt/1pt]
			\rule{0pt}{10pt}Fishr    &                 & 51.8 $\pm$ 0.1                      &                                    & 85.6 $\pm$ 0.5                      &                               & 47.0 $\pm$ 1.4    &                           & 61.5              \\
			Fishr                    & $\checkmark$    & 54.2 $\pm$ 1.0                      & +12.7                              & 85.6 $\pm$ 0.1                      & +1.1                          & 47.7 $\pm$ 1.1    & +0.3                      & 62.5              \\
			\bottomrule                                                                                                                                                                                                                                                     \\
		\end{tabular}}
\end{table}

\begin{table}[ht]
	\centering
	\caption{Overeall OOD generalization performances using test domain validation accuracy with \pairs.}
	\label{CH:PAIR:tab:dobed_select_test_appdx}
	\vskip 0.1in
	\resizebox{\textwidth}{!}{
		\small
		\begin{tabular}{@{}{l}*{8}{c}@{}}
			\toprule
			                         &                 & \multicolumn{2}{c}{{\small\cmnist}} & \multicolumn{2}{c}{{\small \pacs}} & \multicolumn{2}{c}{{\small \terra}} & \multicolumn{1}{c}{{Overall}}                                                                     \\\cmidrule(lr){3-4}\cmidrule(lr){5-6}\cmidrule(lr){7-8}
			                         & \textbf{\pairs} & \textbf{Avg. acc}                   & \textbf{$\Delta$ wr. acc}          & \textbf{Avg. acc}                   & \textbf{$\Delta$ wr. acc}     & \textbf{Avg. acc} & \textbf{$\Delta$ wr. acc} & \textbf{Avg. acc} \\ \midrule
			ERM                      &                 & 57.8 $\pm$ 0.2                      &                                    & 87.0 $\pm$ 0.1                      &                               & 52.9 $\pm$ 0.9    &                           & 65.9              \\  \hdashline[0.5pt/1pt]
			\rule{0pt}{10pt}DANN     &                 & 57.4 $\pm$ 0.8                      &                                    & 84.7 $\pm$ 0.5                      &                               & 50.8 $\pm$ 0.3    &                           & 64.3              \\
			DANN                     & $\checkmark$    & 56.2 $\pm$ 1.1                      & -2.6                               & 85.7 $\pm$ 0.2                      & +2.2                          & 50.7 $\pm$ 0.5    & +0.4                      & 64.2              \\\hdashline[0.5pt/1pt]
			\rule{0pt}{10pt}GroupDRO &                 & 61.3 $\pm$ 0.4                      &                                    & 86.9 $\pm$ 0.0                      &                               & 52.5 $\pm$ 0.2    &                           & 66.9              \\
			GroupDRO                 & $\checkmark$    & 60.1 $\pm$ 0.7                      & -4.3                               & 87.3 $\pm$ 0.2                      & +1.8                          & 52.0 $\pm$ 0.7    & +0.6                      & 66.4              \\\hdashline[0.5pt/1pt]
			\rule{0pt}{10pt}IRM      &                 & 68.1 $\pm$ 1.6                      &                                    & 84.4 $\pm$ 0.5                      &                               & 49.2 $\pm$ 0.6    &                           & 67.2              \\
			IRM                      & $\checkmark$    & 69.0 $\pm$ 1.1                      & +2.9                               & 86.0 $\pm$ 0.4                      & +0.8                          & 50.7 $\pm$ 0.9    & +0.4                      & 68.6              \\\hdashline[0.5pt/1pt]
			\rule{0pt}{10pt}Fishr    &                 & 68.0 $\pm$ 2.9                      &                                    & 87.5 $\pm$ 0.1                      &                               & 53.7 $\pm$ 0.2    &                           & 69.7              \\
			Fishr                    & $\checkmark$    & 68.2 $\pm$ 3.0                      & +0.6                               & 87.4 $\pm$ 0.1                      & +0.6                          & 52.1 $\pm$ 0.7    & -0.5                      & 69.2              \\
			\bottomrule                                                                                                                                                                                                                                                     \\
		\end{tabular}}
\end{table}

\clearpage
\subsubsection{Training domain validation set}
\begin{table}[ht]
	\begin{center}
		\caption{OOD generalization performances with training domain validation set on \cmnist with \pairs.}
		\vskip 0.1in
		\adjustbox{max width=\textwidth}{%
			\begin{tabular}{lcccccc}
				\toprule
				\textbf{Algorithm}       & \textbf{\pairs} & \textbf{+90\%} & \textbf{+80\%} & \textbf{-90\%} & \textbf{Avg} & \textbf{$\Delta$ wr. acc} \\\midrule
				ERM                      &                 & 71.0 $\pm$ 0.5 & 73.4 $\pm$ 0.1 & 10.0 $\pm$ 0.1 & 51.5         &                           \\\hdashline[0.5pt/1pt]
				\rule{0pt}{10pt}DANN     &                 & 71.0 $\pm$ 0.3 & 73.4 $\pm$ 0.1 & 10.0 $\pm$ 0.1 & 51.5         &                           \\
				DANN                     & $\checkmark$    & 71.6 $\pm$ 0.3 & 73.3 $\pm$ 0.2 & 10.9 $\pm$ 0.4 & 51.9         & +0.9                      \\\hdashline[0.5pt/1pt]
				\rule{0pt}{10pt}GroupDRO &                 & 72.6 $\pm$ 0.2 & 73.1 $\pm$ 0.0 & 9.9 $\pm$ 0.1  & 51.8         &                           \\
				GroupDRO                 & $\checkmark$    & 72.7 $\pm$ 0.2 & 73.2 $\pm$ 0.5 & 13.0 $\pm$ 1.5 & 53.0         & +3.1                      \\\hdashline[0.5pt/1pt]
				\rule{0pt}{10pt}IRM      &                 & 72.3 $\pm$ 0.3 & 72.6 $\pm$ 0.4 & 9.9 $\pm$ 0.1  & 51.6         &                           \\
				IRM                      & $\checkmark$    & 67.4 $\pm$ 2.6 & 64.8 $\pm$ 1.4 & 24.2 $\pm$ 1.6 & 52.2         & +14.3                     \\\hdashline[0.5pt/1pt]
				\rule{0pt}{10pt}Fishr    &                 & 72.2 $\pm$ 0.6 & 73.1 $\pm$ 0.3 & 9.9 $\pm$ 0.2  & 51.8         &                           \\
				Fishr                    & $\checkmark$    & 69.1 $\pm$ 2.9 & 70.9 $\pm$ 1.7 & 22.6 $\pm$ 1.4 & 54.2         & +12.7                     \\
				\bottomrule
			\end{tabular}}
	\end{center}
\end{table}

\begin{table}[ht]
	\begin{center}
		\caption{OOD generalization performances with training domain validation set on \pacs with \pairs.}
		\vskip 0.1in
		\adjustbox{max width=\textwidth}{%
			\begin{tabular}{lccccccc}
				\toprule
				\textbf{Algorithm}       & \textbf{\pairs} & \textbf{A}     & \textbf{C}     & \textbf{P}     & \textbf{S}     & \textbf{Avg} & \textbf{$\Delta$ wr. acc} \\\midrule
				ERM                      &                 & 82.6 $\pm$ 1.6 & 79.2 $\pm$ 1.0 & 97.2 $\pm$ 0.5 & 74.9 $\pm$ 2.6 & 83.5         &                           \\\hdashline[0.5pt/1pt]
				\rule{0pt}{10pt}DANN     &                 & 84.7 $\pm$ 1.8 & 75.8 $\pm$ 0.9 & 97.3 $\pm$ 0.1 & 72.3 $\pm$ 1.0 & 82.5         &                           \\
				DANN                     & $\checkmark$    & 86.5 $\pm$ 0.9 & 77.0 $\pm$ 1.8 & 97.0 $\pm$ 0.2 & 73.0 $\pm$ 0.5 & 83.3         & +0.7                      \\\hdashline[0.5pt/1pt]
				\rule{0pt}{10pt}GroupDRO &                 & 83.4 $\pm$ 1.7 & 77.1 $\pm$ 0.3 & 97.6 $\pm$ 0.2 & 78.2 $\pm$ 1.3 & 84.1         &                           \\
				GroupDRO                 & $\checkmark$    & 83.4 $\pm$ 1.7 & 78.3 $\pm$ 0.3 & 97.6 $\pm$ 0.2 & 78.2 $\pm$ 1.3 & 84.4         & +1.1                      \\\hdashline[0.5pt/1pt]
				\rule{0pt}{10pt}IRM      &                 & 82.9 $\pm$ 2.6 & 81.4 $\pm$ 0.1 & 96.7 $\pm$ 0.6 & 73.1 $\pm$ 3.1 & 83.5         &                           \\
				IRM                      & $\checkmark$    & 82.4 $\pm$ 2.3 & 80.5 $\pm$ 0.8 & 97.2 $\pm$ 0.2 & 80.4 $\pm$ 1.3 & 85.1         & +7.3                      \\\hdashline[0.5pt/1pt]
				\rule{0pt}{10pt}Fishr    &                 & 85.3 $\pm$ 1.1 & 80.3 $\pm$ 1.1 & 97.9 $\pm$ 0.3 & 79.1 $\pm$ 1.7 & 85.6         &                           \\
				Fishr                    & $\checkmark$    & 85.4 $\pm$ 1.4 & 80.2 $\pm$ 0.8 & 96.2 $\pm$ 0.7 & 80.5 $\pm$ 0.8 & 85.6         & +1.1                      \\
				\bottomrule
			\end{tabular}}
	\end{center}
\end{table}

\begin{table}[ht]
	\begin{center}
		\caption{OOD generalization performances with training domain validation set on \terra with \pairs.}
		\vskip 0.1in
		\adjustbox{max width=\textwidth}{%
			\begin{tabular}{lccccccc}
				\toprule
				\textbf{Algorithm}       & \textbf{\pairs} & \textbf{L100}  & \textbf{L38}   & \textbf{L43}   & \textbf{L46}   & \textbf{Avg} & \textbf{$\Delta$ wr. acc} \\\midrule
				ERM                      &                 & 46.7 $\pm$ 3.5 & 41.8 $\pm$ 1.0 & 57.4 $\pm$ 1.0 & 39.7 $\pm$ 0.2 & 46.4         &                           \\\hdashline[0.5pt/1pt]
				\rule{0pt}{10pt}DANN     &                 & 46.1 $\pm$ 3.5 & 41.2 $\pm$ 1.0 & 56.7 $\pm$ 0.9 & 35.6 $\pm$ 1.1 & 44.9         &                           \\
				DANN                     & $\checkmark$    & 43.1 $\pm$ 3.8 & 41.1 $\pm$ 0.9 & 55.2 $\pm$ 2.1 & 38.7 $\pm$ 1.9 & 44.5         & +3.1                      \\\hdashline[0.5pt/1pt]
				\rule{0pt}{10pt}GroupDRO &                 & 48.4 $\pm$ 2.9 & 40.3 $\pm$ 3.1 & 57.9 $\pm$ 2.2 & 40.0 $\pm$ 0.5 & 46.6         &                           \\
				GroupDRO                 & $\checkmark$    & 48.4 $\pm$ 2.9 & 40.3 $\pm$ 3.1 & 57.9 $\pm$ 2.2 & 40.0 $\pm$ 0.5 & 46.6         & +0.0                      \\\hdashline[0.5pt/1pt]
				\rule{0pt}{10pt}IRM      &                 & 48.4 $\pm$ 3.8 & 35.6 $\pm$ 2.9 & 55.4 $\pm$ 0.9 & 40.1 $\pm$ 1.4 & 44.9         &                           \\
				IRM                      & $\checkmark$    & 40.4 $\pm$ 7.3 & 38.3 $\pm$ 2.5 & 48.8 $\pm$ 6.3 & 37.0 $\pm$ 0.9 & 41.1         & +1.4                      \\\hdashline[0.5pt/1pt]
				\rule{0pt}{10pt}Fishr    &                 & 49.2 $\pm$ 4.4 & 40.6 $\pm$ 1.4 & 57.9 $\pm$ 1.1 & 40.4 $\pm$ 1.2 & 47.0         &                           \\
				Fishr                    & $\checkmark$    & 51.0 $\pm$ 3.3 & 40.7 $\pm$ 1.3 & 58.2 $\pm$ 0.1 & 40.8 $\pm$ 1.2 & 47.7         & +0.3                      \\
				\bottomrule
			\end{tabular}}
	\end{center}
\end{table}

\clearpage
\subsubsection{Test domain validation set}

\begin{table}[ht]
	\begin{center}
		\caption{OOD generalization performances with test domain validation set on \cmnist with \pairs.}
		\vskip 0.1in
		\adjustbox{max width=\textwidth}{%
			\begin{tabular}{lcccccc}
				\toprule
				\textbf{Algorithm}       & \textbf{\pairs} & \textbf{+90\%} & \textbf{+80\%} & \textbf{-90\%} & \textbf{Avg} & \textbf{$\Delta$ wr. acc} \\\midrule
				ERM                      &                 & 71.7 $\pm$ 0.2 & 72.7 $\pm$ 0.2 & 28.8 $\pm$ 0.8 & 57.8         &                           \\\hdashline[0.5pt/1pt]
				\rule{0pt}{10pt}DANN     &                 & 73.0 $\pm$ 1.2 & 73.3 $\pm$ 0.1 & 25.8 $\pm$ 1.7 & 57.4         &                           \\
				DANN                     & $\checkmark$    & 72.1 $\pm$ 0.3 & 73.2 $\pm$ 0.3 & 23.2 $\pm$ 3.8 & 56.2         & -2.6                      \\\hdashline[0.5pt/1pt]
				\rule{0pt}{10pt}GroupDRO &                 & 73.4 $\pm$ 0.4 & 72.4 $\pm$ 0.0 & 38.1 $\pm$ 0.8 & 61.3         &                           \\
				GroupDRO                 & $\checkmark$    & 73.2 $\pm$ 0.2 & 73.3 $\pm$ 0.3 & 33.8 $\pm$ 2.3 & 60.1         & -4.3                      \\\hdashline[0.5pt/1pt]
				\rule{0pt}{10pt}IRM      &                 & 72.3 $\pm$ 0.3 & 72.5 $\pm$ 0.4 & 59.4 $\pm$ 5.3 & 68.1         &                           \\
				IRM                      & $\checkmark$    & 71.7 $\pm$ 0.4 & 73.1 $\pm$ 0.1 & 62.3 $\pm$ 3.1 & 69.0         & +2.9                      \\\hdashline[0.5pt/1pt]
				\rule{0pt}{10pt}Fishr    &                 & 73.8 $\pm$ 0.5 & 73.6 $\pm$ 0.1 & 56.7 $\pm$ 8.6 & 68.0         &                           \\
				Fishr                    & $\checkmark$    & 73.7 $\pm$ 0.6 & 73.5 $\pm$ 0.2 & 57.3 $\pm$ 8.4 & 68.2         & +0.6                      \\
				\bottomrule
			\end{tabular}}
	\end{center}
\end{table}

\begin{table}[ht]
	\begin{center}
		\caption{OOD generalization performances with test domain validation set on \pacs with \pairs.}
		\vskip 0.1in
		\adjustbox{max width=\textwidth}{%
			\begin{tabular}{lccccccc}
				\toprule
				\textbf{Algorithm}       & \textbf{\pairs} & \textbf{A}     & \textbf{C}     & \textbf{P}     & \textbf{S}     & \textbf{Avg} & \textbf{$\Delta$ wr. acc} \\\midrule
				ERM                      &                 & 86.6 $\pm$ 0.7 & 82.5 $\pm$ 0.8 & 97.3 $\pm$ 0.5 & 81.8 $\pm$ 0.7 & 87.0         &                           \\\hdashline[0.5pt/1pt]
				\rule{0pt}{10pt}DANN     &                 & 86.5 $\pm$ 0.8 & 79.9 $\pm$ 0.4 & 97.1 $\pm$ 0.1 & 75.3 $\pm$ 1.1 & 84.7         &                           \\
				DANN                     & $\checkmark$    & 87.0 $\pm$ 0.2 & 81.4 $\pm$ 0.7 & 96.8 $\pm$ 0.5 & 77.5 $\pm$ 1.3 & 85.7         & +2.2                      \\\hdashline[0.5pt/1pt]
				\rule{0pt}{10pt}GroupDRO &                 & 87.7 $\pm$ 0.4 & 82.1 $\pm$ 0.7 & 98.0 $\pm$ 0.2 & 79.6 $\pm$ 0.7 & 86.9         &                           \\
				GroupDRO                 & $\checkmark$    & 86.7 $\pm$ 0.3 & 83.2 $\pm$ 1.1 & 97.8 $\pm$ 0.1 & 81.4 $\pm$ 0.5 & 87.3         & +1.8                      \\\hdashline[0.5pt/1pt]
				\rule{0pt}{10pt}IRM      &                 & 82.3 $\pm$ 1.5 & 80.8 $\pm$ 0.7 & 95.8 $\pm$ 1.3 & 78.9 $\pm$ 1.4 & 84.4         &                           \\
				IRM                      & $\checkmark$    & 85.3 $\pm$ 0.3 & 81.7 $\pm$ 0.9 & 97.4 $\pm$ 0.3 & 79.7 $\pm$ 1.8 & 86.0         & +0.8                      \\\hdashline[0.5pt/1pt]
				\rule{0pt}{10pt}Fishr    &                 & 88.4 $\pm$ 0.4 & 82.2 $\pm$ 0.7 & 97.7 $\pm$ 0.5 & 81.6 $\pm$ 0.4 & 87.5         &                           \\
				Fishr                    & $\checkmark$    & 87.4 $\pm$ 0.8 & 82.6 $\pm$ 0.5 & 97.5 $\pm$ 0.6 & 82.2 $\pm$ 0.0 & 87.4         & +0.6                      \\
				\bottomrule
			\end{tabular}}
	\end{center}
\end{table}

\begin{table}[ht]
	\begin{center}
		\caption{OOD generalization performances with test domain validation set on \terra with \pairs.}
		\vskip 0.1in
		\adjustbox{max width=\textwidth}{%
			\begin{tabular}{lccccccc}
				\toprule
				\textbf{Algorithm}       & \textbf{\pairs} & \textbf{L100}  & \textbf{L38}   & \textbf{L43}   & \textbf{L46}   & \textbf{Avg} & \textbf{$\Delta$ wr. acc} \\\midrule
				ERM                      &                 & 58.7 $\pm$ 1.7 & 51.3 $\pm$ 1.8 & 59.9 $\pm$ 0.6 & 41.7 $\pm$ 1.0 & 52.9         &                           \\\hdashline[0.5pt/1pt]
				\rule{0pt}{10pt}DANN     &                 & 53.8 $\pm$ 0.5 & 47.4 $\pm$ 1.0 & 59.0 $\pm$ 0.5 & 42.9 $\pm$ 0.3 & 50.8         &                           \\
				DANN                     & $\checkmark$    & 54.4 $\pm$ 1.3 & 46.9 $\pm$ 1.2 & 58.1 $\pm$ 0.2 & 43.3 $\pm$ 0.0 & 50.7         & +0.4                      \\\hdashline[0.5pt/1pt]
				\rule{0pt}{10pt}GroupDRO &                 & 57.3 $\pm$ 0.4 & 50.4 $\pm$ 1.1 & 59.7 $\pm$ 0.7 & 42.8 $\pm$ 0.7 & 52.5         &                           \\
				GroupDRO                 & $\checkmark$    & 55.9 $\pm$ 3.2 & 50.6 $\pm$ 0.7 & 57.9 $\pm$ 0.4 & 43.4 $\pm$ 0.4 & 52.0         & +0.6                      \\\hdashline[0.5pt/1pt]
				\rule{0pt}{10pt}IRM      &                 & 53.6 $\pm$ 0.5 & 47.9 $\pm$ 1.9 & 54.1 $\pm$ 0.9 & 41.3 $\pm$ 0.6 & 49.2         &                           \\
				IRM                      & $\checkmark$    & 59.3 $\pm$ 1.8 & 45.5 $\pm$ 0.6 & 56.4 $\pm$ 1.7 & 41.7 $\pm$ 0.7 & 50.7         & +0.4                      \\\hdashline[0.5pt/1pt]
				\rule{0pt}{10pt}Fishr    &                 & 60.7 $\pm$ 0.8 & 49.4 $\pm$ 0.7 & 59.5 $\pm$ 0.5 & 45.0 $\pm$ 0.5 & 53.7         &                           \\
				Fishr                    & $\checkmark$    & 58.9 $\pm$ 1.0 & 46.4 $\pm$ 1.8 & 58.6 $\pm$ 0.7 & 44.5 $\pm$ 0.8 & 52.1         & -0.5                      \\
				\bottomrule
			\end{tabular}}
	\end{center}
\end{table}

\chapter{Appendices of \feat}\label{APP:FeAT}

\section{Notations}
\label{CH:FeAT:sec:notations_appdx}

We use bold-faced letters for vectors and matrices otherwise for scalar. We use $\| \cdot \|_2$ to denote the Euclidean norm of a vector or the spectral norm of a matrix, while denoting $\| \cdot \|_F$ as the Frobenius norm of a matrix. For a neural network, we denote ${\psi}(x)$ as the activation function. Let ${\bf I}_d$ be the identity matrix with a dimension of $\mathbb{R}^{d \times d}$. When comparing two sequences $\{ a_n \}$ and $\{b_n \}$, we employ standard asymptotic notations such as $O(\cdot)$, $o(\cdot)$, $\Omega(\cdot)$, and $\Theta(\cdot)$ to describe their limiting behavior. Lastly, sequences of integers are denoted as $[n] = \{1,2,\ldots,n \}$.

\begin{table}[ht]
    \caption{Notations for key concepts involved in \feat.}
    \centering
    \resizebox{\textwidth}{!}{
        \begin{tabular}{ll}
            \toprule
            \textbf{Symbols}                      & \textbf{Definitions}                                                                                                                                         \\\midrule
            \(\gX=\R^n\)                          & the input space                                                                                                                                              \\\midrule
            \(\gY=\R\)                            & the label space                                                                                                                                              \\\midrule
            \(\gZ=\R^d\)                          & the latent space                                                                                                                                             \\\midrule
            \(m\)                                 & the hidden dimension                                                                                                                                         \\\midrule
            \(F_j(\cdot)\)                        & the $j$-th filter of the CNN model                                                                                                                           \\\midrule
            \(\rmW_j\)                            & the weights of $j$-th filter of the CNN model, containing $m$ hidden units $\rvw_{j,r}$                                                                      \\\midrule
            \(\psi(\cdot)\)                       & the activation function of the CNN model                                                                                                                     \\\midrule
            \(\varphi\)                           & the featurizer $\varphi:\gX\rightarrow\gZ$ learns a latent representation for each input example                                                             \\\midrule
            \(w\)                                 & the classifier $w:\gZ\rightarrow\gY$                                                                                                                         \\\midrule
            \(w_j\)                               & the classifier learned at $j$-th round                                                                                                                       \\\midrule
            \(f\in\gF\)                           & the predictor $f=w\circ\varphi:\gX\rightarrow\gY$ is composed of a featurizer and classifier                                                                 \\
            \(\)                                  & when $w$ is linear, $f$ can be simply represented via dot product $w\cdot\varphi$                                                                            \\\midrule
            \(\envall\)                           & the set of indices for all environments                                                                                                                      \\\midrule
            \(\envtrain\)                         & the subset of indices of training environments                                                                                                               \\\midrule
            \(e\)                                 & the index set of a specific environment                                                                                                                      \\\midrule
            \(\mathcal{E}_\alpha\)                & the set of environments following the data model as Def.~\ref{CH:FeAT:def:risk_irm}, where each is specified as $(\alpha,\beta_e)$                           \\\midrule
            \(\dataset^e,\dataset_e\)             & the dataset from environment $e$, containing $n_e$ samples $\{\rvx_i^e,y_i^e\}$ considered as i.i.d. from $\mathbb{P}^e$                                     \\
            \midrule
            \(\dataset\)                          & the overall dataset containing $n$ samples from all environments, $\dataset=\{\dataset^e\}_{e\in\envall}$                                                    \\
            \midrule
            \(\dataset^a\)                        & the augmentation set, we use \(\dataset^a_i\) to denote the augmentation set separated at $i$-th round                                                       \\\midrule
            \(\dataset^r\)                        & the retention set, we use \(\dataset^r_i\) to denote the retention set separated at $i$-th round                                                             \\\midrule
            \(G\)                                 & $G=\{G^r,G^a\}$ with $2k-1$ groups at round $k$, where $G^a=\{\dataset_i^a\}_{i=0}^{k-1}$ is the grouped sets,                                               \\\(\) & for new feature augmentation and $G^r=\{\dataset_i^r\}_{i=1}^{k-1}$ is the grouped sets for already learned feature retention\\
            \midrule
            \(L_e\)                               & the empirical risk calculated based on $\dataset^e$, e.g., square loss or logistic loss                                                                      \\\midrule
            \(\ell_\feat\)                        & the \feat objective, including $\ell_{\dataset_i^a}$ the empirical risk at $\dataset^a_i$ and $\ell_{\dataset_i^r}$ at $\dataset^r_i$                        \\\midrule
            \(L_{\irml}(\rmW)\)                   & the \irml loss                                                                                                                                               \\\midrule
            \(\ell'^e\)                           & the first order derivative of $L_e$ with respect to the $i$-th sample from environment $e$                                                                   \\\midrule
            \(C_\irml^e\)                         & $C_\irml^e \triangleq \frac{1}{n_e}\sum_{i=1}^{n_e} {\ell'\big(y^e_i \hat{y}^e_i\big) \cdot y^e_i \hat{y}^e_i}$, a useful quantity to analyze \irml dynamics \\\midrule
            \(\gamma_{j,r}^{inv},\gamma_{j,r,1}\) & the invariant feature learning quantity in Eq.~\ref{CH:FeAT:eq:decompostion}                                                                                 \\\midrule
            \(\gamma_{j,r}^{spu},\gamma_{j,r,2}\) & the spurious feature learning quantity in Eq.~\ref{CH:FeAT:eq:decompostion}                                                                                  \\\midrule
            \(\rho_{j,r,i}(t)\)                   & the noise feature learning quantity in Eq.~\ref{CH:FeAT:eq:decompostion}                                                                                     \\\midrule
            \bottomrule
        \end{tabular}}
\end{table}
\clearpage
\section{Limitations and Future Directions}
As a pioneering work that studies feature learning of ERM and OOD objectives and their interactions in OOD generalization, our theoretical settings are limited to studying the influence of spurious and invariant correlation strengths on spurious and invariant feature learning, based on a two-layer CNN network.
In fact, the feature learning of a network can be influenced by several other factors, such as the difficulty of learning a feature and the capacity of features that a model can learn~\citep{what_shape,toy_model}.
Future works can be built by extending our framework to consider the influence of a broad of factors on feature learning in OOD generalization.

Moreover, as there could exist cases where certain features should not be learned, it is also promising to explore how to prevent the feature learning of undesirable features during the early stages of OOD generalization and to further relieve the optimization dilemma in OOD generalization~\citep{pair}, to improve the robustness against backdoor attacks~\citep{min2023towards}, and its further implications to OOD generalization~\citep{Lin2023SpuriousFD}. Besides, it is also interesting to investigate feature learning for complicated data such as graphs~\citep{huang2023graph}, especially under various graph distribution shifts~\citep{ciga,gala,hao,reactionOOD,drugood}.

\section{Related Work}\label{CH:FeAT:sec:related_work_appdx}
\textbf{On Feature Learning and Generalization.}
Understanding feature learning in deep networks is crucial to understanding their generalization~\citep{mlp,ib,sgd_fl1,sgd_fl2,understand_ensemble,understand_benign}.
Earlier attempts are mostly about empirical probing~\citep{saliency,new_saliency,what_shape,toy_model}.
\citet{what_shape,toy_model,simple_bias} find that the feature learning of a network can be influenced by several other factors, such as the difficulty of learning a feature and the capacity of features that a model can learn.
Although our data model focuses on the correlation perspective, different correlation strengths in fact can simulate the difficulty or the simplicity of learning a feature.

Beyond the empirical probing, \citet{understand_ensemble} proposed a new theoretical framework that characterizes the feature learning process of deep networks, which has been widely adopted to analyze behaviors of deep networks~\citep{understand_ssl,understand_adam,understand_benign}
However, how the learned features from ID data can be generalized to OOD data remains elusive. The only exceptions are \citep{understand_da} and \citep{feat_distort}. \citet{feat_distort} find fine-tuning can distort the pre-trained features while fine-tuning can be considered as a special case in our framework. \citet{understand_da} focus on how data augmentation helps promote good but hard-to-learn features and improve OOD generalization.
\citet{pde} studies feature learning when the group-related features are more predictive for inferring group labels.
In contrast, we study the direct effects of ERM and OOD objectives to feature learning and provide a theoretical explanation to the phenomenon that ERM may have already learned good features~\citep{dare,dfrlearn}.
To the best of our knowledge, we are the \textit{first} to analyze the feature learning of ERM and OOD objectives and their interactions in the general OOD generalization setting.

\textbf{On the correlation between ID and OOD performances.}
The debate about feature learning and generalization under distribution shifts also
extends to the ID and OOD performance correlations along with training or fine-tuning neural nets across a variety of OOD generalization tasks.
\citet{ood_evolution,ood_online,assaying_transfer} found that there often exists a linear dependency between ID and OOD performance under a wide range of models and distribution shifts.
While \citet{feat_distort,robust_finetune} found that fine-tuning pre-trained models often leads to an increased in-distribution but decreased OOD performance.
\citet{idood_inverse} observed cases where ID and OOD performance are inversely correlated. \citet{pair,opt_selection} studied the ID and OOD performance trade-offs from the optimization perspective.

Our work provides theoretical explanations for different correlation behaviors of ID and OOD performance, as well as provides a solution for mitigating the trade-offs in optimization.
Theorem~\ref{CH:FeAT:thm:erm_learn_feat} implies that, in cases where invariant features are more informative than spurious features, the higher ID performance indicates a better fit to invariant features, thus promising a higher OOD performance, aligned with observations in~\citep{ood_evolution,ood_online,assaying_transfer}. While in cases where invariant features are less informative than spurious features, the higher ID performance implies a better fit to spurious features, thus bringing a lower OOD performance~\citep{idood_inverse}.
Similarly, when fine-tuning a pre-trained model, if the model does not learn the features sufficiently well, ID-OOD performance will be in a positive correlation.
However, when spurious correlations are present as easy-to-learn features, ERM can lead to a better fit for spurious features and distort the previously learned invariant features~\citep{feat_distort,robust_finetune,forgetting}.

\textbf{Rich Feature Learning.}
Recently many OOD objectives have been proposed to regularize  ERM such that the model can focus on learning invariant features~\citep{irmv1,vrex,sd,clove,fishr}.
However, due to the intrinsic conflicts of ERM and OOD objectives, it often requires exhaustive hyperparameter tuning of ERM pre-training epochs and regularization weights~\citep{rfc,pair}. Especially, the final OOD performance has a large dependence on the number of pre-training epochs.
To remedy the issue, \citet{rfc} proposed \rfc to construct rich feature representations with plentiful potentially useful features such as network initialization. Although both \rfc and \feat perform DRO on grouped subsets, \rfc rely on multiple initializations of the whole network to capture diverse features from the subsets, and complicated ensembling of the features, which requires much more training epochs for the convergence. In contrast, \feat relieves the requirements by performing direct augmentation-retention on the grouped subsets, and thus obtains better performance.
More crucially, although \rfc and other rich feature learning algorithms such as weight averaging~\citep{diwa,eoa,rfc2} have gained impressive successes in mitigating the dilemma, explanations about the reliance on ERM pre-training and why rich feature learning mitigates the dilemma remain elusive. Our work provides novel theoretical explanations for the success of rich feature learning algorithms for OOD generalization.
Complementary to the empirical observations made by existing works, our work provides the first theoretical explanation for the feature learning of ERM and OOD objectives for OOD generalization.

Besides, there exists a rich literature on learning diverse representations for better generalization. Similar to weight average~\citep{diwa}, \citet{evade_simplicity} propose to train diverse models to resolve simplicity bias.
\citet{diversify} propose to learn diverse solutions for the underspecified learning problem.
\citet{transformer_diverse} propose to regularize attention heads in transformers to learn diverse features. \citet{pro2} propose to learn diverse classifiers for sample efficient domain adaption.

\section{Proofs for theoretical results}

\subsection{Implementation details of the synthetic CNN experiments}
\label{exp:CNN_synthetic}
For linear activation function $\psi(x) = x$, the logit $\hat{y}^e_i$ (which is a function of $\rmW$) of sample $i$ in the environment $e$ can be explicitly written as \[
    \hat{y}^e_i = f(\rmW, \rvx^e_i) = F_{+1}(\rmW_{+1}, \rvx^e_i) - F_{-1}(\rmW_{-1}, \rvx^e_i) = \sum_{j\in \{\pm 1\}} {\frac{j}{m}\sum_{r=1}^m {\left[ \rvw_{j,r}^\top (\rvx^e_{i,1} + \rvx^e_{i,2}) \right]}},
\]
where $\rmW \triangleq \{\rmW_{+1}, \rmW_{-1}\}$ and $\rmW_j \triangleq
    \begin{bmatrix}
        \rvw^\top_{j,1} \\
        \vdots          \\
        \rvw^\top_{j,m}
    \end{bmatrix}$ for $j\in\{\pm 1\}$. We initialized all the network weights as $\mathcal{N}(0, \sigma_0^2)$ and we set $\sigma_0 = 0.01$.

The test dataset $(\rvx, y)$ is generated through
\[
    \rvx_{i,1} = y_i \cdot \rvv_1 + y_i \cdot \Rad(1-\beta_e) \cdot  \rvv_2,  \ \ \quad
    \rvx_{i,2} = \boldsymbol{\xi},
\]
where half of the dataset uses $\Rad(1-\beta_1)$ and the other half uses $\Rad(1-\beta_2)$. Here $\boldsymbol{\xi} \sim \mathcal{N}(0, \sigma_p^2\cdot (\mathbf{I}_d - \mathbf{v}_1\mathbf{v}_1^\top - \mathbf{v}_2\mathbf{v}_2^\top))$ and we chose $\sigma_p = 0.01$.

From the definition of IRMv1, we take derivative wrt. the scalar $1$ of the logit $1\cdot \hat{y}^e_i$. Thus, for environment $e$, the penalty is
\[
    \left(\frac{1}{n_e}\sum_{i=1}^{n_e} {\nabla_{w|w=1}\ell\big(y^e_i (w\cdot\hat{y}^e_i)\big)}\right)^2 = \left(\frac{1}{n_e}\sum_{i=1}^{n_e} {\ell'\big(y^e_i\hat{y}^e_i\big) \cdot y^e_i\hat{y}^e_i}\right)^2.
\]
Then, the IRMv1 objective is (we set $n_1 = n_2 = 2500$ in the simulation)
\[
    L_\irml(\rmW) = \sum_{e\in \mathcal{E}_{tr}}{\frac{1}{n_e}\sum_{i=1}^{n_e}{\ell\big(y^e_i \hat{y}^e_i\big)}} + \lambda \sum_{e\in \mathcal{E}_{tr}}{\left(\frac{1}{n_e}\sum_{i=1}^{n_e} {\ell'\big(y^e_i \hat{y}^e_i\big) \cdot y^e_i \hat{y}^e_i} \right)^2}.
\]

We used constant stepsize GD to minimize $L_\irml(\rmW)$, and we chose $\lambda = 10^{8}$ (heavy regularization setup).

Let $C_\irml^e \triangleq \frac{1}{n_e}\sum_{i=1}^{n_e} {\ell'\big(y^e_i \hat{y}^e_i\big) \cdot y^e_i \hat{y}^e_i}$. The gradient of $L_\irml(\rmW)$ with respect to each $\rvw_{j,r}$ can be explicitly written as
\[
    \begin{aligned}
            & \nabla_{\rvw_{j,r}} L_\irml(\rmW)                                                                                                                                                                                                           \\={} & \sum_{e\in \mathcal{E}_{tr}}{\frac{1}{n_e}\sum_{i=1}^{n_e}{\ell'\big(y^e_i \hat{y}^e_i\big)\cdot y^e_i\cdot \frac{j}{m}(\rvx^e_{i,1} + \rvx^e_{i,2})}}\\
            & + 2\lambda \sum_{e\in \mathcal{E}_{tr}}{
        \frac{C_\irml^e}{n_e} \sum_{i=1}^{n_e} {\Big(\ell''\big(y^e_i \hat{y}^e_i\big)\cdot \hat{y}^e_i\cdot \frac{j}{m}(\rvx^e_{i,1} + \rvx^e_{i,2}) + \ell'\big(y^e_i \hat{y}^e_i\big) \cdot y^e_i\cdot \frac{j}{m}(\rvx^e_{i,1} + \rvx^e_{i,2})\Big)}} \\
        ={} & \sum_{e\in \mathcal{E}_{tr}}{\frac{j}{n_e m}\sum_{i=1}^{n_e}{\ell'\big(y^e_i \hat{y}^e_i\big)\cdot y^e_i\cdot(\rvx^e_{i,1} + \rvx^e_{i,2})}}                                                                                                \\
            & + 2\lambda \sum_{e\in \mathcal{E}_{tr}}{
        \frac{jC_\irml^e}{n_e m} \sum_{i=1}^{n_e} {\ell''\big(y^e_i \hat{y}^e_i\big)\cdot \hat{y}^e_i\cdot (\rvx^e_{i,1} + \rvx^e_{i,2})}}                                                                                                                \\
            & + 2\lambda \sum_{e\in \mathcal{E}_{tr}}{
        \frac{jC_\irml^e}{n_e m} \sum_{i=1}^{n_e} {\ell'\big(y^e_i \hat{y}^e_i\big) \cdot y^e_i\cdot (\rvx^e_{i,1} + \rvx^e_{i,2})}}                                                                                                                      \\
        ={} & \sum_{e\in \mathcal{E}_{tr}}{\frac{j(1 + 2\lambda C_\irml^e)}{n_e m}\sum_{i=1}^{n_e}{\ell'\big(y^e_i \hat{y}^e_i\big)\cdot y^e_i\cdot(\rvx^e_{i,1} + \rvx^e_{i,2})}}                                                                        \\
            & {} + 2\lambda \sum_{e\in \mathcal{E}_{tr}}{
        \frac{jC_\irml^e}{n_e m} \sum_{i=1}^{n_e} {\ell''\big(y^e_i \hat{y}^e_i\big)\cdot \hat{y}^e_i\cdot (\rvx^e_{i,1} + \rvx^e_{i,2})}}.
    \end{aligned}
\]
Observe that $C_\irml^e$ is in fact the scalar gradient
$
    C_\irml^e = \nabla_{w|w=1} L^e_\erm(\rmW)
$ that we want to force zero, whose effect can be understood as a dynamic re-weighting of the ERM gradient. Due to its importance in the analysis and interpretation of \irml, we tracked $C_\irml^e$ in our simulations.

The invariant and spurious feature learning terms that we tracked are the mean of $\langle \rvw_{j,r}, j\rvv_1 \rangle$ and $\langle \rvw_{j,r}, j\rvv_2\rangle$ for $j\in \{\pm 1\}, r\in[m]$, respectively.

\subsection{Proof for Theorem~\ref{CH:FeAT:thm:erm_learn_feat}} \label{CH:FeAT:sec:proof_erm_feat}

\begin{theorem} [Formal statement of Theorem \ref{CH:FeAT:thm:erm_learn_feat}]\label{CH:FeAT:thm:erm_learn_feat_appdx}
    For $\rho>0$, denote $\underline{n} \triangleq \min_{e\in\mathcal{E}_{tr}}{n_e}$, $n\triangleq \sum_{e\in\mathcal{E}_{tr}}{n_e}$, $\epsilon_C \triangleq \sqrt{\frac{2\log{(16/\rho)}}{\underline{n}}}$ and $\delta\triangleq \exp\{O(\underline{n}^{-1})\} - 1$. Define the feature learning terms $\Lambda^t_{j,r} \triangleq \langle \rvw_{j,r}^t, j\rvv_1 \rangle$ and $\Gamma^t_{j,r} \triangleq \langle \rvw_{j,r}^t, j\rvv_2 \rangle$ for $j\in\{\pm 1\}, r\in[m]$. Suppose we run $T$ iterations of GD for the ERM objective. With sufficiently large $\underline{n}$ and $\psi(x)=x$,  assuming that
    \[
        \begin{aligned}
             & \alpha, \beta_1, \beta_2 < \frac{1-\epsilon_C-\delta (\frac{1}{4}+\frac{\epsilon_C}{2})}{2} &  & \text{($\alpha, \beta_1, \beta_2$ are sufficiently smaller than $\frac{1}{2}$)}, \\
             & \alpha > \frac{\beta_1+\beta_2}{2} + \epsilon_C + \frac{\delta (1+\epsilon_C)}{2}           &  & \text{($\alpha$ is sufficiently larger than $\frac{\beta_1+\beta_2}{2}$)},
        \end{aligned}
    \]and choosing
    \[
        \begin{aligned}
            \sigma_0^2 & = O\left(\underline{n}^{-2} \log^{-1}{(m/\rho)}\right),                                                                             \\
            \sigma_p^2 & = O\left(\min\left\{d^{-1/2}\log^{-1/2}{(nm/\rho)}, T^{-1}\eta^{-1} m \left(d+ n\sqrt{d\log(n^2/\rho)}\right)^{-1} \right\}\right),
        \end{aligned}
    \]
    there exists a constant $\eta$, such that for any $j\in\{\pm 1\}, r\in[m]$, with probability at least $1-2\rho$, $\Lambda^t_{j,r}$ and $\Gamma^t_{j,r}$ are converging and the increment of the spurious feature $\Gamma^{t+1}_{j,r} - \Gamma^t_{j,r}$ is larger than that of the invariant feature $\Lambda^{t+1}_{j,r} - \Lambda^t_{j,r}$ at any iteration $t\in\{0,\ldots,T-1\}$.
\end{theorem}

\begin{proof} [Proof of Theorem \ref{CH:FeAT:thm:erm_learn_feat_appdx}]
    We begin with checking the feature learning terms in the ERM stage using constant stepsize GD: $\rmW^{t+1} = \rmW^t - \eta \cdot \nabla_{\rmW} L_\irml(\rmW^t)$. Note that with $\psi(x)=x$ the update rule for each $\rvw_{j,r}, \forall j\in\{+1, -1\}, r\in[m]$ can be written as
    \[
        \begin{aligned}
            \rvw^{t+1}_{j,r} & = \rvw^{t}_{j,r} - \frac{j\eta}{m} \sum_{e\in \mathcal{E}_{tr}}{\frac{1}{n_e}\sum_{i=1}^{n_e}{\ell'\big(y^e_i \hat{y}^e_i\big)\cdot y^e_i\cdot (\rvx^e_{i,1} + \rvx^e_{i,2}})}                                                      \\
                             & = \rvw^{t}_{j,r} - \frac{j\eta}{m} \sum_{e\in \mathcal{E}_{tr}}{\frac{1}{n_e}\sum_{i=1}^{n_e}{\ell'\big(y^e_i \hat{y}^e_i\big)\cdot  ( \Rad(\alpha)_i \cdot  \rvv_1 + \Rad(\beta_e)_i \cdot  \rvv_2 + y^e_i\boldsymbol{\xi}^e_i)}}.
        \end{aligned}
    \]
    Define the quantities of interest (the feature learning terms): $\Lambda^t_{j,r} \triangleq \langle \rvw_{j,r}^t, j\rvv_1 \rangle, \Gamma^t_{j,r} \triangleq \langle \rvw_{j,r}^t, j\rvv_2 \rangle, \Xi^{t,e}_{j,r,i} \triangleq \langle \rvw_{j,r}^t, j \boldsymbol{\xi}^e_i\rangle$. From our data generating procedure (Definition \ref{CH:FeAT:def:risk_irm}), we know that the first two coordinates of $\boldsymbol{\xi}^e_i$ are zero. Thus, we can write down the update rule for each feature learning term as follows.
    \[
        \begin{aligned}
            \Lambda^{t+1}_{j,r}   & = \Lambda^t_{j,r} - \frac{\eta}{m} \sum_{e\in \mathcal{E}_{tr}}{\frac{1}{n_e}\sum_{i=1}^{n_e}{\ell'\big(y^e_i \hat{y}^e_i\big)\cdot  \Rad(\alpha)_i}},                                                                  \\
            \Gamma^{t+1}_{j,r}    & = \Gamma^t_{j,r} - \frac{\eta}{m} \sum_{e\in \mathcal{E}_{tr}}{\frac{1}{n_e}\sum_{i=1}^{n_e}{\ell'\big(y^e_i \hat{y}^e_i\big)\cdot  \Rad(\beta_e)_i}},                                                                  \\
            \Xi^{t+1,e'}_{j,r,i'} & = \Xi^{t, e'}_{j,r, i'} - \frac{\eta}{m} \sum_{e\in \mathcal{E}_{tr}}{\frac{1}{n_e}\sum_{i=1}^{n_e}{\ell'\big(y^e_i \hat{y}^e_i\big)\cdot y_i^e\cdot \langle \boldsymbol{\xi}^e_i, \boldsymbol{\xi}^{e'}_{i'}\rangle}}.
        \end{aligned}
    \]
    More explicitly, we can write
    \begin{align}
        \Lambda^{t+1}_{j,r}   & = \Lambda^t_{j,r} +\frac{\eta}{m} \sum_{e\in \mathcal{E}_{tr}}{\frac{1}{n_e}\sum_{i=1}^{n_e}{\frac{\Rad(\alpha)_i}{1+\exp\{y^e_i \hat{y}^e_i\}}  }}, \label{erm_eq1}                                                                 \\
        \Gamma^{t+1}_{j,r}    & = \Gamma^t_{j,r} + \frac{\eta}{m} \sum_{e\in \mathcal{E}_{tr}}{\frac{1}{n_e}\sum_{i=1}^{n_e}{\frac{\Rad(\beta_e)_i}{1+\exp\{y^e_i \hat{y}^e_i\}}  }},\label{erm_eq2}                                                                 \\
        \Xi^{t+1,e'}_{j,r,i'} & = \Xi^{t, e'}_{j,r, i'} + \frac{\eta}{m} \sum_{e\in \mathcal{E}_{tr}}{\frac{1}{n_e}\sum_{i=1}^{n_e}{\frac{y_i^e\cdot \langle \boldsymbol{\xi}^e_i, \boldsymbol{\xi}^{e'}_{i'}\rangle}{1+\exp\{y^e_i \hat{y}^e_i\}}}}.\label{erm_eq3}
    \end{align}
    Notice that the updates (\ref{erm_eq1}), (\ref{erm_eq2}) for $\Lambda_{j,r}, \Gamma_{j,r}$ are independent of $j,r$. Denoting
    \[
        \begin{aligned}
            \Delta_\Lambda^t & \triangleq \frac{1}{m} \sum_{e\in \mathcal{E}_{tr}}{\frac{1}{n_e}\sum_{i=1}^{n_e}{\frac{\Rad(\alpha)_i}{1+\exp\{y^e_i \hat{y}^e_i\}}}},  \\
            \Delta_\Gamma^t  & \triangleq \frac{1}{m} \sum_{e\in \mathcal{E}_{tr}}{\frac{1}{n_e}\sum_{i=1}^{n_e}{\frac{\Rad(\beta_e)_i}{1+\exp\{y^e_i \hat{y}^e_i\}}}},
        \end{aligned}
    \]
    we can conclude that for any $j\in\{+1,-1\}, r\in[m]$,
    \begin{equation}\label{erm_feature_update}
        \begin{aligned}
            \Lambda^{t+1}_{j,r} & = \Lambda^t_{j,r} +\eta\cdot\Delta_\Lambda^t= \eta\cdot \sum_{k = 0}^t {\Delta_\Lambda^k} + \Lambda^{0}_{j,r}, \\
            \Gamma^{t+1}_{j,r}  & = \Gamma^t_{j,r} +\eta\cdot\Delta_\Gamma^t= \eta\cdot \sum_{k = 0}^t {\Delta_\Gamma^k} + \Gamma^{0}_{j,r}.
        \end{aligned}
    \end{equation}
    Then, we write the logit $\hat{y}^e_i$ as
    \[
        \begin{aligned}
            \hat{y}^e_i ={} & \sum_{j\in \{\pm 1\}} {\frac{j}{m}\sum_{r=1}^m {\left[ \left\langle\rvw^t_{j,r}, y_i^e\cdot\Rad(\alpha)_i \cdot  \rvv_1 + y_i^e\cdot\Rad(\beta_e)_i \cdot  \rvv_2 + \rvx^e_{i,2}\right\rangle\right]}}                                                                   \\
            ={}             & \sum_{j\in \{\pm 1\}} {\frac{j}{m}\sum_{r=1}^m {\left[  jy_i^e\cdot\Rad(\alpha)_i \cdot  \Lambda^t_{j,r} + jy_i^e\cdot\Rad(\beta_e)_i \cdot  \Gamma^t_{j,r} + j\cdot \Xi^{t,e}_{j,r,i}\right]}}                                                                          \\
            ={}             & \sum_{j\in \{\pm 1\}} {\frac{1}{m}\sum_{r=1}^m {\left[  y_i^e\cdot\Rad(\alpha)_i \cdot  \Lambda^t_{j,r} + y_i^e\cdot\Rad(\beta_e)_i \cdot  \Gamma^t_{j,r} + \Xi^{t,e}_{j,r,i}\right]}}                                                                                   \\
            ={}             & y_i^e\cdot\Rad(\alpha)_i \cdot  \sum_{j\in \{\pm 1\}} {\sum_{r=1}^m {\frac{\Lambda^t_{j,r}}{m}}} + y_i^e\cdot\Rad(\beta_e)_i \cdot  \sum_{j\in \{\pm 1\}} {\sum_{r=1}^m {\frac{\Gamma^t_{j,r}}{m}}} + \sum_{j\in \{\pm 1\}} {\sum_{r=1}^m {\frac{\Xi^{t,e}_{j,r,i}}{m}}} \\
            ={}             & y_i^e\cdot\Rad(\alpha)_i \cdot 2\eta \cdot \sum_{k = 0}^{t-1} {\Delta_\Lambda^k} + y_i^e\cdot\Rad(\beta_e)_i \cdot 2\eta \cdot \sum_{k = 0}^{t-1} {\Delta_\Gamma^k}                                                                                                      \\
                            & +
            y_i^e\cdot\Rad(\alpha)_i \cdot  \sum_{j\in \{\pm 1\}} {\sum_{r=1}^m {\frac{\Lambda^{0}_{j,r}}{m}}} + y_i^e\cdot\Rad(\beta_e)_i \cdot  \sum_{j\in \{\pm 1\}} {\sum_{r=1}^m {\frac{ \Gamma^{0}_{j,r}}{m}}} + \sum_{j\in \{\pm 1\}} {\sum_{r=1}^m {\frac{\Xi^{t,e}_{j,r,i}}{m}}}.
        \end{aligned}
    \]
    Denoting $\mathbb{Q}^e_i \triangleq \Rad(\alpha)_i  \sum_{j\in \{\pm 1\}} {\sum_{r=1}^m {\frac{\Lambda^{0}_{j,r}}{m}}} + \Rad(\beta_e)_i  \sum_{j\in \{\pm 1\}} {\sum_{r=1}^m {\frac{ \Gamma^{0}_{j,r}}{m}}} + y^e_i\cdot \sum_{j\in \{\pm 1\}} {\sum_{r=1}^m {\frac{\Xi^{t,e}_{j,r,i}}{m}}}$, we have
    \[
        \hat{y}_i^e = y_i^e\cdot\left(\Rad(\alpha)_i \cdot 2\eta\cdot \sum_{k = 0}^{t-1} {\Delta_\Lambda^k}  + \Rad(\beta_e)_i \cdot 2\eta\cdot \sum_{k = 0}^{t-1} {\Delta_\Gamma^k} + \mathbb{Q}^e_i\right),
    \]
    We need the following concentration lemma to control the scale of $\mathbb{Q}^e_i$, whose proof is given in Appendix \ref{proof_erm_concent_Q}.
    \begin{lemma}\label{lem:erm_concent_Q} Denote $\underline{n} \triangleq \min_{e\in\mathcal{E}_{tr}}{n_e}, n\triangleq \sum_{e\in\mathcal{E}_{tr}}{n_e}$. For $\rho > 0$, if
        \[
            \begin{aligned}
                \sigma_0^2 & = O\left(\underline{n}^{-2} \log^{-1}{(m/\rho)}\right),                                                                             \\
                \sigma_p^2 & = O\left(\min\left\{d^{-1/2}\log^{-1/2}{(nm/\rho)}, T^{-1}\eta^{-1} m \left(d+ n\sqrt{d\log(n^2/\rho)}\right)^{-1} \right\}\right),
            \end{aligned}
        \]
        then with probability at least $1-\rho$, for any $e\in \mathcal{E}_{tr}, i\in [n_e]$, it holds that $\abs{\mathbb{Q}_i^e} = O(\underline{n}^{-1})$.
    \end{lemma}
    Then $\Delta_\Lambda^t$ and $\Delta_\Gamma^t$ can be explicitly written as
    \[
        \begin{aligned}
             & \ \ \ \ \Delta_\Lambda^t  = \\& \sum_{e\in \mathcal{E}_{tr}}{\frac{1}{n_e m}\sum_{i=1}^{n_e}{\frac{\Rad(\alpha)_i}{1+\exp{\left\{\Rad(\alpha)_i \cdot 2\eta\cdot \sum_{k = 0}^{t-1} {\Delta_\Lambda^k}\right\}}\cdot \exp{\left\{\Rad(\beta_e)_i \cdot 2\eta\cdot \sum_{k = 0}^{t-1} {\Delta_\Gamma^k}\right\}} \cdot \exp{\left\{\mathbb{Q}^e_i\right\}}}}},  \\
             & \ \ \ \ \Delta_\Gamma^t  =  \\& \sum_{e\in \mathcal{E}_{tr}}{\frac{1}{n_e m}\sum_{i=1}^{n_e}{\frac{\Rad(\beta_e)_i}{1+\exp{\left\{\Rad(\alpha)_i \cdot 2\eta\cdot \sum_{k = 0}^{t-1} {\Delta_\Lambda^k}\right\}}\cdot \exp{\left\{\Rad(\beta_e)_i \cdot 2\eta\cdot \sum_{k = 0}^{t-1} {\Delta_\Gamma^k}\right\}} \cdot \exp{\left\{\mathbb{Q}^e_i\right\}}}}}.
        \end{aligned}
    \]
    We are going to analyze the convergences of two sequences $\{\Delta_\Gamma^t + \Delta_\Lambda^t\}$ and $\{\Delta_\Gamma^t - \Delta_\Lambda^t\}$. Notice that
    \[
        \begin{aligned}
             & \ \ \ \ \Delta_\Gamma^t + \Delta_\Lambda^t = \\& \sum_{e\in \mathcal{E}_{tr}}{\frac{1}{n_e m}\sum_{i=1}^{n_e}{\frac{\Rad(\beta_e)_i + \Rad(\alpha)_i}{1+\exp{\left\{\Rad(\alpha)_i \cdot 2\eta\cdot \sum_{k = 0}^{t-1} {\Delta_\Lambda^k}\right\}}\cdot \exp{\left\{\Rad(\beta_e)_i \cdot 2\eta\cdot \sum_{k = 0}^{t-1} {\Delta_\Gamma^k}\right\}} \cdot \exp{\left\{\mathbb{Q}^e_i\right\}}}}}, \\
             & \ \ \ \ \Delta_\Gamma^t - \Delta_\Lambda^t = \\& \sum_{e\in \mathcal{E}_{tr}}{\frac{1}{n_e m}\sum_{i=1}^{n_e}{\frac{\Rad(\beta_e)_i -\Rad(\alpha)_i}{1+\exp{\left\{\Rad(\alpha)_i \cdot 2\eta\cdot \sum_{k = 0}^{t-1} {\Delta_\Lambda^k}\right\}}\cdot \exp{\left\{\Rad(\beta_e)_i \cdot 2\eta\cdot \sum_{k = 0}^{t-1} {\Delta_\Gamma^k}\right\}} \cdot \exp{\left\{\mathbb{Q}^e_i\right\}}}}}.
        \end{aligned}
    \]
    We can further write these two terms as
    \[
        \begin{aligned}
            \Delta_\Gamma^t + \Delta_\Lambda^t = {} & \sum_{e\in \mathcal{E}_{tr}}{\frac{2}{n_e m}\sum_{\substack{i\in[n_e]   \\\rad(\beta_e)_i = +1\\ \rad(\alpha)_i=+1}} {\frac{1}{1+\exp{\left\{2\eta\cdot \sum_{k = 0}^{t-1} {(\Delta_\Gamma^k + \Delta_\Lambda^k)}\right\}} \cdot \exp{\left\{\mathbb{Q}^e_i\right\}}}}} \\
                                                    & -\sum_{e\in \mathcal{E}_{tr}}{\frac{2}{n_e m}\sum_{\substack{i\in[n_e]  \\\rad(\beta_e)_i =-1\\ \rad(\alpha)_i=-1}}{\frac{1}{1+\exp{\left\{-2\eta\cdot \sum_{k = 0}^{t-1} {(\Delta_\Gamma^k + \Delta_\Lambda^k)}\right\}} \cdot \exp{\left\{\mathbb{Q}^e_i\right\}}}}}, \\
            \Delta_\Gamma^t - \Delta_\Lambda^t ={}  & \sum_{e\in \mathcal{E}_{tr}}{\frac{2}{n_e m}\sum_{\substack{i\in[n_e]   \\\rad(\beta_e)_i = +1\\ \rad(\alpha)_i=-1}}{\frac{1}{1 + \exp{\left\{2\eta\cdot \sum_{k = 0}^{t-1} {(\Delta_\Gamma^k - \Delta_\Lambda^k)}\right\}} \cdot \exp{\left\{\mathbb{Q}^e_i\right\}}}}} \\
                                                    & - \sum_{e\in \mathcal{E}_{tr}}{\frac{2}{n_e m}\sum_{\substack{i\in[n_e] \\\rad(\beta_e)_i =-1\\ \rad(\alpha)_i=+1}}{\frac{1}{1+\exp{\left\{-2\eta\cdot \sum_{k = 0}^{t-1} {(\Delta_\Gamma^k -\Delta_\Lambda^k)}\right\}} \cdot \exp{\left\{\mathbb{Q}^e_i\right\}}}}}.
        \end{aligned}
    \]
    According to Lemma \ref{lem:erm_concent_Q}, for all $e\in\mathcal{E}_{tr}, i\in [n_e], \rho>0$, letting $\delta\triangleq \exp\{O(\underline{n}^{-1})\} - 1$, we have $1+\delta \geq \exp{\{\mathbb{Q}^e_i\}} \geq (1+\delta)^{-1}$ with probability at least $1-\rho$. Let $C^e_{j\ell} \triangleq |\{i\mid \Rad(\alpha)_i = j, \Rad(\beta_e)_i = \ell, i\in \mathcal{E}_e\}|$ for any $j\in\{\pm 1\},\ell\in \{\pm 1\}, e\in \mathcal{E}_{tr}$, and then define $\overline{C}_{j\ell} \triangleq \sum_{e\in \mathcal{E}_{tr}} {\frac{C^e_{j\ell}}{n_e}}$. We can upper bound and formulate $\Delta_\Gamma^t + \Delta_\Lambda^t$ and $\Delta_\Gamma^t - \Delta_\Lambda^t$ as
    \begin{align}
         & \ \ \ \ \Delta_\Gamma^t + \Delta_\Lambda^t \leq{} \nonumber                                                                                                                                                                                                                                          \\ & \frac{2}{m} \left(\frac{\overline{C}_{+1+1}}{1+\exp{\left\{2\eta\cdot \sum_{k = 0}^{t-1} {(\Delta_\Gamma^k + \Delta_\Lambda^k)}\right\}} \cdot (1+\delta)^{-1}} - \frac{\overline{C}_{-1-1}}{1+\exp{\left\{-2\eta\cdot \sum_{k = 0}^{t-1} {(\Delta_\Gamma^k + \Delta_\Lambda^k)}\right\}} \cdot (1+\delta)}\right) \nonumber  \\
         & ={} \frac{2}{m}\cdot \frac{\overline{C}_{+1+1} (1+\delta) - \overline{C}_{-1-1}\cdot \exp{\left\{2\eta\cdot \sum_{k = 0}^{t-1} {(\Delta_\Gamma^k + \Delta_\Lambda^k)}\right\}}}{1+\delta+\exp{\left\{2\eta\cdot \sum_{k = 0}^{t-1} {(\Delta_\Gamma^k + \Delta_\Lambda^k)}\right\}}}, \label{erm_eq4} \\
         & \ \ \ \ \Delta_\Gamma^t - \Delta_\Lambda^t \leq{} \nonumber                                                                                                                                                                                                                                          \\& \frac{2}{m}\left( \frac{\overline{C}_{-1+1}}{1 + \exp{\left\{2\eta\cdot \sum_{k = 0}^{t-1} {(\Delta_\Gamma^k - \Delta_\Lambda^k)}\right\}} \cdot (1+\delta)^{-1}} - \frac{\overline{C}_{+1-1}}{1+\exp{\left\{-2\eta\cdot \sum_{k = 0}^{t-1} {(\Delta_\Gamma^k -\Delta_\Lambda^k)}\right\}} \cdot (1+\delta)}\right) \nonumber \\
         & ={} \frac{2}{m}\cdot\frac{\overline{C}_{-1+1} (1+\delta) - \overline{C}_{+1-1}\cdot \exp{\left\{2\eta\cdot \sum_{k = 0}^{t-1} {(\Delta_\Gamma^k - \Delta_\Lambda^k)}\right\}}}{1+\delta + \exp{\left\{2\eta\cdot \sum_{k = 0}^{t-1} {(\Delta_\Gamma^k - \Delta_\Lambda^k)}\right\}}}.\label{erm_eq5}
    \end{align}
    Based on similar arguments, we can also establish lower bounds for these two terms,
    \begin{align}
        \Delta_\Gamma^t + \Delta_\Lambda^t \geq{} & \frac{2}{m}\cdot \frac{\overline{C}_{+1+1}  - \overline{C}_{-1-1}(1+\delta)\cdot \exp{\left\{2\eta\cdot \sum_{k = 0}^{t-1} {(\Delta_\Gamma^k + \Delta_\Lambda^k)}\right\}}}{1+\exp{\left\{2\eta\cdot \sum_{k = 0}^{t-1} {(\Delta_\Gamma^k + \Delta_\Lambda^k)}\right\}}\cdot(1+\delta)}, \label{erm_eq6} \\
        \Delta_\Gamma^t - \Delta_\Lambda^t \geq{} & \frac{2}{m}\cdot\frac{\overline{C}_{-1+1} - \overline{C}_{+1-1}(1+\delta)\cdot \exp{\left\{2\eta\cdot \sum_{k = 0}^{t-1} {(\Delta_\Gamma^k - \Delta_\Lambda^k)}\right\}}}{1 + \exp{\left\{2\eta\cdot \sum_{k = 0}^{t-1} {(\Delta_\Gamma^k - \Delta_\Lambda^k)}\right\}\cdot(1+\delta)}}.\label{erm_eq7}
    \end{align}
    The upper and lower bounds (\ref{erm_eq4}), (\ref{erm_eq5}), (\ref{erm_eq6}) and (\ref{erm_eq7}) imply that the convergences of $\{\Delta_\Gamma^t + \Delta_\Lambda^t\}$ and $\{\Delta_\Gamma^t - \Delta_\Lambda^t\}$ are determined by recursive equations of the form $\mathcal{Q}^t = \frac{C_1 - C_2\cdot \exp{\{\eta\sum_{k=0}^{t-1}{\mathcal{Q}^k}\}}}{1 + C_3\cdot\exp{\{\eta\sum_{k=0}^{t-1}{\mathcal{Q}^k}\}}}$. We first establish that with suitably chosen $\eta$, the sequences $\{\Delta_\Gamma^t + \Delta_\Lambda^t\}$ and $\{\Delta_\Gamma^t - \Delta_\Lambda^t\}$ are guaranteed to be positive. Observed that for the $\mathcal{Q}^t$-type recursive equation, the sign of $\mathcal{Q}^0$ is independent of $\eta$, and only determined by the constants $C_1, C_2, C_3$. At iteration $0$, (\ref{erm_eq6}) and (\ref{erm_eq7}) give
    \begin{align}
        \Delta_\Gamma^0 + \Delta_\Lambda^0 \geq{} & \frac{2}{m}\cdot \frac{\overline{C}_{+1+1}  - \overline{C}_{-1-1}(1+\delta)}{2+\delta}, \label{erm_eq8} \\
        \Delta_\Gamma^0 - \Delta_\Lambda^0 \geq{} & \frac{2}{m}\cdot\frac{\overline{C}_{-1+1} - \overline{C}_{+1-1}(1+\delta)}{2+\delta}.\label{erm_eq9}
    \end{align}
    To proceed, we need the following concentration lemma to control the deviations of the constants $\overline{C}_{+1+1}$, $\overline{C}_{+1-1}$, $\overline{C}_{-1+1}$ and $\overline{C}_{-1-1}$ from their expectations, whose proof is given in Appendix \ref{proof_erm_concent_C}.
    \begin{lemma}\label{lem:erm_concent_C}
        For $\rho > 0$, considering two environments and denoting $\epsilon_C \triangleq \sqrt{\frac{2\log{(16/\rho)}}{\underline{n}}}$, with probability at least $1-\rho$, we have
        \begin{equation} \label{erm_constant_bound}
            \begin{aligned}
                \left\lvert\overline{C}_{+1+1} - (1-\alpha)(2-\beta_1-\beta_2)\right\rvert & \leq \epsilon_C, \\
                \left\lvert\overline{C}_{+1-1} - (1-\alpha)(\beta_1+\beta_2)\right\rvert   & \leq \epsilon_C, \\
                \left\lvert\overline{C}_{-1+1} -  \alpha(2-\beta_1-\beta_2)\right\rvert    & \leq \epsilon_C, \\
                \left\lvert\overline{C}_{-1-1} -  \alpha(\beta_1+\beta_2)\right\rvert      & \leq \epsilon_C.
            \end{aligned}
        \end{equation}
    \end{lemma}
    Using Lemma \ref{lem:erm_concent_C}, with probability at least $1-\rho$, the constants $\overline{C}_{+1+1}$, $\overline{C}_{+1-1}$, $\overline{C}_{-1+1}$ and $\overline{C}_{-1-1}$ are close to their expectations.

    Based on our assumptions that
    \[
        \begin{aligned}
             & \alpha, \beta_1, \beta_2 < \frac{1-\epsilon_C-\delta (\frac{1}{4}+\frac{\epsilon_C}{2})}{2} &  & \text{($\alpha, \beta_1, \beta_2$ are sufficiently smaller than $\frac{1}{2}$)}, \\
             & \alpha > \frac{\beta_1+\beta_2}{2} + \epsilon_C + \frac{\delta (1+\epsilon_C)}{2}           &  & \text{($\alpha$ is sufficiently larger than $\frac{\beta_1+\beta_2}{2}$)},
        \end{aligned}
    \]
    it can be verified that with probability at least $1-2\rho$,
    $
        \Delta_\Gamma^0 + \Delta_\Lambda^0 > 0, \Delta_\Gamma^0 - \Delta_\Lambda^0 > 0.
    $

    Then, at iteration $1$, from (\ref{erm_eq6}) and (\ref{erm_eq7}), we see that as long as we require
    \[
        \eta < \min{\left\{\frac{1}{2(\Delta_\Gamma^0 + \Delta_\Lambda^0)} \log{\frac{\overline{C}_{+1+1}}{\overline{C}_{-1-1}(1+\delta)}}, \frac{1}{2(\Delta_\Gamma^0 -\Delta_\Lambda^0)} \log{\frac{\overline{C}_{-1+1}}{\overline{C}_{+1-1}(1+\delta)}}\right\}},
    \]
    it holds that $\Delta_\Gamma^1 + \Delta_\Lambda^1 > 0, \Delta_\Gamma^1 - \Delta_\Lambda^1 > 0$. By recursively applying this argument, we see the requirement for $\eta$ to ensure that $\Delta_\Gamma^t + \Delta_\Lambda^t > 0$ and $\Delta_\Gamma^t - \Delta_\Lambda^t > 0$ for any $t\in\{0, \ldots, T\}$ is
    \begin{equation}\label{erm_eq10}
        \eta < \min{\left\{\frac{1}{2\sum_{k = 0}^{T-1} {(\Delta_\Gamma^k + \Delta_\Lambda^k)}} \log{\frac{\overline{C}_{+1+1}}{\overline{C}_{-1-1}(1+\delta)}}, \frac{1}{2\sum_{k = 0}^{T-1} {(\Delta_\Gamma^k - \Delta_\Lambda^k)}} \log{\frac{\overline{C}_{-1+1}}{\overline{C}_{+1-1}(1+\delta)}}\right\}}.
    \end{equation}
    In other words, for the $\mathcal{Q}^t$-type recursive equation, as long as $\mathcal{Q}^0 \geq 0$, there always exists a sufficiently small $\eta$ to guarantee that the whole sequence $\{\mathcal{Q}^t\}$ is positive. From now on, we will focus on the case where the two sequences $\{\Delta_\Gamma^t + \Delta_\Lambda^t\}$ and $\{\Delta_\Gamma^t - \Delta_\Lambda^t\}$ decrease to an $\epsilon_\Delta>0$ error, i.e., $\min_{t\in \{0, \ldots, T\}} {\{\Delta_\Gamma^t + \Delta_\Lambda^t, \Delta_\Gamma^t - \Delta_\Lambda^t\}} = \epsilon_\Delta$.

    Then, we show that the two sequences $\{\Delta_\Gamma^t + \Delta_\Lambda^t\}$ and $\{\Delta_\Gamma^t - \Delta_\Lambda^t\}$ decrease monotonically, which thus leads to a more refined upper bound for $\eta$ at (\ref{erm_eq10}). Inspect the upper bounds (\ref{erm_eq4}), (\ref{erm_eq5}) at iteration $t+1$, which can be written as
    \[
        \begin{aligned}
             & \ \ \ \ \Delta_\Gamma^{t+1} + \Delta_\Lambda^{t+1} \leq{} \\ & \frac{2}{m}\cdot \frac{\overline{C}_{+1+1}  - \overline{C}_{-1-1}\cdot \exp{\left\{2\eta\cdot \sum_{k = 0}^{t-1} {(\Delta_\Gamma^k + \Delta_\Lambda^k)}\right\}}\cdot \exp{\left\{2\eta\cdot (\Delta_\Gamma^t + \Delta_\Lambda^t)\right\}}(1+\delta)^{-1}}{1+\exp{\left\{2\eta\cdot \sum_{k = 0}^{t-1} {(\Delta_\Gamma^k + \Delta_\Lambda^k)}\right\}}\cdot\exp{\left\{2\eta\cdot (\Delta_\Gamma^t + \Delta_\Lambda^t)\right\}} (1+\delta)^{-1}} \triangleq \spadesuit^{t+1}, \\
             & \ \ \ \ \Delta_\Gamma^{t+1} - \Delta_\Lambda^{t+1} \leq{} \\& \frac{2}{m}\cdot\frac{\overline{C}_{-1+1}  - \overline{C}_{+1-1}\cdot \exp{\left\{2\eta\cdot \sum_{k = 0}^{t-1} {(\Delta_\Gamma^k - \Delta_\Lambda^k)}\right\}}\cdot \exp{\left\{2\eta\cdot (\Delta_\Gamma^t - \Delta_\Lambda^t)\right\}}(1+\delta)^{-1}}{1 + \exp{\left\{2\eta\cdot \sum_{k = 0}^{t-1} {(\Delta_\Gamma^k - \Delta_\Lambda^k)}\right\}}\cdot \exp{\left\{2\eta\cdot (\Delta_\Gamma^t - \Delta_\Lambda^t)\right\}}(1+\delta)^{-1}}\triangleq \clubsuit^{t+1}.
        \end{aligned}
    \]
    Requiring that $\eta>\max{\left\{\frac{1}{\Delta_\Gamma^t + \Delta_\Lambda^t}\log{(1+\delta)}, \frac{1}{\Delta_\Gamma^t - \Delta_\Lambda^t}\log{(1+\delta)}\right\}}, \forall t\in \{0,\ldots, T\} \Rightarrow \eta > \epsilon_\Delta^{-1}\log{(1+\delta)}$, we have
    \[
        \begin{aligned}
            \spadesuit^{t+1} & < \frac{2}{m}\cdot \frac{\overline{C}_{+1+1}  - \overline{C}_{-1-1}\cdot \exp{\left\{2\eta\cdot \sum_{k = 0}^{t-1} {(\Delta_\Gamma^k + \Delta_\Lambda^k)}\right\}}\cdot \exp{\left\{2\eta\cdot (\Delta_\Gamma^t + \Delta_\Lambda^t)\right\}}(1+\delta)^{-1}}{1+\exp{\left\{2\eta\cdot \sum_{k = 0}^{t-1} {(\Delta_\Gamma^k + \Delta_\Lambda^k)}\right\}}\cdot(1+\delta)}   \\
                             & < \Delta_\Gamma^t + \Delta_\Lambda^t,                                                                                                                                                                                                                                                                                                                                      \\
            \clubsuit^{t+1}  & < \frac{2}{m}\cdot\frac{\overline{C}_{-1+1}  - \overline{C}_{+1-1}\cdot \exp{\left\{2\eta\cdot \sum_{k = 0}^{t-1} {(\Delta_\Gamma^k - \Delta_\Lambda^k)}\right\}}\cdot \exp{\left\{2\eta\cdot (\Delta_\Gamma^t - \Delta_\Lambda^t)\right\}}(1+\delta)^{-1}}{1 + \exp{\left\{2\eta\cdot \sum_{k = 0}^{t-1} {(\Delta_\Gamma^k - \Delta_\Lambda^k)}\right\}}\cdot (1+\delta)} \\
                             & < \Delta_\Gamma^t - \Delta_\Lambda^t,
        \end{aligned}
    \]
    where the last inequalities use the lower bounds (\ref{erm_eq6}) and (\ref{erm_eq7}).

    Based on the above discussion and (\ref{erm_eq10}), we can now clarify the requirements of $\eta$ for the sequences $\{\Delta_\Gamma^t + \Delta_\Lambda^t\}$ and $\{\Delta_\Gamma^t - \Delta_\Lambda^t\}$ to be positive and monotonically decreasing:
    \begin{equation}\label{erm_eta_require}
        \begin{aligned}
            \epsilon_\Delta^{-1}\log{(1+\delta)}  < \eta < \min\bigg\{ & \frac{m (2+\delta)}{4T (\overline{C}_{+1+1}(1+\delta)  - \overline{C}_{-1-1})} \log{\frac{\overline{C}_{+1+1}}{\overline{C}_{-1-1}(1+\delta)}},      \\
                                                                       & \frac{m(2+\delta)}{4T (\overline{C}_{-1+1}(1+\delta) - \overline{C}_{+1-1})} \log{\frac{\overline{C}_{-1+1}}{\overline{C}_{+1-1}(1+\delta)}}\bigg\},
        \end{aligned}
    \end{equation}
    which uses the upper bounds (\ref{erm_eq4}) and (\ref{erm_eq5}) at iteration $0$. The constants $\overline{C}_{+1+1}$, $\overline{C}_{+1-1}$, $\overline{C}_{-1+1}$ and $\overline{C}_{-1-1}$ can be substituted using the concentration bounds at (\ref{erm_constant_bound}) to generate an upper bound for $\eta$ that only involves $\alpha, \beta_1, \beta_2, m, \delta, T, \epsilon_C$. Here we omit the precise upper bound for clarity. Note that the left hand side of (\ref{erm_eta_require}) approaches $0$ if $\delta\rightarrow 0$, which means that there exists a constant choice of $\eta$ in (\ref{erm_eta_require}) if $\underline{n}$ is sufficiently large in Lemma \ref{lem:erm_concent_Q} and \ref{lem:erm_concent_C}.

    To conclude, in view of (\ref{erm_feature_update}), the convergences of the sequences $\{\Delta_\Gamma^t + \Delta_\Lambda^t\}$ and $\{\Delta_\Gamma^t - \Delta_\Lambda^t\}$ imply that $\Lambda^t_{j,r}$ and $\Gamma^t_{j,r}$ are converging, and the positive sequence $\{\Delta_\Gamma^t - \Delta_\Lambda^t\}$ indicates that the increment of the spurious feature $\Gamma^{t+1}_{j,r} - \Gamma^t_{j,r}$ is larger than that of the invariant feature $\Lambda^{t+1}_{j,r} - \Lambda^t_{j,r}$ at any iteration $t\in\{0,\ldots,T-1\}$.
\end{proof}
\subsubsection{Proof of Lemma \ref{lem:erm_concent_Q}}
\label{proof_erm_concent_Q}

First, we recall some concentration inequalities for sub-Gaussian random variables. Since $\boldsymbol{\xi}^e_i \sim \mathcal{N}(0, \sigma_p^2\cdot (\mathbf{I}_d - \mathbf{v}_1\mathbf{v}_1^\top - \mathbf{v}_2\mathbf{v}_2^\top))$, for $(i', e') \neq (i, e)$, using Bernstein's inequality for sub-exponential random variables, we have for sufficiently small $a\geq0$,
\[
    \begin{gathered}
        \textup{Pr}\left\{\lvert\langle \boldsymbol{\xi}^e_i, \boldsymbol{\xi}^{e'}_{i'}\rangle\rvert\geq a\right\} \leq 2 \exp\left\{- \frac{a^2}{4\sigma_p^4(d-2)}\right\}, \\
        \textup{Pr}\left\{\big\lvert\lVert \boldsymbol{\xi}^e_i\rVert_2^2 - \sigma_p^2(d-2)\big\rvert \geq a\right\} \leq 2 \exp\left\{- \frac{a^2}{512\sigma_p^4(d-2)}\right\}.
    \end{gathered}
\]
Moreover, for $\xi_r\sim \mathcal{N}(0, \sigma_0^2)$ (indicating the initial weights $\rvw^0_{j,r}$), the standard Gaussian tail gives
\[
    \textup{Pr}\left\{\left\lvert \frac{1}{m}\sum_{r=1}^m {\xi_r}\right\rvert\geq a\right\} \leq 2 \exp\left\{- \frac{ma^2}{2\sigma_0^2}\right\}.
\]

Denote $n\triangleq \sum_{e\in \mathcal{E}_{tr}}{n_e}, \underline{n} \triangleq \min_{e\in \mathcal{E}_{tr}} {n_e}$, by properly choosing $a$ for each tail bound and applying a union bound, we can conclude that for $\rho > 0$, with probability at least $1-\rho$, it holds that $\forall i,e, i', e', r$,
\[
    \begin{aligned}
         & \lvert\langle \boldsymbol{\xi}^e_i, \boldsymbol{\xi}^{e'}_{i'}\rangle\rvert \leq 2\sigma_p^2 \sqrt{(d-2)\log{\frac{8n^2}{\rho}}}, &  & \lVert\boldsymbol{\xi}^e_i\rVert_2^2  \leq \sigma_p^2(d-2) + 16\sigma_p^2 \sqrt{2(d-2)\log{\frac{8n}{\rho}}},                         \\
         & \left\lvert \frac{1}{m}\sum_{r=1}^m {\xi_r}\right\rvert \leq \sigma_0\sqrt{\frac{2}{m}\log{\frac{32m}{\rho}}},                    &  & \lvert\langle \boldsymbol{\xi}_r, \boldsymbol{\xi}^{e'}_{i'}\rangle\rvert \leq 2\sigma_p\sigma_0 \sqrt{(d-2)\log{\frac{16nm}{\rho}}}.
    \end{aligned}
\]

We start with bound the growth of $\Xi^{t,e}_{j,r,i}$. By bounding the update rule (\ref{erm_eq3}), with probability at least $1-\rho$, we have
\[
    \begin{aligned}
        \abs{\Xi^{t+1,e'}_{j,r,i'}} & \leq \abs{\Xi^{t, e'}_{j,r, i'}} + \frac{\eta}{m} \sum_{e\in \mathcal{E}_{tr}}{\frac{1}{n_e}\sum_{i=1}^{n_e}{\frac{1}{1+\exp\{y^e_i \hat{y}^e_i\}} \cdot \lvert\langle \boldsymbol{\xi}^e_i, \boldsymbol{\xi}^{e'}_{i'}\rangle\rvert }}                                                          \\
                                    & \leq \abs{\Xi^{t, e'}_{j,r, i'}} + \frac{\eta}{m} \sum_{e\in \mathcal{E}_{tr}}{\frac{1}{n_e}\sum_{i=1}^{n_e}{\lvert\langle \boldsymbol{\xi}^e_i, \boldsymbol{\xi}^{e'}_{i'}\rangle\rvert }}                                                                                                      \\
                                    & = \abs{\Xi^{0, e'}_{j,r, i'}} + (t+1)\cdot \frac{\eta}{m} \sum_{e\in \mathcal{E}_{tr}}{\frac{1}{n_e}\sum_{i=1}^{n_e}{\lvert\langle \boldsymbol{\xi}^e_i, \boldsymbol{\xi}^{e'}_{i'}\rangle\rvert }}                                                                                              \\
                                    & = \lvert\langle\boldsymbol{\xi}_r, \boldsymbol{\xi}^{e'}_{i'}\rangle\rvert + (t+1)\cdot\left( \frac{\eta}{m n_{e'}} \lVert \boldsymbol{\xi}^{e'}_{i'}\rVert^2_2 + \sum_{(i,e)\neq (i', e')}{\frac{\eta}{mn_e}\lvert\langle \boldsymbol{\xi}^e_i, \boldsymbol{\xi}^{e'}_{i'}\rangle\rvert}\right) \\
                                    & \leq 2\sigma_p\sigma_0 \sqrt{(d-2)\log{\frac{16nm}{\rho}}}                                                                                                                                                                                                                                       \\&\ \ + \frac{T\eta \sigma_p^2}{m \underline{n}}\left((d-2) + 16 \sqrt{2(d-2)\log{\frac{8n}{\rho}}} + 2 n \sqrt{(d-2)\log{\frac{8n^2}{\rho}}} \right).
    \end{aligned}
\]
Then, we can bound $\abs{\mathbb{Q}^e_i}$ as
\[
    \begin{aligned}
        \abs{\mathbb{Q}^e_i} & \leq 2\cdot \left\lvert\frac{1}{m}\sum_{r=1}^m {\xi_r}\right\rvert + 2\cdot\left\lvert\frac{1}{m}\sum_{r=1}^m { \xi_r}\right\rvert +  \frac{2}{m}\sum_{r=1}^m {\abs{\Xi^{t,e}_{j,r,i}}} \\
                             & \leq 4\sigma_0\sqrt{\frac{2}{m}\log{\frac{32m}{\rho}}} + 4\sigma_p\sigma_0 \sqrt{(d-2)\log{\frac{16nm}{\rho}}}                                                                          \\&\ \ + \frac{2T\eta \sigma_p^2}{m \underline{n}}\left((d-2) + 16 \sqrt{2(d-2)\log{\frac{8n}{\rho}}} + 2 n \sqrt{(d-2)\log{\frac{8n^2}{\rho}}} \right).
    \end{aligned}
\]
Thus, with sufficient small $\sigma_0, \sigma_p$, i.e.,
\[
    \begin{aligned}
        \sigma_0^2 & = O\left(\underline{n}^{-2} \log^{-1}{(m/\rho)}\right),                                                                             \\
        \sigma_p^2 & = O\left(\min\left\{d^{-1/2}\log^{-1/2}{(nm/\rho)}, T^{-1}\eta^{-1} m \left(d+ n\sqrt{d\log(n^2/\rho)}\right)^{-1} \right\}\right),
    \end{aligned}
\]
we ensured that $\abs{\mathbb{Q}^e_i} = O(\underline{n}^{-1})$.

\subsubsection{Proof of Lemma \ref{lem:erm_concent_C}}
\label{proof_erm_concent_C}

For $e\in \mathcal{E}_{tr}$, using Hoeffding's inequality, it holds that
\[
    \textup{Pr}\left\{\abs{\frac{1}{n_e} \sum_{i=1}^{n_e} {\mathbf{1}_{\{\Rad(\alpha)_i = +1, \Rad(\beta_e)_i = +1\}} - (1-\alpha)(1 - \beta_e)}} \geq a\right\} \leq 2\exp{\{-2a^2 n_e\}}.
\]
Considering two environments, using a union bound, we can conclude that
\[
    \textup{Pr}\left\{\abs{\overline{C}_{+1+1} - (1-\alpha)(2-\beta_1-\beta_2)} \leq a\right\} \geq 1- 4\exp{\left\{-\frac{a^2 \underline{n}}{2}\right\}}.
\]
Thus, for $\rho>0$, with probability at least $1-\frac{\rho}{4}$, we can conclude that
\[
    \abs{\overline{C}_{+1+1} - (1-\alpha)(2-\beta_1-\beta_2)} \leq \sqrt{\frac{2\log{(16/\rho)}}{\underline{n}}}.
\]
Using the above arguments for other constants $\overline{C}_{+1-1}$, $\overline{C}_{-1+1}$ and $\overline{C}_{-1-1}$, and applying a union bound, we obtain the desired results.

\subsubsection{ERM Feature Learning with Non-Linear Activation Functions}
\label{CH:FeAT:sec:erm_non_linear}

It was numerically observed that in the early stage of (stochastic) GD training, the learning dynamics of neural networks can be mimicked by training a simple linear model \citep{kalimeris2019sgd}. \citet{hu2020surprising} rigorously proved this phenomenon for training two-layer neural network with $\ell_2$ loss function in the Neural Tangent Kernel (NTK) region. We briefly summarize their results here: For a two-layer fully-connected neural network (with fixed second layer $\{v_r\}$):
\begin{equation}\label{erm_non_linear_fc}
    f_{FC}(\rmW, \rvx) \triangleq \frac{1}{\sqrt{m}} \sum_{r = 1}^{m} {v_r \psi\left(\rvw_r^\top \rvx / \sqrt{d}\right)},
\end{equation}
considering the $\ell_2$ training loss
$
    \ell_2(\hat{y}, y) \triangleq \frac{1}{2} (\hat{y} - y)^2
$ and the ERM objective $L_{\textup{ERM}}(\rmW) = \frac{1}{n} \sum_{i=1}^n{\ell_2\big(f_{FC}(\rmW, \rvx_i), y_i\big)}$, when using GD: $\rmW^{t+1} = \rmW^t - \eta \cdot \nabla L_{\textup{ERM}}(\rmW^t)$ to minimize the ERM objective, the following holds.

\begin{theorem}[Theorem 3.2 of \citep{hu2020surprising}]
    Let $\alpha_{nl} \in (0, \frac{1}{4})$ be a fixed constant, and $\psi(\cdot)$ be a smooth (with bounded first and second derivatives) or piece-wise linear activation function. Suppose that $n$ and $m$ satisfy $n = \Omega(d^{1+\alpha_{nl}})$ and $m = \Omega(d^{1+\alpha_{nl}})$. Suppose that $\eta \ll d$. Then there exists a universal constant $c > 0$ such that with high probability, for all $t = O(\frac{d}{\eta}\log{d})$ simultaneously, the learned neural network $f^t_{FC}$ and the linear model $f^t_{lin}$ (defined below) at iteration $t$ are close on average on the training data:
    \[
        \frac{1}{n} \sum_{i=1}^n {\big(f^t_{FC}(\rvx_i) - f^t_{lin}(\rvx_i)\big)^2} = O(d^{-\Omega(\alpha_{nl})}).
    \]
\end{theorem}

The linear model $f_{lin}(\boldsymbol{\beta}, \rvx) = \boldsymbol{\beta}^\top \boldsymbol{R}(\rvx) $ is a linear function of the transformed data $\boldsymbol{R}(\rvx) = \frac{1}{\sqrt{d}}\begin{bmatrix}
        \zeta \rvx \\ \nu
    \end{bmatrix}$, where $\zeta$ and $\nu$ are constants related to $\psi'$ and the dataset distribution (see (5) in \citep{hu2020surprising} for formal definitions).

We show that we can relate our data model to the dataset setup in \citep{hu2020surprising}
, and thus by analyzing the feature learning terms for the  linear model $f_{lin}(\boldsymbol{\beta}, \rvx)$ similar to the analysis\footnote{Note that when $\psi(x)=x$, our CNN model can be viewed as a linear model with re-parameterized weight matrices. Thus, the discussion in Appendix \ref{CH:FeAT:sec:proof_erm_feat} can be viewed as studying the feature learning terms for a linear model with logistic loss function.} in Appendix \ref{CH:FeAT:sec:proof_erm_feat}, we obtain similar results as in Theorem \ref{CH:FeAT:thm:erm_learn_feat_appdx} in the early stage of GD training, but with an error of $O(d^{-\Omega(\alpha_{nl})})$.

Recall that our CNN model is $f (\rmW, \rvx)\!=\!F_{+1}(\rmW_{+1}, \rvx) - F_{-1}(\rmW_{-1}, \rvx)$, where $F_{+1}(\rmW_{+1}, \rvx)$ and $F_{-1}(\rmW_{-1}, \rvx)$ are defined as follows:
\[
    \begin{aligned}
        F_j(\rmW_j, \rvx ) = \frac{1}{m}\sum_{r=1}^m \left[  {\psi}(\rvw_{j,r}^\top \rvx_1) +  {\psi}(\rvw_{j,r}^\top \rvx_2) \right], j\in\{\pm 1\}.
    \end{aligned}
\]
We can cast this CNN model into an instance of the two-layer fully connected neural network defined at (\ref{erm_non_linear_fc}) by specifying the values of $\{v_r = \pm \frac{1}{\sqrt{m}}\}$ and transforming the dataset as $\left\{\sqrt{d}\begin{bmatrix}
        y \cdot \Rad(\alpha) \cdot  \rvv_1 + y \cdot \Rad(\beta) \cdot  \rvv_2 \\0
    \end{bmatrix},
    \sqrt{d}\boldsymbol{\xi}\right\}$. Then by tuning the norms of $\rvv_1, \rvv_2$ and $\boldsymbol{\xi}$, we obtain a dataset that satisfies the input assumptions in \citep{hu2020surprising}. Note that this cast drops the shared variable of our CNN model and thus might lead to a slightly different training dynamic. To fix such gap, we can leverage Proposition~6.4.1 in \citep{hu2021understanding} for the early stage behavior of training a CNN model.

Based on the above ideas, to formalize the convergence results of the feature learning terms in the non-linear case, it remains to re-derive the analysis in Appendix \ref{CH:FeAT:sec:proof_erm_feat}
based on $\ell_2$ loss function, which follows a similar line of proofs and has a simpler dynamic.

\subsection{Proof for Theorem~\ref{CH:FeAT:thm:irmv1_not_learn}} \label{CH:FeAT:sec:proof_irmv1_not_learn}

\begin{theorem} [Restatement of Theorem \ref{CH:FeAT:thm:irmv1_not_learn}]
    \label{CH:FeAT:thm:irmv1_appdx}
    Consider training a CNN model with the same data as in Theorem~\ref{CH:FeAT:thm:erm_learn_feat}, define
    \begin{align*}
        \rvc(t) \triangleq \left[ C^1_\irml(\rmW,t),  C^2_\irml(\rmW,t), \cdots,  C^{|\envtrain|}_\irml(\rmW,t) \right],
    \end{align*} and $\lambda_0 =  \lambda_{\min}(\mathbf{H}^\infty)$, where we define
    \begin{align*}
        \mathbf{H}^\infty_{e,e'} \triangleq \frac{  1}{2m n_e n_{e'}} \sum_{i=1}^{n_e} \psi'(\langle \rvw_{j,r}(0) ,  \mathbf{x}^{e}_{1,i} \rangle ) \mathbf{x}^{e\top}_{1,i} \sum_{i'=1}^{n_{e'}} \psi'(\langle \rvw_{j,r}(0) ,  \rvx_{1,i'}^{e'} \rangle )   \rvx_{1,i'}^{e'}.
    \end{align*}
    Suppose that {activation function is smooth, $\psi'(0) \le \beta$, $|\psi'(x)-\psi'(x')|<\beta|x - x'|$ and Lipschitz $|\psi(0)| < L $, $|\psi(x)-\psi(x')|< L |x - x'|$}. Assume that dimension $d
        =  \Omega(\log(m/\delta))$, network width $m = \Omega(1/\delta)$, regularization factor $\lambda \ge 1/(\sigma_0 \sqrt{|\mathcal{E}_{tr}|}^3 )$, noise variance $\sigma_p = O(d^{-2})$, weight initial scale $\sigma_0 = O( \frac{ |\mathcal{E}_{tr} |^{7/2} \beta^3 L }{ d^{1/2}m^2\lambda^2_0 \log(1/\epsilon)} )$, then with probability at least $1-\delta$, after training time $T = \Omega \left( \frac{ \log(1/\epsilon)}{ \eta  \lambda \lambda_0 } \right) $, we have:
    \begin{align*}
        \| \rvc(T )\|_2 \le  \epsilon, \quad
        { \gamma^{inv}_{j,r}(T) = o(1), \quad \gamma^{spu}_{j,r}(T) = o(1)}.
    \end{align*}
\end{theorem}

Before proving Theorem \ref{CH:FeAT:thm:irmv1_appdx}, we first provide some useful lemmas as follows:

\begin{lemma} [\citep{understand_benign}] \label{lemma:data_innerproducts}
    Suppose that $\delta > 0$ and $d = \Omega( \log(4n / \delta) ) $. Then with probability at least $1 - \delta$,
    \begin{align*}
         & \sigma_p^2 d / 2\leq \| \boldsymbol{\xi}_i \|_2^2 \leq 3\sigma_p^2 d / 2
    \end{align*}
    for all $i,i'\in [n]$.
\end{lemma}

\begin{lemma} [\citep{understand_benign}]
    \label{lemma:initialization_norms} Suppose that $d \geq \Omega(\log(mn/\delta))$, $ m = \Omega(\log(1 / \delta))$. Then with probability at least $1 - \delta$,
    \begin{align*}
         & |\langle \rvw_{j,r}^{(0)}, \mathbf{v}_1 \rangle | \leq \sqrt{2 \log(8m/\delta)} \cdot \sigma_0 \| \mathbf{v}_1 \|_2,    \\
         & |\langle \rvw_{j,r}^{(0)}, \mathbf{v}_2 \rangle | \leq \sqrt{2 \log(8m/\delta)} \cdot \sigma_0 \| \mathbf{v}_2 \|_2,    \\
         & | \langle \rvw_{j,r}^{(0)}, \boldsymbol{\xi}_i \rangle | \leq 2\sqrt{ \log(8mn/\delta)}\cdot \sigma_0 \sigma_p \sqrt{d}
    \end{align*}
    for all $r\in [m]$,  $j\in \{\pm 1\}$ and $i\in [n]$.
\end{lemma}

\begin{lemma}\label{lem:weight_initial}
    Suppose that $\delta > 0$ and $d = \Omega( \log(4m / \delta) ) $. Then with probability at least $1 - \delta$, for all $r \in [m]$ and $j \in \{-1, 1\}$, we have
    \begin{align*}
         & \sigma_0^2 d / 2\leq \| \mathbf{w}_{j,r}(0) \|_2^2 \leq 3\sigma_0^2 d / 2.
    \end{align*}
\end{lemma}

\begin{proof} [Proof of Lemma~\ref{lem:weight_initial}] By Bernstein's inequality, with probability at least $1 - \delta / (2m)$ we have
    \begin{align*}
        \big| \| \mathbf{w}_{j,r}(0) \|_2^2 - \sigma_0^2 d \big| = O(\sigma_0^2 \cdot \sqrt{d \log(4m / \delta)}).
    \end{align*}
    Therefore, as long as $d = \Omega( \log(4m / \delta) )$, we have
    \begin{align*}
        \sigma_0^2 d /2  \leq \| \mathbf{w}_{j,r}(0) \|_2^2 \leq 3\sigma_0^2 d / 2.
    \end{align*}
\end{proof}

\begin{proof} [Proof of Theorem \ref{CH:FeAT:thm:irmv1_appdx}] The proof is by induction method.
    First we show the gradient flow of weights by \irml objective function (\ref{CH:FeAT:eq:irml_cnn}):
    \begin{align*}
        \frac{d \rvw_{j,r}(t)}{dt} & =   - \eta \cdot \nabla_{\rvw_{j,r}} L_{\mathrm{IRMv1}}(\rmW(t))                                                                                                                                                                                                                                                                                                                                                           \\
                                   & =   - \frac{\eta}{nm}
        \sum_{e \in \mathcal{E}_{\mathrm{tr}}} \sum_{i=1}^{n_e} \ell'_i(t)  {\psi}'(\langle \rvw_{j,r}(t) , y_i^e \rvv_i^e \rangle) \cdot j \rvv_i^e  - \frac{\eta}{nm}   \sum_{e \in \mathcal{E}_{\mathrm{tr}}} \sum_{i=1}^{n_e} \ell'_i(t)  {\psi}'(\langle \rvw_{j,r}(t) ,  \boldsymbol{\xi}_i \rangle) \cdot j y_i^e \boldsymbol{\xi}_i                                                                                                                     \\
                                   & \quad - \frac{2 \eta \lambda}{nm}    \sum_{e \in \mathcal{E}_{\mathrm{tr}}}  C_\irml^e \sum_{i=1}^{n_e} \ell''_{i}  \hat{y}_i^e {\psi}'(\langle \rvw_{j,r}(t) , y_i^e \rvv_i^e \rangle)   j y_i^e  \rv^e_i  - \frac{2 \eta \lambda}{nm}  \sum_{e \in \mathcal{E}_{\mathrm{tr}}}  C_\irml^e \sum_{i=1}^{n_e} \ell''_{i}  \hat{y}_i^e   {\psi}'(\langle \rvw_{j,r}(t) ,  \boldsymbol{\xi}_i\rangle )   j  \boldsymbol{\xi}_i \\
                                   & \quad - \frac{2 \eta \lambda}{nm}  \sum_{e \in \mathcal{E}_{\mathrm{tr}}}  C_\irml^e \sum_{i=1}^{n_e} \ell'_i(t)  {\psi}'(\langle \rvw_{j,r}(t) , y_i^e \rvv_i^e \rangle) j \rvv_i^e   - \frac{2 \eta \lambda}{nm}   \sum_{e \in \mathcal{E}_{\mathrm{tr}}}  C_\irml^e \sum_{i=1}^{n_e} \ell'_i(t)  {\psi}'(\langle \rvw_{j,r}(t) ,  \boldsymbol{\xi}_i \rangle) j y_i^e \boldsymbol{\xi}_i                                \\
                                   & = - \frac{\eta}{nm}
        \sum_{e \in \mathcal{E}_{\mathrm{tr}}}   (1+ 2\lambda C_\irml^e(t)) \sum_{i=1}^{n_e} \ell'_i(t)  {\psi}'(\langle \rvw_{j,r}(t) , y_i^e \rvv_i^e \rangle) \cdot j \rvv_i^e                                                                                                                                                                                                                                                                               \\
                                   & \quad   - \frac{\eta}{nm}   \sum_{e \in \mathcal{E}_{\mathrm{tr}}}  (1+ 2\lambda C_\irml^e(t)) \sum_{i=1}^{n_e} \ell'_i(t)  {\psi}'(\langle \rvw_{j,r}(t) ,  \boldsymbol{\xi}_i\rangle ) \cdot j y_i^e \boldsymbol{\xi}_i                                                                                                                                                                                                  \\
                                   & \quad - \frac{2 \eta \lambda}{nm}   \sum_{e \in \mathcal{E}_{\mathrm{tr}}} C_\irml^e \sum_{i=1}^{n_e} \ell''_{i}  \hat{y}_i^e {\psi}'(\langle \rvw_{j,r}(t) , y_i^e \rvv_i^e \rangle)   j y_i^e  \rv^e_i   - \frac{ 2 \eta \lambda}{nm}   \sum_{e \in \mathcal{E}_{\mathrm{tr}}}  C_\irml^e \sum_{i=1}^{n_e} \ell''_{i}  \hat{y}_i^e  {\psi}'(\langle \rvw_{j,r}(t) ,  \boldsymbol{\xi}_i\rangle )  j  \boldsymbol{\xi}_i,
    \end{align*}
    where $C_\irml^e = \frac{1}{n_e} \sum_{i=1}^{n_e} \ell_i'^e \hat{y}_i^e y_i^e$ and $\rvv_i^e = \textrm{Rad}(\alpha)_i \cdot \rvv_1 + \textrm{Rad}(\beta_e)_i \cdot \rvv_2$. Note that $\ell''$ has the opposite sign to $\ell'$.

    Then we look at the dynamics of $C^e_\irml(t)$ according to the gradient flow update rule:
    \begin{align*}
        \frac{d C^e_\irml(\rmW,t)}{d t} & = \sum_{j = \pm 1} \sum_{r=1}^m \left  \langle \frac{\partial C^e_\irml(\rmW,t)}{ \partial \rvw_{j,r}(t) }, \frac{d \rvw_{j,r} (t)}{d t } \right \rangle                                                                  \\
                                        & = \sum_{e'} 2 \lambda C^{e'}_\irml(\rmW,t) \sum_{j} \sum_{r=1}^m \left  \langle \frac{\partial C^e_\irml(\rmW,t)}{ \partial \rvw_{j,r}(t) },\frac{\partial C^{e'}_\irml(\rmW,t)}{ \partial \rvw_{j,r}(t) } \right \rangle \\
                                        & \quad + \sum_{j = \pm 1} \sum_{r=1}^m \left  \langle \frac{\partial C^e_\irml(\rmW,t)}{ \partial \rvw_{j,r}(t) },\frac{\partial L(\rmW,t)}{ \partial \rvw_{j,r}(t) }\right \rangle                                        \\
                                        & = 2 \lambda \sum_{e'}  C^{e'}_\irml(\rmW,t) \cdot \rmH_{e,e'} (t) + \mathbf{g}_e(t),
    \end{align*}
    where we define $\rmH_{e,e'} (t) = \sum_{j} \sum_{r=1}^m \left  \langle \frac{\partial C^e_\irml(\rmW,t)}{ \partial \rvw_{j,r}(t) },\frac{\partial C^{e'}_\irml(\rmW,t)}{ \partial \rvw_{j,r}(t) } \right \rangle $ and $\rvg_e(t) = \sum_{j = \pm 1} \sum_{r=1}^m \left  \langle \frac{\partial C^e_\irml(\rmW,t)}{ \partial \rvw_{j,r}(t) },\frac{\partial L(\rmW,t)}{ \partial \rvw_{j,r}(t) }\right \rangle $. Thus $\rmH(t) $ is an $ |\envtrain|\times |\envtrain| $ matrix. We can write the dynamics of $\rvc(t) = \left[ C^1_\irml(\rmW,t),  C^2_\irml(\rmW,t), \cdots,  C^{|\envtrain|}_\irml(\rmW,t) \right]  $ in a compact way:
    \begin{align} \label{CH:FeAT:eq:c_dynamimcs}
        \frac{d \rvc (t)}{ d t} = 2\lambda \cdot \rmH(t) \rvc(t) + \rvg(t).
    \end{align}
    Our next step is to show $\rmH(t)$ is stable during training. To proceed with the analysis, we write down the expression for $ \frac{\partial C^{e}_\irml(\rmW,t)}{ \partial \rvw_{j,r}(t) }  \in \mathbb{R}^d$:
    \begin{align*}
        \frac{\partial C^{e}_\irml(\rmW(t))}{ \partial \rvw_{j,r}(t) } & = \frac{1  }{n_e m}
        \sum_{i=1}^{n_e} \ell'_i(t)  {\psi}'(\langle \rvw_{j,r}(t) , y_i^e \rvv_i^e \rangle) \cdot j \rvv_i^e   + \frac{1  }{n_e m}  \sum_{i=1}^{n_e} \ell'_i(t)  {\psi}'(\langle \rvw_{j,r}(t) ,  \boldsymbol{\xi}_i \rangle ) \cdot j y_i^e \boldsymbol{\xi}_i                                                                                                                            \\
                                                                       & \quad + \frac{ 1  }{n_e m}     \sum_{i=1}^{n_e} \ell''_{i} \hat{y}_i^e  {\psi}'(\langle \rvw_{j,r}(t) , y_i^e \rvv_i^e \rangle) \cdot j y_i^e  \rvv_i^e   + \frac{1 }{n_e m}   \sum_{i=1}^{n_e} \ell''_{i} \hat{y}_i^e  {\psi}'(\langle \rvw_{j,r}(t) ,  \boldsymbol{\xi}_i \rangle ) \cdot j  \boldsymbol{\xi}_i.
    \end{align*}
    When we consider {non}-linear activation function ${\psi}(x)$, the entry of matrix $\rmH (t) $ can be computed as follows:
    \begin{align*}
         & \rmH_{e,e'} (t) =   \sum_{j} \sum_{r=1}^m \left  \langle \frac{\partial C^e_\irml(\rmW,t)}{ \partial \rvw_{j,r}(t) },\frac{\partial C^{e'}_\irml(\rmW,t)}{ \partial \rvw_{j,r}(t) } \right \rangle                                                                                                                                                                                                                                                                               \\
         & =   \sum_{j} \sum_{r=1}^m  \left(\frac{ 1  }{n_{e} m} \right)  \left(\frac{1  }{n_{e'} m} \right) \bigg[ \sum_{i=1}^{n_e} \ell'_i(t) {\psi'} j \rvv^{e\top}_i \sum_{i'=1}^{n_{e'}} \ell'_{i'}(t){\psi'}   j \rvv_{i'}^{e'}     +  \sum_{i=1}^{n_e}  {\psi'} \ell''_i(t) \hat{y}_i^e(t)   j y_i^e \rvv^{e\top}_i \sum_{i'=1}^{n_{e'}}  {\psi'} \ell''_{i'}(t) \hat{y}_{i'}^{e'}(t)   j y_{i'}^{e'} \rvv_{i'}^{e'}  \bigg]                                                         \\
         & + \sum_{j} \sum_{r=1}^m  \left(\frac{ 1  }{n_{e} m} \right)  \left(\frac{ 1  }{n_{e'} m} \right) \bigg[ \sum_{i=1}^{n_e} {\psi'} \ell''_i(t) \hat y_i^e(t)   j y_i^e \rvv^{e\top}_i \sum_{i'=1}^{n_{e'}}  {\psi'}  \ell'_{i'}(t)  j \rvv_{i'}^{e'}   +  \sum_{i=1}^{n_e}  {\psi'} \ell'_i(t)  j  \rvv^{e\top}_i \sum_{i'=1}^{n_{e'}}   {\psi'}  \ell''_{i'}(t)  \hat{y}_{i'}^{e'}(t)   j \rvv_{i'}^{e'}  \bigg]                                                                  \\
         & + \sum_{j} \sum_{r=1}^m  \left(\frac{ 1 }{n_{e} m} \right)  \left(\frac{ 1  }{n_{e'} m} \right) \bigg[ \sum_{i=1}^{n_e}  {\psi'}  \ell'_i(t)   j y^e_i \boldsymbol{\xi}^{e\top}_i \sum_{i'=1}^{n_{e'}}  {\psi'}  \ell'_{i'}(t)   j y^{e'}_{i'} \boldsymbol{\xi}_{i'}^{e'}   +  \sum_{i=1}^{n_e}  {\psi'}  \ell''_i(t)  \hat{y}_i^e(t) j  \boldsymbol{\xi}^{e\top}_i \sum_{i'=1}^{n_{e'}}  {\psi'}  \ell''_{i'}(t)  \hat{y}_{i'}^{e'}(t)   j  \boldsymbol{\xi}_{i'}^{e'}  \bigg]  \\
         & + \sum_{j} \sum_{r=1}^m  \left(\frac{ 1  }{n_{e} m} \right)  \left(\frac{ 1  }{n_{e'} m} \right) \bigg[ \sum_{i=1}^{n_e}  {\psi'}  \ell''_{i}(t) \hat{y}^{e}_{i}   j  \boldsymbol{\xi}^{e\top}_i \sum_{i'=1}^{n_{e'}}  {\psi'}  \ell'_{i'}(t)   j y^{e'}_{i'} \boldsymbol{\xi}_{i'}^{e'}  + \sum_{i=1}^{n_e}  {\psi'}  y^e_i \ell'_i(t)   j \boldsymbol{\xi}^{e\top}_i \sum_{i'=1}^{n_{e'}}   {\psi'}  \ell''_{i'}(t)     j \boldsymbol{\xi}_{i'}^e \hat{y}^{e'}_{i'}(t)  \bigg] \\
         & \triangleq \rmH^1_{e,e'} (t) + \rmH^2_{e,e'} (t) + \rmH^3_{e,e'} (t) + \rmH^4_{e,e'} (t) + \rmH^5_{e,e'} (t) + \rmH^6_{e,e'} (t) + \rmH^7_{e,e'} (t) + \rmH^8_{e,e'} (t).
    \end{align*}
    The matrix $\mathbf{H}$ is composed of eight elements. In addition, we define
    \begin{align*}
        \rmH^{1,\infty}_{e,e'} & = \sum_{j} \sum_{r=1}^m  \left(\frac{ 1 }{n_{e} m} \right)  \left(\frac{ 1  }{n_{e'} m} \right) \left[ \sum_{i=1}^{n_e}  -\frac{1}{2} {\psi}'(\langle \rvw_{j,r}(0) ,  \mathbf{v}^e_i \rangle )
        j \rvv^{e\top}_i \sum_{i'=1}^{n_{e'}}  -\frac{1}{2} {\psi}'(\langle \rvw_{j,r}(0) , \mathbf{v}^{e'}_{i'} \rangle )  j \rvv_{i'}^{e'}  \right]                                                                                                         \\
                               & =  \frac{ 1}{2 m n_e n_{e'}} \sum_{i=1}^{n_e} {\psi}'(\langle \rvw_{j,r}(0) ,  \mathbf{v}^e_i \rangle ) \rvv^{e\top}_i \sum_{i'=1}^{n_{e'}} {\psi}'(\langle \rvw_{j,r}(0) ,  \mathbf{v}^{e'}_{i'} \rangle )  \rvv_{i'}^{e'}.
    \end{align*}
    Then we can show that:
    \begin{align*}
                   & \left |  \rmH^1_{e,e'}(t) -  \rmH^{1,\infty}_{e,e'} \right |                                                                                                                                                                                                                                                                                                                              \\
        =          & \frac{2}{ m n_e n_{e'}} \left| \sum_{i=1}^{n_e} {\psi}'(t) \ell'_i(t) \rvv^{e\top}_i \sum_{i'=1}^{n_{e'}}  {\psi}'(t)  \ell'_{i'}(t)  \rvv_{i'}^{e'} -  \sum_{i=1}^{n_e} \frac{1}{2}   {\psi}'(0)\rvv^{e\top}_i \sum_{i'=1}^{n_{e'}}  \frac{1}{2} {\psi}'(0)  \rvv_{i'}^{e'} \right|                                                                                                      \\
        =          & \frac{2}{ m n_e n_{e'}} \bigg| \sum_{i=1}^{n_e}  {\psi}'(t) \ell'_i(t) \rvv^{e\top}_i \sum_{i'=1}^{n_{e'}}  {\psi}'(t)  \ell'_{i'}(t)  \rvv_{i'}^{e'} - \sum_{i=1}^{n_e}  {\psi}'(0) \ell'_i(t) \rvv^{e\top}_i \sum_{i'=1}^{n_{e'}}  {\psi}'(0)  \ell'_{i'}(t)  \rvv_{i'}^{e'}                                                                                                            \\
                   & \quad +  \sum_{i=1}^{n_e}  {\psi}'(0) \ell'_i(t) \rvv^{e\top}_i \sum_{i'=1}^{n_{e'}}  {\psi}'(0)  \ell'_{i'}(t)  \rvv_{i'}^{e'} - \sum_{i=1}^{n_e} \frac{1}{2}   {\psi}'(0)\rvv^{e\top}_i \sum_{i'=1}^{n_{e'}}  \frac{1}{2}  {\psi}'(0)  \rvv_{i'}^{e'} \bigg|                                                                                                                            \\
        \le        & \frac{2 }{ m n_e n_{e'}} \left| \sum_{i=1}^{n_e} {\psi}'(t) \ell'_i(t) \rvv^{e\top}_i \sum_{i'=1}^{n_{e'}}  {\psi}'(t) \ell'_{i'}(t)  \rvv_{i'}^{e'} -  \sum_{i=1}^{n_e}   {\psi}'(0) \ell'_{i}(t) \rvv^{e\top}_i \sum_{i'=1}^{n_{e'}}  {\psi}'(t) \ell'_{i'}(t) \rvv_{i'}^{e'} \right|                                                                                                   \\
                   & \quad  +   \frac{2 }{ m n_e n_{e'}} \left| \sum_{i=1}^{n_e} {\psi}'(0) \ell'_i(t) \rvv^{e\top}_i \sum_{i'=1}^{n_{e'}}  {\psi}'(t) \ell'_{i'}(t)  \rvv_{i'}^{e'} -  \sum_{i=1}^{n_e}   {\psi}'(0) \ell'_{i}(t) \rvv^{e\top}_i \sum_{i'=1}^{n_{e'}}  {\psi}'(0) \ell'_{i'}(t) \rvv_{i'}^{e'} \right|                                                                                        \\
                   & \quad +     \frac{2 }{ m n_e n_{e'}} \left| \sum_{i=1}^{n_e} {\psi}'(0) \ell'_i(t) \rvv^{e\top}_i \sum_{i'=1}^{n_{e'}} {\psi}'(0) \ell'_{i'}(t)  \rvv_{i'}^{e'} -  \sum_{i=1}^{n_e}   {\psi}'(0) \ell'_{i} \rvv^{e\top}_i \sum_{i'=1}^{n_{e'}}   {\psi}'(0) \frac{1}{2}  \rvv_{i'}^{e'} \right|                                                                                           \\
                   & \quad + \frac{2  }{ m n_e n_{e'}} \left| \sum_{i=1}^{n_e}  {\psi}'(0) \ell'_i(t) \rvv^{e\top}_i \sum_{i'=1}^{n_{e'}}  {\psi}'(0) \frac{1}{2} \rvv_{i'}^{e'} -  \sum_{i=1}^{n_e}  {\psi}'(0) \frac{1}{2} \rvv^{e\top}_i \sum_{i'=1}^{n_{e'}}   {\psi}'(0) \frac{1}{2}  \rvv_{i'}^{e'} \right|                                                                                              \\
        \le        & \frac{2}{ m n_e n_{e'}} \left| \sum_{i=1}^{n_e} ( {\psi}'(t)- {\psi}'(0)) \ell'_i(t) \rvv^{e\top}_i \sum_{i'=1}^{n_{e'}}  \ell'_{i'}(t)   {\psi}'(t)\rvv_{i'}^{e'} \right|  + \frac{2 }{ m n_e n_{e'}} \left| \sum_{i=1}^{n_e}  {\psi}'(t) \ell'_i(t)   \rvv^{e\top}_i \sum_{i'=1}^{n_{e'}} ( {\psi}'(t)- {\psi}'(0)) \frac{1}{2} \rvv_{i'}^{e'}  \right|                                 \\
                   & + \frac{2}{ m n_e n_{e'}} \left| \sum_{i=1}^{n_e}    {\psi}'(0) \ell'_i(t) \rvv^{e\top}_i \sum_{i'=1}^{n_{e'}}  {\psi}'(0) \left( \ell'_{i'}(t)  + \frac{1}{2} \right)\rvv_{i'}^{e'}  \right|   + \frac{2  }{ m n_e n_{e'}} \left| \sum_{i=1}^{n_e} {\psi}'(0) \left( \ell'_i(t) + \frac{1}{2} \right) \rvv^{e\top}_i \sum_{i'=1}^{n_{e'}} {\psi}'(0) \frac{1}{2} \rvv_{i'}^{e'}  \right| \\
        \triangleq & I_1 + I_2 + I_3 + I_4
    \end{align*}
    where we calculate each item as follows:
    \begin{align*}
        I_1 & = \frac{2}{ m n_e n_{e'}} \left| \sum_{i=1}^{n_e} ( {\psi}'(t)- {\psi}'(0)) \ell'_i(t) \rvv^{e\top}_i \sum_{i'=1}^{n_{e'}}  \ell'_{i'}(t)   {\psi}'(t)\rvv_{i'}^{e'} \right|                                                                         \\
            & \overset{(a)} \le  \frac{2}{ m n_e n_{e'}} \left| \sum_{i=1}^{n_e}   \beta \| \mathbf{w}_{j,r}(t) - \mathbf{w}_{j,r}(0) \|_2  \| \rvv^{e}_i \|_2 \ell'_i(t)  \rvv^{e\top}_i \sum_{i'=1}^{n_{e'}}  \ell'_{i'}(t)   {\psi}'(t)\rvv_{i'}^{e'}   \right| \\
            & \overset{(b)}  \le  \frac{2 \beta^2}{ m n_e n_{e'}}   \sum_{i=1}^{n_e}   \| \mathbf{w}_{j,r}(t) - \mathbf{w}_{j,r}(0) \|_2  \| \rvv^{e}_i \|^2_2 \sum_{i'=1}^{n_{e'}}  \| \mathbf{w}_{j,r}(t) \|_2  \| \rvv_{i'}^{e'} \|^2_2                         \\
            & \overset{(c)}  \le \frac{32 \beta^2 R(R+ \frac{3}{2} \sigma_0 d )}{ m},
    \end{align*}
    where we have used $R \triangleq \| \mathbf{w}_{j,r}(t)  \|_2 $. Besides, inequality (a) results from applying the smoothness property of the activation function and Cauchy-Schwarz inequality; inequality (b) is by smoothness property of the activation function and Cauchy-Schwarz inequality. Besides, we have used $|\ell^e_i| \le 1$ for all $i \in n_e$ and $ e\in\envall$; inequality (c) is by the fact that $\| \mathbf{v}^e_i \|_2 \le 2 $ for all $i \in n_e$ and $ e\in\envall$ and Lemma \ref{lem:weight_initial}.

    Similarly, we calculate the upper bound for $I_2$ as follows:
    \begin{align*}
        I_2 & = \frac{2 }{ m n_e n_{e'}} \left| \sum_{i=1}^{n_e}  {\psi}'(t) \ell'_i(t)   \rvv^{e\top}_i \sum_{i'=1}^{n_{e'}} ( {\psi}'(t)-  {\psi}'(0)) \frac{1}{2} \rvv_{i'}^{e'}  \right| \\
            & \le \frac{32 \beta^2 R(R+ \frac{3}{2} \sigma_0 d )}{ m}.
    \end{align*}
    Next, we give the upper bound of $I_3$:
    \begin{align*}
        I_3 & = \frac{2}{ m n_e n_{e'}} \left| \sum_{i=1}^{n_e}    {\psi}'(0) \ell'_i(t) \rvv^{e\top}_i \sum_{i'=1}^{n_{e'}}  {\psi}'(0) \left( \ell'_{i'}(t)  + \frac{1}{2} \right)\rvv_{i'}^{e'}  \right|                                                                                         \\
            & \le  \frac{2}{ m n_e n_{e'}}  \left| \sum_{i=1}^{n_e}  \beta \| \mathbf{w}_{j,r}(0) \|_2 \| \rvv^{e}_i \|_2 \ell'_i(t) \rvv^{e\top}_i \sum_{i'=1}^{n_{e'}}\beta \| \mathbf{w}_{j,r}(0) \|_2  \|\rvv_{i'}^{e'} \|_2 \left( \ell'_{i'}(t)  + \frac{1}{2} \right)\rvv_{i'}^{e'}  \right| \\
            & \le \frac{64 \beta^2 L R (\frac{3}{2} \sigma_0 d )^2}{ m},
    \end{align*}
    where we have used $\gamma$ which is defined as follows:
    \begin{align*}
        |\hat y^e_i(t)| & =  \left|  \frac{1}{m} \sum_j \sum_{r=1}^m \left[  {\psi}(\rvw_{j,r}^\top(t) \rvx_1) +  {\psi}(\rvw_{j,r}^\top(t) \rvx_2) \right] \right | \\
                        & \overset{(a)}\le  2 L R,
    \end{align*}
    where inequality (a) is by the Lipschitz property of non-linear activation function and we have used the bound for $\ell'_i(t) + \frac{1}{2}$:
    \begin{align*}
        \left| \ell'_i(t) + \frac{1}{2} \right| & = \left|  -  \frac{ \exp(-y_i^e \cdot f(\rmW, \rvx_i,t))}{1 + \exp(-y_i^e \cdot f(\rmW, \rvx_i,t))} + \frac{1}{2} \right|                        \\
                                                & =  \left|  \frac{1}{2} -  \frac{ 1}{1 + \exp(y_i^e \cdot f(\rmW, \rvx_i,t))}  \right|                                                            \\
                                                & \le  \max \left \{ \left|  \frac{1}{2} -  \frac{ 1}{1 + \exp(2LR)}  \right|, \left|  \frac{1}{2} -  \frac{ 1}{1 + \exp(-2LR)}  \right|  \right\} \\          & \le \max \left \{ \left|  \frac{1}{2} -  \frac{ 1}{2 + \frac{7}{4}2LR}  \right|, \left|  \frac{1}{2} -  \frac{ 1}{2 -2LR}  \right|  \right\} = \Theta(LR).
    \end{align*}
    and we provide the bound of $\ell''_i(t) - \frac{1}{4}$:
    \begin{align*}
        \left| \ell''_i(t) - \frac{1}{4} \right| & = \left|   \frac{ \exp(-y_i^e \cdot f(\rmW, \rvx_i,t))}{ (1 + \exp(-y_i^e \cdot f(\rmW, \rvx_i,t)))^2} - \frac{1}{4} \right| \\
                                                 & =  \left|   \frac{1}{ \exp(y_i^e \cdot f(\rmW, \rvx_i,t)) +  2 + \exp(-y_i^e \cdot f(\rmW, \rvx_i,t))} - \frac{1}{4} \right| \\
                                                 & \le   \left|  \frac{1}{4} -  \frac{ 1}{2 + 2 \exp((2LR)^2/2) }  \right|  = \Theta((LR)^2).
    \end{align*}
    Similarly, we give the upper bound of $I_4$:
    \begin{align*}
        I_4 & = \frac{2  }{ m n_e n_{e'}} \left| \sum_{i=1}^{n_e} {\psi}'(0) \left( \ell'_i(t) + \frac{1}{2} \right) \rvv^{e\top}_i \sum_{i'=1}^{n_{e'}} {\psi}'(0) \frac{1}{2} \rvv_{i'}^{e'}  \right|                                                                               \\
            & \le  \frac{2}{ m n_e n_{e'}}  \left| \sum_{i=1}^{n_e}  \beta \| \mathbf{w}_{j,r}(0) \|_2 \| \rvv^{e}_i \|_2 (\ell'_i(t)+\frac{1}{2}) \rvv^{e\top}_i \sum_{i'=1}^{n_{e'}}\beta \| \mathbf{w}_{j,r}(0) \|_2  \|\rvv_{i'}^{e'} \|_2   \frac{1}{2}  \rvv_{i'}^{e'}  \right| \\
            & \le \frac{64 \beta^2 LR \gamma(\frac{3}{2} \sigma_0 d )^2}{ m}.
    \end{align*}
    Together, we obtain the upper bound for $\left |  \rmH^1_{e,e'}(t) -  \rmH^{1,\infty}_{e,e'} \right | $:
    \begin{align*}
        \left |  \rmH^1_{e,e'}(t) -  \rmH^{1,\infty}_{e,e'} \right |  \le \frac{64 \beta^2 R(R+ \frac{3}{2} \sigma_0 d )}{ m} + \frac{128 \beta^2 LR (\frac{3}{2} \sigma_0 d )^2}{ m}.
    \end{align*}

    Then we calculate the upper bound for the residual terms:
    \begin{align*}
        \left |  \rmH^2_{e,e'}(t)   \right | & = \left | \sum_{j} \sum_{r=1}^m  \left(\frac{ 1  }{n_{e} m} \right)  \left(\frac{1  }{n_{e'} m} \right)  \sum_{i=1}^{n_e}  {\psi'} \ell''_i(t) \hat{y}_i^e(t)   j y_i^e \rvv^{e\top}_i \sum_{i'=1}^{n_{e'}}  {\psi'} \ell''_{i'}(t) \hat{y}_{i'}^{e'}(t)   j y_{i'}^{e'} \rvv_{i'}^{e'} \right |   \\
                                             & = \frac{2  }{ m n_e n_{e'}} \left| \sum_{i=1}^{n_e} \psi'(t) \ell''_i(t)  \hat y_i^e(t)  j y_i^e \rvv^{e\top}_i \sum_{i'=1}^{n_{e'}} \psi'(t) \ell''_{i'}(t)  \hat y_{i'}^{e'}(t)  j y_{i'}^{e'} \rvv_{i'}^{e'} \right|                                                                            \\
                                             & \overset{(a)} \le \frac{2\beta^2 }{ m n_e n_{e'}} \left| \sum_{i=1}^{n_e} \| \mathbf{w}_{j,r}(t) \|_2 \|\rvv^{e}_i \|_2 \ell''_i(t)  \hat y_i^e(t) \rvv^{e\top}_i \sum_{i'=1}^{n_{e'}} \| \mathbf{w}_{j,r}(t) \|_2 \|\rvv^{e'}_{i'} \|_2  \ell''_{i'}(t)  \hat y_{i'}^e(t)  \rvv_{i'}^{e'} \right| \\
                                             & \overset{(b)} \le \frac{128 \beta^2  L^2 R^4}{ m },
    \end{align*}
    where inequality (a) is by the smoothness property of the activation function and Cauchy-Schwarz inequality, and inequality (b) is by triangle inequality and the fact that $\| \mathbf{v}^e_i \|_2 \le 2 $ for all $i \in n_e$ and $ e\in\envall$, and $|\ell''_i| \le 1$ for all $i \in [n]$ and Lemma \ref{lem:weight_initial}. Similarly, we further provide the upper bound of residual terms:
    \begin{align*}
        \left |  \rmH^3_{e,e'}(t)   \right | & = \left| \sum_{j} \sum_{r=1}^m  \left(\frac{ 1  }{n_{e} m} \right)  \left(\frac{ 1  }{n_{e'} m} \right) \sum_{i=1}^{n_e} {\psi'} \ell''_i(t) \hat y_i^e(t)   j y_i^e \rvv^{e\top}_i \sum_{i'=1}^{n_{e'}}  {\psi'}  \ell'_{i'}(t)  j \rvv_{i'}^{e'}    \right |     \\
                                             & = \frac{2  }{ m n_e n_{e'}} \left| \sum_{i=1}^{n_e} \psi'(t) \ell''_i(t)  \hat y_i^e(t)  j y_i^e \rvv^{e\top}_i \sum_{i'=1}^{n_{e'}} \psi'(t) \ell'_{i'}(t)     j \rvv_{i'}^{e'} \right|                                                                           \\
                                             & \le \frac{2\beta^2 }{ m n_e n_{e'}} \left| \sum_{i=1}^{n_e} \| \mathbf{w}_{j,r}(t) \|_2 \|\rvv^{e}_i \|_2 \ell''_i(t)  \hat y_i^e(t) \rvv^{e\top}_i \sum_{i'=1}^{n_{e'}} \| \mathbf{w}_{j,r}(t) \|_2 \|\rvv^{e'}_{i'} \|_2  \ell'_{i'}(t)   \rvv_{i'}^{e'} \right| \\
                                             & \le \frac{64\beta^2  LR^3}{ m }.
    \end{align*}
    Similarly, we further have that:
    \begin{align*}
        \left |  \rmH^4_{e,e'}(t)   \right | & = \left| \sum_{j} \sum_{r=1}^m  \left(\frac{ 1  }{n_{e} m} \right)  \left(\frac{ 1  }{n_{e'} m} \right) \sum_{i=1}^{n_e}  {\psi'} \ell'_i(t)  j  \rvv^{e\top}_i \sum_{i'=1}^{n_{e'}}   {\psi'}  \ell''_{i'}(t)  \hat{y}_{i'}^{e'}(t)   j \rvv_{i'}^{e'}   \right |            \\
                                             & = \frac{2}{ m n_e n_{e'}} \left| \sum_{i=1}^{n_e} \psi'(t) \ell'_i(t)   j \rvv^{e\top}_i \sum_{i'=1}^{n_{e'}} \psi'(t) \ell''_{i'}(t)   \hat y_{i'}^{e'}(t)   j \rvv_{i'}^{e'} \right|                                                                                        \\
                                             & \le \frac{2\beta^2 }{ m n_e n_{e'}} \left| \sum_{i=1}^{n_e} \| \mathbf{w}_{j,r}(t) \|_2 \|\rvv^{e}_i \|_2 \ell'_i(t)    \rvv^{e\top}_i \sum_{i'=1}^{n_{e'}} \| \mathbf{w}_{j,r}(t) \|_2 \|\rvv^{e'}_{i'} \|_2  \ell''_{i'}(t) \hat y_{i'}^{e'}(t)   j  \rvv_{i'}^{e'} \right| \\
                                             & \le \frac{64 \beta^2   LR^3}{ m }.
    \end{align*}
    Keep going on, we provide the computation results further:
    \begin{align*}
        \left |  \rmH^5_{e,e'}(t)   \right | & = \left| \sum_{j} \sum_{r=1}^m  \left(\frac{ 1  }{n_{e} m} \right)  \left(\frac{ 1  }{n_{e'} m} \right) \sum_{i=1}^{n_e}  {\psi'}  \ell'_i(t)   j y^e_i \boldsymbol{\xi}^{e\top}_i \sum_{i'=1}^{n_{e'}}  {\psi'}  \ell'_{i'}(t)   j y^{e'}_{i'} \boldsymbol{\xi}_{i'}^{e'}   \right |                             \\
                                             & \overset{(a)}  \le \frac{2\beta^2 }{ m n_e n_{e'}} \left| \sum_{i=1}^{n_e} \| \mathbf{w}_{j,r}(t) \|_2 \|\boldsymbol{\xi}_i \|_2 \ell'_i(t)    \boldsymbol{\xi}^{e\top}_i \sum_{i'=1}^{n_{e'}} \| \mathbf{w}_{j,r}(t) \|_2 \|\boldsymbol{\xi}^{e'}_{i'} \|_2  \ell'_{i'}(t)    \boldsymbol{\xi}_{i'}^{e'} \right| \\
                                             & \overset{(b)} \le \frac{2\beta^2 R^2 \sigma^2_p d}{ m },
    \end{align*}
    where inequality (a) is by smoothness property of non-linear activation function and Cauchy inequality, inequality (b) is by Lemma \ref{lemma:data_innerproducts} and Lemma \ref{lem:weight_initial}. Next, we calculate the
    \begin{align*}
        \left |  \rmH^6_{e,e'}(t)   \right | & = \left| \sum_{j} \sum_{r=1}^m  \left(\frac{ 1  }{n_{e} m} \right)  \left(\frac{ 1  }{n_{e'} m} \right) \sum_{i=1}^{n_e}  {\psi'}  \ell''_i(t)  \hat{y}_i^e(t) j  \boldsymbol{\xi}^{e\top}_i \sum_{i'=1}^{n_{e'}}  {\psi'}  \ell''_{i'}(t)  \hat{y}_{i'}^{e'}(t)   j  \boldsymbol{\xi}_{i'}^{e'}   \right |                          \\
                                             & \le \frac{2\beta^2 }{ m n_e n_{e'}} \left| \sum_{i=1}^{n_e} \| \mathbf{w}_{j,r}(t) \|_2 \|\boldsymbol{\xi}_i \|_2 \ell''_i(t) \hat{y}^e_i    \boldsymbol{\xi}^{e\top}_i \sum_{i'=1}^{n_{e'}} \| \mathbf{w}_{j,r}(t) \|_2 \|\boldsymbol{\xi}^{e'}_{i'} \|_2  \ell''_{i'}(t) \hat{y}_{i'}^{e'}(t)   \boldsymbol{\xi}_{i'}^{e'} \right| \\
                                             & \le \frac{8\beta^2 d L^4 R^2}{ m }.
    \end{align*}
    Similarly, the next $\mathbf{H}$ term can be calculated as follows:
    \begin{align*}
        \left |  \rmH^7_{e,e'}(t)   \right | & = \left| \sum_{j} \sum_{r=1}^m  \left(\frac{ 1  }{n_{e} m} \right)  \left(\frac{ 1  }{n_{e'} m} \right) \sum_{i=1}^{n_e}  {\psi'}  \ell''_{i}(t) \hat{y}^{e}_{i}   j  \boldsymbol{\xi}^{e\top}_i \sum_{i'=1}^{n_{e'}}  {\psi'}  \ell'_{i'}(t)   j y^{e'}_{i'} \boldsymbol{\xi}_{i'}^{e'}  \right |               \\
                                             & \le \frac{2\beta^2 }{ m n_e n_{e'}} \left| \sum_{i=1}^{n_e} \| \mathbf{w}_{j,r}(t) \|_2 \|\boldsymbol{\xi}_i \|_2 \ell''_i(t) \hat{y}^e_i(t)  \boldsymbol{\xi}^{e\top}_i \sum_{i'=1}^{n_{e'}} \| \mathbf{w}_{j,r}(t) \|_2 \|\boldsymbol{\xi}^{e'}_{i'} \|_2  \ell'_{i'}(t)    \boldsymbol{\xi}_{i'}^{e'} \right| \\
                                             & \le \frac{4\beta^2  \sigma^2_p d LR^3}{ m }.
    \end{align*}
    Finally, we have the upper for the last term:
    \begin{align*}
        \left |  \rmH^8_{e,e'}(t)   \right | & = \left| \sum_{j} \sum_{r=1}^m  \left(\frac{ 1  }{n_{e} m} \right)  \left(\frac{ 1  }{n_{e'} m} \right) \sum_{i=1}^{n_e}  {\psi'}  y^e_i \ell'_i(t)   j \boldsymbol{\xi}^{e\top}_i \sum_{i'=1}^{n_{e'}}   {\psi'}  \ell''_{i'}(t)     j \boldsymbol{\xi}_{i'}^e \hat{y}^{e'}_{i'}(t)   \right |                           \\
                                             & \le \frac{2\beta^2 }{ m n_e n_{e'}} \left| \sum_{i=1}^{n_e} \| \mathbf{w}_{j,r}(t) \|_2 \|\boldsymbol{\xi}_i \|_2 \ell'_i(t)    \boldsymbol{\xi}^{e\top}_i \sum_{i'=1}^{n_{e'}} \| \mathbf{w}_{j,r}(t) \|_2 \|\boldsymbol{\xi}^{e'}_{i'} \|_2  \ell''_{i'}(t)   \hat{y}^{e'}_{i'}(t)  \boldsymbol{\xi}_{i'}^{e'}  \right| \\
                                             & \le \frac{4\beta^2   \sigma^2_p d LR^3}{ m }.
    \end{align*}

    To summarize, we have that,
    \begin{align*}
        \left |  \rmH_{e,e'}(t) -  \rmH^\infty_{e,e'} \right | & \le \frac{32 \beta^2 R(R+ \frac{3}{2} \sigma_0 d )}{ m} + \frac{128 \beta^2 (R+\frac{3}{2}\sigma^2_0 d)^2 L^2 R^2}{ m }+\frac{128\beta^2   LR^3}{ m } \\
                                                               & \quad + \frac{2\beta^2 R^2 \sigma^2_p d}{ m } + \frac{8\beta^2 \sigma^2_p d L^2 R^4}{ m } + \frac{4\beta^2   \sigma^2_p d LR^3}{ m }                  \\
                                                               & \le  O\left( \frac{\beta^2 L R }{m} \right).
    \end{align*}
    where we have used $\sigma_p = O(d^{-2})$, $ R = o(1)$, and $\sigma_0 = O( \sqrt{R/d})$. Furthermore, we show that the perturbation term in Equation (\ref{CH:FeAT:eq:c_dynamimcs}) is bounded during training. In particular, we show the complete expression:
    \begin{align*}
        \rvg_{e}(t) & =  \sum_{j = \pm 1} \sum_{r=1}^m \left  \langle \frac{\partial C^e_\irml(\rmW,t)}{ \partial \rvw_{j,r}(t) },\frac{\partial L(\rmW,t)}{ \partial \rvw_{j,r}(t) }\right \rangle                                                                                                                                \\
                    & =  \sum_{j = \pm 1} \sum_{r=1}^m \bigg[ \frac{1  }{n_e m}
        \sum_{i=1}^{n_e} \ell'_i(t)  {\psi}'(\langle \rvw_{j,r}(t) , y_i^e \rvv_i^e \rangle) \cdot j \rvv_i^e  \frac{1  }{n_e m}
        \sum_{i=1}^{n_e} \ell'_i(t)  {\psi}'(\langle \rvw_{j,r}(t) , y_i^e \rvv_i^e \rangle) \cdot j \rvv_i^e                                                                                                                                                                                                                      \\
                    & \qquad + \frac{ 1 }{n_e m}     \sum_{i=1}^{n_e} \ell''_{i} \hat{y}_i^e  {\psi}'(\langle \rvw_{j,r}(t) , y_i^e \rvv_i^e \rangle) \cdot j y_i^e  \rvv_i^e
        \frac{1  }{n_e m}
        \sum_{i=1}^{n_e} \ell'_i(t)  {\psi}'(\langle \rvw_{j,r}(t) , y_i^e \rvv_i^e \rangle) \cdot j \rvv_i^e                                                                                                                                                                                                                      \\
                    & \qquad + \frac{\eta  }{n_e m}  \sum_{i=1}^{n_e} \ell'_i(t)  {\psi}'(\langle \rvw_{j,r}(t) ,  \boldsymbol{\xi}_i \rangle ) \cdot j y_i^e \boldsymbol{\xi}_i \frac{\eta  }{n_e m}  \sum_{i=1}^{n_e} \ell'_i(t)  {\psi}'(\langle \rvw_{j,r}(t) ,  \boldsymbol{\xi}_i \rangle ) \cdot j y_i^e \boldsymbol{\xi}_i \\
                    & \qquad +      \frac{\eta  }{n_e m}  \sum_{i=1}^{n_e} \ell'_i(t)  {\psi}'(\langle \rvw_{j,r}(t) ,  \boldsymbol{\xi}_i \rangle ) \cdot j y_i^e \boldsymbol{\xi}_i \sum_{i=1}^{n_e} \ell''_{i} \hat{y}_i^e  {\psi}'(\langle \rvw_{j,r}(t) ,  \boldsymbol{\xi}_i \rangle ) \cdot j  \boldsymbol{\xi}_i \bigg ]   \\
                    & \triangleq I_1 + I_2 + I_3 + I_4.
    \end{align*}
    Similar to the computation process for matrix $\mathbf{H}$, we adopt a divide and conquer manner:
    \begin{align*}
        |I_1|
         & \le \frac{2\beta^2 }{ m n_e n_{e}} \left| \sum_{i=1}^{n_e} \| \mathbf{w}_{j,r}(t) \|_2 \|\rvv^{e}_i \|_2 \ell'_i(t)   \rvv^{e\top}_i \sum_{i=1}^{n_{e}} \| \mathbf{w}_{j,r}(t) \|_2 \|\rvv^{e}_{i} \|_2  \ell'_{i'}(t)   \rvv_{i}^{e} \right| \\
         & \le \frac{32 \beta^2 (R+\frac{3}{2}\sigma^2_0 d)^2}{ m }.
    \end{align*}
    The techniques used are the same when deriving upper bound for matrix $\mathbf{H}$. Next, we have
    \begin{align*}
        |I_2|
         & \le \frac{2\beta^2 }{ m n_e n_{e}} \left| \sum_{i=1}^{n_e} \| \mathbf{w}_{j,r}(t) \|_2 \|\rvv^{e}_i \|_2 \ell''_i(t) \hat{y}^e_i(t)  \rvv^{e\top}_i \sum_{i=1}^{n_{e}} \| \mathbf{w}_{j,r}(t) \|_2 \|\rvv^{e}_{i} \|_2  \ell'_{i'}(t)   \rvv_{i}^{e} \right| \\
         & \le \frac{64 \beta^2  R^2 LR}{ m }.
    \end{align*}
    The last second term can be calculated as follows:
    \begin{align*}
        |I_3|
         & \le \frac{2\beta^2 }{ m n_e n_{e}} \left| \sum_{i=1}^{n_e} \| \mathbf{w}_{j,r}(t) \|_2 \|\boldsymbol{\xi}_i \|_2 \ell'_i(t)    \boldsymbol{\xi}^{e\top}_i \sum_{i=1}^{n_{e}} \| \mathbf{w}_{j,r}(t) \|_2 \|\boldsymbol{\xi}^{e}_{i} \|_2  \ell'_{i}(t)    \boldsymbol{\xi}_{i}^{e} \right| \\
         & \le \frac{2\beta^2 R^2 \sigma^2_p d}{ m },
    \end{align*}
    Finally, we show the upper bound of last term:
    \begin{align*}
        |I_4|
         & \le \frac{2\beta^2 }{ m n_e n_{e}} \left| \sum_{i=1}^{n_e} \| \mathbf{w}_{j,r}(t) \|_2 \|\boldsymbol{\xi}_i \|_2 \ell''_i(t) \hat{y}^e_i(t)  \boldsymbol{\xi}^{e\top}_i \sum_{i=1}^{n_{e}} \| \mathbf{w}_{j,r}(t) \|_2 \|\boldsymbol{\xi}^{e}_{i} \|_2  \ell'_{i'}(t)    \boldsymbol{\xi}_{i}^{e} \right| \\
         & \le \frac{4\beta^2   \sigma^2_p d LR^3}{ m }.
    \end{align*}
    In a summary, we have the following inequality:
    \begin{align*}
        \left |  \mathbf{g}_{e}(t)  \right | & \le \frac{32 \beta^2 R^2}{ m} +  \frac{64\beta^2   LR^3}{ m }  + \frac{2\beta^2 R^2 \sigma^2_p d}{ m } +   \frac{4\beta^2 R^3 \sigma^2_p d L }{ m } \\
                                             & \le  O\left( \frac{\beta^2 L R^2 }{m} \right),
    \end{align*}
    where we have used $\sigma_p = O(d^{-2})$ and $ R = o(1)$. With all the bounds at hand, we are ready to have the dynamics for $\| \rvc(t)\|^2_2$
    \begin{align}
        \frac{d \| \rvc(t) \|^2_2}{d t} & = - 2 \lambda \mathbf{c}^\top (t) \rmH(t) \mathbf{c}(t) -  \mathbf{c}(t) \mathbf{g}(t)   \le -\lambda_0 \lambda \| \mathbf{c}(t)  \|^2_2, \label{CH:FeAT:eq:c_ex_dynamics}
    \end{align}
    which requires that $\|  \rmH(t) -  \rmH^\infty \|_2 \le \lambda_0 $. This leads to the following inequality:
    \begin{align*}
        \|  \rmH(t) -  \rmH^\infty \|_2 & \le   \|  \rmH(t) -  \rmH^\infty \|_F \le \sum_{i,j} |\rmH_{ij}(t) -  \rmH_{ij}^\infty | \\
                                        & \le \frac{ |\mathcal{E}_{tr}|^2 \beta^2 L R}{m} \le  \lambda_0.
    \end{align*}
    which leads to the conclusion for $R$ as follows:
    \begin{align}
        R \le \frac{\lambda_0 m}{ |\mathcal{E}_{tr}|^2 \beta^2 L}. \label{CH:FeAT:eq:R}
    \end{align}
    Besides, we have the inequality that
    \begin{align}
        \|\mathbf{g} \|_2 \le \frac{\sqrt{|\mathcal{E}_{tr} |} \beta^2 L R}{m} \le \lambda \lambda_0  \| \mathbf{c}(0) \|_2.
    \end{align}
    Combined with Equation (\ref{CH:FeAT:eq:R}), we obtain the condition for $\lambda$ as follows:
    \begin{align} \label{CH:FeAT:eq:lambda}
        \lambda \ge 1/(\sigma_0 \sqrt{|\mathcal{E}_{tr}|}^3 ).
    \end{align}
    By inequality (\ref{CH:FeAT:eq:c_ex_dynamics}), taking the convergence time $T = \Omega \left( \frac{ \log(\sigma_0/\epsilon)}{ \eta  \lambda \lambda_0 } \right) $ we have that:
    \begin{align*}
        \| \mathbf{c}(T) \|_2 \le \epsilon.
    \end{align*}

    According to the gradient descent for IRMV1 objective function, the evolution of coefficients can be expressed as:
    \begin{align*}
        \gamma_{j,r}^{inv}(t+1) & =  \gamma^{inv}_{j,r}(t) - \frac{\eta}{m} \cdot   \sum_{e \in \mathcal{E}_{\mathrm{tr}}}   (1+ 2\lambda C_\irml^e(t)) \frac{1}{n_e}\sum_{i=1}^{n_e}   \ell'_i (t) \psi_i'(t)   \textrm{Rad}(\alpha)_i
        \\&\ \
        -\frac{\eta \lambda}{m}  \cdot  \sum_{e \in \mathcal{E}_{\mathrm{tr}}} 2 C_\irml^e  \frac{1}{n_e}\sum_{i=1}^{n_e} \ell''_{i} \psi_i'(t)   \hat{y}_i^e   \cdot   y_i^e   \textrm{Rad} (\alpha)_i,                                \\
        \gamma_{j,r}^{spu}(t+1) & =  \gamma^{spu}_{j,r}(t) - \frac{\eta}{m} \cdot   \sum_{e \in \mathcal{E}_{\mathrm{tr}}}   (1+ 2\lambda C_\irml^e(t)) \frac{1}{n_e}\sum_{i=1}^{n_e} \ell'_i(t)  \psi_i'(t)    \textrm{Rad}(\beta_e)
        \\&\ \
        -\frac{\eta \lambda}{m}  \cdot  \sum_{e \in \mathcal{E}_{\mathrm{tr}}} 2 C_\irml^e \frac{1}{n_e} \sum_{i=1}^{n_e}  \psi_i'(t)  \ell''_{i} \hat{y}_i^e   \cdot y_i^e  \textrm{Rad}(\beta_e)_i.
    \end{align*}

    Then we have,
    \begin{align*}
        |\gamma_{j,r}^{inv}(t+1)| & \le  |\gamma_{j,r}^{inv}(t)| + \left|\frac{\eta}{m} \cdot   \sum_{e \in \mathcal{E}_{\mathrm{tr}}}   (1+ 2\lambda C_\irml^e(t)) \frac{1}{n_e}\sum_{i=1}^{n_e}  \psi_i'(t)   \ell'_i (t)   \textrm{Rad}(\alpha)_i  \right| \\
                                  & + \left| \frac{\eta \lambda}{m}  \cdot  \sum_{e \in \mathcal{E}_{\mathrm{tr}}} 2 C_\irml^e  \frac{1}{n_e}\sum_{i=1}^{n_e} \ell''_{i}  \psi_i'(t)   \hat{y}_i^e   \cdot   y_i^e   \textrm{Rad} (\alpha)_i \right|          \\
                                  & \le  |\gamma^{inv}_{j,r}(t)| + C \frac{\eta \sqrt{|\mathcal{E}_{tr}|} \lambda \beta R^2 L }{m} \| \rvc(t)\|_2.
    \end{align*}
    Similarly, we have,
    \begin{align*}
        |\gamma_{j,r}^{spu}(t+1)| & \le  |\gamma^{spu}_{j,r,2}(t)| +  C \frac{\eta \sqrt{|\mathcal{E}_{tr}|} \lambda \beta R^2 L }{m} \| \rvc(t)\|_2.
    \end{align*}

    At the time step $T$, the feature learning satisfies that:
    \begin{align*}
        \gamma^{inv}_{j,r}(T) \le C \frac{\eta \sqrt{|\mathcal{E}_{tr}|} \lambda \beta R^2 L T}{m} \|\mathbf{c}(0) \|_2;   \quad \gamma^{spu}_{j,r}(T) \le C \frac{\eta \sqrt{|\mathcal{E}_{tr}|} \lambda \beta R^2 L T}{m} \|\mathbf{c}(0) \|_2.
    \end{align*}

    To make sure that $\gamma^{inv}_{j,r}(T) = o(1)$ and $\gamma^{spu}_{j,r}(T) = o(1)$, we need the following condition:
    \begin{align}
        C \frac{\eta \sqrt{|\mathcal{E}_{tr}|} \lambda \beta R^2 L T}{m} \|\mathbf{c}(0) \|_2 \le d^{-\frac{1}{2}},
    \end{align}
    combined with inequality (\ref{CH:FeAT:eq:R}) and inequality (\ref{CH:FeAT:eq:lambda}), we have:
    \begin{align*}
        \sigma_0 \le \frac{ |\mathcal{E}_{tr} |^{7/2} \beta^3 L }{ d^{1/2}m^2\lambda^2_0 \log(1/\epsilon)}.
    \end{align*}

\end{proof}

\subsection{Proof for Proposition~\ref{pro:irmv1_with_erm_feat}}\label{CH:FeAT:sec:proof_irmv1_with_erm_feat}
\begin{proposition} [Restatement of Proposition \ref{pro:irmv1_with_erm_feat}]\label{CH:FeAT:thm:irmv1_with_erm_feat_appdx}
    Consider training the CNN model with the same data as Theorem~\ref{CH:FeAT:thm:erm_learn_feat},
    suppose that $\psi(x) = x$, $ \gamma_{j,r,1}(t_1 ) =  \gamma_{j,r,1}(t_1-1)$, and $ \gamma_{j,r,2}(t_1 ) =  \gamma_{j,r,2}(t_1-1)$ at the end of ERM pre-train $t_1$ and $\mathcal{E}_{tr} = \{(0.25, 0.1), (0.25, 0.2)\}$. Suppose that $\delta>0$, and $ n > C \log(1/\delta)$, with $C$ being a positive constant, then with a high probability at least $1-\delta$, we have
    \begin{itemize}
        \item $ \sum_e C^e_\irml(t_1) =0 $.
        \item  $\gamma_{j,r,1}(t_1 + 1) >  \gamma_{j,r,1}(t_1) $.
        \item  $\gamma_{j,r,2}(t_1 + 1) <  \gamma_{j,r,2}(t_1) $.
    \end{itemize}
\end{proposition}

\begin{proof} [Proof of Proposition \ref{CH:FeAT:thm:irmv1_with_erm_feat_appdx}] According to the gradient descent for IRMV1 objective function, the evolution of coefficients can be expressed as:
    \begin{align*}
        \gamma_{j,r,1}(t+1) & =  \gamma_{j,r,1}(t) - \frac{\eta}{m} \cdot   \sum_{e \in \mathcal{E}_{\mathrm{tr}}}   (1+ 2\lambda C_\irml^e(t)) \frac{1}{n_e}\sum_{i=1}^{n_e}   \ell'_i (t)   \textrm{Rad}(\alpha)_i
        \\&\ \
        -\frac{\eta \lambda}{m}  \cdot  \sum_{e \in \mathcal{E}_{\mathrm{tr}}} 2 C_\irml^e  \frac{1}{n_e}\sum_{i=1}^{n_e} \ell''_{i}  \hat{y}_i^e   \cdot   y_i^e   \textrm{Rad} (\alpha)_i,                         \\
        \gamma_{j,r,2}(t+1) & =  \gamma_{j,r,2}(t) - \frac{\eta}{m} \cdot   \sum_{e \in \mathcal{E}_{\mathrm{tr}}}   (1+ 2\lambda C_\irml^e(t)) \frac{1}{n_e}\sum_{i=1}^{n_e} \ell'_i(t)    \textrm{Rad}(\beta_e)
        \\&\ \
        -\frac{\eta \lambda}{m}  \cdot  \sum_{e \in \mathcal{E}_{\mathrm{tr}}} 2 C_\irml^e \frac{1}{n_e} \sum_{i=1}^{n_e} \ell''_{i} \hat{y}_i^e   \cdot y_i^e  \textrm{Rad}(\beta_e)_i,
    \end{align*}
    where $\ell''( y_i^e \cdot f(\rmW, \rvx^e_i)) = \frac{ \exp(-y_i^e \cdot f(\rmW, \rvx_i))}{(1 + \exp(-y_i^e \cdot f(\rmW, \rvx_i)))^2} $.

    To simplify the notation, we further define \[A_1^e = \frac{1}{n_e} \sum_{i=1}^{n_e} \ell'_i \mathrm{Rad}(\alpha)_i \] and \[A_2^e = \frac{1}{n_e} \sum_{i=1}^{n_e} \ell''_i \hat{y}_i^e y_i^e \mathrm{Rad}(\alpha)_i \]. Similarly, we define \[B_1^e = \frac{1}{n_e} \sum_{i=1}^{n_e} \ell'_i \mathrm{Rad}(\beta_e)_i \] and \[B_2^e = \frac{1}{n_e} \sum_{i=1}^{n_e} \ell''_i \hat{y}_i^e y_i^e \mathrm{Rad}(\beta_e)_i. \]

    In the limit of $n \rightarrow \infty$, we have:
    \begin{align*}
        \lim_{n \rightarrow \infty}	A^1_1(t_1) & = -1/(1+e^{(\gamma_1 + \gamma_2)})(1- \alpha)(1-\beta_1) - 1/(1+e^{\gamma_1 - \gamma_2})(1- \alpha)\beta_1   \\
                                              & \quad + 1/(1+e^{\gamma_2 - \gamma_1}) \alpha(1-\beta_1) + 1/(1+e^{-\gamma_1 - \gamma_2}) \alpha \beta_1,     \\
        \lim_{n \rightarrow \infty}	A^2_1(t_1) & =  -1/(1+ e^{\gamma_1 + \gamma_2})(1- \alpha)(1-\beta_2) -  1/(1+ e^{\gamma_1 - \gamma_2})(1- \alpha)\beta_2 \\
                                              & \quad + 1/(1+e^{-\gamma_1 + \gamma_2} )\alpha(1-\beta_2) +  1/(1+e^{-\gamma_1 - \gamma_2}) \alpha \beta_2,   \\
        \lim_{n \rightarrow \infty}	B^1_1(t_1) & = -1/(1+ e^{\gamma_1 + \gamma_2})(1- \alpha)(1-\beta_1) + 1/(1+e^{\gamma_1 - \gamma_2})(1- \alpha)\beta_1    \\
                                              & \quad - 1/(1+e^{-\gamma_1 + \gamma_2}) \alpha(1-\beta_1) + 1/(1+ e^{-\gamma_1 - \gamma_2}) \alpha \beta_1,   \\
        \lim_{n \rightarrow \infty}	B^2_1(t_1) & = -1/(1+e^{\gamma_1 + \gamma_2})(1- \alpha)(1-\beta_2) + 1/(1+e^{\gamma_1 - \gamma_2})(1- \alpha)\beta_2     \\
                                              & \quad - 1/(1+e^{-\gamma_1 + \gamma_2}) \alpha(1-\beta_2) + 1/(1+e^{-\gamma_1 - \gamma_2}) \alpha \beta_2.
    \end{align*}
    and,
    \begin{align*}
        \lim_{n \rightarrow \infty}	A^1_2(t_1) & = e^{\gamma_1 + \gamma_2}/(1+e^{\gamma_1 + \gamma_2})^2(1- \alpha)(1-\beta_1)(\gamma_1+\gamma_2) + e^{\gamma_1 - \gamma_2}/(1+e^{\gamma_1 - \gamma_2})^2(1- \alpha)\beta_1(\gamma_1-\gamma_2)           \\
                                              & \quad + e^{-\gamma_1 + \gamma_2}/(1+e^{-\gamma_1 + \gamma_2})^2 \alpha(1-\beta_1)(\gamma_1 - \gamma_2) + e^{-\gamma_1 - \gamma_2}/(1+e^{-\gamma_1 - \gamma_2})^2 \alpha \beta_1(\gamma_1 + \gamma_2),   \\
        \lim_{n \rightarrow \infty}	A^2_2(t_1) & = e^{\gamma_1 + \gamma_2}/(1+e^{\gamma_1 + \gamma_2})^2(1- \alpha)(1-\beta_2)(\gamma_1+\gamma_2) + e^{\gamma_1 - \gamma_2}/(1+e^{\gamma_1 - \gamma_2})^2(1- \alpha)\beta_2(\gamma_1-\gamma_2)           \\
                                              & \quad + e^{-\gamma_1 + \gamma_2}/(1+e^{-\gamma_1 + \gamma_2})^2 \alpha(1-\beta_2)(\gamma_1 - \gamma_2) + e^{-\gamma_1 - \gamma_2}/(1+e^{-\gamma_1 - \gamma_2})^2 \alpha \beta_2(\gamma_1 + \gamma_2),   \\
        \lim_{n \rightarrow \infty}	B^1_2(t_1) & = e^{\gamma_1 + \gamma_2}/(1+e^{\gamma_1 + \gamma_2})^2(1- \alpha)(1-\beta_1)(\gamma_1+\gamma_2) + e^{\gamma_1 - \gamma_2}/(1+e^{\gamma_1 - \gamma_2})^2(1- \alpha)\beta_1(-\gamma_1+\gamma_2)          \\
                                              & \quad +  e^{-\gamma_1 + \gamma_2}/(1+e^{-\gamma_1 + \gamma_2})^2 \alpha(1-\beta_1)(-\gamma_1 + \gamma_2) + e^{-\gamma_1 - \gamma_2}/(1+e^{-\gamma_1 - \gamma_2})^2\alpha \beta_1(\gamma_1 + \gamma_2),  \\
        \lim_{n \rightarrow \infty}	B^2_2(t_1) & =e^{\gamma_1 + \gamma_2}/(1+e^{\gamma_1 + \gamma_2})^2(1- \alpha)(1-\beta_2)(\gamma_1+\gamma_2) +  e^{\gamma_1 - \gamma_2}/(1+e^{\gamma_1 - \gamma_2})^2(1- \alpha)\beta_2(-\gamma_1+\gamma_2)          \\
                                              & \quad +  e^{-\gamma_1 + \gamma_2}/(1+e^{-\gamma_1 + \gamma_2})^2 \alpha(1-\beta_2)(-\gamma_1 + \gamma_2) + e^{-\gamma_1 - \gamma_2}/(1+e^{-\gamma_1 - \gamma_2})^2 \alpha \beta_2(\gamma_1 + \gamma_2).
    \end{align*}

    Because $\mathrm{Rad}(\alpha)_i$ and $\mathrm{Rad}(\beta)_i$ are random variables, applying Hoeffding's inequality, we have with probability at least $1 - \delta$,
    \begin{align*}
        \left| A^1_1(t_1) - \lim_{n \rightarrow \infty} A^1_1(t_1) \right| \le \sqrt{\frac{4 \log(1/\delta)}{n}}.
    \end{align*}
    Similarly, we can apply the concentration bound to other quantities and obtain the same bound.

    By the assumption that $ \gamma_{j,r,1}(t_1 ) =  \gamma_{j,r,1}(t_1-1)$ and $ \gamma_{j,r,2}(t_1 ) =  \gamma_{j,r,2}(t_1-1)$, we have that $ \sum_e A^e_1 (t_1) = \sum_e B^e_1 (t_1) = 0  $:
    \begin{align*}
        \lim_{n \rightarrow \infty} (A_1^1(t_1) + A_1^2(t_1)) & =  -1/(1+e^{\gamma_1 + \gamma_2})(1- \alpha)(2-\beta_1-\beta_2)  - 1/(1+e^{\gamma_1 - \gamma_2})(1- \alpha)(\beta_1+\beta_2)  \\
                                                              & \quad + 1/(1+e^{-\gamma_1 + \gamma_2})\alpha(2-\beta_1-\beta_2) + 1/(1+e^{-\gamma_1 - \gamma_2})\alpha(\beta_1+\beta_2) = 0   \\
        \lim_{n \rightarrow \infty} (B_1^1(t_1) + B_1^2(t_1)) & =   -1/(1+e^{\gamma_1 + \gamma_2})(1- \alpha)(2-\beta_1-\beta_2)  + 1/(1+e^{\gamma_1 - \gamma_2})(1- \alpha)(\beta_1+\beta_2) \\
                                                              & \quad + 1/(1+e^{-\gamma_1 + \gamma_2})\alpha(2-\beta_1-\beta_2) +1/ (1+e^{-\gamma_1 - \gamma_2})\alpha(\beta_1+\beta_2) = 0
    \end{align*}

    Solving the above equations, we have,
    \begin{align*}
        \gamma^\infty_1(t_1) = \frac{1}{2} \log(G_m  G_b ) \quad \gamma^\infty_2(t_1) = \frac{1}{2} \log(G_m /G_b )
    \end{align*}
    where we denote $\gamma^\infty_1(t_1) \triangleq
        \lim_{n \rightarrow \infty} \gamma_1(t_1) $  and $\gamma^\infty_2(t_1) \triangleq
        \lim_{n \rightarrow \infty} \gamma_1(t_2) $, $G_m  = ((1-A) + \sqrt{(A-1)^2+4A)}/(2A)$ and $G_b  = ((1-B) + \sqrt{(B-1)^2+4B)}/(2B)$, with $A = \alpha(\beta_1+\beta_2)/((1-\alpha)(2-\beta_1-\beta_2))$ and
    $B = \alpha(2- \beta_1-\beta_2)/((1-\alpha)*(\beta_1+ \beta_2))$.

    By the convexity of function $f(x) = e^x$, with a constant $C$, we have:
    \begin{align*}
        \left| \gamma_1 - \gamma^\infty_1  \right| <  \left| e^{\gamma_1} - e^{\gamma^\infty_1}  \right| \le  C \left| 1/(1+e^{\gamma_1}) - 1/(1+e^{\gamma^\infty_1})  \right| \le \sqrt{\frac{4 \log(1/\delta)}{n}}, \\
        \left| \gamma_2 - \gamma^\infty_2  \right| <  \left| e^{\gamma_2} - e^{\gamma^\infty_2}  \right| \le  C \left| 1/(1+e^{\gamma_2}) - 1/(1+e^{\gamma^\infty_2})  \right| \le \sqrt{\frac{4 \log(1/\delta)}{n}}.
    \end{align*}

    Then we know that,
    \begin{align*}
        C_\irml^1 = \frac{1}{n_1} \sum_{i=1}^{n_1} \ell'_i \hat{y}_i^1 y_i^1 = \gamma_1 A_1^1 + \gamma_2 B^1_1 \\
        C_\irml^2 = \frac{1}{n_2} \sum_{i=1}^{n_2} \ell'_i \hat{y}_i^2 y_i^2 = \gamma_1 A_1^2 + \gamma_2 B^2_1
    \end{align*}
    Therefore, we have that:
    \begin{align*}
        C_\irml^1 + C_\irml^2 =  0
    \end{align*}
    Then the evolution of coefficients reduces to
    \begin{align*}
        \gamma_{j,r,1}(t+1) & =  \gamma_{j,r,1}(t) - \frac{\eta}{m} \cdot   \sum_{e \in \mathcal{E}_{\mathrm{tr}}}   (1+ 2\lambda C_\irml^e(t)) A^e_1(t)    -\frac{\eta \lambda}{m}  \cdot  \sum_{e \in \mathcal{E}_{\mathrm{tr}}} 2 C_\irml^e  A^e_2(t) \\
        \gamma_{j,r,2}(t+1) & =  \gamma_{j,r,2}(t) - \frac{\eta}{m} \cdot   \sum_{e \in \mathcal{E}_{\mathrm{tr}}}   (1+ 2\lambda C_\irml^e(t)) B^e_1(t)   -\frac{\eta \lambda}{m}  \cdot  \sum_{e \in \mathcal{E}_{\mathrm{tr}}} 2 C_\irml^e B^e_2(t)
    \end{align*}
    Taking the solution of $\gamma_{j,r,1}(t_1)$, $\gamma_{j,r,2}(t_1)$ and value of $\alpha, \beta_1, \beta_2$, we arrive at the conclusion that with a high a probability at least $1- \delta$ and $n > C_1 \log(1/\delta)$ with $C_1$ being a positive constant, we have:
    \begin{align*}
        \gamma_{j,r,1}(t_1 + 1) >  \gamma_{j,r,1}(t_1), \\ \gamma_{j,r,2}(t_1 + 1) <  \gamma_{j,r,2}(t_1).
    \end{align*}

\end{proof}

\subsection{Proof for Corollary~\ref{cor:irmv1_with_bad_feat}}\label{CH:FeAT:sec:proof_irmv1_with_bad_feat}

\begin{corollary} [Restatement of Corollary~\ref{cor:irmv1_with_bad_feat}]
    \label{cor:irmv1_with_bad_feat_appdix}
    Consider training the CNN model with the data generated from Def.~\ref{CH:FeAT:def:risk_irm},
    suppose that $\psi(x) = x$, $ \gamma_{j,r,1}(t_1 ) =  o(1) $, and $ \gamma_{j,r,2}(t_1 ) =  \Theta(1)$ at the end of ERM pre-train $t_1$ and $\mathcal{E}_{tr} = \{(0.25, 0.1), (0.25, 0.2)\}$. Suppose that $\delta>0$, and $ n > C \log(1/\delta)$, with $C$ being a positive constant, then with a high probability at least $1-\delta$, we have
    \begin{align*}
        \gamma_{j,r,1}(t_1 + 1) <  \gamma_{j,r,1}(t_1).
    \end{align*}
\end{corollary}

\begin{proof} [Proof of Corollary \ref{cor:irmv1_with_bad_feat_appdix}]
    Recall that the feature learning update rule:
    \begin{align*}
        \gamma_{j,r,1}(t+1) & =  \gamma_{j,r,1}(t) - \frac{\eta}{m} \cdot   \sum_{e \in \mathcal{E}_{\mathrm{tr}}}   (1+ 2\lambda C_\irml^e(t)) \frac{1}{n_e}\sum_{i=1}^{n_e}   \ell'_i (t)   \textrm{Rad}(\alpha)_i
        \\&\ \
        -\frac{\eta \lambda}{m}  \cdot  \sum_{e \in \mathcal{E}_{\mathrm{tr}}} 2 C_\irml^e  \frac{1}{n_e}\sum_{i=1}^{n_e} \ell''_{i}  \hat{y}_i^e   \cdot   y_i^e   \textrm{Rad} (\alpha)_i,                         \\
        \gamma_{j,r,2}(t+1) & =  \gamma_{j,r,2}(t) - \frac{\eta}{m} \cdot   \sum_{e \in \mathcal{E}_{\mathrm{tr}}}   (1+ 2\lambda C_\irml^e(t)) \frac{1}{n_e}\sum_{i=1}^{n_e} \ell'_i(t)    \textrm{Rad}(\beta_e)    \\&\ \
        -\frac{\eta \lambda}{m}  \cdot  \sum_{e \in \mathcal{E}_{\mathrm{tr}}} 2 C_\irml^e \frac{1}{n_e} \sum_{i=1}^{n_e} \ell''_{i} \hat{y}_i^e   \cdot y_i^e  \textrm{Rad}(\beta_e)_i,
    \end{align*}
    Taking the value of $\gamma_{j,r,1}(t_1)$, $\gamma_{j,r,2}(t_1)$ and, we can conclude that:
    \begin{align*}
        \lim_{n \rightarrow \infty}	A^1_1(t_1) & = -1/(1+e^{ \gamma_2 })(1- \alpha)(1-\beta_1) - 1/(1+e^{ - \gamma_2})(1- \alpha)\beta_1  +
        \\&\ \
        1/(1+e^{\gamma_2}) \alpha(1-\beta_1) + 1/(1+e^{- \gamma_2}) \alpha \beta_1                                                              \\
                                              & = 1/(1+e^{ \gamma_2})(2\alpha - 1)(1-\beta_1) + 1/(1+e^{ -\gamma_2})(2\alpha - 1)(\beta_1)      \\
                                              & = (2\alpha-1) [ 1/(1+e^{ \gamma_2})(1-\beta_2) +1/(1+e^{ -\gamma_2})\beta_1 ) ]                 \\
        \lim_{n \rightarrow \infty}	A^2_1(t_1)
                                              & = 1/(1+e^{ \gamma_2})(2\alpha - 1)(1-\beta_2) + 1/(1+e^{ -\gamma_2})(2\alpha - 1)(\beta_2)      \\
                                              & = (2\alpha-1) [ 1/(1+e^{ \gamma_2})(1-\beta_2) +1/(1+e^{ -\gamma_2})\beta_2 ) ]                 \\
        \lim_{n \rightarrow \infty}	B^1_1(t_1) & = -1/(1+ e^{  \gamma_2})(1- \alpha)(1-\beta_1) + 1/(1+e^{  - \gamma_2})(1- \alpha)\beta_1     -
        \\&\ \
        1/(1+e^{  \gamma_2}) \alpha(1-\beta_1) + 1/(1+ e^{- \gamma_2}) \alpha \beta_1                                                           \\
                                              & = -1/(1+ e^{  \gamma_2})(1-\beta_1) + 1/(1+ e^{ - \gamma_2})\beta_1                             \\
        \lim_{n \rightarrow \infty}	B^2_1(t_1) & = -1/(1+ e^{  \gamma_2})(1- \alpha)(1-\beta_2) + 1/(1+e^{  - \gamma_2})(1- \alpha)\beta_2     -
        \\&\ \
        1/(1+e^{  \gamma_2}) \alpha(1-\beta_2) + 1/(1+ e^{- \gamma_2}) \alpha \beta_2                                                           \\
                                              & = -1/(1+ e^{  \gamma_2})(1-\beta_2) + 1/(1+ e^{ - \gamma_2})\beta_2
    \end{align*}
    On the other hand,
    \begin{align*}
        \lim_{n \rightarrow \infty}	A^1_2(t_1) & = e^{  \gamma_2}/(1+e^{  \gamma_2})^2(1- \alpha)(1-\beta_1)( \gamma_2) + e^{  - \gamma_2}/(1+e^{  - \gamma_2})^2(1- \alpha)\beta_1( -\gamma_2)     \\
                                              & \quad + e^{  + \gamma_2}/(1+e^{  \gamma_2})^2 \alpha(1-\beta_1)(  - \gamma_2) + e^{  - \gamma_2}/(1+e^{ - \gamma_2})^2 \alpha \beta_1(  \gamma_2)  \\
                                              & =  e^{  \gamma_2}/(1+e^{  \gamma_2})^2 (1-2\alpha)(1-\beta_1) + e^{ - \gamma_2}/(1+e^{ - \gamma_2})^2 (2\alpha-1) \beta_1   \gamma_2               \\
        \lim_{n \rightarrow \infty}	A^2_2(t_1) & = e^{ \gamma_2}/(1+e^{ \gamma_2})^2(1- \alpha)(1-\beta_2)( \gamma_2) + e^{  - \gamma_2}/(1+e^{  - \gamma_2})^2(1- \alpha)\beta_2( -\gamma_2)       \\
                                              & \quad + e^{  \gamma_2}/(1+e^{    \gamma_2})^2 \alpha(1-\beta_2)(  - \gamma_2) + e^{  - \gamma_2}/(1+e^{  - \gamma_2})^2 \alpha \beta_2(  \gamma_2) \\
                                              & =  e^{  \gamma_2}/(1+e^{  \gamma_2})^2 (1-2\alpha)(1-\beta_2) + e^{ - \gamma_2}/(1+e^{ - \gamma_2})^2 (2\alpha-1) \beta_2   \gamma_2               \\
        \lim_{n \rightarrow \infty}	B^1_2(t_1) & = e^{ \gamma_2}/(1+e^{  \gamma_2})^2(1- \alpha)(1-\beta_1)( \gamma_2) + e^{\  - \gamma_2}/(1+e^{  - \gamma_2})^2(1- \alpha)\beta_1( \gamma_2)      \\
                                              & \quad +  e^{  \gamma_2}/(1+e^{ \gamma_2})^2 \alpha(1-\beta_1)( \gamma_2) + e^{  - \gamma_2}/(1+e^{  - \gamma_2})^2\alpha \beta_1(  \gamma_2),      \\
        \lim_{n \rightarrow \infty}	B^2_2(t_1) & =e^{  \gamma_2}/(1+e^{  \gamma_2})^2(1- \alpha)(1-\beta_2)( \gamma_2) +  e^{\  - \gamma_2}/(1+e^{  - \gamma_2})^2(1- \alpha)\beta_2(  \gamma_2)    \\
                                              & \quad +  e^{  \gamma_2}/(1+e^{  \gamma_2})^2 \alpha(1-\beta_2)(  \gamma_2) + e^{  - \gamma_2}/(1+e^{  - \gamma_2})^2 \alpha \beta_2(  \gamma_2).
    \end{align*}

    Finally, taking the value of environment of $(\alpha, \beta_1, \beta_2) = (0.25, 0.1, 0.2) $, we conclude that with a high a probability at least $1- \delta$ and $n > C_1 \log(1/\delta)$ with $C_1$ being a positive constant, we have:
    \begin{align*}
        \gamma_{j,r,1}(t_1 + 1) <  \gamma_{j,r,1}(t_1).
    \end{align*}
\end{proof}
\clearpage
\section{More Details about i\feat}
\label{CH:FeAT:sec:ifat_appdx}
As mentioned in Sec.~\ref{CH:FeAT:sec:fat_alg} that, when the featurizer is implemented as a deep net that have a massive amount of parameters,
backpropagating through Algorithm~\ref{alg:fat} can allocate too much memory for propagating with $2K-1$ batches of data.
It is common for many realistic benchmarks such as Camelyon17 and FMoW in wilds benchmark~\citep{wilds} that adopts a DenseNet~\citep{densenet} with $121$ layers as the featurizer.
To relieve the exceeding computational and memory overhead, we propose a lightweight version of \feat, denoted as \feat.
Instead of storing all of historical subsets and classifiers, \feati iteratively use the augmentation and retention sets and historical classifier from only the last round.
In contrast, previous rich feature learning algorithm~\citep{rfc,diwa} incurs a high computational and memory overhead as the round grows.
For example, in RxRx1, we have to reduce the batch size of \rfc to allow the proceeding of rounds $\geq 3$.

We elaborate the detailed algorithmic description of \feati in Algorithm~\ref{alg:ifat_alg}.
\begin{algorithm}[ht]
    \caption{\feat: \featfull }
    \label{alg:ifat_alg}
    \begin{algorithmic}[1]
        \STATE \textbf{Input:} Training data $\train$; the maximum augmentation rounds $K$; predictor $f:=w\circ\varphi$; length of inner training epochs $e$; termination threshold $p$;
        \STATE Initialize groups $G^a\leftarrow {\train}, G^r\leftarrow \{\}$;
        \FOR{$k \in [1,\ldots, K]$}
        \STATE Randomly initialize $w_k$;
        \FOR{$j \in [1,\ldots, e]$}
        \STATE Obtain $\ell_\feat$ with $G$ via Eq.~\ref{CH:FeAT:eq:fat_upd};
        \STATE Update $w_k, \varphi$ with $\ell_\feat$;
        \ENDFOR
        \STATE \texttt{// Early Stop if $f_k=w_k\circ\varphi$ fails to find new features.}
        \IF{Training accuracy of $f_k$ is smaller than $p$}
        \STATE Set $K=k-1$ and terminate the loop;
        \ENDIF
        \IF{$k>1$}
        \STATE \texttt{// Hence it doesnot need to maintain all historical classifiers.}
        \STATE Update $w_k\leftarrow (w_{k-1},w_k)$;
        \ENDIF
        \STATE Split $\train$ into groups $\dataset_k^a,\dataset_k^r$ according to $f_k$;
        \STATE \texttt{// Hence it doesnot need to maintain all historical subsets.}
        \STATE Update groups $G^a \leftarrow \{\dataset_k^a\}, G^r\leftarrow \{\dataset_k^r\}$;
        \ENDFOR
        \STATE \textbf{return} $f=w\circ\varphi$;
    \end{algorithmic}
\end{algorithm}

\section{More Details about the Experiments}
\label{CH:FeAT:sec:exp_appdx}
In this section, we provide more details and the implementation, evaluation and hyperparameter
setups in complementary to the experiments in Sec.~\ref{CH:FeAT:sec:exp}.

\subsection{More details about \cmnist experiments}
\label{CH:FeAT:sec:cmnist_appdx}
\paragraph{Datasets.} In the controlled experiments with \cmnist,
we follow the evaluation settings as previous works~\citep{irmv1,rfc,pair}.
In addition to the original \cmnist with $\envtrain=\{(0.25,0.1),(0.25,0.2)\}$ (denoted as \cmnist-025) where spurious features are better correlated with labels,
we also incorporate the modified one (denoted as \cmnist-01) with $\envtrain=\{(0.1,0.2),(0.1,0.25)\}$ where invariant features are better correlated with labels,
since both cases can happen at real world.

\paragraph{Architecture and optimization.} To ensure a fair comparison, we use $4$-Layer MLP with a hidden dimension of $256$ as the backbone model for all methods,
where we take the first $3$ layers as the featurizer and the last layer as the classifier,
following the common practice~\citep{domainbed,wilds}.
For the optimization of the models,
we use the Adam~\cite{adam} optimizer with a learning rate of $1e-3$
and a weight decay of $1e-3$.
We report the mean and standard deviation of the performances of
different methods with each configuration of hyperparameters $10$ times
with the random seeds from $1$ to $10$.

\paragraph{Implementation of ERM-NF and OOD objectives.} For the common pre-training protocol with ERM, our implementation follows the previous works~\citep{rfc}.
Specifically, we first train the model
with $\{0,50,100,150,200,250\}$ epochs and then apply
the OOD regularization of various objectives with a penalty weight of $\{1e1,1e2,1e3,1e4,1e5\}$.
We adopt the implementations from~\citet{rfc} for various OOD objectives,
including \irml~\citep{irmv1},\vrex~\citep{vrex},IB-IRM~\citep{ib-irm},CLOvE~\citep{clove},IGA~\citep{iga} and Fishr~\citep{fishr}
Besides, we also incorporate the state-of-the-art OOD objective proposed by~\citet{pair} that
is able to resolve both \cmnist-025 and \cmnist-01.

\paragraph{Evaluation of feature learning methods.}
For the sake of fairness in comparison, by default, we train all feature learning methods
by the same number of epochs and rounds (if applicable).
For the implementation \rfc,
we strictly follow the recommended setups provided by~\citet{rfc},~\footnote{\url{https://github.com/TjuJianyu/RFC}}
where we train the model with \rfc by $2$ rounds with $50$ epochs for round $1$, $500$ epochs for round $2$, and $500$ epochs for the synthesize round in \cmnist-025.
While in \cmnist-01, round $1$ contains $150$ epochs, round $2$ contains $400$ epochs and the synthesize round contains $500$ epochs.
For the implementation of \feat, we train the model with $2$ rounds of \feat in \cmnist-025,
and $3$ rounds of \feat in \cmnist-01, where each round contains $150$ epochs.
While for the retain penalty, we find using a fixed number of $0.01$ already achieved
sufficiently good performance.
ERM only contains $1$ round, for which we train the model with $150$ epochs in \cmnist-025
as we empirically find more epochs will incur severe performance degeneration in \cmnist-025.
While in \cmnist-01, we train the model with ERM by $500$ epochs to match up the overall
training epochs of \feat and \rfc.
We provide a detailed distribution of the number of epochs in each round in Table~\ref{tab:cmnist_epoch_appdx}.
\begin{table}[ht]
    \caption{Number of epochs in each round of various feature learning algorithms.}
    \label{tab:cmnist_epoch_appdx}
    \begin{center}
        \begin{small}
            \begin{sc}
                \begin{tabular}{lcccc}
                    \toprule
                    CMNIST-025 & Round-1 & Round-2 & Round-3 & Syn. Round \\\midrule
                    ERM        & 150     & -       & -       & -          \\
                    \rfc       & 50      & 150     & -       & 500        \\
                    \feat      & 150     & 150     & -       & -          \\\midrule
                    CMNIST-01  & Round-1 & Round-2 & Round-3 & Syn. Round \\
                    \midrule
                    ERM        & 500     & -       & -       & -          \\
                    \rfc       & 150     & 400     & -       & 500        \\
                    \feat      & 150     & 150     & 150     & -          \\
                    \bottomrule
                \end{tabular}
            \end{sc}
        \end{small}
    \end{center}
\end{table}
It can be found that, although \rfc costs $2-3$ times of training epochs
more than ERM and \feat, \rfc does not necessarily find better feature representations for OOD training, as demonstrated in Table.~\ref{tab:sythetic}.
In contrast, \feat significantly and consistently learns richer features given both \cmnist-025 and \cmnist-01 than ERM,
which shows the superiority of \feat.

\paragraph{The termination check in \feat.}
A key difference between \feat and previous rich feature learning algorithms is that \feat is able to perform the automatic termination check and learn the desired features stably.
As elaborated in Sec.~\ref{CH:FeAT:sec:fat_alg}, \feat can terminate automatically by inspecting the retention accuracy.
To verify, we list the \feat performances in various subsets of \cmnist-025 and \cmnist-01 at different rounds.
We use a termination accuracy of $130\%$, which trades off the exploration (i.e., training accuracy as $80\%$) and the retention (i.e., retention accuracy as $50\%$) properly.
As shown in Table~\ref{tab:termin_check_appdx}, in \cmnist-025 (\cmnist-01), after \feat learns sufficiently good features at Round $2$ ($3$), respectively, it is not necessary to proceed with Round $3$ ($4$) as it will destroy the already learned features and lead to degenerated retention performance (i.e., the sum of training and retention accuracies is worse than $130\%$.

\begin{table}[ht]
    \caption{Performances in various sets at different \feat rounds.}
    \label{tab:termin_check_appdx}
    \begin{center}
        \begin{scriptsize}
            \begin{sc}
                \begin{tabular}{lcccc}
                    \toprule
                    \cmnist-025    & Round-1         & Round-2         & Round-3                           \\
                    \midrule
                    Training Acc.  & 85.08$\pm$ 0.14 & 71.87$\pm$ 0.96 & 84.93$\pm$ 1.26                   \\
                    Retention Acc. & -               & 88.11$\pm$ 4.28 & 43.82$\pm$ 0.59                   \\
                    OOD Acc.       & 11.08$\pm$ 0.30 & 70.64$\pm$ 0.62 & 10.07$\pm$ 0.26                   \\\midrule
                    \cmnist-01     & Round-1         & Round-2         & Round-3         & Round-4         \\\midrule
                    Training Acc.  & 88.63$\pm$ 0.15 & 74.25$\pm$ 1.23 & 86.07$\pm$ 0.36 & 77.29$\pm$ 0.24 \\
                    Retention Acc. & -               & 85.91$\pm$ 1.78 & 48.05$\pm$ 1.39 & 29.09$\pm$ 1.15 \\
                    OOD Acc.       & 73.50$\pm$ 0.41 & 17.32$\pm$ 2.69 & 85.40$\pm$ 0.54 & 12.48$\pm$ 2.85 \\
                    \bottomrule
                \end{tabular}
            \end{sc}
        \end{scriptsize}
    \end{center}
\end{table}

\subsection{More details about \wilds experiments}
\label{CH:FeAT:sec:wilds_appdx}

In this section, we provide more details about the \wilds datasets used in the experiments
as well as the evaluation setups.

\subsubsection{Dataset description.}
To evaluate the feature learning performance given data from realistic scenarios,
we select $6$ challenging datasets from \wilds~\citep{wilds} benchmark.
The datasets contain various realistic distribution shifts,
ranging from domain distribution shifts, subpopulation shifts and the their mixed.
A summary of the basic information and statistics of the selected \textsc{Wilds} datasets can be found in Table.~\ref{tab:wilds_summary_appdx}, Table.~\ref{tab:wilds_stat_appdx}, respectively.
In the following, we will give a brief introduction to each of the datasets. More details can be found in the \wilds paper~\citep{wilds}.
\begin{table}[ht!]
    \centering
    \caption{A summary of datasets information from \wilds.}\label{tab:wilds_summary_appdx}
    \scalebox{0.7}{
        \begin{tabular}{lllllc}
            \toprule
            \textbf{Dataset}       & \textbf{Data ($x$)} & \textbf{Class information}         & \textbf{Domains}        & \textbf{Metric}        & \textbf{Architecture} \\
            \midrule
            \textsc{Amazon}        & Product reviews     & Star ratings (5 classes)           & 7,676 reviewers         & 10-eth percentile acc. & DistillBERT           \\
            \textsc{Camelyon17}    & Tissue slides       & Tumor (2 classes)                  & 5 hospitals             & Avg. acc.              & DenseNet-121          \\
            \textsc{CivilComments} & Online comments     & Toxicity (2 classes)               & 8 demographic groups    & Wr. group acc.         & DistillBERT           \\
            \textsc{FMoW}          & Satellite images    & Land use (62 classes)              & 16 years x 5 regions    & Wr. group acc.         & DenseNet-121          \\
            \textsc{iWildCam}      & Photos              & Animal species (186 classes)       & 324 locations           & Macro F1               & ResNet-50             \\
            \textsc{RxRx1}         & Cell images         & Genetic treatments (1,139 classes) & 51 experimental batches & Avg. acc               & ResNet-50             \\
            \bottomrule
        \end{tabular}
    }
\end{table}

\begin{table}[ht!]
    \centering
    \caption{A summary of datasets statistics from \wilds.}\label{tab:wilds_stat_appdx}
    \scalebox{0.8}{
        \begin{tabular}{lcccccccc}
            \toprule
            \multirow{2}{*}{Dataset} & \multicolumn{3}{c}{$\#$ Examples} &         & \multicolumn{3}{c}{$\#$ Domains}                            \\
            \cmidrule{2-4}\cmidrule{6-8}
                                     & train                             & val     & test                             &  & train & val   & test  \\
            \midrule
            \textsc{Amazon}          & 1,000,124                         & 100,050 & 100,050                          &  & 5,008 & 1,334 & 1,334 \\
            \textsc{Camelyon17}      & 302,436                           & 34,904  & 85,054                           &  & 3     & 1     & 1     \\
            \textsc{CivilComments}   & 269,038                           & 45,180  & 133,782                          &  & -     & -     & -     \\
            \textsc{FMoW}            & 76,863                            & 19,915  & 22,108                           &  & 11    & 3     & 2     \\
            \textsc{iWildCam}        & 129,809                           & 14,961  & 42,791                           &  & 243   & 32    & 48    \\
            \textsc{RxRx1}           & 40,612                            & 9,854   & 34,432                           &  & 33    & 4     & 14    \\
            \bottomrule
        \end{tabular}}
\end{table}

\textbf{Amazon.}
We follow the \wilds splits and data processing pipeline for the Amazon dataset~\citep{amazon}.
It provides $1.4$ million comments collected from $7,676$ Amazon customers.
The task is to predict the score (1-5 stars) for each review.
The domains $d$ are defined according to the reviewer/customer who wrote the product reviews.
The evaluation metric used for the task is $10$th percentile of per-user accuracies
in the OOD test sets, and the backbone model is a DistilBert~\citep{distillbert}, following the \wilds protocol~\citep{wilds}.

\textbf{Camelyon17.}
We follow the \wilds splits and data processing pipeline for the Camelyon17 dataset~\citep{camelyon}. It provides $450,000$ lymph-node scans from $5$ hospitals. The task in Camelyon17 is to take the input of $96\times96$ medical images to predict whether there exists a tumor tissue in the image. The domains $d$ refers to the index of the hospital where the image was taken. The training data are sampled from the first $3$ hospitals where the OOD validation and test data are sampled from the $4$-th and $5$-th hospital, respectively.
We will use the average accuracy as the evaluation metric and a DenseNet-121~\citep{densenet} as the backbone for the featurizer.

\textbf{CivilComments.}
We follow the \wilds splits and data processing pipeline for the CivilComments dataset~\citep{civil}. It provides $450,000$ comments collected from online articles. The task is to classify whether an online comment text is toxic or non-toxic. The domains $d$ are defined according to the demographic features, including male, female, LGBTQ, Christian, Muslim, other religions, Black, and White. CivilComments is used to study the subpopulation shifts, here we will use the worst group/domain accuracy as the evaluation metric. As for the backbone of the featurizer, we will use a DistillBert~\citep{distillbert} following \wilds~\citep{wilds}.

\textbf{FMoW.}
We follow the \wilds splits and data processing pipeline for the FMoW dataset~\citep{fmow}. It provides satellite images from $16$ years and $5$ regions. The task in FMoW is to classify the images into $62$ classes of building or land use categories. The domain is split according to the year that the satellite image was collected, as well as the regions in the image which could be Africa, America, Asia, Europe or Oceania. Distribution shifts could happen across different years and regions.
The training data contains data collected before $2013$, while the validation data contains images collected within $2013$ to $2015$, and the test data contains images collected after $2015$. The evaluation metric for FMoW is the worst region accuracy and the backbone model for the featurizer is a DenseNet-121~\citep{densenet}.

\textbf{iWildCam.}
We follow the \wilds splits and data processing pipeline for the iWildCam dataset~\citep{iwildcam}. It is consist of  $203,029$ heat or motion-activated photos of animal specifies from 323 different camera traps across different countries around the world. The task of iWildCam is to classify the corresponding animal specifies in the photos. The domains is split according to the locations of the camera traps which could introduce the distribution shifts. We will use the Macro F1 as the evaluation metric and a ResNet-50~\citep{resnet} as the backbone for the featurizer.

\textbf{RxRx1.}
We follow the \wilds splits and data processing pipeline for the RxRx1 dataset~\citep{rxrx1}.
The input is an image of cells taken by fluorescent microscopy. The cells can be genetically perturbed by siRNA and the task of RxRx1 is to predict the class of the corresponding siRNA that have treated the cells.
There exists $1,139$ genetic treatments and the domain shifts are introduced by the experimental batches. We will use the average accuracy of the OOD experimental batches as the evaluation metric and a ResNet-50~\citep{resnet} as the backbone for the featurizer.

\subsubsection{Training and evaluation details.}
\label{CH:FeAT:sec:eval_detail_appdx}
We follow previous works to implement and evaluate different methods used in our experiments~\citep{wilds}.
The information of the referred paper and code is listed as in Table.~\ref{tab:referred_code_appdx}.

\begin{table}[ht]
    \centering
    \small
    \caption{The information of the referred paper and code.}
    \label{tab:referred_code_appdx}
    \resizebox{\textwidth}{!}{
        \begin{tabular}{l|cc}
            \toprule
            Paper                    & Commit                                            & Code                                                             \\
            \midrule
            \wilds\citep{wilds}      & v2.0.0                                            & \url{https://wilds.stanford.edu/}                                \\
            Fish~\citep{fish}        & \texttt{333efa24572d99da0a4107ab9cc4af93a915d2a9} & \url{https://github.com/YugeTen/fish}                            \\
            \rfc~\citep{rfc}         & \texttt{33b9ecad0ce8b3462793a2da7a9348d053c06ce0} & \url{https://github.com/TjuJianyu/RFC}                           \\
            DFR~\citep{dfr,dfrlearn} & \texttt{6d098440c697a1175de6a24d7a46ddf91786804c} & \url{https://github.com/izmailovpavel/spurious_feature_learning} \\
            \bottomrule
        \end{tabular}}
\end{table}

The general hyperparemter setting inherit from the referred codes and papers, and are as listed in Table~\ref{tab:hyper_wilds_appdx}.
We use the same backbone models to implement the featurizer~\citep{resnet,densenet,distillbert}.
By default, we repeat the experiments by $3$ runs with the random seeds of $0,1,2$. While for Camelyon17, we follow the official guide to repeat $10$ times with the random seeds from $0$ to $9$.

\begin{table}[ht]
    \centering
    \small
    \caption{General hyperparameter settings for the experiments on \wilds.}
    \label{tab:hyper_wilds_appdx}
    \resizebox{\textwidth}{!}{
        \begin{tabular}{l|cccccc}
            \toprule
            Dataset              & \sc Amazon & \sc Camelyon17 & \sc  CivilComments            & \sc FMoW               & \sc iWildCam   & \sc RxRx1            \\\midrule
            Num. of seeds        & 3          & 10             & 3                             & 3                      & 3              & 3                    \\
            Learning rate        & 2e-6       & 1e-4           & 1e-5                          & 1e-4                   & 1e-4           & 1e-3                 \\
            Weight decay         & 0          & 0              & 0.01                          & 0                      & 0              & 1e-5                 \\
            Scheduler            & n/a        & n/a            & n/a                           & n/a                    & n/a            & Cosine Warmup        \\
            Batch size           & 64         & 32             & 16                            & 32                     & 16             & 72                   \\
            Architecture         & DistilBert & DenseNet121    & DistilBert                    & DenseNet121            & ResNet50       & ResNet50             \\
            Optimizer            & Adam       & SGD            & Adam                          & Adam                   & Adam           & Adam                 \\
            Domains in minibatch & 5          & 3              & 5                             & 5                      & 10             & 10                   \\
            Group by             & Countries  & Hospitals      & Demographics$\times$ toxicity & Times $\times$ regions & Trap locations & Experimental batches \\
            Training epochs      & 200        & 10             & 5                             & 12                     & 9              & 90                   \\
            \bottomrule
        \end{tabular}}
\end{table}

\paragraph{OOD objective implementations.}
We choose $4$ representative OOD objectives to evaluate the quality of learned features,
including GroupDRO~\citep{groupdro}, \irml~\citep{irmv1}, \vrex~\citep{vrex} and \irmx~\citep{pair}.
We implement the OOD objectives based on the code provided by~\citet{fish}.
For each OOD objective, by default, we follow the \wilds practice to sweep the penalty weights
from the range of $\{1e-2,1e-1,1,1e1,1e2\}$, and perform the model and hyperparameter
selection via the performance in the provided OOD validation set of each dataset.
Due to the overwhelming computational overhead required by large datasets and resource constraints,
we tune the penalty weight in iWildCam according to the performance with seed $0$,
which we empirically find yields similar results as full seed tunning. Besides in Amazon, we adopt the penalty weights tuned from CivilComments since the two datasets share a relatively high similarity, which we empirically find yields similar results as full seed tunning, too. On the other hand, it raises more challenges for feature learning algorithms in iWildCam and Amazon.

\paragraph{Deep Feature Reweighting (DFR) implementations.}
For the implementation of DFR~\citep{dfr,dfrlearn}, we use the code provided in~\citet{dfrlearn}.
By default, DFR considers the OOD validation as an unbiased dataset
and adopts the OOD validation set to learn a new classifier based on the frozen features from the pre-trained featurizer.
We follow the same implementation and evaluation protocol when evaluating feature learning quality in FMoW and CivilComments.
However, since Camelyon17 does not have the desired OOD validation set,
we follow the ``cheating'' protocol as in~\citet{dare} to perform the logistic regression
based the train and test sets.
Note that when ``cheating'', the model is not able to access the whole test sets.
Instead, the logistic regression is conducted on a random split of the concatenated train and test data.
Moreover, for Amazon and iWildCam, we find the original implementation fails to
converge possibly due to the complexity of the task, and the relatively poor feature learning quality.
Hence we implement a new logistic regression based on PyTorch~\citep{pytorch}
optimized with SGD, and perform DFR using ``cheating'' protocol based on the OOD validation set and test set.
Besides, we find neither the two aforementioned implementations or dataset choices
can lead to DFR convergence in RxRx, which we will leave for future investigations.

\paragraph{Feature learning algorithm implementations.}
We implement all the feature learning methods based on the Fish code framework.
For the fairness of comparison, we set all the methods to train the same number of steps
or rounds (if applicable) in \wilds datasets.
The only exception is in RxRx1, where both \rfc and \feat require more steps to converge,
since the initialized featurizer has a relatively large distance from the desired featurizer
in the task.
We did not train the model for much too long epochs as \citet{dfrlearn} find that
it only requires $2-5$ epochs for deep nets to learn high-quality invariant features.
The final model is selected based on the OOD validation accuracy during the training.
Besides, we tune the retain penalty in \feat by searching over $\{1e-2,1e-1,0.5,1,2,10\}$,
and finalize the retain penalty according to the OOD validation performance.
We list the detailed training steps and rounds setups, as well as the used retain penalty in \feat in Table~\ref{tab:wilds_step_appdx}.

\begin{table}[ht]
    \centering
    \caption{Hyperparameter setups of feature learning algorithms for the experiments on \wilds.}
    \label{tab:wilds_step_appdx}
    \resizebox{0.9\textwidth}{!}{
        \begin{tabular}{l|cccccc}
            \toprule
            Dataset              & \sc Amazon & \sc Camelyon17 & \sc  CivilComments & \sc FMoW & \sc iWildCam & \sc RxRx1 \\\midrule
            Overall steps        & 31,000     & 10,000         & 50,445             & 9,600    & 48,000       & 20,000    \\
            Approx. epochs       & 4          & 10             & 3                  & 4        & 10           & 10        \\
            Num. of rounds       & 3          & 2              & 3                  & 2        & 2            & 10        \\
            Steps per round      & 10,334     & 5,000          & 16,815             & 4,800    & 10           & 10        \\
            \feat Retain penalty & 2.0        & 1e-2           & 1e-2               & 1.0      & 0.5          & 10        \\
            \bottomrule
        \end{tabular}}
\end{table}

For ERM, we train the model simply by the overall number of steps, except for RxRx1 where we train the
model by $15,000$ steps following previous setups~\citep{fish}.
\rfc and \feat directly adopt the setting listed in the Table~\ref{tab:wilds_step_appdx}.
Besides, \rfc will adopt one additional round for synthesizing the pre-trained models from different rounds.
Although \citet{rfc} requires \rfc to train the two rounds for synthesizing the learned features,
we empirically find additional training steps in synthesizing will incur overfitting and worse performance.
Moreover, as \rfc requires propagating $2K-1$ batches of the data that may exceed the memory limits,
we use a smaller batch size when training \rfc in iWildCam ($8$) and RxRx1 ($56$).

\subsection{Software and hardware}
\label{CH:FeAT:sec:exp_software_appdx}
We implement our methods with PyTorch~\citep{pytorch}.
For the software and hardware configurations, we ensure the consistent environments for each datasets.
We run all the experiments on Linux servers with NVIDIA V100 graphics cards with CUDA 10.2.

\subsection{Computational analysis}
\label{CH:FeAT:sec:comput_analysis_appdx}
Compared to ERM, the additional computational and memory overhead introduced in \feat mainly lie in the \feat training and partitioning. At each training step, \feat needs $(k-1)$ additional forward and backward propagation, the same as Bonsai, while \feat only needs $\min(1,k-1)$ additional propagation. Besides, Bonsai additionally requires another round of training with $(K-1)$ additional propagation, given $K$ total rounds.

We calculated the computational overhead:
\begin{table}[ht]\centering\small
    \caption{Training and memory overhead of different algorithms.}
    \begin{tabular}{lllll}\toprule
               & Camelyon17            &                & CivilComments      &                \\
               & Training time         & Memory (\%)    & Training time      & Memory (\%)    \\  \midrule
        ERM    & 56.21$\pm$8.29 mins   & 22.56$\pm$0.00 & 24.22$\pm$0.33 hrs & 36.46$\pm$0.00 \\
        Bonsai & 214.55$\pm$1.13 mins  & 51.75$\pm$0.01 & 58.47$\pm$0.91 hrs & 64.43$\pm$0.31 \\
        \feat  & 101.14$\pm$12.79 mins & 51.92$\pm$0.04 & 28.19$\pm$1.15 hrs & 56.21$\pm$0.48 \\ \bottomrule
    \end{tabular}
\end{table}
The results aligned with our discussion. Bonsai requires much more time for the additional synthetic round and much more memory when there are 3 or more rounds. In contrast, \feat achieves the best performance without introducing  too much additional computational overhead.

\subsection{Feature learning analysis}
\label{CH:FeAT:sec:feat_analysis_appdx}
\begin{figure}[t!]
    \vspace{-0.15in}
    \centering
    \subfigure[ERM 150 epochs]{
        \includegraphics[width=0.5\textwidth]{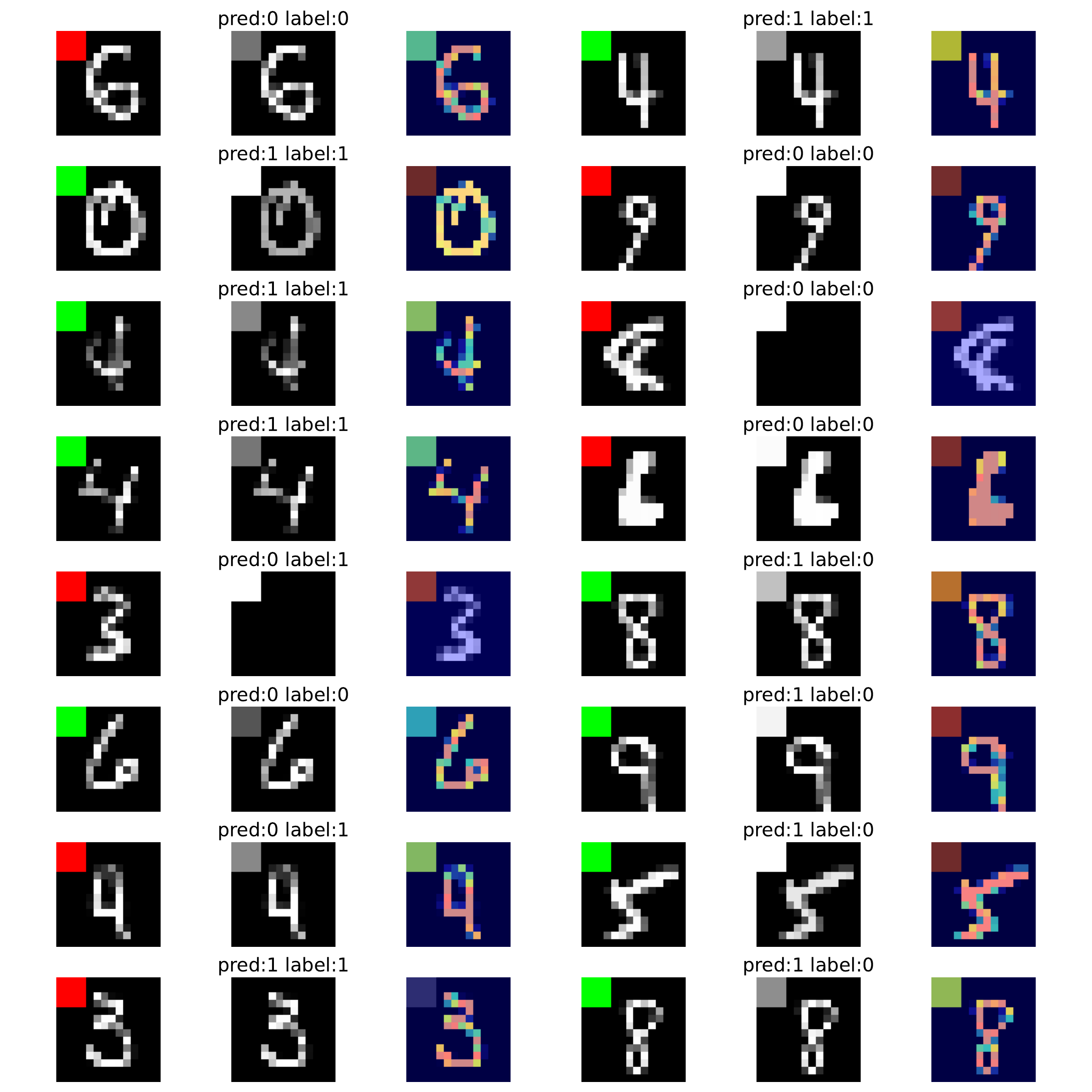}
    }%
    \subfigure[ERM 300 epochs]{
        \includegraphics[width=0.5\textwidth]{Figures/FeAT/visualization/s4_erm_ep300.pdf}
    }%
    \\
    \subfigure[ERM 450 epochs]{
        \includegraphics[width=0.5\textwidth]{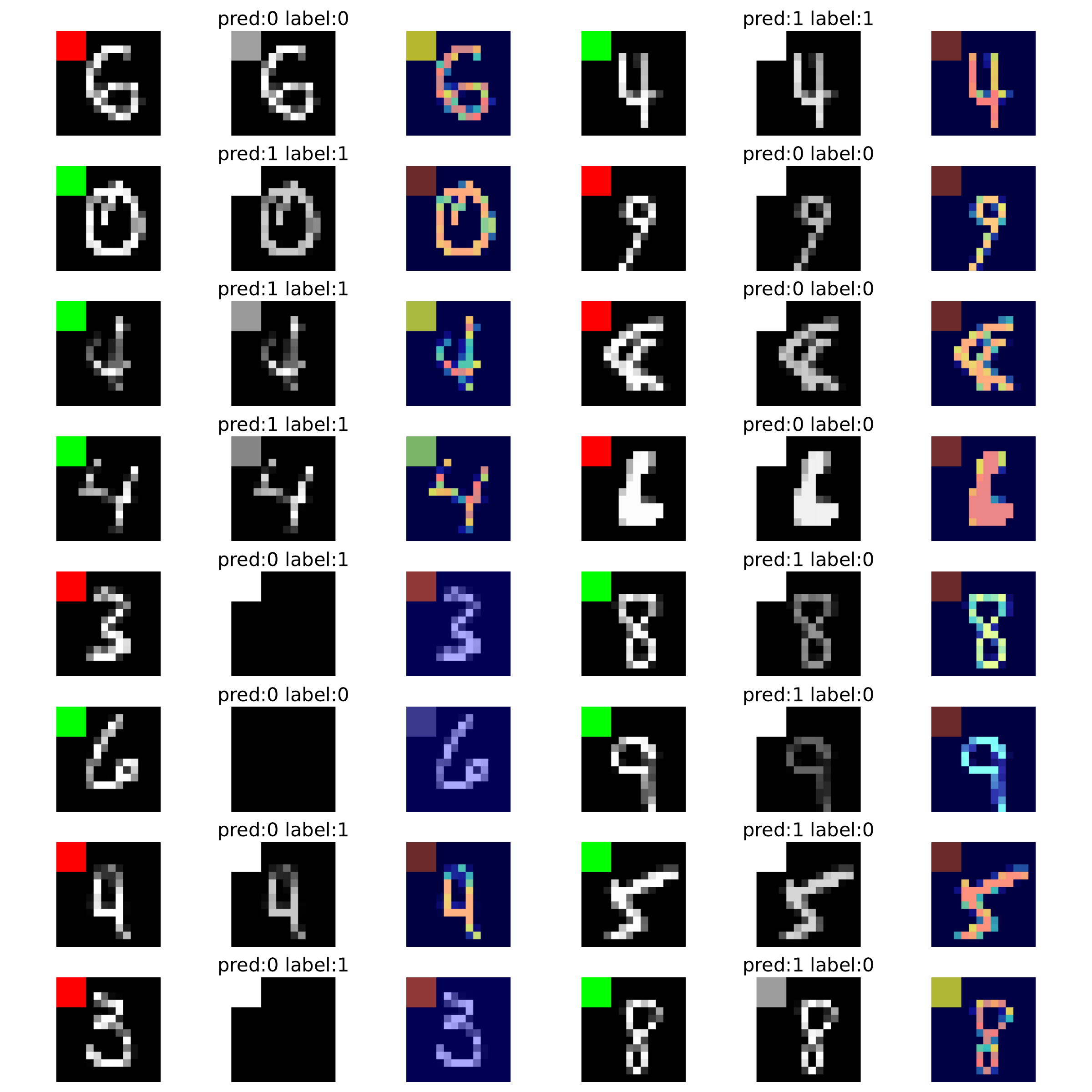}
    }%
    \subfigure[FeAT 2 rounds]{
        \includegraphics[width=0.5\textwidth]{Figures/FeAT/visualization/s4_ifat_r2.pdf}
    }
    \caption[GradCAM visualization on \cmnist-025.]{GradCAM visualization on \cmnist-025, where the shortcuts are now concentrated to a colored path at the up left. Three visualizations are drawn for each sample: the original figure, the gray-colored gradcam, and the gradcam. It can be found that ERM can not properly capture the desired features or even forget certain features with longer training epochs. \feat can stably capture the desired features.}
    \label{CH:FeAT:fig:gradcam_cmnist_appdx}
\end{figure}

We first visualize the feature learning of ERM and \feat on ColoredMNIST-025, as shown in Fig.~\ref{CH:FeAT:fig:gradcam_cmnist_appdx} It can be found that ERM can learn both invariant and spurious features to predict the label, aligned with our theory.

However, ERM focuses more on spurious features and even forgets certain features with longer training epochs, which could be due to multiple reasons such as the simplicity biases of ERM. Hence predictions based on ERM learned features fail to generalize to OOD examples. In contrast, \feat effectively captures the meaningful features for all samples and generalizes to OOD examples well.

\begin{table}[ht]\centering
    \caption{Labels and predictions for the visualized samples.}
    \label{tab:viz_label_appdx}
    \resizebox{\textwidth}{!}{
        \begin{tabular}{@{}lcccclcccc@{}}\toprule
                                                    & \multicolumn{1}{l}{Label} & \multicolumn{1}{l}{ERM} & \multicolumn{1}{l}{Bonsai} & \multicolumn{1}{l}{\feat} &                                    & \multicolumn{1}{l}{Label} & \multicolumn{1}{l}{ERM} & \multicolumn{1}{l}{Bonsai} & \multicolumn{1}{l}{\feat} \\\midrule
            \multirow{4}{*}{\textsc{Camelyon17}}    & 1                         & 1                       & 0                          & 1                         & \multirow{4}{*}{\textsc{iWildCam}} & 113                       & 68                      & 0                          & 113                       \\
                                                    & 1                         & 1                       & 0                          & 1                         &                                    & 113                       & 0                       & 0                          & 113                       \\
                                                    & 1                         & 1                       & 0                          & 1                         &                                    & 36                        & 36                      & 36                         & 36                        \\
                                                    & 1                         & 0                       & 0                          & 0                         &                                    & 36                        & 36                      & 36                         & 36                        \\\midrule
            \multirow{4}{*}{\textsc{FMoW}}          & 40                        & 40                      & 40                         & 40                        & \multirow{4}{*}{\textsc{RxRx1}}    & 1138                      & 812                     & 812                        & 812                       \\
                                                    & 40                        & 40                      & 40                         & 40                        &                                    & 1138                      & 1133                    & 1125                       & 1133                      \\
                                                    & 40                        & 2                       & 29                         & 29                        &                                    & 35                        & 43                      & 1119                       & 143                       \\
                                                    & 40                        & 40                      & 40                         & 40                        &                                    & 35                        & 35                      & 1054                       & 35                        \\\midrule
            \multirow{4}{*}{\textsc{CivilComments}} & toxic                     & toxic                   & toxic                      & toxic                     & \multirow{4}{*}{\textsc{Amazon}}   & 2                         & 3                       & 3                          & 2                         \\
                                                    & toxic                     & toxic                   & toxic                      & toxic                     &                                    & 5                         & 5                       & 5                          & 5                         \\
                                                    & toxic                     & toxic                   & toxic                      & toxic                     &                                    & 3                         & 4                       & 4                          & 4                         \\
                                                    & nontoxic                  & nontoxic                & nontoxic                   & nontoxic                  &                                    & 5                         & 5                       & 5                          & 5                         \\\bottomrule
        \end{tabular}}
\end{table}

\begin{figure}[t!]\centering
    \subfigure[\textsc{CivilComments}]{
        \includegraphics[width=0.8\textwidth]{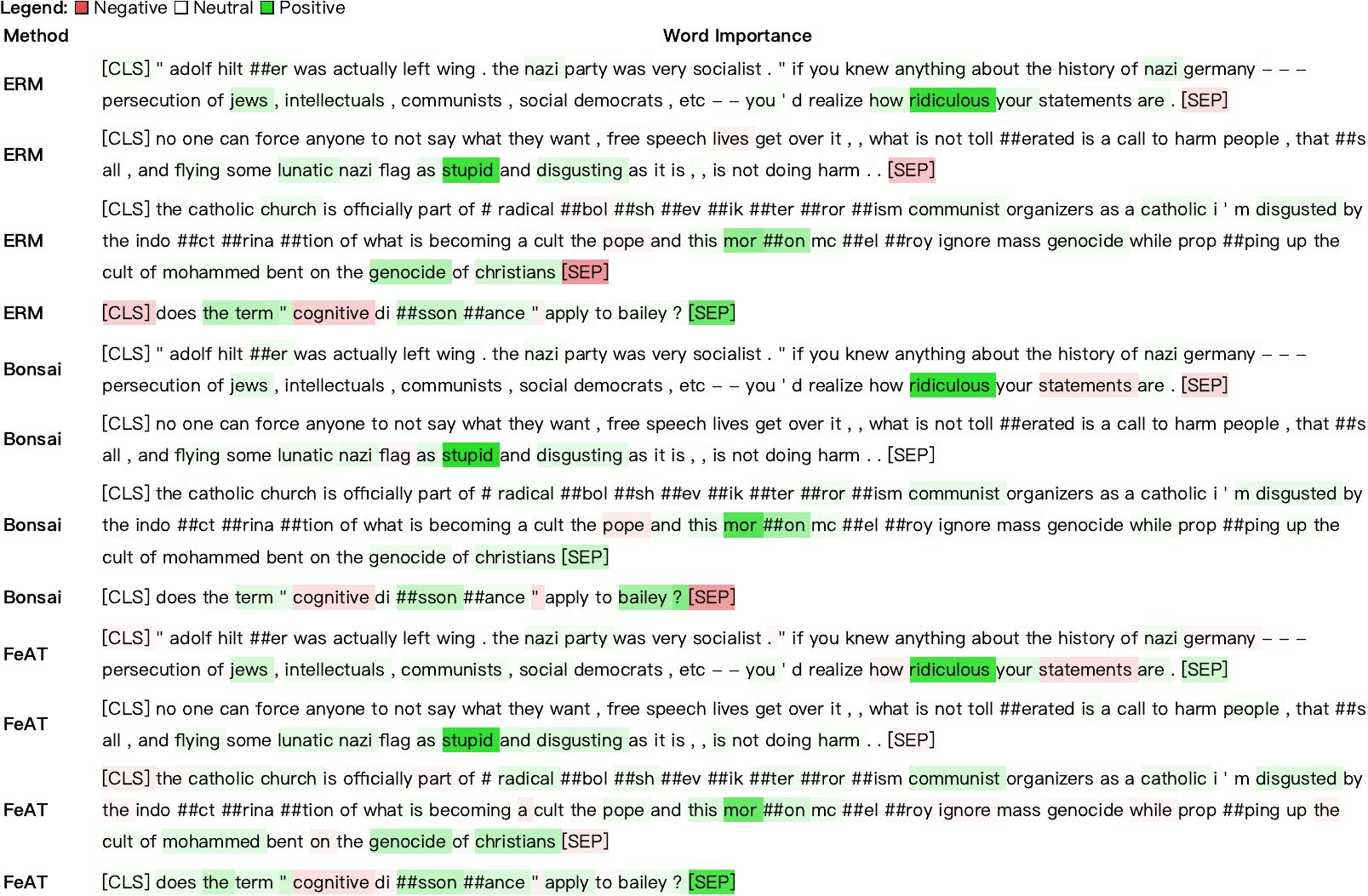}
    }
    \caption[Saliency map of feature learning on \textsc{CivilComments} benchmark.]{Saliency map of feature learning on \textsc{CivilComments} benchmark. The green-colored tokens are the learned features that contributed most to the target class, while the red-colored tokens contributed to the other classes. It can be found that \feat is able to learn more meaningful and diverse features than ERM and Bonsai.}
    \label{CH:FeAT:fig:gradcam_wilds_nlp_appdx}
\end{figure}

\begin{figure}[t!]
    \centering
    \subfigure[\textsc{Amazon}-ERM]{
        \includegraphics[width=0.8\textwidth]{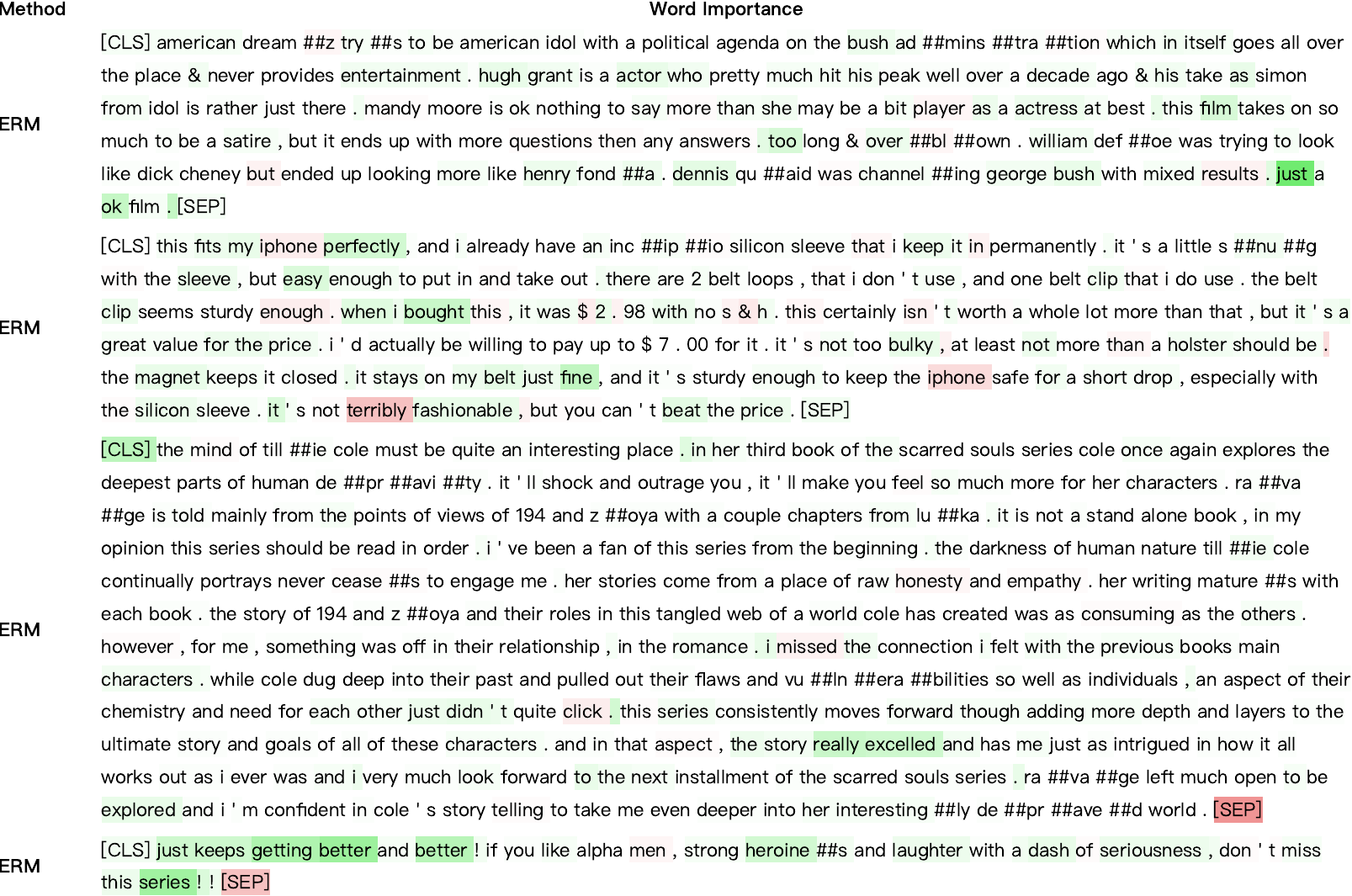}
    }
    \subfigure[\textsc{Amazon}-Bonsai]{
        \includegraphics[width=0.8\textwidth]{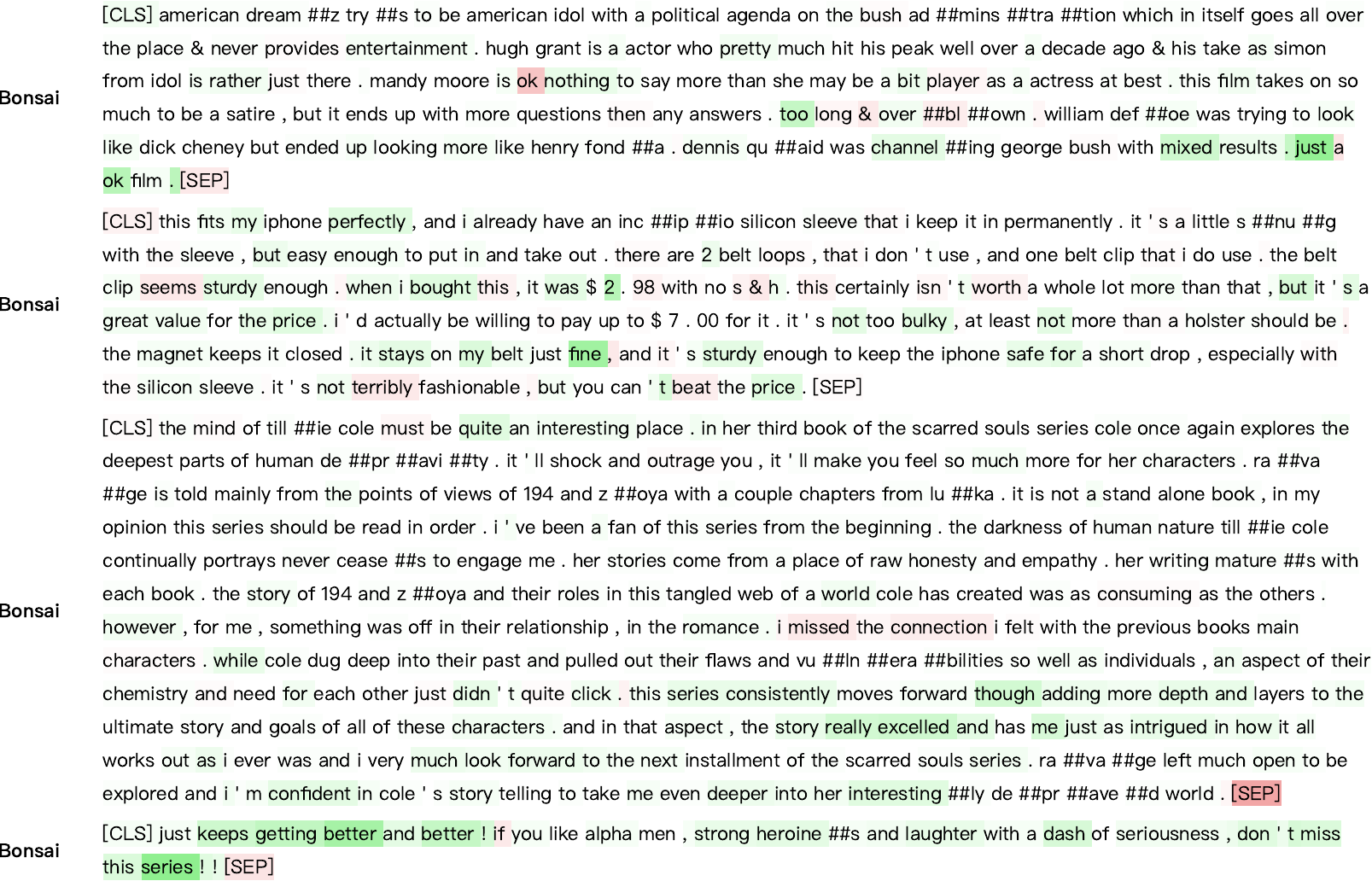}
    }
    \caption[Saliency map of feature learning on \textsc{Amazon} benchmark.]{Saliency map of feature learning on \textsc{Amazon} benchmark. The green-colored tokens are the learned features that contributed most to the target class, while the red-colored tokens contributed to the other classes. It can be found that \feat is able to learn more meaningful and diverse features than ERM and Bonsai.}
    \label{CH:FeAT:fig:gradcam_wilds_nlp2_appdx}
\end{figure}

\begin{figure}[t!]
    \centering
    \subfigure[\textsc{Amazon}-\feat]{
        \includegraphics[width=\textwidth]{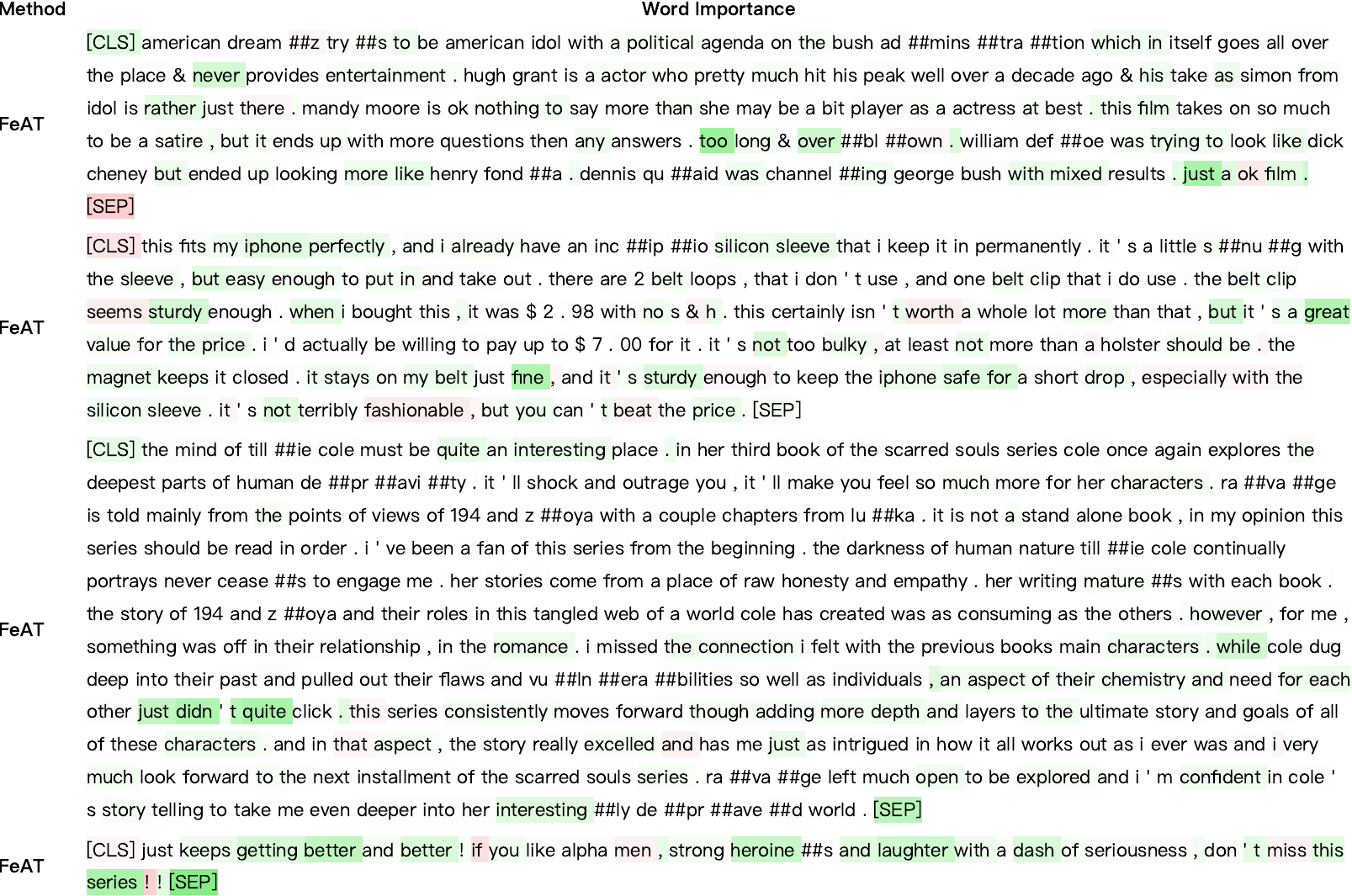}
    }
    \caption[aliency map of feature learning on \textsc{Amazon} benchmark (part 2).]{Saliency map of feature learning on \textsc{Amazon} benchmark (part 2). The green-colored tokens are the learned features that contributed most to the target class, while the red-colored tokens contributed to the other classes. It can be found that \feat is able to learn more meaningful and diverse features than ERM and Bonsai.}
    \label{CH:FeAT:fig:gradcam_wilds_nlp3_appdx}
    \vspace{-2cm}
\end{figure}

\begin{figure}[t!]
    \vspace{-0.15in}
    \centering
    \subfigure[\textsc{Camelyon17}-ERM]{
        \includegraphics[width=0.3\textwidth]{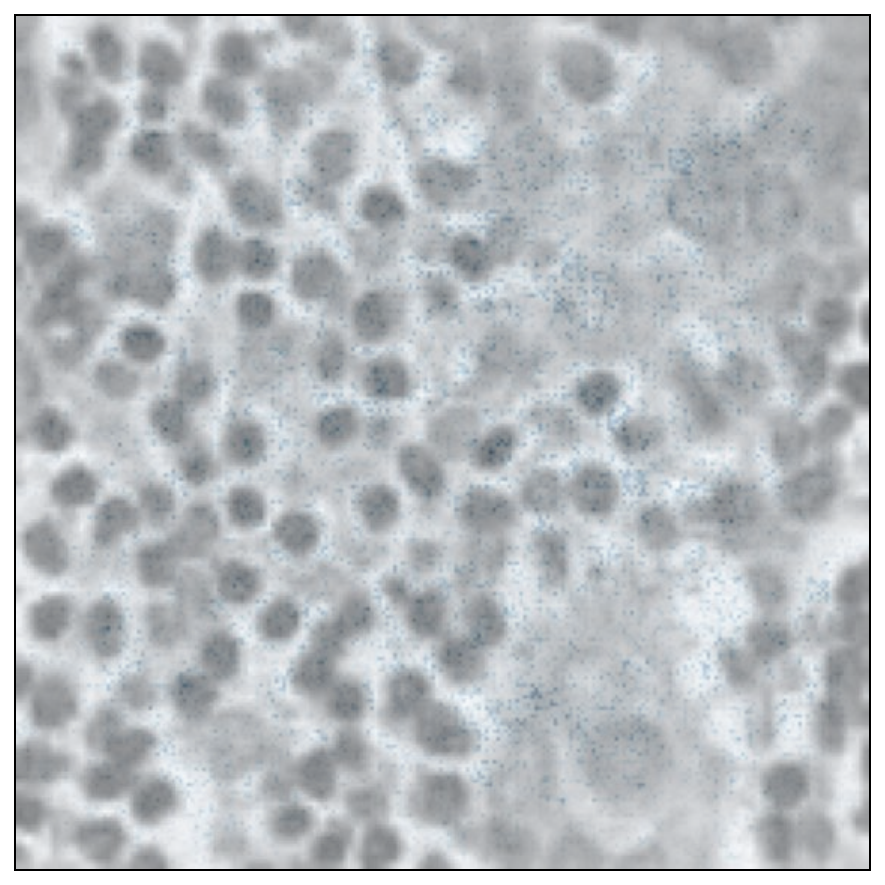}
    }%
    \subfigure[\textsc{Camelyon17}-Bonsai]{
        \includegraphics[width=0.3\textwidth]{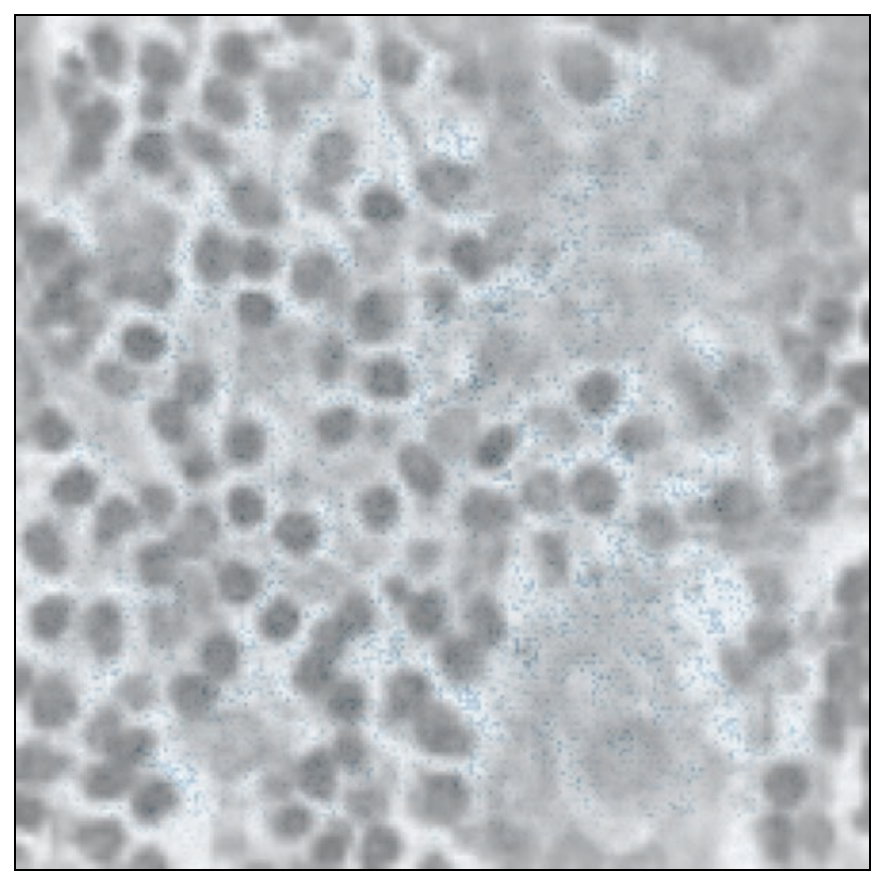}
    }%
    \subfigure[\textsc{Camelyon17}-\feat]{
        \includegraphics[width=0.3\textwidth]{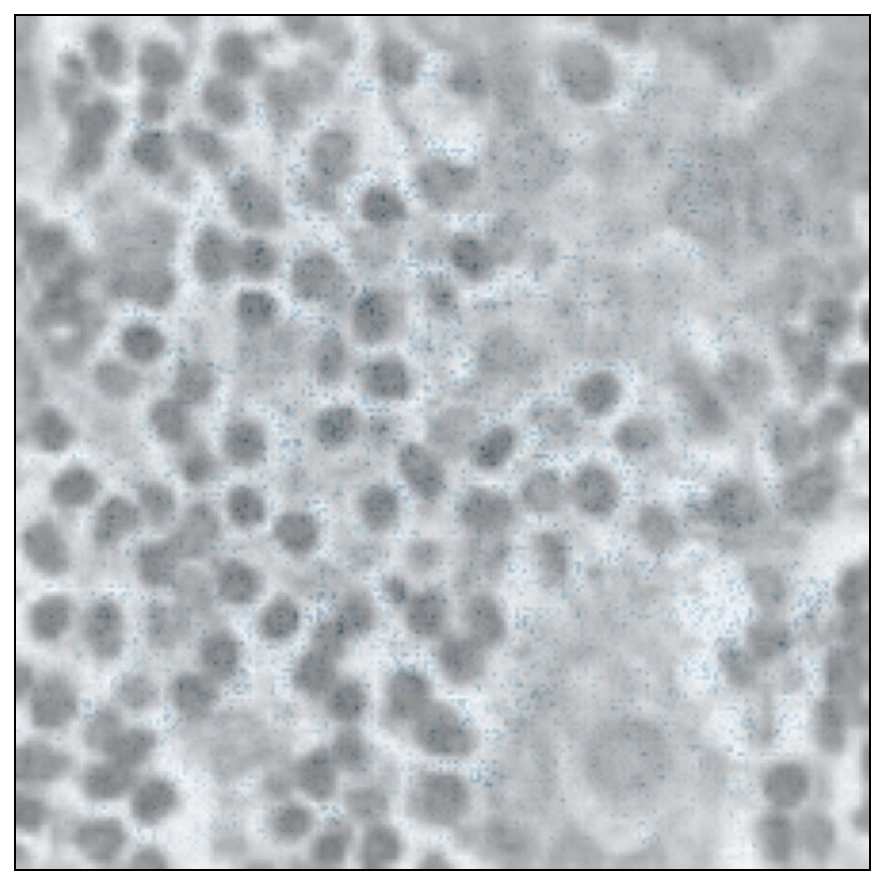}
    }%
    \\
    \subfigure[\textsc{Camelyon17}-ERM]{
        \includegraphics[width=0.3\textwidth]{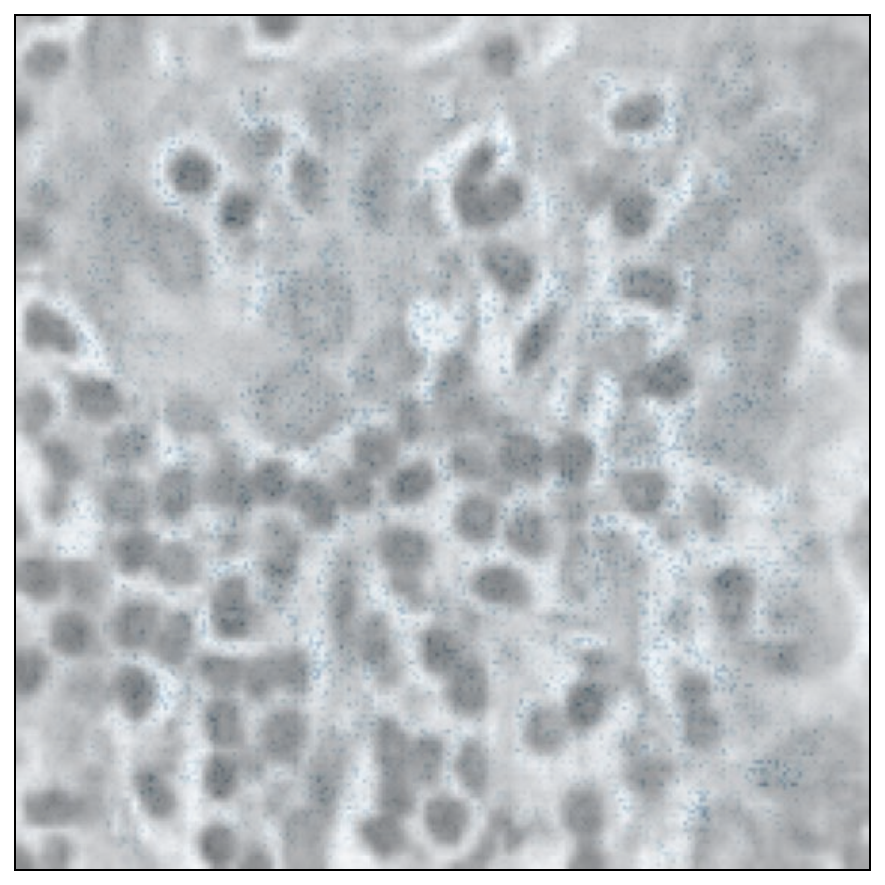}
    }%
    \subfigure[\textsc{Camelyon17}-Bonsai]{
        \includegraphics[width=0.3\textwidth]{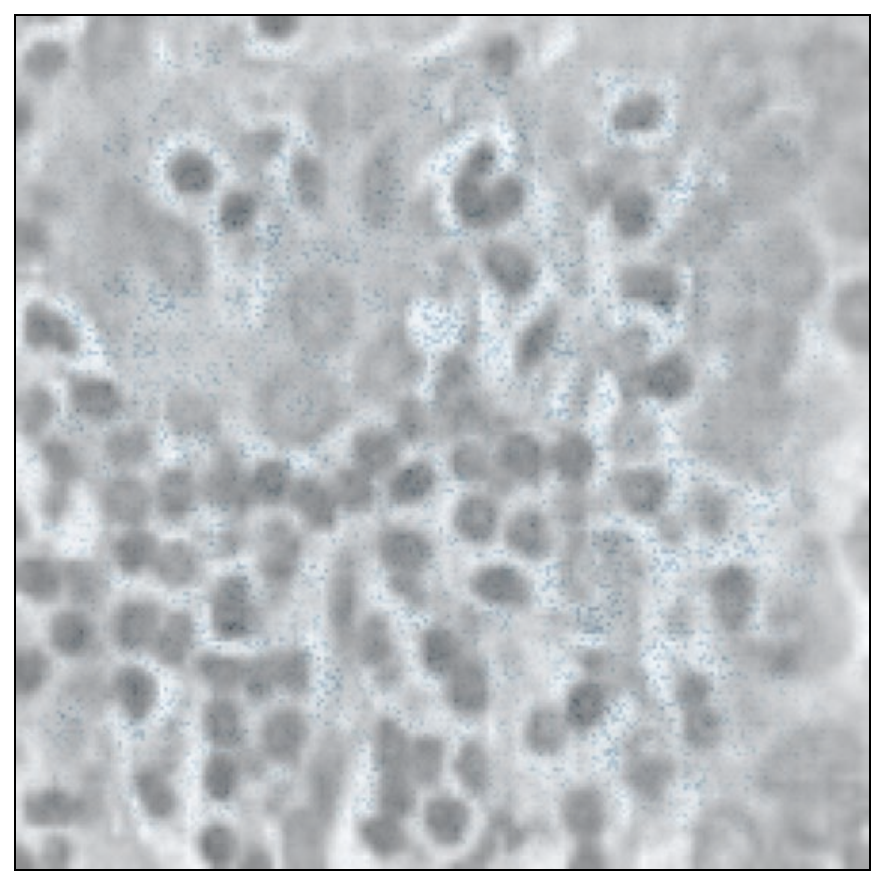}
    }%
    \subfigure[\textsc{Camelyon17}-\feat]{
        \includegraphics[width=0.3\textwidth]{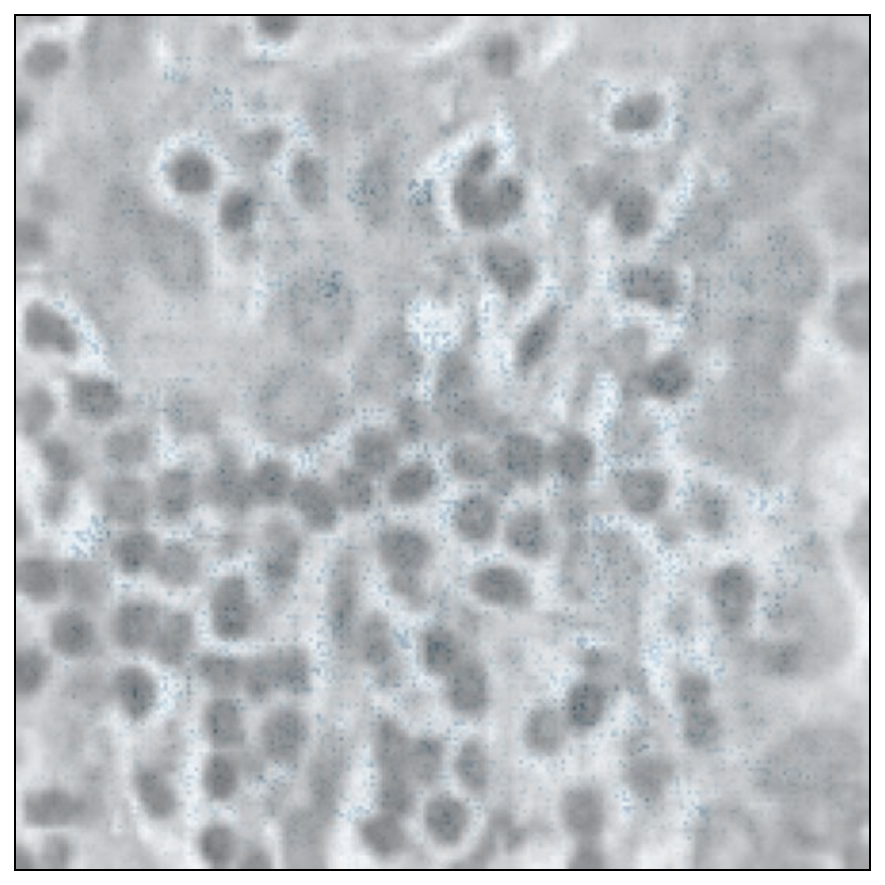}
    }%
    \\
    \subfigure[\textsc{Camelyon17}-ERM]{
        \includegraphics[width=0.3\textwidth]{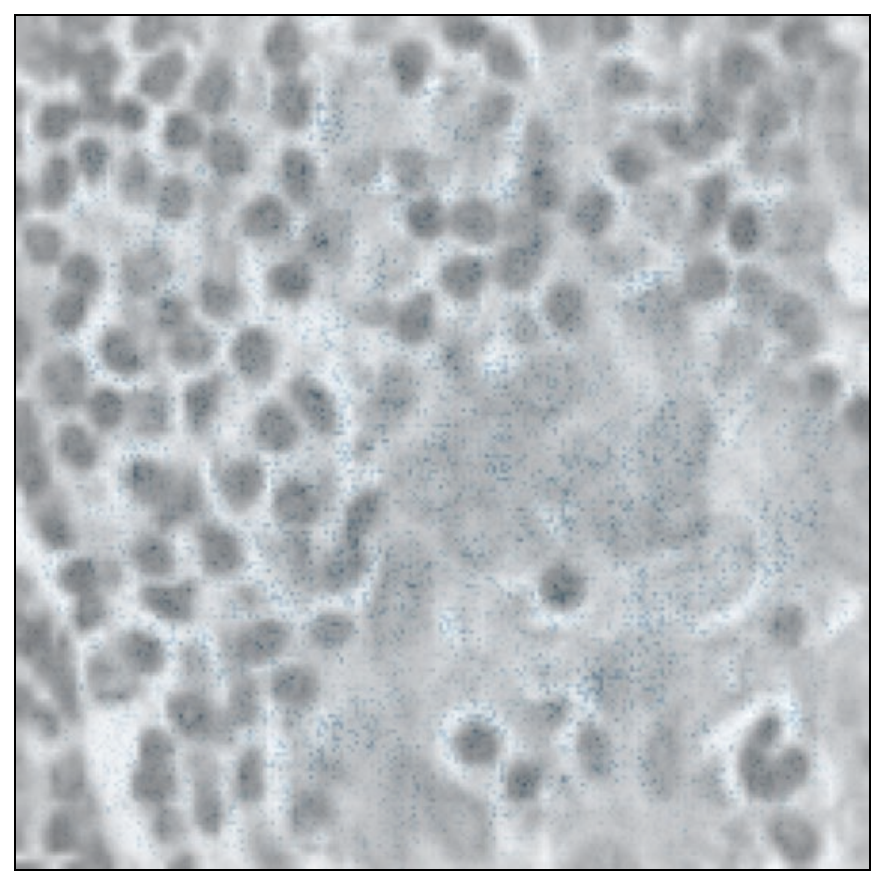}
    }%
    \subfigure[\textsc{Camelyon17}-Bonsai]{
        \includegraphics[width=0.3\textwidth]{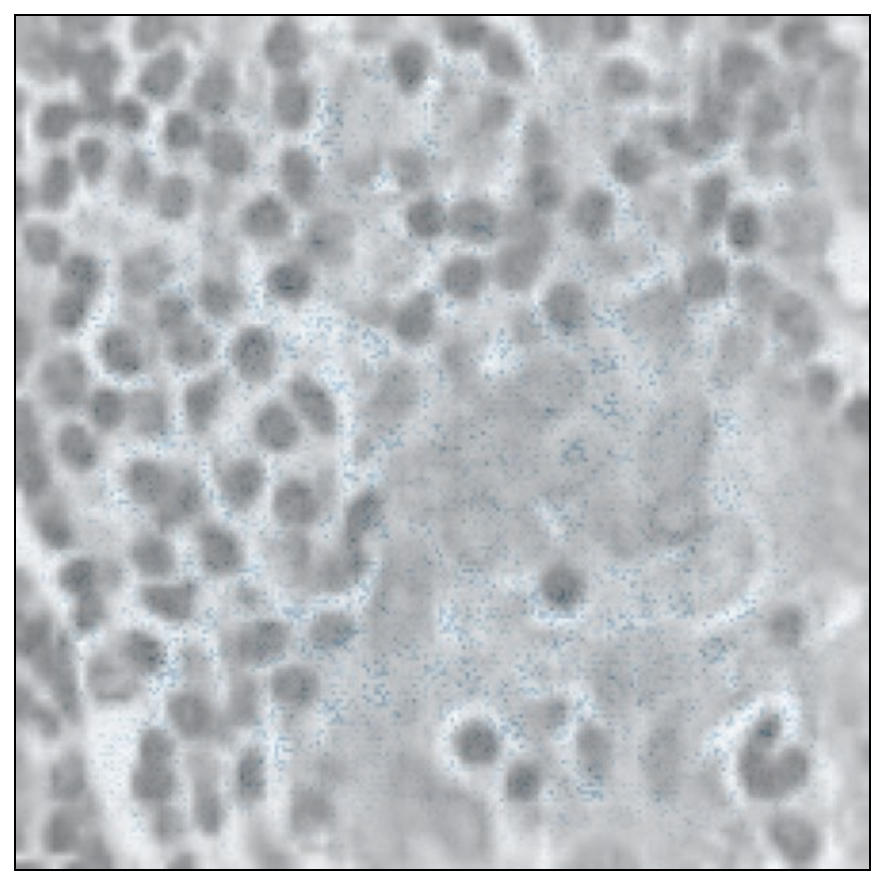}
    }%
    \subfigure[\textsc{Camelyon17}-\feat]{
        \includegraphics[width=0.3\textwidth]{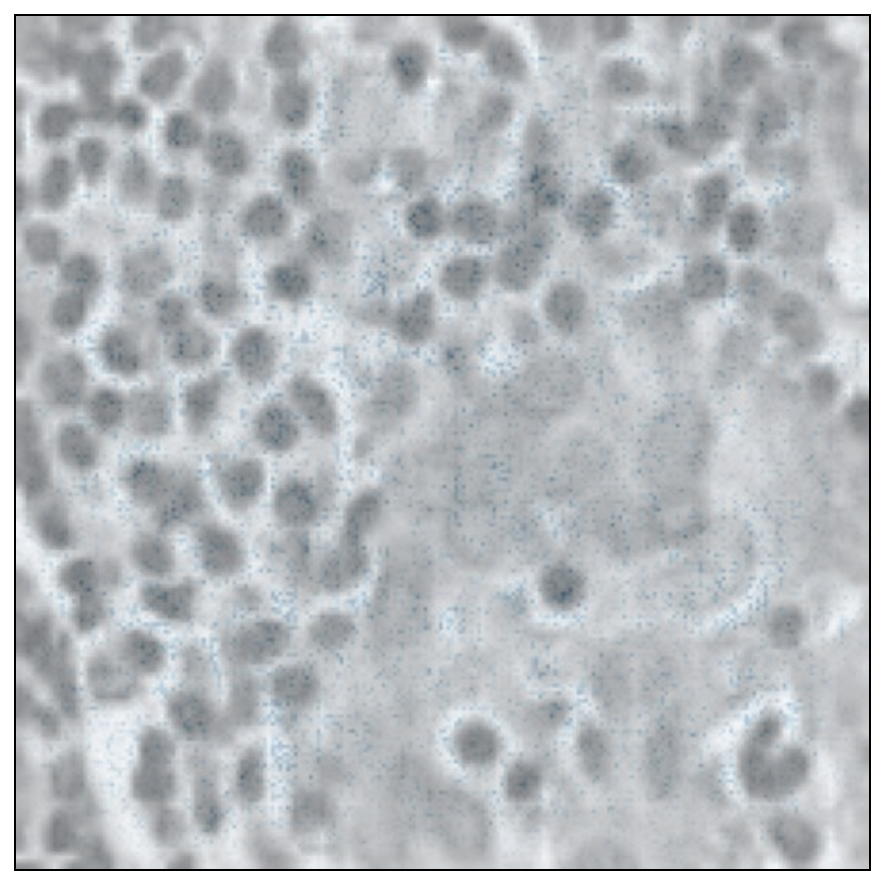}
    }%
    \\
    \subfigure[\textsc{Camelyon17}-ERM]{
        \includegraphics[width=0.3\textwidth]{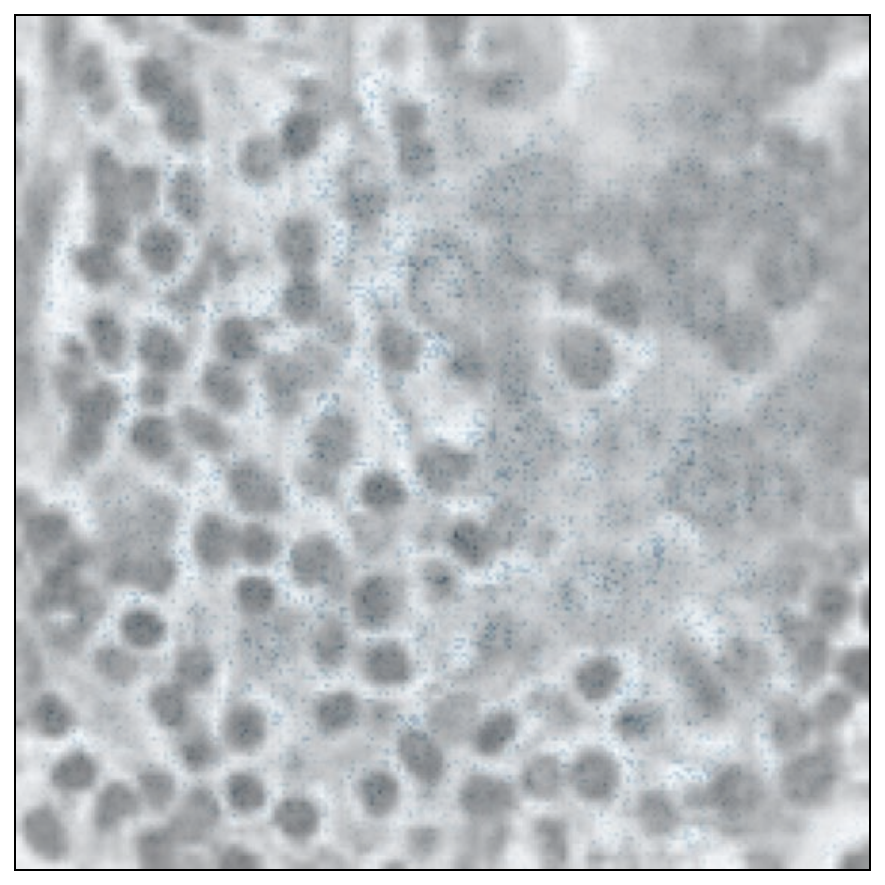}
    }%
    \subfigure[\textsc{Camelyon17}-Bonsai]{
        \includegraphics[width=0.3\textwidth]{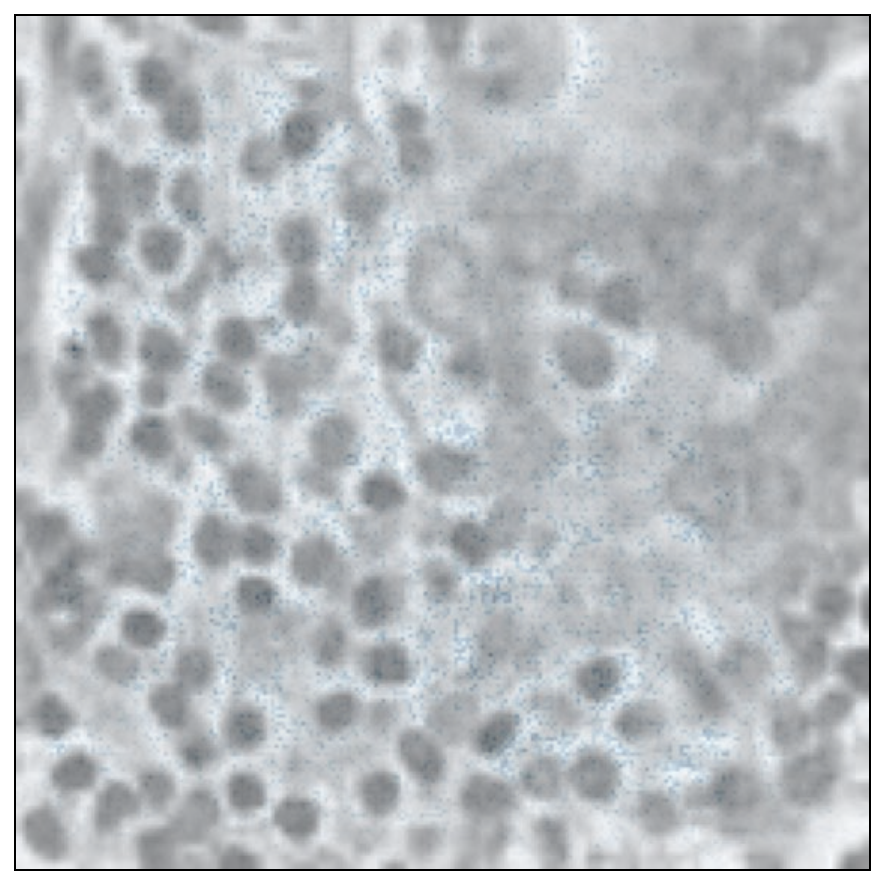}
    }%
    \subfigure[\textsc{Camelyon17}-\feat]{
        \includegraphics[width=0.3\textwidth]{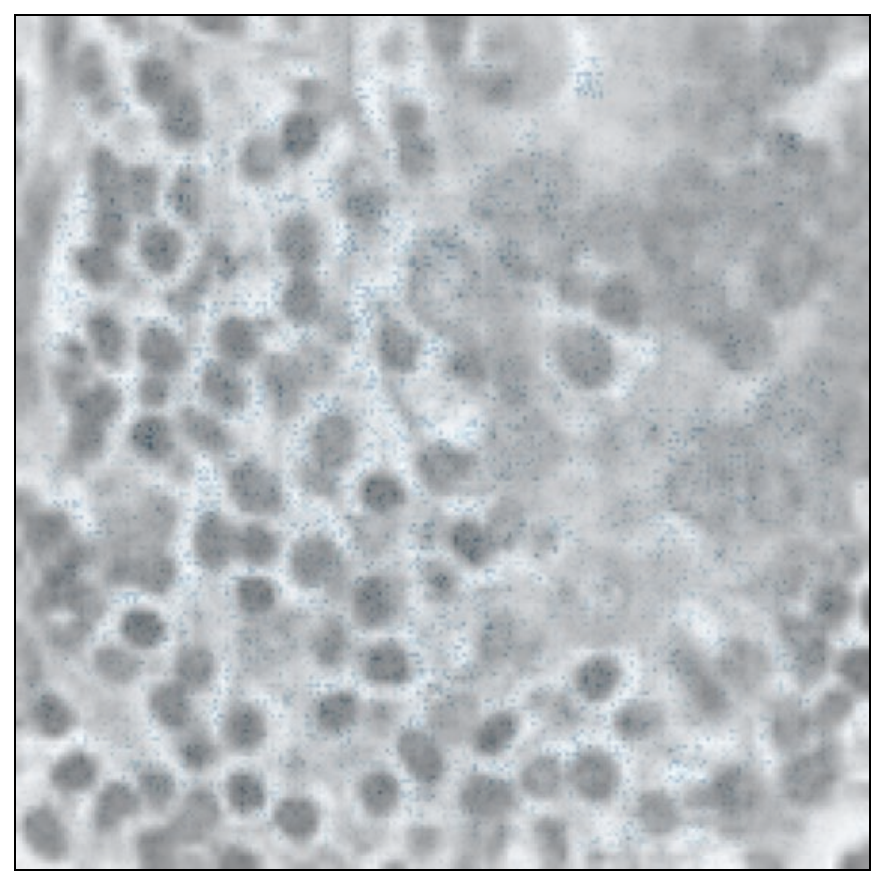}
    }%
    \\
    \vspace{-0.15in}
    \caption[Saliency map of feature learning on \textsc{Camelyon17} benchmark.]{Saliency map of feature learning on \textsc{Camelyon17} benchmark. The blue dots are the salient features.
        A deeper blue color denotes more salient features. It can be found that \feat is able to learn more meaningful and diverse features than ERM and Bonsai.}
    \label{CH:FeAT:fig:gradcam_camelyon17_appdx}
    \vspace{-2cm}
\end{figure}

\begin{figure}[t!]
    \vspace{-0.15in}
    \centering
    \subfigure[\textsc{FMoW}-ERM]{
        \includegraphics[width=0.3\textwidth]{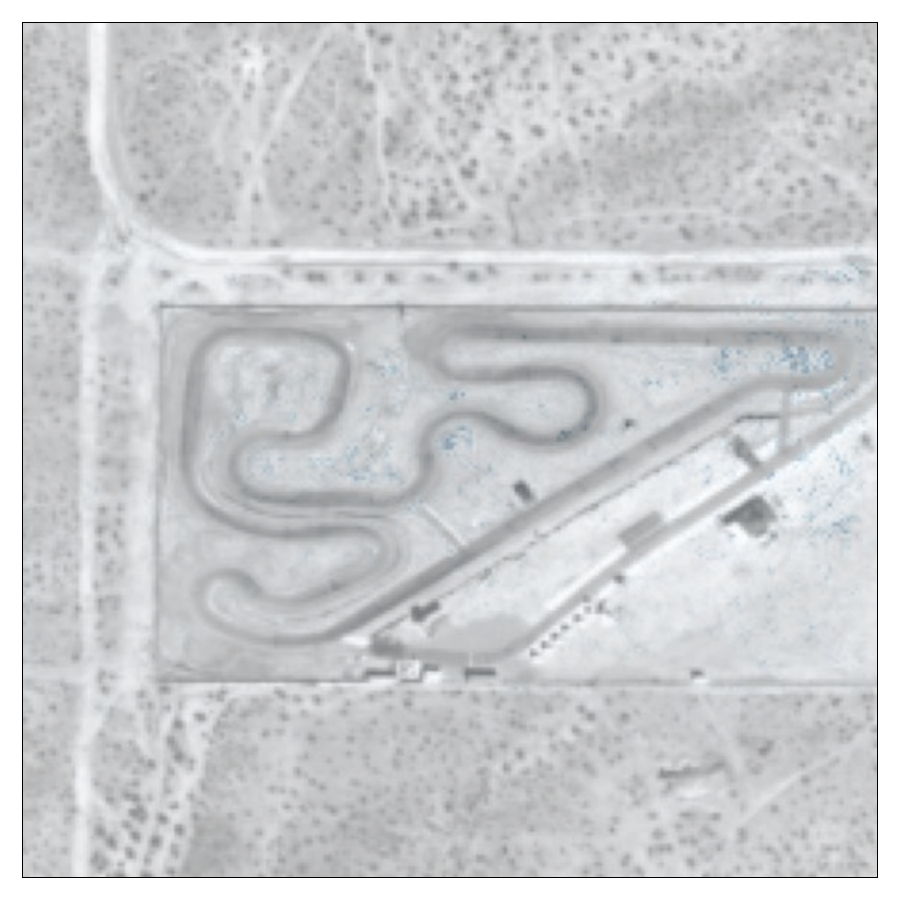}
    }%
    \subfigure[\textsc{FMoW}-Bonsai]{
        \includegraphics[width=0.3\textwidth]{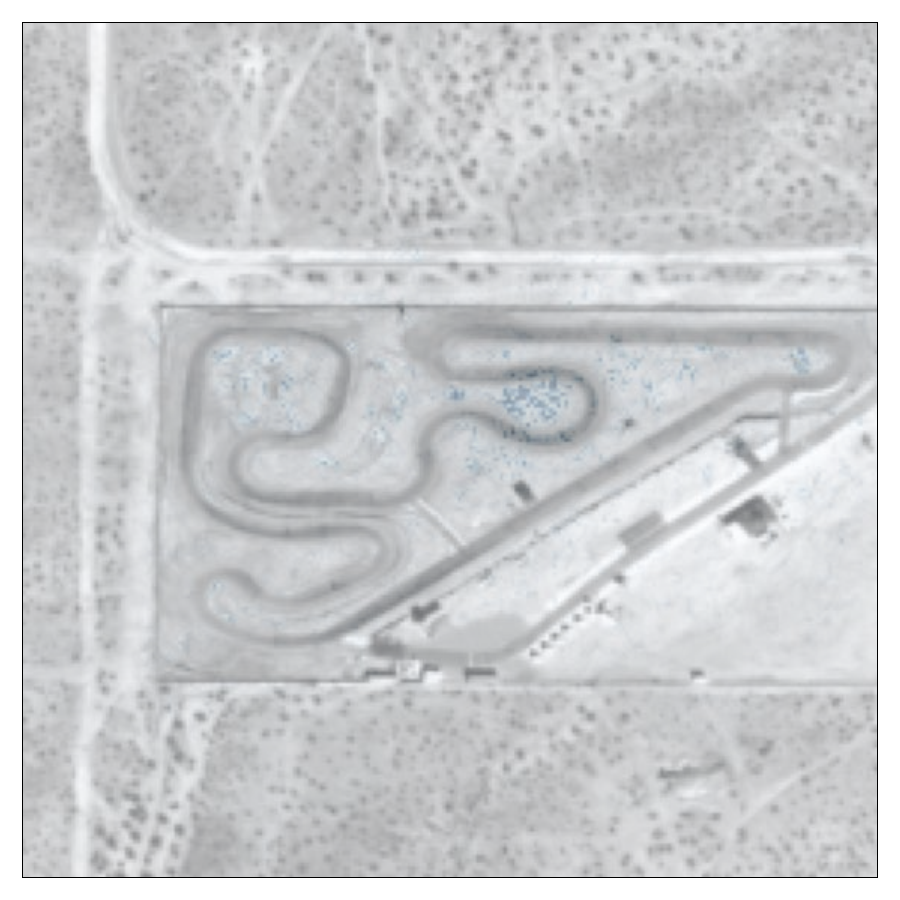}
    }%
    \subfigure[\textsc{FMoW}-\feat]{
        \includegraphics[width=0.3\textwidth]{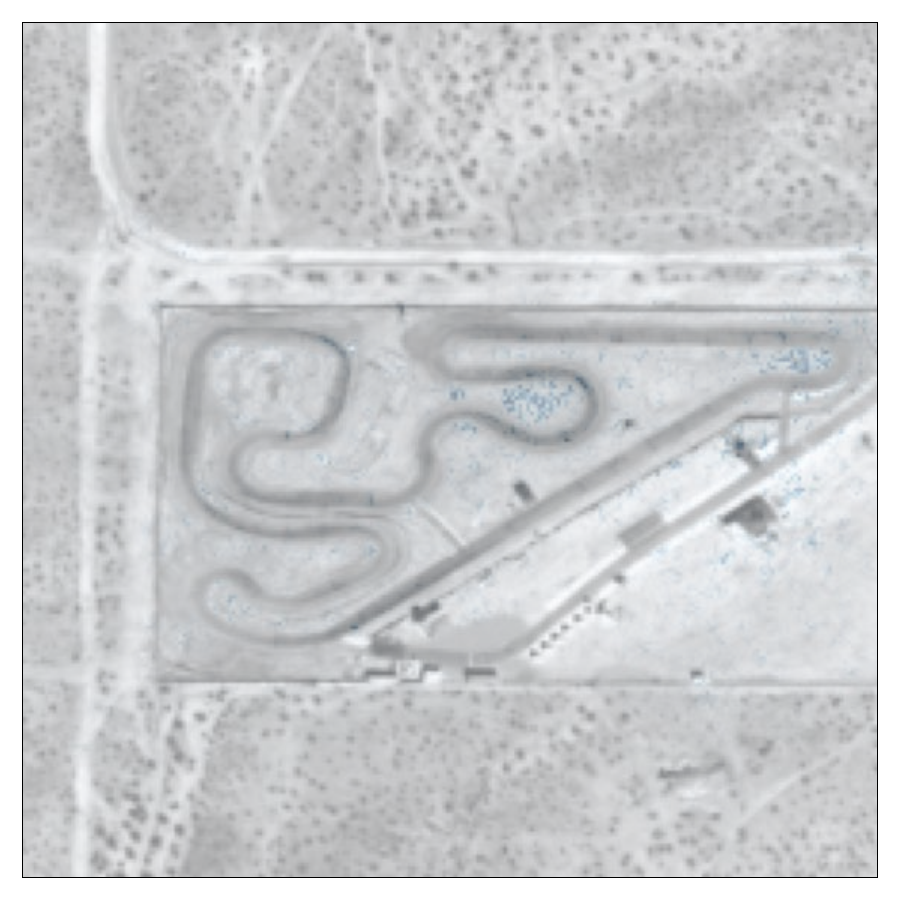}
    }%
    \\
    \subfigure[\textsc{FMoW}-ERM]{
        \includegraphics[width=0.3\textwidth]{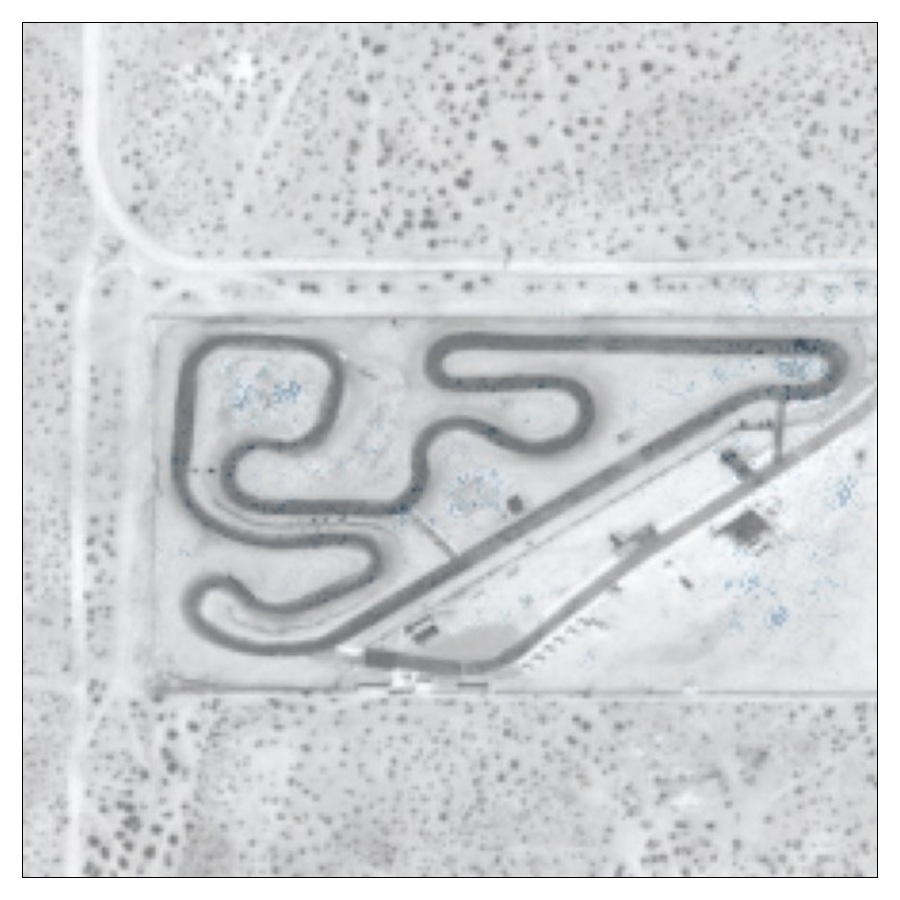}
    }%
    \subfigure[\textsc{FMoW}-Bonsai]{
        \includegraphics[width=0.3\textwidth]{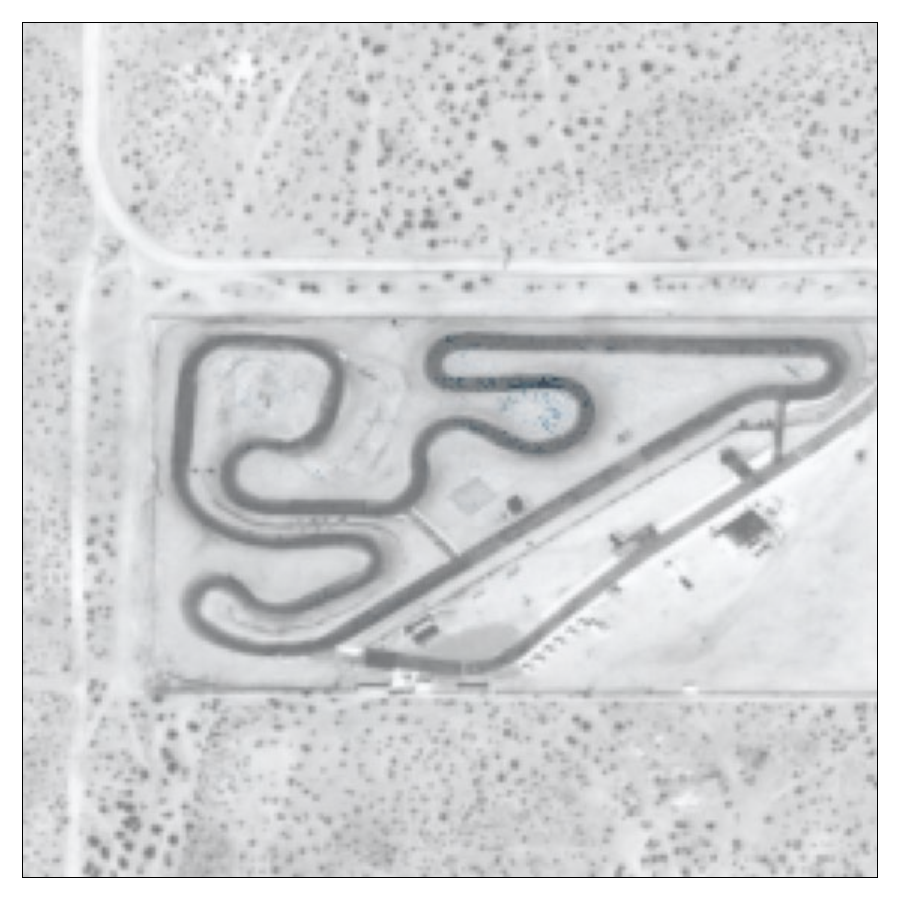}
    }%
    \subfigure[\textsc{FMoW}-\feat]{
        \includegraphics[width=0.3\textwidth]{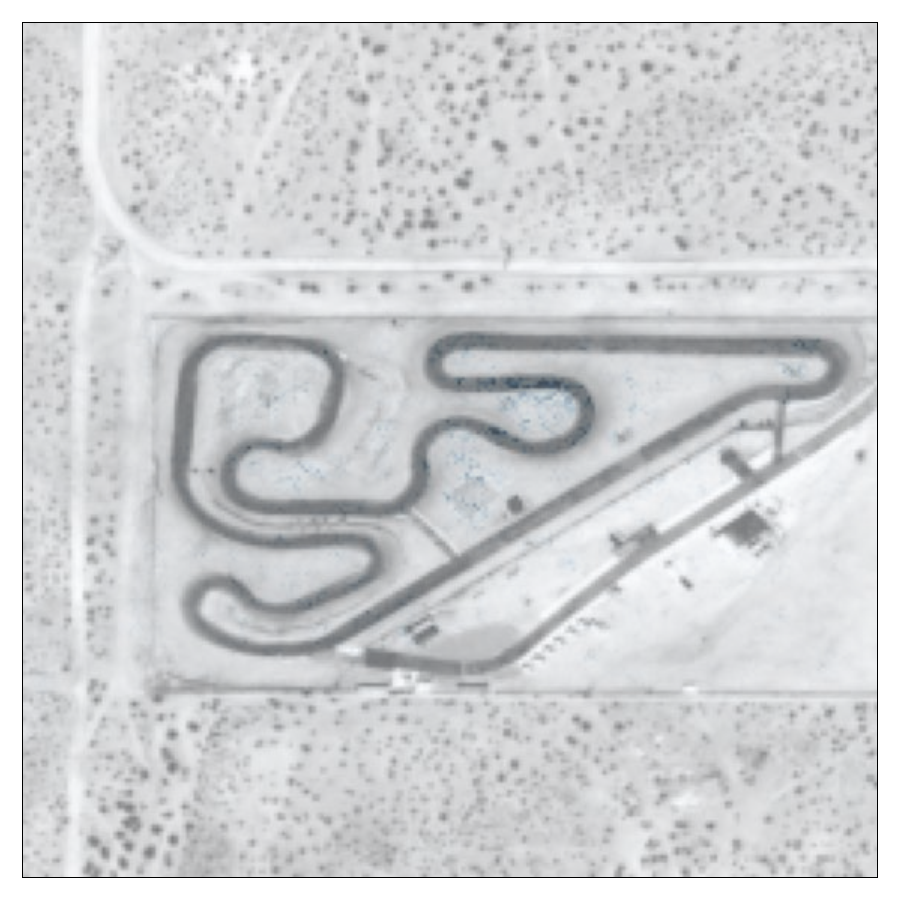}
    }%
    \\
    \subfigure[\textsc{FMoW}-ERM]{
        \includegraphics[width=0.3\textwidth]{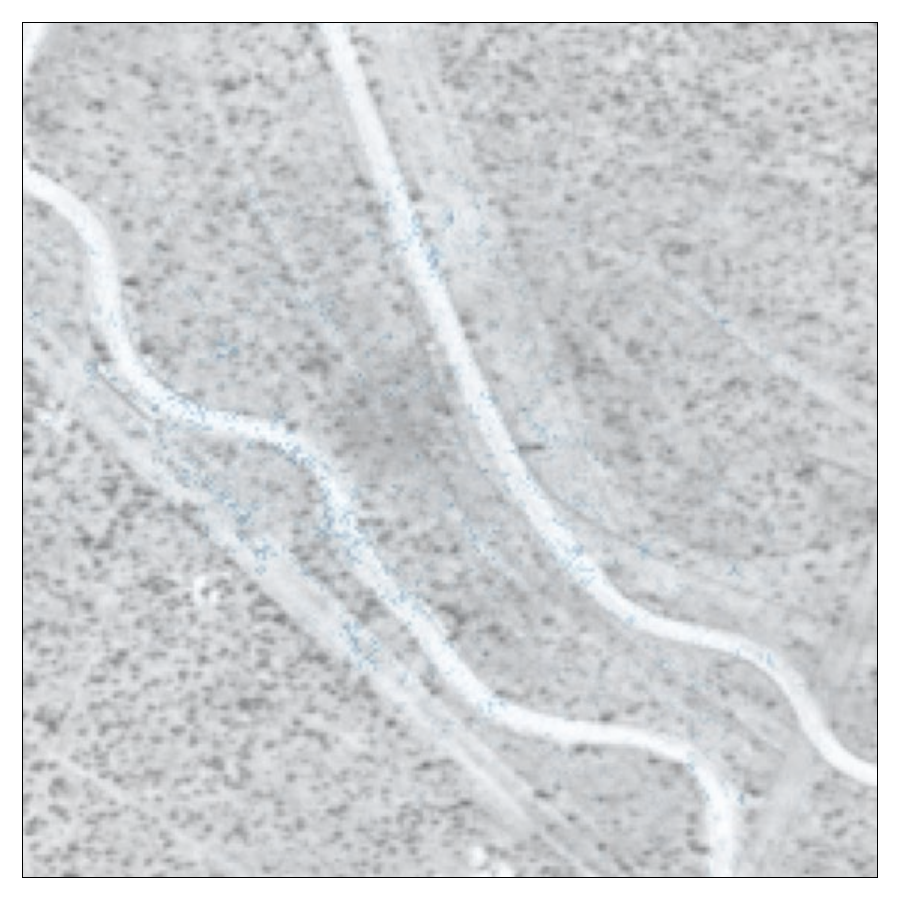}
    }%
    \subfigure[\textsc{FMoW}-Bonsai]{
        \includegraphics[width=0.3\textwidth]{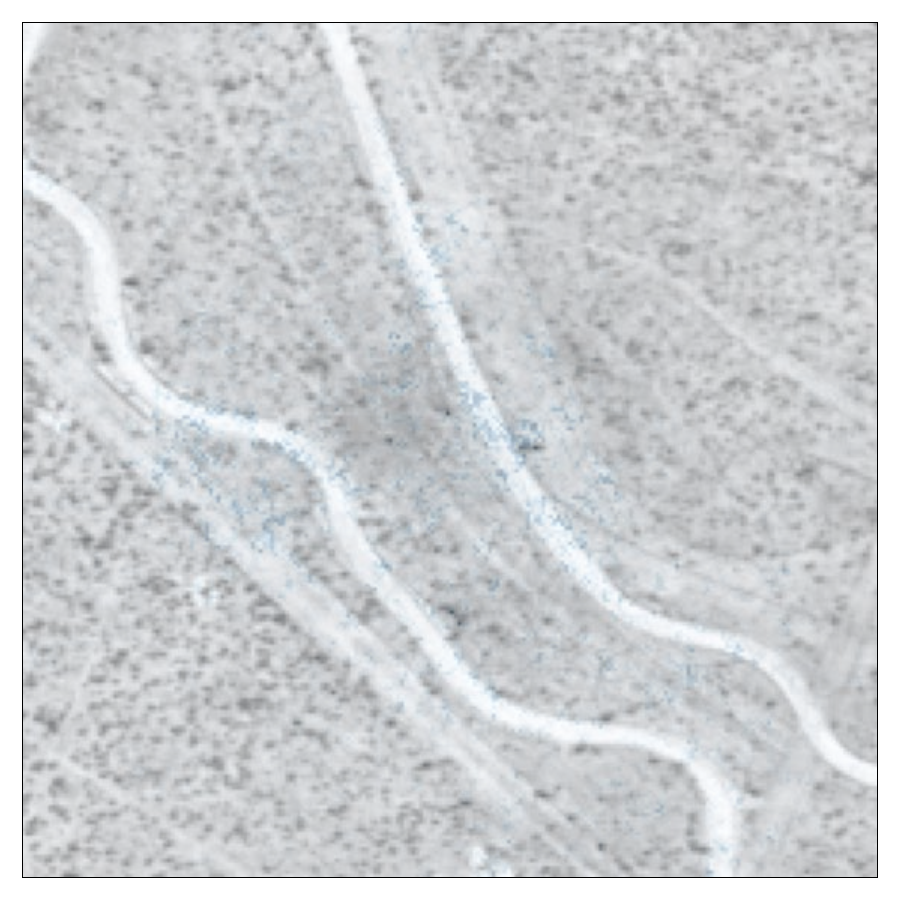}
    }%
    \subfigure[\textsc{FMoW}-\feat]{
        \includegraphics[width=0.3\textwidth]{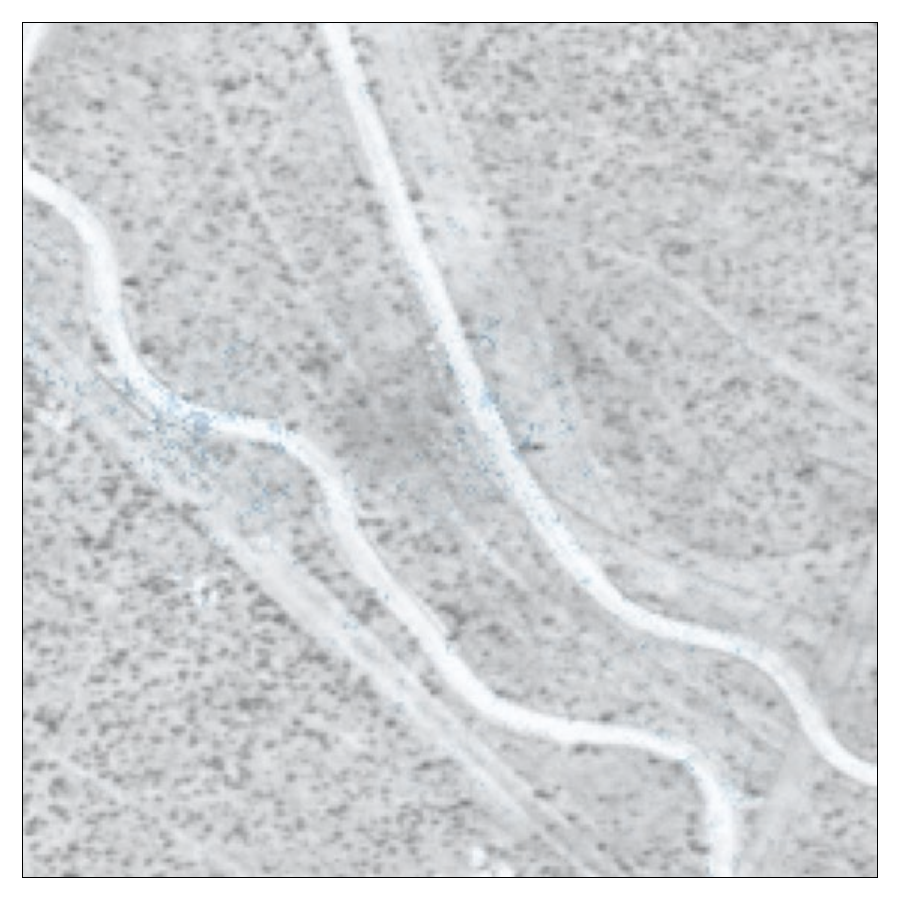}
    }%
    \\
    \subfigure[\textsc{FMoW}-ERM]{
        \includegraphics[width=0.3\textwidth]{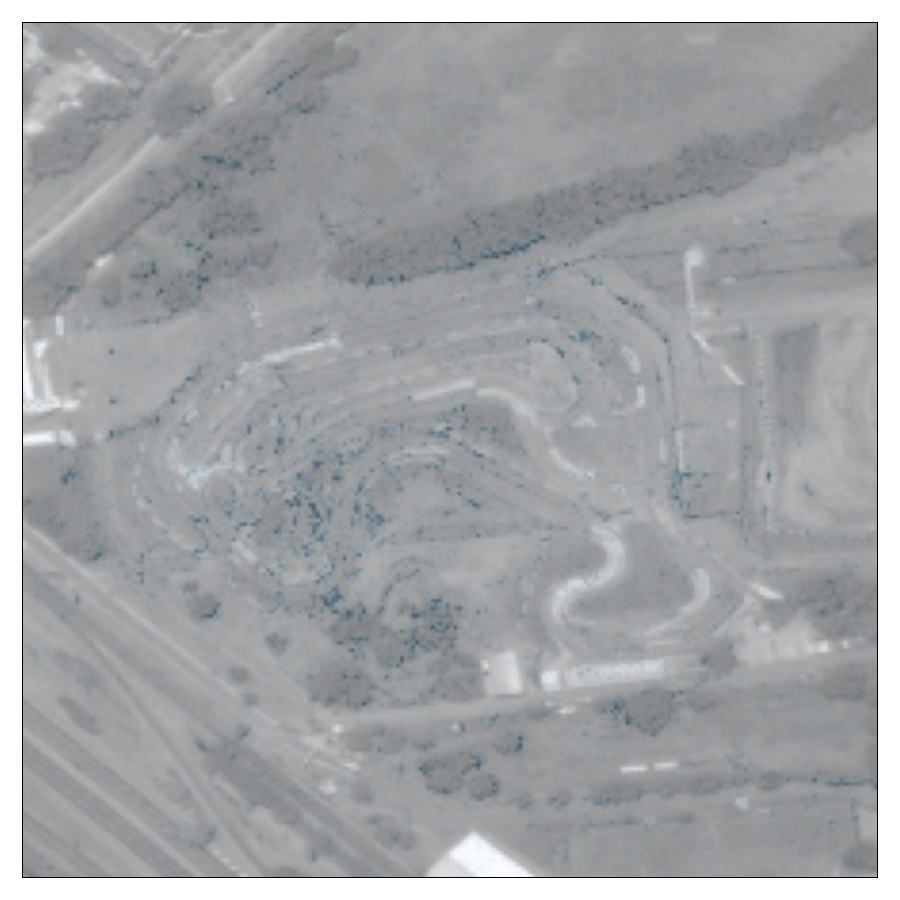}
    }%
    \subfigure[\textsc{FMoW}-Bonsai]{
        \includegraphics[width=0.3\textwidth]{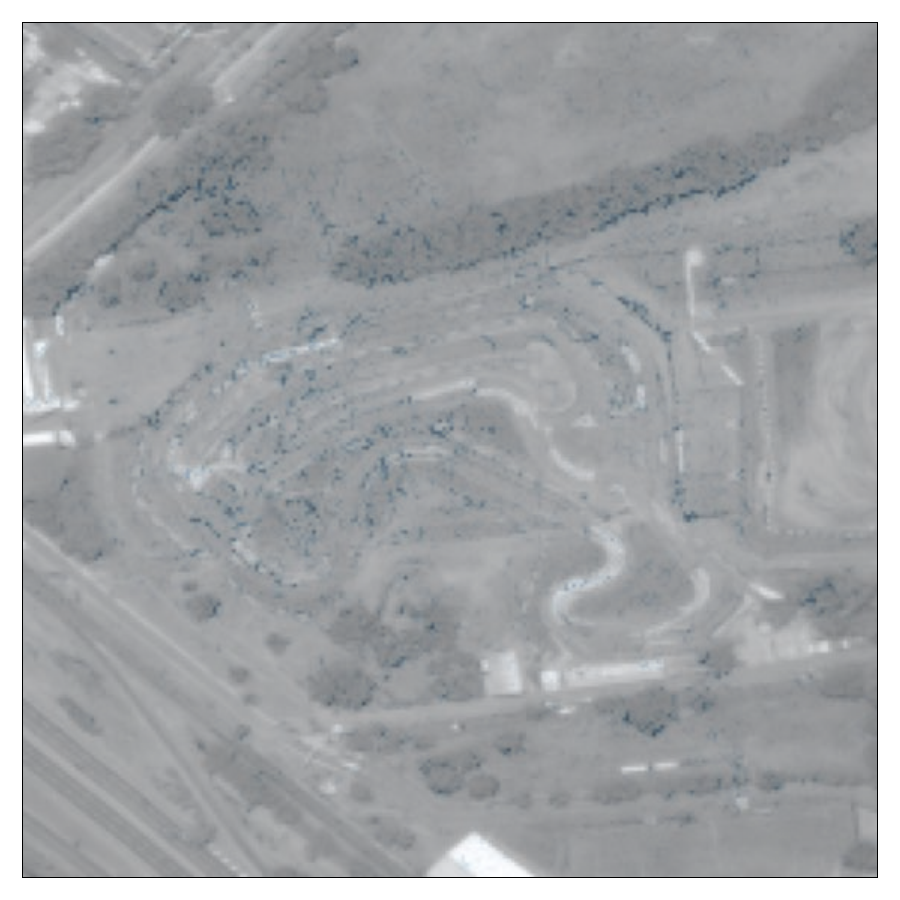}
    }%
    \subfigure[\textsc{FMoW}-\feat]{
        \includegraphics[width=0.3\textwidth]{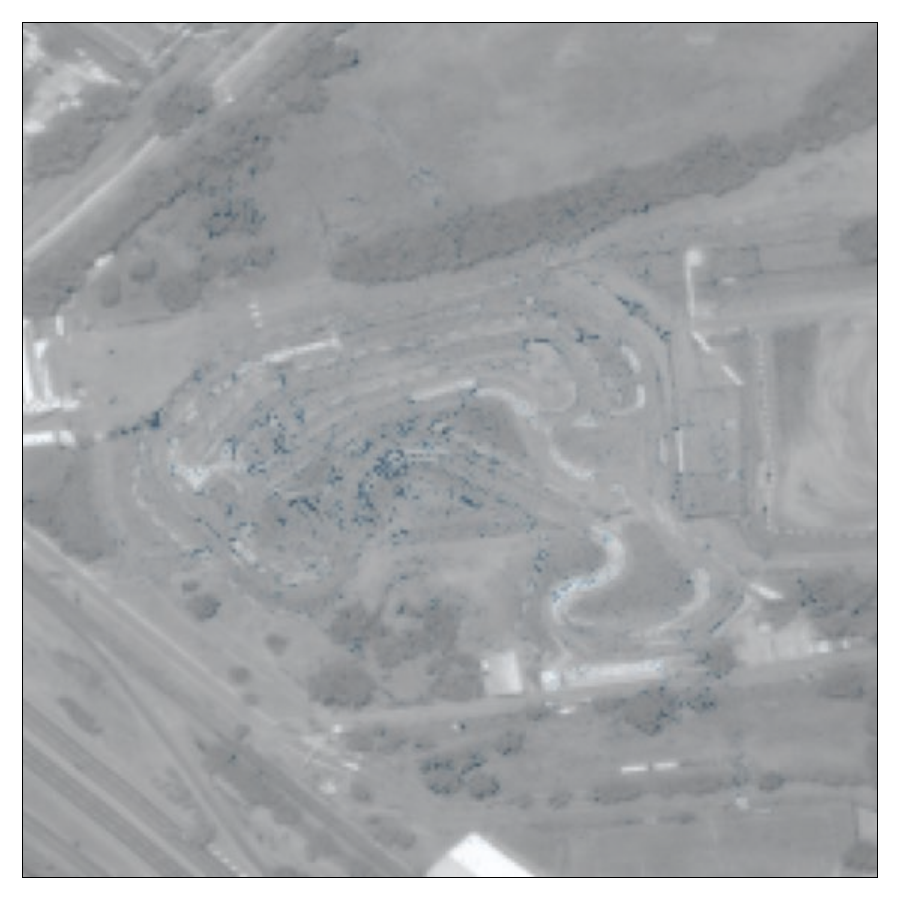}
    }%
    \\
    \vspace{-0.15in}
    \caption[Saliency map of feature learning on \textsc{FMoW} benchmark.]{Saliency map of feature learning on \textsc{FMoW} benchmark. The blue dots are the salient features.
        A deeper blue color denotes more salient features. It can be found that \feat is able to learn more meaningful and diverse features than ERM and Bonsai.}
    \label{CH:FeAT:fig:gradcam_fmow_appdx}
    \vspace{-2cm}
\end{figure}

\begin{figure}[t!]
    \vspace{-0.15in}
    \centering
    \subfigure[\textsc{iWildCam}-ERM]{
        \includegraphics[width=0.3\textwidth]{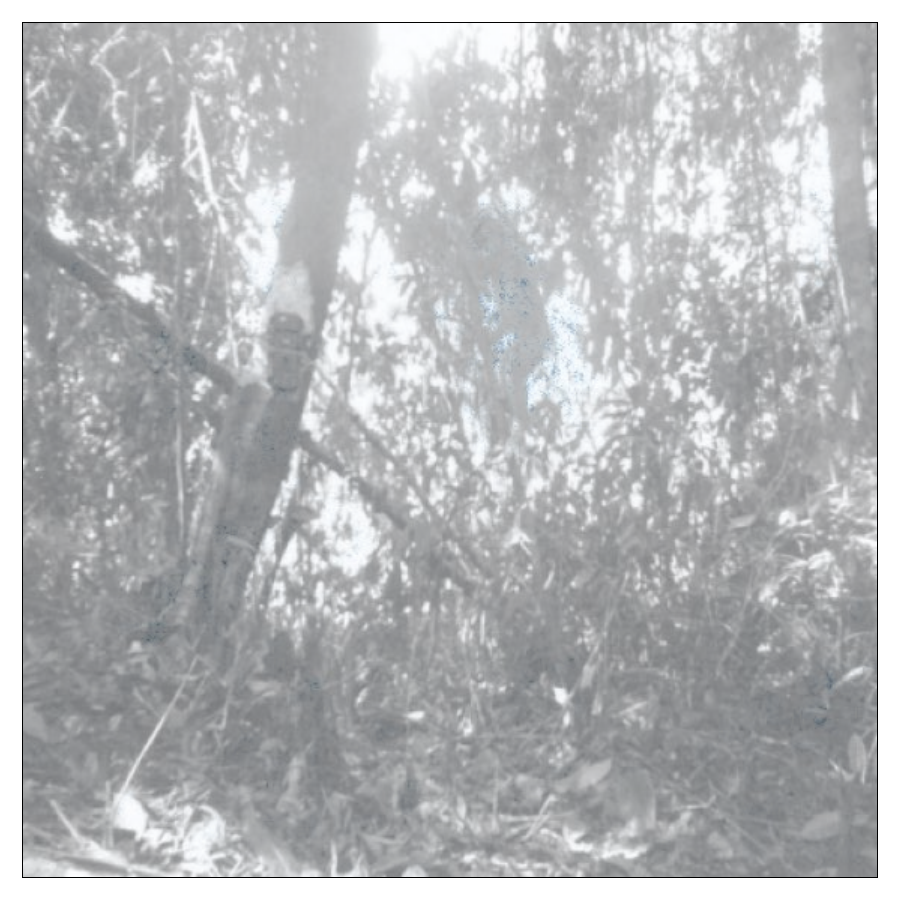}
    }%
    \subfigure[\textsc{iWildCam}-Bonsai]{
        \includegraphics[width=0.3\textwidth]{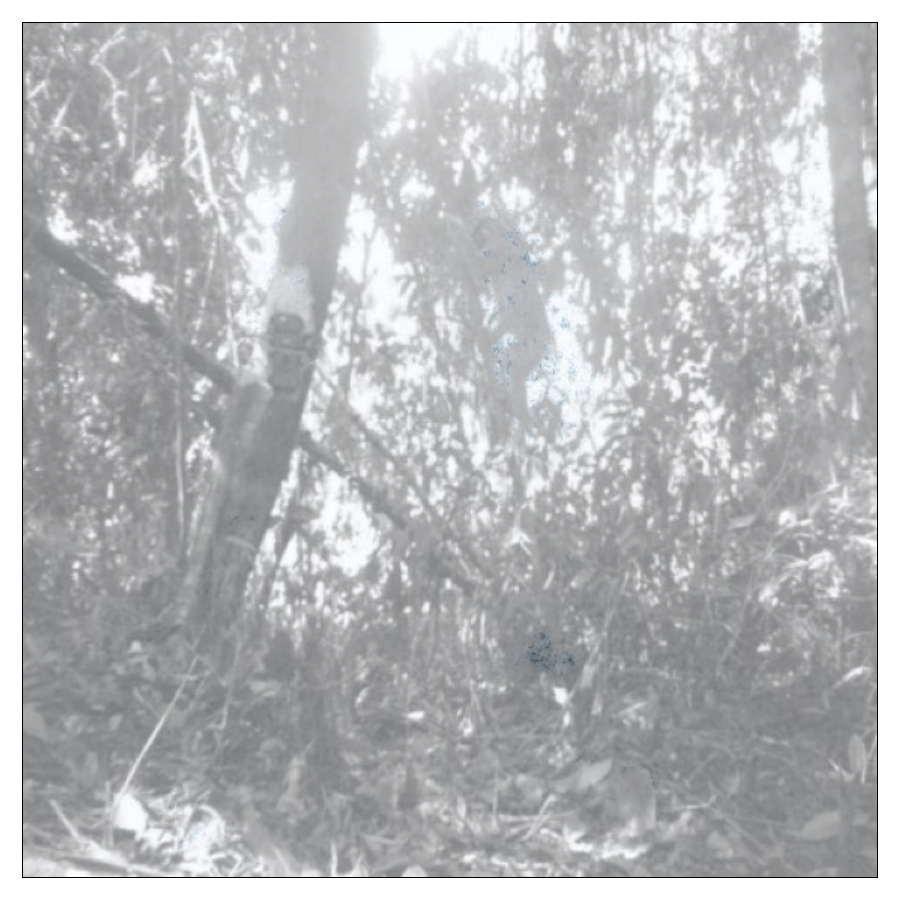}
    }%
    \subfigure[\textsc{iWildCam}-\feat]{
        \includegraphics[width=0.3\textwidth]{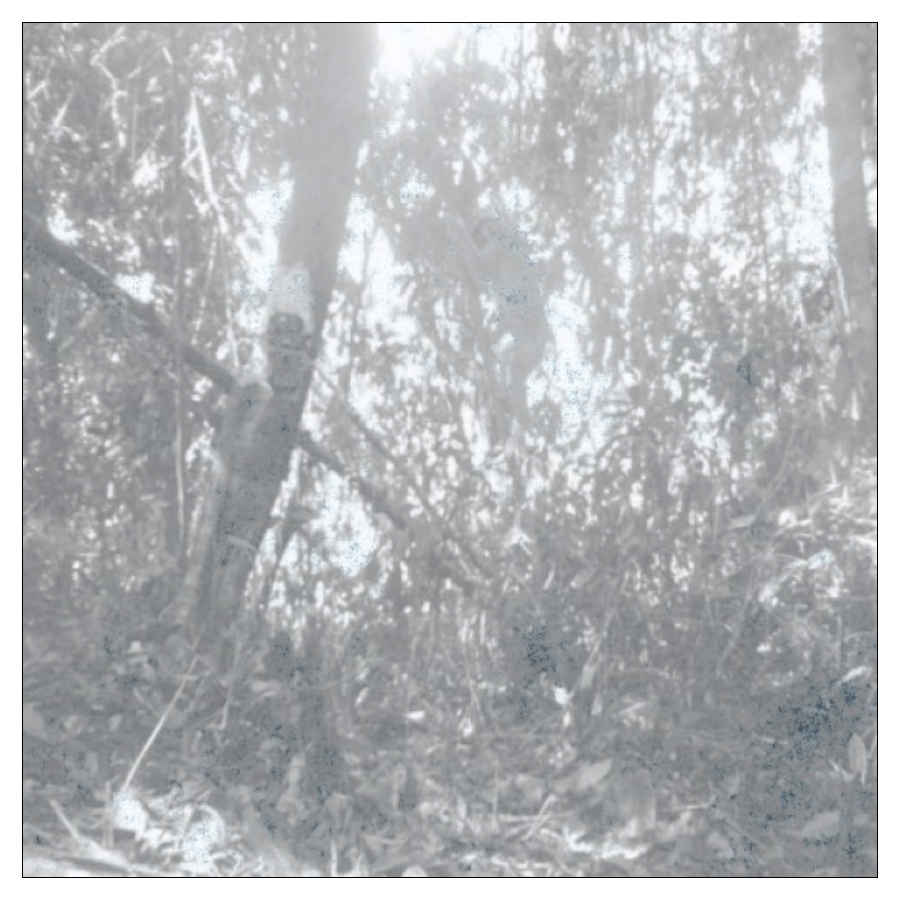}
    }%
    \\
    \subfigure[\textsc{iWildCam}-ERM]{
        \includegraphics[width=0.3\textwidth]{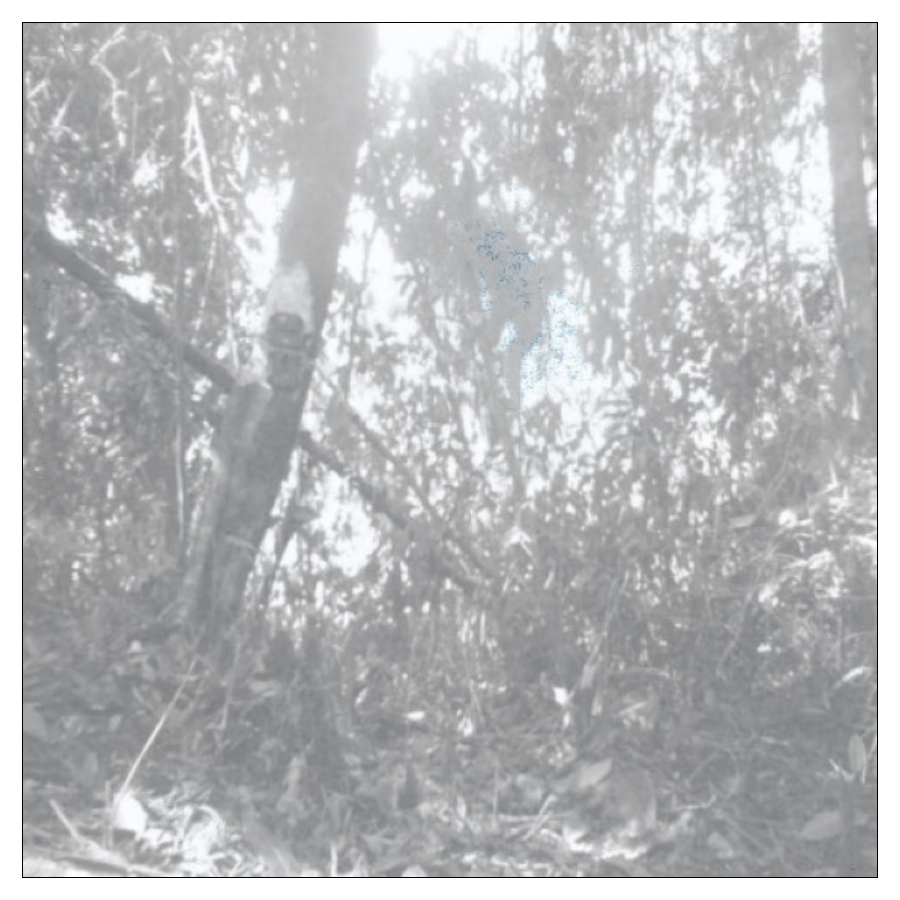}
    }%
    \subfigure[\textsc{iWildCam}-Bonsai]{
        \includegraphics[width=0.3\textwidth]{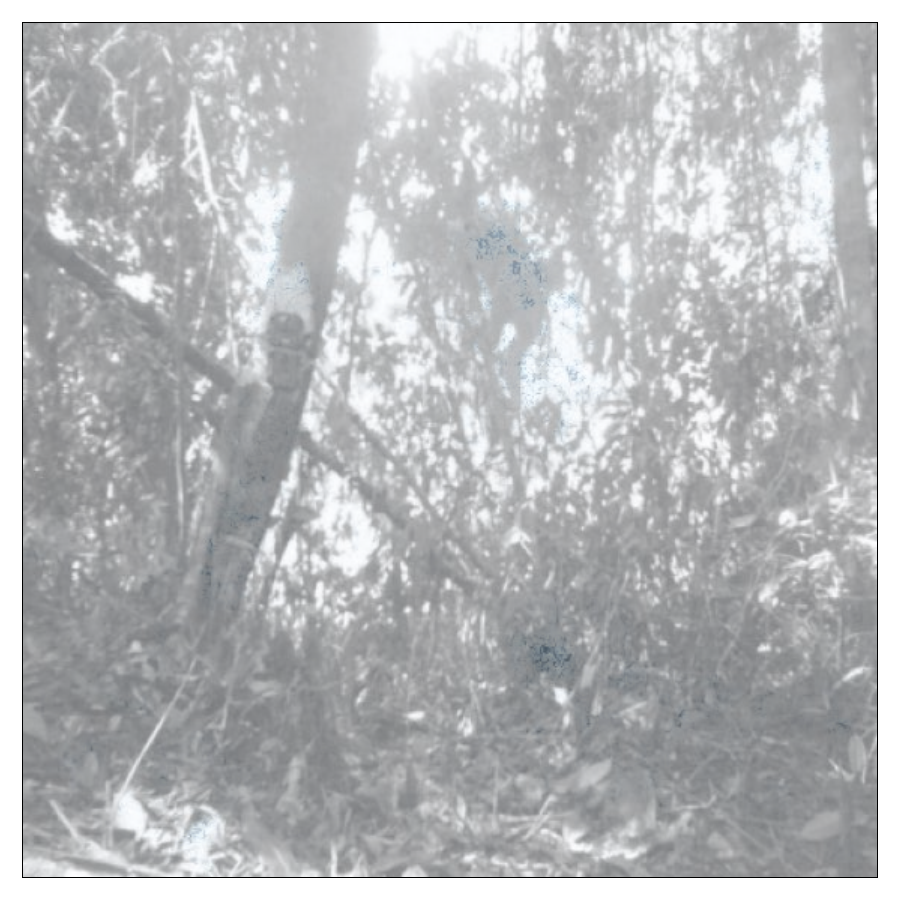}
    }%
    \subfigure[\textsc{iWildCam}-\feat]{
        \includegraphics[width=0.3\textwidth]{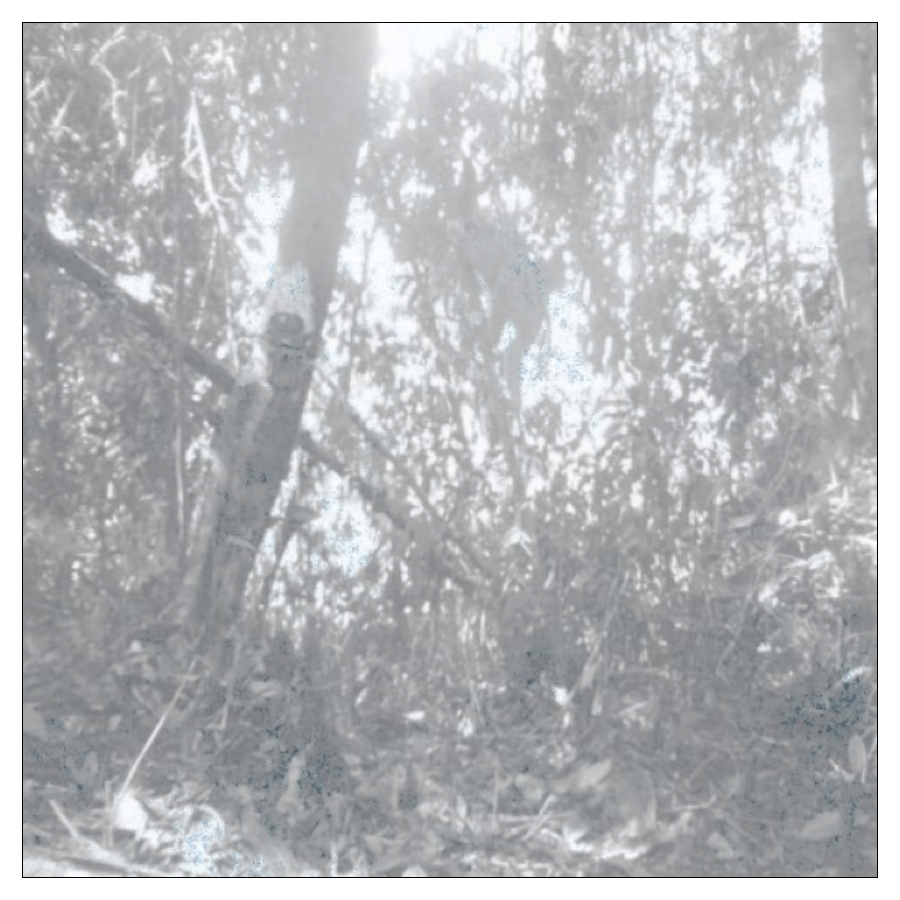}
    }%
    \\
    \subfigure[\textsc{iWildCam}-ERM]{
        \includegraphics[width=0.3\textwidth]{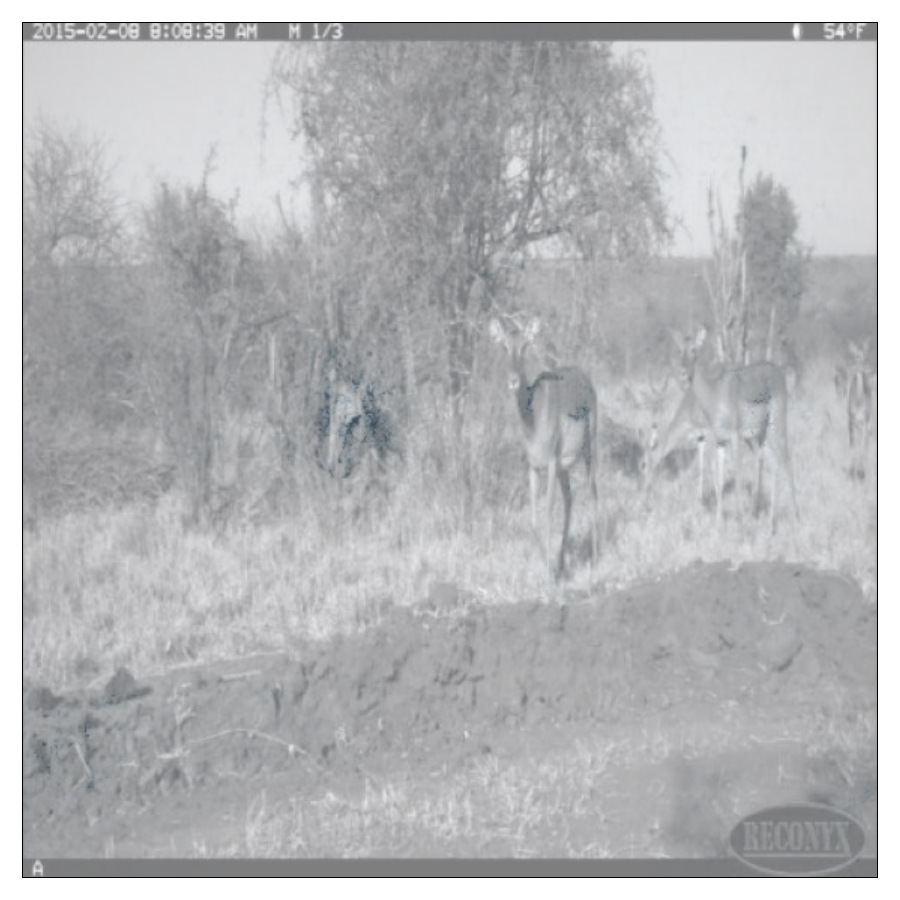}
    }%
    \subfigure[\textsc{iWildCam}-Bonsai]{
        \includegraphics[width=0.3\textwidth]{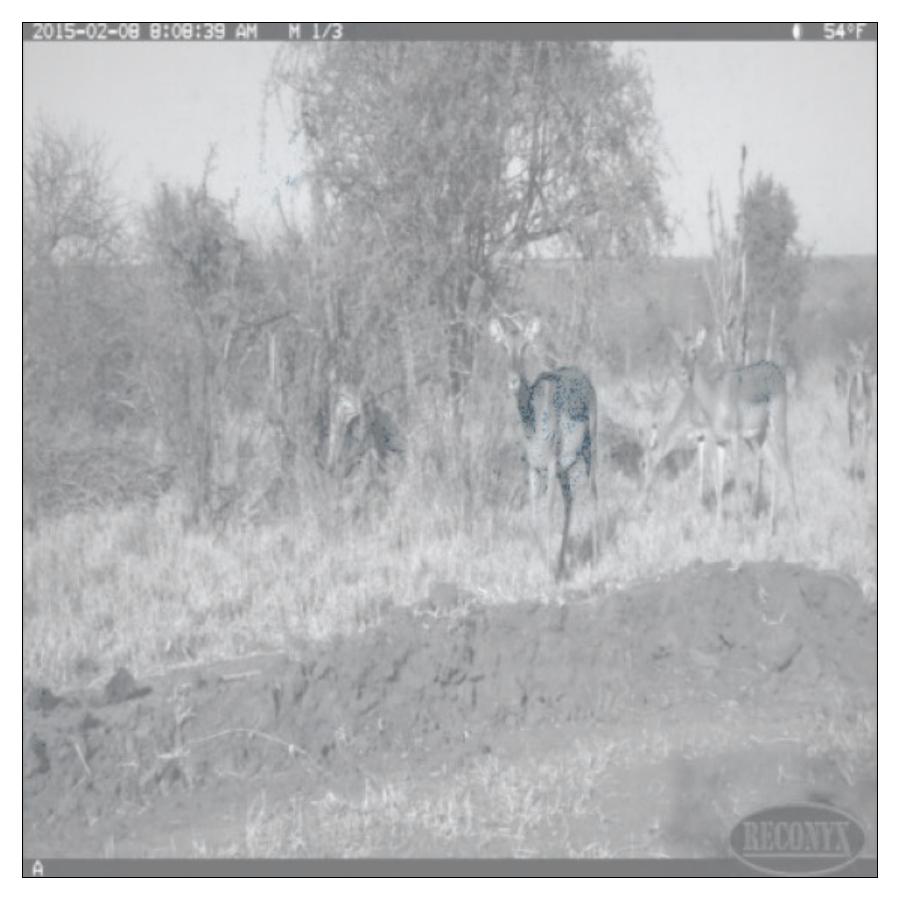}
    }%
    \subfigure[\textsc{iWildCam}-\feat]{
        \includegraphics[width=0.3\textwidth]{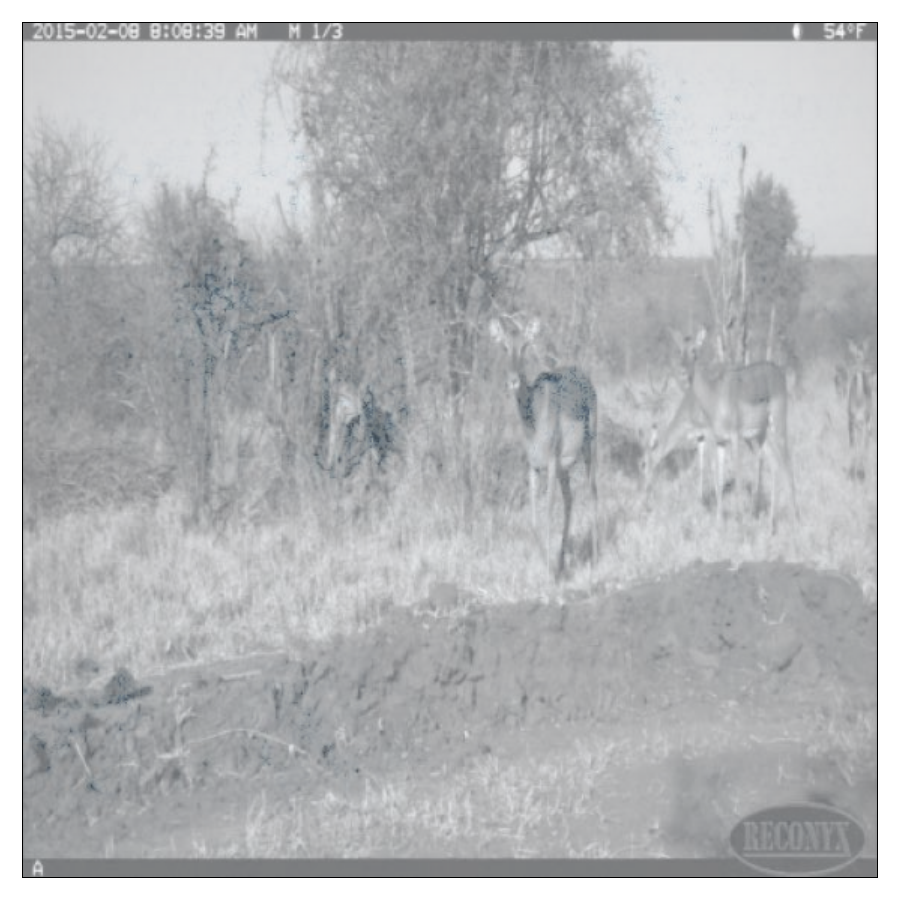}
    }%
    \\
    \subfigure[\textsc{iWildCam}-ERM]{
        \includegraphics[width=0.3\textwidth]{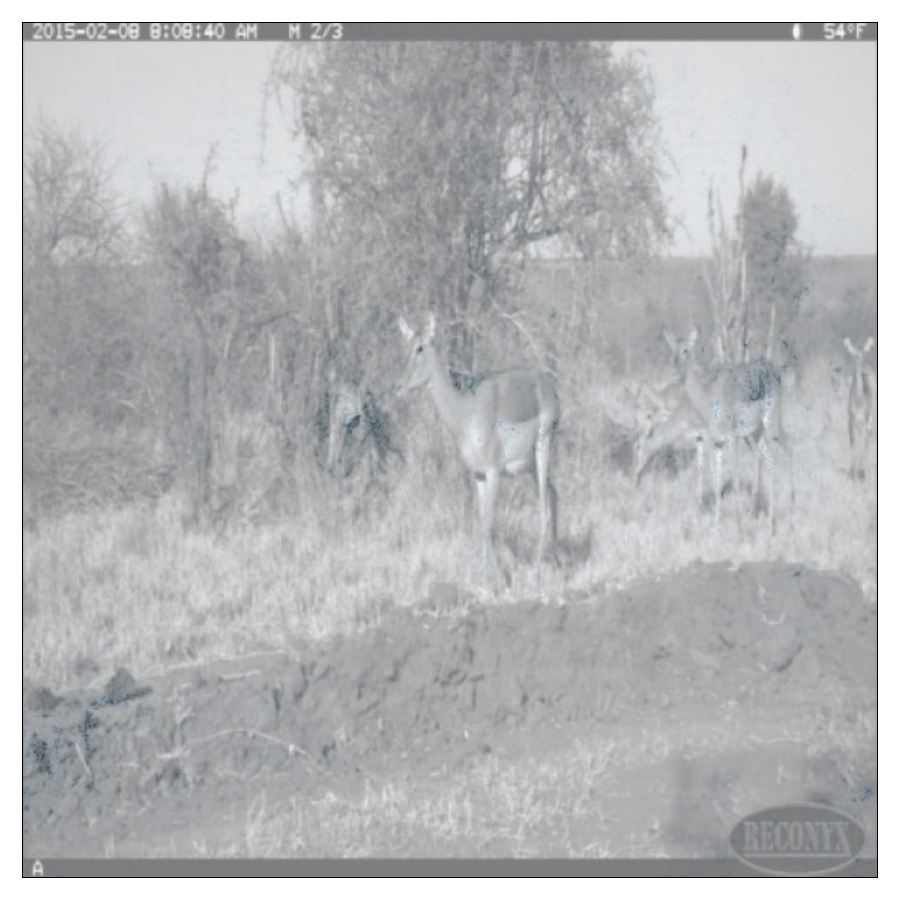}
    }%
    \subfigure[\textsc{iWildCam}-Bonsai]{
        \includegraphics[width=0.3\textwidth]{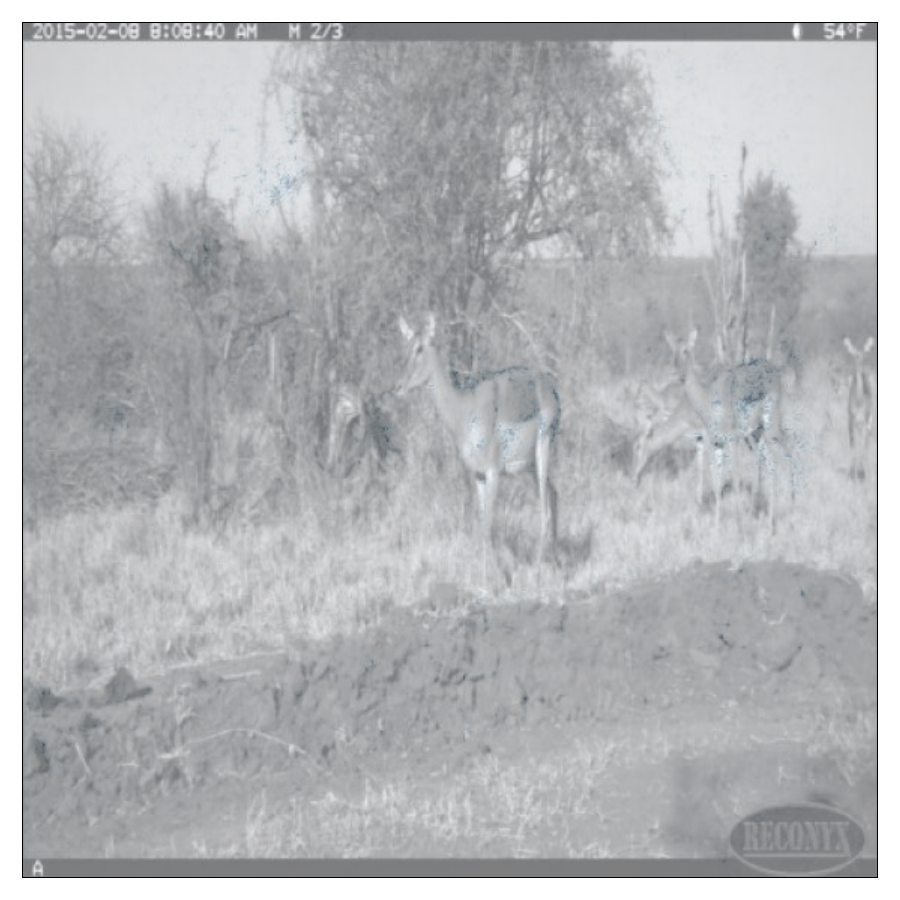}
    }%
    \subfigure[\textsc{iWildCam}-\feat]{
        \includegraphics[width=0.3\textwidth]{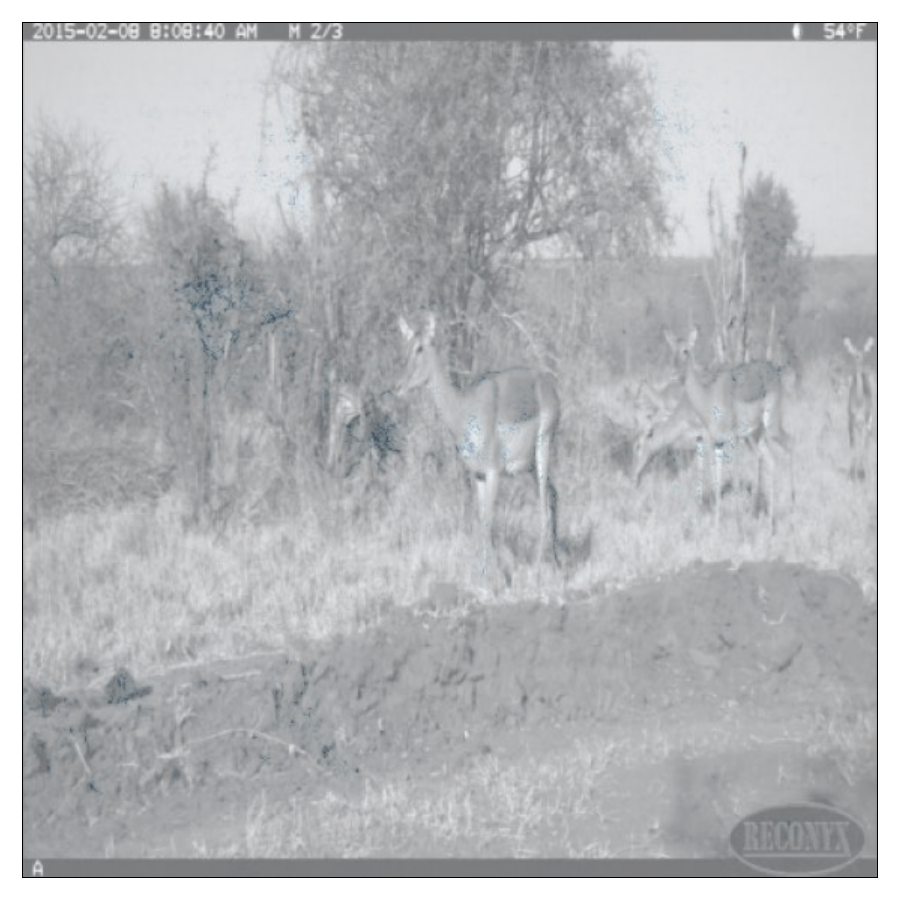}
    }%
    \\
    \caption[Saliency map of feature learning on \textsc{iWildCam} benchmark.]{Saliency map of feature learning on \textsc{iWildCam} benchmark. The blue dots are the salient features.
        A deeper blue color denotes more salient features. It can be found that \feat is able to learn more meaningful and diverse features than ERM and Bonsai.}
    \label{CH:FeAT:fig:gradcam_iwildcam_appdx}
    \vspace{-2cm}
\end{figure}

\begin{figure}[t!]
    \vspace{-0.15in}
    \centering
    \subfigure[\textsc{RxRx1}-ERM]{
        \includegraphics[width=0.3\textwidth]{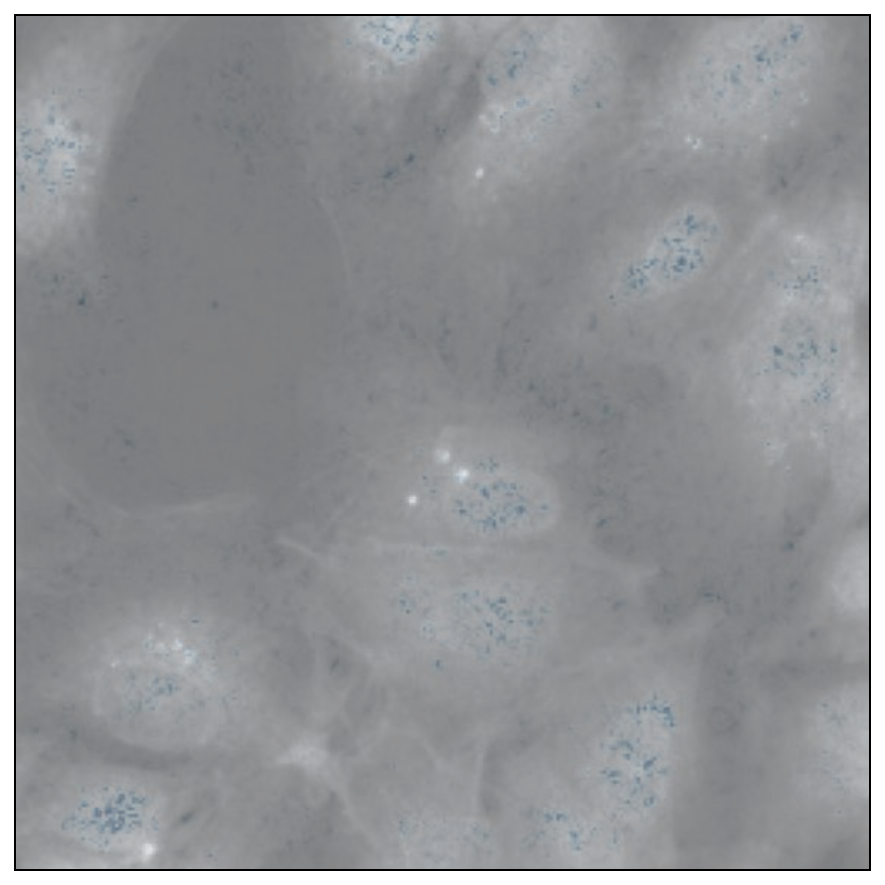}
    }%
    \subfigure[\textsc{RxRx1}-Bonsai]{
        \includegraphics[width=0.3\textwidth]{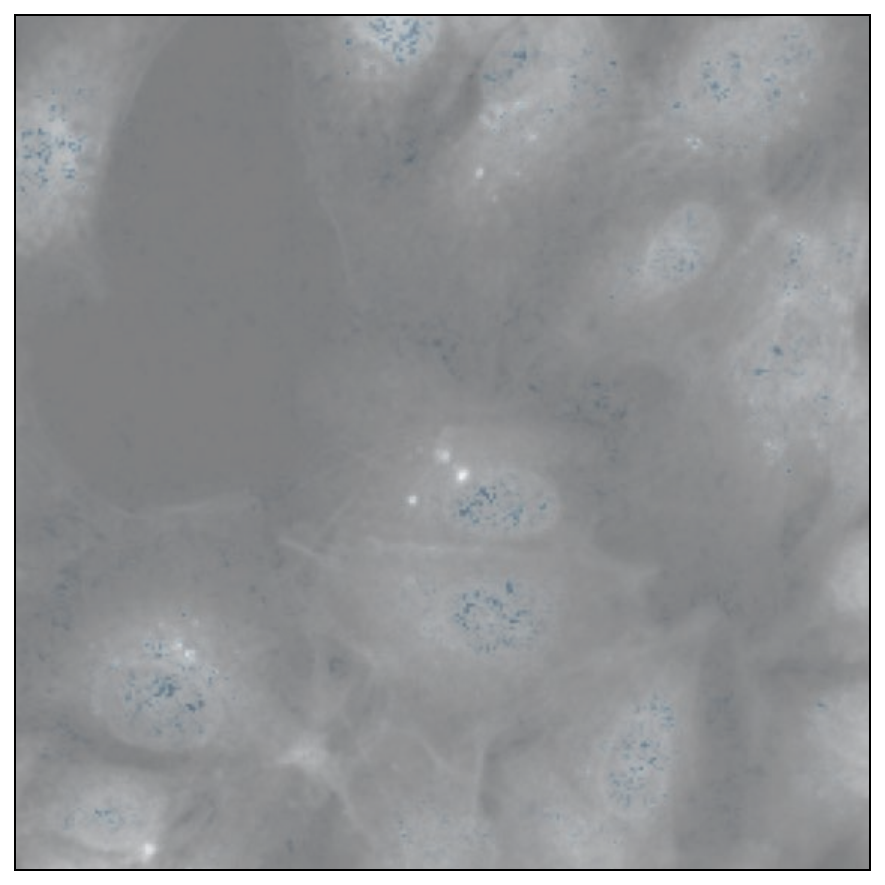}
    }%
    \subfigure[\textsc{RxRx1}-\feat]{
        \includegraphics[width=0.3\textwidth]{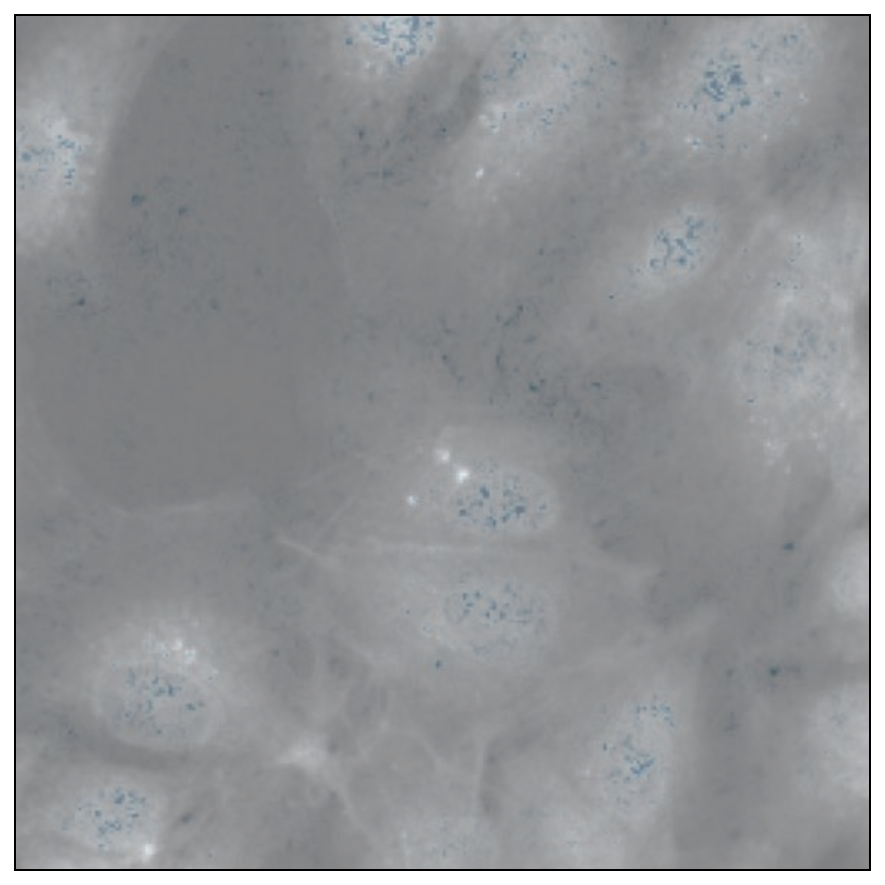}
    }%
    \\
    \subfigure[\textsc{RxRx1}-ERM]{
        \includegraphics[width=0.3\textwidth]{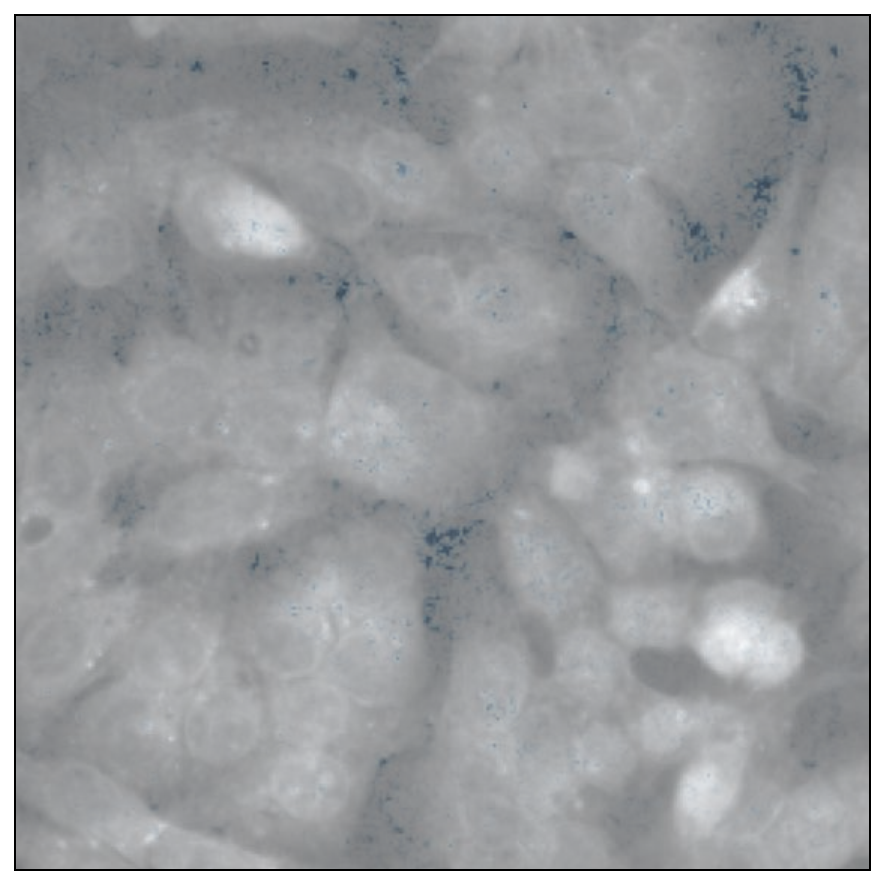}
    }%
    \subfigure[\textsc{RxRx1}-Bonsai]{
        \includegraphics[width=0.3\textwidth]{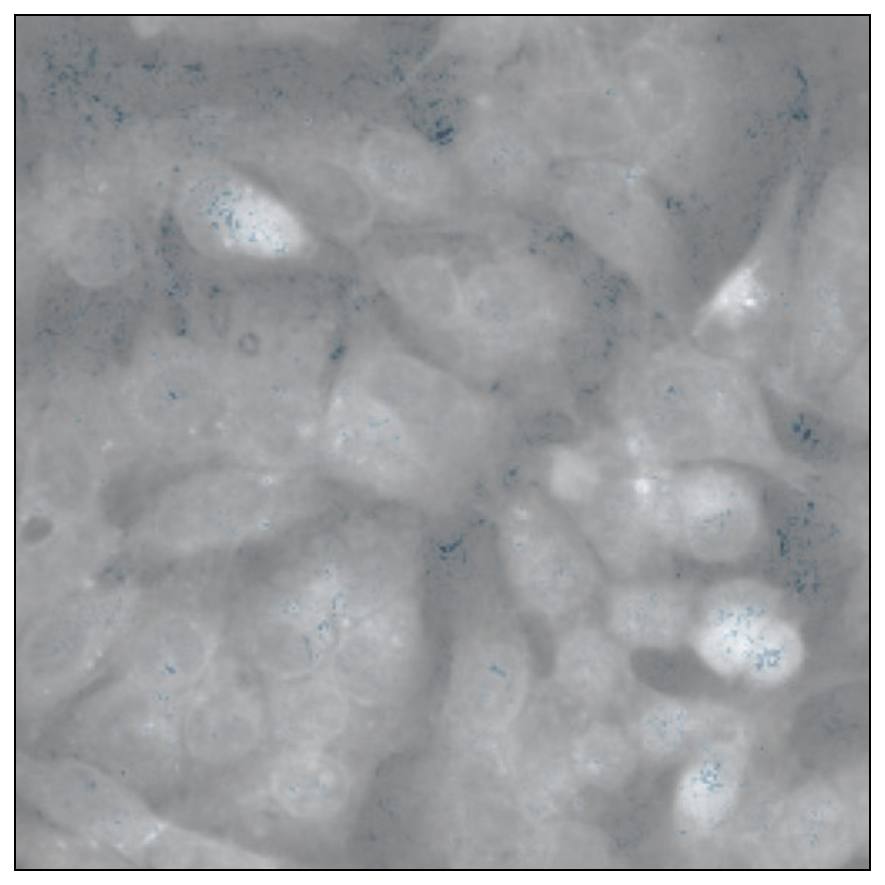}
    }%
    \subfigure[\textsc{RxRx1}-\feat]{
        \includegraphics[width=0.3\textwidth]{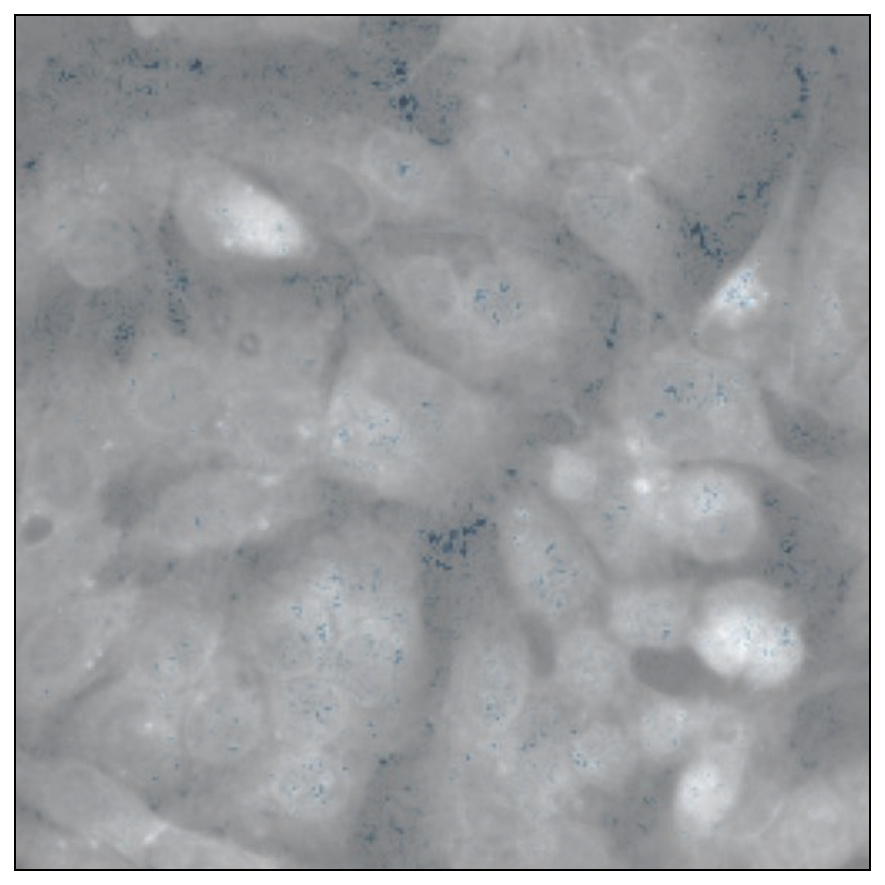}
    }%
    \\
    \subfigure[\textsc{RxRx1}-ERM]{
        \includegraphics[width=0.3\textwidth]{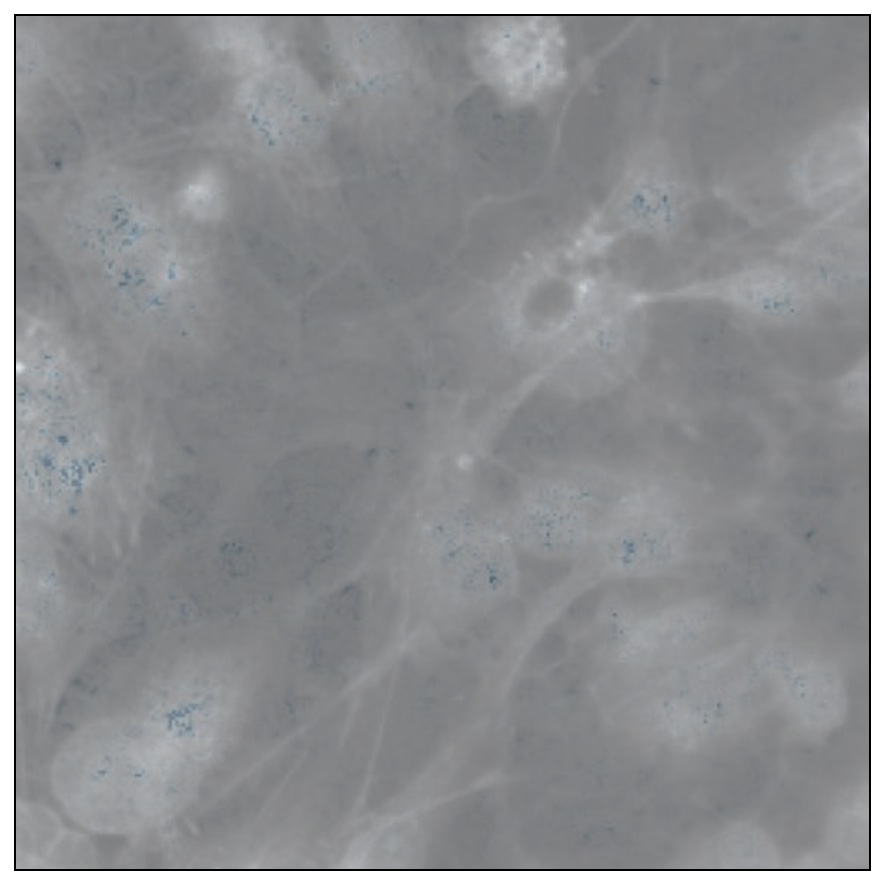}
    }%
    \subfigure[\textsc{RxRx1}-Bonsai]{
        \includegraphics[width=0.3\textwidth]{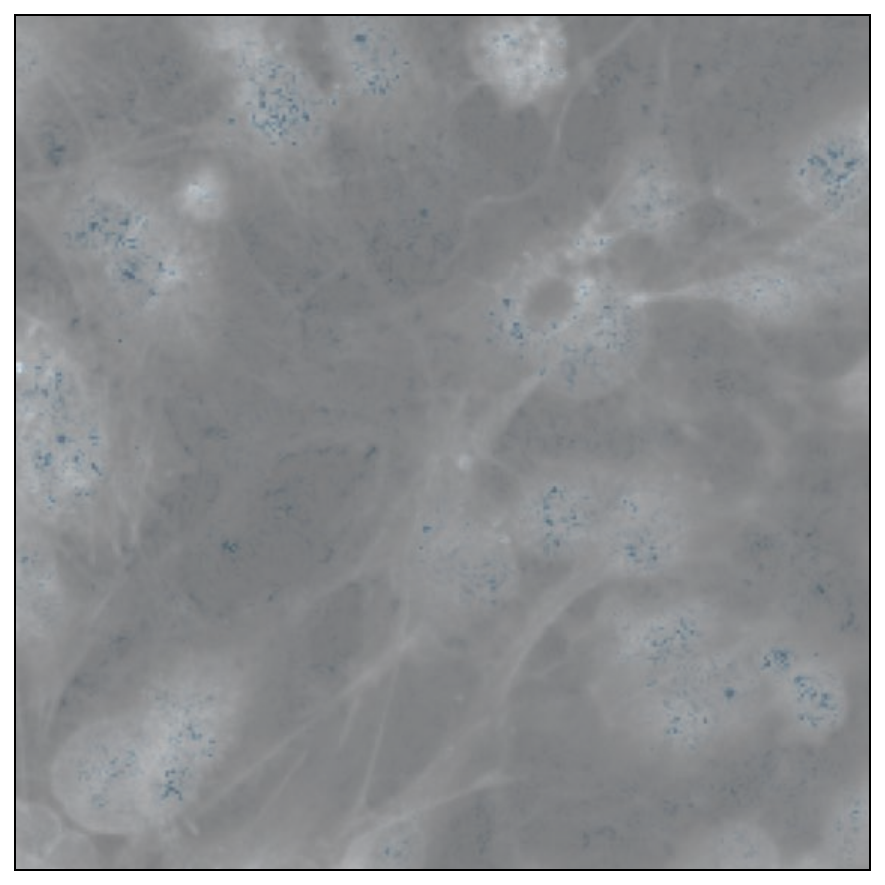}
    }%
    \subfigure[\textsc{RxRx1}-\feat]{
        \includegraphics[width=0.3\textwidth]{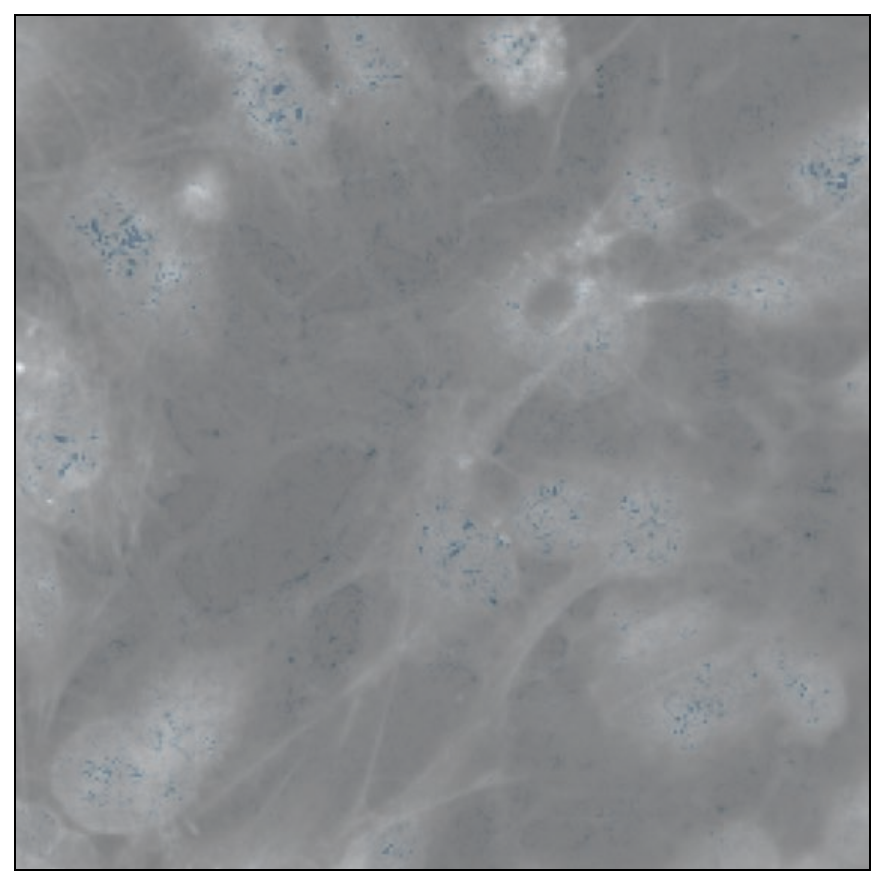}
    }%
    \\
    \subfigure[\textsc{RxRx1}-ERM]{
        \includegraphics[width=0.3\textwidth]{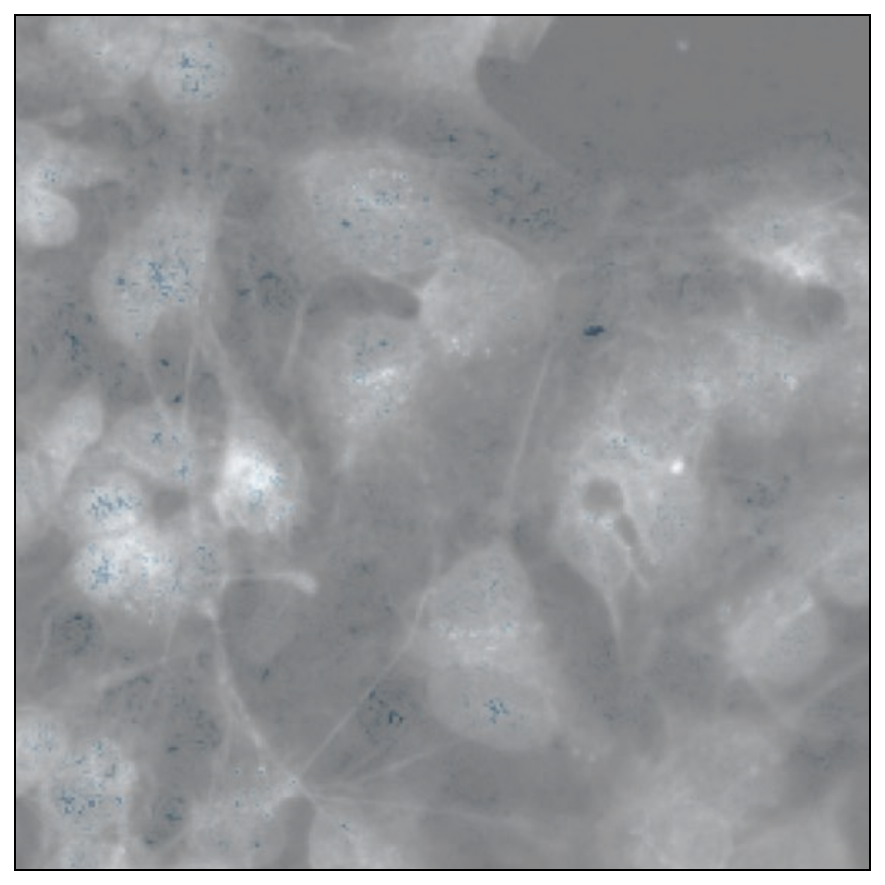}
    }%
    \subfigure[\textsc{RxRx1}-Bonsai]{
        \includegraphics[width=0.3\textwidth]{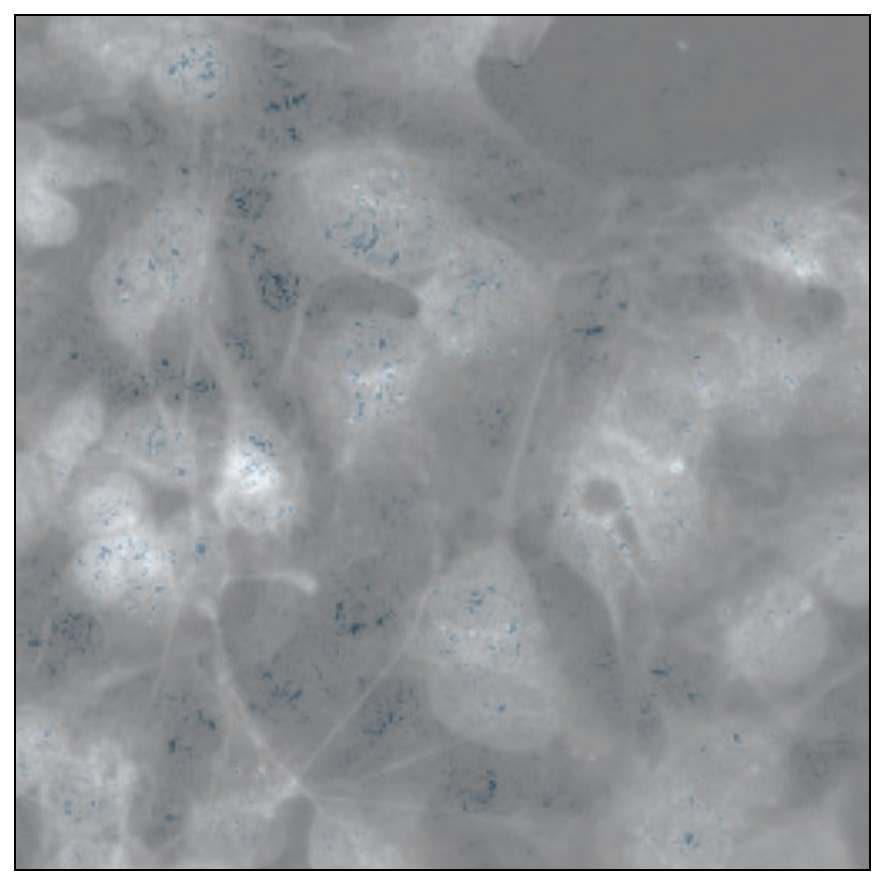}
    }%
    \subfigure[\textsc{RxRx1}-\feat]{
        \includegraphics[width=0.3\textwidth]{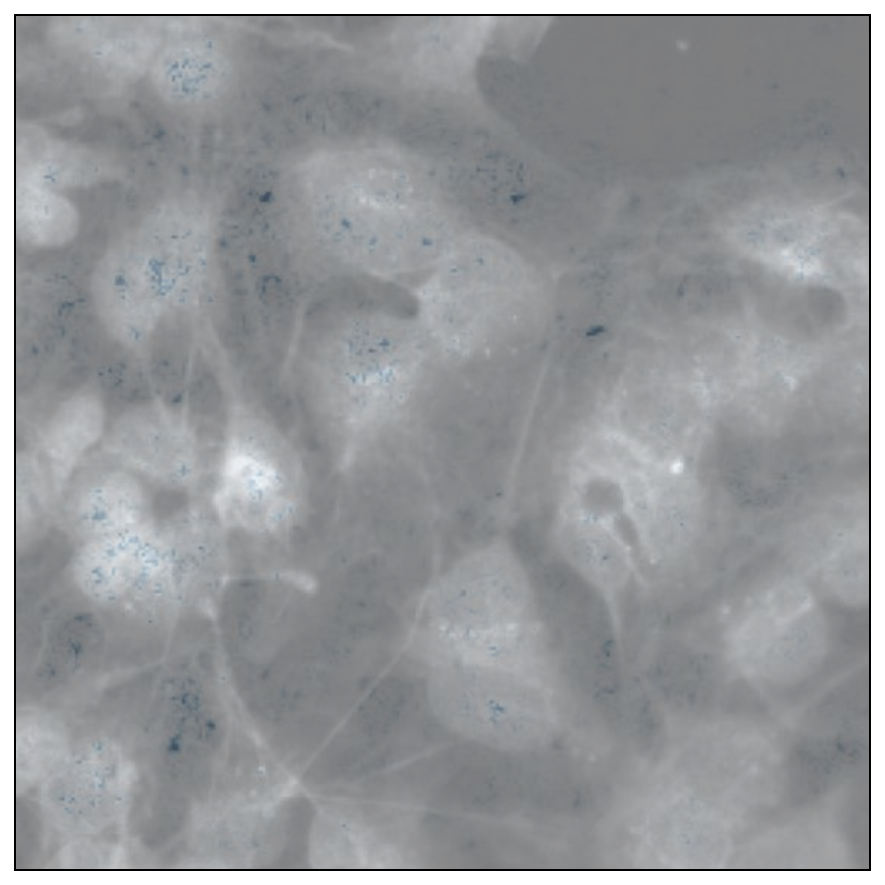}
    }%
    \\
    \caption[Saliency map of feature learning on \textsc{RxRx1} benchmark.]{Saliency map of feature learning on \textsc{RxRx1} benchmark. The blue dots are the salient features.
        A deeper blue color denotes more salient features. It can be found that \feat is able to learn more meaningful and diverse features than ERM and Bonsai.}
    \label{CH:FeAT:fig:gradcam_rxrx_appdx}
    \vspace{-2cm}
\end{figure}

We also visualize the saliency maps of ERM, Bonsai, and \feat on all real-world datasets used in our work with \url{https://github.com/pytorch/captum}.
The visualizations are shown as in Fig.~\ref{CH:FeAT:fig:gradcam_wilds_nlp_appdx} to Fig.~\ref{CH:FeAT:fig:gradcam_rxrx_appdx},
for which the labels and the predictions of different algorithms are given in Table.~\ref{tab:viz_label_appdx}.
It can be found that, across various tasks and data modalities, \feat effectively learns more meaningful and diverse features than ERM and Bonsai, which serve as strong evidence for the consistent superiority of \feat in OOD generalization.

\clearpage
\bibliographystyle{References/ref_style}
\bibliography{References/ood,References/causality,References/gnn,References/pub,References/others}

\end{document}